# Artificial Neural Network and Deep Learning

## Fundamentals and Theory

**Mohamed M. Hammad**

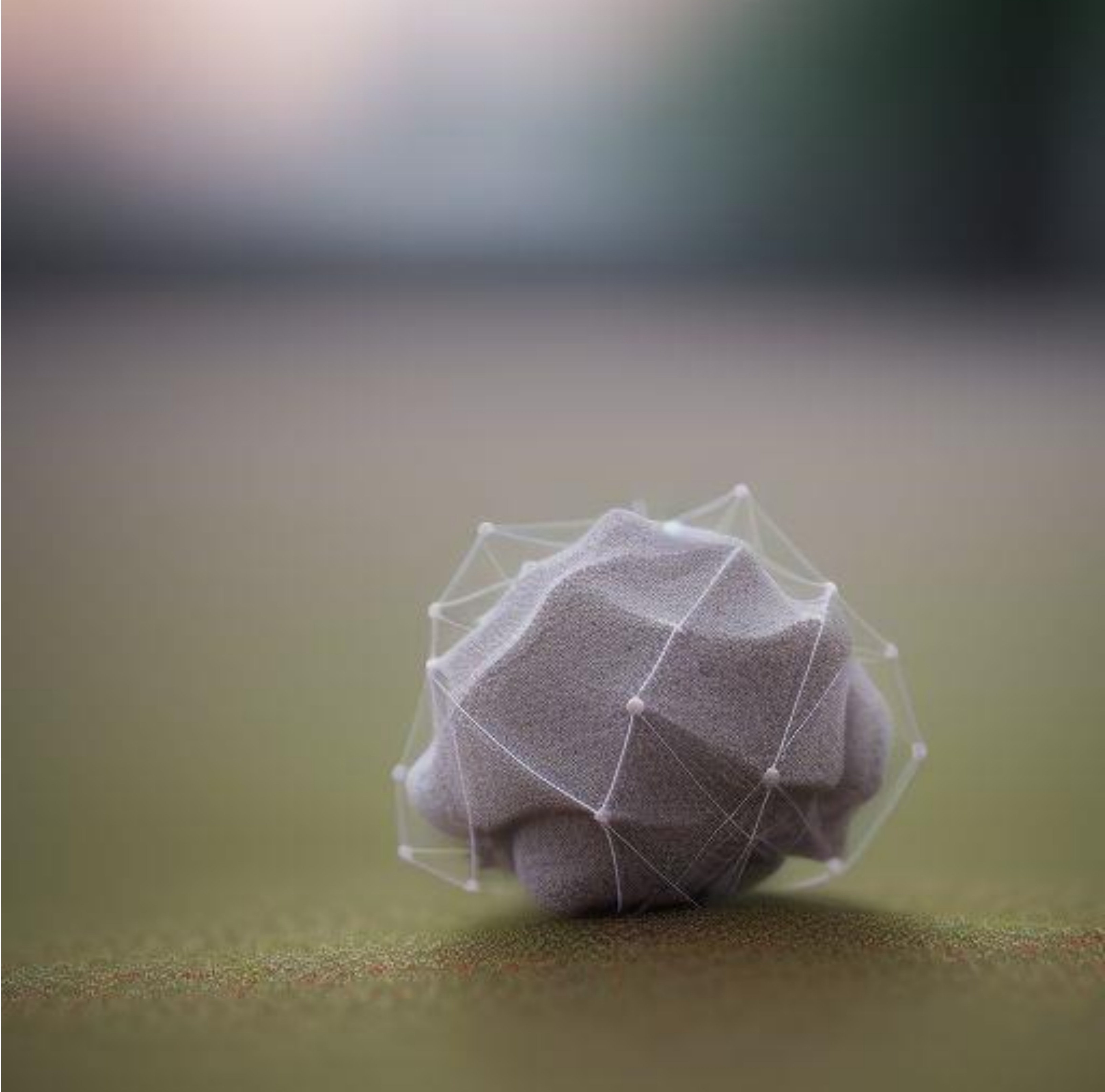

# Artificial Neural Network and Deep Learning: Fundamentals and Theory

**M. M. Hammad**

Department of Mathematics and Computer Science

Faculty of Science

Damanhour University, Egypt

https://orcid.org/0000-0003-0306-9719

m_hammad@sci.dmu.edu.eg

To

my

mother

# Abstract

"Artificial Neural Network and Deep Learning: Fundamentals and Theory" offers a comprehensive exploration of the foundational principles and advanced methodologies in neural networks and deep learning. This book begins with essential concepts in descriptive statistics and probability theory, laying a solid groundwork for understanding data and probability distributions. As the reader progresses, they are introduced to matrix calculus and gradient optimization, crucial for training and fine-tuning neural networks. The book delves into multilayer feed-forward neural networks, explaining their architecture, training processes, and the backpropagation algorithm. Key challenges in neural network optimization, such as activation function saturation, vanishing and exploding gradients, and weight initialization, are thoroughly discussed. The text covers various learning rate schedules and adaptive algorithms, providing strategies to optimize the training process. Techniques for generalization and hyperparameter tuning, including Bayesian optimization and Gaussian processes, are also presented to enhance model performance and prevent overfitting. Advanced activation functions are explored in detail, categorized into sigmoid-based, ReLU-based, ELU-based, miscellaneous, non-standard, and combined types. Each activation function is examined for its properties and applications, offering readers a deep understanding of their impact on neural network behavior. The final chapter introduces complex-valued neural networks, discussing complex numbers, functions, and visualizations, as well as complex calculus and backpropagation algorithms. This chapter provides a comprehensive overview of complex activation functions and their unique properties. This book equips readers with the knowledge and skills necessary to design, and optimize advanced neural network models, contributing to the ongoing advancements in artificial intelligence.

# Preface

The rapid advancements in artificial intelligence over the past few decades have significantly transformed various industries, leading to breakthroughs that were once considered science fiction. Central to these advancements are artificial neural networks and deep learning, which have revolutionized fields such as computer vision, natural language processing, autonomous systems, and beyond. This book, "Artificial Neural Network and Deep Learning: Fundamentals and Theory," aims to provide a comprehensive and in-depth exploration of these technologies, equipping readers with the theoretical knowledge needed to understand, and design advanced neural network models.

A central focus of this book is to introduce readers to the solid mathematical foundations underlying neural networks. This book is designed not only to cater to beginners in the field but also to serve as a valuable reference for seasoned data scientists, machine learning practitioners, biostatisticians, finance professionals, and engineers. Whether they possess prior knowledge of deep learning or seek to fill gaps in their understanding, this book aims to address their needs. We assume that the reader has no prior experience in neural networks and optimization. We tried to provide proofs as simply as possible so that any reader with a background in calculus could easily follow them. It can be used as a textbook for a course spanning one semester.

Each chapter is designed to build on the previous ones, creating a cohesive and comprehensive guide to artificial neural networks and deep learning. Whether you are a student, researcher, or practitioner, this book will equip you with the knowledge and skills to tackle the challenges of modern artificial intelligence and contribute to the ongoing advancements in this exciting field.

The journey begins with Chapter 1: Descriptive Statistics and Probability Theory, where we lay the foundation by discussing essential concepts in statistics and probability. This chapter covers frequency distributions, histograms, and measures of central tendency and dispersion. It further delves into the concepts of symmetry and peakedness, random experiments, sample spaces, and counting techniques. The chapter concludes with a detailed examination of probability interpretations, axioms, conditional probability, and the different types of random variables and distributions, both discrete and continuous. Understanding these fundamental concepts is crucial for anyone looking to delve into machine learning and neural network algorithms, as they provide the statistical framework upon which these models are built.

Moving forward, Chapter 2: Matrix Calculus and Gradient Optimization introduces the mathematical tools necessary for understanding and optimizing neural networks. This chapter covers vectors and matrices using both index and Dirac notations, basics of matrix calculus, and the chain rule. It also explores numerical differentiation, finite difference methods, and optimality criteria for functions of multiple variables. The concepts of gradient descent and other optimization techniques are thoroughly discussed, providing readers with the skills to optimize neural network models effectively.

Chapter 3: Multilayer Feed-Forward Neural Networks dives into the core architecture of neural networks. It explains feed-forward neural networks and forward propagation, presenting the multilayer network as a computational graph. The chapter also covers automatic differentiation, the training process, and loss/cost functions. It introduces the fundamental equations behind backpropagation and discusses various gradient descent algorithms, including batch, stochastic, and mini-batch methods. The chapter concludes with a discussion on linear activation functions and the universal approximation theorems, providing both intuitive and mathematical insights.

In Chapter 4: Challenges in Neural Network Optimization, we address common issues faced when training neural networks. This chapter explores activation function saturation, vanishing and exploding gradients, and weight initialization techniques. It also discusses non-zero centered activation functions and feature scaling methods. Techniques such as normalization, standardization, and batch normalization are presented to help readers understand how to stabilize and improve the training process.

Chapter 5: Learning Rate Schedules and Adaptive Algorithms focuses on methods to adjust and optimize the learning rate during training. Various dynamic learning rate decay methods, including step decay, inverse time decay, and cyclical learning rates, are covered. The chapter also discusses accelerated gradient descent methods, adaptive learning rate algorithms, and Hessian-based methods such as Newton and Marquardt methods. Conjugate direction methods and quasi-Newton approaches are also presented, providing a comprehensive overview of advanced optimization techniques.

Chapter 6: Strategies for Generalization and Hyper-Parameter Tuning explores techniques to prevent overfitting and ensure that models generalize well to new data. This chapter covers statistical learning theory, point estimation, and the bias-variance trade-off. It also discusses training, testing, and validation sets, including cross-validation techniques. Performance measures for classification and regression models are presented, along with various hyperparameter tuning methods such as grid search, random search, and Bayesian optimization. The chapter concludes with a discussion on Gaussian processes and acquisition functions used in Bayesian optimization.

Chapter 7: Regularization Techniques provides an in-depth look at methods to prevent overfitting and improve model performance. This chapter covers penalty-based regularization methods, including L2 (ridge) and L1 regularization, as well as elastic net regularization. Techniques such as early stopping, ensemble methods, and dropout are also discussed, providing readers with a toolkit of methods to improve model robustness.

Chapter 8: Advanced Activation Functions delves into the diverse world of activation functions used in neural networks. This chapter categorizes activation functions into sigmoid-based, ReLU-based, ELU-based, miscellaneous, non-standard, and combined activation functions. Each type is discussed in detail, highlighting their properties, advantages, and applications. The chapter aims to provide a thorough understanding of the role of activation functions in neural network performance and their impact on model behavior.

Chapter 9: Complex Valued Neural Networks introduces the concept of complex-valued neural networks and their unique properties. This chapter covers complex numbers, functions, and their visualizations, along with complex calculus and backpropagation algorithms. It discusses the properties and types of complex activation functions, providing a comprehensive overview of this advanced topic in neural network research.

Finally, we extend our heartfelt thanks to Professor Mohamed Abdalla Darwish, Head of the Department of Mathematics and Computer Science, Faculty of Science, Damanhour University, Egypt, for his unwavering support. We are profoundly grateful to Professor Amr R. El Dhaba for his invaluable discussions and continued encouragement.

We also wish to express our sincere appreciation to our colleagues and friends for their invaluable feedback, thoughtful comments, and constructive suggestions. In particular, we would like to acknowledge Professor Hamed Awad, Dr. Fatma El-Safty, Dr. Hamdy El Shamy, Dr. Mohamed Elhaddad, Mohamed Yahia, Ayman A. Abdelaziz, Eman Farag, Hassan M. Shetawy, Walaa Mansour, Moaz El-Essawey, Aziza Salah, and Eman R. Hendawy for their contributions.

We hope that you find this book informative and inspiring, and that it serves as a valuable resource in your journey through the world of artificial neural networks and deep learning.

*knowledge itself is power - Sir Francis Bacon 1597*

Egypt 2024 M. M. Hammad

# Bridging Theory and Practice: The Dual Approach of This Book

Many books on neural networks and deep learning tend to lean either heavily on theoretical aspects, with dense mathematical formulations, or focus predominantly on computational algorithms, often at the expense of a solid mathematical foundation. This book adopts a balanced approach that bridges the gap between these extremes. By seamlessly integrating theoretical principles with practical implementation, it empowers learners to fully harness the power of neural networks and deep learning in their pursuit of excellence across various domains.

However, attempting to cover both the theoretical concepts and computational algorithms exhaustively within a single volume would be impractical. To ensure a thorough exploration of both aspects, this book is divided into two complementary parts. The first part is titled "Artificial Neural Network and Deep Learning: Fundamentals and Theory." The second part is titled "Neural Network and Deep Learning with Mathematica." For each theoretical chapter in the first part, there is a corresponding chapter in the second part. We strongly recommend that after completing each theoretical chapter, you explore the corresponding practical implementation chapter in the complementary volume. This dual approach will provide you with a well-rounded understanding and the skills necessary to excel in this field.

The book "Neural Network and Deep Learning with Mathematica" adopts a refreshingly code-centric approach, enabling you to solidify your understanding through hands-on practice. Nearly all the concepts introduced are accompanied by illustrative code examples, making the learning experience both practical and tangible. Even the figures in the first part are generated using these code examples, emphasizing the code-first methodology. To ensure accessibility and ease of understanding, the code examples are deliberately crafted in a simple format, prioritizing readability over efficiency and generality. In line with our instructional philosophy, each code example serves a dual purpose: not only does it demonstrate a specific deep learning concept, but it also simultaneously introduces and reinforces Mathematica programming techniques. Readers will learn how to leverage Mathematica to perform complex neural network and deep learning calculations, simulate data, and create visual representations of their findings.

# CONTENTS

## CHAPTER 4: CHALLENGES IN NEURAL NETWORK OPTIMIZATION

## CHAPTER 5: LEARNING RATE SCHEDULES AND ADAPTIVE ALGORITHMS

**CHAPTER 6: STRATEGIES FOR GENERALIZATION AND HYPER-PARAMETER TUNING**

**CHAPTER 7: REGULARIZATION TECHNIQUES**

**CHAPTER 8: ADVANCED ACTIVATION FUNCTIONS**

# Abbreviations:

| | |
|---|---|
| AI | Artificial Intelligence |
| AD | Automatic Differentiation |
| Adam | Adaptive Moment |
| AF | Activation Function |
| ACF | Acquisition Function |
| ACC | Accuracy |
| ALiSA | Adaptive Linearized Sigmoidal Activation |
| APL | Adaptive Piecewise Linear |
| APTF | Amplitude-Phase-Type Function |
| APSF | Amplitude-Phase Sigmoidal Function |
| BO | Bayesian optimization |
| BP | Back Propagation |
| BCE | Binary Cross-Entropy |
| BN | Batch Normalization |
| BDAA | Bi-modal Derivative Adaptive Activation |
| CDF | Cumulative Distribution Function |
| CBP | Complex Backpropagation |
| CVAF | Complex-Valued Activation Function |
| CVNN | Complex-Valued Neural Network |
| CAP-PLS | Complex Amplitude-Phase Piecewise Linear Scaling |
| CAP-ES | Complex Amplitude-Phase Exponential Scaling |
| CAP-ArcTanS | Complex Amplitude-Phase ArcTan Scaling |
| CAP-Swish | Complex Amplitude-Phase Swish |
| CAP-ELU | Complex Amplitude-Phase Exponential Linear Unit |
| CAP-Softplus | Complex Amplitude-Phase Softplus |
| CAP-ErfA | Complex Amplitude-Phase Erf Attenuation |
| CNN | Convolutional Neural Network |
| DAG | Directed Acyclic Graph |
| DNN | Deep Neural Network |
| DSiLU | Derivative Sigmoid-weighted linear unit |
| DReLU | Displaced Rectified Linear Unit |
| ELU | Exponential Linear Unit |
| EMA | Exponential Moving Averages |
| EI | Expected Improvement |
| EReLU | Elastic Rectified Linear Unit |
| EPReLU | Elastic Parametric Rectified Linear Unit |
| EELU | Elastic Exponential Linear Unit |
| ETF | Elementary Transcendental Functions |
| FFNN | Feed-Forward Neural Network |
| FBGD | Full-Batch Gradient Descent |
| FP | False Positive |
| FN | False Negative |
| FC-Swish | Fully Complex Swish |
| FC-Mish | Fully Complex Mish |
| GD | Gradient Descent |
| GC | Gradient Clipping |
| GP | Gaussian Process |
| GELU | Gaussian Error Linear Unit |
| IQR | Interquartile Range |
| IID | Independent and Identically Distributed |
| LN | Layer Normalization |
| LReLU | Leaky Rectified Linear Unit |
| LOOCV | Leave-One-Out Cross-Validation |
| LiSA | Linearized Sigmoidal Activation |
| LuTU | Look-up Table Unit |
| MAD | Mean Absolute Deviation |
| MLP | Multi-Layer Perceptron |
| MSE | Mean Squared Error |
| MTLU | Multi-bin Trainable Linear Unit |
| MoGU | Mixture of Gaussian Unit |
| MeLU | Mexican Rectified Linear Unit |
| MGF | Moment Generating Function |
| MBSGD | Mini-Batch Stochastic Gradient Descent |
| MLE | Maximum Likelihood Estimation |
| NN | Neural Network |
| Nadam | Nesterov-Accelerated Adaptive Moment |
| NAG | Nesterov Accelerated Gradient |
| PDF | Probability Density Function |
| PReLU | Parametric Rectified Linear Unit |
| PCA | Principal Components Analysis |
| PI | Probability of Improvement |
| PPV | Positive Predictive Value |
| PTanh | Penalized Tanh |
| PELU | Parametric Exponential Linear Unit |
| PMF | Probability Mass Function |
| PREU | Parametric Rectified Exponential Unit |
| RMS | Root Mean Square |
| RV | Random Variable |
| RNN | Recurrent Neural Network |
| ReLU | Rectified Linear Unit |
| RReLU | Randomized Rectified Linear Unit |
| RTReLU | Random Translation Rectified Linear Unit |
| ReLTanh | Rectified Linear Tanh |
| REU | Rectified Exponential Unit |
| RVNN | Real Valued Neural Network |
| SGD | Stochastic Gradient Descent |
| SE | Standard Error |
| SRS | Soft-Root-Sign |
| SiLU | Sigmoid-weighted linear unit |
| SReLU | Shifted Rectified Linear Unit |
| SNN | Self-Normalizing Neural Network |
| SELU | Scaled Exponential Linear Unit |
| SLU | SoftPlus Linear Unit |
| SGELU | Symmetrical Gaussian Error Linear Unit |
| Split-STanh | Split-Sigmoidal Tanh |
| SplitPSigmoid | Split-Parametric Sigmoid |
| Split-QAM | Split-Quadrature Amplitude Modulation |
| TP | True Positive |
| TN | True Negative |
| TPR | True Positive Rate |
| TNR | True Negative Rate |
| UAT | Universal Approximation Theorem |
| UCB | Upper Confidence Bound |
| VGP | Vanishing Gradient Problem |
| ZCA | Zero-Phase Component Analysis |

# CHAPTER 1

# DESCRIPTIVE STATISTICS AND PROBABILITY THEORY

In the realm of data analysis and machine learning, a profound synergy exists between descriptive statistics, probability theory, and neural networks (NNs). Descriptive statistics serves as the cornerstone for summarizing and interpreting data, providing key insights into the central tendencies and variability within a dataset. In this chapter, we will explore how descriptive statistics act as the initial lens through which we scrutinize and make sense of raw data, setting the stage for more advanced analyses. On the other hand, probability theory plays a pivotal role in quantifying uncertainty and randomness inherent in data. Understanding the principles of probability is fundamental to grasping the probabilistic nature of many real-world phenomena. This chapter elucidates the symbiotic relationship between probability theory and descriptive statistics, showcasing how probability distributions and statistical measures intertwine to form a comprehensive framework for data analysis.

In the artificial intelligence (AI) applications, probability theory serves as a cornerstone, influencing our approach in two fundamental ways. Firstly, the laws of probability act as guiding principles, shaping the rationale behind how AI systems should reason. In the design phase, we meticulously construct algorithms to compute or approximate diverse expressions derived from probability theory. This intentional integration enables our AI systems to navigate uncertainty, make informed decisions, and exhibit a capacity for adaptive learning. Secondly, the marriage of probability and statistics offers us a lens through which we can theoretically dissect and analyze the behavior of proposed AI systems. This analytical framework allows us to delve into the intricacies of system performance, understand its robustness under varying conditions, and assess the implications of algorithmic choices. By employing probability and statistics in this manner, we gain valuable insights into the theoretical underpinnings of AI, contributing to the refinement and optimization of these intelligent systems.

This chapter is crafted with a specific purpose: to facilitate comprehension for readers whose expertise lies predominantly in software engineering and who may have limited exposure to probability theory and descriptive statistics. Recognizing the diverse backgrounds of our audience, we aim to create a bridge, ensuring that the material presented in this book becomes accessible and digestible for those with a foundational background in software engineering. If you are already familiar with probability theory and descriptive statistics, you may wish to skip this chapter.

The chapter delves into descriptive statistics, starting with the construction and interpretation of frequency distributions. Visual representations such as histograms and frequency polygons are introduced, along with measures of central tendency (mean, median, mode, …) and measures of dispersion (range, variance, standard deviation, …). Additionally, the discussion extends to measures of symmetry and peakedness, including skewness and kurtosis, which aid in characterizing the shape of distributions. The second major section explores probability theory, beginning with the fundamentals of random experiments and sample spaces. Counting techniques, permutations, and combinations are introduced to address probability problems, followed by interpretations and axioms of probability. The chapter proceeds to examine conditional probability and explores discrete random variables and distributions, including probability mass functions. It further covers continuous random variables, probability density functions, and prominent continuous probability distributions.

This chapter is a summary of my book titled "Statistics for Machine Learning with Mathematica Applications". For detailed proofs of theorems, additional examples, and comprehensive explanations, including Mathematica applications, please refer to Ref [1].

## 1.1 Frequency Distributions, Histogram and Frequency Polygons

**Definitions:**

**Population:** A population is the set of all measurements of interest to the investigator.

**Sample:** A sample is a subset of measurements selected from the population of interest.

**Univariate Data:** Univariate data results when a single variable is measured.

**Bivariate Data:** Bivariate data results when two variables are measured.

**Multivariate Data:** Multivariate data results when more than two variables are measured.

**Discrete Variable:** A discrete variable can assume only a finite or countable number of values.

**Continuous Variable:** A continuous variable can assume the infinitely many values corresponding to the points on a line interval.

**Relative-Frequency Distribution:** A tabular arrangement of data by classes together with the corresponding relative class frequencies is called a relative-frequency distribution, percentage distribution, or relative-frequency table.

Once we obtain the sample data values, one way to become acquainted with them is through data visualization techniques such as displaying them in tables or graphically. These visual displays may reveal the patterns of behavior of the variables being studied. The statistical table is a list of the categories along with a measure of how often each value occurred.

General rules for forming frequency distributions are:

1. Determine the largest and smallest numbers in the raw data and thus find the range (the difference between the largest and smallest numbers).
2. Divide the range into a convenient number of class intervals having the same size.
3. Determine the number of observations falling into each class interval; that is, find the class frequencies.

Karl Pearson [2] coined the term "histogram" to refer to an approximate representation of the distribution of numerical data. To create a histogram, the first step is to divide the range of values into intervals or bins, and then tally how many values fall within each interval. These bins are often non-overlapping and consecutive, and they may have either equal or unequal sizes. The resulting histogram can provide an estimate of the distribution of the data.

We can say that histograms and frequency polygons are two graphic representations of frequency distributions.

1. A histogram consists of a set of rectangles having

- bases on a horizontal axis (the $X$ axis), with centers at the class marks (center of class) and lengths equal to the class interval sizes,
- areas proportional to the class frequencies, see Figure 1.1 (a) and (c).

2. A frequency polygon is a line graph of the class frequencies plotted against class marks. It can be obtained by connecting the midpoints of the tops of the rectangles in the histogram, see Figure 1.1 (b) and (d).

The data used to construct a histogram are generated via a function $m_i$ that counts the number of observations that fall into each of the disjoint categories (bins). Thus, if we let $n$ be the total number of observations and $k$ be the total number of bins, the histogram data $m_i$ meet the following condition:

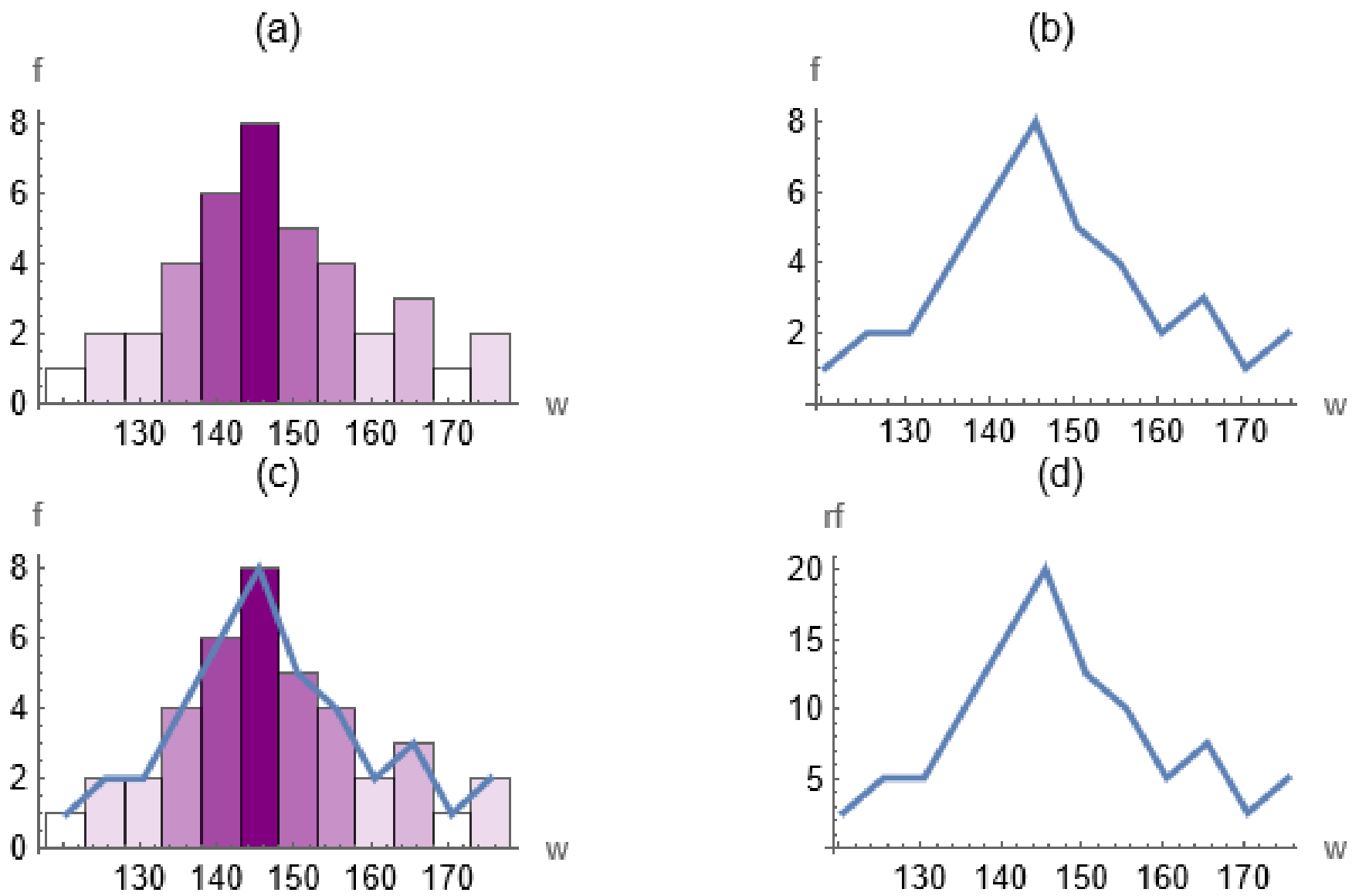

**Figure 1.1.** (a), (b), and (c) are the histogram, frequency polygon and the overlap of histogram and frequency polygon, respectively. However, (d) is the percentage polygon.

$$n = \sum_{i=1}^{k} m_i. \tag{1.1}$$

There is no best number of bins, and different bin sizes can reveal different features of the data. In case a single statistical series of range $R$ with $n$ items is involved in a computation, the optimal class interval may be estimated from the formula [3]:

$$\text{Class interval} = R/(1 + 3.3 \log n)\,, \tag{1.2}$$

or by the square-root choice formula [1]:

$$\text{number of bins} = \lceil \sqrt{n} \rceil, \tag{1.3}$$

which takes the square root of the number of data points in the sample and rounds to the next integer.

**Remarks:**

- Graphic representation of relative-frequency distributions can be obtained from the histogram or frequency polygon simply by changing the vertical scale from frequency to relative frequency, keeping exactly the same diagram. The resulting graphs are called relative-frequency histograms (or percentage histograms) and relative-frequency polygons (or percentage polygons), respectively.
- From the histogram we should be able to identify the center (i.e., the location) of the data, spread of the data, skewness of the data, presence of outliers, presence of multiple modes in the data, and whether the data can be capped with a bell-shaped curve. These properties provide indications of the proper distributional model for the data.
- A sample histogram provides valuable information about the population histogram, the graph that describes the distribution of the entire population. Remember, though, that different samples from the same population will produce different histograms, even if you use the same class boundaries. However, you can expect that the sample and population histograms will be similar. As you add more data to the sample, the two histograms become more alike.

Descriptive statistics are a set of statistical techniques that are used to summarize and describe the main features of a dataset. Descriptive statistics break down into several types (measures of central tendency, measures of dispersion, and measures of symmetry).

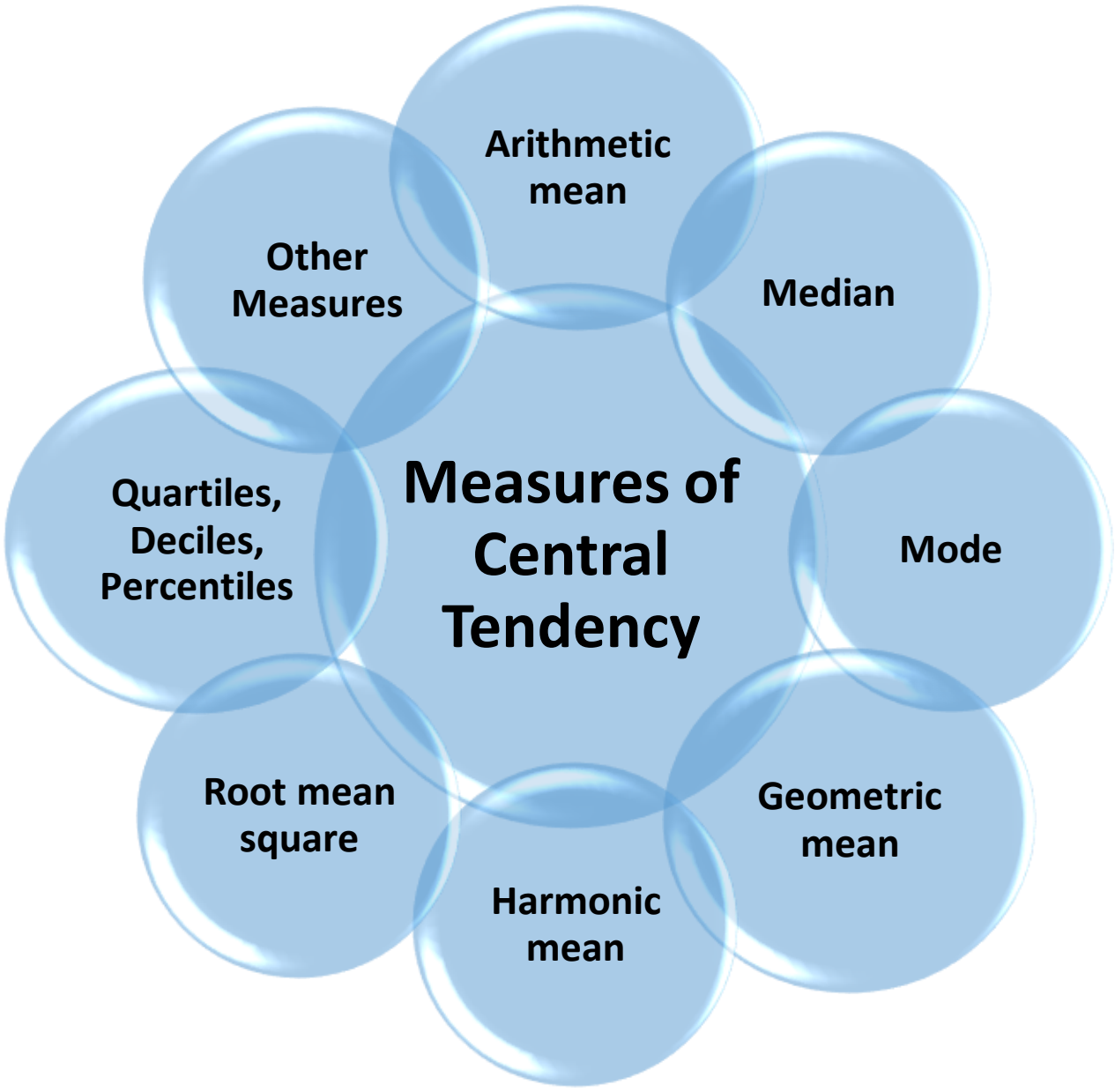

## 1.2 Measures of Central Tendency

An average is a value that is typical, or representative, of a set of data. Since such typical values tend to lie centrally within a set of data arranged according to magnitude, averages are also called measures of central tendency. Several types of measures of central tendency can be defined, such as the arithmetic mean, median, mode, geometric mean, harmonic mean, root mean square, trimmed mean, winsorized mean, quartiles, deciles, and percentiles. Each has advantages and disadvantages, depending on the data and the intended purpose.

**Definition (The Arithmetic Mean):** The arithmetic mean, or briefly the mean, of a set of $N$ numbers $v_1, v_2, v_3, \ldots, v_N$ is denoted by $\bar{v}$ and is defined as [4]:

$$\bar{v} = \frac{v_1 + v_2 + \cdots + v_N}{N} = \frac{\sum_{j=1}^{N} v_j}{N}. \tag{1.4}$$

For instance, the arithmetic mean of the numbers in the set

$$\begin{aligned} h = \{&133, 136, 149, 133, 123, 121, 140, 139, 117, 117, 136, 108, 126, \\ &104, 116, 147, 140, 148, 150, 122, 135, 146, 133, 144, 117, 124, 135, \\ &117, 120, 121, 110, 124, 103, 137, 101, 119, 104, 113, 139, 133\}, \end{aligned} \tag{1.5}$$

is $\bar{v} = 127$. The mean is useful because it shows where the "center of gravity" exists for an observed set of values (see Figure 1.2).

If the numbers $v_1, v_2, v_3, \ldots, v_K$ occur with frequencies $f_1, f_2, f_3, \ldots, f_K$, (grouped data), the arithmetic mean is [5]

$$\begin{aligned} \bar{v} &= \frac{f_1 v_1 + f_2 v_2 + \cdots + f_K v_K}{f_1 + f_2 + \cdots + f_K} \\ &= \frac{\sum_{j=1}^{K} f_j v_j}{\sum_{j=1}^{K} f_j} = \frac{\sum_{j=1}^{K} f_j v_j}{N}, \end{aligned} \tag{1.6}$$

where $\sum_{j=1}^{K} f_j = N$ is the total frequency. For instance, if 5, 8, 6, and 2 occur with frequencies 3, 2, 4, and 1, respectively, the arithmetic mean is $\bar{v} = \frac{(5)(3)+(8)(2)+(6)(4)+(2)(1)}{3+2+4+1} = 5.7$.

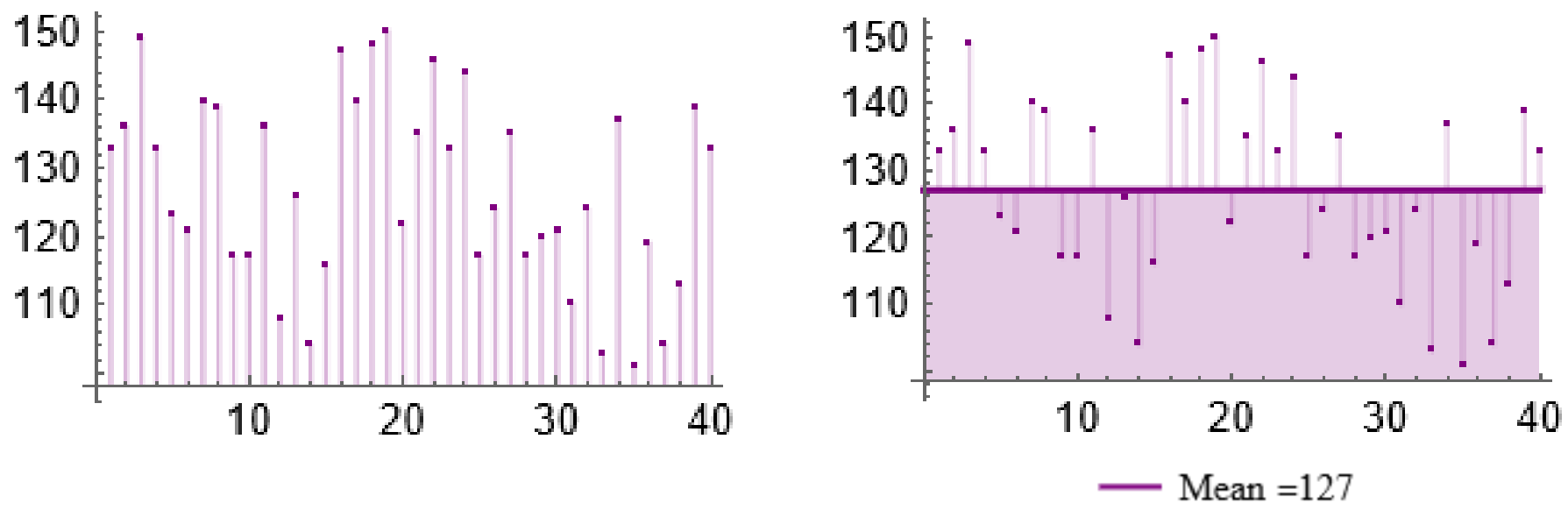

**Figure 1.2.** Left panel: The plot displays the values in the set $h$ (1.5) as a list plot with the area under the points filled in to the axis. Right panel: The plot displays the same set of data, $h$, and a horizontal line at the mean value $\bar{v} = 127$. In the right plot, the data points display as a list plot filled up to the mean value to explain how the values in $h$ are distributed around the mean.

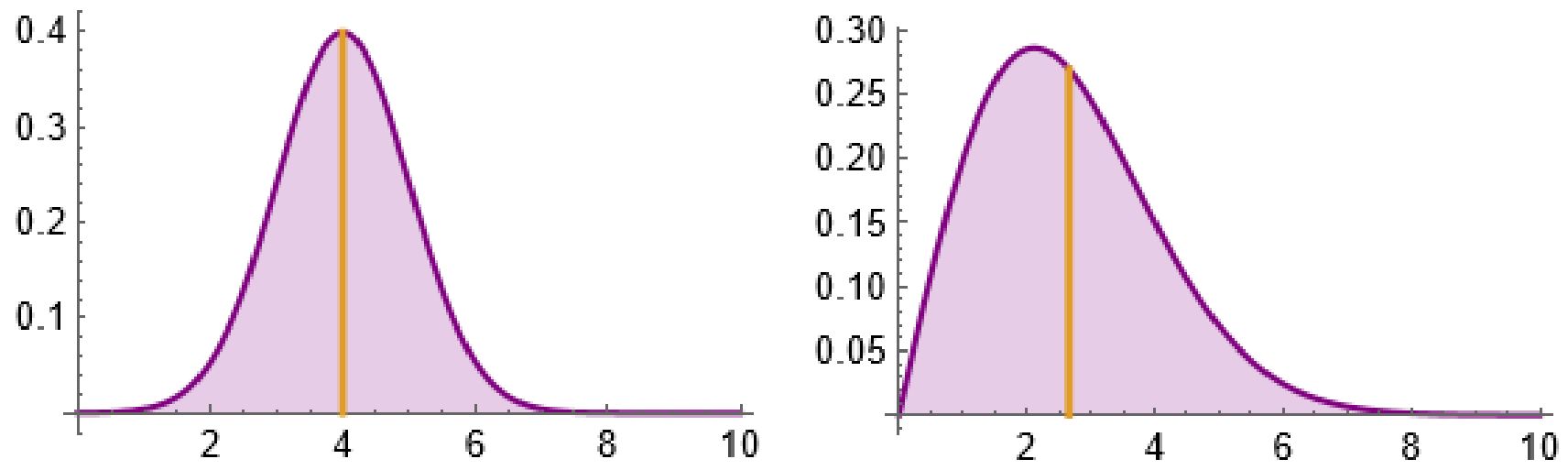

**Figure 1.3.** The position of the mean for symmetric and right-skewed frequency curve.

Sometimes we associate with the numbers $v_1, v_2, v_3, \ldots, v_K$ certain weighting factors (or weights) $w_1, w_2, \ldots, w_K$, depending on the significance or importance attached to the numbers. This can be helpful when we want some values to contribute to the mean more than others. A common example of this is weighting academic exams to give a final grade. If you have three exams and a final exam, and we give each of the three exams 20% weight and the final exam 40% weight of the final grade. The mean for a set of weighted numbers is given by [6],

$$\bar{v} = \frac{w_1 v_1 + w_2 v_2 + \cdots + w_K v_K}{w_1 + w_2 + \cdots + w_K} = \frac{\sum_{j=1}^{K} w_j v_j}{\sum_{j=1}^{K} w_j}, \tag{1.7}$$

and is called the weighted arithmetic mean. For instance, if a final examination in a course is weighted 3 times as much as a quiz and a student has a final examination grade of 85 and quiz grades of 70 and 90, the mean grade is $\bar{v} = \frac{(70)(1)+(90)(1)+(85)(3)}{1+1+3} = 83$.

Figure 1.3 shows the position of the mean for the symmetric and skewed to the right frequency curves.

## Median

**Definition (The Median):** The median of a set of numbers arranged in order of magnitude (i.e., in an array) is either the middle value or the arithmetic mean of the two middle values. Sometimes, the median denotes by $\tilde{v}$.

Loosely speaking, order the values of a data set of size $n$ from smallest to largest. If $n$ is odd, the sample median is the value in position $(n+1)/2$; if $n$ is even, it is the average of the values in positions $n/2$ and $n/2+1$ (see Figure 1.4). For instance, the set of numbers 2, 5, 6, 9, and 11 has a median 6, (rank the $n = 5$ measurements from smallest to largest: 2, 5, 6, 9, 11), however, the set of numbers 2, 9, 11, 5, 6 and 27 has a median $\frac{1}{2}(6+9) = 7.5$, (rank the measurements from smallest to largest: 2, 5, 6, 9, 11, 27) .

The relationship between the median and mean can provide valuable information about the shape of a frequency distribution. Specifically, the positioning of the median and mean can provide insight into the skewness of the distribution. In a symmetric frequency curve, where the data is evenly distributed around the center, the median and mean will be located at the same point. This is because the median represents the center of the data, and the mean is calculated as the sum of all the data points divided by the number of data points, resulting in a value that is also at the center of the data. In contrast, for a right-skewed frequency distribution, the mean will be greater than the median. This occurs because the long tail of the distribution is pulling the mean in that direction, while the median is less affected by extreme values. In this case, the median is a better measure of central tendency than the mean, as it is less influenced by outliers. Figure 1.5 shows the relative positions of the mean and median for the symmetric and skewed to the right frequency curves. For symmetrical curves, the mean and median coincide.

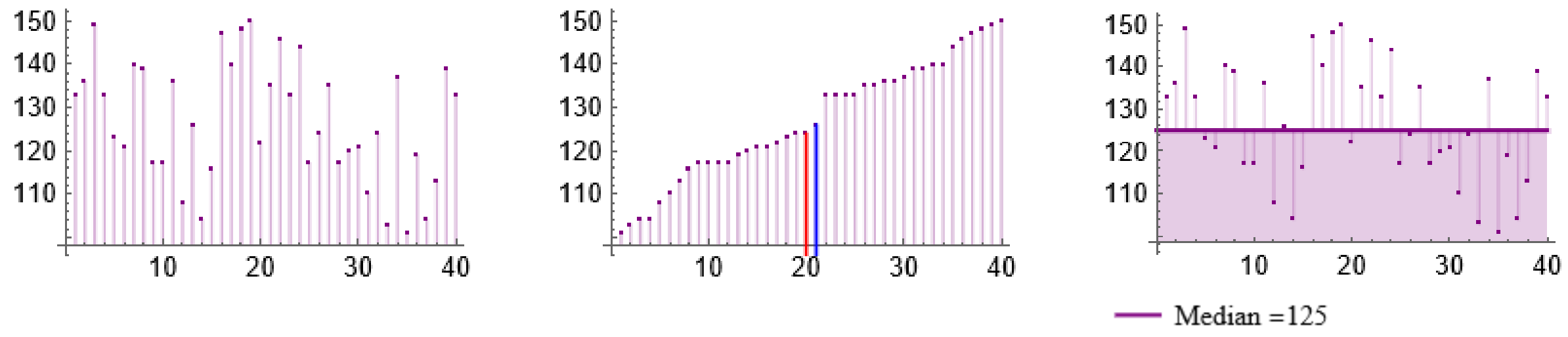

**Figure 1.4.** Left panel: The plot displays the values in the set $h$ (1.5) as a list plot with the area under the points filled in to the axis. Medial panel: The plot displays the same data set, $h$, afer sorting the data points from smallest to largest. The red and blue lines represent positions $n/2 = 20$ and $n/2 + 1 = 21$, respectively. Right panel: The plot displays the same set of data, $h$, and a horizontal line at the median value $\tilde{v} = 125$. In the right plot, the data points display as a list plot filled up to the median value to explain how the values in $h$ are distributed around the median.

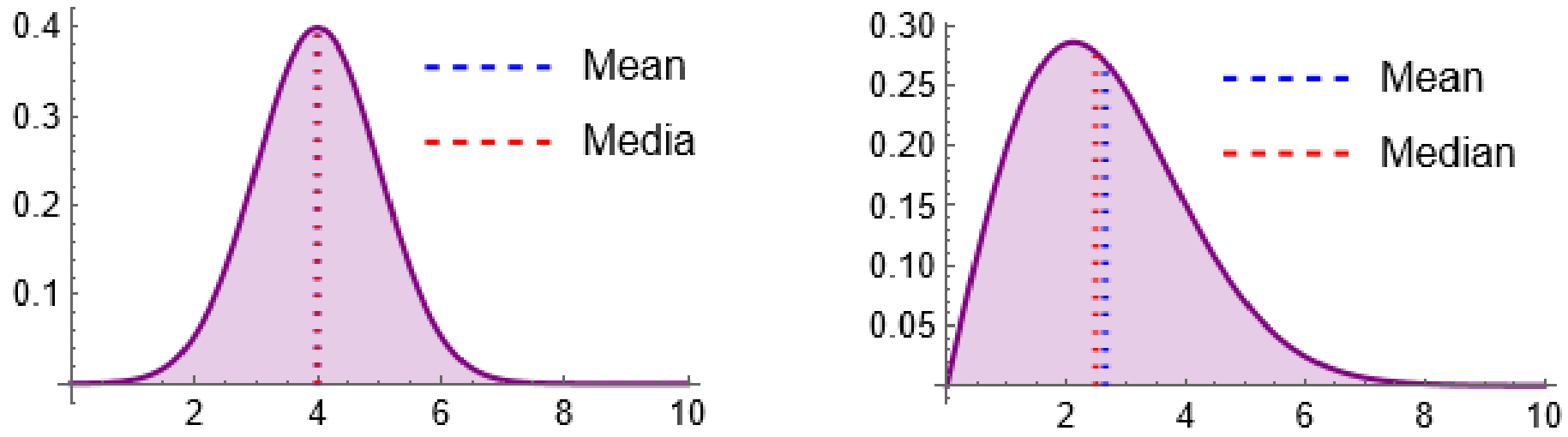

**Figure 1.5.** Relative positions of median and mean for symmetric and right-skewed frequency curve.

### Mode

**Definition (The Mode):** The mode of a set of numbers is that value which occurs with the greatest frequency.

For instance, the set of numbers 2, 4, 7, 8, 8, 8, 10, 10, 11, 12 and 18 has mode 8, however, the set of numbers 2, 3, 8, 11, 12, 14, and 16 has no mode. Set 1, 2, 3, 3, 3, 5, 5, 8, 8, 8, and 9 has two modes, 3 and 8.

**Definition (The Empirical Relation):** For unimodal frequency curves that are moderately skewed (asymmetrical), we have the empirical relation [5],

$$\text{Mean} - \text{mode} = 3(\text{mean} - \text{median}). \tag{1.8}$$

### Geometric Mean

The geometric mean is a mean or average that uses the product of the values of a finite set of real numbers to indicate a central tendency (as opposed to the arithmetic mean, which uses the sum of the values).

**Definition (The Geometric Mean $G$):** The geometric mean $G$ of a set of $N$ positive numbers $v_1, v_2, v_3, \ldots, v_N$ is the $N$th root of the product of the numbers [7]:

$$G = \sqrt[N]{v_1 v_2 v_3 \ldots v_N}. \tag{1.9}$$

For instance, the geometric mean of the numbers 3, 9, and 27 is $G = \sqrt[3]{(3)(9)(27)} = 9$.

**Theorem 1.1:** The geometric mean $G$ of a set of $N$ positive numbers $v_1, v_2, v_3, \dots, v_N$ can also be expressed as the exponential of the arithmetic mean of logarithms

$$G = e^{\frac{1}{N}\sum_{i=}^{N} \ln v_i}. \tag{1.10}$$

### Harmonic Mean

**Definition (The Harmonic Mean *H*):** The harmonic mean $H$ of a set of $N$ numbers $v_1, v_2, v_3, \dots, v_N$ is the reciprocal of the arithmetic mean of the reciprocals of the numbers [7]:

$$H = \frac{1}{\frac{1}{N}\sum_{j=1}^{N} \frac{1}{v_j}}. \tag{1.11}$$

For instance, the harmonic mean of the numbers 2, 4, and 8 is $H = \frac{3}{\frac{1}{2}+\frac{1}{4}+\frac{1}{8}} = 3.43$.

### Root Mean Square

**Definition (The Root Mean Square (RMS) or Quadratic Mean):** The RMS is calculated by taking the square root of the average of the squares of a set of values. The RMS of a set of numbers $v_1, v_2, v_3, \dots, v_N$ is defined by [6]:

$$\text{RMS} = \sqrt{\overline{v^2}} = \sqrt{\frac{\sum_{j=1}^{N} v_j^2}{N}}. \tag{1.12}$$

For instance, the RMS of the set 1, 3, 4, 5, and 7 is $\text{RMS} = \sqrt{\frac{1^2+3^2+4^2+5^2+7^2}{5}} = 4.47$.

### Truncated Mean or Trimmed Mean

**Definition (The Trimmed Mean):** Trimmed mean [8,9] is a statistical measure that is calculated by first removing a certain percentage of the largest and smallest values from a dataset and then taking the arithmetic mean of the remaining values. The percentage of values that are trimmed is typically between 5% and 25%.

### Winsorized Mean

**Definition (The Winsorized Mean):** The Winsorized mean [9] is a statistical method used to reduce the influence of extreme values (outliers) on the calculation of the mean. Instead of completely removing the outliers, the Winsorized mean replaces them with the nearest non-outlying value.

### Quartiles, Deciles, and percentiles

- If a set of data is arranged in order of magnitude, the middle value (or arithmetic mean of the two middle values) that divides the set into two equal parts is the median.
- By extending this idea, quartiles, deciles, and percentiles divide a dataset into equal parts based on their rank or order.
- Quartiles divide a dataset into four equal parts, where the first quartile ($Q_1$) represents the 25th percentile, the second quartile ($Q_2$) represents the 50th percentile (the median), and the third quartile ($Q_3$) represents the 75th percentile. Quartiles are useful in describing the spread of a dataset and identifying outliers (see Figure 1.6).
- Deciles divide a dataset into ten equal parts, where the first decile represents the 10th percentile, the second decile represents the 20th percentile, and so on. Deciles are denoted by $D_1, D_2, \dots, D_9$ (see Figure 1.7).

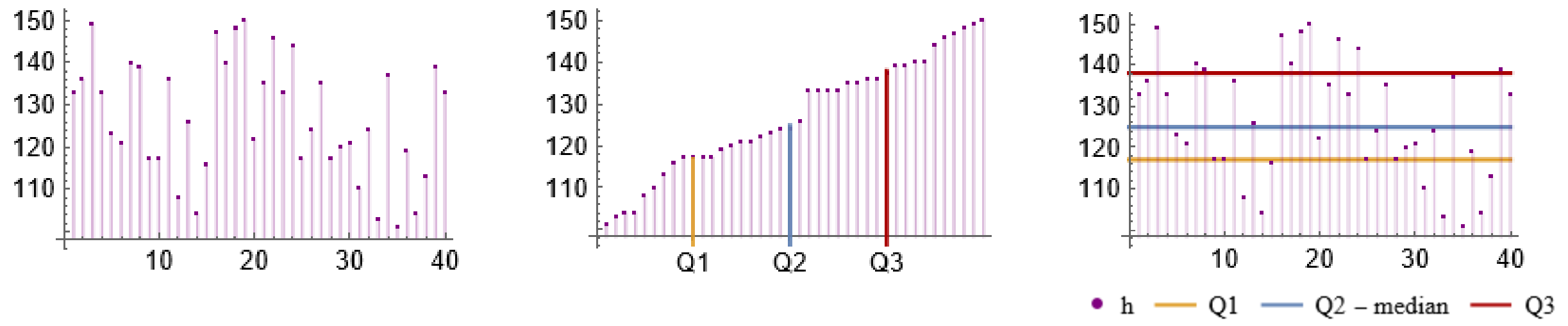

**Figure 1.6.** Left panel: The plot displays the values in the set $h$ (1.5) as a list plot with the area under the points filled in to the axis. Medial panel: The plot displays the same data set, $h$, after sorting the data points from smallest to largest. The $Q1$, $Q2$, and $Q3$ lines represent positions 25%, 50%, and 75% of data, respectively. Right panel: The plot displays the same set of data, $h$, and horizontal lines at the quartiles $Q1 = 117$, $Q2 = 125$, and $Q3 = 138$.

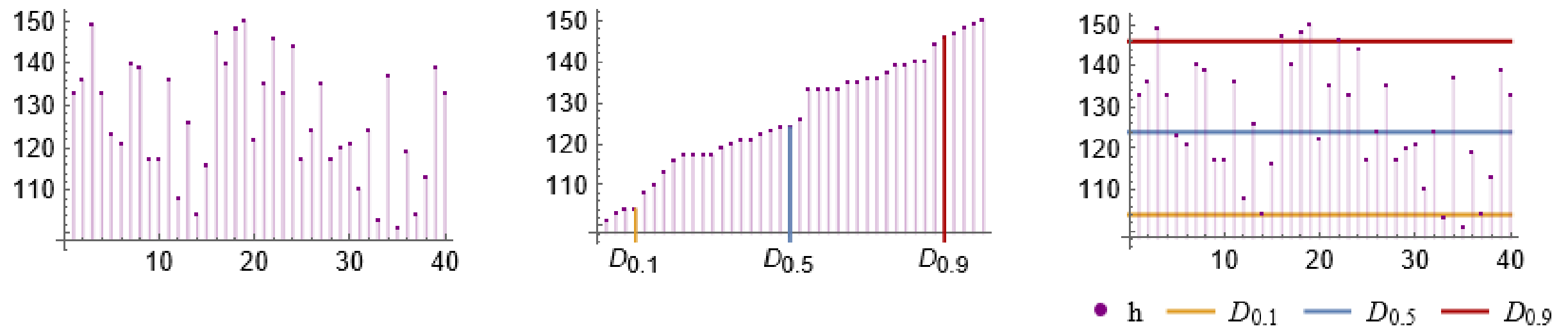

**Figure 1.7.** Left panel: The plot displays the values in the set $h$ (1.5) as a list plot with the area under the points filled in to the axis. Medial panel: The plot displays the same data set, $h$, after sorting the data points from smallest to largest. The $D_{0.1}$, $D_{0.5}$ and $D_{0.9}$ lines represent positions 10%, 50%, and 90% of the data, respectively. Right panel: The plot displays the same set of data, $h$, and horizontal lines at the deciles $D_{0.1} = 104$, $D_{0.5} = 124$ and $D_{0.9} = 146$.

- Percentiles divide a dataset into 100 equal parts, where the first percentile represents the smallest value, and the 100th percentile represents the largest value. For example, the 75th percentile represents the value below which 75% of the data falls. Percentiles are denoted by $P_1, P_2, \ldots, P_{99}$.
- The fifth decile and the 50th percentile correspond to the median.
- The 25th and 75th percentiles correspond to the first and third quartiles, respectively.
- Collectively, quartiles, deciles, percentiles, and other values obtained by equal subdivisions of the data are called quantiles.

An outlier may result from transposing digits when recording a measurement, from incorrectly reading an instrument dial, from a broken piece of equipment, or from other problems. Even when there are no recording errors, a data set may contain one or more measurements that, for one reason or another, are very different from the others in the set. These outliers can cause a distortion in commonly used numerical measures such as mean and standard deviation. In fact, outliers may themselves contain important information not shared with the other measurements in the set. Therefore, isolating outliers, if they are present, is an important first step in analyzing a data set. The box plot and five-number summary are designed exactly for this purpose.

The median and the upper and lower quartiles divide the data into four sets, each containing an equal number of measurements. If we add the largest number (Max) and the smallest number (Min) in the data set to this group, we will have a set of numbers that give insight into the spread, central tendency, and outliers in the data.

The five-number summary consists of the following numerical measures:

$$(\text{Min}, Q_1, \text{Median}, Q_3, \text{Max}).$$

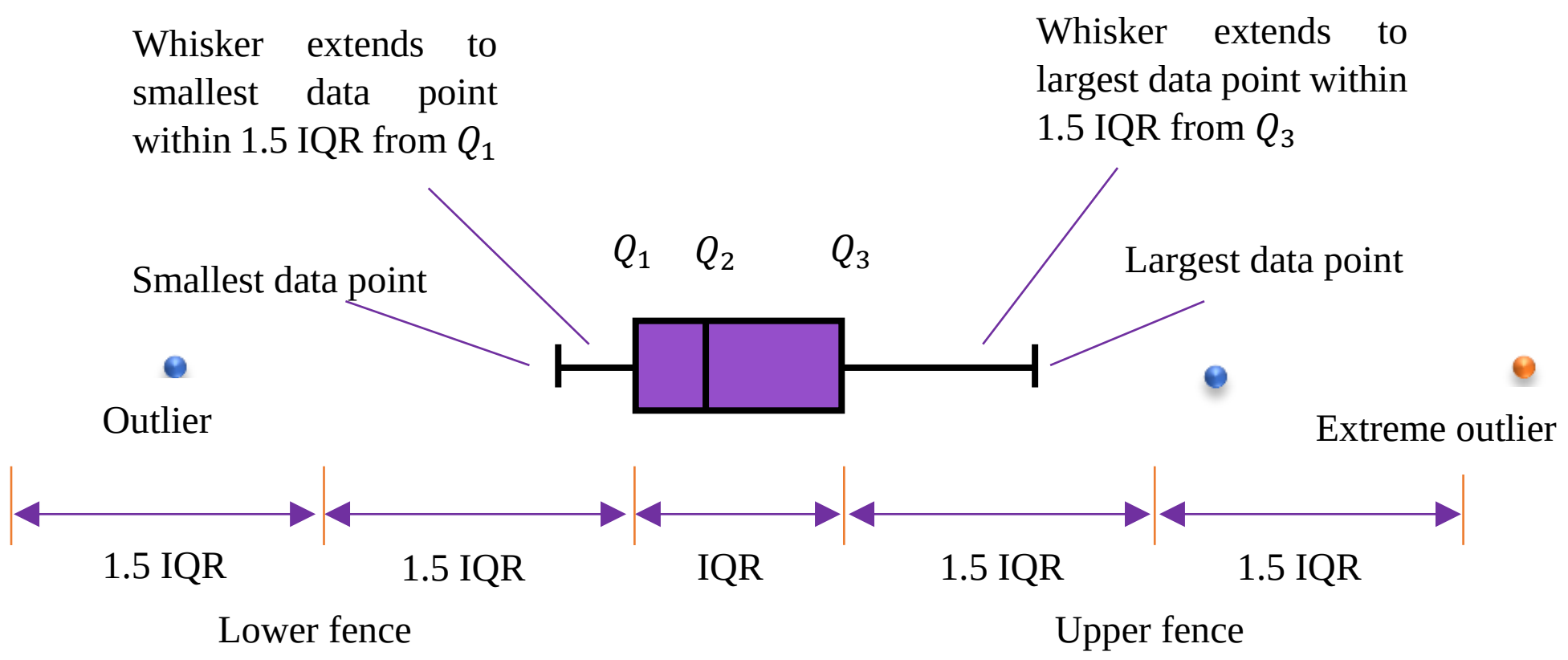

**Figure 1.8.** Description of a box plot.

- The median ($Q_2$) represents the center of the dataset. It divides the data into two equal halves, with 50% of the observations below and 50% above it. It is robust to outliers.
- The range, calculated as the difference between the maximum and minimum values, gives an idea of the overall spread of the data.
- The interquartile range (IQR), calculated as the difference between the third and first quartiles ($Q_3 - Q_1$), represents the spread of the middle 50% of the data and is resistant to outliers.

The box plot, also known as the box-and-whisker plot [8], is a graphical representation of the five-number summary. A box plot is a useful tool for identifying outliers, comparing distributions, and identifying trends in data.

- The median is represented by the vertical line in the box plot.
- The first quartile ($Q_1$) is represented by the left end of the box.
- The third quartile ($Q_3$) is represented by the right end of the box.
- The interquartile range (IQR) is the distance between $Q_1$ and $Q_3$. It is a measure of the spread of data.
- The whiskers extend from the box to the minimum and maximum values.

### Detecting Outliers

Outliers are observations that are beyond the:
- Lower fence: $Q_1 - 1.5$ (IQR).
- Upper fence: $Q_3 + 1.5$ (IQR).

The upper and lower fences are shown in Figure 1.8, but they are not usually drawn on the box plot. Any measurement beyond the upper or lower fence is an outlier; the rest of the measurements, inside the fences, are not unusual. Finally, the box plot marks the range of the data set using "whiskers" to connect the smallest and largest measurements (excluding outliers) to the box.

## 1.3 Measures of Dispersion

### Range

**Definition (The Range):** The range of a set of numbers is the difference between the largest and smallest numbers in the set.

For instance, the range of the set $1, 3, 3, 5, 5, 5, 8, 10, 14$ is $14 - 1 = 13$.

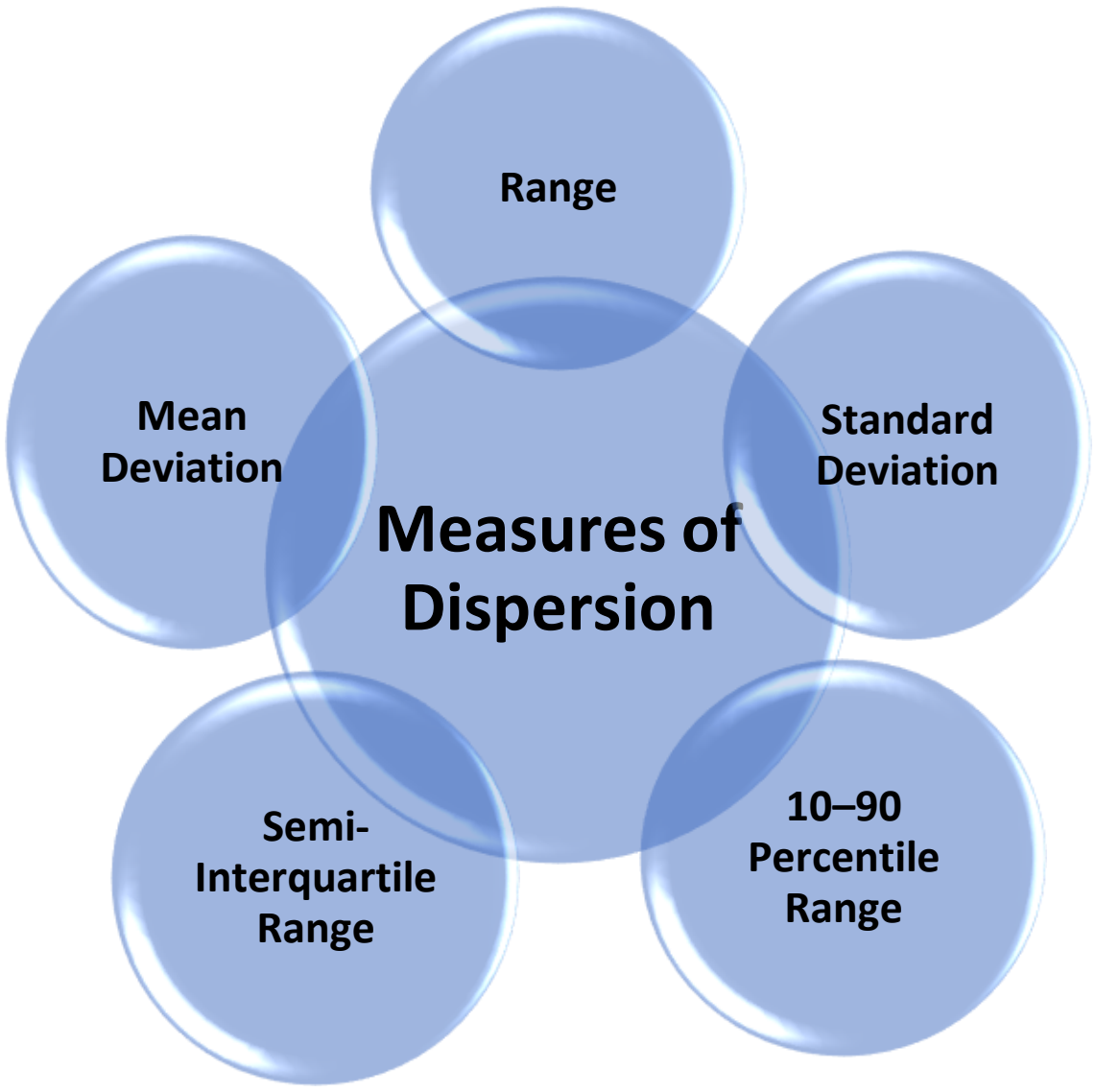

### 10–90 Percentile Range

**Definition (The 10–90 Percentile Range):** The 10–90 percentile range of a set of data is defined by [6]:

$$10 - 90 \text{ percentile range} = P_{90} - P_{10}, \tag{1.13}$$

where $P_{10}$ and $P_{90}$ are the 10th and 90th percentiles for the data.

Let us consider the sets (a) 12, 6, 7, 3, 15, 10, 18, 5 and (b) 9, 3, 8, 8, 9, 8, 9, 18. In both cases, range=(largest number -smallest number)= $18 - 3 = 15$. However, as seen from the arrays of sets (a) and (b),

(a) 3, 5, 6, 7, 10, 12, 15, 18 (b) 3, 8, 8, 8, 9, 9, 9, 18,

there is much more variation, or dispersion, in (a) than in (b). In fact, (b) consists mainly of 8's and 9's. Since the range indicates no difference between the sets, it is not a very good measure of dispersion in this case (where extreme values are present, the range is generally a poor measure of dispersion). An improvement is achieved by throwing out the extreme cases, 3 and 18. Then for set (a) the range is $15 - 5 = 10$, while for set (b) the range is $9 - 8 = 1$, clearly showing that (a) has greater dispersion than (b). However, this is not the way the range is defined. The 10–90 percentile range were designed to improve on the range by eliminating extreme cases.

### Semi-Interquartile Range

**Definition (The Semi-Interquartile Range):** The semi-interquartile range, or quartile deviation, of a set of data is denoted by $Q$ and is defined by [6,10]:

$$Q = \frac{Q_3 - Q_1}{2}, \tag{1.14}$$

where $Q_1$ and $Q_3$ are the first and third quartiles for the data.

For instance, let us consider the semi-interquartile range of the set 6, 47, 49, 15, 43, 41, 7, 39, 43, 41, 36 (or 6, 7, 15, 36, 39, 41, 41, 43, 43, 47, 49). The rank of the median is 6, which means there are five points on each side. Then we need to split the lower half of the data in two again to find the lower quartile. The lower quartile will be the point of rank $(5 + 1)/2 = 3$. The result is $Q_1 = 15$. The second half must also be split in two to find the value of the upper quartile. The rank of the upper quartile will be $6 + 3 = 9$. So $Q_3 = 43$. The interquartile range will be $Q_3 - Q_1 = 28$. The semi-interquartile range is 14 and the range is $49 - 6 = 43$.

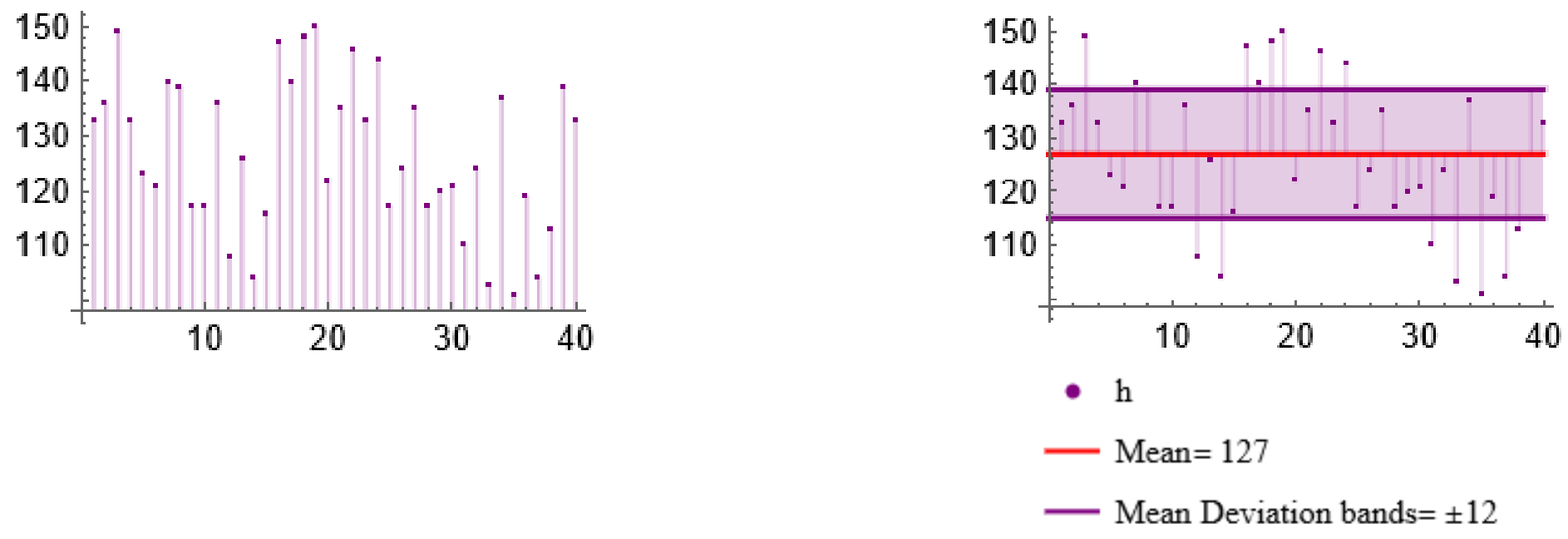

**Figure 1.9.** Left panel: The plot displays the values in the set $h$ (1.5) as a list plot with the area under the points filled in to the axis. Right panel: The plot displays the same set of data, $h$, and horizontal lines at the mean value $\bar{v} = 127$ and mean absolute deviation band. In the right plot, the data points display as a list plot filled up to the mean value to explain how the values in $h$ are distributed around the mean.

### Mean Absolute Deviation

**Definition (The Mean Absolute Deviation):** The mean absolute deviation, or average deviation, of a set of $N$ numbers $v_1, v_2, v_3, \dots, v_N$ is abbreviated (MAD) and is defined by [11]:

$$\text{Mean deviation (MAD)} = \frac{\sum_{j=1}^{N} |v_j - \bar{v}|}{N}, \tag{1.15}$$

where $\bar{v}$ is the arithmetic mean of the numbers and $|v_j - \bar{v}|$ is the absolute value of the deviation of $v_j$ from $\bar{v}$. Simply, it is the average distance between each data point and the mean of the dataset (see Figure 1.9).

For instance, the mean deviation of the set 2, 3, 6, 8, 11 is $\frac{|2-6|+|3-6|+|6-6|+|8-6|+|11-6|}{5} = 2.8$, where we use the mean $\bar{v} = \frac{2+3+6+8+11}{5} = 6$.

### Standard Deviation

In contrast to the range, the standard deviation considers all the observations. Roughly speaking, the standard deviation measures variation by indicating how far, on average, the observations are from the mean. For a data set with a large amount of variation, the observations will, on average, be far from the mean; so the standard deviation will be large. For a data set with a small amount of variation, the observations will, on average, be close to the mean; so the standard deviation will be small.

**Definition (The Standard Deviation of a Population):** The standard deviation of a set of $N$ numbers $v_1, v_2, v_3, \dots, v_N$ is denoted by $S$ and is defined by [5]:

$$S = \sqrt{\frac{\sum_{j=1}^{N} (v_j - \bar{v})^2}{N}}. \tag{1.16}$$

Thus $S$ is sometimes called, the root-mean-square deviation (see Figure 1.10).

Sometimes the standard deviation of a sample's data is defined with $N - 1$ replacing $N$ in the denominators of the expression in (1.16) because the resulting value represents a better estimate of the standard deviation of a population from which the sample is taken. For large values of $N$ (certainly $N > 30$), there is practically no difference between the two definitions.

**Definition (The Standard Deviation for Sample):** The standard deviation of a set of $N$ numbers $v_1, v_2, v_3, \dots, v_N$ is denoted by $S$ and is defined by [10,12]:

$$S = \sqrt{\frac{\sum_{j=1}^{N} (v_j - \bar{v})^2}{N - 1}}. \tag{1.17}$$

Thus $S$ is sometimes called, the root-mean-square deviation.

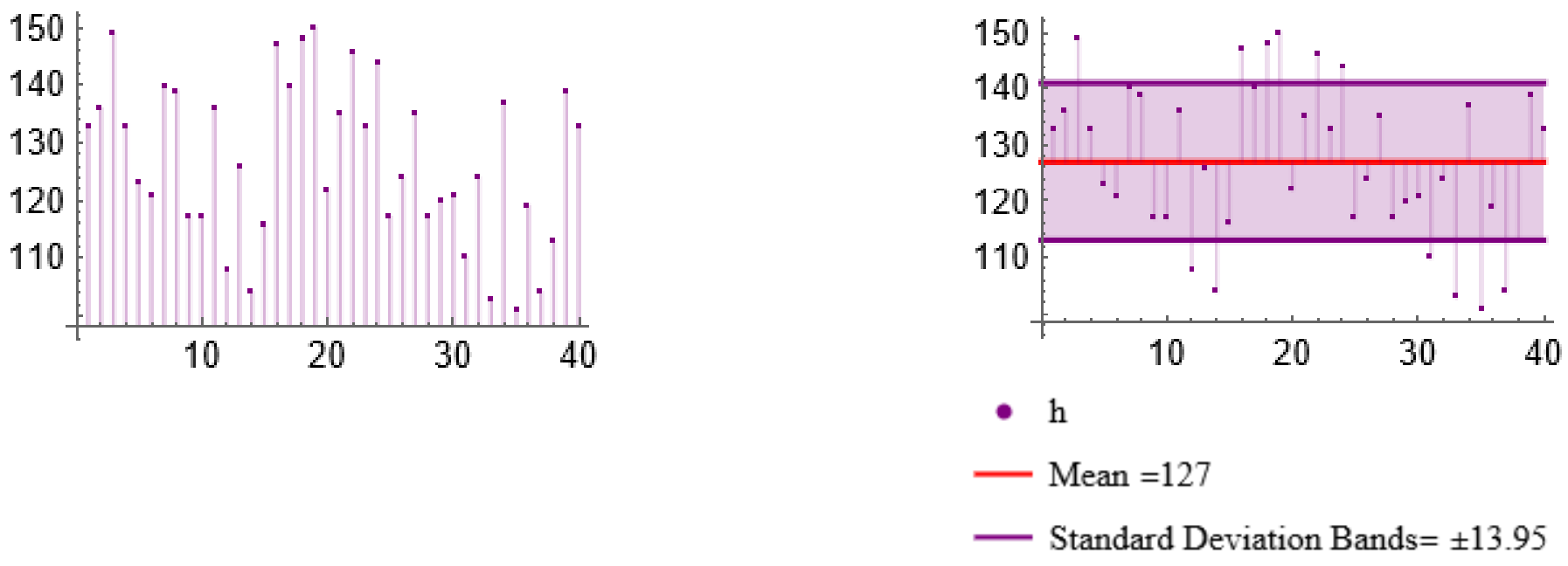

**Figure 1.10.** Left panel: The plot displays the values in the set $h$ (1.5) as a list plot with the area under the points filled in to the axis. Right panel: The plot displays the same set of data, $h$, and horizontal lines at the mean value $\bar{v} = 127$ and standrad deviation band. In the right plot, the data points display as a list plot filled up to the mean value to explain how the values in $h$ are distributed around the mean.

## Variance

Variance is a measure of the spread of data points around the mean. Variance is a useful tool for understanding and comparing data sets. For example, if you have two data sets with the same mean, but different variances, you can tell that the data points in one data set are more spread out than the data points in the other data set.

**Definition (The Variance):** The variance of a set of data is defined as the square of the standard deviation and is thus given by $S^2$ in (1.17).

For instance, the variances of the data set 3, 4, 6, 7, 10 is 6 (the mean is $(3+4+6+7+10)/5 = 6$ and $S^2 = ((-3)^2 + (-2)^2 + (0)^2 + (1)^2 + (4)^2))/5 = 6$).

When it is necessary to distinguish the standard deviation of a population from the standard deviation of a sample drawn from this population, we often use the symbol $S$ for the latter and $\sigma$ for the former. Thus $S^2$ and $\sigma^2$ would represent the sample variance and population variance, respectively.

**Theorem 1.2:** The standard deviation of a set of $N$ numbers $v_1, v_2, v_3, \dots, v_N$ is defined by [6]:

$$S = \sqrt{\frac{\sum_{j=1}^{N} v_j^2}{N} - \left(\frac{\sum_{j=1}^{N} v_j}{N}\right)^2} = \sqrt{\overline{v^2} - \bar{v}^2}, \tag{1.18}$$

where $\overline{v^2}$ denotes the mean of the squares of the various values of $v$, while $\bar{v}^2$ denotes the square of the mean of the various values of $v$.

## Trimmed and Winsorized Variance

One limitation of variance is that it can be affected by outliers or extreme values in the dataset. In such cases, alternative measures such as trimmed variance or winsorized variance may be more appropriate.

**Definition (The Trimmed Variance):** In trimmed variance [9], a certain percentage of the data points at the very top and very bottom of the distribution are trimmed or removed. For example, a 10% trimmed variance would remove the top 10% and bottom 10% of the data points.

**Definition (The Winsorized Variance):** In Winsorized variance [9], a certain percentage of the data points at the very top and very bottom of the distribution are replaced by the value of the nearest data point that is not trimmed. For example, a 10% Winsorized variance would replace the top 10% and bottom 10% of the data points with the value of the 10th and 90th percentiles, respectively.

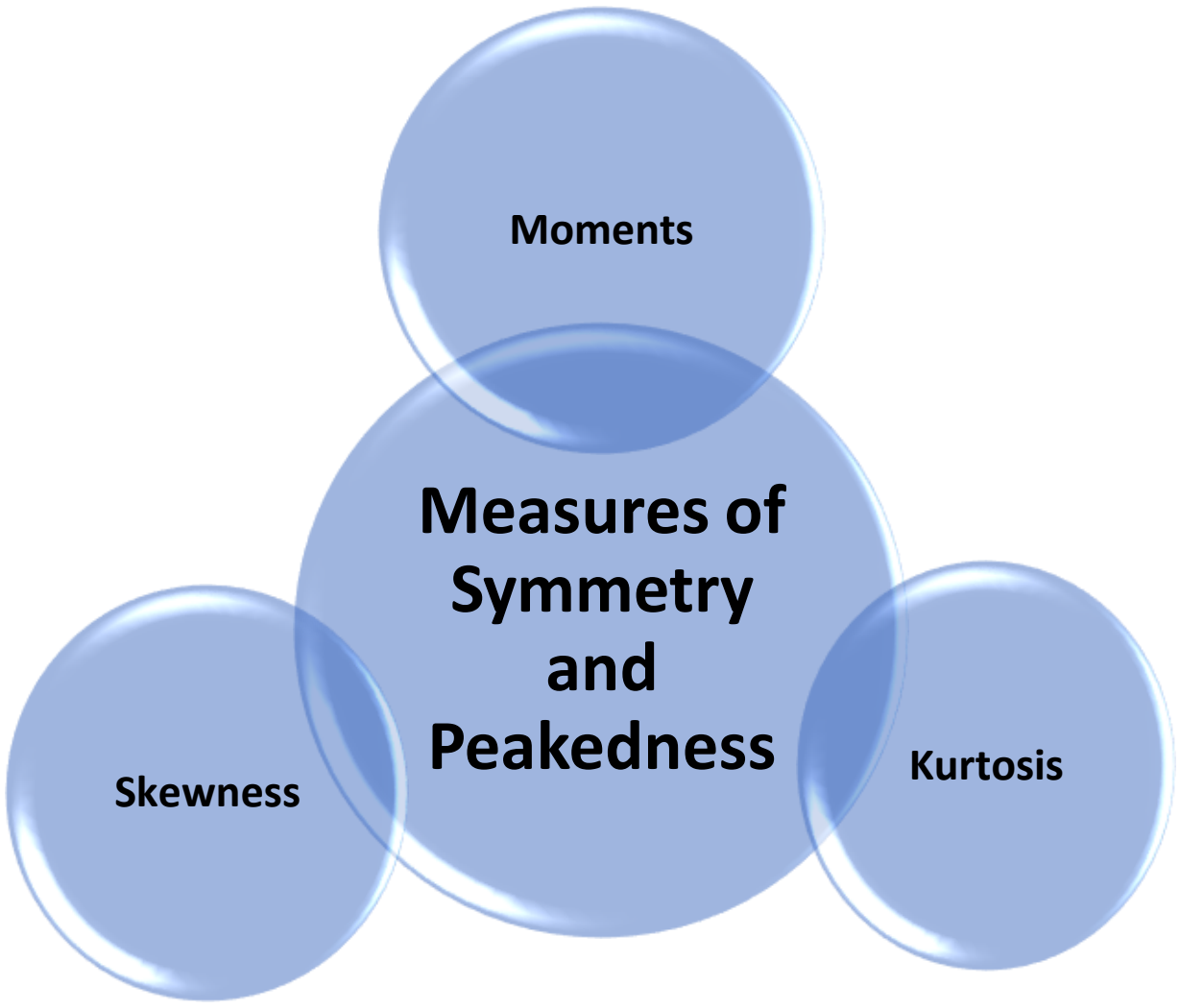

## 1.4 Measures of Symmetry and Peakedness

### Moment

**Definition (The Moment):** If $v_1, v_2, v_3, \dots, v_N$ are the $N$ values assumed by the variable $v$, we define the quantity,

$$\overline{v^r} = \frac{v_1^r + v_2^r + \cdots + v_N^r}{N} = \frac{\sum_{j=1}^{N} v_j^r}{N}, \tag{1.19}$$

called the $r$th moment. The first moment with $r = 1$ is the arithmetic mean $\bar{v}$.

The $r$th moment about the mean $\bar{v}$ is defined as [5]:

$$m_r = \frac{\sum_{j=1}^{N} (v_j - \bar{v})^r}{N}. \tag{1.20}$$

If $r = 1$, then $m_1 = 0$. If $r = 2$, then $m_2 = S^2$, the variance.

The $r$th moment about any origin $A$ is defined as

$$m_r' = \frac{\sum_{j=1}^{N} (v_j - A)^r}{N}. \tag{1.21}$$

If $A = 0$, (1.21) reduces to equation (1.19). If $r = 1$, $m_1' = \frac{\sum_{j=1}^{N} v_j - A}{N} = \frac{\sum_{j=1}^{N} v_j}{N} - \frac{AN}{N} = \bar{v} - A$.

**Theorem 1.3:** The following relations exist between moments about the mean $m_r$ and moments about an arbitrary origin $m_r'$:

$$m_2 = m_2' - m_1'^2, \tag{1.22.1}$$

$$m_3 = m_3' - 3m_1' m_2' + 2m_1'^3, \tag{1.22.2}$$

$$m_4 = m_4' - 4m_1' m_3' + 6m_1'^2 m_2' - 3m_1'^4. \tag{1.22.3}$$

To avoid particular units, we can define the dimensionless moments about the mean as [6]

$$a_r = \frac{m_r}{S^r} = \frac{m_r}{(\sqrt{m_2})^r} = \frac{m_r}{\sqrt{m_2^r}}, \tag{1.23}$$

where $S = \sqrt{m_2}$ is the standard deviation. Since $m_1 = 0$ and $m_2 = S^2$, we have $a_1 = 0$ and $a_2 = 1$.

### Skewness

**Definition (The Skewness):** Skewness is the degree of asymmetry, or departure from symmetry, of a distribution.

Remember, if the frequency curve (smoothed frequency polygon) of a distribution has a longer tail to the right of the central maximum than to the left, the distribution is said to be skewed to the right, or to have positive skewness. If the reverse is true, it is said to be skewed to the left, or to have negative skewness.

For skewed distributions, the mean tends to lie on the same side of the mode as the longer tail. Thus, a measure of the asymmetry is supplied by the difference: mean–mode. This can be made dimensionless if we divide it by a measure of dispersion, such as the standard deviation, leading to the definition:

$$\mathrm{S}_K = \text{Skewness} = \frac{\text{mean} - \text{mode}}{\text{standard deviation}} = \frac{\bar{v} - \text{mode}}{S}. \tag{1.24}$$

To avoid using the mode, we can define

$$\mathrm{S}_K = \text{Skewness} = \frac{3(\text{mean} - \text{median})}{\text{standard deviation}} = \frac{3(\bar{v} - \text{median})}{S}. \tag{1.25}$$

Equations (1.24) and (1.25) are called, respectively, Pearson's first and second coefficients of skewness [5]. The limits for Karl Pearson's coefficient of skewness are $\pm 3$.

Other measures of skewness, (Bowley's Coefficient of Skewness [5]), defined in terms of quartiles and percentiles, are as follows:

$$\text{Quartile coefficient of skewness} = \mathrm{S}_K = \frac{(Q_3 - Q_2) - (Q_2 - Q_1)}{(Q_3 - Q_1)} = \frac{Q_3 - 2Q_2 + Q_1}{Q_3 - Q_1}, \tag{1.26}$$

$$10 - 90 \text{ percentile coefficient of skewness} = \mathrm{S}_K = \frac{(P_{90} - P_{50}) - (P_{50} - P_{10})}{(P_{90} - P_{10})} = \frac{P_{90} - 2P_{50} + P_{10}}{P_{90} - P_{10}}. \tag{1.27}$$

From (1.27) we observe that $\mathrm{S}_K = 0$, if $Q_3 - Q_2 = Q_2 - Q_1$. This implies that for a symmetrical distribution ($\mathrm{S}_K = 0$), median is equidistant from the upper and lower quartiles. Moreover, skewness is positive if:

$$Q_3 - Q_2 > Q_2 - Q_1 \quad \Rightarrow Q_3 + Q_1 > 2Q_2, \tag{1.28}$$

and skewness is negative if

$$Q_3 - Q_2 < Q_2 - Q_1 \quad \Rightarrow Q_3 + Q_1 < 2Q_2. \tag{1.29}$$

We know that for two real positive numbers $a$ and $b$ (i.e., $a > 0$ and $b > 0$), the moduls value of the difference $(a – b)$ is always less than or equal to the modules value of the sum $(a + b)$, i.e.,

$$|a - b| \le |a + b| \Rightarrow \left|\frac{a - b}{a + b}\right| \le 1. \tag{1.30}$$

We also know that $(Q_3 - Q_2)$ and $(Q_2 - Q_1)$ are both non-negative. Thus, taking $a = Q_3 - Q_2$ and $b = Q_2 - Q_1$, in (1.30), we get

$$\left|\frac{(Q_3 - Q_2) - (Q_2 - Q_2)}{(Q_3 - Q_2)}\right| \le 1 \Rightarrow |\,\mathrm{S}_K(\text{Bowley})| \le 1$$
$$\Rightarrow -1 \le \mathrm{S}_K(\text{Bowley}) \le 1. \tag{1.31}$$

Thus, Bowley's coefficient of skewness ranges from $-1$ to 1.

An important measure of skewness uses the third moment about the mean expressed in dimensionless form and is given by [5]:

$$\text{Moment coefficient of skewness} = a_3 = \frac{m_3}{S^3} = \frac{m_3}{(\sqrt{m_2})^3} = \frac{m_3}{\sqrt{m_2^3}}. \tag{1.32}$$

Another measure of skewness is sometimes given by $b_1 = a_3^2$. For perfectly symmetrical curves, such as the normal curve, $a_3$ and $b_1$ are zero.

### Kurtosis

If we know the measures of central tendency, dispersion and skewness, we still cannot form a complete idea about the distribution. In addition to these measures, we should know one more measure which Prof. Karl Pearson calls the "Convexity of curve or Kurtosis". Kurtosis enables us to have an idea about the flatness or peakedness of the curve,

**Definition (The Kurtosis):** Kurtosis is the degree of peakedness of a distribution, usually taken relative to a normal distribution.

Kurtosis is based on the size of a distribution's tails. Positive kurtosis indicates too few observations in the tails, whereas negative kurtosis indicates too many observations in the tail of the distribution.

A distribution having a relatively high peak is called leptokurtic, while one which is flat-topped is called platykurtic. A normal distribution, which is not very peaked or very flat-topped, is called mesokurtic. One measure of kurtosis uses the fourth moment about the mean expressed in dimensionless form and is given by [13]:

$$\text{Moment coefficient of kurtosis} = a_4 = \frac{m_4}{S^4} = \frac{m_4}{m_2^2}, \tag{1.33}$$

which is often denoted by $b_2$. For the normal distribution, $b_2 = a_4 = 3$. For this reason, the kurtosis is sometimes defined by $b_2 - 3$, which is positive for a leptokurtic distribution, negative for a platykurtic distribution, and zero for the normal distribution.

Another measure of kurtosis is based on both quartiles and percentiles and is given by [1, 6]:

$$\kappa = \frac{Q}{P_{90} - P_{10}}, \tag{1.34}$$

where $Q = (Q_3 - Q_1)/2$ is the semi-interquartile range. We refer to $\kappa$ (the lowercase Greek letter kappa) as the percentile coefficient of kurtosis; for the normal distribution, $\kappa$ has the value 0.263.

**Notations:** When it is necessary to distinguish a sample's moments, measures of skewness, and measures of kurtosis from those corresponding to a population of which the sample is a part, it is often the custom to use Latin symbols for the former and Greek symbols for the latter. Thus, if the sample's moments are denoted by $m_r$ and $m_r'$, the corresponding Greek symbols would be $\mu_r$ and $\mu_r'$. Subscripts are always denoted by Latin symbols. Similarly, if the sample's measures of skewness and kurtosis are denoted by $a_3$ and $a_4$, respectively, the population's skewness and kurtosis would be $\alpha_3$ and $\alpha_4$. We already know from that the standard deviation of a sample and of a population are denoted by $S$ and $\sigma$, respectively.

## 1.5 Random Experiment and Sample Space

Statisticians use the word experiment to describe any process that generates a set of data. A simple example of a statistical experiment is the tossing of a coin. In this experiment, there are only two possible outcomes, heads or tails.

**Definition (Random Experiment):** An experiment that can result in different outcomes, even though it is repeated in the same manner every time, is called a random experiment.

Assume that the experiment can be repeated any number of times under identical conditions. Each repetition is called a trial. A (random) experiment satisfies the following three conditions:

1. The set of all possible outcomes is known in advance in each trial;
2. In any particular trial, it is not known which particular outcome will happen; and
3. The experiment can be repeated under identical conditions.

**Definition (Sample Space):** The set of all possible outcomes of a random experiment is called the sample space of the experiment. The sample space is denoted as $S$.

**Definition (Discrete Sample Space):** A sample space is discrete if it consists of a finite or countable infinite set of outcomes.
**Definition (Continuous Sample Space):** A sample space is continuous if it contains an interval (either finite or infinite) of real numbers.

***Example 1.1***

Consider the experiment of tossing a die. If we are interested in the number that shows on the top face, the sample space is

$$S_1 = \{1, 2, 3, 4, 5, 6\}.$$

If we are interested only in whether the number is even or odd, the sample space is simply

$$S_2 = \{\text{even}, \text{odd}\}.$$

Note that, Example 1.1 illustrates the fact that more than one sample space can be used to describe the outcomes of an experiment. In some experiments, it is helpful to list the elements of the sample space systematically utilizing a tree diagram.

***Example 1.2***

An experiment consists of flipping a coin and then flipping it a second time if a head occurs. If a tail occurs on the first flip, then a die is tossed once. To list the elements of the sample space providing the most information, we construct the following tree diagram:

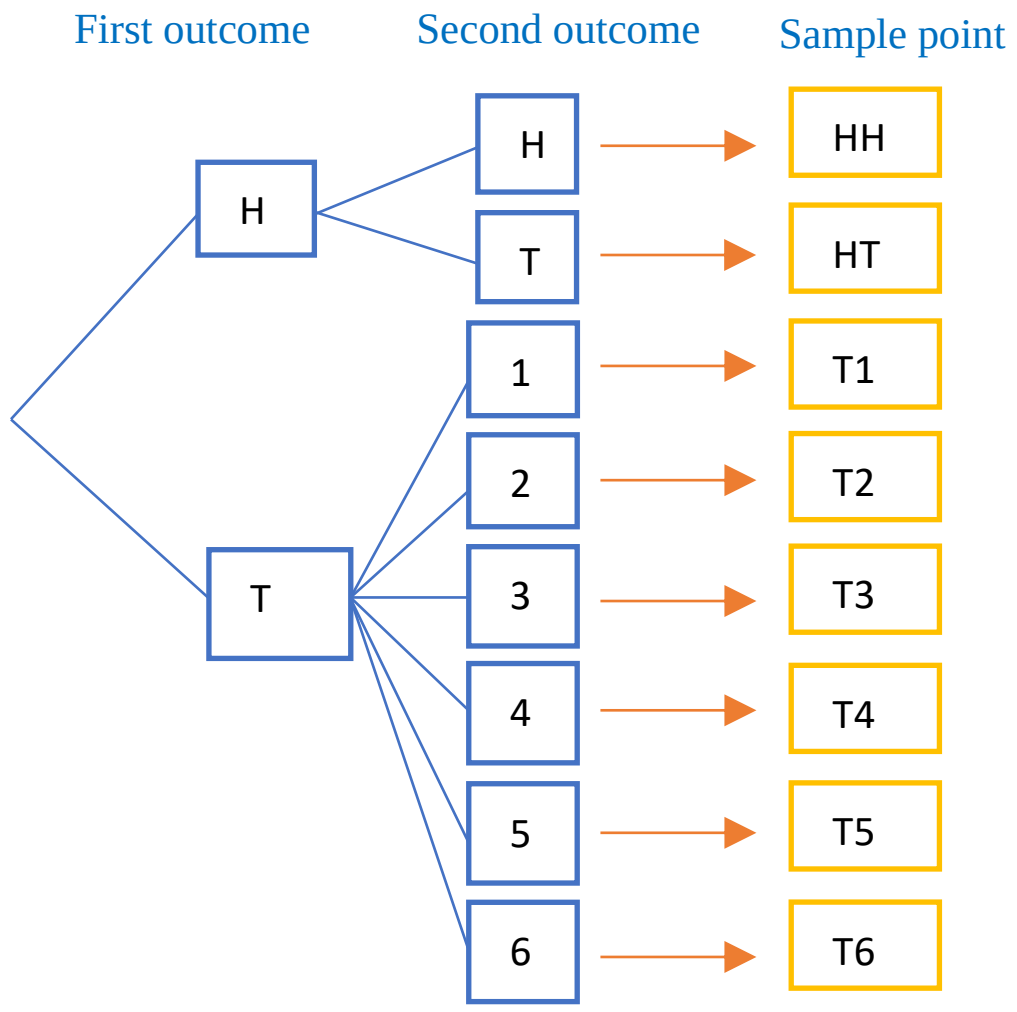

An outcome in $S$ is called a sample point or element. For any given experiment, we may be interested in the occurrence of certain events rather than in the occurrence of a specific element in the sample space. For instance, we may be interested in the event $A$ that the outcome when a die is tossed is divisible by 3. This will occur if the outcome is an element of the subset $A = \{3,6\}$ of the sample space $S_1$ in Example 1.1.

**Definition (Event):** An event is a subset of the sample space of a random experiment.

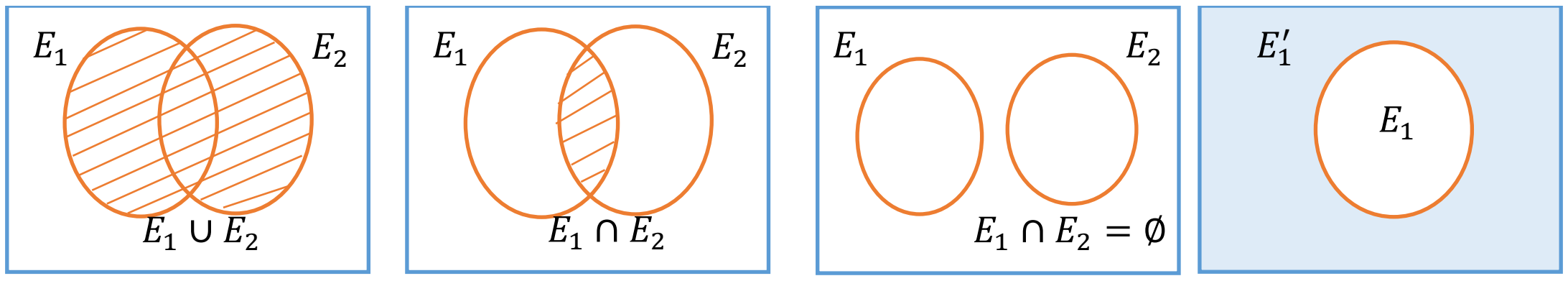

**Figure 1.11.** Venn diagrams.

***Example 1.3***

Define the events $A$ and $B$ for the die-tossing experiment:
$A$: Observe an odd number,
$B$: Observe a number less than 4.

***Solution***

Since event $A$ occurs if the upper face is 1, 3, or 5, it is a collection of three simple events (sample points) and we write $A = \{1,3,5\}$. Similarly, the event $B$ occurs if the upper face is 1, 2, or 3 and is defined as a collection of three simple events: $B = \{1, 2,3\}$.

Some of the basic set operations are summarized here in terms of events:

- The union of two events is the event that consists of all outcomes that are contained in either of the two events. We denote the union as $E_1 \cup E_2$. Note that $E_1 \cup E_2 = E_2 \cup E_1$.
- The intersection of two events is the event that consists of all outcomes that are contained in both of the two events. We denote the intersection as $E_1 \cap E_2$. Note that $E_1 \cap E_2 = E_2 \cap E_1$.
- The distributive law for set operations implies that
$$(A \cup B) \cap C = (A \cap C) \cup (B \cap C) \text{ and } (A \cap B) \cup C = (A \cup C) \cap (B \cup C). \tag{1.35}$$
- The complement of an event in a sample space is the set of outcomes in the sample space that are not in the event. We denote the complement of the event $E$ as $E'$. The notation $E^C$ is also used in other literature to denote the complement.
- The definition of the complement of an event implies that:
$$(E')' = E. \tag{1.36}$$
- DeMorgan's laws imply that
$$(A \cup B)' = A' \cap B' \text{ and } (A \cap B)' = A' \cup B'. \tag{1.37}$$

**Definition (Mutually Exclusive):** Two events are mutually exclusive if, when one event occurs, the other cannot, and vice versa. Hence, two events, denoted as $E_1$ and $E_2$, such that
$$E_1 \cap E_2 = \emptyset, \tag{1.38}$$
are said to be mutually exclusive.

***Example 1.4***

In the die-tossing experiment, Example 1.3, events $A$ and $B$ are not mutually exclusive, because they have two outcomes in common—observing a 1 or a 3. Both events $A$ and $B$ will occur if either 1 or 3 is observed when the experiment is performed. In contrast, the six simple events 1, 2, ..., and 6 form a set of all mutually exclusive outcomes of the experiment. When the experiment is performed once, one and only one of these simple events can occur.

We can use Venn diagrams to represent a sample space and events in a sample space, see for example, Figure 1.11.

## 1.6 Counting Techniques

In a sample space with a large number of outcomes, determining the number of outcomes associated with the events through direct enumeration could be tedious. In this section, we consider three counting techniques (multiplication principle, permutations and combinations) and use them in probability computations.

**Counting Techniques**

The multiplication principle allow us to determine the total number of outcomes in a given scenario. Whether it is arranging objects in a specific order or selecting items from a set, this technique help us systematically count the possibilities.

Permutations refer to the arrangements or orderings of a set of objects. They are used when the order of elements matters.

Combinations, on the other hand, are concerned with the selection of objects without considering their order.

**Definition (Multiplication Principle):** Assume an operation can be described as a sequence of $k$ experiments $A_1$, $A_2$, ., $A_k$ contain, respectively, $n_1$, $n_2$, ., $n_k$ outcomes, such that for each possible outcome of $A_1$ there are $n_2$ possible outcomes for $A_2$, and so on, then there are a total of [12]:

$$n_1 \times n_2 \times \cdots \times n_k, \quad (1.39)$$

possible outcomes for the composite experiment $A_1$, $A_2$, ., $A_k$.

The multiplication principle is particularly useful when dealing with situations where events occur one after another in a sequence. By multiplying the number of possibilities at each step, we can determine the total number of outcomes for the entire sequence. The multiplication principle is often visualized using tree diagrams.

***Example 1.5***

How many sample points are there in the sample space when a pair of dice is thrown once?

***Solution***

(1,1), (2,1), (3,1), (4,1), (5,1), (6,1),
(1,2), (2,2), (3,2), (4,2), (5,2), (6,2),
(1,3), (2,3), (3,3), (4,3), (5,3), (6,3),
(1,4), (2,4), (3,4), (4,4), (5,4), (6,4),
(1,5), (2,5), (3,5), (4,5), (5,5), (6,5),
(1,6), (2,6), (3,6), (4,6), (5,6), (6,6),

The first die can land face-up in any one of $n_1 = 6$ ways. For each of these 6 ways, the second die can also land face-up in $n_2 = 6$ ways. Therefore, the pair of dice can land in $n_1 \times n_2 = 6 \times 6 = 36$ possible ways.

When a random sample of size $k$ is taken from a total of $n$ objects, the total number of ways in which the random sample of size $k$ can be selected depends on the particular sampling method we employ. Here, we will consider four sampling methods:

- Sampling with replacement and the objects are ordered,
- Sampling without replacement and the objects are ordered,
- Sampling without replacement and the objects are not ordered, and
- Sampling with replacement and the objects are not ordered.

### Sampling with Replacement and the Objects Are Ordered

- Sampling with replacement refers to a sampling method where an object or element is selected from a set, and after selection, it is returned to the set before the next selection is made. This means that the same object can be chosen more than once in the sampling process.
- When the objects are ordered, it means that the arrangement or sequence of the selected objects is considered significant. The order in which the objects are selected or arranged affects the outcome.
- When a random sample of size $k$ is taken with replacement from a total of $n$ objects and the objects being ordered, then there are $n^k$ possible ways of selecting $k$-tuples.

| 1 | 2 | … | $k$ |
|---|---|---|---|
| Choice 1 | Choice 1 | … | Choice 1 |
| … | … | … | … |
| Choice $n$ | Choice $n$ | … | Choice $n$ |

For example,

(1) if a die is rolled four times, then the sample space will consist of $6^4$ 4-tuples.

(2) If an urn contains nine balls numbered 1 to 9, and a random sample with replacement of size $k = 6$ is taken, then the sample space $S$ will consist of $9^6$ 6-tuples.

### Sampling without Replacement and the Objects Are Ordered

Sampling without replacement refers to a sampling method where each object or element is selected from a set, and once selected, it is not returned to the set before the next selection is made. This means that each object can only be chosen once in the sampling process.

Consider a set of elements, such as $S = \{a, b, c\}$. A permutation of the elements is an ordered sequence of the elements. For example, $abc$, $acb$, $bac$, $bca$, $cab$, and $cba$ are all of the permutations of the elements of $S$. There are $n_1 = 3$ choices for the first position. No matter which letter is chosen, there are always $n_2 = 2$ choices for the second position. No matter which two letters are chosen for the first two positions, there is only $n_3 = 1$ choice for the last position, giving a total of $n_1 n_2 n_3 = (3)(2)(1) = 6$ permutations.

| 1 | 2 | ... | $n$ |
|---|---|---|---|
| Choice 1 | Choice 1 | ... | Choice 1 |
| ... | ... | ... | |
| | | ... | |
| | Choice $n-1$ | | |
| Choice $n$ | | | |

**Definition (Number of Permutations):** The number of permutations of $n$ different elements is $n!$ [8, 14], where

$$n! = n \times (n-1) \times (n-2) \times \cdots \times 2 \times 1. \tag{1.40}$$

In some situations, we are interested in the number of arrangements of only some of the elements of a set. The following result also follows from the multiplication rule and the previous discussion.

| 1 | 2 | … | $r$ |
|---|---|---|---|
| Choice 1 | Choice 1 | … | Choice 1 |
| … | … | … | … |
| | | | Choice $n-(r-1)$ |
| | | … | |
| | Choice $n-1$ | | |
| Choice $n$ | | | |

**Definition (Permutations of Subsets):** The number of permutations of subsets of $r$ elements selected from a set of $n$ different elements is [8, 14]:

$$P_r^n = n \times (n-1) \times (n-2) \times \cdots \times (n-r+1) = \frac{n!}{(n-r)!}. \tag{1.41}$$

***Example 1.6***

The number of permutations of the four letters $a$, $b$, $c$, and $d$ will be $4! = 24$. We have

$$\{abcd, abdc, acbd, acdb, adbc, adcb, bacd, badc, bdac, bdca, bcad, bcda,$$
$$cabd, cadb, cbad, cbda, cdab, cdba, dabc, dacb, dbac, dbca, dcab, dcba\}.$$

Now consider the number of permutations that are possible by taking two letters at a time from four. These would be

$$\{ab, ac, ad, ba, bc, bd, ca, cb, cd, da, db, dc\}.$$

We have two positions to fill, with $n_1 = 4$ choices for the first and then $n_2 = 3$ choices for the second, for a total of $n_1 n_2 = (4)(3) = 12$ permutations. In general, $n$ distinct objects taken $r$ at a time can be arranged in $n(n-1)(n-2)\cdots(n-r+1)$ ways$= 4!/2! = (4)(3) = 12$.

### Permutations of Similar Objects

When dealing with permutations of similar objects, we often encounter repetitions of elements. This is because there are multiple objects with the same characteristics and arranging them in different orders results in equivalent outcomes.

If the objects are all distinct, then we have seen that the number of permutations without repetition is $n!$. For each of these permutations, we can permute the $n_1$ identical objects of type 1 in $n_1!$ possible ways; since these objects are considered identical, the arrangement is unchanged. Similarly, we can take any of the $n_2!$ permutations of the identical objects of type 2 and obtain the same arrangement. Continuing this argument, we account for these repeated arrangements by dividing by the number of repetitions. This gives the following result for the total number of permutations:

**Definition (Permutations of Similar Objects):** The number of distinct permutations of $n = n_1 + n_2 + \cdots + n_r$ objects of which $n_1$ are of one type, $n_2$ are of a second type, …, and $n_r$ are of an $r$th type, is [8, 14]:

$$\frac{n!}{n_1!\, n_2!\, n_3! \ldots n_r!}. \tag{1.42}$$

### Sampling without Replacement and the Objects Are Not Ordered

When the objects are not ordered, it means that the arrangement or sequence of the selected objects is not considered significant. The order in which the objects are selected or arranged does not affect the outcome.

Note that in a permutation, the order in which each object is selected becomes important. When the order of arrangement is not important for example, if we do not distinguish between $AB$ and $BA$, the arrangement is called a combination.

**Definition (Combinations):** The number of ways in which $r$ objects can be selected (without replacement) from a collection of $n$ distinct objects (the number of combinations is denoted as $\binom{n}{r}$ or $C_r^n$) is [8, 14],

$$C_r^n = \binom{n}{r} = \frac{n!}{r!\,(n-r)!} = \frac{n(n-1)(n-2)\ldots(n-r+1)}{r!}. \tag{1.43}$$

### Sampling with Replacement and the Objects Are Not Ordered

In obtaining an unordered sample of size $r$, with replacement, from a total of $n$ objects, $(r-1)$ replacements will be made before sampling ceases. Hence, the number of possible samples can be obtained by using the formula,

$$\binom{n+r-1}{r} = \frac{(n+r-1)!}{r!\,(n-1)!}. \tag{1.44}$$

**Theorem 1.4:** The total number of distinct $r$ samples from an $n$-element set such that repetition is allowed and order does not matter is the same as the number of distinct solutions to the equation

$$x_1 + x_2 + \ldots + x_n = r, \quad \text{where } x_i \in \{0,1,2,3,\ldots\}. \tag{1.45.1}$$

and is equal to

$$\binom{n+r-1}{r} = \frac{(n+r-1)!}{r!\,(n-1)!}. \tag{1.45.2}$$

## 1.7 Interpretations and Axioms of Probability

Probability is used to quantify the likelihood, or chance, that an outcome of a random experiment will occur. The likelihood of an outcome is quantified by assigning a number from the interval $[0,1]$ to the outcome (or a percentage from 0 to 100%).

### Classical Interpretation

The classical probability rule is applied to compute the probabilities of events for an experiment for which all outcomes are equally likely. For example, head and tail are two equally likely outcomes when a fair coin is tossed once. Each of these two outcomes has the same chance of occurrence.

According to the classical probability rule, to find the probability of a simple event, we divide 1.0 by the total number of outcomes for the experiment. On the other hand, to find the probability of a compound event $E$, we divide the number of outcomes favorable to event $E$ by the total number of outcomes for the experiment.

**Definition (Equally Likely Outcomes):** Whenever a sample space consists of $N$ possible outcomes that are equally likely, the probability of each outcome is $1/N$.

**Definition (Probability of an Event):** For a discrete sample space, the probability of an event $E$, denoted as $P(E)$, equals the sum of the probabilities of the outcomes in $E$.

### Relative Frequency Interpretation

The relative frequency of an event is used as an approximation for the probability of that event. Because relative frequencies are determined by performing an experiment, the probabilities calculated using relative frequencies may change when an experiment is repeated. The probability of an outcome is interpreted as the limiting value of the proportion of times the outcome occurs in $n$ repetitions of the random experiment as $n$ increases beyond all bounds.

If an experiment is repeated $n$ times and an event $A$ is observed $f$ times where $f$ is the frequency, then, according to the relative frequency concept of probability:

$$P(A) = \frac{f}{n} = \frac{\text{Frequency of A}}{\text{Sample size}}. \tag{1.46}$$

### Axioms of Probability

Axioms are the foundational principles upon which probability theory is constructed. The axiomatic approach defines the properties that probabilities must satisfy to be considered valid measures of uncertainty. The Kolmogorov axioms are introduced by Russian mathematician Andrey Kolmogorov in 1933.

**Definition (Kolmogorov Axioms, Axioms of Probability):** Probability is a number that is assigned to each member of a collection of events from a random experiment that satisfies the following properties:
(1) $P(S) = 1$ where $S$ is the sample space
(2) $0 \le P(E) \le 1$ for any event $E$
(3) For two events $E_1$ and $E_2$ with $E_1 \cap E_2 = \emptyset$

$$P(E_1 \cup E_2) = P(E_1) + P(E_2). \tag{1.47}$$

**Remark:**

$$P(\emptyset) = 0, \tag{1.48}$$

and for any event $E$,

$$P(E') = 1 - P(E). \tag{1.49}$$

***Example 1.7***

A coin is tossed twice. What is the probability that at least 1 head occurs?

***Solution***

The sample space for this experiment is

$$S = \{HH, HT, TH, TT\}.$$

If the coin is balanced, each of these outcomes is equally likely to occur. Therefore, we assign a probability of $\omega$ to each sample point. Then $4\omega = 1$, or $\omega = 1/4$. If $A$ represents the event of at least 1 head occurring, then

$$A = \{HH, HT, TH\} \text{ and } P(A) = \frac{1}{4} + \frac{1}{4} + \frac{1}{4} = \frac{3}{4}.$$

Hence, if an experiment can result in any one of $N$ different equally likely outcomes, and if exactly $n$ of these outcomes correspond to event $A$, then the probability of event $A$ is

$$P(A) = \frac{n}{N}. \tag{1.50}$$

### Unions of Events and Addition Rules

Joint events are generated by applying basic set operations to individual events. Unions of events, such as $A \cup B$; intersections of events, such as $A \cap B$; and complements of events, such as $A'$— are common of interest. The probability of a joint event can often be determined from the probabilities of the individual events that it comprises. Basic set operations are also sometimes helpful in determining the probability of a joint event.

**Theorem 1.5 (Probability of a Union):**

$$P(A \cup B) = P(A) + P(B) - P(A \cap B). \tag{1.51}$$

If $A$ and $B$ are mutually exclusive events, ($A \cap B = \emptyset$ and $P(A \cap B) = 0$), then

$$P(A \cup B) = P(A) + P(B). \tag{1.52}$$

For three events, we have

$$P(A \cup B \cup C) = P(A) + P(B) + P(C) - P(A \cap B) - P(A \cap C) - P(B \cap C) + P(A \cap B \cap C). \tag{1.53}$$

Moreover, a collection of events, $E_1, E_2, \ldots, E_k$, is said to be mutually exclusive if for all pairs,

$$E_i \cap E_j = \emptyset. \tag{1.54}$$

For a collection of mutually exclusive events,

$$P(E_1 \cup E_2 \cup \ldots \cup E_k) = P(E_1) + P(E_2) + \cdots + P(E_k). \tag{1.55}$$

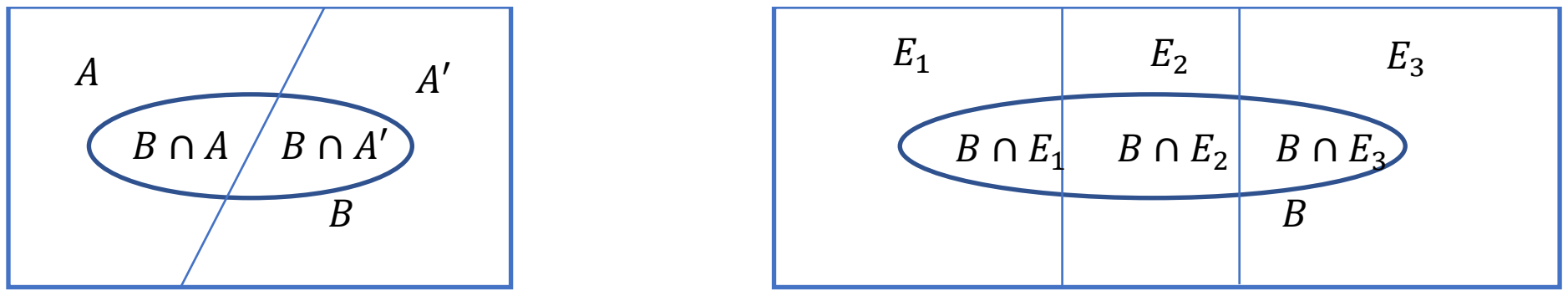

**Figure 1.12.** Left panel: Partitioning an event into two mutually exclusive subsets. Right panel: Partitioning an event into several mutually exclusive subsets.

## 1.8 Conditional Probability

One very important concept in probability theory is conditional probability. In some applications, the practitioner is interested in the probability structure under certain restrictions. The probability of an event $B$ under the knowledge that the outcome will be in event $A$ is denoted as $P(B|A)$ and this is called the conditional probability of $B$ given $A$.

**Definition (Conditional Probability):** The conditional probability of an event $B$ given that an event $A$ has occurred, $P(A) > 0$, is

$$P(B|A) = \frac{P(A \cap B)}{P(A)}. \tag{1.56}$$

This definition can be understood in a special case in which all outcomes of a random experiment are equally likely. If there are $N$ total outcomes,

$$P(A) = (\text{number of outcomes in } A)/N. \tag{1.57}$$

Also,

$$P(A \cap B) = (\text{number of outcomes in } A \cap B)/N. \tag{1.58}$$

Consequently,

$$\frac{P(A \cap B)}{P(A)} = \frac{\text{number of outcomes in } A \cap B}{\text{number of outcomes in } A}. \tag{1.59}$$

Therefore, $P(B|A)$ can be interpreted as the relative frequency of event $B$ among the trials that produce an outcome in event $A$.

**Definition (Multiplication Rule [15]):**

$$P(A \cap B) = P(B|A)P(A) = P(A|B)P(B). \tag{1.60}$$

Thus, the probability that both $A$ and $B$ occur is equal to the probability that $A$ occurs multiplied by the conditional probability that $B$ occurs, given that $A$ occurs.

For any event $B$, we can write $B$ as the union of the part of $B$ in $A$ and the part of $B$ in $A'$. That is,

$$B = (A \cap B) \cup (A' \cap B). \tag{1.61}$$

This result is shown in the Venn diagram in Figure 1.12. Because $A$ and $A'$ are mutually exclusive, $A \cap B$ and $A' \cap B$ are mutually exclusive.

Therefore, the following total probability rule is obtained.

**Definition (Total Probability Rule [15]):** For any events $A$ and $B$,

$$\begin{aligned} P(B) &= P(B \cap A) + P(B \cap A') \\ &= P(B|A)P(A) + P(B|A')P(A'). \end{aligned} \tag{1.62}$$

Moreover, assume $E_1, E_2, \dots, E_k$ are $k$ mutually exclusive and exhaustive sets. Then

$$\begin{aligned} P(B) &= P(B \cap E_1) + P(B \cap E_2) + \cdots + P(B \cap E_k) \\ &= P(B|E_1)P(E_1) + P(B|E_2)P(E_2) + \cdots + P(B|E_k)P(E_k). \end{aligned} \tag{1.63}$$

### Independence

Two events are said to be independent if the occurrence (or non-occurrence) of one event does not affect the probability that the other event will occur. In this case, the conditional probability of $P(B|A)$ might equal $P(B)$, i.e., the knowledge that the outcome of the experiment is in event $A$ does not affect the probability that the outcome is in event $B$. So that, we obtain

$$P(A \cap B) = P(B|A)P(A) = P(B)P(A), \tag{1.64}$$

and

$$P(A|B) = \frac{P(A \cap B)}{P(B)} = \frac{P(A)P(B)}{P(B)} = P(A). \tag{1.65}$$

These conclusions lead to an important definition.

**Definition (Independence, Two Events):** Two events are independent if any one of the following equivalent statements is true:
(1) $P(A|B) = P(A)$,
(2) $P(B|A) = P(B)$,
(3) $P(A \cap B) = P(A)P(B)$.

It is simple to show that independence implies related results such as

$$P(A' \cap B') = P(A')P(B'). \tag{1.66}$$

**Remark:**

A mutually exclusive relationship between two events is based only on the outcomes that compose the events. However, an independence relationship depends on the probability model used for the random experiment. Often, independence is assumed to be part of the random experiment that describes the physical system under study. The concepts of mutually independent events and mutually exclusive events are separate and distinct. The following table contrasts the results for the two cases (provided that the probability of the conditioning event is not zero).

| | If statistically independent | If mutually exclusive |
|---|---|---|
| $P(A\|B) =$ | $P(A)$ | 0 |
| $P(B\|A) =$ | $P(B)$ | 0 |
| $P(A \cap B) =$ | $P(A)P(B)$ | 0 |

**Definition (Independence, Multiple Events):** The events $E_1, E_2, \ldots, E_n$ are independent if and only if [8]

$$P\left(E_{i_1} \cap E_{i_2} \cap \cdots \cap E_{i_k}\right) = P\left(E_{i_1}\right) \times P\left(E_{i_2}\right) \times \cdots \times P\left(E_{i_k}\right). \tag{1.67}$$

**Definition (Pairwise Independent, Multiple Events):** The events $E_1, E_2, \ldots, E_n$ are pairwise independent if and only if [12]

$$P\left(E_{i_j} \cap E_{i_k}\right) = P\left(E_{i_j}\right) \times P\left(E_{i_k}\right) \text{ for each } j \neq k. \tag{1.68}$$

### Bayes Theorem

From the definition of conditional probability,

$$P(A \cap B) = P(A|B)P(B) = P(B \cap A) = P(B|A)P(A). \tag{1.69}$$

Now, considering the second and last terms in the preceding expression, we can write

$$P(A|B) = \frac{P(B|A)P(A)}{P(B)} \text{ for } P(B) > 0. \tag{1.70}$$

This is a useful result that enables us to solve for $P(A|B)$ in terms of $P(B|A)$.

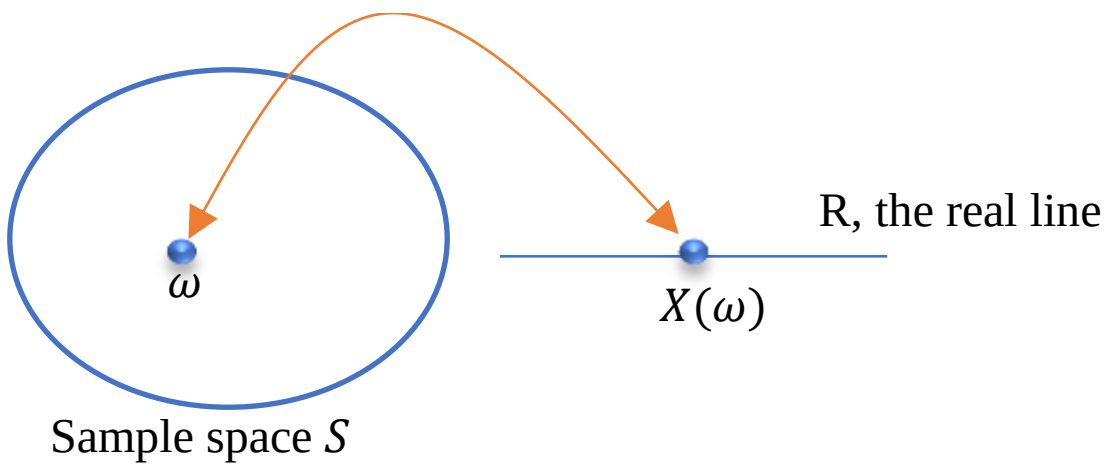

**Figure 1.13.** RV as a function.

**Theorem 1.6 (Bayes Theorem):** If $E_1, E_2, \dots, E_k$ are $k$ mutually exclusive and exhaustive events and $B$ is any event [8],

$$P(E_1|B) = \frac{P(B|E_1)P(E_1)}{P(B|E_1)P(E_1) + P(B|E_2)P(E_2) + \cdots + P(B|E_k)P(E_k)}, \tag{1.71}$$

for $P(B) > 0$

## 1.9 Discrete Random Variables

### Random Variables (RVs)

The concept of a RV allows us to pass from the experimental outcomes to a numerical function of the outcomes, often simplifying the sample space. In simpler terms, a RV is like a function that maps the outcomes of a random event to numerical values. A RV is denoted by an uppercase letter such as $X$. After an experiment is conducted, the measured value of the RV is denoted by a lowercase letter such as $x$. Just like how we assign probabilities to events, we can do the same to RVs. Consider the following example.

***Example 1.8***

In tossing dice, we are often interested in the sum of the two dice and are not really concerned about the values of the individual dice. That is, we may be interested in knowing that the sum is 7 and not be concerned over whether the actual outcome was (1,6), (2,5), (3,4), (4,3), (5,2) or (6,1). Letting $X$ denote the RV that is defined as the sum of two fair dice, then:

$$P(X = 2) = P((1,1)) = 1/36,$$
$$P(X = 3) = P((1,2),(2,1)) = 2/36,$$
$$P(X = 4) = P((1,3),(2,2),(3,1)) = 3/36,$$
$$P(X = 5) = P((1,4),(2,3),(3,2),(4,1)) = 4/36,$$
$$P(X = 6) = P((1,5),(2,4),(3,3),(4,2),(5,1)) = 5/36,$$
$$P(X = 7) = P((1,6),(2,5),(3,4),(4,3),(5,2),(6,1)) = 6/36,$$
$$P(X = 8) = P((2,6),(3,5),(4,4),(5,3),(6,2)) = 5/36,$$
$$P(X = 9) = P((3,6),(4,5),(5,4),(6,3)) = 4/36,$$
$$P(X = 10) = P((4,6),(5,5),(6,4)) = 3/36,$$
$$P(X = 11) = P((5,6),(6,5)) = 2/36,$$
$$P(X = 12) = P((6,6)) = 1/36.$$

In other words, the RV, $X$, can take on any integral value between 2 and 12. Moreover, we must have

$$1 = P(S) = P\left(\bigcup_{x=2}^{12}(X = x)\right) = \sum_{x=2}^{12} P(X = x).$$

**Definition (RV):** A RV is a function that assigns a real number to each outcome in the sample space, $S$, of a random experiment,

$$X(\omega) = x, \tag{1.72}$$

with each outcome $\omega$ in $S$. (See Figure 1.13)

In the above example, the RVs of interest took on a finite number of possible values. RVs whose set of possible values can be written either as a finite sequence $x_1, \ldots, x_n$, or as an infinite sequence $x_1, \ldots$ are said to be discrete.

**Definition (Discrete RVs):** A discrete RV is a RV with a countable finite (or countably infinite) range. Or a discrete RV is a RV whose possible values can be listed.

**Definition (Continuous RVs):** A continuous RV is a RV with an interval (either finite or infinite) of real numbers for its range.

RVs are so important in random experiments that sometimes we essentially ignore the original sample space of the experiment and focus on the probability distribution of the RV. The probability distribution of a RV is a description of the probabilities associated with the possible values of the RV. For a discrete RV, the distribution is often specified by just a list of the possible values along with the probability of each. For other cases, probabilities are expressed in terms of a formula.

**Definition (Probability Mass Function, PMF):** For a discrete RV $X$ with possible values $x_1, x_2, \ldots, x_n$, a PMF, $f$, is a function such that [8]:
(1) $f(x_i) \geq 0$.
(2) $\sum_{i=1}^{n} f(x_i) = 1$.
(3) $f(x_i) = P(X = x_i)$.
PMFs are also commonly referred to as discrete probability distribution.

In general, for any discrete RV with possible values $x_1$, $x_2$, ..., the events $(X = x_1)$, $(X = x_2)$, ... are mutually exclusive. Therefore, $P(X \leq x) = \sum_{x_i \leq x} f(x_i)$. This leads to,

**Definition (Cumulative Distribution Function, CDF):** The CDF of a discrete RV $X = x$, denoted as $F(x)$, is

$$F(x) = P(X \leq x) = \sum_{x_i \leq x} f(x_i). \tag{1.73}$$

For a discrete RV, $X = x$, $F(x)$ satisfies the following properties [8]
(1) $F(x) = P(X \leq x) = \sum_{x_i \leq x} f(x_i)$.
(2) $0 \leq F(x) \leq 1$.
(3) If $x \leq y$, then $F(x) \leq F(y)$.
That is, $F(x)$ is the probability that the RV, $X$, takes on a value that is less than or equal to $x$.

**Remarks:**

- Note that this is analogous to a relative-frequency distribution with probabilities replacing the relative frequencies. Thus, we can think of probability distributions as theoretical or ideal limiting forms of relative-frequency distributions when the number of observations made is very large.
- Notation: We will use the notation $X \sim F$ to signify that $F$ is the CDF of $X$.
- All probability questions about $X$ can be answered in terms of its CDF $F$. For example, suppose we wanted to compute $P(a < X \leq b)$. This can be accomplished by first noting that the event $(X \leq b)$ can be expressed as the union of the two mutually exclusive events $(X \leq a)$ and $(a < X \leq b)$. Therefore, we obtain that

$$P(X \leq b) = P(X \leq a) + P(a < X \leq b), \tag{1.74}$$

or

$$\begin{aligned} P(a < X \leq b) &= P(X \leq b) - P(X \leq a) \\ &= F(b) - F(a). \end{aligned} \tag{1.75}$$

It is often helpful to look at a probability distribution in graphic form. One might plot the points $(x, f(x))$. By joining the points to the $x$ axis either with a dashed or with a solid line, we obtain a PMF plot. The graph of the probability distribution makes it easy to see what values of $X$ are most likely to occur, and it also indicates a symmetry of the probability distribution. Instead of plotting the points $(x, f(x))$, we more frequently construct a probability histogram. To understand that well, let us consider the following example.

***Example 1.9***

Let a pair of fair dice be tossed and let $X$ denote the sum of the points obtained. Then the probability distribution and the CDFs are as shown in Table 1.1. For example, the probability of getting sum 5 is $4/36 = 1/9$; thus in 900 tosses of the dice we would expect 100 tosses to give the sum 5.

**Table 1.1.** Probability distribution and the CDFs.

| $X$ | 2 | 3 | 4 | 5 | 6 | 7 | 8 | 9 | 10 | 11 | 12 |
|---|---|---|---|---|---|---|---|---|---|---|---|
| $P(X)$ | $1/36$ | $2/36$ | $3/36$ | $4/36$ | $5/36$ | $6/36$ | $5/36$ | $4/36$ | $3/36$ | $2/36$ | 1/36 |
| $F(x)$ | $1/36$ | $3/36$ | $6/36$ | $10/36$ | $15/36$ | $21/36$ | $26/36$ | $30/36$ | $33/36$ | $35/36$ | 36/36 |

Notice that the CDF is constant over any half-closed integer interval from 2 to 12. For example, $F(X) = 3/36$ for all $X$ in the interval [3,4). Figure 1.14 represents the probability histogram, PMF and CDF.

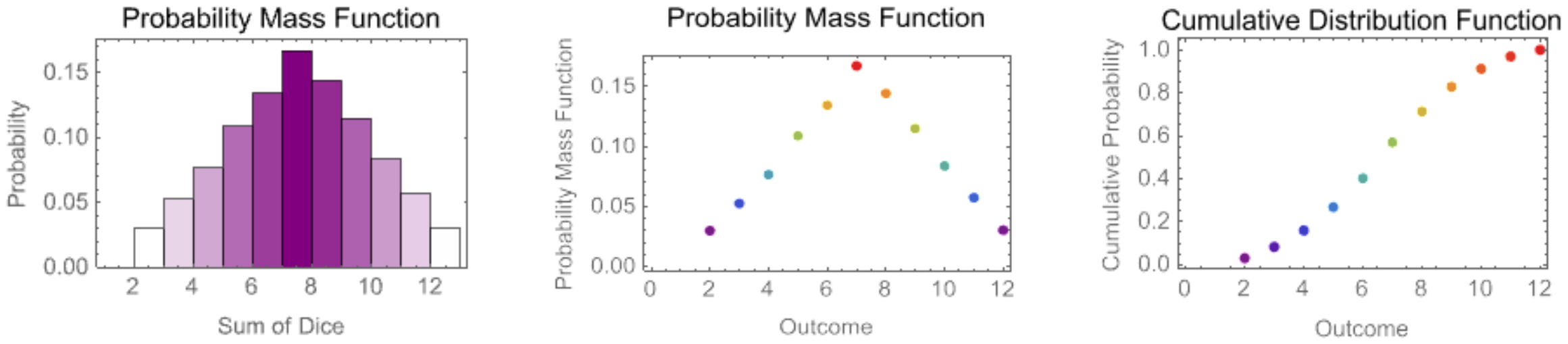

**Figure 1.14.** The probability histogram, PMF and CDF for 10000 samples from a pair of fair dice.

The concept of expectation is easily extended. If $X$ denotes a discrete RV that can assume the values $x_1, x_2, \dots, x_k$ with respective probabilities $P_1, P_2, \dots, P_k$, where $P_1 + P_2 + \dots + P_k = 1$, the mathematical expectation of $X$ (or simply the expectation of $X$), denoted by $\mathbb{E}(X)$, is defined as

$$\mathbb{E}(X) = P_1 x_1 + P_2 x_2 + \dots + P_k x_k = \sum_{j=1}^{k} P_j x_j. \tag{1.76}$$

If the probabilities $P_j$ in this expectation are replaced with the relative frequencies $f_j/N$, where $N = \sum f_j$, the expectation reduces to $\sum_{j=1}^{k} f_j x_j /N$, which is the arithmetic mean $\bar{X}$ of a sample of size $N$ in which $x_1, x_2, \dots, x_k$ appear with these relative frequencies. As $N$ gets larger and larger, the relative frequencies $f_j/N$ approach the probabilities $P_j$. Thus, we are led to the interpretation that $\mathbb{E}(X)$ represents the mean of the population from which the sample is drawn.

**Definition (Mean):** The mean or expected value of the discrete RV $X$, denoted as $\mu$ or $\mathbb{E}(X)$, is [8]

$$\mu = \mathbb{E}(X) = \sum_x x f(x). \tag{1.77}$$

**Definition (Expected Value):** The expected value of a function $h(X)$ of a discrete RV $X$ with PMF $f(x)$ is defined as [13]

$$\mathbb{E}[h(X)] = \sum_x h(x) f(x). \tag{1.78}$$

**Theorem 1.7:** If $a$ and $b$ are constants, then,

$$\mathbb{E}[aX + b] = a\mathbb{E}[X] + b. \tag{1.79}$$

**Remarks:**

- If we take $a = 0$ in Theorem 1.7, we see that,

$$\mathbb{E}[b] = b. \tag{1.80}$$

That is, the expected value of a constant is just its value.

- If we take $b = 0$, then we obtain,

$$\mathbb{E}[aX] = a\mathbb{E}[X]. \tag{1.81}$$

The expected value of a constant multiplied by a RV is just the constant times the expected value of the RV.

- The quantity $\mathbb{E}[X^n]$, $n \geq 1$, is called the $n$th moment of $X$. By (1.78), we note that

$$\mathbb{E}[X^n] = \sum_x x^n P(x). \tag{1.82}$$

**Definition (Variance):** If $X$ is a RV with mean $\mu$, then the variance of $X$, denoted as $\sigma^2$ or $V(X) \equiv \mathrm{Var}(X)$ is defined by [1]

$$\begin{aligned} \sigma^2 = V(X) &= \mathrm{Var}(X) \\ &= \mathbb{E}[(X-\mu)^2] \\ &= \sum_x (x-\mu)^2 f(x). \end{aligned} \tag{1.83}$$

**Definition (Standard Deviation):** The standard deviation of $X$ is

$$\sigma = \sqrt{\sigma^2}. \tag{1.84}$$

An alternative formula for $\mathrm{Var}(X)$ can be derived as follows:

$$\begin{aligned} \mathrm{Var}(X) &= \mathbb{E}[(X-\mu)^2] \\ &= \mathbb{E}[X^2 - 2\mu X + \mu^2] \\ &= \mathbb{E}[X^2] - \mathbb{E}[2\mu X] + \mathbb{E}[\mu^2] \\ &= \mathbb{E}[X^2] - 2\mu\mathbb{E}[X] + \mu^2 = \mathbb{E}[X^2] - \mu^2. \end{aligned}$$

**Theorem 1.8:** The variance of $X$ is equal to the expected value of the square of $X$ minus the square of the expected value of $X$.

$$\mathrm{Var}(X) = \mathbb{E}[X^2] - (\mathbb{E}[X])^2. \tag{1.85}$$

**Theorem 1.9:** For any constants $a$ and $b$,

$$\mathrm{Var}(aX + b) = a^2\mathrm{Var}(X). \tag{1.86}$$

**Remarks:**

- Specifying particular values for $a$ and $b$ in (1.86) leads to some interesting results. For instance, by setting $a = 0$ in (1.86) we obtain

$$\mathrm{Var}(b) = 0. \tag{1.87}$$

That is, the variance of a constant is 0.

- Similarly, by setting $a = 1$, we obtain,

$$\mathrm{Var}(X + b) = \mathrm{Var}(X). \tag{1.88}$$

That is, the variance of a constant plus a RV is equal to the variance of the RV.

- Finally, setting $b = 0$ yields

$$\mathrm{Var}(aX) = a^2\mathrm{Var}(X). \tag{1.89}$$

**Definition (Moment Generating Function, MGF):** The MGF $M_X(t)$ of the RV $X$ is defined for all values $t$ by [1]

$$M_X(t) = \mathbb{E}[e^{tX}] = \sum_x e^{tx}P(x). \tag{1.90}$$

We call $M_X(t)$ the MGF because all of the moments of $X$ can be obtained by successively differentiating $M_X(t)$. Recall that the Maclaurin series of the function $e^{tx}$ is

$$e^{tx} = 1 + tx + \frac{(tx)^2}{2!} + \frac{(tx)^3}{3!} + \cdots + \frac{(tx)^n}{n!} + \cdots. \tag{1.91}$$

By using the fact that the expected value of the sum equals the sum of the expected values, the MGF can be written as

$$\begin{aligned} M_X(t) &= \mathbb{E}[e^{tX}] \\ &= \mathbb{E}\left[1 + tX + \frac{(tX)^2}{2!} + \frac{(tX)^3}{3!} + \cdots + \frac{(tX)^n}{n!} + \cdots\right] \\ &= 1 + \mathbb{E}[tX] + \mathbb{E}\left[\frac{(tX)^2}{2!}\right] + \mathbb{E}\left[\frac{(tX)^3}{3!}\right] + \cdots + \mathbb{E}\left[\frac{(tX)^n}{n!}\right] + \cdots \\ &= 1 + t\mathbb{E}[X] + \frac{t^2}{2!}\mathbb{E}[X^2] + \frac{t^3}{3!}\mathbb{E}[X^3] + \cdots + \frac{t^n}{n!}\mathbb{E}[X^n] + \cdots. \end{aligned} \tag{1.92}$$

Note that $M_X(0) = 1$ for all the distributions. Taking the derivative of $M_X(t)$ with respect to $t$, we obtain

$$\begin{aligned} \frac{d}{dt}M_X(t) &= M_X'(t) \\ &= \mathbb{E}[X] + t\mathbb{E}[X^2] + \frac{t^2}{2!}\mathbb{E}[X^3] + \frac{t^3}{3!}\mathbb{E}[X^3] + \cdots + \frac{t^{n-1}}{(n-1)!}\mathbb{E}[X^n] + \cdots. \end{aligned} \tag{1.93}$$

Evaluating this derivative at $t = 0$, all terms except $\mathbb{E}[X]$ become zero. We have

$$M_X'(0) = \mathbb{E}[X]. \tag{1.94}$$

Or,

$$\begin{aligned} M_X'(t) &= \frac{d}{dt}\mathbb{E}[e^{tX}] \\ &= \mathbb{E}\left[\frac{d}{dt}e^{tX}\right] \\ &= \mathbb{E}[Xe^{tX}]. \end{aligned} \tag{1.95}$$

Hence, $M_X'(0) = \mathbb{E}[X]$. Similarly, taking the second derivative of $M_X(t)$, we obtain

$$M_X''(0) = \mathbb{E}[X^2], \tag{1.96}$$

where,

$$\begin{aligned} M_X''(t) &= \frac{d}{dt}M_X'(t) \\ &= \frac{d}{dt}\mathbb{E}[Xe^{tX}] \\ &= \mathbb{E}\left[\frac{d}{dt}(Xe^{tX})\right] = \mathbb{E}[X^2e^{tX}], \end{aligned} \tag{1.97}$$

and so $M_X''(0) = \mathbb{E}[X^2]$. Continuing in this manner, from the $n$th derivative $M_X^{(n)}(t)$ with respect to $t$, we obtain all the moments to be

$$M_X^{(n)}(0) = \mathbb{E}[X^n], \qquad n = 1,2,3,\dots. \tag{1.98}$$

We summarize these calculations in the following theorem.

**Theorem 1.10:** If $M_X(t)$ exists, then for any positive integer $k$, [5]

$$\left.\frac{d^k}{dt^k}M_X(t)\right|_{t=0} = M_X^{(k)}(0) = \mathbb{E}\left[X^{(k)}\right]. \tag{1.99}$$

The usefulness of the foregoing theorem lies in the fact that, if the MGF can be found, the often difficult process of summation involved in calculating different moments can be replaced by the much easier process of differentiation.

## 1.10 Discrete Distributions

### 1.10.1 Binomial Distribution

A trial with only two possible outcomes is used so frequently as a building block of a random experiment that it is called a Bernoulli trial. It is usually assumed that the trials that constitute the random experiment are independent. This implies that the outcome from one trial has no effect on the outcome to be obtained from any other trial. Furthermore, it is often reasonable to assume that the probability of a success in each trial is constant. The binomial distribution with parameters $n$ and $p$ is the discrete probability distribution of the number of successes in a sequence of $n$ independent experiments, each asking a yes–no question, and each with its own Boolean-valued outcome: success (with probability $p$) or failure (with probability $q = 1 - p$).

**Definition (Binomial Distribution):** The binomial distribution has certain conditions that need to be met for its application. A random experiment consists of $n$ Bernoulli trials such that
(1) The trials are independent.
(2) Each trial results in only two possible outcomes, labeled as "success" and "failure."
(3) The probability of a success in each trial, denoted as $p$, remains constant.

If $p$ is the probability that an event will happen in any single trial (called the probability of a success) and $q = 1 - p$ is the probability that it will fail to happen in any single trial (called the probability of a failure), then the probability that the event will happen exactly $x$ times in $n$ trials (i.e., $x$ successes and $n - x$ failures will occur) is given by (the PMF of $X$ is) [12]

$$f_X(x) = \binom{n}{x} p^x (1-p)^{n-x},\ x = 0,1,\dots,n. \tag{1.100}$$

**Remarks:**

- In $n$ trials if we are getting $x$ successes, then there will be $n - x$ failures. Since the trials are independent and $p$ is same in all trials, probability of getting $x$ successes is $p \times p \times \ldots \times p$ ($x$ times) $= p^x$ and probability of getting $n - x$ failures is $q \times q \times \ldots \times q$ ($n - x$ times) $= q^{n-x}$. Hence, the probability of getting $x$ successes and $n - x$ failures is $p^x q^{n-x}$. The number of ways in which $x$ successes can occur in $n$ trials is $n!/x!(n-x)! = \binom{n}{x}$. Hence, the probability of getting $x$ successes in $n$ trials in any order is given by, $\binom{n}{x} p^x q^{n-x}$. This probability distribution of the RV $X$ is denoted by $X \sim B(n,p)$, see Figure 1.15.
- This discrete probability distribution is often called the binomial distribution since for $x = 0, 1, 2, \dots, n$ it corresponds to successive terms of the binomial formula, or binomial expansion,

  $$(q+p)^n = q^n + \binom{n}{1} q^{n-1}p^1 + \binom{n}{2} q^{n-2}p^2 + \cdots + p^n.$$

  For example,

  $$(q+p)^4 = q^4 + \binom{4}{1} q^3p^1 + \binom{4}{2} q^2p^2 + \binom{4}{3} q^1p^3 + p^4 = q^4 + 4q^3p^1 + 6q^2p^2 + 4q^1p^3 + p^4.$$

- If $n = 1$, the binomial RV reduces to Bernoulli RV, denoted by $B(1,p)$.
- If $X \sim B(n,p)$, then $\sum_{x=0}^{n} f_X(x) = \sum_{x=0}^{n} \binom{n}{x} p^x q^{n-x} = (q+p)^n = 1$.
- Let $X \sim B(n,p)$. Then $X$ gives number of success in $n$ independent trials with probability $p$ for success in each trial. Note that $n - X$ gives number of failures in $n$ independent trials with probability $1 - p = q$ for failure in each trial. Therefore, $n - X \sim B(n,q)$.

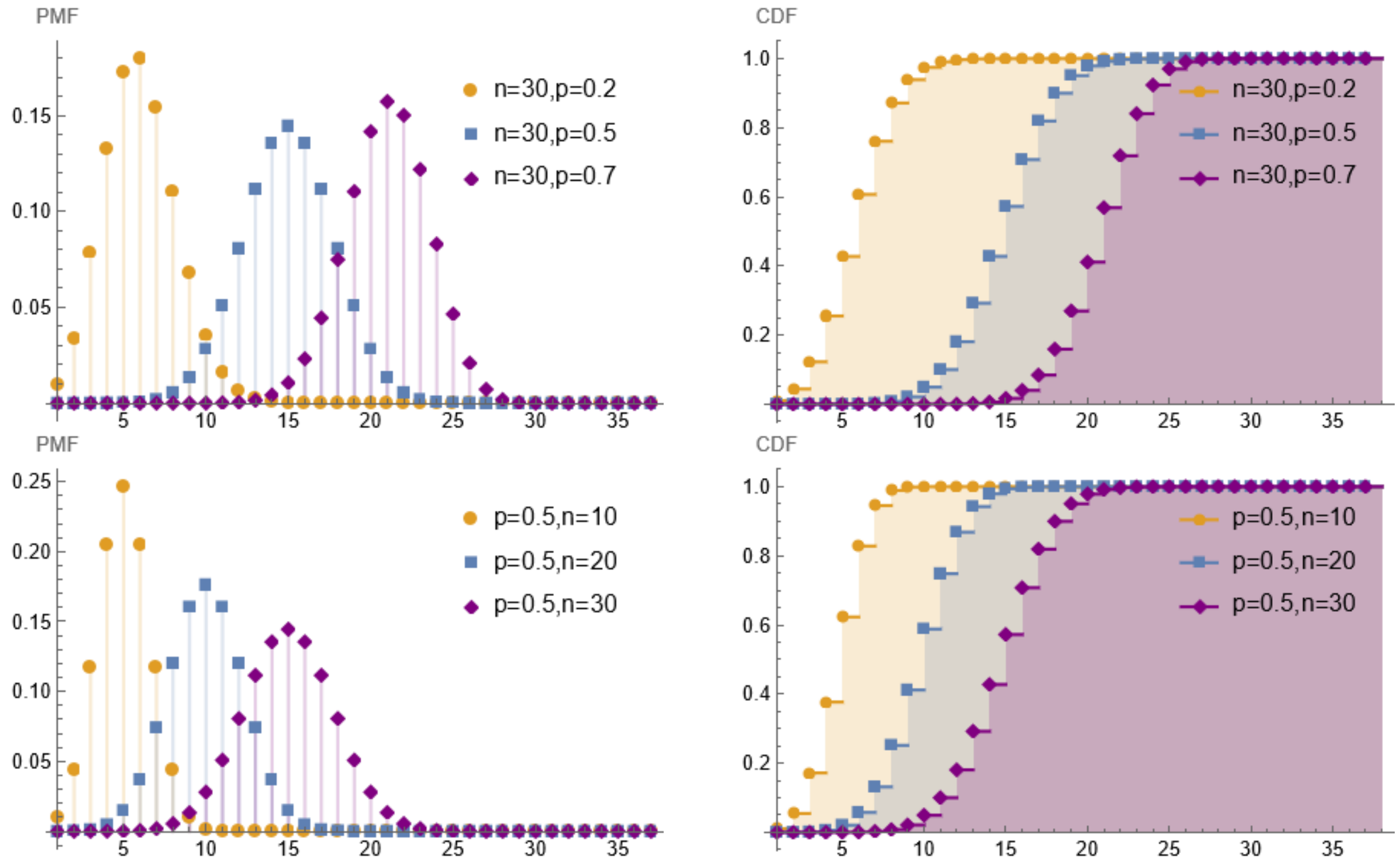

**Figure 1.15.** PMFs (left panels) and CDFs (right panels) of the binomial distribution. The curve of the binomial-distribution depends on the two parameters $p$ and $n$.

**Theorem 1.11:** If $X \sim B(n,p)$, $Y \sim B(m,p)$ and $X$ and $Y$ are independent [8],

$$\mu = \mathbb{E}[(X)] = np, \tag{1.101.1}$$
$$V(X) = nqp, \tag{1.101.2}$$
$$M_X(t) = (q + pe^t)^n, \tag{1.101.3}$$
$$X + Y \sim B(n + m, p). \tag{1.101.4}$$

### 1.10.2 Discrete Uniform Distribution

The discrete uniform distribution is a symmetric probability distribution wherein a finite number of values are equally likely to be observed; every one of $n$ values has equal probability $1/n$. A simple example of the discrete uniform distribution is throwing a fair dice. The possible values are 1, 2, 3, 4, 5, 6, and each time the die is thrown the probability of a given score is $1/6$. If two dice are thrown and their values added, the resulting distribution is no longer uniform because not all sums have equal probability.

#### Parameters of the discrete uniform distribution

- The parameters of the discrete uniform distribution are typically denoted as follows: $a$ represents the lowest value that the RV can take. $b$ represents the highest value that the RV can take. Simply, we have consecutive integers $\{a, a+1, a+2, \dots, b\}$. The PMF of the discrete uniform distribution assigns an equal probability to each value within the range. The PMF is defined as [1]:

  $$P(X = x) = \frac{1}{b - a + 1}, \qquad \text{for} \quad a \le x \le b,$$

  where $X$ is the RV, $x$ is a specific value within the range, and $P(X = x)$ represents the probability of $X$ taking the value $x$.

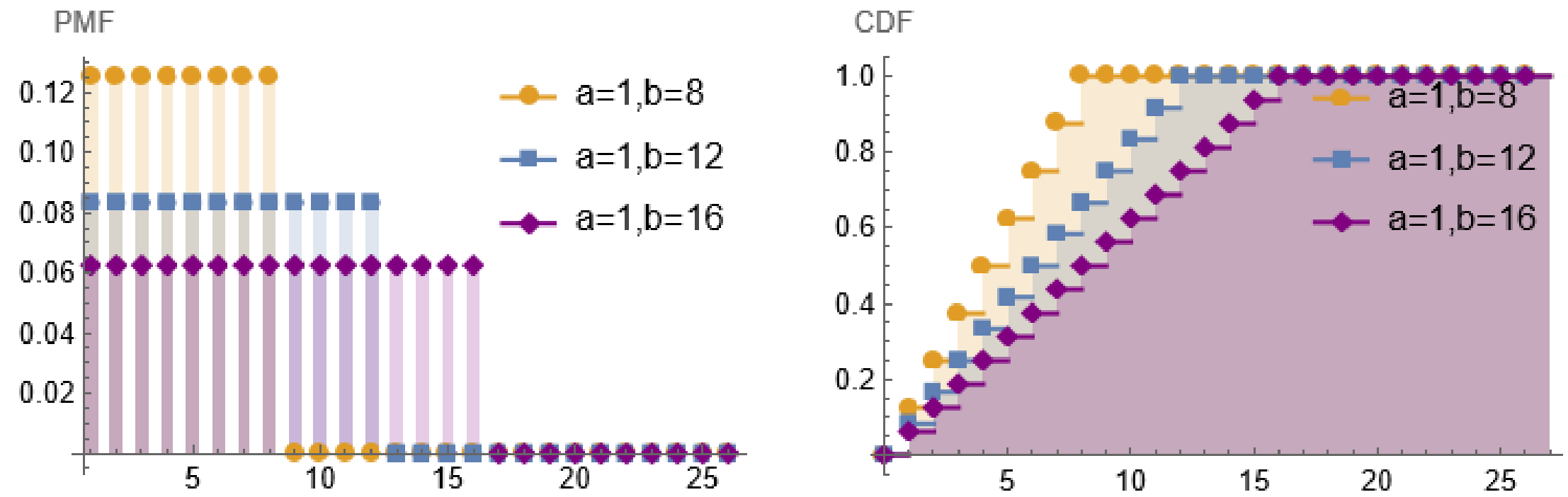

**Figure 1.16.** PMFs (left panel) and CDFs (right panel) of the discrete uniform distribution. The curve of the discrete uniform distribution depends on the parameters $a$ and $b$.

- In some cases, the discrete uniform distribution is defined using the parameter $n$, which represents the total number of possible outcomes or elements in the range. It is important to note that when using $n$ as the parameter, the range of outcomes is implicitly defined as $\{1,2,\ldots,n\}$, and each outcome has an equal probability of $1/n$. The minimum value $a$ can be defined as 1, representing the lowest value in the range. The maximum value $b$ can be defined as $n$, representing the highest value in the range. Using these parameters, the PMF of the discrete uniform distribution becomes:

$$P(X=x)=\frac{1}{n}, \qquad \text{for} \quad 1\le x\le n.$$

**Definition (Discrete Uniform Distribution):** A RV $X$ has a discrete uniform distribution if each of the $n$ values in its range, $\{x_1,x_2,\ldots,x_n\}$, has equal probability (see Figure 1.16). Its PMF is of the form [8],

$$f_X(x)=\begin{cases}\frac{1}{n}; & x=1,2,\ldots n\\ 0; & \text{otherwise.}\end{cases} \tag{1.102.1}$$

Or

$$f_X(x)=\begin{cases}\frac{1}{b-a+1}; & x=a,a+1,a+2,\ldots b,\\ 0; & \text{otherwise.}\end{cases} \tag{1.102.2}$$

**Theorem 1.12:** Suppose that $X$ is a discrete uniform RV on the consecutive integers $\{a,a+1,a+2,\ldots,b\}$, for $a\le b$. The mean of $X$ is [8]

$$\mu=\mathbb{E}[X]=\frac{b+a}{2}. \tag{1.103.1}$$

The variance of $X$ is

$$\sigma^2=\frac{(b-a+1)^2-1}{12}. \tag{1.103.2}$$

Suppose that $X$ is a discrete uniform RV on the consecutive integers $\{1,2,\ldots,n\}$ i.e., in this case $a=1$ and $b=n$. The mean of $X$ is

$$\mathbb{E}[X]=\frac{n+1}{2}. \tag{1.103.3}$$

The variance of $X$ is

$$\text{Var}(X)=\frac{n^2-1}{12}. \tag{1.103.4}$$

The MGF is

$$M_X(t)=\frac{e^t}{n}\frac{e^{nt}-1}{e^t-1}. \tag{1.103.5}$$

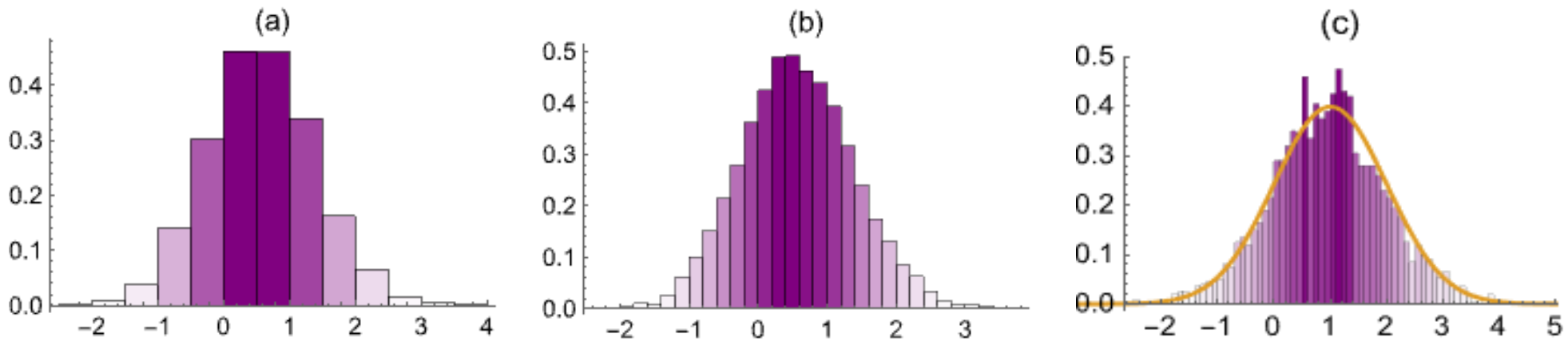

**Figure 1.17.** Relative frequency histograms for increasingly large sample sizes.

## 1.11 Continuous Random Variables

The possible values that a continuous RV can assume are infinite and uncountable. For example, the variable that represents the time taken by a worker to commute from home to work is a continuous RV. Suppose 6 minutes is the minimum time and 120 minutes is the maximum time taken by all workers to commute from home to work. Let $X$ be a continuous RV that denotes the time taken to commute from home to work by a randomly selected worker. Then $X$ can assume any value in the interval 6 to 120 minutes. This interval contains an infinite number of values that are uncountable.

Suppose you have a set of measurements on a continuous RV, and you create a relative frequency histogram to describe their distribution. For a small number of measurements, you could use a small number of classes; then as more and more measurements are collected, you can use more classes and reduce the class width. The outline of the histogram will change slightly, for the most part becoming less and less irregular, as shown in Figure 1.17.a. As the number of measurements becomes very large and the class widths become very narrow, the relative frequency histogram appears more and more like the smooth curve shown in Figure 1.17.c. This smooth curve describes the probability distribution of the continuous RV.

A Probability Density Function (PDF), $f(x)$, can be used to describe the probability distribution of a continuous RV $X$. If an interval is likely to contain a value for $X$, its probability is large, and it corresponds to large values for $f(x)$. A histogram is an approximation to a PDF. For each interval of the histogram, the area of the bar equals the relative frequency (proportion) of the measurements in the interval. The relative frequency is an estimate of the probability that a measurement falls in the interval. Similarly, the area under $f(x)$ over any interval equals the true probability that a measurement falls in the interval.

Remember that, for discrete RVs,

- The sum of all the probabilities $P(x)$ equals 1 and
- The probability that $X$ falls into a certain interval is the sum of all the probabilities in that interval.

Continuous RVs have some parallel characteristics listed next.

- The area under a continuous probability distribution is equal to 1.
- The probability that $X$ will fall into a particular interval—say, from $a$ to $b$—is equal to the area under the curve between the two points $a$ and $b$. Hence, the probability that $X$ is between $a$ and $b$ is determined as the integral of $f(x)$ from $a$ to $b$.

**Definition (Probability Density Function, PDF):** For a continuous RV $X$, a PDF is a function such that [5]

$$f(x) \geq 0, \tag{104.1}$$

$$\int_{-\infty}^{\infty} f(x)dx = 1, \tag{104.2}$$

$$P(a \leq X \leq b) = \int_a^b f(x)dx = \text{area under } f(x) \text{ from } a \text{ to } b \text{ for any } a \text{ and } b. \tag{104.3}$$

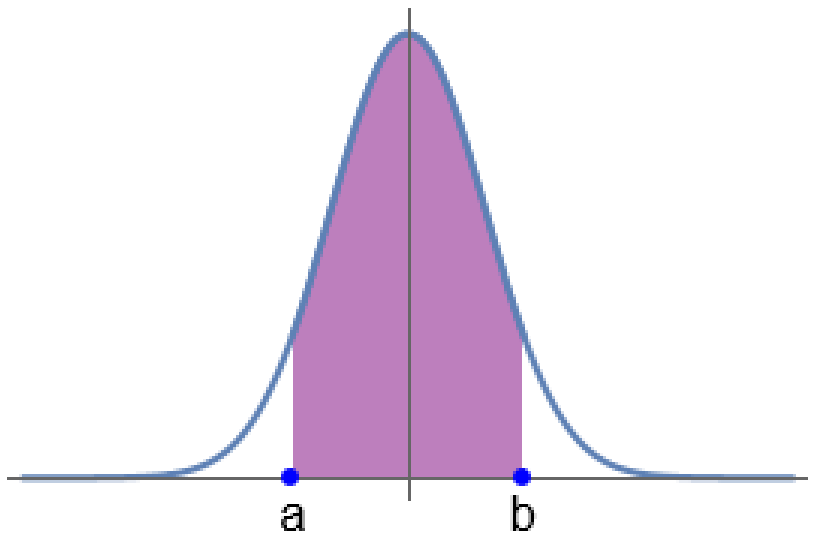

**Figure 1.18.** Probability as an area under a curve.

A PDF provides a simple description of the probabilities associated with a RV. As long as $f(x)$ is nonnegative and $\int_{-\infty}^{\infty} f(x)dx = 1$, $0 \leq P(a < X < b) \leq 1$ so that the probabilities are properly restricted. A PDF is zero for $x$ values that cannot occur (See Figure 1.18).

There is an important difference between discrete and continuous RVs. Consider the probability that $X$ equals some particular value, say, $a$. Since there is no area above a single point, $X = a$, in the probability distribution for a continuous RV, our definition implies that the probability is always zero. For example, if $X$ is the height of a randomly selected female student from a university, then the probability that this student is exactly 66.8 inches tall is zero; that is, $P(X = 66.8) = 0$.

- For a continuous RV $X$ and any value $a$, $P(X = a) = \int_a^a f(x)dx = 0$.
- This implies that
$$P(x \geq a) = P(x > a) \quad \text{and} \quad P(x \leq a) = P(x < a).$$
- If $X$ is a continuous RV, for any $a$ and $b$,
$$P(a \leq X \leq b) = P(a < X \leq b) = P(a \leq X < b) = P(a < X < b).$$
- In other words, the probability that $X$ assumes a value in the interval $a$ to $b$ is the same whether or not the values $a$ and $b$ are included in the interval. For the example on the heights of female students, the probability that a randomly selected female student is between 65 and 68 inches tall is the same as the probability that this female is 65 to 68 inches tall.

**Definition (Cumulative Distribution Function, CDF):** The CDF of a continuous RV $X$ is [5]
$$F(x) = P(X \leq x) = \int_{-\infty}^{x} f(u)du, \tag{1.105.1}$$
for $-\infty < x < \infty$. Given $F(x)$,
$$f(x) = \frac{dF(x)}{dx}, \tag{1.105.2}$$
as long as the derivative exists. That is, the density is the derivative of the CDF.
The importance of the CDF here, just as for discrete RV, is that probabilities of various intervals can be computed from a formula for $F(x)$. If we have the CDF $F(x)$, then
$$P(a \leq X \leq b) = F(b) - F(a). \tag{1.105.3}$$

The mean and variance can also be defined for a continuous RV. Integration replaces summation in the discrete definitions.

**Definition (Mean and Variance):** Suppose that $X$ is a continuous RV with PDF $f(x)$. The mean or expected value of $X$, denoted as $\mu$ or $\mathbb{E}[X]$, is [8]
$$\mu = \mathbb{E}[X] = \int_{-\infty}^{\infty} xf(x)dx. \tag{1.106}$$
The variance of $X$, denoted as $V(X)$ or $\sigma^2$, is
$$\sigma^2 = V(X) = \int_{-\infty}^{\infty} (x-\mu)^2 f(x)dx = \int_{-\infty}^{\infty} x^2 f(x)dx - \mu^2. \tag{1.107}$$
The standard deviation of $X$ is $\sigma = \sqrt{\sigma^2}$.

**Definition (Expected Value):** The expected value of a function $h(X)$ of a continuous RV $X$ with PDF $f(x)$ is defined as

$$\mathbb{E}[h(X)] = \int_{-\infty}^{\infty} h(x)f(x)dx. \tag{1.108}$$

Let us consider the example of the height of adult males in a given population. The height of an individual can be considered a continuous RV because it can take on any value within a certain range (e.g., from 150 cm to 200 cm), and there are infinitely many possible values between any two given heights.

- Let us say we have a population of 10,000 adult males, and we want to study their heights. We can represent the height of each individual as a continuous RV, denoted by the symbol "$X$." In this example, $X$ can take on any value between 150 cm and 200 cm, including decimal values.
- To analyze this continuous RV, we can look at its probability distribution, which describes how likely it is for $X$ to take on different values within the given range. In this case, the probability distribution of the height of adult males might resemble a bell-shaped curve, also known as a normal distribution.
- Using a normal distribution, we can calculate the probability of an individual having a height within a certain range. For example, we can calculate the probability of an adult male being between 170 cm and 180 cm tall. This calculation involves integrating the PDF of the normal distribution over the interval from 170 cm to 180 cm.
- Moreover, continuous RVs allow us to calculate various summary statistics. For instance, we can determine the mean height of the population, $\mu$, which represents the average height of adult males in the given population. We can also calculate the standard deviation, $\sigma$, which measures the variability or spread of heights in the population.

**Definition (MGF):** The MGF of the continuous RV $X$ with PDF $f(x)$ is defined as

$$M_X(t) = \mathbb{E}[e^{tX}] = \int_{-\infty}^{\infty} e^{tx}f(x)dx. \tag{1.109}$$

**Definition (Properties of the MGF):**
1. The MGF of $X$ is unique in the sense that, if two RVs $X$ and $Y$ have the same MGF ($M_X(t) = M_Y(t)$, for $t$ in an interval containing 0), then $X$ and $Y$ have the same distribution.
2. If $X$ and $Y$ are independent, then

$$M_{X+Y} = M_X(t)M_Y(t). \tag{1.110}$$

That is, the MGF of the sum of two independent RVs is the product of the MGFs of the individual RVs. The result can be extended to $n$ RVs.
3. Let $Y = aX + b$. Then

$$M_Y(t) = E\left[e^{(aX+b)t}\right] = E[e^{atX+bt}] = E[e^{bt}e^{atX}] = e^{bt}M_X(at). \tag{1.111}$$

## 1.12 Continuous Distributions

### 1.12.1 Continuous Uniform or Rectangular Distribution

One of the simplest continuous distributions in all of statistics is the continuous uniform distribution. This distribution is characterized by a density function that is "flat," and thus the probability is uniform in a closed interval, say $[a, b]$. It is a family of symmetric probability distributions. The distribution describes an experiment where there is an arbitrary number outcome that lies between certain bounds. The bounds are defined by the parameters, $a$ and $b$. The interval can be either closed (eg. $[a, b]$) or open (eg. $(a, b)$). Therefore, the distribution is often abbreviated $\mathrm{U}(a, b)$. The difference between the bounds defines the interval length; all intervals of the same length on the distribution's support are equally probable.

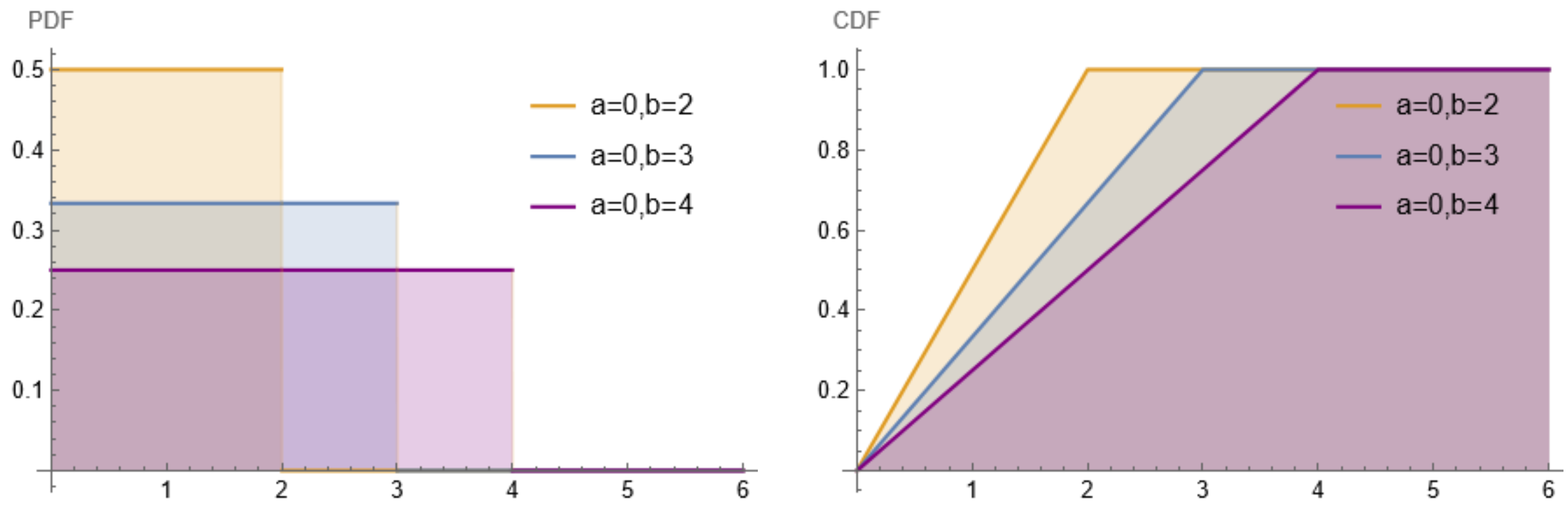

**Figure 1.19.** PMF (left panel) and CDF (right panel) of continuous uniform distribution. The curve of the continuous uniform distribution depends on the two parameters $a$ and $b$.

**Definition (PDF of Continuous Uniform Distribution):** A RV $X$ is said to have a continuous uniform distribution over an interval $[a, b]$ if its PDF is given by [1]

$$f_X(x) = \begin{cases} \dfrac{1}{b-a}; & a \le x \le b, \\ 0; & \text{otherwise.} \end{cases} \tag{1.112}$$

**Definition (CDF of Uniform Continuous Distribution):** The CDF $F(x)$ of $\mathrm{U}(a, b)$ is given by [1]

$$F_X(x) = \begin{cases} 0 \quad ; & x < a\,, \\ \dfrac{x-a}{b-a}; & a \le x \le b, \\ 1 \quad ; & x > b. \end{cases} \tag{1.113}$$

Since $F_X(x)$ is not continuous at $x = a$ and $x = b$, it is not differentiable at these points. Thus

$$\frac{d}{dx}F(x) = f(x) = \frac{1}{b-a} \neq 0,$$

exists everywhere except at the points $x = a$ and $x = b$ and consequently we get the PDF $f(x)$, (see Figure 1.19).

**Remarks:**

- The continuous uniform distribution requires specifying the interval within which the RV can take on values. This interval is usually defined by two parameters, a lower bound and an upper bound, which determine the range of possible outcomes. The width of the interval affects the spread and variability of the distribution.
- The continuous uniform distribution can be seen as the continuous counterpart of the discrete uniform distribution. While the discrete uniform distribution assigns equal probabilities to a finite number of discrete values, the continuous uniform distribution assigns equal probabilities to an infinite number of values within a continuous interval.
- The PDF of a continuous uniform distribution is constant within the interval and zero outside the interval. This reflects the uniformity of the distribution, as there are no peaks or valleys in the probability density.
- The distribution is known as rectangular distribution, since the curve $y = f(x)$ describes a rectangle over the $x$-axis and between the ordinates at $x = a$ and $x = b$.
- The CDF of a continuous uniform distribution is a linear function that increases uniformly from 0 to 1 as the value of the RV ranges from the lower bound to the upper bound. This property makes it easy to calculate probabilities and percentiles associated with specific values.

**Theorem 1.13:** If $X$ is a continuous uniform RV over $a \leq x \leq b$, [1]

$$\mathbb{E}[X] = \frac{a+b}{2}, \quad (1.114.1)$$

$$V(X) = \frac{(b-a)^2}{12}, \quad (1.114.2)$$

$$M_X(t) = \frac{e^{bt} - e^{at}}{t(b-a)}. \quad (1.114.3)$$

### 1.12.2 Normal Distribution

Normal distribution plays a pivotal role in most of the statistical techniques used in applied statistics. The main reason for this is the central limit theorem, according to which normal distribution is found to be the approximation of most of the RVs.

It was first introduced by a French mathematician, Abraham De-Moivre (1667-1754). He obtained it while working on certain problems in the games of chance. Later, two mathematical astronomers Pierre Laplace (1749-1827) and Karl Gauss (1777-1855) developed this distribution independently. They found that it can be used to model errors (the deviation of the observed value from the true value). Hence, this distribution is also known as Gaussian distribution and Laplace's distribution. But it is most commonly known as the normal distribution.

**Definition (Normal Distribution):** A RV $X$ is said to follow a normal distribution with parameters $\mu$ and $\sigma^2$ if its PDF is given by [5]

$$f_X(x) = \frac{1}{\sigma\sqrt{2\pi}} e^{-\frac{(x-\mu)^2}{2\sigma^2}}, \qquad -\infty < x < \infty, \quad (1.115)$$

where $-\infty < \mu < \infty$ and $\sigma > 0$. In this case, we can write $X \sim N(\mu, \sigma^2)$ or $X \sim N(\mu, \sigma)$.

For a normal distribution with density $f$, mean $\mu$ and deviation $\sigma$, the CDF is given by,

$$F_X(x) = \frac{1}{2}\left(1 + \text{Erf}\left(\frac{x-\mu}{\sigma\sqrt{2}}\right)\right), \quad (1.116)$$

where, the error function $\text{Erf}(x)$ is

$$\text{Erf}(x) = \frac{2}{\sqrt{\pi}} \int_0^x e^{-t^2} dt. \quad (1.117)$$

Given the values of the mean, $\mu$, and the standard deviation, $\sigma$, we can find the area under a normal distribution curve for any interval. The value of $\mu$ determines the center of a normal distribution curve on the horizontal axis, and the value of $\sigma$ gives the spread of the normal distribution curve. The three normal distribution curves shown in Figure 1.20 (upper panel) have the same mean but different standard deviations. By contrast, the three normal distribution curves in Figure 1.20 (lower panel) have different means but the same standard deviation.

We list the following properties of the normal curve:

1. The normal curve is symmetrical about the ordinate at $x = \mu$, i.e., $f(\mu + c) = f(\mu - c)$ for any $c$.
2. 50% of the total area under a normal distribution curve lies on the left side of the mean, and 50% lies on the right side of the mean.
3. The mean, median, and mode are identical and occur at $x = \mu$.
4. The mode of the normal curve is at $x = \mu$, and is equal to $\frac{1}{\sigma\sqrt{2\pi}}$.
5. It is unimodal: its first derivative is positive for $x < \mu$ negative for $x > \mu$ and zero only at $x = \mu$.
6. The normal curve extends from $-\infty$ to $+\infty$.
7. In graphical form, normal distribution will appear as a bell curve.
8. For a normal distribution $\beta_1 = 0$ (i.e., symmetric) and $\beta_2 = 3$ (i.e., mesokurtic).
9. $x$-axis is an asymptote to the curve. That is, the curve touches the $x$-axis only at $\pm\infty$.

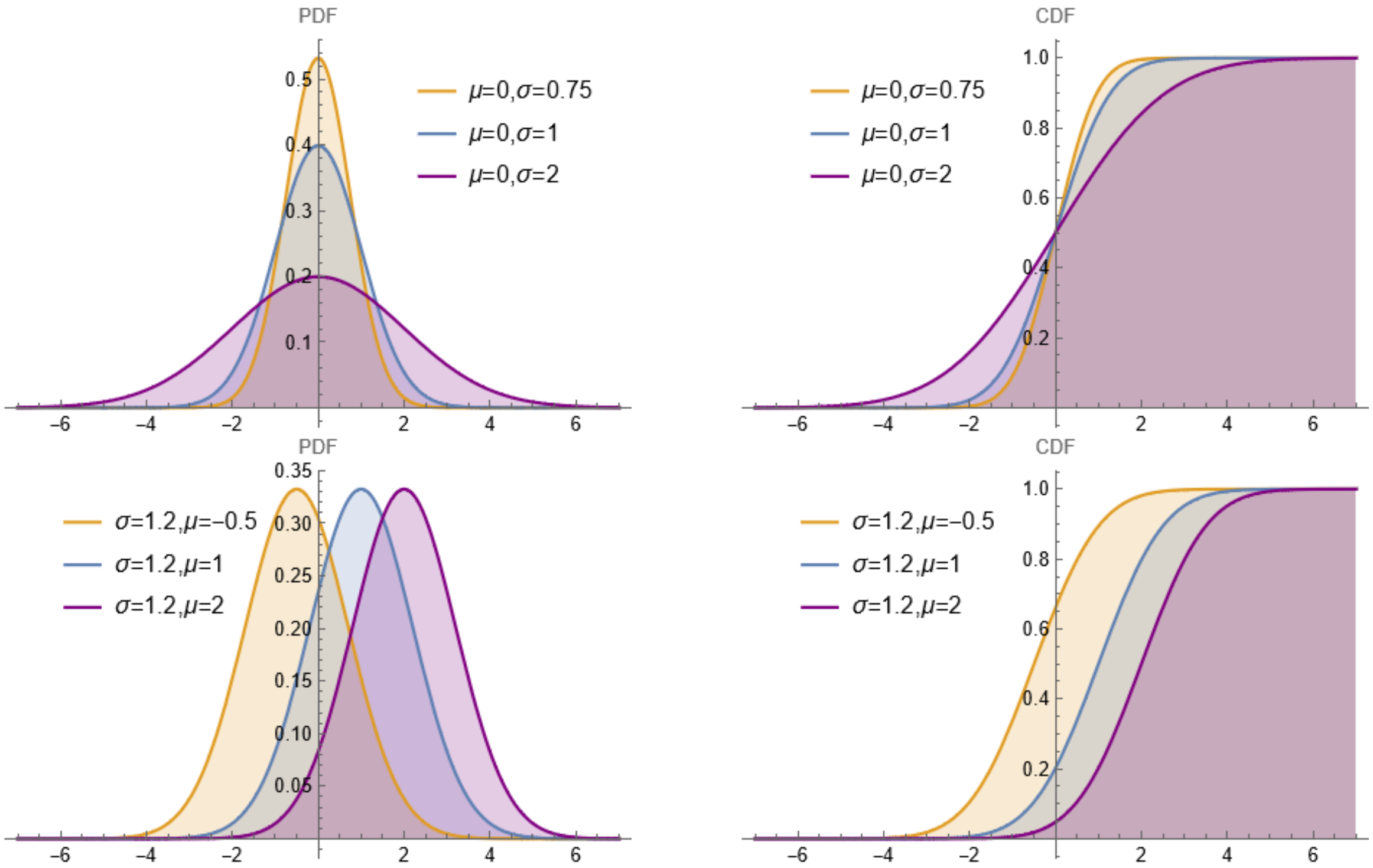

**Figure 1.20.** PMF (left panel) and CDF (right panel) of normal distribution. The curve of the normal distribution depends on the two parameters $\mu$ and $\sigma$.

10. All odd-order central moments are zero. i.e.,
$$\mu_{2r+1} = 0, \qquad r = 0,1,2,\ldots.$$
11. Even order central moments are given by
$$\mu_{2r} = 1.3.5\ldots(2r - 1)\sigma^{2r}, \qquad r = 0,1,2,\ldots.$$
12. The points of inflection of the curve are $x = \mu \pm \sigma$
13. The lower and upper quartiles are equidistant from the median.
14. The area under the normal curve is distributed as:
    (a) 68.27% of the items lie between $\mu - \sigma$ and $\mu + \sigma$.
    $$\text{i.e.}, P(\mu - \sigma \leq X \leq \mu + \sigma) = 0.6827.$$
    (b) 95.45% of the items lie between $\mu - 2\sigma$ and $\mu + 2\sigma$.
    $$\text{i.e.}, P(\mu - 2\sigma \leq X \leq \mu + 2\sigma) = 0.9545.$$
    (c) 99.73% of the items lie between $\mu - 3\sigma$ and $\mu + 3\sigma$.
    $$\text{i.e.}, P(\mu - 3\sigma \leq X \leq \mu + 3\sigma) = 0.9973.$$
    (d) The total area under the curve and above the horizontal axis is equal to 1.

**Theorem 1.14:** The mean and variance of $X \sim N(\mu,\sigma)$ are
$$\mathbb{E}[X] = \mu, \qquad (1.118.1)$$
$$V(X) = \sigma^2. \qquad (1.118.2)$$

**Theorem 1.15:** The MGF of normal distribution $N(\mu,\sigma)$ is given by,
$$M_X(t) = e^{\mu t + \frac{1}{2}t^2\sigma^2}. \qquad (1.119)$$

**Theorem 1.16 (Additive Property):** If $X_1 \sim N(\mu_1,\sigma_1^2)$, $X_2 \sim N(\mu_2,\sigma_2^2)$ and if $X_1$ and $X_2$ are independent, then,
$$X_1 + X_2 \sim N(\mu_1 + \mu_2, \sigma_1^2 + \sigma_2^2). \qquad (1.120)$$

**Theorem 1.17 (Additive Property):** If $X_i$, $i = 1,2,\dots,n$ are $n$ independent normal variates with mean $\mu_i$ and variance $\sigma_i^2$ respectively, then $Y = \sum_{i=1}^{n} X_i$ is normally distributed with mean $\sum_{i=1}^{n} \mu_i$ and variance $\sum_{i=1}^{n} \sigma_i^2$ [5].

$$Y = \sum_{i=1}^{n} X_i \sim N\left(\sum_{i=1}^{n} \mu_i , \sum_{i=1}^{n} \sigma_i^2\right). \tag{1.121}$$

**Theorem 1.18:** If $X_i$, $i = 1,2,\dots,n$ are $n$ independent normal variates with mean $\mu_i$ and variance $\sigma_i^2$ respectively, then their linear combination, $Y = \sum_{i=1}^{n} a_i X_i$, is normally distributed with mean $\sum_{i=1}^{n} a_i \mu_i$ and variance $\sum_{i=1}^{n} a_i^2 \sigma_i^2$ where $a_i$'s are constants.

**Theorem 1.19:** If $X$ is normal RV with mean $\mu$ and variance $\sigma^2$, then for any constants $a$ and $b$, $b \neq 0$, the RV $Y = a + bX$ is also a normal RV with parameters,

$$\mathbb{E}[Y] = \mathbb{E}[a + bX] = a + b\mathbb{E}[X] = a + b\mu, \tag{1.122.1}$$

and variance

$$V(Y) = V(a + bX) = b^2 V(X) = b^2 \sigma^2. \tag{1.122.2}$$

From (1.122.1) and (1.122.2), if $b = \frac{1}{\sigma}$ and $a = -\frac{\mu}{\sigma}$, then

$$Y = a + bX = \frac{X - \mu}{\sigma} = Z,$$

is normal RV with mean $-\frac{\mu}{\sigma} + \frac{1}{\sigma}\mu = 0$ and variance, $\left(\frac{1}{\sigma}\right)^2 \sigma^2 = 1$.

**Theorem 1.20 (Standardizing a Normal RV):** If $X$ is a normal RV with $\mathbb{E}[X] = \mu$ and $V(X) = \sigma^2$, the RV [8]

$$Z = \frac{X - \mu}{\sigma}, \tag{1.123}$$

is a normal RV with $\mathbb{E}[Z] = 0$ and $V(Z) = 1$. $Z$ is called a standard normal RV. We write $Z \sim N(0,1)$.

Creating a new RV by this transformation is referred to as standardizing. The RV $Z$ represents the distance of $X$ from its mean in terms of standard deviations. It is the key step to calculating a probability for an arbitrary normal RV.

**Definition (Standard Normal Distribution):** A RV $Z$ is said to follow standard normal distribution if its PDF is given by

$$f_Z(z) = \frac{1}{\sqrt{2\pi}} e^{-\frac{z^2}{2}}, \qquad -\infty < z < \infty. \tag{1.124}$$

We can see that

$$\mathbb{E}[Z] = \mathbb{E}\left[\frac{X - \mu}{\sigma}\right] = 0, \qquad V(Z) = V\left(\frac{X - \mu}{\sigma}\right) = 1, \tag{1.125}$$

$$M_Z(t) = e^{\frac{t^2}{2}}. \tag{1.126}$$

Suppose $X \sim N(\mu, \sigma^2)$ and we are interested in finding the probability of the variate $X$ lying between two values, say, $a$ and $b$. To determine this, we first make the transformation $Z = \frac{X-\mu}{\sigma}$.

**Definition (Standardizing to Calculate a Probability):** Suppose that $X$ is a normal RV with mean $\mu$ and variance $\sigma^2$. Then, [8]

$$P(X \leq x) = P\left(\frac{X - \mu}{\sigma} \leq \frac{x - \mu}{\sigma}\right) = P(Z \leq z), \tag{1.127}$$

where $Z$ is a standard normal RV, and $z = \frac{x-\mu}{\sigma}$ is the $z$-value obtained by standardizing $X$.

Hence,

$$P(a < X < b) = P\left(\frac{a - \mu}{\sigma} < \frac{X - \mu}{\sigma} < \frac{b - \mu}{\sigma}\right) = P(z_1 < Z < z_2), \tag{1.128}$$

where $z_1 = \frac{a-\mu}{\sigma}$ and $z_2 = \frac{b-\mu}{\sigma}$. Therefore, $P(a < X < b)$ is the area under the standard normal curve between the abscissae $z_1$ and $z_2$.

In conclusion, this chapter has provided a foundational understanding of the key statistical techniques and methodologies essential for data analysis and scientific research. For readers who wish to delve deeper into these topics, it is highly recommended to refer to the comprehensive range of books [1,16-28]. These books offer an extensive exploration of statistical concepts, accompanied by practical implementations in four powerful and versatile programming languages: R, Mathematica, Julia, and Python. Each of these languages has its own strengths and unique features that cater to different aspects of statistical computing and data visualization. By studying these resources, readers can gain valuable insights into the theoretical underpinnings of statistical methods, as well as hands-on experience in applying these techniques to real-world problems across various domains. Whether you are a beginner looking to build a solid foundation or an experienced practitioner seeking to enhance your analytical toolkit, these books serve as an invaluable guide to mastering the art and science of statistics in today's data-driven world.

# CHAPTER 2

# MATRIX CALCULUS AND GRADIENT OPTIMIZATION

In the realm of AI and machine learning, the modern development of NNs stands as a testament to the marriage of sophisticated mathematical frameworks and computational ingenuity. At the heart of this evolution lies the profound influence of matrix calculus [29-37] and gradient optimization techniques [38-40]. These foundational pillars have not only reshaped the landscape of NN design but have also propelled advancements in diverse domains, ranging from computer vision to natural language processing and beyond.

Matrix calculus, with its roots in linear algebra, provides a powerful toolset for analyzing and manipulating multidimensional data structures. It offers a systematic framework for computing derivatives and gradients of functions involving matrices and vectors, enabling efficient optimization in high-dimensional spaces. Matrix calculus is indeed essential for building and training NNs. NNs, especially deep learning models, heavily rely on matrix operations for their computations.

Here's why matrix calculus is crucial for NNs:

- NNs are typically represented and implemented using matrices and vectors. Each layer in a NN can be seen as a matrix operation, where inputs (vectors) are multiplied by weights (matrices) and passed through functions (activation functions).
- The training of NNs often involves optimization algorithms like Gradient Descent (GD). Matrix calculus provides the necessary tools to compute gradients efficiently, enabling the optimization process to update the network parameters (weights) in the direction that minimizes the objective (loss) function.
- Backpropagation is the primary algorithm used to compute gradients efficiently in NNs. It is essentially an application of the chain rule from calculus, which involves matrix multiplication and transposition operations.
- Matrix calculus allows for efficient computation of derivatives and gradients in NNs. This efficiency is crucial for training deep NNs, which may have millions of parameters.
- Most deep learning frameworks handle much of the matrix calculus under the hood. However, understanding the underlying principles of matrix calculus can help in debugging, optimizing, and customizing NN architectures.

Complementing matrix calculus is the arsenal of gradient optimization techniques, which are fundamental to training NNs. By iteratively adjusting model parameters in the direction of steepest descent, these methods seek to minimize a predefined objective function, such as the loss function in supervised learning tasks. From classic algorithms like GD to more advanced variants like Stochastic Gradient Descent (SGD) and Adaptive Moment (Adam) optimization, these techniques play a pivotal role in navigating the vast landscape of model parameter space efficiently and effectively.

In this chapter, we delve into the mathematics of matrix calculus and gradient optimization, explore the algorithms driving their implementation, and examine their relationship in the landscape of modern NNs. Through theoretical insights, we aim to equip the reader with a deeper understanding of these foundational elements and their significance in shaping the landscape of NNs.

This chapter is a part of my book titled " Mathematics for Machine Learning and Data Science: Optimization with Mathematica Applications". For detailed proofs of theorems, additional examples, and comprehensive explanations, including Mathematica applications, please refer to Ref [29].

## 2.1 Vectors and Matrices with Index Notations

Throughout this book, we will distinguish scalars, vectors, and matrices by their typeface. We will let $M_{n\times m}$ denote the space of all real $n\times m$ matrices with $n$ rows and $m$ columns. Such matrices will be denoted using bold capital letters: **A**, **X**, **Y**, etc. An element of $M_{n\times 1}$ or $M_{1\times n}$, that is, a column or row vector, respectively, is denoted with a boldface lowercase letter: **a**, **x**, **y**, etc. An element of matrix **M** is a scalar, denoted with italic letter: $A$, $X$, $Y$, $a$, $x$, $y$, etc.

Let $A_{ij}\in\mathbb{R}$, $i=1,2,\dots,m$, $j=1,2,\dots,n$. The matrix is the ordered rectangular array

$$\mathbf{A}=\begin{pmatrix}A_{11} & A_{12} & \cdots & A_{1n}\\ A_{21} & A_{22} & \cdots & A_{2n}\\ \vdots & \vdots & \ddots & \vdots\\ A_{m1} & A_{m2} & \dots & A_{mn}\end{pmatrix}=A_{ij},\quad i=1,2,\dots,m;\ j=1,2,\dots,n. \tag{2.1}$$

Note the first subscript locates the row in which the typical element lies while the second subscript locates the column. For example, $A_{jk}$ denotes the element lying in the $j^{\text{th}}$ row and $k^{\text{th}}$ column of the matrix **A**.

**Definition:** Let **A** and **B** are $m\times n$ matrices and let the sum $\mathbf{A}+\mathbf{B}$ be $\mathbf{C}=\mathbf{A}+\mathbf{B}$, then **C** is an $m\times n$ matrix, with the element $(i,j)$ given by [29]

$$C_{ij}=A_{ij}+B_{ij}, \tag{2.2}$$

for all $i=1,2,\dots,m$, $j=1,2,\dots,n$.

**Definition:** Let **A** be an $m\times n$ matrix, and let $a$ be a scalar, then $\mathbf{C}=a\mathbf{A}$ is an $m\times n$ matrix, with the element $(i,j)$ given by

$$C_{ij}=aA_{ij}, \tag{2.3}$$

for all $i=1,2,\dots,m$, $j=1,2,\dots,n$.

**Definition:** Let **A** be an $m\times n$ matrix, then $\mathbf{A}^T$ is an $n\times m$ matrix (the transpose of **A**) with the element $(i,j)$ given by

$$(A^T)_{ij}=A_{ji}, \tag{2.4}$$

for all $i=1,2,\dots,m$, $j=1,2,\dots,n$.

***Example 2.1***

Let us consider,

$$\mathbf{A}=\begin{pmatrix}A_{11} & A_{12} & A_{13}\\ A_{21} & A_{22} & A_{23}\\ A_{31} & A_{32} & A_{33}\end{pmatrix}=A_{ij},\quad \mathbf{B}=\begin{pmatrix}B_{11} & B_{12} & B_{13}\\ B_{21} & B_{22} & B_{23}\\ B_{31} & B_{32} & B_{33}\end{pmatrix}=B_{ij}.$$

Hence, we have

$$(A+B)_{ij}=A_{ij}+B_{ij}\Rightarrow \mathbf{A}+\mathbf{B}=\begin{pmatrix}A_{11}+B_{11} & A_{12}+B_{12} & A_{13}+B_{13}\\ A_{21}+B_{21} & A_{22}+B_{22} & A_{23}+B_{23}\\ A_{31}+B_{31} & A_{32}+B_{32} & A_{33}+B_{33}\end{pmatrix},$$

$$(cA)_{ij}=cA_{ij}\Rightarrow c\mathbf{A}=\begin{pmatrix}cA_{11} & cA_{12} & cA_{13}\\ cA_{21} & cA_{22} & cA_{23}\\ cA_{31} & cA_{32} & cA_{33}\end{pmatrix},$$

$$(A^T)_{ij}=A_{ji}\Rightarrow \mathbf{A}^T=\begin{pmatrix}A_{11} & A_{21} & A_{31}\\ A_{12} & A_{22} & A_{32}\\ A_{13} & A_{23} & A_{33}\end{pmatrix}.$$

**Definition:** Let **A** and **B** are $m\times n$ and $n\times p$ matrices, respectively, and let the product **AB** be

$$\mathbf{C}=\mathbf{AB}, \tag{2.5}$$

then **C** is an $m\times p$ matrix, with the element $(i,j)$ given by [30]

$$C_{ij}=\sum_{k=1}^{n}A_{ik}B_{kj}, \tag{2.6}$$

for all $i=1,2,\dots,m$, $j=1,2,\dots,p$.

**Definition:** Let **A** be an $m\times n$ matrix, and **x** be $n\times 1$ vector, then the typical element of the product

$$\mathbf{z} = \mathbf{A}\mathbf{x}, \tag{2.7}$$

is given by

$$z_i = \sum_{k=1}^{n} A_{ik} x_k, \tag{2.8}$$

for all $i = 1,2, \dots, m$.

**Definition:** Let **A** be an $m \times n$ matrix, and **y** be an $m \times 1$ vector, then the typical element of the product

$$\mathbf{z}^T = \mathbf{y}^T \mathbf{A}, \tag{2.9}$$

is given by

$$z_i = \sum_{k=1}^{n} y_k A_{ki}, \tag{2.10}$$

for all $i = 1,2,\dots,n$.

**Definition:** Let **A** be an $m \times n$ matrix, **y** be an $m \times 1$ vector, and **x** be $n \times 1$ vector, then the scalar resulting from the product

$$\alpha = \mathbf{y}^T \mathbf{A}\mathbf{x}, \tag{2.11}$$

is given by

$$\alpha = \sum_{j=1}^{m} \sum_{k=1}^{n} y_j A_{jk} x_k. \tag{2.12}$$

**Definition:** The trace of an $n \times n$ square matrix **A** is defined as

$$\mathrm{tr}(\mathbf{A}) = \sum_{i=1}^{n} A_{ii} = A_{11} + A_{22} + \cdots + A_{nn}. \tag{2.13}$$

For example, let $\mathbf{A} = \begin{pmatrix} 1 & 2 & 3 \\ 4 & 5 & 6 \\ 7 & 8 & 9 \end{pmatrix}$, then $\mathrm{tr}(\mathbf{A}) = 1 + 5 + 9 = 15$.

**Theorem 2.1:**

1- The trace is a linear mapping. That is,

$$\mathrm{tr}(\mathbf{A} + \mathbf{B}) = \mathrm{tr}(\mathbf{A}) + \mathrm{tr}(\mathbf{B}), \tag{2.14}$$

$$\mathrm{tr}(c\mathbf{A}) = c\ \mathrm{tr}(\mathbf{A}), \tag{2.15}$$

$$\mathrm{tr}(\mathbf{A}) = \mathrm{tr}(\mathbf{A}^T), \tag{2.16}$$

for all square matrices **A** and **B**, and all scalars $c$.

2- If **A** and **B** are $m \times n$ and $n \times m$ real matrices, respectively, then

$$\mathrm{tr}(\mathbf{AB}) = \mathrm{tr}(\mathbf{BA}). \tag{2.17}$$

More generally, the trace is invariant under cyclic permutations, that is,

$$\mathrm{tr}(\mathbf{ABC}) = \mathrm{tr}(\mathbf{BCA}) = \mathrm{tr}(\mathbf{CAB}). \tag{2.18}$$

**Proof:**

$$\mathrm{tr}(\mathbf{A} + \mathbf{B}) = \sum_{i=1}^{n} A_{ii} + B_{ii} = \sum_{i=1}^{n} A_{ii} + \sum_{i=1}^{n} B_{ii} = \mathrm{tr}(\mathbf{A}) + \mathrm{tr}(\mathbf{B}),$$

$$\mathrm{tr}(c\mathbf{A}) = \sum_{i=1}^{n} cA_{ii} = c\sum_{i=1}^{n} A_{ii} = \mathrm{c}\ \mathrm{tr}(\mathbf{A}),$$

$$\mathrm{tr}(\mathbf{A}) = \sum_{i=1}^{n} A_{ii} = \mathrm{tr}(\mathbf{A}^T).$$

The trace of a matrix is the sum of its diagonal elements, but transposition leaves the diagonal elements unchanged.

$$\mathrm{tr}(\mathbf{AB}) = \sum_{i=1}^{n} \sum_{j=1}^{m} A_{ij} B_{ji} = \sum_{j=1}^{m} \sum_{i=1}^{n} B_{ji} A_{ij} = \mathrm{tr}(\mathbf{BA}),$$

$$\mathrm{tr}(\mathbf{ABC}) = \sum_{i=1}^{n}\sum_{j=1}^{m}\sum_{k=1}^{l} A_{ij}B_{jk}C_{ki} = \sum_{i=1}^{n}\sum_{j=1}^{m}\sum_{k=1}^{l} B_{jk}C_{ki}A_{ij} = \mathrm{tr}(\mathbf{BCA}).$$

■

**Some Types of Matrices** [32, 35]

**Definition (Identity Matrix):** The identity matrix of size $n$ is the $n \times n$ square matrix with ones on the main diagonal and zeros elsewhere.

$$\mathbf{I} = \begin{pmatrix} 1 & 0 & \dots & 0 \\ 0 & 1 & \dots & \vdots \\ \vdots & \vdots & \ddots & 0 \\ 0 & 0 & \dots & 1 \end{pmatrix}. \tag{2.19}$$

**Definition (Real Matrix):** The conjugate complex of a matrix $\mathbf{M}$ is written as $\mathbf{M}^*$ and the elements of $\mathbf{M}^*$ are the conjugate complexes of the elements of $\mathbf{M}$ i.e.

$$(M^*)_{ij} = \left(M_{ij}\right)^*. \tag{2.20}$$

For a real matrix, all the elements are real and, therefore

$$\mathbf{M} = \mathbf{M}^*, \qquad (M)_{ij} = M_{ij}^*. \tag{2.21}$$

**Definition (Symmetric Matrix):** The transpose of a matrix $\mathbf{M}$, denoted as $\mathbf{M}^T$, is obtained by changing rows into columns (or vice versa). For a symmetric matrix

$$\mathbf{M}^T = \mathbf{M},\ M_{ij} = M_{ji}. \tag{2.22}$$

**Definition (Skew-Symmetric Matrix):** The skew-symmetric matrix satisfies

$$\mathbf{M}^T = -\mathbf{M},\ M_{ij} = -M_{ji}. \tag{2.23}$$

**Definition (Orthogonal Matrix):** An orthogonal matrix satisfies

$$\mathbf{M}\mathbf{M}^T = \mathbf{M}^T\mathbf{M} = \mathbf{I}. \tag{2.24}$$

This leads to the equivalent characterization: a matrix $\mathbf{M}$ is orthogonal if its transpose is equal to its inverse:

$$\mathbf{M}^T = \mathbf{M}^{-1}. \tag{2.25}$$

**Definition (Conjugate Transpose or Hermitian Transpose):** Hermitian transpose of an $m \times n$ complex matrix $\mathbf{M}$ is an $n \times m$ matrix obtained by transposing $\mathbf{M}$ and applying complex conjugate on each entry.

$$\mathbf{M}^H = \mathbf{M}^\dagger = (\mathbf{M}^T)^*. \tag{2.26}$$

For real matrices, the conjugate transpose is just the transpose, $\mathbf{M}^H = \mathbf{M}^\dagger = \mathbf{M}^T$.

**Definition (Unitary Matrix)**: A complex square matrix $\mathbf{M}$ is unitary if its conjugate transpose $\mathbf{M}^\dagger$ is also its inverse, that is, if

$$\mathbf{M}^\dagger\mathbf{M} = \mathbf{M}\mathbf{M}^\dagger = \mathbf{M}\mathbf{M}^{-1} = \mathbf{I}. \tag{2.27}$$

**Definition (Lower and Upper Triangular Matrix):** A matrix of the form

$$\begin{pmatrix} l_{11} & 0 & \dots & 0 \\ l_{21} & l_{22} & \ddots & \vdots \\ \vdots & \vdots & \ddots & 0 \\ l_{n1} & l_{n2} & \dots & l_{nn} \end{pmatrix}, \tag{2.28}$$

is called a lower triangular matrix or left triangular matrix, and analogously a matrix of the form

$$\begin{pmatrix} u_{11} & u_{12} & \dots & u_{1n} \\ 0 & u_{22} & \dots & u_{2n} \\ \vdots & \ddots & \ddots & \vdots \\ 0 & \dots & 0 & u_{nn} \end{pmatrix}, \tag{2.29}$$

is called an upper triangular matrix or right triangular matrix. In the lower triangular matrix, all elements above the diagonal are zeros, in the upper triangular matrix, all the elements below the diagonal are zeros.

For example, the matrix **A** is symmetric, but the matrix **B** is skew-symmetric.

$$\mathbf{A} = \begin{pmatrix} 0 & 1 & -2 \\ 1 & 3 & 0 \\ -2 & 0 & 5 \end{pmatrix}, \quad \mathbf{B} = \begin{pmatrix} 0 & 2 & -45 \\ -2 & 0 & -4 \\ 45 & 4 & 0 \end{pmatrix}.$$

## 2.2 Vectors and Matrices with Dirac Notations

Throughout this book, we have chosen to use Dirac vector/matrix notation, also known as bra-ket notation. This decision was made after careful consideration. While we strongly advocate for index notation when it is suitable, we recognize its limitations. For instance, index notation significantly streamlines the presentation and manipulation of differential geometry. As a general guideline, if your work primarily revolves around differentiation with respect to spatial coordinates, then index notation is usually the preferred choice. In differential geometry, we will be dealing with systems of equations where matrix calculus is relatively straightforward, whereas matrix algebra and arithmetic of the NNs are more intricate. Therefore, we have opted to utilize Dirac notation for its simplicity and efficiency in this particular scenario. This notation system introduced by Dirac has found extensive use in quantum mechanics [41] and other areas of mathematics and theoretical physics, providing a concise and elegant way to represent complex mathematical constructs and operations.

In vector notation, vectorized operations, which operate on entire arrays or matrices at once, leverage highly optimized libraries and hardware acceleration (e.g., GPU) to perform computations much faster than iterative approaches. This notation is concise and abstract, making it easy to understand and manipulate mathematical expressions. Index notation (traditional iterative approaches, loop-based implementations) involves explicitly iterating over the indices of matrices and performing the required multiplications and summations. While index notation is more explicit and directly maps to computational implementations, it can be cumbersome and prone to errors, especially for large matrices. Loop-based implementations provide fine-grained control over memory access and computation, but they may suffer from poor performance due to the overhead of looping constructs and lack of parallelism.

GPU computations are highly efficient for matrix operations, especially for large matrices, due to their massively parallel architecture. They can significantly outperform CPU-based implementations. Vector notation describes operations at a high level, focusing on the relationships between vectors and matrices rather than the specific computational steps needed to perform them. On the other hand, GPU computations leverage the massively parallel architecture of graphics processing units to perform computations on large datasets simultaneously. GPUs excel at executing repetitive tasks in parallel, making them well-suited for matrix operations like multiplication, addition, and convolution. The synergy between GPU computations and vector notation arises from the fact that vector notation naturally lends itself to parallelism. Many operations expressed in vector notation can be efficiently parallelized on a GPU, allowing for faster computation of matrix products and other linear algebra operations. By leveraging vector notation alongside GPU computations, developers can write concise and expressive code that describes complex mathematical operations while taking advantage of the computational power of GPUs to execute those operations efficiently in parallel.

The entities that Dirac called "kets" and "bras" are simply column vectors and row vectors, respectively. In Dirac notation, "braket", ⟨⬚⟩, refers to the combination of "bra" and "ket" elements. Of course, the elements of these vectors are generally complex numbers. In this book, for convenience, we will express ourselves in terms of vectors and matrices of real numbers. Hence, in the language of matrices, these two vectors are related by simply taking the transpose. In summary, Dirac refers to a "bra," which he denoted as $\langle \mathbf{a}|$, a "ket," which he denoted as $|\mathbf{b}\rangle$, and a square matrix $\mathbf{M}$, we can associate these with vectors and matrices (in 3 dimensions) as follows;

$$\langle \mathbf{a}| = \mathbf{a}^T = (a_1, a_2, a_3), \quad |\mathbf{b}\rangle = \mathbf{b} = \begin{pmatrix} b_1 \\ b_2 \\ b_3 \end{pmatrix}, \quad \mathbf{M} = \begin{pmatrix} M_{11} & M_{12} & M_{13} \\ M_{21} & M_{22} & M_{23} \\ M_{31} & M_{32} & M_{33} \end{pmatrix}. \tag{2.30}$$

The product of a bra and a ket, denoted by Dirac as $\langle \mathbf{a}||\mathbf{b}\rangle$ or, more commonly, by omitting one of the middle lines, as $\langle \mathbf{a}|\mathbf{b}\rangle$, is simply a number given by inner products of a row vector and a column vector in the usual way, i.e.,

$$\langle\mathbf{a}|\mathbf{b}\rangle = \mathbf{a}^T\mathbf{b} = (a_1, a_2, a_3)\begin{pmatrix} b_1 \\ b_2 \\ b_3 \end{pmatrix} = a_1b_1 + a_2b_2 + a_3b_3 = \sum_{i=1}^{3} a_ib_i. \tag{2.31}$$

We can also form the product of ket times a bra, which gives a square matrix, as shown below,

$$|\mathbf{b}\rangle\langle\mathbf{a}| = \mathbf{b}\mathbf{a}^T = \begin{pmatrix} b_1 \\ b_2 \\ b_3 \end{pmatrix}(a_1, a_2, a_3) = \begin{pmatrix} b_1a_1 & b_1a_2 & b_1a_3 \\ b_2a_1 & b_2a_2 & b_2a_3 \\ b_3a_1 & b_3a_2 & b_3a_3 \end{pmatrix}. \tag{2.32}$$

The product of square matrix times a ket corresponds to the product of square matrix times a column vector, yielding another column vector (i.e., a ket). Mathematically, if we have a square matrix $\mathbf{M}$ of size $n \times n$ and a ket vector $|\mathbf{b}\rangle$ of size $n \times 1$, their product $\mathbf{M}|\mathbf{b}\rangle = \mathbf{Mb}$ results in another column vector (ket) of size $n \times 1$. The multiplication is performed by taking the dot product of each row of the matrix $\mathbf{M}$ with the column vector $|\mathbf{b}\rangle$ (i.e., using row-column multiplication or inner products of the rows with column). Each element of the resulting vector is obtained by multiplying corresponding elements of the row and the column vectors and summing up the products.

In summation notation, we have

$$\mathbf{M}|\mathbf{b}\rangle = \mathbf{Mb} = \begin{pmatrix} (M_{11} & M_{12} & M_{13}) \\ (M_{21} & M_{22} & M_{23}) \\ (M_{31} & M_{32} & M_{33}) \end{pmatrix}\begin{pmatrix} b_1 \\ b_2 \\ b_3 \end{pmatrix} = \begin{pmatrix} M_{11}b_1 + M_{12}b_2 + M_{13}b_3 \\ M_{21}b_1 + M_{22}b_2 + M_{23}b_3 \\ M_{31}b_1 + M_{32}b_2 + M_{33}b_3 \end{pmatrix} = \begin{pmatrix} \sum_{i=1}^{3} M_{1i}b_i \\ \sum_{i=1}^{3} M_{2i}b_i \\ \sum_{i=1}^{3} M_{3i}b_i \end{pmatrix}. \tag{2.33.1}$$

In Dirac matrix notation (2.33.1) becomes,

$$\mathbf{M}|\mathbf{b}\rangle = \mathbf{Mb} = \begin{pmatrix} \langle\mathbf{m}_{(1)}|\mathbf{b}^{(1)}\rangle \\ \langle\mathbf{m}_{(2)}|\mathbf{b}^{(1)}\rangle \\ \langle\mathbf{m}_{(3)}|\mathbf{b}^{(1)}\rangle \end{pmatrix}, \tag{2.33.2}$$

where,

$$\langle\mathbf{m}_{(1)}| = (M_{11} \quad M_{12} \quad M_{13}), \qquad \langle\mathbf{m}_{(2)}| = (M_{21} \quad M_{22} \quad M_{23}), \qquad \langle\mathbf{m}_{(3)}| = (M_{31} \quad M_{32} \quad M_{33}), \tag{2.34}$$

$$|\mathbf{b}^{(1)}\rangle = \begin{pmatrix} b_1 \\ b_2 \\ b_3 \end{pmatrix}. \tag{2.35}$$

However, we also have column-row multiplication (i.e., a linear combination of the columns using $b_i$). In summation notation, we have

$$\begin{aligned} \mathbf{M}|\mathbf{b}\rangle = \mathbf{Mb} &= \left(\begin{pmatrix} M_{11} \\ M_{21} \\ M_{31} \end{pmatrix}\begin{pmatrix} M_{12} \\ M_{22} \\ M_{32} \end{pmatrix}\begin{pmatrix} M_{13} \\ M_{23} \\ M_{33} \end{pmatrix}\right)\begin{pmatrix} b_1 \\ b_2 \\ b_3 \end{pmatrix} \\ &= \begin{pmatrix} M_{11} \\ M_{21} \\ M_{31} \end{pmatrix}b_1 + \begin{pmatrix} M_{12} \\ M_{22} \\ M_{32} \end{pmatrix}b_2 + \begin{pmatrix} M_{13} \\ M_{23} \\ M_{33} \end{pmatrix}b_3 \\ &= \begin{pmatrix} M_{11}b_1 \\ M_{21}b_1 \\ M_{31}b_1 \end{pmatrix} + \begin{pmatrix} M_{12}b_2 \\ M_{22}b_2 \\ M_{32}b_2 \end{pmatrix} + \begin{pmatrix} M_{13}b_3 \\ M_{23}b_3 \\ M_{33}b_3 \end{pmatrix} \\ &= \begin{pmatrix} M_{11}b_1 + M_{12}b_2 + M_{13}b_3 \\ M_{21}b_1 + M_{22}b_2 + M_{23}b_3 \\ M_{31}b_1 + M_{32}b_2 + M_{33}b_3 \end{pmatrix} = \begin{pmatrix} \sum_{i=1}^{3} M_{1i}b_i \\ \sum_{i=1}^{3} M_{2i}b_i \\ \sum_{i=1}^{3} M_{3i}b_i \end{pmatrix}. \end{aligned} \tag{2.36.1}$$

In Dirac matrix notation (2.36.1) becomes,

$$\begin{aligned}\mathbf{M}|\mathbf{b}\rangle &= \mathbf{Mb}\\ &= |\mathbf{m}^{(1)}\rangle\langle\mathbf{b}_{(1)}| + |\mathbf{m}^{(2)}\rangle\langle\mathbf{b}_{(2)}| + |\mathbf{m}^{(3)}\rangle\langle\mathbf{b}_{(3)}|\\ &= \sum_{i=1}^{3}|\mathbf{m}^{(i)}\rangle\langle\mathbf{b}_{(i)}|,\end{aligned} \tag{2.36.2}$$

where,

$$|\mathbf{m}^{(1)}\rangle = \begin{pmatrix}M_{11}\\M_{21}\\M_{31}\end{pmatrix}, \quad |\mathbf{m}^{(2)}\rangle = \begin{pmatrix}M_{12}\\M_{22}\\M_{32}\end{pmatrix}, \quad |\mathbf{m}^{(3)}\rangle = \begin{pmatrix}M_{13}\\M_{23}\\M_{33}\end{pmatrix}, \tag{2.37}$$

$$\langle\mathbf{b}_{(1)}| = (b_1), \quad \langle\mathbf{b}_{(2)}| = (b_2), \quad \langle\mathbf{b}_{(3)}| = (b_3). \tag{2.38}$$

Thus, $\mathbf{M}|\mathbf{b}\rangle$ is a linear combination of the columns of $\mathbf{M}$. This is fundamental. This thinking leads us to the column space of the matrix $\mathbf{M}$ and the idea of the rank of the matrix. The key idea is to take all combinations of the columns. All real numbers $b_i$ are allowed - the space includes $\mathbf{M}|\mathbf{b}\rangle$ for all vectors $|\mathbf{b}\rangle$. In this way, we get infinitely many output vectors $\mathbf{M}|\mathbf{b}\rangle$.

The product of bra times a square matrix corresponds to the product of a row vector times a square matrix, which is again a row vector (i.e., a "bra"), written as

$$\begin{aligned}\langle\mathbf{a}|\mathbf{M} &= \mathbf{a}^T\mathbf{M}\\ &= (a_1, a_2, a_3)\begin{pmatrix}M_{11} & M_{12} & M_{13}\\ M_{21} & M_{22} & M_{23}\\ M_{31} & M_{32} & M_{33}\end{pmatrix}\\ &= \left(\sum_{1=1}^{3} a_i M_{i1}, \sum_{1=1}^{3} a_i M_{i2}, \sum_{1=1}^{3} a_i M_{i3}\right),\end{aligned} \tag{2.39}$$

or

$$\begin{aligned}\langle\mathbf{a}|\mathbf{M} = \mathbf{a}^T\mathbf{M} &= (a_1, a_2, a_3)\begin{pmatrix}(M_{11} & M_{12} & M_{13})\\ (M_{21} & M_{22} & M_{23})\\ (M_{31} & M_{32} & M_{33})\end{pmatrix}\\ &= a_1(M_{11} \quad M_{12} \quad M_{13}) + a_2(M_{21} \quad M_{22} \quad M_{23}) + a_3(M_{31} \quad M_{32} \quad M_{33})\\ &= (a_1M_{11} \quad a_1M_{12} \quad a_1M_{13}) + (a_2M_{21} \quad a_2M_{22} \quad a_2M_{23}) + (a_3M_{31} \quad a_3M_{32} \quad a_3M_{33})\\ &= \left(\sum_{1=1}^{3} a_i M_{i1}, \sum_{1=1}^{3} a_i M_{i2}, \sum_{1=1}^{3} a_i M_{i3}\right)\\ &= \sum_{i=1}^{3}|\mathbf{a}^{(i)}\rangle\langle\mathbf{m}_{(i)}|,\end{aligned} \tag{2.40}$$

where

$$|\mathbf{a}^{(1)}\rangle = (a_1), \quad |\mathbf{a}^{(2)}\rangle = (a_2), \quad |\mathbf{a}^{(3)}\rangle = (a_3). \tag{2.41}$$

Thus, $\langle\mathbf{a}|\mathbf{M}$ is a linear combination of the rows of $\mathbf{M}$. This thinking leads us to the row space of the matrix $\mathbf{M}$ and the second definition of the rank of a matrix. The key idea is to take all combinations of the rows. All real numbers $a_i$ are allowed - the space includes $\langle\mathbf{a}|\mathbf{M}$ for all vectors $\langle\mathbf{a}|$. In this way, we get infinitely many output vectors $\langle\mathbf{a}|\mathbf{M}$.

Actually, we are seeing the clearest possible example of the first great theorem in linear algebra:

**Theorem 2.2:** The number of independent columns of a matrix equals the number of independent rows [33].

**Definition (Rank of a Matrix):** The rank of a matrix is the dimension of its column or row space.

This rank theorem is true for every matrix. Always columns and rows in linear algebra! The $m$ rows contain the same numbers $M_{ij}$ as the $n$ columns. But different vectors.

Obviously, we can form an ordinary (real) number by taking the compound product of a bra, a square matrix, and a ket, which corresponds to forming the product of a row vector times a square matrix times a column vector. Given a square matrix $\mathbf{M}$ of size $n \times n$, a bra vector $\langle \mathbf{a}|$ of size $1 \times n$, and a ket vector $|\mathbf{b}\rangle$ of size $n \times 1$, the compound product results in a scalar (an ordinary real number). Mathematically, the compound product $\langle \mathbf{a}|\mathbf{M}|\mathbf{b}\rangle$ is computed as the inner product of the row vector $\langle \mathbf{a}|$ and the matrix $\mathbf{M}$ and the product of the column vector $|\mathbf{b}\rangle$:

$$\begin{aligned}\langle \mathbf{a}|\mathbf{M}|\mathbf{b}\rangle &= \mathbf{a}^T\mathbf{M}\mathbf{b} \\ &= (a_1, a_2, a_3)\begin{pmatrix} M_{11} & M_{12} & M_{13} \\ M_{21} & M_{22} & M_{23} \\ M_{31} & M_{32} & M_{33}\end{pmatrix}\begin{pmatrix} b_1 \\ b_2 \\ b_3\end{pmatrix} \\ &= \left(\sum_{i=1}^{3} a_i M_{i1}, \sum_{i=1}^{3} a_i M_{i2}, \sum_{i=1}^{3} a_i M_{i3}\right)\begin{pmatrix} b_1 \\ b_2 \\ b_3\end{pmatrix} \\ &= \sum_{i=1}^{3} a_i M_{i1} b_1 + \sum_{i=1}^{3} a_i M_{i2} b_2 + \sum_{i=1}^{3} a_i M_{i3} b_3 \\ &= \sum_{j=1}^{3}\sum_{i=1}^{3} a_i M_{ij} b_j. \end{aligned} \tag{2.42}$$

Given two square matrices $\mathbf{A}$ and $\mathbf{B}$, both of size $n \times n$, the product $\mathbf{C}$ of these matrices is also a square matrix of size $n \times n$. Mathematically, the product $\mathbf{C}$ is computed by taking the dot product of each row of matrix $\mathbf{A}$ with each column of matrix $\mathbf{B}$, using row-column multiplication. But, we have also column-row multiplication. Mathematically, let:

$$\langle \mathbf{a}_{(1)}| = (A_{11} \quad A_{12} \quad A_{13}), \qquad \langle \mathbf{a}_{(2)}| = (A_{21} \quad A_{22} \quad A_{23}), \qquad \langle \mathbf{a}_{(3)}| = (A_{31} \quad A_{32} \quad A_{33}), \tag{2.43}$$

$$|\mathbf{b}^{(1)}\rangle = \begin{pmatrix} B_{11} \\ B_{21} \\ B_{31}\end{pmatrix}, \qquad |\mathbf{b}^{(2)}\rangle = \begin{pmatrix} B_{12} \\ B_{22} \\ B_{32}\end{pmatrix}, \qquad |\mathbf{b}^{(3)}\rangle = \begin{pmatrix} B_{13} \\ B_{23} \\ B_{33}\end{pmatrix}, \tag{2.44}$$

$$\begin{aligned}\mathbf{AB} &= \begin{pmatrix} A_{11} & A_{12} & A_{13} \\ A_{21} & A_{22} & A_{23} \\ A_{31} & A_{32} & A_{33}\end{pmatrix}\begin{pmatrix} B_{11} & B_{12} & B_{13} \\ B_{21} & B_{22} & B_{23} \\ B_{31} & B_{32} & B_{33}\end{pmatrix} \\ &= \begin{pmatrix} (A_{11} \quad A_{12} \quad A_{13})\begin{pmatrix} B_{11} \\ B_{21} \\ B_{31}\end{pmatrix} & (A_{11} \quad A_{12} \quad A_{13})\begin{pmatrix} B_{12} \\ B_{22} \\ B_{32}\end{pmatrix} & (A_{11} \quad A_{12} \quad A_{13})\begin{pmatrix} B_{13} \\ B_{23} \\ B_{33}\end{pmatrix} \\ (A_{21} \quad A_{22} \quad A_{23})\begin{pmatrix} B_{11} \\ B_{21} \\ B_{31}\end{pmatrix} & (A_{21} \quad A_{22} \quad A_{23})\begin{pmatrix} B_{12} \\ B_{22} \\ B_{32}\end{pmatrix} & (A_{21} \quad A_{22} \quad A_{23})\begin{pmatrix} B_{13} \\ B_{23} \\ B_{33}\end{pmatrix} \\ (A_{31} \quad A_{32} \quad A_{33})\begin{pmatrix} B_{11} \\ B_{21} \\ B_{31}\end{pmatrix} & (A_{31} \quad A_{32} \quad A_{33})\begin{pmatrix} B_{12} \\ B_{22} \\ B_{32}\end{pmatrix} & (A_{31} \quad A_{32} \quad A_{33})\begin{pmatrix} B_{13} \\ B_{23} \\ B_{33}\end{pmatrix}\end{pmatrix} \\ &= \begin{pmatrix} \sum_{p=1}^{3} A_{1p}B_{p1} & \sum_{p=1}^{3} A_{1p}B_{p2} & \sum_{p=1}^{3} A_{1p}B_{p3} \\ \sum_{p=1}^{3} A_{2p}B_{p1} & \sum_{p=1}^{3} A_{2p}B_{p2} & \sum_{p=1}^{3} A_{2p}B_{p3} \\ \sum_{p=1}^{3} A_{3p}B_{p1} & \sum_{p=1}^{3} A_{3p}B_{p2} & \sum_{p=1}^{3} A_{3p}B_{p3}\end{pmatrix} \\ &= \begin{pmatrix} \langle \mathbf{a}_{(1)}|\mathbf{b}^{(1)}\rangle & \langle \mathbf{a}_{(1)}|\mathbf{b}^{(2)}\rangle & \langle \mathbf{a}_{(1)}|\mathbf{b}^{(3)}\rangle \\ \langle \mathbf{a}_{(2)}|\mathbf{b}^{(1)}\rangle & \langle \mathbf{a}_{(2)}|\mathbf{b}^{(2)}\rangle & \langle \mathbf{a}_{(2)}|\mathbf{b}^{(3)}\rangle \\ \langle \mathbf{a}_{(3)}|\mathbf{b}^{(1)}\rangle & \langle \mathbf{a}_{(3)}|\mathbf{b}^{(2)}\rangle & \langle \mathbf{a}_{(3)}|\mathbf{b}^{(3)}\rangle\end{pmatrix}, \end{aligned} \tag{2.45}$$

or

$$
\begin{aligned}
\mathbf{AB} &= \begin{pmatrix} A_{11} \\ A_{21} \\ A_{31} \end{pmatrix} (B_{11} \quad B_{12} \quad B_{13}) + \begin{pmatrix} A_{12} \\ A_{22} \\ A_{32} \end{pmatrix} (B_{21} \quad B_{22} \quad B_{23}) + \begin{pmatrix} A_{13} \\ A_{23} \\ A_{33} \end{pmatrix} (B_{31} \quad B_{32} \quad B_{33}) \\
&= \begin{pmatrix} A_{11}B_{11} & A_{11}B_{12} & A_{11}B_{13} \\ A_{21}B_{11} & A_{21}B_{12} & A_{21}B_{13} \\ A_{31}B_{11} & A_{31}B_{12} & A_{31}B_{13} \end{pmatrix} + \begin{pmatrix} A_{12}B_{21} & A_{12}B_{22} & A_{12}B_{23} \\ A_{22}B_{21} & A_{22}B_{22} & A_{22}B_{23} \\ A_{32}B_{21} & A_{32}B_{22} & A_{32}B_{23} \end{pmatrix} + \begin{pmatrix} A_{13}B_{31} & A_{13}B_{32} & A_{13}B_{33} \\ A_{23}B_{31} & A_{23}B_{32} & A_{23}B_{33} \\ A_{33}B_{31} & A_{33}B_{32} & A_{33}B_{33} \end{pmatrix} \\
&= \begin{pmatrix} \sum_{p=1}^{3} A_{1p}B_{p1} & \sum_{p=1}^{3} A_{1p}B_{p2} & \sum_{p=1}^{3} A_{1p}B_{p3} \\ \sum_{p=1}^{3} A_{2p}B_{p1} & \sum_{p=1}^{3} A_{2p}B_{p2} & \sum_{p=1}^{3} A_{2p}B_{p3} \\ \sum_{p=1}^{3} A_{3p}B_{p1} & \sum_{p=1}^{3} A_{3p}B_{p2} & \sum_{p=1}^{3} A_{3p}B_{p3} \end{pmatrix} \\
&= \sum_{p=1}^{3} \begin{pmatrix} A_{1p}B_{p1} & A_{1p}B_{p2} & A_{1p}B_{p3} \\ A_{2p}B_{p1} & A_{2p}B_{p2} & A_{2p}B_{p3} \\ A_{3p}B_{p1} & A_{3p}B_{p2} & A_{3p}B_{p3} \end{pmatrix} \\
&= \sum_{i=1}^{3} \left|\mathbf{a}^{(i)}\right\rangle \left\langle \mathbf{b}_{(i)}\right|,
\end{aligned}
\tag{2.46}
$$

where

$$
\left|\mathbf{a}^{(1)}\right\rangle = \begin{pmatrix} A_{11} \\ A_{21} \\ A_{31} \end{pmatrix}, \qquad \left|\mathbf{a}^{(2)}\right\rangle = \begin{pmatrix} A_{12} \\ A_{22} \\ A_{32} \end{pmatrix}, \qquad \left|\mathbf{a}^{(3)}\right\rangle = \begin{pmatrix} A_{13} \\ A_{23} \\ A_{33} \end{pmatrix}, \tag{2.47}
$$

$$
\left\langle \mathbf{b}_{(1)}\right| = (B_{11} \quad B_{12} \quad B_{13}), \qquad \left\langle \mathbf{b}_{(2)}\right| = (B_{21} \quad B_{22} \quad B_{23}), \qquad \left\langle \mathbf{b}_{(3)}\right| = (B_{31} \quad B_{32} \quad B_{33}). \tag{2.48}
$$

## 2.3 Basics of Matrix Calculus

In the world of single-variable functions, the options are limited for taking the derivative; for $f: \mathbb{R} \to \mathbb{R}$, $x \to f(x)$, the only derivative of our interest is $\frac{df}{dx}$. But with functions such as $\mathbf{g}(\mathbf{x}) = \mathbf{Ax}$ and $h(\mathbf{x}, \mathbf{A}) = \langle \mathbf{x}|\mathbf{A}|\mathbf{x}\rangle$, we can also consider derivatives such as $\frac{d\mathbf{g}}{d\mathbf{x}}, \frac{d\mathbf{g}}{dx_i}, \frac{dh}{d\mathbf{A}}, \frac{dh}{dA_{ij}}, \frac{dh}{d\mathbf{x}^T}$, etc. In particular, we have the following cases [36]:

| | Scalar | Vector | Matrix |
|---|---|---|---|
| Scalar | $\frac{dy}{dx}$ | $\frac{dy}{d\mathbf{x}}$ | $\frac{dy}{d\mathbf{X}}$ |
| Vector | $\frac{d\mathbf{y}}{dx}$ | $\frac{d\mathbf{y}}{d\mathbf{x}}$ | $\frac{d\mathbf{y}}{d\mathbf{X}}$ |
| Matrix | $\frac{d\mathbf{Y}}{dx}$ | $\frac{d\mathbf{Y}}{d\mathbf{x}}$ | $\frac{d\mathbf{Y}}{d\mathbf{X}}$ |

There are many different versions of definitions, but here we use the denominator-layout notation. Also note that we use $d$ and $\partial$ interchangeably.

### Derivatives of Scalar

We first consider when we take the derivative of a scalar.

1. With respect to a scalar ($\frac{dy}{dx}$): We already know this case. This is simply the single-variable function case.

2. With respect to a vector ($\frac{dy}{d\mathbf{x}}$): An example of this case is when $y = \|\mathbf{x}\|$. This is the gradient we defined. That is, for $\mathbf{x} \in \mathbb{R}^{n\times 1}$,

$$\frac{dy}{d\mathbf{x}} = \begin{pmatrix} \frac{dy}{dx_1} \\ \vdots \\ \frac{dy}{dx_n} \end{pmatrix} \in \mathbb{R}^{n\times 1}. \tag{2.49}$$

We also define what happens when we take the derivative of a scalar with respect to a row vector $\mathbf{x}^T$:

$$\frac{dy}{d\mathbf{x}^T} = \left(\frac{dy}{dx_1}, \dots, \frac{dy}{dx_n}\right) \in \mathbb{R}^{1\times n}. \tag{2.50}$$

3. With respect to a matrix ($\frac{dy}{d\mathbf{X}}$): An example of this case is $y = \sqrt{\sum_{i=1}^{m}\sum_{j=1}^{n}\left|X_{ij}\right|^2}$, where $X_{ij}$ are the entries of the matrix. Expanding on the vector case, for $\mathbf{X} \in \mathbb{R}^{m\times n}$:

$$\frac{dy}{d\mathbf{X}} = \begin{pmatrix} \frac{dy}{dX_{11}} & \cdots & \frac{dy}{dX_{1n}} \\ \vdots & \ddots & \vdots \\ \frac{dy}{dX_{m1}} & \cdots & \frac{dy}{dX_{mn}} \end{pmatrix} \in \mathbb{R}^{m\times n}. \tag{2.51}$$

One thing to notice here is that when we take the derivative of a scalar, we end up with the same shape as the variable we took the derivative with respect to. For example, the shape of $\frac{dy}{d\mathbf{x}}$ is the same as the shape of $\mathbf{x}$. This is a nice property of the denominator-layout notation.

### Derivatives of Vector

Now we expand the scalar case to vectors, i.e., $\frac{d\mathbf{y}}{dx}$, $\frac{d\mathbf{y}}{d\mathbf{x}}$, and $\frac{d\mathbf{y}}{d\mathbf{X}}$. Note that $\mathbf{y}$ here does not necessarily have to be a column vector. The same definitions also apply to row vectors, including the resulting shapes.

1. With respect to a scalar ($\frac{d\mathbf{y}}{dx}$): An example of this case is $d(x\mathbf{v})/dx$ for a scalar $x$ and constant vector $\mathbf{v} \in \mathbb{R}^n$. For $\mathbf{y} \in \mathbb{R}^n$, this is defined as:

$$\frac{d\mathbf{y}}{dx} = \left(\frac{dy_1}{dx}, \dots, \frac{dy_n}{dx}\right) \in \mathbb{R}^{1\times n}. \tag{2.52}$$

2. With respect to a vector ($\frac{d\mathbf{y}}{d\mathbf{x}}$): An example of this case is $\mathbf{y} = \mathbf{A}\mathbf{x}$ for a constant matrix $\mathbf{A}$, and we evaluate $\frac{d\mathbf{y}}{d\mathbf{x}}$. For $\mathbf{y} \in \mathbb{R}^n$ and $\mathbf{x} \in \mathbb{R}^p$, this is defined as

$$\frac{d\mathbf{y}}{d\mathbf{x}} = \left(\nabla y_1(\mathbf{x}), \dots, \nabla y_n(\mathbf{x})\right) = \begin{pmatrix} \frac{dy_1}{dx_1} & \frac{dy_2}{dx_1} & \cdots & \frac{dy_n}{dx_1} \\ \frac{dy_1}{dx_2} & \frac{dy_2}{dx_2} & \cdots & \frac{dy_n}{dx_2} \\ \vdots & \vdots & \ddots & \vdots \\ \frac{dy_1}{dx_p} & \frac{dy_2}{dx_p} & \cdots & \frac{dy_n}{dx_p} \end{pmatrix} \in \mathbb{R}^{p\times n}. \tag{2.53}$$

Consider when $\mathbf{y} = \mathbf{A}\mathbf{x}$ for a constant matrix $\mathbf{A} \in \mathbb{R}^{n\times p}$. Explicit multiplication yields

$$\begin{aligned} \mathbf{y} &= \mathbf{A}\mathbf{x} \\ &= \begin{pmatrix} A_{11} & \cdots & A_{1p} \\ \vdots & \ddots & \vdots \\ A_{n1} & \cdots & A_{np} \end{pmatrix} \begin{pmatrix} x_1 \\ \vdots \\ x_p \end{pmatrix} \\ &= \begin{pmatrix} A_{11}x_1 + A_{12}x_2 + \cdots + A_{1p}x_p \\ \vdots \\ A_{n1}x_1 + A_{n2}x_2 + \cdots + A_{np}x_p \end{pmatrix} = \begin{pmatrix} \sum_{k=1}^{p} A_{1k}x_k \\ \vdots \\ \sum_{k=1}^{p} A_{nk}x_k \end{pmatrix}. \end{aligned} \tag{2.54}$$

This gives $y_i = \sum_{k=1}^{p} A_{ik} x_k$, and therefore $dy_i/dx_j = A_{ij}$. Hence, we have

$$\begin{aligned} \frac{d\mathbf{y}}{d\mathbf{x}} &= \begin{pmatrix} \frac{dy_1}{dx_1} & \frac{dy_2}{dx_1} & \cdots & \frac{dy_n}{dx_1} \\ \frac{dy_1}{dx_2} & \frac{dy_2}{dx_2} & \cdots & \frac{dy_n}{dx_2} \\ \vdots & \vdots & \ddots & \vdots \\ \frac{dy_1}{dx_p} & \frac{dy_2}{dx_p} & \cdots & \frac{dy_n}{dx_p} \end{pmatrix} \\ &= \begin{pmatrix} A_{11} & A_{21} & \dots & A_{n1} \\ A_{12} & A_{22} & \dots & A_{n2} \\ \vdots & \vdots & \ddots & \vdots \\ A_{1p} & A_{2p} & \dots & A_{np} \end{pmatrix} \\ &= \mathbf{A}^T. \end{aligned} \tag{2.55}$$

Hence, we have derived one helpful result:

$$\frac{d(\mathbf{Ax})}{d\mathbf{x}} = \mathbf{A}^T. \tag{2.56}$$

3. With respect to a matrix ($\frac{d\mathbf{y}}{d\mathbf{X}}$): An example of this case is $\mathbf{y} = \mathbf{Xv}$ for a constant vector $\mathbf{v}$, and we evaluate $\frac{d\mathbf{y}}{d\mathbf{X}}$. In general, this encodes three-dimensional information ($dy_i/dX_{jk}$) and is beyond the scope of these lectures. However, we define the following two specific cases:

$$\frac{d(\mathbf{Xv})}{d\mathbf{X}} = \mathbf{v}^T, \qquad \frac{d(\mathbf{v}^T\mathbf{X})}{d\mathbf{X}} = \mathbf{v}, \tag{2.57}$$

for a matrix $\mathbf{X}$ and a constant vector $\mathbf{v}$. Note that the second case is the derivative of a row vector with respect to a matrix.

The fundamental issue is that the derivative of a vector with respect to a vector, i.e., $\frac{d\mathbf{y}}{d\mathbf{x}}$, is often written in two competing ways. If the numerator $\mathbf{y}$ is of size $m$ and the denominator $\mathbf{x}$ of size $n$, then the result can be laid out as either an $m \times n$ matrix or $n \times m$ matrix, i.e., the elements of $\mathbf{y}$ laid out in columns and the elements of $\mathbf{x}$ laid out in rows, or vice versa. This leads to the following possibilities (see Table 2.1):

1- Numerator layout, i.e., lay out corresponding to $\mathbf{y}$ and $\mathbf{x}^T$. This is sometimes known as the Jacobian formulation. This corresponds to the $m \times n$ layout in the previous example.
2- Denominator layout, i.e., lay out corresponding to $\mathbf{y}^T$ and $\mathbf{x}$. This is sometimes known as the Hessian formulation. This corresponds to the $n \times m$ layout in the previous example.

**Table 2.1.** Derivatives of scalar, vector, and matrix based on numerator- and denominator-layout notations.

| Numerator-layout notation | Denominator-layout notation |
|---|---|
| $\frac{dy}{d\mathbf{x}} = \left(\frac{\partial y}{\partial x_1}, \frac{\partial y}{\partial x_2}, \dots, \frac{\partial y}{\partial x_n}\right) \in \mathbb{R}^{1\times n}$ | $\frac{dy}{d\mathbf{x}} = \begin{pmatrix} \frac{\partial y}{\partial x_1} \\ \frac{\partial y}{\partial x_2} \\ \vdots \\ \frac{\partial y}{\partial x_n} \end{pmatrix} \in \mathbb{R}^{n\times 1}$ |

$$\frac{\partial \mathbf{y}}{\partial x} = \begin{pmatrix} \frac{\partial y_1}{\partial x} \\ \frac{\partial y_2}{\partial x} \\ \vdots \\ \frac{\partial y_m}{\partial x} \end{pmatrix} \in \mathbb{R}^{m\times 1} \qquad \frac{\partial \mathbf{y}}{\partial x} = \left(\frac{\partial y_1}{\partial x}, \frac{\partial y_2}{\partial x}, \dots, \frac{\partial y_m}{\partial x}\right) \in \mathbb{R}^{1\times m}$$

$$\frac{d\mathbf{y}}{d\mathbf{x}} = \begin{pmatrix} \frac{\partial y_1}{\partial x_1} & \frac{\partial y_1}{\partial x_2} & \cdots & \frac{\partial y_1}{\partial x_n} \\ \frac{\partial y_2}{\partial x_1} & \frac{\partial y_2}{\partial x_2} & \cdots & \frac{\partial y_2}{\partial x_n} \\ \vdots & \vdots & \ddots & \vdots \\ \frac{\partial y_m}{\partial x_1} & \frac{\partial y_m}{\partial x_2} & \cdots & \frac{\partial y_m}{\partial x_n} \end{pmatrix} \in \mathbb{R}^{m\times n} \qquad \frac{d\mathbf{y}}{d\mathbf{x}} = \begin{pmatrix} \frac{\partial y_1}{\partial x_1} & \frac{\partial y_2}{\partial x_1} & \cdots & \frac{\partial y_m}{\partial x_1} \\ \frac{\partial y_1}{\partial x_2} & \frac{\partial y_2}{\partial x_2} & \cdots & \frac{\partial y_m}{\partial x_2} \\ \vdots & \vdots & \ddots & \vdots \\ \frac{\partial y_1}{\partial x_n} & \frac{\partial y_2}{\partial x_n} & \cdots & \frac{\partial y_m}{\partial x_n} \end{pmatrix} \in \mathbb{R}^{n\times m}$$

$$\frac{dy}{d\mathbf{X}} = \begin{pmatrix} \frac{\partial y}{\partial x_{11}} & \frac{\partial y}{\partial x_{21}} & \cdots & \frac{\partial y}{\partial x_{p1}} \\ \frac{\partial y}{\partial x_{12}} & \frac{\partial y}{\partial x_{22}} & \cdots & \frac{\partial y}{\partial x_{p2}} \\ \vdots & \vdots & \ddots & \vdots \\ \frac{\partial y}{\partial x_{1q}} & \frac{\partial y}{\partial x_{2q}} & \cdots & \frac{\partial y}{\partial x_{pq}} \end{pmatrix} \in \mathbb{R}^{q\times p} \qquad \frac{dy}{d\mathbf{X}} = \begin{pmatrix} \frac{\partial y}{\partial x_{11}} & \frac{\partial y}{\partial x_{12}} & \cdots & \frac{\partial y}{\partial x_{1q}} \\ \frac{\partial y}{\partial x_{21}} & \frac{\partial y}{\partial x_{22}} & \cdots & \frac{\partial y}{\partial x_{2q}} \\ \vdots & \vdots & \ddots & \vdots \\ \frac{\partial y}{\partial x_{p1}} & \frac{\partial y}{\partial x_{p2}} & \cdots & \frac{\partial y}{\partial x_{pq}} \end{pmatrix} \in \mathbb{R}^{p\times q}$$

$$\frac{d\mathbf{Y}}{dx} = \begin{pmatrix} \frac{\partial y_{11}}{\partial x} & \frac{\partial y_{12}}{\partial x} & \cdots & \frac{\partial y_{1n}}{\partial x} \\ \frac{\partial y_{21}}{\partial x} & \frac{\partial y_{22}}{\partial x} & \cdots & \frac{\partial y_{2n}}{\partial x} \\ \vdots & \vdots & \ddots & \vdots \\ \frac{\partial y_{m1}}{\partial x} & \frac{\partial y_{m2}}{\partial x} & \cdots & \frac{\partial y_{mn}}{\partial x} \end{pmatrix} \in \mathbb{R}^{m\times n}$$

## 2.4 Chain Rule

For a single-valued functions $f: \mathbb{R} \to \mathbb{R}$, $g: \mathbb{R} \to \mathbb{R}$, and $h: \mathbb{R} \to \mathbb{R}$, the derivative of $h(x) = f\big(g(x)\big)$ with respect to $x$, is obtained using a chain rule:

$$\frac{dh}{dx} = \frac{dg}{dx}\frac{df}{dg} = \frac{df}{dg}\frac{dg}{dx}. \tag{2.58}$$

The first term is forward mode (inside to outside). The second term is reverse mode (outside to inside). For the multivariable case $h(x) = f(g_1(x), g_2(x))$, the chain rule is extended as

$$\frac{dh}{dx} = \frac{dg_1}{dx}\frac{df}{dg_1} + \frac{dg_2}{dx}\frac{df}{dg_2} = \frac{df}{dg_1}\frac{dg_1}{dx} + \frac{df}{dg_2}\frac{dg_2}{dx}. \tag{2.59}$$

Visually, we can represent the two chain rules in Figure 2.1. This can be thought of as adding all components that contribute to the change of $h$. Building on this, we can extend the chain rule to also work in matrix calculus.

Consider $\mathbf{x} \in \mathbb{R}^p$, $\mathbf{y} \in \mathbb{R}^r$, $\mathbf{z} \in \mathbb{R}^n$ where $\mathbf{z}$ is a function of $\mathbf{y}$, and $\mathbf{y}$ is a function of $\mathbf{x}$; that is, $\mathbf{z} = f(\mathbf{y})$, $\mathbf{y} = g(\mathbf{x})$, and therefore $\mathbf{z} = f(g(\mathbf{x}))$. We can visualize this in Figure 2.2. Note how this figure considers the most general possible case. Now we derive the chain rule for vectors in matrix calculus. Recall that we have previously defined $d\mathbf{z}/d\mathbf{x}$ as

$$\frac{d\mathbf{z}}{d\mathbf{x}} = \begin{pmatrix} \frac{dz_1}{dx_1} & \frac{dz_2}{dx_1} & \cdots & \frac{dz_n}{dx_1} \\ \frac{dz_1}{dx_2} & \frac{dz_2}{dx_2} & \cdots & \frac{dz_n}{dx_2} \\ \vdots & \vdots & \ddots & \vdots \\ \frac{dz_1}{dx_p} & \frac{dz_2}{dx_p} & \cdots & \frac{dz_n}{dx_p} \end{pmatrix} \in \mathbb{R}^{p \times n}. \tag{2.60}$$

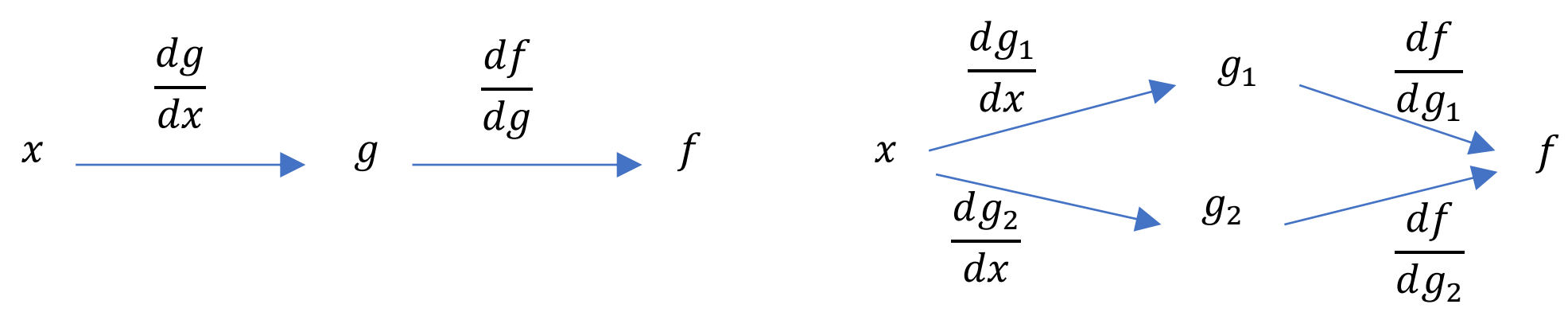

**Figure 2.1.** Chain rules are visualized.

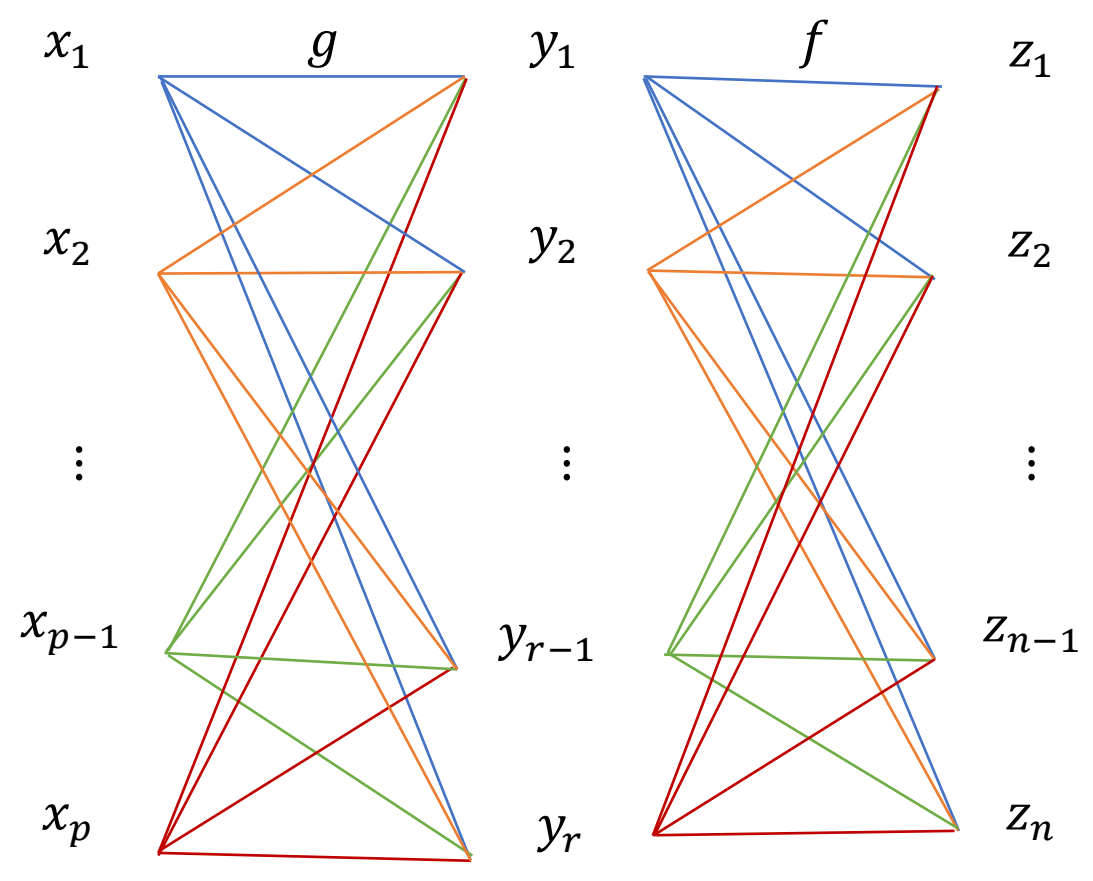

**Figure 2.2.** Visualization $\mathbf{z} = f(g(\mathbf{x}))$, where $\mathbf{z} = f(\mathbf{y})$, $\mathbf{y} = g(\mathbf{x})$.

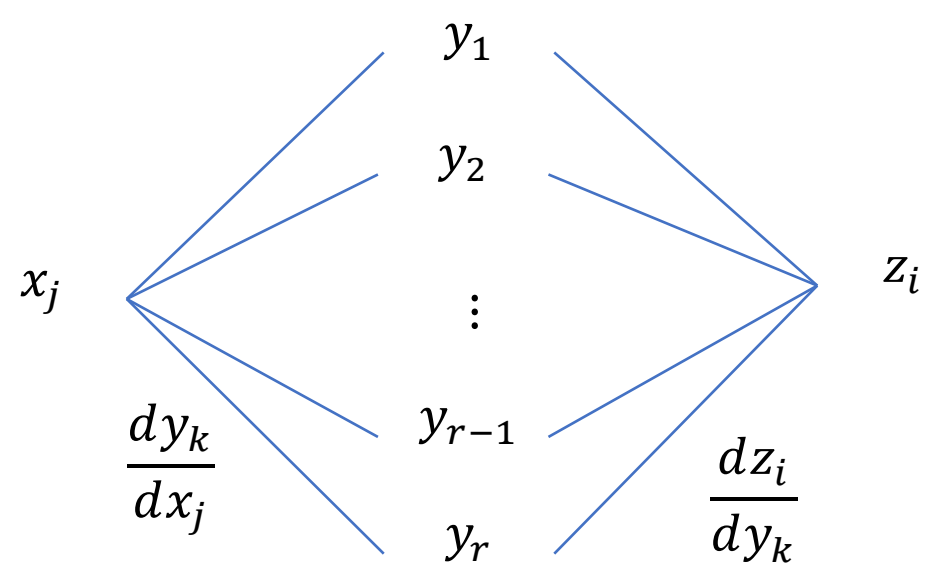

**Figure 2.3.** Chain rule visualized only considering $z_i$ and $x_j$.

By the chain rule,

$$\frac{dz_i}{dx_j} = \sum_{k=1}^{r} \frac{dz_i}{dy_k}\frac{dy_k}{dx_j} = \sum_{k=1}^{r} \frac{dy_k}{dx_j}\frac{dz_i}{dy_k}. \tag{2.61}$$

This directly follows from Figure 2.3, which can be obtained by isolating only $x_j$ and $z_i$ from Figure 2.2:

Apply the scalar chain rule to each element of $d\mathbf{z}/d\mathbf{x}$. By the definition of matrix multiplication, observe that

$$\left(\frac{d\mathbf{z}}{d\mathbf{x}}\right)^T = \begin{pmatrix} \frac{dz_1}{dx_1} & \frac{dz_1}{dx_2} & \cdots & \frac{dz_1}{dx_p} \\ \frac{dz_2}{dx_1} & \frac{dz_2}{dx_2} & \cdots & \frac{dz_2}{dx_p} \\ \vdots & \vdots & \ddots & \vdots \\ \frac{dz_n}{dx_1} & \frac{dz_n}{dx_2} & \cdots & \frac{dz_n}{dx_p} \end{pmatrix} \in \mathbb{R}^{n\times p}$$

$$= \begin{pmatrix} \sum_{k=1}^{r} \frac{dz_1}{dy_k}\frac{dy_k}{dx_1} & \sum_{k=1}^{r} \frac{dz_1}{dy_k}\frac{dy_k}{dx_2} & \cdots & \sum_{k=1}^{r} \frac{dz_1}{dy_k}\frac{dy_k}{dx_n} \\ \sum_{k=1}^{r} \frac{dz_2}{dy_k}\frac{dy_k}{dx_1} & \sum_{k=1}^{r} \frac{dz_2}{dy_k}\frac{dy_k}{dx_2} & \cdots & \sum_{k=1}^{r} \frac{dz_2}{dy_k}\frac{dy_k}{dx_n} \\ \vdots & \vdots & \ddots & \vdots \\ \sum_{k=1}^{r} \frac{dz_p}{dy_k}\frac{dy_k}{dx_1} & \sum_{k=1}^{r} \frac{dz_p}{dy_k}\frac{dy_k}{dx_2} & \cdots & \sum_{k=1}^{r} \frac{dz_p}{dy_k}\frac{dy_k}{dx_n} \end{pmatrix}. \tag{2.62.1}$$

Hence, we have

$$\left(\frac{d\mathbf{z}}{d\mathbf{x}}\right)^T = \begin{pmatrix} \frac{dz_1}{dy_1} & \frac{dz_1}{dy_2} & \cdots & \frac{dz_1}{dy_r} \\ \frac{dz_2}{dy_1} & \frac{dz_2}{dy_2} & \cdots & \frac{dz_2}{dy_r} \\ \vdots & \vdots & \ddots & \vdots \\ \frac{dz_n}{dy_1} & \frac{dz_n}{dy_2} & \cdots & \frac{dz_n}{dy_r} \end{pmatrix} \begin{pmatrix} \frac{dy_1}{dx_1} & \frac{dy_1}{dx_2} & \cdots & \frac{dy_1}{dx_p} \\ \frac{dy_2}{dx_1} & \frac{dy_2}{dx_2} & \cdots & \frac{dy_2}{dx_p} \\ \vdots & \vdots & \ddots & \vdots \\ \frac{dy_r}{dx_1} & \frac{dy_r}{dx_2} & \cdots & \frac{dy_r}{dx_p} \end{pmatrix}, \tag{2.62.2}$$

and

$$\left(\frac{d\mathbf{z}}{d\mathbf{x}}\right)^T = \left(\frac{d\mathbf{z}}{d\mathbf{y}}\right)^T \left(\frac{d\mathbf{y}}{d\mathbf{x}}\right)^T. \tag{2.62.3}$$

Taking the transpose of both sides, we have that the chain rule extends to

$$\frac{d\mathbf{z}}{d\mathbf{x}} = \frac{d\mathbf{y}}{d\mathbf{x}}\frac{d\mathbf{z}}{d\mathbf{y}}. \tag{2.63}$$

Note the matrix multiplication order; $d\mathbf{y}/d\mathbf{x}$ comes first. The order did not matter for the scalar case, but we need to be mindful of the order for the matrix case.

The key idea for this derivation was to manipulate the matrices cleverly and use the scalar chain rule. When other types of derivatives are involved, this chain rule may change; some derivatives may be transposed, and the multiplication order may change. The chain rules also vary depending on how the derivatives are defined. However, the scalar chain rule must hold no matter what.

In the following, we represent some important derivative formulas in matrix calculus.

1- Suppose that $\mathbf{x}$ is an $n \times 1$ vector and $f(\mathbf{x})$ is a scalar function of the elements of $\mathbf{x}$. Then

$$\frac{\partial f}{\partial \mathbf{x}} = \begin{pmatrix} \frac{\partial f}{\partial x_1} \\ \vdots \\ \frac{\partial f}{\partial x_n} \end{pmatrix} \in \mathbb{R}^{n\times 1}. \tag{2.64}$$

2- Suppose $\mathbf{x}$ and $\mathbf{y}$ are $n$-element column vectors. Then

$$\mathbf{x}^T\mathbf{y} = \langle \mathbf{x}|\mathbf{y}\rangle = x_1y_1 + x_2y_2 + \cdots + x_ny_n. \tag{2.65}$$

Hence, we have

$$\frac{\partial(\mathbf{x}^T\mathbf{y})}{\partial \mathbf{x}} = \frac{\partial\langle \mathbf{x}|\mathbf{y}\rangle}{\partial \mathbf{x}} = \begin{pmatrix} \frac{\partial\langle \mathbf{x}|\mathbf{y}\rangle}{\partial x_1} \\ \vdots \\ \frac{\partial\langle \mathbf{x}|\mathbf{y}\rangle}{\partial x_n} \end{pmatrix} = \begin{pmatrix} \frac{\partial(x_1y_1 + x_2y_2 + \cdots + x_ny_n)}{\partial x_1} \\ \vdots \\ \frac{\partial(x_1y_1 + x_2y_2 + \cdots + x_ny_n)}{\partial x_n} \end{pmatrix} = \begin{pmatrix} y_1 \\ \vdots \\ y_n \end{pmatrix} = \mathbf{y}. \tag{2.66}$$

3- Also, we can obtain

$$\frac{\partial(\mathbf{x}^T\mathbf{y})}{\partial \mathbf{y}} = \frac{\partial\langle \mathbf{x}|\mathbf{y}\rangle}{\partial \mathbf{y}} = \begin{pmatrix} \frac{\partial\langle \mathbf{x}|\mathbf{y}\rangle}{\partial y_1} \\ \vdots \\ \frac{\partial\langle \mathbf{x}|\mathbf{y}\rangle}{\partial y_n} \end{pmatrix} = \begin{pmatrix} \frac{\partial(x_1y_1 + x_2y_2 + \cdots + x_ny_n)}{\partial y_1} \\ \vdots \\ \frac{\partial(x_1y_1 + x_2y_2 + \cdots + x_ny_n)}{\partial y_n} \end{pmatrix} = \begin{pmatrix} x_1 \\ \vdots \\ x_n \end{pmatrix} = \mathbf{x}. \tag{2.67}$$

4- Now we will compute the partial derivative of a quadratic form $\mathbf{x}^T\mathbf{A}\mathbf{x}$ with respect to a vector. First, write the quadratic form as follows

$$\begin{aligned} \mathbf{x}^T\mathbf{A}\mathbf{x} = \langle \mathbf{x}|\mathbf{A}|\mathbf{x}\rangle &= (x_1, x_2, \dots, x_n)\begin{pmatrix} A_{11} & \cdots & A_{1n} \\ \vdots & \ddots & \vdots \\ A_{n1} & \cdots & A_{nn} \end{pmatrix}\begin{pmatrix} x_1 \\ \vdots \\ x_n \end{pmatrix} \\ &= \left(\sum_{i=1}^{\boldsymbol{n}} x_iA_{i1}, \sum_{i=1}^{\boldsymbol{n}} x_iA_{i2}, \dots, \sum_{i=1}^{\boldsymbol{n}} x_iA_{in}\right)\begin{pmatrix} x_1 \\ \vdots \\ x_n \end{pmatrix} = \sum_{j=1}^{n}\sum_{i=1}^{n} x_ix_jA_{ij}. \end{aligned} \tag{2.68}$$

Now take the partial derivative of the quadratic as follows:

$$\begin{aligned} \frac{\partial\mathbf{x}^T\mathbf{A}\mathbf{x}}{\partial \mathbf{x}} = \frac{\partial\langle \mathbf{x}|\mathbf{A}|\mathbf{x}\rangle}{\partial \mathbf{x}} &= \begin{pmatrix} \frac{\partial\langle \mathbf{x}|\mathbf{A}|\mathbf{x}\rangle}{\partial x_1} \\ \vdots \\ \frac{\partial\langle \mathbf{x}|\mathbf{A}|\mathbf{x}\rangle}{\partial x_n} \end{pmatrix} = \begin{pmatrix} \frac{\partial\left(\sum_{j=1}^{n}\sum_{i=1}^{\boldsymbol{n}} x_ix_jA_{ij}\right)}{\partial x_1} \\ \vdots \\ \frac{\partial\left(\sum_{j=1}^{n}\sum_{i=1}^{\boldsymbol{n}} x_ix_jA_{ij}\right)}{\partial x_n} \end{pmatrix} \\ &= \begin{pmatrix} \frac{\partial\sum_{j=1}^{n} x_1x_jA_{1j}}{\partial x_1} + \frac{\partial\sum_{i=1}^{n} x_ix_1A_{i1}}{\partial x_1} \\ \vdots \\ \frac{\partial\sum_{j=1}^{n} x_nx_jA_{nj}}{\partial x_n} + \frac{\partial\sum_{i=1}^{n} x_ix_nA_{in}}{\partial x_n} \end{pmatrix} \\ &= \begin{pmatrix} \sum_{j=1}^{n} x_jA_{1j} + \sum_{i=1}^{n} x_iA_{i1} \\ \vdots \\ \sum_{j=1}^{n} x_jA_{nj} + \sum_{i=1}^{n} x_iA_{in} \end{pmatrix} \\ &= \begin{pmatrix} \sum_{j=1}^{n} x_jA_{1j} \\ \vdots \\ \sum_{j=1}^{n} x_jA_{nj} \end{pmatrix} + \begin{pmatrix} \sum_{i=1}^{n} x_iA_{i1} \\ \vdots \\ \sum_{i=1}^{n} x_iA_{in} \end{pmatrix} = \mathbf{A}\mathbf{x} + \mathbf{A}^T\mathbf{x} = (\mathbf{A} + \mathbf{A}^T)\mathbf{x}. \end{aligned} \tag{2.69}$$

5- If $\mathbf{A}$ is symmetric, then $\mathbf{A} = \mathbf{A}^T$ and the above expression simplifies to

$$\frac{\partial\langle \mathbf{x}|\mathbf{A}|\mathbf{x}\rangle}{\partial \mathbf{x}} = 2\mathbf{A}\mathbf{x}. \tag{2.70}$$

6- Suppose $\mathbf{A}$ is an $m \times n$ matrix, $\mathbf{x}$ is an $n \times 1$ vector, and $\mathbf{y} = \mathbf{A}\mathbf{x}$;

$$\frac{\partial \mathbf{y}}{\partial \mathbf{x}} = \frac{\partial \mathbf{A}\mathbf{x}}{\partial \mathbf{x}} = \begin{pmatrix} \frac{dy_1}{dx_1} & \frac{dy_2}{dx_1} & \cdots & \frac{dy_m}{dx_1} \\ \frac{dy_1}{dx_2} & \frac{dy_2}{dx_2} & \cdots & \frac{dy_m}{dx_2} \\ \vdots & \vdots & \ddots & \vdots \\ \frac{dy_1}{dx_n} & \frac{dy_2}{dx_n} & \cdots & \frac{dy_m}{dx_n} \end{pmatrix} = \begin{pmatrix} A_{11} & A_{21} & \cdots & A_{m1} \\ A_{12} & A_{22} & \cdots & A_{m2} \\ \vdots & \vdots & \ddots & \vdots \\ A_{1n} & A_{2n} & \cdots & A_{mn} \end{pmatrix} = \mathbf{A}^T. \tag{2.71}$$

7- Now we suppose that $\mathbf{A}$ is an $m \times n$ matrix, $\mathbf{B}$ is an $n \times n$ matrix, and we want to compute the partial derivative of $\mathrm{Tr}(\mathbf{ABA}^T)$ with respect to $\mathbf{A}$. First, compute $\mathbf{ABA}^T$ as follows:

$$\begin{aligned} \mathbf{ABA}^T &= \begin{pmatrix} A_{11} & \cdots & A_{1n} \\ \vdots & \ddots & \vdots \\ A_{m1} & \cdots & A_{mn} \end{pmatrix} \begin{pmatrix} B_{11} & \cdots & B_{1n} \\ \vdots & \ddots & \vdots \\ B_{n1} & \cdots & B_{nn} \end{pmatrix} \begin{pmatrix} A_{11} & \cdots & A_{m1} \\ \vdots & \ddots & \vdots \\ A_{1n} & \cdots & A_{mn} \end{pmatrix} \\ &= \begin{pmatrix} \sum_{k,j} A_{1k}B_{kj}A_{1j} & \cdots & \sum_{k,j} A_{1k}B_{kj}A_{mj} \\ \vdots & \ddots & \vdots \\ \sum_{k,j} A_{mk}B_{kj}A_{1j} & \cdots & \sum_{k,j} A_{mk}B_{kj}A_{mj} \end{pmatrix}. \end{aligned} \tag{2.72}$$

From this we see that the trace of $\mathbf{ABA}^T$ is given as

$$\mathrm{Tr}(\mathbf{ABA}^T) = \sum_{i,j,k} A_{ik}B_{kj}A_{ij}. \tag{2.73}$$

Its partial derivative with respect to $\mathbf{A}$ can be computed as

$$\begin{aligned} \frac{\partial \mathrm{Tr}(\mathbf{ABA}^T)}{\partial \mathbf{A}} &= \begin{pmatrix} \frac{\partial \sum_{i,j,k} A_{ik}B_{kj}A_{ij}}{\partial A_{11}} & \cdots & \frac{\partial \sum_{i,j,k} A_{ik}B_{kj}A_{ij}}{\partial A_{1n}} \\ \vdots & \ddots & \vdots \\ \frac{\partial \sum_{i,j,k} A_{ik}B_{kj}A_{ij}}{\partial A_{m1}} & \cdots & \frac{\partial \sum_{i,j,k} A_{ik}B_{kj}A_{ij}}{\partial A_{mn}} \end{pmatrix} \\ &= \begin{pmatrix} \frac{\partial \sum_j A_{11}B_{1j}A_{1j}}{\partial A_{11}} + \frac{\partial \sum_k A_{1k}B_{k1}A_{11}}{\partial A_{11}} & \cdots & \frac{\partial \sum_j A_{1n}B_{nj}A_{1j}}{\partial A_{1n}} + \frac{\partial \sum_k A_{1k}B_{kn}A_{1n}}{\partial A_{1n}} \\ \vdots & \ddots & \vdots \\ \frac{\partial \sum_j A_{m1}B_{1j}A_{mj}}{\partial A_{m1}} + \frac{\partial \sum_k A_{mk}B_{k1}A_{m1}}{\partial A_{m1}} & \cdots & \frac{\partial \sum_j A_{mn}B_{nj}A_{mj}}{\partial A_{mn}} + \frac{\partial \sum_k A_{mk}B_{kn}A_{mn}}{\partial A_{mn}} \end{pmatrix} \\ &= \begin{pmatrix} \sum_j A_{1j}B_{1j} + \sum_k A_{1k}B_{k1} & \cdots & \sum_j A_{1j}B_{nj} + \sum_k A_{1k}B_{kn} \\ \vdots & \ddots & \vdots \\ \sum_j A_{mj}B_{1j} + \sum_k A_{mk}B_{k1} & \cdots & \sum_j A_{mj}B_{nj} + \sum_k A_{mk}B_{kn} \end{pmatrix} \\ &= \begin{pmatrix} \sum_j A_{1j}B_{1j} & \cdots & \sum_j A_{1j}B_{nj} \\ \vdots & \ddots & \vdots \\ \sum_j A_{mj}B_{1j} & \cdots & \sum_j A_{mj}B_{nj} \end{pmatrix} + \begin{pmatrix} \sum_k A_{1k}B_{k1} & \cdots & \sum_k A_{1k}B_{kn} \\ \vdots & \ddots & \vdots \\ \sum_k A_{mk}B_{k1} & \cdots & \sum_k A_{mk}B_{kn} \end{pmatrix} \\ &= \mathbf{AB}^T + \mathbf{AB} \\ &= \mathbf{A}(\mathbf{B}^T + \mathbf{B}). \end{aligned} \tag{2.74}$$

8- If $\mathbf{B}$ is symmetric, then this can be simplified to

$$\frac{\partial \mathrm{Tr}(\mathbf{ABA}^T)}{\partial \mathbf{A}} = 2\mathbf{AB}. \tag{2.75}$$

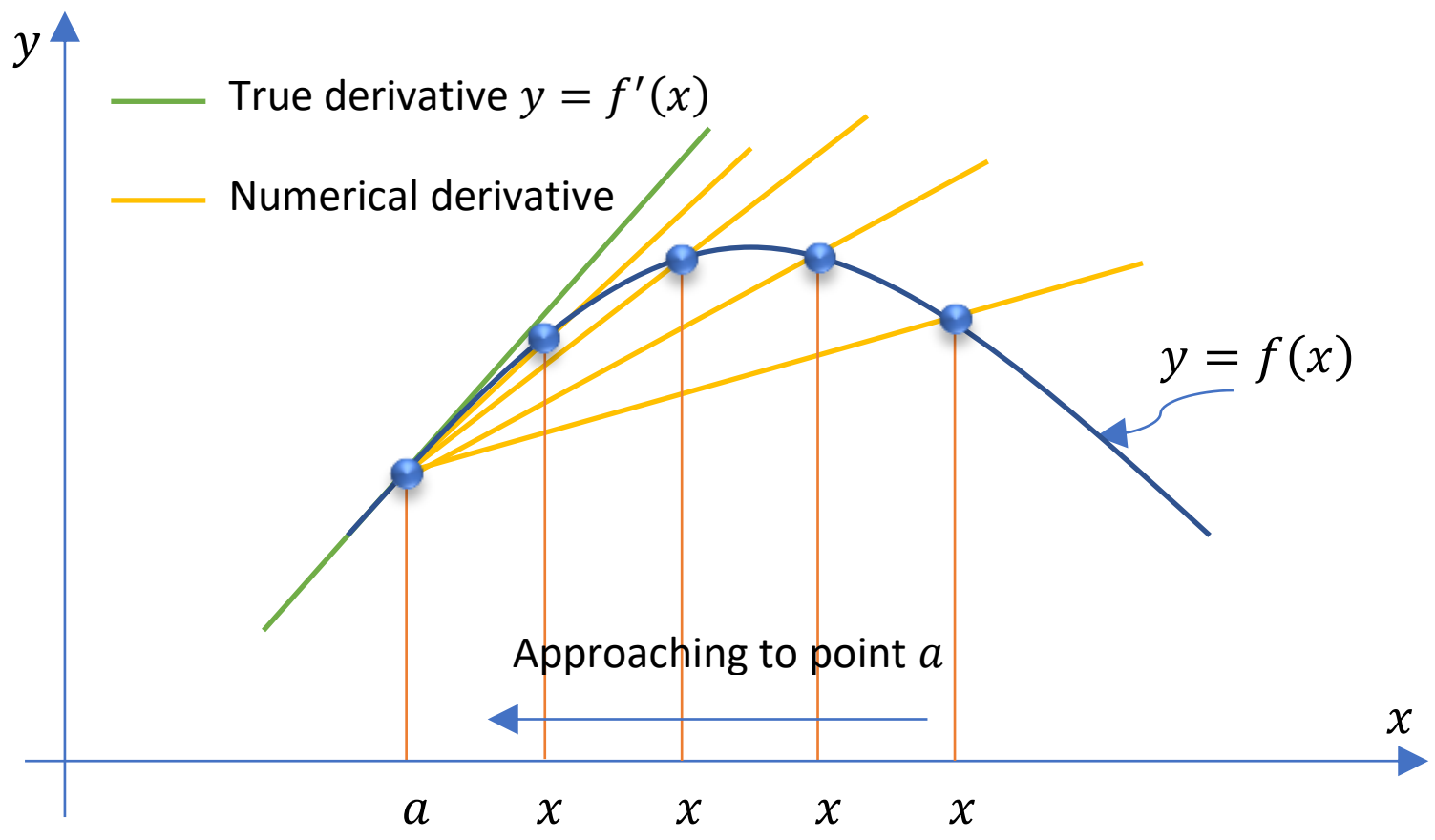

**Figure 2.4.** Definition of the derivative.

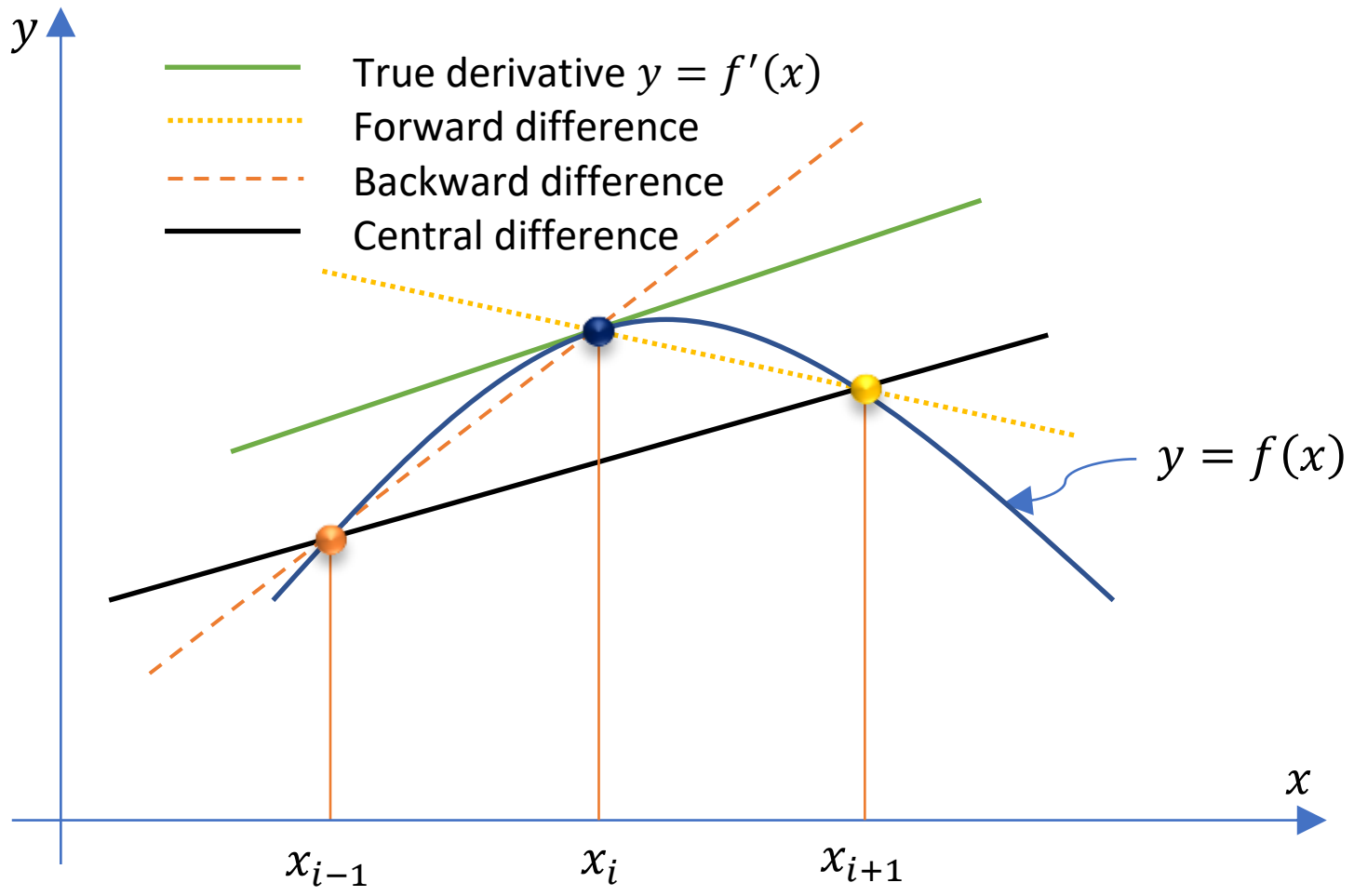

**Figure 2.5.** Finite difference approximation of derivative.

## 2.5 Numerical Differentiation and Finite Difference Methods

The process of estimating derivatives numerically is referred to as numerical differentiation. Estimates can be derived in different ways from function evaluations. This section discusses finite difference methods.

The derivative $f^{(1)}(x)$ of a function $f(x)$ at the point $x = a$ is defined by:

$$\left.\frac{df(x)}{dx}\right|_{x=a} = f^{(1)}(a) = \lim_{x\to a} \frac{f(x) - f(a)}{x - a}. \tag{2.76}$$

Graphically, the definition is illustrated in Figure 2.4. The derivative is the value of the slope of the tangent line to the function at $x = a$. In finite difference approximations of the derivative, values of the function at different points in the neighborhood of the point $x = a$ are used for estimating the slope. The forward, backward, and central finite difference formulas are the simplest finite difference approximations of the derivative. In these approximations, illustrated in Figure 2.5, the derivative at the point $x_i$ is calculated from the values of two points. The derivative is estimated as the value of the slope of the line that connects the two points.

1- The forward difference is the slope of the line that connects points $(x_i, f(x_i))$ and $(x_{i+1}, f(x_{i+1}))$:

$$\left.\frac{df}{dx}\right|_{x=x_i} = \frac{f(x_{i+1}) - f(x_i)}{x_{i+1} - x_i}. \tag{2.77}$$

2- The backward difference is the slope of the line that connects points $(x_{i-1}, f(x_{i-1}))$ and $(x_i, f(x_i))$:

$$\left.\frac{df}{dx}\right|_{x=x_i} = \frac{f(x_i) - f(x_{i-1})}{x_i - x_{i-1}}. \tag{2.78}$$

3- The central difference is the slope of the line that connects points $(x_{i-1}, f(x_{i-1}))$ and $(x_{i+1}, f(x_{i+1}))$:

$$\left.\frac{df}{dx}\right|_{x=x_i} = \frac{f(x_{i+1}) - f(x_{i-1})}{x_{i+1} - x_{i-1}}. \tag{2.79}$$

The forward, backward, and central difference formulas, as well as many other finite difference formulas for approximating derivatives, can be derived by using Taylor series expansion [42]. One advantage of using Taylor series expansion for deriving the formulas is that it also provides an estimate for the truncation error in the approximation. All the formulas derived in this section are for the case where the points are equally spaced.

### Two-Point Forward Difference Formula for First Derivative

The value of a function at a point $x_{i+1}$ can be approximated by a Taylor series in terms of the value of the function and its derivatives at a point $x_i$:

$$f(x_{i+1}) = f(x_i) + f^{(1)}(x_i)h + \frac{f^{(2)}(x_i)}{2!}h^2 + \frac{f^{(3)}(x_i)}{3!}h^3 + \frac{f^{(4)}(x_i)}{4!}h^4 + \cdots, \tag{2.80}$$

where $h = x_{i+1} - x_i$ is the spacing between the points. By using a two-term Taylor series expansion with a remainder, (2.80) can be rewritten as:

$$f(x_{i+1}) = f(x_i) + f^{(1)}(x_i)h + \frac{f^{(2)}(\xi)}{2!}h^2, \tag{2.81}$$

where $\xi$ is a value of $x$ between $x_i$ and $x_{i+1}$. Solving (2.81) for $f^{(1)}(x_i)$ yields:

$$f^{(1)}(x_i) = \frac{f(x_{i+1}) - f(x_i)}{h} - \frac{f^{(2)}(\xi)}{2!}h. \tag{2.82}$$

An approximate value of the derivative $f^{(1)}(x_i)$ can now be calculated if the second term on the right-hand side of (2.82) is ignored. Ignoring this second term introduces a truncation (discretization) error. Since this term is proportional to $h$, the truncation error is said to be on the order of $h$ (written as $O(h)$):

$$\text{truncation error } = \frac{f^{(2)}(\xi)}{2!}h = O(h). \tag{2.83}$$

It should be pointed out here that the magnitude of the truncation error is not really known since the value of $f^{(2)}(\xi)$ is not known. Nevertheless, (2.83) is valuable since it implies that a smaller $h$ gives a smaller error. Using the notation of (2.83), the approximated value of the first derivative is:

$$f^{(1)}(x_i) = \frac{f(x_{i+1}) - f(x_i)}{h} + O(h). \tag{2.84}$$

### Two-Point Backward Difference Formula for First Derivative

The backward difference formula can also be derived by the application of Taylor series expansion. The value of the function at a point $x_{i-1}$ is approximated by a Taylor series:

$$f(x_{i-1}) = f(x_i) - f^{(1)}(x_i)h + \frac{f^{(2)}(x_i)}{2!}h^2 - \frac{f^{(3)}(x_i)}{3!}h^3 + \frac{f^{(4)}(x_i)}{4!}h^4 - \cdots, \tag{2.85}$$

where $h = x_i - x_{i-1}$. By using a two-term Taylor series expansion with a remainder, (2.85) can be rewritten as:

$$f(x_{i-1}) = f(x_i) - f^{(1)}(x_i)h + \frac{f^{(2)}(\xi)}{2!}h^2, \tag{2.86}$$

where $\xi$ is a value of $x$ between $x_{i-1}$ and $x_i$. Solving (2.86) for $f^{(1)}(x_i)$ yields:

**Table 2.2.** List of difference formulas that can be used for numerical evaluation of first derivatives.

| | First Derivative | |
|---|---|---|
| Method | Formula | Truncation Error |
| Two-point forward difference | $f'(x_i) = \dfrac{f(x_{i+1}) - f(x_i)}{h}$ | $O(h)$ |
| Three-point forward difference | $f'(x_i) = \dfrac{-3f(x_i) + 4f(x_{i+1}) - f(x_{i+2})}{2h}$ | $O(h^2)$ |
| Two-point backward difference | $f'(x_i) = \dfrac{f(x_i) - f(x_{i-1})}{h}$ | $O(h)$ |
| Three-point backward difference | $f'(x_i) = \dfrac{f(x_{i-2}) - 4f(x_{i-1}) + 3f(x_i)}{2h}$ | $O(h^2)$ |
| Two-point central difference | $f'(x_i) = \dfrac{f(x_{i+1}) - f(x_{i-1})}{2h}$ | $O(h^2)$ |
| Four-point central difference | $f'(x_i) = \dfrac{f(x_{i-2}) - 8f(x_{i-1}) + 8f(x_{i+1}) - f(x_{i+2})}{12h}$ | $O(h^4)$ |

$$f^{(1)}(x_i) = \frac{f(x_i) - f(x_{i-1})}{h} - \frac{f^{(2)}(\xi)}{2!}h. \tag{2.87}$$

An approximate value of the derivative, $f^{(1)}(x_i)$, can be calculated if the second term on the right-hand side of (2.87) is ignored. This yield:

$$f^{(1)}(x_i) = \frac{f(x_i) - f(x_{i-1})}{h} + O(h). \tag{2.88}$$

### Two-Point Central Difference Formula for First Derivative

The central difference formula can be derived by using three terms in the Taylor series expansion and a remainder. The value of the function at a point $x_{i+1}$ in terms of the value of the function and its derivatives at a point $x_i$ is given by:

$$f(x_{i+1}) = f(x_i) + f^{(1)}(x_i)h + \frac{f^{(2)}(x_i)}{2!}h^2 + \frac{f^{(3)}(\xi_1)}{3!}h^3, \tag{2.89}$$

where $\xi_1$ is a value of $x$ between $x_i$ and $x_{i+1}$. The value of the function at a point $x_{i-1}$ is given by:

$$f(x_{i-1}) = f(x_i) - f^{(1)}(x_i)h + \frac{f^{(2)}(x_i)}{2!}h^2 - \frac{f^{(3)}(\xi_2)}{3!}h^3, \tag{2.90}$$

where $\xi_2$ is a value of $x$ between $x_{i-1}$ and $x_i$. In the last two equations, the spacing of the intervals is taken to be equal so that $h = x_{i+1} - x_i = x_i - x_{i-1}$. Subtracting (2.90) from (2.89) gives:

$$f(x_{i+1}) - f(x_{i-1}) = 2f^{(1)}(x_i)h + \frac{f^{(3)}(\xi_1)}{3!}h^3 + \frac{f^{(3)}(\xi_2)}{3!}h^3. \tag{2.91}$$

An estimate for the first derivative is obtained by solving (2.91) for $f^{(1)}(x_i)$ while neglecting the remainder terms, which introduces a truncation error, which is of the order of $h^2$:

$$f^{(1)}(x_i) = \frac{f(x_{i+1}) - f(x_{i-1})}{2h} + O(h^2). \tag{2.92}$$

A comparison of (2.84), (2.88), and (2.92) shows that in the forward and backward difference approximation, the truncation error is of the order $h$, while in the central difference approximation, the truncation error is of the order $h^2$. This indicates that the central difference approximation gives a more accurate approximation of the derivative.

### Finite Difference Formulas for the Second Derivative

The same approach used to develop finite difference formulas for the first derivative can be used to develop expressions for higher-order derivatives. For example, for points $x_{i+1}$, and $x_{i-1}$, the four-term Taylor series expansion with a remainder is

$$f(x_{i+1}) = f(x_i) + f^{(1)}(x_i)h + \frac{f^{(2)}(x_i)}{2!}h^2 + \frac{f^{(3)}(x_i)}{3!}h^3 + \frac{f^{(4)}(\xi_1)}{4!}h^4, \tag{2.93}$$

$$f(x_{i-1}) = f(x_i) - f^{(1)}(x_i)h + \frac{f^{(2)}(x_i)}{2!}h^2 - \frac{f^{(2)}(x_i)}{3!}h^3 + \frac{f^{(4)}(\xi_2)}{4!}h^4, \tag{2.94}$$

where $\xi_1$ is a value of $x$ between $x_i$ and $x_{i+1}$, and $\xi_2$ is a value of $x$ between $x_i$ and $x_{i-1}$. Adding (2.93) and (2.94) gives

$$f(x_{i+1}) - f(x_{i-1}) = 2f(x_i) + 2\frac{f^{(2)}(x_i)}{2!}h^2 + \frac{f^{(4)}(\xi_1)}{4!}h^4 + \frac{f^{(4)}(\xi_2)}{4!}h^4. \tag{2.95}$$

**Table 2.3.** List of difference formulas that can be used for numerical evaluation of second derivatives.

| Method | Second Derivative Formula | Truncation Error |
|---|---|---|
| Three-point forward difference | $f''(x_i) = \dfrac{f(x_i) - 2f(x_{i+1}) + f(x_{i+2})}{h^2}$ | $O(h)$ |
| Four-point forward difference | $f''(x_i) = \dfrac{2f(x_i) - 5f(x_{i+1}) + 4f(x_{i+2}) - f(x_{i+3})}{h^2}$ | $O(h^2)$ |
| Three-point backward difference | $f''(x_i) = \dfrac{f(x_{i-2}) - 2f(x_{i-1}) + f(x_i)}{h^2}$ | $O(h)$ |
| Four-point backward difference | $f''(x_i) = \dfrac{-f(x_{i-3}) + 4f(x_{i-2}) - 5f(x_{i-1}) + 2f(x_i)}{h^2}$ | $O(h^2)$ |
| Three-point central difference | $f''(x_i) = \dfrac{f(x_{i-1}) - 2f(x_i) + f(x_{i+1})}{h^2}$ | $O(h^2)$ |
| Five-point central difference | $f''(x_i) = \dfrac{-f(x_{i-2}) + 16f(x_{i-1}) - 30f(x_i) + 16f(x_{i+1}) - f(x_{i+2})}{12h^2}$ | $O(h^4)$ |

An estimate for the second derivative can be obtained by solving (2.95) for $f^{(2)}(x_i)$ while neglecting the remainder terms. This introduces a truncation error of the order $h^2$.

$$f^{(2)}(x_i) = \frac{f(x_{i-1}) - 2f(x_i) + f(x_{i+1})}{h^2} + O(h^2). \tag{2.96}$$

Equation (2.96) is the three-point central difference formula with a truncation error of $O(h^2)$. The same procedure can be used to develop the three-point forward and backward difference formulas for the second derivative. Tables 2.2 and 2.3 list difference formulas of various accuracy that can be used for numerical evaluation of first and second derivatives.

### Numerical Partial Differentiation

For a function of several independent variables, the partial derivative of the function with respect to one of the variables represents the rate of change of the value of the function with respect to this variable, while all the other variables are kept constant. For a function $f(x, y)$ with two independent variables, the partial derivatives with respect to $x$ and $y$ at the point $(a, b)$ are defined as

$$\left.\frac{\partial f(x,y)}{\partial x}\right|_{\substack{x=a\\y=b}} = \lim_{x\to a}\frac{f(x,b) - f(a,b)}{x-a}, \tag{2.97}$$

$$\left.\frac{\partial f(x,y)}{\partial y}\right|_{\substack{x=a\\y=b}} = \lim_{y\to b}\frac{f(a,y) - f(a,b)}{y-b}. \tag{2.98}$$

This means that the finite difference formulas that are used for approximating the derivatives of functions with one independent variable can be adopted for calculating partial derivatives. The formulas are applied for one of the variables, while the other variables are kept constant. For example, consider a function of two independent variables $f(x, y)$ specified as a set of discrete $m \cdot n$ points $(x_1, y_1)$, $(x_1, y_2)$, ..., $(x_n, y_m)$. The spacing between the points in each direction is constant such that $h_x = x_{i+1} - x_i$ and $h_y = y_{i+1} - y_i$. Figure 2.6 shows a case where $n = 5$ and $m = 4$. An approximation for the partial derivative at a point $(x_i, y_i)$ with the two-point forward difference formula is

$$\left.\frac{\partial f}{\partial x}\right|_{\substack{x=x_i\\y=y_i}} = \frac{f(x_{i+1}, y_i) - f(x_i, y_i)}{h_x}, \tag{2.99}$$

$$\left.\frac{\partial f}{\partial y}\right|_{\substack{x=x_i\\y=y_i}} = \frac{f(x_i, y_{i+1}) - f(x_i, y_i)}{h_y}. \tag{2.100}$$

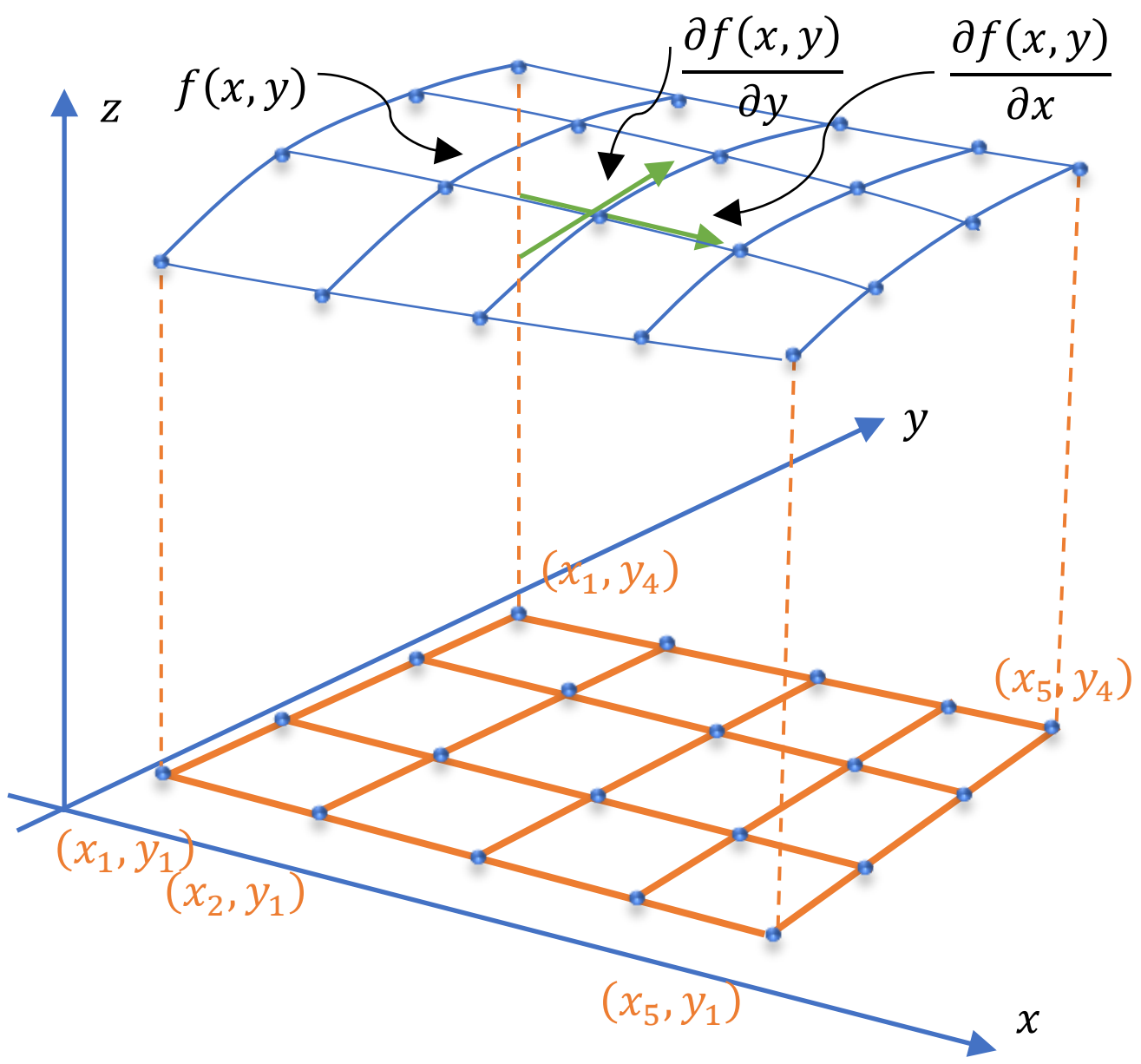

**Figure 2.6.** A function with two independent variables.

In the same way, the two-point backward and central difference formulas are

$$\left.\frac{\partial f}{\partial x}\right|_{\substack{x=x_i\\y=y_i}} = \frac{f(x_i, y_i) - f(x_{i-1}, y_i)}{h_x}, \tag{2.101}$$

$$\left.\frac{\partial f}{\partial y}\right|_{\substack{x=x_i\\y=y_i}} = \frac{f(x_i, y_i) - f(x_i, y_{i-1})}{h_y}, \tag{2.102}$$

$$\left.\frac{\partial f}{\partial x}\right|_{\substack{x=x_i\\y=y_i}} = \frac{f(x_{i+1}, y_i) - f(x_{i-1}, y_i)}{2h_x}, \tag{2.103}$$

$$\left.\frac{\partial f}{\partial x}\right|_{\substack{x=x_i\\y=y_i}} = \frac{f(x_i, y_{i+1}) - f(x_i, y_{i-1})}{2h_y}. \tag{2.104}$$

Moreover, the second partial derivatives with the three-point central difference formula are

$$\left.\frac{\partial^2 f}{\partial x^2}\right|_{\substack{x=x_i\\y=y_i}} = \frac{f(x_{i-1}, y_i) - 2f(x_i, y_i) + f(x_{i+1}, y_i)}{h_x^2}, \tag{2.105}$$

$$\left.\frac{\partial^2 f}{\partial y^2}\right|_{\substack{x=x_i\\y=y_i}} = \frac{f(x_i, y_{i-1}) - 2f(x_i, y_i) + f(x_i, y_{i+1})}{h_y^2}. \tag{2.106}$$

The second order mixed four-point central finite difference formula is

$$\left.\frac{\partial^2 f}{\partial x \partial y}\right|_{\substack{x=x_i\\y=y_i}} = \frac{[f(x_{i+1}, y_{i+1}) - f(x_{i-1}, y_{i+1})] - [f(x_{i+1}, y_{i-1}) - f(x_{i-1}, y_{i-1})]}{2h_x \cdot 2h_y}. \tag{2.107}$$

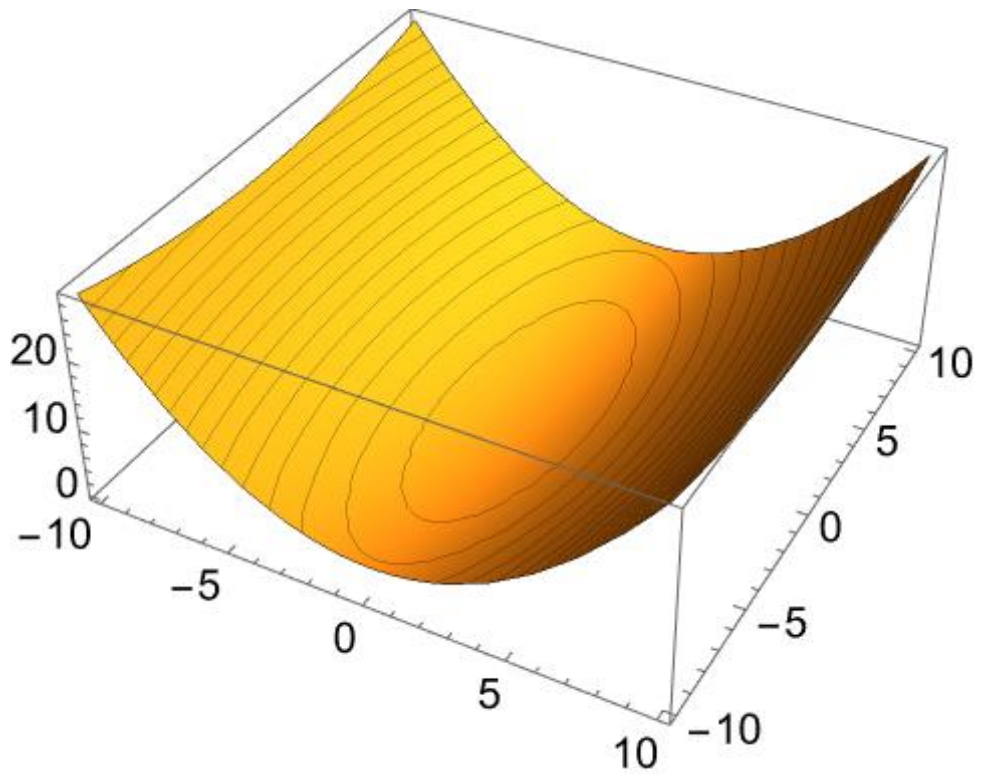

**Figure 2.7.** The graph of $f|\mathbf{x}\rangle = \frac{x^2}{4} + \frac{y^2}{25}$.

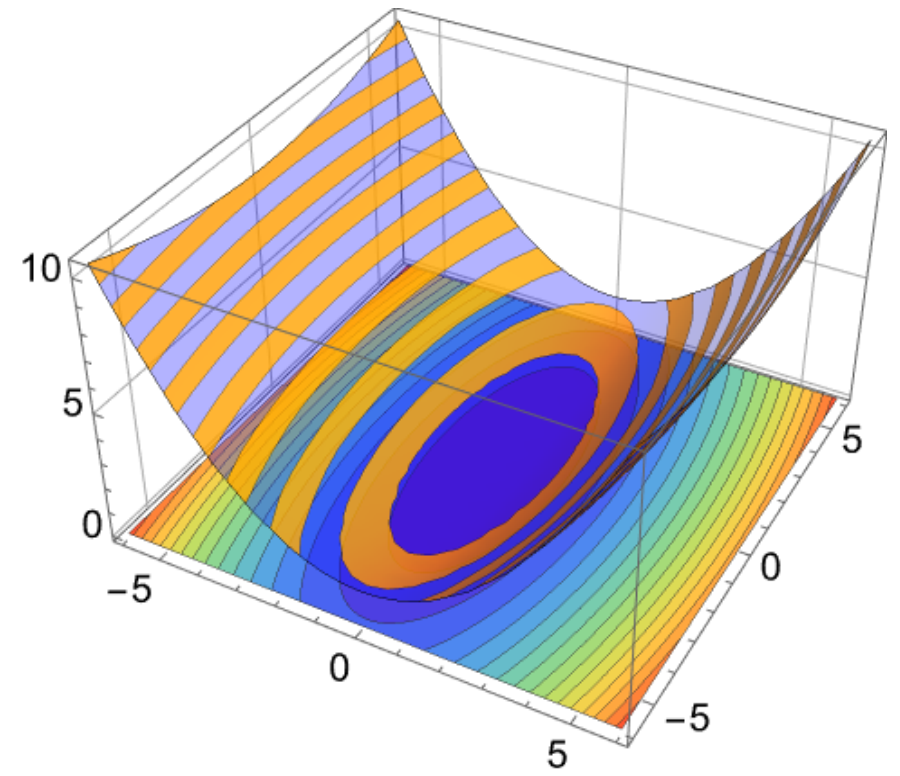

**Figure 2.8.** The level curve C with equation $f|\mathbf{x}\rangle = k$ is the projection of the trace of $f$ in the plane $z = k$ onto the $xy$-plane.

## 2.6 Optimality Criteria of the Two-Variable Functions

This section elucidates the necessary and sufficient conditions for determining the minimum of an unconstrained function involving multiple variables. It delves into fundamental concepts such as tangent planes, directional derivatives, and the Taylor series expansion of multivariable functions. Initially, our attention is directed towards functions of two variables, emphasizing the pivotal role of visualizing these functions in three dimensions. Visualizing functions with two variables in three dimensions are crucial for gaining a deeper understanding. Subsequently, we extend these foundational principles to the scenario of functions with $n$-variables.

One way to visualize a function of two variables is through its graph. The graph of $f$ is the surface with equation $z = f|\mathbf{x}\rangle$, $|\mathbf{x}\rangle = (x, y)^T$. See, for example, Figure 2.7. Another method for visualizing functions is a contour map on which points of constant elevation are joined to form contour lines or level curves.

**Definition (Level Curves):** The level curves of a function of two variables are the curves in the $xy$-plane with equations $f|\mathbf{x}\rangle = k$, where $k$ is a constant in the range of $f$.

More generally, a contour line for a function of two variables is a curve connecting points where the function has the same particular value (a constant value). It is a plane section of the three-dimensional graph of the function $f|\mathbf{x}\rangle$ parallel to the $xy$-plane. The surface is steep where the level curves are close together. It is somewhat flatter where they are farther apart, see Figure 2.8.

***Example 2.2***

Sketch a contour map for the surface described by $f|\mathbf{x}\rangle = x^2 + y^2$, using the level curves corresponding to $k = 0$, 1, 4, 9, and 16.

***Solution***

The level curve of $f$ corresponding to each value of $k$ is a circle $x^2 + y^2 = k$ of radius $\sqrt{k}$, centered at the origin, see Figure 2.9. For example, if $k = 4$, the level curve is the circle with the equation $x^2 + y^2 = 4$, centered at the origin and having a radius 2.

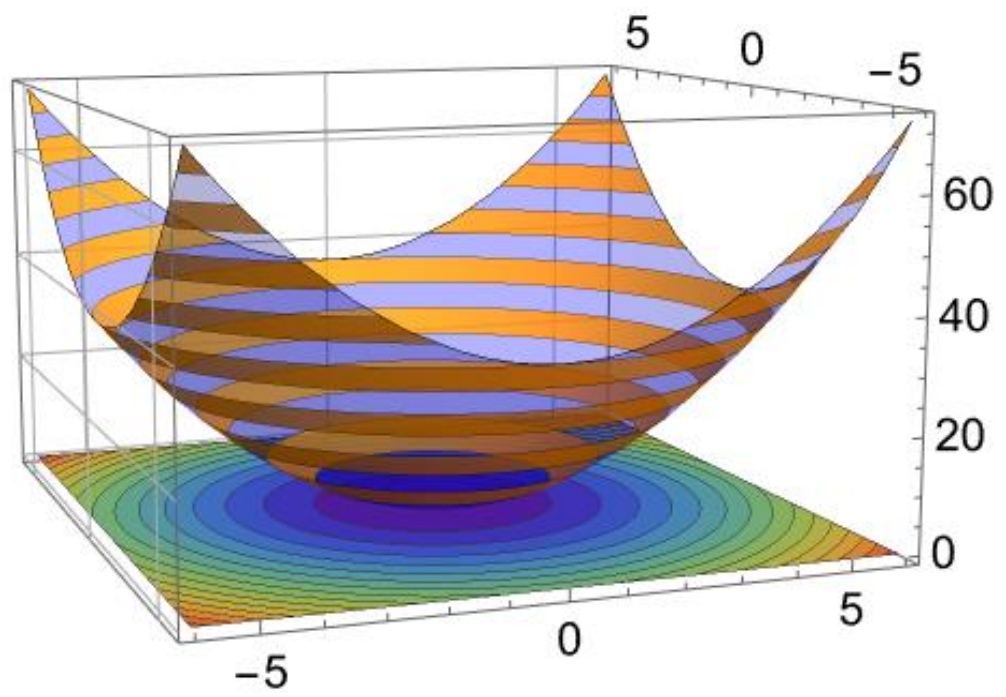

**Figure 2.9.** The level curve of the function $f|\mathbf{x}\rangle = x^2 + y^2$.

### 2.6.1 Taylor series for the two-variable functions and linearization

It will be helpful to review the Taylor series for a function of one variable and see how it extends to functions of more than one variable. Recall that the Taylor series for a function $f(x)$, based at a point $x = a$, is given by the following, where we assume that $f$ is analytic:

$$f(x) = f(a) + f'(a)(x-a) + \frac{1}{2!}f''(a)(x-a)^2 + \cdots. \tag{2.108}$$

Therefore, we can approximate $f$ using a constant:

$$f(x) \approx f(a), \tag{2.109}$$

or using a linear approximation (which is the tangent line to $f$ at $a$):

$$f(x) \approx f(a) + f'(a)(x-a), \tag{2.110}$$

or using a quadratic approximation:

$$f(x) \approx f(a) + f'(a)(x-a) + \frac{1}{2!}f''(a)(x-a)^2. \tag{2.111}$$

We can do something similar if $f$ depends on more than one variable as follows [43].

**Definition (Tangent Plane):** Provided $f$ is differentiable at $(a, b)$, the approximation $f(x, y)$:

$$f(x,y) \approx f(a,b) + f_x(a,b)(x-a) + f_y(a,b)(y-b), \tag{2.112.1}$$

is called the linear approximation or the tangent plane approximation of $f(x, y)$. In ket (column) notation

$$f|\mathbf{x}\rangle \approx f|\hat{\mathbf{x}}\rangle + f_x|\hat{\mathbf{x}}\rangle(x-a) + f_y|\hat{\mathbf{x}}\rangle(y-b), \tag{2.112.2}$$

where $|\mathbf{x}\rangle = (x, y)^T$, and $|\hat{\mathbf{x}}\rangle = (a, b)^T$.

**Remarks:**

- Since the tangent plane lies close to the surface near the point at which they meet, $z$-values on the tangent plane are close to values of $f|\mathbf{x}\rangle$ for points near $|\hat{\mathbf{x}}\rangle$. See Figure 2.10.
- We are thinking of $a$ and $b$ as fixed, so the expression on the right side of (2.112) is linear in $x$ and $y$. The right side of this approximation gives the local linearization of $f$ near $x = a$, $y = b$.

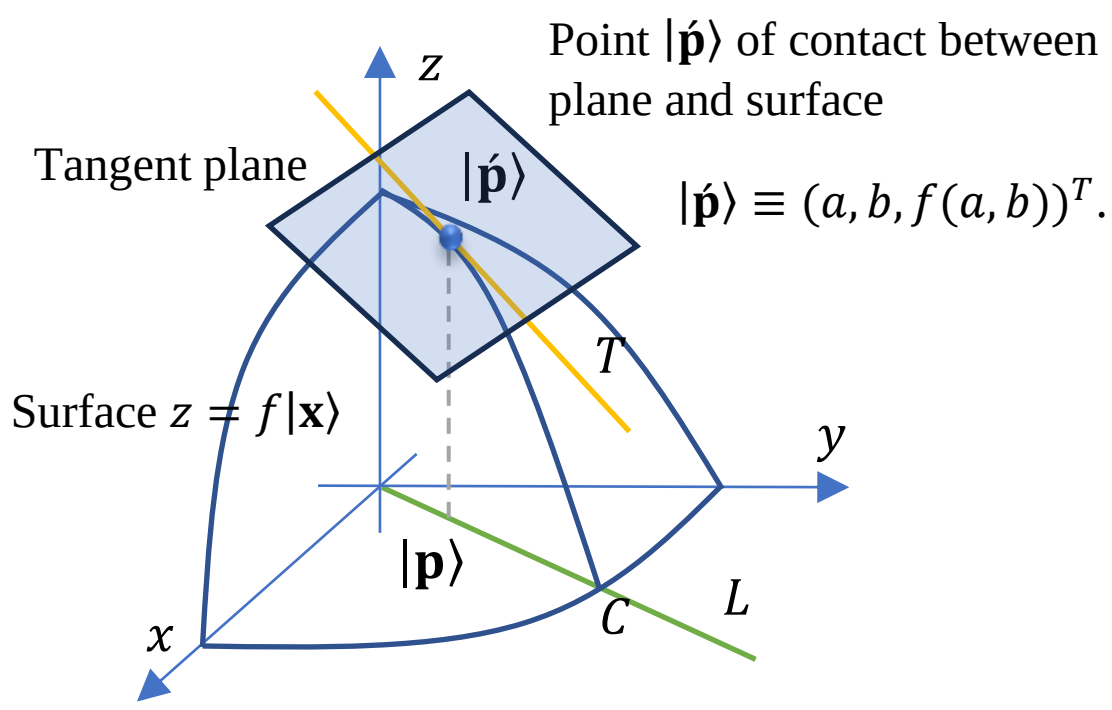

**Figure 2.10.** The tangent plane to the surface $z = f(x,y)$ at the point $(a,b)$

- For a function of one variable, local linearity means that as we zoom in on the graph, it looks like a straight line. As we zoom in on the graph of a two-variable function, the graph usually looks like a plane.

If we want to go further with a second-order (quadratic) approximation, it looks very similar. First, if $z = f(x,y)$ at $(a,b)$, the quadratic approximation looks like this:

$$f(x,y) \approx f(a,b) + f_x(a,b)(x-a) + f_y(a,b)(y-b) \\ + \frac{1}{2}\left(f_{xx}(a,b)(x-a)^2 + 2f_{xy}(a,b)(x-a)(y-b) + f_{yy}(a,b)(y-b)^2\right), \tag{2.113.1}$$

where we assume that $f_{xy}(a,b) = f_{yx}(a,b)$. In ket notation, we have

$$f|\mathbf{x}\rangle \approx f|\hat{\mathbf{x}}\rangle + f_x|\hat{\mathbf{x}}\rangle(x-a) + f_y|\hat{\mathbf{x}}\rangle(y-b) \\ + \frac{1}{2}\left[f_{xx}|\hat{\mathbf{x}}\rangle(x-a)^2 + 2f_{xy}|\hat{\mathbf{x}}\rangle(x-a)(y-b) + f_{yy}|\hat{\mathbf{x}}\rangle(y-b)^2\right], \tag{2.113.2}$$

where $|\mathbf{x}\rangle = (x,y)^T$, and $|\hat{\mathbf{x}}\rangle = (a,b)^T$. Moreover, the gradient of $f$, in 2D, is a vector of first partial derivatives:

$$\nabla f = \begin{pmatrix} f_x \\ f_y \end{pmatrix}, \tag{2.114}$$

and the $2 \times 2$ matrix of second partial derivatives is the Hessian matrix

$$\mathbf{H}_f = \begin{pmatrix} f_{xx} & f_{yx} \\ f_{xy} & f_{yy} \end{pmatrix}. \tag{2.115}$$

Using these notations, the linear approximation to $f$ at $|\hat{\mathbf{x}}\rangle = (a,b)^T$ is

$$f|\mathbf{x}\rangle \approx f|\hat{\mathbf{x}}\rangle + \langle \nabla f(\hat{\mathbf{x}})|\mathbf{x} - \hat{\mathbf{x}}\rangle, \tag{2.116}$$

and the quadratic approximation to $f$ is:

$$f(\mathbf{x}) \approx f|\hat{\mathbf{x}}\rangle + \langle \nabla f(\hat{\mathbf{x}})|\mathbf{x} - \hat{\mathbf{x}}\rangle + \frac{1}{2}\langle \mathbf{x} - \hat{\mathbf{x}}|\mathbf{H}_f(\hat{\mathbf{x}})|\mathbf{x} - \hat{\mathbf{x}}\rangle. \tag{2.117}$$

### 2.6.2 The Directional Derivative of Two-Variable Function

Suppose that $f$ is a function defined by the equation $z = f|\mathbf{x}\rangle$, $|\mathbf{x}\rangle = (x,y)^T$ and let $|\mathbf{p}\rangle \equiv (a,b)^T$ be a point in the domain $D$ of $f$. Furthermore, let $|\mathbf{u}\rangle$ be a unit vector having a specified direction. Then the vertical plane containing the line $L$ passing through $|\mathbf{p}\rangle$ and having the same direction as $|\mathbf{u}\rangle$ will intersect the surface $z = f|\mathbf{x}\rangle$ along a curve $C$ (see Figure 2.11). Intuitively, we see that the rate of change of $z$ at the point $|\mathbf{p}\rangle$ with respect to the distance measured along $L$ is given by the slope of the tangent line $T$ to the curve $C$ at the point $|\acute{\mathbf{p}}\rangle \equiv (a, b, f(a,b))^T$. To find the slope of $T$, first, observe that $|\mathbf{u}\rangle$ may be specified by writing $|\mathbf{u}\rangle = (u_1, u_2)^T$ for appropriate components $u_1$ and $u_2$.

Equivalently, we may specify $|\mathbf{u}\rangle$ by giving the angle $\theta$ that it makes with the positive $x$-axis, in which case $u_1 = \cos\theta$ and $u_2 = \sin\theta$ (see Figure 2.12).

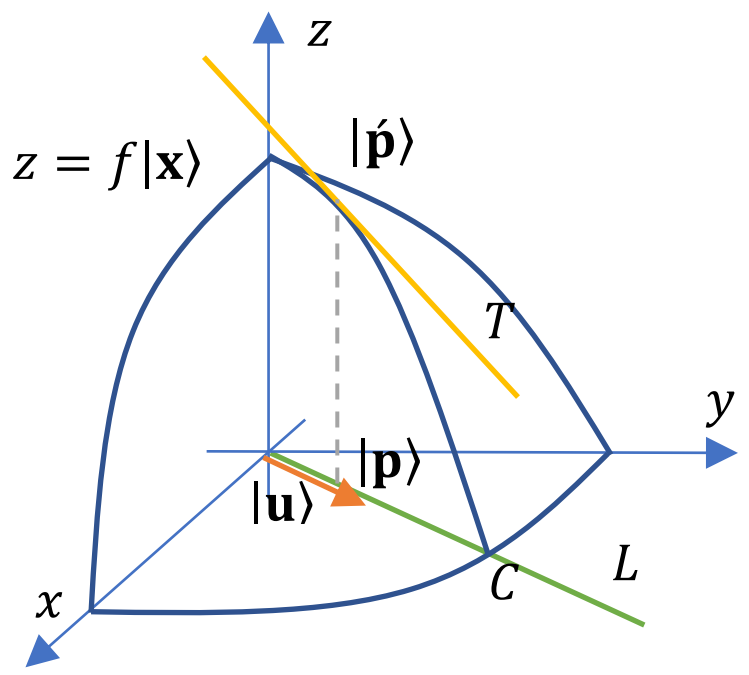

**Figure 2.11.** The rate of change of $z$ at $|\mathbf{p}\rangle$ with respect to the distance measured along $L$ is given by the slope of $T$.

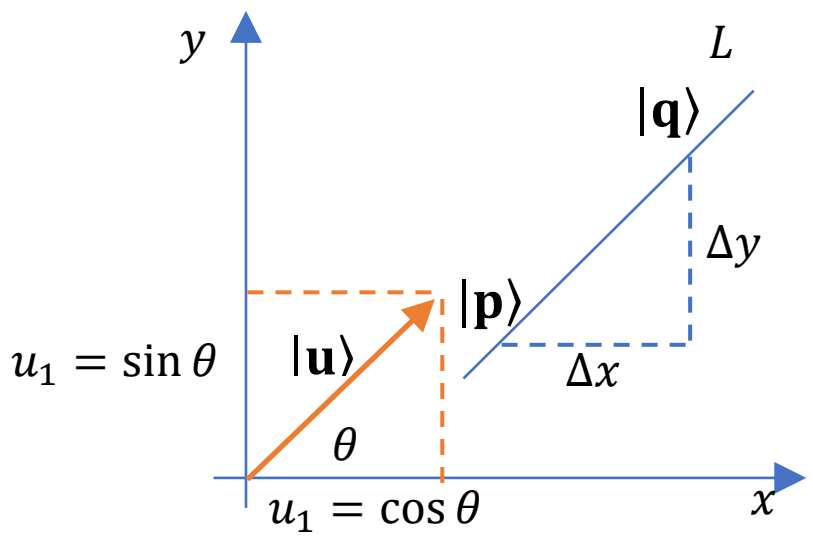

**Figure 2.12.** Any direction in the plane can be specified in terms of a unit vector $|\mathbf{u}\rangle$.

Next, let $|\mathbf{q}\rangle \equiv (a + \Delta x, b + \Delta y)^T$ be any point distinct from $|\mathbf{p}\rangle$ lying on the line $L$ passing through $|\mathbf{p}\rangle$ and having the same direction as $|\mathbf{u}\rangle$ (Figure 2.12). Since the vector $|\mathbf{pq}\rangle$ is parallel to $|\mathbf{u}\rangle$, it must be a scalar multiple of $|\mathbf{u}\rangle$. In other words, there exists a nonzero number $h$ such that

$$|\mathbf{pq}\rangle = h|\mathbf{u}\rangle = (hu_1, hu_2)^T. \tag{2.118}$$

But $|\mathbf{pq}\rangle$ is also given by $(\Delta x, \Delta y)^T$, and therefore,

$$\Delta x = hu_1, \qquad \Delta y = hu_2, \qquad h = \sqrt{(\Delta x)^2 + (\Delta y)^2}. \tag{2.119}$$

So, the point $|\mathbf{q}\rangle$ can be expressed as $|\mathbf{q}\rangle \equiv (a + hu_1, b + hu_2)^T$. Therefore, the slope of the secant line $S$ passing through the points $|\acute{\mathbf{p}}\rangle$ and $|\acute{\mathbf{q}}\rangle$ (see Figure 2.13) is given by

$$\frac{\Delta z}{h} = \frac{f|\mathbf{q}\rangle - f|\mathbf{p}\rangle}{h} = \frac{f(a + hu_1, b + hu_2)^T - f(a, b)^T}{h}. \tag{2.120}$$

Observe that (2.120) also gives the average rate of change of $z = f|\mathbf{x}\rangle$ from $|\mathbf{p}\rangle \equiv (a, b)^T$ to $|\mathbf{q}\rangle \equiv (a + \Delta x, b + \Delta y)^T = (a + hu_1, b + hu_2)^T$ in the direction of $|\mathbf{u}\rangle$.

If we let $h$ approach zero in (2.120), we see that the slope of the secant line $S$ approaches the slope of the tangent line at $|\acute{\mathbf{p}}\rangle$. Also, the average rate of change of $z$ approaches the (instantaneous) rate of change of $z$ at $|\mathbf{p}\rangle$ in the direction of $|\mathbf{u}\rangle$. This limit, whenever it exists, is called the directional derivative of $f$ at $|\mathbf{p}\rangle \equiv (a, b)^T$ in the direction of $|\mathbf{u}\rangle$. Since the point $|\mathbf{p}\rangle$ is arbitrary, we can replace it by $|\mathbf{p}\rangle \equiv (x, y)^T$ and define the directional derivative of $f$ at any point as follows.

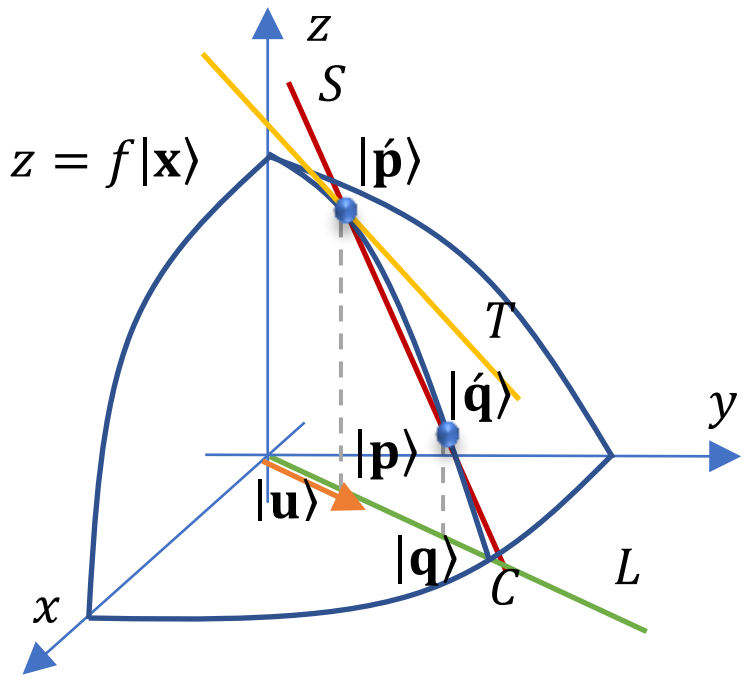

**Figure 2.13.** The secant line passes through the points $|\acute{\mathbf{p}}\rangle$ and $|\acute{\mathbf{q}}\rangle$ on the curve $C$.

**Definition (Directional Derivative):** Let $f$ be a function of $|\mathbf{x}\rangle = (x, y)^T$, and let $|\mathbf{u}\rangle = (u_1, u_2)^T$ be a unit vector. Then the directional derivative of $f$ at $|\mathbf{x}\rangle$ in the direction of $|\mathbf{u}\rangle$ is

$$D_{|\mathbf{u}\rangle} f|\mathbf{x}\rangle = \lim_{h\to 0} \frac{f|\mathbf{x} + h\mathbf{u}\rangle - f|\mathbf{x}\rangle}{h}, \tag{2.121.1}$$

$$D_{|\mathbf{u}\rangle} f|\mathbf{x}\rangle = \lim_{h\to 0} \frac{f(x + hu_1, y + hu_2)^T - f(x, y)^T}{h}, \tag{2.121.2}$$

if this limit exists.

**Remark:**

If $|\mathbf{u}\rangle = |\mathbf{i}\rangle$ ($u_1 = 1$ and $u_2 = 0$), then (2.121) gives

$$D_{|\mathbf{i}\rangle} f|\mathbf{x}\rangle = \lim_{h\to 0} \frac{f(x + h, y)^T - f(x, y)^T}{h} = f_x|\mathbf{x}\rangle. \tag{2.122}$$

That is, the directional derivative of $f$ in the $x$-direction is the partial derivative of $f$ in the $x$-direction. Similarly, we can show that $D_{|\mathbf{j}\rangle} f|\mathbf{x}\rangle = f_y|\mathbf{x}\rangle$.

**Theorem 2.3:** If $f$ is a differentiable function of $|\mathbf{x}\rangle = (x, y)^T$, then $f$ has a directional derivative in the direction of any unit vector $|\mathbf{u}\rangle = (u_1, u_2)^T$ and

$$D_{|\mathbf{u}\rangle} f|\mathbf{x}\rangle = f_x|\mathbf{x}\rangle u_1 + f_y|\mathbf{x}\rangle u_2. \tag{2.123}$$

**Proof:**

Fix the point $|\hat{\mathbf{x}}\rangle = (a, b)^T$. Using local linearity, we have

$$\Delta f = f|\mathbf{x}\rangle - f|\hat{\mathbf{x}}\rangle \approx f_x|\hat{\mathbf{x}}\rangle \Delta x + f_y|\hat{\mathbf{x}}\rangle \Delta y = f_x|\hat{\mathbf{x}}\rangle h u_1 + f_y|\hat{\mathbf{x}}\rangle h u_2.$$

Thus, dividing by $h$ gives

$$\frac{\Delta f}{h} \approx f_x|\hat{\mathbf{x}}\rangle u_1 + f_y|\hat{\mathbf{x}}\rangle u_2.$$

This approximation becomes exact as $h \to 0$, so we have the following formula:

$$D_{|\mathbf{u}\rangle} f|\hat{\mathbf{x}}\rangle = f_x|\hat{\mathbf{x}}\rangle\, u_1 + f_y|\hat{\mathbf{x}}\rangle\, u_2.$$

Finally, since $|\hat{\mathbf{x}}\rangle = (a, b)^T$ is arbitrary, we may replace it by $|\mathbf{x}\rangle = (x, y)^T$ and the result follows.

■

### 2.6.3 The Gradient and Tangent Planes of Functions of Two- and Three-Variables

The directional derivative $D_{|\mathbf{u}\rangle} f|\mathbf{x}\rangle$ can be written as the dot product of the unit vector $|\mathbf{u}\rangle = (u_1, u_2)^T$ and the gradient vector $\nabla f = (f_x, f_y)^T$. We have,

$$D_{|\mathbf{u}\rangle} f|\mathbf{x}\rangle = f_x|\mathbf{x}\rangle u_1 + f_y|\mathbf{x}\rangle u_2 = \langle \nabla f(\mathbf{x})|\mathbf{u}\rangle. \tag{2.124}$$

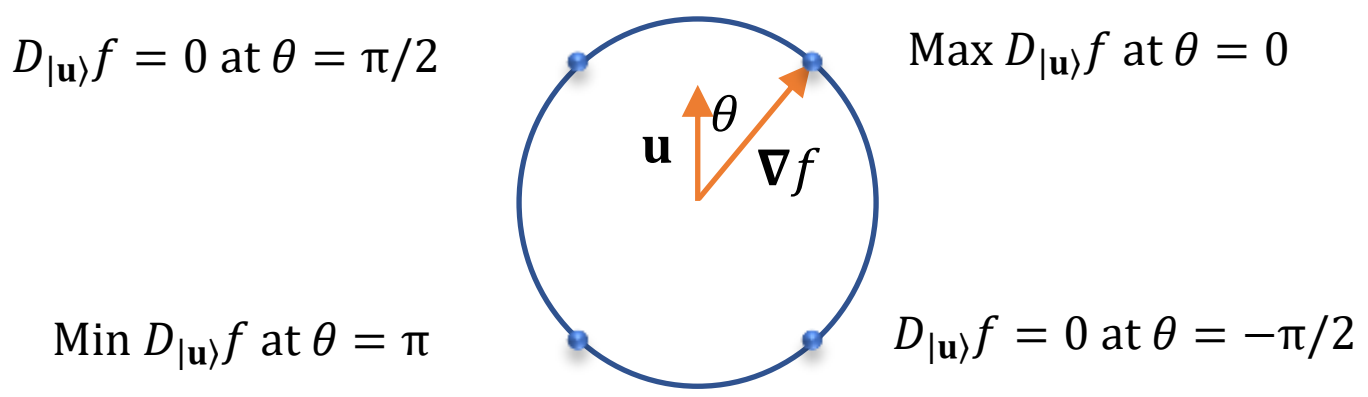

**Figure 2.14.** Values of the directional derivative at different angles to the gradient.

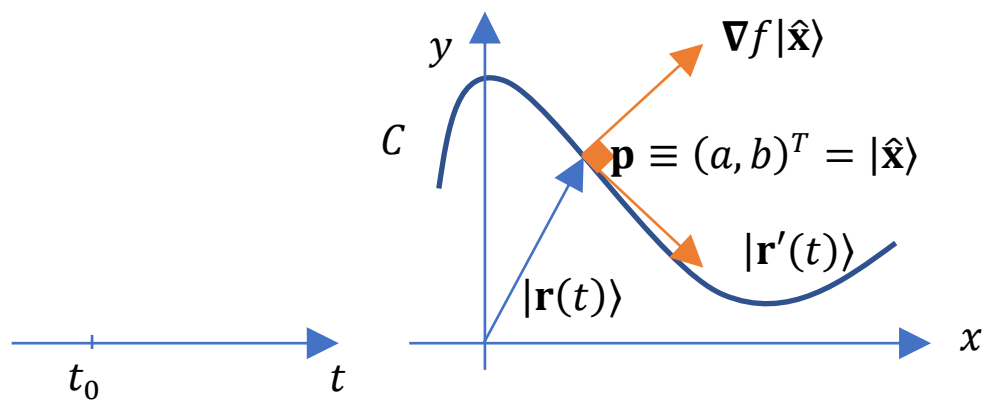

**Figure 2.15.** The curve may be represented by $|\mathbf{r}(t)\rangle = (x, y)^T = (g(t), h(t))^T$.

**Theorem 2.4:** If $f$ is a differentiable function of $|\mathbf{x}\rangle = (x, y)^T$, then $f$ has a directional derivative in the direction of any unit vector $|\mathbf{u}\rangle$, and

$$D_{|\mathbf{u}\rangle}f|\mathbf{x}\rangle = \langle\nabla f(\mathbf{x})|\mathbf{u}\rangle. \tag{2.125}$$

Suppose $\theta$ is the angle between the vectors $\nabla f|\mathbf{x}\rangle$ and $|\mathbf{u}\rangle$. At the point $(a, b)^T$, we have

$$\langle\nabla f(\mathbf{x})|\mathbf{u}\rangle = \|\nabla f\|\|\mathbf{u}\| \cos\theta = \|\nabla f\| \cos\theta. \tag{2.126}$$

Imagine that $\nabla f$ is fixed and that $|\mathbf{u}\rangle$ can rotate. (see Figure 2.14). The maximum value of $D_{|\mathbf{u}\rangle}f$ occurs when $\cos\theta = 1$, so $\theta = 0$ and $|\mathbf{u}\rangle$ is pointing in the direction of $\nabla f|\mathbf{x}\rangle$ . Then

$$\text{Max}\left[D_{|\mathbf{u}\rangle}f\right] = \|\nabla f\|. \tag{2.127}$$

The minimum value of $D_{|\mathbf{u}\rangle}f$ occurs when $\cos\theta = -1$, so $\theta = \pi$ and $|\mathbf{u}\rangle$ is pointing in the direction opposite to $\nabla f$. Then

$$\text{Min}\left[D_{|\mathbf{u}\rangle}f\right] = -\|\nabla f\|. \tag{2.128}$$

Hence, we have

**Theorem 2.5:** Suppose $f$ is differentiable at the point $|\mathbf{x}\rangle = (x, y)^T$.
1. If $\nabla f|\mathbf{x}\rangle = |\mathbf{0}\rangle$, then $D_{|\mathbf{u}\rangle}f|\mathbf{x}\rangle = 0$ for every $|\mathbf{u}\rangle$.
2. The maximum value of $D_{|\mathbf{u}\rangle}f|\mathbf{x}\rangle$ is $|\nabla f(\mathbf{x})|$, and this occurs when $|\mathbf{u}\rangle$ has the same direction as $\nabla f$.
3. The minimum value of $D_{|\mathbf{u}\rangle}f|\mathbf{x}\rangle$ is $-|\nabla f(\mathbf{x})|$, and this occurs when $|\mathbf{u}\rangle$ has the direction of $-\nabla f$.

**Remarks:**

1. Property (2) of Theorem 2.5 tells us that $f$ increases most rapidly in the direction of $\nabla f$. This direction is called the direction of the steepest ascent.
2. Property (3) of Theorem 2.5 says that $f$ decreases most rapidly in the direction of $-\nabla f$. This direction is called the direction of the steepest descent.

We can now give the geometric interpretation of the gradient in 2D. Suppose that the curve $C$ is represented by the vector function

$$|\mathbf{r}(t)\rangle = (g(t), h(t))^T, \tag{2.129}$$

where $g$ and $h$ are differentiable functions, $a = g(t_0)$ and $b = h(t_0)$, and $t_0$ lies in the parameter interval (Figure 2.15). Since the point $(x, y)^T = (g(t), h(t))^T$ lies on $C$, we have

$$f|\mathbf{x}\rangle = f|\mathbf{r}(t)\rangle = f(g(t), h(t))^T = c, \tag{2.130}$$

for all $t$ in the parameter interval.

Differentiating both sides of this equation with respect to $t$ and using the chain rule for a function of two variables, we obtain

$$\frac{\partial f}{\partial x}\frac{dx}{dt} + \frac{\partial f}{\partial y}\frac{dy}{dt} = 0. \tag{2.131}$$

Using $\nabla f|\mathbf{x}\rangle = \left(\frac{\partial f}{\partial x}, \frac{\partial f}{\partial y}\right)^T$ and $|\mathbf{r}'(t)\rangle = \left(\frac{dx}{dt}, \frac{dy}{dt}\right)^T$, we can write (2.131) in the form

$$\langle \nabla f(\mathbf{x}) | \mathbf{r}'(t)\rangle = 0. \tag{2.132}$$

In particular, when $t = t_0$, i.e., $|\hat{\mathbf{x}}\rangle = (a, b)^T$, we have

$$\langle \nabla f(\hat{\mathbf{x}}) | \mathbf{r}'(t_0)\rangle = 0. \tag{2.133}$$

Thus, if $|\mathbf{r}'(t_0)\rangle \neq |\mathbf{0}\rangle$, the vector $\nabla f|\hat{\mathbf{x}}\rangle$ is orthogonal to the tangent vector $|\mathbf{r}'(t_0)\rangle$ at $(a, b)^T$. Loosely speaking, we have demonstrated the following:

**Theorem 2.6.1:** $\nabla f$ is orthogonal to the level curve $f|\mathbf{x}\rangle = c$ at point $|\hat{\mathbf{x}}\rangle$.

***Example 2.3***

Let $f|\mathbf{x}\rangle = x^2 - y^2$. Find the level curve of $f$ passing through the point $|\hat{\mathbf{x}}\rangle = (5,3)^T$. Also, find the gradient of $f$ at that point, and make a sketch of both the level curve and the gradient vector.

***Solution***

Since $f|\hat{\mathbf{x}}\rangle = 25 - 9 = 16$, the required level curve is the hyperbola $x^2 - y^2 = 16$. The gradient of $f$ at any point $|\mathbf{x}\rangle = (x, y)^T$ is

$$\nabla f|\mathbf{x}\rangle = (2x, -2y)^T,$$

and, in particular, the gradient of $f$ at $|\hat{\mathbf{x}}\rangle$ is

$$\nabla f|\hat{\mathbf{x}}\rangle = (10, -6)^T.$$

The level curve and $\nabla f|\mathbf{x}\rangle$ are shown in Figure 2.16.

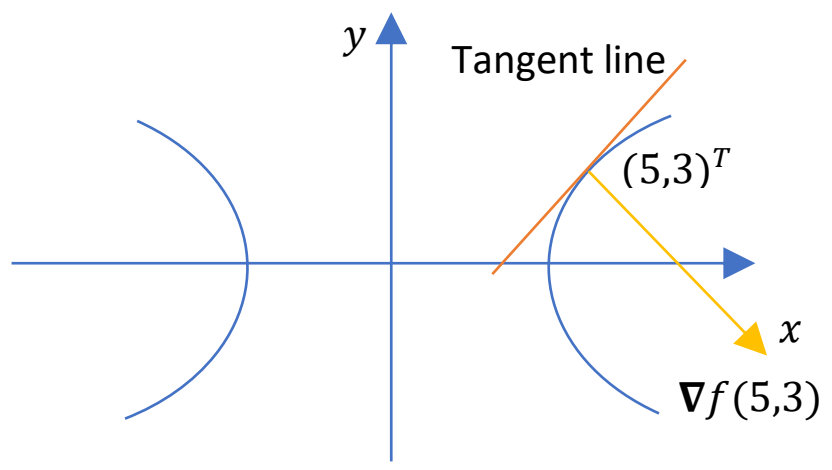

**Figure 2.16.** The gradient $\nabla f|\mathbf{x}\rangle$ is orthogonal to the level curve $x^2 - y^2 = 16$ at $|\hat{\mathbf{x}}\rangle = (5,3)^T$.

### 2.6.4 Functions of Three Variables and Level Surfaces

A function $F$ of three variables is a rule that assigns to each ordered triple $|\mathbf{x}\rangle = (x, y, z)^T$ in a domain $D = \{|\mathbf{x}\rangle = (x, y, z)^T : x, y, z \in \mathbb{R}\}$ a unique real number $w$ denoted by $F|\mathbf{x}\rangle = w$. For example, $F|\mathbf{x}\rangle = xyz$. Since the graph of a function of three variables is composed of the points $(x, y, z, w)^T$, where $w = F|\mathbf{x}\rangle$, lying in four-dimensional space, we cannot draw the graphs of such functions. But by examining the level surfaces, which are the surfaces with equations $F|\mathbf{x}\rangle = k$, $k$ a constant, we are often able to gain some insight into the nature of $F$.

***Example 2.4***

Find the level surfaces of the function defined by $F|\mathbf{x}\rangle = x^2 + y^2 + z^2$.

***Solution***

The required level surfaces of $F$ are the graphs of the equations $x^2 + y^2 + z^2 = k$, where $k \geq 0$. These surfaces are concentric spheres of radius $\sqrt{k}$ centered at the origin (see Figure 2.17). Observe that $F$ has the same value for all points $(x, y, z)^T$ lying on any such sphere.

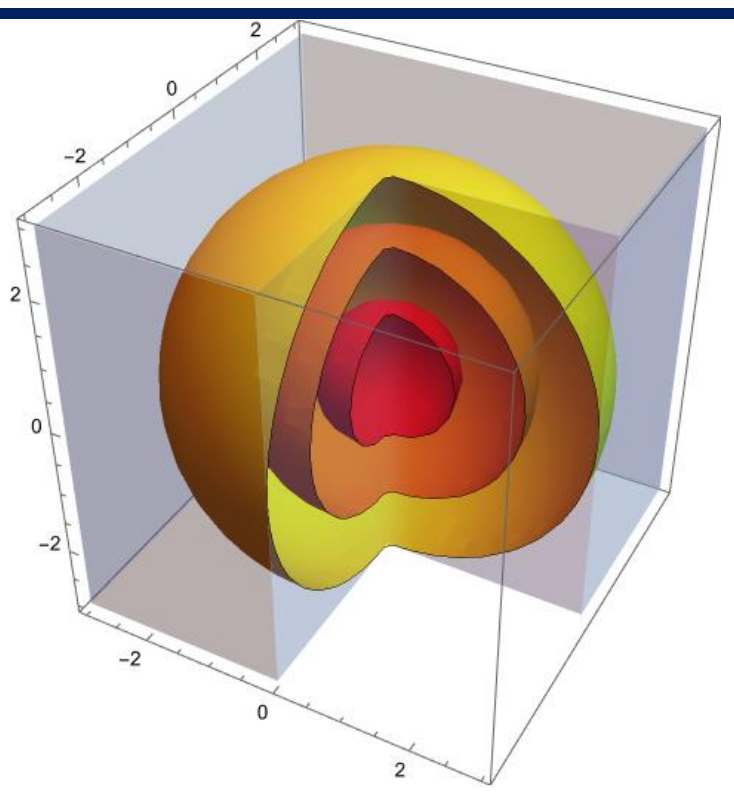

**Figure 2.17.** The level surfaces of $F|\mathbf{x}\rangle = x^2 + y^2 + z^2$ corresponding to $k = 1, 4, 9$.

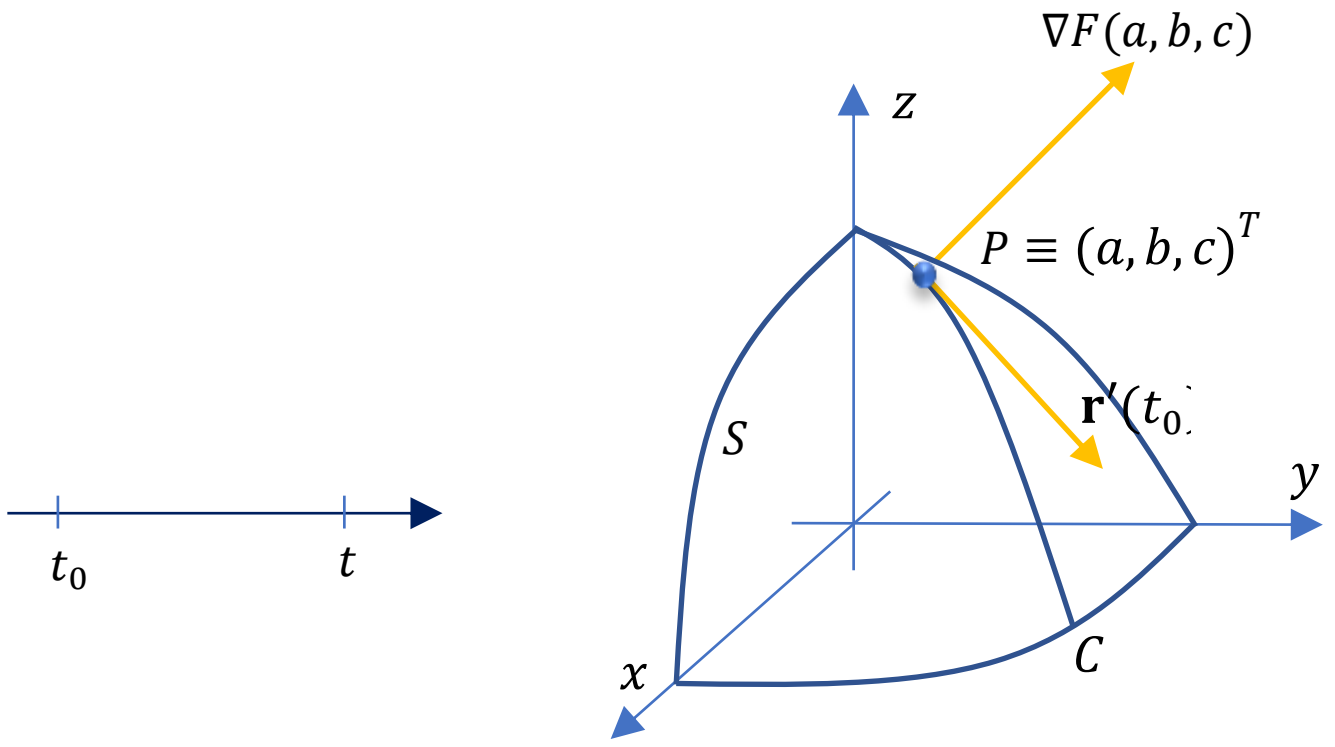

**Figure 2.18.** The curve $C$ is described by $|\mathbf{r}(t)\rangle = (f(t), g(t), h(t))^T$ with $P \equiv |\hat{\mathbf{x}}\rangle = (a, b, c)^T$ corresponding to $t_0$.

Suppose that $F|\mathbf{x}\rangle = k$ is the level surface $S$ of a differentiable function $F$ defined by $T = F|\mathbf{x}\rangle$, $|\mathbf{x}\rangle = (x, y, z)^T$. Suppose that $|\hat{\mathbf{x}}\rangle \equiv (a, b, c)^T$ is a point on $S$ and let $C$ be a smooth curve on $S$ passing through $|\hat{\mathbf{x}}\rangle$. Then $C$ can be described by the vector function $|\mathbf{r}(t)\rangle = (f(t), g(t), h(t))^T$ where $f(t_0) = a$, $g(t_0) = b$ , $h(t_0) = c$, and $t_0$ is a point in the parameter interval (see Figure 2.18).

Since the point $|\mathbf{x}\rangle = (x, y, z)^T = (f(t), g(t), h(t))^T$ lies on $S$, we have, $F|\mathbf{x}\rangle = F(f(t), g(t), h(t))^T = k$, for all $t$ in the parameter interval. If $|\mathbf{r}\rangle$ is differentiable, then we can use the chain rule to differentiate both sides of this equation to obtain,

$$\frac{\partial F}{\partial x}\frac{dx}{dt} + \frac{\partial F}{\partial y}\frac{dy}{dt} + \frac{\partial F}{\partial z}\frac{dz}{dt} = 0. \tag{2.134}$$

This is the same as

$$(F_x, F_y, F_z) \cdot \begin{pmatrix} \frac{dx}{dt} \\ \frac{dy}{dt} \\ \frac{dz}{dt} \end{pmatrix} = 0, \tag{2.135}$$

or, in an even more abbreviated form, $\langle \nabla F | \mathbf{r}'(t)\rangle = 0$. At $t = t_0$, i.e., $|\hat{\mathbf{x}}\rangle = (a, b, c)^T$, we have $\langle \nabla F(\hat{\mathbf{x}}) | \mathbf{r}'(t_0)\rangle = 0$. This shows that if $|\mathbf{r}'(t_0)\rangle \neq |\mathbf{0}\rangle$, then the gradient vector $\nabla F|\hat{\mathbf{x}}\rangle$ is orthogonal to the tangent vector $|\mathbf{r}'(t_0)\rangle$ to $C$ at $|\hat{\mathbf{x}}\rangle$ (see Figure 2.18). Since this argument holds for any differentiable curve passing through $|\hat{\mathbf{x}}\rangle$ on $C$, we have shown

that $\nabla F|\hat{\mathbf{x}}\rangle$ is orthogonal to the tangent vector of every curve on $S$ passing through $|\hat{\mathbf{x}}\rangle$. Thus, we have demonstrated the following result.

**Theorem 2.6.2: $\nabla F$** is orthogonal to the level surface $F|\mathbf{x}\rangle = 0$ at $|\hat{\mathbf{x}}\rangle$.
The gradient $\nabla F|\hat{\mathbf{x}}\rangle$ is orthogonal to the tangent vector of every curve on $S$ passing through $\mathbf{p} \equiv |\hat{\mathbf{x}}\rangle = (a, b, c)^T$ (Figure 2.18). This suggests that we define the tangent plane to $S$ at $\mathbf{p}$ to be the plane passing through $\mathbf{p}$ and containing all these tangent vectors. Equivalently, the tangent plane should have $\nabla F|\hat{\mathbf{x}}\rangle$ as a normal vector.

**Definition (Normal Line):** Let $\mathbf{p} \equiv |\hat{\mathbf{x}}\rangle = (a, b, c)^T$ be a point on the surface $S$ described by $F|\mathbf{x}\rangle = 0$, where $F$ is differentiable at $\mathbf{p}$, and suppose that $\nabla F|\hat{\mathbf{x}}\rangle \neq |\mathbf{0}\rangle$. Then the tangent plane to $S$ at $\mathbf{p}$ is the plane that passes through $\mathbf{p}$ and has normal vector $\nabla F|\hat{\mathbf{x}}\rangle$. The normal line to $S$ at $\mathbf{p}$ is the line that passes through $\mathbf{p}$ and has the same direction as $\nabla F|\hat{\mathbf{x}}\rangle$.
The equation of the tangent plane is

$$(x-a)F_x|\hat{\mathbf{x}}\rangle + (y-b)F_y|\hat{\mathbf{x}}\rangle + (z-c)F_z|\hat{\mathbf{x}}\rangle = 0. \tag{2.136}$$

## 2.7 Optimality Criteria (General Case $n$-Variables)

The present section will largely consist of a generalization of the results of Section 2.6 to the case of $n$-variables [38,44-48]. We have,

$$\text{optimize: } z = f|\mathbf{x}\rangle, \qquad \text{where } |\mathbf{x}\rangle = (x_1, x_2, \ldots, x_n)^T. \tag{2.137}$$

The optimality criteria can be derived using the definition of a locally optimal point and the Taylor series expansion of a multivariable function.

**Definition ($\epsilon$-Neighborhood):** An $\epsilon$-neighborhood ($\epsilon > 0$) around $|\hat{\mathbf{x}}\rangle$ is the set of all vectors $|\mathbf{x}\rangle$ such that

$$\begin{aligned}\|\mathbf{x}-\hat{\mathbf{x}}\| &= \langle \mathbf{x}-\hat{\mathbf{x}}|\mathbf{x}-\hat{\mathbf{x}}\rangle \\ &= (x_1-\hat{x}_1)^2 + (x_2-\hat{x}_2)^2 + \cdots + (x_n-\hat{x}_n)^2 \leq \epsilon^2.\end{aligned} \tag{2.138}$$

In geometrical terms, an $\epsilon$-neighborhood around $|\hat{\mathbf{x}}\rangle$ is the interior and boundary of an $n$-dimensional sphere of radius $\epsilon$ centered at $|\hat{\mathbf{x}}\rangle$.

**Definition (Local and Global Minimizer):** An objective function $f|\mathbf{x}\rangle$ has a local minimizer at $|\hat{\mathbf{x}}\rangle$ if there exists an $\epsilon$-neighborhood around $|\hat{\mathbf{x}}\rangle$ such that $f|\mathbf{x}\rangle \geq f|\hat{\mathbf{x}}\rangle$ for all $|\mathbf{x}\rangle$ in this $\epsilon$-neighborhood at which the function is defined. If the condition is met for every positive $\epsilon$ (no matter how large), then $f|\mathbf{x}\rangle$ has a global minimizer at $|\hat{\mathbf{x}}\rangle$.

**Definition (Saddle Point):** A saddle point or minimax point is a point on the surface of the graph of a function where the slopes (derivatives) in orthogonal directions are all zero (a critical point) but which is not a local extremum of the function.

In the most general terms, a saddle point for a smooth function (whose graph is a curve, surface, or hypersurface) is a stationary point such that the curve/surface/etc. in the neighborhood of that point is not entirely on any side of the tangent space at that point. (see Figure 2.19)
Remember, the gradient vector at any point $|\hat{\mathbf{x}}\rangle$ is represented by $\nabla f|\hat{\mathbf{x}}\rangle$ which is an $n$-dimensional vector given as follows:

$$\nabla f = \begin{pmatrix} \dfrac{\partial f}{\partial x_1} \\ \dfrac{\partial f}{\partial x_2} \\ \vdots \\ \dfrac{\partial f}{\partial x_n} \end{pmatrix}. \tag{2.139}$$

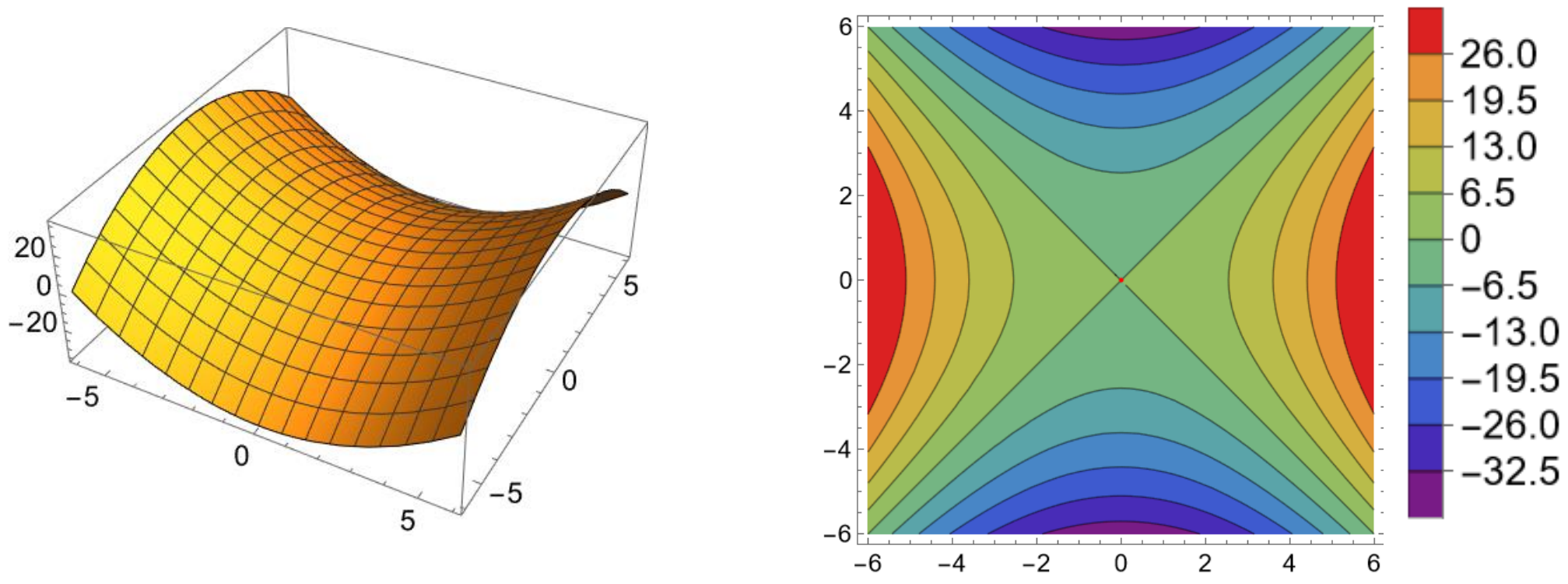

**Figure 2.19.** A saddle point on the graph of $z = x^2 - y^2$.

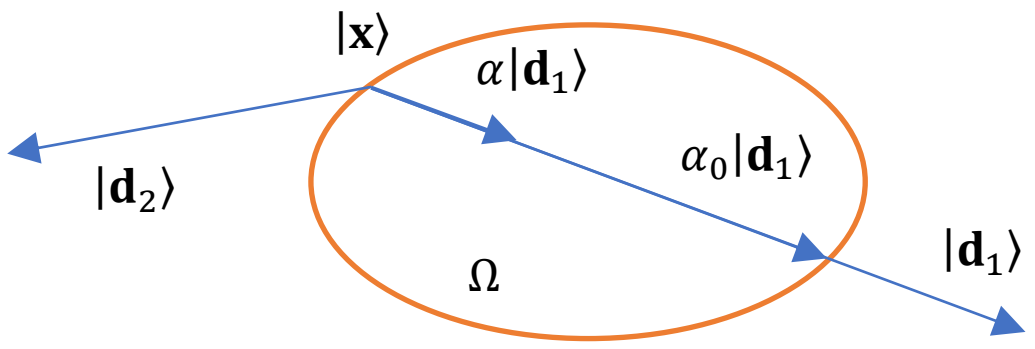

**Figure 2.20.** $|\mathbf{d}_1\rangle$ is a feasible direction, however $|\mathbf{d}_2\rangle$ is not a feasible direction.

The first-order partial derivatives can be calculated numerically using finite difference. The second-order derivatives in multivariable functions form a matrix, $\nabla^2 f|\hat{\mathbf{x}}\rangle$ (the Hessian matrix) given as follows:

$$\mathbf{H}_f|\hat{\mathbf{x}}\rangle \equiv \nabla^2 f|\hat{\mathbf{x}}\rangle = \begin{pmatrix} \frac{\partial^2 f}{\partial x_1^2} & \frac{\partial^2 f}{\partial x_1 \partial x_2} & \cdots & \frac{\partial^2 f}{\partial x_1 \partial x_n} \\ \frac{\partial^2 f}{\partial x_2 \partial x_1} & \frac{\partial^2 f}{\partial x_2^2} & \cdots & \vdots \\ \cdots & \cdots & \ddots & \vdots \\ \frac{\partial^2 f}{\partial x_n \partial x_1} & \frac{\partial^2 f}{\partial x_n \partial x_2} & \cdots & \frac{\partial^2 f}{\partial x_n^2} \end{pmatrix} = \left(\frac{\partial^2 f}{\partial x_i \partial x_j}\right), \quad (i, j = 1, 2, \ldots, n). \tag{2.140}$$

The second-order partial derivatives can also be calculated numerically using finite difference.

**Theorem 2.7 (Generalized of Theorem 2.5):** For small displacements from $|\hat{\mathbf{x}}\rangle$ in various directions, the direction of maximum increase in $f|\hat{\mathbf{x}}\rangle$ is the direction of the vector $\nabla f|\hat{\mathbf{x}}\rangle$.

**Proof:**

For any fixed vector $|\hat{\mathbf{x}}\rangle$ and any unit vector $|\mathbf{u}\rangle$, the directional derivative,

$$D_{|\mathbf{u}\rangle} f|\hat{\mathbf{x}}\rangle = \langle \nabla f(\hat{\mathbf{x}})|\mathbf{u}\rangle,$$

gives the rate of change of $f|\mathbf{x}\rangle$ at $|\hat{\mathbf{x}}\rangle$ in the direction of $|\mathbf{u}\rangle$. Since

$$\langle \nabla f|\mathbf{u}\rangle = \|\nabla f\|\|\mathbf{u}\| \cos\theta = \|\nabla f\| \cos\theta,$$

the greatest increase in $f|\mathbf{x}\rangle$ occurs when $\theta = 0$, i.e., when $|\mathbf{u}\rangle$ is in the same direction as $\nabla f$. Therefore, any (small) movement from $|\hat{\mathbf{x}}\rangle$ in the direction of $\nabla f|\hat{\mathbf{x}}\rangle$ will, initially, increase the function over $f|\hat{\mathbf{x}}\rangle$ as rapidly as possible.

■

**Definition (Feasible Direction):** A vector $|\mathbf{d}\rangle \in \mathbb{R}^n$, $|\mathbf{d}\rangle \neq |\mathbf{0}\rangle$, is a feasible direction at $|\mathbf{x}\rangle \in \Omega$ if there exists $\alpha_0 > 0$ such that $|\mathbf{x}\rangle + \alpha|\mathbf{d}\rangle \in \Omega$ for all $\alpha \in [0, \alpha_0]$. See Figure 2.20.

**Theorem 2.8 (First-Order Necessary Condition):** Let $\Omega$ be a subset of $\mathbb{R}^n$ and $f \in C^1$ a real-valued function on $\Omega$. If $|\mathbf{x}^*\rangle$ is a local minimizer of $f$ over $\Omega$, then for any feasible direction $|\mathbf{d}\rangle$ at $|\mathbf{x}^*\rangle$, we have

$$\langle \mathbf{d}|\nabla f(\mathbf{x}^*)\rangle \geq 0. \tag{2.141}$$

If $|\mathbf{x}^*\rangle$ is located in the interior of $\Omega$, then

$$\nabla f(\mathbf{x}^*) = |\mathbf{0}\rangle. \tag{2.142}$$

**Proof:**

If $|\mathbf{d}\rangle$ is a feasible direction at $|\mathbf{x}^*\rangle$, then we have

$$|\mathbf{x}\rangle = |\mathbf{x}^*\rangle + \alpha|\mathbf{d}\rangle \in \Omega.$$

From the Taylor theorem,

$$f|\mathbf{x}\rangle = f|\mathbf{x}^*\rangle + \alpha\langle \mathbf{d}|\nabla f(\mathbf{x}^*)\rangle + O(\alpha).$$

If

$$\langle \mathbf{d}|\nabla f(\mathbf{x}^*)\rangle < 0,$$

then as $\alpha \to 0$

$$\alpha\langle \mathbf{d}|\nabla f(\mathbf{x}^*)\rangle + O(\alpha) < 0,$$

and so

$$f|\mathbf{x}\rangle < f|\mathbf{x}^*\rangle.$$

This contradicts the assumption that $|\mathbf{x}^*\rangle$ is a local minimizer. So, a necessary condition for $|\mathbf{x}^*\rangle$ to be a minimizer is $\langle \mathbf{d}|\nabla f(\mathbf{x}^*)\rangle \geq 0$.

If $|\mathbf{x}^*\rangle$ is in the interior of $\Omega$, vectors exist in all directions which are feasible. Thus, a direction $|\mathbf{d}\rangle = |\mathbf{d}_1\rangle$ yields

$$\langle \mathbf{d}_1|\nabla f(\mathbf{x}^*)\rangle \geq 0.$$

Similarly, for a direction $|\mathbf{d}\rangle = -|\mathbf{d}_1\rangle$

$$-\langle \mathbf{d}_1|\nabla f(\mathbf{x}^*)\rangle \geq 0.$$

In this case, a necessary condition for $|\mathbf{x}^*\rangle$ to be a local minimizer is

$$\nabla f(\mathbf{x}^*) = |\mathbf{0}\rangle.$$

■

**Theorem 2.9 (Second-Order Necessary Condition):** Let $\Omega$ be a subset of $\mathbb{R}^n$ and $f \in C^2$ a real-valued function on $\Omega$, $|\mathbf{x}^*\rangle$ is a local minimizer of $f$ over $\Omega$, and $|\mathbf{d}\rangle$ a feasible direction at $|\mathbf{x}^*\rangle$. If $\langle \mathbf{d}|\nabla f(\mathbf{x}^*)\rangle = 0$, then

$$\langle \mathbf{d}|\mathbf{H}(\mathbf{x}^*)|\mathbf{d}\rangle \geq 0, \tag{2.143}$$

where $\mathbf{H}$ is the Hessian of $f$.

If $|\mathbf{x}^*\rangle$ is a local minimizer in the interior of $\Omega$, then $\nabla f(\mathbf{x}^*) = |\mathbf{0}\rangle$ and $\langle \mathbf{d}|\mathbf{H}(\mathbf{x}^*)|\mathbf{d}\rangle \geq 0$ for all $|\mathbf{d}\rangle \neq |\mathbf{0}\rangle$.

(This condition is equivalent to stating that $\mathbf{H}(\mathbf{x}^*)$ is positive semidefinite.)

**Proof:**

If $|\mathbf{d}\rangle$ is a feasible direction at $|\mathbf{x}^*\rangle$, then we have

$$|\mathbf{x}\rangle = |\mathbf{x}^*\rangle + \alpha|\mathbf{d}\rangle \in \Omega.$$

From the Taylor theorem,

$$f|\mathbf{x}\rangle = f|\mathbf{x}^*\rangle + \alpha\langle \mathbf{d}|\nabla f(\mathbf{x}^*)\rangle + \frac{1}{2}\alpha^2\langle \mathbf{d}|\mathbf{H}(\mathbf{x}^*)|\mathbf{d}\rangle + O(\alpha^2).$$

If

$$\langle \mathbf{d}|\mathbf{H}(\mathbf{x}^*)|\mathbf{d}\rangle < 0,$$

then as $\alpha \to 0$

$$\frac{1}{2}\alpha^2\langle \mathbf{d}|\mathbf{H}(\mathbf{x}^*)|\mathbf{d}\rangle + O(\alpha^2) < 0,$$

and so

$$f|\mathbf{x}\rangle < f|\mathbf{x}^*\rangle.$$

This contradicts the assumption that $|\mathbf{x}^*\rangle$ is a local minimizer. So that, if $\langle \mathbf{d}|\nabla f(\mathbf{x}^*)\rangle = 0$ then $\langle \mathbf{d}|\mathbf{H}(\mathbf{x}^*)|\mathbf{d}\rangle \geq 0$.

If $|\mathbf{x}^*\rangle$ is a local minimizer in the interior of $\Omega$, then all vectors $|\mathbf{d}\rangle$ are feasible directions and, therefore, the second part holds.

■

***Example 2.5***

Let $f|\mathbf{x}\rangle = x_1^2 - x_2^2$. The Theorem 2.8 requires that $\nabla f|\mathbf{x}\rangle = (2x_1, -2x_2)^T = |\mathbf{0}\rangle$. Thus, $|\mathbf{x}\rangle = (0,0)^T$ satisfies the Theorem 2.8. The Hessian matrix of $f$ is

$$\mathbf{H} = \begin{pmatrix} 2 & 0 \\ 0 & -2 \end{pmatrix}.$$

The Hessian matrix is indefinite; that is, for some $|\mathbf{d}_1\rangle \in \mathbb{R}^2$ we have $\langle \mathbf{d}_1|\mathbf{H}|\mathbf{d}_1\rangle > 0$, e.g., $|\mathbf{d}_1\rangle = (1,0)^T$ and, for some $|\mathbf{d}_2\rangle$, we have $\langle \mathbf{d}_2|\mathbf{H}|\mathbf{d}_2\rangle < 0$, e.g., $|\mathbf{d}_2\rangle = (0,1)^T$. Thus, $|\mathbf{x}\rangle = (0,0)^T$ does not satisfy Theorem 2.9, and hence it is not a minimizer. Figure 2.21 shows the graph of $f|\mathbf{x}\rangle = x_1^2 - x_2^2$.

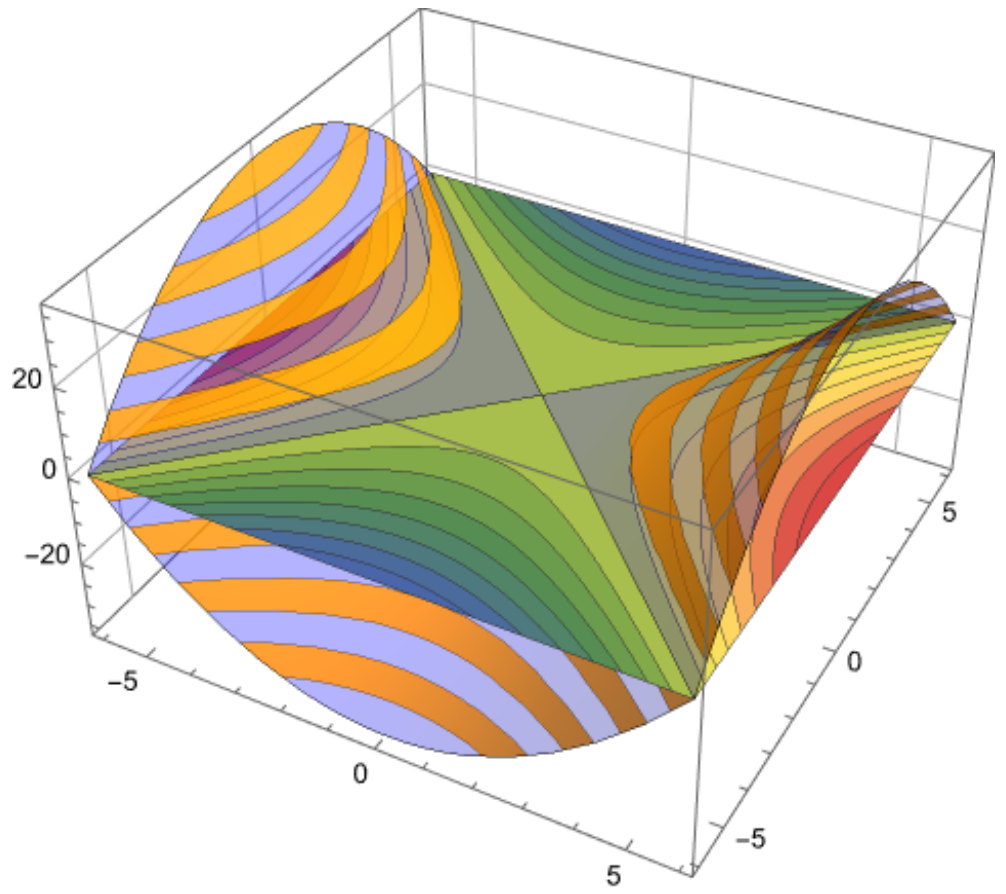

**Figure 2.21.** 3D and level curves of the function $f|\mathbf{x}\rangle = x_1^2 - x_2^2$.

**Theorem 2.10 (Second-Order Necessary Condition):** Let $\Omega$ be a subset of $\mathbb{R}^n$ and $f \in C^2$ a real-valued function on $\Omega$, $|\mathbf{x}^*\rangle$ is located in the interior of $\Omega$, then the conditions

1. $\nabla f(\mathbf{x}^*) = |\mathbf{0}\rangle$
2. $\mathbf{H}(\mathbf{x}^*)$ is positive definite

are sufficient for $|\mathbf{x}^*\rangle$ to be a strict local minimizer.

**Proof:**

If $|\mathbf{d}\rangle$ is a feasible direction at $|\mathbf{x}^*\rangle$, then we have

$$|\mathbf{x}\rangle = |\mathbf{x}^*\rangle + \alpha|\mathbf{d}\rangle \in \Omega.$$

From the Taylor theorem,

$$f|\mathbf{x}\rangle = f|\mathbf{x}^*\rangle + \alpha\langle\mathbf{d}|\nabla f(\mathbf{x}^*)\rangle + \frac{1}{2}\alpha^2\langle\mathbf{d}|\mathbf{H}(\mathbf{x}^*)|\mathbf{d}\rangle + O(\alpha^2),$$

and if condition (1) is satisfied, we have,

$$f|\mathbf{x}\rangle = f|\mathbf{x}^*\rangle + \frac{1}{2}\alpha^2\langle\mathbf{d}|\mathbf{H}(\mathbf{x}^*)|\mathbf{d}\rangle + O(\alpha^2).$$

Now, if condition (2) is satisfied, then

$$\frac{1}{2}\alpha^2\langle\mathbf{d}|\mathbf{H}(\mathbf{x}^*)|\mathbf{d}\rangle + O(\alpha^2) > 0, \qquad \text{as } \|\mathbf{d}\| \to 0.$$

Therefore,

$$f|\mathbf{x}\rangle > f|\mathbf{x}^*\rangle.$$

that is, $|\mathbf{x}^*\rangle$ is a strict local minimizer.

■

## 2.8 Gradient Descent (GD) Algorithm

Obtaining analytical solutions for optimization problems involving multivariable functions is often much more challenging than for single-variable functions. This difficulty arises primarily due to the complexity of multivariable functions and the increased number of variables involved. In many cases, the objective function may not have a closed-form expression, or even if it does, finding analytical solutions for minimizing it can be intractable or even impossible. As a result, numerical optimization methods become indispensable for approximating solutions to optimization problems involving multivariable functions.

Numerical optimization methods, such as GD, Newton's method, and various derivative-free optimization algorithms, are commonly used to find approximate solutions to multivariable optimization problems [44-48]. These methods iteratively explore the parameter space, adjusting the parameters to minimize (or maximize) the objective function while satisfying certain constraints and tolerances.

While these numerical methods do not guarantee finding the global optimum (especially in non-convex and high-dimensional spaces), they are effective in finding local minima within specified tolerances. By adjusting parameters such as the step size, initial guess, and termination criteria, practitioners can tailor these numerical methods to their specific optimization problem and computational resources.

GD is a first-order optimization algorithm used to minimize a function. It is commonly used in machine learning and deep learning for optimizing the parameters of a model to minimize a loss function. The basic idea behind GD is to iteratively minimize a function by adjusting its parameters in the direction of the steepest descent (i.e., the negative gradient) of the function with respect to those parameters.

Here is a simplified explanation of how GD works (see Figures 2.22 and 2.23):

1. Start with initial values for the parameters of the function to be optimized.
2. Compute the gradient of the function with respect to the parameters. The gradient points in the direction of the steepest increase of the function.
3. Adjust the parameters in the opposite direction of the gradient to minimize the function. This adjustment is done by subtracting a fraction of the gradient (scaled by a learning rate) from the current parameter values.
4. Continue steps 2 and 3 until a stopping criterion is met, such as reaching a maximum number of iterations or when the improvement in the function value is below a certain threshold.

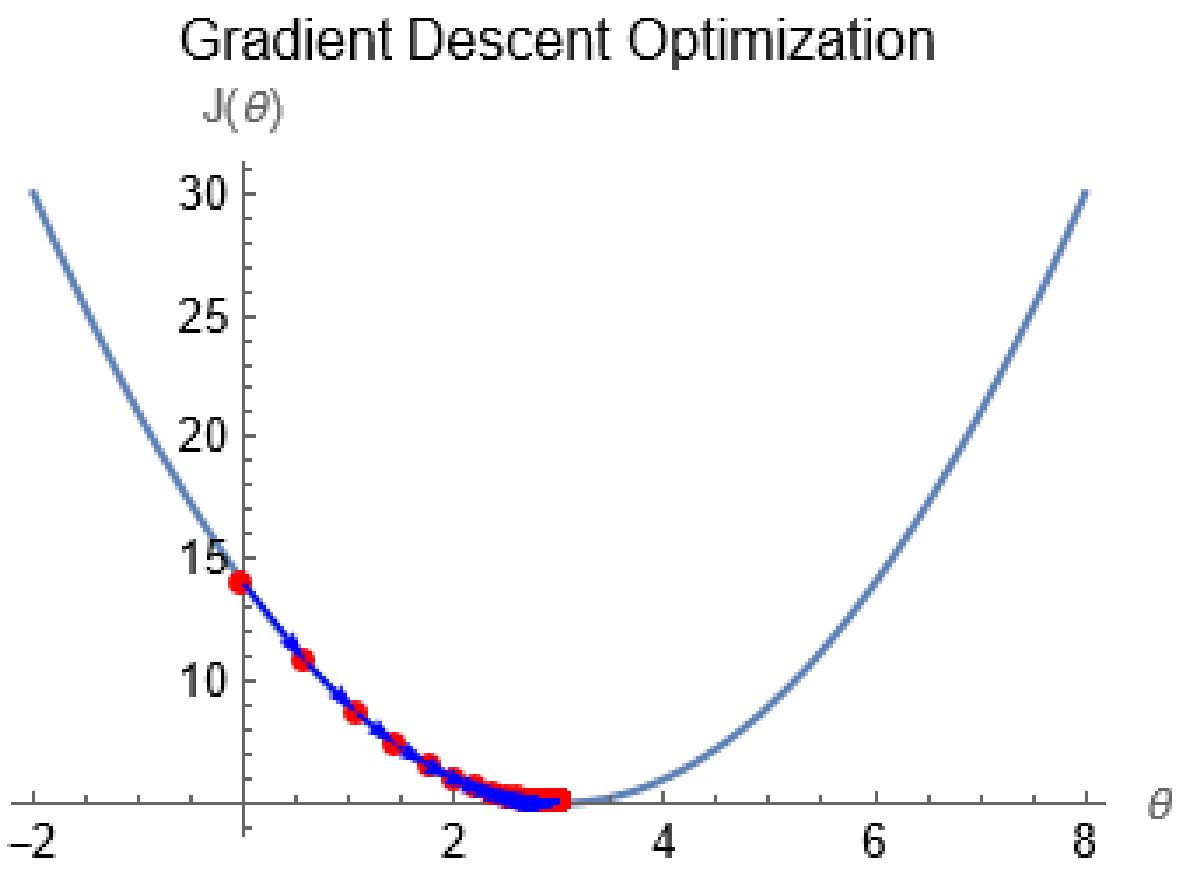

**Figure 2.22.** The figure depicts the GD optimization process applied to a quadratic loss function, $J(\theta) = (\theta - 3)$^2 + 5). The plot shows the loss function $J(\theta)$ over the range $\theta$ from $-2$ to 8. The red points represent the iterative updates of the parameter $\theta$ as it converges towards the optimal value, starting from an initial guess of $\theta_0 = 0$. Blue arrows illustrate the path of optimization, indicating the direction and steps taken by the GD algorithm with a learning rate $\alpha = 0.1$. The optimized parameter $\theta$ is found to be 3, resulting in a minimum loss of 5.

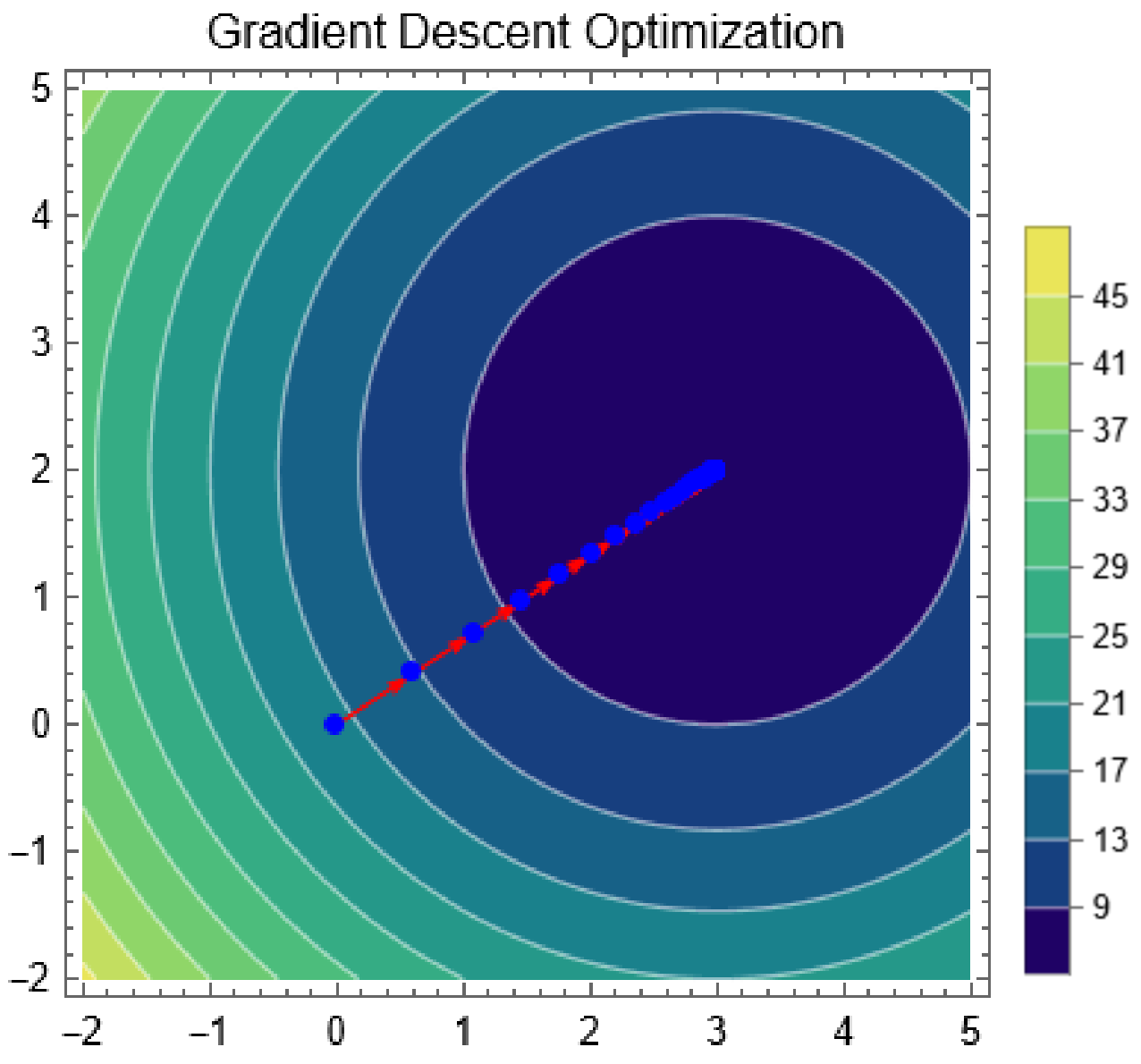

**Figure 2.23.** The figure illustrates the GD optimization process on a contour plot of the loss function $J(x, y) = (x - 3)$^2 + $(y - 2)$^2 + 5). The contour plot shows the level curves of the loss function over the range $x$ and $y$ from $-2$ to 5. The color gradient transitions from blue to green to yellow, representing increasing values of the loss function. The blue points indicate the iterative updates of the parameters $x$ and $y$ as they converge towards the optimal values starting from an initial guess of $(x_0, y_0) = (0,0)$. The red arrows trace the optimization path, showing the direction and steps taken by the GD algorithm with a learning rate $\alpha = 0.1$. The algorithm iteratively minimizes the loss function, moving towards the point $(x, y) = (3,2)$, where the loss function achieves its minimum value of 5.

Suppose we have a loss function $J(\boldsymbol{\theta})$, where $\boldsymbol{\theta}$ represents the parameters of our model. The goal of GD is to find the values of $\boldsymbol{\theta}$ that minimize $J(\boldsymbol{\theta})$. The gradient of $J(\boldsymbol{\theta})$ with respect to $\boldsymbol{\theta}$, denoted by $\nabla J(\boldsymbol{\theta})$, gives us the direction of the steepest ascent of the function.

The update rule for GD is given by:

$$\boldsymbol{\theta} = \boldsymbol{\theta} - \alpha \nabla J(\boldsymbol{\theta}), \tag{2.144}$$

where $\alpha$ is the learning rate, a hyperparameter that controls the size of the steps taken in the parameter space. It is usually a small positive value.

To understand this update rule, let's break it down:

1. We compute the gradient $\nabla J(\boldsymbol{\theta})$, which is a vector containing the partial derivatives of the loss function $J(\boldsymbol{\theta})$ with respect to each parameter $\theta_i$.

$$\nabla J(\boldsymbol{\theta}) = \left(\frac{\partial J(\boldsymbol{\theta})}{\partial \theta_1}, \frac{\partial J(\boldsymbol{\theta})}{\partial \theta_2}, \dots, \frac{\partial J(\boldsymbol{\theta})}{\partial \theta_n}\right). \tag{2.145}$$

2. We update each parameter $\theta_i$ by subtracting a fraction of the gradient $\nabla J(\boldsymbol{\theta})$ from its current value. This fraction is determined by the learning rate $\alpha$.

$$\theta_i = \theta_i - \alpha \frac{\partial J(\boldsymbol{\theta})}{\partial \theta_i}. \tag{2.146}$$

3. This process is repeated iteratively until convergence criteria are met, such as reaching a specified number of iterations or achieving a sufficiently small change in the loss function.

Now, let us discuss a bit about the mathematical intuition behind GD: The negative gradient $-\nabla J(\boldsymbol{\theta})$ points in the direction of the steepest descent of the loss function. By subtracting this gradient from the current parameter values, we move closer to the minimum of the loss function, see Figures 2.22 and 2.23. The learning rate $\alpha$ controls the size of the steps taken in the parameter space. If the learning rate is too small, the convergence may be slow. On the other hand, if it is too large, we might overshoot the minimum and oscillate around it or even diverge.

GD is a cornerstone of optimization in machine learning and is used not only in training NNs but also in many other algorithms and models. However, it has some limitations, such as sensitivity to the choice of learning rate and potential convergence to local minima in non-convex optimization problems. Various techniques, such as learning rate schedules and momentum, are often employed to address these challenges.

# CHAPTER 3

# MULTILAYER FEED-FORWARD NEURAL NETWORKS

In the rapidly evolving landscape of AI and machine learning [49-70], Feed-Forward Neural Networks (FFNNs) stand out as foundational structures, powering a multitude of applications across various domains. This chapter serves as a comprehensive introduction to the inner workings of FFNNs, delving into essential concepts and mechanisms crucial for understanding their functionality and effectiveness. We will start by looking at the structure of a FFNN, followed by how they are trained and used for making predictions. We will also take a brief look at the loss functions that should be used in different settings, the Activation Functions (AFs) used within a neuron, and the different types of optimizers that could be used for training.

The training process is a crucial phase in the development of FFNNs, enabling them to learn from data and improve their performance over time. The training procedure can be broken down into two main components, each playing a distinct role in the network's learning process (forward propagation and back propagation).

We begin with an exploration of forward propagation in NNs, elucidating how input data traverse through the network's layers to produce output predictions. Through a step-by-step examination, readers will grasp the fundamental principles underlying the propagation of information within these intricate systems.

Subsequently, attention shifts towards understanding multilayer networks as computational graphs, offering insights into the structural organization of NNs and their representation as interconnected nodes and edges. This graphical depiction lays the groundwork for comprehending the computational processes that drive NN operations.

A pivotal aspect of NN training is automatic differentiation and its main modes [71-74]. Automatic differentiation illustrates the mechanisms through which gradients are computed efficiently, enabling the network to adapt and optimize its parameters during the learning process. Readers will gain a nuanced understanding of automatic differentiation's role in facilitating gradient-based optimization techniques.

The chapter further explores the training process and loss/cost functions integral to optimizing NNs. From defining objectives through appropriate loss functions to navigating the landscape of optimization algorithms, this section equips readers with the necessary tools to effectively train and fine-tune NN models.

We shift our focus to the backward pass, often referred to as Back Propagation (BP). During this phase, the network evaluates the error or the disparity between its predictions (outputs) and the actual target values. This discrepancy serves as a guide for adjusting the network's weights to minimize the error and enhance its accuracy. BP involves traversing the network in reverse, updating weights based on the calculated error and its gradients. We provide a detailed examination of the four fundamental equations that underpin BP. By dissecting these mathematical foundations, we explain the principles that govern how NNs learn from data, adjust their internal representations, and iteratively refine their predictive capabilities.

Finally, we explore the Universal Approximation Theorem (UAT) [75-99], which asserts that a FFNN with a single hidden layer can approximate any continuous function on a compact subset of $\mathbb{R}^n$. This theorem underscores the remarkable flexibility and potential of NNs in modeling complex, non-linear relationships.

Building on these theoretical foundations, this chapter aims to demystify the complex workings of FFNNs, equipping readers with the knowledge to confidently navigate the intricacies of modern NNs. This chapter is the basis for most of the remaining chapters where NNs will be used in the subsequent discussions.

## 3.1 Feed-Forward Neural Networks

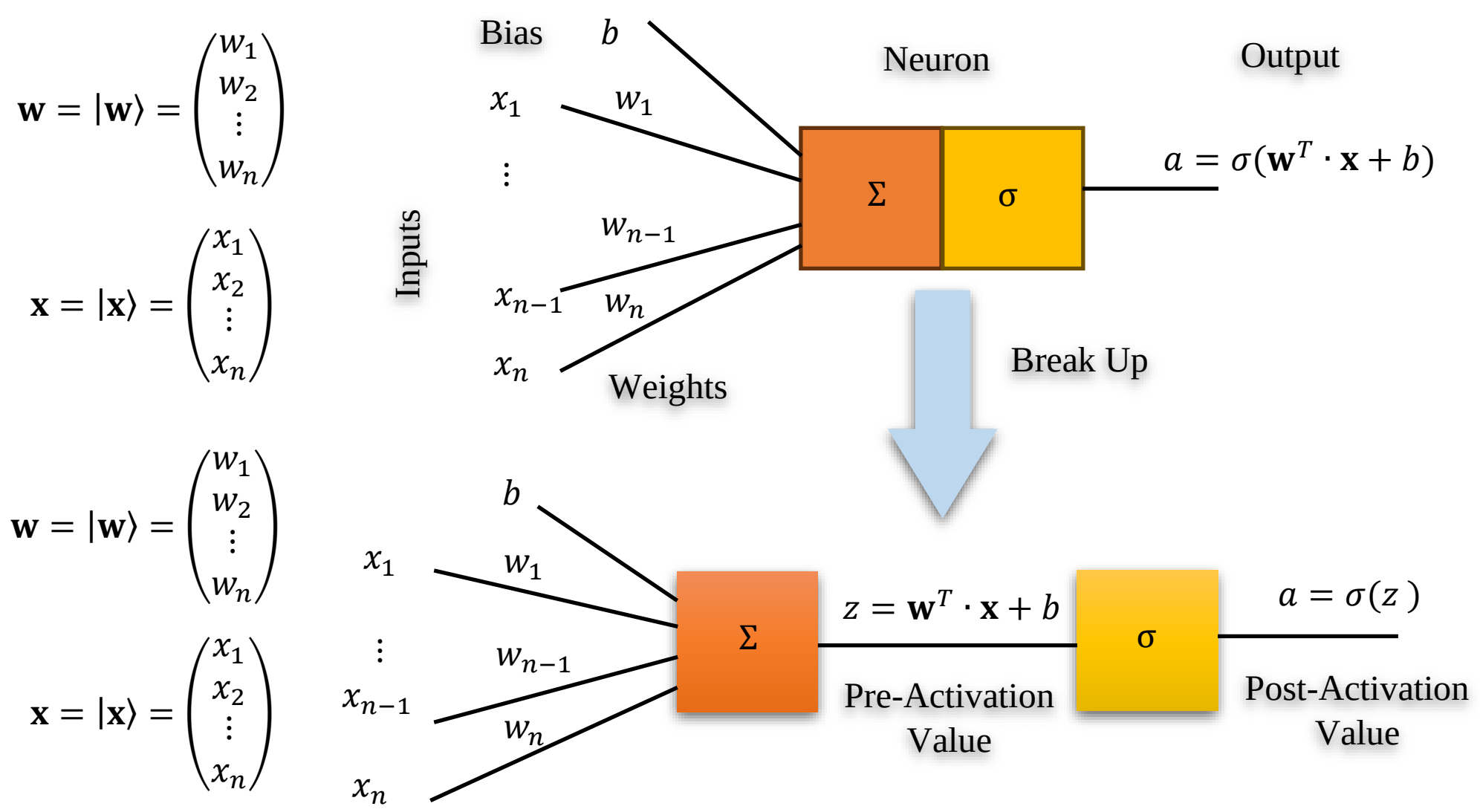

**Figure 3.1.** pre- and post- activation values within a neuron.

### Unit/ Node/Neuron Computations

NNs are computational models inspired by the structure and functioning of the human brain. They consist of interconnected units (also known as nodes or neurons) organized into layers, see Figure 3.1 and Figure 3.2. Each node, or neuron, Figure 3.1, in a NN is a basic processing unit that receives one or more inputs, performs mathematical operations on these inputs, and produces an output. Each connection between nodes has an associated weight, which determines the strength of the connection. Additionally, each node has a bias term. The weights and biases are adjusted during the training process to minimize the difference between the predicted output and the actual output.

One single neuron has an $n$-dimensional input vector and has one single output signal. Mathematically, a neuron is a function that takes as input a vector $\mathbf{x} \in \mathbb{R}^n$, $\mathbf{x} = |\mathbf{x}\rangle = (x_1, x_2, \dots, x_n)^T$ and produces a scalar. A unit is parameterized by a weight vector $\mathbf{w} \in \mathbb{R}^n$, $\mathbf{w} = |\mathbf{w}\rangle = (w_1, w_2, \dots, w_n)^T$, where $w_i$ is the weight associated with input $x_i$, and a bias term denoted by $b$, see Figure 3.1. Neurons apply an AF, σ, to the weighted sum of their inputs. The basic operations of a neuron can be represented mathematically as follows:

$$z = \sum_{i=1}^{n} w_i x_i + b = \mathbf{w}^T \cdot \mathbf{x} + b = \langle \mathbf{w}|\mathbf{x}\rangle + b, \quad (3.1.1)$$

$$a = \sigma(z). \quad (3.1.2)$$

The break-up of the neuron computations into two separate values is shown in Figure 3.1. A neuron really computes two functions within the node, which is why we have incorporated the summation symbol Σ as well as the activation symbol σ within a neuron.

- Before applying the AF, $\sigma$, the node computes a value referred to as the pre-activation value ($z$). This value is calculated as the weighted sum of the inputs ($x_i$) plus the bias ($b$), (3.1.1).
- The pre-activation value is then passed through an AF $\sigma$, denoted as $\sigma(z)$. Common AFs include Sigmoid ($\sigma(z) = 1/(1 + e^{-z})$ ), and hyperbolic tangent ($\tanh(z)$). The result of applying the AF is the post-activation value $a$. This is the final output of the node. The AF allows NNs to model complex relationships in data that may not be captured by a simple linear transformation.

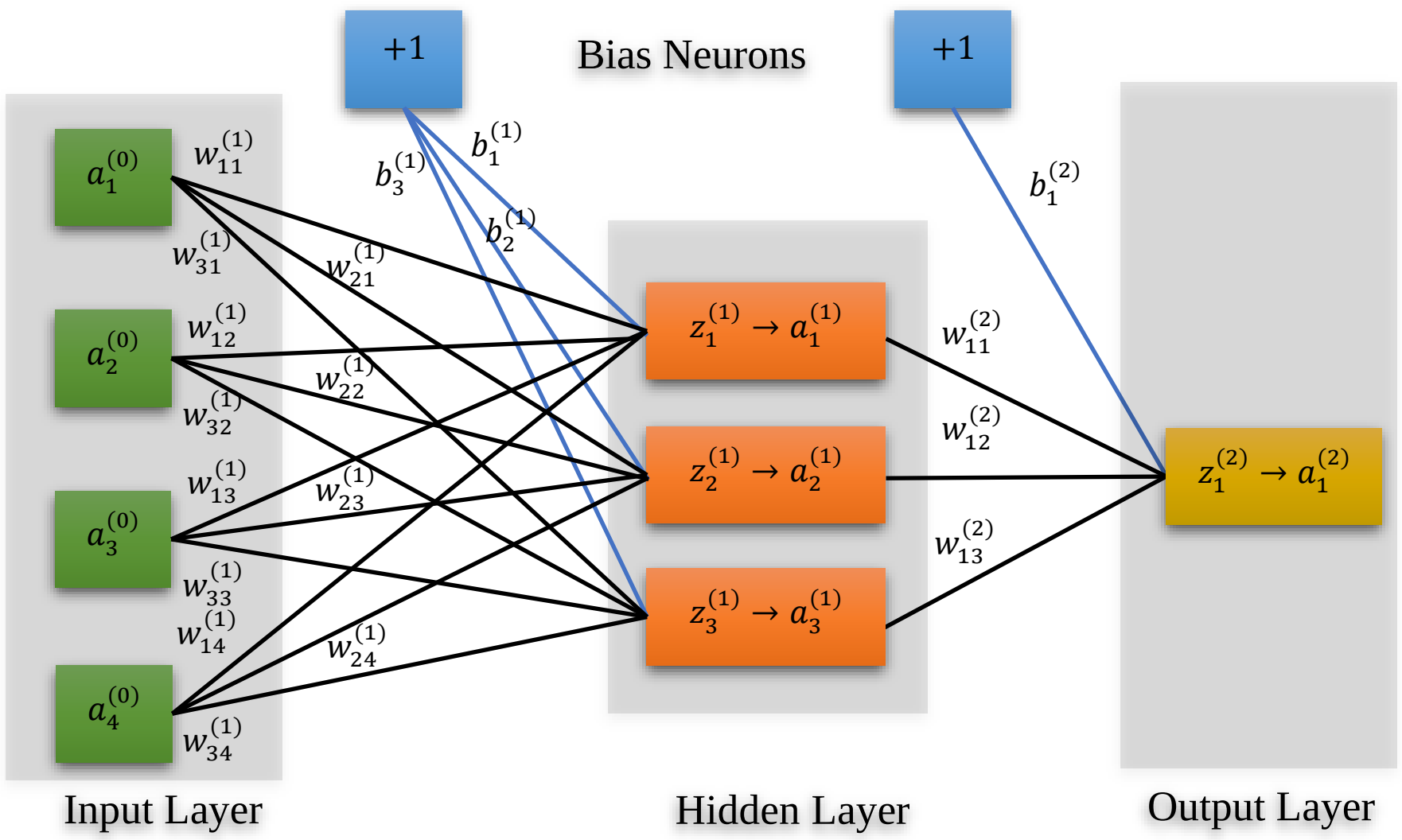

**Figure 3.2.** FFNN having one input layer (4 neurons), only one hidden layer (3 neurons) and one output layer (1 neuron).

An important point that emerges from Figure 3.1 is that one could treat a node with a nonlinear activation as two separate computational nodes, one for the linear transformation $z = \mathbf{w}^T \cdot \mathbf{x} + b$ and the other for the nonlinear transformation $\sigma(z)$. This conceptual separation can simplify analytical results and make it easier to understand and analyze the behavior of NNs in certain scenarios.

### Feed-Forward Neural Networks

In artificial NNs, a layer is a fundamental building block that helps organize and structure the network. Each layer typically consists of a set of neurons or nodes that perform specific operations on the data. The layers work together to process input data and produce output. The basic architecture of a FFNN consists of three main types of layers: the input layer, one or more hidden layers, and the output layer. A FFNN is a type of artificial NN where the information moves in one direction— the input data is fed into the input layer, and it passes through the network layer by layer until it reaches the output layer, producing the final output. It is one of the simplest NNs. There are no cycles or loops in the network, the data flows in a forward direction. Figure 3.2 shows a FFNN having only one hidden layer. A default architecture of FFNNs assumes fully connected, also known as a fully connected layer or dense layer. It is a type of NN architecture where each node in one layer is connected to every node in the next layer. The connections are represented by weights, and each connection has its own weight parameter.

One way of representing a network of neural interconnections is as a Directed Acyclic Graph (DAG). A graph is a set of vertices or nodes (representing basic computing elements) and a set of edges (representing the connections between the nodes), where we assume that both sets are of finite size. In a directed graph (or digraph), the edges are assigned an orientation so that numerical information flows along each edge in a particular direction. An acyclic graph is one in which no loops or feedback are allowed.

Each layer type has its specific purpose and contributes to the overall functionality and performance of the NN:

- Input Layer:
  - The input layer is responsible for receiving the initial raw input data.
  - Each node in the input layer represents a feature or attribute of the input data.
  - The number of nodes in the input layer is determined by the dimensionality of the input data.
- Hidden Layers:
  - Between the input and output layers, there can be one or more hidden layers.
  - These layers perform most of the computation in the NN.

- Each neuron in a hidden layer takes inputs from the previous layer, applies a transformation (usually a weighted sum followed by an AF), and passes the result to the next layer.
  - The number of hidden layers and the number of nodes in each hidden layer are design choices that depend on the complexity of the problem.
- Output Layer:
  - The output layer produces the final results of the NN's computation.
  - The number of nodes in the output layer depends on the nature of the task.

**Remarks:**

- The number of units in a layer is referred to as the width of the layer. The width of each layer need not be the same, but the dimension should be aligned, as we shall see later.
- The number of layers is referred to as the depth of the network. This is where the notion of "deep" (as in "deep learning") comes from.
- The input vector of the NN is the input vector of the neurons in the first layer. In this input vector of the network, there are $(n-1)$ externally applied signals and one bias input signal. The input vector of the neurons in the second layer consists of all outputs of the neurons in the first layer and of one bias input signal. Other layers in the multi-layer network receive their inputs only from all neurons in the previous layer and from one bias signal source.
- The output of the last layer is the output of the network and is the prediction generated based on the input.
- Bias neurons can be used both in the hidden layers and in the output layers.
- Note that there is some confusion in the literature regarding the terminology for counting the number of layers in NNs. Thus, the network in Figure 3.2 may be described as a 3-layer network (which counts the number of layers of units, and treats the inputs as units) or sometimes as a single-hidden-layer network (which counts the number of layers of hidden units). The input layer is often considered as a pass-through layer without adaptive weights in many NN architectures. Its purpose is to provide the initial input to the network. Hence, we recommend a terminology in which Figure 3.2 is called a two-layer network, because it is the number of layers of adaptive weights (computation layers) that is important for determining the network properties.
- At an abstract level, a NN can be thought of as a function $f_w: \mathbf{x} \to \mathbf{y}$, which takes as input $\mathbf{x} \in \mathbb{R}^n$ and produces as output $\mathbf{y} \in \mathbb{R}^m$, and the behavior of which is parameterized by $w$, where $w$ is simply a collection of all the weights and biases for all the units in the network.
- The term "Multi-Layer Perceptron" (MLP) is often used interchangeably with FFNN. Convolutional Neural Networks (CNNs) and Recurrent Neural Networks (RNNs) are special cases of FFNNs.

Therefore, the architecture of a FFNN is defined by specifying the number of layers, the number of nodes in each layer, and the AFs.

### Activation Functions

AFs play a crucial role in artificial NNs by introducing non-linearity, allowing them to model complex relationships between inputs and outputs [50] and [63]. Here are some common AFs used in NNs:

Sigmoid Function:

$$\sigma_{\text{Sigmoid}}(z) = \frac{1}{1+e^{-z}}. \tag{3.2}$$

Hyperbolic Tangent (Tanh):

$$\sigma_{\text{Tanh}}(z) = \frac{e^z - e^{-z}}{e^z + e^{-z}}. \tag{3.3}$$

Rectified Linear Unit (ReLU):

$$\sigma_{\text{ReLU}}(z) = \max(z, 0). \tag{3.4}$$

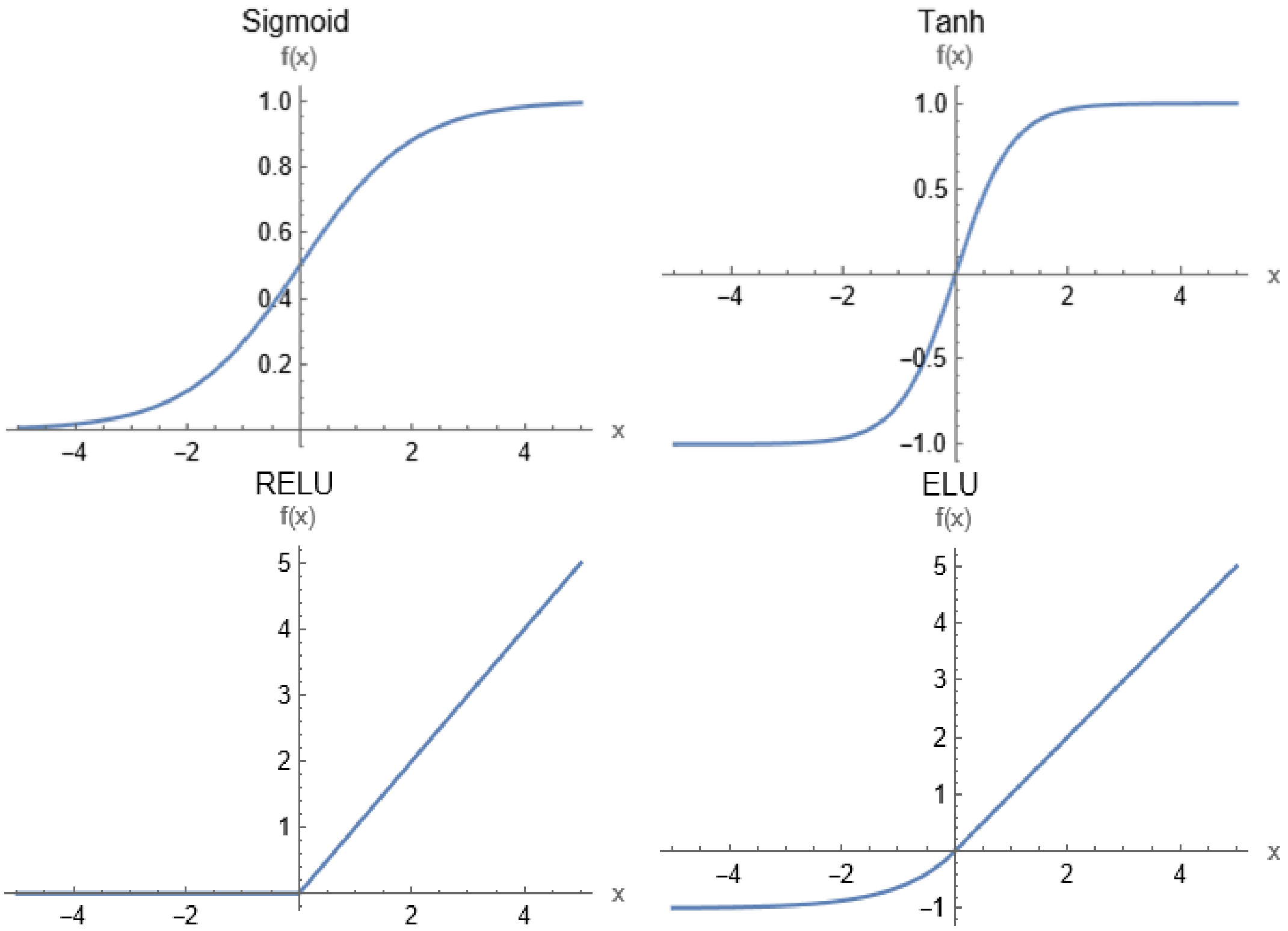

**Figure 3.3.** Comparison of popular AFs: Sigmoid, Tanh, ReLU, and ELU. Each function exhibits distinct characteristics essential for shaping the behavior of neurons within NNs.

Exponential Linear Unit (ELU):

$$\sigma_{\mathrm{ELU}}(z) = \begin{cases} z, & z > 0 \\ \alpha(e^{z} - 1), & z \leq 0 \end{cases}. \tag{3.5}$$

Softmax Function:

$$\sigma_{\mathrm{Softmax}}(z_i) = a_i = \frac{e^{z_i}}{\sum_j e^{z_j}} \quad \text{for } i = 1,2,\ldots,n. \tag{3.6}$$

These functions play pivotal roles in shaping the activation patterns of neurons, influencing the learning dynamics within NNs, see Figure 3.3.

- Most NN architectures maintain uniformity in the AF used within a layer. Throughout the layers of the network, the same AF is typically applied, except for the output layer. This consistency facilitates the training process and allows for a smooth flow of information through the network.
- While this assumption holds for many traditional architectures, it is essential to note that there are variations and advancements in NN architectures, such as skip connections, attention mechanisms, and architectures like transformers, where different parts of the network might use different AFs or have more complex interactions. These advancements aim to improve the learning capabilities of NNs, especially in handling complex tasks and capturing long-range dependencies in data.

The choice of AFs is crucial and is often guided by the characteristics of the data and the underlying assumptions about the distribution of target variables. Different types of problems and data may require different AFs to ensure that the network can effectively learn and represent the patterns in the data. This adaptability in choosing AFs contributes to the flexibility and applicability of NNs across various tasks and domains. The choice of AF at the output layer depends on the nature of the task.

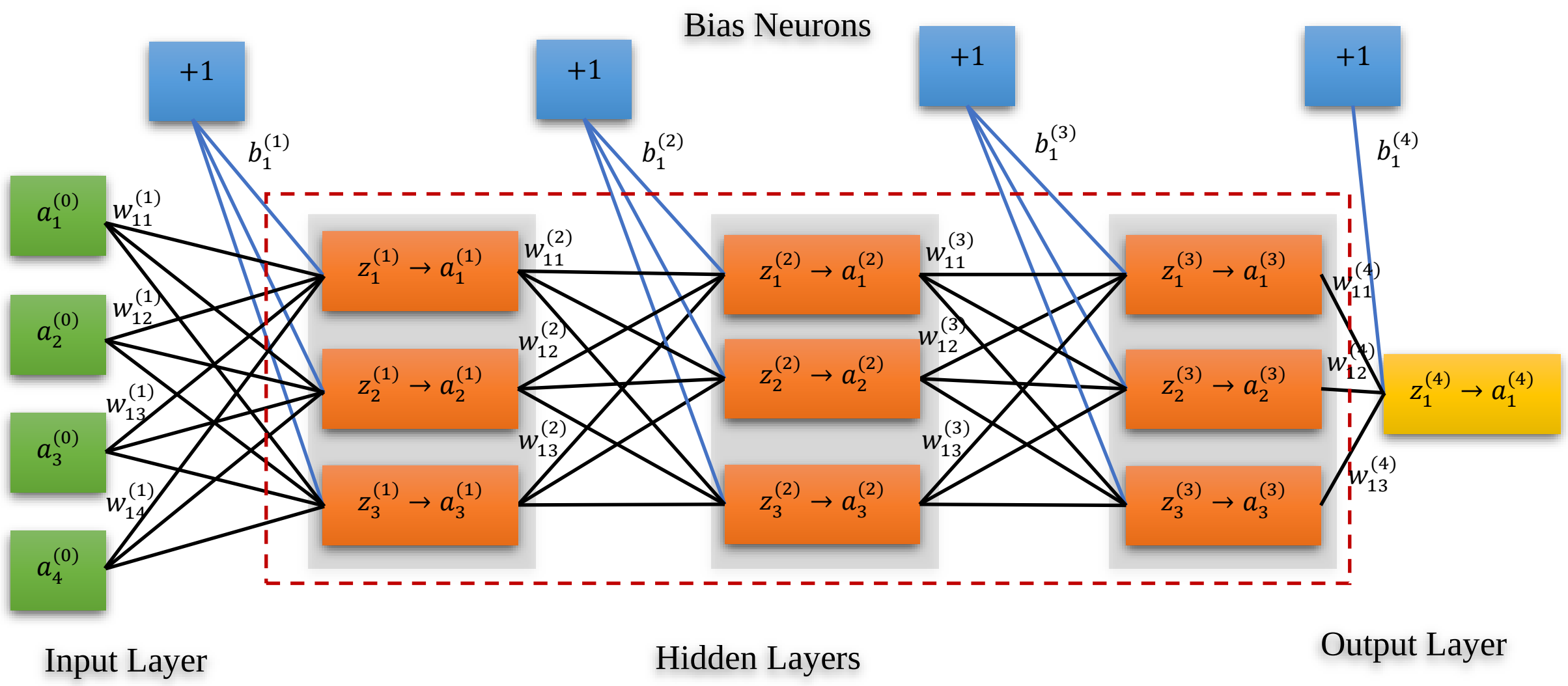

**Figure 3.4.** Scalar notations and architecture (the basic architecture of a FFNN with input layer, three hidden layers and output layer.) Schematic of a simple FFNN with information flowing from the left to right. The inputs $a_i^{(0)}$ are multiplied by weights, added together with the bias, and passed to the function $\sigma$ to get the intermediate values $z_k^{(1)}$. The $z_k^{(1)}$ are then combined with a bias and fed to the next layer.

- In regression problems, where the goal is to predict a continuous numerical value, the identity function is commonly used as the AF for the output layer. This allows the network to directly output the predicted values without any transformation.
- In binary classification problems, where the goal is to classify instances into one of two classes, the Logistic Sigmoid function is often used as the AF for the output layer. This function squashes the output between 0 and 1, representing probabilities for the two classes.
- In multiclass classification problems, where there are more than two classes, the softmax AF is commonly used. It converts the raw output scores into probability distributions over multiple classes, ensuring that the sum of the probabilities across all classes is 1, for more detailed information about the Softmax AF, please refer to Chapter 8.

The choice of AF in hidden layers is discussed in detail in the next chapters.

### Deep Neural Networks

The real power of NNs arises when the network has more than a single hidden layer. NNs with multiple hidden layers are called Deep Neural Network (DNN). For example, a NN with 3 hidden layers is shown in Figure 3.4. This network has 3 hidden units per hidden layer. The term "deep" in DNNs refers to the depth of the network, which is determined by the number of hidden layers. Deeper networks can capture more complex patterns and representations in the data, allowing them to learn hierarchical features. This ability to learn intricate and abstract features makes DNNs powerful for tasks such as image recognition, natural language processing, and more.

The increasing number of connections between nodes in deeper networks also means that there are more weights and biases to be optimized during the training process. While this may pose challenges in terms of computational complexity and training time, advancements in hardware and optimization algorithms have enabled the successful training of DNNs.

Training the FFNN consists of two parts: working out what the outputs are for the given inputs and the current weights, and then updating the weights according to the error, which is a function of the difference between the outputs and the targets. These are generally known as going forwards and backwards through the network.

## 3.2 Forward Propagation in Neural Networks

Forward propagation is the process of passing input data through the NN to calculate an output, and it sets the stage for the subsequent steps in the training process. The term "forward" signifies the direction in which information flows through the network, from the input layer to the output layer. The following is a step-by-step explanation of the forward propagation process:

1. The process begins with the input layer, where the network receives the raw input data.
2. The weighted sum of inputs and biases is calculated for each node in the first hidden layer.
3. This value is then passed through an AF.
4. The process is repeated for each hidden layer in the network. The output from the previous layer serves as the input to the next layer.
5. The final hidden layer's output is passed through output layer to produce the network's final output. The type of AF used in the output layer depends on the nature of the problem.
6. The output is then compared to the true target values using a loss function, which measures the difference between the predicted output and the actual output.

The forward propagation process is integral to both training and making predictions with NNs. During training, the weights and biases are adjusted using optimization algorithms (e.g., GD) to minimize the loss function, enabling the network to learn from the data. Once the network is trained, it can be used to make predictions on new, unseen data by simply performing forward propagation with the learned weights and biases.

Before we delve into the intricacies of the mathematics, let us begin with a useful observation: the layers of a fully connected NN can be thought of as vector functions:

$$\mathbf{a} = \sigma(\mathbf{z}), \tag{3.7}$$

where the input to the layer is $\mathbf{z}$ and the output is $\mathbf{a}$. These are both vectors; each node in a layer produces a single scalar output, which, when grouped, becomes $\mathbf{a}$, a vector representing the output of the layer.

AFs, such as the Sigmoid function, can be applied element-wise to vector arguments. This means that the AF is independently applied to each element of the vector, producing a new vector of the same length where each element is the result of applying the AF to the corresponding element of the input vector. For example, let us say you have a vector $\mathbf{z} \in \mathbb{R}^n$, $\mathbf{z} = |\mathbf{z}\rangle = (z_1, z_2, \dots, z_n)^T$ and you want to apply AF $\sigma$ element-wise to each element of the vector. The resulting vector $\mathbf{a}$ would be:

$$\mathbf{a} = |\mathbf{a}\rangle = \sigma(\mathbf{z}) = \big(\sigma(z_1), \sigma(z_2), \dots, \sigma(z_n)\big)^T. \tag{3.8}$$

This element-wise application is a common operation in NNs, where the AFs are typically applied independently to the elements of the input vectors in each layer.

Hence, one can represent NN architectures with vector variables. In vector-based representations, instead of representing individual units in a layer separately, the entire layer is treated as a vector which is represented as one rectangle. The use of rectangles is a simplification and abstraction to convey the idea of a layer with multiple units without explicitly drawing individual circles or squares for each unit, which can become impractical and cluttered in diagrams for large networks. This can make it more convenient for mathematical operations and expressions. For example, the architectural diagram in Figure 3.4 (with scalar units) has been transformed into a vector-based neural architecture in Figure 3.5.

**Remarks:**

- In traditional NN diagrams, each unit in a layer is represented individually, often as circles or squares for neurons.
- When transitioning to a vector-based representation, the entire layer is treated as a single vector. This means that the activations of all units in the layer are combined into a single mathematical entity.

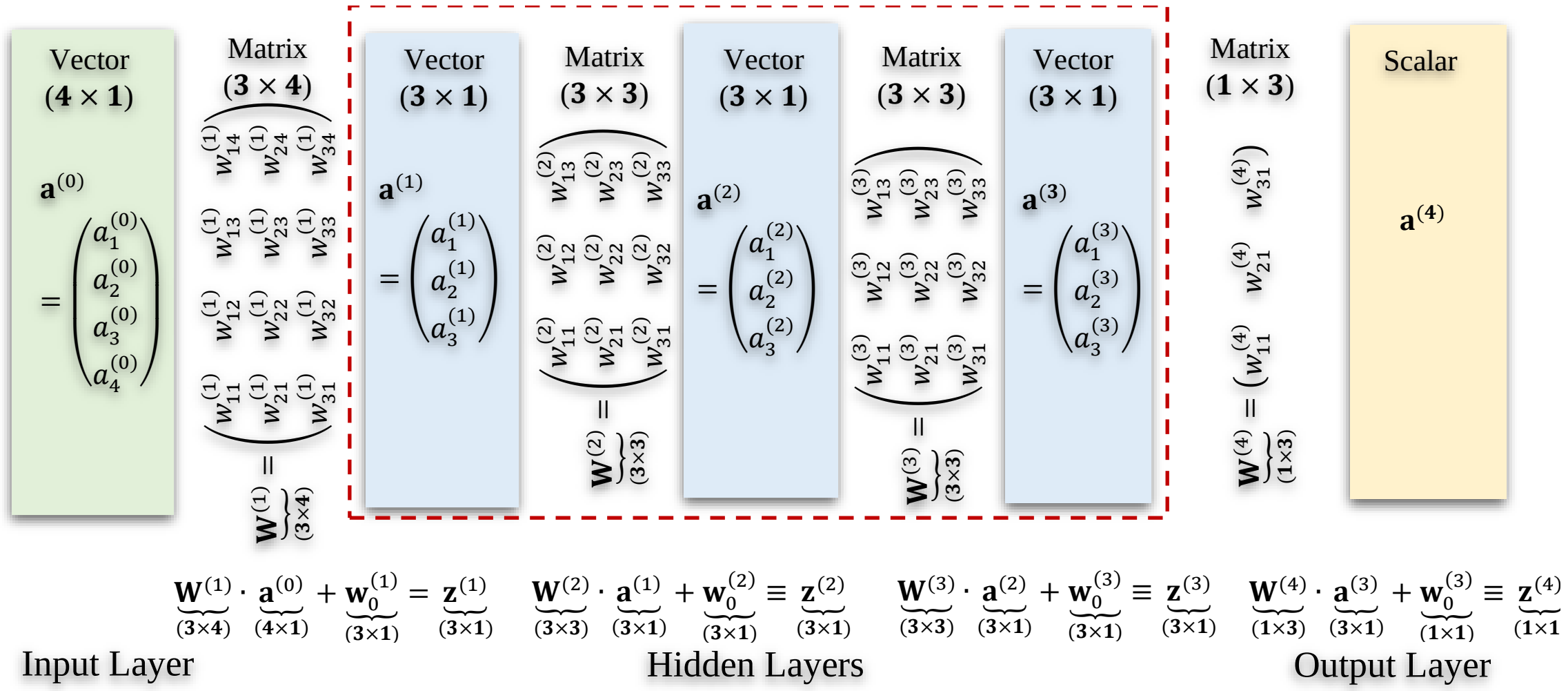

**Figure 3.5.** Vector notations and architecture (the basic architecture of a FFNN with input layer, three hidden layers and output layer.)

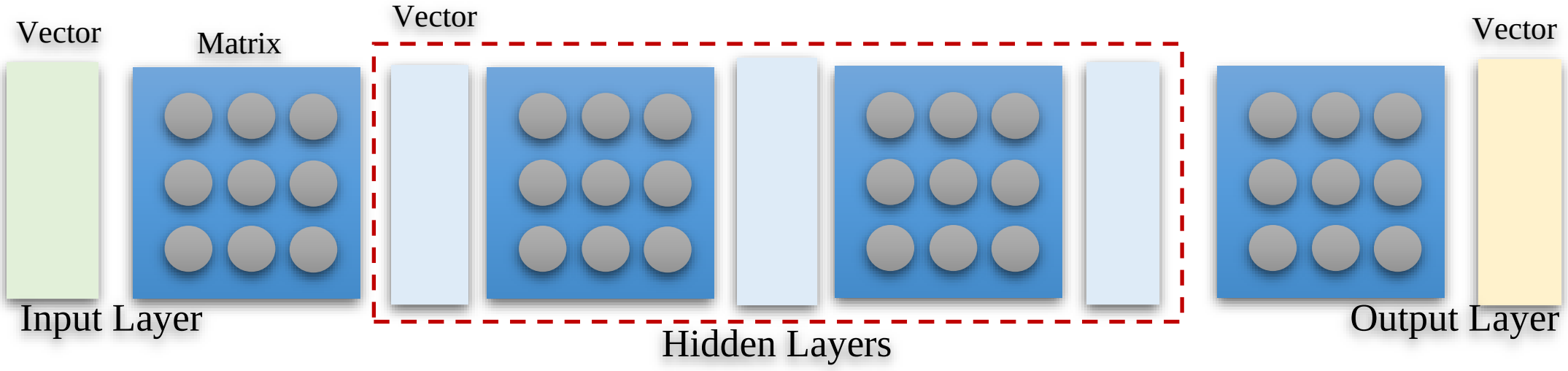

**Figure 3.6.** Vector notations and architecture (the basic architecture of a FFNN with input layer, three hidden layers and output layer.)

- The connections between layers are represented as matrices. Each element of the matrix corresponds to the weight of the connection between a pair of neurons. In other words, when you move from a scalar-based representation to a vector-based representation, you group the units in each layer into vectors. The weights connecting these vectors are then represented as matrices, see Figure 3.6.
- For example, in a FFNN, the connection between the input layer and the first hidden layer can be represented by a weight matrix $\mathbf{W}^{(1)}$.
- When performing the mathematical operation of multiplying a matrix ($\mathbf{W}^{(1)}$) with a vector (input), it is crucial to ensure that the dimensions are compatible for matrix multiplication.
- In the weight matrix $\mathbf{W}^{(1)}$, the connection matrix is shown as a $3 \times 4$ matrix, see Figure 3.5.
- It is important to ensure that the dimensions align correctly for matrix multiplication, following the rule that the number of columns in the first matrix must match the number of rows in the second matrix (vector).
- Representing NN architectures with vector variables and matrices simplifies the mathematical expressions involved in forward and backward passes.
- It also facilitates the use of linear algebra libraries for efficient implementation of NN operations, as matrix multiplication can be optimized.
- In the NNs, the vector-centric operations often includes operations like matrix multiplication and element-wise vector operations. Matrix multiplication is a fundamental vector-centric operation used in the computation of weighted sums in fully connected layers of NNs. Element-wise vector operations include operations like element-wise addition, subtraction, multiplication, and division. Vector-centric operations are often more computationally efficient, especially when parallel processing capabilities are leveraged, as modern hardware (GPUs) is designed to handle such operations efficiently.

- With vector-centric operations, the entire NN can be expressed as a single path which greatly simplifies the topology of the NN graph. However, all functions in the vector-centric view of NNs are vector-to-vector functions. Therefore, one needs to use the vector-to-vector derivatives and a corresponding chain rule. In the vector-centric view, one wants to compute derivatives with respect to entire layers of nodes.

To describe forward propagation in a DNN, we need to define some notation for the weights and biases. The input layer has $n_0$ neurons. In this work, we adopt a systematic notation to enhance clarity and cohesion in our mathematical expressions. To maintain a consistent representation throughout the model, we denote the vector input data at the input layer as $\mathbf{a}^{(0)}$. This choice is made to unify the notation with $\mathbf{a}^{(l)}$, which represents the output of hidden layers $(l)$. By aligning the input layer notation with that of hidden layers, we aim to streamline the understanding of our mathematical formulations. This notation strategy serves to clearly convey the layer index, providing a cohesive framework for both input and hidden layers in our models. We consider the input layer as layer number 0 of length $n_0$. Explicitly, the input values are denoted as $\mathbf{a}^{(0)}$, where $\mathbf{a}^{(0)}$ is an $n_0$-dimensional vector:

$$\mathbf{a}^{(0)} = \left|\mathbf{a}^{(0)}\right\rangle = \begin{pmatrix} a_1^{(0)} \\ a_2^{(0)} \\ \vdots \\ a_{n_0}^{(0)} \end{pmatrix}. \tag{3.9}$$

Each hidden layer $(l)$ has $n_l$ units, and the output layer $(L)$ has $n_L$ units. For each layer $l$, the activation values $\mathbf{a}^{(l)}$ are calculated using the weighted sum of the inputs followed by an AF:

$$\mathbf{z}^{(l)} = \mathbf{W}^{(l)} \cdot \mathbf{a}^{(l-1)} + \mathbf{b}^{(l)}, \quad \text{or} \quad \left|\mathbf{z}^{(l)}\right\rangle = \mathbf{W}^{(l)}\left|\mathbf{a}^{(l-1)}\right\rangle + \left|\mathbf{b}^{(l)}\right\rangle, \tag{3.10}$$

$$\mathbf{a}^{(l)} = \sigma^{(l)}\left(\mathbf{z}^{(l)}\right), \quad \text{or} \quad \left|\mathbf{a}^{(l)}\right\rangle = \sigma^{(l)}\left|\mathbf{z}^{(l)}\right\rangle, \tag{3.11}$$

where: $\mathbf{W}^{(l)}$ is the weight matrix for layer $l$, $\mathbf{b}^{(l)}$ is the bias vector for layer $l$, and $\sigma^{(l)}$ is the AF for layer $l$. The output of the NN is the activation of the last layer: $\mathbf{a}^{(L)}$. The equations above now explicitly include the bias term $\mathbf{b}^{(l)}$ in the calculation of the weighted sum $\mathbf{z}^{(l)}$. Let

$$\mathbf{W}^{(l)} = \begin{pmatrix} w_{11}^{(l)} & w_{12}^{(l)} & \dots & w_{1n_{l-1}}^{(l)} \\ w_{21}^{(l)} & w_{22}^{(l)} & \dots & w_{2n_{l-1}}^{(l)} \\ \vdots & \vdots & \ddots & \vdots \\ w_{n_l1}^{(l)} & w_{n_l2}^{(l)} & \dots & w_{n_ln_{l-1}}^{(l)} \end{pmatrix}, \quad \mathbf{a}^{(l)} = \left|\mathbf{a}^{(l)}\right\rangle = \begin{pmatrix} a_1^{(l)} \\ a_2^{(l)} \\ \vdots \\ a_{n_l}^{(l)} \end{pmatrix}, \text{and} \quad \mathbf{b}^{(l)} = \left|\mathbf{b}^{(l)}\right\rangle = \begin{pmatrix} b_1^{(l)} \\ b_2^{(l)} \\ \vdots \\ b_{n_l}^{(l)} \end{pmatrix}, \tag{3.12}$$

where $w_{ij}^{(l)}$ represents the weight connecting the $j$-th neuron in layer $l-1$ to the $i$-th neuron in layer $l$. The dimension of $\mathbf{W}^{(l)}$ would be $n_l \times n_{l-1}$. When indexing weights, we will typically use $i$ to indicate the destination and $j$ to indicate source - remember weights are indexed (destination, source). We have

$$\begin{aligned} \mathbf{z}^{(l)} &= \mathbf{W}^{(l)} \cdot \mathbf{a}^{(l-1)} + \mathbf{b}^{(l)} \\ &= \mathbf{W}^{(l)}\left|\mathbf{a}^{(l-1)}\right\rangle + \left|\mathbf{b}^{(l)}\right\rangle \\ &= \begin{pmatrix} w_{11}^{(l)} & w_{12}^{(l)} & \dots & w_{1n_{l-1}}^{(l)} \\ w_{21}^{(l)} & w_{22}^{(l)} & \dots & w_{2n_{l-1}}^{(l)} \\ \vdots & \vdots & \ddots & \vdots \\ w_{n_l1}^{(l)} & w_{n_l2}^{(l)} & \dots & w_{n_ln_{l-1}}^{(l)} \end{pmatrix} \begin{pmatrix} a_1^{(l-1)} \\ a_2^{(l-1)} \\ \vdots \\ a_{n_{l-1}}^{(l-1)} \end{pmatrix} + \begin{pmatrix} b_1^{(l)} \\ b_2^{(l)} \\ \vdots \\ b_{n_l}^{(l)} \end{pmatrix} \\ &= \begin{pmatrix} w_{11}^{(l)} a_1^{(l-1)} + \dots + w_{1n_{l-1}}^{(l)} a_{n_{l-1}}^{(l-1)} + b_1^{(l)} \\ w_{21}^{(l)} a_1^{(l-1)} + \dots + w_{2n_{l-1}}^{(l)} a_{n_{l-1}}^{(l-1)} + b_2^{(l)} \\ \vdots \\ w_{n_l1}^{(l)} a_1^{(l-1)} + \dots + w_{n_ln_{l-1}}^{(l)} a_{n_{l-1}}^{(l-1)} + b_{n_l}^{(l)} \end{pmatrix} \end{aligned} \tag{3.13.1}$$

$$\mathbf{z}^{(l)} = \begin{pmatrix} \sum_{j=1}^{n_{l-1}} w_{1j}^{(l)} a_j^{(l-1)} + b_1^{(l)} \\ \sum_{j=1}^{n_{l-1}} w_{2j}^{(l)} a_j^{(l-1)} + b_2^{(l)} \\ \vdots \\ \sum_{j=1}^{n_{l-1}} w_{n_l j}^{(l)} a_j^{(l-1)} + b_{n_l}^{(l)} \end{pmatrix} = \begin{pmatrix} z_1^{(l)} \\ z_2^{(l)} \\ \vdots \\ z_{n_l}^{(l)} \end{pmatrix}, \tag{3.13.2}$$

where

$$z_i^{(l)} = \sum_{j=1}^{n_{l-1}} w_{ij}^{(l)} a_j^{(l-1)} + b_i^{(l)}, \tag{3.13.3}$$

and

$$\mathbf{a}^{(l)} = |\mathbf{a}^{(l)}\rangle = \sigma^{(l)}|\mathbf{z}^{(l)}\rangle = \sigma^{(l)} \begin{pmatrix} z_1^{(l)} \\ z_2^{(l)} \\ \vdots \\ z_{n_l}^{(l)} \end{pmatrix} = \begin{pmatrix} \sigma^{(l)}(z_1^{(l)}) \\ \sigma^{(l)}(z_2^{(l)}) \\ \vdots \\ \sigma^{(l)}(z_{n_l}^{(l)}) \end{pmatrix} = \begin{pmatrix} a_1^{(l)} \\ a_2^{(l)} \\ \vdots \\ a_{n_l}^{(l)} \end{pmatrix}. \tag{3.13.4}$$

The output of the NN is the activation of the last layer: $\mathbf{a}^{(L)}$.

In addition to our systematic notation for vector input data, we introduce a consistent representation for bias terms within our model architecture. Traditionally, bias terms are denoted as $b_i^{(l)}$, where $(i)$ is the index of the neuron in layer $(l)$. However, for the sake of clarity and uniformity in our notation, we adopt the convention of using $(w_{i0}^{(l)})$ to represent the bias term associated with the neuron $(i)$ in layer $(l)$ and using $(\mathbf{w}_0^{(l)})$ to represent the bias vector associated with the neurons in layer $(l)$. This choice aligns with our overarching goal of establishing a cohesive and intuitive notation system. Now, both weights and bias terms are seamlessly integrated into our notation, promoting a clearer understanding of the mathematical expressions within each layer of our model. Hence,

$$\mathbf{b}^{(l)} = \mathbf{w}_0^{(l)} = \begin{pmatrix} w_{10}^{(l)} \\ w_{20}^{(l)} \\ \vdots \\ w_{n_l 0}^{(l)} \end{pmatrix}, \tag{3.14}$$

and

$$\begin{aligned} \mathbf{z}^{(l)} &= \mathbf{W}^{(l)} \cdot \mathbf{a}^{(l-1)} + \mathbf{b}^{(l)} \\ &= \mathbf{W}^{(l)} \cdot \mathbf{a}^{(l-1)} + \mathbf{w}_0^{(l)}. \end{aligned} \tag{3.15}$$

Now that you have an idea of how each neuron is computed in the NN, you might have realized that explicitly writing out the computation on each node in each layer can be a daunting task. We generally do not express NNs in terms of the computation that happens on each node. We instead express them in terms of layers and because each layer has multiple nodes, we can write the equations in terms of vectors and matrices. Let us examine the preceding forward propagation computations in their complete matrix form for the NN illustrated in Figures 3.4 and 3.5. For the sake of simplicity, we will analyze it layer by layer. To simplify the view and to properly understand what is happening, we will now denote $\mathbf{z}^{(1)} = \mathbf{W}^{(1)} \cdot \mathbf{a}^{(0)} + \mathbf{w}_0^{(1)}$ and $\mathbf{a}^{(1)} = \sigma(\mathbf{z}^{(1)})$.

Calculate $\mathbf{z}^{(1)}$ and $\mathbf{a}^{(1)}$ as follows:

$$\mathbf{z}^{(1)} = |\mathbf{z}^{(1)}\rangle = \underbrace{\begin{pmatrix} z_1^{(1)} \\ z_2^{(1)} \\ z_3^{(1)} \end{pmatrix}}_{3\times1} = \underbrace{\begin{pmatrix} w_{11}^{(1)} & w_{12}^{(1)} & w_{13}^{(1)} & w_{14}^{(1)} \\ w_{21}^{(1)} & w_{22}^{(1)} & w_{23}^{(1)} & w_{24}^{(1)} \\ w_{31}^{(1)} & w_{32}^{(1)} & w_{33}^{(1)} & w_{34}^{(1)} \end{pmatrix}}_{3\times4} \underbrace{\begin{pmatrix} a_1^{(0)} \\ a_2^{(0)} \\ a_3^{(0)} \\ a_4^{(0)} \end{pmatrix}}_{4\times1} + \underbrace{\begin{pmatrix} w_{10}^{(1)} \\ w_{20}^{(1)} \\ w_{30}^{(1)} \end{pmatrix}}_{3\times1}, \tag{3.16.1}$$

$$\mathbf{a}^{(1)} = |\mathbf{a}^{(1)}\rangle = \underbrace{\begin{pmatrix} a_1^{(1)} \\ a_2^{(1)} \\ a_3^{(1)} \end{pmatrix}}_{3\times 1} = \sigma(\mathbf{z}^{(1)}) = \underbrace{\begin{pmatrix} \sigma(z_1^{(1)}) \\ \sigma(z_2^{(1)}) \\ \sigma(z_3^{(1)}) \end{pmatrix}}_{3\times 1}. \tag{3.16.2}$$

Calculate $\mathbf{z}^{(2)}$ and $\mathbf{a}^{(2)}$ as follows:

$$\mathbf{z}^{(2)} = |\mathbf{z}^{(2)}\rangle = \underbrace{\begin{pmatrix} z_1^{(2)} \\ z_2^{(2)} \\ z_3^{(2)} \end{pmatrix}}_{3\times 1} = \underbrace{\begin{pmatrix} w_{11}^{(2)} & w_{12}^{(2)} & w_{13}^{(2)} \\ w_{21}^{(2)} & w_{22}^{(2)} & w_{23}^{(2)} \\ w_{31}^{(2)} & w_{32}^{(2)} & w_{33}^{(2)} \end{pmatrix}}_{3\times 3} \underbrace{\begin{pmatrix} a_1^{(1)} \\ a_2^{(1)} \\ a_3^{(1)} \end{pmatrix}}_{3\times 1} + \underbrace{\begin{pmatrix} w_{10}^{(2)} \\ w_{20}^{(2)} \\ w_{30}^{(2)} \end{pmatrix}}_{3\times 1}, \tag{3.17.1}$$

$$\mathbf{a}^{(2)} = |\mathbf{a}^{(2)}\rangle = \underbrace{\begin{pmatrix} a_1^{(2)} \\ a_2^{(1)} \\ a_3^{(2)} \end{pmatrix}}_{3\times 1} = \sigma(\mathbf{z}^{(2)}) = \underbrace{\begin{pmatrix} \sigma(z_1^{(2)}) \\ \sigma(z_2^{(2)}) \\ \sigma(z_3^{(2)}) \end{pmatrix}}_{3\times 1}. \tag{3.17.2}$$

Calculate $\mathbf{z}^{(3)}$ and $\mathbf{a}^{(3)}$ as follows:

$$\mathbf{z}^{(3)} = |\mathbf{z}^{(3)}\rangle = \underbrace{\begin{pmatrix} z_1^{(3)} \\ z_2^{(3)} \\ z_3^{(3)} \end{pmatrix}}_{3\times 1} = \underbrace{\begin{pmatrix} w_{11}^{(3)} & w_{12}^{(3)} & w_{13}^{(3)} \\ w_{21}^{(3)} & w_{22}^{(3)} & w_{23}^{(3)} \\ w_{31}^{(3)} & w_{32}^{(3)} & w_{33}^{(3)} \end{pmatrix}}_{3\times 3} \underbrace{\begin{pmatrix} a_1^{(2)} \\ a_2^{(2)} \\ a_3^{(2)} \end{pmatrix}}_{3\times 1} + \underbrace{\begin{pmatrix} w_{10}^{(3)} \\ w_{20}^{(3)} \\ w_{30}^{(3)} \end{pmatrix}}_{3\times 1}, \tag{3.18.1}$$

$$\mathbf{a}^{(3)} = |\mathbf{a}^{(3)}\rangle = \underbrace{\begin{pmatrix} a_1^{(3)} \\ a_2^{(3)} \\ a_3^{(3)} \end{pmatrix}}_{3\times 1} = \sigma(\mathbf{z}^{(3)}) = \underbrace{\begin{pmatrix} \sigma(z_1^{(3)}) \\ \sigma(z_2^{(3)}) \\ \sigma(z_3^{(3)}) \end{pmatrix}}_{3\times 1}. \tag{3.18.2}$$

Calculate $\mathbf{z}^{(4)}$ and $\mathbf{a}^{(4)}$ as follows:

$$\mathbf{z}^{(4)} = |\mathbf{z}^{(4)}\rangle = \underbrace{\begin{pmatrix} z_1^{(4)} \end{pmatrix}}_{1\times 1} = \underbrace{\begin{pmatrix} w_{11}^{(4)} & w_{21}^{(4)} & w_{31}^{(4)} \end{pmatrix}}_{1\times 3} \underbrace{\begin{pmatrix} a_1^{(3)} \\ a_2^{(3)} \\ a_3^{(3)} \end{pmatrix}}_{3\times 1} + \underbrace{w_{10}^{(4)}}_{1\times 1}, \tag{3.19.1}$$

$$\mathbf{a}^{(4)} = |\mathbf{a}^{(4)}\rangle = \underbrace{\begin{pmatrix} a_1^{(4)} \end{pmatrix}}_{1\times 1} = \sigma(\mathbf{z}^{(4)}) = \underbrace{\sigma(z_1^{(4)})}_{1\times 1}. \tag{3.19.2}$$

Those are all the operations that take place in our FFNN illustrated in Figures 3.4 and 3.5. we have slightly tweaked the preceding notation by putting in brackets and writing $\mathbf{a}^{(4)}$ as a vector, even though it is clearly a scalar. This was only done to keep the flow and to avoid changing the notation.

In linear algebra, when a matrix or vector is multiplied by another matrix, the resulting matrix or vector may have different dimensions. This multiplication operation can be seen as a mapping, where points in one space are transformed to points in another space. In NN, various operations are carried out on input vectors. These operations involve matrix multiplications, activations functions, and possibly biases. The composition of these operations essentially represents a series of transformations on the input data. The input vector is transformed through the layers of the network, involving matrix multiplications and non-linear activations, ultimately mapping the input from one Euclidean space to the output in another Euclidean space. Hence, NNs can be conceptualized as mappings between different Euclidean spaces, where the layers of the network act as transformation functions. Matrix multiplication is a key operation in this process, contributing to the mapping of input data to output predictions.

Using this observation, we can generalize and write the following:

$$\mathcal{N}: \mathbb{R}^{n_0} \to \mathbb{R}^{n_L}, \tag{3.20}$$

where $\mathcal{N}$ represents the NN, which is a function composed of multiple layers. $\mathbb{R}^{n_0}$ denotes the input space, where $n_0$ is the dimensionality of the input layer. This is the space of all possible input vectors. $\mathbb{R}^{n_L}$ represents the output space, where $n_L$ is the dimensionality of the output layer. This is the space of all possible output vectors. The function $\mathcal{N}$ encapsulates the entire architecture of the NN, which consists of an input layer, hidden layers, and an output layer. Each layer involves matrix multiplication with weights, addition of biases, and AFs. Now, we can summarize our NN of Figure 3.4 in the following equations:

$$\mathcal{N}: \mathbb{R}^4 \to \mathbb{R}^1. \tag{3.21.1}$$

$$\mathcal{N}(\mathbf{a}^{(0)}) = \underbrace{\overbrace{\sigma^{(4)}\left(\underbrace{\overbrace{\mathbf{W}^{(4)}}^{1\times3}\underbrace{\overbrace{\left[\sigma^{(3)}\left(\underbrace{\overbrace{\mathbf{W}^{(3)}}^{3\times3}\underbrace{\overbrace{\left[\sigma^{(2)}\left(\underbrace{\overbrace{\mathbf{W}^{(2)}}^{3\times3}\underbrace{\overbrace{\left[\sigma^{(1)}\left(\underbrace{\overbrace{\overbrace{\mathbf{W}^{(1)}}^{3\times4}\overbrace{\mathbf{a}^{(0)}}^{4\times1}+\overbrace{\mathbf{w}_0^{(1)}}^{3\times1}}^{3\times1}}_{\mathbf{z}^{(1)}}\right)\right]}^{3\times1}}_{\mathbf{a}^{(1)}}+\overbrace{\mathbf{w}_0^{(2)}}^{3\times1}}_{\mathbf{z}^{(2)}}\right)\right]}^{3\times1}}_{\mathbf{a}^{(2)}}+\overbrace{\mathbf{w}_0^{(3)}}^{3\times1}}_{\mathbf{z}^{(3)}}\right)\right]}^{3\times1}}_{\mathbf{a}^{(3)}}+\overbrace{\mathbf{w}_0^{(4)}}^{1\times1}}_{\mathbf{z}^{(4)}}\right)}^{1\times1}}_{\mathbf{a}^{(4)}}. \tag{3.21.2}$$

Adding hidden layers to a FFNN creates a DNN. The depth of a NN is determined by the number of hidden layers it has. Each hidden layer in a DNN introduces additional transformations to the input data through a combination of linear operations and non-linear AFs. The ability to perform multiple nested transformations allows DNNs to learn more complex and abstract representations from the input data. This increased flexibility is particularly beneficial for tasks that involve hierarchical and intricate patterns, as the network can capture and model these patterns across multiple layers.

**Remarks:**

- A multilayer NN is a computational model that successively applies parameterized multivariate functions in a nested composition fashion, enabling it to learn and represent intricate patterns in data. The overall function computed from the inputs to the outputs can be controlled very closely by the choice of parameters. Learning occurs by adjusting the parameters (weights and biases) through training, where the network learns to map input data to the desired output.
- In traditional machine learning algorithms, such as linear regression or decision trees, the prediction function can often be expressed in a closed form. These algorithms typically have a clear and interpretable structure, making it easier to understand how each feature contributes to the final prediction. The relationships between features are usually explicit and can be described using mathematical equations.
- On the other hand, the structure of a NN, with its nested composition of many interconnected nodes and non-linear AFs, introduces a high degree of complexity. The network learns to represent hierarchical and intricate patterns in the data, but expressing this learned function in a concise and interpretable form becomes challenging. The weights and biases associated with each connection in the network are adjusted during training, making the overall function highly parameterized and dynamic.
- This lack of a closed-form expression in NNs is often referred to as the "black box" nature of deep learning models. While they can achieve remarkable performance in various tasks, understanding the exact nature of their learned representations may require additional tools, such as visualization techniques, to make the models more transparent and explainable.

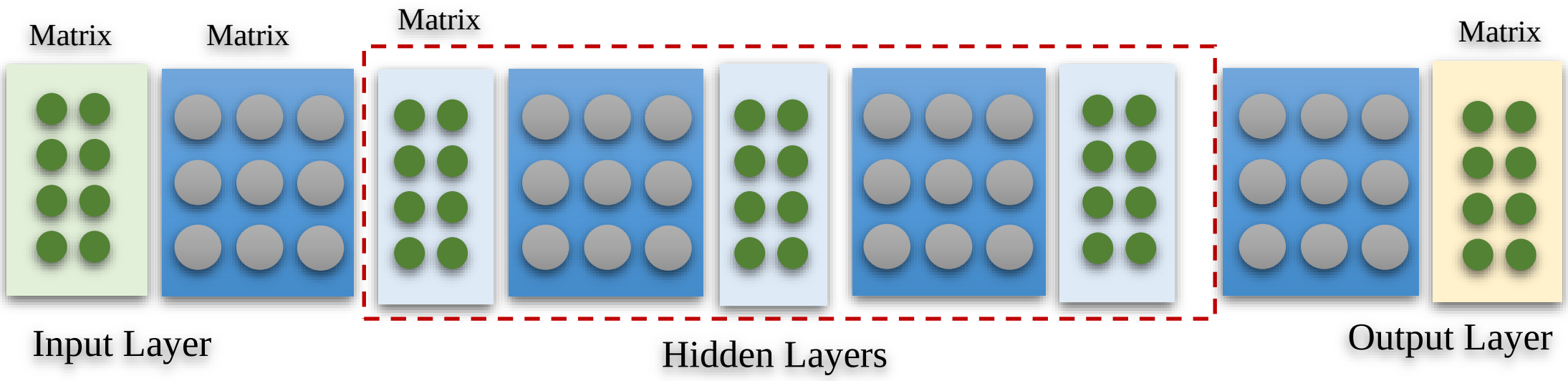

**Figure 3.7.** Matrix notations and architecture (For all training examples, the input vector becomes input matrix.).

### Forward Propagation for All Training Examples

Let $\mathbf{X}$ be the matrix representing the input data for all training examples, where each column corresponds to a different example:

$$\mathbf{X} = \begin{pmatrix} | & | & & | \\ \mathbf{x}^{(1)} & \mathbf{x}^{(2)} & & \mathbf{x}^{(m)} \\ | & | & & | \end{pmatrix}. \tag{3.22}$$

Here, $\mathbf{x}^{(i)}$ represents the input vector for the $i$-th training example, and $m$ is the number of training examples, see Figure 3.7. $\mathbf{X}$ is $(n_0 \times m)$ matrix, where $n_0$ is the number of input features, and $m$ is the number of training examples. The entire forward propagation process can be expressed in vectorized form for all training examples (if you are working with a batch of examples):

$$\mathbf{Z}^{(l)} = \mathbf{W}^{(l)} \cdot \mathbf{A}^{(l-1)} + \mathbf{W}_0^{(l)}, \tag{3.23}$$

$$\mathbf{A}^{(l)} = \sigma\left(\mathbf{Z}^{(l)}\right), \tag{3.24}$$

where: $\mathbf{A}^{(0)} = \mathbf{X}$ is the input data, $\mathbf{A}^{(l)}$ is the $(n_l \times m)$ activation matrix for layer $l$, and $\mathbf{Z}^{(l)}$ is the $(n_l \times m)$ weighted sum matrix for layer $l$. For each layer $l$, $\mathbf{W}^{(l)}$ is $(n_l \times n_{l-1})$ weight matrix. $\mathbf{W}_0^{(l)}$ is $(n_l \times m)$ bias matrix for layer $l$. The final output of the network is given by $\mathbf{A}^{(L)}$, and its dimensions depend on the number of output units: $(n_L \times m)$ activation matrix for the output layer.

### Forward Propagation and Bias Term (for AF $\sigma(1) \neq 1$)

To incorporate the bias term into the weight matrix and the activation vector, we can introduce a dummy input neuron with a fixed value of 1. This way, the bias term becomes a part of the weight matrix, and the activation vector includes the bias term implicitly. Let us denote the augmented weight matrix for layer $l$ as $\widehat{\mathbf{W}}^{(l)}$ and the augmented activation vector as $\hat{\mathbf{a}}^{(l)}$. The equations are as follows. The augmented weight matrix becomes

$$\widehat{\mathbf{W}}^{(l)} = \begin{pmatrix} w_{10}^{(l)} & w_{11}^{(l)} & w_{12}^{(l)} & \cdots & w_{1n_{l-1}}^{(l)} \\ w_{20}^{(l)} & w_{21}^{(l)} & w_{22}^{(l)} & \cdots & w_{2n_{l-1}}^{(l)} \\ \vdots & \vdots & \vdots & \ddots & \vdots \\ w_{n_l 0}^{(l)} & w_{n_l 1}^{(l)} & w_{n_l 2}^{(l)} & \cdots & w_{n_l n_{l-1}}^{(l)} \end{pmatrix}, \tag{3.25}$$

where $w_{i0}^{(l)}$ is the bias term for the $i$-th neuron in layer $l$, and $w_{ij}^{(l)}$ represents the weight connecting the $j$-th neuron in layer $l-1$ to the $i$-th neuron in layer $l$. The augmented activation vector becomes

$$\hat{\mathbf{a}}^{(l)} = \begin{pmatrix} 1 \\ a_1^{(l)} \\ a_2^{(l)} \\ \vdots \\ a_{n_l}^{(l)} \end{pmatrix}. \tag{3.26}$$

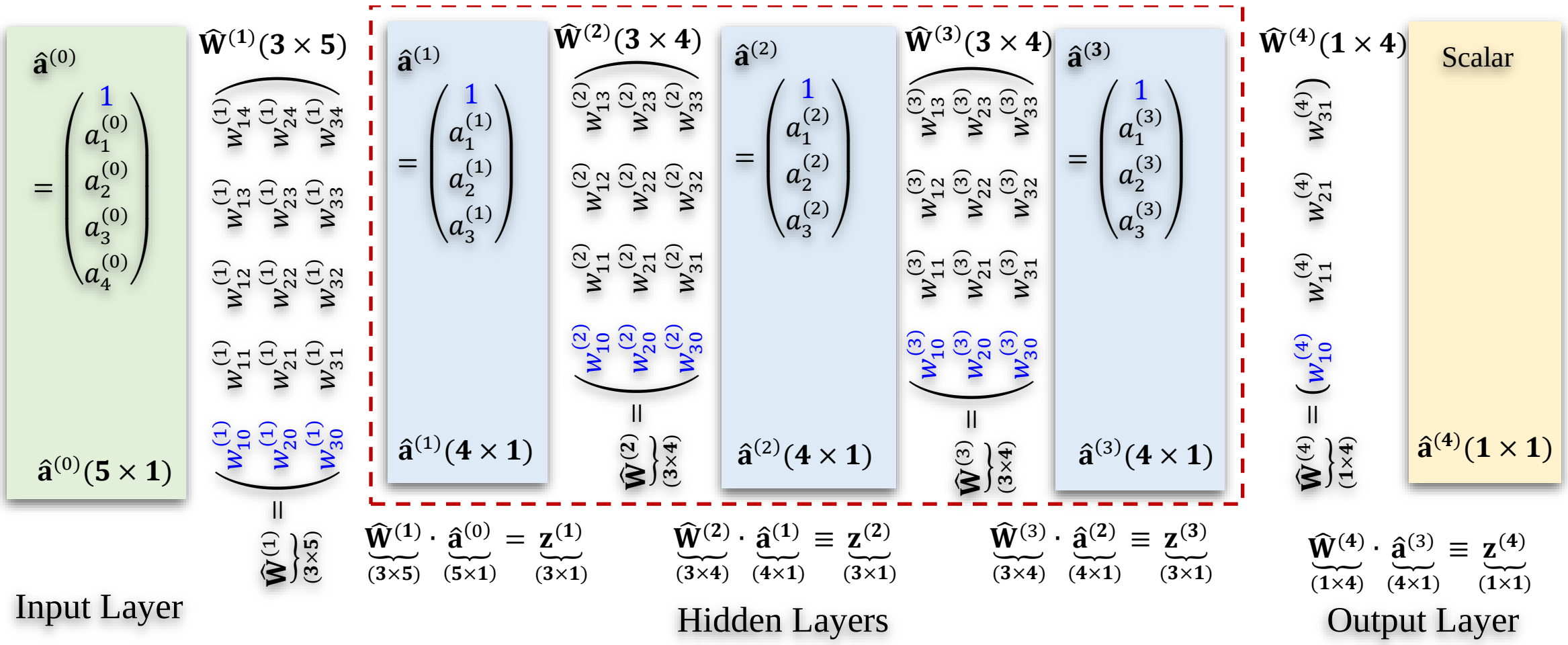

**Figure 3.8.** Vector notations and architecture (the basic architecture of a FFNN with input layer, three hidden layers and output layer.) (without bias term in the case $\sigma(1) \neq 1$ )

Here, the first element is fixed at 1, representing the bias term. With these augmentations, the equations for forward propagation become:

$$\mathbf{z}^{(l)} = \widehat{\mathbf{W}}^{(l)} \cdot \hat{\mathbf{a}}^{(l-1)}, \tag{3.27}$$

$$\mathbf{a}^{(l)} = \sigma\left(\mathbf{z}^{(l)}\right). \tag{3.28}$$

This way, the bias term is treated as just another weight in the weight matrix, and the activation vector includes the bias implicitly, see Figure 3.8. Now, it is clear, why we use the notation $w_{i0}^{(l)}$ with bias term. This formulation simplifies the implementation and helps maintain consistency in the mathematical expressions.

Moreover, let us express the entire forward propagation process in vectorized form for all training examples. Introduce a row of ones to represent the bias term in the input layer:

$$\widehat{\mathbf{X}} = \begin{pmatrix} 1 & 1 & \dots & 1 \\ \mathbf{x}^{(1)} & \mathbf{x}^{(2)} & \dots & \mathbf{x}^{(m)} \end{pmatrix}. \tag{3.29}$$

Now, $\widehat{\mathbf{X}}$ includes the bias term. $\widehat{\mathbf{X}}$ is $(n_0 + 1) \times m$ matrix (the input layer with an additional row for the bias term).

Simillary,

$$\widehat{\mathbf{A}}^{(l)} = \begin{pmatrix} 1 & 1 & \dots & 1 \\ \mathbf{a}_1^{(l)} & \mathbf{a}_2^{(l)} & \dots & \mathbf{a}_m^{(l)} \end{pmatrix}. \tag{3.30}$$

Here, $\mathbf{a}_i^{(l)}$ represents the activation vector of layer $(l)$ for the $i$-th training example, and $m$ is the number of training examples. For each layer $l$, the weighted sum $\mathbf{Z}^{(l)}$ and activation $\mathbf{A}^{(l)}$ can be calculated as follows:

$$\mathbf{Z}^{(l)} = \widehat{\mathbf{W}}^{(l)} \cdot \widehat{\mathbf{A}}^{(l-1)}, \tag{3.31}$$

$$\mathbf{A}^{(l)} = \sigma\left(\mathbf{Z}^{(l)}\right). \tag{3.32}$$

$\widehat{\mathbf{W}}^{(l)}$ is $n_l \times (n_{l-1} + 1)$ weight matrix for layer $l$, including bias weights. $\mathbf{Z}^{(l)}$ is $n_l \times m$ weighted sum matrix for layer $l$. $\widehat{\mathbf{A}}^{(l)}$ is $(n_l + 1) \times m$ activation matrix for layer $l$, including the bias term. The final output of the network is given by $\widehat{\mathbf{A}}^{(L)}$, and its dimensions depend on the number of output units: $\widehat{\mathbf{A}}^{(L)}$ is $(n_L + 1) \times m$ activation matrix for the output layer.

### Forward Propagation and Bias Term (for AF $\sigma(1) = 1$)

For example if we use ReLU AF, $\sigma_{\text{ReLU}}(1) = \max(1,0) = 1$. In this case, let us denote the augmented weight matrix for layer $l$ as $\overline{\mathbf{W}}^{(l)}$ and the augmented activation vector as $\bar{\mathbf{a}}^{(l)}$. The equations are as follows. The augmented weight matrix becomes

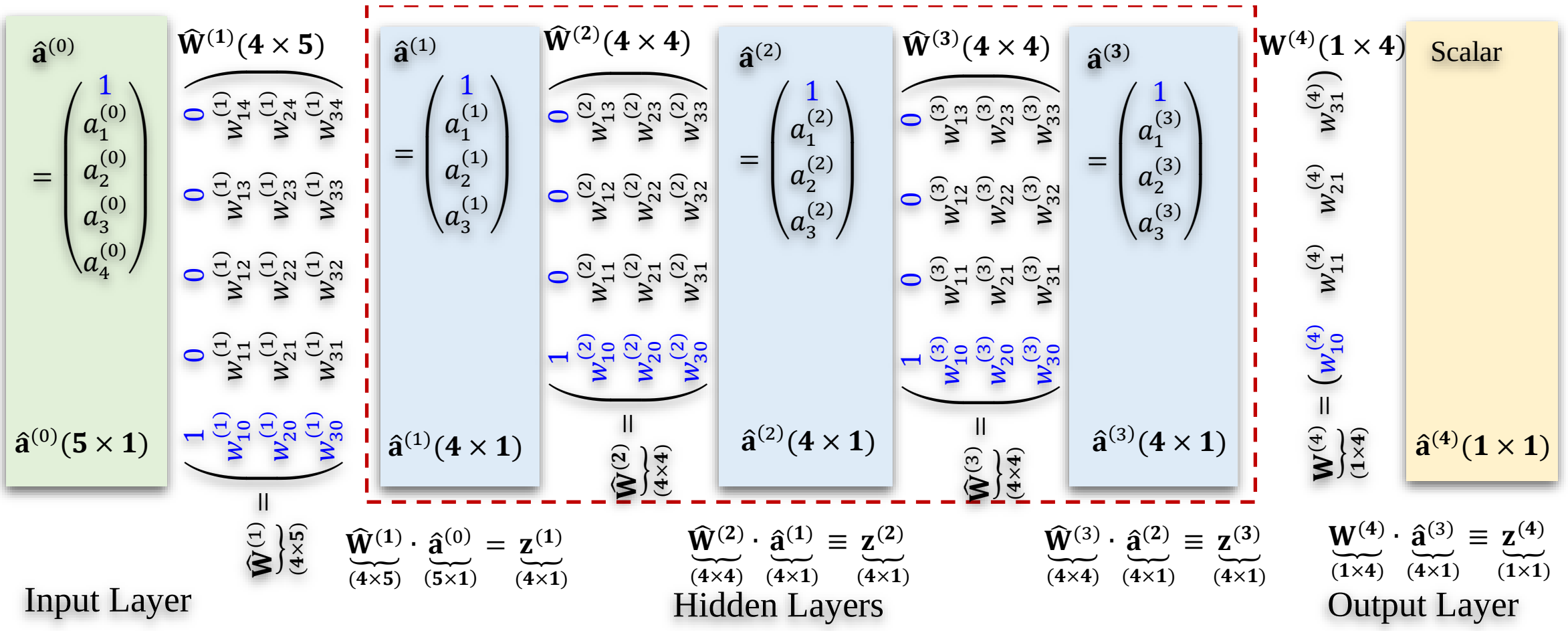

**Figure 3.9.** Vector notations and architecture (the basic architecture of a FFNN with input layer, three hidden layers and output layer.) (without bias term in the case $\sigma(1) = 1$ )

$$\overline{\mathbf{W}}^{(l)} = \begin{pmatrix} 1 & 0 & 0 & \dots & 0 \\ w_{10}^{(l)} & w_{11}^{(l)} & w_{12}^{(l)} & \dots & w_{1n_{l-1}}^{(l)} \\ w_{20}^{(l)} & w_{21}^{(l)} & w_{22}^{(l)} & \dots & w_{2n_{l-1}}^{(l)} \\ \vdots & \vdots & \vdots & \ddots & \vdots \\ w_{n_l0}^{(l)} & w_{n_l1}^{(l)} & w_{n_l2}^{(l)} & \dots & w_{n_ln_{l-1}}^{(l)} \end{pmatrix} = \begin{pmatrix} 1 & \mathbf{0}^T \\ \mathbf{w}_0^{(l)} & \mathbf{W}^{(l)} \end{pmatrix}, \tag{3.33}$$

where $w_{i0}^{(l)}$ is the bias term for the $i$-th neuron in layer $l$, $w_{ij}^{(l)}$ represents the weight connecting the $j$-th neuron in layer $l-1$ to the $i$-th neuron in layer $l$, $\mathbf{0}^T$ is the transpose of zero vector with dimension $1 \times n_{l-1}$, $\mathbf{w}_0^{(l)}$ represent the bias vector associated with the neurons in layer $l$ and $\mathbf{W}^{(l)}$ is the weight matrix for layer $l$ with $n_l \times n_{l-1}$ dimension. The augmented activation vector becomes

$$\hat{\mathbf{a}}^{(l)} = \begin{pmatrix} 1 \\ a_1^{(l)} \\ a_2^{(l)} \\ \vdots \\ a_{n_l}^{(l)} \end{pmatrix} = \begin{pmatrix} 1 \\ \mathbf{a}^{(l)} \end{pmatrix}. \tag{3.34}$$

Here, the first element is fixed at 1, representing the bias term. For each layer of the neural net, the top entry of activation vector for layer $l$ is now fixed at 1. After multiplying that activation vector by $\overline{\mathbf{W}}^{(l)}$, the top entry of activation vector on the next layer is still 1, because $\sigma(1) = \sigma_{\text{ReLU}}(1) = 1$. With these augmentations, the equations for forward propagation become:

$$\hat{\mathbf{z}}^{(l)} = \overline{\mathbf{W}}^{(l)} \cdot \hat{\mathbf{a}}^{(l-1)}, \tag{3.35}$$

$$\hat{\mathbf{a}}^{(l)} = \sigma\left(\hat{\mathbf{z}}^{(l)}\right), \tag{3.36}$$

and final output of the network is given by

$$\hat{\mathbf{z}}^{(L)} = \widehat{\mathbf{W}}^{(L)} \cdot \hat{\mathbf{a}}^{(L-1)}, \tag{3.37}$$

$$\hat{\mathbf{a}}^{(L)} = \sigma\left(\hat{\mathbf{z}}^{(L)}\right), \tag{3.38}$$

where $\overline{\mathbf{W}}^{(l)}$ is $(n_l + 1) \times (n_{l-1} + 1)$ weight matrix for layer $l$, including bias weights. $\hat{\mathbf{z}}^{(l)}$ is $(n_l + 1) \times 1$ weighted sum vector for layer $l$. $\hat{\mathbf{a}}^{(l)}$ is $(n_l + 1) \times 1$ activation vector for layer $l$, including the bias term. $\widehat{\mathbf{W}}^{(L)}$ is the $n_L \times (n_{L-1} + 1)$ weight matrix for layer $L$ i.e., without the row $(1 \quad 0 \quad \dots \quad 0)$, because there is no existence of next layer. This way, the bias term is treated as just another weight in the weight matrix, and the activation vector includes the bias implicitly, see Figure 3.9. Finally, let us express the entire forward propagation process in vectorized form for all training examples. Introduce a row of ones to represent the bias term:

$$\widehat{\mathbf{A}}^{(l)} = \begin{pmatrix} 1 & 1 & \dots & 1 \\ \mathbf{a}_1^{(l)} & \mathbf{a}_2^{(l)} & \dots & \mathbf{a}_m^{(l)} \end{pmatrix}. \tag{3.39}$$

Now, $\widehat{\mathbf{A}}^{(l)}$ includes the bias term. $\widehat{\mathbf{A}}^{(l)}$ is $(n_l + 1) \times m$ matrix (the layer with an additional row for the bias term). Here, $\mathbf{a}_i^{(l)}$ represents the activation vector of layer $(l)$ for the $i$-th training example, and $m$ is the number of training examples. For each layer $l$, the weighted sum $\widehat{\mathbf{Z}}^{(l)}$ and activation $\widehat{\mathbf{A}}^{(l)}$ can be calculated as follows:

$$\widehat{\mathbf{Z}}^{(l)} = \overline{\mathbf{W}}^{(l)} \cdot \widehat{\mathbf{A}}^{(l-1)}, \qquad \widehat{\mathbf{A}}^{(l)} = \sigma\big(\widehat{\mathbf{Z}}^{(l)}\big), \qquad \widehat{\mathbf{Z}}^{(L)} = \widehat{\mathbf{W}}^{(L)} \cdot \widehat{\mathbf{A}}^{(L-1)}, \qquad \widehat{\mathbf{A}}^{(L)} = \sigma\big(\widehat{\mathbf{Z}}^{(L)}\big). \tag{3.40}$$

$\overline{\mathbf{W}}^{(l)}$ is $(n_l + 1) \times (n_{l-1} + 1)$ weight matrix for layer $l$, including bias weights. $\widehat{\mathbf{Z}}^{(l)}$ is $(n_l + 1) \times m$ weighted sum matrix for layer $l$. $\widehat{\mathbf{A}}^{(l)}$ is $(n_l + 1) \times m$ activation matrix for layer $l$, including the bias term.

Now, we can summarize our NN of Figure 3.9 in the following equations:

$$\mathcal{N}: \mathbb{R}^4 \to \mathbb{R}^1, \tag{3.41.1}$$

$$\mathcal{N}\big(\mathbf{a}^{(0)}\big) = \underbrace{\overbrace{\sigma^{(4)}\Bigg(\underbrace{\overbrace{\overbrace{\mathbf{W}^{(4)}}^{1\times4} \cdot \underbrace{\overbrace{\Bigg[\sigma^{(3)}\Bigg(\underbrace{\overbrace{\overbrace{\mathbf{W}^{(3)}}^{4\times4} \cdot \underbrace{\overbrace{\Bigg[\sigma^{(2)}\Bigg(\underbrace{\overbrace{\mathbf{W}^{(2)}}^{4\times4} \cdot \underbrace{\overbrace{\Bigg[\sigma^{(1)}\Bigg(\underbrace{\overbrace{\mathbf{W}^{(1)}}^{4\times5} \cdot \overbrace{\mathbf{a}^{(0)}}^{5\times1}}_{\mathbf{z}^{(1)}}\Bigg)\Bigg]}^{4\times1}}_{\mathbf{a}^{(1)}}}_{\mathbf{z}^{(2)}}\Bigg)\Bigg]}^{4\times1}}_{\mathbf{a}^{(2)}}}^{4\times1}}_{\mathbf{z}^{(3)}}\Bigg)\Bigg]}^{4\times1}}_{\mathbf{a}^{(3)}}}^{1\times1}}_{\mathbf{z}^{(4)}}\Bigg)}^{1\times1}}_{\mathbf{a}^{(4)}}. \tag{3.41.2}$$

## 3.3 Multilayer Network as a Computational Graph

As you increase the depth and width of a NN, the composition of functions becomes more complex, leading to an exponentially growing expression for the overall function. This is due to the fact that each additional node or edge in the NN introduces new terms and dependencies, leading to an exponential increase in the mathematical representation. Writing out this function in closed form quickly becomes impractical, and even if you could write it down, the resulting expression would be incredibly long and difficult to manage. For example, the global function evaluated by the NN of Figure 3.4 is as follows:

$$\begin{aligned}
&\mathcal{N}\big(\mathbf{a}^{(0)}\big) \\
&= \sigma^{(4)}\Big(\mathbf{W}^{(4)} \cdot \Big[\sigma^{(3)}\Big(\mathbf{W}^{(3)} \cdot \Big[\sigma^{(2)}\Big(\mathbf{W}^{(2)} \cdot \Big[\sigma^{(1)}\Big(\mathbf{W}^{(1)} \cdot \mathbf{a}^{(0)} + \mathbf{w}_0^{(1)}\Big)\Big] + \mathbf{w}_0^{(2)}\Big)\Big] + \mathbf{w}_0^{(3)}\Big)\Big] + \mathbf{w}_0^{(4)}\Big) \\
&= \sigma^{(4)}\left(\underbrace{\begin{pmatrix} w_{11}^{(4)} & w_{21}^{(4)} & w_{31}^{(4)} \end{pmatrix}}_{1\times3} \cdot \left[\sigma^{(3)}\left(\underbrace{\begin{pmatrix} w_{11}^{(3)} & w_{12}^{(3)} & w_{13}^{(3)} \\ w_{21}^{(3)} & w_{22}^{(3)} & w_{23}^{(3)} \\ w_{31}^{(3)} & w_{32}^{(3)} & w_{33}^{(3)} \end{pmatrix}}_{3\times3} \cdot \left[\sigma^{(2)}\left(\underbrace{\begin{pmatrix} w_{11}^{(2)} & w_{12}^{(2)} & w_{13}^{(2)} \\ w_{21}^{(2)} & w_{22}^{(2)} & w_{23}^{(2)} \\ w_{31}^{(2)} & w_{32}^{(2)} & w_{33}^{(2)} \end{pmatrix}}_{3\times3}\right.\right.\right.\right. \\
&\quad \left.\left.\left.\left. \cdot \left[\sigma^{(1)}\left(\underbrace{\begin{pmatrix} w_{11}^{(1)} & w_{12}^{(1)} & w_{13}^{(1)} & w_{14}^{(1)} \\ w_{21}^{(1)} & w_{22}^{(1)} & w_{23}^{(1)} & w_{24}^{(1)} \\ w_{31}^{(1)} & w_{32}^{(1)} & w_{33}^{(1)} & w_{34}^{(1)} \end{pmatrix}}_{3\times4} \underbrace{\begin{pmatrix} a_1^{(0)} \\ a_2^{(0)} \\ a_3^{(0)} \\ a_4^{(0)} \end{pmatrix}}_{4\times1} + \underbrace{\begin{pmatrix} w_{10}^{(1)} \\ w_{20}^{(1)} \\ w_{30}^{(1)} \end{pmatrix}}_{3\times1}\right)\right] + \underbrace{\begin{pmatrix} w_{10}^{(2)} \\ w_{20}^{(2)} \\ w_{30}^{(2)} \end{pmatrix}}_{3\times1}\right)\right] + \underbrace{\begin{pmatrix} w_{10}^{(3)} \\ w_{20}^{(3)} \\ w_{30}^{(3)} \end{pmatrix}}_{3\times1}\right)\right] + \underbrace{w_{10}^{(4)}}_{1\times1}\right) \\
&= \mathbf{a}^{(4)},
\end{aligned} \tag{3.42.1}$$

where, for example $\sigma^{(1)}(z)=\sigma^{(2)}(z)=\sigma^{(3)}(z)=\sigma^{(4)}(z)=1/(1+\exp(-z))$ are Sigmoid functions. Since, the AFs must be applied element-wise to vector arguments, the inner term of (3.42.1) becomes

$$\sigma^{(1)}\left(\underbrace{\begin{pmatrix} w_{11}^{(1)} & w_{12}^{(1)} & w_{13}^{(1)} & w_{14}^{(1)} \\ w_{21}^{(1)} & w_{22}^{(1)} & w_{23}^{(1)} & w_{24}^{(1)} \\ w_{31}^{(1)} & w_{32}^{(1)} & w_{33}^{(1)} & w_{34}^{(1)} \end{pmatrix}}_{3\times 4}\underbrace{\begin{pmatrix} a_1^{(0)} \\ a_2^{(0)} \\ a_3^{(0)} \\ a_4^{(0)} \end{pmatrix}}_{4\times 1}+\underbrace{\begin{pmatrix} w_{10}^{(1)} \\ w_{20}^{(1)} \\ w_{30}^{(1)} \end{pmatrix}}_{3\times 1}\right)$$
$$=\begin{pmatrix} \sigma^{(1)}\left(w_{11}^{(1)}a_1^{(0)}+w_{12}^{(1)}a_2^{(0)}+w_{13}^{(1)}a_3^{(0)}+w_{14}^{(1)}a_4^{(0)}+w_{10}^{(1)}\right) \\ \sigma^{(1)}\left(w_{21}^{(1)}a_1^{(0)}+w_{22}^{(1)}a_2^{(0)}+w_{23}^{(1)}a_3^{(0)}+w_{24}^{(1)}a_4^{(0)}+w_{20}^{(1)}\right) \\ \sigma^{(1)}\left(w_{31}^{(1)}a_1^{(0)}+w_{32}^{(1)}a_2^{(0)}+w_{33}^{(1)}a_3^{(0)}+w_{34}^{(1)}a_4^{(0)}+w_{30}^{(1)}\right) \end{pmatrix}$$
$$=\begin{pmatrix} \dfrac{1}{1+e^{-\left(w_{11}^{(1)}a_1^{(0)}+w_{12}^{(1)}a_2^{(0)}+w_{13}^{(1)}a_3^{(0)}+w_{14}^{(1)}a_4^{(0)}+w_{10}^{(1)}\right)}} \\ \dfrac{1}{1+e^{-\left(w_{21}^{(1)}a_1^{(0)}+w_{22}^{(1)}a_2^{(0)}+w_{23}^{(1)}a_3^{(0)}+w_{24}^{(1)}a_4^{(0)}+w_{20}^{(1)}\right)}} \\ \dfrac{1}{1+e^{-\left(w_{31}^{(1)}a_1^{(0)}+w_{32}^{(1)}a_2^{(0)}+w_{33}^{(1)}a_3^{(0)}+w_{34}^{(1)}a_4^{(0)}+w_{30}^{(1)}\right)}} \end{pmatrix}. \tag{3.42.2}$$

Trying to find the derivative of complete composition function $\mathcal{N}\left(\mathbf{a}^{(0)}\right)$ algebraically becomes increasingly tedious as we increase the complexity of the NN.

Viewing a NN as a computational graph is a useful abstraction that helps in understanding and implementing NNs. The computational graph provides a clear and structured representation of the NN's operations, making it easier to apply the chain rule and perform efficient calculations for finding the derivative of the composition functions by using BP and automatic differentiation, as we will see in the next sections.

Computational graph separates the big computation into small steps, and we can find the derivative of each step (each computation) on the graph. Then the chain rule gives the derivatives of the final output. This is an incredibly efficient improvement on the computation of the derivative. At first it seems unbelievable, that reorganizing the computations can make such an enormous difference. In the end, you have to compute derivatives for each step and multiply by the chain rule. But the method does work.

**Definition (Directed Acyclic Computational Graph):** A directed acyclic computational graph is a directed acyclic graph of nodes, where each node contains a variable. Edges might be associated with learnable parameters. A variable in a node is either fixed externally (for input nodes with no incoming edges), or it is a computed as a function of the variables in the tail ends of edges incoming into the node and the learnable parameters on the incoming edges.

**Remarks:**

- We have been using graphs all along to represent NNs. In the following, we will use graphs to represent expressions instead.
- A graph is a collection of nodes (vertices) and edges connecting these nodes. In a directed graph, each edge has a specific direction, indicating a relationship between the connected nodes.
- Acyclic graph means that there are no cycles or loops in the graph. In other words, you cannot start at a node and follow a sequence of edges to return to the same node.
- Each node in the graph contains a variable. This variable can represent various quantities or data, depending on the context of the computational graph.
- For input nodes with no incoming edges, variables are fixed externally. This means that the values of these variables are provided from an external source rather than being computed within the graph.

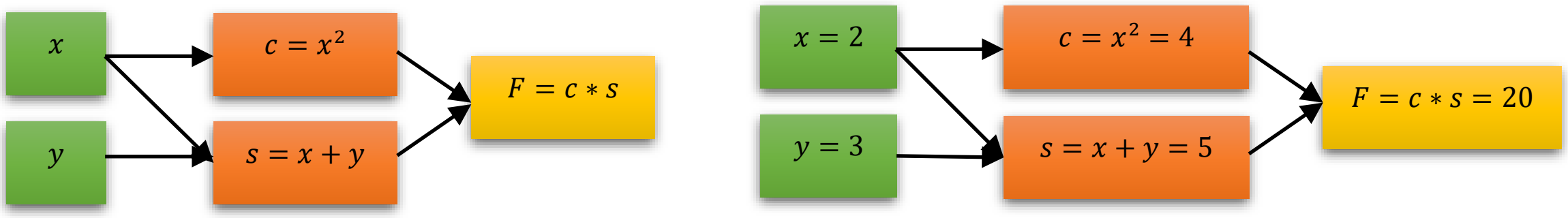

**Figure 3.10.** The computational graph for $F = x^2(x + y)$.

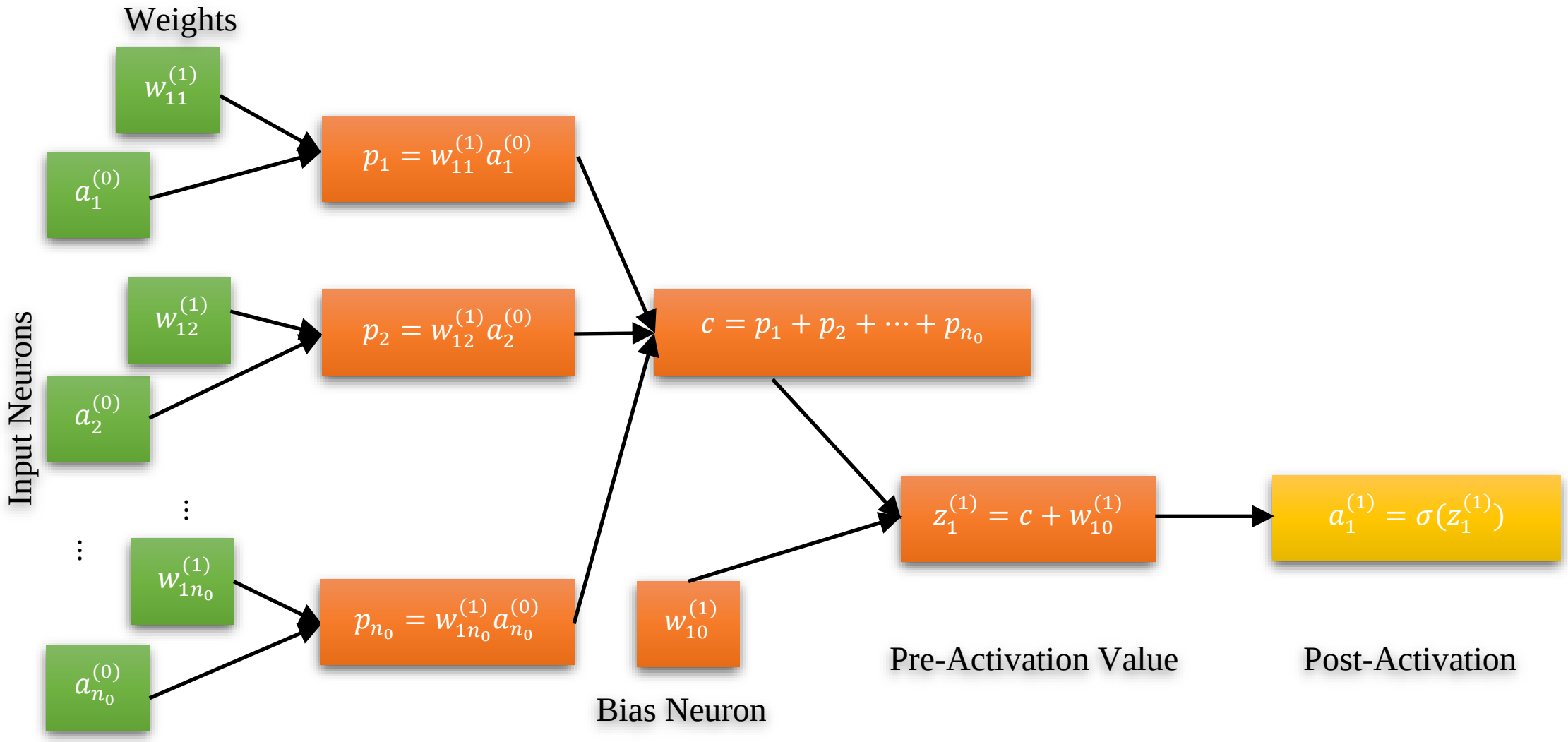

**Figure 3.11.** The computational graph for a neuron described in the text.

- Variables in nodes, other than input nodes, are computed as functions of the variables in the tail ends of edges and the learnable parameters associated with those edges. The "tail ends of edges" refer to the nodes from which the incoming edges originate.
- Edges in the graph may be associated with learnable parameters. These parameters are values that are adjusted during the learning process, typically through optimization algorithms like GD.
- The definition does not impose any restrictions on the type of function computed in a node. This means that various types of mathematical operations or functions can be applied at each node in the graph. Although it is common to use a combination of a linear function and an AF in a NN.
- Unlike the typical layered arrangement in NNs, where nodes are organized into input, hidden, and output layers, the definition allows for a more flexible arrangement of nodes. This flexibility enables the representation of various computational models that may not follow a strict layer-wise structure.
- The key restriction is that the graph must be acyclic. This constraint ensures that computations can be performed in a well-defined way without encountering infinite loops. The acyclic nature is crucial for the effective use of computational graphs in iterative optimization algorithms, such as BP in training NNs.
- Suppose $F(x, y)$ is a function of two variables $x$ and $y$. Those inputs are the first two nodes in the computational graph. A typical step in the computation -an edge in the graph- is one of the operations of arithmetic (addition, subtraction, multiplication, ...). The final output is the function $F(x, y)$. Our example will be $F = x^2(x + y)$. Figure 3.10 is the graph that computes $F$ with intermediate nodes $c = x^2$ and $s = x + y$. When we have inputs $x$ and $y$, for example $x = 2$ and $y = 3$, the edges lead to $c = 4$ and $s = 5$ and $F = 20$. This agrees with the algebra that we normally crowd into one line: $F = x^2(x + y) = 2^2(2 + 3) = 20$.
- In Figure 3.11, you will find the computational graph for the neuron described previously.

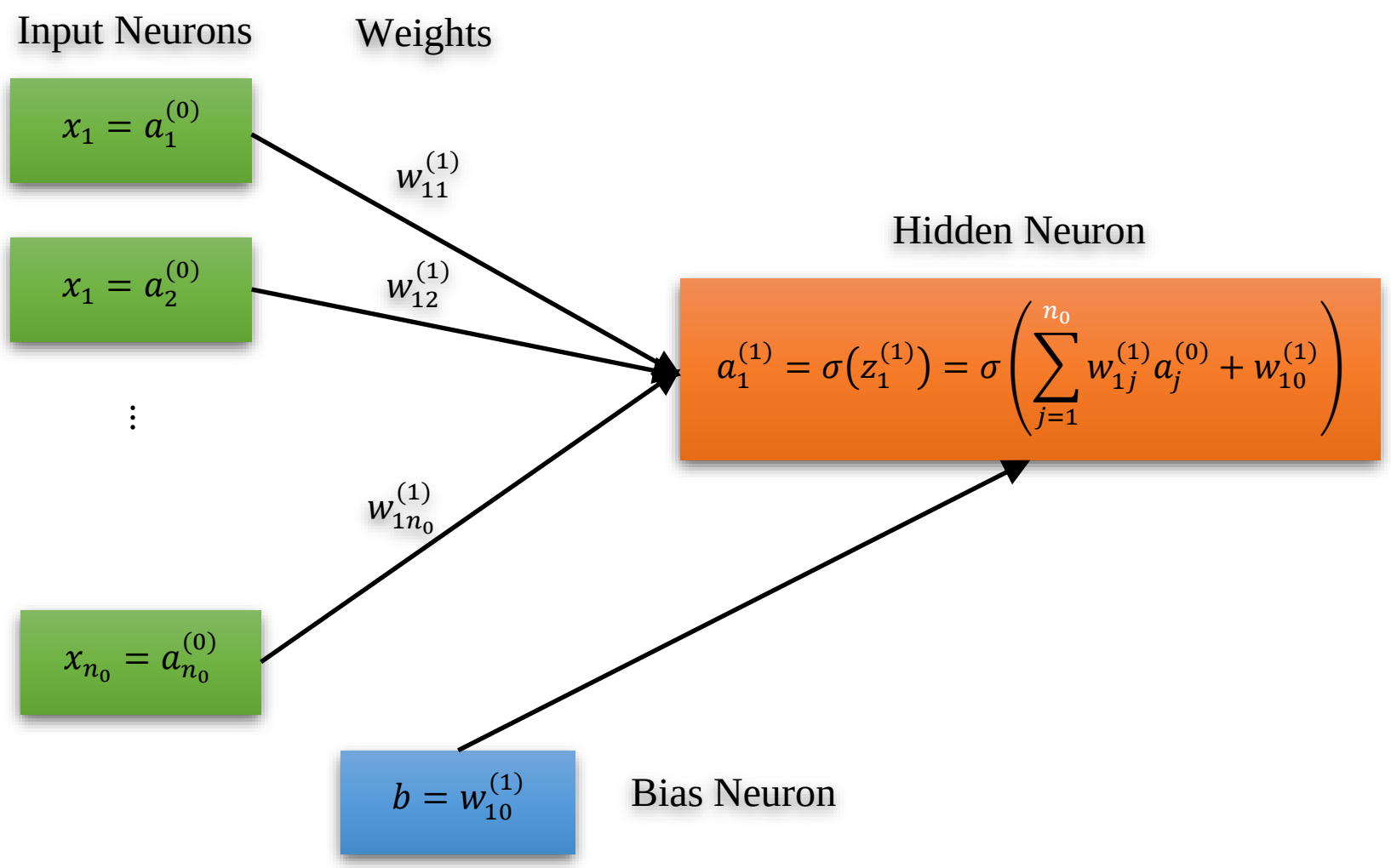

**Figure 3.12.** The neuron representation mostly used by practitioners.

Figure 3.11 is not what you usually find in, NNs books. It is rather complicated and not very practical to use, especially when you want to draw networks with many neurons. In the literature, you can find numerous representations for neurons. In this book, we will use the one shown in Figure 3.12, because it is widely used and is easy to understand. Figure 3.12 must be interpreted in the following way:

- The weights' names are written along the arrow. This means that before passing the inputs to the central node, the input first will be multiplied by the relative weight, as labeled on the arrow.
- The central node will perform several calculations at the same time. First, it will sum the inputs (the $w_{1j}^{(1)} a_j^{(0)}$ for $j = 1, 2, \dots, n_0$), then sum to the result the bias $w_{10}^{(1)}$, and, finally, apply to the resulting value the AF.

## 3.4 Automatic Differentiation and Its Main Modes

Automatic Differentiation (AD) is a technique used in computational mathematics and computer science for efficiently and accurately evaluating derivatives of functions [71-74]. It plays a crucial role in machine learning, optimization, and scientific computing. AD is also known as algorithmic differentiation or autodiff. BP is really a special case of automatic differentiation. It may be somewhat easier to understand the basic idea of BP by seeing the more general algorithm first. For that reason, we will first talk about autodiff.

**The core Idea** [71]**:** Automatic differentiation exploits the fact that all numerical computations, no matter how complicated, are ultimately compositions of a finite set of elementary arithmetic operations (addition, subtraction, multiplication, division, etc.) and elementary functions (exponential, logarithmic, trigonometric, etc.), for which derivatives are known, and combining the derivatives of the constituent operations through the chain rule gives the derivative of the overall composition. By doing so, AD can automatically and accurately calculate gradients, even for functions with intricate compositions and nested operations.

For example. Consider the scalar function

$$f(x) = \exp(\exp(x) + \exp(x)^2) + \sin(\exp(x) + \exp(x)^2). \tag{3.43}$$

By introducing intermediate variables, you can break down complex expressions into simpler steps, making it easier to apply the chain rule and compute derivatives. Say

$$a = \exp(x), \quad b = a^2, \quad c = a + b, \quad d = \exp(c), \quad e = \sin(c), \tag{3.44.1}$$

and finally, we have

$$f = d + e. \tag{3.44.2}$$

Fundamental to automatic differentiation is the decomposition of differentials provided by the chain rule of partial derivatives of composite functions. For the simple composition

$$\begin{aligned} y &= g_3\left(g_2\left(g_1(x)\right)\right) \\ &= g_3\left(g_2\left(g_1(v_0)\right)\right) \\ &= g_3\left(g_2(v_1)\right) \\ &= g_3(v_2) \\ &= v_3, \end{aligned} \tag{3.45}$$

where

$$v_0 = x, \qquad v_1 = g_1(v_0), \qquad v_2 = g_2(v_1), \qquad v_3 = g_3(v_2) = y, \tag{3.46}$$

the chain rule gives

$$\begin{aligned} \frac{\partial y}{\partial x} &= \frac{\partial y}{\partial v_2}\left(\frac{\partial v_2}{\partial v_1}\frac{\partial v_1}{\partial x}\right) \\ &= \left(\frac{\partial y}{\partial v_2}\frac{\partial v_2}{\partial v_1}\right)\frac{\partial v_1}{\partial x}. \end{aligned} \tag{3.47}$$

Usually, two distinct modes of automatic differentiation are presented.

**Forward Mode:**

- Forward accumulation (also called bottom-up, forward mode, or tangent mode)
- In forward mode, the computation proceeds from the input variables to the output variables.
- Forward accumulation specifies that one traverses the chain rule from inside to outside (that is, first compute $\partial v_1/\partial x$ and then $\partial v_2/\partial v_1$ and at last $\partial y/\partial v_2$.
- More succinctly, forward accumulation computes the recursive relation: $\frac{\partial v_i}{\partial x} = \frac{\partial v_i}{\partial v_{i-1}}\frac{\partial v_{i-1}}{\partial x}$ with $v_3 = y$.
- In forward accumulation AD, one first fixes the independent variable with respect to which differentiation is performed and computes the derivative of each sub-expression recursively. This involves repeatedly substituting the derivative of the inner functions in the chain rule:

$$\begin{aligned} \frac{\partial y}{\partial x} &= \frac{\partial y}{\partial v_{n-1}}\frac{\partial v_{n-1}}{\partial x} \\ &= \frac{\partial y}{\partial v_{n-1}}\left(\frac{\partial v_{n-1}}{\partial v_{n-2}}\frac{\partial v_{n-2}}{\partial x}\right) \\ &= \frac{\partial y}{\partial v_{n-1}}\left(\frac{\partial v_{n-1}}{\partial v_{n-2}}\left(\frac{\partial v_{n-2}}{\partial v_{n-3}}\frac{\partial v_{n-3}}{\partial x}\right)\right) \\ &= \cdots. \end{aligned} \tag{3.48}$$

- In forward accumulation, the quantity of interest is the tangent, denoted with a dot $\dot{v}_i$; it is a derivative of a subexpression with respect to a chosen independent variable

$$\dot{v}_i = \frac{\partial v_i}{\partial x}. \tag{3.49}$$

- Forward accumulation is particularly useful when the number of input variables is much smaller than the number of output variables (for functions $f: \mathbb{R}^n \to \mathbb{R}^m$ with $n \ll m$).

**Reverse Mode (or Backward Mode):**

- Reverse accumulation (also called top-down, reverse mode, or adjoint mode)
- In reverse mode, the computation proceeds backward from the output variables to the input variables.
- Reverse accumulation has the traversal from outside to inside (first compute $\partial y/\partial v_2$ and then $\partial v_2/\partial v_1$ and at last $\partial v_1/\partial x$.

- More succinctly, reverse accumulation computes the recursive relation: $\frac{\partial y}{\partial v_i} = \frac{\partial y}{\partial v_{i+1}} \frac{\partial v_{i+1}}{\partial v_i}$ with $v_0 = x$.
- In reverse accumulation AD, the dependent variable to be differentiated is fixed and the derivative is computed with respect to each sub-expression recursively. The derivative of the outer functions is repeatedly substituted in the chain rule:

$$\begin{aligned}\frac{\partial y}{\partial x} &= \frac{\partial y}{\partial v_1} \frac{\partial v_1}{\partial x} \\ &= \left(\frac{\partial y}{\partial v_2} \frac{\partial v_2}{\partial v_1}\right) \frac{\partial v_1}{\partial x} \\ &= \left(\left(\frac{\partial y}{\partial v_3} \frac{\partial v_3}{\partial v_2}\right) \frac{\partial v_2}{\partial v_1}\right) \frac{\partial v_1}{\partial x} \\ &= \cdots. \end{aligned} \tag{3.50}$$

- In reverse accumulation, the quantity of interest is the adjoint, denoted with a bar $\bar{v}_i$; it is a derivative of a dependent variable with respect to a subexpression

$$\bar{v}_i = \frac{\partial y}{\partial v_i}. \tag{3.51}$$

- Reverse accumulation is especially efficient when the number of output variables is much smaller than the number of input variables (for functions $f: \mathbb{R}^n \to \mathbb{R}^m$ with $n \gg m$).

**Examples**

Let us consider two illustrative examples, to clarify and deepen understanding of the forward and reverse modes. We adopt the three-part notation used by [71], where a function $f: \mathbb{R}^n \to \mathbb{R}^m$ is constructed using intermediate variables $v_i$ such that

- variables $v_{i-n} = x_i$, $i = 1, \ldots, n$ are the input variables,
- variables $v_i$, $i = 1, \ldots, l$ are the working (intermediate) variables, and
- variables $y_{m-i} = v_{l-i}$, $i = m-1, \ldots, 0$ are the output variables.

**Example of Forward Mode**

Step by step forward mode

- Express your function as a composition of elementary operations (addition, multiplication, etc.).
- Choose the variables with respect to which you want to compute derivatives (seeds) (usually input variables).
- Execute the computation graph in a forward direction, evaluating both the function values and the derivatives of the elementary operations with respect to the seed.
- Store intermediate values (function values and derivatives) at each node in the computation graph.
- Use the chain rule to propagate the derivatives through the computation graph.
- Combine the stored intermediate values to compute the derivative of the overall function with respect to each input variable.
- Extract the derivatives of interest. These derivatives represent the gradient of the function with respect to the chosen input variables.

On the left hand side of Table 3.1 we see the representation of the computation $y = f(x_1, x_2) = \ln(x_1) + x_1 x_2 - \sin(x_2)$ as an evaluation trace of elementary operations—also called a Wengert list and graphically represented in Figure 3.13. For computing the derivative of $f$ with respect to $x_1$, we start by associating with each intermediate variable $v_i$ a derivative $\dot{v}_i = \frac{\partial v_i}{\partial x_1}$. Applying the chain rule to each elementary operation in the forward primal trace, we generate the corresponding tangent (derivative) trace, given on the right-hand side in Table 3.1. Evaluating the primal $v_i$ in lockstep with their corresponding tangents $\dot{v}_i$ gives us the required derivative in the final variable $\dot{v}_5 = \frac{\partial y}{\partial x_1}$.

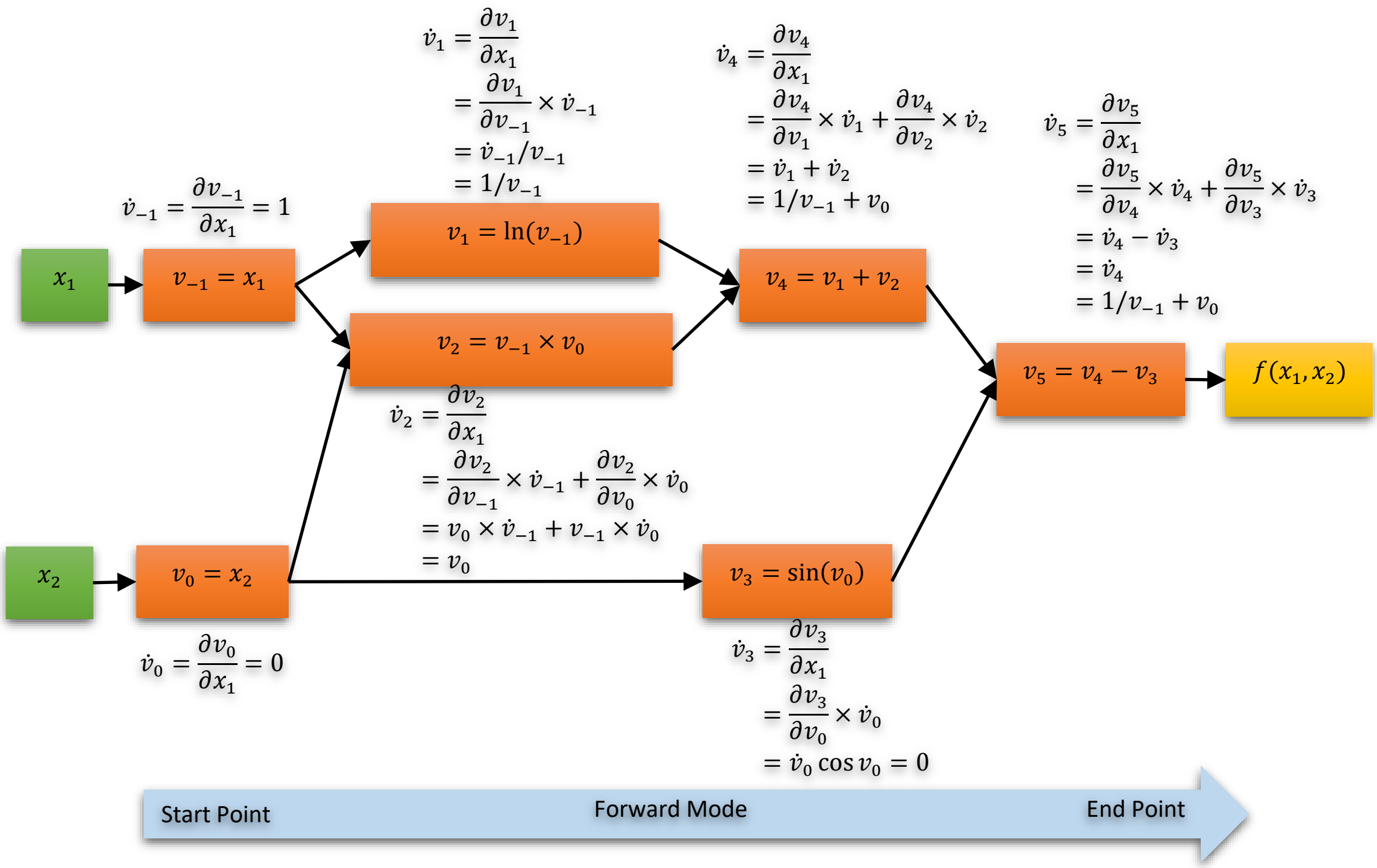

**Figure 3.13.** Example of forward accumulation with computational graph of $y = f(x_1, x_2) = \ln(x_1) + x_1x_2 - \sin(x_2)$. The choice of the independent variable to which differentiation is performed affects the seed values $\dot{v}_{-1}$ and $\dot{v}_0$. Given interest in the derivative of this function with respect to $x_1$, the seed values should be set to: $\dot{v}_{-1} = \dot{x}_1 = 1$, $\dot{v}_0 = \dot{x}_2 = 0$.

**Table 3.1.** Forward mode AD example, with $y = f(x_1, x_2) = \ln(x_1) + x_1x_2 - \sin(x_2)$ evaluated at $(x_1, x_2) = (2,5)$ and setting $\dot{x}_1 = 1$ to compute $\partial y/\partial x_1$, see Figure 3.13. The original forward evaluation of the primals on the left is augmented by the tangent operations on the right, where each line complements the original directly to its left.

| Forward Primal Trace | | Forward Tangent (Derivative) Trace | |
|---|---|---|---|
| $v_{-1} = x_1$ | $= 2$ | $\dot{v}_{-1} = \dot{x}_1$ | $= 1$ |
| $v_0 = x_2$ | $= 5$ | $\dot{v}_0 = \dot{x}_2$ | $= 0$ |
| $v_1 = \ln v_{-1}$ | $= \ln 2$ | $\dot{v}_1 = \dot{v}_{-1}/v_{-1}$ | $= 1/2$ |
| $v_2 = v_{-1} \times v_0$ | $= 2 \times 5$ | $\dot{v}_2 = \dot{v}_{-1} \times v_0 + v_{-1} \times \dot{v}_0$ | $= 1 \times 5 + 0 \times 2$ |
| $v_3 = \sin v_0$ | $= \sin 5$ | $\dot{v}_3 = \dot{v}_0 \cos v_0$ | $= 0 \times \cos 5$ |
| $v_4 = v_1 + v_2$ | $= 0.693 + 10$ | $\dot{v}_4 = \dot{v}_1 + \dot{v}_2$ | $= 0.5 + 5$ |
| $v_5 = v_4 - v_3$ | $= 10.693 + 0.959$ | $\dot{v}_5 = \dot{v}_4 - \dot{v}_3$ | $= 5.5 - 0$ |
| $y = v_5$ | $= 11.652$ | $\dot{y} = \dot{v}_5$ | $= 5.5$ |

### Example of Reverse Mode

Returning to the example $y = f(x_1, x_2) = \ln(x_1) + x_1x_2 - \sin(x_2)$, see Figure 3.14. In Table 3.2, we see the adjoint statements on the right-hand side, and original elementary operation on the left-hand side. In simple terms, we are interested in computing the contribution $\bar{v}_i = \frac{\partial y}{\partial v_i}$ of the change in each variable $v_i$ to the change in the output $y$. After the forward pass on the left-hand side, we run the reverse pass of the adjoints on the right-hand side, starting with $\bar{v}_5 = \bar{y} = \frac{\partial y}{\partial y} = 1$. In the end we get the derivatives $\frac{\partial y}{\partial x_1} = \bar{x}_1$ and $\frac{\partial y}{\partial x_2} = \bar{x}_2$ in just one reverse pass.

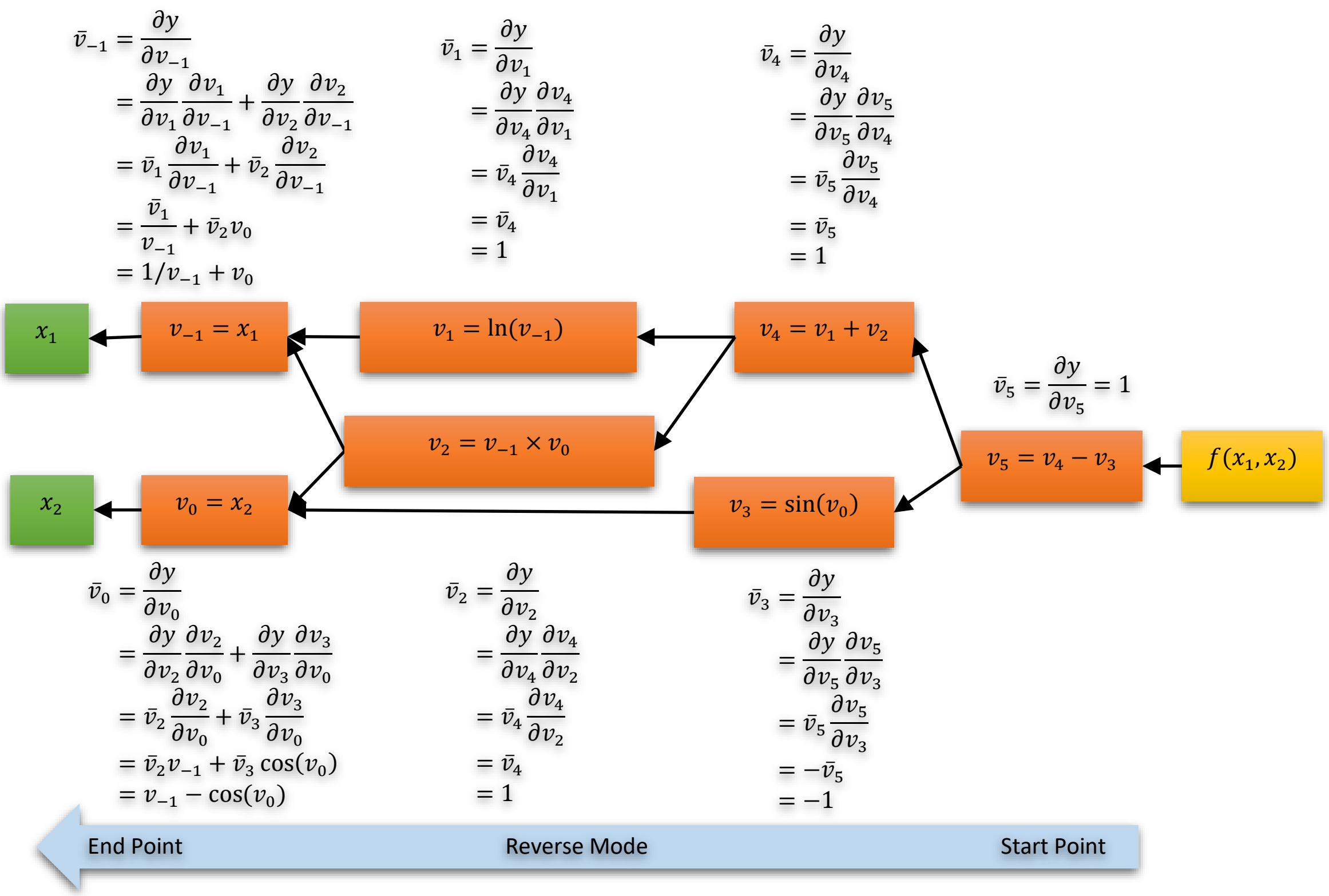

**Figure 3.14.** Example of reverse accumulation with computational graph of $y = f(x_1, x_2) = \ln(x_1) + x_1x_2 - \sin(x_2)$. Reverse accumulation evaluates the function first and calculates the derivatives with respect to all independent variables in an additional pass. The adjoint, $\bar{v}_i$, is a derivative of a dependent variable with respect to a subexpression $\bar{v}_i = \frac{\partial y}{\partial v_i}$.

**Table 3.2.** Reverse mode AD example, with $y = f(x_1, x_2) = \ln(x_1) + x_1x_2 - \sin(x_2)$ evaluated at $(x_1, x_2) = (2,5)$. After the forward evaluation of the primals on the left, the adjoint operations on the right are evaluated (see Figure 3.14). Note that both $\partial y/\partial x_1$ and $\partial y/\partial x_2$ are computed in the same reverse pass, starting from the adjoint $\bar{v}_5 = \bar{y} = \partial y/\partial y = 1$.

| Forward Primal Trace | Reverse Adjoint (Derivative) Trace |
|---|---|
| $v_{-1} = x_1 \quad = 2$<br>$v_0 = x_2 \quad = 5$ | $\bar{v}_5 = \bar{y} \quad = 1$<br>$\bar{v}_4 = \bar{v}_5\frac{\partial v_5}{\partial v_4} = \bar{v}_5 \times (1) \quad = 1$ |
| $v_1 = \ln v_{-1} \quad = \ln 2$ | $\bar{v}_3 = \bar{v}_5\frac{\partial v_5}{\partial v_3} = \bar{v}_5 \times (-1) \quad = -1$ |
| $v_2 = v_{-1} \times v_0 = 2 \times 5$ | $\bar{v}_1 = \bar{v}_4\frac{\partial v_4}{\partial v_1} = \bar{v}_4 \times 1 \quad = 1$ |
| $v_3 = \sin v_0 \quad = \sin 5$ | $\bar{v}_2 = \bar{v}_4\frac{\partial v_4}{\partial v_2} = \bar{v}_4 \times 1 \quad = 1$ |
| $v_4 = v_1 + v_2 \quad = 0.693 + 10$ | $\bar{v}_0 = \bar{v}_3\frac{\partial v_3}{\partial v_0} + \bar{v}_2\frac{\partial v_2}{\partial v_0} = \bar{v}_3 \cos v_0 + \bar{v}_2 v_{-1} \quad = 1.716$ |
| $v_5 = v_4 - v_3 \quad = 10.693 + 0.959$ | $\bar{v}_{-1} = \bar{v}_2\frac{\partial v_2}{\partial v_{-1}} + \bar{v}_1\frac{\partial v_1}{\partial v_{-1}} = \bar{v}_2 v_0 + \frac{\bar{v}_1}{v_{-1}} = 5.5$ |
| $y = v_5 \quad = 11.652$ | $\bar{x}_2 = \bar{v}_0 \quad = 1.716$<br>$\bar{x}_1 = \bar{v}_{-1} \quad = 5.5$ |

**Remarks:**

- Reverse accumulation traverses the chain rule from outside to inside. The example function is scalar-valued, and thus there is only one seed for the derivative computation, and only one sweep of the computational graph is needed to calculate the (two-component) gradient. This is only half the work when compared to forward accumulation, but reverse accumulation requires the storage of the intermediate variables $v_i$ as well as the instructions that produced them in a data structure known as a "tape" or a Wengert list, which may consume significant memory if the computational graph is large.
- Forward mode requires a new graph for each input $x_i$, to compute the partial derivative $\frac{\partial y}{\partial x_i}$.
- The reverse mode starts with the output $y$. It computes the derivatives with respect to both inputs. The computations go backward through the graph. That means it does not follow the empty line that started with $\partial x_2/\partial x_1 = 0$ in the forward graph for $x_1$-derivatives. And it would not follow the empty line $\partial x_1/\partial x_2 = 0$ in the forward graph for $x_2$-derivatives.
- A larger and more realistic problem with $N$ inputs will have $N$ forward graphs, each with $N - 1$ empty lines (because the $N$ inputs are independent). The derivative of $x_i$ with respect to every other input $x_j$ is $\partial x_i/\partial x_j = 0$. Instead of $N$ forward graphs from $N$ inputs, we will have one backward graph from one output. This is the success of reverse mode.
- Because machine learning practice principally involves the gradient of a scalar-valued objective with respect to a large number of parameters, this establishes the reverse mode, as opposed to the forward mode, as the mainstay technique in the form of the BP algorithm.

In NN, the difficulty arises when trying to compute the derivatives of the loss function with respect to the parameters for the purpose of gradient-based optimization, such as GD. Closed-form solutions, which would provide an explicit expression for these derivatives, are often elusive due to the intricate nature of the composition of functions in the graph. The reverse mode of automatic differentiation or BP addresses this issue by recursively applying the chain rule. In a computational graph representing a NN, the chain rule needs to be applied repeatedly. This is because the output is typically a complex function of many intermediate variables, and the chain rule helps in breaking down the derivatives with respect to each variable. Instead of explicitly writing out the entire function and differentiating it with respect to each parameter, BP allows the algorithm to traverse the computational graph backward, computing the derivatives at each step. This way, the complexity of the computation is managed automatically, and you can efficiently calculate the gradients needed for optimization without having to deal with the explicit closed form. In essence, BP leverages the structure of the computational graph to compute gradients efficiently, making it feasible to train deep and wide NNs without needing to write out and differentiate the entire function manually. The BP algorithm is a cornerstone of modern deep learning.

## 3.5 Training Process and Loss/Cost Functions

Loss functions, also known as cost functions or objective functions, play a crucial role in training NNs. These functions measure how well the model is performing by quantifying the difference between the predicted values and the actual ground truth values. The loss is essentially a measure of the model's error. The choice of a specific loss function depends on the nature of the problem you are trying to solve. Different tasks, such as regression or classification, may require different loss functions.

### Mean Squared Error

The quadratic loss function, also known as squared error loss, or mean squared error (MSE) is indeed commonly used, especially in regression problems and when employing least squares techniques. The mathematical formula is:

$$\mathcal{L}_{\text{MSE}} = \frac{1}{n}\sum_{j=1}^{n}\left(y_j - a_j\right)^2, \tag{3.52.1}$$

or

$$\mathcal{L}_{\text{MSE}} = c\sum_{j=1}^{n}\left(y_j - a_j\right)^2, \tag{3.52.2}$$

where $n$ is the number of instances in the dataset, $y_j$ is the actual value and $a_j$ is the predicted value, for the $j$-th instance and $c$ is a constant (the value of the constant makes no difference to a decision and can be ignored by setting it equal to 1). The derivative with respect to $a_i$ is (in the case $c = 1/2$):

$$\begin{aligned}\frac{\partial\mathcal{L}_{\text{MSE}}}{\partial a_i} &= \frac{1}{2}\frac{\partial}{\partial a_i}\sum_{j=1}^{n}\left(y_j - a_j\right)^2\\ &= \frac{1}{2}\frac{\partial}{\partial a_i}(y_i - a_i)^2\\ &= a_i - y_i.\end{aligned} \tag{3.53}$$

The summation term over $j = 1$ to $n$ is dropped because the derivatives $\frac{\partial}{\partial a_i}\left(y_j - a_j\right)^2$ will be zero for all $j$ except for the case when $j = i$. The quadratic loss has several desirable properties:

- The quadratic form simplifies mathematical operations, making it easier to find analytical solutions and derivatives. This is particularly advantageous in optimization algorithms.
- The loss is symmetric with respect to errors above and below the target. An error of $y_i - a_i$ and $a_i - y_i$ both contribute the same amount to the loss.
- The use of quadratic loss is well-suited for problems where the goal is to minimize the average squared difference between predicted and actual values. However, it is sensitive to outliers, and if your data contains outliers, other loss functions like Huber loss or Tukey's bisquare loss might be more robust.

**Binary Cross-Entropy (Logistic Loss):**

Binary Cross-Entropy (BCE), also known as logistic loss, is a loss function commonly used in binary classification problems. It is particularly associated with problems where the goal is to classify instances into one of two classes, often denoted as 0 or 1. The loss is calculated for each instance, and the overall loss is the average over all instances in the dataset. The formula for BCE loss is as follows:

$$\mathcal{L}_{\text{BCE}} = -\frac{1}{n}\sum_{j=1}^{n} y_j\log\left(a_j\right) + \left(1 - y_j\right)\log\left(1 - a_j\right), \tag{3.54}$$

where $n$ is the number of instances in the dataset, $y_j$ is the true label for the $j$-th instance (either 0 or 1) and $a_j$ is the predicted probability that the $j$-th instance belongs to class 1. This loss function is commonly used in logistic regression and NN models with a Sigmoid AF in the output layer for binary classification. The derivative with respect to $a_i$ is:

$$\begin{aligned}\frac{\partial}{\partial a_i}\mathcal{L}_{\text{BCE}} &= \frac{\partial}{\partial a_i}\left\{\frac{1}{n}\sum_{j=1}^{n}\left(-y_j\log\left(a_j\right) - \left(1 - y_j\right)\log\left(1 - a_j\right)\right)\right\}\\ &= \frac{\partial}{\partial a_i}\left\{\frac{1}{n}\left(-y_i\log(a_i) - (1 - y_i)\log(1 - a_i)\right)\right\}\\ &= \frac{1}{n}\left(-y_i\frac{1}{a_i} - (1 - y_i)\frac{(-1)}{(1 - a_i)}\right)\\ &= -\frac{1}{n}\left(\frac{y_i}{a_i} - \frac{1 - y_i}{1 - a_i}\right)\\ &= \frac{1}{n}\frac{a_i - y_i}{a_i(1 - a_i)}.\end{aligned} \tag{3.55}$$

The power of NNs lies in their ability to learn and optimize the parameters jointly through a process called training. The basic training process for a NN using GD and a loss function can describes as follows:

- During training, the network is presented with input data along with the corresponding target outputs. Then, the computations are performed in the forward direction to determine the predicted values of the outputs. If these predicted values are different from the observed values in the training data, compute the loss value. The loss function measures the difference between the predicted output of the NN and the true output (ground truth) for a given input. Mathematically, it is often denoted as $\mathcal{L}_i$, where the subscript $i$ indicates the specific training instance. The overall objective during training is to minimize the average or total loss over all training instances, $\mathcal{L}$. This is often expressed as the average of the individual losses across the entire training dataset. The objective function $\mathcal{L}$, which represents the goal of the optimization, is the sum or average of the individual losses:

$$\mathcal{L} = \frac{1}{n}\sum_{i=1}^{n} \mathcal{L}_i, \tag{3.56}$$

  where $n$ is the number of training instances.
- Symbolically, we can write the loss function as $\mathcal{L}(\boldsymbol{\theta})$, where $\boldsymbol{\theta}$ is a vector of all the weights and biases in the network. Our goal is to move through the space that the loss function defines to find the minimum, the specific $\boldsymbol{\theta}$ leading to the smallest loss, $\mathcal{L}$. The gradient of the loss function with respect to the parameters $\boldsymbol{\theta}$ of the model tells us how the loss changes concerning each parameter. GD is an optimization algorithm used to minimize the objective function in NNs. It works by iteratively adjusting the weights of the NN in the direction opposite to the gradient of the objective function with respect to the weights (iteratively moving towards the minimum of the function). This adjustment is proportional to the learning rate, a hyperparameter that determines the step size in the weight update, review Chapter 2.
- Therefore, to train a NN via GD, we need to know how each weight and bias value contributes to the loss function; that is, we need to know $\partial\mathcal{L}/\partial w$ and $\partial\mathcal{L}/\partial b$, for some weight $w$ and bias $b$.
- The weights of the NN are updated in the direction that reduces the loss. The general update rule for a weight $w_{ij}$ is given by:

$$w_{ij\,\text{new}} = w_{ij\,\text{old}} - \alpha\frac{\partial\mathcal{L}}{\partial w_{ij}}, \qquad \Delta w_{ij} = -\alpha\frac{\partial\mathcal{L}}{\partial w_{ij}}, \tag{3.57}$$

  where $\alpha$ is a free parameter ($\alpha$ is the "learning rate" that we set prior to training; it lets us scale our step size according to the problem at hand), and $\partial\mathcal{L}/\partial w_{ij}$ is the partial derivative of the objective function with respect to the weight $w_{ij}$. This iterative process of computing gradients, updating weights, and repeating the process is performed until the network's performance converges to an acceptable level or a predefined stopping criterion is met. This is the essence of supervised learning in NNs.
- This phase is referred to as the backwards phase. The rationale for calling it a "backwards phase" is that derivatives of the loss with respect to weights near the output (where the loss function is computed) are easier to compute and are computed first. The derivatives become increasingly complex as we move towards edge weights away from the output (in the backwards direction).
- The goal of BP is to compute the partial derivatives $\partial\mathcal{L}/\partial w$ and $\partial\mathcal{L}/\partial b$ of the cost function $\mathcal{L}$ with respect to any weight $w$ or bias $b$ in the network.
- Remember from Chapter 2, the univariate and multivariate chain rules allow you to find the derivative of a composite function. Let $f(u)$ be a function of $u$, and $u = g(x)$ be another function of $x$. The composite function is $h(x) = f(g(x))$. The univariate chain rule can be expressed as:

$$\frac{\partial h(x)}{\partial x} = \frac{\partial f(u)}{\partial u}\frac{\partial u}{\partial x}, \tag{3.58}$$

  or

$$\frac{\partial f(g(x))}{\partial x} = \frac{\partial f(g(x))}{\partial g(x)}\frac{\partial g(x)}{\partial x}. \tag{3.59}$$

This rule essentially states that the derivative of the composite function $h(x)$ with respect to $x$ is the product of the derivative of $f$ with respect to its immediate variable $u$, and the derivative of $g(x)$ with respect to $x$. The derivative $df(u)/du$ represents the local gradient of $f$ with respect to its immediate argument $u$, and $du/dx$ represents the local gradient of $g$ with respect to its immediate argument $x$. In the NNs, this univariate chain rule is applied iteratively during the backward pass in BP. Each layer of the network corresponds to a function, and the chain rule is used to compute the gradients with respect to the parameters of each layer by combining local gradients at each step. The multivariate chain rule is defined as follows:

$$\frac{\partial f(g_1(x),\dots,g_k(x))}{\partial x} = \sum_{i=1}^{k} \frac{\partial f(g_1(x),\dots,g_k(x))}{\partial g_i(x)} \frac{\partial g_i(x)}{\partial x}. \tag{3.60}$$

- In the training of a NN, an epoch (or round) refers to one complete pass through the entire training dataset. During each epoch, the NN processes every training example in the dataset, calculates the loss, performs BP to compute gradients. For each epoch, the parameters are updated based on the accumulated gradients of the loss with respect to the weights, see Example 3.1 for clarification.
- The number of epochs is a hyperparameter that determines how many times the learning algorithm will work through the entire training dataset. Choosing the right number of epochs is important. Too few epochs may result in the model not capturing the underlying patterns in the data, while too many epochs may lead to overfitting, where the model learns the training data too well but fails to generalize to new, unseen data.
- By learning the optimal parameters, NNs can capture complex patterns and relationships in data, allowing them to generalize well to unseen examples. The joint optimization of parameters enables NNs to create highly nonlinear and expressive compositions of simple functions, making them effective for a wide range of tasks, such as image recognition, natural language processing, and more. This is what makes NNs more powerful than their individual building blocks or basic parametric models.

***Example 3.1***

Let us define a simple NN, one that accepts two input values, has two nodes in its hidden layer, and has a single output node, as shown in Figure 3.15. We will use Sigmoid AFs $\sigma(z) = \frac{1}{1+e^{-z}}$ in the hidden layer. Notice that output has identity AF $\sigma(z) = z$. To train the network, we will use a squared-error loss function, $\mathcal{L}_{\text{MSE}} = \frac{1}{2}\left(y - a_1^{(2)}\right)^2$, where $y$ is the actual value and $a_1^{(2)}$ is the predicted value of the network for the input associated with $y$, namely $a_1^{(0)}$ and $a_2^{(0)}$.

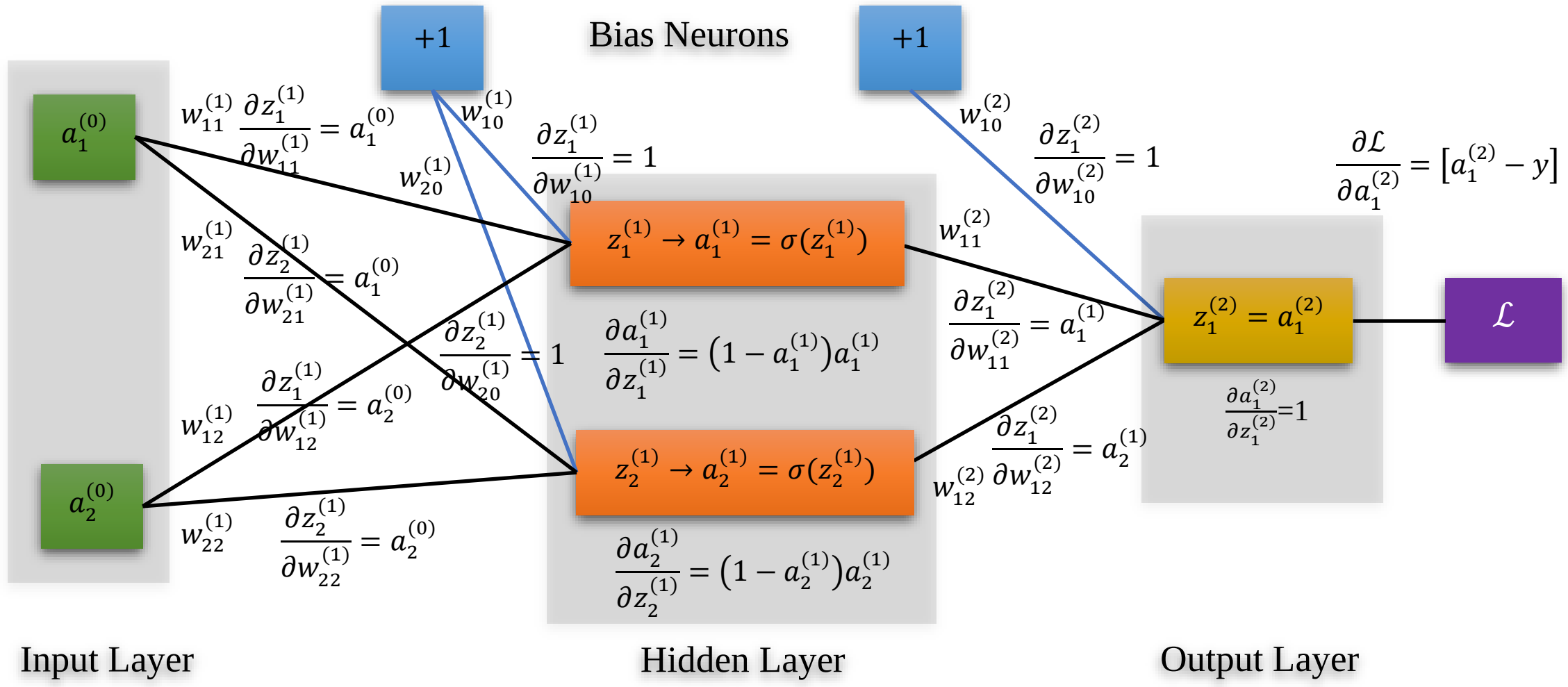

**Figure 3.15.** A simple NN.

***Solution***

The forward pass involves calculating the activations at each layer. The equations for the forward pass are as follows:

$$\begin{aligned} z_1^{(1)} &= w_{11}^{(1)} a_1^{(0)} + w_{12}^{(1)} a_2^{(0)} + w_{10}^{(1)}, \\ a_1^{(1)} &= \sigma\left(z_1^{(1)}\right) \\ &= \frac{1}{1 + e^{-\left(w_{11}^{(1)} a_1^{(0)} + w_{12}^{(1)} a_2^{(0)} + w_{10}^{(1)}\right)}}, \end{aligned}$$

$$\begin{aligned} z_2^{(1)} &= w_{21}^{(1)} a_1^{(0)} + w_{22}^{(1)} a_2^{(0)} + w_{20}^{(1)}, \\ a_2^{(1)} &= \sigma\left(z_2^{(1)}\right) \\ &= \frac{1}{1 + e^{-\left(w_{21}^{(1)} a_1^{(0)} + w_{22}^{(1)} a_2^{(0)} + w_{20}^{(1)}\right)}}, \end{aligned}$$

$$\begin{aligned} z_1^{(2)} &= w_{11}^{(2)} a_1^{(1)} + w_{12}^{(2)} a_2^{(1)} + w_{10}^{(2)}, \\ a_1^{(2)} &= \sigma\left(z_1^{(2)}\right) \\ &= w_{11}^{(2)} a_1^{(1)} + w_{12}^{(2)} a_2^{(1)} + w_{10}^{(2)}. \end{aligned}$$

If the label associated with the training example $\mathbf{a}^{(0)} = \left(a_1^{(0)}, a_2^{(0)}\right)^T$ is $y$, and the output of the network for this input is $a_1^{(2)}$, then we can define a loss function that measures the difference between the predicted output and the actual label, $\mathcal{L}_{\text{MSE}} = \frac{1}{2}\left(y - a_1^{(2)}\right)^2$. The argument to the loss function is $a_1^{(2)}$; $y$ is a fixed constant.

The loss function can be considered a function of the weights and biases, denoted as $\boldsymbol{\theta}$ (which includes $w_{11}^{(1)}$, $w_{12}^{(1)}$, $w_{21}^{(1)}$, $w_{22}^{(1)}$, $w_{11}^{(2)}$, $w_{12}^{(2)}$, $w_{10}^{(1)}$, $w_{20}^{(1)}$ and $w_{10}^{(2)}$), along with the constant input vector $\mathbf{a}^{(0)} = \left(a_1^{(0)}, a_2^{(0)}\right)^T$ and the associated label $y$. Mathematically, this can be expressed as:

$$\mathcal{L} = \mathcal{L}_{\text{MSE}}\left(\boldsymbol{\theta}; \mathbf{a}^{(0)}, y\right).$$

Here, $\boldsymbol{\theta}$ represents the set of weights and biases, and $\mathcal{L}$ is the loss function that measures the error between the predicted output and the true label. The input vector $\mathbf{a}^{(0)}$ and label $y$ are considered constants in this context, as they are fixed for a particular training example.

During the training process, the goal is to minimize this loss function with respect to the weights and biases ($\boldsymbol{\theta}$). This involves the forward pass to compute the predicted output $a_1^{(2)}$ given the current set of weights and biases, followed by the backward pass, BP, to calculate the gradients of the loss with respect to $\boldsymbol{\theta}$. These gradients are then used in an optimization algorithm to update the weights and biases iteratively. To perform gradient-based optimization, you need to compute the partial derivatives of the loss function with respect to each weight and bias in the network.

We also need an expression for the derivative of our AF, the Sigmoid. The derivative of the sigmoid is

$$\begin{aligned} \frac{d}{dz}\sigma(z) &= \frac{d}{dz}\frac{1}{1 + e^{-z}} \\ &= \frac{d}{dz}\frac{e^z}{1 + e^z} \\ &= \frac{e^z(1 + e^z) - e^z e^z}{(1 + e^z)^2} \\ &= \frac{e^z\left((1 + e^z) - e^z\right)}{(1 + e^z)^2} \\ &= \frac{1}{(1 + e^z)}\frac{e^z}{(1 + e^z)} \\ &= \left(1 - \frac{e^z}{(1 + e^z)}\right)\frac{e^z}{(1 + e^z)} \\ &= \left(1 - \sigma(x)\right)\sigma(x). \end{aligned}$$

The derivative of the Sigmoid function can be written in terms of the Sigmoid itself: This is a convenient property when performing BP in NNs because during the forward pass, you have already calculated $\sigma(z)$ as the activation of the neuron. So, you can reuse this value during the backward pass to efficiently compute the derivative without recalculating the Sigmoid.

BP involves working backward from the output layer to the input layer, applying the chain rule at each step to compute the partial derivatives. The expression for the partial derivative for the specified parameter $a_1^{(2)}$ is:

$$\begin{aligned}\frac{\partial \mathcal{L}}{\partial a_1^{(2)}} &= \frac{\partial}{\partial a_1^{(2)}}\left(\frac{1}{2}\left(y - a_1^{(2)}\right)^2\right) \\ &= a_1^{(2)} - y.\end{aligned}$$

Recall $y$ is the label for the current training example, and we compute $a_1^{(2)}$ during the forward pass as the output of the network. So, when working through the BP algorithm, you can replace $\frac{\partial \mathcal{L}}{\partial a_1^{(2)}}$ with $\left(a_1^{(2)} - y\right)$ in the relevant expressions, making the calculations more straightforward. This simplification is a common step in the BP process.

Output layer:

$$\begin{aligned}\frac{\partial \mathcal{L}}{\partial w_{11}^{(2)}} &= \frac{\partial \mathcal{L}}{\partial a_1^{(2)}}\frac{\partial a_1^{(2)}}{\partial w_{11}^{(2)}} = \left(a_1^{(2)} - y\right)\left(a_1^{(1)}\right), \\ \frac{\partial \mathcal{L}}{\partial w_{12}^{(2)}} &= \frac{\partial \mathcal{L}}{\partial a_1^{(2)}}\frac{\partial a_1^{(2)}}{\partial w_{12}^{(2)}} = \left(a_1^{(2)} - y\right)\left(a_2^{(1)}\right), \\ \frac{\partial \mathcal{L}}{\partial w_{10}^{(2)}} &= \frac{\partial \mathcal{L}}{\partial a_1^{(2)}}\frac{\partial a_1^{(2)}}{\partial w_{10}^{(2)}} = \left(a_1^{(2)} - y\right)(1).\end{aligned}$$

First neuron in hidden layer:

$$\begin{aligned}\frac{\partial \mathcal{L}}{\partial w_{11}^{(1)}} &= \frac{\partial \mathcal{L}}{\partial a_1^{(2)}}\frac{\partial a_1^{(2)}}{\partial a_1^{(1)}}\frac{\partial a_1^{(1)}}{\partial z_1^{(1)}}\frac{\partial z_1^{(1)}}{\partial w_{11}^{(1)}} = \left(a_1^{(2)} - y\right)\left(w_{11}^{(2)}\right)\left(\left(1 - a_1^{(1)}\right)a_1^{(1)}\right)\left(a_1^{(0)}\right), \\ \frac{\partial \mathcal{L}}{\partial w_{12}^{(1)}} &= \frac{\partial \mathcal{L}}{\partial a_1^{(2)}}\frac{\partial a_1^{(2)}}{\partial a_1^{(1)}}\frac{\partial a_1^{(1)}}{\partial z_1^{(1)}}\frac{\partial z_1^{(1)}}{\partial w_{12}^{(1)}} = \left(a_1^{(2)} - y\right)\left(w_{11}^{(2)}\right)\left(\left(1 - a_1^{(1)}\right)a_1^{(1)}\right)\left(a_2^{(0)}\right), \\ \frac{\partial \mathcal{L}}{\partial w_{10}^{(1)}} &= \frac{\partial \mathcal{L}}{\partial a_1^{(2)}}\frac{\partial a_1^{(2)}}{\partial a_1^{(1)}}\frac{\partial a_1^{(1)}}{\partial z_1^{(1)}}\frac{\partial z_1^{(1)}}{\partial w_{10}^{(1)}} = \left(a_1^{(2)} - y\right)\left(w_{11}^{(2)}\right)\left(\left(1 - a_1^{(1)}\right)a_1^{(1)}\right)(1).\end{aligned}$$

Second neuron in hidden layer:

$$\begin{aligned}\frac{\partial \mathcal{L}}{\partial w_{21}^{(1)}} &= \frac{\partial \mathcal{L}}{\partial a_1^{(2)}}\frac{\partial a_1^{(2)}}{\partial a_2^{(1)}}\frac{\partial a_2^{(1)}}{\partial z_2^{(1)}}\frac{\partial z_2^{(1)}}{\partial w_{21}^{(1)}} = \left(a_1^{(2)} - y\right)\left(w_{12}^{(2)}\right)\left(\left(1 - a_2^{(1)}\right)a_2^{(1)}\right)\left(a_1^{(0)}\right), \\ \frac{\partial \mathcal{L}}{\partial w_{22}^{(1)}} &= \frac{\partial \mathcal{L}}{\partial a_1^{(2)}}\frac{\partial a_1^{(2)}}{\partial a_2^{(1)}}\frac{\partial a_2^{(1)}}{\partial z_2^{(1)}}\frac{\partial z_2^{(1)}}{\partial w_{22}^{(1)}} = \left(a_1^{(2)} - y\right)\left(w_{12}^{(2)}\right)\left(\left(1 - a_2^{(1)}\right)a_2^{(1)}\right)\left(a_2^{(0)}\right), \\ \frac{\partial \mathcal{L}}{\partial w_{20}^{(1)}} &= \frac{\partial \mathcal{L}}{\partial a_1^{(2)}}\frac{\partial a_1^{(2)}}{\partial a_2^{(1)}}\frac{\partial a_2^{(1)}}{\partial z_2^{(1)}}\frac{\partial z_2^{(1)}}{\partial w_{20}^{(1)}} = \left(a_1^{(2)} - y\right)\left(w_{12}^{(2)}\right)\left(\left(1 - a_2^{(1)}\right)a_2^{(1)}\right)(1),\end{aligned}$$

where we use $\frac{\partial a_1^{(1)}}{\partial z_1^{(1)}} = \left(1 - \sigma\left(z_1^{(1)}\right)\right)\sigma\left(z_1^{(1)}\right) = \left(1 - a_1^{(1)}\right)a_1^{(1)}$ and $\frac{\partial a_2^{(1)}}{\partial z_2^{(1)}} = \left(1 - a_2^{(1)}\right)a_2^{(1)}$.

In the context of updating the bias term ($w_{10}^{(2)}$), the update rule for each epoch can be written as follows:

$$w_{10}^{(2)} = w_{10}^{(2)} - \alpha\frac{1}{m}\sum_{i=1}^{m}\left.\frac{\partial \mathcal{L}}{\partial w_{10}^{(2)}}\right|_{\mathbf{a}_i^{(0)}}.$$

Here: $w_{10}^{(2)}$ is the current value of the bias term. $\alpha$ is the learning rate. $m$ is the number of samples in the training set. $\mathcal{L}$ is the loss function. $\left.\frac{\partial \mathcal{L}}{\partial w_{10}^{(2)}}\right|_{\mathbf{a}_i^{(0)}}$ is the partial derivative of the loss with respect to $w_{10}^{(2)}$ evaluated for the $i$-th

training sample. The summation term is the accumulation of the partial derivatives of the loss with respect to $w_{10}^{(2)}$ over all training samples. This accumulation represents the overall contribution of the bias term to the loss across the entire training set. The division by $m$ is to take the average contribution over all samples, and the entire term $\frac{1}{m}\sum_{i=1}^{m}\frac{\partial\mathcal{L}}{\partial w_{10}^{(2)}}\Big|_{\mathbf{a}_i^{(0)}}$ is the amount by which the bias term is adjusted during the update step. This process is repeated for each parameter in the NN (weights and biases) during each epoch to gradually improve the model's performance. The learning rate ($\alpha$) controls the size of the step taken in the direction of the steepest decrease in the loss landscape.

Output layer:

$$w_{11}^{(2)} = w_{11}^{(2)} - \alpha\frac{1}{m}\sum_{i=1}^{m}\frac{\partial\mathcal{L}}{\partial w_{11}^{(2)}}\Bigg|_{\mathbf{a}_i^{(0)}} = w_{11}^{(2)} - \alpha\frac{1}{m}\sum_{i=1}^{m}\left(\left(a_1^{(2)} - y\right)\left(a_1^{(1)}\right)\right)\Big|_{\mathbf{a}_i^{(0)}},$$

$$w_{12}^{(2)} = w_{12}^{(2)} - \alpha\frac{1}{m}\sum_{i=1}^{m}\frac{\partial\mathcal{L}}{\partial w_{12}^{(2)}}\Bigg|_{\mathbf{a}_i^{(0)}} = w_{12}^{(2)} - \alpha\frac{1}{m}\sum_{i=1}^{m}\left(\left(a_1^{(2)} - y\right)\left(a_2^{(1)}\right)\right)\Big|_{\mathbf{a}_i^{(0)}},$$

$$w_{10}^{(2)} = w_{10}^{(2)} - \alpha\frac{1}{m}\sum_{i=1}^{m}\frac{\partial\mathcal{L}}{\partial w_{10}^{(2)}}\Bigg|_{\mathbf{a}_i^{(0)}} = w_{10}^{(2)} - \alpha\frac{1}{m}\sum_{i=1}^{m}\left(a_1^{(2)} - y\right)\Big|_{\mathbf{a}_i^{(0)}}.$$

First neuron in hidden layer:

$$w_{11}^{(1)} = w_{11}^{(1)} - \alpha\frac{1}{m}\sum_{i=1}^{m}\frac{\partial\mathcal{L}}{\partial w_{11}^{(1)}}\Bigg|_{\mathbf{a}_i^{(0)}} = w_{11}^{(1)} - \alpha\frac{1}{m}\sum_{i=1}^{m}\left(\left(a_1^{(2)} - y\right)\left(w_{11}^{(2)}\right)\left(\left(1 - a_1^{(1)}\right)a_1^{(1)}\right)\left(a_1^{(0)}\right)\right)\Big|_{\mathbf{a}_i^{(0)}},$$

$$w_{12}^{(1)} = w_{12}^{(1)} - \alpha\frac{1}{m}\sum_{i=1}^{m}\frac{\partial\mathcal{L}}{\partial w_{12}^{(1)}}\Bigg|_{\mathbf{a}_i^{(0)}} = w_{12}^{(1)} - \alpha\frac{1}{m}\sum_{i=1}^{m}\left(\left(a_1^{(2)} - y\right)\left(w_{11}^{(2)}\right)\left(\left(1 - a_1^{(1)}\right)a_1^{(1)}\right)\left(a_2^{(0)}\right)\right)\Big|_{\mathbf{a}_i^{(0)}},$$

$$w_{10}^{(1)} = w_{10}^{(1)} - \alpha\frac{1}{m}\sum_{i=1}^{m}\frac{\partial\mathcal{L}}{\partial w_{10}^{(1)}}\Bigg|_{\mathbf{a}_i^{(0)}} = w_{10}^{(1)} - \alpha\frac{1}{m}\sum_{i=1}^{m}\left(\left(a_1^{(2)} - y\right)\left(w_{11}^{(2)}\right)\left(\left(1 - a_1^{(1)}\right)a_1^{(1)}\right)\right)\Big|_{\mathbf{a}_i^{(0)}}.$$

Second neuron in hidden layer:

$$w_{21}^{(1)} = w_{21}^{(1)} - \alpha\frac{1}{m}\sum_{i=1}^{m}\frac{\partial\mathcal{L}}{\partial w_{21}^{(1)}}\Bigg|_{\mathbf{a}_i^{(0)}} = w_{21}^{(1)} - \alpha\frac{1}{m}\sum_{i=1}^{m}\left(\left(a_1^{(2)} - y\right)\left(w_{12}^{(2)}\right)\left(\left(1 - a_2^{(1)}\right)a_2^{(1)}\right)\left(a_1^{(0)}\right)\right)\Big|_{\mathbf{a}_i^{(0)}},$$

$$w_{22}^{(1)} = w_{22}^{(1)} - \alpha\frac{1}{m}\sum_{i=1}^{m}\frac{\partial\mathcal{L}}{\partial w_{22}^{(1)}}\Bigg|_{\mathbf{a}_i^{(0)}} = w_{22}^{(1)} - \alpha\frac{1}{m}\sum_{i=1}^{m}\left(\left(a_1^{(2)} - y\right)\left(w_{12}^{(2)}\right)\left(\left(1 - a_2^{(1)}\right)a_2^{(1)}\right)\left(a_2^{(0)}\right)\right)\Big|_{\mathbf{a}_i^{(0)}},$$

$$w_{20}^{(1)} = w_{20}^{(1)} - \alpha\frac{1}{m}\sum_{i=1}^{m}\frac{\partial\mathcal{L}}{\partial w_{20}^{(1)}}\Bigg|_{\mathbf{a}_i^{(0)}} = w_{20}^{(1)} - \alpha\frac{1}{m}\sum_{i=1}^{m}\left(\left(a_1^{(2)} - y\right)\left(w_{12}^{(2)}\right)\left(\left(1 - a_2^{(1)}\right)a_2^{(1)}\right)\right)\Big|_{\mathbf{a}_i^{(0)}}.$$

## 3.6 The Four Fundamental Equations Behind Backpropagation

The key idea behind BP is to update the weights and biases of the network in the direction that reduces the error or cost function. To achieve this, the algorithm computes the partial derivatives of the cost function with respect to the weights ($\partial\mathcal{L}/\partial w_{jk}^{(l)}$) and biases ($\partial\mathcal{L}/\partial w_{j0}^{(l)}$) in each layer of the NN. The derivatives $\partial\mathcal{L}/\partial w_{jk}^{(l)}$ and $\partial\mathcal{L}/\partial w_{j0}^{(l)}$, represent how much the cost function changes with respect to the weights and biases, respectively.

BP is indeed based on four fundamental equations that help in computing the error and the gradient of the cost function. These equations are derived using the chain rule and are applied during the backward pass of the training process. The auxiliary variable $\boldsymbol{\delta}^{(l)}$ is commonly used in BP to represent the derivative of the loss ($\mathcal{L}$) with respect to the weighted sum of inputs $\mathbf{z}^{(l)}$ for layer $l$, i.e., the auxiliary variable $\boldsymbol{\delta}^{(l)}$ represents the rate of change of the loss with respect to the pre-activation output of layer $l$. Here, $l$ can take values from 0 to the final layer $L$. In mathematical terms:

$$\boldsymbol{\delta}^{(l)} = \frac{\partial \mathcal{L}}{\partial \mathbf{z}^{(l)}}. \tag{3.61}$$

Let us revisit and update our notations, opting to reintegrate $\mathbf{b}^{(l)}$ as the biases for layer $l$, $\mathbf{z}^{(l)}$ as the weighted input to layer $l$, $\mathbf{a}^{(l)}$ as the output of layer $l$ after applying the AF, $\mathcal{L}$ as the cost function, $\mathbf{W}^{(l)}$ as the weights for layer $l$, as commonly done in most literature. Our strategy involves initially presenting the equations with a straightforward example to demonstrate their application in a spatial case, Example 3.2. Following this, we will provide a comprehensive proof for the general case. The four fundamental equations are as follows:

**1. Error in the Output Layer $\boldsymbol{\delta}^{(L)}$:**

The components of $\delta^{(L)}$ are given by

$$\delta_j^{(L)} = \frac{\partial \mathcal{L}}{\partial a_j^{(L)}} \sigma'\left(z_j^{(L)}\right). \tag{3.62}$$

The first term on the right, $\partial \mathcal{L}/\partial a_j^{(L)}$, just measures how fast the cost is changing as a function of the $j$-th output activation. The second term on the right, $\sigma'\left(z_j^{(L)}\right)$, measures how fast the AF $\sigma$ is changing at $z_j^{(L)}$. Notice that everything in (3.62) is easily computed. In particular, we compute $z_j^{(L)}$ while computing the behaviour of the network, and it is easy to compute $\sigma'\left(z_j^{(L)}\right)$. The exact form of $\partial \mathcal{L}/\partial a_j^{(L)}$ will, of course, depend on the form of the cost function. However, provided the cost function is known there should be little trouble computing $\partial \mathcal{L}/\partial a_j^{(L)}$. For example, if we're using the quadratic cost function then $\mathcal{L} = \frac{1}{2}\sum_j \left(y_j - a_j^{(L)}\right)^2$, and so $\partial \mathcal{L}/\partial a_j^{(L)} = \left(a_j^{(L)} - y_j\right)$, which obviously is easily computable. Equation (3.62) is a component-wise expression for $\delta^{(L)}$. It is a perfectly good expression, but not the matrix-based form we want for BP. However, it is easy to rewrite the equation in a matrix-based form, as

$$\boldsymbol{\delta}^{(L)} = \boldsymbol{\nabla}_a \mathcal{L} \odot \sigma'\left(\mathbf{z}^{(L)}\right), \tag{3.63}$$

where $\boldsymbol{\nabla}_a \mathcal{L}$ is the gradient of the cost function with respect to the output activations, $\sigma'\left(\mathbf{z}^{(L)}\right)$ is the derivative of the AF at the output layer, and $\odot$ denotes element-wise multiplication (Hadamard product between two vectors). It is basically a vector of elementwise products of corresponding vector elements. Thus,

$$\mathbf{a} = \begin{pmatrix} a_0 \\ a_1 \\ \vdots \\ a_n \end{pmatrix}, \quad \mathbf{b} = \begin{pmatrix} b_0 \\ b_1 \\ \vdots \\ b_n \end{pmatrix}, \quad \mathbf{a} \odot \mathbf{b} = \begin{pmatrix} a_0 b_0 \\ a_1 b_1 \\ \vdots \\ a_n b_n \end{pmatrix}. \tag{3.64}$$

$\boldsymbol{\nabla}_a \mathcal{L}$ is a vector whose components are the partial derivatives $\partial \mathcal{L}/\partial a_j^{(L)}$. You can think of $\boldsymbol{\nabla}_a \mathcal{L}$ as expressing the rate of change of $\mathcal{L}$ with respect to the output activations. As an example, in the case of the quadratic cost we have $\boldsymbol{\nabla}_a \mathcal{L} = \left(\mathbf{a}^{(L)} - \mathbf{y}\right)$, and so the fully matrix-based form of (3.63) becomes

$$\boldsymbol{\delta}^{(L)} = \left(\mathbf{a}^{(L)} - \mathbf{y}\right) \odot \sigma'\left(\mathbf{z}^{(L)}\right). \tag{3.65}$$

**2. Error in Terms of the Next Layer's Error $\boldsymbol{\delta}^{(l)}$ for Hidden Layers:**

$$\boldsymbol{\delta}^{(l)} = \left(\left(\mathbf{W}^{(l+1)}\right)^T \boldsymbol{\delta}^{(l+1)}\right) \odot \sigma'\left(\mathbf{z}^{(l)}\right), \tag{3.66}$$

where $\mathbf{W}^{(l+1)}$ is the weight matrix connecting layer $l$ to layer $l+1$, $\boldsymbol{\delta}^{(l+1)}$ is the error in layer $(l+1)$, and $\sigma'\left(\mathbf{z}^{(l)}\right)$ is the derivative of the AF at layer $l$. By combining (3.63) with (3.66) we can compute the error $\boldsymbol{\delta}^{(l)}$ for any layer in the network. We start by using (3.63) to compute $\boldsymbol{\delta}^{(L)}$, then apply (3.66) to compute $\boldsymbol{\delta}^{(L-1)}$, then (3.66) again to compute $\boldsymbol{\delta}^{(L-2)}$, and so on, all the way back through the network.

**3. Gradient of the Cost Function with Respect to Biases $\frac{\partial \mathcal{L}}{\partial \mathbf{b}^{(l)}}$:**

In component form,

$$\frac{\partial \mathcal{L}}{\partial b_j^{(l)}} = \delta_j^{(l)}. \tag{3.67}$$

In vector form,

$$\frac{\partial \mathcal{L}}{\partial \mathbf{b}^{(l)}} = \boldsymbol{\delta}^{(l)}. \tag{3.68}$$

The gradient of the cost function with respect to the biases in layer $(l)$ is simply the error in layer $(l)$.

**4. Gradient of the Cost Function with Respect to Weights $\frac{\partial \mathcal{L}}{\partial \mathbf{W}^{(l)}}$:**

In component form,

$$\frac{\partial \mathcal{L}}{\partial w_{jk}^{(l)}} = a_k^{(l-1)} \delta_j^{(l)}. \tag{3.69}$$

In vector form,

$$\frac{\partial \mathcal{L}}{\partial \mathbf{W}^{(l)}} = \boldsymbol{\delta}^{(l)} \left(\mathbf{a}^{(l-1)}\right)^T. \tag{3.70}$$

The gradient of the cost function with respect to the weights in layer $(l)$ is the outer product of the output activations of layer $(l-1)$ and the error in layer $(l)$.

These equations provide the necessary information to update the weights and biases in the network using a GD optimization algorithm, allowing the model to learn from the training data.

***Example 3.2***

Let us go through the BP process for a simple NN with a single neuron per layer and using MSE loss, see Figure 3.16. We will now prove the four fundamental equations in this simple case. For simplicity, we will denote the weights and biases without subscripts, as there is only one weight and one bias between two successive layers. We will use superscripts to indicate layer IDs.

***Solution***

Let us establish a few important equations for this NN in Figure 3.16.

Forward propagation, for arbitrary layer $l \in \{0, L\}$:

$$z^{(l)} = w^{(l)} a^{(l-1)} + b^{(l)},$$
$$a^{(l)} = \sigma\left(z^{(l)}\right).$$

Here we are working with a single training data instance, $x_i$ whose output is $y_i$.

$$\mathcal{L} = \frac{1}{2}\left(a^{(L)} - y_i\right)^2.$$

**Partial derivative of loss with respect to weight and bias in terms of auxiliary variable, for last layer $L$:** Using the chain rule for partial derivatives

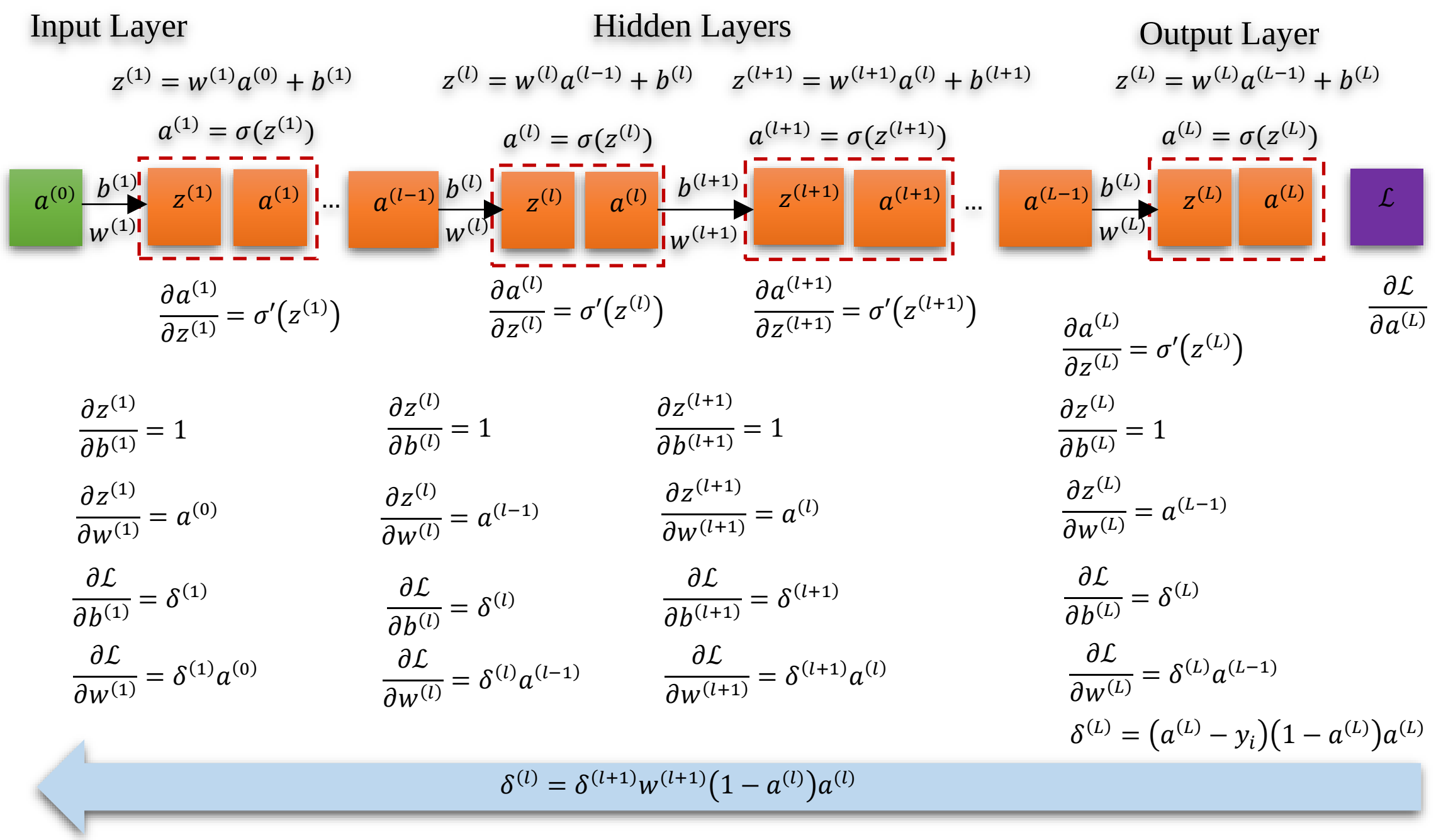

**Figure 3.16.** A simple NN (single neuron per layer).

$$\frac{\partial \mathcal{L}}{\partial w^{(L)}} = \frac{\partial \mathcal{L}}{\partial z^{(L)}}\frac{\partial z^{(L)}}{\partial w^{(L)}} = \delta^{(L)}\frac{\partial z^{(L)}}{\partial w^{(L)}},$$

where

$$\delta^{(L)} = \frac{\partial \mathcal{L}}{\partial z^{(L)}}.$$

And using the forward propagation equations,

$$\frac{\partial z^{(L)}}{\partial w^{(L)}} = a^{(L-1)}.$$

Together, they lead to

$$\frac{\partial \mathcal{L}}{\partial w^{(L)}} = \delta^{(L)}a^{(L-1)}.$$

Similarly,

$$\frac{\partial \mathcal{L}}{\partial b^{(L)}} = \frac{\partial \mathcal{L}}{\partial z^{(L)}}\frac{\partial z^{(L)}}{\partial b^{(L)}} = \delta^{(L)}.$$

**Auxiliary variable, for last layer *L*:** Using the chain rule for partial derivatives

$$\delta^{(L)} = \frac{\partial \mathcal{L}}{\partial z^{(L)}} = \frac{\partial \mathcal{L}}{\partial a^{(L)}}\frac{\partial a^{(L)}}{\partial z^{(L)}} = \left(a^{(L)} - y_i\right)\frac{\partial \sigma\left(z^{(L)}\right)}{\partial z^{(L)}}.$$

Using derivative of sigmoid, we get

$$\begin{aligned}\delta^{(L)} &= \left(a^{(L)} - y_i\right)\frac{\partial \sigma\left(z^{(L)}\right)}{\partial z^{(L)}} \\ &= \left(a^{(L)} - y_i\right)\left(1 - \sigma\left(z^{(L)}\right)\right)\sigma\left(z^{(L)}\right) \\ &= \left(a^{(L)} - y_i\right)\left(1 - a^{(L)}\right)a^{(L)}.\end{aligned}$$

**Partial Derivative of loss with respect to weight and bias in terms of auxiliary variable, for arbitrary layer *l*:** Using the chain rule for partial derivatives

$$\frac{\partial \mathcal{L}}{\partial w^{(l)}} = \frac{\partial \mathcal{L}}{\partial z^{(l)}}\frac{\partial z^{(l)}}{\partial w^{(l)}}.$$

Using the definition of the auxiliary variable and the forward propagation, this leads to

$$\frac{\partial \mathcal{L}}{\partial w^{(l)}} = \delta^{(l)} a^{(l-1)}.$$

Similarly,

$$\frac{\partial \mathcal{L}}{\partial b^{(l)}} = \frac{\partial \mathcal{L}}{\partial z^{(l)}} \frac{\partial z^{(l)}}{\partial b^{(l)}} = \delta^{(l)}.$$

**Auxiliary variable, for arbitrary layer $l$:** Using the chain rule for partial derivatives

$$\delta^{(l)} = \frac{\partial \mathcal{L}}{\partial z^{(l)}} = \frac{\partial \mathcal{L}}{\partial z^{(l+1)}} \frac{\partial z^{(l+1)}}{\partial a^{(l)}} \frac{\partial a^{(l)}}{\partial z^{(l)}}.$$

Using the definition of the auxiliary variable and the forward propagation, this leads to

$$\delta^{(l)} = \frac{\partial \mathcal{L}}{\partial z^{(l+1)}} \frac{\partial z^{(l+1)}}{\partial a^{(l)}} \frac{\partial a^{(l)}}{\partial z^{(l)}} = \frac{\partial \mathcal{L}}{\partial z^{(l+1)}} \frac{\partial z^{(l+1)}}{\partial a^{(l)}} \frac{\partial \sigma\left(z^{(l)}\right)}{\partial z^{(l)}} = \delta^{(l+1)} w^{(l+1)} \left(1 - a^{(l)}\right) a^{(l)}.$$

We initialize the system with some values of weights $w^{(l)}$ and biases $b^{(l)}$. Using those we can evaluate layer 1 outputs. We can evaluate, $z^{(1)}$ and $a^{(1)}$ easily (since all the inputs are known)

$$z^{(1)} = w^{(1)} a^{(0)} + b^{(1)},$$
$$a^{(1)} = \sigma\left(z^{(1)}\right).$$

But, if we have $z^{(1)}$ and $a^{(1)}$ , using them we can evaluate $z^{(2)}$ and $a^{(2)}$. And we can proceed all the way upto layer $L$ in this fashion to obtain $a^{(L)}$ which is the grand output of the NN. In other words, we can iteratively evaluate the outputs of successive layers using only the outputs from previous layer. No other layers need to be known. At any given iteration, we have to only keep the previous layer in memory and we can build the current layer from that. A similar trick can be applied to evaluate the auxiliary variables too, except there we have go backwards. We can evaluate the auxiliary variable for last layer, $\delta^{(L)}$. But once we have $\delta^{(L)}$, we can evaluate $\delta^{(L-1)}$. From that we can evaluate $\delta^{(L-2)}$. We can proceed in this fashion all they way till layer 0, evaluating successively $\delta^{(L)}, \delta^{(L-1)}, \cdots,$ $\delta^{(0)}$. Every time we evaluate a $\delta^{(l)}$, we can also evaluate the $\frac{\partial \mathcal{L}}{\partial w^{(l)}}$ and $\frac{\partial \mathcal{L}}{\partial b^{(l)}}$ for the same layer. Thus, starting from the last layer, we can update the weights and biases of all layers till layer 0 in this fashion. This is BP.

### Proof of the Four Fundamental Equations Behind BP for an Arbitrary Network

Now we will study a more generic network. A NN with $L + 1$ layers consists of an input layer, hidden layers, and an output layer, with each layer having a specific number of neurons and connections, and the network learns by adjusting weights and biases during training. The ultimate goal is to evaluate the partial derivatives of the loss with respect to the weights and biases. The overall strategy will be as follows.

- Derive expressions to compute the auxiliary variables for the output layer of the NN. This involves calculating how much the loss function changes concerning the output of the network.
- Derive an expression to compute the auxiliary variables for an arbitrary hidden layer ($l$) based on the auxiliary variables of the next layer ($l + 1$). This process involves propagating the error backward through the network.
- Use the derived expressions to compute auxiliary variables for each layer, starting from the last layer and moving backward through the network until reaching the input layer (layer 0).
- Derive expressions to compute the partial derivatives of the loss with respect to the weights and biases for each layer. This step involves applying the chain rule to find how the loss changes concerning the weights and biases at each layer. This gives us everything we need.
- Use the computed partial derivatives to update the weights and biases of the NN in a way that minimizes the loss. This step is done using an optimization algorithm, such as GD.

Forward propagation through this network has already been described in Section 3.2 and can be succinctly represented by (3.13.1) and (3.13.3). We will repeat them here for handy reference. On the left side, the scalar equations - for one neuron at a time - are presented. On the right-hand side, the vector equations - for entire layer - are presented. They are equivalent.

$$z_j^{(l)} = \sum_{k=0}^{m} w_{jk}^{(l)} a_k^{(l-1)} + b_j^{(l)}, \qquad \mathbf{z}^{(l)} = \mathbf{W}^{(l)}\mathbf{a}^{(l-1)} + \mathbf{b}^{(l)}, \tag{3.71}$$

$$a_j^{(l)} = \sigma\left(z_j^{(l)}\right), \qquad \mathbf{a}^{(l)} = \sigma\left(\mathbf{z}^{(l)}\right). \tag{3.72}$$

We define auxiliary variables as,

$$\delta_j^{(l)} = \frac{\partial \mathcal{L}}{\partial z_j^{(l)}}. \tag{3.73}$$

Applying the chain rule, we can re-express the partial derivative above in terms of partial derivatives with respect to the output activations,

$$\delta_j^{(L)} = \sum_k \frac{\partial \mathcal{L}}{\partial a_k^{(L)}} \frac{\partial a_k^{(L)}}{\partial z_j^{(L)}}, \tag{3.74}$$

where the sum is over all neurons $k$ in the output layer. Of course, the output activation $a_k^{(L)}$ of the $k$-th neuron depends only on the weighted input $z_j^{(L)}$ for the $j$-th neuron when $k = j$. And so $\partial a_k^{(L)} / \partial z_j^{(L)}$ vanishes when $k \neq j$. As a result, we can simplify the previous equation to (in component form)

$$\delta_j^{(L)} = \frac{\partial \mathcal{L}}{\partial a_j^{(L)}} \frac{\partial a_j^{(L)}}{\partial z_j^{(L)}} = \frac{\partial \mathcal{L}}{\partial a_j^{(L)}} \sigma'\left(z_j^{(L)}\right). \tag{3.75.1}$$

In vector form, we have

$$\underbrace{\boldsymbol{\delta}^{(L)}}_{n_L\times 1} = \underbrace{\boldsymbol{\nabla}_a \mathcal{L}}_{n_L\times 1} \odot \underbrace{\sigma'\left(\underbrace{\mathbf{z}^{(L)}}_{n_L\times 1}\right)}_{n_L\times 1} = \underbrace{\boldsymbol{\nabla}_a \mathcal{L}}_{n_L\times 1} \odot \underbrace{\sigma'\left|\underbrace{\mathbf{z}^{(L)}}_{n_L\times 1}\right\rangle}_{n_L\times 1}. \tag{3.75.2}$$

(3.75.1) and (3.75.2) are identical. The former is a scalar equation expressing individual auxiliary variables of the last layer. The latter is a vector equation expressing all the auxiliary variables of the last layer together. We can compute these directly if we have performed a forward pass and have its results, the $a_j^{(L)}$'s available along with the training data ground truth.

Note that, for the activation layer, if the input has $n_L$-elements, then the output also has $n_L$ elements. Therefore, the relationship $\sigma'\left(\mathbf{z}^{(L)}\right)$ should map a $n_L$ element vector to another $n_L$-element vector. The term, $\boldsymbol{\nabla}_a \mathcal{L}$ is a $n_L$-element vector. Finally, the Hadamard product between the two also outputs a $n_L$-element vector, as needed.

Auxiliary variable for arbitrary layer $l$: Using chain rule of partial differentiation

$$\begin{aligned}\delta_j^{(l)} &= \sum_k \frac{\partial \mathcal{L}}{\partial z_k^{(l+1)}} \frac{\partial z_k^{(l+1)}}{\partial z_j^{(l)}} \\ &= \sum_k \frac{\partial z_k^{(l+1)}}{\partial z_j^{(l)}} \delta_k^{(l+1)},\end{aligned} \tag{3.76}$$

where

$$\frac{\partial \mathcal{L}}{\partial z_k^{(l+1)}} = \delta_k^{(l+1)}. \tag{3.77}$$

To evaluate the first term on the last line, note that

$$z_k^{(l+1)} = \sum_j w_{kj}^{(l+1)} a_j^{(l)} + b_k^{(l+1)}$$
$$= \sum_j w_{kj}^{(l+1)} \sigma\left(z_j^{(l)}\right) + b_k^{(l+1)}. \tag{3.78}$$

Differentiating, we obtain

$$\frac{\partial z_k^{(l+1)}}{\partial z_j^{(l)}} = w_{kj}^{(l+1)} \sigma'\left(z_j^{(l)}\right). \tag{3.79}$$

Substituting back into (3.76) we obtain (in component form)

$$\delta_j^{(l)} = \sum_k w_{kj}^{(l+1)} \delta_k^{(l+1)} \sigma'\left(z_j^{(l)}\right). \tag{3.80.1}$$

In vector form, we have

$$\underbrace{\boldsymbol{\delta}^{(l)}}_{n_l\times 1} = \left(\underbrace{\left(\mathbf{W}^{(l+1)}\right)^T}_{n_l\times n_{l+1}} \underbrace{\boldsymbol{\delta}^{(l+1)}}_{n_{l+1}\times 1}\right) \odot \underbrace{\sigma'\left(\underbrace{\mathbf{z}^{(l)}}_{n_l\times 1}\right)}_{n_l\times 1} = \left(\underbrace{\left(\mathbf{W}^{(l+1)}\right)^T}_{n_l\times n_{l+1}} \underbrace{\left|\boldsymbol{\delta}^{(l+1)}\right\rangle}_{n_{l+1}\times 1}\right) \odot \underbrace{\sigma'\,\underbrace{\left|\mathbf{z}^{(l)}\right\rangle}_{n_l\times 1}}_{n_l\times 1}. \tag{3.80.2}$$

(3.80.1) and (3.80.2) allow us to evaluated $\boldsymbol{\delta}^{(l)}$'s from the $\boldsymbol{\delta}^{(l+1)}$'s if the results of forward propagation are available. We have already shown that the auxiliary variables for the last layer are directly computable from the activations of that layer. Hence, these to evaluate all the preceding layers' auxiliary variables.

For the fully connected layer, we have an $n_l$-element input, $\mathbf{z}^{(l)}$; an $n_{l+1} \times n_l$ element weight matrix, $\mathbf{W}^{(l+1)}$. So, we need to generate an $n_l$-element vector, $\boldsymbol{\delta}^{(l)}$, from the $n_{l+1}$-element error term, $\boldsymbol{\delta}^{(l+1)}$. Multiplying the transpose of the weight matrix, an $n_l \times n_{l+1}$ element matrix, by the error term does result in an $n_l$-element vector, since $n_l \times n_{l+1}$ by $n_{l+1} \times 1$ is $n_l \times 1$, an $n_l$-element column vector.

Note that

$$\frac{\partial \mathbf{z}^{(l+1)}}{\partial \mathbf{z}^{(l)}} = \frac{\partial}{\partial \mathbf{z}^{(l)}}\left(\mathbf{W}^{(l+1)} \sigma\left(\mathbf{z}^{(l)}\right) + \mathbf{b}^{(l+1)}\right)$$
$$= \left(\mathbf{W}^{(l+1)}\right)^T \sigma'\left(\mathbf{z}^{(l)}\right). \tag{3.81}$$

The result is $\left(\mathbf{W}^{(l+1)}\right)^T$, not $\mathbf{W}^{(l+1)}$, because the derivative of a matrix times a vector in denominator notation is the transpose of the matrix rather than the matrix itself, see Chapter 2 for details.

Now we are going to express the partial derivatives of loss with respect to weights and biases in terms of those. This will provide us with the gradients we need to update the weights and biases along negative gradient which is the optimal move to minimize loss.

$$\frac{\partial \mathcal{L}}{\partial w_{jk}^{(l)}} = \frac{\partial \mathcal{L}}{\partial z_j^{(l)}} \frac{\partial z_j^{(l)}}{\partial w_{jk}^{(l)}} = \delta_j^{(l)} a_k^{(l-1)}, \tag{3.82.1}$$

$$\underbrace{\frac{\partial \mathcal{L}}{\partial \mathbf{W}^{(l)}}}_{n_l\times n_{l-1}} = \underbrace{\boldsymbol{\delta}^{(l)}}_{n_l\times 1} \underbrace{\left(\mathbf{a}^{(l-1)}\right)^T}_{1\times n_{l-1}} = \underbrace{\left|\boldsymbol{\delta}^{(l)}\right\rangle}_{n_l\times 1} \underbrace{\left\langle \mathbf{a}^{(l-1)}\right|}_{1\times n_{l-1}}. \tag{3.82.2}$$

(3.82.1) and (3.82.2) are equivalent. The first one is scalar and pertains to individual weights in layer $l$. The second one describes the entire layer. The weight matrix, $\mathbf{W}^{(l)}$, is an $n_l \times n_{l-1}$ element matrix. Therefore, the $\nabla_{\mathbf{W}^{(l)}}\mathcal{L}$, also must be an $n_l \times n_{l-1}$ matrix. We know $\boldsymbol{\delta}^{(l)}$ is an $n_l$-element column vector, and the transpose of $\mathbf{a}^{(l-1)}$ is an $n_{l-1}$-element row vector. The outer product of the two is an $n_l \times n_{l-1}$ matrix, as required.

Similarly,

$$\frac{\partial \mathcal{L}}{\partial b_j^{(l)}} = \frac{\partial \mathcal{L}}{\partial z_j^{(l)}} \frac{\partial z_j^{(l)}}{\partial b_j^{(l)}} = \delta_j^{(l)}, \tag{3.83.1}$$

$$\underbrace{\frac{\partial \mathcal{L}}{\partial \mathbf{b}^{(l)}}}_{n_l\times 1} = \underbrace{\boldsymbol{\delta}^{(l)}}_{n_l\times 1} = \underbrace{|\boldsymbol{\delta}^{(l)}\rangle}_{n_l\times 1}. \tag{3.83.2}$$

(3.83.1) and (3.83.2) are equivalent. The first one is scalar and pertains to individual biases in layer $l$. The second one describes the entire layer.

Now we are going to express the partial derivatives of loss with respect to weights and biases in vectorized form for all training examples (if you are working with a batch of examples):

$$\left.\underbrace{\frac{\partial \mathcal{L}}{\partial \mathbf{W}^{(l)}}}_{n_l\times n_{l-1}}\right|_{\text{average}} = \frac{1}{m}\sum_{i=1}^{m} \underbrace{\boldsymbol{\delta}_i^{(l)}}_{n_l\times 1} \underbrace{\left(\mathbf{a}_i^{(l-1)}\right)^T}_{1\times n_{l-1}} = \frac{1}{m}\sum_{i=1}^{m} \underbrace{|\boldsymbol{\delta}_i^{(l)}\rangle}_{n_l\times 1} \underbrace{\langle \mathbf{a}_i^{(l-1)}|}_{1\times n_{l-1}}, \tag{3.83.3}$$

$$\left.\underbrace{\frac{\partial \mathcal{L}}{\partial \mathbf{b}^{(l)}}}_{n_l\times 1}\right|_{\text{average}} = \frac{1}{m}\sum_{i=1}^{m} \underbrace{\boldsymbol{\delta}_i^{(l)}}_{n_l\times 1} = \frac{1}{m}\sum_{i=1}^{m} \underbrace{|\boldsymbol{\delta}_i^{(l)}\rangle}_{n_l\times 1}, \tag{3.83.4}$$

where $m$ is the number of training examples and $\mathcal{L} = \frac{1}{m}\sum_{i=1}^{m}\mathcal{L}_i$ is the average error across all training examples.

Moreover, the entire BP propagation process can be expressed in matrixized form for all training examples:

$$\underbrace{\boldsymbol{\Delta}^{(L)}}_{n_L\times m} = \begin{pmatrix} | & \dots & | \\ \boldsymbol{\delta}_1^{(L)} & \dots & \boldsymbol{\delta}_m^{(L)} \\ | & \dots & | \end{pmatrix} = \underbrace{\frac{\partial \mathcal{L}}{\partial \mathbf{A}^{(L)}}}_{n_L\times m} \odot \underbrace{\sigma'\Big(\underbrace{\mathbf{Z}^{(L)}}_{n_L\times m}\Big)}_{n_L\times m} = \underbrace{\frac{\partial \mathcal{L}}{\partial \mathbf{A}^{(L)}}}_{n_L\times m} \odot \underbrace{\sigma'\,\underbrace{|\mathbf{Z}^{(L)}\rangle}_{n_L\times m}}_{n_L\times m}, \tag{3.84.1}$$

$$\underbrace{\boldsymbol{\Delta}^{(l)}}_{n_l\times m} = \begin{pmatrix} | & \dots & | \\ \boldsymbol{\delta}_1^{(l)} & \dots & \boldsymbol{\delta}_m^{(l)} \\ | & \dots & | \end{pmatrix} = \Bigg(\underbrace{\left(\mathbf{W}^{(l+1)}\right)^T}_{n_l\times n_{l+1}} \underbrace{\boldsymbol{\Delta}^{(l+1)}}_{n_{l+1}\times m}\Bigg) \odot \underbrace{\sigma'\Big(\underbrace{\mathbf{Z}^{(l)}}_{n_l\times m}\Big)}_{n_l\times m} = \Bigg(\underbrace{\left(\mathbf{W}^{(l+1)}\right)^T}_{n_l\times n_{l+1}} \underbrace{|\boldsymbol{\Delta}^{(l+1)}\rangle}_{n_{l+1}\times m}\Bigg) \odot \underbrace{\sigma'\,\underbrace{|\mathbf{Z}^{(l)}\rangle}_{n_l\times m}}_{n_l\times m}, \tag{3.84.2}$$

$$\left.\underbrace{\frac{\partial \mathcal{L}}{\partial \mathbf{W}^{(l)}}}_{n_l\times n_{l-1}}\right|_{\text{average}} = \frac{1}{m} \underbrace{\boldsymbol{\Delta}^{(l)}}_{n_l\times m} \underbrace{\left(\mathbf{A}^{(l-1)}\right)^T}_{m\times n_{l-1}} = \frac{1}{m} \underbrace{|\boldsymbol{\Delta}^{(l)}\rangle}_{n_l\times m} \underbrace{\langle \mathbf{A}^{(l-1)}|}_{m\times n_{l-1}}, \tag{3.84.3}$$

$$\left.\underbrace{\frac{\partial \mathcal{L}}{\partial \mathbf{b}^{(l)}}}_{n_l\times 1}\right|_{\text{average}} = \frac{1}{m}\sum_{i=1}^{m} \underbrace{\boldsymbol{\delta}_i^{(l)}}_{n_l\times 1} = \frac{1}{m}\sum_{i=1}^{m} \underbrace{|\boldsymbol{\delta}_i^{(l)}\rangle}_{n_l\times 1}, \tag{3.84.4}$$

where $m$ is the number of training examples. $\mathbf{A}^{(0)} = \mathbf{X}$ is the input data, $\mathbf{A}^{(l)}$ is the $(n_l \times m)$ activation matrix for layer $l$, and $\mathbf{Z}^{(l)}$ is the $(n_l \times m)$ weighted sum matrix for layer $l$. For each layer $l$, $\mathbf{W}^{(l)}$ is $(n_l \times n_{l-1})$ weight matrix. $\boldsymbol{\Delta}^{(l)}$ is $(n_l \times m)$ Auxiliary matrix for arbitrary layer $l$. The final output of the network is given by $\mathbf{A}^{(L)}$, and its dimensions depend on the number of output units, $(n_L \times m)$ activation matrix for the output layer.

**Special case:**

Using the loss function $\mathcal{L} = \frac{1}{2}\left\|\mathbf{a}^{(L)} - \bar{\mathbf{y}}\right\|^2$ and Sigmoid AF, the four fundamental equations of BP are (in component form),

$$\delta_j^{(L)} = \left(a_j^{(L)} - \bar{y}_j\right) a_j^{(L)}\left(1 - a_j^{(L)}\right), \tag{3.85.1}$$

$$\delta_j^{(l)} = \delta_k^{(l+1)} w_{kj}^{(l+1)} a_j^{(l)}\left(1 - a_j^{(l)}\right), \tag{3.85.2}$$

$$\frac{\partial \mathcal{L}}{\partial w_{jk}^{(l)}} = \frac{\partial \mathcal{L}}{\partial z_j^{(l)}} \frac{\partial z_j^{(l)}}{\partial w_{jk}^{(l)}} = \delta_j^{(l)} a_k^{(l-1)}, \tag{3.85.3}$$

$$\frac{\partial \mathcal{L}}{\partial b_j^{(l)}} = \frac{\partial \mathcal{L}}{\partial z_j^{(l)}} \frac{\partial z_j^{(l)}}{\partial b_j^{(l)}} = \delta_j^{(l)}. \tag{3.85.4}$$

In vector form, we have (see Algorithm 3.1)

$$\boldsymbol{\delta}^{(L)} = \left(\mathbf{a}^{(L)} - \bar{\mathbf{y}}\right) \odot \mathbf{a}^{(L)} \odot \left(1 - \mathbf{a}^{(L)}\right), \tag{3.86.1}$$

$$\boldsymbol{\delta}^{(l)} = \left(\left(\mathbf{W}^{(l+1)}\right)^T \boldsymbol{\delta}^{(l+1)}\right) \odot \mathbf{a}^{(l)} \odot \left(1 - \mathbf{a}^{(l)}\right), \tag{3.86.2}$$

$$\nabla_{\mathbf{W}^{(l)}} \mathcal{L} = \boldsymbol{\delta}^{(l)} \left(\mathbf{a}^{(l-1)}\right)^T, \tag{3.86.3}$$

$$\nabla_{\mathbf{b}^{(l)}} \mathcal{L} = \boldsymbol{\delta}^{(l)}. \tag{3.86.4}$$

***Algorithm 3.1*** *Training a NN*

Initialize **W**, **b** with random values
while $\mathcal{L}$ > threshold do

**Forward pass**

for $l \leftarrow 1$ to $L$ do
$\mathbf{z}^{(l)} = \mathbf{W}^{(l)}\mathbf{a}^{(l-1)} + \mathbf{b}^{(l)}$
$\mathbf{a}^{(l)} = \sigma\left(\mathbf{z}^{(l)}\right)$
end for

**Loss**

$\mathcal{L} = \frac{1}{2}\left\|\mathbf{a}^{(L)} - \bar{\mathbf{y}}\right\|^2$

**Gradients for the last layer**

$\boldsymbol{\delta}^{(L)} = \left(\mathbf{a}^{(L)} - \bar{\mathbf{y}}\right) \odot \mathbf{a}^{(L)} \odot \left(1 - \mathbf{a}^{(L)}\right)$
$\nabla_{\mathbf{W}^{(L)}} \mathcal{L} = \boldsymbol{\delta}^{(L)}\left(\mathbf{a}^{(L-1)}\right)^T$
$\nabla_{\mathbf{b}^{(L)}} \mathcal{L} = \boldsymbol{\delta}^{(L)}$

**Gradients for the remaining layers**

for $l \leftarrow L - 1$ to 1 do
$\boldsymbol{\delta}^{(l)} = \left(\left(\mathbf{W}^{(l+1)}\right)^T \boldsymbol{\delta}^{(l+1)}\right) \odot \mathbf{a}^{(l)} \odot \left(1 - \mathbf{a}^{(l)}\right)$
$\nabla_{\mathbf{W}^{(l)}} \mathcal{L} = \boldsymbol{\delta}^{(l)}\left(\mathbf{a}^{(l-1)}\right)^T$
$\nabla_{\mathbf{b}^{(l)}} \mathcal{L} = \boldsymbol{\delta}^{(l)}$
end for

**Parameter Update**

$\mathbf{W}^{(l)} = \mathbf{W}^{(l)} - \alpha \boldsymbol{\nabla}_{\mathbf{W}^{(l)}} \mathcal{L}$
$\mathbf{b}^{(l)} = \mathbf{b}^{(l)} - \alpha \boldsymbol{\nabla}_{\mathbf{b}^{(l)}} \mathcal{L}$

end while

## 3.7 Batch, Stochastic, and Mini-Batch Gradient Descent

### 3.7.1 Full-Batch Gradient Descent (Gradient Descent)

How can we apply Full-Batch Gradient Descent (FBGD) (traditional GD) to learn in a NN? The idea is to use GD to find the weights $\mathbf{W}^{(l)}$ and biases $\mathbf{b}^{(l)}$ which minimize the cost function. In GD, one tries to minimize the cost function of the NN by moving the parameters along the negative direction of the gradient of the loss over all points, because this is the direction of the steepest descent. Most machine learning problems can be recast as optimization problems over a linearly additive sum of the loss functions on the individual training data points. However, the loss over the entire data set is really defined as the sum (average) of the losses over individual training points. One can write the loss function of a NN as the sum of point-specific losses:

$$\mathcal{L} = \frac{1}{m}\sum_{i=1}^{m} \mathcal{L}_{\mathbf{x}_i}, \tag{3.87}$$

where $m$ is the number of all training inputs $\{\mathbf{x}_i\}$, $i \in 1,..m$, and $\mathcal{L}_{\mathbf{x}_i}$ is the loss contributed by the $i$th training point. To understand what the problem is, let's look back at the quadratic cost function. Notice that this cost function is an average over costs $\mathcal{L}_{\mathbf{x}_i} \equiv \frac{\left(y_i - a^{(L)}\right)^2}{2}$ for individual training examples. In practice, to compute the gradient $\nabla\mathcal{L}$ we need to compute the gradients $\nabla\mathcal{L}_{\mathbf{x}_i}$ separately for each training input, $i$, and then average them,

$$\nabla\mathcal{L} = \frac{1}{m}\sum_{i=1}^{m} \nabla\mathcal{L}_{\mathbf{x}_i}. \tag{3.88}$$

Therefore, in traditional GD, one would try to perform GD steps (in vector notation) such as the following:

$$\mathbf{W}^{(l)} = \mathbf{W}^{(l)} - \alpha \frac{\partial \mathcal{L}}{\partial \mathbf{W}^{(l)}}, \tag{3.89.1}$$

$$\mathbf{b}^{(l)} = \mathbf{b}^{(l)} - \alpha \frac{\partial \mathcal{L}}{\partial \mathbf{b}^{(l)}}. \tag{3.89.2}$$

Correspondingly, it is easy to show that the true update of the NN should be the following:

$$\mathbf{W}^{(l)} = \mathbf{W}^{(l)} - \alpha \frac{1}{m}\sum_{i=1}^{m} \frac{\partial \mathcal{L}_{\mathbf{x}_i}}{\partial \mathbf{W}^{(l)}}, \tag{3.90.1}$$

$$\mathbf{b}^{(l)} = \mathbf{b}^{(l)} - \alpha \frac{1}{m}\sum_{i=1}^{m} \frac{\partial \mathcal{L}_{\mathbf{x}_i}}{\partial \mathbf{b}^{(l)}}. \tag{3.90.2}$$

By repeatedly applying this update rule we can "roll down the hill", and hopefully find a minimum of the cost function.

In other words, the NN is designed to be a batch algorithm. All of the training examples are presented to the NN, and the average sum of the cost functions of all training examples is then computed, and this is used to update the weights. Thus, there is only one set of weight updates for each epoch (pass through all the training examples). This means that we only update the weights once for each iteration of the algorithm, which means that the weights are moved in the direction that most of the inputs want them to move, rather than being pulled around by each input individually.

The basic intuition behind FBGD can be illustrated by a hypothetical scenario. A person is stuck in the mountains and is trying to get down (i.e., trying to find the global minimum). There is heavy fog such that visibility is extremely low. Therefore, the path down the mountain is not visible, so he must use local information to find the minimum. He can use the method of FBGD, which involves looking at the steepness of the hill at its current position, and then proceeding in the direction with the steepest descent (i.e., downhill).

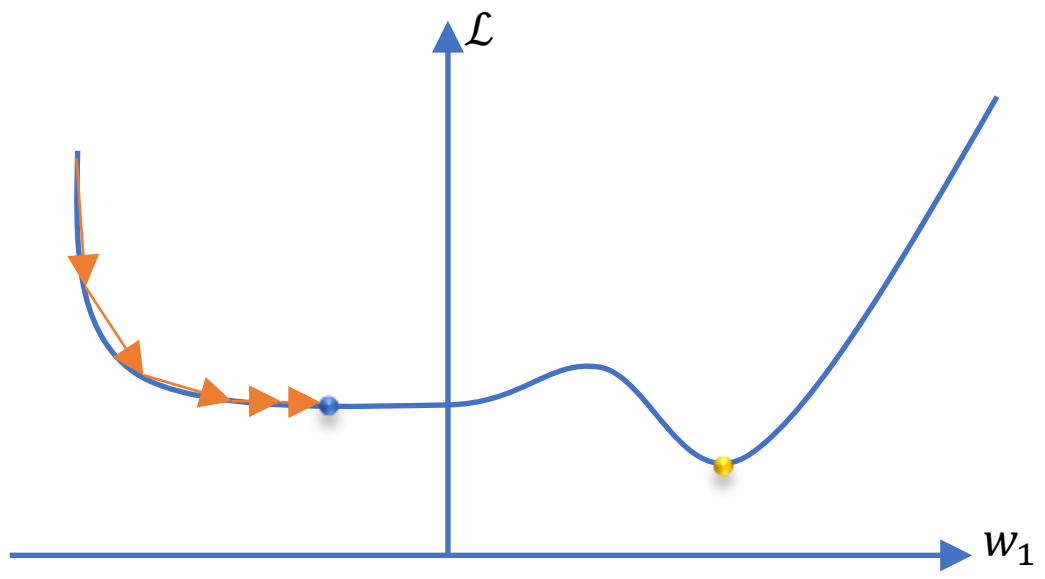

**Figure 3.17.** FBGD is sensitive to saddle points which can lead to premature convergence.

- Envision the mountain landscape as the cost function surface, where the goal is to find the lowest point, representing the minimum of the cost function. The mountains symbolize the high-dimensional space of possible parameter values in a NN model. Each point in this space corresponds to a set of parameters that define the model.
- The person in the mountains represents the optimization algorithm, specifically FBGD in this context.
- The low visibility indicates that the person (algorithm) can't see the entire cost function surface. Instead, he relies on local information obtained from a subset of the landscape. In machine learning, this local information corresponds to the gradient of the cost function at the current parameter values. The gradient provides information about the slope or steepness of the terrain at the current location.
- In FBGD, the person doesn't just evaluate the steepness of the hill at one point but considers the entire landscape. This corresponds to using the entire dataset (a batch of data) to compute the gradient.
- The person assesses the steepest descent by considering the average steepness of the entire landscape in the batch. In FBGD, the algorithm calculates the average gradient of the cost function with respect to the model parameters using all data points in the batch.
- The person then moves downhill based on the average gradient, aiming to reach lower elevations. In FBGD, the algorithm updates the model parameters in the direction opposite to the average gradient, seeking to decrease the average value of the cost function over the entire batch.
- FBGD continues this process iteratively, considering the entire dataset in each iteration.
- Through iterations, FBGD aims to converge to the global minimum of the cost function, taking advantage of the comprehensive information provided by the entire dataset in each step.

Unfortunately, when the number of training inputs is very large this can take a long time, and learning thus occurs slowly. Note that the intermediate activations and derivatives for each training instance would need to be maintained simultaneously over hundreds of thousands of NN nodes. This can be exceedingly large in most practical settings. It is, therefore, impractical to simultaneously run all examples through the network to compute the gradient with respect to the entire data set in one shot. As the training set size grows to billions of examples, the time to take a single gradient step becomes prohibitively long. If we are using FBGD (which looks at all the data), we take steps leading us in the correct direction. But each step is expensive, so we can only take a few steps.

Moreover, for a simple quadratic error surface (cost function surface), FBGD works quite well. But in most cases, our error surface may be a lot more complicated. Let us consider the scenario in Figure 3.17. We have only a single weight, and we use random initialization and FBGD to find its optimal setting. The error surface, however, has a flat region (also known as saddle point in high-dimensional spaces), and if we get unlucky, we might find ourselves getting stuck while performing FBGD.

**Remarks:**

- Since FBGD uses the entire dataset to compute the gradient, it requires storing the entire dataset in memory. This might be a limitation for very large datasets.
- FBGD assumes that the cost function is smooth and continuous. If the cost function has many local minima, FBGD might get stuck in a suboptimal solution.

- FBGD continues iterating until certain convergence criteria are met. This could be a predefined number of iterations or a threshold for the change in the cost function. Commonly, practitioners monitor the cost function and stop training when the change is smaller than a predefined tolerance.
- The key advantage of FBGD is that it utilizes the entire dataset to compute the gradient in each iteration, which leads to a more stable convergence.
- One drawback is that it may be slow when dealing with large datasets or when the model is complex. To address this, variations such as stochastic gradient descent and mini-batch stochastic gradient descent are often used, where the gradient is computed using only a subset of the data in each iteration. These methods provide a compromise between the stability of FBGD and the computational efficiency of GD with respect to individual data points.

### 3.7.2 Stochastic Gradient Descent

Stochastic Gradient Descent (SGD) is an extension of the GD algorithm, sometimes also called iterative or on-line GD. A recurring problem in machine learning is that large training sets are necessary for good generalization, but large training sets are also more computationally expensive. One rarely uses all the points at a single time, and one might stochastically select a point in order to update the parameters to reduce that point-specific loss. In a NN, this process is natural because the simple methods we have introduced so far process the points one at a time in forwards and backwards propagation in order to iteratively minimize point-specific losses. In SGD, all updates to the weights of a NN are performed in point-specific fashion. The point-at-a-time update introduced so far is really a practical approximation of the true update by updating with the use of $\mathcal{L}_{\mathbf{x}_j}$ rather than $\mathcal{L}$:

$$\mathbf{W}^{(l)} = \mathbf{W}^{(l)} - \alpha \frac{\partial \mathcal{L}_{\mathbf{x}_i}}{\partial \mathbf{W}^{(l)}}, \tag{3.91.1}$$

$$\mathbf{b}^{(l)} = \mathbf{b}^{(l)} - \alpha \frac{\partial \mathcal{L}_{\mathbf{x}_i}}{\partial \mathbf{b}^{(l)}}. \tag{3.91.2}$$

Assuming a random ordering of points, each update can be viewed as a probabilistic approximation of the true update. The main advantages of SGD are that it is fast and memory efficient, albeit at the expense of accuracy. The main issue with SGD is that the point-at-a-time approach can sometimes behave in an unstable way, because individual points in the training data might be mislabeled or have other errors.

**Remarks:**

- The SGD algorithm is the sequential algorithm, where the errors are computed and the weights updated after each input. This is not guaranteed to be as efficient in learning, but it is simpler to program when using loops, and it is therefore much more common.
- When we use SGD, we perform $m$ updates per epoch, so we get more updates or steps for a fixed number of epochs. But because of the stochastic or random behavior of using just one data point for each update, the steps we take are noisy. They don't always head in the correct direction. But the larger total number of steps eventually gets us closer to the answer.
- Since each gradient is calculated based on a single training example, the error surface is noisier than in FBGD, which can also have the advantage that SGD can escape shallow local minima more readily. The SGD approach is illustrated by Figure 3.18, where instead of a single static error surface, our error surface is dynamic. As a result, descending on this stochastic surface significantly improves our ability to navigate flat regions.
- In the SGD algorithm, the order of the weight updates can matter, which is why the algorithm includes a suggestion about randomizing the order of the input vectors at each iteration (which is why we want to shuffle the training set for every epoch to prevent cycles). This can significantly improve the speed with which the algorithm learns.
- In SGD implementations, the fixed learning rate $\alpha$ is often replaced by an adaptive learning rate that decreases over time.

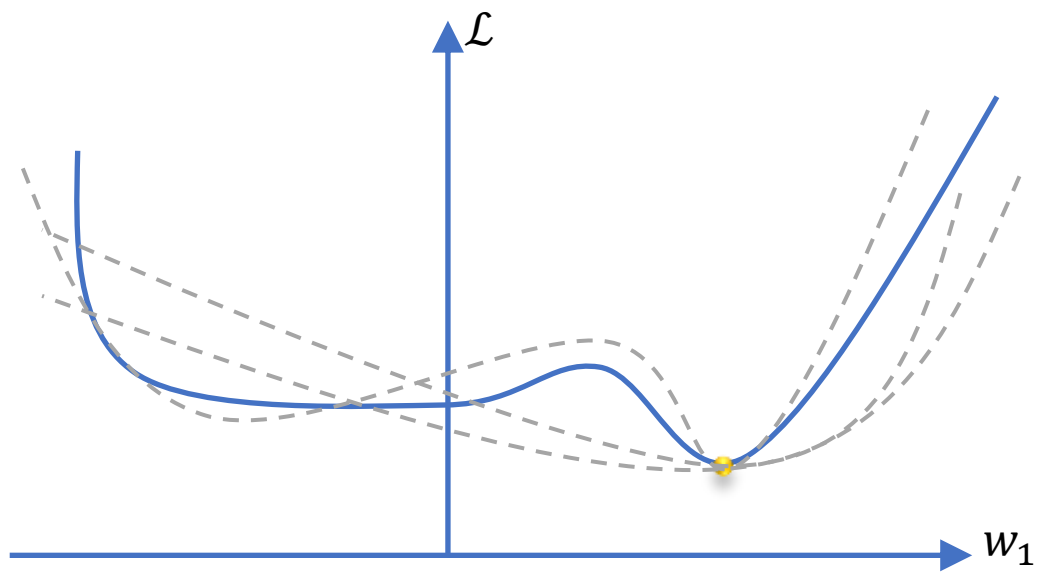

**Figure 3.18.** The SGD error surface fluctuates with respect to the batch error surface, enabling saddle point avoidance.

- Another advantage of SGD is that we can use it for online learning. In online learning, our model is trained on-the-fly as new training data arrives. This is especially useful if we are accumulating large amounts of data—for example, customer data in typical web applications. Using online learning, the system can immediately adapt to changes and the training data can be discarded after updating the model if storage space in an issue.
- The frequent updates allow an easy check on how the model learning is going. (You don't have to wait until all the datasets have been considered.)

The basic intuition behind SGD can be illustrated by a hypothetical scenario.

- Imagine our person in the mountains facing unpredictable weather changes. Sometimes the fog lifts momentarily, allowing for clearer visibility, but it can also return suddenly. This uncertainty corresponds to the stochastic nature of SGD.
- In SGD, visibility is limited even more. The person now relies on brief moments of clarity (randomly selected individual data points) to gather information about the steepness of the terrain.
- The information obtained from a single data point is noisier compared to the average gradient computed over the entire batch. It might provide a good estimate or introduce some randomness.
- When visibility improves, the person quickly assesses the steepness and takes a step downhill. However, this step is now based on information from only one data point.
- SGD updates model parameters more frequently but with a higher level of uncertainty compared to FBGD.
- Due to the stochastic nature, the person takes rapid, less precise steps downhill. This enables the algorithm to explore the landscape more dynamically and escape potential local minima.
- SGD might not consistently move directly toward the global minimum because of the noisy information obtained from individual data points. Instead, it meanders around, allowing for exploration.
- The person adapts to the changing weather conditions by adjusting the step size and direction more frequently. Similarly, SGD adjusts its learning rate dynamically based on the noisiness of individual data points.
- SGD trades off some accuracy for increased efficiency. It may not precisely follow the steepest descent, but the frequent, quick updates make it computationally less expensive, especially with large datasets.

### 3.7.3 Mini-Batch Stochastic Gradient Descent

The insight of mini-batch stochastic gradient descent (MBSGD) is that the gradient is an expectation. The expectation may be approximately estimated using a small set of samples. The MBSGD can be used to speed up learning. The idea is to estimate the gradient $\nabla\mathcal{L}$ by computing $\mathcal{L}_{\mathbf{x}_j}$ for a small sample of randomly chosen training inputs. By averaging over this small sample, it turns out that we can quickly get a good estimate of the true gradient $\nabla\mathcal{L}$, and this helps speed up gradient descent, and thus learning.

Specifically, on each step of the algorithm, we can sample a minibatch of examples $B = \{\mathbf{x}_1, \mathbf{x}_2, \dots, \mathbf{x}_{m'}\}$ drawn uniformly from the training set. The minibatch size $m'$ is typically chosen to be a relatively small number of examples, ranging from one to a few hundred. Crucially, $m'$ is usually held fixed as the training set size $m$ grows. We may fit a training set with billions of examples using updates computed on only a hundred examples. Provided the sample size $m'$ is large enough we expect that the average value of the $\nabla\mathcal{L}_{\mathbf{x}_j}$ will be roughly equal to the average over all $\nabla\mathcal{L}$, that is,

$$\frac{1}{m'}\sum_{j=1}^{m'}\nabla\mathcal{L}_{\mathbf{x}_j} \approx \frac{1}{m}\sum_{i=1}^{m}\nabla\mathcal{L}_i = \nabla\mathcal{L}, \tag{3.92}$$

where the second sum is over the entire set of training data. Swapping sides we get

$$\nabla\mathcal{L} \approx \frac{1}{m'}\sum_{j=1}^{m'}\nabla\mathcal{L}_{\mathbf{x}_j}, \tag{3.93}$$

confirming that we can estimate the overall gradient by computing gradients just for the randomly chosen mini-batch.

To connect this explicitly to learning in NNs, suppose $\mathbf{W}^{(l)}$ and $\mathbf{b}^{(l)}$ denote the weights and biases in our NN. Then MBSGD works by picking out a randomly chosen mini-batch of training inputs, and training with those,

$$\mathbf{W}^{(l)} = \mathbf{W}^{(l)} - \alpha\frac{1}{m'}\sum_{j=1}^{m'}\frac{\partial\mathcal{L}_{\mathbf{x}_j}}{\partial\mathbf{W}^{(l)}}, \tag{3.94.1}$$

$$\mathbf{b}^{(l)} = \mathbf{b}^{(l)} - \alpha\frac{1}{m'}\sum_{j=1}^{m'}\frac{\partial\mathcal{L}_{\mathbf{x}_j}}{\partial\mathbf{b}^{(l)}}, \tag{3.94.2}$$

where the sums are over all the training examples $\mathbf{x}_j$ in the current mini-batch. Then we pick out another randomly chosen mini-batch and train with those. And so on, until we have exhausted the training inputs, which is said to complete an epoch of training. At that point, we start over with a new training epoch.

**Remarks:**

- The idea of a MBSGD is to find some happy middle ground between the FBGD and SGD, by splitting the training set into random batches, estimating the gradient based on one of the subsets of the training set, performing a weight update, and then using the next subset to estimate a new gradient and using that for the weight update, until all of the training set have been used. The training set is then randomly shuffled into new batches and the next iteration takes place. If the batches are small, then there is often a reasonable degree of error in the gradient estimate, and so the optimization has the chance to escape from local minima, albeit at the cost of heading in the wrong direction.
- Incidentally, it's worth noting that conventions vary about the scaling of the cost function and of mini-batch updates to the weights and biases. In (3.87) we scaled the overall cost function by a factor $1/m$. People sometimes omit the $1/m$, summing over the costs of individual training examples instead of averaging. This is particularly useful when the total number of training examples isn't known in advance. This can occur if more training data is being generated in real-time, for instance. And, similarly, the mini-batch update rules (3.94.1) and (3.94.2) sometimes omit the $1/m'$ term out the front of the sums. Conceptually this makes little difference since it's equivalent to rescaling the learning rate $\alpha$. But when doing detailed comparisons of different work it's worth watching out for.
- We can think of MBSGD as being like political polling: it's much easier to sample a small mini-batch than it is to apply GD to the full batch, just as carrying out a poll is easier than running a full election. For example, if we have a training set of size $n = 60{,}000$, and choose a mini-batch size of (say) $m' = 10$, this means we'll get a factor of 6,000 speedup in estimating the gradient! Of course, the estimate won't be perfect – there will be statistical fluctuations – but it doesn't need to be perfect: all we really care about is moving in a general direction that will help decrease $\mathcal{L}$, and that means we don't need an exact computation of the gradient. In practice, MBSGD is a commonly used and powerful technique for learning in NNs.

- The data used in machine learning problems often have a high level of redundancy in terms of the knowledge captured by different training points, and therefore the gradient obtained from a sample of points is usually quite accurate. Furthermore, the weights are often incorrect to such a degree at the beginning of the learning process that even a small sample of points can be used to create an excellent estimate of the gradient. This observation provides a practical foundation for the success of MBSGD, which often exhibits the best trade-off between stability, speed, and memory requirements.
- From an implementation perspective, a mini-batch can be represented by a matrix because each individual training example is an array of inputs and an array of arrays becomes a matrix. Similarly, the weights for a single neuron can be arranged as an array, and we can arrange the weights for all neurons in a layer as a matrix. Computing the inputs to all activation functions for all neurons in the layer for all input examples in the mini-batch is then reduced to a single matrix-matrix multiplication. In other words, when using MBSGD, the outputs of each layer are matrices instead of vectors, and forward propagation requires the multiplication of the weight matrix with the activation matrix (3.84). The same is true for backward propagation in which matrices of gradients are maintained. Hence, mini-batch learning allows us to replace the for-loop over the training samples in SGD by vectorized operations (specifically matrix–matrix multiplications), which can further improve the computational efficiency of our learning algorithm. In this scenario, we also have the option of sending the vectorized computations to GPUs if they are present. The GPUs do a good job of computing a mini-batch in parallel.
- Note that, the SGD is a more extreme version of the MBSGD. The idea is to use just one piece of data to estimate the gradient, and to pick that piece of data uniformly at random from the training set. So, a single input vector is chosen from the training set, and the output and hence the error for that one vector computed, and this is used to estimate the gradient and so update the weights. A new random input vector (which could be the same as the previous one) is then chosen and the process is repeated.
- The advantage over FBGD is that convergence is reached faster via mini-batches because of the more frequent weight updates. Applying MBSGD has also been shown to lead to smoother convergence because the gradient is computed at each step it uses more training examples to compute the gradient.
- As the mini-batch size increases the gradient computed is closer to the "true" gradient of the entire training set. This also gives us the advantage of better computational efficiency. Ideally, each mini-batch trained on should contain an example of each class to reduce the sampling error when estimating the gradient for the entire training set.
- The use of MBSGD introduces a new hyperparameter that must be tuned: the batch size (number of observations in the mini-batch).
- The relationship between how fast our algorithm can learn the model is typically U-shaped (batch size versus training speed). This means that initially as the batch size becomes larger, the training time will decrease. Eventually, we'll see the training time begin to increase when we exceed a certain batch size that is too large.
- The model update frequency is higher than with FBGD but lower than SGD. Therefore, allows for a more robust convergence.
- Finally, don't forget that the number of parameter updates per epoch is reduced when we increase the mini-batch size (where an epoch refers to one full pass of the training data). The number of parameter updates per epoch is just the total number of examples in our training set divided by the mini-batch size. For example, if $20$ arbitrary data points are selected out of 100 training data points, and the MBSGD is applied to the training data, in this case, a total of five weight adjustments are performed to complete the training process for all the data points ($5 = 100/20$). Figure 3.19 shows how the mini-batch scheme selects training data and calculates the weight update. In the FBGD, the number of training cycles of the NN equals an epoch, as shown in Figure 3.19. This makes perfect sense because the batch method utilizes all of the data for one training process. When we have $m$ training data points in total, the number of training processes per epoch is greater than one, which corresponds to the MBSGD, and equal to $m$, which corresponds to the SGD.
- The MBSGD, when it selects an appropriate number of data points, obtains the benefits from both methods: speed and the local-minima avoidance afforded by the SGD and stability from the FBGD. For this reason, it is often utilized in deep learning, which manipulates a significant amount of data.

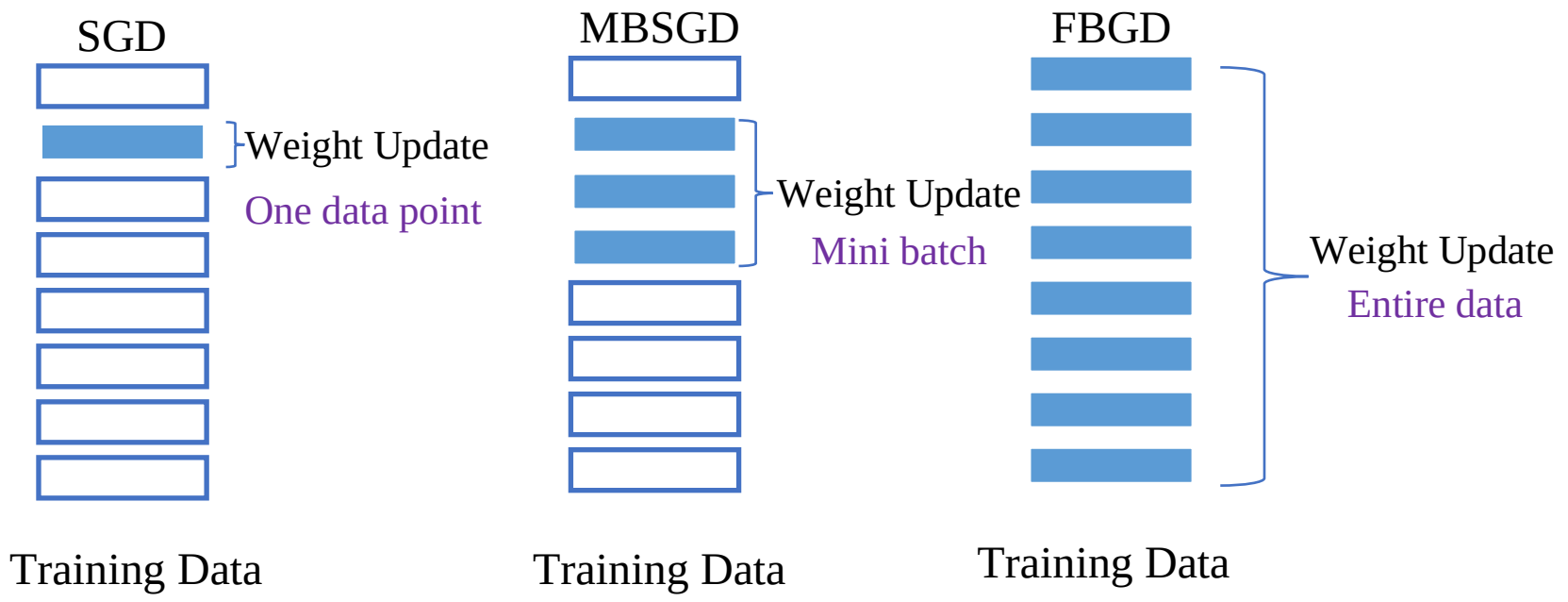

**Figure 3.19.** Comparison of SGD, MBSGD, and FBGD algorithms based on batch size.

In principle (with appropriate tuning), a NN can learn with any minibatch size. In practice, we need to choose a mini-batch size that balances the following: Memory requirements, computational efficiency, and optimization efficiency. Regarding computational efficiency, modern deep learning libraries parallelize learning at the level of mathematical operations such as matrix multiplications and vector operations (additions, element-wise multiplications, etc.). This means that too small a mini-batch size results in poor hardware utilization (especially on GPUs), and too large a mini-batch size can be inefficient—again, we average gradients over all examples in the mini-batch. For performance (this is most important in the case of GPUs), we should use a multiple of 32 for the batch size, or multiples of 16, 8, 4, or 2 if multiples of 32 can't be used. The reason for this is simple: memory access and hardware design are better optimized for operating on arrays with dimensions that are powers of two, compared to other sizes. We should also consider the powers of two when setting our layer sizes. For example, we should use a layer size of 128 over size 125, or size 256 over size 250, and so on. Regarding optimization efficiency, it's sufficient to note that we cannot choose the mini-batch size totally in isolation from the other hyperparameters such as learning rate—a larger mini-batch size means smoother gradients (i.e., more accurate/consistent gradients), which, in conjunction with appropriate tuning, allow for faster learning for a given number of parameter updates. The trade-off of course is that each parameter update will take longer to compute. Using a larger mini-batch size might help our network to learn in some difficult cases, such as for noisy or imbalanced datasets.

In summary, given the dataset contains 1280 points and each mini-batch contains 128 points, we can calculate the number of mini-batches needed to cover the entire dataset: Number of mini-batches = Total number of data points / Mini-batch size= 1280 /128= 10. So, in each epoch of training, there are 10 mini-batches, with each mini-batch containing 128 data points. This setup ensures that every data point is seen exactly once in each epoch.

- The training data is shuffled to introduce randomness, and then divided into mini-batches, each containing 128 data points. This step ensures that the model sees different subsets of data in each epoch, aiding generalization. The sampling is done without replacement, meaning that at the end of an epoch each point data has been seen by the algorithm only once.
- Each mini-batch is passed through the network using forward propagation. The input data is processed layer by layer until the output layer produces predicted values, denoted as $\hat{y}$.
- The predicted values ($\hat{y}$) are compared against the true labels ($y$) using a cost function. This function evaluates how well the model is performing on the current mini-batch.
- With the cost calculated, the model updates its parameters (weights and biases) using GD. The gradients of the parameters with respect to the cost are computed through BP. The adjustments made to the parameters are scaled by the learning rate hyperparameter ($\alpha$).
- After completing training on all mini-batches once (i.e., after 10 cycles in this case), the first epoch of training concludes. The training process then proceeds to the next epoch, where the entire dataset is replenished, shuffled, and divided into mini-batches again.
- Training continues for multiple epochs until the desired number is reached. Each epoch involves iterating through the entire dataset with multiple mini-batches and updating the model parameters accordingly. This iterative process allows the model to gradually improve its performance over successive epochs.

## 3.8 Linear Activation Function

Before concluding this chapter, we must ask an important question: why do NNs work so well? One of the fundamental reasons for the effectiveness of NNs is their ability to introduce non-linearity. Non-linearity is essential because it allows NNs to model complex patterns and relationships within data. Without non-linearity, NNs would be limited to modeling linear functions, severely restricting their expressive power. In this section, we will delve into this point in detail. In the next section, we will explore other reasons for the success of NNs.

The linear AF, also known as the identity AF, is one of the simplest AFs used in NNs. It is defined as: $\sigma_{\text{Linear}}(x) = c\,x$, for $c = 1$, i.e., identity function. The function is infinitely smooth, but all derivatives beyond the second derivative are zero. The range of the function is $[-\infty, \infty]$.

At its most basic level, a NN is a computational graph that performs compositions of simpler functions to provide a more complex function. Much of the power of deep learning arises from the fact that the repeated composition of functions has significant expressive power. However, not all base functions are equally good at achieving this goal. In fact, the nonlinear squashing functions used in NNs are not arbitrarily chosen but are carefully designed because of certain types of properties. For example, imagine a situation in which the identity AF is used in each layer, so that only linear functions are computed. In such a case, the resulting NN is no stronger than a single-layer, linear NN.

**Theorem 3.1:** A multi-layer NN that uses only the identity AF in all its layers reduces to a single-layer NN.

**Proof:**

Consider a NN containing $L$ hidden layers, and therefore contains a total of $(L+1)$ computational layers (including the output layer). The corresponding $(L+1)$ weight matrices between successive layers are denoted by $\mathbf{W}^{(1)} \dots \mathbf{W}^{(L+1)}$. Let $\mathbf{x}$ be the $d$-dimensional column vector corresponding to the input, $\mathbf{a}^{(1)} \dots \mathbf{a}^{(L)}$ be the post-activation column vectors corresponding to the hidden layers, and $\mathbf{O}$ be the $m$-dimensional column vector corresponding to the output.

**Case 1: (without biases)**

We have the following recurrence conditions for multi-layer NNs:

$$\mathbf{a}^{(1)} = \sigma\left(\mathbf{W}^{(1)} \cdot \mathbf{x}\right) = \mathbf{W}^{(1)} \cdot \mathbf{x},$$

$$\mathbf{a}^{(p+1)} = \sigma\left(\mathbf{W}^{(p+1)} \cdot \mathbf{a}^{(p)}\right) = \mathbf{W}^{(p+1)} \cdot \mathbf{a}^{(p)} \quad \forall\, p \in \{1 \dots L-1\},$$

$$\mathbf{O} = \sigma\left(\mathbf{W}^{(L+1)} \cdot \mathbf{a}^{(L)}\right) = \mathbf{W}^{(L+1)} \cdot \mathbf{a}^{(L)}.$$

In all the cases above, the AF $\sigma(\cdot)$ has been set to the identity function. Then, by eliminating the hidden layer variables, we obtain the following:

$$\begin{aligned}
\mathbf{O} &= \sigma\left(\mathbf{W}^{(L+1)} \cdot \mathbf{a}^{(L)}\right) \\
&= \mathbf{W}^{(L+1)} \cdot \mathbf{a}^{(L)} \\
&= \mathbf{W}^{(L+1)} \cdot \left(\mathbf{W}^{(L)} \cdot \mathbf{a}^{(L-1)}\right) \\
&= \mathbf{W}^{(L+1)} \cdot \left(\mathbf{W}^{(L)} \cdot \left(\mathbf{W}^{(L-1)} \cdot \mathbf{a}^{(L-2)}\right)\right) \\
&= \mathbf{W}^{(L+1)} \cdot \left(\mathbf{W}^{(L)} \cdot \left(\mathbf{W}^{(L-1)} \cdot \left(\mathbf{W}^{(L-2)} \cdot \mathbf{a}^{(L-3)}\right)\right)\right) \\
&= \mathbf{W}^{(L+1)} \cdot \left(\mathbf{W}^{(L)} \cdot \left(\mathbf{W}^{(L-1)} \cdot \left(\mathbf{W}^{(L-2)} \cdot \left(\dots \left(\mathbf{W}^{(2)} \cdot \left(\mathbf{W}^{(1)} \cdot \mathbf{x}\right)\right)\right)\right)\right)\right) \\
&= \mathbf{W}^{(L+1)} \cdot \mathbf{W}^{(L)} \cdot \mathbf{W}^{(L-1)} \cdot \dots \cdot \mathbf{W}^{(2)} \cdot \mathbf{W}^{(1)} \cdot \mathbf{x} \\
&= \left(\prod_{i=1}^{L+1} \mathbf{W}^{(i)}\right) \cdot \mathbf{x}
\end{aligned}$$

Let

$$\bar{\mathbf{W}} = \prod_{i=1}^{L+1} \mathbf{W}^{(i)}.$$

Finally, we can express this as a single-layer NN:

$$\mathbf{O} = \bar{\mathbf{W}} \cdot \mathbf{x}$$

**Case 2: (with biases)**

We have the following recurrence conditions for multi-layer NNs:

$$\mathbf{a}^{(1)} = \sigma\left(\mathbf{W}^{(1)} \cdot \mathbf{x} + \mathbf{b}^{(1)}\right) = \mathbf{W}^{(1)} \cdot \mathbf{x} + \mathbf{b}^{(1)},$$

$$\mathbf{a}^{(p+1)} = \sigma\left(\mathbf{W}^{(p+1)} \cdot \mathbf{a}^{(p)} + \mathbf{b}^{(p+1)}\right) = \mathbf{W}^{(p+1)} \cdot \mathbf{a}^{(p)} + \mathbf{b}^{(p+1)} \quad \forall\, p \in \{1 \ldots\ L-1\},$$

$$\mathbf{O} = \sigma\left(\mathbf{W}^{(L+1)} \cdot \mathbf{a}^{(L)} + \mathbf{b}^{(L+1)}\right) = \mathbf{W}^{(L+1)} \cdot \mathbf{a}^{(L)} + \mathbf{b}^{(L+1)}.$$

In all the cases above, the AF $\sigma(\cdot)$ has been set to the identity function. Then, by eliminating the hidden layer variables, we obtain the following:

$$\begin{aligned}
\mathbf{O} &= \sigma\left(\mathbf{W}^{(L+1)} \cdot \mathbf{a}^{(L)} + \mathbf{b}^{(L+1)}\right) \\
&= \mathbf{W}^{(L+1)} \cdot \mathbf{a}^{(L)} + \mathbf{b}^{(L+1)} \\
&= \mathbf{W}^{(L+1)} \cdot \left(\mathbf{W}^{(L)} \cdot \mathbf{a}^{(L-1)} + \mathbf{b}^{(L)}\right) + \mathbf{b}^{(L+1)} \\
&= \mathbf{W}^{(L+1)} \cdot \left(\mathbf{W}^{(L)} \cdot \left(\mathbf{W}^{(L-1)} \cdot \mathbf{a}^{(L-2)} + \mathbf{b}^{(L-1)}\right) + \mathbf{b}^{(L)}\right) + \mathbf{b}^{(L+1)} \\
&= \mathbf{W}^{(L+1)} \cdot \left(\mathbf{W}^{(L)} \cdot \left(\mathbf{W}^{(L-1)} \cdot \left(\mathbf{W}^{(L-2)} \cdot \mathbf{a}^{(L-3)} + \mathbf{b}^{(L-2)}\right) + \mathbf{b}^{(L-1)}\right) + \mathbf{b}^{(L)}\right) + \mathbf{b}^{(L+1)} \\
&= \mathbf{W}^{(L+1)} \cdot \left(\mathbf{W}^{(L)} \cdot \left(\mathbf{W}^{(L-1)} \cdot \left(\mathbf{W}^{(L-2)} \cdot \left(\ldots\left(\mathbf{W}^{(2)} \cdot \left(\mathbf{W}^{(1)} \cdot \mathbf{x} + \mathbf{b}^{(1)}\right) + \mathbf{b}^{(2)}\right)\ldots\right) + \mathbf{b}^{(L-2)}\right) + \mathbf{b}^{(L-1)}\right) + \mathbf{b}^{(L)}\right) \\
&\qquad + \mathbf{b}^{(L+1)} \\
&= \mathbf{b}^{(L+1)} + \mathbf{W}^{(L+1)} \cdot \mathbf{b}^{(L)} + \mathbf{W}^{(L+1)}\mathbf{W}^{(L)} \cdot \mathbf{b}^{(L-1)} + \mathbf{W}^{(L+1)}\mathbf{W}^{(L)}\mathbf{W}^{(L-1)} \cdot \mathbf{b}^{(L-2)} + \cdots + \mathbf{W}^{(L+1)}\mathbf{W}^{(L)}\mathbf{W}^{(L-1)} \cdot \ldots \\
&\qquad \cdot \mathbf{W}^{(2)} \cdot \mathbf{b}^{(1)} + \mathbf{W}^{(L+1)}\mathbf{W}^{(L)}\mathbf{W}^{(L-1)} \cdot \ldots \cdot \mathbf{W}^{(2)} \cdot \mathbf{W}^{(1)} \cdot \mathbf{x} \\
&= \left(\prod_{i=1}^{L+1} \mathbf{W}^{(i)}\right) \cdot \mathbf{x} + \sum_{i=1}^{L} \left(\left(\prod_{j=i+1}^{L+1} \mathbf{W}^{(j)}\right) \cdot \mathbf{b}^{(i)}\right) + \mathbf{b}^{(L+1)}.
\end{aligned}$$

Here, $\prod_{i=1}^{L+1} \mathbf{W}^{(i)}$ represents the product of all weight matrices from $i = 1$ to $L + 1$ (i.e., new weight matrix), and $\sum_{i=1}^{L}\left(\left(\prod_{j=i+1}^{L+1} \mathbf{W}^{(j)}\right) \cdot \mathbf{b}^{(i)}\right) + \mathbf{b}^{(L+1)}$ represents the sum of all the bias terms (i.e., new bias). Let

$$\bar{\mathbf{W}} = \prod_{i=1}^{L+1} \mathbf{W}^{(i)},$$

$$\bar{\mathbf{b}} = \sum_{i=1}^{L} \left(\left(\prod_{j=i+1}^{L+1} \mathbf{W}^{(j)}\right) \cdot \mathbf{b}^{(i)}\right) + \mathbf{b}^{(L+1)}.$$

Finally, we can express this as a single-layer NN:

$$\mathbf{O} = \bar{\mathbf{W}} \cdot \mathbf{x} + \bar{\mathbf{b}}.$$

So, using only the identity AF in all layers collapses the entire NN into a single-layer NN with a combined weight matrix and bias vector.

■

In the special case, where all layers use identity activation and the final layer uses a single output with sign activation for prediction, the multilayer NN reduces to the perceptron. This result is summarized below.

**Lemma 3.1:** Consider a multilayer NN in which all hidden layers use identity activation and the single output node with linear activation that uses the perceptron criterion as the loss function. This NN reduces to the single-layer perceptron.

Since multilayer NNs perform repeated compositions of functions of the form $f(g(\cdot))$, the above result can also be stated as follows:

**Lemma 3.2:** The composition of linear functions is always a linear function. The repeated composition of simple nonlinear functions can be a very complex nonlinear function.

## 3.9 Universal Approximation Theorems (UATs)

Now, returning to the same question: Why do NNs work so well? One of the fundamental reasons for the effectiveness of NNs is their ability to approximate a wide variety of functions, a property known as the UAT. The UAT is a cornerstone of NN theory, providing a theoretical foundation for their ability to model complex and varied functions. This property, combined with the power of non-linear AFs and hierarchical feature learning, explains why NNs are such effective tools for a wide range of machine learning tasks. By leveraging universal approximation, NNs can adapt to and learn from complex data, making them indispensable in modern AI.

The UAT [75-77] states that a FFNN with a single hidden layer with any "squashing" AF (such as the Logistic Sigmoid AF) containing a finite number of neurons can approximate any continuous function on a closed and bounded subset of $\mathbb{R}^n$ to arbitrary accuracy, given a sufficiently large number of neurons in the hidden layer. This makes NNs powerful tools for approximating complex, non-linear relationships in data.

In simpler terms, imagine you have a function, say a curve on a graph, that you want to represent or approximate using a NN. The UAT says that no matter how complicated or wiggly that curve is, you can find a NN architecture (specifically, a single hidden layer NN) that can closely mimic or approximate that curve.

- Let us say you have a function $f(\mathbf{x})$ that takes some input $\mathbf{x}$ and produces an output. This could be anything, like predicting house prices based on features or recognizing handwritten digits.
- The theorem talks about a specific kind of NN - one with a single hidden layer. The hidden layer contains a certain number of neurons, and there are weights and biases associated with each connection between neurons.
- The theorem assumes a specific kind of AF that is non-constant, and bounded. Common choices include Sigmoid or hyperbolic tangent functions.
- With the right combination of weights, biases, and AFs, you can configure the NN so that its output closely approximates your original function $f(\mathbf{x})$. The closeness is determined by a small value $\epsilon$, which you can make as small as you want (meaning you can get as close as you desire).

**Remarks:**

- While the original theorems were first stated in terms of units with AFs that saturate for both very negative and very positive arguments, UATs have also been proved for a wider class of AFs, which includes the now commonly used ReLU [78].
- The UAT means that regardless of what function we are trying to learn, we know that a large FFNN will be able to represent this function. We are not guaranteed, however, that the training algorithm will be able to learn that function. Even if the NN is able to represent the function, learning can fail for two different reasons. First, the optimization algorithm used for training may not be able to find the value of the parameters that corresponds to the desired function. Second, the training algorithm might choose the wrong function as a result of overfitting.

- FFNNs provide a universal system for representing functions in the sense that, given a function, there exists a FFNN that approximates the function. There is no universal procedure for examining a training set of specific examples and choosing a function that will generalize to points not in the training set.
- You will notice that it is unclear exactly how many neurons are needed in the hidden layer for it to be able to approximate any function. This could vary greatly, depending on the function we want it to learn. You just have to experiment by training NNs with different numbers of hidden nodes and then choosing the one that gives the best results.
- Ref [79] provides some bounds on the size of a single-layer NN needed to approximate a broad class of functions.
- It is important to note that the UAT does not provide specific guidance on how to find the weights and biases during the training process. The practical success of NNs in learning complex mappings comes from the training algorithms and techniques used in practice, such as BP and GD.
- By now, you might be thinking that if FFNNs have been around since the late 1960s, why has it taken nearly 60 years for them to take off and be used as widely as they are today? This is because the computing power that was available 60 years ago was nowhere near as powerful as what is available today, nor was the same amount of data that is available now available back then. So, because of the lack of results that NNs were able to achieve back then, they faded into obscurity. Because of this, as well as the UAT, researchers at the time had not looked deeper than into a couple of layers.

Most UATs can be parsed into two classes. The first quantifies the approximation capabilities of NNs with an arbitrary number of artificial neurons ("arbitrary width" case) and the second focuses on the case with an arbitrary number of hidden layers, each containing a limited number of artificial neurons ("arbitrary depth" case). In addition to these two main classes, there are also UATs for NNs with bounded number of hidden layers and a limited number of neurons in each layer ("bounded depth and bounded width" case).

- The study of the expressive power of NNs investigates what class of functions NNs can/cannot represent or approximate. Classical results in this field are mostly focused on shallow NNs. An example of such results is the UATs [75-77], which shows that a NN with fixed depth and arbitrary width can approximate any continuous function on a compact set, up to arbitrary accuracy, if the AF is continuous and nonpolynomial.
- After the advent of deep learning, researchers started to investigate the benefit of depth in the expressive power of NNs, in an attempt to understand the success of DNNs. This has led to interesting results showing the existence of functions that require the NN to be extremely wide for shallow NNs to approximate, while being easily approximated by deep and narrow NNs [80-83]. If the budget number of the neuron is fixed, the deeper NNs have better expression power [84,85].
- In search of a deeper understanding of the depth in NNs, a dual scenario of the classical UAT has also been studied [86-89]. Instead of bounded depth and arbitrary width studied in classical results, the dual problem studies whether universal approximation is possible with a NN of bounded width and arbitrary depth. A very interesting characteristic of this setting is that there exists a critical threshold on the width that allows a NN to be a universal approximator.
- Lu et al. [86] first showed that the ReLU NNs have the UAT for $L^1$ functions from $\mathbb{R}^{d_x}$ to $\mathbb{R}$ if the width is larger than $d_x + 4$, and the UAT disappears if the width is less than $d_x$, where $d_x$ is the input dimension. This implies that the minimum width required for universal approximation lies between $d_x + 1$ and $d_x + 4$. Further research, [87-89], improved the minimum width bound for ReLU NNs. Particularly, Park et al. [90, 91] revealed that the minimum width is $\max(d_x + 1, d_y)$ for the $L^p(\mathbb{R}^{d_x}, \mathbb{R}^{d_y})$ UAT of ReLU NNs, where $d_x, d_y$ are the input and output dimensions, respectively.
- The bounded depth and bounded width case was first studied by Maiorov and Pinkus in 1999 [92]. They showed that there exists an analytic sigmoidal AF such that two hidden layer NNs with bounded number of units in hidden layers are universal approximators.
- Moreover, it was constructively proved in 2018 [93] that single hidden layer FFNNs with the fixed weight 1 and two neurons in the hidden layer can approximate any continuous function on a compact subset of the real line.

One of the earlier and most popular works on approximation properties of NNs is by George Cybenko [75]. The result proves that one-layer NNs with Sigmoid-like AFs can approximate any continuous function on the unit hypercube $I_d = [0,1]^d$. Let us discuss the main mathematics points of this result. Let $\mathbf{W}$ be a $d \times m$ weight matrix, $\mathbf{b}$ an $m \times 1$ bias vector, and $\boldsymbol{\alpha}$ an $m \times 1$ weight vector, which we write as:

$$\mathbf{W} = \begin{pmatrix} \uparrow & \square & \uparrow \\ \mathbf{w}_1 & \dots & \mathbf{w}_m \\ \downarrow & \square & \downarrow \end{pmatrix}, \qquad \mathbf{b} = \begin{pmatrix} b_1 \\ \vdots \\ b_m \end{pmatrix}, \qquad \boldsymbol{\alpha} = \begin{pmatrix} \alpha_1 \\ \vdots \\ \alpha_m \end{pmatrix}. \tag{3.95}$$

Our one-layer NNs take the form:

$$f(\mathbf{x}; \mathbf{W}, \mathbf{b}, \boldsymbol{\alpha}) = \sum_{k=1}^{m} \alpha_k \sigma(\langle \mathbf{x}, \mathbf{w}_k \rangle + b_k), \tag{3.96}$$

where $\alpha_k, b_k \in \mathbb{R}$ and $\mathbf{w}_k \in \mathbb{R}^d$ and $\sigma: \mathbb{R} \to \mathbb{R}$ is a sigmoidal function,

$$\sigma_{\text{sigmoidal}}(z) = \begin{cases} 1, & z \to \infty \\ 0, & z \to -\infty \end{cases}. \tag{3.97}$$

We will initially consider labeling functions of the form

$$y = F(\mathbf{x}), \qquad F \in \mathbf{C}([0,1]^d), \tag{3.98}$$

where $\mathbf{C}([0,1]^d)$ consists of all continuous functions $F: [0,1]^d \to \mathbb{R}$. We have the following theorem.

**Theorem 3.2 (UAT, Cybenko 1989)**: Let $\sigma$ be any continuous sigmoidal function and $I_d$ denotes the $d$-dimensional unit hypercube $I_d = [0,1]^d$. The space of continuous functions on $I_d$ is denoted by $\mathbf{C}(I_d)$. Then finite sums of the form

$$f(\mathbf{x}; \mathbf{W}, \mathbf{b}, \boldsymbol{\alpha}) = \sum_{k=1}^{m} \alpha_k \sigma(\langle \mathbf{x}, \mathbf{w}_k \rangle + b_k), \tag{3.99}$$

are dense in $\mathbf{C}(I_d)$**.** In other words, given any $F \in \mathbf{C}(I_d)$ and an $\epsilon > 0$, there is a one layer NN $f(\mathbf{x}; \mathbf{W}, \mathbf{b}, \boldsymbol{\alpha})$ of the form (3.99) with $m$, $\mathbf{W} \in \mathbb{R}^{d \times m}$, $\mathbf{b} \in \mathbb{R}^m$, and $\boldsymbol{\alpha} \in \mathbb{R}^m$ depending on $d$, $F$, and $\epsilon$, for which

$$|f(\mathbf{x}; \mathbf{W}, \mathbf{b}, \boldsymbol{\alpha}) - F(\mathbf{x})| < \epsilon, \qquad \text{for all } \mathbf{x} \in [0,1]^d. \tag{3.100}$$

Cybenko's result relied upon the Kolmogorov-Arnold representation theorem (or superposition theorem). This theorem states that every multivariate continuous function can be represented as a superposition of continuous functions of one variable.

**Theorem 3.3 (Kolmogorov-Arnold Representation Theorem)**: Any continuous function $f: [0,1]^n \to \mathbb{R}$ can be written as

$$f(\mathbf{x}) = f(x_1, x_2, \dots, x_n) = \sum_{q=0}^{2n} \Phi_q \left( \sum_{p=1}^{n} \phi_{q,p}(x_p) \right), \tag{3.101}$$

where $\phi_{q,p}: [0,1] \to \mathbb{R}$ and $\Phi_q: \mathbb{R} \to \mathbb{R}$.

## 3.10 Intuitive and Illustrative Approach of UAT

The universality theorems are well-known by people who use NNs. Most of the explanations available are quite technical. For instance, one of the original papers proving the result did so using the Hahn-Banach theorem, the Riesz representation theorem, and some Fourier analysis. If you are a mathematician the argument is not difficult to follow, but it is not so easy for most people. That's a pity since the underlying reasons for universality are simple and beautiful. In the following, we will see the simple and intuitive illustrative of UAT.

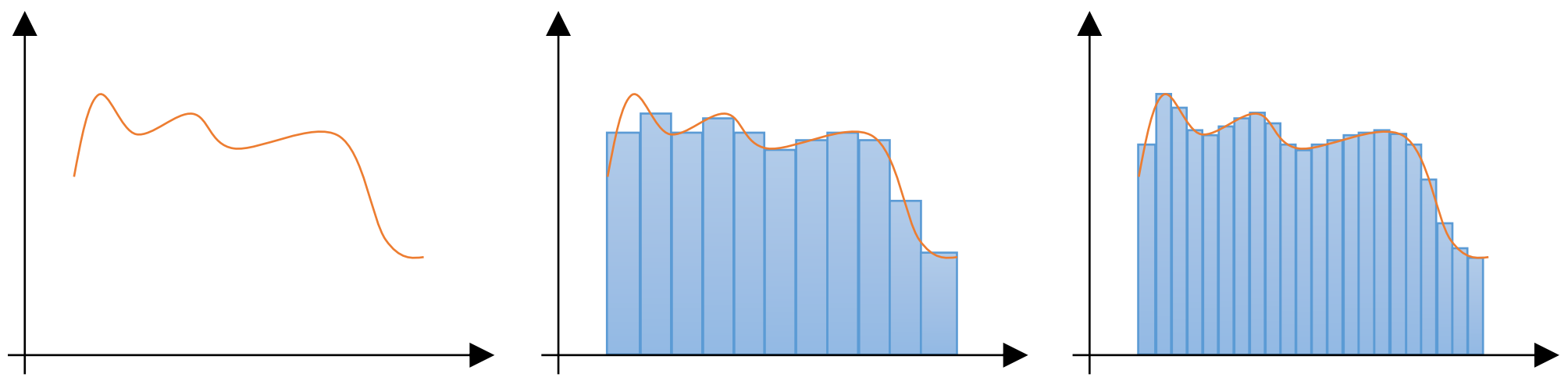

**Figure 3.20.** This figure demonstrates the process of approximating a complex function by breaking it into multiple smaller parts, each represented by a simpler rectangular function. The approach involves dividing the function's domain into a series of subintervals and approximating the function within each subinterval using a rectangular segment. The height of each rectangle corresponds to the function's value within that subinterval. The method aims to approximate the relationship between the input variable $x$ and the output variable $y$ as closely as possible to the true function. By increasing the number of rectangular functions used, the approximation becomes more accurate, effectively capturing the nuances of the original function.

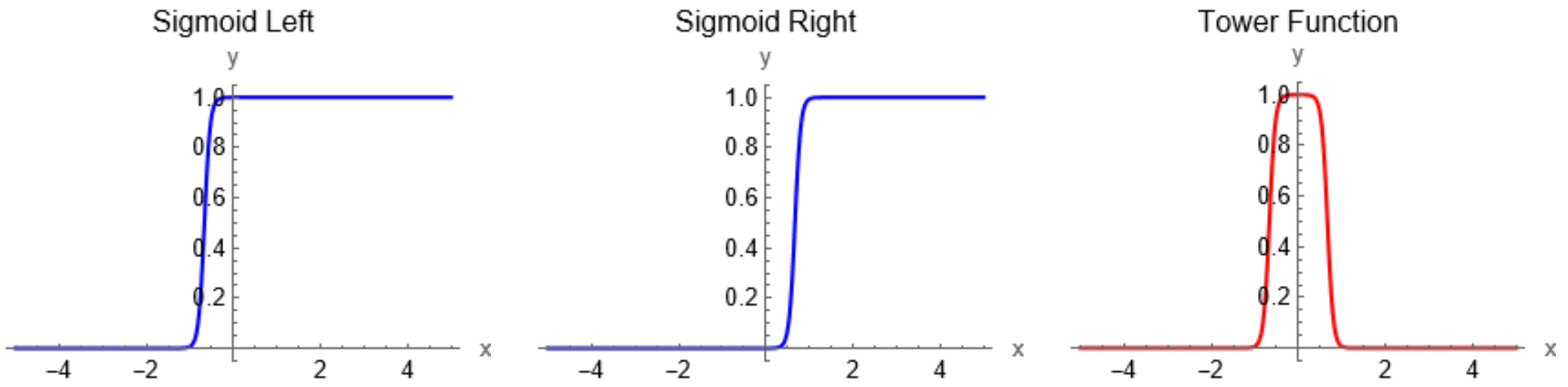

**Figure 3.21.** This set of plots illustrates the construction and visualization of the "rectangular towers function," which is derived by subtracting two sigmoid functions with different shifts and peaks. Left panel: This plot illustrates the left-shifted sigmoid function, defined by a steep slope and a shift parameter. The left sigmoid, shown in blue, peaks earlier along the $x$-axis. Middle panel: This plot illustrates the right-shifted sigmoid function, also defined by a steep slope and a shift parameter. The right sigmoid, shown in blue, peaks later along the $x$-axis. Right panel: This plot shows the rectangular towers function, which is created by subtracting the right-shifted sigmoid from the left-shifted sigmoid. The resulting function forms a shape reminiscent of rectangular towers, depicted in red. The $x$-axis represents the input variable $x$, while the $y$-axis represents the function value.

### 3.10.1 One-Dimensional Case

- To understand why the UATs are true, let us start by understanding how to construct a NN that approximates a function with just one input, $x$, and one output $y$. For example, Figure 3.20 represents the true relationship between input and output. Once we have understood this special case it is actually pretty easy to extend to functions with many inputs and many outputs.
- To solve this problem, we will break this function into multiple smaller parts so that each part is represented by a simpler function. By combining the series of smaller functions (rectangular) we can approximate the relation between $x$ and $y$ as close to the true relationship possible. The more functions that we choose in this method the better will be the approximation.
- A typical shifted Logistic-Sigmoid AF equation is given as follows.

$$\sigma_{\text{Logistic}}(z) = \frac{1}{1 + e^{\alpha x + \beta}}. \tag{3.102}$$

With an increase in $\alpha$ the function becomes more steeper like the step function. More positive values of $\beta$ move the curves towards the left from the original curve. Hence, by changing these values we can create different versions of Sigmoid which we can superimpose on each other to obtain tower-like structures. Now, let us take two Sigmoid functions having a very steep slope, Figure 3.21, and notice that they have different places at which they peak. The left Sigmoid peaks just before zero and the right sigmoid peaks just after zero. If we subtract these two functions the net effect is going to be a rectangular (towers) output (local function).

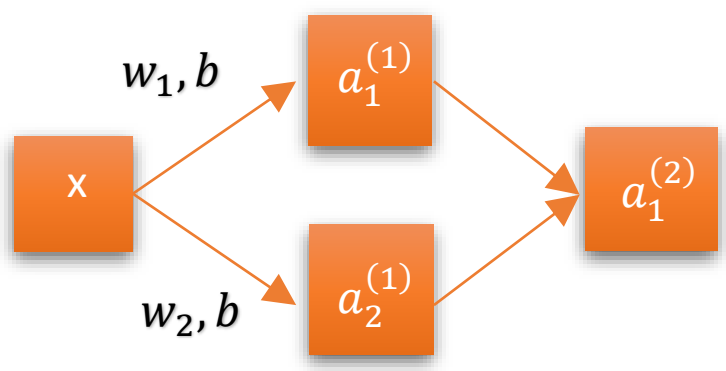

**Figure 3.22.** Connectivity needed to obtain one tower-like structure.

By appropriate selection of parameters, this local function can be made of any width, centered on any point, and with abrupt or smooth edges. If we can get the series of these towers, then we can approximate any true function between input and output.

- Can we come up with a NN to represent this operation of subtracting one Sigmoid function from another? If we have an input $x$ and it is passed through the two Sigmoid neurons and the output from these two neurons is combined in another neuron with correct weights, then we will get our tower. i.e., parameters are tuned in sigmoidal AFs to create such approximation towers, Figure 3.22. Now you can see that we have our building block ready which is a connection of three sigmoid neurons. If we can construct many such building blocks and add all of them up, we can approximate any complex true relationship between input and output. Theoretically, there is no limit to the accuracy of NNs as per this interpretation.

### 3.10.2 Two-Dimensional Case

- In two dimensions we need to localize a function about a point on the plane. A pair of Sigmoid functions define a ridge in the input plane. However, the ridge is of infinite length, see Figure 3.23 (top right).
- By adding another pair of Sigmoid functions, another ridge, at an angle with the first, can be produced. A two-dimensional local bump is created where the ridges intersect, Figure 3.23 (bottom left).
- The infinite tails fanning out from the bump can be removed by applying another Sigmoid, with an appropriate threshold, Figure 3.23 (bottom right).
- The corresponding network consists of two input neurons, a hidden layer of four neurons, and an output layer of one neuron. The sigmoidal function is applied at both the hidden and the output layers. The weights and the thresholds determine the orientation, position, and shape of the bump.
- The same method of construction applies in higher dimensions. The number of hidden neurons required to define a hypercube centered at a point in $\mathbb{R}^n$ is $2n$, Figure 3.24.

Mathematically, we consider input connects to hidden layer 1 ($H_1$), hidden layer 1 to hidden layer 2 ($H_2$), and hidden layer 2 to output. Therefore:

$$O_l = \sum_{k\in H_2} w_{kl} \cdot \sigma_{\text{Logistic}}\left(\sum_{i\in H_1} w_{ki} \cdot \sigma_{\text{Logistic}}\left(\sum_j w_{ij}x_j + b_j\right) + b_k\right) + b_l. \tag{3.103}$$

Recall that the output neurons a linear computing element so that only two $\sigma_{\text{Logistic}}$ occur in formula (3.103), due to the two nonlinear hidden layers. For ease in later analysis, let us rewrite this formula as

$$O_l = \sum_{k\in H_2} w_{kl} \cdot \sigma_{\text{Logistic}}(S_k + b_k) + b_l, \tag{3.104.1}$$

where,

$$S_k = \sum_{i\in H_1} w_{ki} \cdot \sigma_{\text{Logistic}}\left(\sum_j w_{ij}x_j + b_j\right). \tag{3.104.2}$$

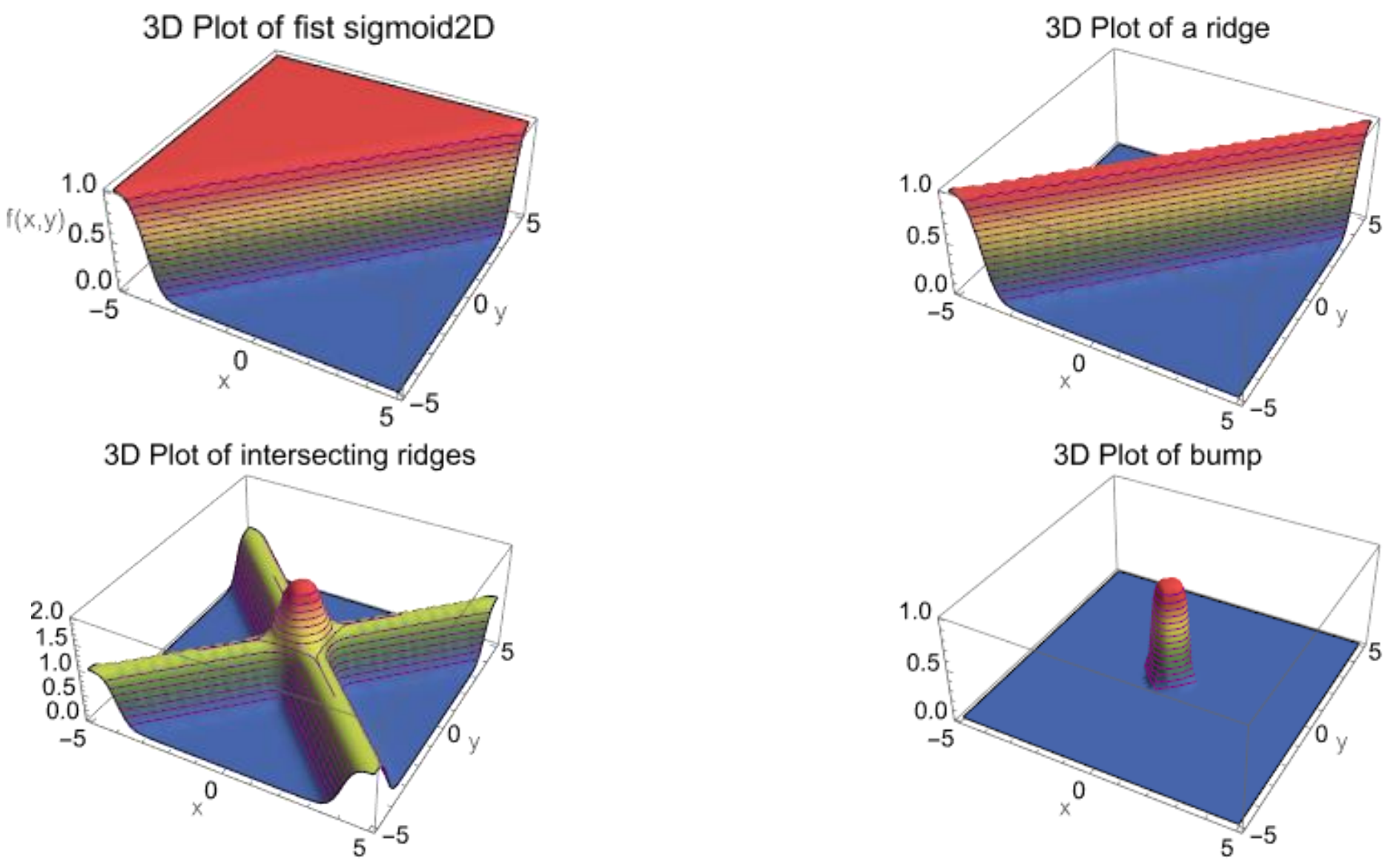

**Figure 3.23.** This set of 3D plots illustrates the process of generating and combining two-dimensional sigmoid functions to create ridged surfaces and localized bumps. The approach involves summing sigmoid functions to obtain surfaces with specific characteristics and further refining these surfaces to achieve desired features. Top left panel: This plot shows the surface generated by a single two-dimensional sigmoid function. The sigmoid function produces a smooth, S-shaped surface. Top right panel: This plot shows a ridged surface generated by subtracting two two-dimensional sigmoid functions. The resulting surface features a distinct ridge, demonstrating the effect of combining sigmoid functions with opposing parameters. Bottom left panel: This plot depicts the intersection of two ridged surfaces, created by adding two perpendicular ridged surfaces generated by different pairs of sigmoid functions. The intersection of these ridges forms a complex surface with intersecting ridges. Bottom right panel: This plot illustrates a localized bump achieved by applying a two-dimensional sigmoid function to the sum of the two ridged surfaces from the previous plots. The sigmoid function depresses local minima and saddles to zero while pushing the central maximum toward 1, resulting in a prominent bump.

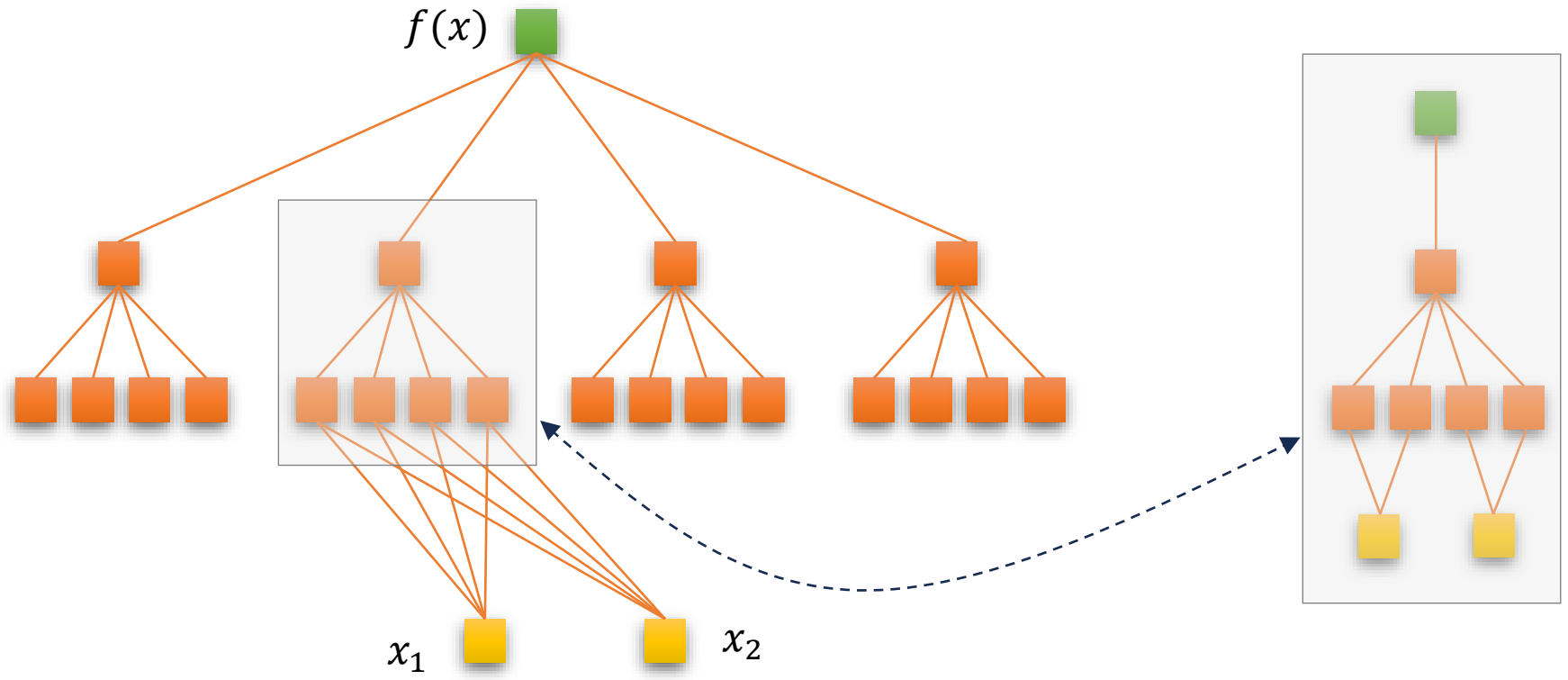

**Figure 3.24.** Network structure for function approximation. The boxed assembly of units defines a local function. The output unit provides a linear combination of all the local function. The boxed area illustrates the necessary connectivity to obtain one bump in the function approximation. Each unit within this assembly contributes to the formation of a localized bump in the function being approximated. Add four more neurons to hidden layer 1, and one more neurn to hidden layer 2, for each additional bump. Each bump represents a local function that can capture fine details and variations in the target function. By increasing the number of bumps, the network can better approximate regions with higher complexity or more intricate patterns. More bumps allow the network to approximate the target function with higher resolution, capturing smaller and more localized features that a network with fewer bumps might miss. With more bumps, the network can minimize approximation error by closely fitting the function's true values at multiple points, leading to a more accurate overall approximation.

**Lemma 3.3**: The special form of (3.104) is actually a general representation for quite arbitrary surfaces.

**Proof:**

To prove that (3.104) is a reasonable representation for surfaces we first point out that surfaces may be approximated by adding up a series of "bumps" that are appropriately placed. An example of this occurs in familiar Fourier analysis, where wave trains of suitable frequency and amplitude are added together to approximate curves (or surfaces). Each half period of each wave of fixed wavelength is a "bump," and one adds all the bumps together to form the approximant. Let us now see how (3.104) may be interpreted as adding together bumps of specified heights and positions. First, consider $S_k$ which is a sum of $\sigma_{\text{Logistic}}(\,)$ functions. In Figure 3.23 (top left), we plot an example of such a $\sigma_{\text{Logistic}}(\,)$ function for the case of two inputs. The orientation of this sigmoidal surface is determined by $w_{ij}$, the position by $b_j$ and height by $w_{ki}$. Now consider another $\sigma_{\text{Logistic}}(\,)$ function that occurs in $S_k$. The $b_j$ of the second $\sigma_{\text{Logistic}}(\,)$ function is chosen to displace it from the first, the $w_{ij}$ is chosen so that it has the same orientation as the first, and $w_{ki}$ is chosen to have the opposite sign to the first. These two $\sigma_{\text{Logistic}}(\,)$ functions occur in $S_k$ and so to determine their contribution to $S_k$, we sum them together and plot the result in Figure 3.23 (top right). The result is a ridged surface. Since our goal is to obtain localized bumps, we select another pair of $\sigma_{\text{Logistic}}(\,)$ functions in $S_k$, add them together to get a ridged surface perpendicular to the first ridged surface, and then add the two perpendicular ridged surfaces together to see the contribution to $S_k$. The result is plotted in Figure 3.23 (bottom left). We see that this almost worked, in so much as one obtains a local maxima by this procedure. However, there are also saddle-like configurations at the corners which corrupt the bump we were trying to obtain. Note that one way to fix this is to take $\sigma_{\text{Logistic}}(S_k + b_k)$ which will, if $b_k$ is chosen appropriately, depress the local minima and saddles to zero while simultaneously sending the central maximum towards 1. The result is plotted in Figure 3.23 (bottom right) and is the sought-after bump.

Furthermore, note that the necessary $\sigma_{\text{Logistic}}(\,)$ function is supplied by (3.104). Therefore (3.104) is a procedure to obtain localized bumps of arbitrary height and position. For two inputs, the $k$-th bump is obtained by using four $\sigma_{\text{Logistic}}(\,)$ functions from $S_k$ (two $\sigma_{\text{Logistic}}(\,)$ functions for each ridged surface and two ridged surfaces per bump) and then taking $\sigma_{\text{Logistic}}(\,)$ of the result in (3.104.1). The height of the $k$-th bump is determined by $w_{lk}$ in (3.104.1) and the $k$ bumps are added together by that equation as well. The general network architecture which corresponds to the above procedure of adding two $\sigma_{\text{Logistic}}(\,)$ functions together to form a ridge, two perpendicular ridges together to form a pseudo-bump, and the final $\sigma_{\text{Logistic}}(\,)$ to form the final bump is represented in Figure 3.24. To obtain any number of bumps one adds more neurons to the hidden layers by repeatedly using the connectivity of boxed assembly in Figure 3.24 as a template (i.e., four neurons per bump in hidden layer 1, and one neuron per bump in hidden layer 2).

■

One never needs more than two layers, or any other type of connectivity than that already schematically specified by boxed assembly in Figure 3.24. The accuracy of the approximation depends on the number of bumps, which in turn is specified by the number of neurons per layer.

### 3.10.3 Mathematical Details

Let us give more mathematical details. To approximate a continuous function $f$ over a compact subset $K$ of $\mathbb{R}^d$ using a step function, you can use a method called piecewise constant approximation. The idea is to divide $K$ into smaller subintervals or cells and assign a constant value to $f$ on each cell. The values are usually chosen to be the average or some representative value of $f$ on each cell. The general steps are:

- Divide the compact set $K$ into smaller subintervals or cells. This can be done in various ways, such as using a regular grid or adapting the partition based on the behavior of the function. For example, if $K$ is a square in $\mathbb{R}^2$, you might create a grid of smaller squares within $K$.
- Choose a representative value for $f$ on each cell. Common choices include the average value, the value at the center of the cell, or any other representative value that makes sense in the context of the function.

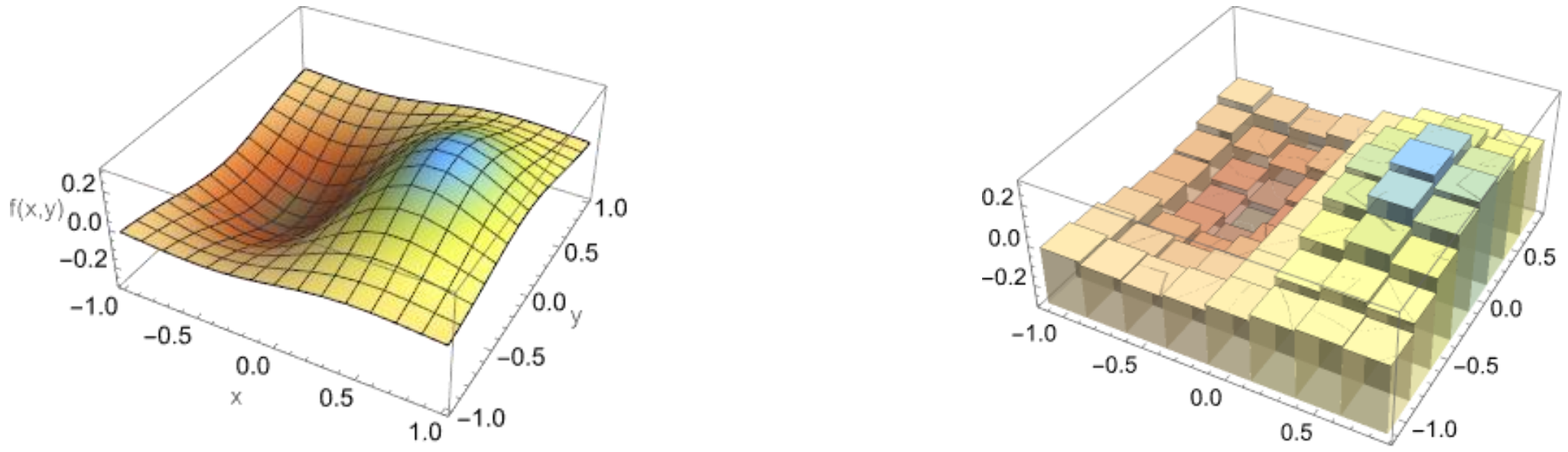

**Figure 3.25.** Left panel: This figure shows a three-dimensional plot of the function $f(x,y) = x\, e^{-2x^2-2y^2}$ which demonstrates how the function varies over the x-y plane. Right panel: This figure presents a three-dimensional plot of the function $\hat{f}(x,y)$, which is realized by the network. It shows how the network approximates the function over a grid, resulting in a piecewise-constant surface. The surface appears to be a discretized or piecewise constant approximation.

- Define the step function $g$ by assigning the chosen values on each cell. The step function $g$ is a piecewise constant function that approximates $f$ over $K$.
- Calculate the difference between the step function $g$ and the original function $f$. Common measures include maximum absolute error, $\max_{\text{over } K} |f(\mathbf{x}) - g(\mathbf{x})|$.

This process is a form of piecewise constant interpolation, and the accuracy of the approximation depends on the size of the cells and the choice of representative values. As the size of the cells decreases, and the number of cells increases, the step function converges to the original function in a sense. For example, in 2D case:

- Divide the region $K$ into smaller rectangles using a regular grid. For example, if $K$ is a square $[a,b] \times [c,d]$, create a grid of smaller rectangles within this square.
- Choose a representative value for $f(x,y)$ on each rectangle. Again, you can use the average value or another representative value. For a rectangle with corners $(x_1,y_1)$, $(x_2,y_1)$, $(x_2,y_2)$, $(x_1,y_2)$, you might use the average value: $c_{ij} = \frac{1}{(x_2-x_1)(y_2-y_1)} \int_{x_1}^{x_2} \int_{y_1}^{y_2} f(x,y)dydx$.
- The step function $g(x,y)$ is defined as $g(x,y) = c_{ij}$ for $(x,y)$ in the $i$-th row and $j$-th column of rectangles.
- Measure the error using the chosen metric. For example, you might calculate the maximum absolute error, $\max_{\text{over } K} |f(x,y) - g(x,y)|$.

Step functions are often used to discuss approximation properties of FFNNs [94]. Assume that the FFNN hidden layer neurons have the following step AF:

$$\text{step}(x) = \begin{cases} 1, & x \geq 0 \\ 0, & x < 0 \end{cases}. \tag{3.105}$$

Let us consider the hypercube in $d$-dimensional space with intervals $[a_i, b_i)$ for each dimension $i$, $I = [a_1,b_1) \times [a_2,b_2) \times \ \ldots \times [a_d,b_d)$. Hyper-step function is

$$\text{hyperstep}_I(\mathbf{x}) = \begin{cases} 1, & \mathbf{x} \in I \\ 0, & \text{otherwise} \end{cases}, \qquad \mathbf{x} = (x_1, x_2, \ldots, x_d)^T. \tag{3.106}$$

If our purpose is to approximate a continuous function $f$ over a compact subset $K$ of $\mathbb{R}^d$ then this task can be easily fulfilled by splitting the domain $K$ into a set of $I_1, I_2, \ldots, I_n$ hypercubes and writing

$$\hat{f}(\mathbf{x}) = \sum_{i=1}^{n} f(\mathbf{z}_i)\, \text{hyperstep}_{I_i}(\mathbf{x}), \tag{3.107}$$

where $\mathbf{z}_i$ is the center of $I_i$, the $i$-th hypercube. It is simple to observe that this expression can be interpreted in terms of NNs as shown in Figure 3.24 (with step AF). For example, let us assume that $f(x,y) = x\, e^{-2x^2-2y^2}$, see Figure 3.25 (left panel), and the domain $K$, where we wish to model $f$, is the square $-1 \leq x \leq 1$ and $-1 \leq y \leq 1$. Moreover, suppose that $n = 64$ and $I_1, \ldots, I_{64}$ are the $0.25 \times 0.25$ squares defined by a uniform grid on $K$. Figure 3.25 (right panel) shows the output of the NN that is defined in (3.107).

# CHAPTER 4

# CHALLENGES IN NEURAL NETWORK OPTIMIZATION

NNs have revolutionized various fields, from image recognition to natural language processing, by enabling machines to learn complex patterns and make decisions with human-like accuracy. However, despite their remarkable capabilities, training NNs effectively remains a challenging task. This chapter delves into the myriad challenges encountered during the optimization process of NNs, exploring key concepts such as AF saturation, vanishing and exploding gradients, weight initialization methods, non-zero centered AFs, feature scaling techniques, and normalization methods like Batch Normalization (BN) and Layer Normalization (LN).

One of the fundamental challenges in NN optimization is AF saturation. AFs play a crucial role in introducing non-linearity to the model, enabling it to learn complex patterns. However, certain AFs, such as the Sigmoid or hyperbolic tangent (Tanh) functions, tend to saturate when the input values are too large or too small, leading to vanishing gradients and hampering the learning process.

The vanishing and exploding gradients problem is another significant hurdle faced during NN training. In DNNs, gradients can diminish exponentially or explode during BP, making it challenging to update the weights effectively. This phenomenon hinders the convergence of the model and affects its ability to generalize to unseen data.

While non-zero centered AFs like ReLU and its variants indeed alleviate the vanishing gradient problem (VGP), they can introduce another issue known as "zig-zag" updates. These updates occur when the neuron's output oscillates between positive and negative values, causing the GD updates to zig-zag back and forth. This oscillatory behavior can potentially slow down the learning process, as the network struggles to converge towards the optimal solution.

Effective weight initialization is crucial for mitigating the issues of vanishing and exploding gradients. Common weight initialization techniques, such as random initialization, aim to set initial weights to small values to prevent saturation and maintain stable gradients during training.

Kaiming or He initialization is a popular weight initialization method designed specifically for DNNs. It initializes the weights of each layer based on the number of input units, effectively addressing the vanishing and exploding gradients problem and promoting faster convergence.

Xavier or Glorot initialization is another widely used technique for weight initialization. It sets the initial weights using a uniform or normal distribution, scaled based on the number of input and output units, thus ensuring stable gradients and facilitating smoother training.

Feature scaling techniques, including standardization, normalization, and whitening, are essential for preprocessing input data and enhancing the convergence of NNs. These methods ensure that input features are on similar scales, preventing certain features from dominating the learning process.

BN and LN are techniques that normalize the activations of each layer, effectively stabilizing the learning process and accelerating convergence. These techniques mitigate the effects of internal covariate shift and facilitate smoother optimization of NNs.

In summary, addressing the challenges in NN optimization, such as AF saturation, vanishing and exploding gradients, weight initialization, and normalization techniques, is crucial for training efficient and robust models capable of tackling complex tasks effectively. This chapter will explore these challenges in detail, along with strategies to overcome them, thereby providing insights into the intricate process of optimizing NNs.

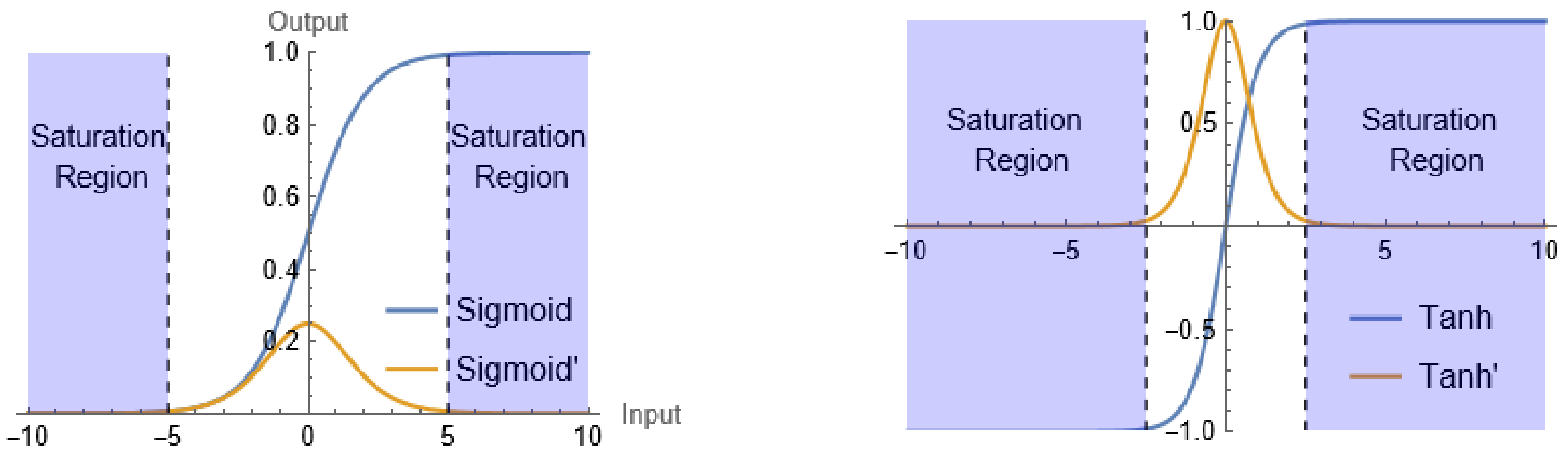

**Figure 4.1.** Comparing saturation regions. Left panel: Sigmoid AF and its derivative. Right panel: Tanh AF and its derivative.

## 4.1 Activation Function Saturation

AFs used in NN hidden and output layers are usually chosen to be non-linear and bounded. Non-linearity of the hidden units allows a NN to approximate any non-linear mapping between inputs and outputs provided that enough neurons are used in the hidden layer. Upper and lower bounds ensure that the signal does not grow uncontrollably as it propagates from one layer to the next. Functions with a sigmoidal curve such as the Logistic function and the hyperbolic tangent are often used as AFs. Sigmoidal functions exhibit linear behavior in the active range determined by the function slope, and saturate (approach asymptotes) for large positive and negative input values. Thus, if the magnitude of the input signal lies outside the active range of a Sigmoid, the output signal will be close to an asymptotic value. The net input signal of a hidden unit (pre-activation) is a weighted sum of inputs from the previous NN layer:

$$\text{net} = \sum_{i=1}^{I} w_i z_i, \tag{4.1}$$

where $I$ is the number of incoming connections, including the bias, $w_i$ is the weight of the $i$-th connection, and $z_i$ is the $i$-th input signal. Now, suppose that $w_i$ for a certain $i$ is set to a very large value, causing net to always lie outside of the sigmoid's active range. Clearly, the output will be very close to either the lower or the upper asymptotic value of the sigmoid, depending on the net's sign. In such case, we speak of hidden unit saturation: the phenomena when a NN unit is reduced to a binary state, predominantly outputting values close to the asymptotic ends of the AF.

**Definition (AF Saturation)**: AF saturation refers to a situation where the output of an AF becomes very insensitive to small changes in the input (the change in the output becomes very small or negligible), see Figure 4.1.

**Definition (AF Saturation)**: AF saturation occurs when bounded AFs are used in the hidden layer. A node is considered to be saturated when the majority of its outputs are close to the bounds of its AF.

In other words, once a certain input range is reached, further changes in input do not cause significant changes in output, leading to vanishing gradients. This can be problematic during the training of a NN because the gradients during BPbecome extremely small, which makes it difficult for the model to learn and update its weights effectively. Looking at the Logistic AF, Figure 4.1, you can see that when inputs become large (negative or positive), the function saturates at 0 or 1. Similarly, the Tanh function saturates at $-1$ or 1. In these regions, the gradients (derivatives) of the AFs approach zero. When the gradients are close to zero, the updates to the weights during the training process become very small, leading to slow or stalled learning.

Sigmoidal AFs saturate when the value of net as defined in (4.1) lies outside the function's active range. As follows from (4.1), the value of net is determined by three variables:

1. The total number of inputs and biases, $I$.
2. The input values, $z_i$, for $1 \leq i \leq I$.
3. The weight values, $w_i$, for $1 \leq i \leq I$.

The number of inputs is defined by the problem at hand and the NN architecture chosen for that problem. This book is concerned with static architectures only, thus $I$ should be treated as a constant rather than a variable. The values of $z_i$ are not modifiable by the training algorithm. The only variable from (4.1) that is directly modifiable by the training algorithm is $w_i$. It is important to note that $w$ is usually adjusted by the training algorithm such that the error of the NN is minimised, and both the error surface and the function approximation modelled by the NN depend on the AF chosen. Thus, choosing an appropriate AF is crucial to algorithm's success, and will have a significant effect on the saturation levels.

There are a few ways to evaluate the saturation of AFs during training:

- Monitor the gradients during BP. If you notice that the gradients are consistently very small, it may indicate saturation.
- Check how much the weights are updating during training. If the weights are not changing significantly, it could be a sign of saturation.
- Examine the output distributions of neurons in different layers. If the outputs are consistently pushed to extreme values (close to 0 or 1 for Sigmoid, $-1$ or 1 for Tanh), it suggests saturation.

A number of methods for evaluating the saturation of the AFs within a NN have been discussed in the literature. An obvious way to check for the presence of saturation is to examine the AF outputs in the non-linear layers of a NN. If the activation outputs on the given data set are concentrated around the asymptotic ends, then saturation is present. Raw activation output data can be analysed graphically to approximate the extent of saturation. Glorot and Bengio [95] graphed the AF output ranges over algorithm iterations to observe the level of saturation. In both [95] and [96] frequency distributions were used to compare the spread of AF outputs across a selection of different AFs. This provides a quick and convenient way to examine the saturation of a NN and even provides a means of comparing the saturation of different NNs.

A numeric measure of saturation, which allows for automated comparisons and statistical testing would be ideal. Many numerical measures of saturation are available in the literature [97-99]. One of the numerical saturation measure is [97]

$$\varrho_h = \frac{\sum_{p=1}^{P}\sum_{j=1}^{H}\left|\mathrm{net}_{jp}\right|}{PH}, \tag{4.2}$$

where $h$ is the hidden layer number, $P$ is the number of examples, $H$ is the size of hidden layer $h$, and $\mathrm{net}_{jp}$ is the net input to hidden node $j$ on pattern $p$. Note that, $\varrho_h$ cannot be used to compare saturation between two different AFs. The saturation measure $\varrho_h$ effectively measures the growth of the input signal magnitudes. Saturation occurs when the value of net lies outside the active range of the AF. Thus, monitoring the growth of average net gives an indication of the extent of saturation present in the given hidden layer.

It's crucial to note that AF saturation can manifest due to various factors, including:

- Large weights can contribute to AF saturation in NNs because they can cause the input to the AF to become large. When the input becomes large, the Sigmoid function approaches its asymptotes, leading to saturation.
- Biases also affect the input to the AF. Large biases can shift the input towards the saturation regions.
- If the input data to the NN tends to be large or has a wide distribution, it can lead to saturation of the AF.

Some techniques to address AF saturation include:

- Properly initializing the weights can help prevent saturation problems. Techniques like He initialization and Xavier initialization are commonly used (We will explore weight initialization in depth in Sections 4.3, 4.4).
- Choosing appropriate AFs can also help. Until 2010 (paper [95] by Glorot and Bengio) most people had assumed that if mother nature had chosen to use roughly Sigmoid AFs in biological neurons, they must be an excellent choice. But it turns out that other AFs behave much better in deep NNs—in particular, the ReLU AF, mostly because it does not saturate for positive values (and because it is fast to compute).

- Normalizing the inputs of each layer helps in mitigating internal covariate shift, which can alleviate saturation issues. (We will provide a comprehensive examination of BN in Section 4.7).

## 4.2 Vanishing and Exploding Gradients Problems

The Vanishing Gradient Problem, VGP, is a challenge that arises during the training of DNNs, particularly in the context of gradient-based optimization methods like BP. The BP algorithm works by going from the output layer to the input layer, propagating the error gradient along the way. Once the algorithm has computed the gradient of the cost function with respect to each parameter in the NN, it uses these gradients to update each parameter with a gradient descent step. Unfortunately, gradients often get smaller and smaller as the algorithm progresses down to the lower layers. As a result, the gradient descent update leaves the lower layers' connection weights virtually unchanged, and training never converges to a good solution.

**Definition (VGP)**: During the algorithm, gradients are computed and used to update the weights of the NN. The VGP occurs when the gradients become extremely small as they are backpropagated to the earlier layers of the NN. Consequently, the weights of the neurons in these layers are updated very little, if at all, which makes it difficult for these layers to learn meaningful representations from the data [55].

In some cases, the opposite can happen: the gradients can grow bigger and bigger until layers get insanely large weight updates and the algorithm diverges.

**Definition (Exploding Gradient Problem)**: The exploding gradients problem is the counterpart to the VGP and occurs when the gradients during the training of a NN become extremely large. This can lead to numerical instability during the training process, causing the weights to update to very large values. As a result, the parameters of the model can become so large that they may overflow, causing the NN to produce unreliable and incorrect predictions.

More generally, DNNs suffer from unstable gradients; different layers may learn at widely different speeds.

Let us consider the mathematical details of a simple NN with just a single neuron in each layer to gain insight into why the VGP occurs. In this case, we will look at a NN with three hidden layers, with the following architecture:

1. Input Layer: This layer receives the input features, $x = a^{(0)}$, where $x$ is the input.
2. Hidden Layer 1: This layer has a single neuron, $z^{(1)} = w^{(1)} \cdot a^{(0)} + b^{(1)}$, $a^{(1)} = \sigma_{\text{Logistic}}(z^{(1)})$.
3. Hidden Layer 2: Another layer with a single neuron, $z^{(2)} = w^{(2)} \cdot a^{(1)} + b^{(2)}$, $a^{(2)} = \sigma_{\text{Logistic}}(z^{(2)})$.
4. Hidden Layer 3: A third layer with a single neuron, $z^{(3)} = w^{(3)} \cdot a^{(2)} + b^{(3)}$, $a^{(3)} = \sigma_{\text{Logistic}}(z^{(3)})$.
5. Output Layer: This layer produces the final output, $z^{(4)} = w^{(4)} \cdot a^{(3)} + b^{(4)}, a^{(4)} = \sigma_{\text{Logistic}}(z^{(4)})$.

where $w^{(i)}$ and $b^{(i)}$ represent the weight and bias of layer $i$, $(i = 1, \dots, 4)$. $\sigma_{\text{Logistic}}$ is the Logistic Sigmoid AF. The neural architecture is illustrated in Figure 4.2.

Understanding the VGP in a simple NN helps highlight the challenges that arise as networks become deeper, making it harder for gradients to effectively reach the early layers and update their weights. During the training of the NN using BP algorithm, we use a loss function $\mathcal{L}$ to measure the error between the predicted output $a^{(4)}$ and the true output $y$. The goal is to minimize this loss by adjusting the weights and biases. The weights are adjusted based on the gradients of the loss function with respect to these weights. The gradients are calculated by applying the chain rule during BP. The derivative of the loss with respect to the weights in the first hidden layer is denoted as $\frac{\partial \mathcal{L}}{\partial w^{(1)}}$. According to the chain rule, we have [55]:

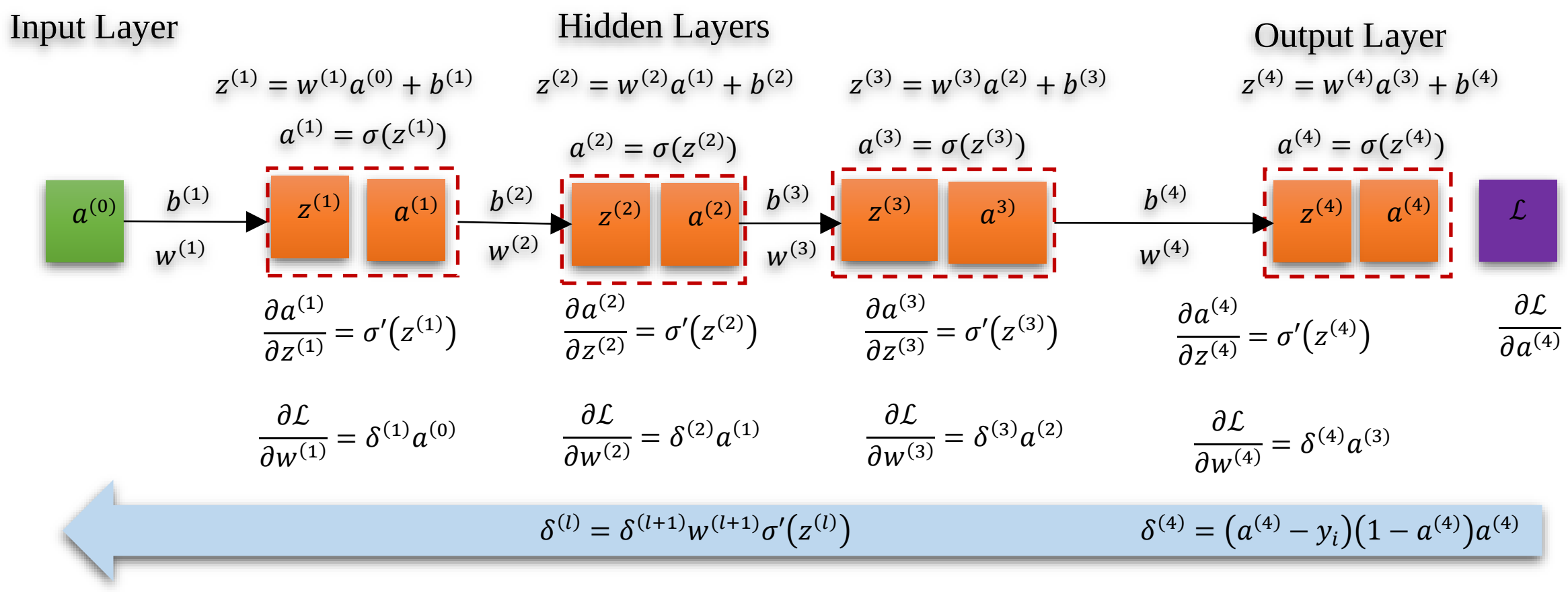

**Figure 4.2.** A simple NN (three hidden layers).

$$\begin{aligned}
\frac{\partial \mathcal{L}}{\partial w^{(1)}} &= \frac{\partial \mathcal{L}}{\partial a^{(4)}} \frac{\partial a^{(4)}}{\partial z^{(4)}} \frac{\partial z^{(4)}}{\partial a^{(3)}} \frac{\partial a^{(3)}}{\partial z^{(3)}} \frac{\partial z^{(3)}}{\partial a^{(2)}} \frac{\partial a^{(2)}}{\partial z^{(2)}} \frac{\partial z^{(2)}}{\partial a^{(1)}} \frac{\partial a^{(1)}}{\partial z^{(1)}} \frac{\partial z^{(1)}}{\partial w^{(1)}} \\
&= \frac{\partial \mathcal{L}}{\partial a^{(4)}} \left( \frac{\partial z^{(4)}}{\partial a^{(3)}} \frac{\partial a^{(4)}}{\partial z^{(4)}} \right) \left( \frac{\partial z^{(3)}}{\partial a^{(2)}} \frac{\partial a^{(3)}}{\partial z^{(3)}} \right) \left( \frac{\partial z^{(2)}}{\partial a^{(1)}} \frac{\partial a^{(2)}}{\partial z^{(2)}} \right) \left( \frac{\partial a^{(1)}}{\partial z^{(1)}} \right) \left( \frac{\partial z^{(1)}}{\partial w^{(1)}} \right) \\
&= \frac{\partial \mathcal{L}}{\partial a^{(4)}} \left( w^{(4)} \sigma'_{\text{Logistic}}(z^{(4)}) \right) \cdot \left( w^{(3)} \sigma'_{\text{Logistic}}(z^{(3)}) \right) \cdot \left( w^{(2)} \sigma'_{\text{Logistic}}(z^{(2)}) \right) \cdot \sigma'_{\text{Logistic}}(z^{(1)}) \cdot x
\end{aligned} \tag{4.3}$$

Now, let us focus on the terms involving the Sigmoid AF ($\sigma_{\text{Logistic}}$):

$$\frac{\partial a^{(i)}}{\partial z^{(i)}} = \sigma'_{\text{Logistic}}(z^{(i)}) = \sigma_{\text{Logistic}}(z^{(i)}) \left(1 - \sigma_{\text{Logistic}}(z^{(i)})\right) = a^{(i)}(1 - a^{(i)}). \tag{4.4}$$

The derivative reaches a maximum at $\sigma'_{\text{Logistic}}(0) = 1/4$, i.e., $z^{(i)} = 0$, and it decreases as $a^{(i)} = \sigma_{\text{Logistic}}(z^{(i)})$ approaches 0 or 1. If we use standard approach to initializing the weights in the network, then we will choose the weights using a Gaussian with mean 0 and standard deviation 1. So, the weights will usually satisfy $|w^{(i)}| < 1$. Putting these observations together, we see that the terms $w^{(i)} \cdot \sigma'_{\text{Logistic}}(z^{(i)})$ will usually satisfy $\left|w^{(i)} \cdot \sigma'_{\text{Logistic}}(z^{(i)})\right| < 1/4$.

If we have many layers (as in the case of DNNs), these small derivatives get multiplied together during BP, leading to vanishing gradients. As a result, the gradient updates for the weights in the early layers become extremely small, and these weights are adjusted very slowly, if at all. Hence, these neurons fail to learn meaningful representations of the input data, and the network struggles to capture complex patterns and relationships in the data.

$$w_{\text{new}}^{(1)} = w_{\text{old}}^{(1)} - \alpha \frac{\partial \mathcal{L}}{\partial w^{(1)}}. \tag{4.5}$$

In other words, the basic idea underlying the VGP, is that the gradients of the loss function with respect to the weights in different layers is very different in magnitude. For example, the (magnitude of the) partial derivative with respect to a weight in the final layer might be $10^6$ times the magnitude of the partial derivative with respect to a weight in the first layer (using 10 layers). Therefore, a gradient descent update that changes a parameter in the final layer by 0.2 units will (negligibly) change the weight in the first layer by $2 \times 10^{-6}$.

It is easy to generalize the above argument to cases in which NN contains $n$ layers (not only three), Figure 4.3. Using the chain rule for partial derivatives, auxiliary variable, for arbitrary layer $l$ becomes,

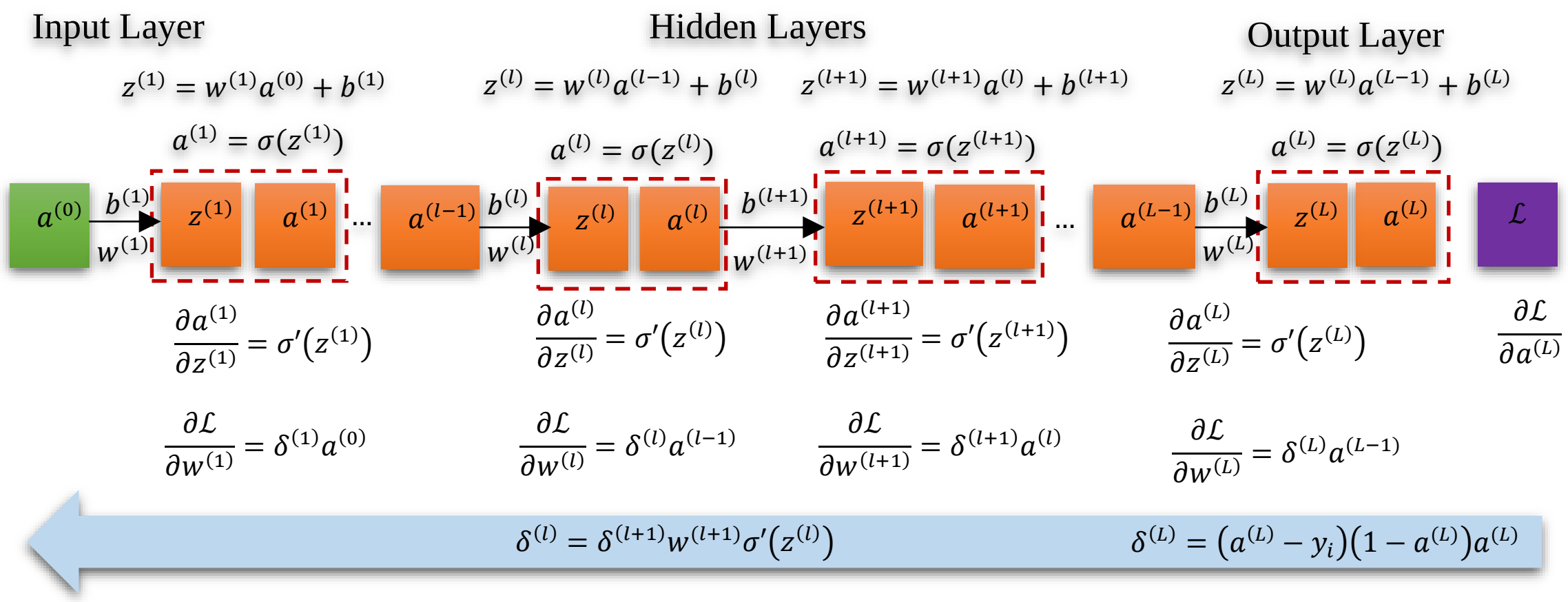

**Figure 4.3.** A simple NN (single neuron per layer).

$$\delta^{(l)} = \frac{\partial \mathcal{L}}{\partial z^{(l)}} = \frac{\partial \mathcal{L}}{\partial z^{(l+1)}} \frac{\partial z^{(l+1)}}{\partial a^{(l)}} \frac{\partial a^{(l)}}{\partial z^{(l)}}. \tag{4.6}$$

Using the definition of the auxiliary variable and the forward propagation, this leads to

$$\begin{aligned}\delta^{(l)} &= \frac{\partial \mathcal{L}}{\partial z^{(l)}} \\ &= \frac{\partial \mathcal{L}}{\partial z^{(l+1)}} \frac{\partial z^{(l+1)}}{\partial a^{(l)}} \frac{\partial a^{(l)}}{\partial z^{(l)}} \\ &= \frac{\partial \mathcal{L}}{\partial z^{(l+1)}} \frac{\partial z^{(l+1)}}{\partial a^{(l)}} \frac{\partial \sigma\left(z^{(l)}\right)}{\partial z^{(l)}} \\ &= \delta^{(l+1)} w^{(l+1)} \sigma'\left(z^{(l)}\right) \\ &= \delta^{(l+1)} w^{(l+1)} \left(1 - a^{(l)}\right) a^{(l)}.\end{aligned} \tag{4.7}$$

Since the absolute value of $w^{(l+1)}$ is expected to be 1 (assume that the weights are initialized from a standard normal distribution.), it follows that each weight update will (typically) cause the value of $\frac{\partial \mathcal{L}}{\partial z^{(l)}} = \delta^{(l)}$ to be less than 0.25 that of $\frac{\partial \mathcal{L}}{\partial z^{(l+1)}} = \delta^{(l+1)}$. Therefore, after moving by about $r$ layers, this value will typically be less than $(0.25)^r$. Just to get an idea of the magnitude of this drop, if we set $r = 10$, then the gradient update magnitudes drop to $9.53674 \times 10^{-7} \approx 10^{-6}$ of their original values!

It is also noteworthy that even though we have shown the derivatives with respect to the hidden layers, the derivatives with respect to weights are direct multiples of those with respect to hidden layers; the trends will be similar.

$$\frac{\partial \mathcal{L}}{\partial w^{(l)}} = \delta^{(l)} a^{(l-1)}. \tag{4.8}$$

Therefore, when BP algorithm is used, the earlier layers will receive very small updates compared to the later layers.

Although we have used an oversimplified example here with only one node in each layer, it is easy to generalize the argument to cases in which multiple nodes are available in each layer. Of course, this is an informal argument, not a rigorous proof that the VGP will occur. There are several possible escape clauses. In particular, we might wonder whether the weights $w_i$ could grow during training. If they do, it is possible the terms $w^{(i)} \cdot \sigma'_{\text{Logistic}}\left(z^{(i)}\right)$ in the product will no longer satisfy $\left|w^{(i)} \cdot \sigma'_{\text{Logistic}}\left(z^{(i)}\right)\right| < 1/4$. Indeed, if the terms get large enough – greater than 1 – then we will no longer have a VGP. Instead, the gradient will actually grow exponentially as we move backward through the layers. Instead of a VGP, we will have an exploding gradient problem.

There are two steps to getting an exploding gradient. First, we choose all the weights in the NN to be large, say $w^{(1)} = w^{(2)} = w^{(3)} = w^{(4)} = 100$. Second, we will choose the biases so that the $\sigma'_{\text{Logistic}}\left(z^{(i)}\right)$ terms are not too small. That is actually pretty easy to do: all we need do is choose the biases to ensure that the weighted input to each neuron is $z^{(i)} = 0$ (and so $\sigma'_{\text{Logistic}}\left(z^{(i)}\right) = 1/4$). So, for instance, we want $z^{(1)} = w^{(1)} \cdot x + b^{(1)} = 0$. We can achieve this by setting $b^{(1)} = -100x$. We can use the same idea to select the other biases. When we do this, we see that all the terms $w^{(i)} \cdot \sigma'_{\text{Logistic}}\left(z^{(i)}\right)$ are equal to $100 \cdot 1/4 = 25$. With these choices we get an exploding gradient.

The VGP is most commonly associated with deep FFNNs and RNNs, although it can affect other types of NNs as well. In summary, it occurs for the following reasons:

- Some AFs, such as the Sigmoid or Tanh (these functions squash their input to a very small range 0 to 1 or $-1$ to 1), have derivatives that are close to zero for most of their input range. This means that when gradients are backpropagated through layers that use these functions, the gradients can become very small, causing the VGP.
- The VGP is particularly pronounced in DNNs with many layers, where the gradients have to pass through multiple nonlinear AFs. As gradients are backpropagated through multiple layers, their values can decrease exponentially, making it challenging to update the weights of the early layers effectively.
- Poor weight initialization strategies can exacerbate the VGP. If the initial weights are too small, it can make the gradients even smaller as they propagate through the NN.

The VGP was empirically observed long ago, and it was one of the reasons DNNs were mostly abandoned in the early 2000s. It was not clear what caused the gradients to be so unstable when training a DNN, but some light was shed in a 2010 paper by Xavier Glorot and Yoshua Bengio [95]. In short, the authors showed that the combination of the popular Logistic Sigmoid AF and the weight initialization technique that was most popular at the time, the variance of the outputs of each layer is much greater than the variance of its inputs. Going forward in the NN, the variance keeps increasing after each layer until the AF saturates at the top layers. This saturation is actually made worse by the fact that the Logistic function has a mean of 0.5, not 0. Looking at the Logistic AF, you can see that when inputs become large (negative or positive), the function saturates at 0 or 1, with a derivative extremely close to 0.

To mitigate the VGP, several techniques have been developed, including:

- Initialization techniques like Xavier/Glorot initialization [95] and He initialization [100] set the initial weights in a way that helps prevent the VGP to some extent.
- ReLU and its variants, such as Leaky ReLU (LReLU) and Parametric ReLU (PReLU), have become popular because they have non-zero gradients for a large input range, reducing the VGP.
- BN [101] normalizes the inputs to each layer, making the optimization process more stable and mitigating the VGP.
- Gradient Clipping (GC): This technique involves setting a threshold on the gradient values during BP algorithm to prevent them from becoming too small or too large.
- Skip connections and residual networks (ResNets) allow gradients to flow more directly through the NN, helping to alleviate the VGP in very DNNs.

These techniques, individually or in combination, help address the VGP and enable the training of deeper and more complex NNs, which has been crucial in the advancement of deep learning.

## 4.3 Weight Initialization Taxonomy

When training NNs, the weights of the connections between neurons are initialized with certain values before the training process begins. The choice of these initial weights plays a crucial role in determining the success of the training process. Here are some key points related to the impact of initial weights:

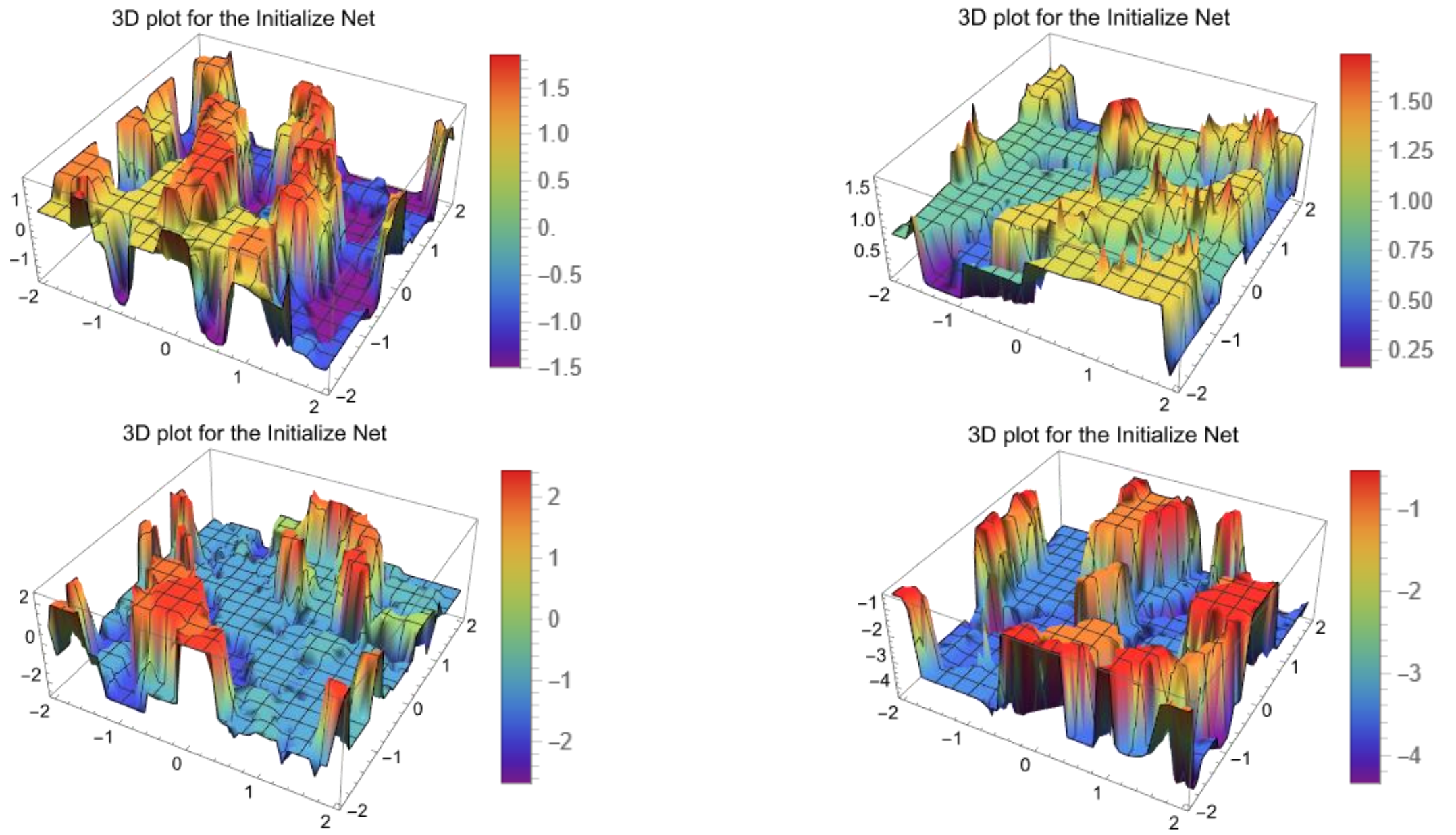

**Figure 4.4.** Visualization of four randomly initialized NNs' responses to input vectors. Each 3D Plot represents the output of one initialized network, highlighting the diversity in responses among networks due to varied initialization parameters.

- The choice of initial weights can affect whether the optimization algorithm converges at all. Some poor choices may lead to numerical instability or difficulties in finding a suitable minimum for the cost function, causing the algorithm to fail to converge.
- Even if the algorithm converges, the initial weights influence how quickly it reaches convergence. Well-chosen initial weights can lead to faster convergence, reducing the time and computational resources required for training.
- The choice of initial weights can influence whether the algorithm converges to a local minimum or a global minimum in the cost function landscape. Different initializations might result in the algorithm getting stuck in different local minima, affecting the overall performance of the model.
- The choice of initial weights can mitigate issues like vanishing or exploding gradients. If weights are too small, gradients may vanish during BP, hindering learning. Conversely, if weights are too large, gradients may explode, making it challenging to find an optimal solution.

Initial weights have this impact because the loss surface of a DNN is a complex, high-dimensional, and non-convex landscape with many local minima. So, the point where the weights start on this loss surface determines the local minimum to which they converge, the better the initialization, the better the model.

An illustrative example was constructed using a fundamental NN designed to accept 2-dimensional vector inputs and generate 1-dimensional vector outputs. As depicted in Figure 4.4, we showcase four representatives, independently initialized instances of this network using 3D plots. Each 3D plot represents the output of one initialized network for varying input values. Due to the randomness in the initialization process, each network has different starting points (weights and biases). Consequently, the responses of these networks to input vectors exhibit diversity, visually depicted in the distinct shapes and patterns of the 3D plots. By observing how the initialized networks respond to inputs, you gain insights into how initialization influences the learning process. Networks that start with different initial parameters may converge to different local minima during training, resulting in varied learned representations.

The field of NN optimization is indeed complex and not fully understood, making the development of optimal initialization methods a challenging task. Here are some additional insights into the difficulties and ongoing research in this area:

- The optimization landscape of deep NNs is highly non-convex, and its theoretical understanding is limited. While some initialization methods aim to achieve desirable properties in the initial state, it's challenging to predict how these properties will evolve during the optimization process.
- NN optimization is a non-stationary process, meaning the properties of the optimization landscape can change during training. What might be a good initialization strategy at the beginning may not necessarily remain effective as training progresses.
- Achieving good optimization performance (convergence, speed, avoiding local minima) during training does not guarantee good generalization to unseen data. Some initialization strategies that expedite optimization may lead to overfitting or poor generalization, and vice versa.
- The choice of AFs in a NN also interacts with the initialization strategy. Some AFs may be more sensitive to certain weight initializations, and this complex interplay adds another layer of difficulty in designing universally effective strategies.
- A successful initialization strategy for one type of NN or task may not necessarily generalize well to other scenarios. This lack of transferability complicates the development of universally applicable initialization methods.

Perhaps the only property known with complete certainty is that the initial parameters need to "break symmetry" between different units. If two hidden units with the same AF have the same initial parameters and are connected to the same inputs, they will always update in the same way during training, leading to symmetric behavior, see Section 4.4.1 for details. This symmetry can persist throughout training, limiting the capacity of the network to learn diverse features and representations. To overcome this issue, it is crucial to initialize the parameters of the NN in a way that breaks the symmetry among hidden units. This is typically done by using random initialization methods. Random initialization from a high-entropy distribution over a high-dimensional space is computationally cheaper and unlikely to assign any units to compute the same function as each other. By starting each unit with different initial parameters, the learning algorithm can update them differently during training, allowing the network to capture a broader range of features and patterns in the data. This ensures that different hidden units respond differently to the same input patterns during forward propagation. During BP, having distinct gradients for each unit allows the model to update their parameters independently, avoiding a situation where all units are consistently updated in the same way.

Both Gaussian and uniform distributions are commonly used for weight initialization. The choice between them depends on the specific characteristics of the problem and the AFs used. Gaussian distributions are often preferred when using AFs like Tanh, while uniform distributions may be suitable for ReLU activations.

The scale of the initial weight distribution determines the range of values from which the initial weights are drawn, and it plays a crucial role in how well the network learns during optimization. The scale of the initial weights influences the magnitude of gradients during BP. If the weights are too small, the gradients can become vanishingly small, leading to slow convergence or difficulty in learning. On the other hand, if the weights are too large, it may result in exploding gradients, causing numerical instability during training. These competing factors determine the ideal initial scale of the weights.

Researchers have proposed weight initialization strategies achieving either or some of the following:

- Faster convergence
- Avoiding getting stuck in false minima
- Depth independent performance
- Learning meaningful representations from data
- Better accuracy or error rate

It is possible to define the taxonomy of weight initialization strategies based on pre-training. The broad categories for iterative training mechanisms are initialization techniques without pre-training and initialization techniques with pre-training, as shown in Figure 4.5. Each of these categories and subcategories might have different techniques and methods associated with them, and the choice often depends on the specific characteristics of the task and the available data.

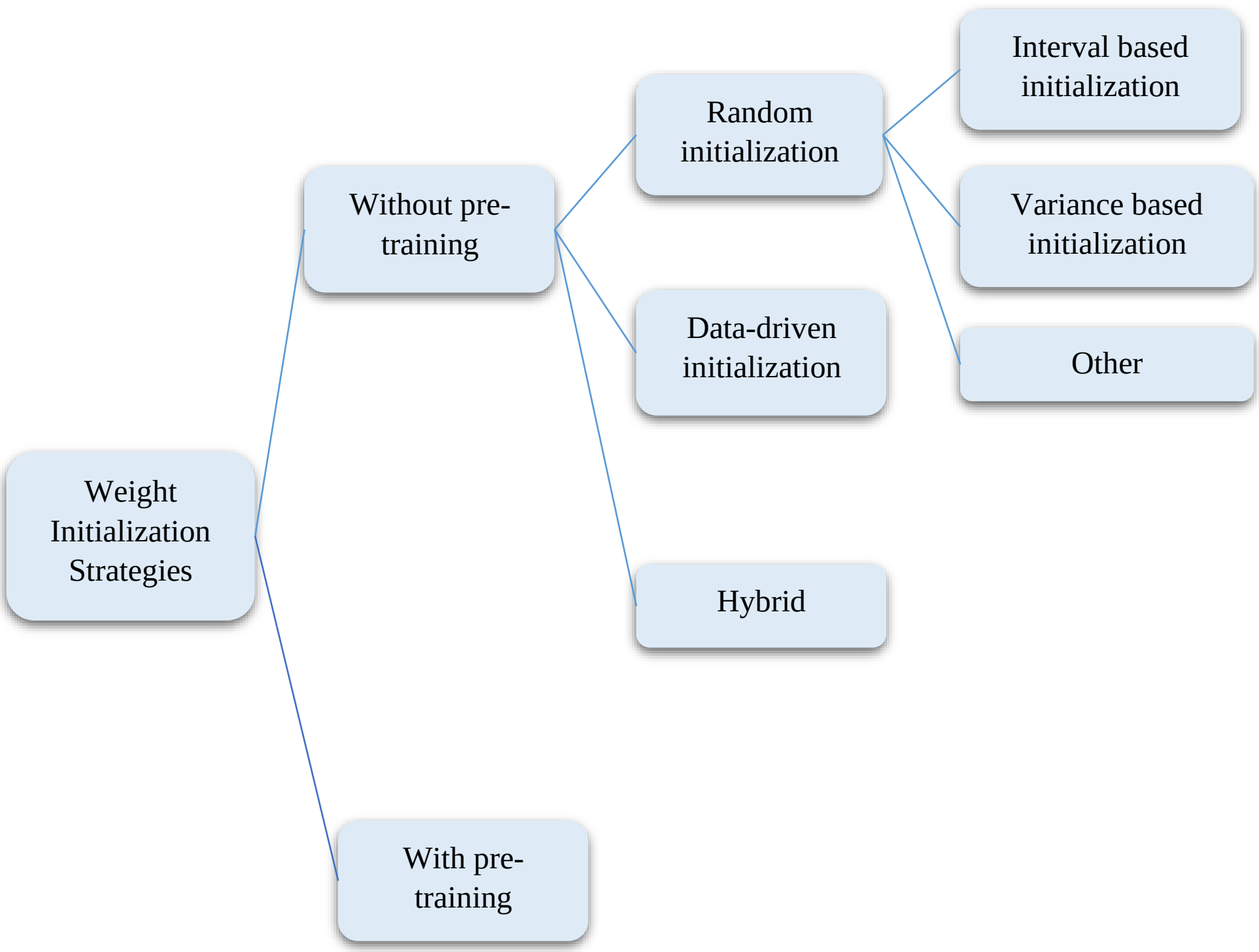

**Figure 4.5.** Categorization of weight initialization strategies in literature for iterative training mechanism.

- Initialization without pre-training:
    - Random initialization: This category explores the initialization techniques which select weights from random distributions.
        - Interval based initialization: This category explores the initialization techniques which select initial weights within an optimal derived range.
        - Variance scaling based initialization: This category explores various initialization schemes in which the weights are scaled such that the variance of inputs and output activations is maintained.
        - Other: The random initialization techniques other than interval based and variance scaling based have been listed in this category.
    - Data-driven initialization: This category deals with the initialization of network weights derived from the training data.
    - Hybrid: This category explores initialization techniques that fall as the combination of both random and data-driven initialization approaches.
- Initialization with pre-training: This category explores the methods where the initialization point is defined by an unsupervised approach for a supervised training algorithm.

Commonly utilized weight initialization techniques include [95, 100, 102-106] zero initialization, random initialization, Xavier/Glorot initialization, He initialization, LeCun initialization, and orthogonal initialization. Zero initialization assigns all weights and biases to zero, which can lead to symmetry issues and hinder learning. Random initialization, on the other hand, initializes weights and biases from a random distribution, with careful consideration of the initialization scale. Xavier/Glorot initialization maintains activation and gradient variances by scaling the initialization based on the number of inputs and outputs. He initialization, a variant of Xavier, is tailored for ReLU AFs to prevent neuron saturation. LeCun initialization method is specifically designed for networks with hyperbolic tangent (Tanh) AF. It initializes the weights using a Gaussian distribution with zero mean and variance of 1/(number of inputs). Orthogonal initialization initializes weights as orthogonal matrices, particularly beneficial for RNNs. Each technique addresses specific challenges in NN training, aiming to promote stable and efficient learning processes.

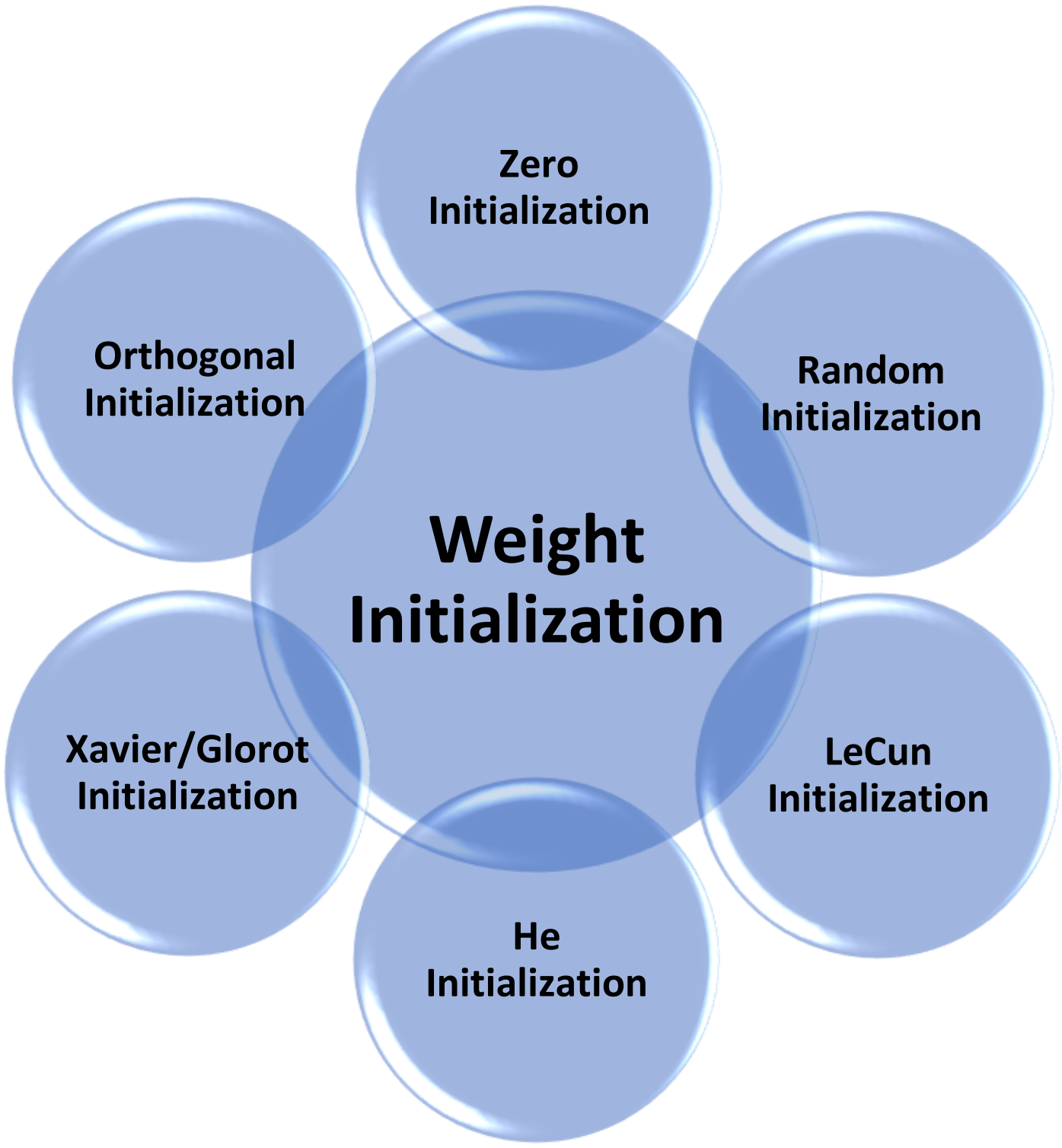

## 4.4 Common Weight Initialization Methods

### 4.4.1 Zero or Constant Initialization (Symmetric Behavior)

This section will explain why initializing all the weights to a zero or constant value is suboptimal. Let's consider a NN with two inputs and one hidden layer with two neurons, and initialize the weights and biases to zero, as shown in Figure 4.6.

For this NN, $z_1^{(1)}$ and $z_2^{(1)}$ are given by the following equations:

$$z_1^{(1)} = w_{11}^{(1)} a_1^{(0)} + w_{12}^{(1)} a_2^{(0)}, \tag{4.9.1}$$
$$z_2^{(1)} = w_{21}^{(1)} a_1^{(0)} + w_{22}^{(1)} a_2^{(0)}. \tag{4.9.2}$$

Now, if all weights are initialized to the same value, say $w$, $w_{11}^{(1)} = w_{12}^{(1)} = w_{21}^{(1)} = w_{22}^{(1)} = w$, then the output for all neurons in that layer becomes:

$$z_1^{(1)} = z_2^{(1)} = w\left(a_1^{(0)} + a_2^{(0)}\right) = z, \tag{4.10}$$

and

$$a_1^{(1)} = a_2^{(1)} = \sigma(z) = a, \tag{4.11}$$

where $\sigma$ denote the AF. All the neurons in the first hidden layer will have the same value. Generally, when the weights are initialized to the same value and the same AF is used across neurons in a layer, the output of the layer (post-activation) becomes identical for each neuron. This, in turn, leads to identical pre-activations for the neurons in the next layer, and this pattern continues through subsequent layers. This causes the neurons in each layer to perform the same as one neuron does during training, which makes the network fail to learn different features from the input.

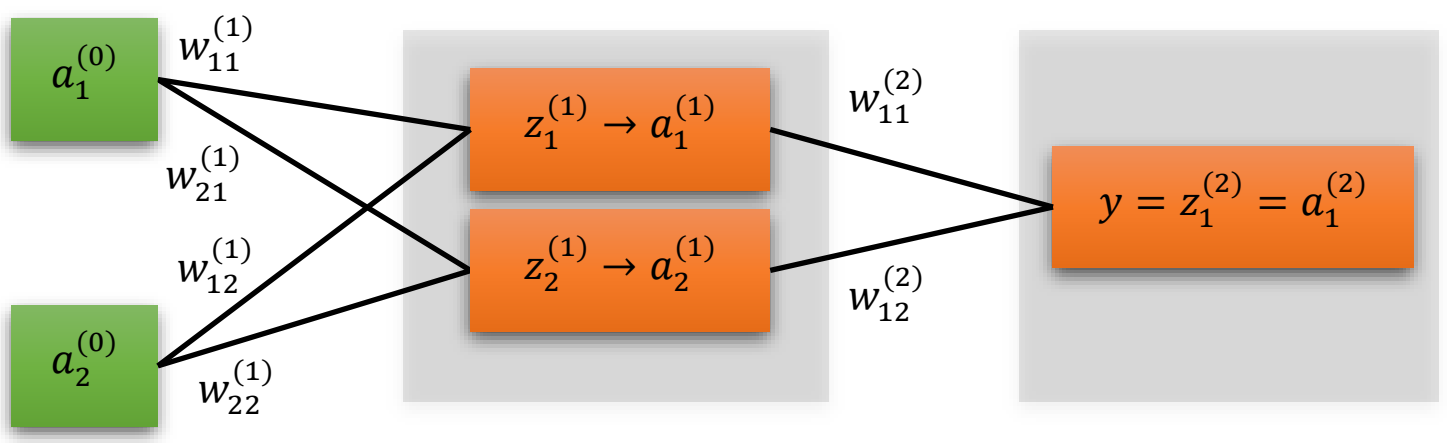

**Figure 4.6.** A simple NN with one hidden layer, with the biases set to zero.

During BP, when we compute the gradient of the loss function with respect to the weights, $\frac{\partial \mathcal{L}}{\partial w}$, the gradient for each weight is the same because all weights are initialized to the same value:

$$\frac{\partial \mathcal{L}}{\partial w_{11}^{(1)}} = \frac{\partial \mathcal{L}}{\partial y} \cdot \frac{\partial y}{\partial a_1^{(1)}} \cdot \frac{\partial a_1^{(1)}}{\partial z_1^{(1)}} \cdot \frac{\partial z_1^{(1)}}{\partial w_{11}^{(1)}} = \frac{\partial \mathcal{L}}{\partial y} \cdot \frac{\partial y}{\partial a_1^{(1)}} \cdot \frac{\partial a_1^{(1)}}{\partial z_1^{(1)}} \cdot a_1^{(0)}, \tag{4.12.1}$$

$$\frac{\partial \mathcal{L}}{\partial w_{21}^{(1)}} = \frac{\partial \mathcal{L}}{\partial y} \cdot \frac{\partial y}{\partial a_2^{(1)}} \cdot \frac{\partial a_2^{(1)}}{\partial z_2^{(1)}} \cdot \frac{\partial z_2^{(1)}}{\partial w_{21}^{(1)}} = \frac{\partial \mathcal{L}}{\partial y} \cdot \frac{\partial y}{\partial a_2^{(1)}} \cdot \frac{\partial a_2^{(1)}}{\partial z_2^{(1)}} \cdot a_1^{(0)}, \tag{4.12.2}$$

since $z_1^{(1)} = z_2^{(1)}$ and $a_1^{(1)} = a_2^{(1)}$,

$$\frac{\partial \mathcal{L}}{\partial w_{11}^{(1)}} = \frac{\partial \mathcal{L}}{\partial w_{21}^{(1)}} = \frac{\partial \mathcal{L}}{\partial w}. \tag{4.13}$$

Therefore, the weight update rule during GD becomes:

$$w_{\text{new}} = w_{\text{old}} - \alpha \frac{\partial \mathcal{L}}{\partial w}. \tag{4.14}$$

After the first update, the weights $w_{11}^{(1)}$ and $w_{21}^{(1)}$ move away from $w$ but are equal. Similarly,

$$\frac{\partial \mathcal{L}}{\partial w_{12}^{(1)}} = \frac{\partial \mathcal{L}}{\partial y} \cdot \frac{\partial y}{\partial a_1^{(1)}} \cdot \frac{\partial a_1^{(1)}}{\partial z_1^{(1)}} \cdot \frac{\partial z_1^{(1)}}{\partial w_{12}^{(1)}} = \frac{\partial \mathcal{L}}{\partial y} \cdot \frac{\partial y}{\partial a_1^{(1)}} \cdot \frac{\partial a_1^{(1)}}{\partial z_1^{(1)}} \cdot a_2^{(0)}, \tag{4.15.1}$$

$$\frac{\partial \mathcal{L}}{\partial w_{22}^{(1)}} = \frac{\partial \mathcal{L}}{\partial y} \cdot \frac{\partial y}{\partial a_2^{(1)}} \cdot \frac{\partial a_2^{(1)}}{\partial z_2^{(1)}} \cdot \frac{\partial z_2^{(1)}}{\partial w_{22}^{(1)}} = \frac{\partial \mathcal{L}}{\partial y} \cdot \frac{\partial y}{\partial a_2^{(1)}} \cdot \frac{\partial a_2^{(1)}}{\partial z_2^{(1)}} \cdot a_2^{(0)}. \tag{4.15.2}$$

Since $z_1^{(1)} = z_2^{(1)}$ and $a_1^{(1)} = a_2^{(1)}$,

$$\frac{\partial \mathcal{L}}{\partial w_{12}^{(1)}} = \frac{\partial \mathcal{L}}{\partial w_{22}^{(1)}}. \tag{4.16}$$

The strength of NNs lies in their ability to learn complex, non-linear mappings from input to output. They are referred to as universal function approximators because, theoretically, they can approximate any continuous function given the right architecture and training. However, when weights flowing into neurons in a layer stay equal (as in the case of symmetry issues), it severely limits the capacity of the network. If all neurons in a layer are learning the same features, the network becomes akin to a shallower model, as it's not effectively leveraging the representational capacity of multiple neurons. Each neuron should ideally be able to capture different aspects or features of the input data. Breaking symmetry through proper weight initialization is crucial to allow each neuron to specialize in learning different patterns. This diversity in learned features across neurons enhances the overall expressive power of the network, enabling it to capture intricate relationships in the data.

### 4.4.2 Random Initialization

As we see in the previous section, it is widely acknowledged that initializing weights to all zeros or any constant value is not a viable option, as it leads to symmetry issues and prevents the model from effectively learning diverse features. Consequently, the next logical approach is to initialize the weights to random values. This random initialization introduces diversity and breaks the symmetry, allowing the network to learn unique patterns and representations.

However, the question arises: does random initialization indeed work? The problem of random initialization is that it may lead to vanishing or exploding layer outputs in the DNN, depending on the variance of Gaussian distribution.

Let us initialize the weights with small random values drawn from a standard normal distribution, characterized by zero mean and a negligible variance of 0.0001. However, this strategy may lead to diminishing activations as we traverse deeper into the NN. This issue arises because during BP, the gradients flowing into a neuron are proportional to the activation it receives. When the weights are initialized to small values, the magnitudes of activations become diminutive, resulting in vanishingly small gradients. Consequently, neurons fail to learn effectively. On the contrary, initializing weights to larger random values from a standard normal distribution with zero mean and unit variance may lead to a different challenge. The Sigmoid and Tanh AFs tend to saturate when the weights have a large magnitude, causing activations to approach saturation. In this state, gradients during BP approach zero, impeding effective learning.

To address this dilemma, finding the optimal scaling factor, or equivalently, determining the optimal variance for the weight distribution becomes crucial. By selecting an appropriate scaling factor, we can position the network in the intermediate region between vanishing and saturating activations. This optimized initialization allows for a balanced flow of gradients, facilitating effective learning throughout the NN.

One possible approach to initialize the weights is to generate random values from a standard Gaussian distribution with zero mean and standard deviation 1. Typically, this will result in random values that are both positive and negative. However, when this method is applied to a NN with specific input patterns, (training input vector, **x**, with half of the input neurons are set to 1 and the other half are set to 0), saturation issues may arise, impacting the learning process. While this specific scenario may be simplified or specialized, the underlying argument holds broader implications that can be applied more generally across various NNs.

Consider a NN with $n$ input neurons. We initialize the weights $w_i$ with normalized Gaussian random variables, i.e.,

$$w_i \sim N(0, 1), \tag{4.17}$$

independently for each $i$. Now, let us examine the impact of this initialization when the training input **x** has a specific structure. Suppose **x** is the training input vector with $n$ elements,

$$\mathbf{x} = (1, 1, \dots, 1, 0, 0, \dots, 0)^T, \tag{4.18}$$

with $n/2$ elements set to 1 and the remaining $n/2$ set to 0. The weighted sum $z$ for a neuron connected to the input neurons:

$$z = \sum_{i=1}^{n} w_i x_i + b. \tag{4.19}$$

For simplicity, let the bias term $b = 0$. Due to the structure of **x**, only the weights corresponding to the active input neurons (set to 1) will contribute to the sum. $n/2$ terms in this sum vanish, because the corresponding input $x_i$ is zero. $z = \sum_{i=1}^{n/2} w_i$. Since $w_i$ follows a standard Gaussian distribution, the weighted sum $z$ becomes a sum of independent Gaussian random variables. By the properties of the Gaussian distribution, the sum of independent Gaussians is also Gaussian,

$$z \sim N\left(0, \sqrt{n/2}\right). \tag{4.20}$$

The standard deviation of $z$ is $\sqrt{n/2}$, which becomes large as $n$ increases. That is, $z$ has a very broad Gaussian distribution, not sharply peaked at all. AFs like Sigmoid saturate for extreme input values, approaching 0 or 1. Given the large standard deviation of $z$ due to the normalized Gaussian initialization, the weighted sum often falls within the saturation region during training. This causes the neuron's output to be consistently pushed toward the extremes, leading to activation saturation. Activation saturation can lead to vanishing gradients during BP, hindering the learning process.

Figure 4.7 showcases the saturation behavior of the Sigmoid function, approaching 0 or 1 for extreme inputs. The dashed lines indicate the saturation region, and the blue shaded areas represent instances where the weighted sum often falls within this saturation region during training. The standard deviation of $z$ increases with the square root of $n/2$, resulting in a broader Gaussian distribution. If $n = 200$, we have $\sigma = \sqrt{200/2} = 10$. That is, $z$ has a very broad Gaussian distribution, not sharply peaked at all, see the case $N(0,10)$ in the Figure 4.7. When initializing the weights of a NN, selecting an appropriate scaling factor for the variance becomes crucial. The goal is to position the network in the intermediate region between vanishing and saturating activations.

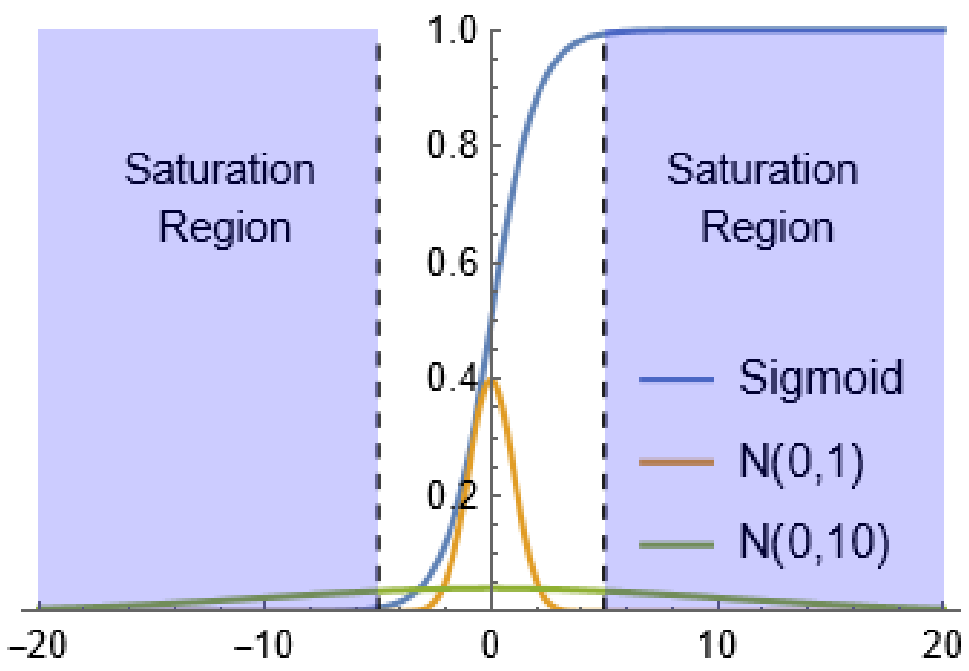

**Figure 4.7.** Visualization of sigmoid AF and normal distributions. The sigmoid function (labeled 'Sigmoid') is compared with two normal distributions: $N(0,1)$ and $N(0,10)$, over the range $-20$ to $20$. The plot highlights the saturation region of the sigmoid function, where gradients approach zero, with blue shading.

The problem with the standard Gaussian initialization is that it is not sensitive to the number of inputs to a specific neuron. For example, if one neuron has only 2 inputs and another has 100 inputs, the output of the former is far more sensitive to the average weight because of the additive effect of more inputs.

In addressing the challenge of saturation and the associated learning slowdown, a potential solution lies in optimizing the initialization of weights. For a neuron with $n_{in}$ input weights, a refined approach involves initializing these weights as Gaussian random variables with a mean of 0 and a standard deviation of $1/\sqrt{n_{in}}$. This deliberate adjustment effectively compresses the Gaussian distribution, diminishing the likelihood of neuron saturation during training. Under this initialization scheme, the weighted sum retains its Gaussian nature, with a mean of 0. However, the crucial distinction lies in the substantially sharper peak of the distribution. Suppose as mentioned earlier, that $n/2$ of the inputs are zero, and $n/2$ are 1. In this case, it is straightforward to demonstrate that $z$ has a Gaussian distribution with a mean of 0 and a standard deviation of $\sqrt{(1/n_{in})(n_{in}/2)} = \sqrt{1/2} = 0.707$. This distribution is much more sharply peaked than the previous scenario, see the case $N(0,1)$ in the Figure 4.7.

In fact, it doesn't much matter how we initialize the biases, provided we avoid the problem with saturation. Some people continue to initialize the biases as Gaussian random variables with a mean of 0 and a standard deviation of 1. But since it's unlikely to make much difference.

### 4.4.3 LeCun Initialization

LeCun initialization [106] aims to prevent the vanishing or explosion of the gradients during the BP by solving the growing variance with the number of inputs and by setting constant variance. Consider the hypothesis that we are in a linear regime at the initialization. The "linear regime" of the Sigmoid or Tanh functions refers to the region around the origin (input values close to 0) where the function behaves approximately like a linear function. Weights should be chosen randomly but in such a way that the AF (Sigmoid or Tanh) is primarily activated in its linear region. Intermediate weights that range over the AF's linear region have the advantage that (1) the gradients are large enough that learning can proceed and (2) the network will learn the linear part of the mapping before the more difficult nonlinear part.

Let us begin by assuming that the initialized elements in $\mathbf{W}^{(l)}$ be Independent and Identically Distributed (IID). The input features variances are the same ($\mathrm{Var}\left[a_i^{(0)}\right] = \mathrm{Var}\left[a^{(0)}\right]$), where $\mathrm{Var}\left[a^{(0)}\right]$ is the shared scalar variance of all inputs. The inputs to our system have undergone normalization, resulting in a dataset with zero mean.

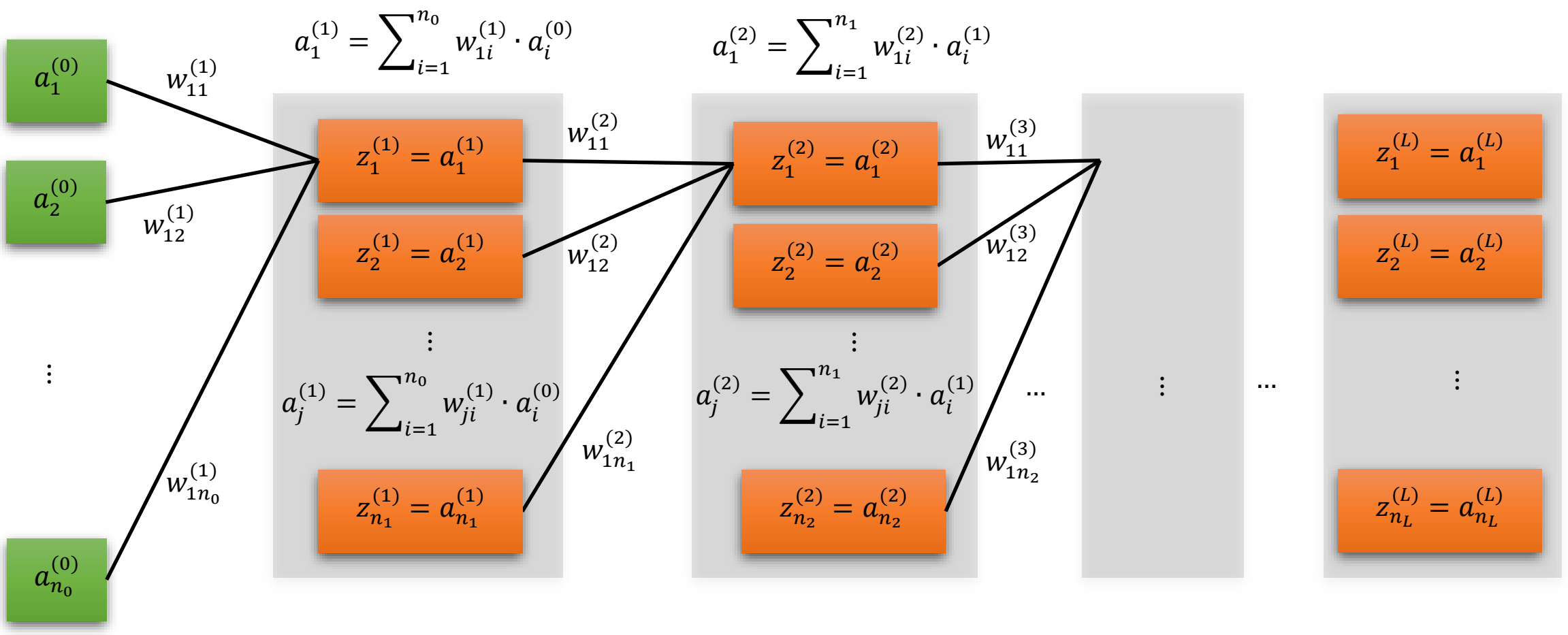

**Figure 4.8.** Diagrammatic representation of a NN (ignore the explicit effect of activations).

Now, as we delve into the intricacies of our model, it becomes crucial to address the question of the variance associated with the weights. The weights in our model are drawn from a distribution with a mean of zero, which aligns with the normalized nature of our inputs. However, a pivotal aspect remains unresolved: what should the fixed variance of this weight distribution be? This question serves as a gateway to a more in-depth analysis of our model's behavior and performance.

To compute the optimal variance, we will use $a_1^{(1)} \approx z_1^{(1)}$ (we ignore the explicit effect of activations, linear regime) as the first input to the first neuron in the second hidden layer, Figure 4.8,

$$a_1^{(1)} = \sum_{i=1}^{n_0} w_{1i}^{(1)} \cdot a_i^{(0)}. \tag{4.21}$$

Note that, in probability and statistics, a random variable is a variable whose values are determined by the outcome of a random experiment. Random variables are traditionally denoted by uppercase letters, such as $X$. After the experiment is conducted, and a specific outcome is observed, the value of the random variable is denoted by a lowercase letter, such as $x$. For example, if $X$ represents the random variable "the outcome of rolling a six-sided die," then $x$ could be a specific outcome like 3, indicating that the die landed on the number 3. The distinction between uppercase (random variable) and lowercase (measured value) letters helps to clarify whether we are referring to the general concept of a random variable or a specific observed value from a particular instance of the random experiment. To avoid confusion with matrices, which are denoted using bold uppercase letters in our notation, we will use a bar above the variable to represent random variables. Then, we have

$$\bar{a}_1^{(1)} = \sum_{i=1}^{n_0} \bar{w}_{1i}^{(1)} \cdot \bar{a}_i^{(0)}, \tag{4.22}$$

$$\text{Var}\left[\bar{a}_1^{(1)}\right] = \text{Var}\left[\sum_{i=1}^{n_0} \bar{w}_{1i}^{(1)} \cdot \bar{a}_i^{(0)}\right]. \tag{4.23}$$

Note that, if $X$ and $Y$ are two independent random variables, then:

$$\text{Var}[X+Y] = \text{Var}[X] + \text{Var}[Y], \tag{4.24}$$

and

$$\text{Var}[XY] = \text{Var}[X] \cdot \text{Var}[Y] + \text{Var}[X] \cdot (\mathbb{E}[Y])^2 + \text{Var}[Y] \cdot (\mathbb{E}[X])^2. \tag{4.25}$$

Assumption: The weights, $\bar{w}_{1i}^{(1)}$, and inputs, $\bar{a}_i^{(0)}$, are uncorrelated random variables with means $\mathbb{E}\left[\bar{w}_{1i}^{(1)}\right]$, $\mathbb{E}\left[\bar{a}_i^{(0)}\right]$ and variances $\text{Var}\left[\bar{w}_{1i}^{(1)}\right]$, $\text{Var}\left[\bar{a}_i^{(0)}\right]$. So, we have

$$\begin{aligned}\text{Var}\left(\bar{a}_1^{(1)}\right) &= \sum_{i=1}^{n_0} \text{Var}\left[\bar{w}_{1i}^{(1)} \cdot \bar{a}_i^{(0)}\right] \\ &= \sum_{i=1}^{n_0} \left\{\text{Var}\left[\bar{w}_{1i}^{(1)}\right] \cdot \text{Var}\left[\bar{a}_i^{(0)}\right] + \text{Var}\left[\bar{w}_{1i}^{(1)}\right] \cdot \left(\mathbb{E}\left[\bar{a}_i^{(0)}\right]\right)^2 + \text{Var}\left[\bar{a}_i^{(0)}\right] \cdot \left(\mathbb{E}\left[\bar{w}_{1i}^{(1)}\right]\right)^2\right\}.\end{aligned} \tag{4.26}$$

Substituting $\mathbb{E}\left[\bar{w}_{1i}^{(1)}\right] = 0$ and $\mathbb{E}\left[\bar{a}_i^{(0)}\right] = 0$ in the above equation:

$$\text{Var}\left[\bar{a}_1^{(1)}\right] = \sum_{i=1}^{n_0} \text{Var}\left[\bar{w}_{1i}^{(1)}\right] \cdot \text{Var}\left[\bar{a}_i^{(0)}\right]. \tag{4.27}$$

Let $\text{Var}\left[\bar{a}_i^{(0)}\right] = \text{Var}\left[\bar{a}^{(0)}\right]$ (we assumed that the inputs features variances are the same) and $\text{Var}\left[\bar{w}_{1i}^{(1)}\right] = \text{Var}\left[\bar{w}^{(1)}\right]$, where $\text{Var}\left[\bar{w}^{(1)}\right]$ is the shared scalar variance of all weights at layer 1.

$$\text{Var}\left[\bar{a}_1^{(1)}\right] = n_0 \cdot \text{Var}\left[\bar{w}^{(1)}\right] \cdot \text{Var}\left[\bar{a}^{(0)}\right]. \tag{4.28}$$

Similarly, the obtained result can be extended to any neuron within layer 1, as illustrated below:

$$\text{Var}\left[\bar{a}_j^{(1)}\right] = n_0 \cdot \text{Var}\left[\bar{w}^{(1)}\right] \cdot \text{Var}\left[\bar{a}^{(0)}\right]. \tag{4.29}$$

Let's compute the variance of $\bar{a}_j^{(2)}$ in the second hidden layer.

$$\begin{aligned}\text{Var}\left(\bar{a}_j^{(2)}\right) &= \text{Var}\left[\sum_{i=1}^{n_1} \bar{w}_{ji}^{(2)} \cdot \bar{a}_i^{(1)}\right] \\ &= \sum_{i=1}^{n_1} \text{Var}\left[\bar{w}_{ji}^{(2)} \cdot \bar{a}_i^{(1)}\right] \\ &= \sum_{i=1}^{n_1} \left\{\text{Var}\left[\bar{w}_{ji}^{(2)}\right] \cdot \text{Var}\left[\bar{a}_i^{(1)}\right] + \text{Var}\left[\bar{w}_{ji}^{(2)}\right] \cdot \left(\mathbb{E}\left[\bar{a}_i^{(1)}\right]\right)^2 + \text{Var}\left[\bar{a}_i^{(1)}\right] \cdot \left(\mathbb{E}\left[\bar{w}_{ji}^{(2)}\right]\right)^2\right\} \\ &= \sum_{i=1}^{n_1} \text{Var}\left[\bar{w}_{ji}^{(2)}\right] \cdot \text{Var}\left[\bar{a}_i^{(1)}\right] \\ &= n_1 \cdot \text{Var}\left[\bar{a}^{(1)}\right] \cdot \text{Var}\left[\bar{w}^{(2)}\right] \\ &= n_1 \cdot \left\{n_0 \cdot \text{Var}\left[\bar{a}^{(0)}\right] \cdot \text{Var}\left[\bar{w}^{(1)}\right]\right\} \cdot \text{Var}\left[\bar{w}^{(2)}\right] \\ &= \left(n_0 \cdot \text{Var}\left[\bar{w}^{(1)}\right]\right) \cdot \left(n_1 \cdot \text{Var}\left[\bar{w}^{(2)}\right]\right) \cdot \text{Var}\left[\bar{a}^{(0)}\right],\end{aligned} \tag{4.30}$$

where $\mathbb{E}\left[\bar{w}_{ji}^{(2)}\right] = 0$, and $\text{Var}\left[\bar{w}_{ji}^{(2)}\right] = \text{Var}\left[\bar{w}^{(2)}\right]$, $\text{Var}\left[\bar{w}^{(2)}\right]$ is the shared scalar variance of all weights at layer 2, $\text{Var}\left[\bar{a}_i^{(1)}\right] = \text{Var}\left[\bar{a}^{(1)}\right]$ and

$$\begin{aligned}\mathbb{E}\left[\bar{a}_i^{(1)}\right] &= \mathbb{E}\left[\sum_{k=1}^{n_0} \bar{w}_{1k}^{(1)} \cdot \bar{a}_k^{(0)}\right] \\ &= \sum_{k=1}^{n_0} \mathbb{E}\left[\bar{w}_{1k}^{(1)} \cdot \bar{a}_k^{(0)}\right] \\ &= \sum_{k=1}^{n_0} \mathbb{E}\left[\bar{w}_{1k}^{(1)}\right] \cdot \mathbb{E}\left[\bar{a}_k^{(0)}\right] = 0,\end{aligned} \tag{4.31}$$

where $\mathbb{E}\left[\bar{a}_k^{(0)}\right] = 0$, and we used $\mathbb{E}[Z] = \mathbb{E}[X + Y] = \mathbb{E}[X] + \mathbb{E}[Y]$ and $\mathbb{E}[X \cdot Y] = \mathbb{E}[X] \cdot \mathbb{E}[Y] = \mathbb{E}[Y] \cdot \mathbb{E}[X]$, since $X$ and $Y$ are independent from each other.

By induction, the variance of input to neuron 1 in the hidden layer $l$ is given by:

$$\begin{aligned}\text{Var}\left[\bar{a}_j^{(l)}\right] &= \left(n_0 \cdot \text{Var}\left[\bar{w}^{(1)}\right]\right) \cdot \left(n_1 \cdot \text{Var}\left[\bar{w}^{(2)}\right]\right) \cdot \ldots \cdot \left(n_{l-1} \cdot \text{Var}\left[\bar{w}^{(l)}\right]\right) \cdot \text{Var}\left[\bar{a}^{(0)}\right] \\ &= \left(\prod_{i=1}^{l}\left(n_{i-1} \cdot \text{Var}\left[\bar{w}^{(i)}\right]\right)\right) \cdot \text{Var}\left[\bar{a}^{(0)}\right].\end{aligned} \tag{4.32}$$

We write $\text{Var}\left[\bar{w}^{(i)}\right]$ for the shared scalar variance of all weights at layer $i$. To ensure that the activations have the same variance as the features $\text{Var}\left[\bar{a}_j^{(l)}\right] = \text{Var}\left[\bar{a}^{(0)}\right]$, and the quantity $n_{i-1} \cdot \text{Var}\left[\bar{w}^{(i)}\right]$ neither vanishes nor grows exponentially (leading to instability in the training process), we need $n_{i-1} \cdot \text{Var}\left[\bar{w}^{(i)}\right] = 1$.

$$\text{Var}\left[\bar{w}^{(i)}\right] = \sigma^2 = \frac{1}{n_{i-1}}, \tag{4.33}$$

and

$$\sigma = \frac{1}{\sqrt{n_{i-1}}}. \tag{4.34}$$

In a layer $l$, the number of inputs will be the number of neurons of the preceding layer $l-1$. So, we will have $n_{in}^{(l)} = n_{l-1}$. As a result, we have

$$w^{(l)} \sim N\left(0,1/\sqrt{n_{in}^{(l)}}\right). \tag{4.35}$$

### 4.4.4 Kaiming/He Initialization

Kaiming initialization, also known as He initialization, is a technique for initializing the weights of NN layers introduced by Kaiming He and his colleagues [100]. It is designed to address the vanishing or exploding gradient problem during the training of DNNs. The Kaiming Initialization is commonly used with AFs like ReLU and its variants, such as LReLU. Mathematically, the initialization can be expressed as follows:

$$w^{(l)} \sim N\left(0, \sqrt{2/n_{in}^{(l)}}\right). \tag{4.36}$$

In a layer $l$, the number of inputs will be the number of neurons of the preceding layer $l-1$ (it is related to the number of connections entering each neuron). So, we will have $n_{in}^{(l)} = n_{l-1}$. This initialization strategy is derived based on the assumption that the weights are IID.

Kaiming initialization is designed to maintain consistent activation variances across layers during both forward and backward passes. In forward propagation, Kaiming initialization ensures that the variance of activations remains roughly constant as data flows through the NN layers. Similarly, during backward propagation, Kaiming initialization plays a vital role. Although the principles are akin to forward propagation, there are nuances specific to the backward pass. It involves the proper initialization of weights to facilitate the backflow of gradients, allowing for efficient and stable training. By addressing the challenges of vanishing or exploding gradients in both directions, Kaiming initialization contributes to the overall stability and convergence of the NN. Let's explore each case separately to understand their similarities and distinctions.

#### Forward-propagation

For each layer $l$, the response is written as:

$$\mathbf{z}^{(l)} = \mathbf{W}^{(l)} \cdot \mathbf{a}^{(l-1)} + \mathbf{b}^{(l)} = \mathbf{W}^{(l)} \cdot \mathbf{x}^{(l)} + \mathbf{b}^{(l)}, \tag{4.37}$$

where $\mathbf{x}^{(l)} = \mathbf{a}^{(l-1)}$ is a $n_{l-1} \times 1$ vector that represents the activations of the previous layer $l-1$, $n_{l-1}$ is the number of activations of layer $l-1$, so we have

$$\mathbf{x}^{(l)} = \mathbf{a}^{(l-1)} = \sigma\left(\mathbf{z}^{(l-1)}\right). \tag{4.38}$$

$\mathbf{W}^{(l)}$ is a matrix of all the connections from layer $l-1$ to layer $l$. $\mathbf{b}^{(l)}$ is the vector of biases of layer $l$ (that are usually initialized to 0). $\mathbf{z}^{(l)}$ is pre-activation vector of layer $l$. We have:

$$\underbrace{\mathbf{z}^{(l)}}_{n_l \times 1} = \begin{pmatrix} z_1^{(l)} \\ z_2^{(l)} \\ \vdots \\ z_{n_l}^{(l)} \end{pmatrix} = \underbrace{\mathbf{W}^{(l)}}_{n_l \times n_{l-1}} \cdot \underbrace{\mathbf{x}^{(l)}}_{n_{l-1} \times 1} = \begin{pmatrix} \sum_{j=1}^{n_{l-1}} w_{1j}^{(l)} x_j^{(l)} \\ \sum_{j=1}^{n_{l-1}} w_{2j}^{(l)} x_j^{(l)} \\ \vdots \\ \sum_{j=1}^{n_{l-1}} w_{n_l j}^{(l)} x_j^{(l)} \end{pmatrix}. \tag{4.39}$$

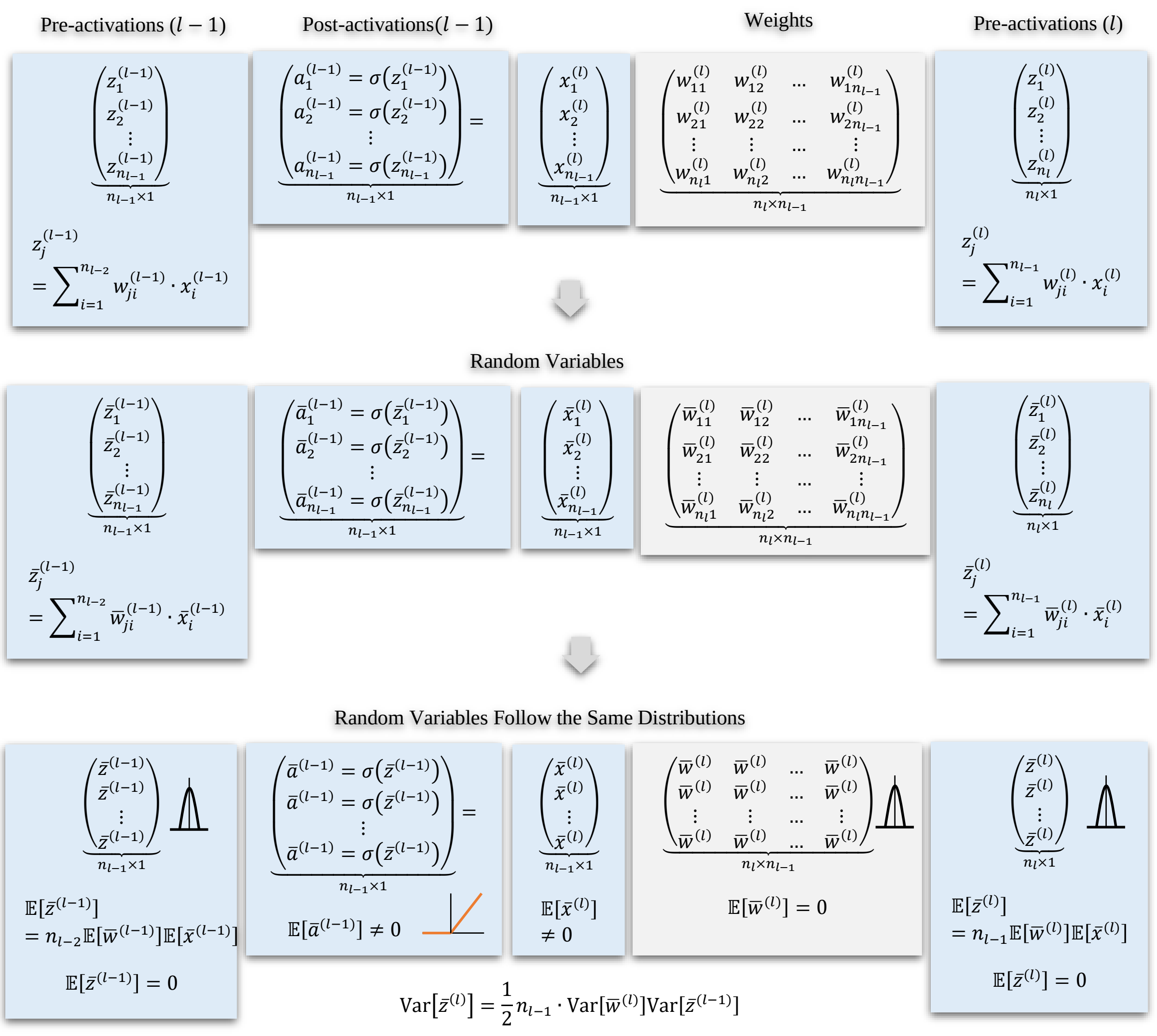

**Figure 4.9.** Diagrammatic representation of NN architecture and vector notations across two layers $(l-1, l)$. This visualization illustrates the flow of information through these layers, showcasing the vectors utilized in the computations. The elements in $\mathbf{x}^{(l)}$ are mutually independent and share the same distribution. Similarly, the initialized elements in $\mathbf{W}^{(l)}$ are mutually independent and share the same distribution. Random variables of pre-activation, post-activation and weights are distinctly marked with a bar placed above the variable symbol, serving to differentiate them from deterministic variables within the network.

The Kaiming initialization strategy is derived based on the following assumptions:

- The weights are IID, i.e., the initialized elements in $\mathbf{W}^{(l)}$ are mutually independent and share the same distribution. $\mathbf{W}^{(l)} = \left(w_{ij}^{(l)}\right)$, $1 \le i \le n_l, 1 \le j \le n_{l-1}$, with all $w_{ij}^{(l)}$ following the same distribution. As a result, let $\bar{w}_{ij}^{(l)} = \bar{w}^{(l)}$ represent the random variable of each element in $\mathbf{W}^{(l)}$.
- Likewise, the elements in $\mathbf{x}^{(l)}$ are mutually independent and share the same distribution, i.e., $\mathbf{x}^{(l)} = \left(x_j^{(l)}\right)$, $1 \le j \le n_{l-1}$, with all following the same distribution. Hence, let $\bar{x}_j^{(l)} = \bar{x}^{(l)}$ represent the random variable of each element in $\mathbf{x}^{(l)}$, see Figure 4.9.
- $\mathbf{x}^{(l)}$ and $\mathbf{W}^{(l)}$ are independent of each other.

Now, we take the variance of (4.39):

$$\mathrm{Var}\left[\bar{z}_k^{(l)}\right] = \mathrm{Var}\left[\sum_{j=1}^{n_{l-1}} \bar{w}_{kj}^{(l)} \bar{x}_j^{(l)}\right] = \sum_{j=1}^{n_{l-1}} \mathrm{Var}\left[\bar{w}_{kj}^{(l)} \bar{x}_j^{(l)}\right]. \tag{4.40}$$

And because all the $\bar{w}_{kj}^{(l)}$ and $\bar{x}_j^{(l)}$ follow the same distribution (respectively), i.e., $\bar{w}_{kj}^{(l)} = \bar{w}^{(l)}$ and $\bar{x}_j^{(l)} = \bar{x}^{(l)}$ . Hence:

$$\begin{aligned}\mathrm{Var}\left[\bar{z}_k^{(l)}\right] &= \sum_{j=1}^{n_{l-1}} \mathrm{Var}\left[\bar{w}_{kj}^{(l)} \bar{x}_j^{(l)}\right] \\ &= n_{l-1}\mathrm{Var}\left[\bar{w}^{(l)}\bar{x}^{(l)}\right].\end{aligned} \tag{4.41}$$

Following the exact same reasoning, $\bar{z}_j^{(l)} = \bar{z}^{(l)}$, we obtain:

$$\mathrm{Var}\left[\bar{z}^{(l)}\right] = n_{l-1}\mathrm{Var}\left[\bar{w}^{(l)}\bar{x}^{(l)}\right]. \tag{4.42}$$

We now make one more assumption: that the random variable $\bar{w}^{(l)}$ has zero mean. After all, setting the distribution of $\bar{w}^{(l)}$ is what we're trying to do here, so we can choose one that has zero mean. We can now further simplify (4.41) as follow (we also use once more the fact that $w_l$ and $x_l$ are independent):

$$\begin{aligned}\mathrm{Var}\left[\bar{z}^{(l)}\right] &= n_{l-1}\mathrm{Var}\left[\bar{w}^{(l)}\bar{x}^{(l)}\right] \\ &= n_{l-1}\left(\underbrace{\mathbb{E}\left[\left(\bar{w}^{(l)}\right)^2\right]}_{\mathrm{Var}\left[\bar{w}^{(l)}\right]}\mathbb{E}\left[\left(\bar{x}^{(l)}\right)^2\right] - \underbrace{\left(\mathbb{E}\left[\bar{w}^{(l)}\right]\right)^2}_{=0}\left(\mathbb{E}\left[\bar{x}^{(l)}\right]\right)^2\right) \\ &= n_{l-1}\cdot\mathrm{Var}\left[\bar{w}^{(l)}\right]\mathbb{E}\left[\left(\bar{x}^{(l)}\right)^2\right],\end{aligned} \tag{4.43}$$

where we use $\mathrm{Var}[XY] = \mathbb{E}[X^2]\mathbb{E}[Y^2] - (\mathbb{E}[X])^2(\mathbb{E}[Y])^2$ and $\mathrm{Var}[X] = \mathbb{E}[X^2] - (\mathbb{E}[X])^2$. In the above formulae, $\mathbb{E}\left[\left(\bar{x}^{(l)}\right)^2\right]$ refers to the expectation of the square of $\bar{x}^{(l)}$.

Why doesn't $\bar{x}^{(l)}$ have zero mean? That's because it is the result of the ReLU of the previous layer ($x^{(l)} = \max(0, z^{(l-1)})$ and thus it does not have zero mean. That is also one point on which LeCun and Kaiming differ: LeCun doesn't take into account the AF (remember, we ignore the explicit effect of activations in LeCun initialization, linear regime). As a reminder, here's the definition of the ReLU AF:

$$\mathrm{ReLU}(x) = \begin{cases} x, & x \geq 0 \\ 0, & x < 0 \end{cases} = \max(x, 0). \tag{4.44}$$

Now let's see what to do with $\mathbb{E}\left[\left(\bar{x}^{(l)}\right)^2\right] = \mathbb{E}\left[\left(\max\left(0, \bar{z}^{(l-1)}\right)\right)^2\right]$. We assume that $\bar{w}^{(l-1)}$ has a symmetric distribution around 0 and that $b^{l-1} = 0$. Then $\bar{z}^{(l-1)}$ will also have a symmetric distribution around 0, and will have zero mean (because $\mathbb{E}\left[\bar{w}^{(l-1)}\right] = 0$, and $\bar{w}^{(l-1)}$ and $\bar{x}^{(l-1)}$ are independent). We have

$$\begin{aligned}\mathbb{E}\left[\bar{z}^{(l-1)}\right] &= \mathbb{E}\left[n_{l-2}\bar{w}^{(l-1)}\bar{x}^{(l-1)}\right] \\ &= n_{l-2}\mathbb{E}\left[\bar{w}^{(l-1)}\right]\mathbb{E}\left[\bar{x}^{(l-1)}\right] \\ &= 0.\end{aligned} \tag{4.45}$$

Because $\mathbb{E}\left[\bar{w}^{(l-1)}\right] = 0$ and is distributed symmetrically around 0:

$$\begin{aligned}\mathbb{P}\left(\bar{z}^{(l-1)} > 0\right) &= \mathbb{P}\left(\bar{w}^{(l-1)}\bar{x}^{(l-1)} > 0\right) \\ &= \mathbb{P}\left(\left(\bar{w}^{(l-1)} > 0 \text{ and } \bar{x}^{(l-1)} > 0\right) \text{ or } \left(\bar{w}^{(l-1)} < 0 \text{ and } \bar{x}^{(l-1)} < 0\right)\right) \\ &= \mathbb{P}\left(\bar{w}^{(l-1)} > 0\right)\mathbb{P}\left(\bar{x}^{(l-1)} > 0\right) + \mathbb{P}\left(\bar{w}^{(l-1)} < 0\right)\mathbb{P}\left(\bar{x}^{(l-1)} < 0\right) \\ &= \frac{1}{2}\mathbb{P}\left(\bar{x}^{(l-1)} > 0\right) + \frac{1}{2}\mathbb{P}\left(\bar{x}^{(l-1)} < 0\right) \\ &= \frac{1}{2}\mathbb{P}\left(\bar{x}^{(l-1)} > 0\right) + \frac{1}{2}\left(1 - \mathbb{P}\left(\bar{x}^{(l-1)} > 0\right)\right) \\ &= 1/2.\end{aligned} \tag{4.46}$$

The random variable $\bar{x}^{(l)} = \max\left(0, \bar{z}^{(l-1)}\right)$ has a finite list of possible outcomes, $\{0, \bar{z}^{(l-1)}\}$, each of which (respectively) has probability $\{1/2, 1/2\}$ of occurring. Now that we have established that $\bar{z}^{(l-1)}$ is indeed centered on 0 and symmetric, we can compute the expectation of $x_l^2$:

Case 1: $\bar{z}^{(l-1)} \geq 0$

$$\mathbb{E}\left[\left(\bar{x}^{(l)}\right)^2\right] = \mathbb{E}\left[\left(\max\left(0, \bar{z}^{(l-1)}\right)\right)^2\right] = \mathbb{E}\left[\left(\bar{z}^{(l-1)}\right)^2\right]. \tag{4.47}$$

Case 1: $\bar{z}^{(l-1)} < 0$

$$\mathbb{E}\left[\left(\bar{x}^{(l)}\right)^2\right] = \mathbb{E}\left[\left(\max\left(0, \bar{z}^{(l-1)}\right)\right)^2\right] = \mathbb{E}[0] = 0. \tag{4.48}$$

The expectation of the square of the max function can be written as a weighted average of the two cases:

$$\left(\max\left(0, \bar{z}^{(l-1)}\right)\right)^2 = \frac{1}{2}\left(\bar{z}^{(l-1)}\right)^2 + \frac{1}{2}(0) = \frac{1}{2}\left(\bar{z}^{(l-1)}\right)^2. \tag{4.49}$$

The factor of $1/2$ arises because, in a symmetric distribution like $\bar{z}^{(l-1)}$, half of the values are negative and half are non-negative. Therefore, we take the weighted average as shown above. We have

$$\begin{aligned}
\mathbb{E}\left[\left(\bar{x}^{(l)}\right)^2\right] &= \mathbb{E}\left[\left(\max\left(0, \bar{z}^{(l-1)}\right)\right)^2\right] \\
&= \mathbb{E}\left[\frac{1}{2}\left(\bar{z}^{(l-1)}\right)^2\right] \\
&= \frac{1}{2}\mathbb{E}\left[\left(\bar{z}^{(l-1)}\right)^2\right] \\
&= \frac{1}{2}\mathrm{Var}\left[\bar{z}^{(l-1)}\right].
\end{aligned} \tag{4.50}$$

Plugging this back into (4.43), we get:

$$\mathrm{Var}\left[\bar{z}^{(l)}\right] = \frac{1}{2} n_{l-1} \cdot \mathrm{Var}\left[\bar{w}^{(l)}\right]\mathrm{Var}\left[\bar{z}^{(l-1)}\right]. \tag{4.51}$$

We now have a recurrence equation between the activations at layer $l$ and the activations at layer $l-1$. Starting from the last layer $L$, we can thus form the following product:

$$\begin{aligned}
\mathrm{Var}\left[\bar{z}^{(L)}\right] &= \frac{1}{2} n_{L-1} \cdot \mathrm{Var}\left[\bar{w}^{(L)}\right]\mathrm{Var}\left[\bar{z}^{(L-1)}\right] \\
&= \frac{1}{2} n_{L-1} \cdot \mathrm{Var}\left[\bar{w}^{(L)}\right]\left(\frac{1}{2} n_{L-2} \cdot \mathrm{Var}\left[\bar{w}^{(L-1)}\right]\mathrm{Var}\left[\bar{z}^{(L-2)}\right]\right) \\
&= \left(\frac{1}{2} n_{L-1} \cdot \mathrm{Var}\left[\bar{w}^{(L)}\right]\right)\left(\frac{1}{2} n_{L-2} \cdot \mathrm{Var}\left[\bar{w}^{(L-1)}\right]\right)\left(\frac{1}{2} n_{L-3} \cdot \mathrm{Var}\left[\bar{w}^{(L-2)}\right]\mathrm{Var}\left[\bar{z}^{(L-3)}\right]\right) \\
&= \left(\frac{1}{2} n_{L-1} \cdot \mathrm{Var}\left[\bar{w}^{(L)}\right]\right)\left(\frac{1}{2} n_{L-2} \cdot \mathrm{Var}\left[\bar{w}^{(L-1)}\right]\right)\left(\frac{1}{2} n_{L-3} \cdot \mathrm{Var}\left[\bar{w}^{(L-2)}\right]\right) \ldots \left(\frac{1}{2} n_1 \right. \\
&\qquad \left. \cdot \mathrm{Var}\left[\bar{w}^{(2)}\right]\mathrm{Var}\left[\bar{z}^{(1)}\right]\right) \\
&= \left(\frac{1}{2} n_{L-1} \cdot \mathrm{Var}\left[\bar{w}^{(L)}\right]\right)\left(\frac{1}{2} n_{L-2} \cdot \mathrm{Var}\left[\bar{w}^{(L-1)}\right]\right)\left(\frac{1}{2} n_{L-3} \cdot \mathrm{Var}\left[\bar{w}^{(L-2)}\right]\right) \ldots \left(\frac{1}{2} n_1 \right. \\
&\qquad \left. \cdot \mathrm{Var}\left[\bar{w}^{(2)}\right]\right)\mathrm{Var}\left[\bar{z}^{(1)}\right] \\
&= \mathrm{Var}\left[\bar{z}^{(1)}\right]\prod_{l=2}^{L}\frac{1}{2} n_{l-1} \cdot \mathrm{Var}\left[\bar{w}^{(l)}\right].
\end{aligned} \tag{4.52}$$

In a layer $l$, the number of inputs will be the number of neurons of the preceding layer $l-1$. So, we will have $n_{in}^{(l)} = n_{l-1}$. As a result, we have

$$\mathrm{Var}\left[\bar{z}^{(L)}\right] = \mathrm{Var}\left[\bar{z}^{(1)}\right]\prod_{l=2}^{L}\frac{1}{2} n_{in}^{l} \cdot \mathrm{Var}\left[\bar{w}^{(l)}\right]. \tag{4.53}$$

This formula is the one that lets us see what could go wrong without a proper initialization, and thus how to design the right one. The product is key. Indeed, if we have a lot of layers (so if $L$ is large), we see that the variance of the last layer $\bar{z}^{(L)}$ could be very small (if $\frac{1}{2}n_{in}^{l}\cdot\mathrm{Var}\left[\bar{w}^{(l)}\right]$ is below 1) or very large ( if $\frac{1}{2}n_{in}^{l}\cdot\mathrm{Var}\left[\bar{w}^{(l)}\right]$ is above 1). The proper value for what's inside that product should thus be 1, and that is exactly the sufficient condition the Kaiming paper (and initialization) takes:

$$\frac{1}{2}n_{in}^{l}\cdot\mathrm{Var}\left[\bar{w}^{(l)}\right]=1,\forall l. \tag{4.54}$$

The Kaiming paper accordingly suggests to initialize the weights of layer with a zero-mean Gaussian distribution with variance of

$$\mathrm{Var}\left[\bar{w}^{(l)}\right]=\frac{2}{n_{l-1}}=\frac{2}{n_{in}^{l}}, \tag{4.55}$$

and null biases. For the first layer ($l=1$), we should have $n_{in}^{1}\cdot\mathrm{Var}\left[\bar{w}^{(1)}\right]=1$ because there is no ReLU applied on the input signal. But the factor $1/2$ does not matter if it just exists on one layer. So, we also adopt (4.54) in the first layer for simplicity.

### Backward-propagation

The backward-propagation is very similar to the forward-propagation one. The only difference is that we're dealing with gradients, and that the flow is now from the last layers to the first ones. Accordingly, we'll have a very similar but still slightly different result.

The auxiliary variable $\boldsymbol{\delta}^{(l)}$ is commonly used in BP to represent the derivative of the loss ($\mathcal{L}$) with respect to the weighted sum of inputs $\mathbf{z}^{(l)}$ for layer $l$, i.e., the auxiliary variable $\boldsymbol{\delta}^{(l)}$ represents the rate of change of the loss with respect to the pre-activation output of layer $l$. Here, $l$ can take values from 0 to the final layer $L$. In mathematical terms:

$$\begin{aligned}\delta_j^{(l)}&=\frac{\partial\mathcal{L}}{\partial z_j^{(l)}}\\&=\sum_k w_{kj}^{(l+1)}\frac{\partial\mathcal{L}}{\partial z_k^{(l+1)}}\sigma'\left(z_j^{(l)}\right)\\&=\sum_k w_{kj}^{(l+1)}\delta_k^{(l+1)}\sigma'\left(z_j^{(l)}\right),\end{aligned} \tag{4.56.1}$$

and

$$\begin{aligned}\underbrace{\boldsymbol{\delta}^{(l)}}_{n_l\times1}&=\frac{\partial\mathcal{L}}{\partial\mathbf{z}^{(l)}}\\&=\left(\underbrace{\left(\mathbf{W}^{(l+1)}\right)^T}_{n_l\times n_{l+1}}\underbrace{\frac{\partial\mathcal{L}}{\partial\mathbf{z}^{(l+1)}}}_{n_{l+1}\times1}\right)\odot\underbrace{\sigma'\left(\underbrace{\mathbf{z}^{(l)}}_{n_l\times1}\right)}_{n_l\times1}\\&=\underbrace{\left(\underbrace{\left(\mathbf{W}^{(l+1)}\right)^T}_{n_l\times n_{l+1}}\underbrace{\boldsymbol{\delta}^{(l+1)}}_{n_{l+1}\times1}\right)}_{n_l\times1}\odot\underbrace{\sigma'\left(\underbrace{\mathbf{z}^{(l)}}_{n_l\times1}\right)}_{n_l\times1}.\end{aligned} \tag{4.56.2}$$

Then we have:

$$\mathrm{Var}\left[\bar{\delta}_j^{(l)}\right]=\mathrm{Var}\left[\sum_{k=1}^{n_{l+1}}\bar{w}_{kj}^{(l+1)}\bar{\delta}_k^{(l+1)}\sigma'\left(\bar{z}_j^{(l)}\right)\right]. \tag{4.57}$$

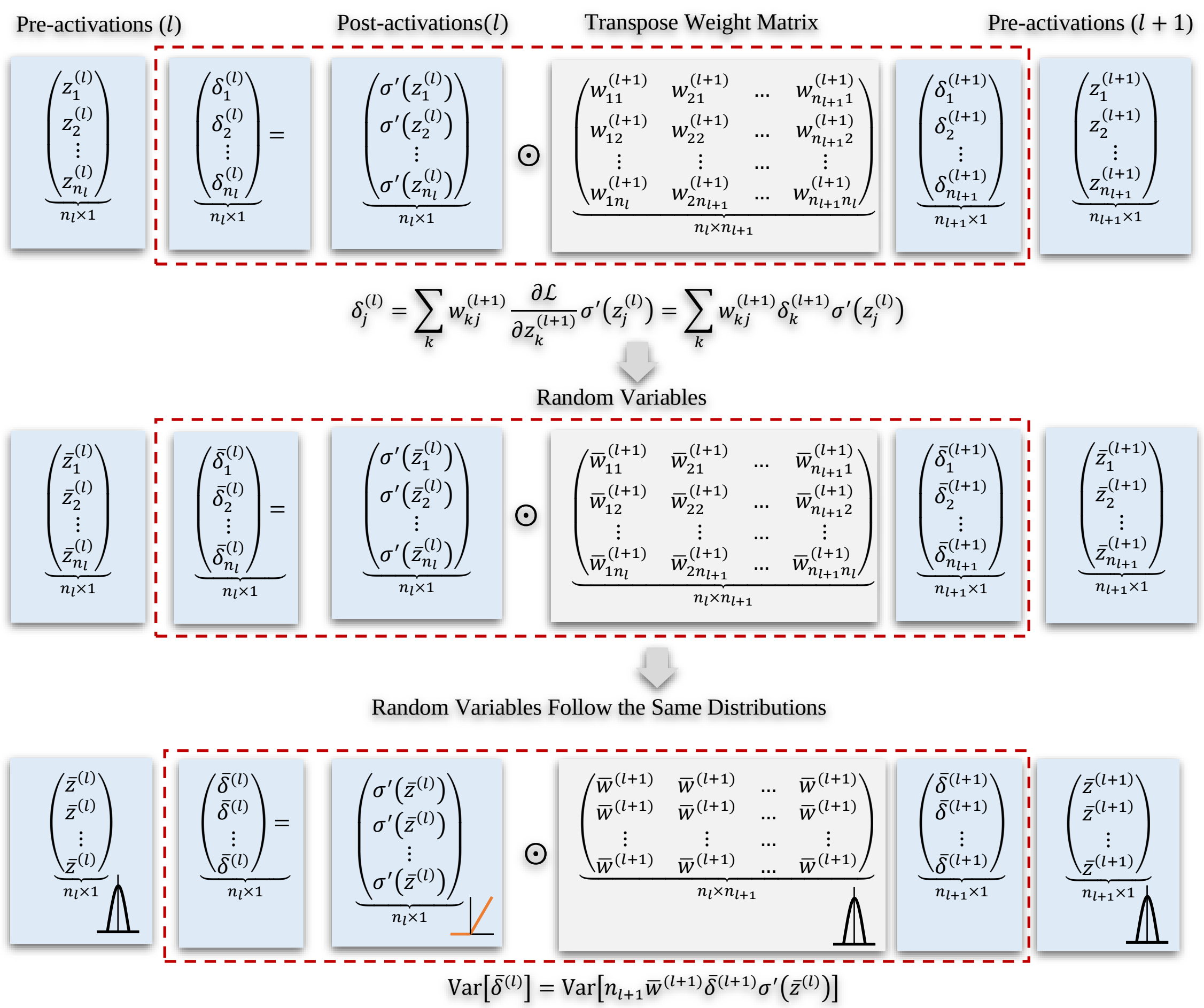

**Figure 4.10.** Diagrammatic representation of NN architecture and vector notations across two layers $(l, l+1)$. This visualization illustrates the flow of information through these layers, showcasing the vectors utilized in the computations. The elements in $\boldsymbol{\delta}^{(l+1)}$ are mutually independent and share the same distribution. Similarly, the initialized elements in $\left(\mathbf{W}^{(l+1)}\right)^T$ are mutually independent and share the same distribution. Random variables of pre-activation, post-activation, auxiliary variable and weights are distinctly marked with a bar placed above the variable symbol, serving to differentiate them from deterministic variables within the network.

We also make very similar assumptions to the forward-propagation case: $\left(\mathbf{W}^{(l+1)}\right)^T$ and $\boldsymbol{\delta}^{(l+1)}$ contain random variables that have the same distribution, respectively. Accordingly, we note $\bar{w}^{(l+1)}$ and $\bar{\delta}^{(l+1)}$ random variables that follow the same distribution as the ones in $\left(\mathbf{W}^{(l+1)}\right)^T$ and $\boldsymbol{\delta}^{(l+1)}$, respectively, i.e., $\bar{\delta}_j^{(l)} = \bar{\delta}^{(l)}$, $\bar{w}_{kj}^{(l+1)} = \bar{w}^{(l+1)}$, $\bar{\delta}_k^{(l+1)} = \bar{\delta}^{(l+1)}$, and $\bar{z}_j^{(l)} = \bar{z}^{(l)}$, see Figure 4.10. We also assume that $\bar{w}^{(l+1)}$ and $\bar{\delta}^{(l+1)}\sigma'(\bar{z}^{(l)})$ are independent of each other. Finally, $\bar{w}^{(l+1)}$ has a symmetric distribution around 0 so $\bar{\delta}^{(l)}$ has zero mean for all $l$.

$$
\begin{aligned}
\mathbb{E}\left[\bar{\delta}^{(l)}\right] &= \mathbb{E}\left[\sum_{k=1}^{n_{l+1}} \bar{w}^{(l+1)}\bar{\delta}^{(l+1)}\sigma'\left(\bar{z}^{(l)}\right)\right] \\
&= \mathbb{E}\left[n_{l+1}\bar{w}^{(l+1)}\bar{\delta}^{(l+1)}\sigma'\left(\bar{z}^{(l)}\right)\right] \\
&= n_{l+1}\mathbb{E}\left[\bar{w}^{(l+1)}\bar{\delta}^{(l+1)}\sigma'\left(\bar{z}^{(l)}\right)\right] \\
&= n_{l+1}\mathbb{E}\left[\bar{w}^{(l+1)}\right]\mathbb{E}\left[\bar{\delta}^{(l+1)}\sigma'\left(\bar{z}^{(l)}\right)\right] \\
&= 0.
\end{aligned} \tag{4.58}
$$

We can now further simplify (4.57) as follow:

$$\begin{aligned}\text{Var}\left[\bar{\delta}^{(l)}\right] &= \text{Var}\left[\sum_{k=1}^{n_{l+1}} \bar{w}^{(l+1)}\bar{\delta}^{(l+1)}\sigma'\left(\bar{z}^{(l)}\right)\right] \\ &= \text{Var}\left[n_{l+1}\bar{w}^{(l+1)}\bar{\delta}^{(l+1)}\sigma'\left(\bar{z}^{(l)}\right)\right] \\ &= n_{l+1}\text{Var}\left[\bar{w}^{(l+1)}\bar{\delta}^{(l+1)}\sigma'\left(\bar{z}^{(l)}\right)\right].\end{aligned} \tag{4.59}$$

In the case of ReLU, which we use as the AF here, the derivative is:

$$\text{ReLU}'(x) = \begin{cases}1, & x \geq 0 \\ 0, & x < 0\end{cases}. \tag{4.60}$$

One last assumption: $\sigma'\left(\bar{z}^{(l)}\right)$ and $\bar{\delta}^{(l+1)}$ are independent from each other. Thus, taking into account that ReLU 'halves' its input, at least if the input is symmetric around 0, we have:

$$\mathbb{E}\left[\bar{\delta}^{(l+1)}\sigma'\left(\bar{z}^{(l)}\right)\right] = \frac{1}{2}\mathbb{E}\left[\bar{\delta}^{(l+1)}\right] = 0, \tag{4.61}$$

and because $\bar{\delta}^{(l+1)}\sigma'\left(\bar{z}^{(l)}\right)$ has a zero mean, and that $\left(\sigma'\left(\bar{z}^{(l)}\right)\right)^2 = \sigma'\left(\bar{z}^{(l)}\right)$, by taking the expectation of the square of the equation $\bar{\delta}^{(l+1)}\sigma'\left(\bar{z}^{(l)}\right)$, we get:

$$\begin{aligned}\mathbb{E}\left[\left(\bar{\delta}^{(l+1)}\sigma'\left(\bar{z}^{(l)}\right)\right)^2\right] &= \text{Var}\left[\bar{\delta}^{(l+1)}\sigma'\left(\bar{z}^{(l)}\right)\right] \\ &= \mathbb{E}\left[\left(\bar{\delta}^{(l+1)}\right)^2\left(\sigma'\left(\bar{z}^{(l)}\right)\right)^2\right] \\ &= \frac{1}{2}\mathbb{E}\left[\left(\bar{\delta}^{(l+1)}\right)^2\right] = \frac{1}{2}\text{Var}\left[\bar{\delta}^{(l+1)}\right].\end{aligned} \tag{4.62}$$

To remind you, we do have $\mathbb{E}\left[\left(\bar{\delta}^{(l+1)}\right)^2\right] = \text{Var}\left[\bar{\delta}^{(l+1)}\right]$ because $\bar{\delta}^{(l+1)}$ has zero mean as we just demonstrated.

We now have everything we need to compute the variance of (4.59), with the exact same reasoning we had with the forward case:

$$\text{Var}[XY] = \mathbb{E}[X^2]\mathbb{E}[Y^2] - (\mathbb{E}[X])^2(\mathbb{E}[Y])^2, \tag{4.63}$$

and

$$\begin{aligned}\text{Var}\left[\bar{\delta}^{(l)}\right] &= n_{l+1}\text{Var}\left[\bar{w}^{(l+1)}\bar{\delta}^{(l+1)}\sigma'\left(\bar{z}^{(l)}\right)\right] \\ &= n_{l+1}\left\{\underbrace{\mathbb{E}\left[\left(\bar{w}^{(l+1)}\right)^2\right]}_{\text{Var}\left[\bar{w}^{(l+1)}\right]}\underbrace{\mathbb{E}\left[\left(\bar{\delta}^{(l+1)}\sigma'\left(\bar{z}^{(l)}\right)\right)^2\right]}_{\frac{1}{2}\text{Var}\left[\bar{\delta}^{(l+1)}\right]} - \left(\underbrace{\mathbb{E}\left[\bar{w}^{(l+1)}\right]}_{0}\right)^2\left(\mathbb{E}\left[\bar{\delta}^{(l+1)}\sigma'\left(\bar{z}^{(l)}\right)\right]\right)^2\right\} \\ &= \frac{1}{2}n_{l+1}\text{Var}\left[\bar{w}^{(l+1)}\right]\text{Var}\left[\bar{\delta}^{(l+1)}\right].\end{aligned} \tag{4.64}$$

As before, we have a recurrence equation. This time it's on $\bar{\delta}^{(l)}$, but we can turn it into a product over all the $L$ layers all the same:

$$\begin{aligned}\text{Var}\left[\bar{\delta}^{(1)}\right] &= \left(\frac{1}{2}n_2\text{Var}\left[\bar{w}^{(2)}\right]\right)\text{Var}\left[\bar{\delta}^{(2)}\right] \\ &= \left(\frac{1}{2}n_2\text{Var}\left[\bar{w}^{(2)}\right]\right)\left(\frac{1}{2}n_3\text{Var}\left[\bar{w}^{(3)}\right]\right)\text{Var}\left[\bar{\delta}^{(3)}\right] \\ &= \left(\frac{1}{2}n_2\text{Var}\left[\bar{w}^{(2)}\right]\right)\left(\frac{1}{2}n_3\text{Var}\left[\bar{w}^{(3)}\right]\right)\left(\frac{1}{2}n_4\text{Var}\left[\bar{w}^{(4)}\right]\right)\text{Var}\left[\bar{\delta}^{(4)}\right] \\ &= \left(\frac{1}{2}n_2\text{Var}\left[\bar{w}^{(2)}\right]\right)\left(\frac{1}{2}n_3\text{Var}\left[\bar{w}^{(3)}\right]\right)\left(\frac{1}{2}n_4\text{Var}\left[\bar{w}^{(4)}\right]\right)\ldots\left(\frac{1}{2}n_L\text{Var}\left[\bar{w}^{(L)}\right]\right)\text{Var}\left[\bar{\delta}^{(L)}\right] \\ &= \text{Var}\left[\bar{\delta}^{(L)}\right]\prod_{l=2}^{L}\frac{1}{2}n_l\text{Var}\left[\bar{w}^{(l)}\right].\end{aligned} \tag{4.65}$$

Again, as before, this product is key to understanding why the right initialization is so important: if not set carefully, the gradient can explode or vanish, depending on whether $\frac{1}{2}n_l\mathrm{Var}\left[\overline{w}^{(l)}\right]$ is over or below 1 (strictly). Setting that number to 1 is important:

$$\frac{1}{2}n_l\mathrm{Var}\left[\overline{w}^{(l)}\right] = 1, \forall l. \tag{4.66}$$

In a layer $l$, the number of outputs will be the number of neurons in the current layer $l$. So, we will have $n_{out}^{(l)} = n_l$. As a result, we have

$$\mathrm{Var}\left[\overline{w}^{(l)}\right] = \frac{2}{n_l} = \frac{2}{n_{out}^{(l)}}. \tag{4.67}$$

It is important to note that it is sufficient to use either (4.55) or (4.67) alone.

A random variable that is uniformly distributed in an interval centered around zero has zero mean. So, we can sample the weights from a uniform distribution with variance $\frac{2}{n_l}$. But how do we find the endpoints of the interval? A continuous random variable $A$ that is uniformly distributed in the interval $[-a, a]$ has zero mean and variance of $a^2/3$. We can also draw the weights from a uniform distribution for Kaiming initialization. Using

$$\mathrm{Var}[w] = \frac{a^2}{3}, \tag{4.68}$$

and,

$$\mathrm{Var}[w] = \frac{2}{n_{l-1}}, \tag{4.69}$$

we have

$$\frac{a^2}{3} = \frac{2}{n_{l-1}}, \qquad a = \pm\sqrt{\frac{6}{n_{l-1}}}. \tag{4.70}$$

For a layer that uses a ReLU AF, the Kaiming initialization with a uniform distribution, $U$, scales the weights by

$$w \sim U\left[-\sqrt{\frac{6}{n_{l-1}}}, \sqrt{\frac{6}{n_{l-1}}}\right]. \tag{4.71}$$

### 4.4.5 Xavier/Glorot Initialization

The Xavier and Kaiming papers follow a very similar reasoning, that differs a tiny bit at the end. The only difference is that the Kaiming paper takes into account the AF, whereas Xavier does not. As the AF was assumed to be linear by Xavier (or at least approximated to be linear with a derivative of 1 around 0), it's not taken into account, and thus the $1/2$ that comes from ReLU isn't there. The Xavier initialization formula [95] in the forward case is hence:

$$\forall l, \qquad n_{l-1}\mathrm{Var}\left[\overline{w}^{(l)}\right] = 1. \tag{4.72}$$

In the backward case, it is:

$$\forall l, \qquad n_l\mathrm{Var}\left[\overline{w}^{(l)}\right] = 1. \tag{4.73}$$

Note how both constraints are satisfied when all layers have the same width. As a compromise between these two constraints, we might want to have

$$\forall l, \qquad \mathrm{Var}\left[\overline{w}^{(l)}\right] = \frac{2}{n_{l-1} + n_l}. \tag{4.74}$$

We can also draw the weights from a uniform distribution for Xavier initialization. Using

$$\mathrm{Var}[w] = \frac{(2a)^2}{12} = \frac{a^2}{3}, \tag{4.75}$$

and

$$\mathrm{Var}[w] = \frac{2}{n_{l-1} + n_l}, \tag{4.76}$$

we have,

$$\frac{a^2}{3} = \frac{2}{n_{l-1} + n_l}, \qquad a = \pm\sqrt{\frac{6}{n_{l-1} + n_l}}. \tag{4.77}$$

For a layer that uses a tanh AF, the Kaiming initialization with a uniform distribution scales the weights by

$$w \sim U\left[-\sqrt{\frac{6}{n_{l-1} + n_l}}, \sqrt{\frac{6}{n_{l-1} + n_l}}\right] = U\left[-\sqrt{\frac{6}{n_{in}^{(l)} + n_{out}^{(l)}}}, \sqrt{\frac{6}{n_{in}^{(l)} + n_{out}^{(l)}}}\right]. \tag{4.78}$$

In discussing initialization strategies, it is crucial to touch upon sparse initialization. Sparse initialization presents an alternative method to weight initialization that mitigates some of the drawbacks of scaling rules such as $\sqrt{1/n}$, especially in large networks. Instead of scaling all weights uniformly, sparse initialization aims to maintain a certain number of non-zero weights per unit, regardless of the size of the layer. Sparse initialization can help in achieving more diversity among the units at initialization time, as each unit can have a unique set of non-zero weights. By keeping the number of non-zero weights constant per unit, sparse initialization prevents individual weight magnitudes from becoming extremely small, which can help in maintaining gradient flow during training. Sparse initialization imposes a strong prior on the weights, particularly those chosen to have large Gaussian values. This can affect the expressiveness of the network and might not be suitable for all types of architectures or tasks.

## 4.5 Non-zero Centered Activation Function

Consider a NN with activation of node $j$ given by

$$a_j = \sigma_j(z_j);\ z_j = \sum_{i \in A_j} w_{ij}\, a_i, \tag{4.79}$$

where $\sigma_j$ is a nonlinear AF, $w_{ij}$ are the synaptic weights, and $A_j$ denotes the set of anterior nodes feeding their activity $a_i$ into node $j$. The weights $w_{ij}$ of the NN given are typically optimized by GD in some cost function $\mathcal{L}$. With a local step size $\alpha_{ij}$ for each weight, this results in the weight update equation [107]

$$\begin{aligned} \Delta w_{ij} &= -\alpha_{ij}\frac{\partial \mathcal{L}}{\partial w_{ij}} \\ &= -\alpha_{ij}\frac{\partial \mathcal{L}}{\partial z_j}\frac{\partial z_j}{\partial w_{ij}} \\ &= \alpha_{ij}\delta_j a_i, \qquad\qquad \delta_j = -\frac{\partial \mathcal{L}}{\partial z_j}. \end{aligned} \tag{4.80}$$

**Theorem 4.1:** Convergence is usually faster if the average of each input variable over the training set is close to zero.

To see this, consider the extreme case where all the inputs are positive [107]. Weights to a particular node in the first weight layer are updated by an amount proportional to $\delta \cdot \mathbf{x}$ where $\delta$ is the (scalar) error at that node and $\mathbf{x}$ is the input vector (see (4.79) and (4.80)). When all of the components of an input vector are positive, all of the updates of weights that feed into a node will be the same sign (i.e. sign($\delta$)). As a result, these weights can only all decrease or all increase together for a given input pattern. However, the ideal gradient weight update might be one where some weights increase while the other weights decrease. If the gradient update is positive, these weights become too positive in the current iteration. In the next iteration, the gradient may be negative as well as large to remedy these increased weights, which might end up overshooting the weights. This can cause a zig zag patter in search of minima, which can slow down the training.

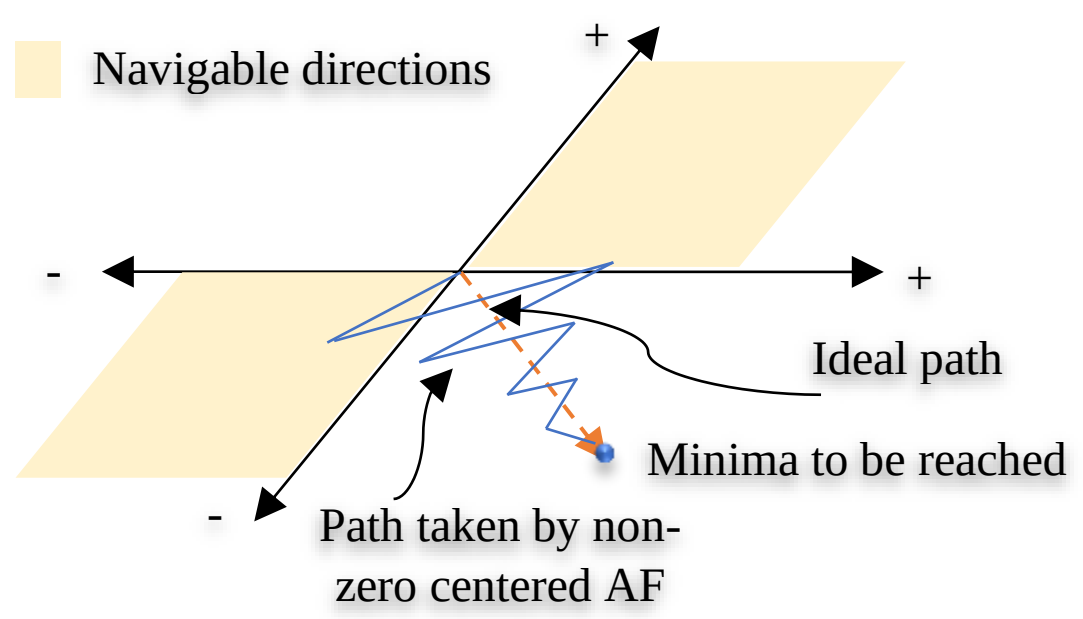

**Figure 4.11.** Zigzagging behavior can slow down convergence.

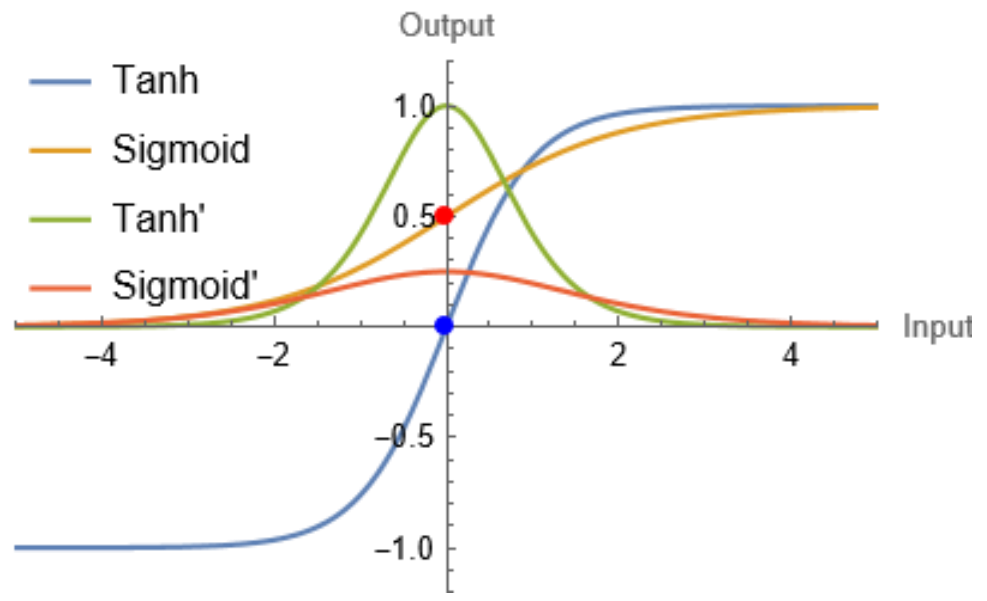

**Figure 4.12.** Comparison of zero-centered (Tanh) and non-zero-centered (sigmoid) AFs and their derivatives.

In the above example, the inputs were all positive. However, in general, any shift of the average input away from zero will bias the updates in a particular direction and thus slow down learning. Therefore, it is good to shift the inputs so that the average over the training set is close to zero. This heuristic should be applied at all layers which means that we want the average of the outputs of a node to be close to zero because these outputs are the inputs to the next layer. This is not possible with non-zero centered activation.

Let us visualize this in a simple two-dimensional weight space (see Figure 4.11). The weight vector $\mathbf{w} = (w_1, w_2)$ can only change direction by zigzagging. Since, both $w_1$ and $w_2$ will be updated in the same direction at each step (these weights can only all positive or all negative together for a given input pattern). If the weights are consistently updated in the same direction, the weight vector will keep moving along a straight line. To change direction, it needs to oscillate back and forth between the positive and negative weight vector. This zigzagging behavior can slow down convergence.

**Definition (Zero-Centered AF)**: A zero-centered AF is one where the average value of the function is around zero. In other words, when you take the sum of all the output values of the AF over a large dataset, it should be close to zero.

By having a zero-centered AF, Figure 4.12, the positive and negative values in the output have a balanced distribution around zero. This helps to avoid the problem of consistently positive or negative gradients. It can contribute to more stable and efficient training of NNs, as the weight updates will not consistently push the weights in one direction.

## 4.6 Feature Scaling

In addition to initializing the weights properly, it is important to preprocess the input data. Feature scaling is a preprocessing step commonly applied to the input features of a dataset before feeding them into a machine learning algorithm. It involves transforming the features so that they have a similar scale or distribution. The feature processing methods used for NN training are not very different from those in other machine learning algorithms.

There are several reasons why feature scaling is important:

- Feature scaling ensures that all features are on a similar scale, preventing certain features from dominating others solely based on their magnitude. When features have vastly different scales, the optimization process can be skewed towards features with larger magnitudes, causing it to take longer to reach the minimum. This normalization makes the algorithm treat all features equally during training, preventing bias towards features with larger scales.
- Many machine learning algorithms, such as GD -based optimization methods, converge faster when features are scaled. Without scaling, the optimization process may take longer to reach convergence or may even fail to converge. Scaling facilitates a smoother and more efficient optimization process.
- Properly scaled features can lead to improved model performance. Scaling can help the algorithm better identify meaningful patterns in the data, leading to more accurate predictions. It can also prevent numerical instability issues that may arise when working with features of vastly different scales.
- Scaling features can enhance the interpretability of the model. When features are on a similar scale, it becomes easier to interpret the importance of each feature based on their coefficients or feature importance scores. This facilitates a better understanding of the model's decision-making process.
- Feature scaling aids in stabilizing regularization techniques, such as L1 or L2 regularization. Regularization penalizes large coefficients in the model, and features with larger scales may receive disproportionately higher penalties. Scaling ensures that regularization is applied uniformly across all features, preventing bias towards certain features.
- Feature scaling is essential for ensuring the stability, efficiency, and effectiveness of machine learning algorithms across various tasks and datasets. It helps address issues related to data variability, numerical stability, and algorithm sensitivity, ultimately leading to more reliable and accurate models.

The importance of feature scaling can be illustrated by a simple example. Suppose we have two features:

- Feature 1: Measured on a scale from 1 to 10.
- Feature 2: Measured on a scale from 1 to 100,000.

Consider a simple linear regression model with the squared error loss function. When we compute the error for each data point, the error contribution from Feature 2 will be much larger compared to Feature 1 due to its larger scale. Consequently, during the optimization process (e.g., GD), the algorithm will prioritize minimizing the errors associated with Feature 2 over those associated with Feature 1. This behavior can lead to several issues:

1. The algorithm may focus more on reducing errors associated with Feature 2, potentially neglecting the errors associated with Feature 1. As a result, the model may not effectively capture the relationship between Feature 1 and the target variable.
2. Because the algorithm is primarily focused on minimizing errors associated with Feature 2, it may take longer to converge to the optimal solution, especially if the errors associated with Feature 2 are large.
3. Even if the algorithm converges, the resulting model may have suboptimal performance, as it may not effectively utilize the information from Feature 1 due to its smaller scale.

By scaling both features to a similar range (e.g., between 0 and 1 or standardizing them with a mean of 0 and a standard deviation of 1), we can address these issues:

1. Scaling ensures that both features contribute equally to the optimization process. The algorithm will not prioritize one feature over the other based solely on its scale.

2. With scaled features, the optimization process is likely to converge faster since the errors associated with all features are on a similar scale.
3. A model trained on scaled features is more likely to capture the relationships between all features and the target variable effectively, leading to better overall performance.

Centering and scaling are fundamental techniques used in feature scaling. Centering involves transforming the feature values so that they are centered around a specific point. This point is often the origin $(0)$ or the mean of the feature values. Centering around the mean involves subtracting the mean of the feature from each data point. This ensures that the mean of the feature becomes 0. Centering is useful for removing biases in the data and making it more symmetric around a central point. Scaling refers to rescaling the feature values so that they have a consistent scale or variance across the dataset. One common scaling technique is to rescale the feature values such that they have a variance of 1. Another scaling technique is to rescale the feature values such that they fall within a specific range, such as $[0, 1]$ or $[-1, 1]$.

When the variances of some features are much larger than others, it can lead to instability during optimization. Features with larger variances may dominate the learning process, causing the optimization algorithm to prioritize minimizing errors associated with those features over others. This can result in slow convergence or oscillations in the optimization process. Standardization, by scaling the features to have a variance of 1, ensures that all features contribute equally to the optimization process. It mitigates the problem of varying sensitivity of the loss function to different weights, promoting more stable and efficient optimization.

Centering the features around the mean ensures that the feature values have different signs. This is beneficial because it prevents all weights pointing into the same neuron from being constrained to move in the same direction during optimization. When feature values have different signs, it allows the weights associated with those features to move independently and adjust in different directions. This helps avoid the "zigzagging" behavior, where weights oscillate back and forth during optimization, which can slow down convergence and hinder the learning process.

By combining standardization (scaling to equalize variances) and centering (ensuring different signs of feature values), machine learning algorithms can achieve more stable and efficient optimization, leading to faster convergence and better performance on the task at hand.

Normalization and standardization are two common techniques used to bring different features onto the same scale, but they serve slightly different purposes and operate in different ways.

### 4.6.1 Normalization (Min-Max Scaling)

Normalization typically refers to rescaling the features to a range of $[0,1]$. This is achieved by subtracting the minimum value of the feature and then dividing by the range (i.e., the maximum value minus the minimum value). Mathematically, the new value $x^{(i)}_{\text{normalized}}$ of a sample $x^{(i)}$ can be calculated as follows [56]:

$$x^{(i)}_{\text{normalized}} = \frac{x^{(i)} - \min(x)}{\max(x) - \min(x)}, \tag{4.81}$$

where $\min(x)$ and $\max(x)$ are the smallest and the largest value in a feature column, respectively. This ensures that all feature values fall within the $[0,1]$ range, making it suitable for algorithms that require input features to be on a similar scale, such as NNs.

Let $x$ be a feature with $n$ data points. The original feature values are denoted as $x^{(1)}, x^{(2)}, \ldots, x^{(n)}$. Min-max scaling (normalization) transforms each original feature value $x^{(i)}$ to its corresponding normalized value $x^{(i)}_{\text{normalized}}$ using the (4.81). Now, let's consider the extreme cases: When $x^{(i)} = \min(x)$, we have:

$$x^{(i)}_{\text{normalized}} = \frac{\min(x) - \min(x)}{\max(x) - \min(x)} = 0. \tag{4.82}$$

**Table 4.1.** Comparison of Feature Scaling Techniques: Normalization (Min-Max Scaling) vs. Standardization (Z-score Normalization) for a Sample Dataset of Numbers 0 to 5.

| input | Standardized | Normalized |
|---|---|---|
| 0.0 | −1.33631 | 0.0 |
| 1.0 | −0.801784 | 0.2 |
| 2.0 | −0.267261 | 0.4 |
| 3.0 | 0.267261 | 0.6 |
| 4.0 | 0.801784 | 0.8 |
| 5.0 | 1.33631 | 1.0 |

When $x^{(i)} = \max(x)$, we have:

$$x^{(i)}_{\text{normalized}} = \frac{\max(x) - \min(x)}{\max(x) - \min(x)} = 1. \tag{4.83}$$

Hence, $x^{(i)}_{\text{normalized}}$ is bounded between 0 and 1 for all $i$ (where $1 \leq i \leq n$).

Min-max scaling is more sensitive to outliers, because a single outlier (small or large) affects the min-max values and can have a big effect on the normalization.

### 4.6.2 Standardization (Z-score Normalization)

Standardization, on the other hand, transforms the features to have a mean of 0 and a standard deviation of 1. To do this, first we need to calculate the mean and standard deviation for each feature. Standardization is achieved by subtracting the mean of the feature and then dividing by the standard deviation. Mathematically, for each feature [56]:

$$x^{(i)}_{\text{standardized}} = \frac{x^{(i)} - \mu_x}{\sigma_x}, \tag{4.84}$$

where, $\mu_x$ is the sample mean of a particular feature column and $\sigma_x$ is the corresponding standard deviation. Equation (4.84) will center the feature value on 0 with a standard deviation of 1.0, producing values in the range of $[-1, 1]$. Table 4.1 illustrates the difference between standardization and normalization for a simple sample dataset consisting of numbers 0 to 5.

Let's denote the original data point along the $j$th feature dimension as $x_{ij}$, where $i$ represents the sample index. After subtracting the mean $\mu_j$ from each data point, the centered data point $\hat{x}_{ij}$ can be expressed as:

$$\hat{x}_{ij} = x_{ij} - \mu_j. \tag{4.85}$$

Now, let's calculate the mean of the centered data along the $j$th feature dimension:

$$\text{Mean}(\hat{x}_j) = \frac{1}{n}\sum_{i=1}^{n} \hat{x}_{ij}. \tag{4.86}$$

Substituting the expression for $\hat{x}_{ij}$, we get:

$$\begin{aligned}\text{Mean}(\hat{x}_j) &= \frac{1}{n}\sum_{i=1}^{n}(x_{ij} - \mu_j)\\ &= \frac{1}{n}\sum_{i=1}^{n} x_{ij} - \frac{1}{n}\sum_{i=1}^{n}\mu_j\\ &= \mu_j - \frac{1}{n}n\mu_j\\ &= 0.\end{aligned} \tag{4.87}$$

Therefore, the mean of the centered data along the $j$th feature dimension is indeed zero.

Now, let's prove that dividing each centered feature by its standard deviation results in unit variance. For our centered feature $\hat{x}_{ij}$, its variance $\mathrm{Var}[\hat{x}_{ij}]$ can be expressed as:

$$\mathrm{Var}[\hat{x}_{ij}] = \mathbb{E}\left[\left(\hat{x}_{ij} - \mu_j\right)^2\right]. \tag{4.88}$$

Since we've centered the data, $\mu_j = 0$, and we have:

$$\mathrm{Var}[\hat{x}_{ij}] = \mathbb{E}[\hat{x}_{ij}{}^2]. \tag{4.89}$$

The standard deviation $\sigma_j$ of the centered feature is the square root of its variance:

$$\sigma_j = \sqrt{\mathrm{Var}[\hat{x}_{ij}]} = \sqrt{\mathbb{E}[\hat{x}_{ij}{}^2]}. \tag{4.90}$$

Now, if we divide each centered feature by its standard deviation $\sigma_j$, we get:

$$\bar{x}_{ij} = \frac{\hat{x}_{ij}}{\sigma_j}. \tag{4.91}$$

Let's calculate the variance of $\bar{x}_{ij}$:

$$\mathrm{Var}[\bar{x}_{ij}] = \mathbb{E}\left[\left(\frac{\hat{x}_{ij}}{\sigma_j}\right)^2\right] = \mathbb{E}\left[\frac{\hat{x}_{ij}^2}{\sigma_j^2}\right] = \frac{1}{\sigma_j^2}\mathbb{E}[\hat{x}_{ij}^2] = \frac{1}{\sigma_j^2}\mathrm{Var}[\hat{x}_{ij}] = \frac{1}{\sigma_j^2}\sigma_j^2 = 1. \tag{4.92}$$

Therefore, dividing each centered feature by its standard deviation indeed results in unit variance.

Standardization is indeed a practical and widely used technique in machine learning for several reasons:

- By centering the feature columns at mean 0, standardization ensures that the mean of each feature is effectively 0. This is beneficial for algorithms that assume zero-centered data, such as principal component analysis or algorithms based on GD optimization, as it helps prevent biases in the model.
- Standardization also scales the feature columns to have a standard deviation of 1. This results in features that are on a comparable scale, making it easier for the algorithm to learn the weights during training. It helps stabilize the optimization process, particularly in algorithms that use gradient-based optimization techniques.
- Many machine learning models, especially those using GD optimization, initialize the weights to small random values close to 0. By standardizing the features to have a normal distribution, it becomes easier for the algorithm to learn meaningful patterns in the data and adjust the weights accordingly.
- Standardization maintains useful information about outliers by preserving the spread of the data. Since it scales features based on their standard deviation, outliers have less influence on the scaled values compared to min-max scaling. This makes the algorithm less sensitive to outliers and improves its robustness.

### 4.6.3 Covariance matrix and whitening

Covariance measures the degree to which two random variables change together [1]. When the covariance between two variables is positive, it indicates that as one random variable tends to increase, the other random variable also tends to increase, and vice versa. This suggests a positive linear relationship between the variables. Conversely, when the covariance is negative, it means that as one random variable tends to increase, the other random variable tends to decrease, and vice versa. This indicates a negative linear relationship between the variables. The sign of the covariance thus provides information about the direction or tendency of the linear relationship between the variables. The covariance between two random variables $X$ and $Y$ with finite second moments is defined as the expected value (or mean) of the product of their deviations from their individual expected values.

**Definition (Covariance):** The covariance of $X$ and $Y$, denoted by $\mathrm{Cov}(X,Y)$ or $\sigma_{XY}$, is defined by [1]

$$\begin{aligned}\mathrm{Cov}(X,Y) &= \sigma_{XY}\\ &= \mathbb{E}[(X-\mathbb{E}[X])(Y-\mathbb{E}[Y])]\\ &= \mathbb{E}[XY]-\mathbb{E}[X]\mathbb{E}[Y].\end{aligned} \tag{4.93}$$

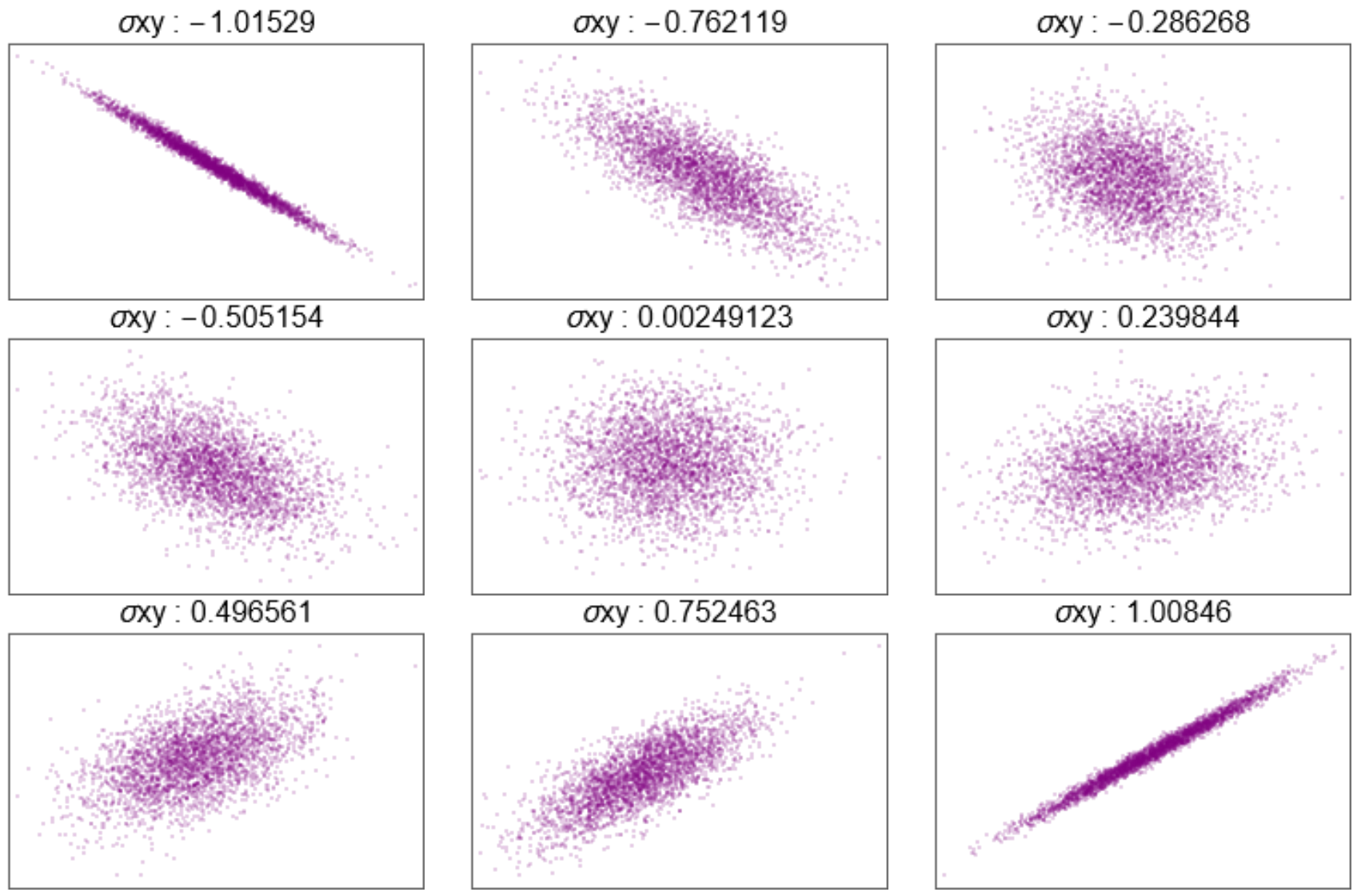

**Figure 4.13.** The sign of the covariance of two RVs $X$ and $Y$.

**Proof:**

$$\begin{aligned}
\mathrm{Cov}(X,Y) &= \mathbb{E}[(X-\mathbb{E}[X])(Y-\mathbb{E}[Y])] \\
&= \mathbb{E}\big[XY-\mathbb{E}[X]Y-X\mathbb{E}[Y]+\mathbb{E}[X]\mathbb{E}[Y]\big] \\
&= \mathbb{E}[XY]-\mathbb{E}[X]E[Y]-\mathbb{E}[X]\mathbb{E}[Y]+\mathbb{E}[X]\mathbb{E}[Y] \\
&= \mathbb{E}[XY]-\mathbb{E}[X]\mathbb{E}[Y].
\end{aligned}$$

■

The resulting covariance value can provide insight into the relationship between the variables:

- A positive covariance ($\mathrm{Cov} > 0$) indicates that when $X$ tends to be above its mean ($X-\mu_X > 0$), $Y$ tends to be above its mean ($Y-\mu_Y > 0$), and vice versa. It suggests a positive relationship between the variables, meaning that they tend to move together in the same direction. For example, if $X$ represents hours of study and $Y$ represents exam scores, a positive covariance would imply that as study time increases, exam scores also tend to increase. See Figure 4.13.
- A negative covariance ($\mathrm{Cov} < 0$) implies that when $X$ tends to be above its mean, $Y$ tends to be below its mean (and vice versa). It indicates a negative relationship between the variables, suggesting that they tend to move in opposite directions. For instance, in the context of temperature and sales of winter clothing, a negative covariance would indicate that as temperatures rise ($X$ increases), sales of winter clothing decrease ($Y$ decreases).
- A covariance of zero ($\mathrm{Cov} = 0$) suggests that changes in one variable are unrelated to changes in the other.

**Definition (Uncorrelated RVs):** If $\mathrm{Cov}(X,Y) = 0$, then we say that $X$ and $Y$ are uncorrelated. From (4.93), we see that $X$ and $Y$ are uncorrelated if

$$\mathbb{E}[XY] = \mathbb{E}[X]\mathbb{E}[Y]. \tag{4.94}$$

Note that, the variance is a special case of the covariance in which the two variables are identical (that is, in which one variable always takes the same value as the other):

$$\begin{aligned}
\mathrm{Cov}(X,X) &= \mathbb{E}[XX]-\mathbb{E}[X]\mathbb{E}[X] \\
&= \mathbb{E}[X^2]-(\mathbb{E}[X])^2 \\
&= \sigma_X^2 \\
&= \mathrm{Var}(X).
\end{aligned} \tag{4.95}$$

In the following discuss, boldfaced unsubscripted $\boldsymbol{X}$ and $\boldsymbol{Y}$ are used to refer to random vectors, and Roman subscripted $X_i$ and $Y_i$ are used to refer to scalar random variables. If the entries in the column vector

$$\boldsymbol{X} = |\boldsymbol{X}\rangle = (X_1, X_2, \dots, X_n)^T, \tag{4.96}$$

are random variables, each with finite variance and expected value, then the covariance matrix $\boldsymbol{\Sigma}$ is the matrix whose $(i, j)$ entry is the covariance

$$\begin{aligned}\Sigma_{ij} &= \mathrm{Cov}(X_i, X_j) \\ &= \mathbb{E}\big[(X_i - \mathbb{E}[X_i])(X_j - \mathbb{E}[X_j])\big].\end{aligned} \tag{4.97}$$

It is the matrix of covariances between the scalar components of the vector $\boldsymbol{X}$. The covariance matrix between two vectors is

$$\begin{aligned}\mathrm{Cov}(\boldsymbol{X}, \boldsymbol{Y}) &= \boldsymbol{\Sigma}_{\boldsymbol{XY}} \\ &= \mathbb{E}[(\boldsymbol{X} - \mathbb{E}[\boldsymbol{X}])(\boldsymbol{Y} - \mathbb{E}[\boldsymbol{Y}])^T] \\ &= \mathbb{E}[|\boldsymbol{X} - \mathbb{E}[\boldsymbol{X}]\rangle\langle \boldsymbol{Y} - \mathbb{E}[\boldsymbol{Y}]|],\end{aligned} \tag{4.98}$$

and

$$\begin{aligned}\mathrm{Cov}(\boldsymbol{X}, \boldsymbol{X}) &= \boldsymbol{\Sigma}_{\boldsymbol{XX}} \\ &= \mathbb{E}[(\boldsymbol{X} - \mathbb{E}[\boldsymbol{X}])(\boldsymbol{X} - \mathbb{E}[\boldsymbol{X}])^T] \\ &= \mathbb{E}[|\boldsymbol{X} - \mathbb{E}[\boldsymbol{X}]\rangle\langle \boldsymbol{X} - \mathbb{E}[\boldsymbol{X}]|].\end{aligned} \tag{4.99}$$

**Remarks:**

- $\boldsymbol{\Sigma}$ is real valued: $\sigma_{ij} = \Sigma_{ij} \in \mathbb{R}$.
- $\boldsymbol{\Sigma}$ is symmetric: $\Sigma_{ij} = \Sigma_{ji}$ or $\sigma_{ij} = \sigma_{ji}$.
- The diagonal of $\boldsymbol{\Sigma}$ contains $\sigma_{ii} = \Sigma_{ii} = \mathrm{Var}[X_i] = \sigma_i^2$, i.e. the variances of the components of $\boldsymbol{X}$.
- Let $\boldsymbol{X}$ be a random vector with covariance matrix $\boldsymbol{\Sigma}$, and let $\mathbf{A}$ be a matrix that can act on $\mathbf{X}$ on the left. The covariance matrix of the matrix-vector product $\mathbf{A}\boldsymbol{X}$ is:

$$\begin{aligned}\mathrm{Cov}(\mathbf{A}\boldsymbol{X}, \mathbf{A}\boldsymbol{X}) &= \mathbb{E}[(\mathbf{A}\boldsymbol{X})(\mathbf{A}\boldsymbol{X})^T] - \mathbb{E}[\mathbf{A}\boldsymbol{X}]\mathbb{E}[(\mathbf{A}\boldsymbol{X})^T] \\ &= \mathbb{E}[(\mathbf{A}\boldsymbol{X})(\boldsymbol{X}^T\mathbf{A}^T)] - \mathbb{E}[\mathbf{A}\boldsymbol{X}]\mathbb{E}[(\boldsymbol{X}^T\mathbf{A}^T)] \\ &= \mathbf{A}\mathbb{E}[\boldsymbol{X}\boldsymbol{X}^T]\mathbf{A}^T - \mathbf{A}\mathbb{E}[\boldsymbol{X}]\mathbb{E}[\boldsymbol{X}^T]\mathbf{A}^T \\ &= \mathbf{A}(\mathbb{E}[\boldsymbol{X}\boldsymbol{X}^T] - \mathbb{E}[\boldsymbol{X}]\mathbb{E}[\boldsymbol{X}^T])\mathbf{A}^T \\ &= \mathbf{A}\boldsymbol{\Sigma}\mathbf{A}^T.\end{aligned} \tag{4.100}$$

Recall from linear matrix algebra that any real symmetric matrix has real eigenvalues and a complete set of orthogonal eigenvectors. These can be obtained by orthogonal eigen decomposition. Applying eigenvalue decomposition to the covariance matrix yields

$$\boldsymbol{\Sigma} = \boldsymbol{\Phi}\boldsymbol{\Lambda}\boldsymbol{\Phi}^T, \tag{4.101}$$

where $\boldsymbol{\Phi}$ is an orthogonal matrix containing the eigenvectors of the covariance matrix and

$$\boldsymbol{\Lambda} = \begin{pmatrix} \lambda_1 & 0 & \dots & 0 \\ 0 & \lambda_2 & \dots & 0 \\ \vdots & \vdots & \ddots & \vdots \\ 0 & \dots & \dots & \lambda_d \end{pmatrix}, \tag{4.102}$$

contains the corresponding eigenvalues $\lambda_i$. The eigenvalues represent the amount of variance captured by each eigenvector, and the eigenvectors represent the directions of maximum variance in the data. By sorting the eigenvalues in descending order, one can identify the principal components of the data—the directions along which the data varies the most.

Importantly, the eigenvalues of a covariance matrix are not only real-valued but are by construction further constrained to be non-negative. This can be seen by computing the quadratic form $\boldsymbol{z}^T\boldsymbol{\Sigma}\boldsymbol{z} = \langle \boldsymbol{z}|\boldsymbol{\Sigma}|\boldsymbol{z}\rangle$ where $\boldsymbol{z} = |\boldsymbol{z}\rangle$ is a non-random vector. For any non-zero $\boldsymbol{z}$

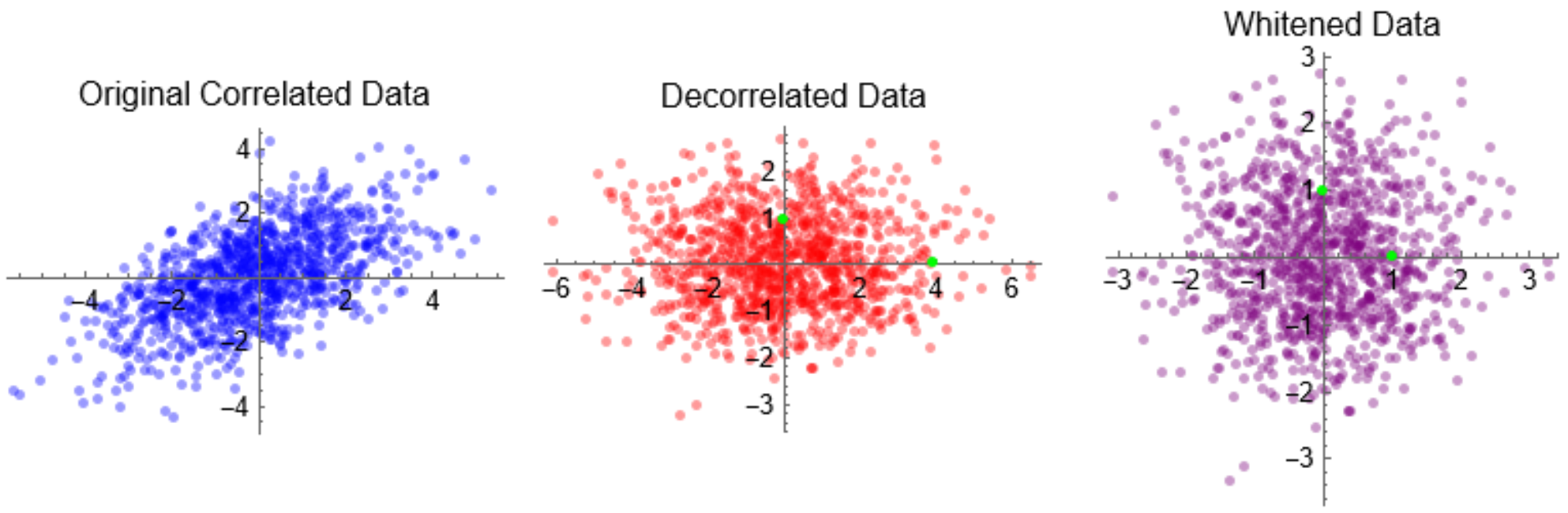

**Figure 4.14.** Visualization of whitening transformation applied to multivariate normal data. (Left) Original correlated data (blue), (Middle) Decorrelated data (red), with green points indicating its covariance, and (Right) Whitened data (purple), with green points indicating its covariance. Each plot illustrates a step in the transformation process, revealing the decorrelation and normalization achieved through whitening.

$$\begin{aligned} \boldsymbol{z}^T\boldsymbol{\Sigma}\boldsymbol{z} &= \boldsymbol{z}^T\mathbb{E}[(\boldsymbol{X}-\boldsymbol{\mu})(\boldsymbol{X}-\boldsymbol{\mu})^T]\boldsymbol{z} \\ &= \mathbb{E}[\boldsymbol{z}^T(\boldsymbol{X}-\boldsymbol{\mu})(\boldsymbol{X}-\boldsymbol{\mu})^T\boldsymbol{z}] \\ &= \mathbb{E}\left[\left(\boldsymbol{z}^T(\boldsymbol{X}-\boldsymbol{\mu})\right)^2\right] \geq 0, \end{aligned} \tag{4.103}$$

and the eigenvalues $\lambda_i \geq 0$. Therefore, the covariance matrix $\boldsymbol{\Sigma}$ is always positive semi-definite.

**Theorem 4.2:** Let $\boldsymbol{A}$ be a positive semidefinite and symmetric matrix. Then there is exactly one positive semidefinite and symmetric matrix $\boldsymbol{B}$ such that $\boldsymbol{A} = \boldsymbol{B}\boldsymbol{B}$. Note that there can be more than one non-symmetric and positive semidefinite matrix $\boldsymbol{B}$ such that $\boldsymbol{A} = \boldsymbol{B}^T\boldsymbol{B}$.

The whitening transformation, also known as sphering transformation, is a technique used in statistics and machine learning to preprocess data in order to make it more amenable to analysis or modeling. The main goal of whitening is to remove correlations between features and to normalize the scale of the features.

- Imagine you have a dataset with several features (dimensions). These features may be correlated with each other, meaning that changes in one feature are associated with changes in another. For example, in a dataset containing height and weight measurements, there might be a correlation between the two variables - taller individuals tend to weigh more. By applying whitening, you aim to remove this correlation, effectively making the features independent of each other. Independent features can simplify the learning process because each feature provides unique information to the model.
- Another aspect of whitening is to normalize the scale of the features. Different features may have different scales, which can make it difficult to compare them or to apply certain algorithms that assume all features are on the same scale. Whitening scales the features so that they all have the same variance. This means that each feature has equal importance in terms of its contribution to the overall variance of the data. This means that the spread of values along each feature is standardized, making them comparable in scale. This step is crucial because it ensures that no single feature dominates the analysis simply because of its scale.

**Definition (Whitening Transformation):** A whitening transformation or sphering transformation is a linear transformation that transforms a vector of random variables with a known covariance matrix into a vector of new random variables whose covariance is the identity matrix, meaning that they are uncorrelated, and each have variance 1. (Figure 4.14)

There are two steps to the whitening transformation procedure. First, we must decorrelate the components of the vector. This produces a data point drawn from a distribution with a diagonal covariance matrix. Finally, we must scale the different components so that they have unit variance, see Figure 4.14. In order to do this, we need the eigen decomposition of the original covariance matrix $\boldsymbol{\Sigma}$. This is either known or it can be estimated from your data set. We assume that $\boldsymbol{\Sigma}$ is positive definite. We also assume that $\boldsymbol{X}$ has zero mean. If not, then subtract the mean (estimated or otherwise) from your samples such that

$$\mathbb{E}[\boldsymbol{X}] = 0, \tag{4.104}$$
$$\boldsymbol{\Sigma} = \mathbb{E}[\boldsymbol{X}\boldsymbol{X}^T]. \tag{4.105}$$

Whitening transformation:

$$\underbrace{\boldsymbol{Y}}_{\boldsymbol{n}\times\boldsymbol{1} \text{ random vector}} = \underbrace{\mathbf{W}}_{\boldsymbol{n}\times\boldsymbol{n} \text{ whitening matrix}} \cdot \underbrace{\boldsymbol{X}}_{\boldsymbol{n}\times\boldsymbol{1} \text{ random vector}}. \tag{4.106}$$

In general, the whitening matrix $\mathbf{W}$ needs to satisfy a constraint:

$$\text{Cov}[\boldsymbol{Y}] = \mathbf{I}_n, \tag{4.107}$$

where identity matrix of size $n$, $\mathbf{I}_n$, is the square matrix with ones on the main diagonal and zeros elsewhere. So, we need,

$$\begin{aligned}\text{Cov}[\mathbf{W}\boldsymbol{X}] &= \mathbb{E}[\mathbf{W}\boldsymbol{X}(\mathbf{W}\boldsymbol{X})^T] \\ &= \mathbb{E}[\mathbf{W}\boldsymbol{X}(\boldsymbol{X}^T\mathbf{W}^T)] \\ &= \mathbf{W}\mathbb{E}[\boldsymbol{X}\boldsymbol{X}^T]\mathbf{W}^T \\ &= \mathbf{W}\boldsymbol{\Sigma}\mathbf{W}^T = \mathbf{I}_n.\end{aligned} \tag{4.108}$$

As a result,

$$\mathbf{W}\boldsymbol{\Sigma}\mathbf{W}^T\mathbf{W} = \mathbf{I}_n\mathbf{W} = \mathbf{W}. \tag{4.109}$$

Hence, whitening matrix $\mathbf{W}$ is a matrix that satisfying the condition

$$\mathbf{W}^T\mathbf{W} = \boldsymbol{\Sigma}^{-1}. \tag{4.110}$$

A general way to specify a valid whitening matrix is

$$\mathbf{W} = \mathbf{Q}\boldsymbol{\Sigma}^{-1/2}, \tag{4.111}$$

where $\mathbf{Q}$ is an orthogonal matrix and $\boldsymbol{\Sigma}^{-1/2}$ is square root symmetric matrix such that $\boldsymbol{\Sigma}^{-\frac{1}{2}}\boldsymbol{\Sigma}^{-\frac{1}{2}} = \boldsymbol{\Sigma}^{-1}$, Theorem 4.2. Recall that an orthogonal matrix $\mathbf{Q}$ has the property that

$$\mathbf{Q}^{-1} = \mathbf{Q}^T, \tag{4.112}$$

and as a consequence

$$\mathbf{Q}^T\mathbf{Q} = \mathbf{Q}\mathbf{Q}^T = \mathbf{I}. \tag{4.113}$$

As a result, the above $\mathbf{W}$ satisfies the whitening constraint:

$$\begin{aligned}\mathbf{W}^T\mathbf{W} &= \left(\mathbf{Q}\boldsymbol{\Sigma}^{-\frac{1}{2}}\right)^T\left(\mathbf{Q}\boldsymbol{\Sigma}^{-\frac{1}{2}}\right) \\ &= \left(\left(\boldsymbol{\Sigma}^{-\frac{1}{2}}\right)^T\mathbf{Q}^T\right)\left(\mathbf{Q}\boldsymbol{\Sigma}^{-\frac{1}{2}}\right) \\ &= \left(\boldsymbol{\Sigma}^{-\frac{1}{2}}\right)^T(\mathbf{Q}^T\mathbf{Q})\boldsymbol{\Sigma}^{-\frac{1}{2}} \\ &= \underbrace{\left(\boldsymbol{\Sigma}^{-\frac{1}{2}}\right)^T}_{\boldsymbol{\Sigma}^{-\frac{1}{2}}}\underbrace{\left(\mathbf{Q}^T\mathbf{Q}\right)}_{\mathbf{I}}\boldsymbol{\Sigma}^{-\frac{1}{2}} \quad \left[\boldsymbol{\Sigma}^{-\frac{1}{2}} \text{ is symmetric matrix}, \boldsymbol{\Sigma}^{-\frac{1}{2}} = \left(\boldsymbol{\Sigma}^{-\frac{1}{2}}\right)^T\right] \\ &= \boldsymbol{\Sigma}^{-\frac{1}{2}}\boldsymbol{\Sigma}^{-\frac{1}{2}} = \boldsymbol{\Sigma}^{-1}.\end{aligned} \tag{4.114}$$

Note the converse is also true: any whitening matrix, i.e. any $\mathbf{W}$ satisfying the whitening constraint, can be written in the above form as

$$\mathbf{Q} = \mathbf{W}\boldsymbol{\Sigma}^{1/2}, \tag{4.115}$$

is orthogonal by construction. Hence, instead of choosing $\mathbf{W}$, we choose the orthogonal matrix $\mathbf{Q}$:

- Recall that orthogonal matrices geometrically represent rotations (plus reflections).
- It is now clear that there are infinitely many whitening procedures, because there are infinitely many rotations! This also means we need to find ways to choose among whitening procedures.

- Commonly used choices are $\mathbf{Q} = \mathbf{I}$ (Mahalanobis or Zero-Phase Component Analysis (ZCA) whitening) or the eigen-system of $\mathbf{\Sigma}$ (Principal Components Analysis (PCA) whitening).

Note that the matrix square root, $\mathbf{\Sigma}^{-1/2}$, doesn't mean the matrix whose elements are the square roots of the elements of $\mathbf{\Sigma}^{-1}$; rather, it is the matrix which, when matrix-multiplied by itself, will give $\mathbf{\Sigma}^{-1}$. So, for some square matrix $\mathbf{A}$, matrix $\mathbf{B}$ is a square root of $\mathbf{A}$ if $\mathbf{BB} = \mathbf{A}$. The matrix square root can be computed multiple ways. We will denote the square root of $\mathbf{\Sigma}$ as $\mathbf{\Sigma}^{1/2}$ and the square root of $\mathbf{\Sigma}^{-1}$ as $\mathbf{\Sigma}^{-1/2}$. A couple properties of the matrix square root are worth mentioning:

- A matrix square root is symmetric.
- $\mathbf{\Sigma}^{-1/2} = \left(\mathbf{\Sigma}^{1/2}\right)^{-1}$.

Mathematically, the whitening transformation involves linearly transforming the data such that the resulting features are uncorrelated and have unit variance. One common approach to achieve this is by using the eigenvalue decomposition of the covariance matrix of the data. This decomposition provides a set of orthogonal vectors (principal components) along which the data varies the most, and their associated eigenvalues represent the variance along each of these directions. By rescaling the data by the square root of the inverse of these eigenvalues, you effectively decorrelate the features and normalize their variance.

The decomposition of covariance matrix can be derived from the fundamental property of eigenvectors:

$$\mathbf{\Sigma}\boldsymbol{e} = \lambda\boldsymbol{e}, \tag{4.116}$$

where $\boldsymbol{e}$ is eigenvector of the matrix $\mathbf{\Sigma}$ with eigenvalue $\lambda$. We have

$$\mathbf{\Sigma\Phi} = \mathbf{\Phi\Lambda}, \tag{4.117}$$

where $\mathbf{\Phi}$ is a matrix whose $i$th column is the eigenvector $\boldsymbol{e}_i$ of $\mathbf{\Sigma}$, and $\mathbf{\Lambda}$ is the diagonal matrix whose diagonal elements are the corresponding eigenvalues, $\Lambda_{ii} = \lambda_i$. From (4.101), we have

$$\mathbf{\Sigma} = \mathbf{\Phi\Lambda\Phi}^{-1}. \tag{4.118}$$

Hence, the covariance matrix can be expressed as follows,

$$\mathbf{\Sigma} = \mathbf{\Phi}\mathbf{\Lambda}^{\frac{1}{2}}\mathbf{\Lambda}^{\frac{1}{2}}\mathbf{\Phi}^{-1}. \tag{4.119}$$

Recall from linear matrix algebra that any real symmetric matrix has real eigenvalues and a complete set of orthogonal eigenvectors. The columns of $\mathbf{\Phi}$ are orthonormal so that $\mathbf{\Phi}^{-1} = \mathbf{\Phi}^T$. Let

$$\overline{\boldsymbol{Y}} = \mathbf{\Phi}^T\boldsymbol{X}. \tag{4.120}$$

Then $\overline{\boldsymbol{Y}}$ is a random vector with a decorrelated multivariates Gaussian distribution with variances $\lambda_i$ on the diagonal of its covariance matrix. The multiplication by $\mathbf{\Phi}^T$ rotates the data points in the original feature space. The columns of $\mathbf{\Phi}$ are the eigenvectors of the covariance matrix $\mathbf{\Sigma}$. Since these eigenvectors are orthogonal, multiplying by $\mathbf{\Phi}^T$ effectively rotates the data such that the axes align with the directions of maximum variance. The next step then is quite obvious. Let

$$\boldsymbol{Y} = \mathbf{\Lambda}^{-\frac{1}{2}}\overline{\boldsymbol{Y}} = \mathbf{\Lambda}^{-\frac{1}{2}}\mathbf{\Phi}^T\boldsymbol{X}. \tag{4.121}$$

$\boldsymbol{Y}$ is then random vector with the standard multivariate distribution. The multiplication by $\mathbf{\Lambda}^{-\frac{1}{2}}$ scales the rotated data along each principal component axis by the inverse square root of its corresponding eigenvalue. This scaling operation effectively normalizes the variance along each axis, ensuring that the variance along each direction is equal. After this scaling operation, the transformed data has unit variance along each axis and is decorrelated. By combining rotation and scaling, the transformed data $\boldsymbol{Y}$ achieves a state where the axes align with the directions of maximum variance, and the variance along each axis is equal. This transformation facilitates subsequent analysis by simplifying the data structure and making it amenable to techniques that assume uncorrelated features with equal variance.

Well, we can derive the covariance matrix of $\boldsymbol{Y}$. Recall that its mean is the zero vector so its covariance is

$$\begin{aligned}
\mathbb{E}[\overline{\boldsymbol{Y}}\overline{\boldsymbol{Y}}^T] &= \mathbb{E}[(\boldsymbol{\Phi}^T\boldsymbol{X})(\boldsymbol{\Phi}^T\boldsymbol{X})^T] \\
&= \mathbb{E}[(\boldsymbol{\Phi}^T\boldsymbol{X})(\boldsymbol{X}^T\boldsymbol{\Phi})] \\
&= \boldsymbol{\Phi}^T\mathbb{E}[\boldsymbol{X}\boldsymbol{X}^T]\boldsymbol{\Phi} \\
&= \boldsymbol{\Phi}^T\boldsymbol{\Sigma}\boldsymbol{\Phi} \\
&= \boldsymbol{\Phi}^T(\boldsymbol{\Phi}\boldsymbol{\Lambda}\boldsymbol{\Phi}^{-1})\boldsymbol{\Phi} \\
&= (\boldsymbol{\Phi}^T\boldsymbol{\Phi})(\boldsymbol{\Lambda})(\boldsymbol{\Phi}^{-1}\boldsymbol{\Phi}) \\
&= \boldsymbol{\Lambda}. \qquad (4.122)
\end{aligned}$$

This proves what we previously stated; the components of $\overline{\boldsymbol{Y}}$ are uncorrelated since the covariance matrix is diagonal. Using the fact that diagonal matrix $\boldsymbol{\Lambda}^{\frac{1}{2}}$ is symmetric, $\boldsymbol{\Lambda}^{-\frac{1}{2}} = \left(\boldsymbol{\Lambda}^{-\frac{1}{2}}\right)^T$, the covariance matrix of $\boldsymbol{Y}$ is

$$\begin{aligned}
\mathbb{E}[\boldsymbol{Y}\boldsymbol{Y}^T] &= \mathbb{E}\left[\left(\boldsymbol{\Lambda}^{-\frac{1}{2}}\overline{\boldsymbol{Y}}\right)\left(\boldsymbol{\Lambda}^{-\frac{1}{2}}\overline{\boldsymbol{Y}}\right)^T\right] \\
&= \mathbb{E}\left[\left(\boldsymbol{\Lambda}^{-\frac{1}{2}}\overline{\boldsymbol{Y}}\right)\left(\overline{\boldsymbol{Y}}^T\left(\boldsymbol{\Lambda}^{-\frac{1}{2}}\right)^T\right)\right] \\
&= \mathbb{E}\left[\left(\left(\boldsymbol{\Lambda}^{-\frac{1}{2}}\right)^T\overline{\boldsymbol{Y}}\right)\left(\overline{\boldsymbol{Y}}^T\boldsymbol{\Lambda}^{-\frac{1}{2}}\right)\right] \\
&= \left(\boldsymbol{\Lambda}^{-\frac{1}{2}}\right)^T\mathbb{E}[\overline{\boldsymbol{Y}}\overline{\boldsymbol{Y}}^T]\boldsymbol{\Lambda}^{-\frac{1}{2}} \\
&= \left(\boldsymbol{\Lambda}^{-\frac{1}{2}}\right)^T\boldsymbol{\Lambda}\boldsymbol{\Lambda}^{-\frac{1}{2}} \\
&= \left(\boldsymbol{\Lambda}^{-\frac{1}{2}}\boldsymbol{\Lambda}^{\frac{1}{2}}\right)\left(\boldsymbol{\Lambda}^{\frac{1}{2}}\boldsymbol{\Lambda}^{-\frac{1}{2}}\right) \\
&= \mathbf{I}. \qquad (4.123)
\end{aligned}$$

This concludes the proof of how the linear transform given by matrix pre-multiplication by $\boldsymbol{\Lambda}^{-\frac{1}{2}}\boldsymbol{\Phi}^T$ produces whitened data.

By whitening the data, we make the distribution of the data isotropic, meaning it has the same shape in all directions. Therefore, even if the original data was elongated or stretched along certain axes due to its elliptical shape, after whitening, the scatter plot will have a roughly symmetric (spherical) shape.

General overview of how data whitening is typically performed:

1. Compute the mean vector of the dataset by taking the average of each feature across all samples.
2. Subtract the mean vector from each data point to center the dataset around the origin.
3. Calculate the covariance matrix of the centered data. The covariance matrix provides information about the relationships between different features in the dataset.
4. Perform an eigenvalue decomposition (or singular value decomposition) on the covariance matrix to obtain its eigenvectors and eigenvalues. Eigenvectors represent the directions of maximum variance in the data, while eigenvalues represent the magnitude of variance along those directions.
5. Transform the centered data into the space spanned by the eigenvectors of the covariance matrix. This is done by multiplying the centered data matrix by the matrix of eigenvectors (rotation matrix). This rotation (transformation) matrix is used to decorrelate the data.
6. Construct a scaling (transformation) matrix by using the inverse square root of the diagonal matrix formed by the eigenvalues. Multiply the decorrelated data by the scaling matrix to obtain the whitened data.

### 4.6.4 Mahalanobis Distance

Let's explore the challenge of gauging the probability that a test point located in an N-dimensional Euclidean space is a member of a particular set. We have sample points that are clearly identified as part of this set. We typically start by examining the characteristics of the sample points we already know belong to that set. One common intuition is to consider the centroid or center of mass of these sample points. This center serves as a reference point for determining the likelihood of other points belonging to the set.

However, merely looking at the distance between a test point and the center of mass might not be sufficient. We also need to understand how spread out the set is in space. A simple way to capture this notion is by estimating the standard deviation of the distances of the sample points from the center of mass. If a test point falls within the standard deviation from the center, it might be considered highly probable to belong to the set. As the distance increases beyond this threshold, the likelihood of the test point belonging to the set decreases. To quantify this intuition, we define the normalized distance between the test point and the set using the formula:

$$\frac{\|\mathbf{x}-\boldsymbol{\mu}\|_2}{\sigma}, \tag{4.124}$$

where $\mathbf{x}$ is the test point, $\boldsymbol{\mu}$ is the sample mean (or center of mass), and $\sigma$ is the standard deviation of distances of the sample points from the center. By plugging this formula into the normal distribution, we can derive the probability of the test point belonging to the set.

However, this approach assumes that the distribution of sample points around the center of mass is spherical. When dealing with spherical distributions, the spread of the data is uniform in all directions from the center. In other words, regardless of the direction from the center of the distribution, the spread of data remains the same. This uniformity implies that the probability of a test point belonging to the set depends solely on its distance from the center of mass. In a spherical distribution, if a test point is located at a certain distance from the center, its likelihood of belonging to the set is the same regardless of the direction from which it is measured. For example, if you imagine a three-dimensional spherical distribution, any point located at a certain radius from the center has an equal chance of belonging to the set, regardless of whether it lies along the x-axis, y-axis, or z-axis.

On the other hand, when dealing with non-spherical distributions, such as ellipsoids, the spread of the data in different directions varies. This means that the probability of a test point belonging to the set not only depends on its distance from the center (centroid) but also on the direction in which it lies relative to the shape of the distribution. Consider an ellipsoidal distribution in two dimensions, for instance. Imagine an elongated ellipse rather than a circle. Now, suppose you have a test point located at the same distance from the center of mass in both the major axis direction (long axis of the ellipse) and the minor axis direction (short axis of the ellipse). In the major axis direction, the ellipsoid is wider, meaning that points in this direction are more spread out. As a result, even though the test point is at the same distance from the center, it might be more probable for it to belong to the set because it's within the "expected range" given the spread of data along that axis. Conversely, in the minor axis direction, the ellipsoid is narrower, indicating that points in this direction are less spread out. Thus, the same distance from the center might be less likely to fall within the expected range of the set because the data is more concentrated along this axis. This concept extends to higher dimensions as well. In an N-dimensional space, the ellipsoidal distribution represents the spread of data along different axes. Therefore, the probability of a test point belonging to the set is influenced not only by its distance from the center but also by the alignment of that distance with the principal axes of the ellipsoid.

To address this, non-spherical distributions, such as ellipsoids, must be whitened first. When data is whitened, it effectively removes any elongation or stretching along specific axes by transforming the data into a new set of orthogonal axes (principal components). As a result, the scatter plot tends to exhibit a more symmetrical shape, especially compared to the original data, which might have had elongation or skewness due to correlations between features. This symmetrical shape simplifies the analysis and makes it easier to apply distance metrics. Once the data is whitened, Euclidean distance can be used effectively as a measure of similarity between points. In the transformed space, the Euclidean distance reflects the true geometric distance between points without any bias introduced by the original coordinate system.

Now let us discuss more mathematical details. The Mahalanobis [108-114] distance is a measure of the distance between a point $P$ and a distribution $D$, introduced by P. C. Mahalanobis in 1936, taking into account the covariance structure of the distribution (or data). Given a probability distribution $D$ on $\mathbb{R}^n$, with mean $\boldsymbol{\mu} = (\mu_1, \mu_2, \dots \mu_n)^T$ and positive-definite covariance matrix $\boldsymbol{\Sigma}$, the Mahalanobis distance of a point $\mathbf{x} = (x_1, x_2, \dots x_n)^T$ from $D$ is

$$d_M(\mathbf{x}, D) = \sqrt{(\mathbf{x}-\boldsymbol{\mu})^T \boldsymbol{\Sigma}^{-1} (\mathbf{x}-\boldsymbol{\mu})}. \tag{4.125}$$

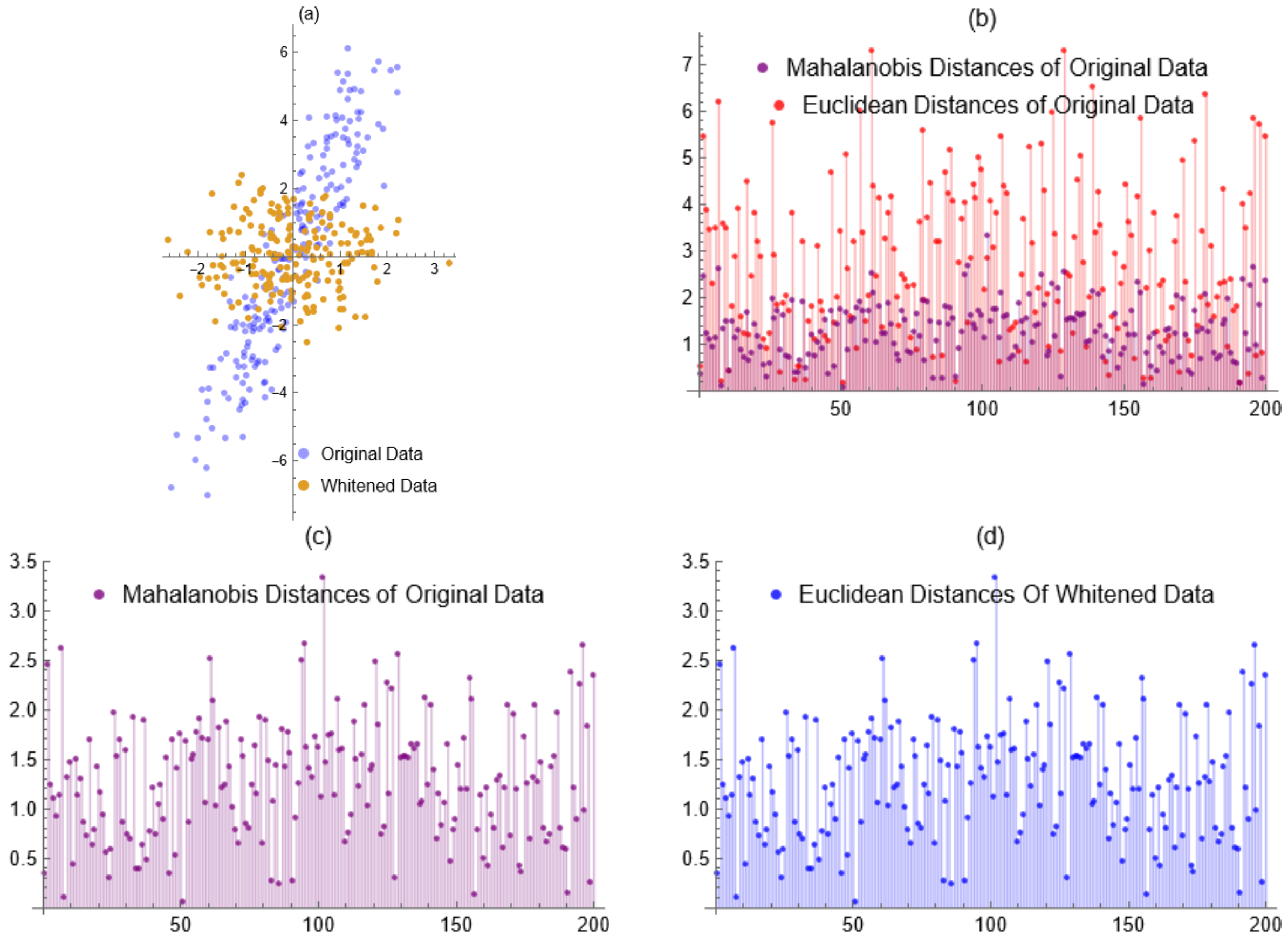

**Figure 4.15.** Figure (a) displays the original dataset generated from a bivariate normal distribution alongside the whitened dataset. The whitened data is obtained by removing the mean and scaling it using the inverse square root of the covariance matrix. Figure (b) presents a combination of Euclidean distances of the original dataset from the mean point and Mahalanobis distances of the original dataset. The plot allows for a comparison between these two-distance metrics and provides insight into data variability and distribution characteristics. Figure (c) illustrates the Mahalanobis distances of the original dataset. Mahalanobis distances account for the covariance structure of the data and are represented by the shaded area. Figure (d) presents the Euclidean distances of the whitened dataset from the origin. Notably, the plot of Euclidean distances of the whitened dataset is identical to the plot of Mahalanobis distances of the original dataset.

Given two points $\mathbf{x}$ and $\mathbf{y}$ the Mahalanobis distance between them with respect to $D$ is

$$d_M(\mathbf{x}, \mathbf{y}, D) = \sqrt{(\mathbf{x}-\mathbf{y})^T \boldsymbol{\Sigma}^{-1} (\mathbf{x}-\mathbf{y})}, \tag{4.126}$$

which means that $d_M(\mathbf{x}, D) = d_M(\mathbf{x}, \boldsymbol{\mu}, D)$. If the covariance matrix $\boldsymbol{\Sigma}$ is the identity matrix (i.e., no correlations between dimensions), then the Mahalanobis distance reduces to the Euclidean distance. Using whitening constraint $\mathbf{W}^T\mathbf{W} = \boldsymbol{\Sigma}^{-1}$, we have

$$\begin{aligned} d_M(\mathbf{x}, \mathbf{y}, D) &= \sqrt{(\mathbf{x}-\mathbf{y})^T \boldsymbol{\Sigma}^{-1} (\mathbf{x}-\mathbf{y})} \\ &= \sqrt{(\mathbf{x}-\mathbf{y})^T \mathbf{W}^T \mathbf{W} (\mathbf{x}-\mathbf{y})} \\ &= \sqrt{\left(\mathbf{W}(\mathbf{x}-\mathbf{y})\right)^T \mathbf{W}(\mathbf{x}-\mathbf{y})} \\ &= \|\mathbf{W}(\mathbf{x}-\mathbf{y})\|. \end{aligned} \tag{4.127}$$

That is, the Mahalanobis distance is the Euclidean distance after a whitening transformation, see Figure 4.15. The Mahalanobis distance takes into account the covariance structure of the data (the scale and correlation of the variables), providing a more accurate measure of dissimilarity compared to the Euclidean distance, especially when dealing with high-dimensional data with correlated features. Moreover,

$$
\begin{aligned}
d_M(\mathbf{x}, D) &= \sqrt{(\mathbf{x}-\boldsymbol{\mu})^T\boldsymbol{\Sigma}^{-1}(\mathbf{x}-\boldsymbol{\mu})} \\
&= \sqrt{(\mathbf{x}-\boldsymbol{\mu})^T\boldsymbol{\Sigma}^{-\frac{1}{2}}\boldsymbol{\Sigma}^{-\frac{1}{2}}(\mathbf{x}-\boldsymbol{\mu})} \\
&= \sqrt{(\mathbf{x}-\boldsymbol{\mu})^T\left(\boldsymbol{\Sigma}^{-\frac{1}{2}}\right)^T\boldsymbol{\Sigma}^{-\frac{1}{2}}(\mathbf{x}-\boldsymbol{\mu})} \\
&= \sqrt{\left(\boldsymbol{\Sigma}^{-\frac{1}{2}}(\mathbf{x}-\boldsymbol{\mu})\right)^T\left(\boldsymbol{\Sigma}^{-\frac{1}{2}}(\mathbf{x}-\boldsymbol{\mu})\right)} \\
&= \left\|\boldsymbol{\Sigma}^{-\frac{1}{2}}(\mathbf{x}-\boldsymbol{\mu})\right\| = \|\mathbf{y}\|,
\end{aligned} \tag{4.128}
$$

where the transformed vector $\mathbf{y}$ is given by Mahalanobis whitening transformation, remember $\mathbf{Q} = \mathbf{I}$ (4.108),

$$
\mathbf{y} = \boldsymbol{\Sigma}^{-1/2}(\mathbf{x}-\boldsymbol{\mu}). \tag{4.129}
$$

Next, let's consider the covariance of $\mathbf{y}$

$$
\begin{aligned}
\mathbb{E}[\mathbf{y}\mathbf{y}^T] &= \mathbb{E}\left[\left(\boldsymbol{\Sigma}^{-\frac{1}{2}}(\mathbf{x}-\boldsymbol{\mu})\right)\left(\boldsymbol{\Sigma}^{-\frac{1}{2}}(\mathbf{x}-\boldsymbol{\mu})\right)^T\right] \\
&= \mathbb{E}\left[\left(\boldsymbol{\Sigma}^{-\frac{1}{2}}(\mathbf{x}-\boldsymbol{\mu})\right)\left((\mathbf{x}-\boldsymbol{\mu})^T\left(\boldsymbol{\Sigma}^{-\frac{1}{2}}\right)^T\right)\right] \\
&= \boldsymbol{\Sigma}^{-\frac{1}{2}}\mathbb{E}[(\mathbf{x}-\boldsymbol{\mu})(\mathbf{x}-\boldsymbol{\mu})^T]\left(\boldsymbol{\Sigma}^{-\frac{1}{2}}\right)^T \\
&= \boldsymbol{\Sigma}^{-\frac{1}{2}}\mathbb{E}[(\mathbf{x}-\boldsymbol{\mu})(\mathbf{x}-\boldsymbol{\mu})^T]\boldsymbol{\Sigma}^{-\frac{1}{2}} \\
&= \boldsymbol{\Sigma}^{-\frac{1}{2}}\boldsymbol{\Sigma}\boldsymbol{\Sigma}^{-\frac{1}{2}} \\
&= \boldsymbol{\Sigma}^{-\frac{1}{2}}\boldsymbol{\Sigma}^{\frac{1}{2}}\boldsymbol{\Sigma}^{\frac{1}{2}}\boldsymbol{\Sigma}^{-\frac{1}{2}} = \mathbf{I}.
\end{aligned} \tag{4.130}
$$

So not only the Mahalanobis distance is the Euclidean distance after a whitening transformation, but the transformed vector $\mathbf{y}$ also has unit variance in all directions (its covariance is whitened and normalized).

The matrix square root, $\boldsymbol{\Sigma}^{-1/2}$, can be computed multiple ways. Using $\boldsymbol{\Sigma} = \boldsymbol{\Phi}\boldsymbol{\Lambda}\boldsymbol{\Phi}^T$, we have

$$
\boldsymbol{\Sigma}^{\frac{1}{2}} = \boldsymbol{\Phi}\boldsymbol{\Lambda}^{\frac{1}{2}}\boldsymbol{\Phi}^T, \tag{4.131}
$$

since,

$$
\boldsymbol{\Sigma}^{\frac{1}{2}}\boldsymbol{\Sigma}^{\frac{1}{2}} = \left(\boldsymbol{\Phi}\boldsymbol{\Lambda}^{\frac{1}{2}}\boldsymbol{\Phi}^T\right)\left(\boldsymbol{\Phi}\boldsymbol{\Lambda}^{\frac{1}{2}}\boldsymbol{\Phi}^T\right) = \boldsymbol{\Phi}\boldsymbol{\Lambda}^{\frac{1}{2}}(\boldsymbol{\Phi}^T\boldsymbol{\Phi})\boldsymbol{\Lambda}^{\frac{1}{2}}\boldsymbol{\Phi}^T = \boldsymbol{\Phi}\boldsymbol{\Lambda}^{\frac{1}{2}}\boldsymbol{\Lambda}^{\frac{1}{2}}\boldsymbol{\Phi}^T = \boldsymbol{\Phi}\boldsymbol{\Lambda}\boldsymbol{\Phi}^T = \boldsymbol{\Sigma}, \tag{4.132}
$$

where $\boldsymbol{\Phi}$ is the matrix of eigenvectors of $\boldsymbol{\Sigma}$, and $\boldsymbol{\Lambda}^{\frac{1}{2}}$ is the diagonal matrix of the square roots of eigenvalues. As a result, we have

$$
\boldsymbol{\Sigma}^{-\frac{1}{2}} = \left(\boldsymbol{\Sigma}^{\frac{1}{2}}\right)^{-1} = \left(\boldsymbol{\Phi}\boldsymbol{\Lambda}^{\frac{1}{2}}\boldsymbol{\Phi}^T\right)^{-1} = \left(\boldsymbol{\Phi}\boldsymbol{\Lambda}^{\frac{1}{2}}\boldsymbol{\Phi}^{-1}\right)^{-1} = (\boldsymbol{\Phi}^{-1})^{-1}\boldsymbol{\Lambda}^{-\frac{1}{2}}\boldsymbol{\Phi}^{-1} = \boldsymbol{\Phi}\boldsymbol{\Lambda}^{-\frac{1}{2}}\boldsymbol{\Phi}^T. \tag{4.133}
$$

To prove that $\boldsymbol{\Sigma}^{-\frac{1}{2}}$ is the matrix square root of the covariance matrix $\boldsymbol{\Sigma}^{-1}$, we need to show that $\boldsymbol{\Sigma}^{-\frac{1}{2}}\boldsymbol{\Sigma}^{-\frac{1}{2}} = \boldsymbol{\Sigma}^{-1}$.

$$
\begin{aligned}
\boldsymbol{\Sigma}^{-\frac{1}{2}}\boldsymbol{\Sigma}^{-\frac{1}{2}} &= \left(\boldsymbol{\Phi}\boldsymbol{\Lambda}^{-\frac{1}{2}}\boldsymbol{\Phi}^T\right)\left(\boldsymbol{\Phi}\boldsymbol{\Lambda}^{-\frac{1}{2}}\boldsymbol{\Phi}^T\right) \\
&= \boldsymbol{\Phi}\boldsymbol{\Lambda}^{-\frac{1}{2}}(\boldsymbol{\Phi}^T\boldsymbol{\Phi})\boldsymbol{\Lambda}^{-\frac{1}{2}}\boldsymbol{\Phi}^T \\
&= \boldsymbol{\Phi}\boldsymbol{\Lambda}^{-\frac{1}{2}}(\mathbf{I})\boldsymbol{\Lambda}^{-\frac{1}{2}}\boldsymbol{\Phi}^T \\
&= \boldsymbol{\Phi}\boldsymbol{\Lambda}^{-1}\boldsymbol{\Phi}^T \\
&= ((\boldsymbol{\Phi}^T)^{-1}\boldsymbol{\Lambda}\boldsymbol{\Phi}^{-1})^{-1} \\
&= (\boldsymbol{\Phi}\boldsymbol{\Lambda}\boldsymbol{\Phi}^T)^{-1} = \boldsymbol{\Sigma}^{-1}.
\end{aligned} \tag{4.134}
$$

Thus, $\boldsymbol{\Sigma}^{-\frac{1}{2}} = \boldsymbol{\Phi}\boldsymbol{\Lambda}^{-\frac{1}{2}}\boldsymbol{\Phi}^T$ is indeed the matrix square root of the covariance matrix $\boldsymbol{\Sigma}^{-1}$. Moreover, $\boldsymbol{\Sigma}^{-\frac{1}{2}} = \boldsymbol{\Phi}\boldsymbol{\Lambda}^{-\frac{1}{2}}\boldsymbol{\Phi}^T$ is whitening matrix.

## 4.7 Batch Normalization

BN stands out as a remarkable advancement in the optimization of DNNs. Surprisingly, it diverges from the typical optimization algorithms. Rather, it operates as a technique for adaptive reparametrization, driven by the challenges encountered in training excessively deep models.

Although using He initialization along with any variant of ReLU AF or Xavier initialization along with Tanh AF can significantly reduce the danger of the vanishing/exploding gradients problems at the beginning of training, it doesn't guarantee that they won't come back during training. Normalizing the inputs does not necessarily prevent saturation of neurons for hidden layers. To address these problems Ioffe and Szegedy [101] introduced the BN technique. The idea is to normalize values inside of the network as well and thereby prevent hidden neurons from becoming saturated and reduce the vanishing/exploding gradients problems.

Fixed distribution of inputs to a sub-network would have positive consequences for the layers outside the subnetwork, as well. Consider a layer with a Sigmoid AF $\mathbf{a} = \sigma(\mathbf{Wu} + \mathbf{b})$ where $\mathbf{u}$ is the layer input, the weight matrix $\mathbf{W}$ and bias vector $\mathbf{b}$ are the layer parameters to be learned, and $\sigma(z) = 1/(1 + \exp(-z))$. As $|z|$ increases, $\sigma'(z)$ tends to zero. This means that for all dimensions of $\mathbf{z} = \mathbf{Wu} + \mathbf{b}$ except those with small absolute values, the gradient flowing down to $\mathbf{u}$ will vanish and the model will train slowly. However, since $\mathbf{z}$ is affected by $\mathbf{W}$, $\mathbf{b}$, and the parameters of all the layers below, changes to those parameters during training will likely move many dimensions of $\mathbf{z}$ into the saturated regime of the nonlinearity and slow down the convergence. This effect is amplified as the network depth increases. If we could ensure that the distribution of nonlinearity inputs remains more stable as the network trains, then the optimizer would be less likely to get stuck in the saturated regime, and the training would accelerate.

By fixing the distribution of the layer inputs as the training progresses, we expect to improve the training speed. It has been long known [106] that the network training converges faster if its inputs are whitened – i.e., linearly transformed to have zero means and unit variances, and decorrelated. As each layer observes the inputs produced by the layers below, it would be advantageous to achieve the same whitening of the inputs of each layer. By whitening the inputs to each layer, we would take a step towards achieving the fixed distributions of inputs.

**Definition (Internal Covariate Shift)**: Internal covariate shift refers to the change in the distribution of network activations (or inputs) as the parameters of the preceding layers change during training [115]. In simpler terms, it's the phenomenon where the distribution of the input to each layer of a NN changes as the network learns.

This shift can occur due to various reasons:

- As the parameters of the NN are updated during training, the activations of each layer change accordingly. This change can lead to a shift in the distribution of activations, making it challenging for subsequent layers to learn effectively.
- During training, as the model learns from different mini-batches of data, the distribution of the input data to the network can vary. This variation can also contribute to internal covariate shift, particularly if the mini-batches are not representative of the overall dataset.

When the distribution of inputs keeps changing due to internal covariate shift, it can slow down the convergence of the training process for several reasons:

- Fluctuations in the distribution of inputs can lead to unstable gradients, making it challenging for the optimization algorithm to converge to an optimal solution.
- Internal covariate shift can exacerbate the vanishing or exploding gradient problems, where gradients in deeper layers become too small or too large, hindering effective learning.
- If the inputs to each layer keep changing significantly, it becomes difficult for the network to learn meaningful representations of the data, as the learned features may not generalize well to unseen examples.

By ensuring that the mean and variance of the inputs to each layer remain stable during training, BN helps alleviate the issues caused by internal covariate shift, leading to faster and more stable training of DNNs.

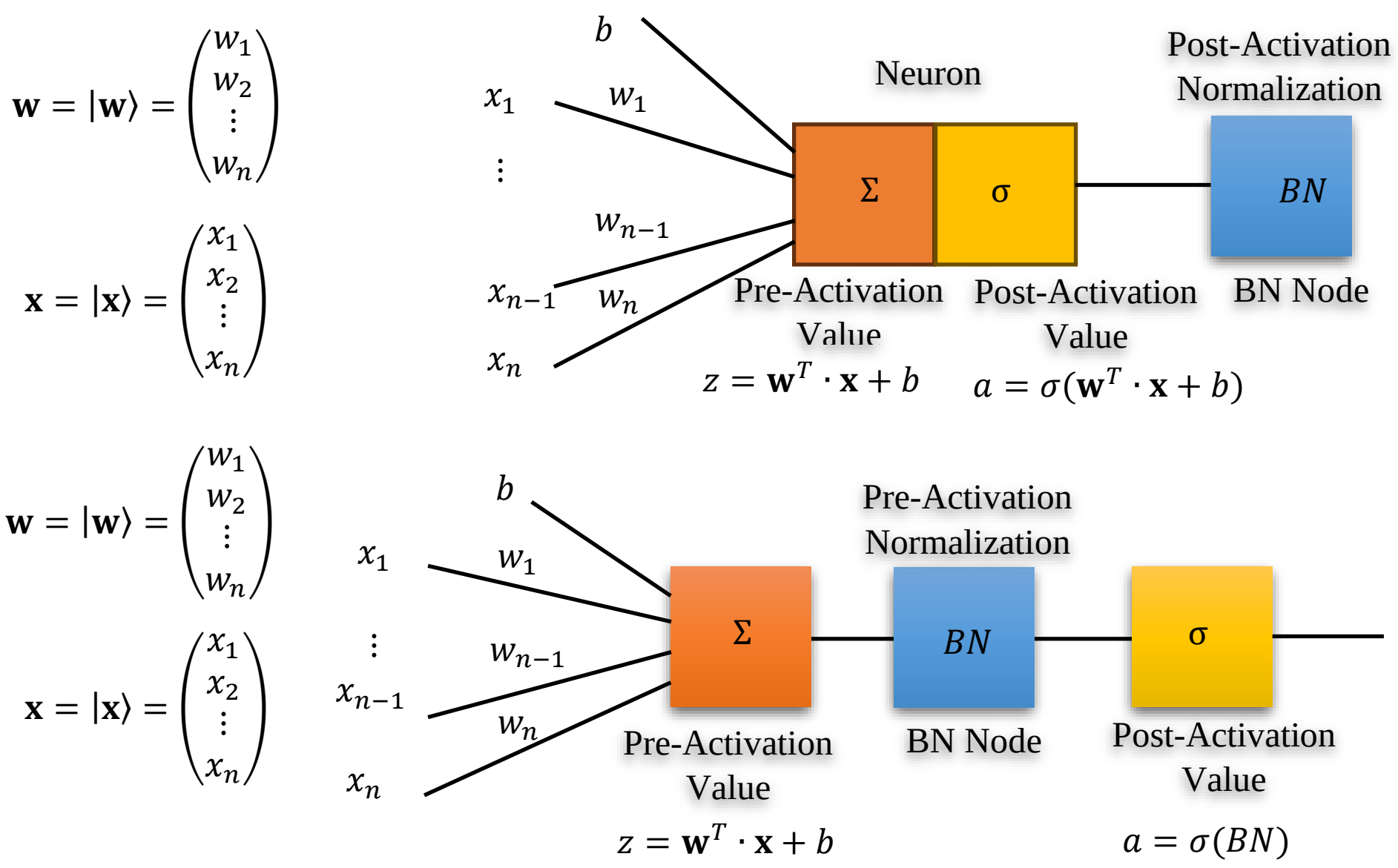

**Figure 4.16.** The different choices in BN. Top panel: BN is applied to the output of the AF. Bottom panel: BN as presented by Loffe and Szegedy (2015). The layer of neurons is broken up into two parts. The first part is the weighted sums for all neurons. BN is applied to these weighted sums. The AF is applied to the output of the BN operation.

There are two choices for the placement of a normalization layer (set of BN nodes) within a NN architecture.

- Normalization after Activation (Post-Activation Normalization): In this approach, normalization is applied immediately after the AF is applied to the linearly transformed inputs (the normalization is performed on the post-activation values). This approach is depicted in Figure 4.16, (top).
- Normalization before Activation (Pre-Activation Normalization): Alternatively, normalization can be applied after the linear transformation of the inputs but before the AF is applied. (In this case, normalization is performed on the pre-activation values). This approach is illustrated in Figure 4.16, (bottom).

### During Forward Propagation

The main idea behind BN is to normalize the pre-activations of each layer across a mini-batch of data. Let us consider the scenario where the input of a BN node is represented as $z_i^{(r)}$, corresponding to the $r$-th element of the batch fed into the $i$-th unit. Each $z_i^{(r)}$ is derived through a linear transformation defined by the vector $\mathbf{w}_i$ (and biases if present). Considering a specific batch of $m$ instances, where the pre-activations' values are denoted as $z_i^{(1)}, z_i^{(2)}, \ldots z_i^{(m)}$, Figure 4.17 and Figure 4.18, the initial step involves computing the mean $\mu_i$ and standard deviation $\sigma_i$ for the $i$-th hidden unit. Subsequently, these values are scaled and shifted using the parameters $\gamma_i$ and $\beta_i$ to generate the final pre-activation output of the $i$th node. For each hidden unit $i$, the mean $\mu_i$ and the variance $\sigma_i^2$ are computed over the pre-activations of the $i$th unit across the batch of $m$ instances. Mathematically, the mean $\mu_i$ for the $i$th hidden unit is calculated as the average of $z_i^{(r)}$ across all training samples, which can be expressed as:

$$\mu_i = \frac{1}{m}\sum_{r=1}^{m} z_i^{(r)} \quad \forall i, \tag{4.135.1}$$

and variance $\sigma_i^2$ for the $i$th hidden unit can be written as

$$\sigma_i^2 = \frac{1}{m}\sum_{r=1}^{m}\left(z_i^{(r)} - \mu_i\right)^2 + \epsilon \quad \forall i, \tag{4.135.2}$$

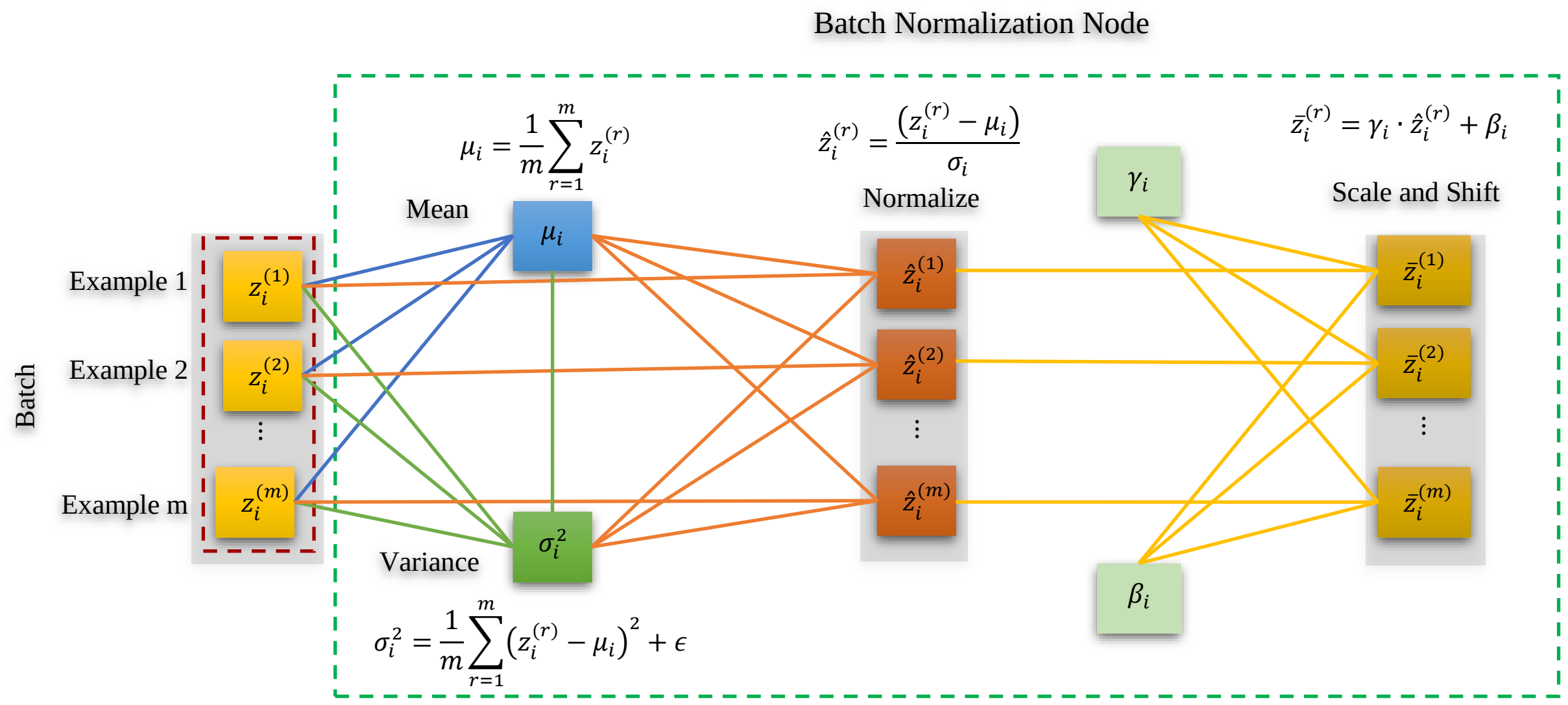

**Figure 4.17.** Flow of computation through BN node.

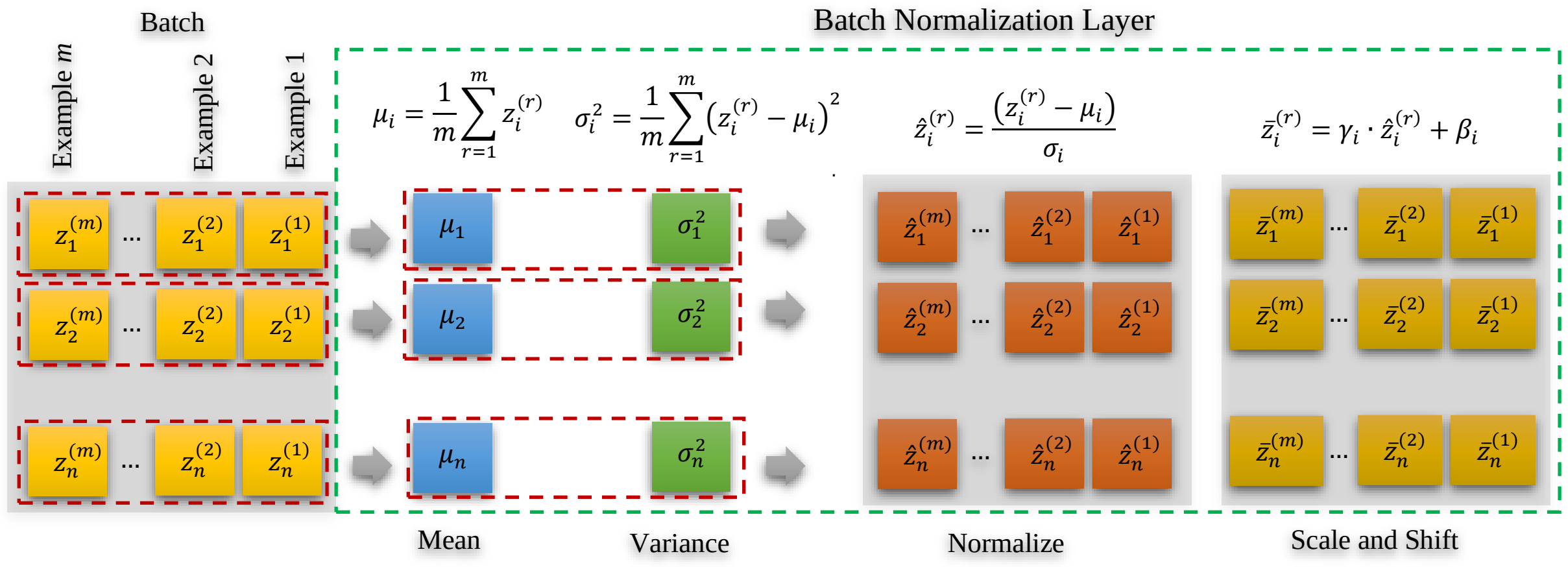

**Figure 4.18.** Diagrammatic representation of BN layer.

where $z_i^{(r)}$ represents the pre-activation of the $i$th unit for the $r$th instance in the batch. Adding a small value $\epsilon$ to the variance $\sigma_i^2$ is a regularization technique in BN to avoid division by zero and to prevent potential numerical instability when the variance is close to zero. This regularization term is typically a small positive constant, such as $10^{-5}$ or $10^{-6}$, chosen to be sufficiently small to not significantly affect the overall scaling behavior of the network but large enough to prevent division by zero or very small variances. Including $\epsilon$ in the computation of the variance $\sigma_i^2$ ensures that even if all activations for a particular unit are the same (resulting in zero variance), the denominator in the normalization step doesn't become zero. This is crucial for numerical stability during training.

Each pre-activation $z_i^{(r)}$ is normalized using the computed mean $\mu_i$ and standard deviation $\sigma_i$. Normalized pre-activations $\hat{z}_i^{(r)}$ for the $i$th hidden unit can be expressed as:

$$\hat{z}_i^{(r)} = \frac{z_i^{(r)} - \mu_i}{\sigma_i} \quad \forall i, r. \tag{4.135.3}$$

After normalizing the pre-activations $z_i^{(r)}$, the normalized pre-activations $\hat{z}_i^{(r)}$ are then scaled using the parameter $\gamma_i$ and shifted using the parameter $\beta_i$ to create the final outputs of BN node, $\bar{z}_i^{(r)}$. These parameters enable the network to learn a new mean and variance for each feature dimension, effectively allowing the model to adjust the normalized values to better suit the task at hand.

$$\bar{z}_i^{(r)} = \gamma_i \cdot \hat{z}_i^{(r)} + \beta_i \quad \forall i, r, \tag{4.135.4}$$

where $\gamma_i$ and $\beta_i$ are learnable parameters specific to the $i$th hidden unit.

The scale parameter $\gamma_i$ allows the network to scale the normalized values. It acts as a learnable scaling factor, allowing the network to increase or decrease the magnitude of the normalized activations as needed. This parameter enables the network to adaptively adjust the range of values for each feature dimension.

The shift parameter $\beta_i$ allows the network to shift the normalized values. It acts as a learnable bias term, allowing the network to shift the mean of the normalized activations to a new value. This parameter enables the network to adaptively adjust the center of the distribution for each feature dimension.

**Remarks:**

- Note that BN node has only one input, and its job is to perform the normalization and scaling.
- $\bar{z}_i^{(r)}$ is the pre-activation output of the $i$th node, when the $r$th batch instance passes through it. This value would otherwise have been set to $z_i^{(r)}$, if we had not applied BN.
- BN introduces two learnable parameters to each node it is applied to: $\gamma_i$ and $\beta_i$. In the final stage of BN, the normalized pre-activations undergo a linear transformation by scaling with $\gamma_i$ and offsetting with $\beta_i$. Conceptually, $\gamma_i$ corresponds to the standard deviation, while $\beta_i$ corresponds to the mean. Interestingly, this operation is the exact inverse of the normalization process applied initially! Initially, the pre-activations values are normalized using the batch mean and batch standard deviation, whereas $\gamma_i$ and $\beta_i$ are updated through SGD during training. The BN node initializes with $\gamma_i = 1$ and $\beta_i = 0$, ensuring that at the beginning of training, this linear transformation has no effect, allowing BN to normalize the pre-activations as intended. However, as the network progresses in learning, it may determine that adjusting the pre-activations of a particular layer by denormalizing is beneficial for minimizing the overall cost. Therefore, if BN proves unhelpful, the network can learn to deactivate it on a layer-by-layer basis. Since $\gamma_i$ and $\beta_i$ are continuous variables, the network can flexibly decide the degree to which it wishes to denormalize the outputs, depending on what yields the lowest cost. This adaptive mechanism enables the network to dynamically adjust its normalization strategy, enhancing its efficiency and effectiveness in minimizing the cost.
- In other words, although the normalization process occurs at each unit, the learning process itself determines the best normalization that is required to maximize the performance of the model (loss minimization). Therefore, it has the capability to nullify the effects of the normalization if it is not necessary for some feature, or it can also use the normalization effects. The important point to remember is that, when BN is used, the learning algorithm will learn to use normalization optimally.
- The variables $\gamma_i$ and $\beta_i$ are learned parameters that allow the new variable to have any mean and standard deviation. At first glance, this may seem useless—why did we set the mean to 0, and then introduce a parameter that allows it to be set back to any arbitrary value $\beta$? The answer is that the new parametrization can represent the same family of functions of the input as the old parametrization, but the new parametrization has different learning dynamics. In the old parametrization, the mean was determined by a complicated interaction between the parameters in the preceding layers. In the new parametrization, the mean of $\gamma_i \cdot \hat{z}_i^{(r)} + \beta_i$ is determined solely by $\beta_i$. The new parametrization is much easier to learn with GD.
- One might wonder whether it might make sense to simply set each $\beta_i$ to 0 and each $\gamma_i$ to 1, but doing so reduces the representation power of the network. For example, if we make this transformation, then the Sigmoid units will be operating within their linear regions, especially if the normalization is performed just before activation. Recall, multilayer networks do not gain power from depth without nonlinear activations. Therefore, allowing some "wiggle" with these parameters and learning them in a data-driven manner makes sense.
- Note that, the parameter $\beta_i$ plays the role of a learned bias variable, and therefore we do not need additional bias units in these layers.

- In many cases, if you add a BN layer as the very first layer of your NN, you do not need to standardize your training set; the BN layer will do it for you (well, approximately, since it only looks at one batch at a time, and it can also rescale and shift each input feature).

The whole BN operation is summarized step by step in vector notations as follows [101]:

$$\boldsymbol{\mu}_B = \frac{1}{m_B}\sum_{r=1}^{m_B} \mathbf{z}^{(r)}, \tag{4.136.1}$$

$$\boldsymbol{\sigma}_B^2 = \frac{1}{m_B}\sum_{r=1}^{m_B} \left(\mathbf{z}^{(r)} - \boldsymbol{\mu}_B\right)^2, \tag{4.136.2}$$

$$\hat{\mathbf{z}}^{(r)} = \frac{\mathbf{z}^{(r)} - \boldsymbol{\mu}_B}{\sqrt{\boldsymbol{\sigma}_B^2 + \boldsymbol{\epsilon}}}, \tag{4.136.3}$$

$$\bar{\mathbf{z}}^{(r)} = \boldsymbol{\gamma} \odot \hat{\mathbf{z}}^{(r)} + \boldsymbol{\beta}, \tag{4.136.4}$$

where

- $\boldsymbol{\mu}_B$ is the vector of input means, evaluated over the whole mini-batch $B$ (it contains one mean per input).
- $\boldsymbol{\sigma}_B$ is the vector of input standard deviations, also evaluated over the whole mini-batch (it contains one standard deviation per input).
- $m_B$ is the number of instances in the mini-batch.
- $\hat{\mathbf{z}}^{(r)}$ is the vector of zero-centered and normalized inputs for instance $r$.
- $\boldsymbol{\gamma}$ is the scale parameter vector for the layer (it contains one scale parameter per input).
- $\odot$ represents element-wise multiplication.
- $\boldsymbol{\beta}$ is the shift (offset) parameter vector for the layer (it contains one offset parameter per input).
- $\boldsymbol{\epsilon}$ is a vector of a tiny number that avoids division by zero (typically $10^{-5}$). This is called a smoothing term.
- $\bar{\mathbf{z}}^{(r)}$ is the output of the BN operation. It is a rescaled and shifted version of the inputs.
- The arithmetic is element-wise, so each element $z_i^{(r)}$ of $\mathbf{z}^{(r)}$ is normalized by subtracting $\mu_i$ and dividing by $\sigma_i$.

### During Backpropagation

When BN is applied, there are several implications for the BP algorithm, which is used to compute the gradients of the loss function with respect to the parameters of the network. Here are some changes in the BP algorithm caused by BN:

- During BP, gradients are calculated with respect to the parameters of the normalization layer. This includes scaling and shifting parameters used in the normalization process.
- BN introduces additional parameters (mean and variance) per feature map or neuron. During BP, gradients with respect to these parameters need to be computed.

**Step 1: Derivatives with respect to parameters**

Assuming that we have already propagated backward up to the output of the BN node, thereby possessing each $\partial\mathcal{L}/\partial\bar{z}_i^{(r)}$ available, the derivatives with respect to the two parameters, $\beta_i$ and $\gamma_i$, can be computed as follows. Since $\beta_i$ is used to calculate all the outputs $\bar{z}_i^{(r)}$ where $r = \{1, \dots, m\}$, the gradients will be summed during BP, (using the multivariable chain rule), see Figure 4.17:

$$\frac{\partial\mathcal{L}}{\partial\beta_i} = \sum_{r=1}^{m} \frac{\partial\mathcal{L}}{\partial\bar{z}_i^{(r)}}\frac{\partial\bar{z}_i^{(r)}}{\partial\beta_i} = \sum_{r=1}^{m} \frac{\partial\mathcal{L}}{\partial\bar{z}_i^{(r)}}. \tag{4.137}$$

Similarly, $\gamma_i$ is used to calculate all the outputs $\bar{z}_i^{(r)}$ where $r = \{1, \dots, m\}$, the gradients will be summed during BP:

$$\frac{\partial \mathcal{L}}{\partial \gamma_i} = \sum_{r=1}^{m} \frac{\partial \mathcal{L}}{\partial \bar{z}_i^{(r)}} \frac{\partial \bar{z}_i^{(r)}}{\partial \gamma_i} = \sum_{r=1}^{m} \frac{\partial \mathcal{L}}{\partial \bar{z}_i^{(r)}} \hat{z}_i^{(r)}, \tag{4.138}$$

where $\partial \mathcal{L}/\partial \bar{z}_i^{(r)}$ is the gradient of the loss with respect to the normalized outputs of BN node, and we used,

$$\frac{\partial \bar{z}_i^{(r)}}{\partial \beta_i} = \frac{\partial}{\partial \beta_i}\left(\gamma_i \cdot \hat{z}_i^{(r)} + \beta_i\right) = 1, \tag{4.139.1}$$

$$\frac{\partial \bar{z}_i^{(r)}}{\partial \gamma_i} = \frac{\partial}{\partial \gamma_i}\left(\gamma_i \cdot \hat{z}_i^{(r)} + \beta_i\right) = \hat{z}_i^{(r)}. \tag{4.139.2}$$

**Step 2: Derivative with respect to pre-activation**

We compute the derivative of the loss with respect to the pre-activation $z_i^{(r)}$ before normalization. One can compute the value of $\frac{\partial \mathcal{L}}{\partial z_i^{(r)}}$ in terms of $\hat{z}_i^{(r)}$, $\mu_i$, and $\sigma_i$, by observing that $z_i^{(r)}$ can be written as a function of only $\hat{z}_i^{(r)}$, $\mu_i$, and $\sigma_i^2$ : $\left(z_i^{(r)} = \hat{z}_i^{(r)}\sigma_i + \mu_i\right)$, see Figure 4.17. Observe that $\mu_i$, and $\sigma_i$ are not treated as constants, but as variables because they depend on the batch at hand. Therefore, we have the following:

$$\begin{aligned}\frac{\partial \mathcal{L}}{\partial z_i^{(r)}} &= \frac{\partial \mathcal{L}}{\partial \hat{z}_i^{(r)}} \frac{\partial \hat{z}_i^{(r)}}{\partial z_i^{(r)}} + \frac{\partial \mathcal{L}}{\partial \mu_i} \frac{\partial \mu_i}{\partial z_i^{(r)}} + \frac{\partial \mathcal{L}}{\partial \sigma_i^2} \frac{\partial \sigma_i^2}{\partial z_i^{(r)}} \\ &= \frac{\partial \mathcal{L}}{\partial \hat{z}_i^{(r)}}\left(\frac{1}{\sigma_i}\right) + \frac{\partial \mathcal{L}}{\partial \mu_i}\left(\frac{1}{m}\right) + \frac{\partial \mathcal{L}}{\partial \sigma_i^2}\left(\frac{2\left(z_i^{(r)} - \mu_i\right)}{m}\right),\end{aligned} \tag{4.140}$$

where, we used

$$\frac{\partial \hat{z}_i^{(r)}}{\partial z_i^{(r)}} = \frac{\partial}{\partial z_i^{(r)}}\left(\frac{z_i^{(r)} - \mu_i}{\sigma_i}\right) = \frac{1}{\sigma_i}, \tag{4.141.1}$$

$$\frac{\partial \mu_i}{\partial z_i^{(r)}} = \frac{\partial}{\partial z_i^{(r)}}\left(\frac{1}{m}\sum_{q=1}^{m} z_i^{(q)}\right) = \frac{1}{m}\frac{\partial}{\partial z_i^{(r)}}\left(z_i^{(r)}\right) = \frac{1}{m}, \tag{4.141.2}$$

$$\frac{\partial \sigma_i^2}{\partial z_i^{(r)}} = \frac{\partial}{\partial z_i^{(r)}}\left(\frac{1}{m}\sum_{q=1}^{m}\left(z_i^{(q)} - \mu_i\right)^2 + \epsilon\right) = \frac{1}{m}\frac{\partial}{\partial z_i^{(r)}}\left(z_i^{(r)} - \mu_i\right)^2 = \frac{2\left(z_i^{(r)} - \mu_i\right)}{m}. \tag{4.141.3}$$

We need to evaluate each of the three partial derivatives on the right-hand side of (4.140). The partial derivatives of the loss with respect to the $\hat{z}_i^{(r)}$ can be computed as follows:

$$\begin{aligned}\frac{\partial \mathcal{L}}{\partial \hat{z}_i^{(r)}} &= \frac{\partial \mathcal{L}}{\partial \bar{z}_i^{(r)}} \frac{\partial \bar{z}_i^{(r)}}{\partial \hat{z}_i^{(r)}} \\ &= \gamma_i \frac{\partial \mathcal{L}}{\partial \bar{z}_i^{(r)}} \quad \left[\text{Since } \bar{z}_i^{(r)} = \gamma_i \cdot \hat{z}_i^{(r)} + \beta_i\right].\end{aligned} \tag{4.142}$$

Therefore, using (4.142) and (4.140), we have the following:

$$\frac{\partial \mathcal{L}}{\partial z_i^{(r)}} = \frac{\partial \mathcal{L}}{\partial \bar{z}_i^{(r)}}\left(\frac{\gamma_i}{\sigma_i}\right) + \frac{\partial \mathcal{L}}{\partial \mu_i}\left(\frac{1}{m}\right) + \frac{\partial \mathcal{L}}{\partial \sigma_i^2}\left(\frac{2\left(z_i^{(r)} - \mu_i\right)}{m}\right). \tag{4.143}$$

**Step 3: Derivatives with respect to mean and variance**

Again, using multivariable chain rule we add the gradients coming from $\hat{z}_i^{(r)}$ to compute the gradient with respect to $\sigma_i^2$. The partial derivatives of the loss with respect to the variance can be computed as follows:

$$\begin{aligned}\frac{\partial \mathcal{L}}{\partial \sigma_i^2} &= \sum_{j=1}^{m} \frac{\partial \mathcal{L}}{\partial \hat{z}_i^{(j)}} \frac{\partial \hat{z}_i^{(j)}}{\partial \sigma_i^2} \\ &= -\frac{1}{2\sigma_i^3} \sum_{j=1}^{m} \frac{\partial \mathcal{L}}{\partial \hat{z}_i^{(j)}} \left(z_i^{(j)} - \mu_i\right) \\ &= -\frac{1}{2\sigma_i^3} \sum_{j=1}^{m} \frac{\partial \mathcal{L}}{\partial \bar{z}_i^{(j)}} \gamma_i \left(z_i^{(j)} - \mu_i\right) \\ &= -\frac{\gamma_i}{2\sigma_i^3} \sum_{j=1}^{m} \frac{\partial \mathcal{L}}{\partial \bar{z}_i^{(j)}} \left(z_i^{(j)} - \mu_i\right),\end{aligned} \tag{4.144}$$

where,

$$\frac{\partial \hat{z}_i^{(j)}}{\partial \sigma_i^2} = \frac{\partial \hat{z}_i^{(j)}}{\partial \sigma_i} \frac{\partial \sigma_i}{\partial \sigma_i^2} = \frac{\partial}{\partial \sigma_i}\left[\frac{z_i^{(j)} - \mu_i}{\sigma_i}\right] \frac{\partial \sigma_i}{\partial \sigma_i^2} = \left[-\frac{z_i^{(j)} - \mu_i}{\sigma_i^2}\right]\left(\frac{1}{2\sigma_i}\right) = -\frac{z_i^{(j)} - \mu_i}{2\sigma_i^3}. \tag{4.145}$$

Since $\mu_i$ is used to calculate not only $\hat{z}_i^{(r)}$ but also $\sigma_i^2$, we add the respective gradients. The partial derivatives of the loss with respect to the mean can be computed as follows:

$$\begin{aligned}\frac{\partial \mathcal{L}}{\partial \mu_i} &= \sum_{j=1}^{m} \frac{\partial \mathcal{L}}{\partial \hat{z}_i^{(j)}} \frac{\partial \hat{z}_i^{(j)}}{\partial \mu_i} + \frac{\partial \mathcal{L}}{\partial \sigma_i^2} \frac{\partial \sigma_i^2}{\partial \mu_i} \\ &= -\frac{1}{\sigma_i} \sum_{j=1}^{m} \frac{\partial \mathcal{L}}{\partial \hat{z}_i^{(j)}} \\ &= -\frac{\gamma_i}{\sigma_i} \sum_{j=1}^{m} \frac{\partial \mathcal{L}}{\partial \bar{z}_i^{(j)}},\end{aligned} \tag{4.146}$$

where, we used

$$\frac{\partial \sigma_i^2}{\partial \mu_i} = \frac{\partial}{\partial \mu_i}\left[\frac{1}{m} \sum_{j=1}^{m} \left(z_i^{(j)} - \mu_i\right)^2 + \epsilon\right] = \frac{1}{m} \sum_{j=1}^{m} -2\left(z_i^{(j)} - \mu_i\right) = -\frac{2}{m} \sum_{j=1}^{m} \left(z_i^{(j)} - \mu_i\right) = 0, \tag{4.147.1}$$

$$\frac{\partial \hat{z}_i^{(j)}}{\partial \mu_i} = \frac{\partial}{\partial \mu_i}\left[\frac{z_i^{(j)} - \mu_i}{\sigma_i}\right] = -\frac{1}{\sigma_i}. \tag{4.147.2}$$

By substituting the derivatives of the loss with respect to mean and variance into the expression for $\frac{\partial \mathcal{L}}{\partial z_i^{(r)}}$, (4.143), we establish a recursive relationship between the gradient before normalization, $\frac{\partial \mathcal{L}}{\partial z_i^{(r)}}$, and after normalization, $\frac{\partial \mathcal{L}}{\partial \bar{z}_i^{(r)}}$.

$$\begin{aligned}\frac{\partial \mathcal{L}}{\partial z_i^{(r)}} &= \frac{\partial \mathcal{L}}{\partial \bar{z}_i^{(r)}}\left(\frac{\gamma_i}{\sigma_i}\right) + \frac{\partial \mathcal{L}}{\partial \mu_i}\left(\frac{1}{m}\right) + \frac{\partial \mathcal{L}}{\partial \sigma_i^2}\left(\frac{2\left(z_i^{(r)} - \mu_i\right)}{m}\right) \\ &= \frac{\partial \mathcal{L}}{\partial \bar{z}_i^{(r)}}\left(\frac{\gamma_i}{\sigma_i}\right) + \left[-\frac{\gamma_i}{\sigma_i} \sum_{j=1}^{m} \frac{\partial \mathcal{L}}{\partial \bar{z}_i^{(j)}}\right]\left(\frac{1}{m}\right) + \left[-\frac{\gamma_i}{2\sigma_i^3} \sum_{j=1}^{m} \frac{\partial \mathcal{L}}{\partial \bar{z}_i^{(j)}} \left(z_i^{(j)} - \mu_i\right)\right]\left(\frac{2\left(z_i^{(r)} - \mu_i\right)}{m}\right) \\ &= \frac{m\gamma_i}{m\sigma_i} \frac{\partial \mathcal{L}}{\partial \bar{z}_i^{(r)}} - \frac{\gamma_i}{m\sigma_i} \sum_{j=1}^{m} \frac{\partial \mathcal{L}}{\partial \bar{z}_i^{(j)}} - \frac{\gamma_i\left(z_i^{(r)} - \mu_i\right)}{m\sigma_i^3} \sum_{j=1}^{m} \frac{\partial \mathcal{L}}{\partial \bar{z}_i^{(j)}} \left(z_i^{(j)} - \mu_i\right) \\ &= \frac{\gamma_i}{m\sigma_i}\left\{m \frac{\partial \mathcal{L}}{\partial \bar{z}_i^{(r)}} - \sum_{j=1}^{m} \frac{\partial \mathcal{L}}{\partial \bar{z}_i^{(j)}} - \frac{\left(z_i^{(r)} - \mu_i\right)}{\sigma_i^2} \sum_{j=1}^{m} \frac{\partial \mathcal{L}}{\partial \bar{z}_i^{(j)}} \left(z_i^{(j)} - \mu_i\right)\right\}.\end{aligned} \tag{4.148}$$

This provides a full view of the BP of the loss through the batch-normalization layer corresponding to the BN node. The other aspects of BP remain similar to the traditional case.

### During Test Time

During training, BN standardizes its inputs using the mean and standard deviation computed within each batch, and then it rescales and offsets them using learned parameters ($\gamma_i$ and $\beta_i$). However, at test time, when making predictions for individual instances (i.e., during inference, when the model is deployed for making predictions, the model might process data one example at a time) or when dealing with small batches where batch statistics may not be reliable, we cannot directly compute the mean and standard deviation for normalization.

One solution could be to wait until the end of training, then run the whole training set through the NN and compute the mean and standard deviation of each input of the BN layer, and then treated as constants during testing time. These "final" input means and standard deviations could then be used instead of the batch input means and standard deviations when making predictions. However, this approach may be computationally expensive and not always practical. Mathematically, once the network has been trained, we use the normalization

$$\hat{z} = \frac{z - \mathbb{E}[z]}{\sqrt{\mathrm{Var}[z] + \epsilon}}, \tag{4.149}$$

using the population, rather than mini-batch, statistics. Neglecting $\epsilon$, these normalized activations have the same mean 0 and variance 1 as during training. In (4.149), we use the unbiased variance and mean estimated values (process multiple training mini-batches (B), each of size $m$, and average over them:)

$$\mathrm{Var}[z] = \mathbb{E}_{\mathrm{B}}\left[\frac{m}{m-1}\sigma_{\mathrm{B}}^2\right], \tag{4.150.1}$$

$$\mathbb{E}[z] = \mathbb{E}_{\mathrm{B}}[\mu_{\mathrm{B}}], \tag{4.150.2}$$

where the expectation is over training mini-batches of size $m$ and $\sigma_{\mathrm{B}}^2$ and $\mu_{\mathrm{B}}$ are their sample variances and means. Then,

$$\begin{aligned}
\bar{z} &= \gamma \cdot \hat{z} + \beta \\
&= \gamma \frac{z - \mathbb{E}[z]}{\sqrt{\mathrm{Var}[z] + \epsilon}} + \beta \\
&= \frac{\gamma z}{\sqrt{\mathrm{Var}[z] + \epsilon}} - \frac{\gamma \mathbb{E}[z]}{\sqrt{\mathrm{Var}[z] + \epsilon}} + \beta \\
&= \frac{\gamma}{\sqrt{\mathrm{Var}[z] + \epsilon}} z + \left(\beta - \frac{\gamma \mathbb{E}[z]}{\sqrt{\mathrm{Var}[z] + \epsilon}}\right).
\end{aligned} \tag{4.151}$$

Instead, most implementations of BN, including those in frameworks like Mathematica, estimate these final statistics during training by using a moving average (or an exponentially weighted average) of the layer's input means and standard deviations computed during training across multiple mini-batches. The moving average acts as a good proxy for the mean and variance of the data. It is much more efficient because the calculation is incremental — we have to remember only the most recent moving average. Since the means and variances are fixed during inference, the normalization is a simple linear transformation during inference.

During training, BN computes batch-wise statistics for mean $\mu_{\mathrm{B}}$ and variance $\sigma_{\mathrm{B}}^2$ for each hidden unit. These batch statistics are then used for normalization within each mini-batch. However, to ensure stability and generalization, BN also estimates population statistics (mean $\mu$ and variance $\sigma^2$ across the entire training dataset) by maintaining exponential moving averages (EMA) of the batch statistics across training steps. Initially, at the start of training, the EMA values are typically initialized to zero or some other appropriate value. As the training progresses, for each mini-batch processed, the mean and variance are calculated. These calculated mean and variance values are then used to update the EMA estimates using the EMA update equations. The EMA update equations are as follows:

$$\mu_{\mathrm{EMA;new}} = \alpha \cdot \mu_{\mathrm{EMA;old}} + (1-\alpha) \cdot \mu_{\mathrm{B}}, \tag{4.152.1}$$

$$\sigma_{\mathrm{EMA;new}}^2 = \alpha \cdot \sigma_{\mathrm{EMA;old}}^2 + (1-\alpha) \cdot \sigma_{\mathrm{B}}^2, \tag{4.152.2}$$

where $\mu_{\mathrm{B}}$ and $\sigma_{\mathrm{B}}^2$ are the calculated mean and variance for the current mini-batch and $\alpha$ is a decay rate (momentum) which controls the contribution of new batch statistics to the estimation of population statistics, typically set to a value close to 1 (e.g., 0.9). After training is completed, the final EMA values represent the aggregated statistics over the entire training dataset. During inference, these final EMA values are used for normalization. The final EMA estimates of the mean ($\mu_{\text{EMA}}$) and variance ($\sigma_{\text{EMA}}^2$) are used in the normalization formula:

$$\hat{z} = \frac{z - \mu_{\text{EMA}}}{\sqrt{\sigma_{\text{EMA}}^2 + \epsilon}}. \tag{4.153}$$

So, during inference, $\mu_{\text{EMA}}$ and $\sigma_{\text{EMA}}^2$ indeed refer to the final estimates of the mean and variance, which are obtained after continuous updates during training using the EMA technique. Using population statistics maintains consistency in the normalization process across training and inference, leading to more reliable model performance.

**Remarks:**

1. To sum up, four parameter vectors are learned in each batch-normalized layer: $\boldsymbol{\gamma}$ (the output scale vector) and $\boldsymbol{\beta}$ (the output offset vector) are learned through regular BP, and $\boldsymbol{\mu}$ (the final input mean vector) and $\boldsymbol{\sigma}$ (the final input standard deviation vector) are estimated using an exponential moving average. Note that, $\boldsymbol{\mu}$ and $\boldsymbol{\sigma}$ are estimated during training, but they are used only after training.
2. Ioffe and Szegedy demonstrated that BN considerably improved all the DNN they experimented with, leading to a huge improvement in the ImageNet classification task (ImageNet is a large database of images classified into many classes, commonly used to evaluate computer vision systems).
3. The vanishing gradients problem was strongly reduced to the point that they could use saturating AFs such as the Tanh and even the Logistic AF.
4. The networks were also much less sensitive to the weight initialization.
5. The authors were able to use much larger learning rates —because there are no extreme values in the normalized activations—significantly speeding up the learning process.
6. BN allows layers to learn more independently from each other, because large values in one layer won’t excessively influence the calculations in the next layer.
7. In addition to improving training stability and convergence speed, BN also acts as a form of regularization, reducing the need for other regularization techniques such as dropout. (Regularization is covered in the next chapter, but suffice it to say here that regularization helps a network generalize to data it hasn’t encountered previously.)
8. BN's effectiveness can be influenced by the choice of batch size. In general, larger batch sizes tend to provide more accurate estimates of the mean and variance, leading to more stable training. However, very small batch sizes might not capture the true statistical properties of the data well enough, potentially degrading the performance of BN.
9. While BN offers numerous benefits, it also introduces additional computational complexity to the model due to the need to compute mean and variance statistics and apply normalization during both training and inference. Moreover, there is a runtime penalty: the NN makes slower predictions due to the extra computations required at each layer. In some cases, this complexity might be undesirable, especially in resource-constrained environments.
10. Noteworthy is that after the initial idea was published, subsequent work indicated that the reason BN works is different than the initial explanation [116-118]. Batch Norm impacts network training in a fundamental way: it makes the landscape of the corresponding optimization problem significantly more smooth. The smoother optimization landscape induced by BN ensures that the gradients are more predictive. This means that the gradients provide more reliable information about the direction of steepest descent, leading to more effective updates of the network parameters. With BN, the smoother optimization landscape allows for the use of a larger range of learning rates during training. This flexibility in choosing learning rates accelerates network convergence, enabling faster training of DNNs.
11. Since its introduction, several adaptations and extensions of BN have been proposed to address specific challenges or improve its performance further. This includes techniques like LN [119], Group Normalization [120], Instance Normalization [121], Switchable Normalization [122], and Weight Normalization [123], each with its own advantages and suitable use cases.

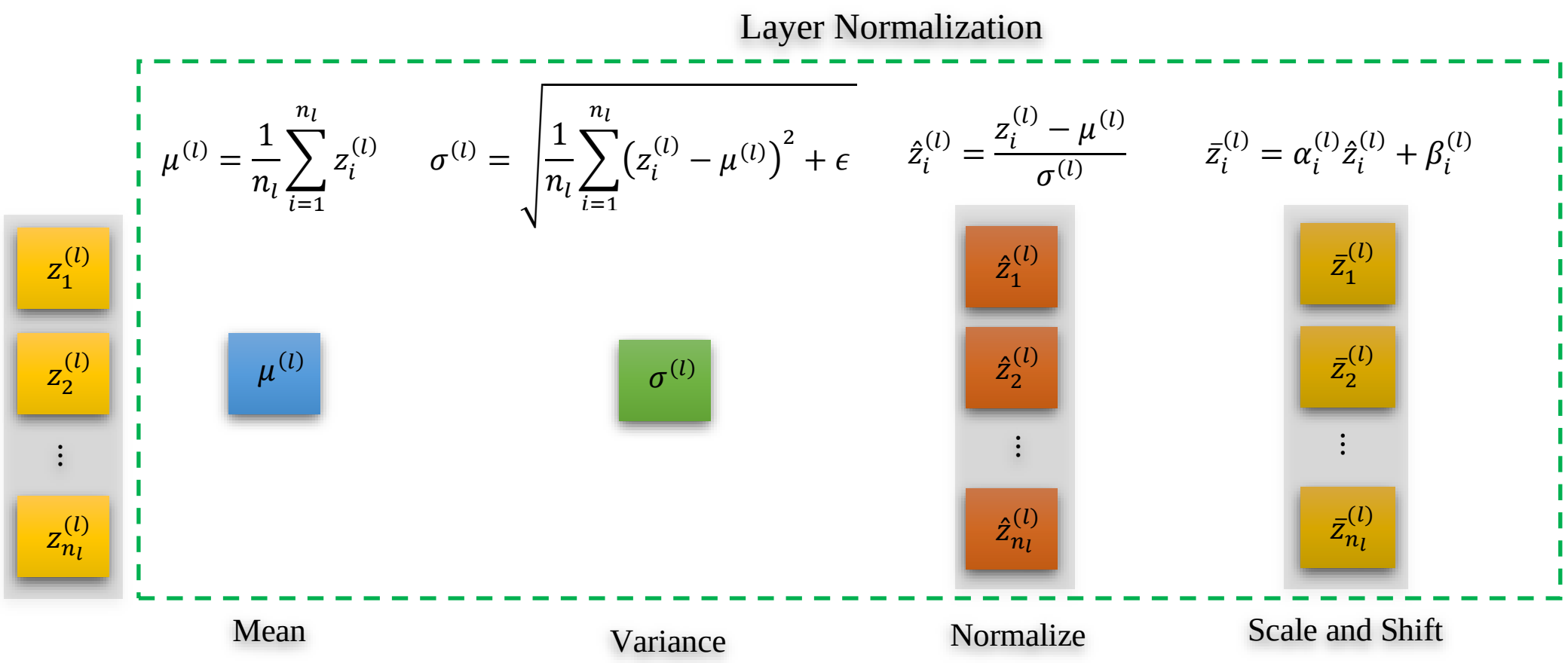

**Figure 4.19.** Diagrammatic representation of LN.

Although BN can be adapted to RNNs, a more effective approach is LN [119]. LN is a normalization technique introduced as an alternative to BN, particularly useful in scenarios where the concept of a "batch" might not be applicable or meaningful. Unlike BN, which operates across the batch dimension, LN normalizes across the features dimension (e.g., across the units in a fully connected layer). LN operates similarly to BN but normalizes the activations within each layer instead of across mini-batches. This ensures that the mean and variance of the summed inputs within each layer are fixed, reducing the internal covariate shift and stabilizing the training process. The normalization process can be represented as follows:

$$\mu^{(l)} = \frac{1}{n_l}\sum_{i=1}^{n_l} z_i^{(l)}, \tag{4.154.1}$$

$$\sigma^{(l)} = \sqrt{\frac{1}{n_l}\sum_{i=1}^{n_l}\left(z_i^{(l)} - \mu^{(l)}\right)^2 + \epsilon}, \tag{4.154.2}$$

$$\hat{z}_i^{(l)} = \frac{z_i^{(l)} - \mu^{(l)}}{\sigma^{(l)}}, \tag{4.154.3}$$

$$\bar{z}_i^{(l)} = \gamma_i^{(l)}\hat{z}_i^{(l)} + \beta_i^{(l)}, \tag{4.154.4}$$

where $n_l$ denotes the number of hidden units in a layer. $\gamma_i^{(l)}$ and $\beta_i^{(l)}$ are learned parameters, see Figure 4.19. These parameters are analogous to the parameters $\gamma_i$ and $\beta_i$ on BN. The purpose of these parameters is to re-scale the normalized values and add bias in a learnable way. Under LN, all the hidden units in a layer share the same normalization terms $\mu^{(l)}$ and $\sigma^{(l)}$, but different training cases have different normalization terms. Unlike BN, LN does not impose any constraint on the size of a mini-batch and it can be used in the pure online regime with batch size 1.

# CHAPTER 5

# LEARNING RATE SCHEDULES AND ADAPTIVE ALGORITHMS

In deep learning, the optimization process lies at the heart of training NNs. Central to this optimization is the notion of the learning rate, a crucial hyperparameter that dictates the step size in the parameter space during optimization. Selecting an appropriate learning rate and its schedule significantly impacts the convergence speed and the final performance of the model. In this chapter, we delve into various learning rate schedules and adaptive algorithms that play pivotal roles in optimizing NNs. By understanding these techniques, practitioners can fine-tune their training processes and achieve better results in their machine-learning endeavors.

The chapter initiates an exploration of learning rate schedules, which entail predefined strategies for altering the learning rate throughout the training process. Learning rate schedules include step decay, inverse time decay, exponential decay, polynomial decay with warm restart, cyclical learning rate, stochastic gradient descent with warm restarts (cosine decay), exponential decay sine wave learning rate, Hessian-aware learning rate decay, etc., each with its own advantages and drawbacks. Understanding and appropriately implementing these schedules are crucial for achieving optimal convergence without encountering issues such as slow convergence or overshooting the minima.

Transitioning from learning rate schedules, we explore accelerated gradient descent, a variant of the traditional GD algorithm designed to speed up convergence. Two popular accelerated gradient descent algorithms are SGD with momentum and Nesterov accelerated gradient descent. In both cases, the momentum term helps the optimization algorithm to continue moving in the same direction or accelerate in the relevant direction, even if the gradient changes direction frequently or the surface of the loss function is highly irregular. Both methods help in smoothing out the updates, which is particularly useful when dealing with noisy or high-variance gradients common in SGD. The momentum term helps navigate through the irregularities of the loss surface more effectively, leading to a smoother and often faster path to the minimum. By leveraging past gradients, these methods can speed up convergence, reducing the time and computational resources needed for training. This leads to faster convergence and better overall performance in training NNs and other machine-learning models.

Following the discussion on accelerated gradient descent, we turn our attention to adaptive learning rate algorithms. Adaptive learning rate algorithms adaptively adjust the learning rate during training based on past gradients, and other relevant metrics. These algorithms aim to strike a balance between the benefits of using large learning rates for fast convergence and the stability provided by smaller learning rates to prevent overshooting or oscillations. Popular examples include AdaGrad, RMSProp, AdaDelta, Adam, AdaMax, Nadam, and AMSGRAD, each with its own approach to adaptively scale the learning rates for individual model parameters.

Lastly, we explore second-order optimization methods. Second-order optimization methods, such as Newton, Marquardt, and variants like conjugate gradient (Hestenes-Stiefel formula, Polak-Ribiere formula, Fletcher-Reeves formula), and quasi-Newton (rank one correction, DFP, and BFGS), etc., leverage information from the second derivatives of the loss function to guide the optimization process. By incorporating curvature information, these methods can converge faster and more accurately than first-order methods like GD. However, they often come with higher computational costs and memory requirements due to the need to compute and store second-order derivatives or their approximations.

In the subsequent sections, we explore each of these methodologies in depth, elucidating their strengths, weaknesses, and best practices for their application in optimizing NNs. Through a comprehensive understanding of these techniques, practitioners can navigate the complex landscape of optimization algorithms to achieve superior model performance.

## 5.1 Dynamic Learning Rate Decay

The learning rate is a crucial hyperparameter in the training of NNs, and its proper selection significantly impacts the performance and convergence of the model. Here are some key aspects highlighting the importance of the learning rate:

- The learning rate determines the size of the steps taken during optimization. A well-chosen learning rate can help the model converge to a solution faster. Too high of a learning rate may cause the model to overshoot the optimal solution, while too low of a learning rate may result in slow convergence.
- A proper learning rate can contribute to the stability of the optimization process. If the learning rate is too high, it may lead to oscillations or divergence, making the training process unstable. On the other hand, a too-low learning rate may cause the optimization to get stuck in local minima or plateaus.
- A well-tuned learning rate helps the optimization algorithm navigate through the loss landscape, avoiding getting stuck in local minima. It allows the algorithm to explore different regions of the parameter space and find the global minimum or a good approximation.
- In some cases, gradients computed during training can be noisy or have high variance. The learning rate acts as a regularization parameter, helping to smooth out the optimization process and reduce the impact of noisy gradients.
- A well-chosen learning rate can contribute to efficient resource utilization, as it allows the model to converge with fewer training iterations, saving computational time and resources.

Keeping the learning rate constant throughout the training process is a decision that might not be optimal in many cases. Using a low learning rate in the initial stages will result in the algorithm taking an extended period to approach an optimal solution. Conversely, employing a high initial learning rate enables the algorithm to quickly reach a reasonably good solution initially. However, if the elevated learning rate is sustained, the algorithm may either oscillate around the solution for an extended duration or diverge in an unstable manner. In both scenarios, maintaining a constant learning rate is not considered optimal. The idea is based on the observation that different stages of training can have different characteristics.

- In the early stages of training, when the model parameters are far from optimal, using large learning rates can help the model quickly navigate the loss landscape and converge faster.
- However, in the middle stages of training, gradually reducing the learning rate can help the model converge more accurately as it approaches the optimal parameters. Smaller learning rates help to fine-tune the model and make smaller adjustments to the parameters.
- In the late stages of training, very small learning rates may be used to carefully refine the model parameters, avoid overshooting the minimum, and approach the optimal solution.

It's essential to strike a balance between having a learning rate that is large enough to facilitate quick convergence in the early stages of training and small enough to allow for precise adjustments near the minimum. By adjusting the learning rate dynamically, you can potentially achieve faster convergence in the initial stages and more precise fine-tuning as the training progresses. It's worth mentioning that the optimal learning rate can depend on the specific problem, the model architecture, and the dataset. Therefore, experimentation and monitoring of the training process are essential to find the most effective learning rate strategy for a given task.

Techniques such as learning rate schedules, and adaptive methods have been developed to automate or guide the learning rate selection process. We have nothing to do with adaptive learning rate methods as they automatically update (decrease or increase) the learning rate value during training! We just need to select an optimizer that has an adaptive learning rate. Some commonly used adaptive learning rate methods are Adagrad, RMSprop, Adam, Adadelta, Nadam, and AMSGrad. More details about adaptive learning rate methods will be discussed in Sections 5.2 and 5.3. Here, we focus on the learning rate schedule techniques. We want to have a learning rate that starts (relatively) big and then decreases with the iterations. Allowing the learning rate to decay over time can naturally achieve the desired learning rate adjustment.

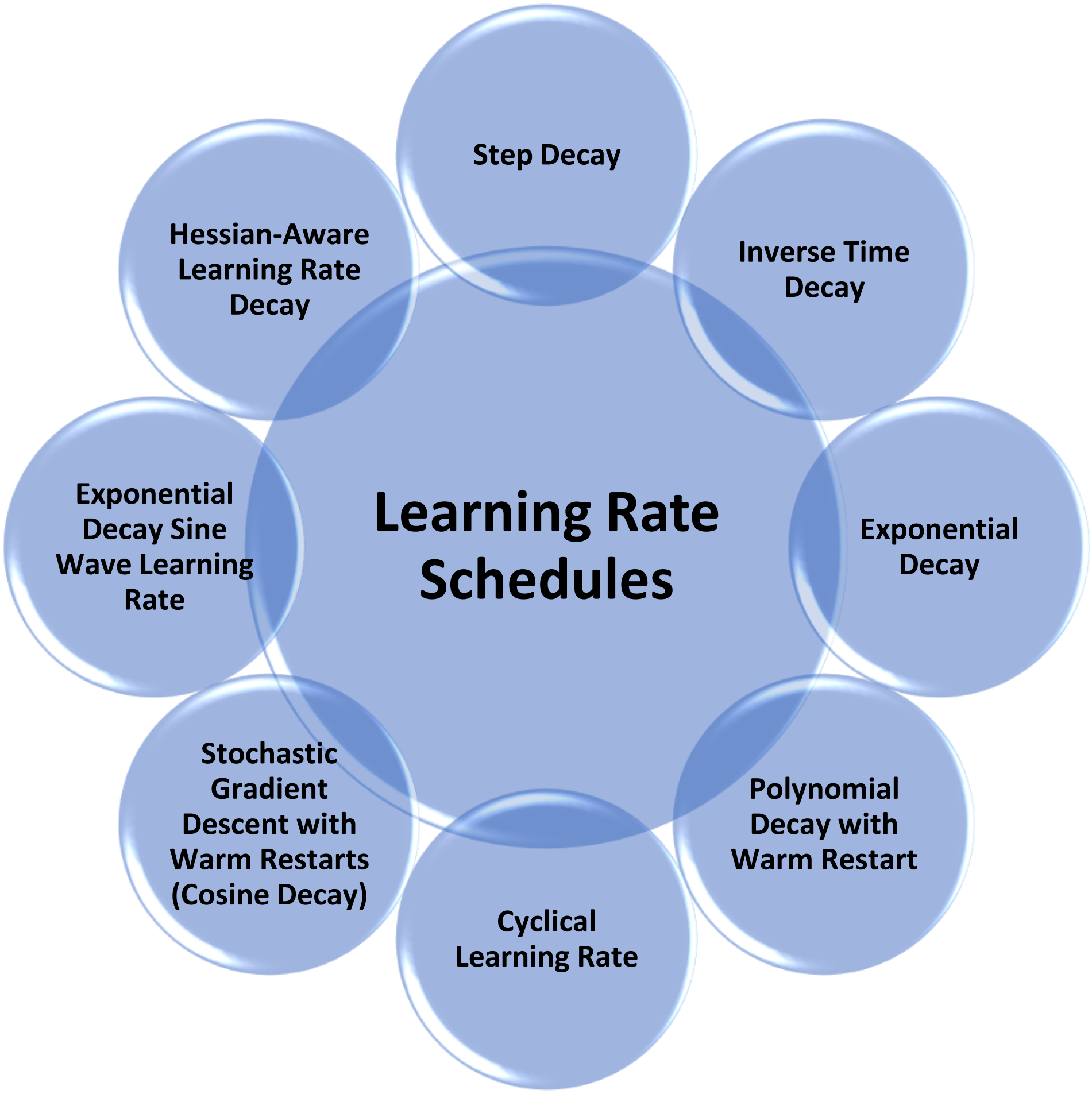

Understanding the terms iteration, epoch, and their relationship is crucial for properly configuring hyperparameters, including learning rate decay, during the training of NNs. The terms iteration and epoch (round in Mathematica) are interconnected yet distinct concepts. An epoch typically refers to one complete pass through the entire training dataset. During an epoch, the model sees each example in the dataset exactly once. Therefore, if you say you are training a model for 10 epochs, it means the model has gone through the entire dataset 10 times. The number of epochs is a hyperparameter that you can set before training. On the other hand, an iteration generally refers to one update of the model's weights based on a batch of training examples. In mini-batch stochastic gradient descent, an iteration occurs for every mini-batch of data processed. The model's parameters are updated after processing each mini-batch. So, technically, iterations are not the same as epochs. An epoch consists of multiple iterations, each corresponding to the processing of one mini-batch. The number of iterations per epoch is determined by the size of your dataset and the size of the mini-batches used during training. For example, if you have 1000 samples and a batch size of 50, it will take 20 iterations to complete one epoch $(1000/50 = 20)$.

Before delving into the intricate details of various learning rate schedules [124-136], let's start with a quick overview. Step decay involves reducing the learning rate by a constant factor after a fixed number of epochs. Inverse time decay adjusts the rate based on a fixed schedule over time. Exponential decay exponentially decreases the learning rate during training. Polynomial decay with warm restart combines polynomial decay with occasional restarts to escape local minima. Cyclical learning rate cyclically varies between minimum and maximum values, aiding convergence. SGD with warm restarts (cosine decay) employs cosine annealing to cyclically adjust the learning rate. Exponential decay sine wave learning rate combines exponential decay with a sine wave for a more dynamic learning rate schedule. Hessian-aware learning rate decay utilizes information from the Hessian matrix to adaptively adjust the learning rate during training. Each of these schedules serves specific purposes, and their effectiveness depends on factors such as the problem domain and model architecture. Note that, for each algorithm that decreases dynamically the learning rate will introduce new hyperparameters that you must optimize, adding some complexity to your model selection process.

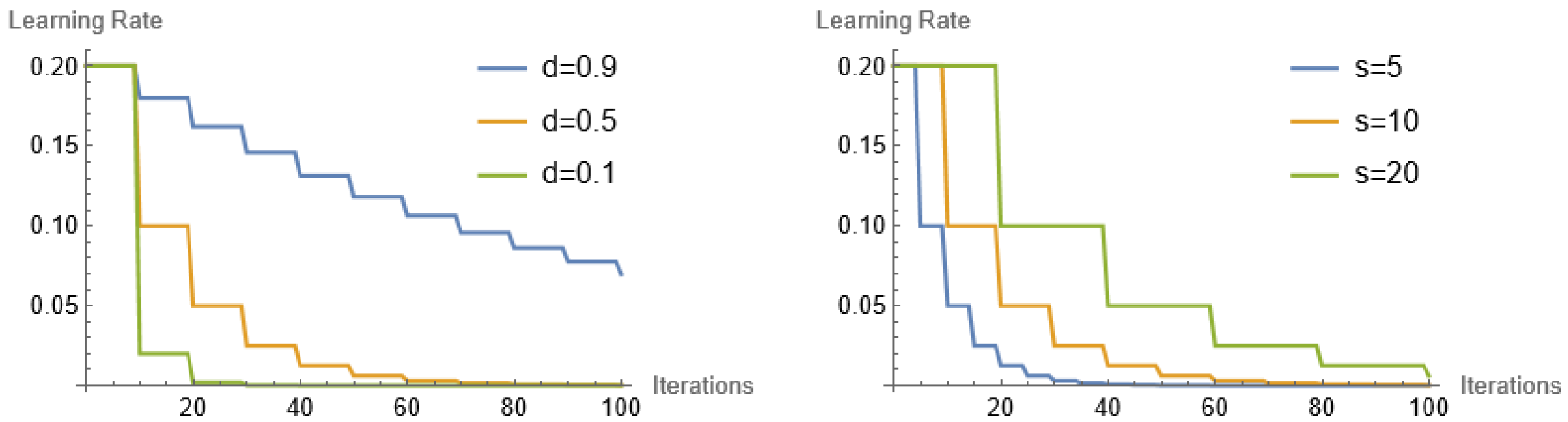

**Figure 5.1.** Left panel: Step decay learning rate schedule for different drop factors ($d = 0.1, 0.5, 0.9$) using step decay algorithm $s = 10$. Right panel: Step decay learning rate schedule with varied step sizes ($s = 5, 10, 20$) using $d = 0.5$.

Now, let's delve into the details of various learning rate schedules to understand their characteristics and applications.

### 5.1.1 Step Decay

Step decay is a simple learning rate decay strategy where the learning rate is reduced by a fixed factor after a certain number of training iterations. Mathematically, it can be written as [124, 125]

$$\alpha_t = \alpha_0 d^{\left\lfloor \frac{t}{s} \right\rfloor}, \tag{5.1}$$

where $\alpha_t$ is the learning rate at iteration $t$, $\alpha_0$ is the initial learning rate, $d$ is the constant factor (drop factor or decay factor) by which the learning rate drops each time (we can tune it), $s$ is the step size, indicating after how many iterations the learning rate should be decayed. For example, if the step size is 10, the learning rate will be adjusted every 10 iterations. The floor function $\lfloor x \rfloor$ rounds down the value of $x$ to the nearest integer. This means that the learning rate will remain constant for the first $s$ iterations and then decrease by a factor of $d$ every $s$ iterations.

Figure 5.1 (left) illustrates the step decay learning rate schedule for three different values of the drop factor, $d$. The initial learning rate is set to 0.2, and the step size is defined as 10 iterations. Each curve corresponds to a specific drop factor value, influencing the rate of learning rate decay. The plot demonstrates the impact of different drop factors ($d = 0.9$, $d = 0.5$, $d = 0.1$) on the learning rate over 100 iterations. As the iteration progresses, the learning rate undergoes step-wise decreases based on the specified drop factor. Figure 5.1 (right) presents the step decay learning rate schedule with three different step sizes $s$. The initial learning rate is set to 0.2, and a constant drop factor of 0.5 is applied. The plot showcases the influence of different step sizes ($s = 5$, $s = 10$, $s = 20$) on the learning rate decay over 100 iterations. As the iteration progresses, the learning rate undergoes step-wise decreases based on the specified step size.

It is important to have an idea of how fast the learning rate is decreasing.

- Fast decay (small $d$): If $d$ is a small value (close to 0), the learning rate will decrease rapidly. This might lead to a situation where the learning rate becomes very small early in the training process. While fast decay can help convergence in some cases, too rapid a decrease may hinder convergence or cause the model to converge to suboptimal solutions.
- Moderate decay (intermediate $d$): Choosing an intermediate value for $d$ can provide a balance between fast and slow decay. This allows the model to adapt to the data and make meaningful updates to the weights without decreasing the learning rate too quickly.
- Slow decay (large $d$): A large value for $d$ (close to 1) results in a slower learning rate decay.

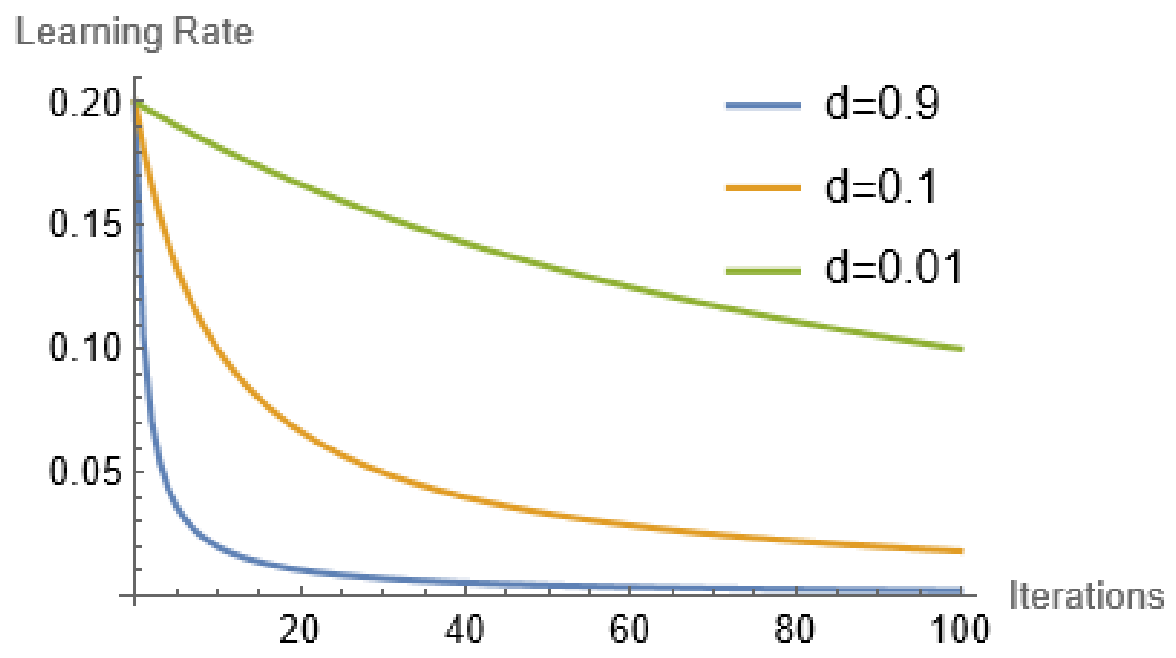

**Figure 5.2.** Inverse time decay learning rate schedule.

- The key is to find a balance that prevents the learning rate from becoming too small too quickly, as this can impede convergence. Experimentation and monitoring the model's performance during training are essential to selecting appropriate values for hyperparameters such as the initial learning rate, step size, and drop factor.
- The step size, $s$, in step decay plays a crucial role in determining how frequently the learning rate is adjusted during training.
- If the step size is small, the learning rate will be adjusted more frequently. This can lead to rapid decay of the learning rate, potentially causing the learning rate to become very small early in the training process. Too small a step size might hinder convergence as the model may not have sufficient time to explore the solution space effectively.
- A larger step size results in less frequent adjustments to the learning rate. This can slow down the decay process, allowing the model to converge more gradually and potentially reach a more stable solution. However, setting the step size too large might slow down the training process and result in suboptimal convergence.

### 5.1.2 Inverse Time Decay

Inverse time decay is a method where the learning rate decreases over time following an inverse time schedule. The formula for the learning rate at each iteration can be expressed as [58, 69]:

$$\alpha_t = \frac{\alpha_0}{1 + d \cdot t}, \tag{5.2}$$

where $\alpha_t$ is the learning rate at iteration $t$, $\alpha_0$ is the initial learning rate, and $d$ is the decay rate parameter. As the iteration number increases, the denominator becomes larger, causing the learning rate to decrease. This decay schedule is particularly useful in scenarios where the model may converge quickly in the initial iterations but needs smaller learning rates for fine-tuning as training progresses.

Figure 5.2 illustrates the learning rate decay over iterations using the inverse time decay strategy for three different values of the decay rate parameter $d$. The curves correspond to decay rates of 0.9, 0.1, and 0.01, showcasing the diverse impact of $d$ on the learning rate. As the number of iterations increases, the inverse time decay function dynamically adjusts the learning rate, resulting in distinct decay patterns. A higher $d$ value leads to a rapid decrease in the learning rate, while a lower $d$ causes a slower decline.

In TensorFlow, the learning rate decay formula for the inverse time decay is indeed expressed as follows [126, 127]:

$$\alpha_t = \frac{\alpha_0}{1 + d \cdot \frac{t}{d_s}}, \tag{5.3}$$

where $d_s$ is the decay steps parameter (the number of iterations after which the learning rate needs to be updated). Note that the TensorFlow formula for inverse time decay is consistent with the initial formula when the decay rate $d$ is appropriately scaled by decay steps $d_s$.

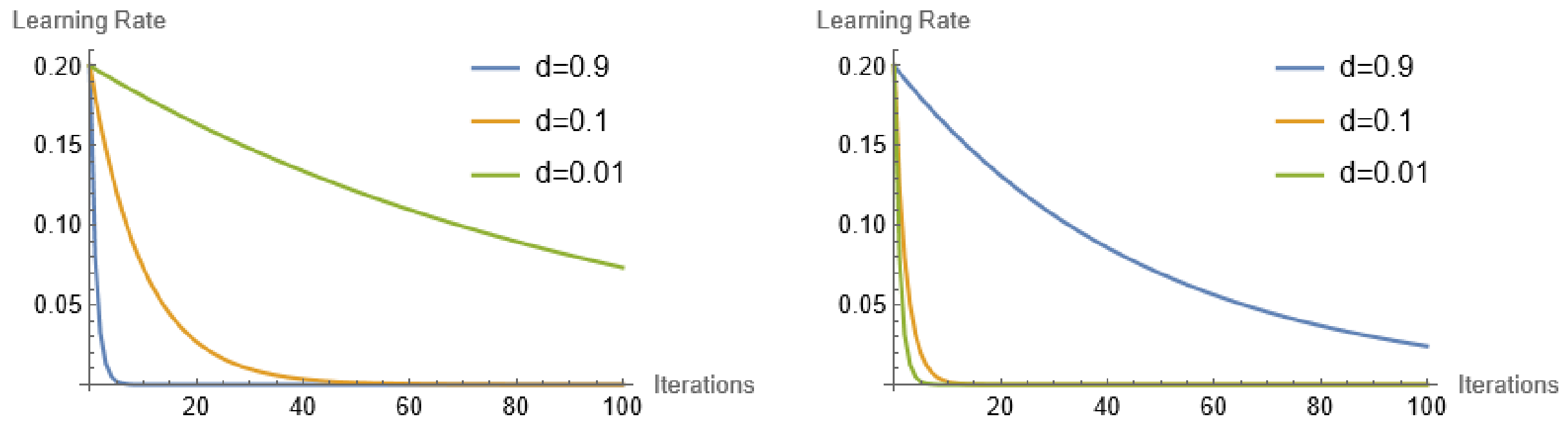

**Figure 5.3.** Left panel: Decrease in the learning rate for three values of $d = 0.01, 0.1, 0.9$ using exponential decay algorithm (standard formula). Right panel: Decrease in the learning rate for three values of $d = 0.01, 0.1, 0.9$ and $d_s = 5$ (TensorFlow formula).

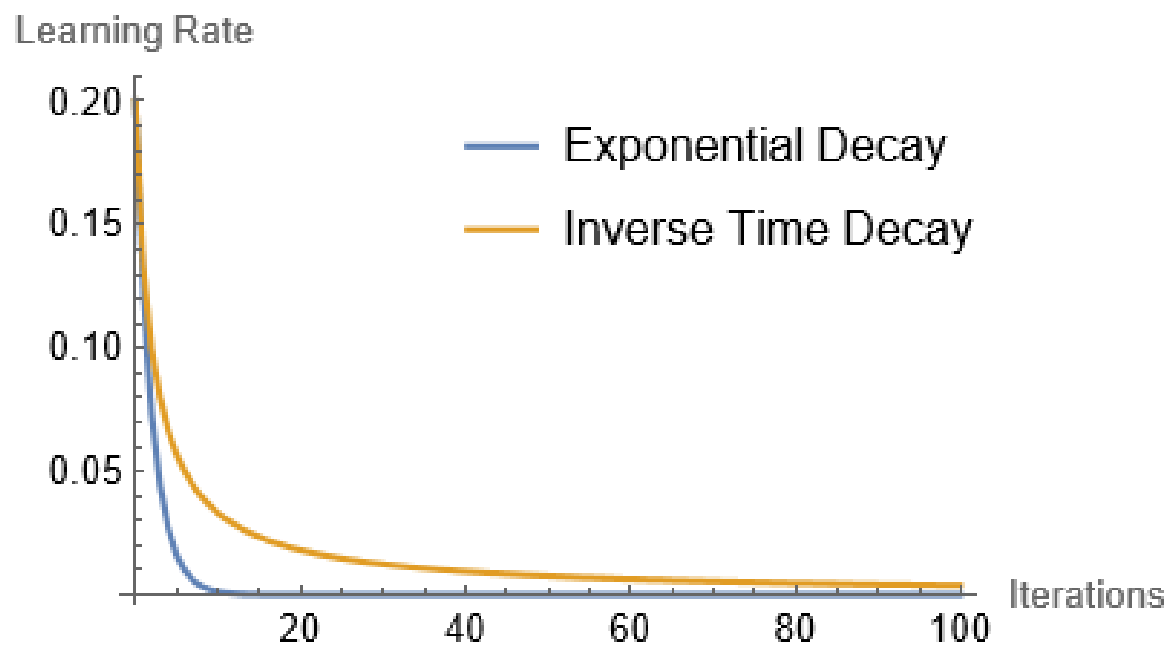

**Figure 5.4.** Comparative analysis of exponential decay and inverse time decay learning rate. Fixed values of 0.2 and 0.5 are employed for the initial learning rate and decay rate, respectively.

### 5.1.3 Exponential Decay

Another way of reducing the learning rate is according to the formula called exponential decay [69,126]:

$$\alpha_t = \alpha_0 e^{-d \cdot t}. \tag{5.4}$$

The learning rate decreases exponentially over time during the training. In TensorFlow, the formula for exponential decay is indeed expressed as follows [128]:

$$\alpha_t = \alpha_0 d^{-\frac{t}{d_s}}, \tag{5.5}$$

where the notations carry the same meaning as before. This formulation is different as it introduces the decay rate directly into an exponentiation term. Figure 5.3 (left) illustrates the impact of different decay rates, $d$, on the learning rate schedule using the standard exponential decay function. The curves represent learning rate decay for three distinct values of $d$: 0.9, 0.1, and 0.01. As the number of iterations progresses, the learning rate decreases exponentially, showcasing the sensitivity of the decay rate parameter in shaping the learning rate behavior. Utilizing the TensorFlow formula, Figure 5.3 (right) showcases the learning rate decay for three different values of the decay rate parameter, $d$, with a fixed number of decay steps, $d_s = 5$.

Figure 5.4 investigates the behavior of learning rates generated by inverse time decay and exponential decay. Fixed values of 0.2 and 0.5 are employed for the initial learning rate and decay rate, respectively. The curves showcase the distinct decay patterns inherent in each method over 100 iterations. Notably, the learning rates produced by inverse time decay tend to surpass those generated by exponential decay, particularly with increasing iterations. This divergence is attributed to the inherent characteristics of each decay function. Inverse time decay, characterized by the term $(1 + d \cdot t)$, results in a slower decrease in the learning rate (more gradually). Conversely, the exponential decay function, governed by the term $e^{-d \cdot t}$, experiences a more rapid decline as the iteration count grows. The exponential term approaches zero, causing the learning rate to decrease swiftly.

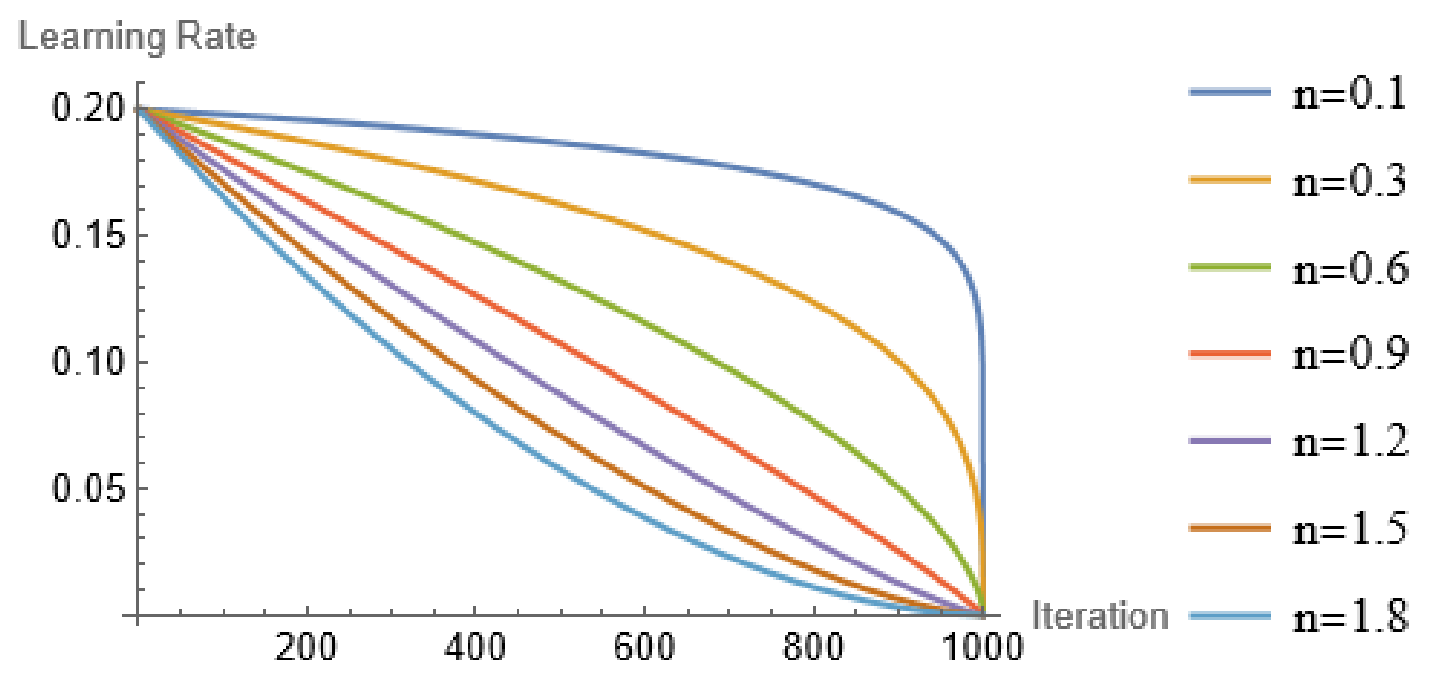

**Figure 5.5.** Polynomial decay learning rate curves.

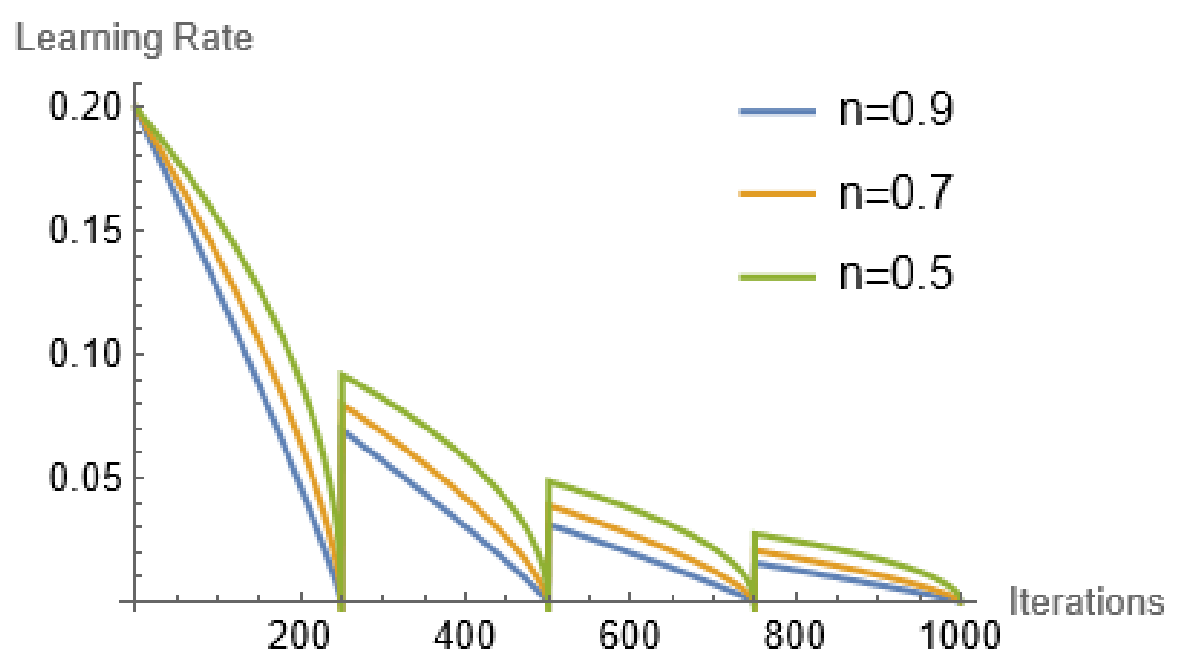

**Figure 5.6.** Polynomial learning rate with warm restarts for different power values.

### 5.1.4 Polynomial Learning Rate Policy with Warm Restart

The basic form of the polynomial learning rate policy is given by [130]

$$\alpha_t = \alpha_0 \left(1 - \frac{t}{T_t}\right)^n, \tag{5.6}$$

where $\alpha_0$ represents the initial learning rate, $t$ signifies the number of iterations, and $T_t$ denotes the total number of iterations. $T_t$ is calculated as the product of the total number of epochs and the number of iterations per epoch. The power term, $n$, within the formula, serves as a critical determinant, shaping the decay characteristics of the learning rate, as visually depicted in Figure 5.5. The figure illustrates the behavior of learning rate curves generated by the polynomial decay function for varying values of the power parameter $n$. The initial learning rate is set to 0.2, and the total number of iterations is fixed at 1000. As $n$ increases, the learning rate decay becomes more pronounced, showcasing the influence of the power term on the decay pattern. Each curve represents a distinct power value, ranging from 0.1 to 1.8, providing insights into the impact of the power parameter on the adaptability and convergence characteristics of the learning rate during the training process.

The polynomial decay formula elucidates the interplay between the initial learning rate, iteration count, total iterations, and the power term. The choice of a power value such as 0.9, as endorsed by previous works [131], suggests a deliberate selection to achieve a particular decay pattern conducive to effective model training.

Figure 5.6 illustrates the dynamic learning rate strategy utilizing polynomial decay with warm restarts. The learning rate is systematically adjusted during training, incorporating four distinct warm restarts at specified fractions of the total training iterations. The initial learning rate, $\alpha_0 = 0.2$, is progressively reduced by 35% at each restart, promoting adaptability and convergence. The plot showcases three learning rate curves for distinct power values: ($n = 0.9$, $n = 0.7$, and $n = 0.5$) over a total of 1000 iterations. The legend provides clarity on the corresponding power values, elucidating the influence of the power term on the learning rate dynamics. The warm restarts occur at 25%, 50%, and 75% of the total iterations, contributing to a dynamic and effective learning rate schedule. These strategies aim to enhance training performance and model convergence.

### 5.1.5 Cyclical Learning Rates (CLR) (Triangular Learning Rate Policy)

The essence of the CLR policy, particularly the triangular learning rate policy, lies in the observation that varying the learning rate within a certain range can have both short-term and long-term benefits during the training process. Increasing the learning rate may have a short-term negative effect, as it might cause the optimization process to oscillate or diverge. However, despite the short-term challenges, a higher learning rate can contribute to escaping local minima and exploring different regions of the loss landscape, leading to longer-term beneficial effects. Instead of adopting a fixed or exponentially decreasing learning rate, the idea is to let the learning rate vary cyclically within a range of values. That is, one sets minimum and maximum boundaries, and the learning rate cyclically varies between these bounds. This cyclic variation aims to balance the trade-off between the exploration of diverse regions and the exploitation of promising areas in the loss landscape. The learning rate is varied using a triangular window, specifically a linearly increasing and then a linearly decreasing function. This choice is motivated by its simplicity and effectiveness in incorporating the idea of cycling the learning rate within a range. Other functional forms, such as parabolic or sinusoidal windows, were experimented with and produced equivalent results, but the triangular window was adopted due to its simplicity which is illustrated in Figure 5.7. Despite the different functional forms, the results were found to be equivalent, suggesting that the key factor is the cyclical variation within a range, rather than the specific shape of the window.

The cyclical nature of CLR, especially the triangular learning rate policy, provides a mechanism for models to navigate plateau regions more effectively. Traditional learning rate schedules may struggle to navigate plateau regions in the loss landscape, where the gradient is small, leading to slow convergence or getting stuck. Other mechanisms don't give assurity that the model will not get stuck in a plateau region because we might not have that value of learning rate which can take it away from that region. CLR addresses this issue by cyclically varying the learning rate. The cycling between minimum and maximum values allows the model to escape from plateau regions, as the learning rate is increased during part of the cycle. The cyclic variation ensures that, if the model gets stuck, the increasing learning rate during the cycle will help it move away from the plateau. By facilitating both exploration and exploitation, CLR often leads to quicker convergence during the training process. This adaptability makes CLR particularly useful in scenarios where traditional learning rate schedules may struggle to find an appropriate learning rate for efficient optimization. An intuitive understanding of why CLR methods work comes from considering the loss function topology. Dauphin et al. [132] argue that the difficulty in minimizing the loss arises from saddle points rather than poor local minima. Saddle points have small gradients that slow the learning process. However, increasing the learning rate allows more rapid traversal of saddle point plateaus.

In the CLR mechanism, we vary the learning rate between two boundary values i.e. upper bound and lower bound. The upper bound and lower bound values are calculated by the learning rate range test. In the learning rate range test, we plot a curve between loss and learning rate. The range of the learning rate where we get a smooth decline in the curve is chosen as the optimal range for varying the learning rate. The learning rate is varied in the optimal range using the following equations [133]:

$$\alpha_t = \alpha_{\min} + (\alpha_{\max} - \alpha_{\min}) \times \max(0, (1 - x)), \tag{5.7}$$

where

$$x = \left| \frac{t}{s} - 2c + 1 \right|, \tag{5.8}$$

and $c$ can be calculated as,

$$c = \left\lfloor 1 + \frac{t}{2s} \right\rfloor, \tag{5.9}$$

where $\alpha_{\min}$ is the specified lower (i.e., base) learning rate, $t$ is the number of iterations of training, and $\alpha_t$ is the computed learning rate. $s$ is half the period or cycle length and $\alpha_{\max}$ is the maximum learning rate boundary. These equations vary the learning rate linearly between the minimum, $\alpha_{\min}$, and the maximum $\alpha_{\max}$.

Figure 5.7 illustrates a CLR schedule implemented with the triangular policy. The learning rate undergoes cyclic variations between a base learning rate $\alpha_{\min}$ of 0.001 and a maximum learning rate $\alpha_{\max}$ of 0.01. The half-period or cycle length is set to 5 iterations. The plot showcases the evolution of the learning rate across 100 iterations.

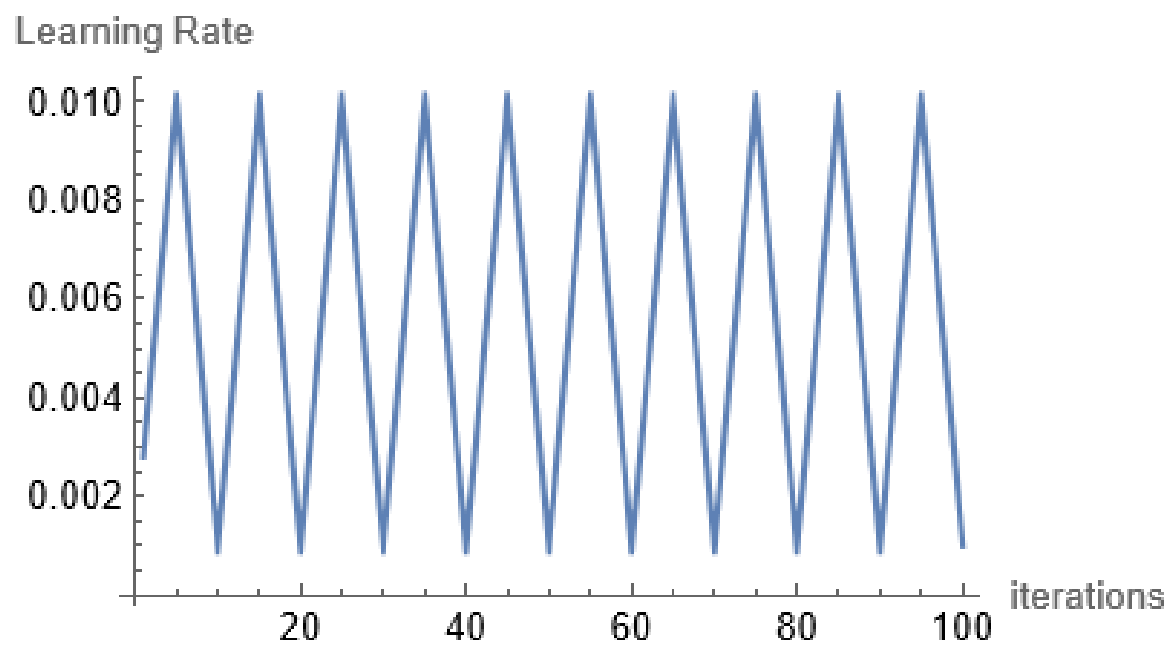

**Figure 5.7.** Cyclical learning rate schedule using triangular policy.

### 5.1.6 Stochastic Gradient Descent with Warm Restarts (SGDR) (Cosine Annealing)

Loshchilov and Hutter [134] introduced SGDRs as a learning rate schedule for training DNNs. The schedule is designed to allow the learning rate to oscillate between a minimum value $\alpha_{min}^{i}$ and a maximum value $\alpha_{max}^{i}$ following a cosine curve, and it periodically restarts to the initial learning rate. This allows the optimization process to escape from local minima and explore different regions of the loss landscape. This warm restart is based on a parameter $T_i$, where $T_i$ is the period for the learning rate variation. The learning rate at the $t$-th iterations in SGDR is given by the following expression [134, 135]:

$$\alpha_t = \alpha_{min}^{i} + \frac{1}{2}\left(\alpha_{max}^{i} - \alpha_{min}^{i}\right)\left(1 + \cos\left(\frac{T_{cur}}{T_i}\pi\right)\right), \tag{5.10}$$

where $\alpha_{min}^{i}$ and $\alpha_{max}^{i}$ are the lower and upper bounds respectively for the learning rate. $T_{cur}$ represents how many iterations have been performed since the last restart and a warm restart is simulated once $T_i$ iterations are performed. The period $T_i$ is determined by $T_i = T_{mult} \times T_{i-1}$, where $T_{mult}$ is a user-defined factor that controls the lengthening of the period after each restart.

The cosine function $\cos\left(\frac{T_{cur}}{T_i}\pi\right)$ oscillates between $-1$ and 1 as $T_{cur}$ varies from 0 to $T_i$. The factor $\frac{T_{cur}}{T_i}$ is scaled by $\pi$ to ensure it completes a full oscillation from $-1$ to 1 over the interval $[0, T_i]$. The entire cosine function is scaled by $\frac{1}{2}\left(\alpha_{max}^{i} - \alpha_{min}^{i}\right)$ and then shifted by $\alpha_{min}^{i}$. This scaling and shifting ensure that the result of the cosine function smoothly varies between $\alpha_{max}^{i}$ and $\alpha_{min}^{i}$. At $T_{cur} = 0$, the cosine function outputs 1, so $\alpha_t = \alpha_{max}^{i}$ . At $T_{cur} = T_i$, the cosine function outputs $-1$, so $\alpha_t = \alpha_{min}^{i}$. The cosine annealing results in a smooth transition of the learning rate from the maximum to the minimum value over $T_i$ iterations.

If $T_{mult} = 1$, the formula for updating the period $T_i$ after each warm restart becomes: $T_i = T_{i-1}$. In other words, if $T_{mult} = 1$, the period $T_i$ will remain the same after each warm restart. This means that the learning rate schedule will not lengthen, and the same period $T_i$ will be used throughout the training process. The period $T_i$ determines how many iterations the learning rate oscillates between $\alpha_{min}^{i}$ and $\alpha_{max}^{i}$ before a warm restart occurs. If $T_{mult}$ is set to 1, the period remains constant, and the learning rate schedule will exhibit the same cyclical behavior in each cycle without any lengthening of the period.

Adjusting both $\alpha_{max}^{i}$ and $\alpha_{min}^{i}$ at every new restart could be an interesting avenue for exploration, and it might offer additional flexibility during the training process. However, for the sake of simplicity, you can choose to keep them constant across restarts. By keeping these values constant, the implementation of SGDR becomes more straightforward and reduces the number of hyperparameters that need to be tuned. This choice simplifies the training process and makes it more accessible, especially when fine-tuning and experimentation with different configurations are involved. In this case, SGDR is given by the following expression:

$$\alpha_t = \alpha_{min} + \frac{1}{2}\left(\alpha_{max} - \alpha_{min}\right)\left(1 + \cos\left(\frac{T_{cur}}{T_i}\pi\right)\right). \tag{5.11}$$

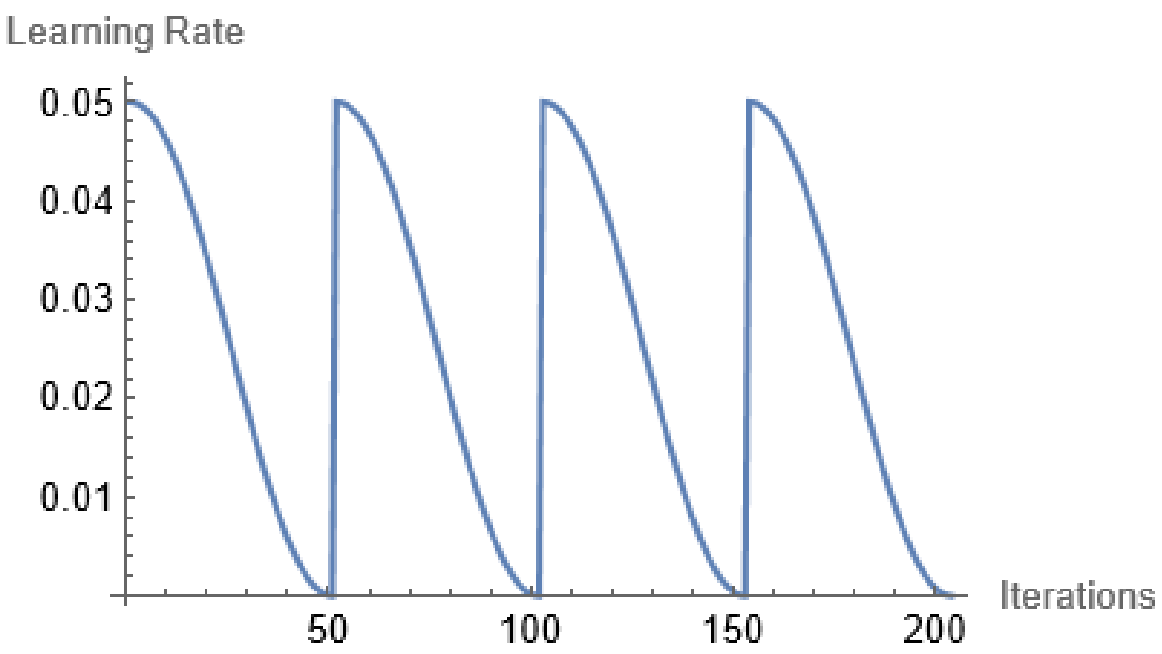

**Figure 5.8.** Cosine Annealing Learning Rate Schedule.

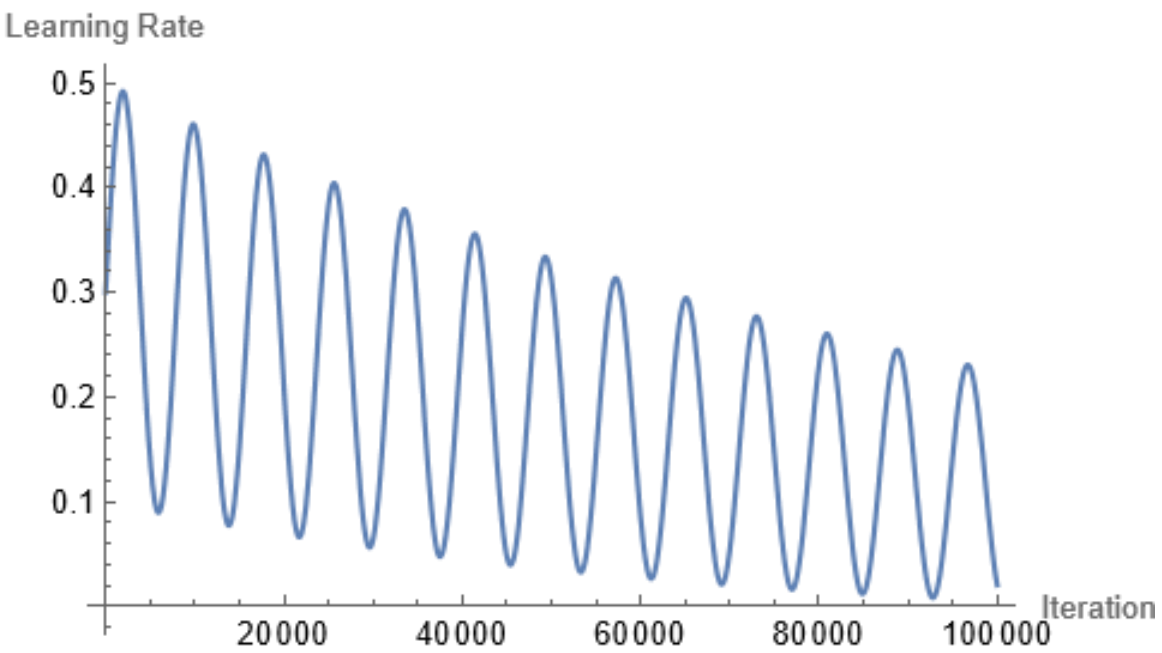

**Figure 5.9.** Exponential decay with sine wave oscillation learning rate schedule.

SGDR's empirical success on various tasks and datasets suggests that the combination of cosine annealing and warm restarts contributes to better convergence and generalization. Figure 5.8 shows an instance of this learning rate schedule. The figure illustrates the learning rate schedule using the cosine annealing policy. The learning rate smoothly varies between a minimum learning rate ($\alpha_{min} = 0$) and a maximum learning rate ($\alpha_{max} = 0.05$) over multiple cycles. Each cycle consists of $50$ iterations, and a total of $4$ cycles are depicted.

### 5.1.7 Exponential Decay Sine Wave Learning Rate

The exponential decay sine wave learning rate is given by the following expression [136]:

$$\alpha_t = \alpha_0 e^{-\frac{\rho t}{T}} \left(\sin\left(\tau \frac{t}{2\pi b}\right) + e^{-\frac{\rho t}{T}} + 0.5\right), \tag{5.12}$$

where $t$ represents the iteration number and $\alpha_0$ denotes base learning rate. $T$ represents total iterations and $b$ is the number of batches per epoch. $\rho$ and $\tau$ control the decay and oscillation nature of the learning rate, respectively. The exponential decay term causes the learning rate to decrease over time, and the sine wave term introduces periodic oscillations. The frequency and amplitude of oscillations are controlled by $\tau$ and $\frac{1}{2\pi b}$, respectively.

Figure 5.9 showcases a dynamic learning rate schedule utilizing an exponential decay function modulated by a sine wave oscillation. The initial learning rate $\alpha_0$ is set to 0.2, and the decay rate $\rho$ is 0.6. The learning rate undergoes an oscillatory pattern defined by a sine wave with a frequency parameter $\tau$ and a total of 10 batches per epoch, $b$. The total number of iterations is set to $100{,}000$. The plot demonstrates the intricate interplay between exponential decay and sinusoidal oscillation, resulting in a fluctuating learning rate.

### 5.1.8 Hessian-Aware Learning Rate Adjustment

In comparison to small-batch training, large-batch training involves tuning the initial learning rate to relatively larger values. The effectiveness of this strategy has been thoroughly examined through both theoretical analyses and empirical studies. The large learning rate induces a higher level of noise during training (high-noise regime), acting as a regularizer and enhancing the generalization performance of the model. Contrary to our initial expectations, this regularization effect diminishes when the learning rate is decayed prematurely, i.e., before the model is sufficiently trained under high-noise conditions. This unexpected observation prompts a deeper investigation into the dynamics of learning rate decay in the context of large-batch training.

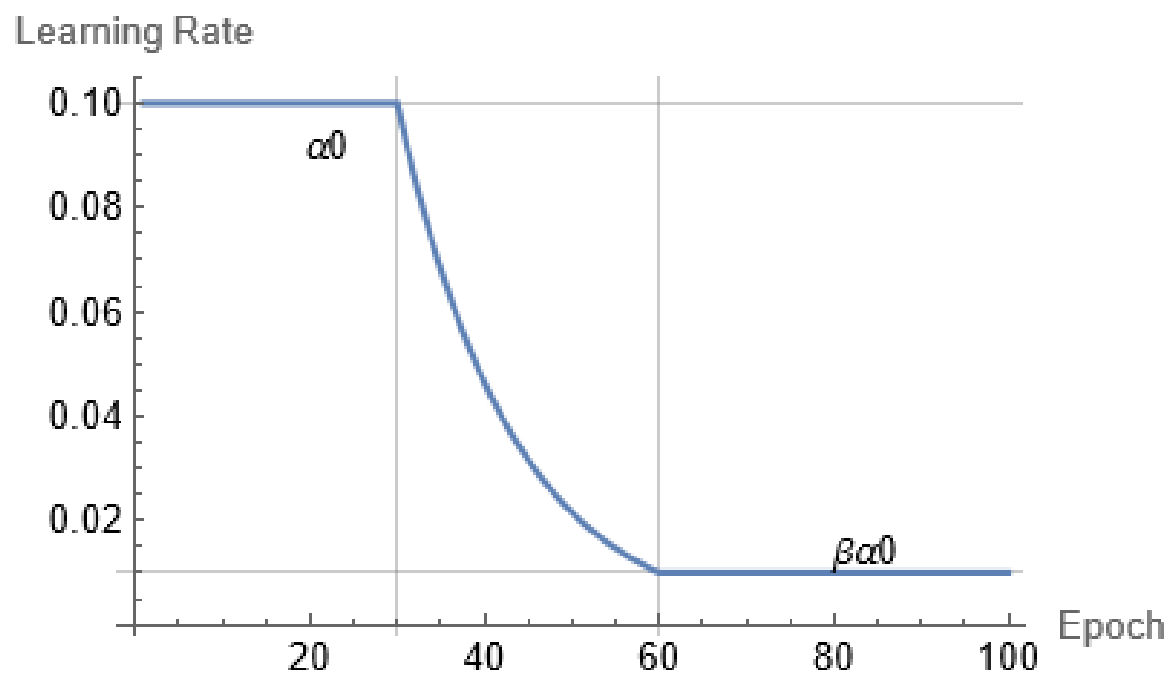

**Figure 5.10.** Hessian-aware learning rate schedule.

After achieving sufficient training under high-noise conditions, the objective is to enforce convergence by appropriately reducing the noise scale. Commonly, learning rate decay is employed for this purpose, with the epoch budget often partitioned using static schedules. Static decay schedules, such as those reducing the learning rate by a fixed factor at predefined percentages of the epoch budget, may not be optimal for achieving convergence under reduced noise conditions. Such decay schedules do not consider the impact of the reduced noise scale on the generalization performance. By dynamically adjusting the decay schedule based on the evolving noise scale, we can achieve enhanced convergence and improved generalization performance.

On the other hand, the minimum learning rate is identified as a pivotal parameter that requires careful tuning to achieve optimal generalization performance.

In [137], instead of continuously reducing the learning rate throughout the entire training process, the authors propose a different perspective. The learning rate decay is considered a one-time transition of the noise scale within the given epoch budget. This implies that the learning rate adjustment is not a continuous process but occurs only once during training. The learning rate is gradually decayed between two specific epochs, namely epoch $n_0$ and epoch $n_1$. This suggests that the change in the learning rate is applied smoothly over this interval. After the gradual decay phase, the model is trained using a specific learning rate denoted as $\beta\alpha_0$ for the remaining epochs $T - n_1$. This indicates that the learning rate is kept constant (with a different scale represented by $\beta$) for the latter part of the training process.

The learning rate decays from $\alpha_0$ to $\beta\alpha_0$ is governed by the equation:

$$\alpha_{t+1} = \alpha_t \times \beta^{1/(n_1-n_0)}. \tag{5.13}$$

As $n_1$ is determined, the above equation automatically determines the decay factor based on the given $n_0$ and $\beta$. This equation defines how the learning rate decreases over the specified epoch range $[n_0, n_1]$. The $n_1$ can be adjusted between $n_0$ and $T$. If $n_1 = n_0$, the learning rate is immediately reduced to $\beta\alpha_0$. This setting dramatically reduces the noise scale but may lead to slow convergence and insufficient minimization of the training loss within the given epoch budget. On the other hand, if $n_1 = T$, the learning rate is gradually reduced in all the remaining $T - n_0$ epochs. This is similar to polynomial decay starting at epoch $n_0$ and is generally more efficient in minimizing training loss. However, the solution may become sharper as the model is trained for more epochs using a learning rate. Therefore, $n_1$ should be carefully tuned to make a good trade-off between the statistical efficiency of the loss minimization and the generalization performance.

Figure 5.10 illustrates the Hessian-aware learning rate schedule. The initial learning rate $\alpha_0$ is set to 0.1, and the decay factor $\beta$ is set to 0.1. The decay starts at epoch 30 ($n_0 = 30$) and continues until $n_1 = 60$, after which the learning rate remains constant. The plot demonstrates the gradual decrease in the learning rate during the decay phase, highlighting the impact of the decay factor on the rate of decrease. The gridlines indicate the transition points at epochs $n_0$ and $n_1$, emphasizing the initial learning rate $\alpha_0$ and the final decayed rate $\beta\alpha_0$.

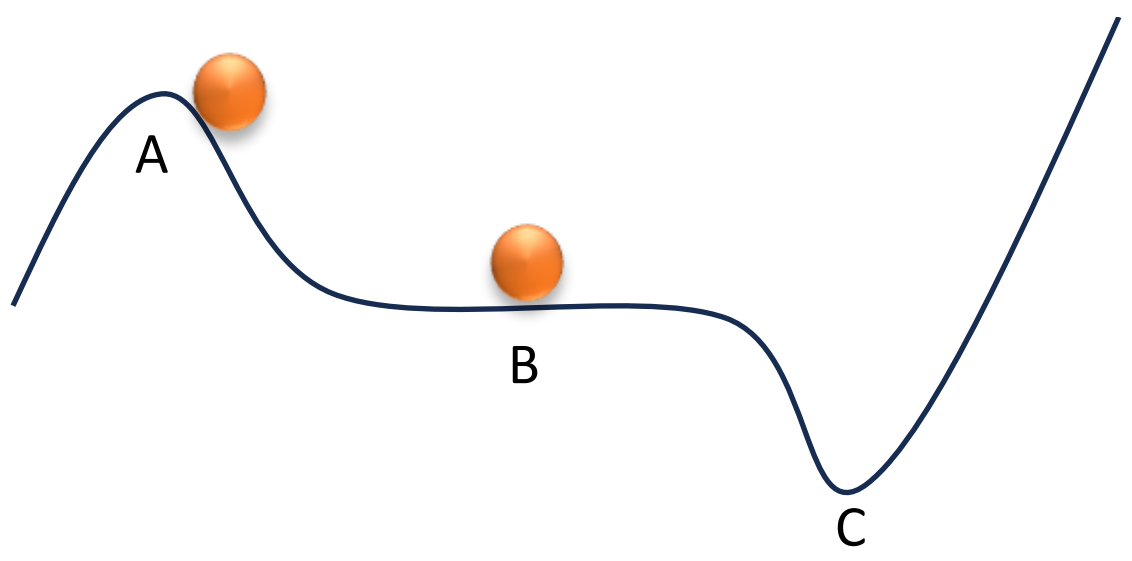

**Figure 5.11.** With gradient descent, the Loss function decreases rapidly along the slope AB as the gradient along this slope is high. But as soon as it reaches point B the gradient becomes very low. The weight updates around B is very small. Even after many iterations, the cost moves very slowly before getting stuck at a point where the gradient eventually becomes zero. Now, imagine you have a ball rolling from point A. The ball starts rolling down slowly and gathers some momentum across the slope AB. When the ball reaches point B, it has accumulated enough momentum to push itself across the plateau region B and finally following slope BC to land at the global minima C.

## 5.2 Accelerated Gradient Descent

### 5.2.1. Gradient Descent with Momentum

The gradient descent with momentum, proposed by Boris Polyak in 1964 [138], often referred to as the "Heavy Ball Method," is an optimization algorithm used for minimizing functions. The name "Heavy Ball Method" comes from the analogy of a heavy ball rolling on a surface. The momentum term simulates the inertia of the ball, helping it roll past small bumps and accelerate in the direction of the steepest descent. It enhances traditional GD by adding a momentum term, which helps accelerate convergence, especially in the presence of high curvature, small gradients, or noisy gradients.

- In regions where the function being optimized has high curvature, the gradient (which indicates the direction of steepest ascent) can rapidly change direction. Traditional GD algorithms rely on following the gradient direction to iteratively approach the optimum. However, in such regions, the rapid changes in gradient direction can cause oscillations or slow convergence because the algorithm may overcorrect or take small steps due to the instability caused by these rapid changes.
- If there are small flat regions (where the gradient is zero) surrounded by downhill segments, traditional GD algorithms may get stuck in these regions since they cannot detect the direction of descent. This can lead to stagnation in optimization progress, as the algorithm fails to escape these flat regions, see Figure 5.11.
- When gradients are small, traditional GD updates may be too small to make meaningful progress towards the optimal solution. This can result in slow convergence or premature convergence to suboptimal solutions, especially if the algorithm is unable to effectively navigate these regions with small gradients.
- In scenarios where gradients are noisy or contain errors, traditional GD updates can be erratic, leading to instability and hindering convergence. Traditional GD algorithms may struggle to differentiate between meaningful gradient signals and noise, resulting in suboptimal optimization performance.

Imagine you're trying to navigate a hilly terrain to reach the lowest point (representing the minimum of a function). You're blindfolded and can only feel the slope beneath your feet (representing the gradient of the function at your current position). Now, let's compare traditional GD with gradient descent with momentum.

Traditional GD: In this scenario, you take a step in the direction of the steepest slope you can feel. Without any additional momentum, you rely solely on the current information (the slope at your current position) to decide your next step. If the slope suddenly changes or there are small bumps along the way, you adjust your direction accordingly based solely on the local information (the current slope). You might find yourself zigzagging or getting stuck in local valleys (local minima) or on plateaus (saddle points).

Gradient Descent with Momentum: Now, let's add momentum to the mix. Instead of just feeling the slope at your current position, you remember the direction and speed of your previous steps. As you take each step, you factor in the direction and speed of your previous movements. This helps smooth out your path and maintain a more consistent direction of descent. Imagine you're rolling a heavy ball downhill. The ball carries momentum from its previous motion, so it doesn't immediately respond to small bumps or fluctuations in the terrain. As you move, you keep a memory of your past motions, allowing you to smooth out your path and maintain a consistent direction of movement. If the terrain has irregularities or sudden changes in slope, the momentum helps you push through them, maintaining your overall progress towards the lowest point.

In this analogy:

- The terrain: Represents the landscape of the optimization function, with valleys, hills, and plateaus corresponding to local minima, maxima, and saddle points.
- Your steps: Correspond to the parameter updates in the optimization algorithm.
- The momentum: Represents the inertia or memory from past updates. It helps smooth out your path and maintain momentum towards the optimal solution, akin to rolling a heavy ball downhill.

The addition of momentum to GD is indeed one of the most convenient and powerful ideas in deep learning and optimization. The basic idea behind momentum-based descent is to introduce a momentum term that accelerates the GD algorithm in the relevant direction and dampens oscillations in irrelevant directions. This momentum term is a moving average of past gradients and helps the optimization algorithm build up speed in directions where the gradient consistently points, while also smoothing out the oscillations caused by noise or curvature in the loss landscape.

In gradient descent with momentum, the update rule is given by two equations, the velocity update equation and parameter update equation.

**Velocity Update Equation:**

Remember, the key idea behind GD is to iteratively update the parameters $\mathbf{w}$ by moving them in the direction of the steepest descent of the cost function. This direction is given by the negative of the gradient of the cost function with respect to the parameters, denoted as $\nabla_{\mathbf{w}}\mathcal{L}(\mathbf{w})$. To update the velocity vector $\mathbf{v}$ using the gradient $\nabla_{\mathbf{w}}\mathcal{L}(\mathbf{w})$, we employ the following formula [69]:

$$\mathbf{v} = -\alpha\nabla_{\mathbf{w}}\mathcal{L}(\mathbf{w}). \tag{5.14.1}$$

Subsequently, the parameter vector $\mathbf{w}$ is updated using the velocity vector $\mathbf{v}$ as follows:

$$\mathbf{w} = \mathbf{w} + \mathbf{v}. \tag{5.14.2}$$

Here, $\alpha$ is the learning rate. We are using the matrix calculus convention where the derivative of a scalar $\mathcal{L}(\mathbf{w})$ with respect to a column vector $\mathbf{w}$ is denoted as the column vector $\nabla_{\mathbf{w}}\mathcal{L}(\mathbf{w})$:

$$\nabla_{\mathbf{w}}\mathcal{L}(\mathbf{w}) = \frac{\partial}{\partial \mathbf{w}}\mathcal{L}(\mathbf{w}) = \begin{pmatrix} \frac{\partial \mathcal{L}(\mathbf{w})}{\partial w_1} \\ \vdots \\ \frac{\partial \mathcal{L}(\mathbf{w})}{\partial w_d} \end{pmatrix}. \tag{5.14.3}$$

In momentum-based descent, the update rule for the velocity vector $\mathbf{v}$ is given

$$\mathbf{v} = \beta\mathbf{v} - \alpha\nabla_{\mathbf{w}}\mathcal{L}(\mathbf{w}). \tag{5.15.1}$$

Or, at iteration $t$, the velocity vector is updated using the equation:

$$\mathbf{v}_{t+1} = \beta\mathbf{v}_t - \alpha\nabla_{\mathbf{w}_t}\mathcal{L}(\mathbf{w}_t). \tag{5.15.2}$$

Here: $\beta$ is the momentum coefficient, typically a value between 0 and 1. $\mathbf{v}$ represents the velocity vector, which is updated at each iteration. $\alpha$ is the learning rate, determining the step size of each iteration. $\nabla_{\mathbf{w}}\mathcal{L}(\mathbf{w})$ is the gradient of the objective function $\mathcal{L}$ with respect to the parameters $\mathbf{w}$.

Let's start with $t = 0$, where $\mathbf{v}_0 = 0$, and then proceed with the iterative updates of the velocity vector to illustrate how the equation represents an exponential moving average, EMA, of the past gradients. The velocity vector is updated using the equation:

$$\begin{aligned}\mathbf{v}_1 &= \beta\mathbf{v}_0 - \alpha\nabla_{\mathbf{w}_0}\mathcal{L}(\mathbf{w}_0)\\ &= -\alpha\nabla_{\mathbf{w}_0}\mathcal{L}(\mathbf{w}_0).\end{aligned} \tag{5.16.1}$$

At $t = 1$, the velocity vector is updated using the equation:

$$\begin{aligned}\mathbf{v}_2 &= \beta\mathbf{v}_1 - \alpha\nabla_{\mathbf{w}_1}\mathcal{L}(\mathbf{w}_1)\\ &= -\alpha\beta\nabla_{\mathbf{w}_0}\mathcal{L}(\mathbf{w}_0) - \alpha\nabla_{\mathbf{w}_1}\mathcal{L}(\mathbf{w}_1)\\ &= -\alpha\left(\beta\nabla_{\mathbf{w}_0}\mathcal{L}(\mathbf{w}_0) + \nabla_{\mathbf{w}_1}\mathcal{L}(\mathbf{w}_1)\right).\end{aligned} \tag{5.16.2}$$

At $t = 2$, the velocity vector is updated using the equation:

$$\begin{aligned}\mathbf{v}_3 &= \beta\mathbf{v}_2 - \alpha\nabla_{\mathbf{w}_2}\mathcal{L}(\mathbf{w}_2)\\ &= -\alpha\beta\left(\beta\nabla_{\mathbf{w}_0}\mathcal{L}(\mathbf{w}_0) + \nabla_{\mathbf{w}_1}\mathcal{L}(\mathbf{w}_1)\right) - \alpha\nabla_{\mathbf{w}_2}\mathcal{L}(\mathbf{w}_2)\\ &= -\alpha\left(\beta^2\nabla_{\mathbf{w}_0}\mathcal{L}(\mathbf{w}_0) + \beta\nabla_{\mathbf{w}_1}\mathcal{L}(\mathbf{w}_1) + \nabla_{\mathbf{w}_2}\mathcal{L}(\mathbf{w}_2)\right).\end{aligned} \tag{5.16.3}$$

For general iteration $t$, the update equation becomes:

$$\mathbf{v}_t = -\alpha\left(\beta^{t-1}\nabla_{\mathbf{w}_0}\mathcal{L}(\mathbf{w}_0) + \beta^{t-2}\nabla_{\mathbf{w}_1}\mathcal{L}(\mathbf{w}_1) + \cdots + \nabla_{\mathbf{w}_{t-1}}\mathcal{L}(\mathbf{w}_{t-1})\right). \tag{5.16.4}$$

In each iteration, the velocity vector $\mathbf{v}_t$ is influenced by the current gradient $\boldsymbol{g}_t = \nabla_{\mathbf{w}_{t-1}}\mathcal{L}(\mathbf{w}_{t-1})$ as well as an accumulation of past gradients, with each past gradient being weighted by $\beta$ to a progressively lesser extent. In other words, we are taking a weighted sum of the gradients from past iterations. Since $\beta < 1$, this means weights decrease with age (long past iterations have less influence).

The term $\beta\mathbf{v}$, (5.15.1), represents the exponentially decaying moving average of past gradients. The momentum coefficient $\beta$ controls the decay rate, where a smaller $\beta$ places more emphasis on recent gradients, while a larger $\beta$ includes more past gradients in the average. Each past gradient contributes to the moving average with decreasing weight over time, with more recent gradients having a higher influence than older ones. This exponentially decaying nature ensures that recent information is prioritized while still retaining some memory of past gradients. If $\beta = 0.1$ with $t = 4$, the gradient at $t = 3$ will contribute $100\%$ of its value, the gradient at $t = 2$ will contribute $10\%$ of its value, gradient at $t = 1$ will only contribute $1\%$ of its value and gradient at $t = 0$ will only contribute $0.1\%$ of its value. If $\beta = 0.9$ with $t = 4$, the gradient at $t = 3$ will contribute $100\%$ of its value, the gradient at $t = 2$ will contribute $90\%$ of its value, gradient at $t = 1$ will only contribute $81\%$ of its value and gradient at $t = 0$ will only contribute $0.729\%$ of its value.

The accumulated moving average of past gradients influences the direction of movement in subsequent iterations. The algorithm continues to move in the direction of this moving average, in addition to the influence of the current gradient $\nabla_{\mathbf{w}}\mathcal{L}(\mathbf{w})$. This ensures that the algorithm retains some memory of the past gradients and continues to move in their direction, helping to smooth out the optimization path and maintain momentum towards the optimal solution.

The momentum term ($\beta\mathbf{v}$) represents the inertia or memory of your previous steps, while the gradient term ($\nabla_{\mathbf{w}}\mathcal{L}(\mathbf{w})$) indicates the steepness of the slope at your current position. In physics, inertia is the tendency of an object to resist changes in its velocity. Similarly, in optimization, the momentum term introduces inertia into the optimization process. When you walk down a hill with momentum, your momentum from previous steps keeps you moving in the same direction, resisting sudden changes in your speed or direction. This inertia helps to smooth out your descent and maintain a more consistent velocity. The momentum term can also be likened to a memory of your previous steps. As you walk down the hill, your momentum carries information about the direction and speed of your past movements. This memory helps you to adapt your current movement based on your recent experience. For example, if you encountered a slight incline a few steps ago, your momentum would reflect that, influencing your current movement to continue in the same general direction. The momentum term combines both inertia and memory aspects. It resists sudden changes in velocity, akin to inertia, while also incorporating information about past movements, akin to memory. This combination allows the optimization process to navigate through the optimization landscape more efficiently by smoothing out fluctuations and maintaining a more stable direction of movement.

Indeed, the equation, (5.15.1), combines the momentum from previous steps ($\beta\mathbf{v}$) with the influence of the current slope ($\alpha\nabla_{\mathbf{w}}\mathcal{L}(\mathbf{w})$). The momentum term keeps you moving in the direction of your previous movement, while the gradient term adjusts your velocity based on the steepness of the current slope. This combined velocity determines how you adjust your speed and direction as you continue walking down the hill.

**Parameter Update Equation:**

$$\mathbf{w} = \mathbf{w} + \mathbf{v}, \tag{5.17.1}$$

or

$$\mathbf{w}_{t+1} = \mathbf{w}_t + \mathbf{v}_{t+1}. \tag{5.17.2}$$

Your position on the hill ($\mathbf{w}$) represents where you are in the optimization process, and your velocity ($\mathbf{v}$) indicates how you're moving down the hill. Updating your position ($\mathbf{w}$) involves adding your velocity ($\mathbf{v}$) to your current position. This adjustment determines your new position on the hill, taking into account both your momentum from previous steps and the current direction and speed of your descent.

**Remarks:**

- Note that

$$\mathbf{v}_{t+1} = \beta\mathbf{v}_t - \alpha\nabla_{\mathbf{w}_t}\mathcal{L}(\mathbf{w}_t), \tag{5.18.1}$$

$$\begin{aligned}\mathbf{w}_{t+1} &= \mathbf{w}_t + \mathbf{v}_{t+1}\\ &= \mathbf{w}_t - \alpha\nabla_{\mathbf{w}_t}\mathcal{L}(\mathbf{w}_t) + \beta\mathbf{v}_t.\end{aligned} \tag{5.18.2}$$

- Moreover, the above updates can be rewritten as [109]

$$\mathbf{v}_{t+1} = \mathbf{w}_{t+1} - \mathbf{w}_t \Rightarrow \beta\mathbf{v}_t = \beta(\mathbf{w}_t - \mathbf{w}_{t-1}), \tag{5.19.1}$$

and

$$\begin{aligned}\mathbf{w}_{t+1} &= \mathbf{w}_t - \alpha\nabla_{\mathbf{w}_t}\mathcal{L}(\mathbf{w}_t) + \beta\mathbf{v}_t\\ &= \mathbf{w}_t - \alpha\nabla_{\mathbf{w}_t}\mathcal{L}(\mathbf{w}_t) + \beta(\mathbf{w}_t - \mathbf{w}_{t-1}),\end{aligned} \tag{5.19.2}$$

where $\mathbf{v}_t = (\mathbf{w}_t - \mathbf{w}_{t-1})$ is called the momentum.
- Now, we have two coefficients to choose- the step size $\alpha$ and momentum coefficient $\beta$.
- The step size $\alpha$ determines the size of the updates made to the parameters in each iteration of the optimization algorithm. A larger $\alpha$ leads to larger updates, which can help accelerate convergence but may risk overshooting the optimal solution or causing instability. Conversely, a smaller $\alpha$ leads to smaller updates, which may slow down convergence but can provide more stability and robustness to noise in the gradients.
- The momentum coefficient $\beta$ controls the influence of the momentum term in the optimization process. A larger $\beta$ gives more weight to the momentum term, allowing the optimizer to retain more information from past updates and smooth out the optimization path. This can help accelerate convergence and overcome local minima. Conversely, a smaller $\beta$ reduces the influence of the momentum term, making the optimization process more similar to traditional GD. This may be preferable in some cases, especially if the optimization landscape is simple or if there is significant noise in the gradients.
- While the learning rate $\alpha$ is often adapted over time (e.g., through learning rate schedules or techniques like AdaGrad, RMSprop, or Adam), adapting $\beta$ over time is less common. Typically, $\beta$ starts with a small value and is later raised. This approach allows for a gradual increase in reliance on past gradients as optimization progresses, potentially improving convergence stability and robustness.
- Although adapting $\beta$ over time can be beneficial, it is generally considered less critical compared to adapting the learning rate $\alpha$. The learning rate has a direct impact on the step size of parameter updates and can significantly affect convergence speed and optimization performance. Properly tuning the learning rate

**Algorithm 5.1:** SGD with momentum

Initialize Parameters:

Initialize the parameters $\mathbf{w}$ (the vector you're optimizing) to some initial values. Set the learning rate $\alpha$, momentum coefficient $\beta$, and the initial velocity $\mathbf{v}_0$ to zero or small random values.

Repeat Until Convergence:

For each iteration $t = 1,2, \ldots$:

- Sample Data: Randomly select a mini-batch of data examples $(\boldsymbol{x}^{(i)}, \boldsymbol{y}^{(i)})$ from the training dataset.
- Compute Gradient: Compute the gradient of the objective function

$$\boldsymbol{g} = \nabla_{\mathbf{w}}\left(\frac{1}{m}\sum_{i=1}^{m}\mathcal{L}\left(\boldsymbol{x}^{(i)}, \boldsymbol{y}^{(i)}; \mathbf{w}\right)\right),$$

  where $\mathcal{L}\left(\boldsymbol{x}^{(i)}, \boldsymbol{y}^{(i)}; \mathbf{w}\right)$ is the loss function (e.g., MSE loss, cross-entropy loss) for the example $(\boldsymbol{x}^{(i)}, \boldsymbol{y}^{(i)})$
- Update Velocity: Update the velocity $\mathbf{v}$ using the previous velocity and the current gradient:

$$\mathbf{v} = \beta\mathbf{v} - \alpha\boldsymbol{g}.$$

- Update Parameters: Update the parameters $\mathbf{w}$ using the velocity $\mathbf{v}$:

$$\mathbf{w} = \mathbf{w} + \mathbf{v}.$$

- Repeat: Continue this process until convergence criteria are met (e.g., maximum number of iterations reached, small gradient norm, etc.).

schedule or employing adaptive learning rate methods is often prioritized over adaptation of $\beta$ for most optimization tasks.

- The choice of both $\alpha$ and $\beta$ is crucial for effective optimization. The optimal values of $\alpha$ and $\beta$ may depend on factors such as the characteristics of the optimization landscape, the amount of noise in the gradients, and the computational resources available. Experimentation and tuning are often required to find the best combination of coefficients for a particular optimization problem.
- The SGD algorithm with momentum is given in Algorithm 5.1.

### 5.2.2 Accelerated Gradient Descent with Nesterov Momentum (Nesterov Accelerated Gradient)

Intuitively, traditional momentum-based gradient descent methods can be likened to a ball rolling downhill towards a valley, where the objective is to reach the bottom (the minimum of the loss function). Momentum helps the ball gain speed as it descends, which is analogous to the algorithm making larger updates to parameters based on past gradients. However, one of the challenges with traditional momentum-based gradient descent is indeed the potential for overshooting the minima, especially when the algorithm gains momentum and the updates become increasingly large (the loss decreases during the initial phase of optimization and then, when we are close to the minima, an update actually overshoots the minima and increases the loss.). Picture a ball rolling down a hill with increasing velocity—it might overshoot the valley it's trying to settle in. Similarly, in optimization, this overshooting can lead to moving past the optimal point, causing the loss function to increase instead of decrease. This phenomenon can lead to oscillations around the optimal solution or even divergence from it. Imagine the ball rolling back and forth at the bottom of the valley, never quite settling into a stable position. This oscillatory behavior can prevent the algorithm from converging to the optimal solution. This is the phenomenon that Nesterov's accelerated gradient-based optimization tries to tackle.

Sutskever et al. (2013) [139] introduced a variant of the momentum algorithm that was inspired by Nesterov's accelerated gradient method (Nesterov, 1983, 2004) [140, 141], commonly referred to as Nesterov Accelerated Gradient (NAG). The update rules in this case are given by

$$\mathbf{v} = \beta\mathbf{v} - \alpha\frac{\partial\mathcal{L}(\mathbf{w} + \beta\mathbf{v})}{\partial\mathbf{w}}, \quad (5.20.1)$$

$$\mathbf{w} = \mathbf{w} + \mathbf{v}, \quad (5.20.2)$$

or

$$\mathbf{v}_{t+1} = \beta\mathbf{v}_t - \alpha\boldsymbol{\nabla}\mathcal{L}(\mathbf{w}_t + \beta\mathbf{v}_t), \quad (5.20.3)$$

$$\mathbf{w}_{t+1} = \mathbf{w}_t + \mathbf{v}_{t+1}. \quad (5.20.4)$$

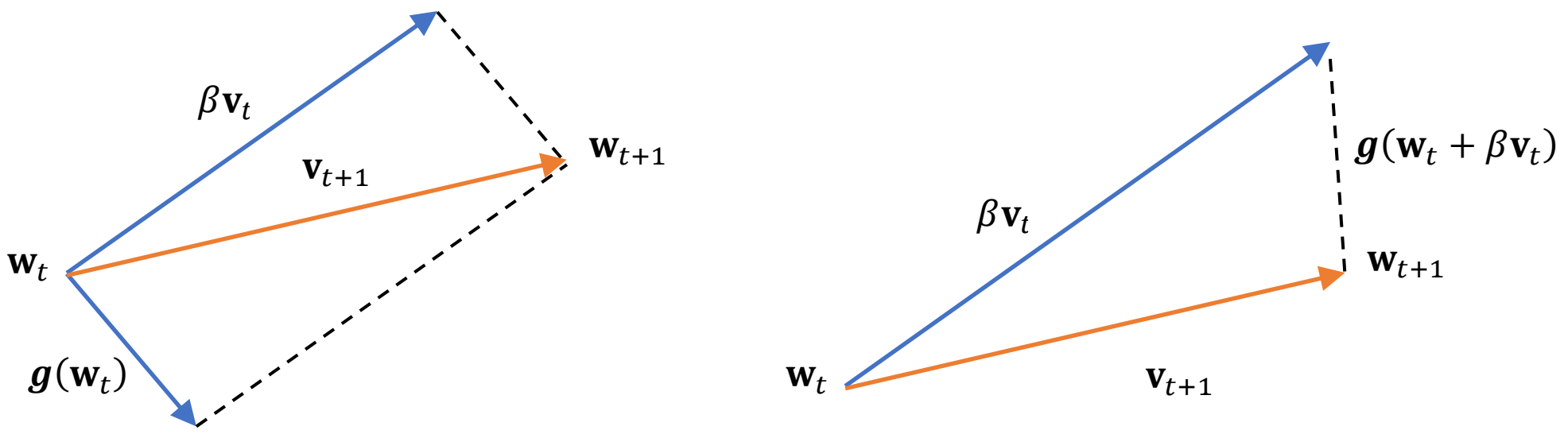

**Figure 5.12.** Left panel: Classical momentum. Right panel: Nesterov accelerated gradient.

Figure 5.12 illustrates the above idea. Moreover, the above updates can be rewritten as [33]

$$\mathbf{v}_{t+1} = \mathbf{w}_{t+1} - \mathbf{w}_t \Rightarrow \beta\mathbf{v}_t = \beta(\mathbf{w}_t - \mathbf{w}_{t-1}), \tag{5.20.5}$$

$$\begin{aligned}\mathbf{w}_{t+1} &= \mathbf{w}_t + \mathbf{v}_{t+1} \\ &= \mathbf{w}_t - \alpha\nabla\mathcal{L}(\mathbf{w}_t + \beta\mathbf{v}_t) + \beta\mathbf{v}_t \\ &= \mathbf{w}_t - \alpha\nabla\mathcal{L}\big(\mathbf{w}_t + \beta(\mathbf{w}_t - \mathbf{w}_{t-1})\big) + \beta(\mathbf{w}_t - \mathbf{w}_{t-1}).\end{aligned} \tag{5.20.6}$$

In more details, the update may be computed as follows:

$$\mathbf{v} = \beta\mathbf{v} - \alpha\nabla_{\mathbf{w}'}\left(\frac{1}{m}\sum_{i=1}^{m}\mathcal{L}\big(\boldsymbol{x}^{(i)}, \boldsymbol{y}^{(i)}; \mathbf{w}' = \mathbf{w} + \beta\mathbf{v}\big)\right), \tag{5.20.7}$$

$$\mathbf{w} = \mathbf{w} + \mathbf{v}, \tag{5.20.8}$$

where the parameters $\beta$ and $\alpha$ play a similar role as in the standard momentum method. The complete Nesterov momentum algorithm is presented in Algorithm 5.2.

The key distinction between Nesterov momentum and standard momentum lies in when the gradient is evaluated during the update step. In standard momentum, the gradient is evaluated at the current position of the parameters. The momentum term is then applied to the gradient to adjust the velocity of the update, and this adjusted velocity is used to update the parameters. However, in Nesterov momentum, the gradient is evaluated after the current velocity is applied to the parameters. This means that instead of evaluating the gradient at the current position, it's evaluated slightly ahead in the direction of the current velocity. This allows the algorithm to anticipate the next position of the parameters and adjust the velocity accordingly. By doing so, Nesterov momentum can reduce the likelihood of overshooting the minimum and improve convergence speed compared to standard momentum methods. In a sense, Nesterov momentum "looks ahead" to where the parameters are likely to be after applying the current velocity, which helps it make more accurate updates and navigate towards the minimum more efficiently.

Why does this help? Well, consider the following

- When the algorithm is far away from the minimum, the gradient at the estimated destination (based on the current velocity) is similar to the gradient at the current point. This means that the algorithm progresses towards the minimum in a manner similar to standard momentum-based gradient descent.
- As the algorithm gets closer to the minimum and the current update potentially takes it past the minimum, the gradient at the estimated destination will point in the opposite direction, indicating that the algorithm is starting to climb back up the loss surface. This is visualized as the algorithm traversing down one side of a cone-shaped loss surface and then beginning to climb up the other side.
- In Nesterov momentum, the gradient at the estimated destination and the previous step's gradient are combined with a weighted average. Since the gradients are now in opposite directions (due to the overshooting), they tend to cancel each other out in some dimensions. This cancellation results in a smaller magnitude of the combined gradient, leading to a smaller step size.
- The smaller step size obtained from the weighted average of opposing gradients helps mitigate the overshooting phenomenon. Instead of taking a large step that could potentially overshoot the minimum, the algorithm takes a smaller, more controlled step towards the minimum, thus improving convergence behavior.

**Algorithm 5.2:** Nesterov Accelerated Gradient (NAG):

Initialize Parameters:
Initialize the parameters $\mathbf{w}$ (the vector you're optimizing) to some initial values. Set the learning rate $\alpha$, momentum coefficient $\beta$, and the initial velocity $\mathbf{v}_0$ to zero or small random values.

Repeat Until Convergence:
For each iteration $t = 1,2, \ldots$:

- Sample Data: Randomly select a mini-batch of data examples $(\boldsymbol{x}^{(i)}, \boldsymbol{y}^{(i)})$ from the training dataset.
- Compute Gradient Ahead: Compute the gradient of the objective function with respect to the parameters $\mathbf{w}$, but at a point slightly ahead in the direction of the momentum. This is done by first updating the parameters with the momentum term:
$$\mathbf{w}' = \mathbf{w} + \beta\mathbf{v}.$$
Then compute the gradient at $\mathbf{w}'$:
$$\boldsymbol{g}' = \nabla_{\mathbf{w}'}\left(\frac{1}{m}\sum_{i=1}^{m}\mathcal{L}\left(\boldsymbol{x}^{(i)}, \boldsymbol{y}^{(i)}; \mathbf{w}'\right)\right),$$
where $\mathcal{L}\left(\boldsymbol{x}^{(i)}, \boldsymbol{y}^{(i)}; \mathbf{w}'\right)$ is the loss function (e.g., MSE loss, cross-entropy loss) for the example $(\boldsymbol{x}^{(i)}, \boldsymbol{y}^{(i)})$
- Update Velocity: Update the velocity $\boldsymbol{v}_t$ using the previous velocity and the current gradient:
$$\mathbf{v} = \beta\mathbf{v} - \alpha\boldsymbol{g}'.$$
- Update Parameters: Update the parameters $\mathbf{w}$ using the velocity $\mathbf{v}$:
$$\mathbf{w} = \mathbf{w} + \mathbf{v}.$$
- Repeat: Continue this process until convergence criteria are met (e.g., maximum number of iterations reached, small gradient norm, etc.).
- This algorithm is similar to standard momentum-based SGD, but it computes the gradient at a "lookahead" position $\mathbf{w}'$ before updating the velocity.

In other words, in the analogy of the rolling ball, Nesterov momentum essentially allows the algorithm to "anticipate" the reversal in gradient direction, akin to applying the brakes on the rolling ball as it approaches the bottom of the bowl. By evaluating the gradient slightly ahead in the direction of the current velocity, Nesterov momentum provides a lookahead mechanism that can detect when the algorithm is approaching the minimum. When the lookahead gradient indicates that the gradient direction is about to reverse, it signals the algorithm to adjust its velocity accordingly. This adjustment helps prevent overshooting the minimum and facilitates smoother convergence towards it.

Imagine that as you're rolling down the hill, you're not just blindly reacting to the terrain as it comes. Instead, you're peeking ahead to anticipate how the slope will change based on your momentum. If there's a sudden dip or rise just beyond your current position, you adjust your descent accordingly, taking into account both the current slope and the anticipated change due to your momentum. This forward-looking approach allows you to navigate the terrain with even greater finesse. You're not just reacting to the immediate slope beneath your feet; you're also considering how your momentum will affect your future path. This enables you to smoothly adapt to irregularities or sudden changes in the terrain, maintaining your overall progress towards the lowest point more effectively than ever before.

Indeed, in the convex batch gradient case, Nesterov momentum can significantly improve the rate of convergence compared to standard momentum-based methods. Specifically, Nesterov showed in his 1983 paper that the rate of convergence of the excess error can be accelerated from $O(1/k)$ to $O(1/k^2)$ after $k$ steps, where $k$ is the number of iterations.

However, in the stochastic gradient case, Nesterov momentum does not provide the same improvement in the rate of convergence. Stochastic gradient methods involve computing gradients using only a subset of the data (or even just a single data point) at each iteration, which introduces additional noise into the optimization process. In this setting, the conditions that enable Nesterov momentum to achieve accelerated convergence in the batch gradient case are not satisfied. The noise inherent in stochastic gradients complicates the optimization process, making it more difficult for Nesterov momentum to anticipate the direction of the next update accurately.

## 5.3 Adaptive Learning Rates Algorithms

The basic idea in the momentum methods of the previous section is to leverage the consistency in the gradient direction of certain parameters in order to speed up the updates. This goal can also be achieved more explicitly by having different learning rates for different parameters. Adaptive learning rate methods are techniques used in optimization algorithms to adjust the learning rate dynamically during the training process where instead of using a single learning rate for all parameters, individual learning rates are assigned based on the behavior of each parameter during training. The idea is that parameters with large partial derivatives or gradients tend to oscillate and zigzag during optimization, which can lead to slow convergence if a fixed learning rate is used. On the other hand, parameters with small partial derivatives exhibit more consistent behavior but may move slowly towards the optimum. The intuition behind using different learning rates for different parameters lies in the understanding that not all parameters behave the same during optimization. Some parameters may have large gradients and require smaller learning rates to prevent oscillations or overshooting, while others with smaller gradients may benefit from larger learning rates to speed up convergence. By adjusting the learning rates for each parameter individually, the optimization algorithm can effectively navigate through the parameter space, accelerating convergence and improving overall performance.

Imagine you're trying to navigate through a dense forest to reach a distant destination. Now, consider that the terrain of the forest is not uniform. Some areas have smooth paths with gentle slopes, while others are rocky and steep. In traditional navigation (akin to using a fixed learning rate in GD), you might adopt a constant pace regardless of the terrain. This approach could lead to inefficiencies or even setbacks. Adaptive learning rate methods, on the other hand, would be like adjusting your pace dynamically based on the conditions you encounter. For instance:

- When you encounter a steep incline, you slow down to conserve energy and ensure stability, preventing yourself from slipping or falling (analogous to using a smaller learning rate when gradients are large).
- Conversely, on a smooth downhill slope, you might increase your speed to make faster progress without exerting too much effort (analogous to using a larger learning rate when gradients are small).
- If you notice that certain types of obstacles require more careful navigation, such as thick underbrush or hidden roots, you adapt your approach to be more cautious and deliberate in those areas (analogous to adjusting learning rates based on the behavior of each parameter during training).

By adapting your pace to the specific challenges of the terrain, you optimize your journey through the forest, conserving energy where needed and making swift progress when possible. Similarly, in machine learning, assigning individual learning rates based on parameter behavior optimizes the training process, ensuring smooth convergence towards the optimal model.

Choosing the right optimization algorithm for training deep learning models is often a nuanced decision that depends on various factors including the dataset, model architecture, computational resources, and the specific problem being addressed. There's no one-size-fits-all answer, and the choice often involves experimentation and empirical validation. Currently, the most popular optimization algorithms actively in use include SGD, SGD with momentum, RMSProp, RMSProp with momentum, AdaDelta, AdaMax, Nadam, AMSGRAD and Adam [142-155]. The choice of which algorithm to use, at this point, seems to depend largely on the user's familiarity with the algorithm (for ease of hyperparameter tuning). In practice, it's common to start with Adam or RMSProp with momentum as they tend to perform well across many scenarios. However, fine-tuning and experimentation are crucial to finding the best-performing optimization algorithm for a specific problem. Additionally, it's important to monitor the training process and adjust hyperparameters accordingly during training.

**Remarks:**

- The performance of optimizers exhibits significant variability across different tasks.
- In practice, it's essential to consider factors such as computational resources, time constraints, and the specific requirements of your task when deciding whether to tune hyperparameters of a single fixed optimizer or evaluate multiple optimizers with default parameters. Additionally, conducting thorough experimentation and validation on your particular dataset and task is crucial to determine the most effective approach.

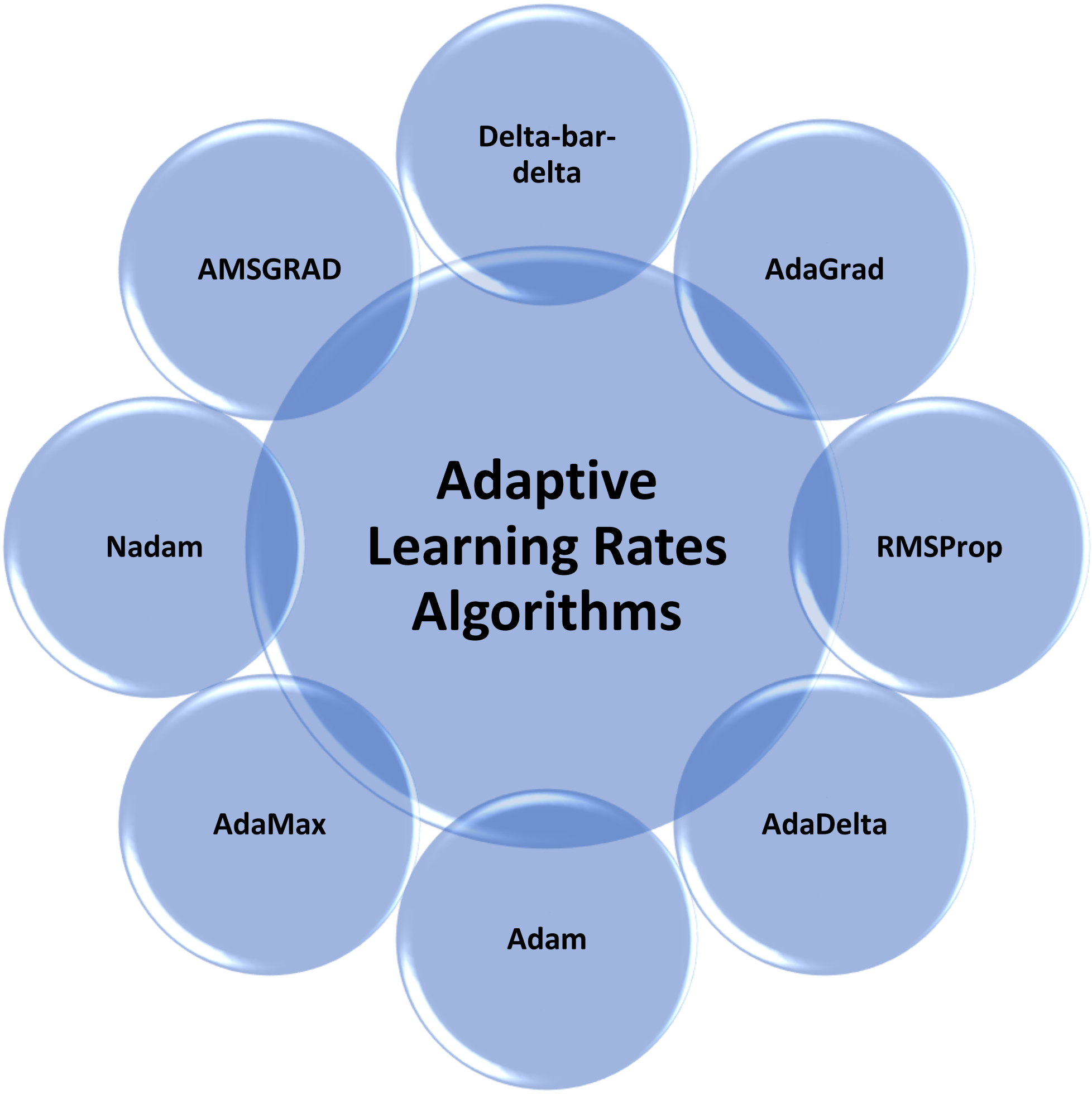

The growing literature now lists hundreds of optimization methods. Thus, we have opted to narrow our focus to eight optimizers, which currently stand as the most popular choices within the community. These do not necessarily reflect the "best" methods, but are either commonly used by practitioners and researchers, or have recently generated attention. Our selection is focused on first-order optimization methods, both due to their prevalence for non-convex optimization problems in deep learning as well as to simplify the comparison.

### 5.3.1 Delta-Bar-Delta Algorithm

The delta-bar-delta method [146] is an early attempt to adaptively adjust learning rates based on the consistency of gradient directions. The method monitors whether the sign of each partial derivative of the loss function with respect to a particular model parameter remains consistent or changes. If the sign of the partial derivative remains consistent, it suggests that the current direction is correct for minimizing the loss. In such cases, the method increases the partial derivative in that direction. Conversely, if the sign of the partial derivative flips frequently, it indicates that the current direction may not be optimal. In such cases, the method decreases the partial derivative in that direction. The delta-bar-delta method was primarily designed for full batch gradient descent, where gradients are computed using the entire dataset. In this setting, the method's approach to tracking gradient signs and adjusting derivatives based on their consistency could be effective. However, the method may not perform well with SGD, where gradients are computed using small batches of data. In SGD, fluctuations in gradients due to mini-batch sampling can lead to noisy updates, potentially amplifying errors when adjusting derivatives based on their signs.

**Algorithm 5.3:** Delta-Bar-Delta

- Initialization: Initialize the weights of the model $w_t$ and the learning rates $\alpha_t$ for each weight. Set the parameters: $\beta$ (exponential moving average parameter), $\kappa$ (constant factor for increasing learning rate), and $\phi$ (proportion for decreasing learning rate).
- For each iteration $t$ until convergence:
    1. Compute the partial derivative of the loss function $\mathcal{L}_t$ with respect to each weight $w_t$:
    $$\delta_t = \frac{\partial \mathcal{L}_t}{\partial w_t}.$$
    2. Update the exponential moving average of derivatives:
    $$\bar{\delta}_t = \beta \bar{\delta}_{t-1} + (1-\beta)\delta_t.$$
    3. For each weight $w_t$:
    If $\delta_t$ and $\bar{\delta}_{t-1}$ have the same sign: Increment the learning rate by a constant factor $\kappa$,
    $$\alpha_{t+1} = \alpha_t + \kappa.$$
    If $\delta_t$ and $\bar{\delta}_{t-1}$ have opposite signs: Decrement the learning rate by a proportion $\phi$ of its current value,
    $$\alpha_{t+1} = \alpha_t - \phi\alpha_t.$$
    Otherwise: Keep the learning rate unchanged,
    $$\alpha_{t+1} = \alpha_t.$$
    4. Update each weight using the updated learning rate and the corresponding partial derivative:
    $$w_{t+1} = w_t - \alpha_{t+1}\delta_t.$$
- Repeat steps 1-4 for a predefined number of iterations or until convergence criteria are met.

Note: The choice of parameters $\beta$, $\kappa$, and $\phi$ can impact the performance of the algorithm and may require tuning.

Mathematically, the weight update rule of delta-bar-delta method is given by

$$w_{t+1} = w_t - \alpha_{t+1}\frac{\partial \mathcal{L}_t}{\partial w_t}, \tag{5.21}$$

where $w_t$ is the value of a single weight at time $t$ and $\alpha_t$ is the learning rate value corresponding to $w_t$ at time $t$.

The learning rate update rule is defined as follows [146]:

$$\Delta\alpha_t = \begin{cases} \kappa, & \bar{\delta}_{t-1}\delta_t > 0 \\ -\phi\alpha_t, & \bar{\delta}_{t-1}\delta_t < 0, \\ 0, & \text{Otherwise} \end{cases} \tag{5.22.1}$$

where

$$\delta_t = \frac{\partial \mathcal{L}_t}{\partial w_t}, \tag{5.22.2}$$

and

$$\bar{\delta}_t = \beta\bar{\delta}_{t-1} + (1-\beta)\delta_t. \tag{5.22.3}$$

In these equations, $\delta_t$ is the partial derivative of the error with respect to $w$ at time $t$ and $\bar{\delta}_t$ is an exponential average of the current and past derivatives with $\beta$ as the exponential average coefficient. According to the delta-bar-delta algorithm, if the current derivative of a weight and the exponential average of the weight's previous derivatives possess the same sign, then the learning rate for that weight is incremented by a constant, $\kappa$. If the current derivative of a weight and the exponential average of the weight's previous derivatives possess opposite signs, then the learning rate for that weight is decremented by a proportion, $\phi$, of its current value. The delta-bar-delta rule increments learning rates linearly, but decrements them exponentially. Incrementing linearly prevents the learning rates from becoming too large too fast. Decrementing exponentially ensures that the rates are always positive and allows them to be decreased rapidly. The complete delta-bar-delta method is presented in Algorithm 5.3.

### 5.3.2 Adaptive Gradient Algorithm (AdaGrad)

In the AdaGrad algorithm [147], one keeps track of the aggregated squared magnitude of the partial derivative with respect to each parameter over the course of the algorithm. For each parameter $w_i$, AdaGrad accumulates the squared magnitude of the partial derivative of the loss function with respect to the parameter $w_i$ up to the current iteration [69]:

$$r_i \leftarrow r_i + \left(\frac{\partial \mathcal{L}}{\partial w_i}\right)^2; \quad \forall i, \tag{5.23.1}$$

where, $r_i$ is the aggregated squared magnitude of the partial derivative with respect to the parameter $w_i$. $\frac{\partial \mathcal{L}}{\partial w_i}$ is the partial derivative of the loss function with respect to $w_i$. The update rules for the $i$th parameter $w_i$ is as follows [69]:

$$w_i \leftarrow w_i - \frac{\alpha}{\sqrt{r_i}}\frac{\partial \mathcal{L}}{\partial w_i}; \quad \forall i. \tag{5.23.2}$$

If desired, one can use $\sqrt{r_i} + \epsilon$ in the denominator instead of $\sqrt{r_i}$ to avoid ill-conditioning. Here, $\epsilon$ is a small constant, such as $10^{-8}$, added for numerical stability and $\alpha$ is the learning rate.

By scaling the learning rates inversely with the square root of the sum of historical squared gradients, parameters with large partial derivatives of the loss, indicating steep slopes in the parameter space, will experience a rapid decrease in their learning rates. This means that AdaGrad effectively slows down the updates for parameters that are changing rapidly, helping to prevent overshooting and oscillations during optimization. Conversely, parameters with small partial derivatives of the loss, indicating gentle slopes in the parameter space, will experience a relatively smaller decrease in their learning rates. This allows AdaGrad to make more substantial updates to parameters in flat or slowly changing regions of the parameter space, facilitating progress in those directions. The net effect of AdaGrad's adaptive learning rate scheme is that it promotes greater progress in the more gently sloped directions of the parameter space while providing stability and controlled updates in regions with steep gradients. This property can be particularly useful in optimizing complex and nonlinear objective functions, where the landscape of the parameter space may exhibit varying degrees of curvature and steepness. The complete AdaGrad method is presented in Algorithm 5.4.

While AdaGrad offers several advantages, such as automatic adjustment of learning rates and effective handling of sparse data, it also has some limitations and potential issues:

1. The accumulation of squared gradients from the beginning of training in the AdaGrad algorithm can indeed lead to a premature and excessive decrease in the effective learning rate. This issue arises because AdaGrad accumulates the squared gradients over time in the denominator of the adaptive learning rate update rule. As training progresses, the sum of squared gradients increases, causing the adaptive learning rates to shrink. Parameters associated with steep gradients may end up with learning rates that decrease too quickly, slowing down their updates and potentially causing convergence issues. In the context of deep learning, where optimization landscapes are often highly nonlinear and complex, this premature decrease in the effective learning rate can be particularly problematic. It may lead to slow convergence, suboptimal solutions, or difficulties in escaping from saddle points or local minima. To address this limitation, subsequent adaptive optimization algorithms such as RMSProp, Adam, and AdaDelta were developed. These algorithms employ strategies to control the accumulation of squared gradients and prevent the learning rates from decreasing too rapidly over time. By adapting the learning rates more effectively, these algorithms can often achieve faster convergence and better performance in training DNNs compared to AdaGrad.
2. Another critical issue with AdaGrad is the reliance on historical information that can become stale over time. As training progresses, the accumulated squared gradients reflect gradients from earlier stages of training, potentially leading to outdated scaling factors for the learning rates. This staleness in the scaling factors can introduce inaccuracies into the optimization process. Parameters may end up being scaled incorrectly based on outdated information, which can slow down convergence or lead to suboptimal solutions. To address this problem, adaptive optimization algorithms, such as RMSProp and Adam, utilize exponential moving averages instead of accumulating squared gradients directly. Exponential averaging allows these methods to give more weight to recent gradients while still incorporating historical information, thereby reducing the impact of stale information on the scaling factors.

**Algorithm 5.4:** AdaGrad

Initialization: Initialize a vector $\mathbf{r}$ of the same dimensionality as the parameter vector $\mathbf{w}$ with zeros, $\mathbf{r} = \mathbf{0}$. Initialize the weights of the model $\mathbf{w}$ and the global learning rate $\alpha$. Small constant $\epsilon$, perhaps $10^{-8}$, for numerical stability.

For each iteration of the optimization process:

1. Sample a minibatch of $m$ examples from the training set $\{\boldsymbol{x}^{(1)}, \ldots, \boldsymbol{x}^{(m)}\}$ with corresponding targets $\boldsymbol{y}^{(i)}$.
2. Compute the gradient of the loss function with respect to the parameters:
$$\boldsymbol{g} \leftarrow \nabla_{\mathbf{w}} \left( \frac{1}{m} \sum_{i=1}^{m} \mathcal{L}\left(\boldsymbol{x}^{(i)}, \boldsymbol{y}^{(i)}; \mathbf{w}\right) \right).$$
3. Update the squared gradient accumulator $\mathbf{r}$ element-wise by adding the square of the current gradient:
$$\mathbf{r} \leftarrow \mathbf{r} + \boldsymbol{g} \odot \boldsymbol{g},$$
where $\odot$ denotes element-wise multiplication.
4. Take the element-wise square root of the accumulated squared gradients:
$$\mathbf{s} \leftarrow \sqrt{\mathbf{r}} + \epsilon,$$
where $\epsilon$ is a small constant added for numerical stability.
5. Update the parameters using the adaptive learning rate:
$$\mathbf{w} \leftarrow \mathbf{w} - \frac{\alpha}{\mathbf{s}} \odot \boldsymbol{g}.$$
6. Repeat the iterative update process for a predetermined number of iterations or until convergence criteria are met.

3. AdaGrad, while effective in adapting learning rates based on the historical gradients, can suffer from issues related to sensitivity to initial conditions and the continual decrease of learning rates throughout training. If the initial gradients are large, the accumulated sum in the denominator will be large, the learning rates will be low for the remainder of training, resulting in very small effective learning rates. This can hinder the learning process, especially at the beginning of training when the model parameters are far from optimal.
4. AdaGrad does not incorporate momentum, which can be beneficial for escaping shallow local minima or saddle points. The absence of momentum may result in slower convergence, especially in optimization landscapes with flat regions or plateaus.
5. Empirical studies have shown that while AdaGrad performs well for certain types of deep learning models and optimization problems, it may not be as effective for others.

### 5.3.3 Root Mean Square Propagation (RMSProp)

RMSProp [148] is another popular optimization algorithm, particularly in the realm of deep learning. It addresses some of the limitations of AdaGrad, specifically the decaying learning rate problem. In RMSProp, rather than accumulating squared gradients directly, exponential averaging is used to estimate the squared gradients. This approach helps to prevent the scaling factor from increasing indefinitely, thus avoiding the premature slowdown of progress during training. AdaGrad is designed to converge rapidly when applied to convex functions due to its adaptive learning rate mechanism, which prioritizes updates in directions with smaller historical gradients.

RMSProp computes an adaptive learning rate for each parameter by dividing the learning rate by the exponentially decaying average of squared gradients. This allows the algorithm to adjust the step size for each parameter based on the magnitude of its gradients. Mathematically, update the running estimate of squared gradients for the $i$th parameter $w_i$ using exponential averaging as the following

**Algorithm 5.5:** Original RMSProp

Initialization: Initialize a vector $\mathbf{r}$ of the same dimensionality as the parameter vector $\mathbf{w}$ with zeros, $\mathbf{r} = \mathbf{0}$. Initialize the weights of the model $\mathbf{w}$ and the global learning rate $\alpha$. Small constant $\epsilon$, perhaps $10^{-8}$, for numerical stability. Initialize decay rate $\rho$.

For each iteration of the optimization process:

1. Sample a minibatch of $m$ examples from the training set $\{\boldsymbol{x}^{(1)}, \dots, \boldsymbol{x}^{(m)}\}$ with corresponding targets $\boldsymbol{y}^{(i)}$.
2. Compute the gradient of the loss function with respect to the parameters:

$$\boldsymbol{g} \leftarrow \nabla_{\mathbf{w}} \left( \frac{1}{m} \sum_{i=1}^{m} \mathcal{L}\left(\boldsymbol{x}^{(i)}, \boldsymbol{y}^{(i)}; \mathbf{w}\right) \right).$$

3. Update the squared gradient accumulator $\mathbf{r}$ element-wise by adding the square of the current gradient:

$$\mathbf{r} \leftarrow \rho \mathbf{r} + (1 - \rho) \boldsymbol{g} \odot \boldsymbol{g},$$

   where $\odot$ denotes element-wise multiplication.

4. Take the element-wise square root of the accumulated squared gradients:

$$\mathbf{s} \leftarrow \sqrt{\mathbf{r}} + \epsilon,$$

   where $\epsilon$ is a small constant added for numerical stability.

5. Update the parameters using the adaptive learning rate:

$$\mathbf{w} \leftarrow \mathbf{w} - \frac{\alpha}{\mathbf{s}} \odot \boldsymbol{g}.$$

6. Repeat the iterative update process for a predetermined number of iterations or until convergence criteria are met.

$$r_i \leftarrow \rho r_i + (1 - \rho) \left( \frac{\partial \mathcal{L}}{\partial w_i} \right)^2; \quad \forall i. \tag{5.24.1}$$

The square-root of this value for each parameter is used to normalize its gradient. Then, the following update is used for (global) learning rate $\alpha$:

$$w_i \leftarrow w_i - \frac{\alpha}{\sqrt{r_i}} \frac{\partial \mathcal{L}}{\partial w_i}; \quad \forall i. \tag{5.24.2}$$

If desired, one can use $\sqrt{r_i} + \epsilon$ in the denominator instead of $\sqrt{r_i}$ to avoid ill-conditioning. Here, $\epsilon$ is a small positive value such as $10^{-7}$.

In RMSProp, the importance of ancient or stale gradients decays exponentially over time due to the use of exponential averaging. This means that older gradients have less influence on the current update, allowing RMSProp to adapt more effectively to changes in the optimization landscape. This addresses one of the shortcomings of AdaGrad, where the accumulation of historical gradients can lead to a slowdown in learning.

RMSProp can benefit from incorporating concepts of momentum within the optimization algorithm. By combining momentum with RMSProp, the algorithm can exhibit smoother convergence and improved robustness, particularly in the presence of noisy or ill-conditioned optimization landscapes. RMSProp is shown in its standard form in Algorithm 5.5 and combined with Nesterov momentum in Algorithm 5.6.

RMSProp introduces a new hyperparameter $\rho$, the decay rate, which controls the length scale of the moving average of squared gradients. This hyperparameter determines how much weight is given to past squared gradients relative to the current gradient. A larger value of $\rho$ gives more weight to recent gradients, while a smaller value allows the algorithm to consider a longer history of gradients.

**Algorithm 5.6:** RMSProp with Nesterov Momentum

Initialization:

Initialize a vector $\mathbf{r}$ of the same dimensionality as the parameter vector $\mathbf{w}$ with zeros, $\mathbf{r} = \mathbf{0}$. Initialize the weights of the model $\mathbf{w}$ and the global learning rate $\alpha$. Small constant $\epsilon$, perhaps $10^{-8}$, for numerical stability. Initialize velocity $\mathbf{v}$, momentum coefficient $\beta$ and decay rate $\rho$.

For each iteration of the optimization process:

1. Sample a minibatch of $m$ examples from the training set $\{\boldsymbol{x}^{(1)}, \ldots, \boldsymbol{x}^{(m)}\}$ with corresponding targets $\boldsymbol{y}^{(i)}$.
2. Compute the gradient of the loss function with respect to the parameters:
$$\boldsymbol{g} \leftarrow \nabla_{\mathbf{w}'} \left( \frac{1}{m} \sum_{i=1}^{m} \mathcal{L}\left(\boldsymbol{x}^{(i)}, \boldsymbol{y}^{(i)}; \mathbf{w}' = \mathbf{w} + \beta\mathbf{v}\right) \right).$$
3. Update the squared gradient accumulator $\mathbf{r}$ element-wise by adding the square of the current gradient:
$$\mathbf{r} \leftarrow \rho\mathbf{r} + (1-\rho)\boldsymbol{g} \odot \boldsymbol{g},$$
where $\odot$ denotes element-wise multiplication.
4. Take the element-wise square root of the accumulated squared gradients:
$$\mathbf{s} \leftarrow \sqrt{\mathbf{r}} + \epsilon,$$
where $\epsilon$ is a small constant added for numerical stability.
5. Compute velocity update:
$$\mathbf{v} \leftarrow \beta\mathbf{v} - \frac{\alpha}{\mathbf{s}} \odot \boldsymbol{g}.$$
6. Update the parameters using:
$$\mathbf{w} \leftarrow \mathbf{w} + \mathbf{v}.$$
7. Repeat the iterative update process for a predetermined number of iterations or until convergence criteria are met.

Note that the partial derivative of the loss function is computed at a shifted point, as is common in the Nesterov method.

Empirically, RMSProp has been shown to be effective and practical for optimizing DNNs. RMSProp's adaptive learning rate mechanism, combined with its straightforward implementation and effectiveness in practice, has made it one of the go-to optimization methods for deep learning practitioners.

### 5.3.4 AdaDelta

AdaDelta algorithm is a variant of the SGD optimization algorithm, designed to address some of the limitations of traditional SGD methods, such as the need to manually tune learning rates and the inherent instability of the learning process, especially for tasks with highly varying gradients or sparse data. AdaDelta algorithm introduced by Matthew D. Zeiler in 2012 [149]. AdaDelta eliminates the need for a learning rate hyperparameter by adaptively scaling the step sizes.

The key idea behind AdaDelta is to maintain two state variables: the accumulated squared gradients (a running average of past squared gradients, using similar update as RMSProp) and the accumulated squared parameter updates (also a running average). These state variables are used to compute the step size for each parameter update, effectively adjusting the learning rate dynamically during training.

Accumulated squared gradients ($r_i$): This variable keeps track of the squared gradients of the parameter $w_i$ over time. It acts as a running average of the past squared gradients. By accumulating these squared gradients, AdaDelta can adaptively adjust the step sizes based on the magnitudes of the gradients. This allows AdaDelta to handle varying gradients more effectively.

$$r_i \leftarrow \rho r_i + (1-\rho)\left(\frac{\partial \mathcal{L}}{\partial w_i}\right)^2 ; \quad \forall i. \tag{5.25.1}$$

Accumulated squared parameter updates ($\delta_i$): Similar to $r_i$, this variable maintains a running average of the squared parameter updates. It tracks how much each parameter has been updated over time and adjusts the step sizes accordingly. By considering the historical updates, AdaDelta can control the step sizes relative to the magnitude of the parameter changes. This helps in stabilizing the learning process and preventing large fluctuations in the parameter updates.

$$\delta_i \leftarrow \rho \delta_i + (1-\rho)(\Delta w_i)^2; \quad \forall i. \tag{5.25.2}$$

These two state variables are crucial for the adaptivity of AdaDelta. By using them to compute the step sizes for parameter updates, AdaDelta can effectively adjust the learning rates for individual parameters based on the past gradients and updates, without the need for a manually specified learning rate parameter.

The update rule for AdaDelta can be summarized as follows:

$$w_i \leftarrow w_i - \frac{\sqrt{\delta_i + \epsilon}}{\sqrt{r_i + \epsilon}} \frac{\partial \mathcal{L}}{\partial w_i}; \quad \forall i. \tag{5.25.3}$$

AdaDelta indeed doesn't have a parameter $\alpha$ for the learning rate. Instead, it adapts the step size using accumulated statistics of gradients and parameter updates. This lack of a learning rate parameter is one of the key features of AdaDelta and distinguishes it from traditional optimization algorithms like SGD, where selecting an appropriate learning rate can be crucial and often requires manual tuning.

Now, let's explore the update rule for AdaDelta to understand its intricacies. In traditional optimization methods like SGD, Momentum, and AdaGrad, the parameter updates ($\Delta w$) are not directly related to the units of the parameters. Instead, the parameter update, $\Delta w$, is proportional to the gradient, $g$, which in turn is proportional to the partial derivative of the cost function $\partial \mathcal{L}/\partial w$. Assuming the cost function is unitless, the units of the gradient are 1/units of $w$, and consequently, the units of the parameter update are also 1/units of $w$. This means that the parameter update does not have the same units as the parameters themselves. ADAGRAD also does not have correct units since the update involves ratios of gradient quantities, hence the update is unitless. In contrast, second-order optimization methods, like Newton's method, do indeed ensure that the parameter updates have the correct units. In Newton's method, the parameter update, $\Delta w$, is proportional to the inverse of the Hessian matrix, $H^{-1}$, multiplied by the gradient $g$. The Hessian matrix relates to the second derivatives of the cost function with respect to the parameters, $\partial^2 \mathcal{L}/\partial w^2$. So, the units of $\Delta w$ are indeed proportional to the units of $\partial \mathcal{L}/\partial w$ divided by the units of $\partial^2 \mathcal{L}/\partial w^2$, resulting in the correct units for the parameter updates.

$$\Delta w \propto H^{-1} g \propto \frac{\frac{\partial \mathcal{L}}{\partial w}}{\frac{\partial^2 \mathcal{L}}{\partial w^2}} \propto \text{units of } w. \tag{5.26.1}$$

Since second order methods are correct, we rearrange Newton's method (assuming a diagonal Hessian) for the inverse of the second derivative to determine the quantities involved:

$$\Delta w = \frac{\partial \mathcal{L}/\partial w}{\partial^2 \mathcal{L}/\partial w^2} \qquad \Rightarrow \qquad \frac{\Delta w}{\partial \mathcal{L}/\partial w} = \frac{1}{\partial^2 \mathcal{L}/\partial w^2}. \tag{5.26.2}$$

Assuming the cost function, $\mathcal{L}$, is unitless, second order methods such as Newton's method have the correct units for the parameter updates. This unit consistency is a significant advantage of second-order optimization methods. By incorporating information about the curvature of the cost function, these methods ensure that the parameter updates align with the units of the parameters themselves. This is particularly important when the units of the parameters have meaningful interpretations.

**Algorithm 5.7:** AdaDelta

Initialization: Initialize accumulation variables $\mathbb{E}[\boldsymbol{g}^2]_0 = \mathbf{0}$, $\mathbb{E}[\Delta\mathbf{w}^2]_0 = \mathbf{0}$. Initialize the weights of the model $\mathbf{w}$. Small constant $\epsilon$, perhaps $10^{-8}$, for numerical stability. Initialize decay rate $\rho$.

For each iteration of the optimization process:

1. Sample a minibatch of $m$ examples from the training set $\{\boldsymbol{x}^{(1)}, \dots, \boldsymbol{x}^{(m)}\}$ with corresponding targets $\boldsymbol{y}^{(i)}$.
2. Compute the gradient of the loss function with respect to the parameters:
$$\boldsymbol{g}_t = \nabla_{\mathrm{w}}\left(\frac{1}{m}\sum_{i=1}^{m}\mathcal{L}\left(\boldsymbol{x}^{(i)}, \boldsymbol{y}^{(i)}; \mathbf{w}\right)\right).$$
3. Accumulate Squared Gradient:
$$\mathbb{E}[\boldsymbol{g}^2]_t = \rho\mathbb{E}[\boldsymbol{g}^2]_{t-1} + (1-\rho)\boldsymbol{g}_t^2.$$
4. Compute Update:
$$\Delta\mathbf{w}_t = -\frac{\mathrm{RMS}[\Delta\mathbf{w}]_{t-1}}{\mathrm{RMS}[\boldsymbol{g}]_t}\boldsymbol{g}_t,$$
where $\mathrm{RMS}[\boldsymbol{g}]_t = \sqrt{\mathbb{E}[\boldsymbol{g}^2]_t + \epsilon}$ and $\mathrm{RMS}[\Delta\mathbf{w}]_{t-1} = \sqrt{\mathbb{E}[\Delta\mathbf{w}^2]_{t-1} + \epsilon}$.
5. Accumulate Updates:
$$\mathbb{E}[\Delta\boldsymbol{w}^2]_t = \rho\mathbb{E}[\Delta\boldsymbol{w}^2]_{t-1} + (1-\rho)\Delta\boldsymbol{w}_t^2.$$
6. Update the parameters using:
$$\mathbf{w}_{t+1} = \mathbf{w}_t + \Delta\mathbf{w}_t.$$
7. Repeat the iterative update process for a predetermined number of iterations or until convergence criteria are met.

Building upon above observations, Zeiler introduced the AdaDelta algorithm. In the update equations of both the AdaGrad and RMSProp algorithms, the denominator already accounts for the accumulated squared gradients (5.24.2) and the running estimate of squared gradients for each parameter, $w_i$, through exponential averaging (5.24.1). However, in the AdaDelta algorithm, the author introduces a measure of the change in parameter $w_i$, denoted as $\delta_i$, in the numerator (5.25.3). This consideration is essential for maintaining consistency between the units of the parameters and their updates in AdaDelta. By carefully examining the units of the involved parameters, AdaDelta ensures that the parameter updates align with the parameters themselves. The complete AdaDelta method is presented in Algorithm 5.7.

### 5.3.5 Adaptive Moment (Adam)

Adam [150] stands for "Adaptive Moment Estimation". It is an extension of the SGD algorithm that computes adaptive learning rates for each parameter. It combines ideas from two other popular optimization algorithms: RMSProp and momentum optimization.

1. Adam maintains two moving averages, namely the exponentially decaying average of past gradients (the first moment), and the exponentially decaying average of past squared gradients (the second moment). These are initialized as vectors of zeros. At each iteration, Adam computes the current gradient of the loss function with respect to the parameters. It then updates the moving averages of the gradients and squared gradients.
   - Compute the gradient of the loss function with respect to the parameters:
$$\boldsymbol{g}_t \leftarrow \nabla_{\boldsymbol{w}_{t-1}}\mathcal{L}_t(\boldsymbol{w}_{t-1}). \tag{5.27}$$
   - Update the biased first moment estimate:
$$\boldsymbol{m}_t \leftarrow \beta_1\boldsymbol{m}_{t-1} + (1-\beta_1)\boldsymbol{g}_t. \tag{5.28.1}$$
   - Update the biased second raw moment estimate:
$$\mathbf{v}_t \leftarrow \beta_2\mathbf{v}_{t-1} + (1-\beta_2)\boldsymbol{g}_t^2, \tag{5.28.2}$$

where $\boldsymbol{m}_t$ and $\mathbf{v}_t$ are the first and second moment estimates at time step $t$. $\beta_1$ and $\beta_2$ are the exponential decay rates for the first and second moments, typically set to values like 0.9 and 0.999, respectively. $\boldsymbol{g}_t^2$ indicates the elementwise square $\boldsymbol{g}_t \odot \boldsymbol{g}_t$

2. Since the moving averages are initialized as zeros, they are biased towards zero at the beginning. Adam applies bias correction by scaling these averages by factors to account for their initialization biases.
   - Compute bias-corrected first moment estimate:

$$\widehat{\boldsymbol{m}}_t \leftarrow \frac{\boldsymbol{m}_t}{1-\beta_1^t}. \tag{5.29.1}$$

   - Compute bias-corrected second moment estimate:

$$\hat{\mathbf{v}}_t \leftarrow \frac{\mathbf{v}_t}{1-\beta_2^t}, \tag{5.29.2}$$

   where $\beta_1^t$ and $\beta_2^t$ denote $\beta_1$ and $\beta_2$ to the power $t$.
3. Finally, Adam computes the update for each parameter using the corrected estimates of the first and second moments, along with a learning rate parameter.

$$\Delta \mathbf{w}_t \leftarrow -\frac{\alpha}{\sqrt{\hat{\mathbf{v}}_t}+\epsilon}\widehat{\boldsymbol{m}}_t, \tag{5.30.1}$$

   where $\alpha$ is the learning rate, good default settings for the tested machine learning problems are $\alpha = 0.001$. $\epsilon$ is a small constant to avoid division by zero, typically set to a small value like $10^{-8}$.
4. Update the parameters:

$$\boldsymbol{w}_t \leftarrow \boldsymbol{w}_{t-1} + \Delta \boldsymbol{w}_t. \tag{5.30.2}$$

Note that the efficiency of algorithm can, at the expense of clarity, be improved upon by changing the order of computation, e.g. by replacing the last three lines in the loop with the following lines:

$$\alpha_t = \alpha \frac{\sqrt{1-\beta_2^t}}{1-\beta_1^t}, \tag{5.31}$$

and

$$\boldsymbol{w}_t = \boldsymbol{w}_{t-1} - \frac{\alpha_t \boldsymbol{m}_t}{\sqrt{\mathbf{v}_t}+\hat{\epsilon}}. \tag{5.32}$$

Now, we will derive the term for the second moment estimate; the derivation for the first moment estimate is completely analogous. Let $\boldsymbol{g}$ be the gradient of the stochastic objective $\mathcal{L}$, and we wish to estimate its second raw moment (uncentered variance) using an exponential moving average of the squared gradient, with decay rate $\beta_2$. Let $\boldsymbol{g}_1, \ldots, \boldsymbol{g}_T$ be the gradients at subsequent timesteps, each a draw from an underlying gradient distribution $\boldsymbol{g}_t \sim p(\boldsymbol{g}_t)$. Let us initialize the exponential moving average as $\mathbf{v}_0 = \mathbf{0}$ (a vector of zeros). First note that the update at timestep $t$ of the exponential moving average $\mathbf{v}_t = \beta_2 \mathbf{v}_{t-1} + (1-\beta_2)\boldsymbol{g}_t^2$ (where $\boldsymbol{g}_t^2$ indicates the elementwise square $\boldsymbol{g}_t \odot \boldsymbol{g}_t$) can be written as a function of the gradients at all previous timesteps:

$$\begin{aligned}
\mathbf{v}_t &= \beta_2 \mathbf{v}_{t-1} + (1-\beta_2)\boldsymbol{g}_t^2 \\
&= \beta_2(\beta_2 \mathbf{v}_{t-2} + (1-\beta_2)\boldsymbol{g}_{t-1}^2) + (1-\beta_2)\boldsymbol{g}_t^2 \\
&= \beta_2^2 \mathbf{v}_{t-2} + \beta_2(1-\beta_2)\boldsymbol{g}_{t-1}^2 + (1-\beta_2)\boldsymbol{g}_t^2 \\
&= \beta_2^2 (\beta_2 \mathbf{v}_{t-3} + (1-\beta_2)\boldsymbol{g}_{t-2}^2) + \beta_2(1-\beta_2)\boldsymbol{g}_{t-1}^2 + (1-\beta_2)\boldsymbol{g}_t^2 \\
&= \beta_2^3 \mathbf{v}_{t-3} + \beta_2^2(1-\beta_2)\boldsymbol{g}_{t-2}^2 + \beta_2(1-\beta_2)\boldsymbol{g}_{t-1}^2 + (1-\beta_2)\boldsymbol{g}_t^2 \\
&= \beta_2^3 \mathbf{v}_{t-3} + \{\beta_2^2\boldsymbol{g}_{t-2}^2 + \beta_2 \boldsymbol{g}_{t-1}^2 + \boldsymbol{g}_t^2\}(1-\beta_2) \\
&= \beta_2^4 \mathbf{v}_{t-4} + \{\beta_2^3\boldsymbol{g}_{t-3}^2 + \beta_2^2\boldsymbol{g}_{t-2}^2 + \beta_2 \boldsymbol{g}_{t-1}^2 + \boldsymbol{g}_t^2\}(1-\beta_2) \\
&= \beta_2^t \mathbf{v}_0 + \{\beta_2^{t-1}\boldsymbol{g}_1^2 + \cdots + \beta_2^3\boldsymbol{g}_{t-3}^2 + \beta_2^2\boldsymbol{g}_{t-2}^2 + \beta_2 \boldsymbol{g}_{t-1}^2 + \boldsymbol{g}_t^2\}(1-\beta_2) \\
&= 0 + \sum_{i=1}^{t} \beta_2^{t-i}\boldsymbol{g}_i^2(1-\beta_2) \\
&= (1-\beta_2)\sum_{i=1}^{t} \beta_2^{t-i}\boldsymbol{g}_i^2.
\end{aligned} \tag{5.33}$$

We wish to know how $\mathbb{E}[\mathbf{v}_t]$, the expected value of the exponential moving average at timestep $t$, relates to the true second moment $\mathbb{E}[\boldsymbol{g}_t^2]$, so we can correct for the discrepancy between the two. Taking expectations of the left-hand and right-hand sides of (5.33):

$$
\begin{aligned}
\mathbb{E}[\mathbf{v}_t] &= \mathbb{E}\left[(1-\beta_2)\sum_{i=1}^{t}\beta_2^{t-i}\boldsymbol{g}_i^2\right] \\
&= (1-\beta_2)\sum_{i=1}^{t}\beta_2^{t-i}\mathbb{E}[\boldsymbol{g}_i^2] \\
&= \mathbb{E}[\boldsymbol{g}_t^2](1-\beta_2)\sum_{i=1}^{t}\beta_2^{t-i} + \zeta \\
&= \mathbb{E}[\boldsymbol{g}_t^2](1-\beta_2)\left(\frac{1-\beta_2^t}{1-\beta_2}\right) + \zeta \\
&= \mathbb{E}[\boldsymbol{g}_t^2](1-\beta_2^t) + \zeta, \qquad (5.34)
\end{aligned}
$$

where $\zeta = 0$ if the true second moment $\mathbb{E}[\boldsymbol{g}_i^2]$ is stationary; otherwise $\zeta$ can be kept small since the exponential decay rate $\beta_1$ can (and should) be chosen such that the exponential moving average assigns small weights to gradients too far in the past. What is left is the term $(1-\beta_2^t)$ which is caused by initializing the running average with zeros. In Adam algorithm we therefore divide by this term to correct the initialization bias.

In case of sparse gradients, for a reliable estimate of the second moment one needs to average over many gradients by choosing a small value of $\beta_2$; however, it is exactly this case of small $\beta_2$ where a lack of initialisation bias correction would lead to initial steps that are much larger.

Adam has several hyperparameters, including the learning rate $\alpha$, $\beta_1$ (exponential decay rate for the first moment estimate), $\beta_2$ (exponential decay rate for the second moment estimate), and $\epsilon$ (a small constant added to the denominator to prevent division by zero). These hyperparameters need to be tuned for optimal performance based on the specific problem and dataset.

Adam is known for its effectiveness in training DNNs and is widely used in practice due to its good convergence properties and ease of use. However, it's not necessarily always the best choice for every problem, and researchers continue to develop new optimization algorithms to address different challenges in training deep learning models.

**Remarks:**

An optimization method closely related to Adam is RMSProp. There are a few important differences between RMSProp with momentum and Adam:

1. First, in Adam, momentum is incorporated directly as an estimate of the first-order moment (with exponential weighting) of the gradient. The most straightforward way to add momentum to RMSProp is to apply momentum to the rescaled gradients. The use of momentum in combination with rescaling does not have a clear theoretical motivation.
2. Second, Adam includes bias corrections to the estimates of both the first-order moments (the momentum term) and the (uncentered) second-order moments to account for their initialization at the origin (see Algorithm 5.8). RMSProp also incorporates an estimate of the (uncentered) second-order moment; however, it lacks the correction factor. Thus, unlike in Adam, the RMSProp second-order moment estimate may have high bias early in training. Adam is generally regarded as being fairly robust to the choice of hyperparameters, though the learning rate sometimes needs to be changed from the suggested default.

**Algorithm 5.8:** Adam

$\boldsymbol{g}_t^2$ indicates the elementwise square $\boldsymbol{g}_t \odot \boldsymbol{g}_t$. Good default settings for the tested machine learning problems are $\alpha = 0.001, \beta_1 = 0.9, \beta_2 = 0.999$ and $\epsilon = 10^{-8}$. All operations on vectors are element-wise. With $\beta_1^t$ and $\beta_2^t$ means $\beta_1$ and $\beta_2$ to the power $t$.

Require: α: Stepsize. $\beta_1$, $\beta_2 \in [0, 1)$: Exponential decay rates for the moment estimates. $\mathcal{L}$: Stochastic objective function with parameters $\boldsymbol{w}$.

Initialization: $\boldsymbol{w}_0$: Initial parameter vector. $\boldsymbol{m}_0 \leftarrow \mathbf{0}$ (Initialize 1st moment vector). $\mathbf{v}_0 \leftarrow \mathbf{0}$ (Initialize 2nd moment vector). $t \leftarrow 0$ (Initialize timestep).

For each iteration of the optimization process:

1. Sample a minibatch of $m$ examples from the training set $\{\boldsymbol{x}^{(1)}, \dots, \boldsymbol{x}^{(m)}\}$ with corresponding targets $\boldsymbol{y}^{(i)}$.
2. Compute the gradient of the loss function with respect to the parameters:
$$\boldsymbol{g}_t \leftarrow \nabla_{\mathbf{w}} \left( \frac{1}{m} \sum_{i=1}^{m} \mathcal{L}\left(\boldsymbol{x}^{(i)}, \boldsymbol{y}^{(i)}; \mathbf{w}_{t-1}\right) \right).$$
3. Update biased first moment estimate:
$$\boldsymbol{m}_t \leftarrow \beta_1 \boldsymbol{m}_{t-1} + (1 - \beta_1)\boldsymbol{g}_t.$$
4. Update the biased second raw moment estimate:
$$\mathbf{v}_t \leftarrow \beta_2 \mathbf{v}_{t-1} + (1 - \beta_2)\boldsymbol{g}_t^2.$$
5. Compute bias-corrected first moment estimate:
$$\widehat{\boldsymbol{m}}_t \leftarrow \frac{\boldsymbol{m}_t}{1 - \beta_1^t}.$$
6. Compute bias-corrected second moment estimate:
$$\hat{\mathbf{v}}_t \leftarrow \frac{\mathbf{v}_t}{1 - \beta_2^t}.$$
7. Compute the update for each parameter:
$$\Delta \mathbf{w}_t \leftarrow -\frac{\alpha}{\sqrt{\hat{\mathbf{v}}_t} + \epsilon} \widehat{\boldsymbol{m}}_t.$$
8. Update the parameters:
$$\boldsymbol{w}_t \leftarrow \boldsymbol{w}_{t-1} + \Delta \boldsymbol{w}_t.$$
9. Repeat the iterative update process for a predetermined number of iterations or until convergence criteria are met.

### 5.3.6 AdaMax

AdaMax [150] is a variant of the Adam optimization algorithm, designed to address some of its limitations. While Adam stands for "Adaptive Moment Estimation," AdaMax stands for "Adaptive Moments with Maximum Norm." Both algorithms are widely used in training DNNs.

The key difference between AdaMax and Adam lies in the way the second moment is computed. While Adam calculates the second moment as the exponentially decaying average of the square of gradients, AdaMax computes the exponentially weighted infinity norm (maximum norm) of the gradients. This adjustment is intended to make AdaMax less sensitive to large gradients, which can lead to more stable training in some cases. AdaMax is shown in its standard form in Algorithm 5.9.

AdaMax is particularly useful when dealing with very sparse data and in cases where adaptive learning rates are required. It has been shown to perform well in a variety of deep learning tasks and is commonly used alongside other optimization algorithms like RMSprop and SGD with momentum.

**Algorithm 5.9:** AdaMax

Like Adam, AdaMax requires initializing parameters such as learning rate $\alpha$, exponential decay rates for the moment estimates $\beta_1$ and $\beta_2$, and a small constant to prevent division by zero (usually denoted as $\epsilon$). Good default settings for the tested machine learning problems are $\alpha = 0.002$, $\beta_1 = 0.9$ and $\beta_2 = 0.999$. With $\beta_1^t$ we denote $\beta_1$ to the power $t$. Here, $(\alpha/(1-\beta_1^t))$ is the learning rate with the bias-correction term for the first moment. All operations on vectors are element-wise.

Initialization: $\boldsymbol{w}_0$: Initial parameter vector. $\boldsymbol{m}_0 \leftarrow \mathbf{0}$ (Initialize 1st moment vector). $\mathbf{u}_0 \leftarrow \mathbf{0}$ (Initialize 2nd moment vector). $t \leftarrow 0$ (Initialize timestep).

Iterations: During each iteration of training:

- Compute gradients w.r.t. stochastic objective at timestep $t$.
$$\boldsymbol{g}_t \leftarrow \nabla_w \mathcal{L}_t(\boldsymbol{w}_{t-1}),$$
where $\mathcal{L}$ stochastic objective function with parameters $\boldsymbol{w}$.
- The first moment $\boldsymbol{m}_t$ is calculated similarly to Adam:
$$\boldsymbol{m}_t \leftarrow \beta_1 \boldsymbol{m}_{t-1} + (1-\beta_1)\boldsymbol{g}_t.$$
- Unlike Adam, which uses the $L^2$ norm of past gradients, AdaMax utilizes the infinity norm. Update the exponentially moving average of the absolute gradients (the exponentially weighted infinity norm):
$$\mathbf{u}_t \leftarrow \max(\beta_2 \mathbf{u}_{t-1}, |\boldsymbol{g}_t|).$$
This update mechanism guarantees that $\mathbf{u}_t$ always captures the maximum observed gradient magnitude up to time step $t$. If the current gradient magnitude $|\boldsymbol{g}_t|$ is larger than $\beta_2 \mathbf{u}_{t-1}$, $\mathbf{u}_t$ is updated to $|\boldsymbol{g}_t|$, otherwise, it remains $\beta_2 \mathbf{u}_{t-1}$.
- Update parameters:
$$\boldsymbol{w}_t \leftarrow \boldsymbol{w}_{t-1} - \frac{\alpha}{1-\beta_1^t}\frac{\boldsymbol{m}_t}{\mathbf{u}_t + \epsilon}.$$

We'll now derive the algorithm. Let, in case of the $L^p$ norm, the stepsize at time $t$ be inversely proportional to $(\mathbf{v}_t)^{\frac{1}{p}}$, where:

$$\begin{aligned}\mathbf{v}_t &= \beta_2^p \mathbf{v}_{t-1} + \left(1-\beta_2^p\right)|\boldsymbol{g}_t|^p \\ &= \left(1-\beta_2^p\right)\sum_{i=1}^{t} \beta_2^{p(t-i)} |\boldsymbol{g}_i|^p.\end{aligned} \tag{5.35}$$

Note that the decay term is here equivalently parameterized as $\beta_2^p$ instead of $\beta_2$. Now let $p \to \infty$, and define $\mathbf{u}_t = \lim_{p\to\infty} (\mathbf{v}_t)^{\frac{1}{p}}$, then:

$$\begin{aligned}\mathbf{u}_t &= \lim_{p\to\infty} (\mathbf{v}_t)^{\frac{1}{p}} \\ &= \lim_{p\to\infty}\left(\left(1-\beta_2^p\right)\sum_{i=1}^{t}\beta_2^{p(t-i)}|\boldsymbol{g}_i|^p\right)^{\frac{1}{p}} \\ &= \lim_{p\to\infty}\left[\left(1-\beta_2^p\right)^{\frac{1}{p}}\left(\sum_{i=1}^{t}\beta_2^{p(t-i)}|\boldsymbol{g}_i|^p\right)^{\frac{1}{p}}\right] \\ &= \lim_{p\to\infty}\left[\left(1-\beta_2^p\right)^{\frac{1}{p}}\right]\lim_{p\to\infty}\left(\sum_{i=1}^{t}\beta_2^{p(t-i)}|\boldsymbol{g}_i|^p\right)^{\frac{1}{p}}.\end{aligned} \tag{5.36.1}$$

Note that $\lim_{p\to\infty}\left(1-\beta_2^p\right)^{1/p}$ converges to 1 as $p$ approaches infinity. To prove that we can use the fact that $\beta_2$ is between 0 and 1, which implies that $\beta_2^p$ approaches 0 as $p$ tends to infinity.

$$\begin{aligned}\mathbf{v}_t &= \lim_{p\to\infty}\left(\sum_{i=1}^{t}\beta_2^{p(t-i)}|\boldsymbol{g}_i|^p\right)^{\frac{1}{p}}\\ &= \lim_{p\to\infty}\left(\sum_{i=1}^{t}\left(\beta_2^{(t-i)}|\boldsymbol{g}_i|\right)^p\right)^{\frac{1}{p}}\\ &= \lim_{p\to\infty}\left(\left(\beta_2^{(t-1)}|\boldsymbol{g}_1|\right)^p+\dots+(\beta_2^1|\boldsymbol{g}_{t-1}|)^p+(|\boldsymbol{g}_t|)^p\right)^{\frac{1}{p}}\\ &= \max\left(\beta_2^{(t-1)}|\boldsymbol{g}_1|,\dots,\beta_2|\boldsymbol{g}_{t-1}|,|\boldsymbol{g}_t|\right).\end{aligned} \tag{5.36.2}$$

Which corresponds to the simple recursive formula:

$$\mathbf{u}_t = \max\left(\beta_2\mathbf{u}_{t-1},|\boldsymbol{g}_t|\right), \tag{5.37}$$

with initial value $\mathbf{u}_0 = 0$. Where, we use

$$\begin{aligned}\mathbf{u}_t &= \max\left(\beta_2\mathbf{u}_{t-1},|\boldsymbol{g}_t|\right)\\ &= \max\left(\beta_2\max\left(\beta_2\mathbf{u}_{t-2},|\boldsymbol{g}_{t-1}|\right),|\boldsymbol{g}_t|\right)\\ &= \max\left(\max\left(\beta_2^2\mathbf{u}_{t-2},\beta_2|\boldsymbol{g}_{t-1}|\right),|\boldsymbol{g}_t|\right)\\ &= \max\left(\max\left(\beta_2^2\max\left(\beta_2\mathbf{u}_{t-3},|\boldsymbol{g}_{t-2}|\right),\beta_2|\boldsymbol{g}_{t-1}|\right),|\boldsymbol{g}_t|\right)\\ &= \max\left(\max\left(\max\left(\beta_2^3\mathbf{u}_{t-3},\beta_2^2|\boldsymbol{g}_{t-2}|\right),\beta_2|\boldsymbol{g}_{t-1}|\right),|\boldsymbol{g}_t|\right)\\ &= \max\left(\beta_2^{(t-1)}|\boldsymbol{g}_1|,\dots,\beta_2^3|\boldsymbol{g}_{t-3}|,\beta_2^2|\boldsymbol{g}_{t-2}|,\beta_2|\boldsymbol{g}_{t-1}|,|\boldsymbol{g}_t|\right).\end{aligned} \tag{5.38}$$

In the expression, $\sum_{i=1}^{t}\left(\beta_2^{(t-i)}|\boldsymbol{g}_i|\right)^p$ , (5.36.2), as $p$ tends to infinity, each term within the sum behaves differently based on the magnitude of $\beta_2^{(t-i)}|\boldsymbol{g}_i|$. Let's consider two scenarios. If $\beta_2^{(t-i)}|\boldsymbol{g}_i|$ is relatively small, when $p$ is large, raising a small value to a large power makes it approach zero rapidly. Thus, smaller terms within the sum will converge to zero faster than larger terms. If $\beta_2^{(t-i)}|\boldsymbol{g}_i|$ is relatively large, raising a larger value to a large power results in a value that grows without bound. Therefore, larger terms within the sum will contribute significantly more to the overall value of the sum as $p$ tends to infinity. Since the expression involves a summation, the overall behavior of the sum is dominated by the term with the largest magnitude, as it contributes the most to the sum. This term represents the maximum absolute value among all terms in the sequence $\beta_2^{(t-i)}|\boldsymbol{g}_i|$. Therefore, as $p$ tends to infinity, the sum is essentially determined by the largest term, which corresponds to the maximum absolute value among all terms in the sequence $\beta_2^{(t-i)}|\boldsymbol{g}_i|$. This observation is crucial for understanding why the expression converges to the maximum absolute value.

### 5.3.7 Nadam

The Nadam algorithm [151], short for Nesterov-accelerated Adaptive Moment Estimation, is a variant of the popular Adam optimization algorithm commonly used in training NNs. Nadam combines the adaptive learning rates of Adam with Nesterov momentum.

Sutskever et al. [139] introduced the NAG algorithm, a variant of gradient descent designed to expedite convergence by incorporating momentum. They proposed a modification to the gradient computation within the SGD algorithm, outlined as follows:

$$\boldsymbol{g}_t \leftarrow \nabla_{\boldsymbol{w}_{t-1}}\mathcal{L}_t(\boldsymbol{w}_{t-1}-\mu\boldsymbol{m}_{t-1}). \tag{5.39}$$

Moreover, they suggested updating the momentum term $\boldsymbol{m}_t$ as follows:

**Algorithm 5.10:** Nadam

Require: $\alpha_0, \ldots, \alpha_T$ ; $\beta_0, \ldots, \beta_T$ ; $\nu$; $\epsilon$: Hyperparameters
$\boldsymbol{m}_0$; $\boldsymbol{n}_0 \leftarrow \boldsymbol{0}$ (first/second moment vectors)
while $\boldsymbol{w}_t$ not converged do

1. Compute the gradient of the loss function with respect to the parameters:
$$\boldsymbol{g}_t \leftarrow \boldsymbol{\nabla}_{\boldsymbol{w}_{t-1}} \mathcal{L}_t(\boldsymbol{w}_{t-1}).$$
2. Update the biased first moment $\boldsymbol{m}_t$ estimate:
$$\boldsymbol{m}_t \leftarrow \beta_t \boldsymbol{m}_{t-1} + (1-\beta_t)\boldsymbol{g}_t.$$
3. Update the biased second raw moment estimate:
$$\boldsymbol{n}_t \leftarrow \nu \boldsymbol{n}_{t-1} + (1-\nu)\boldsymbol{g}_t^2.$$
4. Compute bias-corrected first moment estimate:
$$\widehat{\boldsymbol{m}}_t \leftarrow \frac{\beta_{t+1}\boldsymbol{m}_t}{(1-\prod_{i=1}^{t+1}\beta_i)} + \frac{(1-\beta_t)\boldsymbol{g}_t}{(1-\prod_{i=1}^{t}\beta_i)}.$$
5. Compute bias-corrected second moment estimate:
$$\widehat{\boldsymbol{n}}_t \leftarrow \frac{\boldsymbol{n}_t}{1-\nu^t}.$$
6. Update parameters:
$$\boldsymbol{w}_t \leftarrow \boldsymbol{w}_{t-1} - \frac{\alpha_t}{\sqrt{\widehat{\boldsymbol{n}}_t + \epsilon}}\widehat{\boldsymbol{m}}_t.$$

end while

$$\boldsymbol{m}_t \leftarrow \beta \boldsymbol{m}_{t-1} + \alpha_t \boldsymbol{g}_t. \tag{5.40}$$

Finally, the parameter update step was defined as:

$$\boldsymbol{w}_t \leftarrow \boldsymbol{w}_{t-1} - \boldsymbol{m}_t. \tag{5.41}$$

Dozat [151], in his development of the Nadam algorithm, proposed a refinement to the gradient computation of the NAG algorithm. Instead of updating parameters solely with the momentum step to compute the gradient (5.39), then reverting that step to restore the original parameter state, and subsequently applying the momentum step again during the actual update (5.41), he recommended applying the momentum step of timestep $t+1$ only once, during the update of the previous timestep $t$ instead of $t+1$. This is expressed as:

$$\boldsymbol{g}_t \leftarrow \boldsymbol{\nabla}_{\boldsymbol{w}_{t-1}} \mathcal{L}_t(\boldsymbol{w}_{t-1}), \tag{5.42.1}$$
$$\boldsymbol{m}_t \leftarrow \beta_t \boldsymbol{m}_{t-1} + \alpha_t \boldsymbol{g}_t, \tag{5.42.2}$$
$$\boldsymbol{w}_t \leftarrow \boldsymbol{w}_{t-1} - (\beta_{t+1}\boldsymbol{m}_t + \alpha_t \boldsymbol{g}_t). \tag{5.42.3}$$

This modification simplifies the implementation by reducing the number of steps required to apply the momentum term. In the modified NAG algorithm, both the momentum step and the gradient step, in parameter update equation, depend directly on the current gradient $\boldsymbol{g}_t$. It is important to note that, while this modification simplifies implementation, it still relies on some level of intuitiveness compared to the original NAG algorithm.

Note that the update at timestep $t$ of the exponential moving average $\boldsymbol{m}_t = \beta_t\, \boldsymbol{m}_{t-1} + (1-\beta_t)\boldsymbol{g}_t$ for Adam algorithm can be written as a function of the gradients at all previous timesteps (where, $\beta_t$ is used instead of $\beta_1$):

$$\begin{aligned}
\boldsymbol{m}_t &= \beta_t\, \boldsymbol{m}_{t-1} + (1-\beta_t)\boldsymbol{g}_t \\
&= \beta_t(\beta_{t-1}\, \boldsymbol{m}_{t-2} + (1-\beta_{t-1})\boldsymbol{g}_{t-1}) + (1-\beta_t)\boldsymbol{g}_t \\
&= \beta_t\beta_{t-1}\, \boldsymbol{m}_{t-2} + \beta_t(1-\beta_{t-1})\boldsymbol{g}_{t-1} + (1-\beta_t)\boldsymbol{g}_t \\
&= \beta_t\beta_{t-1}\, (\beta_{t-2}\, \boldsymbol{m}_{t-3} + (1-\beta_{t-2})\boldsymbol{g}_{t-2}) + \beta_t(1-\beta_{t-1})\boldsymbol{g}_{t-1} + (1-\beta_t)\boldsymbol{g}_t \\
&= \beta_t\beta_{t-1}\beta_{t-2}\, \boldsymbol{m}_{t-3} + \beta_t\beta_{t-1}(1-\beta_{t-2})\boldsymbol{g}_{t-2} + \beta_t(1-\beta_{t-1})\boldsymbol{g}_{t-1} + (1-\beta_t)\boldsymbol{g}_t \\
&= \beta_t \ldots \beta_1\, \boldsymbol{m}_0 + \{\beta_t \ldots \beta_2(1-\beta_1)\boldsymbol{g}_1 \ldots + \beta_t\beta_{t-1}(1-\beta_{t-2})\boldsymbol{g}_{t-2} + \beta_t(1-\beta_{t-1})\boldsymbol{g}_{t-1}\} \\
&\qquad + (1-\beta_t)\boldsymbol{g}_t \\
&= 0 + \{\beta_t\beta_{t-1} \ldots \beta_2(1-\beta_1)\boldsymbol{g}_1 + \cdots + \beta_t\beta_{t-1}(1-\beta_{t-2})\boldsymbol{g}_{t-2} + \beta_t(1-\beta_{t-1})\boldsymbol{g}_{t-1}\} + (1-\beta_t)\boldsymbol{g}_t \\
&= \sum_{j=1}^{t-1}\left\{\left(\prod_{i=j+1}^{t}\beta_i\right)(1-\beta_j)\boldsymbol{g}_j\right\} + (1-\beta_t)\boldsymbol{g}_t.
\end{aligned} \tag{5.43}$$

Taking expectations of the left-hand and right-hand sides of (5.43):

$$\begin{aligned}
\mathbb{E}[\boldsymbol{m}_t] &= \mathbb{E}\left[\sum_{j=1}^{t-1}\left\{\left(\prod_{i=j+1}^{t}\beta_i\right)(1-\beta_j)\boldsymbol{g}_j\right\} + (1-\beta_t)\boldsymbol{g}_t\right] \\
&= \sum_{j=1}^{t-1}\left\{\left(\prod_{i=j+1}^{t}\beta_i\right)(1-\beta_j)\mathbb{E}[\boldsymbol{g}_j]\right\} + (1-\beta_t)\mathbb{E}[\boldsymbol{g}_t] \\
&= \mathbb{E}[\boldsymbol{g}_t]\sum_{j=1}^{t-1}\left\{\prod_{i=j+1}^{t}\beta_i - \prod_{i=j+1}^{t}\beta_i\beta_j\right\} + (1-\beta_t)\mathbb{E}[\boldsymbol{g}_t] + \zeta \\
&= \mathbb{E}[\boldsymbol{g}_t]\left(\sum_{j=1}^{t-1}\left\{-\prod_{i=j}^{t}\beta_i + \prod_{i=j+1}^{t}\beta_i\right\} + (1-\beta_t)\right) + \zeta \\
&= \mathbb{E}[\boldsymbol{g}_t]\left(-\prod_{i=1}^{t}\beta_i + \prod_{i=t}^{t}\beta_i + (1-\beta_t)\right) + \zeta \\
&= \mathbb{E}[\boldsymbol{g}_t]\left(-\prod_{i=1}^{t}\beta_i + \beta_t + (1-\beta_t)\right) + \zeta \\
&= \mathbb{E}[\boldsymbol{g}_t]\left(1-\prod_{i=1}^{t}\beta_i\right) + \zeta,
\end{aligned} \tag{5.44}$$

where

$$\begin{aligned}
&\sum_{j=1}^{t-1}\left\{-\prod_{i=j}^{t}\beta_i + \prod_{i=j+1}^{t}\beta_i\right\} \\
&= \left\{-\prod_{i=1}^{t}\beta_i + \prod_{i=2}^{t}\beta_i\right\} + \left\{-\prod_{i=2}^{t}\beta_i + \prod_{i=3}^{t}\beta_i\right\} + \left\{-\prod_{i=3}^{t}\beta_i + \prod_{i=4}^{t}\beta_i\right\} + \cdots + \left\{-\prod_{i=t-1}^{t}\beta_i + \prod_{i=t}^{t}\beta_i\right\} \\
&= -\prod_{i=1}^{t}\beta_i + \prod_{i=t}^{t}\beta_i \\
&= -\prod_{i=1}^{t}\beta_i + \beta_t,
\end{aligned} \tag{5.45}$$

and $\zeta = 0$ if the true first moment $\mathbb{E}[\boldsymbol{g}_i]$ is stationary. What is left is the term $(1 - \prod_{i=1}^{t}\beta_i)$ which is caused by initializing the running average with zeros. In the Nadam algorithm, division by this term corrects the initialization bias.

Based on the above discussion, Dozat [151] employs the same approach with Adam's momentum. Initially, Adam's update step is reformulated in terms of $\boldsymbol{m}_{t-1}$ and $\boldsymbol{g}_t$.

$$\boldsymbol{w}_t = \boldsymbol{w}_{t-1} - \alpha_t\left(\frac{\beta_t\boldsymbol{m}_{t-1}}{(1-\prod_{i=1}^{t}\beta_i)} + \frac{(1-\beta_t)\boldsymbol{g}_t}{(1-\prod_{i=1}^{t}\beta_i)}\right). \tag{5.46.1}$$

Then, the subsequent momentum step is substituted for the current one, while ensuring to address the initialization bias appropriately.

$$\boldsymbol{w}_t = \boldsymbol{w}_{t-1} - \alpha_t\left(\frac{\beta_{t+1}\boldsymbol{m}_{t}}{(1-\prod_{i=1}^{t+1}\beta_i)} + \frac{(1-\beta_t)\boldsymbol{g}_t}{(1-\prod_{i=1}^{t}\beta_i)}\right). \tag{5.46.2}$$

When Adam is modified in this manner, it results in Algorithm 5.10.

### 5.3.8 AMSGRAD

AMSGRAD [152] is an adaptive learning rate algorithm designed for training DNNs. The aim of AMSGRAD is to devise a new optimization strategy that combines the practical benefits of ADAM and RMSProp while guaranteeing convergence. Algorithm 5.11 presents the pseudocode for the algorithm.

Several proposed stochastic optimization methods that have been successfully used in training DNNs such as RMSPROP, ADAM, ADADELTA, NADAM, etc are based on using gradient updates scaled by square roots of exponential moving averages of squared past gradients. In many applications, it has been empirically observed that these algorithms fail to converge to an optimal solution (or a critical point in nonconvex settings). One cause for such failures is the exponential moving average used in the algorithms. The reliance on a short-term memory of gradients, typical of exponential moving averages, can lead to significant convergence issues. While the introduction of exponential averages was motivated by the necessity to prevent learning rates from diminishing to infinitesimal values during training, as seen in the Adagrad algorithm, this approach poses challenges in different scenarios. The limitation of updates to essentially the past few gradients can hinder convergence. In general, any algorithm that relies on an essentially fixed sized window of past gradients to scale the gradient updates will suffer from converge to highly suboptimal solutions.

In addressing this issue, Reddi et al. (2018) [152] proposed AMSGrad algorithm, which offers a solution by utilizing the maximum of past squared gradients, $\mathbf{v}_t$, rather than relying on exponential averages. This departure from the conventional approach provides a means to overcome the limitations imposed by short-term memory of gradients, thereby enhancing convergence properties and improving the effectiveness of stochastic optimization in various contexts. In other words, instead of directly utilizing $\mathbf{v}_t$, the proposed modification involves employing the previous $\hat{\mathbf{v}}_{t-1}$ if it larger than the current one. This adjustment ensures that the algorithm prioritizes the maximum of past squared gradients over the current one, thus integrating a form of memory that favors stability and consistent progress in optimization.

**Remarks:**

- In AMSGRAD, the maximum of all $\mathbf{v}_t$ up to the present time step is maintained and denoted as $\hat{\mathbf{v}}_t$. This ensures that the learning rate is always normalized by the maximum of the second moment estimates encountered so far, rather than just the current $\mathbf{v}_t$ as in ADAM. This adjustment prevents the learning rate from increasing over time, which can lead to oscillations or instability in the optimization process. By using the maximum of all $\mathbf{v}_t$ for normalizing the running average of the gradient, AMSGRAD results in a non-increasing step size and avoids the pitfalls of ADAM and RMSPROP.
- To gain a deeper understanding of the updates made by AMSGRAD, it's valuable to contrast its behavior with that of ADAM and ADAGRAD. Let's consider a specific time step $t$ and a particular coordinate $i$ within the range of dimensions $[d]$. If at this point, the squared gradient magnitude $v_{t-1,i}$ exceeds the actual gradient squared $g_{t,i}^2$ and is greater than zero, here's how the three optimization algorithms behave: ADAM tends to aggressively amplify the learning rate. However, this approach can sometimes prove detrimental to the overall performance of the optimization process. ADAGRAD, conversely, tends to slightly decrease the learning rate over time. This can often result in suboptimal performance, particularly in practice, because the accumulation of gradients over an extended period may excessively diminish the learning rate, impeding effective progress. AMSGRAD, on the other hand, takes a more balanced approach. It neither aggressively increases nor decreases the learning rate.
- In practice, a constant $\beta_{1,t}$ is typically used in AMSGRAD, similar to ADAM. This exponential decay rate controls the contribution of past gradients to the running average and helps smooth out the optimization process.
- In the update rule of AMSGRAD, Algorithm 5.11, the debiasing step, which is typically present in ADAM to correct for the bias introduced by the first and second moment estimates being initialized to zero, is omitted. This simplification doesn't affect the overall performance of AMSGrad significantly and helps streamline the algorithm for practical implementation.

**Algorithm 5.11:** AMSGRAD

Require: $\boldsymbol{w}_1 \in \mathcal{F}$, where $\mathcal{F} \in \mathbb{R}^d$ is the feasible set of points (i.e. the parameters of the model to be learned), $\alpha_1, \ldots, \alpha_T$ ; $\beta_{1,1}, \ldots, \beta_{1,T}$ and $\beta_2$.

Set $\boldsymbol{m}_0 = \mathbf{0}, \mathbf{v}_0 = \mathbf{0}$ (first/second moment vectors) and $\hat{\mathbf{v}}_0 = \mathbf{0}$.

for $t = 1$ to $T$ do

1. Compute the gradient of the loss function with respect to the parameters:
$$\boldsymbol{g}_t = \nabla_{\boldsymbol{w}_{t-1}} \mathcal{L}_t(\boldsymbol{w}_{t-1}).$$
2. Update the biased first moment $\boldsymbol{m}_t$ estimate:
$$\boldsymbol{m}_t = \beta_{1,t}\boldsymbol{m}_{t-1} + (1 - \beta_{1,t})\boldsymbol{g}_t.$$
3. Update the biased second raw moment estimate:
$$\mathbf{v}_t = \beta_2 \mathbf{v}_{t-1} + (1 - \beta_2)\boldsymbol{g}_t^2,$$
$$\hat{\mathbf{v}}_t = \max(\hat{\mathbf{v}}_{t-1}, \mathbf{v}_t) \ \text{(element} - \text{wise maximum)},$$
$$\hat{\mathbf{V}}_t = \operatorname{diag}(\hat{\mathbf{v}}_t).$$
4. Update parameters:
$$\boldsymbol{w}_{t+1} = \boldsymbol{w}_t - \frac{\alpha_t}{\sqrt{\hat{\mathbf{v}}_t}}\boldsymbol{m}_t,$$

end for

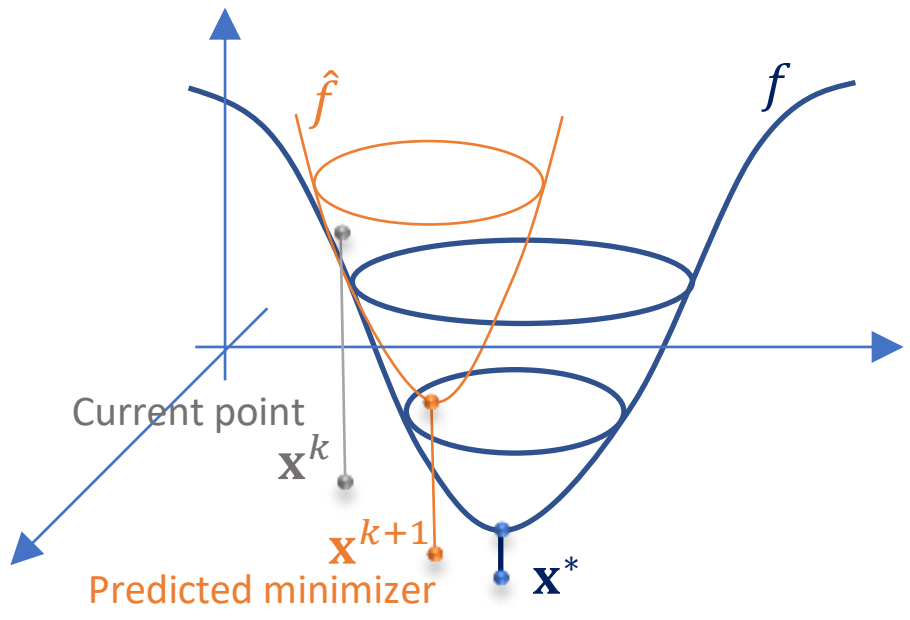

**Figure 5.13.** Quadratic approximation to the objective function using first and second derivatives.

## 5.4 Newton and Marquardt Methods (Hessian-Based Methods)

This section focuses on the second-order approximations that use the Hessian in multivariate optimization to direct the search [39, 44, 144, 153].

### 5.4.1. Newton Method

Recall that the method of steepest descent uses only first derivatives (gradients) in selecting a suitable search direction. This strategy is not always the most effective. If higher derivatives are used, the resulting iterative algorithm may perform better than the steepest descent method. The Newton method (sometimes called the Newton-Raphson method) uses first and second derivatives and, indeed, does perform better than the steepest descent method if the initial point is close to the minimizer. The idea behind this method is as follows. Given a starting point, we construct a quadratic approximation to the objective function that matches the first and second derivative values at that point. We then minimize the approximate (quadratic) function instead of the original objective function. We use the minimizer of the approximate function as the starting point in the next step and repeat the procedure iteratively. If the objective function is quadratic, then the approximation is exact, and the method yields the true minimizer in one step. If, on the other hand, the objective function is not quadratic, then the approximation will provide only an estimate of the position of the true minimizer. Figure 5.13 illustrates the above idea.

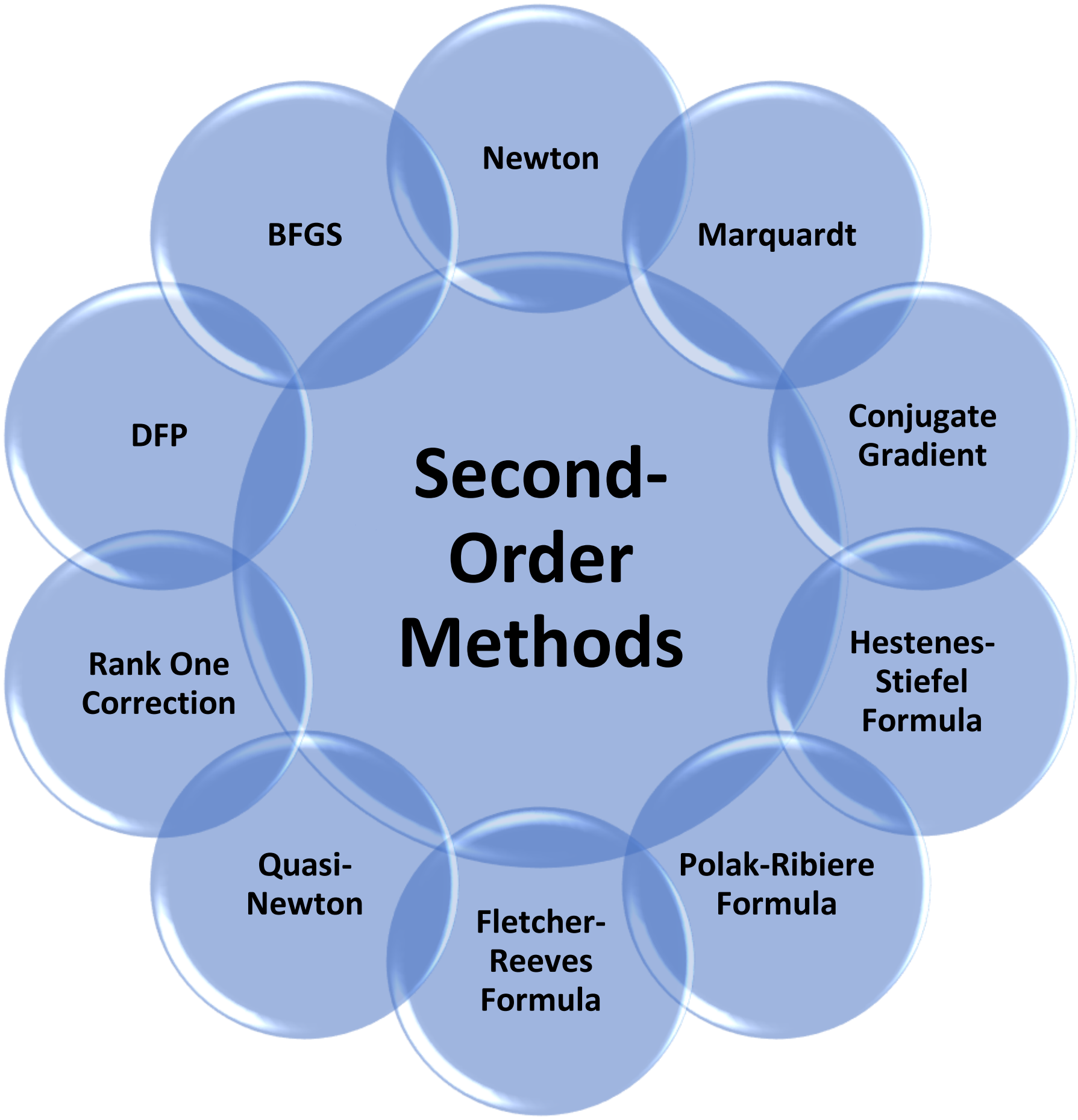

Consider the Taylor expansion of the objective:

$$f|\mathbf{x}\rangle = f|\mathbf{x}_k\rangle + \langle\nabla f(\mathbf{x}_k)|\Delta\mathbf{x}\rangle + \frac{1}{2}\langle\Delta\mathbf{x}|\nabla^2 f(\mathbf{x}_k)|\Delta\mathbf{x}\rangle + O(\Delta\mathbf{x}^3). \tag{5.47}$$

We form a quadratic approximation to $f|\mathbf{x}\rangle$ by dropping terms of order 3 and above:

$$\hat{f}(\mathbf{x};\mathbf{x}_k) = f|\mathbf{x}_k\rangle + \langle\nabla f(\mathbf{x}_k)|\Delta\mathbf{x}\rangle + \frac{1}{2}\langle\Delta\mathbf{x}|\nabla^2 f(\mathbf{x}_k)|\Delta\mathbf{x}\rangle, \tag{5.48}$$

where $\hat{f}(\mathbf{x};\mathbf{x}_k)$ denotes an approximating function constructed at $|\mathbf{x}_k\rangle$ which is itself a function of $|\mathbf{x}\rangle$. Now let us use this quadratic approximation of $f|\mathbf{x}\rangle$ to form an iteration sequence by forcing $|\mathbf{x}_{k+1}\rangle$ the next point in the sequence, to be a point where the gradient of the approximation is zero. Therefore,

$$\nabla\hat{f}(\mathbf{x};\mathbf{x}_k) = \nabla f|\mathbf{x}_k\rangle + \langle\nabla^2 f(\mathbf{x}_k)|\Delta\mathbf{x}\rangle = 0, \tag{5.49}$$

so, the search direction becomes

$$|\Delta\mathbf{x}\rangle = |\mathbf{s}_k\rangle = |\mathbf{x}_{k+1}\rangle - |\mathbf{x}_k\rangle = -\frac{\nabla f|\mathbf{x}_k\rangle}{\nabla^2 f|\mathbf{x}_k\rangle}. \tag{5.50}$$

Accordingly, this successive quadratic approximation scheme produces Newton's optimization method:

$$|\mathbf{x}_{k+1}\rangle = |\mathbf{x}_k\rangle - \frac{\nabla f|\mathbf{x}_k\rangle}{\nabla^2 f|\mathbf{x}_k\rangle} = |\mathbf{x}_k\rangle - \mathbf{H}_k^{-1}|\mathbf{g}_k\rangle, \tag{5.51}$$

where, the Hessian is $\mathbf{H}_k = \mathbf{H}_f|\mathbf{x}_k\rangle = \nabla^2 f|\mathbf{x}_k\rangle$ and $|\mathbf{g}_k\rangle = \nabla f|\mathbf{x}_k\rangle$. The complete Newton method is presented in Algorithm 5.12.

| **Algorithm 5.12:** Newton | |
|---|---|
| Step 1: | Choose a maximum number of iterations $M$ to be performed, an initial point $\|\mathbf{x}_0\rangle$, two termination parameters $\epsilon_1$, $\epsilon_2$, and set $k = 0$. |
| Step 2: | Compute $\mathbf{g}_k$ and $\mathbf{H}_k$. If $\|\mathbf{g}_k\| \leq \epsilon_1$, Terminate; If $\mathbf{H}_k$ is not a positive definite, force it to become a positive definite. Compute $\mathbf{H}_k^{-1}$ and $\|\mathbf{d}_k\rangle = -\mathbf{H}_k^{-1}\|\mathbf{g}_k\rangle$ . |
| Step 3: | Perform a unidirectional search to find $\alpha^{(k)}$ using $\epsilon_2$ such that $f\|\mathbf{x}_{k+1}\rangle = f(\|\mathbf{x}_k\rangle + \alpha_k\|\mathbf{d}_k\rangle)$, is minimum. |
| Step 4: | Set $\|\mathbf{x}_{k+1}\rangle = \|\mathbf{x}_k\rangle + \alpha_k\|\mathbf{d}_k\rangle$ $f_{k+1} = f\|\mathbf{x}_{k+1}\rangle$ |
| Step 5: | Is $\left\|\alpha^{(k)}\mathbf{d}_k\right\| \leq \epsilon_1$? If yes, then output $\|\mathbf{x}^*\rangle = \|\mathbf{x}_{k+1}\rangle$ and $f\|\mathbf{x}^*\rangle = f_{k+1}$, and Terminate; Else set $k = k + 1$ and go to Step 2. |

This solution exists if and only if the following conditions hold: (a) The Hessian is nonsingular. (b) The approximation in (5.47) is valid. It can also be shown that if the matrix $\mathbf{H}_k^{-1}$ is positive-semidefinite, the direction $|\mathbf{s}_k\rangle$ must be descent. But if the matrix $\mathbf{H}_k^{-1}$ is not positive-semidefinite, the direction $|\mathbf{s}_k\rangle$ may or may not be descent, depending on whether the quantity $\langle\mathbf{g}_k|\mathbf{H}_k^{-1}|\mathbf{g}_k\rangle$ is positive or not. Thus, the above search direction may not always guarantee a decrease in the function value in the vicinity of the current point. But the second-order optimality condition suggests that $\nabla^2 f|\mathbf{x}^*\rangle$ be positive-definite for the minimum point. Thus, it can be assumed that the matrix $\nabla^2 f|\mathbf{x}^*\rangle$ is positive-definite in the vicinity of the minimum point (i.e., for $\|\mathbf{x} - \mathbf{x}^*\| < \varepsilon$), and the method is suitable and efficient when the initial point is close to the optimum point.

**Modified Newton Method**

Any quadratic function has a Hessian, which is constant for any $|\mathbf{x}\rangle$. If the function has a minimum and the second-order sufficiency conditions for a minimum hold, then $\mathbf{H}$ is positive definite and, therefore, nonsingular at any point $|\mathbf{x}\rangle$. Since any quadratic function is represented exactly by the quadratic approximation of the Taylor series, the solution in (5.50) exists. Furthermore, for any point $|\mathbf{x}\rangle$ one iteration will yield the solution. For nonquadratic functions, the Newton step will often be large when $|\mathbf{x}_0\rangle$ is far from $|\mathbf{x}^*\rangle$, and there is the real possibility of divergence. It is possible to modify the method in a logical and simple way to ensure descent by adding a line search as in the Cauchy (steepest-descent) method. That is, we form the sequence of iterates

$$|\mathbf{x}_{k+1}\rangle = |\mathbf{x}_k\rangle + \alpha_k|\mathbf{d}_k\rangle = |\mathbf{x}_k\rangle - \alpha_k \frac{\nabla f|\mathbf{x}_k\rangle}{\nabla^2 f|\mathbf{x}_k\rangle}, \tag{5.52}$$

by choosing $\alpha_k$ such that

$$f|\mathbf{x}_{k+1}\rangle \to \min, \tag{5.53}$$

which ensures that

$$f|\mathbf{x}_{k+1}\rangle \leq f|\mathbf{x}_k\rangle, \tag{5.54}$$

where,

$$|\mathbf{d}_k\rangle = -\frac{\nabla f|\mathbf{x}_k\rangle}{\nabla^2 f|\mathbf{x}_k\rangle}. \tag{5.55}$$

This is the modified Newton method, and we find it reliable and efficient when the first and second derivatives are accurately and inexpensively calculated.

***Example 5.1***

Use Newton method to minimize the function:

$$f|\mathbf{x}\rangle = (x_1 + 10x_2)^2 + 5(x_3 - x_4)^2 + (x_2 - 2x_3)^4 + 10(x_1 - x_4)^4.$$

Use as the starting point $|\mathbf{x}_0\rangle = (3, -1, 0, 1)^T$. Perform three iterations.

***Solution***

Note that $f|\mathbf{x}_0\rangle = 215$. We have

$$\nabla f|\mathbf{x}\rangle = \begin{pmatrix} 2(x_1+10x_2)+40(x_1-x_4)^3 \\ 20(x_1+10x_2)+4(x_2-2x_3)^3 \\ 10(x_3-x_4)-8(x_2-2x_3)^3 \\ -10(x_3-x_4)-40(x_1-x_4)^3 \end{pmatrix},$$

and $\mathbf{H}_f|\mathbf{x}\rangle$ is given by

$$\begin{pmatrix} 2+120(x_1-x_4)^2 & 20 & 0 & -120(x_1-x_4)^2 \\ 20 & 200+12(x_2-2x_3)^2 & -24(x_2-2x_3)^2 & 0 \\ 0 & -24(x_2-2x_3)^2 & 10+48(x_2-2x_3)^2 & -10 \\ -120(x_1-x_4)^2 & 0 & -10 & 10+120(x_1-x_4)^2 \end{pmatrix}.$$

**Iteration 1.**

$$|\mathbf{g}_0\rangle = (306,-144,-2,-310)^T,$$

$$\mathbf{H}_0 = \begin{pmatrix} 482 & 20 & 0 & -480 \\ 20 & 212 & -24 & 0 \\ 0 & -24 & 58 & -10 \\ -480 & 0 & -10 & 490 \end{pmatrix},$$

$$\mathbf{H}_0^{-1} = \begin{pmatrix} .1126 & -.0089 & .0154 & .1106 \\ -.0089 & .0057 & .0008 & -.0087 \\ .0154 & .0008 & .0203 & .0155 \\ .1106 & -.0087 & .0155 & .1107 \end{pmatrix},$$

$$\mathbf{H}_0^{-1}|\mathbf{g}_0\rangle = (1.4127,-0.8413,-0.2540,0.7460)^T.$$

Hence,

$$|\mathbf{x}_1\rangle = |\mathbf{x}_0\rangle - \mathbf{H}_0^{-1}|\mathbf{g}_0\rangle = (1.5873,-0.1587,0.2540,0.2540)^T, \qquad f|\mathbf{x}_1\rangle = 31.8.$$

**Iteration 2.**

$$|\mathbf{g}_1\rangle = (94.81,-1.179,2.371,-94.81)^T,$$

$$\mathbf{H}_1 = \begin{pmatrix} 215.3 & 20 & 0 & -213.3 \\ 20 & 205.3 & -10.67 & 0 \\ 0 & -10.67 & 31.34 & -10 \\ -213.3 & 0 & -10 & 223.3 \end{pmatrix},$$

$$\mathbf{H}_1^{-1}|\mathbf{g}_1\rangle = (0.5291,-0.0529,-0.0846,0.0846)^T,$$

Hence,

$$|\mathbf{x}_2\rangle = |\mathbf{x}_1\rangle - \mathbf{H}_1^{-1}|\mathbf{g}_1\rangle = (1.0582,-0.1058,0.1694,0.1694)^T, \qquad f|\mathbf{x}_2\rangle = 6.28.$$

**Iteration 3.**

$$|\mathbf{g}_2\rangle = (28.09,-0.3475,0.7031,-28.08)^T,$$

$$\mathbf{H}_2 = \begin{pmatrix} 96.80 & 20 & 0 & -94.80 \\ 20 & 202.4 & -4.744 & 0 \\ 0 & -4.744 & 19.49 & -10 \\ -94.80 & 0 & -10 & 104.80 \end{pmatrix},$$

$$|\mathbf{x}_3\rangle = (0.7037,-0.0704,0.1121,0.1111)^T, \qquad f|\mathbf{x}_3\rangle = 1.24.$$

### 5.4.2. Marquardt Method and Modification of the Hessian

If the Hessian is not positive definite in any iteration of the Newton algorithm, it is forced to become positive definite in Step 2 of the algorithm. This modification of $\mathbf{H}_k = \nabla^2 f|\mathbf{x}_k\rangle$, can be accomplished in one of several ways. One approach is to replace the matrix $\mathbf{H}_k$ by the $n \times n$ identity matrix $\mathbf{I}_n$ whenever it becomes nonpositive definite [154]. Since $\mathbf{I}_n$ is positive definite, the problem of a nonsingular $\mathbf{H}_k$ is eliminated.

Another approach would be to let

| **Algorithm 5.13:** Marquardt | |
|---|---|
| Step 1: | Choose a starting point, $\lvert\mathbf{x}_0\rangle$, the maximum number of iterations, $M$, and a termination parameter, $\epsilon$. Set $k = 0$ and $\lambda^{(0)} = 10^4$ (a large number). |
| Step 2: | Calculate $\nabla f\lvert\mathbf{x}_k\rangle$ and $\mathbf{H}^{(k)}$. |
| Step 3: | If $\lVert\nabla f(\mathbf{x}_k)\rVert \le \epsilon$ or $k \ge M$? Terminate; Else go to Step 4. |
| Step 4: | Calculate $\mathbf{s}^{(k)} = -\left[\mathbf{H}^{(k)} + \lambda\mathbf{I}\right]^{-1}\nabla f\lvert\mathbf{x}_k\rangle$. Set $\lvert\mathbf{x}_{k+1}\rangle = \lvert\mathbf{x}_k\rangle + \lvert\mathbf{s}_k\rangle$. |
| Step 5: | Is $f\lvert\mathbf{x}_{k+1}\rangle < f\lvert\mathbf{x}_k\rangle$? If yes, go to Step 6; Else go to Step 7. |
| Step 6: | Set $\lambda^{(k+1)} = \frac{1}{2}\lambda^{(k)}$, $k = k + 1$, and go to Step 2. |
| Step 7: | Set $\lambda^{(k)} = 2\lambda^{(k)}$ and go to Step 4. |

$$\widehat{\mathbf{H}}_k = \frac{1}{1+\beta}(\mathbf{H}_k + \beta\mathbf{I}_n), \tag{5.56}$$

where $\beta$ is a positive scalar that is slightly larger than the absolute value of the most negative eigenvalue of $\mathbf{H}_k$ so as to assure the positive definiteness of $\mathbf{H}_k$ in (5.56). If $\beta$ is large, then

$$\widehat{\mathbf{H}}_k \approx \mathbf{I}_n, \tag{5.57}$$

and from (5.55)

$$\lvert\mathbf{d}_k\rangle \approx -\nabla f\lvert\mathbf{x}_k\rangle. \tag{5.58}$$

In effect, the modification in (5.56) converts the Newton method into the steepest-descent method. A nonpositive definite $\mathbf{H}_k$ is likely to arise at points far from the solution where the steepest-descent method is most effective in reducing the value of $f\lvert\mathbf{x}\rangle$. Therefore, the modification in (5.56) leads to an algorithm that combines the complementary convergence properties of the Newton and steepest-descent methods.

**Marquardt Method**

The steepest-descent method works well when the initial point is far away from the minimum point, and the Newton method works well when the initial point is near the minimum point. In any given problem, it is usually not known whether the chosen initial point is away from the minimum or close to the minimum, but wherever be the minimum point, a method can be devised to take advantage of both these methods.

The Marquardt method [46-48, 153] combines steepest-descent and Newton methods in a convenient manner that exploits the strengths of both but does require second-order information. Marquardt specified the search direction to be

$$\lvert\mathbf{s}_k\rangle = -\left[\mathbf{H}_f(\mathbf{x}_k) + \lambda_k\mathbf{I}\right]^{-1}\nabla f\lvert\mathbf{x}_k\rangle, \tag{5.59}$$

and set $\alpha_k = +1$ in $\lvert\mathbf{x}_{k+1}\rangle = \lvert\mathbf{x}_k\rangle + \alpha_k\lvert\mathbf{s}_k\rangle$, since $\lambda$ is used to control both the direction of the search and the length of the step. To begin the search, let $\lambda_0$ be a large constant, say $10^4$, such that

$$\left[\mathbf{H}_f(\mathbf{x}_0) + \lambda_0\mathbf{I}\right]^{-1} \cong [\lambda_0\mathbf{I}]^{-1} = \frac{\mathbf{I}}{\lambda_0}. \tag{5.60}$$

We notice from (5.59) that as $\lambda$ decreases from a large value to zero, $\lvert\mathbf{s}\rangle$ goes from the gradient to the Newton direction. Hence, in the Marquardt method, the Cauchy method is initially followed. Thereafter, the Newton method is adopted. The complete Marquardt method is presented in Algorithm 5.13.

The application of Newton's method for training large NNs is constrained by the considerable computational burden it imposes. The number of elements in the Hessian is squared in the number of parameters, so with $k$ parameters (where even relatively modest networks can boast millions of parameters), Newton's method would require the inversion of a $k \times k$ matrix—with a computational complexity of $O(k^3)$. Moreover, the dynamic nature of NN training, where parameters are continuously updated, compounds the computational challenge. At each iteration of training, the inverse Hessian must be recomputed, further amplifying the computational overhead. Consequently, the practical application of Newton's method is largely confined to networks with a very limited number of parameters. In the subsequent sections, we explore alternative methodologies that strive to harness some of the benefits offered by Newton's method while circumventing the computational obstacles it presents.

## 5.5 Conjugate Direction Method

### 5.5.1 Conjugate Direction for the Quadratic Function

The conjugate direction method is an optimization algorithm used to find the minimum of a function. The method is an extension of the steepest descent method, where instead of always moving in the direction of the negative gradient, it moves in a series of conjugate directions with respect to a certain matrix. A finite set of distinct nonzero vectors $\{|\mathbf{d}_0\rangle, |\mathbf{d}_1\rangle, \dots, |\mathbf{d}_m\rangle\}$ is said to be conjugate with respect to a real symmetric matrix $\mathbf{H}$, if

$$\langle \mathbf{d}_i|\mathbf{H}|\mathbf{d}_j\rangle = 0 \quad \text{for all } i \neq j. \tag{5.61}$$

The $\mathbf{H}$-conjugate directions $|\mathbf{d}_0\rangle, |\mathbf{d}_1\rangle, \dots, |\mathbf{d}_m\rangle$ form a set of linearly independent vectors. Moreover, if $\mathbf{H}$ is the symmetric positive definite matrix, then the $\mathbf{H}$-conjugate vectors $|\mathbf{d}_i\rangle, i = 0,1,2,\cdots,n-1$ form a basis for $\mathbb{R}^n$. To see this fact, let $\alpha_i$, $i = 0,1,\dots,k$ be scalar, such that,

$$\alpha_0|\mathbf{d}_0\rangle + \alpha_1|\mathbf{d}_1\rangle + \cdots + \alpha_k|\mathbf{d}_k\rangle = \sum_{i=0}^{k} \alpha_i|\mathbf{d}_i\rangle = |\mathbf{0}\rangle.$$

Hence

$$\sum_{i=0}^{k} \alpha_i\langle \mathbf{d}_j|\mathbf{H}|\mathbf{d}_i\rangle = \alpha_j\langle \mathbf{d}_j|\mathbf{H}|\mathbf{d}_j\rangle = \langle \mathbf{d}_j|\mathbf{H}|\mathbf{0}\rangle = 0.$$

Since $\mathbf{H}$ is a symmetric positive definite matrix and $|\mathbf{d}_j\rangle \neq |\mathbf{0}\rangle$, we get $\alpha_j = 0$, that is,

$$\alpha_0 = 0, \alpha_1 = 0, \dots, \alpha_k = 0.$$

As a result, the set of distinct nonzero vectors $\{|\mathbf{d}_0\rangle, |\mathbf{d}_1\rangle, \dots, |\mathbf{d}_k\rangle\}$ are linearly independent.

***Example 5.2***

Construct a set of $\mathbf{H}$-conjugate vectors $|\mathbf{d}_0\rangle, |\mathbf{d}_1\rangle, |\mathbf{d}_2\rangle$, where $|\mathbf{d}_0\rangle = (1,0,0)^T$ and the matrix $\mathbf{H} = \begin{pmatrix} 3 & 0 & 1 \\ 0 & 4 & 1 \\ 1 & 1 & 3 \end{pmatrix}$.

***Solution***

First, note that the matrix $\mathbf{H}$ is positive definite. Let $|\mathbf{d}_1\rangle = (d_{11}, d_{12}, d_{13})^T$, and $|\mathbf{d}_2\rangle = (d_{21}, d_{22}, d_{23})^T$. We require $\langle \mathbf{d}_0|\mathbf{H}|\mathbf{d}_1\rangle = 0$. We have

$$\langle \mathbf{d}_0|\mathbf{H}|\mathbf{d}_1\rangle = (1,0,0)\begin{pmatrix} 3 & 0 & 1 \\ 0 & 4 & 1 \\ 1 & 1 & 3 \end{pmatrix}\begin{pmatrix} d_{11} \\ d_{12} \\ d_{13} \end{pmatrix} = (3,0,1)\begin{pmatrix} d_{11} \\ d_{12} \\ d_{13} \end{pmatrix} = 3d_{11} + d_{13}.$$

Let $d_{11} = 1$, $d_{12} = 0$ and $d_{13} = -3$. Then $|\mathbf{d}_1\rangle = (1,0,-3)^T$, and thus $\langle \mathbf{d}_0|\mathbf{H}|\mathbf{d}_1\rangle = 0$.
To find the third vector $|\mathbf{d}_2\rangle = (d_{21}, d_{22}, d_{23})^T$, which would be $\mathbf{H}$-conjugate with $|\mathbf{d}_0\rangle$ and $|\mathbf{d}_1\rangle$, we require $\langle \mathbf{d}_0|\mathbf{H}|\mathbf{d}_2\rangle = 0$ and $\langle \mathbf{d}_1|\mathbf{H}|\mathbf{d}_2\rangle = 0$. We have

$$\langle \mathbf{d}_0|\mathbf{H}|\mathbf{d}_2\rangle = (1,0,0)\begin{pmatrix} 3 & 0 & 1 \\ 0 & 4 & 1 \\ 1 & 1 & 3 \end{pmatrix}\begin{pmatrix} d_{21} \\ d_{22} \\ d_{23} \end{pmatrix} = (3,0,1)\begin{pmatrix} d_{21} \\ d_{22} \\ d_{23} \end{pmatrix} = 3d_{21} + d_{23} = 0,$$

and

$$\langle \mathbf{d}_1|\mathbf{H}|\mathbf{d}_2\rangle = (1,0,-3)\begin{pmatrix} 3 & 0 & 1 \\ 0 & 4 & 1 \\ 1 & 1 & 3 \end{pmatrix}\begin{pmatrix} d_{21} \\ d_{22} \\ d_{23} \end{pmatrix} = (0,-3,-8)\begin{pmatrix} d_{21} \\ d_{22} \\ d_{23} \end{pmatrix} = -3d_{22} - 8d_{23} = 0.$$

If we take $|\mathbf{d}_2\rangle = (1,8,-3)^T$, then the resulting set of vectors is mutually conjugate.

**Algorithm 5.14:** Conjugate direction method for the quadratic function

| | |
|---|---|
| Step 1: | Given an initial point $|\mathbf{x}_0\rangle \in \mathbb{R}^n$ and a tolerance value $0 < \varepsilon < 1$. |
| Step 2: | Choose $\mathbf{H}$-conjugate directions $|\mathbf{d}_0\rangle, |\mathbf{d}_1\rangle, \dots, |\mathbf{d}_{n-1}\rangle$. |
| Step 3: | Set the iteration $k = 0$. |
| Step 4: | Compute $|\mathbf{g}_k\rangle$ the gradient of the function at $|\mathbf{x}_k\rangle$. |
| Step 5: | Compute $\alpha_k$ by the formula $\alpha_k = -\frac{\langle \mathbf{g}_k|\mathbf{d}_k\rangle}{\langle \mathbf{d}_k|\mathbf{H}|\mathbf{d}_k\rangle}$. |
| Step 6: | Compute a new point $|\mathbf{x}_{k+1}\rangle$ by the formula $|\mathbf{x}_{k+1}\rangle = |\mathbf{x}_k\rangle + \alpha_k|\mathbf{d}_k\rangle$. |
| Step 7: | If $\|\mathbf{g}_{k+1}\| < \varepsilon$, then $|\mathbf{x}_{k+1}\rangle = |\mathbf{x}^*\rangle$, which is the minimizer. |
| Step 8: | Set $k = k + 1$ and go to step 4. |

An equation of the form

$$f|\mathbf{x}\rangle = \sum_{i=1}^{n}\sum_{j=1}^{n} A_{ij}x_ix_j + \sum_{i=1}^{n} b_ix_i + c = 0, \tag{5.62}$$

where $A_{ij}$, $b_i$ and $c$ are real constants, is called a quadratic equation in $n$ variables $x_1, x_2, \dots, x_n$. In matrix form, it can be written as

$$f|\mathbf{x}\rangle = \langle \mathbf{x}|\mathbf{A}|\mathbf{x}\rangle + \langle \mathbf{b}|\mathbf{x}\rangle + c = 0, \tag{5.63}$$

where $\mathbf{A} = (A_{ij})$, $|\mathbf{x}\rangle = (x_1 \dots x_n)^T$ and $|\mathbf{b}\rangle = (b_1 \dots b_n)^T$ in $\mathbb{R}^n$. A linear form is a polynomial of degree 1 in $n$ variables $x_1, x_2, \dots, x_n$ of the form

$$\langle \mathbf{b}|\mathbf{x}\rangle = \sum_{i=1}^{n} b_ix_i, \tag{5.64}$$

where $|\mathbf{x}\rangle = (x_1 \dots x_n)^T$ and $|\mathbf{b}\rangle = (b_1 \dots b_n)^T$ in $\mathbb{R}^n$. A quadratic form is a (homogeneous) polynomial of degree 2 in $n$ variables $x_1, x_2, \dots, x_n$ of the form

$$q|\mathbf{x}\rangle = \langle \mathbf{x}|\mathbf{A}|\mathbf{x}\rangle = (x_1 \dots x_n)(A_{ij})\begin{pmatrix} x_1 \\ \vdots \\ x_n \end{pmatrix} = \sum_{i=1}^{n}\sum_{j=1}^{n} A_{ij}x_ix_j, \tag{5.65}$$

where $|\mathbf{x}\rangle = (x_1 \dots x_n)^T \in \mathbb{R}^n$ and $\mathbf{A} = (A_{ij})$ is a real $n \times n$ matrix.

Consider the quadratic objective function $f: \mathbb{R}^n \to \mathbb{R}$ defined by

$$f|\mathbf{x}\rangle = \frac{1}{2}\langle \mathbf{x}|\mathbf{H}|\mathbf{x}\rangle + \langle \mathbf{b}|\mathbf{x}\rangle + c, \tag{5.66}$$

where $\mathbf{H}$ is an $n \times n$ symmetric positive definite matrix, $|\mathbf{b}\rangle$ is an $n \times 1$ vector, and $c$ is a real number. The function $f$ has a gradient vector $|\mathbf{g}\rangle$ given by: $\mathbf{g}|\mathbf{x}\rangle = \nabla f|\mathbf{x}\rangle = \mathbf{H}|\mathbf{x}\rangle + |\mathbf{b}\rangle$. If $f$ has a strict global minimizer $|\mathbf{x}^*\rangle$ in $\mathbb{R}^n$, then $\mathbf{g}|\mathbf{x}^*\rangle = |\mathbf{0}\rangle$. For a quadratic function of $n$ variables, the conjugate direction method reaches the solution after $n$ steps, as in the following theorem. The complete conjugate direction method for the quadratic function is presented in Algorithm 5.14.

**Theorem 5.1 (Conjugate Direction Algorithm):** Let $\{|\mathbf{d}_i\rangle\}_{i=0}^{n-1}$ be a set of nonzero $\mathbf{H}$-conjugate directions. For any $|\mathbf{x}_0\rangle \in \mathbb{R}^n$ the sequence $\{|\mathbf{x}_k\rangle\}$ generated according to

$$|\mathbf{x}_{k+1}\rangle = |\mathbf{x}_k\rangle + \alpha_k|\mathbf{d}_k\rangle, \qquad k \geq 0, \tag{5.67}$$

with

$$\alpha_k = -\frac{\langle \mathbf{g}_k|\mathbf{d}_k\rangle}{\langle \mathbf{d}_k|\mathbf{H}|\mathbf{d}_k\rangle}, \tag{5.68}$$

and

$$|\mathbf{g}_k\rangle = \mathbf{H}|\mathbf{x}_k\rangle + |\mathbf{b}\rangle, \tag{5.69}$$

converges to a unique solution, $|\mathbf{x}^*\rangle$ after $n$ steps, that is, $|\mathbf{x}_n\rangle = |\mathbf{x}^*\rangle$.

**Proof:**

Since the $|\mathbf{d}_k\rangle$ are linearly independent, then there exist constants $\alpha_i$, $i = 0,1,\ldots,n-1$ such that,

$$|\mathbf{x}^*\rangle - |\mathbf{x}_0\rangle = \alpha_0|\mathbf{d}_0\rangle + \alpha_1|\mathbf{d}_1\rangle + \cdots + \alpha_{n-1}|\mathbf{d}_{n-1}\rangle = \sum_{i=0}^{n-1} \alpha_i|\mathbf{d}_i\rangle.$$

We multiply by $\mathbf{H}$ and take the scalar product with $|\mathbf{d}_k\rangle$ to find

$$\begin{aligned}\langle \mathbf{d}_k|\mathbf{H}|\mathbf{x}^* - \mathbf{x}_0\rangle &= \langle \mathbf{d}_k|\mathbf{H}\sum_{i=0}^{n-1} \alpha_i|\mathbf{d}_i\rangle \\ &= \sum_{i=0}^{n-1} \alpha_i\langle \mathbf{d}_k|\mathbf{H}|\mathbf{d}_i\rangle = \alpha_k\langle \mathbf{d}_k|\mathbf{H}|\mathbf{d}_k\rangle,\end{aligned}$$

or

$$\alpha_k = \frac{\langle \mathbf{d}_k|\mathbf{H}|\mathbf{x}^* - \mathbf{x}_0\rangle}{\langle \mathbf{d}_k|\mathbf{H}|\mathbf{d}_k\rangle}.$$

Using $|\mathbf{x}_{k+1}\rangle = |\mathbf{x}_k\rangle + \alpha_k|\mathbf{d}_k\rangle$, the iterative process from $|\mathbf{x}_0\rangle$ up to $|\mathbf{x}_k\rangle$ gives

$$\begin{aligned}&|\mathbf{x}_1\rangle = |\mathbf{x}_0\rangle + \alpha_0|\mathbf{d}_0\rangle,\\ &|\mathbf{x}_2\rangle = |\mathbf{x}_1\rangle + \alpha_1|\mathbf{d}_1\rangle = |\mathbf{x}_0\rangle + \alpha_0|\mathbf{d}_0\rangle + \alpha_1|\mathbf{d}_1\rangle,\\ &|\mathbf{x}_3\rangle = |\mathbf{x}_2\rangle + \alpha_2|\mathbf{d}_2\rangle = |\mathbf{x}_0\rangle + \alpha_0|\mathbf{d}_0\rangle + \alpha_1|\mathbf{d}_1\rangle + \alpha_2|\mathbf{d}_2\rangle,\\ &\ldots\ldots\ldots\\ &|\mathbf{x}_k\rangle = |\mathbf{x}_0\rangle + \sum_{i=0}^{k-1} \alpha_i|\mathbf{d}_i\rangle.\end{aligned}$$

Moreover, by the $\mathbf{H}$ -orthogonality of the $|\mathbf{d}_k\rangle$ it follows that

$$\langle \mathbf{d}_k|\mathbf{H}|\mathbf{x}_k - \mathbf{x}_0\rangle = \langle \mathbf{d}_k|\mathbf{H}\sum_{i=0}^{k-1} \alpha_i|\mathbf{d}_i\rangle = \sum_{i=0}^{k-1} \alpha_i\langle \mathbf{d}_k|\mathbf{H}|\mathbf{d}_i\rangle = 0,$$

and

$$\langle \mathbf{d}_k|\mathbf{H}|\mathbf{x}_k\rangle = \langle \mathbf{d}_k|\mathbf{H}|\mathbf{x}_0\rangle.$$

Hence, we get

$$\begin{aligned}\alpha_k &= \frac{\langle \mathbf{d}_k|\mathbf{H}|\mathbf{x}^* - \mathbf{x}_0\rangle}{\langle \mathbf{d}_k|\mathbf{H}|\mathbf{d}_k\rangle}\\ &= \frac{\langle \mathbf{d}_k|\mathbf{H}|\mathbf{x}^*\rangle - \langle \mathbf{d}_k|\mathbf{H}|\mathbf{x}_0\rangle}{\langle \mathbf{d}_k|\mathbf{H}|\mathbf{d}_k\rangle}\\ &= \frac{\langle \mathbf{d}_k|\mathbf{H}|\mathbf{x}^*\rangle - \langle \mathbf{d}_k|\mathbf{H}|\mathbf{x}_k\rangle}{\langle \mathbf{d}_k|\mathbf{H}|\mathbf{d}_k\rangle} = \frac{\langle \mathbf{d}_k|\mathbf{H}|\mathbf{x}^* - \mathbf{x}_k\rangle}{\langle \mathbf{d}_k|\mathbf{H}|\mathbf{d}_k\rangle}.\end{aligned}$$

From $|\mathbf{g}_k\rangle = \mathbf{H}|\mathbf{x}_k\rangle + |\mathbf{b}\rangle$, we have $\mathbf{H}|\mathbf{x}_k\rangle = |\mathbf{g}_k\rangle - |\mathbf{b}\rangle$, and since $|\mathbf{g}_k\rangle = |\mathbf{0}\rangle$ at minimizer $|\mathbf{x}^*\rangle$ , we get $\mathbf{H}|\mathbf{x}^*\rangle = -|\mathbf{b}\rangle$. Therefore,

$$\begin{aligned}\alpha_k &= \frac{\langle \mathbf{d}_k|\mathbf{H}|\mathbf{x}^* - \mathbf{x}_k\rangle}{\langle \mathbf{d}_k|\mathbf{H}|\mathbf{d}_k\rangle}\\ &= \frac{\langle \mathbf{d}_k|\mathbf{H}\mathbf{x}^* - \mathbf{H}\mathbf{x}_k\rangle}{\langle \mathbf{d}_k|\mathbf{H}|\mathbf{d}_k\rangle}\\ &= \frac{\langle \mathbf{d}_k|-\mathbf{b} - \mathbf{H}\mathbf{x}_k\rangle}{\langle \mathbf{d}_k|\mathbf{H}|\mathbf{d}_k\rangle} = -\frac{\langle \mathbf{d}_k|\mathbf{b} + \mathbf{H}\mathbf{x}_k\rangle}{\langle \mathbf{d}_k|\mathbf{H}|\mathbf{d}_k\rangle} = -\frac{\langle \mathbf{d}_k|\mathbf{g}_k\rangle}{\langle \mathbf{d}_k|\mathbf{H}|\mathbf{d}_k\rangle}.\end{aligned}$$

Finally, we have

$$|\mathbf{x}^*\rangle = |\mathbf{x}_0\rangle + \alpha_0|\mathbf{d}_0\rangle + \alpha_1|\mathbf{d}_1\rangle + \cdots + \alpha_{n-1}|\mathbf{d}_{n-1}\rangle$$

$$= |\mathbf{x}_0\rangle + \sum_{i=0}^{n-1} \alpha_i |\mathbf{d}_i\rangle$$
$$= |\mathbf{x}_0\rangle + |\mathbf{x}_n\rangle - |\mathbf{x}_0\rangle = |\mathbf{x}_n\rangle.$$

■

***Example 5.3***

Let $f: \mathbb{R}^2 \to \mathbb{R}$ be defined by

$$f|\mathbf{x}\rangle = \frac{1}{2}\langle \mathbf{x}|\mathbf{H}|\mathbf{x}\rangle + \langle \mathbf{b}|\mathbf{x}\rangle + c,$$

where

$$\mathbf{H} = \begin{pmatrix} 2 & 0 \\ 0 & 4 \end{pmatrix}, |\mathbf{b}\rangle = \begin{pmatrix} -1 \\ 1 \end{pmatrix}, c = 1, |\mathbf{x}_0\rangle = \begin{pmatrix} 0 \\ 0 \end{pmatrix},$$
$$|\mathbf{d}_0\rangle = \begin{pmatrix} 1 \\ 0 \end{pmatrix}, |\mathbf{d}_1\rangle = \begin{pmatrix} 0 \\ 1 \end{pmatrix}.$$

Verify that the iteration $|\mathbf{x}_2\rangle$ generated from the general conjugate direction method is the minimizer of $f$. Take $\varepsilon = 0.0001$.

***Solution***

We can write $f|\mathbf{x}\rangle$ as follows

$$f|\mathbf{x}\rangle = \frac{1}{2}(x_1, x_2)\begin{pmatrix} 2 & 0 \\ 0 & 4 \end{pmatrix}\begin{pmatrix} x_1 \\ x_2 \end{pmatrix} - (-1,1)\begin{pmatrix} x_1 \\ x_2 \end{pmatrix} + 1$$
$$= \frac{1}{2}(2x_1, 4x_2)\begin{pmatrix} x_1 \\ x_2 \end{pmatrix} + x_1 - x_2 + 1$$
$$= \frac{1}{2}(2x_1^2 + 4x_2^2) + x_1 - x_2 + 1$$
$$= x_1^2 + 2x_2^2 + x_1 - x_2 + 1.$$

Now, the gradient vector $\mathbf{g}|\mathbf{x}\rangle$ is

$$\mathbf{g}|\mathbf{x}\rangle = \left(\frac{\partial f}{\partial x_1}, \frac{\partial f}{\partial x_2}\right)^T = (2x_1 + 1, 4x_2 - 1)^T,$$

so that

$$\mathbf{g}_0 = \mathbf{g}|\mathbf{x}_0\rangle = (1, -1)^T.$$

The values of $\alpha_0$ and $|\mathbf{x}_1\rangle$ are

$$\alpha_0 = -\frac{\langle \mathbf{g}_0|\mathbf{d}_0\rangle}{\langle \mathbf{d}_0|\mathbf{H}|\mathbf{d}_0\rangle} = -\frac{(1,-1)\begin{pmatrix} 1 \\ 0 \end{pmatrix}}{(1,0)\begin{pmatrix} 2 & 1 \\ 1 & 2 \end{pmatrix}\begin{pmatrix} 1 \\ 0 \end{pmatrix}} = -\frac{1}{2},$$
$$|\mathbf{x}_1\rangle = |\mathbf{x}_0\rangle + \alpha_0|\mathbf{d}_0\rangle = \begin{pmatrix} 0 \\ 0 \end{pmatrix} - \frac{1}{2}\begin{pmatrix} 1 \\ 0 \end{pmatrix} = \begin{pmatrix} -\frac{1}{2} \\ 0 \end{pmatrix},$$

and therefore

$$\mathbf{g}_1 = \mathbf{g}|\mathbf{x}_1\rangle = (0, -1)^T \Rightarrow \|\mathbf{g}_1\| = 1 > \varepsilon.$$

Compute $\alpha_1$ and $|\mathbf{x}_2\rangle$ by

$$\alpha_1 = -\frac{\langle \mathbf{g}_1|\mathbf{d}_1\rangle}{\langle \mathbf{d}_1|\mathbf{H}|\mathbf{d}_1\rangle} = -\frac{(0,-1)\begin{pmatrix} 0 \\ 1 \end{pmatrix}}{(0,1)\begin{pmatrix} 2 & 0 \\ 0 & 4 \end{pmatrix}\begin{pmatrix} 0 \\ 1 \end{pmatrix}} = -\frac{-1}{4} = \frac{1}{4},$$
$$|\mathbf{x}_2\rangle = |\mathbf{x}_1\rangle + \alpha_1|\mathbf{d}_1\rangle = \begin{pmatrix} -\frac{1}{2} \\ 0 \end{pmatrix} + \frac{1}{4}\begin{pmatrix} 0 \\ 1 \end{pmatrix} = \begin{pmatrix} -\frac{1}{2} \\ 0 \end{pmatrix} + \begin{pmatrix} 0 \\ \frac{1}{4} \end{pmatrix} = \begin{pmatrix} -\frac{1}{2} \\ \frac{1}{4} \end{pmatrix},$$

and

$$\mathbf{g}_2 = \mathbf{g}|\mathbf{x}_2\rangle = \begin{pmatrix} 0 \\ 0 \end{pmatrix} \Rightarrow \|\mathbf{g}_2\| = 0 < \varepsilon.$$

Therefore, $|\mathbf{x}_2\rangle$ is the minimizer of $f$. Since $f$ is a quadratic function of two variables, therefore it converges to the minimum point$|\mathbf{x}_2\rangle$ and the minimum value of the objective function $f|\mathbf{x}^*\rangle = 5/8$ in two iterations.

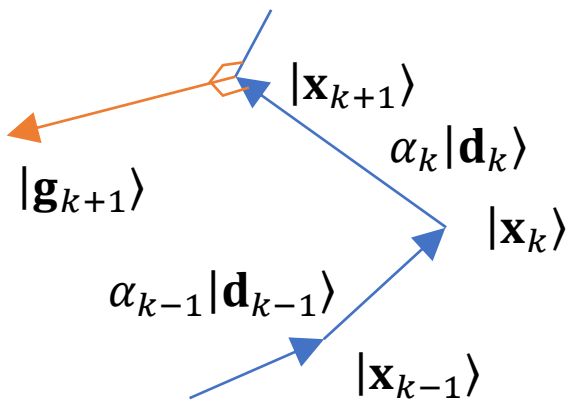

**Figure 5.14.** Illustration of Theorem 5.2.

**Algorithm 5.15:** Conjugate gradient algorithm for quadratic function.

| | |
|---|---|
| Step 1: | Choose a starting point $|\mathbf{x}_0\rangle \in \mathbb{R}^n$, tolerance value $0 < \epsilon < 1$, set $k = 0$. |
| Step 2: | Compute $|\mathbf{g}_0\rangle$, which is the gradient vector at the point $|\mathbf{x}_0\rangle$. If $|\mathbf{g}_0\rangle = |\mathbf{0}\rangle$ stop else go to step 3. |
| Step 3: | Set $|\mathbf{d}_0\rangle = -|\mathbf{g}_0\rangle$. |
| Step 4: | Compute $\mathbf{H}$, which is the Hessian matrix for the function $f|\mathbf{x}\rangle$ at $|\mathbf{x}_k\rangle$. |
| Step 5: | Compute $\alpha_k$ as: $\alpha_k = -\frac{\langle \mathbf{g}_k|\mathbf{d}_k\rangle}{\langle \mathbf{d}_k|\mathbf{H}|\mathbf{d}_k\rangle}$. |
| Step 6: | Compute $|\mathbf{x}_{k+1}\rangle$ as: $|\mathbf{x}_{k+1}\rangle = |\mathbf{x}_k\rangle + \alpha_k|\mathbf{d}_k\rangle$. |
| Step 7: | Compute $|\mathbf{g}_{k+1}\rangle$. If $\|\mathbf{g}_{k+1}\| < \varepsilon$ stop, else go to step 8. |
| Step 8: | Compute $\beta_k$ by $\beta_k = \frac{\langle \mathbf{g}_{k+1}|\mathbf{H}|\mathbf{d}_k\rangle}{\langle \mathbf{d}_k|\mathbf{H}|\mathbf{d}_k\rangle}$ ($\beta_k$ is chosen such that $|\mathbf{d}_k\rangle$ is $\mathbf{H}$-conjugate with $|\mathbf{d}_{k+1}\rangle$). |
| Step 9: | Compute $|\mathbf{d}_{k+1}\rangle = -|\mathbf{g}_{k+1}\rangle + \beta_k|\mathbf{d}_k\rangle$. |
| Step 10: | Set $k = k + 1$, and go to step 4. |

**Theorem 5.2 (Orthogonality of Gradient to a Set of Conjugate Directions):** In the conjugate direction algorithm,

$$\langle \mathbf{g}_{k+1}|\mathbf{d}_i\rangle = 0, \tag{5.70}$$

for all $k$, $0 \leq k \leq n-1$, and $0 \leq i \leq k$.

Theorem 5.2 says that under given conditions, the gradient vector at $|\mathbf{x}_{k+1}\rangle$ is orthogonal to all the preceding directions $|\mathbf{d}_i\rangle$, $i = 0,1,\cdots,k$. Figure 5.14 illustrates this statement.

### 5.5.2 Conjugate Gradient Method

The conjugate gradient method [155] is the modification of the steepest descent method. It uses conjugate directions to minimize a quadratic function, but the steepest descent method uses local gradients to minimize the quadratic function. The conjugate gradient method does not repeat any previous search directions and converges in $n$ iterations for $n$ variables of quadratic functions. The complete conjugate gradient method for the quadratic function is presented in Algorithm 5.15.

Consider the quadratic function $f(\mathbf{x}) = \frac{1}{2}\langle \mathbf{x}|\mathbf{H}|\mathbf{x}\rangle - \langle \mathbf{x}|\mathbf{b}\rangle$ where $\mathbf{H}$ is an $n \times n$ symmetric positive definite matrix, and $|\mathbf{b}\rangle$ is an $n \times 1$ vector. Our first search direction from an initial point $|\mathbf{x}_0\rangle$ is in the direction of the steepest descent; that is $|\mathbf{d}_0\rangle = -|\mathbf{g}_0\rangle$. Thus

$$|\mathbf{x}_1\rangle = |\mathbf{x}_0\rangle + \alpha_0|\mathbf{d}_0\rangle, \tag{5.71}$$

where $\alpha_0 = \min_{\alpha \geq 0} f(|\mathbf{x}_0\rangle + \alpha|\mathbf{d}_0\rangle) = -\frac{\langle \mathbf{g}_0|\mathbf{d}_0\rangle}{\langle \mathbf{d}_0|\mathbf{H}|\mathbf{d}_0\rangle}$. In the next step, we search for a direction $|\mathbf{d}_1\rangle$ that is $\mathbf{H}$-conjugate to $|\mathbf{d}_0\rangle$. We choose $|\mathbf{d}_1\rangle$ as a linear combination of $|\mathbf{g}_1\rangle$ and $|\mathbf{d}_0\rangle$ that can be expressed as $|\mathbf{d}_1\rangle = -|\mathbf{g}_1\rangle + \beta_0|\mathbf{d}_0\rangle$.

In general, at the $(k+1)$th step, we have $|\mathbf{x}_{k+1}\rangle = |\mathbf{x}_k\rangle + \alpha_k|\mathbf{d}_k\rangle$, where

$$\alpha_k = -\frac{\langle \mathbf{g}_k|\mathbf{d}_k\rangle}{\langle \mathbf{d}_k|\mathbf{H}|\mathbf{d}_k\rangle}, \tag{5.72}$$

and we choose $|\mathbf{d}_{k+1}\rangle$ to be a linear combination of $|\mathbf{g}_{k+1}\rangle$ and $|\mathbf{d}_k\rangle$, which can be presented

$$|\mathbf{d}_{k+1}\rangle = -|\mathbf{g}_{k+1}\rangle + \beta_k|\mathbf{d}_k\rangle, \quad k = 0,1,2,\cdots. \tag{5.73}$$

The coefficient $\beta_k$, $k = 0,1,\cdots$, are chosen in such a way that $|\mathbf{d}_{k+1}\rangle$ is $\mathbf{H}$-conjugate to $|\mathbf{d}_0\rangle, |\mathbf{d}_1\rangle, \cdots, |\mathbf{d}_k\rangle$. This is accomplished by choosing $\beta_k$ as follows:

$$\beta_k = \frac{\langle \mathbf{g}_{k+1}|\mathbf{H}|\mathbf{d}_k\rangle}{\langle \mathbf{d}_k|\mathbf{H}|\mathbf{d}_k\rangle}. \tag{5.74}$$

***Example 5.4***

Find the minimizer of the quadratic function

$$f|\mathbf{x}\rangle = \frac{1}{2}\langle \mathbf{x}|\mathbf{H}|\mathbf{x}\rangle - \langle \mathbf{x}|\mathbf{b}\rangle,$$

where

$$\mathbf{H} = \begin{pmatrix} 3 & 0 & \sqrt{3} \\ 0 & 4 & 2 \\ \sqrt{3} & 2 & 3 \end{pmatrix} \text{ and } \boldsymbol{b} = \begin{pmatrix} 2 \\ 0 \\ 1 \end{pmatrix},$$

by using the conjugate gradient algorithm, where $|\mathbf{x}_0\rangle = (0,0,0)^T$, and $\varepsilon = 0.0001$.

***Solution***

1. Find the gradient vector of the function $f|\mathbf{x}\rangle$:

$$\mathbf{g}|\mathbf{x}\rangle = \mathbf{H}|\mathbf{x}\rangle - \mathbf{b} = \begin{pmatrix} 3 & 0 & \sqrt{3} \\ 0 & 4 & 2 \\ \sqrt{3} & 2 & 3 \end{pmatrix}\begin{pmatrix} x_1 \\ x_2 \\ x_3 \end{pmatrix} - \begin{pmatrix} 2 \\ 0 \\ 1 \end{pmatrix} = \begin{pmatrix} 3x_1 + \sqrt{3}x_3 \\ 4x_2 + 2x_3 \\ \sqrt{3}x_1 + 2x_2 + 3x_3 \end{pmatrix} - \begin{pmatrix} 2 \\ 0 \\ 1 \end{pmatrix} = \begin{pmatrix} 3x_1 + \sqrt{3}x_3 - 2 \\ 4x_2 + 2x_3 \\ \sqrt{3}x_1 + 2x_2 + 3x_3 - 1 \end{pmatrix}.$$

2. Find $|\mathbf{g_0}\rangle$: $|\mathbf{g_0}\rangle = \mathbf{g}|\mathbf{x}_0\rangle = (-2,0,-1)^T$
3. Find $|\mathbf{d_0}\rangle$: $|\mathbf{d_0}\rangle = -|\mathbf{g_0}\rangle = (2,0,1)^T$
4. Find $\alpha_0$:

$$\alpha_0 = -\frac{\langle \mathbf{g_0}|\mathbf{d}_0\rangle}{\langle \mathbf{d}_0|\mathbf{H}|\mathbf{d}_0\rangle} = -\frac{(-2,0,-1)\begin{pmatrix} 2 \\ 0 \\ 1 \end{pmatrix}}{(2,0,1)\begin{pmatrix} 3 & 0 & \sqrt{3} \\ 0 & 4 & 2 \\ \sqrt{3} & 2 & 3 \end{pmatrix}\begin{pmatrix} 2 \\ 0 \\ 1 \end{pmatrix}} = -\frac{-5}{15 + 4\sqrt{3}} = 0.228017.$$

5. Find $|\mathbf{x}_1\rangle$:

$$|\mathbf{x}_1\rangle = |\mathbf{x}_0\rangle + \alpha_0|\mathbf{d}_0\rangle = \begin{pmatrix} 0 \\ 0 \\ 0 \end{pmatrix} + 0.228017\begin{pmatrix} 2 \\ 0 \\ 1 \end{pmatrix} = \begin{pmatrix} 0.456034 \\ 0 \\ 0.228017 \end{pmatrix}.$$

6. Find $|\mathbf{g_1}\rangle$: $|\mathbf{g_1}\rangle = \begin{pmatrix} -0.237053 \\ 0.456034 \\ 0.473872 \end{pmatrix}$ where $\|\mathbf{g_1}\| = 0.699082 > \varepsilon$.

7. Find $\beta_0$:

$$\beta_0 = \frac{\langle \mathbf{g}_1|\mathbf{H}|\mathbf{d}_0\rangle}{\langle \mathbf{d}_0|\mathbf{H}|\mathbf{d}_0\rangle} = \frac{(-0.237053,0.456034,0.473872)\begin{pmatrix} 3 & 0 & \sqrt{3} \\ 0 & 4 & 2 \\ \sqrt{3} & 2 & 3 \end{pmatrix}\begin{pmatrix} 2 \\ 0 \\ 1 \end{pmatrix}}{(2,0,1)\begin{pmatrix} 3 & 0 & \sqrt{3} \\ 0 & 4 & 2 \\ \sqrt{3} & 2 & 3 \end{pmatrix}\begin{pmatrix} 2 \\ 0 \\ 1 \end{pmatrix}} = \frac{2.14232}{15 + 4\sqrt{3}} = 0.097697.$$

8. Find $\boldsymbol{d}^{(1)}$:

$$|\mathbf{d}_1\rangle = -|\mathbf{g}_1\rangle + \beta_0|\mathbf{d}_0\rangle = \begin{pmatrix} 0.237053 \\ -0.456034 \\ -0.473872 \end{pmatrix} + 0.097697\begin{pmatrix} 2 \\ 0 \\ 1 \end{pmatrix} = \begin{pmatrix} 0.237053 \\ -0.456034 \\ -0.473872 \end{pmatrix} + \begin{pmatrix} 0.195394 \\ 0 \\ 0.097697 \end{pmatrix} = \begin{pmatrix} 0.432447 \\ -0.456034 \\ -0.376175 \end{pmatrix}.$$

9. Find $\alpha_1$:

$$\alpha_1 = -\frac{\langle \mathbf{g}_1|\mathbf{d}_1\rangle}{\langle \mathbf{d}_1|\mathbf{H}|\mathbf{d}_1\rangle} = -\frac{(-0.2371, 0.4560, 0.4739)\begin{pmatrix}0.4325\\-0.4560\\-0.3762\end{pmatrix}}{(0.4325, -0.4560, -0.3762)\begin{pmatrix}3 & 0 & \sqrt{3}\\0 & 4 & 2\\\sqrt{3} & 2 & 3\end{pmatrix}\begin{pmatrix}0.4325\\-0.4560\\-0.3762\end{pmatrix}} = -\frac{-0.4887}{1.9401} = 0.2519.$$

10. Find $\boldsymbol{x}^{(2)}$:

$$|\mathbf{x}_2\rangle = |\mathbf{x}_1\rangle + \alpha_1|\mathbf{d}_1\rangle = \begin{pmatrix}0.4560\\0\\0.2280\end{pmatrix} + 0.2519\begin{pmatrix}0.4325\\-0.4560\\-0.3762\end{pmatrix} = \begin{pmatrix}0.4560\\0\\0.2280\end{pmatrix} + \begin{pmatrix}0.1089\\-0.1149\\-0.0948\end{pmatrix} = \begin{pmatrix}0.5649\\-0.1149\\0.1332\end{pmatrix}.$$

11. Find $|\mathbf{g}_2\rangle$:

$$|\mathbf{g}_2\rangle = \begin{pmatrix}-0.0743\\-0.1930\\0.1486\end{pmatrix}, \text{ with } \|\mathbf{g}_2\| = 0.2547 > \varepsilon.$$

12. Find $\beta_1$:

$$\beta_1 = \frac{\langle \mathbf{g}_2|\mathbf{H}|\mathbf{d}_1\rangle}{\langle \mathbf{d}_1|\mathbf{H}|\mathbf{d}_1\rangle} = \frac{(-0.0743, -0.1930, 0.1486)\begin{pmatrix}3 & 0 & \sqrt{3}\\0 & 4 & 2\\\sqrt{3} & 2 & 3\end{pmatrix}\begin{pmatrix}0.4325\\-0.4560\\-0.3762\end{pmatrix}}{(0.4325, -0.4560, -0.3762)\begin{pmatrix}3 & 0 & \sqrt{3}\\0 & 4 & 2\\\sqrt{3} & 2 & 3\end{pmatrix}\begin{pmatrix}0.4325\\-0.4560\\-0.3762\end{pmatrix}} = \frac{0.2575}{1.9401} = 0.1327.$$

13. Find $\boldsymbol{d}^{(2)}$:

$$|\mathbf{d}_2\rangle = -|\mathbf{g}_2\rangle + \beta_1|\mathbf{d}_1\rangle = \begin{pmatrix}0.0743\\0.1930\\-0.1486\end{pmatrix} + 0.1327\begin{pmatrix}0.4325\\-0.4560\\-0.3762\end{pmatrix} = \begin{pmatrix}0.0743\\0.1930\\-0.1486\end{pmatrix} + \begin{pmatrix}0.0574\\-0.0605\\-0.0499\end{pmatrix} = \begin{pmatrix}0.1317\\0.1325\\-0.1985\end{pmatrix}.$$

14. Find $\alpha_2$:

$$\alpha_2 = -\frac{\langle \mathbf{g}_2|\mathbf{d}_2\rangle}{\langle \mathbf{d}_2|\mathbf{H}|\mathbf{d}_2\rangle} = -\frac{(-0.0743, -0.1930, 0.1486)\begin{pmatrix}0.1317\\0.1325\\-0.1985\end{pmatrix}}{(0.1317, 0.1325, -0.1985)\begin{pmatrix}3 & 0 & \sqrt{3}\\0 & 4 & 2\\\sqrt{3} & 2 & 3\end{pmatrix}\begin{pmatrix}0.1317\\0.1325\\-0.1985\end{pmatrix}} = -\frac{-0.0648}{0.0447} = 1.4508.$$

15. Find $\boldsymbol{x}^{(3)}$:

$$|\mathbf{x}_3\rangle = |\mathbf{x}_2\rangle + \alpha_2|\mathbf{d}_2\rangle = \begin{pmatrix}0.5649\\-0.1149\\0.1333\end{pmatrix} + 1.4508\begin{pmatrix}0.1317\\0.1325\\-0.1985\end{pmatrix} = \begin{pmatrix}0.5649\\-0.1149\\0.1333\end{pmatrix} + \begin{pmatrix}0.1910\\0.1922\\-0.2880\end{pmatrix} = \begin{pmatrix}0.7560\\0.0774\\-0.1547\end{pmatrix}.$$

Note that

$$|\mathbf{g}_3\rangle = \begin{pmatrix}0.00004\\-0.00002\\0.00001\end{pmatrix}, \text{ with } \|\mathbf{g}_3\| = 0.000042 < \varepsilon.$$

**Theorem 5.3 (Convergence of Conjugate Gradient Method):** (a) If $\mathbf{H}$ is a positive definite matrix, then for any initial point $|\mathbf{x}_0\rangle$ and an initial direction

$$|\mathbf{d}_0\rangle = -|\mathbf{g}_0\rangle = -|\mathbf{b} + \mathbf{H}\mathbf{x}_0\rangle, \tag{5.75}$$

the sequence generated by the recursive relation

$$|\mathbf{x}_{k+1}\rangle = |\mathbf{x}_k\rangle + \alpha_k|\mathbf{d}_k\rangle, \tag{5.76}$$

where

$$\alpha_k = -\frac{\langle \mathbf{g}_k|\mathbf{d}_k\rangle}{\langle \mathbf{d}_k|\mathbf{H}|\mathbf{d}_k\rangle}, \tag{5.77.1}$$

$$|\mathbf{g}_k\rangle = |\mathbf{b} + \mathbf{H}\mathbf{x}_k\rangle, \tag{5.77.2}$$

$$|\mathbf{d}_{k+1}\rangle = -|\mathbf{g}_{k+1}\rangle + \beta_k|\mathbf{d}_k\rangle, \tag{5.77.3}$$

$$\beta_k = \frac{\langle \mathbf{g}_{k+1}|\mathbf{H}|\mathbf{d}_k\rangle}{\langle \mathbf{d}_k|\mathbf{H}|\mathbf{d}_k\rangle}, \quad (5.77.4)$$

converges to a unique solution $|\mathbf{x}^*\rangle$ for the minimization of the quadratic function $f|\mathbf{x}\rangle = a + \langle \mathbf{b}|\mathbf{x}\rangle + \frac{1}{2}\langle \mathbf{x}|\mathbf{H}|\mathbf{x}\rangle$.
(b) The gradient $|\mathbf{g}_k\rangle$ is orthogonal to $\{|\mathbf{g}_0\rangle, |\mathbf{g}_1\rangle, \dots, |\mathbf{g}_{k-1}\rangle\}$, i.e.,

$$\langle \mathbf{g}_k|\mathbf{g}_i\rangle = 0, \quad 0 \le i < k. \quad (5.78)$$

**Theorem 5.4:** In the conjugate gradient algorithm, the directions $|\mathbf{d}_0\rangle, |\mathbf{d}_1\rangle, \dots, |\mathbf{d}_{n-1}\rangle$ are **H**-conjugate.

**Remarks:**

1- The expression for $\alpha_k$ in Theorem 5.3 can be simplified as follows. From (5.77.3)

$$-\langle \mathbf{g}_k|\mathbf{d}_k\rangle = \langle \mathbf{g}_k|\mathbf{g}_k\rangle - \beta_{k-1}\langle \mathbf{g}_k|\mathbf{d}_{k-1}\rangle, \quad (5.79)$$

where

$$\langle \mathbf{g}_k|\mathbf{d}_{k-1}\rangle = 0, \quad (5.80)$$

according to Theorem 5.2. Hence

$$-\langle \mathbf{g}_k|\mathbf{d}_k\rangle = \langle \mathbf{g}_k|\mathbf{g}_k\rangle, \quad (5.81)$$

and, therefore, the expression for $\alpha_k$ in (5.77.1) is modified as

$$\alpha_k = \frac{\langle \mathbf{g}_k|\mathbf{g}_k\rangle}{\langle \mathbf{d}_k|\mathbf{H}|\mathbf{d}_k\rangle}. \quad (5.82)$$

2- The expression for $\beta_k$ in Theorem 5.3 can also be simplified as follows. We have,

$$\mathbf{H}|\mathbf{d}_k\rangle = \frac{1}{\alpha_k}|\mathbf{g}_{k+1} - \mathbf{g}_k\rangle, \quad (5.83)$$

and so

$$\langle \mathbf{g}_{k+1}|\mathbf{H}|\mathbf{d}_k\rangle = \frac{1}{\alpha_k}(\langle \mathbf{g}_{k+1}|\mathbf{g}_{k+1}\rangle - \langle \mathbf{g}_{k+1}|\mathbf{g}_k\rangle). \quad (5.84)$$

Let $S(|\mathbf{d}_1\rangle, |\mathbf{d}_2\rangle, \dots, |\mathbf{d}_k\rangle)$ be the subspace spanned by $|\mathbf{d}_1\rangle, |\mathbf{d}_2\rangle, \dots, |\mathbf{d}_k\rangle$, we can prove that

$$|\mathbf{g}_k\rangle \in S(|\mathbf{d}_1\rangle, |\mathbf{d}_2\rangle, \dots, |\mathbf{d}_k\rangle), \quad (5.85)$$

or

$$|\mathbf{g}_k\rangle = \sum_{i=0}^{k} a_i|\mathbf{d}_i\rangle, \quad (5.86)$$

and as a result

$$\langle \mathbf{g}_{k+1}|\mathbf{g}_k\rangle = \sum_{i=0}^{k} a_i\langle \mathbf{g}_{k+1}|\mathbf{d}_i\rangle = 0, \quad (5.87)$$

by virtue of Theorem 5.2. Therefore,

$$\beta_k = \frac{\langle \mathbf{g}_{k+1}|\mathbf{H}|\mathbf{d}_k\rangle}{\langle \mathbf{d}_k|\mathbf{H}|\mathbf{d}_k\rangle} = \frac{1}{\alpha_k}\langle \mathbf{g}_{k+1}|\mathbf{g}_{k+1}\rangle \frac{\alpha_k}{\langle \mathbf{g}_k|\mathbf{g}_k\rangle} = \frac{\langle \mathbf{g}_{k+1}|\mathbf{g}_{k+1}\rangle}{\langle \mathbf{g}_k|\mathbf{g}_k\rangle}. \quad (5.88)$$

3- Hence, the iterative scheme of the conjugate gradient method is

$$|\mathbf{x}_{k+1}\rangle = |\mathbf{x}_k\rangle + \alpha_k|\mathbf{d}_k\rangle, \quad (5.89.1)$$

$$\alpha_k = \frac{\langle \mathbf{g}_k|\mathbf{g}_k\rangle}{\langle \mathbf{d}_k|\mathbf{H}|\mathbf{d}_k\rangle}, \quad (5.89.2)$$

$$|\mathbf{d}_{k+1}\rangle = -|\mathbf{g}_{k+1}\rangle + \beta_k|\mathbf{d}_k\rangle, \quad (5.89.3)$$

$$\beta_k = \frac{\langle \mathbf{g}_{k+1}|\mathbf{g}_{k+1}\rangle}{\langle \mathbf{g}_k|\mathbf{g}_k\rangle}. \quad (5.89.4)$$

### 5.5.3 Method of Conjugate for General Functions

The conjugate gradient algorithm can be extended to general nonlinear functions [157] by interpreting $f|\mathbf{x}\rangle = \frac{1}{2}\langle \mathbf{x}|\mathbf{H}|\mathbf{x}\rangle - \langle \mathbf{x}|\mathbf{b}\rangle$ as a second-order Taylor series approximation of the objective function. Near the solution,

| **Algorithm 5.16:** Hestenes-Stiefel, Polak-Ribiere, or Fletcher-Reeves Algorithm for General Nonlinear Functions | |
|---|---|
| Step 1: | Choose a starting point $\|\mathbf{x}_0\rangle \in \mathbb{R}^n$ and a tolerance $0 < \varepsilon < 1$. Set $k = 0$. |
| Step 2: | Compute the gradient $\|\mathbf{g}_0\rangle$ of $f\|\mathbf{x}\rangle$ at $\|\mathbf{x}_0\rangle$. |
| Step 3: | Set $\|\mathbf{d}_0\rangle = -\|\mathbf{g}_0\rangle$. |
| Step 4: | Compute $\alpha_k$ as: $\alpha_k$ by using line search $\alpha_k = \min_{\alpha\geq 0} f\|\mathbf{x}_k + \alpha\mathbf{d}_k\rangle$. |
| Step 5: | Compute $\|\mathbf{x}_{k+1}\rangle$ as: $\|\mathbf{x}_{k+1}\rangle = \|\mathbf{x}_k\rangle + \alpha_k\|\mathbf{d}_k\rangle$. |
| Step 6: | Compute $\|\mathbf{g}_{k+1}\rangle$. If $\\|\mathbf{g}_{k+1}\\| < \varepsilon$ stop, else go to the next step. |
| Step 7: | Compute $\beta_k$, (5.92), (5.95), or (5.97). |
| Step 8: | Compute $\|\mathbf{d}_{k+1}\rangle$ as: $\|\mathbf{d}_{k+1}\rangle = -\|\mathbf{g}_{k+1}\rangle + \beta_k\|\mathbf{d}_k\rangle$. |
| Step 9: | Set $k = k + 1$ and go to step 4. |

such functions behave approximately as quadratics. For a quadratic, the matrix **H**, the Hessian of the quadratic, is constant. However, for a general nonlinear function, the Hessian is a matrix that has to be reevaluated at each iteration of the algorithm. This can be computationally very expensive. Thus, an efficient implementation of the conjugate gradient algorithm that eliminates the Hessian evaluation at each step is desirable.

Observe that **H** appears only in the computation of the scalars $\alpha_k$ and $\beta_k$. Because $\alpha_k = \min_{\alpha\geq 0} f(|\mathbf{x}_k\rangle + \alpha|\mathbf{d}_k\rangle)$ the closed-form formula for $\alpha_k$ in the algorithm can be replaced by a numerical line search procedure. Therefore, we only need to concern ourselves with the formula for $\beta_k$. Here, we will discuss three well-known formulas for $\beta_k$.

### The Hestenes-Stiefel Formula

Recall that $\beta_k = \frac{\langle \mathbf{g}_{k+1}|\mathbf{H}|\mathbf{d}_k\rangle}{\langle \mathbf{d}_k|\mathbf{H}|\mathbf{d}_k\rangle}$. We have, $|\mathbf{x}_{k+1}\rangle = |\mathbf{x}_k\rangle + \alpha_k|\mathbf{d}_k\rangle$. Then, multiply both sides by **H** yields

$$\mathbf{H}|\mathbf{x}_{k+1}\rangle = \mathbf{H}|\mathbf{x}_k\rangle + \alpha_k\mathbf{H}|\mathbf{d}_k\rangle, \tag{5.90}$$

which equivalent to

$$\mathbf{H}|\mathbf{x}_{k+1}\rangle - |\mathbf{b}\rangle = \mathbf{H}|\mathbf{x}_k\rangle - |\mathbf{b}\rangle + \alpha_k\mathbf{H}|\mathbf{d}_k\rangle. \tag{5.91}$$

Since $|\mathbf{g}_k\rangle = \mathbf{H}|\mathbf{x}_k\rangle - |\mathbf{b}\rangle$ and $|\mathbf{g}_{k+1}\rangle = \mathbf{H}|\mathbf{x}_{k+1}\rangle - |\mathbf{b}\rangle$, we get $|\mathbf{g}_{k+1}\rangle = |\mathbf{g}_k\rangle + \alpha_k\mathbf{H}|\mathbf{d}_k\rangle$. Then $\mathbf{H}|\mathbf{d}_k\rangle = \frac{1}{\alpha_k}|\mathbf{g}_{k+1} - \mathbf{g}_k\rangle$. Substituting this into the original equation for $\beta_k$ gives

$$\beta_k = \frac{\langle \mathbf{g}_{k+1}|\mathbf{g}_{k+1} - \mathbf{g}_k\rangle}{\langle \mathbf{d}_k|\mathbf{g}_{k+1} - \mathbf{g}_k\rangle}, \tag{5.92}$$

which is called the Hestenes-Stiefel formula [157]. The complete Hestenes-Stiefel method is presented in Algorithm 5.16.

### The Polak-Ribiere Formula

Starting from Hestenes-Stiefel formula

$$\beta_k = \frac{\langle \mathbf{g}_{k+1}|\mathbf{g}_{k+1} - \mathbf{g}_k\rangle}{\langle \mathbf{d}_k|\mathbf{g}_{k+1}\rangle - \langle \mathbf{d}_k|\mathbf{g}_k\rangle}. \tag{5.93}$$

Since $\langle \mathbf{d}_k|\mathbf{g}_{k+1}\rangle = 0$ and $|\mathbf{d}_k\rangle = -|\mathbf{g}_k\rangle + \beta_{k-1}|\mathbf{d}_{k-1}\rangle$. Then, multiplying both sides by $\langle \mathbf{g}_k|$ implies

$$\begin{aligned}\langle \mathbf{g}_k|\mathbf{d}_k\rangle &= -\langle \mathbf{g}_k|\mathbf{g}_k\rangle + \beta_{k-1}\langle \mathbf{g}_k|\mathbf{d}_{k-1}\rangle \\ &= -\langle \mathbf{g}_k|\mathbf{g}_k\rangle,\end{aligned} \tag{5.94}$$

and thus, the expression for $\beta_k$ becomes

$$\beta_k = \frac{\langle \mathbf{g}_{k+1}|\mathbf{g}_{k+1} - \mathbf{g}_k\rangle}{\langle \mathbf{g}_k|\mathbf{g}_k\rangle}, \tag{5.95}$$

which is known as the Polak-Ribiere formula.

### The Fletcher-Reeves Formula

Starting with Polak-Ribiere formula

$$\beta_k = \frac{\langle \mathbf{g}_{k+1}|\mathbf{g}_{k+1}\rangle - \langle \mathbf{g}_{k+1}|\mathbf{g}_k\rangle}{\langle \mathbf{g}_k|\mathbf{g}_k\rangle}. \tag{5.96}$$

Since $\langle \mathbf{g}_{k+1}|\mathbf{g}_k\rangle = 0$, so that

$$\beta_k = \frac{\langle \mathbf{g}_{k+1}|\mathbf{g}_{k+1}\rangle}{\langle \mathbf{g}_k|\mathbf{g}_k\rangle}, \tag{5.97}$$

which is called the Fletcher-Reeves formula.

Practitioners have consistently reported favorable outcomes when employing the nonlinear conjugate gradients algorithm for training NNs. However, it's commonly observed that initializing the optimization process with a few iterations of SGD before transitioning to nonlinear conjugate gradients can prove beneficial. This hybrid approach leverages the advantages of both methods, enhancing convergence and efficiency. Moreover, although the (nonlinear) conjugate gradients algorithm has conventionally been conceptualized as a batch method, minibatch versions have been used successfully for training NNs.

## 5.6 The Basic Quasi-Newton Approach

In Section 5.5, multidimensional optimization methods were considered, in which the search for the minimizer is carried out by using a set of conjugate directions. An important feature of some of these methods is that explicit expressions for the second derivatives of the objective function $f|\mathbf{x}\rangle$ are not required. Another class of methods that do not require explicit expressions for the second derivatives is the class of quasi-Newton methods [29,46-48,153]. Quasi-Newton methods rank among the most efficient methods available and are used very extensively in numerous applications. Quasi-Newton methods, like most other methods, are developed for the convex quadratic problem and are then extended to the general problem. The foundation of these methods is the classical Newton method described in Section 5.4. In the Newton method, we are required to compute the inverse of the Hessian at every iteration, which is a very expensive computation. Furthermore, Newton's method fails to converge to the minimizer $|\mathbf{x}^*\rangle$ for the following cases.

1. The search direction $|\mathbf{d}_k\rangle$ is orthogonal to the gradient vector $|\mathbf{g}_k\rangle$.
2. The inverse of the Hessian matrix exists, and it is not positive definite.
3. The search direction $|\mathbf{d}_k\rangle$ is so large that $f|\mathbf{x}_{k+1}\rangle > f|\mathbf{x}_k\rangle$.
4. The inverse of the Hessian matrix does not exist.

These drawbacks of Newton's method gave the motivation to develop the quasi-Newton methods. The basic principle in quasi-Newton methods is that the direction of search is based on an $n \times n$ direction matrix $\mathbf{S}$ which serves the same purpose as the inverse Hessian in the Newton method.

Recall that the idea behind Newton's method is to locally approximate the function $f$ being minimized, at every iteration, by a quadratic function. The basic recursive formula representing Newton's method is

$$|\mathbf{x}_{k+1}\rangle = |\mathbf{x}_k\rangle - \mathbf{H}_k^{-1}|\mathbf{g}_k\rangle. \tag{5.98}$$

However, if the initial point is not sufficiently close to the solution, then the algorithm may not possess the descent property (i.e., $f|\mathbf{x}_{k+1}\rangle \nless f|\mathbf{x}_k\rangle$ for some $k$). To avoid this and guarantee that the algorithm has descent property, we modify the original algorithm as follows:

$$|\mathbf{x}_{k+1}\rangle = |\mathbf{x}_k\rangle - \alpha_k \mathbf{H}_k^{-1}|\mathbf{g}_k\rangle, \tag{5.99}$$

where $\alpha_k$ is chosen to ensure that $f|\mathbf{x}_{k+1}\rangle < f|\mathbf{x}_k\rangle$. For example, we may choose $\alpha_k = \arg\min_{\alpha\geq 0} f|\mathbf{x}_k - \alpha \mathbf{H}_k^{-1}\mathbf{g}_k\rangle$. We can then determine an appropriate value of $\alpha_k$ by performing a line search in the direction $-\mathbf{H}_k^{-1}|\mathbf{g}_k\rangle$. Note that although the line search is simply the minimization of the real variable function $\phi_k(\alpha) = f(|\mathbf{x}_k\rangle - \alpha \mathbf{H}_k^{-1}|\mathbf{g}_k\rangle)$, it is not a trivial problem to solve because we need to evaluate $\mathbf{H}_k$ and solve the equation $\mathbf{H}_k|\mathbf{d}_k\rangle = -|\mathbf{g}_k\rangle$.

***Example 5.5***

Let $f: \mathbb{R}^2 \to \mathbb{R}$ be a function defined by

$$f|\mathbf{x}\rangle = x_1 - x_2 + 2x_1^2 + 2x_1x_2 + x_2^2, \qquad |\mathbf{x}\rangle = (x_1, x_2)^T.$$

Use Newton method with line search to find the minimizer of the objective function $f$. Take $|\mathbf{x}_0\rangle = (0,0)^T$ and $\varepsilon = 0.0001$.

***Solution***

First, we find the gradient vector $|\mathbf{g}\rangle$ as:

$$|\mathbf{g}\rangle = \nabla f|\mathbf{x}\rangle = \left(\frac{\partial f}{\partial x_1}, \frac{\partial f}{\partial x_2}\right)^T = (1 + 4x_1 + 2x_2, -1 + 2x_1 + 2x_2)^T.$$

Now, find the Hessian matrix $\mathbf{H}$ as follows

$$\mathbf{H} = \mathbf{H}_f|\mathbf{x}\rangle = \nabla^2 f|\mathbf{x}\rangle = \begin{pmatrix} \frac{\partial^2 f}{\partial x_1^2} & \frac{\partial^2 f}{\partial x_1 \partial x_2} \\ \frac{\partial^2 f}{\partial x_2 \partial x_1} & \frac{\partial^2 f}{\partial x_2^2} \end{pmatrix} = \begin{pmatrix} 4 & 2 \\ 2 & 2 \end{pmatrix}.$$

**Iteration 1.**

$$|\mathbf{g}_0\rangle = (1, -1)^T,$$
$$\mathbf{H}_0^{-1} = \begin{pmatrix} 1/2 & -1/2 \\ -1/2 & 1 \end{pmatrix},$$
$$|\mathbf{d}_0\rangle = -\mathbf{H}_0^{-1}|\mathbf{g}_0\rangle = (-1, 3/2)^T.$$

Hence

$$|\mathbf{x}_1\rangle = |\mathbf{x}_0\rangle + \alpha_0|\mathbf{d}_0\rangle = (-\alpha_0, 3\alpha_0/2)^T,$$
$$f|\mathbf{x}_1\rangle = -\frac{5}{2}\alpha_0 + \frac{5}{4}\alpha_0^2,$$

so that, $\frac{df|\mathbf{x}_1\rangle}{d\alpha_0} = -\frac{5}{2} + \frac{5}{2}\alpha_0$. Now, to find the minimizer of $f|\mathbf{x}_1\rangle$, set $\frac{df|\mathbf{x}_1\rangle}{d\alpha_0} = 0$, which implies $\alpha_0 = 1$. Note that $\frac{d^2 f|\mathbf{x}_1\rangle}{d\alpha_0^2} = \frac{5}{2} > 0$, so that $\alpha_0 = 1$ is the minimizer of $f|\mathbf{x}_1\rangle$, which yield

$$|\mathbf{x}_1\rangle = (-1, 3/2)^T,$$
$$|\mathbf{g}_1\rangle = (0,0)^T.$$

Since $\|\mathbf{g}_1\| = \|\nabla f|\mathbf{x}_1\rangle\| = 0 < \varepsilon$. Thus, $|\mathbf{x}_1\rangle = (-1, 3/2)^T$ is the minimizer of $f|\mathbf{x}\rangle$.

### 5.6.1. Approximating the Inverse Hessian (Computation of $\mathbf{S}_k$)

The idea of the quasi-Newton method is to approximate the inverse of Hessian by some other matrix which should be positive definite. This saves the work of computation of second derivatives and avoids the difficulties associated with the loss of positive definiteness. Positive definiteness is essential; otherwise, the search direction $|\mathbf{d}_k\rangle$ is not guaranteed to be a descent direction (i.e., $f|\mathbf{x}_{k+1}\rangle \nless f|\mathbf{x}_k\rangle$.) So, we approximate $\mathbf{H}_k^{-1}$ by another matrix $\mathbf{S}_k$, using only the first partial derivatives of the objective function $f$, [29].

Let $f|\mathbf{x}\rangle \in C^2$ be a function in $\mathbb{R}^n$ and assume that the gradients of $f|\mathbf{x}\rangle$ at points $|\mathbf{x}_k\rangle$ and $|\mathbf{x}_{k+1}\rangle$ are $|\mathbf{g}_k\rangle$ and $|\mathbf{g}_{k+1}\rangle$, respectively. Now, suppose that

$$|\boldsymbol{\gamma}_k\rangle = |\mathbf{g}_{k+1}\rangle - |\mathbf{g}_k\rangle, \tag{5.100}$$

and

$$|\boldsymbol{\delta}_k\rangle = |\mathbf{x}_{k+1}\rangle - |\mathbf{x}_k\rangle. \tag{5.101}$$

Consider the Taylor expansion of the objective:

$$f|\mathbf{x}_{k+1}\rangle = f|\mathbf{x}_k\rangle + \langle\nabla f(\mathbf{x}_k)|\boldsymbol{\delta}_k\rangle + \frac{1}{2}\langle\boldsymbol{\delta}_k|\nabla^2 f(\mathbf{x}_k)|\boldsymbol{\delta}_k\rangle + \cdots, \tag{5.102}$$

then the Taylor series gives the elements of $|\mathbf{g}_{k+1}\rangle$ as

$$|\mathbf{g}_{k+1}\rangle = \nabla f|\mathbf{x}_{k+1}\rangle = \nabla f|\mathbf{x}_k\rangle + \nabla^2 f(\mathbf{x}_k)|\boldsymbol{\delta}_k\rangle + \frac{1}{2}\langle\boldsymbol{\delta}_k|\nabla^3 f(\mathbf{x}_k)|\boldsymbol{\delta}_k\rangle + \cdots. \tag{5.103}$$

Now if the objective function $f|\mathbf{x}\rangle$ is quadratic, the second derivative of $f|\mathbf{x}\rangle$ are constant (independent of $|\mathbf{x}\rangle$) and, in turn, the second derivatives of $|\mathbf{g}_k\rangle$ are zero (i.e., $\nabla^3 f|\mathbf{x}_k\rangle = 0$). Thus

$$\nabla f|\mathbf{x}_{k+1}\rangle = \nabla f|\mathbf{x}_k\rangle + \nabla^2 f(\mathbf{x}_k)|\boldsymbol{\delta}_k\rangle, \tag{5.104}$$

or

$$|\mathbf{g}_{k+1}\rangle = |\mathbf{g}_k\rangle + \mathbf{H}|\boldsymbol{\delta}_k\rangle, \tag{5.105}$$

where $\mathbf{H}$ is the Hessian of $f|\mathbf{x}\rangle$. Hence, we have

$$|\boldsymbol{\gamma}_k\rangle = \mathbf{H}|\boldsymbol{\delta}_k\rangle. \tag{5.106}$$

If the gradient is evaluated sequentially at $n+1$ points, say, $|\mathbf{x}_0\rangle, |\mathbf{x}_1\rangle, \dots, |\mathbf{x}_n\rangle$ such that the changes in $|\mathbf{x}\rangle$, namely,

$$\begin{aligned} |\boldsymbol{\delta}_0\rangle &= |\mathbf{x}_1 - \mathbf{x}_0\rangle, \\ |\boldsymbol{\delta}_1\rangle &= |\mathbf{x}_2 - \mathbf{x}_1\rangle, \\ &\vdots \\ |\boldsymbol{\delta}_{n-1}\rangle &= |\mathbf{x}_n - \mathbf{x}_{n-1}\rangle, \end{aligned} \tag{5.107}$$

form a set of linearly independent vectors; then sufficient information is obtained to determine $\mathbf{H}$ uniquely. To illustrate this fact, $n$ equations of (5.106) can be re-arranged as

$$(|\boldsymbol{\gamma}_0\rangle, |\boldsymbol{\gamma}_1\rangle, \dots, |\boldsymbol{\gamma}_{n-1}\rangle) = \mathbf{H}(|\boldsymbol{\delta}_0\rangle, |\boldsymbol{\delta}_1\rangle, \dots, |\boldsymbol{\delta}_{n-1}\rangle), \tag{5.108}$$

and, therefore,

$$\mathbf{H} = (|\boldsymbol{\gamma}_0\rangle, |\boldsymbol{\gamma}_1\rangle, \dots, |\boldsymbol{\gamma}_{n-1}\rangle)(|\boldsymbol{\delta}_0\rangle, |\boldsymbol{\delta}_1\rangle, \dots, |\boldsymbol{\delta}_{n-1}\rangle)^{-1}. \tag{5.109}$$

The solution exists if $|\boldsymbol{\delta}_0\rangle, |\boldsymbol{\delta}_1\rangle, \dots, |\boldsymbol{\delta}_{n-1}\rangle$ form a set of linearly independent vectors.

The strategy of the quasi-Newton methods is as follows.

1- Assume that a positive definite real symmetric matrix $\mathbf{S}_k$ is available for the minimization of the quadratic problem $f|\mathbf{x}\rangle = a + \langle\mathbf{b}|\mathbf{x}\rangle + \frac{1}{2}\langle\mathbf{x}|\mathbf{H}|\mathbf{x}\rangle$, which is an approximation of $\mathbf{H}^{-1}$.

2- Compute a quasi-Newton direction

$$|\mathbf{d}_k\rangle = -\mathbf{S}_k|\mathbf{g}_k\rangle. \tag{5.110}$$

3- Find $\alpha_k$, the value of $\alpha$ that minimizes $f|\mathbf{x}_k + \alpha\mathbf{d}_k\rangle$, by differentiating $f|\mathbf{x}_k - \alpha\mathbf{S}_k\mathbf{g}_k\rangle$ with respect to $\alpha$ and then setting the result to zero, the value of $\alpha$ that minimizes $f|\mathbf{x}_k - \alpha\mathbf{S}_k\mathbf{g}_k\rangle$ can be deduced as

$$\alpha_k = \frac{\langle\mathbf{g}_k|\mathbf{S}_k|\mathbf{g}_k\rangle}{\langle\mathbf{g}_k|\mathbf{S}_k\mathbf{H}\mathbf{S}_k|\mathbf{g}_k\rangle} = \frac{\langle\mathbf{g}_k|\mathbf{S}_k|\mathbf{g}_k\rangle}{\langle\mathbf{S}_k\mathbf{g}_k|\mathbf{H}|\mathbf{S}_k\mathbf{g}_k\rangle}, \tag{5.111}$$

where $\mathbf{S}_k$ and $\mathbf{H}$ are positive definite and $|\mathbf{g}_k\rangle = |\mathbf{b}\rangle + \mathbf{H}|\mathbf{x}_k\rangle$ is the gradient of $f|\mathbf{x}\rangle$ at $|\mathbf{x}\rangle = |\mathbf{x}_k\rangle$.

4- Determine a change in $|\mathbf{x}\rangle$ as

$$|\boldsymbol{\delta}_k\rangle = \alpha_k|\mathbf{d}_k\rangle, \tag{5.112}$$

and deduce a new point $|\mathbf{x}_{k+1}\rangle$, using (5.101).

5- The change in the gradient, $|\boldsymbol{\gamma}_k\rangle$, can be computed using (5.100) (by computing the gradient at points $|\mathbf{x}_k\rangle$ and $|\mathbf{x}_{k+1}\rangle$.)

6- Apply a correction to $\mathbf{S}_k$ and generate

$$\mathbf{S}_{k+1} = \mathbf{S}_k + \mathbf{C}_k, \tag{5.113}$$

where $\mathbf{C}_k$ is an $n \times n$ correction matrix that can be computed from available data. Note that the equations that the matrices $\mathbf{S}_k$ are required to satisfy, do not determine those matrices uniquely. Thus, we have some freedom in the way we compute the $\mathbf{S}_k$ matrix.

Let $\mathbf{S}_0, \mathbf{S}_1, \mathbf{S}_2, \dots$ be successive approximations of the inverse $\mathbf{H}_k^{-1}$ of the Hessian. On applying the above procedure iteratively starting with an initial point $|\mathbf{x}_0\rangle$ and an initial positive definite matrix $\mathbf{S}_0$, say, $\mathbf{S}_0 = \mathbf{I}_n$, the sequences $|\boldsymbol{\delta}_0\rangle, |\boldsymbol{\delta}_1\rangle, \dots, |\boldsymbol{\delta}_k\rangle, |\boldsymbol{\gamma}_0\rangle, |\boldsymbol{\gamma}_1\rangle, \dots, |\boldsymbol{\gamma}_k\rangle$, and $\mathbf{S}_1, \mathbf{S}_2, \dots, \mathbf{S}_{k+1}$ can be generated. If the approximation $\mathbf{S}_{k+1}$ of the Hessian satisfy

$$\mathbf{S}_{k+1}|\boldsymbol{\gamma}_i\rangle = |\boldsymbol{\delta}_i\rangle, \quad 0 \le i \le k, \tag{5.114}$$

then for $n$ steps $(k = n - 1)$, we can write

$$\begin{aligned} \mathbf{S}_n|\boldsymbol{\gamma}_0\rangle &= |\boldsymbol{\delta}_0\rangle, \\ \mathbf{S}_n|\boldsymbol{\gamma}_1\rangle &= |\boldsymbol{\delta}_1\rangle, \\ &\vdots \\ \mathbf{S}_n|\boldsymbol{\gamma}_{n-1}\rangle &= |\boldsymbol{\delta}_{n-1}\rangle. \end{aligned} \tag{5.115}$$

This set of equations can be represented as

$$\mathbf{S}_n\left(|\boldsymbol{\gamma}_0\rangle, |\boldsymbol{\gamma}_1\rangle, \dots, |\boldsymbol{\gamma}_{n-1}\rangle\right) = \left(|\boldsymbol{\delta}_0\rangle, |\boldsymbol{\delta}_1\rangle, \dots, |\boldsymbol{\delta}_{n-1}\rangle\right), \tag{5.116}$$

or

$$\mathbf{S}_n = \left(|\boldsymbol{\delta}_0\rangle, |\boldsymbol{\delta}_1\rangle, \dots, |\boldsymbol{\delta}_{n-1}\rangle\right)\left(|\boldsymbol{\gamma}_0\rangle, |\boldsymbol{\gamma}_1\rangle, \dots, |\boldsymbol{\gamma}_{n-1}\rangle\right)^{-1}, \tag{5.117}$$

and from (5.109) and (5.117), we find that if $\left(|\boldsymbol{\gamma}_0\rangle, |\boldsymbol{\gamma}_1\rangle, \dots, |\boldsymbol{\gamma}_{n-1}\rangle\right)$ is nonsingular, then $\mathbf{H}^{-1}$ is determined uniquely after $n$ steps, via

$$\mathbf{S}_n = \mathbf{H}^{-1} = \left(|\boldsymbol{\delta}_0\rangle, |\boldsymbol{\delta}_1\rangle, \dots, |\boldsymbol{\delta}_{n-1}\rangle\right)\left(|\boldsymbol{\gamma}_0\rangle, |\boldsymbol{\gamma}_1\rangle, \dots, |\boldsymbol{\gamma}_{n-1}\rangle\right)^{-1}. \tag{5.118}$$

As a consequence, we conclude that if $\mathbf{S}_n$ satisfies the equations $\mathbf{S}_n|\boldsymbol{\gamma}_i\rangle = |\boldsymbol{\delta}_i\rangle$, $0 \le i \le n - 1$, then the algorithm $|\mathbf{x}_{k+1}\rangle = |\mathbf{x}_k\rangle - \alpha_k\mathbf{S}_k|\mathbf{g}_k\rangle$, $\alpha_k = \arg\min_{\alpha\ge 0} f|\mathbf{x}_k - \alpha\mathbf{S}_k\mathbf{g}_k\rangle$, is guaranteed to solve problems with quadratic objective functions in $n + 1$ steps, because the update $|\mathbf{x}_{n+1}\rangle = |\mathbf{x}_n\rangle - \alpha_n\mathbf{S}_n|\mathbf{g}_n\rangle$ is equivalent to Newton's algorithm.

Hence, quasi-Newton algorithms have the form.

$$|\mathbf{x}_{k+1}\rangle = |\mathbf{x}_k\rangle + \alpha_k|\mathbf{d}_k\rangle, \tag{5.119.1}$$

$$|\mathbf{d}_k\rangle = -\mathbf{S}_k|\mathbf{g}_k\rangle, \tag{5.119.2}$$

$$\alpha_k = \arg\min_{\alpha\ge 0} f|\mathbf{x}_k + \alpha\mathbf{d}_k\rangle, \tag{5.119.3}$$

where the matrices $\mathbf{S}_0, \mathbf{S}_1, \dots$ are symmetric.

### 5.6.2 The Rank One Correction Formula

The general formula for correcting the matrix $\mathbf{S}_k$ has been defined in (5.113), where $\mathbf{C}_k$ is considered to be the correction matrix added to $\mathbf{S}_k$. To derive a rank one formula [157-160], we choose a scaled outer product of a vector $|\mathbf{z}\rangle$ for $\mathbf{C}_k$ as

$$\mathbf{C}_k = \beta_k|\mathbf{z}_k\rangle\langle\mathbf{z}_k|. \tag{5.120}$$

Hence, the update formula is

$$\mathbf{S}_{k+1} = \mathbf{S}_k + \beta_k|\mathbf{z}_k\rangle\langle\mathbf{z}_k|, \tag{5.121}$$

where $\beta_k \in \mathbb{R}$ and $|\mathbf{z}_k\rangle \in \mathbb{R}^n$. Note that, the rank of the outer product $|\mathbf{z}_k\rangle\langle\mathbf{z}_k|$ equal 1.

$$\begin{aligned} \operatorname{rank} |\mathbf{z}_k\rangle\langle\mathbf{z}_k| &= \operatorname{rank}\left[\begin{pmatrix} z_{1,k} \\ \vdots \\ z_{n,k} \end{pmatrix}\left(z_{1,k}, \dots, z_{n,k}\right)\right] = \operatorname{rank}\begin{pmatrix} z_{1,k}z_{1,k} & z_{1,k}z_{2,k} & \cdots & z_{1,k}z_{n,k} \\ z_{2,k}z_{1,k} & z_{2,k}z_{2,k} & \cdots & z_{2,k}z_{n,k} \\ \vdots & \vdots & \ddots & \vdots \\ z_{n,k}z_{1,k} & z_{n,k}z_{2,k} & \cdots & z_{n,k}z_{n,k} \end{pmatrix} \\ &= \operatorname{rank}\begin{pmatrix} z_{1,k}\left(z_{1,k}, z_{2,k}, \dots, z_{n,k}\right) \\ z_{2,k}\left(z_{1,k}, z_{2,k}, \dots, z_{n,k}\right) \\ \vdots \\ z_{n,k}\left(z_{1,k}, z_{2,k}, \dots, z_{n,k}\right) \end{pmatrix} = 1. \end{aligned} \tag{5.122}$$

Our goal now is to determine $\beta_k$ and $|\mathbf{z}_k\rangle$, given $\mathbf{S}_k$, $|\boldsymbol{\gamma}_k\rangle$, $|\boldsymbol{\delta}_k\rangle$, so that the required relationship $\mathbf{S}_{k+1}|\boldsymbol{\gamma}_i\rangle = |\boldsymbol{\delta}_i\rangle$, $i = 0, \dots, k$, is satisfied. To begin, let us first consider the relation,

$$\mathbf{S}_{k+1}|\boldsymbol{\gamma}_k\rangle = |\boldsymbol{\delta}_k\rangle. \tag{5.123}$$

In other words, given $\mathbf{S}_k$, $|\boldsymbol{\gamma}_k\rangle$, and $|\boldsymbol{\delta}_k\rangle$, we wish to find $\beta_k$ and $|\mathbf{z}_k\rangle$ to ensure that

$$\mathbf{S}_{k+1}|\boldsymbol{\gamma}_k\rangle = (\mathbf{S}_k + \beta_k|\mathbf{z}_k\rangle\langle\mathbf{z}_k|)|\boldsymbol{\gamma}_k\rangle = |\boldsymbol{\delta}_k\rangle. \tag{5.124}$$

From (5.123)

$$|\boldsymbol{\delta}_k\rangle = \mathbf{S}_k|\boldsymbol{\gamma}_k\rangle + \beta_k|\mathbf{z}_k\rangle\langle\mathbf{z}_k|\boldsymbol{\gamma}_k\rangle, \tag{5.125}$$

and

$$\langle\boldsymbol{\gamma}_k|(|\boldsymbol{\delta}_k\rangle - \mathbf{S}_k|\boldsymbol{\gamma}_k\rangle) = \beta_k\langle\boldsymbol{\gamma}_k|\mathbf{z}_k\rangle\langle\mathbf{z}_k|\boldsymbol{\gamma}_k\rangle = \beta_k\langle\mathbf{z}_k|\boldsymbol{\gamma}_k\rangle^2. \tag{5.126}$$

From (5.123), we have

$$|\boldsymbol{\delta}_k\rangle - \mathbf{S}_k|\boldsymbol{\gamma}_k\rangle = \beta_k|\mathbf{z}_k\rangle\langle\mathbf{z}_k|\boldsymbol{\gamma}_k\rangle = (\beta_k\langle\mathbf{z}_k|\boldsymbol{\gamma}_k\rangle)|\mathbf{z}_k\rangle, \tag{5.127.1}$$
$$(|\boldsymbol{\delta}_k\rangle - \mathbf{S}_k|\boldsymbol{\gamma}_k\rangle)^T = (\beta_k|\mathbf{z}_k\rangle\langle\mathbf{z}_k|\boldsymbol{\gamma}_k\rangle)^T = \beta_k\langle\boldsymbol{\gamma}_k|\mathbf{z}_k\rangle\langle\mathbf{z}_k| = (\beta_k\langle\mathbf{z}_k|\boldsymbol{\gamma}_k\rangle)\langle\mathbf{z}_k|, \tag{5.127.2}$$

since $\langle\mathbf{z}_k|\boldsymbol{\gamma}_k\rangle$ is a scalar. Hence

$$(|\boldsymbol{\delta}_k\rangle - \mathbf{S}_k|\boldsymbol{\gamma}_k\rangle)(|\boldsymbol{\delta}_k\rangle - \mathbf{S}_k|\boldsymbol{\gamma}_k\rangle)^T = \beta_k\langle\mathbf{z}_k|\boldsymbol{\gamma}_k\rangle^2\beta_k|\mathbf{z}_k\rangle\langle\mathbf{z}_k|, \tag{5.128}$$

and from (5.125) and (5.127), we have

$$\begin{aligned}\mathbf{C}_k &= \beta_k|\mathbf{z}_k\rangle\langle\mathbf{z}_k| \\ &= \frac{(|\boldsymbol{\delta}_k\rangle - \mathbf{S}_k|\boldsymbol{\gamma}_k\rangle)(|\boldsymbol{\delta}_k\rangle - \mathbf{S}_k|\boldsymbol{\gamma}_k\rangle)^T}{\beta_k\langle\mathbf{z}_k|\boldsymbol{\gamma}_k\rangle^2} \\ &= \frac{(|\boldsymbol{\delta}_k\rangle - \mathbf{S}_k|\boldsymbol{\gamma}_k\rangle)(|\boldsymbol{\delta}_k\rangle - \mathbf{S}_k|\boldsymbol{\gamma}_k\rangle)^T}{\langle\boldsymbol{\gamma}_k|(|\boldsymbol{\delta}_k\rangle - \mathbf{S}_k|\boldsymbol{\gamma}_k\rangle)} \\ &= \frac{|\boldsymbol{\delta}_k - \mathbf{S}_k\boldsymbol{\gamma}_k\rangle\langle\boldsymbol{\delta}_k - \mathbf{S}_k\boldsymbol{\gamma}_k|}{\langle\boldsymbol{\gamma}_k|\boldsymbol{\delta}_k - \mathbf{S}_k\boldsymbol{\gamma}_k\rangle}.\end{aligned} \tag{5.129}$$

With the correction matrix known, $\mathbf{S}_{k+1}$ can be deduced from (5.122) as

$$\begin{aligned}\mathbf{S}_{k+1} &= \mathbf{S}_k + \mathbf{C}_k \\ &= \mathbf{S}_k + \frac{|\boldsymbol{\delta}_k - \mathbf{S}_k\boldsymbol{\gamma}_k\rangle\langle\boldsymbol{\delta}_k - \mathbf{S}_k\boldsymbol{\gamma}_k|}{\langle\boldsymbol{\gamma}_k|\boldsymbol{\delta}_k - \mathbf{S}_k\boldsymbol{\gamma}_k\rangle}.\end{aligned} \tag{5.130}$$

This formula is known as the unique rank one update formula for $\mathbf{S}_{k+1}$. To implement (5.130), an initial symmetric positive definite matrix is selected for $\mathbf{S}_0$ at the start of the algorithm, and the next point $|\mathbf{x}_1\rangle$ is computed using (5.119). Then the new matrix $\mathbf{S}_1$ is computed using (5.130) and the new point $|\mathbf{x}_2\rangle$ is determined from (5.119). This iterative process is continued until convergence is achieved. If $\mathbf{S}_k$ is symmetric, (5.130) ensures that $\mathbf{S}_{k+1}$ is also symmetric. However, there is no guarantee that $\mathbf{S}_{k+1}$ remains positive definite even if $\mathbf{S}_k$ is positive definite. This might lead to a breakdown of the procedure, especially when used for the optimization of nonquadratic functions.

There are two problems associated with the rank one method. First, a positive definite matrix $\mathbf{S}_k$ may not yield a positive definite matrix $\mathbf{S}_{k+1}$. Second, the denominator in formula (5.130) may become zero, and the method will break down since $\mathbf{S}_{k+1}$ will become undefined. From (5.130), we can write

$$\begin{aligned}\langle\boldsymbol{\gamma}_i|\mathbf{S}_{k+1}|\boldsymbol{\gamma}_i\rangle &= \langle\boldsymbol{\gamma}_i|\mathbf{S}_k|\boldsymbol{\gamma}_i\rangle + \langle\boldsymbol{\gamma}_i|\frac{|\boldsymbol{\delta}_k - \mathbf{S}_k\boldsymbol{\gamma}_k\rangle\langle\boldsymbol{\delta}_k - \mathbf{S}_k\boldsymbol{\gamma}_k|}{\langle\boldsymbol{\gamma}_k|\boldsymbol{\delta}_k - \mathbf{S}_k\boldsymbol{\gamma}_k\rangle}|\boldsymbol{\gamma}_i\rangle \\ &= \langle\boldsymbol{\gamma}_i|\mathbf{S}_k|\boldsymbol{\gamma}_i\rangle + \frac{\langle\boldsymbol{\gamma}_i|\boldsymbol{\delta}_k - \mathbf{S}_k\boldsymbol{\gamma}_k\rangle\langle\boldsymbol{\delta}_k - \mathbf{S}_k\boldsymbol{\gamma}_k|\boldsymbol{\gamma}_i\rangle}{\langle\boldsymbol{\gamma}_k|\boldsymbol{\delta}_k - \mathbf{S}_k\boldsymbol{\gamma}_k\rangle} \\ &= \langle\boldsymbol{\gamma}_i|\mathbf{S}_k|\boldsymbol{\gamma}_i\rangle + \frac{[\langle\boldsymbol{\gamma}_i|\boldsymbol{\delta}_k\rangle - \langle\boldsymbol{\gamma}_i|\mathbf{S}_k|\boldsymbol{\gamma}_k\rangle][\langle\boldsymbol{\delta}_k|\boldsymbol{\gamma}_i\rangle - \langle\boldsymbol{\gamma}_k|\mathbf{S}_k|\boldsymbol{\gamma}_i\rangle]}{\langle\boldsymbol{\gamma}_k|\boldsymbol{\delta}_k - \mathbf{S}_k\boldsymbol{\gamma}_k\rangle} \\ &= \langle\boldsymbol{\gamma}_i|\mathbf{S}_k|\boldsymbol{\gamma}_i\rangle + \frac{[\langle\boldsymbol{\gamma}_i|\boldsymbol{\delta}_k\rangle - \langle\boldsymbol{\gamma}_i|\mathbf{S}_k|\boldsymbol{\gamma}_k\rangle]^2}{\langle\boldsymbol{\gamma}_k|\boldsymbol{\delta}_k - \mathbf{S}_k\boldsymbol{\gamma}_k\rangle}.\end{aligned} \tag{5.131}$$

Hence, if $\mathbf{S}_k$ is positive definite, a sufficient condition for $\mathbf{S}_{k+1}$ to be positive and definite is

$$\langle \boldsymbol{\gamma}_k | \boldsymbol{\delta}_k - \mathbf{S}_k \boldsymbol{\gamma}_k \rangle > 0. \tag{5.132}$$

***Example 5.6***

Let

$$f|\mathbf{x}\rangle = a + \langle \mathbf{b}|\mathbf{x}\rangle + \frac{1}{2}\langle \mathbf{x}|\mathbf{H}|\mathbf{x}\rangle, \ \ |\mathbf{x}\rangle \in \mathbb{R}^2,$$

with $a = 7$, $|\mathbf{b}\rangle = \begin{pmatrix} -1 \\ 1 \end{pmatrix}$, $\mathbf{H} = \begin{pmatrix} 1 & 0 \\ 0 & 2 \end{pmatrix}$, and $|\mathbf{x}_0\rangle = \begin{pmatrix} 0 \\ 0 \end{pmatrix}$. Use the rank one correction method to generate two $\mathbf{H}$-conjugate directions.

***Solution***

We first compute the gradient $|\mathbf{g}\rangle = \boldsymbol{\nabla} f|\mathbf{x}\rangle$ and evaluate it at $|\mathbf{x}_0\rangle$,

$$|\mathbf{g}_0\rangle = |\mathbf{b}\rangle + \mathbf{H}|\mathbf{x}_0\rangle = \begin{pmatrix} -1 \\ 1 \end{pmatrix}.$$

It is a nonzero vector, so we proceed with the first iteration.

**Iteration 1.**

Let $\mathbf{S}_0 = \mathbf{I}_2$. Then,

$$|\mathbf{d}_0\rangle = -\mathbf{S}_0|\mathbf{g}_0\rangle = \begin{pmatrix} 1 \\ -1 \end{pmatrix}.$$

The step size $\alpha_0$ is

$$\alpha_0 = -\frac{\langle \mathbf{g}_0|\mathbf{d}_0\rangle}{\langle \mathbf{d}_0|\mathbf{H}|\mathbf{d}_0\rangle} = 2/3.$$

Hence,

$$|\mathbf{x}_1\rangle = |\mathbf{x}_0\rangle + \alpha_0|\mathbf{d}_0\rangle = \begin{pmatrix} 2/3 \\ -2/3 \end{pmatrix}.$$

We evaluate the gradient $|\mathbf{g}\rangle = \boldsymbol{\nabla} f|\mathbf{x}\rangle$ at $|\mathbf{x}_1\rangle$ to obtain

$$|\mathbf{g}_1\rangle = |\mathbf{b}\rangle + \mathbf{H}|\mathbf{x}_1\rangle = \begin{pmatrix} -1/3 \\ -1/3 \end{pmatrix}.$$

It is a nonzero vector, so we proceed with the second iteration.

**Iteration 2.**

Compute $\mathbf{S}_1$,

$$\mathbf{S}_1 = \mathbf{S}_0 + \frac{|\boldsymbol{\delta}_0 - \mathbf{S}_0\boldsymbol{\gamma}_0\rangle\langle \boldsymbol{\delta}_0 - \mathbf{S}_0\boldsymbol{\gamma}_0|}{\langle \boldsymbol{\gamma}_0|\boldsymbol{\delta}_0 - \mathbf{S}_0\boldsymbol{\gamma}_0\rangle}.$$

To find $\mathbf{S}_1$ we need to compute

$$|\boldsymbol{\delta}_0\rangle = |\mathbf{x}_1\rangle - |\mathbf{x}_0\rangle = \begin{pmatrix} 2/3 \\ -2/3 \end{pmatrix},$$

$$|\boldsymbol{\gamma}_0\rangle = |\mathbf{g}_1\rangle - |\mathbf{g}_0\rangle = \begin{pmatrix} 2/3 \\ -4/3 \end{pmatrix}.$$

Using the above, we determine,

$$|\boldsymbol{\delta}_0\rangle - \mathbf{S}_0|\boldsymbol{\gamma}_0\rangle = \begin{pmatrix} 0 \\ 2/3 \end{pmatrix},$$

$$\langle \boldsymbol{\gamma}_0|(|\boldsymbol{\delta}_0\rangle - \mathbf{S}_0|\boldsymbol{\gamma}_0\rangle) = -8/9.$$

Then, we obtain

$$\mathbf{S}_1 = \begin{pmatrix} 1 & 0 \\ 0 & 1/2 \end{pmatrix},$$

and

$$|\mathbf{d}_1\rangle = -\mathbf{S}_1|\mathbf{g}_1\rangle = \begin{pmatrix} 1/3 \\ 1/6 \end{pmatrix}.$$

Note that $\langle \mathbf{d}_0|\mathbf{H}|\mathbf{d}_1\rangle = 0$, that is, $|\mathbf{d}_0\rangle$ and $|\mathbf{d}_1\rangle$ are $\mathbf{H}$-conjugate.

**Algorithm 5.17:**
Quasi-Newton Algorithms:
The Rank One Correction.
The Davidon Fletcher Powell (DFP) Method
The Broyden Fletcher Goldfarb Shanno (BFGS)

Step 1: Input $|\mathbf{x}_0\rangle$ and initialize the tolerance $\varepsilon$.
Step 2: Set $k = 0$ and select a real symmetric positive definite matrix $\mathbf{S}_0$ (put $\mathbf{S}_0 = \mathbf{I}_n$). Compute $|\mathbf{g}_0\rangle$.

Step 3: If $|\mathbf{g}_k\rangle = |\mathbf{0}\rangle$, stop; else, $|\mathbf{d}_k\rangle = -\mathbf{S}_k|\mathbf{g}_k\rangle$.
Compute $\alpha_k = \arg\min_{\alpha\geq 0} f|\mathbf{x}_k + \alpha\mathbf{d}_k\rangle$, the value of $\alpha$ that minimizes $f|\mathbf{x}_k + \alpha\mathbf{d}_k\rangle$, using a line search.
Set $|\boldsymbol{\delta}_k\rangle = \alpha_k|\mathbf{d}_k\rangle$ and update $|\mathbf{x}_{k+1}\rangle = |\mathbf{x}_k\rangle + |\boldsymbol{\delta}_k\rangle$.

Step 4: If $\|\boldsymbol{\delta}_k\|_2 < \varepsilon$, output $|\mathbf{x}^*\rangle = |\mathbf{x}_{k+1}\rangle$ and $f|\mathbf{x}^*\rangle = f|\mathbf{x}_{k+1}\rangle$, and stop.
Step 5: Compute $|\mathbf{g}_{k+1}\rangle$ and set $|\boldsymbol{\gamma}_k\rangle = |\mathbf{g}_{k+1}\rangle - |\mathbf{g}_k\rangle$.
**The rank one correction method**
Compute $\mathbf{S}_{k+1}$ by the formula

$$\mathbf{S}_{k+1} = \mathbf{S}_k + \frac{|\boldsymbol{\delta}_k - \mathbf{S}_k\boldsymbol{\gamma}_k\rangle\langle\boldsymbol{\delta}_k - \mathbf{S}_k\boldsymbol{\gamma}_k|}{\langle\boldsymbol{\gamma}_k|\boldsymbol{\delta}_k - \mathbf{S}_k\boldsymbol{\gamma}_k\rangle}.$$

**The DFP method**
Compute $\mathbf{S}_{k+1}$ by the formula

$$\mathbf{S}_{k+1} = \mathbf{S}_k + \frac{|\boldsymbol{\delta}_k\rangle\langle\boldsymbol{\delta}_k|}{\langle\boldsymbol{\delta}_k|\boldsymbol{\gamma}_k\rangle} - \frac{|\mathbf{S}_k\boldsymbol{\gamma}_k\rangle\langle\mathbf{S}_k\boldsymbol{\gamma}_k|}{\langle\boldsymbol{\gamma}_k|\mathbf{S}_k|\boldsymbol{\gamma}_k\rangle}.$$

**The BFGS method**
Compute $\mathbf{S}_{k+1}$ by the formula

$$\mathbf{S}_{k+1} = \mathbf{S}_k + \left(1 + \frac{\langle\boldsymbol{\gamma}_k|\mathbf{S}_k|\boldsymbol{\gamma}_k\rangle}{\langle\boldsymbol{\gamma}_k|\boldsymbol{\delta}_k\rangle}\right)\frac{|\boldsymbol{\delta}_k\rangle\langle\boldsymbol{\delta}_k|}{\langle\boldsymbol{\delta}_k|\boldsymbol{\gamma}_k\rangle} - \frac{|\mathbf{S}_k\boldsymbol{\gamma}_k\rangle\langle\boldsymbol{\delta}_k| + |\boldsymbol{\delta}_k\rangle\langle\mathbf{S}_k\boldsymbol{\gamma}_k|}{\langle\boldsymbol{\gamma}_k|\boldsymbol{\delta}_k\rangle}.$$

Step 6: Set $k = k + 1$ and go to step 3.
In step 3, the value of $\alpha_k$ is obtained by using a line search to render the algorithm more amenable to nonquadratic problems. However, for convex quadratic problems, $\alpha_k$ should be calculated by using (5.111).

### 5.6.3 The Rank Two Correction Formula

The rank two update formulas guarantee both symmetry and positive definiteness of the matrix $\mathbf{S}_{k+1}$ and are more robust in minimizing general nonlinear functions, hence are preferred in practical applications. The Davidon Fletcher Powell (DFP) and Broyden Fletcher Goldfarb Shanno (BFGS) iterative methods are described in detail in the following sections.

### The DFP Method

The DFP algorithm [158,161] is superior to the rank one correction algorithm because it preserves the positive definiteness of the matrix $\mathbf{S}_k$. In rank two updates, we choose the update matrix $\mathbf{C}_k$ as the sum of two “rank one” updates as

$$\mathbf{C}_k = \beta_{1k}|\mathbf{z}_{1,k}\rangle\langle\mathbf{z}_{1,k}| + \beta_{2k}|\mathbf{z}_{2,k}\rangle\langle\mathbf{z}_{2,k}|, \tag{5.133}$$

where $\beta_{1k}, \beta_{2k} \in \mathbb{R}$ and the vectors $|\mathbf{z}_{1,k}\rangle, |\mathbf{z}_{2,k}\rangle \in \mathbb{R}^n$. Therefore, (5.113) and (5.133) lead to the new update formula

$$\mathbf{S}_{k+1} = \mathbf{S}_k + \beta_{1k}|\mathbf{z}_{1,k}\rangle\langle\mathbf{z}_{1,k}| + \beta_{2k}|\mathbf{z}_{2,k}\rangle\langle\mathbf{z}_{2,k}|. \tag{5.134}$$

By forcing (5.133) to satisfy the quasi-Newton condition, $|\boldsymbol{\delta}_k\rangle = \mathbf{S}_{k+1}|\boldsymbol{\gamma}_k\rangle$, we obtain

$$|\boldsymbol{\delta}_k\rangle = \left(\mathbf{S}_k + \beta_{1k}|\mathbf{z}_{1,k}\rangle\langle\mathbf{z}_{1,k}| + \beta_{2k}|\mathbf{z}_{2,k}\rangle\langle\mathbf{z}_{2,k}|\right)|\boldsymbol{\gamma}_k\rangle \tag{5.135}$$

$$= \mathbf{S}_k|\boldsymbol{\gamma}_k\rangle + \beta_{1k}\langle \mathbf{z}_{1,k}|\boldsymbol{\gamma}_k\rangle|\mathbf{z}_{1,k}\rangle + \beta_{2k}\langle \mathbf{z}_{2,k}|\boldsymbol{\gamma}_k\rangle|\mathbf{z}_{2,k}\rangle,$$

where $\langle \mathbf{z}_{1,k}|\boldsymbol{\gamma}_k\rangle$ and $\langle \mathbf{z}_{2,k}|\boldsymbol{\gamma}_k\rangle$ can be identified as scalars. Although the vectors $|\mathbf{z}_{1,k}\rangle$ and $|\mathbf{z}_{2,k}\rangle$ in (5.134) are not unique, the following choices can be made to satisfy (5.134):

$$\begin{aligned} |\mathbf{z}_{1,k}\rangle &= |\boldsymbol{\delta}_k\rangle, \\ |\mathbf{z}_{2,k}\rangle &= \mathbf{S}_k|\boldsymbol{\gamma}_k\rangle, \\ \beta_{1k} &= \frac{1}{\langle \mathbf{z}_{1,k}|\boldsymbol{\gamma}_k\rangle}, \\ \beta_{2k} &= -\frac{1}{\langle \mathbf{z}_{2,k}|\boldsymbol{\gamma}_k\rangle}. \end{aligned} \tag{5.136}$$

So, the correction matrix (5.133) becomes

$$\begin{aligned} \mathbf{C}_k &= \frac{1}{\langle \mathbf{z}_{1,k}|\boldsymbol{\gamma}_k\rangle}|\boldsymbol{\delta}_k\rangle\langle \mathbf{z}_{1,k}| - \frac{1}{\langle \mathbf{z}_{2,k}|\boldsymbol{\gamma}_k\rangle}\mathbf{S}_k|\boldsymbol{\gamma}_k\rangle\langle \mathbf{z}_{2,k}| \\ &= \frac{|\boldsymbol{\delta}_k\rangle\langle\boldsymbol{\delta}_k|}{\langle\boldsymbol{\delta}_k|\boldsymbol{\gamma}_k\rangle} - \frac{(\mathbf{S}_k|\boldsymbol{\gamma}_k\rangle)(\mathbf{S}_k|\boldsymbol{\gamma}_k\rangle)^T}{(\mathbf{S}_k|\boldsymbol{\gamma}_k\rangle)^T|\boldsymbol{\gamma}_k\rangle} \\ &= \frac{|\boldsymbol{\delta}_k\rangle\langle\boldsymbol{\delta}_k|}{\langle\boldsymbol{\delta}_k|\boldsymbol{\gamma}_k\rangle} - \frac{\mathbf{S}_k|\boldsymbol{\gamma}_k\rangle\langle\boldsymbol{\gamma}_k|\mathbf{S}_k}{\langle\boldsymbol{\gamma}_k|\mathbf{S}_k|\boldsymbol{\gamma}_k\rangle} \\ &= \frac{|\boldsymbol{\delta}_k\rangle\langle\boldsymbol{\delta}_k|}{\langle\boldsymbol{\delta}_k|\boldsymbol{\gamma}_k\rangle} - \frac{|\mathbf{S}_k\boldsymbol{\gamma}_k\rangle\langle\mathbf{S}_k\boldsymbol{\gamma}_k|}{\langle\boldsymbol{\gamma}_k|\mathbf{S}_k|\boldsymbol{\gamma}_k\rangle}, \end{aligned} \tag{5.137}$$

and thus, the rank two update formula (5.134) can be expressed as

$$\begin{aligned} \mathbf{S}_{k+1} &= \mathbf{S}_k + \mathbf{C}_k \\ &= \mathbf{S}_k + \frac{|\boldsymbol{\delta}_k\rangle\langle\boldsymbol{\delta}_k|}{\langle\boldsymbol{\delta}_k|\boldsymbol{\gamma}_k\rangle} - \frac{|\mathbf{S}_k\boldsymbol{\gamma}_k\rangle\langle\mathbf{S}_k\boldsymbol{\gamma}_k|}{\langle\boldsymbol{\gamma}_k|\mathbf{S}_k|\boldsymbol{\gamma}_k\rangle}, \end{aligned} \tag{5.138}$$

where the correction is an $n \times n$ symmetric matrix of rank two. This equation is known as the DFP formula. The validity of (5.137) can be demonstrated by post-multiplying both sides by $|\boldsymbol{\gamma}_k\rangle$, that is,

$$\mathbf{S}_{k+1}|\boldsymbol{\gamma}_k\rangle = \mathbf{S}_k|\boldsymbol{\gamma}_k\rangle + \frac{|\boldsymbol{\delta}_k\rangle\langle\boldsymbol{\delta}_k|\boldsymbol{\gamma}_k\rangle}{\langle\boldsymbol{\delta}_k|\boldsymbol{\gamma}_k\rangle} - \frac{\mathbf{S}_k|\boldsymbol{\gamma}_k\rangle\langle\boldsymbol{\gamma}_k|\mathbf{S}_k|\boldsymbol{\gamma}_k\rangle}{\langle\boldsymbol{\gamma}_k|\mathbf{S}_k|\boldsymbol{\gamma}_k\rangle}.$$

Since $\langle\boldsymbol{\delta}_k|\boldsymbol{\gamma}_k\rangle$ and $\langle\boldsymbol{\gamma}_k|\mathbf{S}_k|\boldsymbol{\gamma}_k\rangle$ are scalars, we have

$$\mathbf{S}_{k+1}|\boldsymbol{\gamma}_k\rangle = |\boldsymbol{\delta}_k\rangle,$$

as required.

Using the iteration formula $|\mathbf{x}_{k+1}\rangle = |\mathbf{x}_k\rangle + \alpha_k|\mathbf{d}_k\rangle$, such that $|\mathbf{d}_k\rangle$ related to $|\boldsymbol{\delta}_k\rangle = |\mathbf{x}_{k+1}\rangle - |\mathbf{x}_k\rangle$ by the formula $|\boldsymbol{\delta}_k\rangle = \alpha_k|\mathbf{d}_k\rangle$. Thus (5.138) can be expressed as

$$\mathbf{S}_{k+1} = \mathbf{S}_k + \alpha_k\frac{|\mathbf{d}_k\rangle\langle\mathbf{d}_k|}{\langle\mathbf{d}_k|\boldsymbol{\gamma}_k\rangle} - \frac{|\mathbf{S}_k\boldsymbol{\gamma}_k\rangle\langle\mathbf{S}_k\boldsymbol{\gamma}_k|}{\langle\boldsymbol{\gamma}_k|\mathbf{S}_k|\boldsymbol{\gamma}_k\rangle}. \tag{5.139}$$

Note that, (5.130) and (5.137) are known as inverse update formulas since these equations approximate the inverse of the Hessian matrix of $f$. The complete DFP method is presented in Algorithm 5.17.

***Example 5.7***

Locate the minimizer of

$$f|\mathbf{x}\rangle = \langle\mathbf{b}|\mathbf{x}\rangle + \frac{1}{2}\langle\mathbf{x}|\mathbf{H}|\mathbf{x}\rangle, \quad |\mathbf{x}\rangle \in \mathbb{R}^2,$$

with $\mathbf{H} = \begin{pmatrix} 4 & 1 \\ 1 & 2 \end{pmatrix}$ and $|\mathbf{b}\rangle = \begin{pmatrix} 1 \\ -1 \end{pmatrix}$. Use the initial point $|\mathbf{x}_0\rangle = \begin{pmatrix} 0 \\ 0 \end{pmatrix}$ and $\mathbf{S}_0 = \mathbf{I}_2$.

***Solution***

Compute the gradient $|\mathbf{g}_0\rangle$,

$$|\mathbf{g}_0\rangle = |\mathbf{b}\rangle + \mathbf{H}|\mathbf{x}_0\rangle = \begin{pmatrix} 1 \\ -1 \end{pmatrix}.$$

It is a nonzero vector, so we proceed with the first iteration.

**Iteration 1.** Given that $\mathbf{S}_0 = \mathbf{I}_2$. Then,

$$|\mathbf{d}_0\rangle = -\mathbf{S}_0|\mathbf{g}_0\rangle = \begin{pmatrix} -1 \\ 1 \end{pmatrix}.$$

Because $f$ is a quadratic function,

$$\alpha_0 = \arg\min_{\alpha\geq 0} f|\mathbf{x}_0 + \alpha\mathbf{d}_0\rangle = -\frac{\langle \mathbf{g}_0|\mathbf{d}_0\rangle}{\langle \mathbf{d}_0|\mathbf{H}|\mathbf{d}_0\rangle} = \frac{1}{2}.$$

Therefore,

$$|\mathbf{x}_1\rangle = |\mathbf{x}_0\rangle + \alpha_0|\mathbf{d}_0\rangle = \begin{pmatrix} -1/2 \\ 1/2 \end{pmatrix}.$$

We then compute the gradient $|\mathbf{g}_1\rangle$

$$|\mathbf{g}_1\rangle = |\mathbf{b}\rangle + \mathbf{H}|\mathbf{x}_1\rangle = \begin{pmatrix} -1/2 \\ -1/2 \end{pmatrix}.$$

It is a nonzero vector, so we proceed with the second iteration.

**Iteration 2.** Compute

$$|\boldsymbol{\delta}_0\rangle = |\mathbf{x}_1\rangle - |\mathbf{x}_0\rangle = \begin{pmatrix} -1/2 \\ 1/2 \end{pmatrix},$$

$$|\boldsymbol{\gamma}_0\rangle = |\mathbf{g}_1\rangle - |\mathbf{g}_0\rangle = \begin{pmatrix} -3/2 \\ 1/2 \end{pmatrix}.$$

Using the above, we determine,

$$|\boldsymbol{\delta}_0\rangle\langle\boldsymbol{\delta}_0| = \begin{pmatrix} 1/4 & -1/4 \\ -1/4 & 1/4 \end{pmatrix},$$

$$\langle\boldsymbol{\delta}_0|\boldsymbol{\gamma}_0\rangle = 1,$$

$$\mathbf{S}_0|\boldsymbol{\gamma}_0\rangle = \begin{pmatrix} -3/2 \\ 1/2 \end{pmatrix}.$$

Thus,

$$\mathbf{S}_0|\boldsymbol{\gamma}_0\rangle\langle\boldsymbol{\gamma}_0|\mathbf{S}_0 = \begin{pmatrix} -3/2 \\ 1/2 \end{pmatrix}(-3/2, 1/2) = \begin{pmatrix} 9/4 & -3/4 \\ -3/4 & 1/4 \end{pmatrix},$$

$$\langle\boldsymbol{\gamma}_0|\mathbf{S}_0|\boldsymbol{\gamma}_0\rangle = (-3/2, 1/2)\begin{pmatrix} 1 & 0 \\ 0 & 1 \end{pmatrix}\begin{pmatrix} -3/2 \\ 1/2 \end{pmatrix} = 2.5.$$

Using the above, we now compute $\mathbf{S}_1$:

$$\mathbf{S}_1 = \mathbf{S}_0 + \frac{|\boldsymbol{\delta}_0\rangle\langle\boldsymbol{\delta}_0|}{\langle\boldsymbol{\delta}_0|\boldsymbol{\gamma}_0\rangle} - \frac{\mathbf{S}_0|\boldsymbol{\gamma}_0\rangle\langle\boldsymbol{\gamma}_0|\mathbf{S}_0}{\langle\boldsymbol{\gamma}_0|\mathbf{S}_0|\boldsymbol{\gamma}_0\rangle} = \begin{pmatrix} 0.35 & 0.05 \\ 0.05 & 1.15 \end{pmatrix}.$$

We now compute $|\mathbf{d}_1\rangle = -\mathbf{S}_1|\mathbf{g}_1\rangle = \begin{pmatrix} 0.2 \\ 0.6 \end{pmatrix}$ and

$$\alpha_1 = \arg\min_{\alpha\geq 0} f|\mathbf{x}_1 + \alpha\mathbf{d}_1\rangle = -\frac{\langle \mathbf{g}_1|\mathbf{d}_1\rangle}{\langle \mathbf{d}_1|\mathbf{H}|\mathbf{d}_1\rangle} = 0.357143.$$

Hence, the new update

$$|\mathbf{x}_2\rangle = |\mathbf{x}_1\rangle + \alpha_1|\mathbf{d}_1\rangle = \begin{pmatrix} -0.428571 \\ 0.714286 \end{pmatrix} = |\mathbf{x}^*\rangle,$$

because $f$ is a quadratic function of two variables.

Note that we have $\langle\mathbf{d}_0|\mathbf{H}|\mathbf{d}_1\rangle = \langle\mathbf{d}_1|\mathbf{H}|\mathbf{d}_0\rangle = 0$; that is, $|\mathbf{d}_0\rangle$ and $|\mathbf{d}_1\rangle$ are $\mathbf{H}$-conjugate directions.

### The BFGS Method

Recall the updating formulas for the approximation of the inverse $\mathbf{H}^{-1}$ of the Hessian matrix which were based on the following equations:

$$\mathbf{S}_{k+1}|\boldsymbol{\gamma}_i\rangle = |\boldsymbol{\delta}_i\rangle, \quad 0 \leq i \leq k. \tag{5.140}$$

We then formulated update formulas for the approximations to the inverse $\mathbf{H}^{-1}$ of the Hessian. It is possible to derive a family of direct update formulas in which approximations to the Hessian matrix $\mathbf{H}$ are considered instead of approximating $\mathbf{H}^{-1}$. To do this, let $\mathbf{E}_k$ be our estimate of $\mathbf{H}_k$ at the $k$th step. We need $\mathbf{E}_{k+1}$ to satisfy the following set of equations:

$$|\boldsymbol{\gamma}_i\rangle = \mathbf{E}_{k+1}|\boldsymbol{\delta}_i\rangle,\ 0 \le i \le k. \tag{5.141}$$

Note that, (5.140) and (5.141) are similar; the only difference is that $|\boldsymbol{\delta}_i\rangle$ and $|\boldsymbol{\gamma}_i\rangle$ are interchanged. Thus, given any update formula for $\mathbf{S}_k$, a corresponding update formula for $\mathbf{E}_k$ can be found by interchanging the roles of $\mathbf{E}_k$ and $\mathbf{S}_k$ and of $|\boldsymbol{\gamma}_k\rangle$ and $|\boldsymbol{\delta}_k\rangle$. In particular, the BFGS update [157,162-166] for $\mathbf{E}_k$ corresponding to the DFP update for $\mathbf{S}_k$. Formulas related in this way are said to be dual or complementary. Recall the DFP update for the approximation $\mathbf{S}_k$ of the inverse $\mathbf{H}^{-1}$ Hessian from (5.138) as

$$\mathbf{S}_{k+1}^{\mathrm{DFP}} = \mathbf{S}_k + \frac{|\boldsymbol{\delta}_k\rangle\langle\boldsymbol{\delta}_k|}{\langle\boldsymbol{\delta}_k|\boldsymbol{\gamma}_k\rangle} - \frac{|\mathbf{S}_k\boldsymbol{\gamma}_k\rangle\langle\mathbf{S}_k\boldsymbol{\gamma}_k|}{\langle\boldsymbol{\gamma}_k|\mathbf{S}_k|\boldsymbol{\gamma}_k\rangle}. \tag{5.142}$$

We apply the complementarity concept in the above equation. In other words, the procedure used in deriving (5.130) and (5.138) can be followed by using $\mathbf{E}_k$, $|\boldsymbol{\delta}_k\rangle$, and $|\boldsymbol{\gamma}_k\rangle$ in place of $\mathbf{S}_k$, $|\boldsymbol{\gamma}_k\rangle$, and $|\boldsymbol{\delta}_k\rangle$, respectively. This leads to the rank two update formula, similar to (5.138), known as the BFGS update of $\mathbf{E}_k$:

$$\mathbf{E}_{k+1} = \mathbf{E}_k + \frac{|\boldsymbol{\gamma}_k\rangle\langle\boldsymbol{\gamma}_k|}{\langle\boldsymbol{\gamma}_k|\boldsymbol{\delta}_k\rangle} - \frac{|\mathbf{E}_k\boldsymbol{\delta}_k\rangle\langle\mathbf{E}_k\boldsymbol{\delta}_k|}{\langle\boldsymbol{\delta}_k|\mathbf{E}_k|\boldsymbol{\delta}_k\rangle}, \tag{5.143}$$

that represents an update equation for the approximation $\mathbf{E}_k$ of the Hessian itself. In practical computations, (5.143) is rewritten more conveniently in terms of $\mathbf{S}_k$, using the following formulas

$$\mathbf{S}_{k+1}^{\mathrm{BFGS}} = (\mathbf{E}_{k+1})^{-1} = \left(\mathbf{E}_k + \frac{|\boldsymbol{\gamma}_k\rangle\langle\boldsymbol{\gamma}_k|}{\langle\boldsymbol{\gamma}_k|\boldsymbol{\delta}_k\rangle} - \frac{|\mathbf{E}_k\boldsymbol{\delta}_k\rangle\langle\mathbf{E}_k\boldsymbol{\delta}_k|}{\langle\boldsymbol{\delta}_k|\mathbf{E}_k|\boldsymbol{\delta}_k\rangle}\right)^{-1}, \tag{5.144}$$

or

$$\mathbf{S}_{k+1}^{\mathrm{BFGS}} = \mathbf{S}_k + \left(1 + \frac{\langle\boldsymbol{\gamma}_k|\mathbf{S}_k|\boldsymbol{\gamma}_k\rangle}{\langle\boldsymbol{\gamma}_k|\boldsymbol{\delta}_k\rangle}\right)\frac{|\boldsymbol{\delta}_k\rangle\langle\boldsymbol{\delta}_k|}{\langle\boldsymbol{\delta}_k|\boldsymbol{\gamma}_k\rangle} - \frac{|\mathbf{S}_k\boldsymbol{\gamma}_k\rangle\langle\boldsymbol{\delta}_k| + |\boldsymbol{\delta}_k\rangle\langle\mathbf{S}_k\boldsymbol{\gamma}_k|}{\langle\boldsymbol{\gamma}_k|\boldsymbol{\delta}_k\rangle}. \tag{5.145}$$

This represents the BFGS formula for updating $\mathbf{S}_k$. Numerical experience indicates that the BFGS method is the best unconstrained quasi-Newton method and is less influenced by errors in finding $\alpha_k$ compared to the DFP method.

***Example 5.8***

Use the BFGS method to minimize

$$f|\mathbf{x}\rangle = a + \langle\mathbf{b}|\mathbf{x}\rangle + \frac{1}{2}\langle\mathbf{x}|\mathbf{H}|\mathbf{x}\rangle,\ \ |\mathbf{x}\rangle \in \mathbb{R}^2,$$

where

$$\mathbf{H} = \begin{pmatrix} 5 & -1 \\ -1 & 4 \end{pmatrix}, \qquad |\mathbf{b}\rangle = \begin{pmatrix} 0 \\ -1 \end{pmatrix}, \qquad a = -1.$$

Use the initial point $|\mathbf{x}_0\rangle = \begin{pmatrix} 0 \\ 0 \end{pmatrix}$ and take $\mathbf{S}_0 = \mathbf{I}_2$. Verify that $\mathbf{S}_2^{\mathrm{BFGS}} = \mathbf{H}^{-1}$.

***Solution***

We compute the gradient $|\mathbf{g}_0\rangle$

$$|\mathbf{g}_0\rangle = |\mathbf{b}\rangle + \mathbf{H}|\mathbf{x}_0\rangle = \begin{pmatrix} 0 \\ -1 \end{pmatrix}.$$

It is a nonzero vector, so we proceed with the first iteration.

**Iteration 1.** Given that $\mathbf{S}_0 = \mathbf{I}_2$. Then,

$$|\mathbf{d}_0\rangle = -\mathbf{S}_0|\mathbf{g}_0\rangle = \begin{pmatrix} 0 \\ 1 \end{pmatrix}.$$

The objective function is quadratic, and hence we can use the following formula to compute $\alpha_0$:

$$\alpha_0 = \arg\min_{\alpha \ge 0} f|\mathbf{x}_0 + \alpha\mathbf{d}_0\rangle = -\frac{\langle\mathbf{g}_0|\mathbf{d}_0\rangle}{\langle\mathbf{d}_0|\mathbf{H}|\mathbf{d}_0\rangle} = \frac{1}{4}.$$

Therefore,

$$|\mathbf{x}_1\rangle = |\mathbf{x}_0\rangle + \alpha_0|\mathbf{d}_0\rangle = \begin{pmatrix} 0 \\ 1/4 \end{pmatrix}.$$

We then compute the gradient $|\mathbf{g}_1\rangle$

$$|\mathbf{g}_1\rangle = |\mathbf{b}\rangle + \mathbf{H}|\mathbf{x}_1\rangle = \begin{pmatrix} -1/4 \\ 0 \end{pmatrix}.$$

It is a nonzero vector, so we proceed with the second iteration.

**Iteration 2.** To compute $\mathbf{S}_1^{\text{BFGS}}$, we need to evaluate the following quantities:

$$|\boldsymbol{\delta}_0\rangle = |\mathbf{x}_1\rangle - |\mathbf{x}_0\rangle = \begin{pmatrix} 0 \\ 1/4 \end{pmatrix},$$

$$|\boldsymbol{\gamma}_0\rangle = |\mathbf{g}_1\rangle - |\mathbf{g}_0\rangle = \begin{pmatrix} -1/4 \\ 1 \end{pmatrix}.$$

Using the above, we now compute $\mathbf{S}_1^{\text{BFGS}}$:

$$\mathbf{S}_1^{\text{BFGS}} = \mathbf{S}_0 + \left(1 + \frac{\langle\boldsymbol{\gamma}_0|\mathbf{S}_0|\boldsymbol{\gamma}_0\rangle}{\langle\boldsymbol{\gamma}_0|\boldsymbol{\delta}_0\rangle}\right)\frac{|\boldsymbol{\delta}_0\rangle\langle\boldsymbol{\delta}_0|}{\langle\boldsymbol{\delta}_0|\boldsymbol{\gamma}_0\rangle} - \frac{\mathbf{S}_0|\boldsymbol{\gamma}_0\rangle\langle\boldsymbol{\delta}_0| + |\boldsymbol{\delta}_0\rangle\langle\boldsymbol{\gamma}_0|\mathbf{S}_0}{\langle\boldsymbol{\gamma}_0|\boldsymbol{\delta}_0\rangle} = \begin{pmatrix} 1 & 0.25 \\ 0.25 & 0.3125 \end{pmatrix}.$$

We now compute $|\mathbf{d}_1\rangle$

$$|\mathbf{d}_1\rangle = -\mathbf{S}_1^{\text{BFGS}}|\mathbf{g}_1\rangle = \begin{pmatrix} 0.25 \\ 0.0625 \end{pmatrix},$$

and

$$\alpha_1 = \arg\min_{\alpha\geq 0} f|\mathbf{x}_1 + \alpha\mathbf{d}_1\rangle = -\frac{\langle\mathbf{g}_1|\mathbf{d}_1\rangle}{\langle\mathbf{d}_1|\mathbf{H}|\mathbf{d}_1\rangle} = 0.210526.$$

Therefore, the new update

$$|\mathbf{x}_2\rangle = |\mathbf{x}_1\rangle + \alpha_1|\mathbf{d}_1\rangle = \begin{pmatrix} 0.052632 \\ 0.263158 \end{pmatrix}.$$

Because our objective function $f$ is a quadratic on $\mathbb{R}^2$, $|\mathbf{x}_2\rangle$ is the minimizer. Note that the gradient at $|\mathbf{x}_2\rangle$ is $|\mathbf{g}_2\rangle = |\mathbf{0}\rangle$.

To verify that $\mathbf{S}_2^{\text{BFGS}} = \mathbf{H}^{-1}$, we compute

$$|\boldsymbol{\delta}_1\rangle = |\mathbf{x}_2\rangle - |\mathbf{x}_1\rangle = \begin{pmatrix} 0.0526316 \\ 0.0131579 \end{pmatrix},$$

$$|\boldsymbol{\gamma}_1\rangle = |\mathbf{g}_2\rangle - |\mathbf{g}_1\rangle = \begin{pmatrix} 1/4 \\ 0 \end{pmatrix}.$$

Hence,

$$\mathbf{S}_2^{\text{BFGS}} = \mathbf{S}_1 + \left(1 + \frac{\langle\boldsymbol{\gamma}_1|\mathbf{S}_1|\boldsymbol{\gamma}_1\rangle}{\langle\boldsymbol{\gamma}_1|\boldsymbol{\delta}_1\rangle}\right)\frac{|\boldsymbol{\delta}_1\rangle\langle\boldsymbol{\delta}_1|}{\langle\boldsymbol{\delta}_1|\boldsymbol{\gamma}_1\rangle} - \frac{\mathbf{S}_1|\boldsymbol{\gamma}_1\rangle\langle\boldsymbol{\delta}_1| + |\boldsymbol{\delta}_1\rangle\langle\boldsymbol{\gamma}_1|\mathbf{S}_1}{\langle\boldsymbol{\gamma}_1|\boldsymbol{\delta}_1\rangle} = \begin{pmatrix} 0.210526 & 0.052632 \\ 0.052632 & 0.263158 \end{pmatrix}.$$

Note that indeed $\mathbf{S}_2^{\text{BFGS}}\mathbf{H} = \mathbf{H}\mathbf{S}_2^{\text{BFGS}} = \mathbf{I}_2$, and hence $\mathbf{S}_2^{\text{BFGS}} = \mathbf{H}^{-1}$.

# CHAPTER 6

# STRATEGIES FOR GENERALIZATION AND HYPER-PARAMETER TUNING

In this chapter, we delve into some of the most critical concepts and methodologies that form the backbone of machine learning, guiding the development of models that are not only powerful but also robust and generalizable across various scenarios. We focus on the aspects of NN training that determine how effectively a model can generalize from training data to unseen data. Generalization is the ultimate test of a NN's performance, assessing its ability to apply learned patterns to new datasets.

- We begin by addressing the fundamental challenges of overfitting and generalization. Overfitting occurs when the NN learns the details and noise in the training data to an extent that it negatively impacts the performance of NN on new data. Conversely, generalization refers to the model's ability to apply what it has learned to unseen data. We will discuss strategies to balance this, including the importance of a robust model architecture.
- Next, we explore the bias-variance trade-off, a pivotal concept that helps in diagnosing the performance of machine learning algorithms. Bias refers to errors due to overly simplistic assumptions in the learning algorithm. Variance refers to errors from sensitivity to small fluctuations in the training set. High bias can cause a model to miss the relevant relations between features and target outputs (underfitting), whereas high variance can cause modeling the random noise in the training data (overfitting).
- To assess a model's generalization, we will split the data into three sets: training, validation, and testing. The training set is used to train the model, the validation set is used to tune the model's hyperparameters and prevent overfitting, and the test set is used to evaluate the model's performance as it simulates real-world, unseen data.
- As we measure the success of our models, performance measures come into play. Common metrics include accuracy, precision, recall, the F1 score for classification tasks, and mean squared error or mean absolute error for regression tasks. We will explore how these metrics can guide hyperparameter tuning.
- The practice of tuning hyperparameters is essential for optimizing model performance. Techniques such as grid search and random search are popular methods for exploring the hyperparameter space. Grid search evaluates the model across a grid of hyperparameter combinations, while random search randomly selects combinations, offering a balance between exploration and exploitation.
- Gaussian processes (GPs) are a probabilistic model used in machine learning to predict the distribution of possible outcomes rather than just the best estimate. GPs are particularly useful for understanding model uncertainty, which can be leveraged for Bayesian optimization (BO) in hyperparameter tuning.
- Further enhancing our toolkit for hyperparameter optimization, we will introduce tuning hyperparameters with BO. BO is a strategy for the global optimization of objective functions that are noisy, expensive to evaluate, or have no closed form. It is particularly useful for tuning hyperparameters in scenarios where evaluations are costly or time-consuming. BO uses past evaluation results to form a probabilistic model mapping hyperparameters to a probability of a score on the objective function.
- Lastly, we will discuss Acquisition Functions (ACFs). ACFs in BO are used to select the next set of hyperparameters to evaluate. Common ACFs include Expected Improvement (EI), Probability of Improvement (PI), and Upper Confidence Bound (UCB). These functions help in deciding which hyperparameter settings are likely to yield improvements over the best current observations.

This chapter aims to provide a comprehensive understanding of these areas, equipping you with the knowledge to effectively train, evaluate, and optimize NNs for robust generalization capabilities.

## 6.1 Overfitting and Generalization

NNs have a high capacity for learning intricate details and nuances present in the training data. However, the great power of NNs is also their greatest weakness; they may end up capturing noise rather than learning the underlying true patterns that generalize well to unseen data [69]. Overfitting occurs when a model learns to memorize the training data rather than generalize from it. In the context of NNs, this means the model captures noise or random fluctuations in the training data as if they were meaningful patterns. As a result, the model performs very well on the training data but fails to generalize to new, unseen data. Practically, this can lead to a NN providing excellent performance metrics (such as accuracy or loss) when evaluated on the training dataset (that it is built on) but performing poorly when applied to new, unseen data, which is typically evaluated using a separate test dataset.

Generalization is indeed a crucial concept in machine learning, including NNs. It refers to the ability of a model to perform well on unseen data or data that it hasn't been trained on. Achieving good generalization is often the primary goal in machine learning tasks because it indicates that the model has learned the underlying patterns in the data rather than memorizing specific instances. When a model generalizes well, it can make accurate predictions or classifications for new, previously unseen examples. This is essential for the model to be useful in real-world applications where it encounters data that it hasn't encountered during training.

For a helpful analogy, imagine you're preparing for a test by repeatedly studying a set of practice questions. You've memorized each question and its exact answer. When it comes time for the actual test, you find that the questions are slightly different from what you studied, and you struggle to answer them correctly. In this analogy: The practice questions are like the training data for a NN. Memorizing the exact answers to the practice questions is akin to overfitting. When the test questions differ slightly from the practice questions, your memorization doesn't help, just as an overfitted NN struggles with unseen test instances. Extreme overfitting, or memorization, occurs when you've essentially memorized the training data without truly understanding the underlying concepts.

Let's consider another scenario where you're teaching a child to classify animals. You show them pictures of cats, dogs, birds, fish, and so on. Initially, the child might try to memorize each specific instance of every animal they see. However, as they encounter more and more animals, they start to notice common features that define each category. For example, they might notice that cats and dogs have fur, while birds have feathers, and fish have scales. As the child continues to learn, they start to abstract these common features and associate them with each animal category. Eventually, they don't need to see every single variation of a cat or a dog to recognize one; instead, they can identify common features like fur, four legs, and a tail, and classify an animal as a cat or a dog based on those features.

Similarly, in a NN, during training, the model is exposed to various examples of different categories (e.g., cats, dogs, birds). Initially, it might try to memorize each training example. However, as training progresses, the model starts to identify common features (e.g., fur, feathers, scales) that define each category. These common features become the abstract representations the model uses for classification during inference. Overfitting in this context would be akin to the child memorizing every single animal they see during training, including irrelevant details or outliers. This could lead to poor performance when trying to classify new animals that weren't seen during training. To prevent overfitting, the model needs to focus on learning the underlying patterns and features that generalize well across different examples of each category. We do not want our network to remember every training instance. Rather, we want the network to form abstractions that will enable it to recognize the object during inferencing, even though the exact likeness of the object instance encountered during inferencing was never seen during training. If however, the network has too much expressive power (too many neurons or equivalently too many weights) relative to the number of training instances - the network can and often will rote remember the training instances.

On the other hand, underfitting occurs when the model is too simple to capture the underlying structure of the data, resulting in poor performance not only on the training data but also on unseen data (testing data). In essence, the model fails to learn the relevant patterns and relationships present in the training data. To address underfitting, one common approach is indeed to increase the complexity of the model, such as by adding more layers, more neurons, or using a more complex architecture altogether. By doing so, the model gains more capacity to learn intricate patterns and relationships in the data, hopefully leading to improved performance on both training and testing datasets.

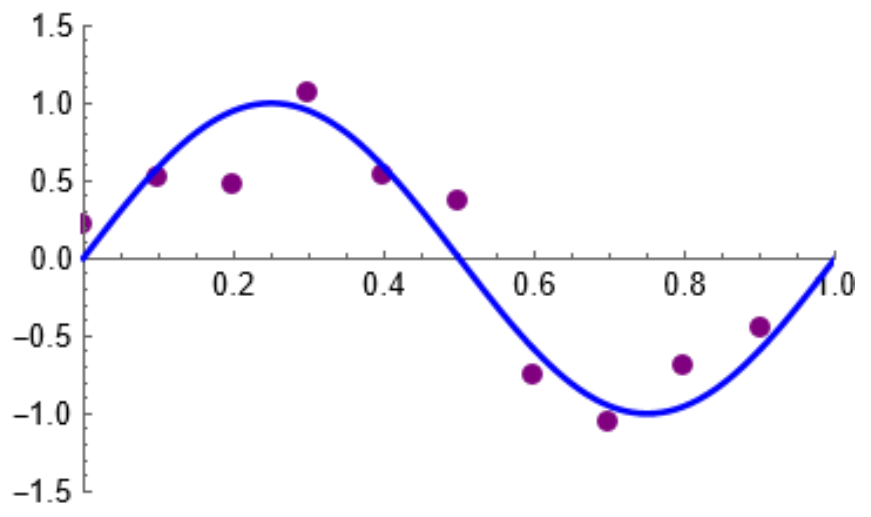

**Figure 6.1.** Training Data Points with Original Function Curve: This figure illustrates the training data points (shown in purple) generated from a sinusoidal function with added random noise. Additionally, the original sinusoidal function curve (shown in blue) is overlaid for reference.

For a helpful analogy of underfitting, imagine you're trying to teach a child how to identify shapes. You start by showing them only circles and squares, and you ask the child to classify objects as either circles or squares based on their shapes. Now, if you stop your teaching there and never introduce the child to any other shapes, they might develop a very simplistic understanding of shapes. They might believe that all objects are either circles or squares, and they might struggle to classify anything beyond those two shapes. This scenario mirrors underfitting in machine learning: the child's understanding of shapes is too simplistic to accurately classify a wide range of shapes, just as an underfit model's understanding of the data is too simplistic to accurately capture its complexity.

In order to understand the problem of generalization, polynomial regression is indeed a great example to explain the concepts of overfitting and generalization in machine learning. Polynomial regression is a type of regression analysis in which the relationship between the independent variable $x$ and the dependent variable $y$ is modeled as an $n$th degree polynomial. While polynomial regression can capture more complex relationships between variables compared to linear regression, it also introduces the risk of overfitting and challenges in generalization. In the context of polynomial regression, higher-degree polynomials can lead to highly flexible models that can closely fit the training data, even if the underlying relationship is not truly polynomial. This can result in a model that performs well on the training data but poorly on unseen data.

Let's consider a training set consisting of $N$ observations of $x$, represented as $\mathbf{x} \equiv (x_1, \ldots, x_N)^T$, along with corresponding observations of $y$, denoted as $\mathbf{y} \equiv (y_1, \ldots, y_N)^T$. The training data is visualized in Figure 6.1, which shows a plot with 10 data points. The input data points ($\mathbf{x}$) are generated by selecting values uniformly distributed in the range $[0,1]$. Then, the corresponding output data points ($\mathbf{y}$) are derived by computing the values of the function $\sin(2\pi x)$ for each $x$, with the addition of a slight level of random noise following a Gaussian distribution to each point. These noisy points are denoted as $y_n$. The objective of utilizing this training set is to predict the value of the target variable ($\hat{y}$) for a new value of the input variable ($\hat{x}$). This is essentially a regression task, where the goal is to uncover the underlying function $\sin(2\pi x)$ that generates the output variable $y$ based on the input variable $x$.

To fit our data, we will employ a polynomial function of the form [49]:

$$y(x, \mathbf{w}) = w_0 + w_1 x + w_2 x^2 + \cdots + w_M x^M, \tag{6.1}$$

where $M$ is the order of the polynomial. This model uses $(M+1)$ parameters $w_0 \ldots w_M$ in order to explain pairs $(x, y)$ available to us. The polynomial coefficients $w_0, \ldots, w_M$ are collectively denoted by the vector $\mathbf{w}$. One could implement this model by using a NN with $M$ inputs corresponding to $x$, $x^2 \ldots x^M$, and a single bias neuron whose coefficient is $w_0$. These coefficients will be determined by fitting the polynomial to our training data. This fitting process involves minimizing an error function:

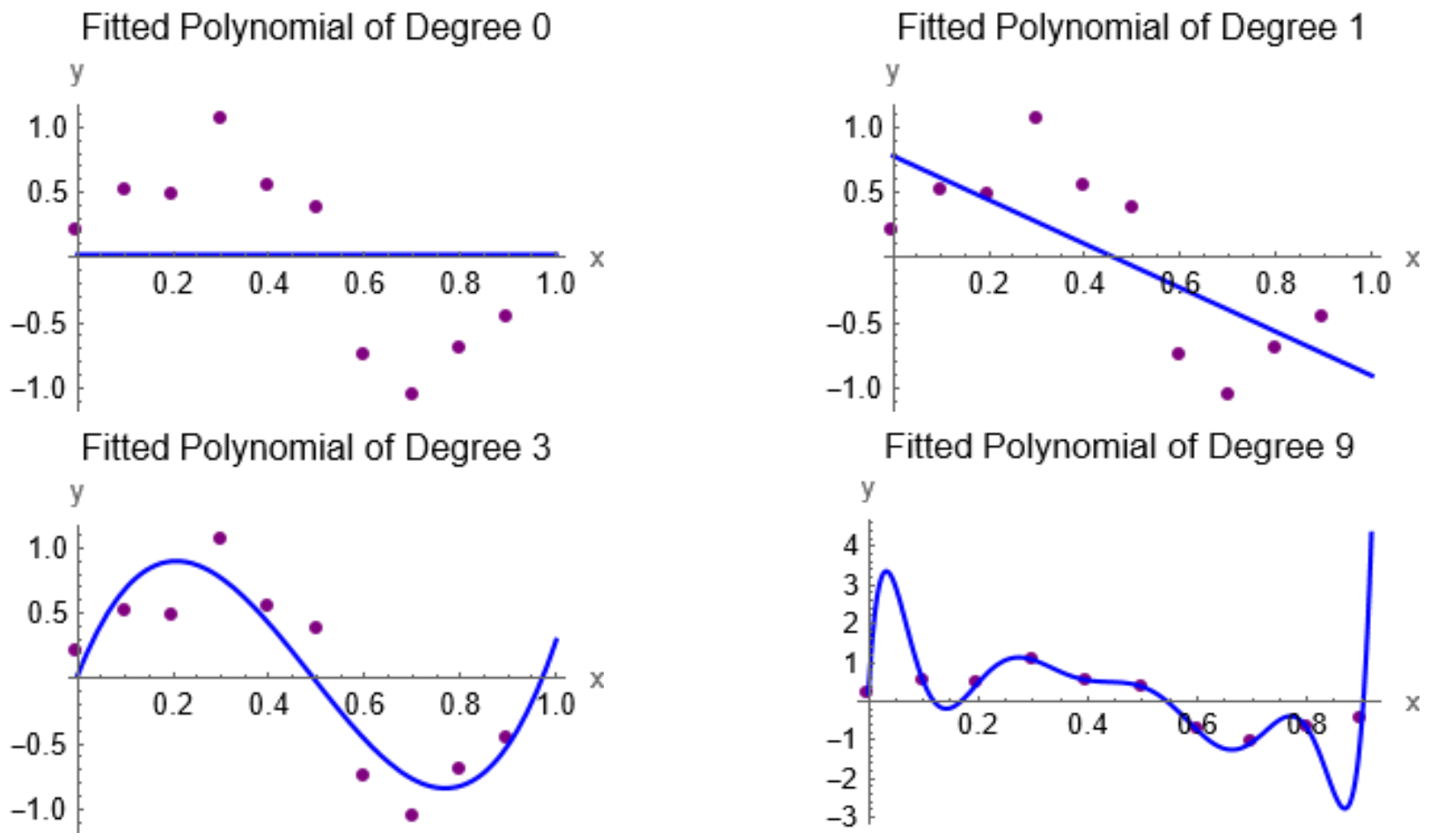

**Figure 6.2.** Comparison of Polynomial Fitting: The plot illustrates the fitting of polynomials of varying orders ($M = 0$, $M = 1$, $M = 3$, and $M = 9$) to a dataset (depicted in purple) sampled from the function $y = \sin(2\pi x)$ with added random noise. While lower-order polynomials struggle to capture the underlying sinusoidal trend (underfitting), excessively high-order polynomials perfectly fit the training data but exhibit erratic behavior (overfitting).

$$\mathcal{L}(\mathbf{w}) = \frac{1}{2}\sum_{n=1}^{N}(y(x_n, \mathbf{w}) - y_n)^2. \tag{6.2}$$

This function measures the discrepancy between the predicted values of the polynomial and the actual training data. We can solve the curve fitting problem by choosing the value of $\mathbf{w}$ for which $\mathcal{L}(\mathbf{w})$ is as small as possible. Since the error function is quadratic in the coefficients $\mathbf{w}$, its derivatives with respect to these coefficients will be linear in the elements of $\mathbf{w}$. Consequently, the minimization process leads to a unique solution, denoted as $\mathbf{w}^*$, which can be determined analytically. The resulting polynomial is given by the function $y(x, \mathbf{w}^*)$.

The problem of choosing the appropriate order, $M$, of a polynomial when fitting data is crucial in machine learning and statistical modeling. In Figure 6.2, we show four examples of the results of fitting polynomials having orders $M =$ 0, 1, 3, and 9 to the data set shown in Figure 6.1. Which curve should we trust? The line that gets almost no training example correct? Or the complicated curve that hits every single point in the dataset? Selecting an overly low order may lead to underfitting, while selecting an overly high order may result in overfitting. Underfitting occurs when the model is too simple to capture the underlying structure of the data, leading to poor performance in both training and generalization. This is evident when using low-order polynomials like $M = 0$ or $M = 1$, where the fitted curves fail to capture the sinusoidal nature of the data. On the other hand, overfitting happens when the model is too complex, capturing noise or fluctuations in the training data rather than the underlying trend. This typically occurs when using excessively high-order polynomials, such as $M = 9$. The polynomial passes exactly through each data point, and $\mathcal{L}(\mathbf{w}^*) = 0$. While these polynomials perfectly fit the training data, they exhibit erratic behavior and do not generalize well (the fitted curve oscillates wildly) to unseen data. The third-order ($M = 3$) polynomial seems to give the best fit to the function $\sin(2\pi x)$ of the examples shown in Figure 6.2.

This leads to an interesting point about training and evaluating machine learning models. By building a very complex model, it's quite easy to perfectly fit our training dataset because we give our model enough degrees of freedom to contort itself to fit the observations in the training set. But when we evaluate such a complex model on new data, it performs poorly. This becomes an even more significant issue in deep learning, where our NNs have large numbers of layers containing many neurons. The number of connections in these models is astronomical, reaching the millions. As a result, overfitting is commonplace.

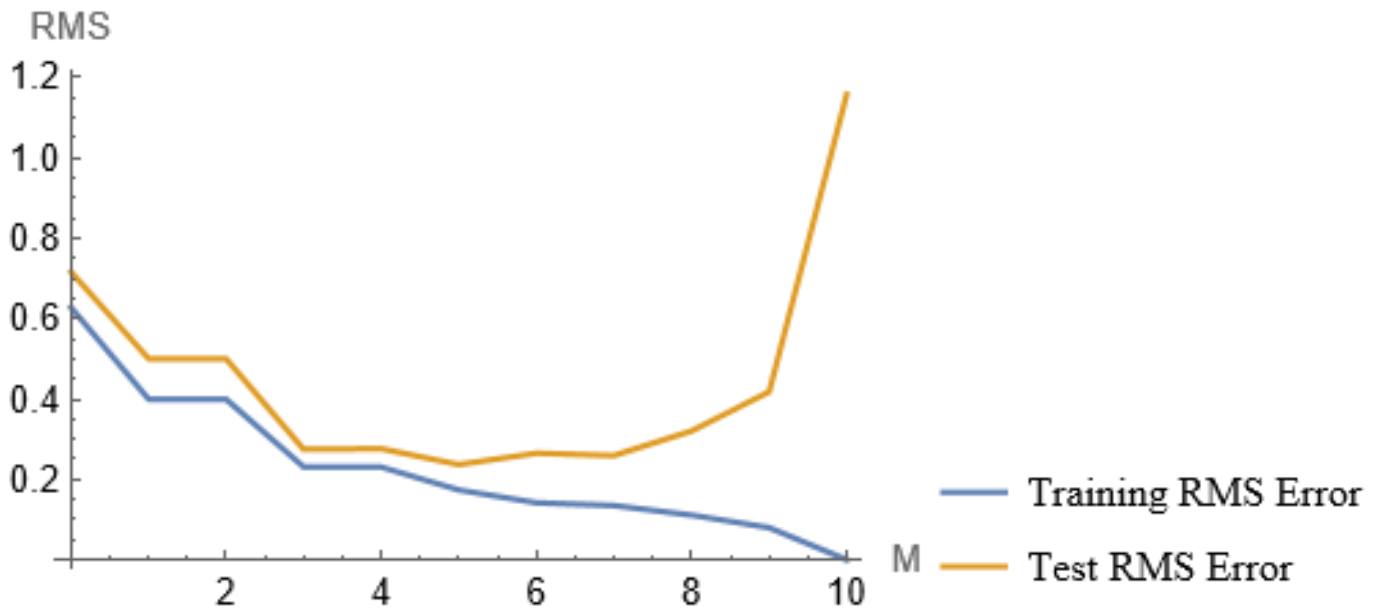

**Figure 6.3.** Training and Test RMS Errors vs. Polynomial Degree ($M$): This figure illustrates the RMS errors for polynomial regression models of varying degrees ($M$) evaluated on both training and test datasets. The plot shows the trade-off between model complexity (polynomial degree) and generalization performance, aiding in the selection of optimal model complexity to avoid underfitting or overfitting. The RMS errors for both the training and test datasets are plotted against the polynomial degrees, providing insights into the model's performance on seen and unseen data.

**Table 6.1.** Coefficients ($\mathbf{w}^*$) for polynomials of varying orders. This table displays the estimated coefficients for polynomial models of different orders. It's noteworthy to observe the significant increase in coefficient magnitudes as polynomial orders increase.

| | $M = 0$ | $M = 1$ | $M = 3$ | $M = 9$ |
|---|---|---|---|---|
| $w_0^*$ | $+0.0209197$ | $+0.780035$ | $+0.0221099$ | $+0.213963$ |
| $w_1^*$ | | $-1.68692$ | $+9.33508$ | $+236.151$ |
| $w_2^*$ | | | $-28.6867$ | $-5876.84$ |
| $w_3^*$ | | | $+19.6234$ | $+57575$ |
| $w_4^*$ | | | | $-294786$ |
| $w_5^*$ | | | | $+877743$ |
| $w_6^*$ | | | | $-1.57544 * 10^6$ |
| $w_7^*$ | | | | $+1.67889 * 10^6$ |
| $w_8^*$ | | | | $-977501$ |
| $w_9^*$ | | | | $+239311$ |

To gain a quantitative understanding of the relationship between the generalization performance and $M$, we employ a distinct test dataset consisting of 100 data points. These points are generated using the identical procedure employed for the training set, albeit with new selections for the random noise values incorporated into the target values. For every $M$ under consideration, we proceed to assess the residual value of $\mathcal{L}(\mathbf{w}^*)$ given by (6.2) for the training dataset. Concurrently, we evaluate $\mathcal{L}(\mathbf{w}^*)$ for the test dataset. When choosing the value of $M$, it's crucial to strike a balance between flexibility and generalization. We note from Figure 6.3 that:

- Small values of $M$ often lead to significant test set errors as the resulting polynomials lack the flexibility needed to accurately capture the oscillations inherent in functions like $\sin(2\pi x)$.
- In contrast, values of $M$ within the range $3 \leq M \leq 8$ tend to yield favorable outcomes, producing small test set errors while providing reasonable representations of the underlying function. These values of $M$ strike a balance between model complexity and generalization capability, effectively capturing the essential features of the generating function.
- However, when $M = 9$, something interesting happens. The training set error diminishes to zero because the high-degree polynomial can perfectly fit the training data. However, despite the perfect fit to the training data, the test set error becomes very large, indicating poor generalization performance.
- The substantial test set error for $M = 9$ serves as a clear indicator of overfitting. While the polynomial achieves flawless fitting to the training data, it struggles to generalize to unseen data. This is evident from the wild oscillations observed in the corresponding function $y(x, \mathbf{w}^*)$. Overfitting highlights the importance of not only capturing the nuances of the training data but also ensuring the model's ability to generalize effectively beyond it.

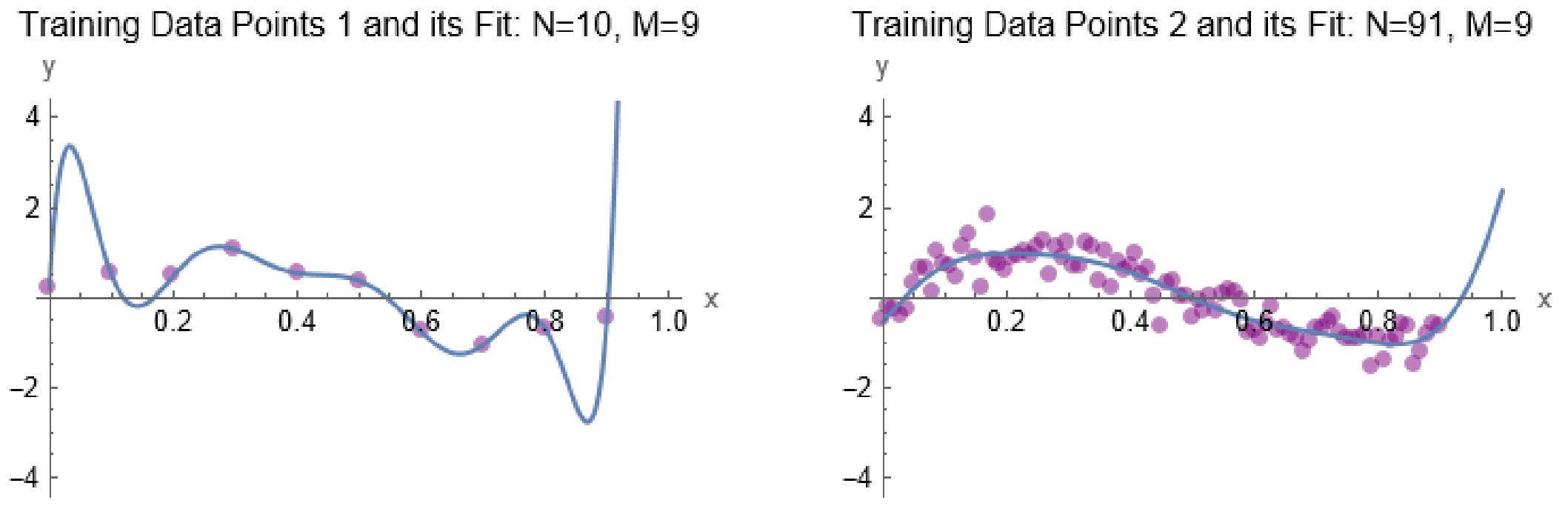

**Figure 6.4.** Left panel: This figure illustrates a dataset containing 10 data points ($N = 10$) generated with a step size of 0.1. A polynomial curve of degree 9 ($M = 9$) is fitted to this dataset, capturing the relationship between the input variable ($x$) and the target variable ($y$). The plot displays the original data points in purple along with the fitted polynomial curve. Right panel: In this figure, a larger dataset consisting of 91 data points ($N = 91$) is depicted, generated with a finer step size of 0.01. Similarly, a polynomial curve of degree 9 ($M = 9$) is fitted to this dataset, aiming to capture the underlying relationship between $x$ and $y$ with increased data density. The plot shows the original data points in purple along with the fitted polynomial curve. We see that increasing the size of the data set reduces the over-fitting problem.

Examining the values of the coefficients $\mathbf{w}^*$ derived from polynomials of varying orders, as outlined in Table 6.1, provides valuable insight into the problem at hand. It's evident that with an increase in $M$, the coefficients tend to escalate in magnitude. Specifically, for the polynomial of order $M = 9$, the coefficients have been meticulously adjusted to match each data point precisely. This adjustment manifests in the form of large positive and negative values, resulting in the polynomial function perfectly aligning with the given data points. However, amidst these data points, the function exhibits significant oscillations. Intuitively, this phenomenon can be explained by understanding that as we opt for more flexible polynomials with higher $M$ values, they progressively adapt to the random noise present in the target values. Consequently, they become finely attuned to this noise, which, in turn, leads to the observed oscillations between data points.

It is also interesting to examine the behavior of a given model as the size of the data set is varied, as shown in Figure 6.4. We see that, for a given model complexity, the overfitting problem becomes less severe as the size of the data set increases. Another way to say this is that the larger the data set, the more complex (in other words more flexible) the model that we can afford to fit to the data.

Finding the right balance between model complexity and training data size is crucial for building models that generalize well to unseen data [49]. It's essential to choose models with appropriate complexity for the given task and ensure an adequate amount of training data to support generalization.

- Having more training instances generally improves the generalization power of the model. This is because a larger and more diverse dataset provides more information for the model to learn from, helping it to capture the underlying patterns in the data more accurately.
- Increasing the complexity of the model, such as adding more layers or parameters in a NN, can reduce its generalization power. This is because overly complex models may start to memorize the training data instead of learning the underlying relationships, leading to overfitting.
- When a significant amount of training data is available, overly simple models may struggle to capture complex relationships between features and the target variable. In such cases, a more complex model may be necessary to achieve better performance. However, it's essential to strike a balance, as overly complex models can still suffer from overfitting, especially when training data is limited.
- A common guideline is that the total number of training data points should be at least 2 to 3 times larger than the number of parameters in the model. This guideline helps ensure that the model has enough data to learn from without overfitting the training set. However, the precise number of data instances required depends on the specific characteristics of the dataset and the complexity of the model.

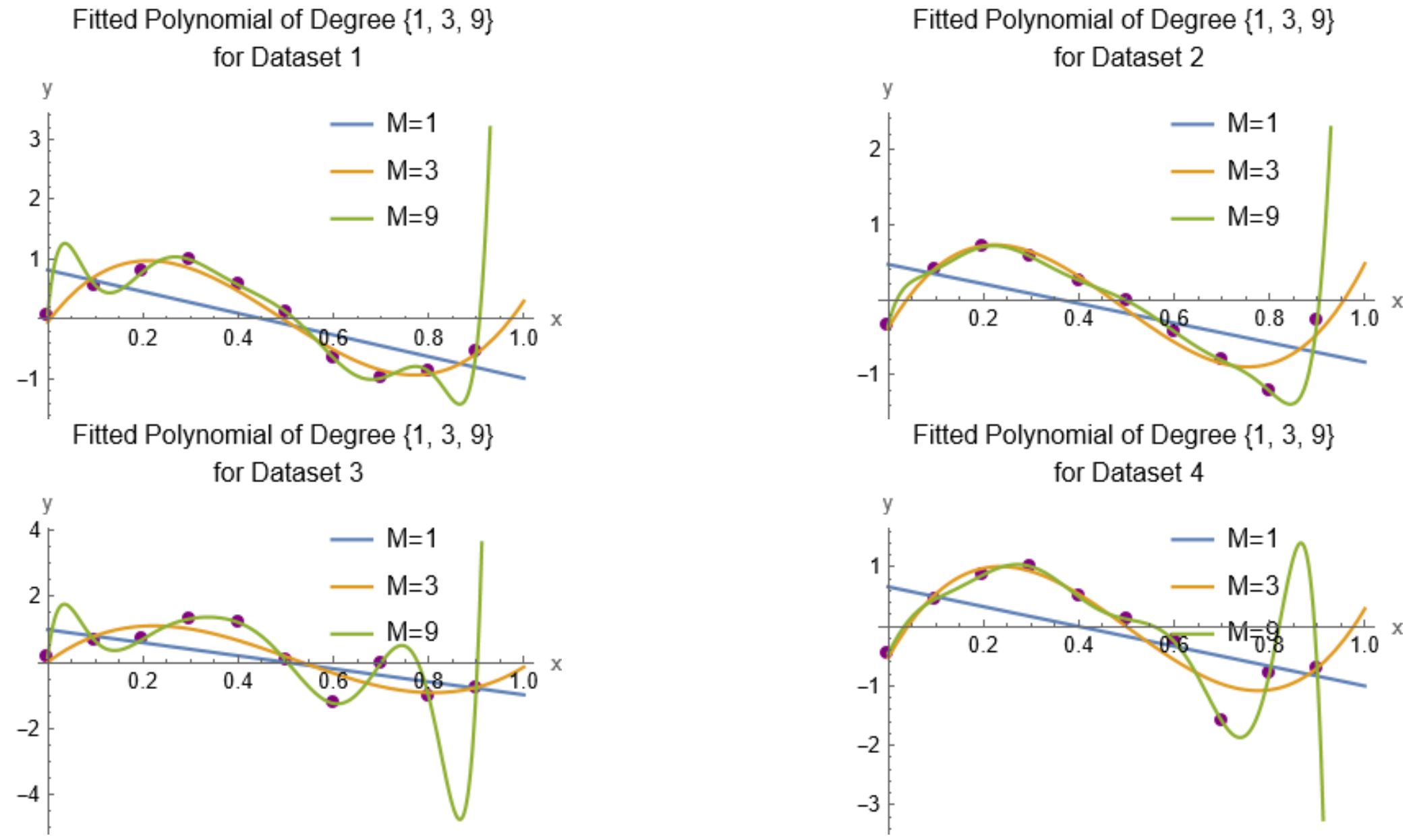

**Figure 6.5.** Comparison of Polynomial Regression Models on Sinusoidal Data with Varying Noise Levels: Fitted polynomials of degrees 1, 3, and 9 visualized alongside training data sets with noise levels ranging from 0.1 to 0.4. Each subplot represents a dataset, illustrating the trade-off between model complexity and accuracy in approximating the underlying sinusoidal function. The polynomial model with degree 9 exhibits significant variability across different training datasets.

- Models with a larger number of parameters are said to have high capacity, meaning they have the potential to learn complex patterns from data. However, high-capacity models require a larger amount of data to generalize well to unseen test data. Insufficient data can lead to overfitting, where the model performs well on the training set but poorly on new data.

On the one hand, the polynomial model is adept at closely modeling the true data distribution. However, it exhibits significant variability across different training datasets, see Figure 6.5. Consequently, while the linear model consistently provides similar predictions for the same test instance, the polynomial model yields disparate predictions depending on the choice of training data. This unpredictability in the polynomial model's behavior is undesirable for practitioners, who anticipate consistent predictions regardless of the training dataset used. This variance in predictions for identical test instances, but differing training datasets, characterizes the model's variance. Models with high variance tend to memorize random artifacts of the training data, leading to inconsistencies and inaccuracies in predicting unseen test instances. It's worth noting that a polynomial model with a higher degree inherently possesses greater power than a linear model, as the higher-order coefficients can be adjusted to enhance complexity. However, this enhanced power cannot be fully realized when the dataset is limited. This dilemma, balancing the model's power against its performance on limited data, encapsulates the bias-variance trade-off.

Identifying signs of overfitting is crucial for ensuring the reliability of a model's predictions. Several tell-tale indicators include:

- Inconsistent predictions across data sets: When a model trained on different datasets yields vastly different predictions for the same test instance, it suggests overfitting. This phenomenon indicates that the model is memorizing specific nuances of the training data rather than learning generalizable patterns applicable to unseen test instances.
- Large discrepancy in error rates: Another prominent sign of overfitting is a significant gap between the error rates of predicting training instances and those of unseen test instances. A considerable disparity between these error rates implies that the model has overly adapted to the intricacies of the training data, failing to generalize effectively to new data points.

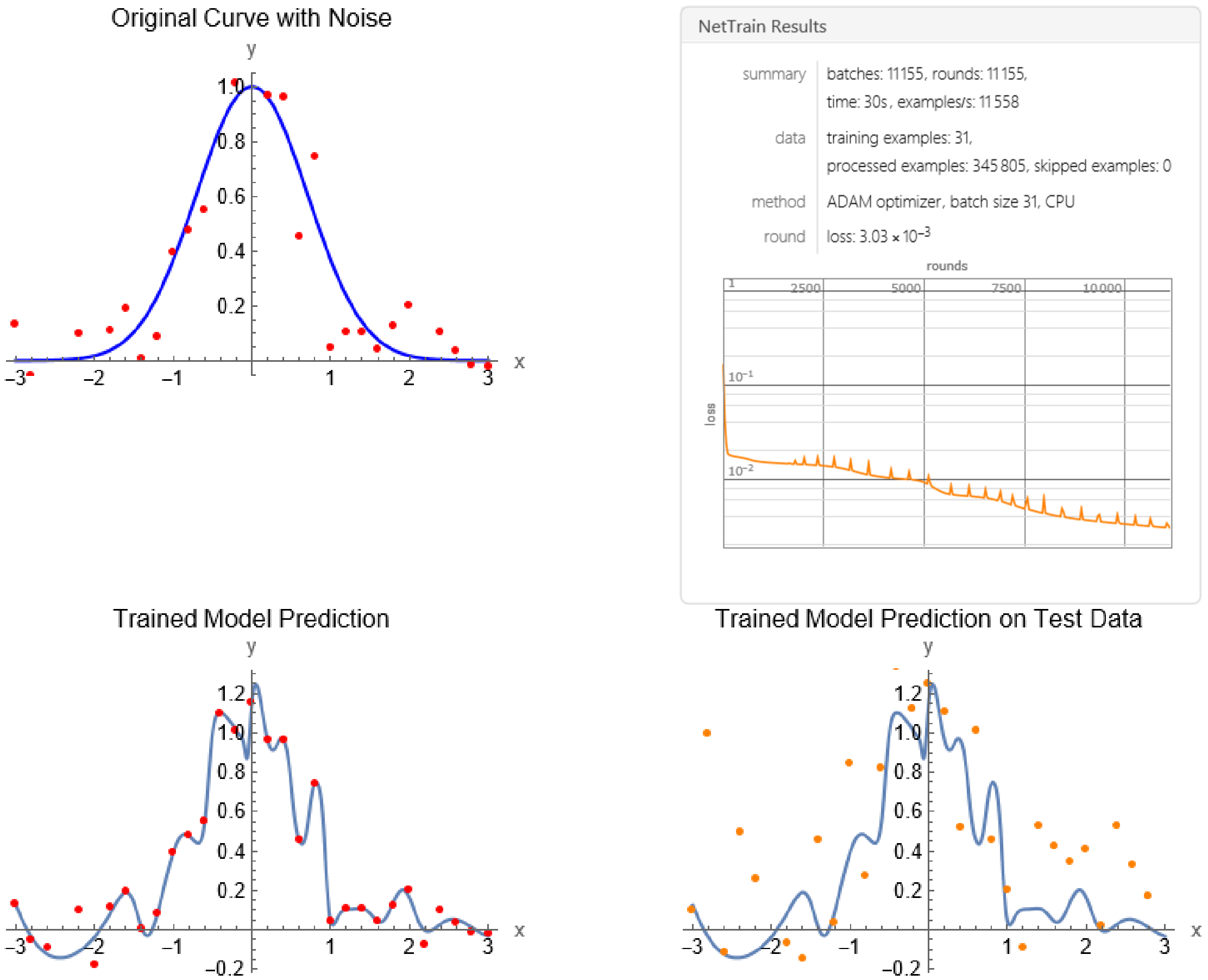

**Figure 6.6.** The top-left subplot displays the original mathematical function (blue curve) along with the noisy training data (red dots). The top-right subplot displays loss per round. The bottom-left subplot demonstrates the trained model's predictions overlaid on the training data. The bottom-right subplot shows the trained model's predictions on a separate test dataset (orange dots), highlighting the disparity between the model's performance on the training and test sets, indicative of potential overfitting.

Now, let's see how this looks in the context of a NN. We create a synthetic training dataset by taking noisy samples from a Gaussian curve, see Figure 6.6. Next, we train a net on those samples. The net has a much higher capacity than needed, meaning that it can model functions that are far more complex than necessary to fit the Gaussian curve. Say we have a NN with two hidden layers, each with 150 neurons, and uses the hyperbolic tangent, Tanh, AF for both hidden layers. The output layer has a single neuron, which suggests that the network is used for regression tasks where the goal is to predict a continuous output value. Because we know the form of the true model, it is visually obvious when overfitting occurs: the trained net produces a function that is quite different from the Gaussian, as it has "learned the noise" in the original data. To see this, we plot the function learned by the net alongside the original data, Figure 6.6 (bottom left). The quantitative way to demonstrate that overfitting has occurred is to test the net on data that comes from the same underlying distribution but that was not used to train the net. The fitted net is visually not a good explanation for the test samples, Figure 6.6 (bottom right). It's already quite apparent from these images that as the number of connections in our network increases, so does our propensity to overfit the data. Note that, we can similarly see the phenomenon of overfitting as we make our NNs deep.

This leads to four major observations.

- First, the machine learning engineer is always working with a direct trade-off between overfitting and model complexity. If the model isn't complex enough, it may not be powerful enough to capture all of the useful information necessary to solve a problem. However, if our model is very complex (especially if we have a limited amount of data at our disposal), we run the risk of overfitting. Deep learning takes the approach of solving complex problems with complex models and taking additional countermeasures to prevent overfitting. We'll see a lot of these measures in Chapter 7.

- Second, it is misleading to evaluate a model using the data we used to train it. We almost never train our model on the entire dataset. Instead, we split up our data into a training set and a test set. This enables us to make a fair evaluation of our model by directly measuring how well it generalizes on new data it has not yet seen.
- Third, it's quite likely that while we're training our data, there's a point in time when instead of learning useful features, we start overfitting to the training set. To avoid that, we want to be able to stop the training process as soon as we start overfitting to prevent poor generalization. To do this, we divide our training process into epochs. At the end of each epoch, we want to measure how well our model is generalizing. To do this, we use an additional validation set. At the end of an epoch, the validation set will tell us how the model does on data it has yet to see. If the accuracy on the training set continues to increase while the accuracy on the validation set stays the same (or decreases), it's a good sign that it's time to stop training because we're overfitting.
- Fourth, even though a natural way of avoiding overfitting is to simply build smaller networks (with fewer units and parameters), it has often been observed that it is better to build large networks and then regularize them to avoid overfitting. This is because large NNs can capture complex patterns in data due to their increased number of parameters and capacity to represent intricate relationships. Regularization techniques, Chapter 7, such as $L_1$ and $L_2$ regularizations are employed to prevent overfitting by imposing constraints on the network's parameters during training. Regularization encourages simpler models and reduces the risk of fitting noise in the training data. By starting with a large network and applying regularization, we strike a balance between allowing the model to capture complex patterns in the data while preventing overfitting. This approach enables the model to adapt its complexity based on the available data and the complexity of the underlying patterns. By giving the model, the option to decide its complexity level through regularization, we avoid the risk of underfitting, where the model is too simplistic to capture the underlying patterns in the data.

## 6.2 Statistical Learning Theory and Point Estimation

Typically, in the process of training a machine learning model, we begin with a designated training set, from which we derive insights and patterns to inform the model's learning. Throughout this training phase, we iteratively compute an error measure on the training set, commonly referred to as the training error. Our objective is to systematically minimize this training error through various optimization techniques. However, what distinguishes machine learning from mere optimization is our ultimate aim to ensure that the model's performance extends beyond the confines of the training data. We aspire for the model to exhibit robustness and effectiveness when confronted with previously unseen inputs. This aspiration is encapsulated by the concept of generalization error, often interchangeably termed as the test error [32]. In the realm of statistical learning theory, the challenge of improving performance on the test set despite only having access to the training set is addressed through a series of statistical principles and assumptions.

Central to these considerations is the concept of a data-generating process, which dictates how training and test data are produced. This process is governed by a probability distribution over datasets, representing the inherent variability and structure in the data. To facilitate analysis and modeling, we often rely on a set of assumptions collectively known as the IID (Independent and Identically Distributed) assumptions. Under the IID assumptions, we postulate that examples within each dataset are independent of one another, and both the training and test sets are drawn from the same underlying probability distribution. This alignment in distribution between the training and test sets allows us to describe the data-generating process using a single probability distribution, denoted as pdata. Consequently, every instance within both the training and test sets is generated from this shared distribution. This probabilistic framework, augmented by the IID assumptions, forms the cornerstone of statistical learning theory. It enables us to rigorously study the relationship between training error and test error, shedding light on the factors influencing model performance and generalization capabilities. By leveraging these principles and assumptions, we can devise strategies to optimize model performance on unseen data, even in the absence of direct observation [32].

One immediate connection we can observe between training error and test error is that the expected training error of a randomly selected model is equal to the expected test error of that model. Suppose we have a probability distribution $P(x, y)$ and we sample from it repeatedly to generate the training set and the test set. Indeed, the expected training set error is exactly the same as the expected test set error, because both expectations are formed using the same dataset sampling process. The only difference between the two conditions is the name we assign to the dataset we sampled.

In this section, we introduce several foundational concepts crucial for understanding machine learning models' behavior. These include parameter estimation, bias, and variance. Understanding these concepts is essential for formally characterizing notions of generalization, underfitting, and overfitting.

**Definition (Sampling Theory):** Sampling theory studies relationships between a population and samples drawn from the population.

Sampling theory [1] provides a framework for understanding how well a sample can represent an entire population. Sampling theory is useful in estimating unknown population quantities (such as population mean and variance), often called population parameters or briefly parameters, from knowledge of corresponding sample quantities (such as sample mean and variance), often called sample statistics or briefly statistics. In the context of machine learning, the training set is analogous to a sample, and the entire dataset (including potentially unseen future data) is like the population. In other words, just as sampling theory allows us to estimate population parameters from sample statistics, machine learning models use training data (sample) to learn parameters that generalize to new data (population). In statistics, the reliability of estimates (like mean or variance) is assessed through techniques like confidence intervals or hypothesis tests. In machine learning, model performance is evaluated using metrics like test error, and methods like cross-validation help ensure that the model's performance estimates are reliable. By understanding the principles of sampling theory, we can better appreciate the challenges and strategies in machine learning related to training and testing models.

If a number of samples, each of the same size $n$, is drawn from a given population (either with or without replacement) and for each sample, the value of the statistic (such as the mean and the standard deviation) is calculated, a series of values of a statistic will be obtained. If the number of samples is large, this may be arranged into a frequency table. The frequency distribution of the statistic that will be obtained if the number of samples, each of the same size, were large is called the sampling distribution of a statistic. If, for example, the particular statistic used is the sample mean, then the distribution is called the sampling distribution of means, or the sampling distribution of the mean. Similarly, we could have sampling distributions of standard deviations, variances, medians, etc. For each sampling distribution, we can compute the mean, standard deviation, etc. Thus, we can speak of the mean and standard deviation of the sampling distribution of means, etc.

**Definition (Sampling Distribution):** The probability distribution of a statistic is called a sampling distribution.

**Remarks:**

- The standard deviation of a sampling distribution of a statistic is often called its Standard Error (SE).
- The standard deviation measures the dispersion or amount of variability of individual data values from its mean. While standard error measures how far the value of the statistic is likely to be from the true parameter value. For example, the standard error of the sample mean measures how far the sample mean of the data is likely to be from the true population mean.
- There is an important way to find the sampling distribution of a statistic: use a simulation to approximate the distribution. That is, draw a large number of samples of size $n$, calculate the value of the statistic for each sample, and tabulate the results in a relative frequency histogram. When the number of samples is large, the histogram will be very close to the theoretical sampling distribution.

***Example 6.1***

A population consists of $N = 5$ numbers: 3, 6, 9, 12, 15. If a random sample of size $n = 3$ is selected without replacement, find the sampling distributions for the sample mean $\bar{X}$ and the sample median $m$.

***Solution***

The population contains five distinct numbers, and each is equally likely, with probability $p(X) = 1/5$. We can easily find the population mean and median as

$$\mu = \frac{3 + 6 + 9 + 12 + 15}{5}, \qquad M = 9.$$

To find the sampling distribution, we need to know what values of mean, $\bar{X}$, and median, $m$, can occur when the sample is taken. There are 10 possible random samples of size $n = 3$ and each is equally likely, with a probability $1/10$. These samples, along with the calculated values of $\bar{X}$ and $m$ for each, are listed in Table 6.2.

**Table 6.2.** Summary of sample data with calculated of $\bar{X}$ (sample mean) and $m$ (sample median).

| Sample | Sample elements | $\bar{X}$ | $m$ |
|---|---|---|---|
| 1 | 3,6,9 | 6 | 6 |
| 2 | 3,6,12 | 7 | 6 |
| 3 | 3,6,15 | 8 | 6 |
| 4 | 3,9,12 | 8 | 9 |
| 5 | 3,9,15 | 9 | 9 |
| 6 | 3,12,15 | 10 | 12 |
| 7 | 6,9,12 | 9 | 9 |
| 8 | 6,9,15 | 10 | 9 |
| 9 | 6,12,15 | 11 | 12 |
| 10 | 9,12,15 | 12 | 12 |

Using the values in Table 6.2, we can find the sampling distribution of $\bar{X}$ and $m$, shown in Table 6.3 and graphed in Figure 6.7.

**Table 6.3.** Sampling distributions of $\bar{X}$ and $m$.

| $\bar{X}$ | 6 | 7 | 8 | 9 | 10 | 11 | 12 | $m$ | 6 | 9 | 12 |
|---|---|---|---|---|---|---|---|---|---|---|---|
| $p(\bar{X})$ | 0.1 | 0.1 | 0.2 | 0.2 | 0.2 | 0.1 | 0.1 | $p(m)$ | 0.3 | 0.4 | 0.3 |

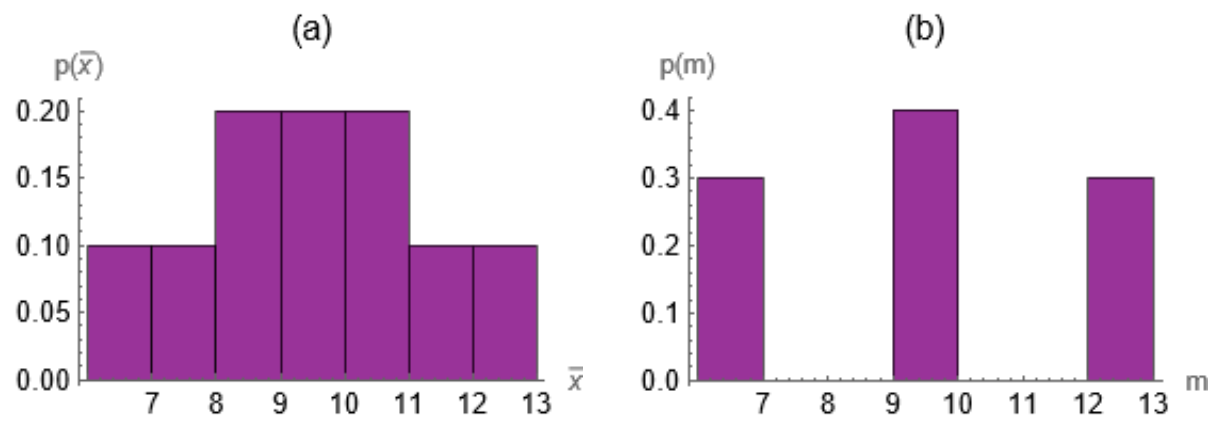

**Figure 6.7.** The sampling distribution of $\bar{X}$ and $m$.

**Definition (Point Estimate):** An estimate of a population parameter given by a single number is called a point estimate of the parameter.

**Definition (Interval Estimate):** An estimate of a population parameter given by two numbers between which the parameter may be considered to lie is called an interval estimate of the parameter.

In a practical situation, there may be several statistics that could be used as point estimators for a population parameter. To decide which of several choices is best, you need to know how the estimator behaves in repeated sampling, described by its sampling distribution.

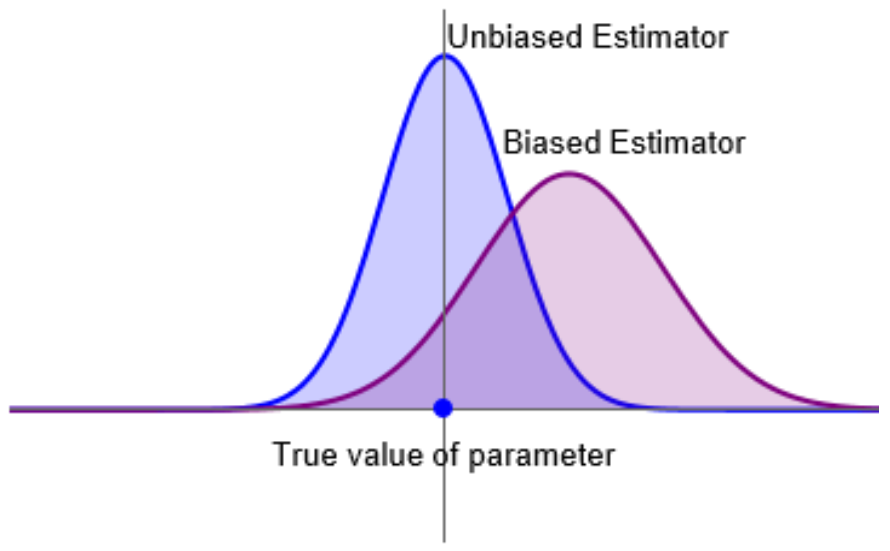

**Figure 6.8.** Distributions for biased and unbiased estimators.

### 6.2.1. Biased /Efficient Estimators

Sampling distributions provide information that can be used to select the best estimator. What characteristics would be valuable?

First, the sampling distribution of the point estimator should be centered over the true value of the parameter to be estimated. That is, the estimator should not constantly underestimate or overestimate the parameter of interest. Such an estimator is said to be unbiased.

**Definition (Unbiased and Biased Estimator):** If the mean of the sampling distribution of a statistic equals the corresponding population parameter, the statistic is called an unbiased estimator of the parameter; otherwise, it is called a biased estimator. The corresponding values of such statistics are called unbiased or biased estimates, respectively. The bias of an estimator is defined as [32]:

$$\text{bias}(\widehat{\boldsymbol{\theta}}_m) = \mathbb{E}(\widehat{\boldsymbol{\theta}}_m) - \boldsymbol{\theta}, \quad (6.3)$$

where the expectation is over the data (seen as samples from a random variable) and $\boldsymbol{\theta}$ is the true underlying value of $\boldsymbol{\theta}$ used to define the data-generating distribution. An estimator $\widehat{\boldsymbol{\theta}}_m$ is said to be unbiased if $\text{bias}(\widehat{\boldsymbol{\theta}}_m) = 0$, which implies that $\mathbb{E}(\widehat{\boldsymbol{\theta}}_m) = \boldsymbol{\theta}$.

The sampling distributions for an unbiased estimator and a biased estimator are shown in Figure 6.8. The sampling distribution for the biased estimator is shifted to the right of the true value of the parameter.

***Example 6.2***

The mean of the sampling distribution of means $\mu_{\bar{X}}$ is $\mu$, the population mean. Hence the sample mean $\bar{X}$ is an unbiased estimate of the population mean $\mu$.

***Example 6.3***

The mean of the sampling distribution of variances is

$$\mu_{S^2} = \frac{n-1}{n}\sigma^2,$$

where $\sigma^2$ is the population variance and $n$ is the sample size. Thus, the sample variance $S^2$ is a biased estimate of the population variance $\sigma^2$. By using the modified variance

$$\hat{S}^2 = \frac{n}{n-1}S^2,$$

we find $\mu_{\hat{S}^2} = \sigma^2$, so that $\hat{S}^2$ is an unbiased estimate of $\sigma^2$. However, $\hat{S}$ is a biased estimate of $\sigma$.

In the language of expectation, we could say that a statistic is unbiased if its expectation equals the corresponding population parameter. Thus, in the above examples, $\bar{X}$ and $\hat{S}^2$ are unbiased since $\mathbb{E}[\bar{X}] = \mu$ and $\mathbb{E}[\hat{S}^2] = \sigma^2$.

A second important characteristic is that the spread (as measured by the variance) of the estimator sampling distribution should be as small as possible. This ensures that, with a high probability, an individual estimate will fall close to the true value of the parameter. The sampling distributions for two unbiased estimators, one with a small variance and the other with a larger variance, are shown in Figure 6.9. Naturally, you would prefer the estimator with the smaller variance because the estimates tend to lie closer to the true value of the parameter than in the distribution with the larger variance.

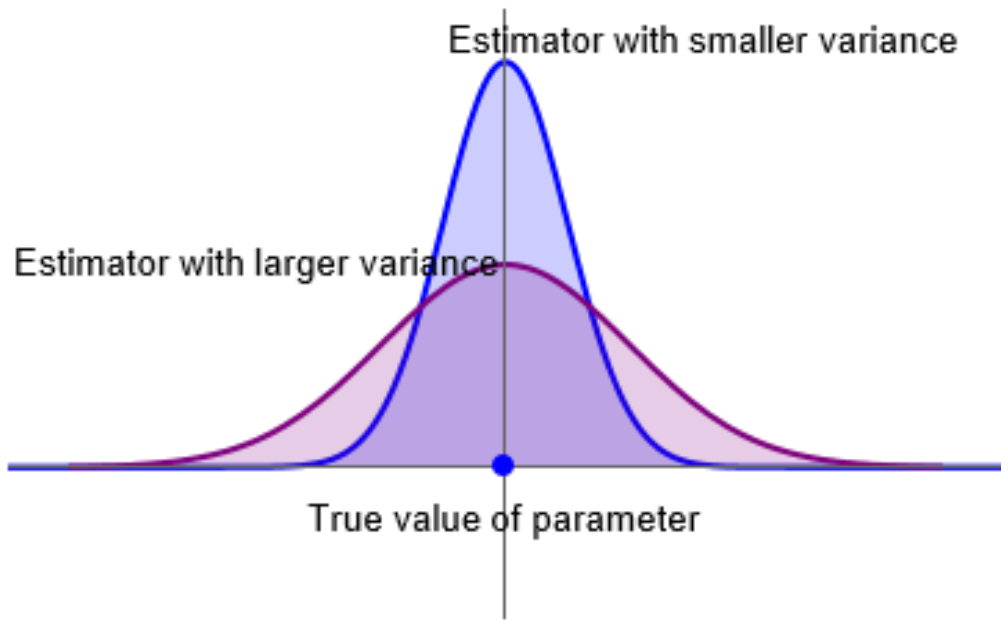

**Figure 6.9.** Comparison of estimator variability.

**Definition (Efficient and Inefficient Estimator):** If the sampling distributions of two statistics have the same mean (or expectation), then the statistic with the smaller variance is called an efficient estimator of the mean, while the other statistic is called an inefficient estimator. The corresponding values of the statistics are called efficient and inefficient estimates.

***Example 6.4***

The sampling distributions of the mean and median both have the same mean, namely, the population mean. However, the variance of the sampling distribution of means is smaller than the variance of the sampling distribution of medians. Hence the sample mean gives an efficient estimate of the population mean, while the sample median gives an inefficient estimate of it. Of all statistics estimating the population mean, the sample mean provides the best (or most efficient) estimate.

Our goal in statistics (and machine learning) is to make inferences about the population based on our sample data. Consider Figure 6.10 where the center represents the true value (population mean), we aim to predict or estimate. Surrounding this center, measurements or data points (sample outcomes) are depicted to show how different statistical models estimate this true value.

- High Bias: This is illustrated by measurements that consistently miss the true value, clustered away from the center. This represents a systematic error in the estimator's (model) assumptions, making it unable to capture the underlying population characteristics accurately, hence consistently estimating an erroneous outcome.
- High Variance: Illustrated by a wide scatter of measurements around their mean (not necessarily the true value), high variance indicates that the estimator (model) is sensitive to small fluctuations in the sample. This results in different samples leading to significantly different estimates, which may or may not be close to the population mean.
- Low Bias: Measurements are centered around the true value, indicating that the estimator (model) accurately estimates the population mean. The low bias suggests that the model's assumptions and structure properly capture the underlying patterns in the population.
- Low Variance: The tight clustering of measurements around their mean shows that the estimator (model) yields similar results across different samples. This consistency indicates that the estimator (model) is not overly sensitive to sample-specific fluctuations and reliably predicts the population mean.

An estimator (model) with high bias overlooks important complexities in the population, while an estimator (model) with high variance overfits the peculiarities of its sample. Ideally, a model should achieve low bias to ensure accuracy and low variance to ensure reliability across different samples. Figure 6.10 illustrates the dual challenges of bias and variance in statistical modeling. Recognizing and balancing these aspects is essential for developing models that are both accurate and reliable, providing dependable insights into the broader population based on sample data [63].

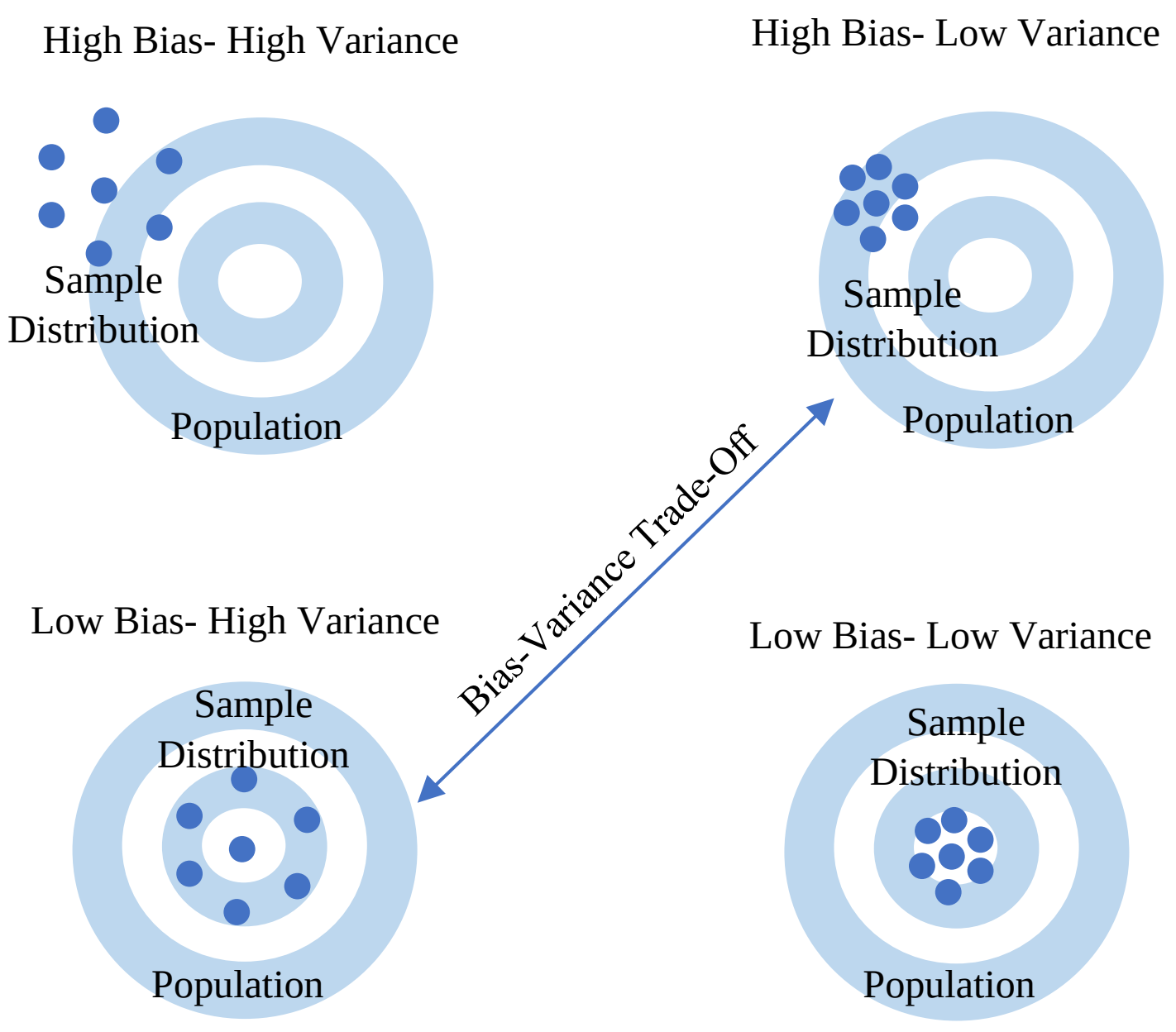

**Figure 6.10.** Statistical Insights into Bias and Variance.

Accuracy and precision are terms often used in the context of measurement, while bias and variance are concepts commonly discussed in the context of machine learning. Accuracy and precision are two measures of observational error. Accuracy refers to how close measurements or observations are to the true or actual value. When measurements are accurate, it means there's a small error between the average of these measurements and the true value. In a visual representation, accurate data points are centered around the true value, even if they are spread out. Precision, on the other hand, indicates how close measurements are to each other, regardless of whether they are near the true value. Precise measurements have less scatter among them, showing high repeatability or reproducibility. In simpler terms, given a statistical sample or set of data points from repeated measurements of the same quantity, the sample or set can be said to be accurate if its average is close to the true value of the quantity being measured, while the set can be said to be precise if their standard deviation is relatively small.

### 6.2.2. Maximum Likelihood Estimation (MLE)

Any statistic used to estimate the value of an unknown parameter $\theta$ is called an estimator of $\theta$. The observed value of the estimator is called the estimate. For instance, as we shall see, the usual estimator of the mean of a normal population, based on a sample $X_1, \ldots, X_n$ from that population, is the sample mean $\bar{X} = \sum_i X_i/n$. If a sample of size 3 yields the data $x_1 = 2$, $x_2 = 3$, $x_3 = 4$, then the estimate of the population mean, resulting from the estimator $\bar{X}$, is the value 3.

MLE is a method of estimating the parameters of an assumed probability distribution, given some observed data. The MLE method is based on the principle that the values of the parameters should be chosen to make the observed data most probable. This is achieved by maximizing a likelihood function so that, under the assumed statistical model, the observed data is most probable.

Let us suppose we have a statistical model with a set of parameters $\boldsymbol{\theta}$ ($\boldsymbol{\theta}$ is a vector of parameters) and a sample of observed data $X = \{x_1, x_2, \ldots, x_n\}$. Let $f(x_1, x_2, \ldots, x_n; \boldsymbol{\theta})$ denote the joint PMF of the RVs $X_1, \ldots, X_n$ when they are discrete, and let it be their joint PDF when they are jointly continuous RVs. Because $\boldsymbol{\theta}$ is assumed unknown, we also write $f$ as a function of $\boldsymbol{\theta}$. The likelihood function, denoted by $L(X|\boldsymbol{\theta})$, is a measure of how likely the observed data is under the given parameter values. Now since $L(X|\boldsymbol{\theta}) = f(x_1, x_2, \ldots, x_n; \boldsymbol{\theta})$ represents the likelihood that the values $x_1, \ldots, x_n$ will be observed when $\boldsymbol{\theta}$ is the true value of the parameter, it would seem that a reasonable estimate of $\boldsymbol{\theta}$ would be that value yielding the largest likelihood of the observed values. The goal of MLE is to find the values

**Procedure 6.1:** MLE

1. Start by defining the probability distribution that you believe represents the data you are working with. This distribution could be Gaussian (normal), binomial, Poisson, etc., depending on the nature of your data.
2. The likelihood function represents the probability of observing your data given a set of parameters. It is derived from the probability distribution you defined in step 1. The likelihood function is typically denoted as $L(X|\boldsymbol{\theta}) = f(x_1;\boldsymbol{\theta})f(x_2;\boldsymbol{\theta})\ldots f(x_n;\boldsymbol{\theta})$, where $f(x_i;\boldsymbol{\theta})$ is the PDF or PMF of the model and $\boldsymbol{\theta}$ represents the parameters of the distribution.
3. To simplify the calculations, it is common to take the natural logarithm of the likelihood function. This step does not change the location of the maximum, as the logarithm is a monotonic function.
4. Differentiate the logarithm of the likelihood function with respect to the parameters $\boldsymbol{\theta}$. This step helps find the maximum point in the parameter space.
5. Set the derivative obtained in step 4 to zero and solve for the parameters. This identifies the values of $\boldsymbol{\theta}$ that maximize the likelihood function.
6. Calculate the second derivative of the log-likelihood function with respect to the parameters. This is known as the Hessian matrix. Evaluate the second derivative at the values of $\boldsymbol{\theta}$ obtained in step 5. Verify that the Hessian matrix is negative definite or negative semi-definite. This condition ensures that the maximum point found in step 5 is indeed a maximum and not a minimum or saddle point. You can use also the Mathematica `FindMaximum` function to find values of $\boldsymbol{\theta}$ that maximize the likelihood function.
7. Solve the equations obtained from step 5 to obtain the MLEs for the parameters of the distribution.

of $\boldsymbol{\theta}$ that maximize the likelihood function. The likelihood function is typically defined as:

$$L(X|\boldsymbol{\theta}) = f(x_1;\boldsymbol{\theta})f(x_2;\boldsymbol{\theta})\ldots f(x_n;\boldsymbol{\theta}), \tag{6.4}$$

where $f(x_i;\boldsymbol{\theta})$ is the PDF or PMF of the model. To find the MLEs, we seek the value of $\boldsymbol{\theta}$ that maximizes the likelihood function. In determining the value of $\boldsymbol{\theta}$, it is often useful to use the fact that $f(x_1, x_2, \ldots, x_n;\boldsymbol{\theta})$ and $\log[f(x_1, x_2, \ldots, x_n;\boldsymbol{\theta})]$ have their maximum at the same value of $\boldsymbol{\theta}$. The log-likelihood function, denoted by $\ell(X|\boldsymbol{\theta})$, is:

$$\ell(X|\boldsymbol{\theta}) = \log[L(X|\boldsymbol{\theta})] = \log[f(x_1;\boldsymbol{\theta})] + \log[f(x_2;\boldsymbol{\theta})] + \ldots + \log[f(x_n;\boldsymbol{\theta})]. \tag{6.5}$$

Once we have the log-likelihood function, we differentiate it with respect to $\boldsymbol{\theta}$, set the derivative to zero, and solve for $\boldsymbol{\theta}$. Procedure 6.1 represents how MLE typically works.

A few more important aspects and properties of MLEs are:

- Consistency:
  Under certain regularity conditions, MLEs are consistent, meaning that as the sample size increases, the estimated parameter values converge to the true values. This property ensures that the estimates become more accurate with larger amounts of data.
- Efficiency:
  MLEs are often asymptotically efficient, which means that they achieve the smallest possible asymptotic variance among all consistent estimators. In simpler terms, MLEs tend to have smaller SEs compared to other estimators, making them more precise.
- Computational methods:
  In some cases, finding the MLE analytically may be challenging or impossible. In such situations, numerical optimization algorithms, such as the Newton-Raphson method or the expectation-maximization algorithm, are commonly used to find the MLEs.
- Applications:
  MLE is a versatile method used in various fields, including statistics, econometrics, machine learning, and many other areas of research. It is employed to estimate parameters in a wide range of models, including linear regression, logistic regression, survival analysis, mixed-effects models, and more.

## 6.3 The Bias-Variance Trade-Off (Decomposition Theorem)

The bias-variance tradeoff is a central problem in supervised learning. In statistics and machine learning, the bias-variance tradeoff describes the relationship between a model's complexity, the accuracy of its predictions, and how well it can make predictions on previously unseen data that were not used to train the model. Ideally, one wants to choose a model that both accurately captures the regularities in its training data, but also generalizes well to unseen data. Unfortunately, it is typically impossible to do both simultaneously.

The basic intuition behind the bias-variance tradeoff can be illustrated by a hypothetical scenario. Let us compare it to planning a road trip to an unfamiliar destination, focusing on how you might choose your route. Imagine planning your trip using a very basic map that offers only major highways and lacks detail about current road conditions or points of interest. This route is straightforward, perhaps too much so. It's akin to following a set of overly simplistic instructions that don't account for variable elements such as traffic jams or roadwork. In machine learning, a high-bias scenario occurs when a model is too simple, effectively ignoring the finer details and complexities of the data it trains on. Just as consistently following the same major highway might get you to your destination, it may not be the most efficient or enjoyable journey. This represents a model that doesn’t adapt well to the underlying data, potentially leading to underfitting. Now, consider using a highly detailed, dynamic map that reacts to every minor change in driving conditions. This map might reroute you frequently, taking you down less known paths or scenic routes based on real-time updates about slight delays or minor detours. While this approach adapts meticulously to current conditions, it can lead to inconsistency and unpredictability in your journey. Similarly, in machine learning, a high variance model pays excessive attention to the training data, including noise and outliers, which can result in a model that fits this particular set of data very well but fails to perform adequately on new, unseen datasets.

The key to a successful road trip, and to machine learning, is using a navigation tool that effectively balances detail and flexibility. This ideal map would provide enough information to steer clear of major issues like traffic congestion and closed roads but wouldn’t be so sensitive as to suggest a new route for every temporary slowdown. The goal is to choose a path that’s both reliable and adaptable, avoiding the extremes of sticking rigidly to major highways or changing course at every new piece of information. In machine learning, this balance allows a model to capture the essential patterns in the data, adjusting to significant anomalies or trends without overfitting the minutiae. Just as a well-planned route leads to a smooth and efficient road trip, a well-tuned model reaches the best generalization performance, making accurate predictions on new, unseen data. By navigating the bias-variance trade-off effectively, data scientists, like skilled travelers, can ensure that their models reach their destination not only successfully but also efficiently.

Mathematically, suppose that we have a training set consisting of a set of points $x_1, \ldots, x_n$ and real values $y_i$ associated with each point $x_i$. We assume that the data is generated by a function $f(x_i)$ such as

$$y_i = f(x_i) + \epsilon_i, \tag{6.6}$$

where the noise, $\epsilon_i$, has zero mean, $\mathbb{E}[\epsilon_i] = 0$. We want to find a function $\hat{f}(x_i; D)$, that approximates the true function $f(x_i)$, as well as possible, using some learning algorithm based on a training dataset (sample) $D = \{(x_1, y_1), \ldots, (x_n, y_n)\}$. We make "as well as possible" precise by measuring the MSE between $y$ and

$$\hat{y}_i = \hat{f}(x_i; D), \tag{6.7}$$

we want $(y_i - \hat{y}_i)^2$ to be minimal, both for $\{x_1, \ldots, x_n\}$ and for points outside of our sample. Of course, we cannot hope to do so perfectly, since the $y_i$ contain noise $\epsilon_i$; this means we must be prepared to accept an irreducible error in any function we come up with. The MSE of the learning algorithm $\hat{f}(\cdot, D)$ is defined over the set of instances $x_1 \ldots x_t$ as follows [69]:

$$\text{MSE} = \frac{1}{t}\sum_{i=1}^{t}(\hat{y}_i - y_i)^2 = \frac{1}{t}\sum_{i=1}^{t}\left(\hat{f}(x_i, D) - f(x_i) - \epsilon_i\right)^2. \tag{6.8}$$

**Theorem 6.1 (Bias-Variance Decomposition):** The expectation of the MSE can be decomposed into three components: the squared bias, the variance, and the irreducible error (noise)

$$\mathbb{E}[\text{MSE}] = \underbrace{\frac{1}{t}\sum_{i=1}^{t}\{f(x_i) - \mathbb{E}[\hat{f}(x_i, D)]\}^2}_{\text{Bias}^2} + \underbrace{\frac{1}{t}\sum_{i=1}^{t}\mathbb{E}\left[\{\hat{f}(x_i, D) - \mathbb{E}[\hat{f}(x_i, D)]\}^2\right]}_{\text{Variance}} + \underbrace{\frac{1}{t}\sum_{i=1}^{t}\mathbb{E}[\epsilon_i^2]}_{\text{Noise}}. \quad (6.9)$$

The expectation ranges over different choices of the training set.

**Proof:**

To accurately estimate error in a manner that does not depend on a particular choice of training dataset, it is best to calculate the expected error across various training dataset selections. For notational convenience, we abbreviate $f = f(x_i)$ and $\hat{f} = \hat{f}(x_i, D)$. We have

$$\begin{aligned}
\mathbb{E}[\text{MSE}] &= \frac{1}{t}\sum_{i=1}^{t}\mathbb{E}\left[(\hat{f} - f - \epsilon_i)^2\right] \\
&= \frac{1}{t}\sum_{i=1}^{t}\mathbb{E}\left[(\hat{f} - f)^2 + \epsilon_i^2 - 2(\hat{f} - f)\epsilon_i\right] \\
&= \frac{1}{t}\sum_{i=1}^{t}\mathbb{E}\left[(\hat{f} - f)^2\right] + \mathbb{E}[\epsilon_i^2] - 2\,\mathbb{E}[\hat{f} - f]\,\underbrace{\mathbb{E}[\epsilon_i]}_{0} \\
&= \frac{1}{t}\sum_{i=1}^{t}\mathbb{E}\left[(\hat{f} - f)^2\right] + \frac{1}{t}\sum_{i=1}^{t}\mathbb{E}[\epsilon_i^2] \\
&= \frac{1}{t}\sum_{i=1}^{t}\mathbb{E}\left[\{(f - \mathbb{E}[\hat{f}]) + (\mathbb{E}[\hat{f}] - \hat{f})\}^2\right] + \frac{1}{t}\sum_{i=1}^{t}\mathbb{E}[\epsilon_i^2] \\
&= \frac{1}{t}\sum_{i=1}^{t}\mathbb{E}\left[(f - \mathbb{E}[\hat{f}])^2 + (\mathbb{E}[\hat{f}] - \hat{f})^2 + 2(f - \mathbb{E}[\hat{f}])(\mathbb{E}[\hat{f}] - \hat{f})\right] + \frac{1}{t}\sum_{i=1}^{t}\mathbb{E}[\epsilon_i^2] \\
&= \frac{1}{t}\sum_{i=1}^{t}\mathbb{E}\left[(f - \mathbb{E}[\hat{f}])^2\right] + \mathbb{E}[2(f - \mathbb{E}[\hat{f}])(\mathbb{E}[\hat{f}] - \hat{f})] + \mathbb{E}\left[(\mathbb{E}[\hat{f}] - \hat{f})^2\right] + \frac{1}{t}\sum_{i=1}^{t}\mathbb{E}[\epsilon_i^2] \\
&= \frac{1}{t}\sum_{i=1}^{t}\mathbb{E}\left[(f - \mathbb{E}[\hat{f}])^2\right] + \frac{2}{t}\sum_{i=1}^{t}\mathbb{E}[(f - \mathbb{E}[\hat{f}])(\mathbb{E}[\hat{f}] - \hat{f})] + \frac{1}{t}\sum_{i=1}^{t}\mathbb{E}\left[(\mathbb{E}[\hat{f}] - \hat{f})^2\right] + \frac{1}{t}\sum_{i=1}^{t}\mathbb{E}[\epsilon_i^2] \\
&= \frac{1}{t}\sum_{i=1}^{t}\mathbb{E}\left[(f - \mathbb{E}[\hat{f}])^2\right] + \frac{2}{t}\sum_{i=1}^{t}(\mathbb{E}[f] - \mathbb{E}[\hat{f}])\underbrace{(\mathbb{E}[\hat{f}] - \mathbb{E}[\hat{f}])}_{0} + \frac{1}{t}\sum_{i=1}^{t}\mathbb{E}\left[(\mathbb{E}[\hat{f}] - \hat{f})^2\right] + \frac{1}{t}\sum_{i=1}^{t}\mathbb{E}[\epsilon_i^2] \\
&= \underbrace{\frac{1}{t}\sum_{i=1}^{t}(f - \mathbb{E}[\hat{f}])^2}_{\text{Bias}^2} + \underbrace{\frac{1}{t}\sum_{i=1}^{t}\mathbb{E}\left[(\hat{f} - \mathbb{E}[\hat{f}])^2\right]}_{\text{Variance}} + \underbrace{\frac{1}{t}\sum_{i=1}^{t}\mathbb{E}[\epsilon_i^2]}_{\text{Noise}}.
\end{aligned}$$

Where $\mathbb{E}[f] = f$, since $f$ does not depend on the data.

■

The bias-variance trade-off states that the total error of a model can be decomposed into three components: bias, variance, and irreducible error.

Bias: Bias in a machine learning model refers to the error introduced by approximating a real-world problem, which typically involves some degree of complexity, with a simpler model. This error is systematic in nature, even if the model has access to an infinite source of training data, the bias cannot be removed. High bias can lead a model to miss relevant relations between features and target outputs (underfitting). Essentially, a high-bias model is overly

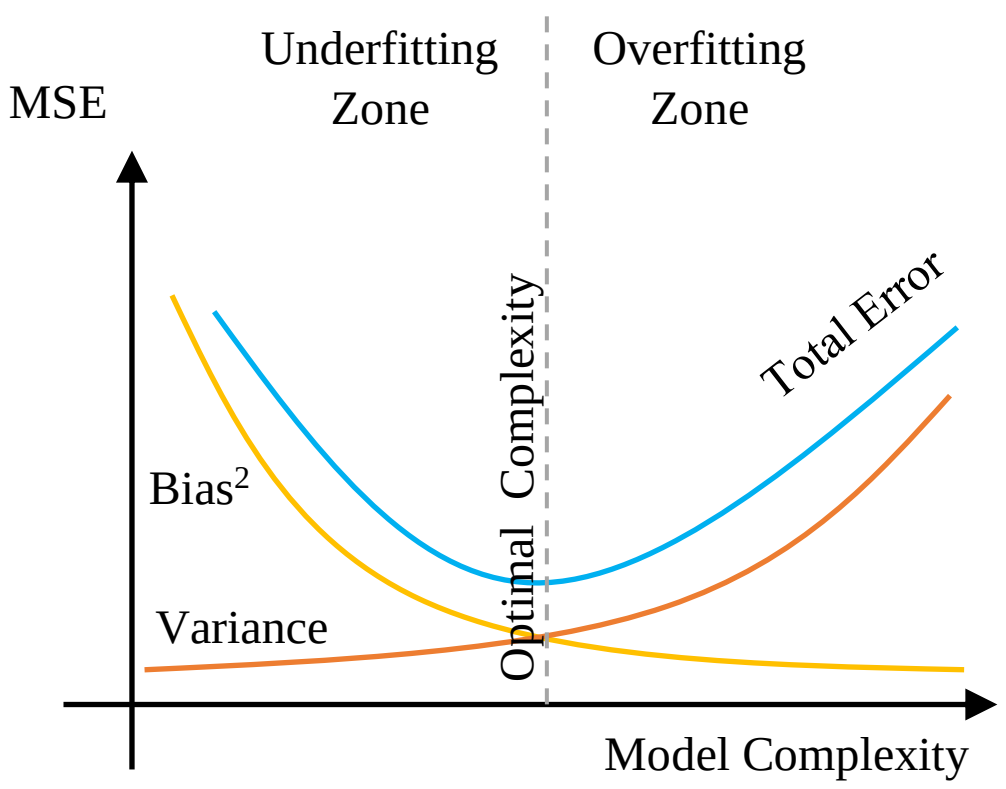

**Figure 6.11.** The variation of Bias and Variance with the model complexity. This is similar to the concept of overfitting and underfitting. More complex models overfit while the simplest models underfit.

simplistic—it doesn't learn enough from the training data. For instance, consider the scenario depicted in Figure 6.5 of a linear model attempting to fit a dataset with a slight curvature. No matter how much training data is available, a linear model inherently lacks the flexibility to capture the curved trend in the data fully. This limitation results in a consistent prediction error at specific points, regardless of the training sample used.

Variance: Variance in a machine learning model refers to the model's sensitivity to small fluctuations in the training dataset. High variance is typically indicative of a model that can adapt too well to the training data, capturing random noise as if it were true underlying patterns, which is known as overfitting. This is particularly problematic in models with many parameters relative to the available data. A model with high variance will yield different predictions for the same test instance when trained on different subsets of data. For example, in Figure 6.5, consider how the polynomial model behaves compared to the linear model. The linear model, due to its simplicity, provides similar predictions at the same point regardless of variations in the training data. In contrast, the polynomial model, which is more complex and has higher degrees of freedom, shows considerable variability in its predictions at the same point depending on the specific training data used. These predictions by the polynomial model can often be not just diverse, but also far from the actual value, demonstrating the model's high variance. Thus, in Figure 6.5, the polynomial model is characterized by higher variance compared to the linear model.

Noise: Noise refers to random fluctuations and errors inherently present in the data, which cannot be attributed to any predictable or systematic error by the model. It represents the irreducible error that even the most perfect model cannot eliminate. This error can originate from various sources, including measurement errors, incomplete data coverage, or the chaotic nature of the system being modeled. In the context of Figure 6.5, noise is illustrated by the scatter of data points around the true model curve. If there were no noise, every data point would align perfectly with the curved line, which depicts the true underlying relationship. Instead, the scatter of points indicates the presence of noise, as each point deviates from this curve in a seemingly random manner. This deviation highlights the inevitable discrepancies that occur due to noise in real-world data.

The bias-variance trade-off suggests that there is a trade-off between bias and variance [32]. As you reduce bias (by increasing model complexity), you typically increase variance, and vice versa. The goal is to find the right balance that minimizes the total error, which is the sum of bias, variance, and irreducible error. This trade-off between bias and variance with increasing model complexity is illustrated in Figure 6.11. The optimal model complexity is the point where the trade-off between bias and variance is balanced, leading to the lowest total error. At this point, the model is complex enough to capture the underlying patterns in the data but not so complex that it overfits the noise. Finding the optimal model complexity often involves techniques such as model selection, hyperparameter tuning, and cross-validation. By systematically evaluating the model's performance across different levels of complexity, one can identify the model that achieves the best balance between bias and variance for a given dataset.

## 6.4 Training, Testing, and Validation Sets

There are several practical issues in the training of NN models that one must be careful of because of the bias-variance trade-off. The first of these issues is associated with model tuning and hyperparameter choice. For example, if one tuned the NN with the same data that were used to train it, one would not obtain very good results because of overfitting. Therefore, the hyperparameters are tuned on a separate held-out set from the one on which the weight parameters on the NN are learned.

Given a labeled data set, one needs to use this resource for training, tuning, and testing the accuracy of the model. Clearly, one cannot use the entire resource of labeled data for model building (i.e., learning the weight parameters). For example, using the same data set for both model building and testing grossly overestimates the accuracy. Furthermore, the portion of the data set used for model selection and parameter tuning also needs to be different from that used for model building. A common mistake is to use the same data set for both parameters tuning and final evaluation (testing). A given data set should always be divided into three parts defined according to the way in which the data are used [56-70]:

1. Training data: The training data is used to build the model by adjusting its parameters (such as weights and biases) based on the patterns present in the data. This process involves iterative optimization techniques, such as gradient descent, to minimize the difference between the model's predictions and the actual target values. Multiple models may be constructed using different hyperparameters (e.g., learning rate, number of layers, etc) and architectures (e.g., NN structures). These variations allow for the exploration of different modeling approaches to find the one that best fits the data. After training multiple models, the next step is to select the best-performing algorithm. This process is known as model selection. However, it's crucial to note that the evaluation of these algorithms is not performed on the training data itself. Instead of evaluating model performance on the training data, a separate validation data set is used. This ensures that the evaluation is conducted on unseen data, preventing the model from memorizing the training set and overfitting.
2. Validation data: The validation data set is crucial for model selection and parameter tuning. Just like the test data set, the validation data set provides an independent sample of data that the model hasn't seen during training. This allows for a fair assessment of model performance without the risk of overfitting the training data. After training multiple models on the training data, each with different hyperparameters and architectures, the performance of these models is evaluated using the validation set. This evaluation typically involves measuring metrics such as accuracy, precision, recall, or F1 score, depending on the nature of the problem. The validation set is used to tune hyperparameters and other design choices that impact the performance of the model. For example, hyperparameters like learning rate, or the number of units in a NN layer can significantly affect the model's performance. By testing different combinations of these parameters on the validation set, the optimal configuration can be determined. Essentially, one could regard validation data as a test dataset employed for fine-tuning the parameters of the algorithm. In a broader sense, one could argue that the validation data is indeed part of the overall training process because it does influence the final model indirectly through the parameter tuning and model selection steps. Note that, it's essential to avoid using the test set for model selection, as doing so may lead to overfitting of the test data.
3. Testing data: The testing data set is reserved for evaluating the performance of the final, tuned model. Unlike the validation set, which is used iteratively for parameter tuning and model selection, the testing data set is only used once at the very end of the model development process. It's crucial that the testing data set remains unseen during the process of parameter tuning and model selection to prevent overfitting. If the model is tuned based on its performance on the testing data, it may lead to an overly optimistic assessment of its generalization performance. Using the testing data to adjust the model after evaluating its performance can contaminate the results. Any adjustments made to the model based on insights gained from the testing data compromise the integrity of the evaluation process and may lead to biased performance estimates. Adhering to the principle of using the testing data set only once is a stringent but essential requirement in machine learning evaluation. This principle ensures that the final evaluation of the model's performance is unbiased and reflects its true generalization ability.

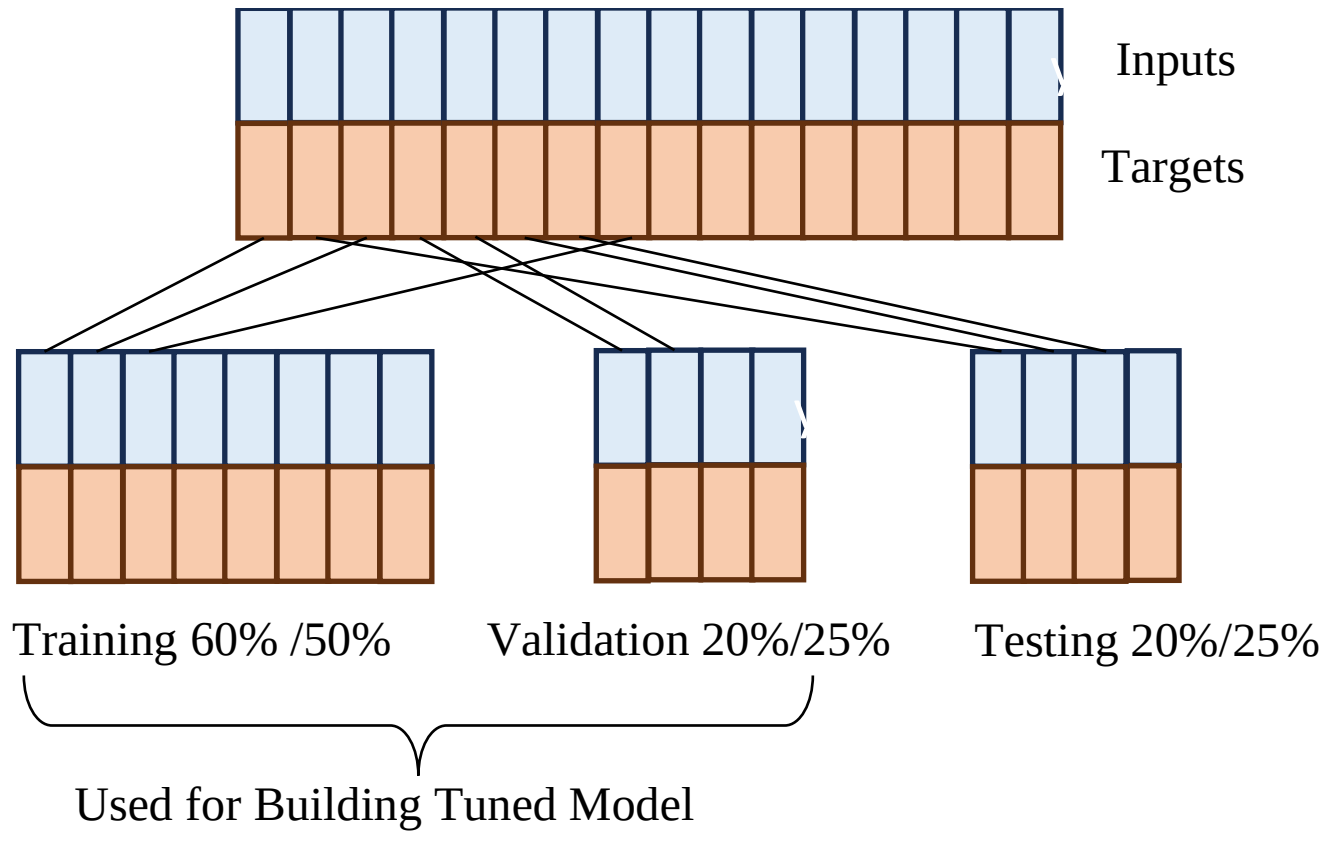

**Figure 6.12.** The dataset is split into different sets, some for training, some for validation, and some for testing.

The division of the labeled data set into training, validation, and test data is shown in Figure 6.12. This division is becoming expensive in data, especially since for supervised learning it all has to have target values attached, and it is not always easy to get accurate labels.

Clearly, each algorithm is going to need some reasonable amount of data to learn from. Precise needs may vary, but generally, the more data the algorithm encounters, the higher the likelihood it has encountered examples of each possible type of input. However, it's important to note that more data also increases the computational time required for learning. Generally, the exact proportion of training to testing to validation data is up to you, but it is typical to do something like 50:25:25% if you have plenty of data, and 60:20:20 if you don't. This ratio was established in an era when data sets were relatively small, and there was a need to strike a balance between model training, evaluation, and validation. However, it's essential to recognize that this ratio should not be viewed as a strict rule. The appropriateness of the division depends on the size and nature of the data set, as well as the specific requirements of the machine learning task. In the modern era, where large labeled data sets are increasingly available, there may be less need for a strict division of data. With ample data available for model training, it becomes more feasible to allocate a larger proportion of the data for this purpose. When a very large data set is available, it makes sense to use as much of it for model building as possible. This is because even a modest number of examples in the validation and testing sets can provide accurate estimates of model performance.

Ultimately, the division of data should be flexible and adaptable to the specific characteristics of the data set and the requirements of the machine learning task at hand. It's important to strike a balance between having enough data for training, validation, and testing while ensuring that each set is representative of the overall dataset to achieve reliable model performance.

How you split your dataset into training and testing sets can significantly impact the performance and generalization of your machine-learning model. The order in which the data points are presented might introduce bias or skewness in the learning process, especially if the data points are organized in a specific way, such as by class labels. Many datasets are presented with the first set of data points being in class 1, the next in class 2, and so on. If you pick the first few points to be the training set, the next the test set, etc., then the results are going to be pretty bad, since the training did not see all the classes.

To mitigate this issue, one common approach is to randomly shuffle the dataset before splitting it into training and testing sets. This ensures that the data points are distributed randomly across different classes and avoids any bias introduced by the ordering of the data. Another method is to randomly assign each data point to either the training or testing set, Figure 6.12, ensuring that each set contains a representative sample from all classes. This approach helps in achieving a more balanced distribution of data across different sets and reduces the risk of biased learning.

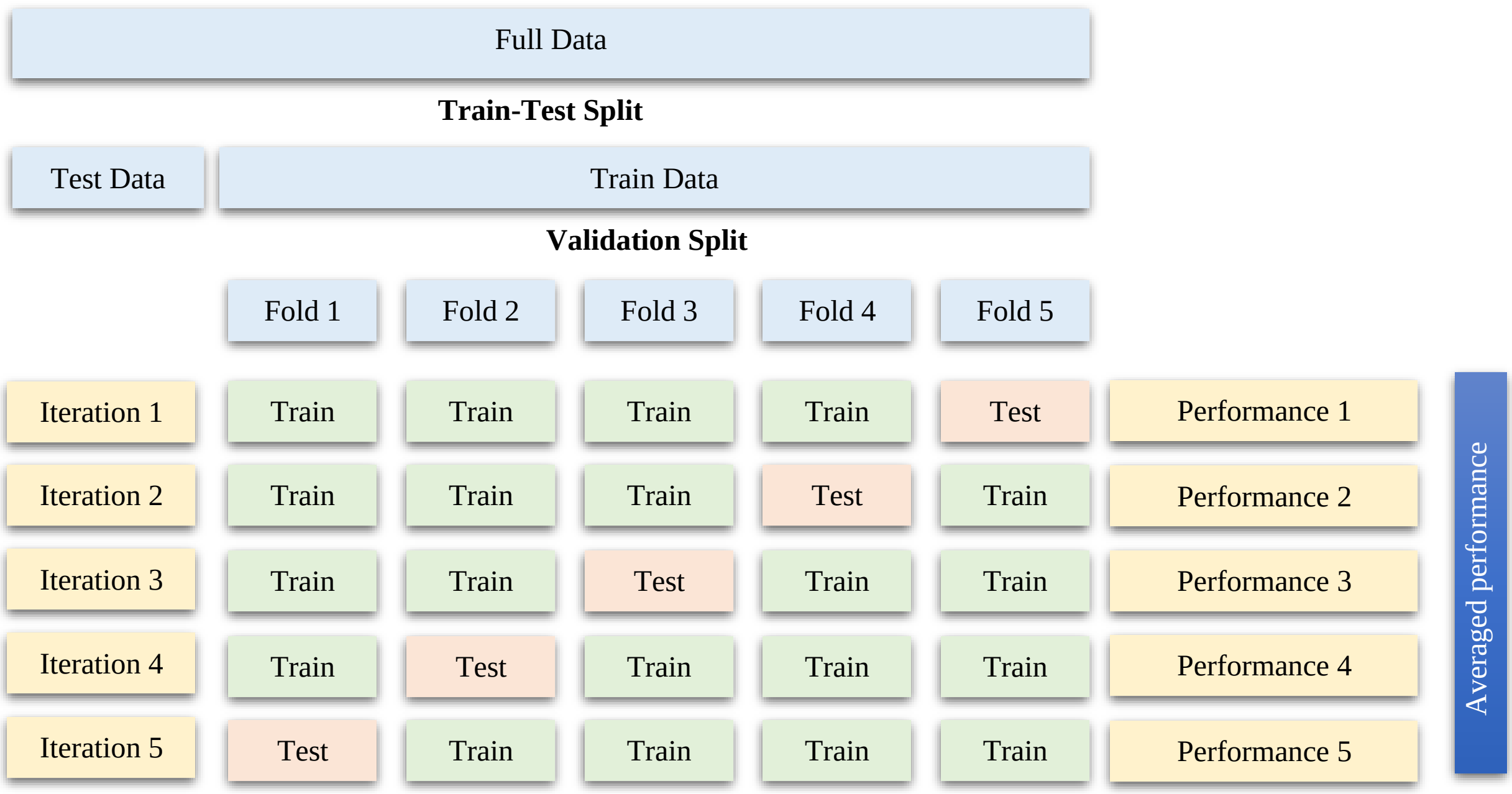

**Figure 6.13.** Five-fold cross-validation diagram. The dataset was divided into five parts, four of them were taken as training data in turn, and one was used as validation data for validation. The average value of the performance of the five-group validation results is calculated as an estimate of the model accuracy.

### 6.4.1. Cross-Validation

One unfortunate effect of introducing the validation set is that we can now use only 60% of the original data to train the weights in our network. This can be a problem if we have a limited amount of training data to begin with. We can address this problem using a technique known as cross-validation, which avoids holding out parts of the dataset to be used as validation data but at the expense of additional computation. We focus on one of the most popular cross-validation techniques, known as $k$-fold cross-validation. We start by splitting our data into a training set and a test set, using something like an 80/20 split. The test set is not used for training or hyperparameter tuning but is used only in the end to establish how good the final model is. We further split our training dataset into $k$ similarly sized pieces known as folds, where a typical value for $k$ is a number between 5 and 10.

We can now use these folds to create $k$ instances of a training set and validation set by using $k - 1$ folds for training and 1 fold for validation. That is, in the case of $k = 5$, we have five alternative instances of training/validation sets. The first one uses folds 1, 2, 3, and 4 for training and fold 5 for validation, the second instance uses folds 1, 2, 3, and 5 for training and fold 4 for validation, and so on.

Let us now use these five instances of train/validation sets to both train the weights of our model and tune the hyperparameters. Instead of training each configuration once, we train each configuration $k$ times with our $k$ different instances of train/validation data. Each of these $k$ instances of the same model is trained from scratch, without reusing weights that were learned by a previous instance. That is, for each configuration, we now have $k$ measures of how well the configuration performs. We now compute the average of these measures for each configuration to arrive at a single number for each configuration that is then used to determine the best-performing configuration.

Now that we have identified the best configuration (the best set of hyperparameters), we again start training this model from scratch, but this time we use all of the $k$ folds as training data. When we finally are done training this best-performing configuration on all the training data, we can run the model on the test dataset to determine how well it performs on not-yet-seen data. This process comes with additional computational cost because we must train each configuration $k$ times instead of a single time. Procedure 6.2 represents how $k$-fold cross-validation typically works. The overall process is illustrated in Figure 6.13.

**Procedure 6.2:** $k$-Fold Cross-Validation

1. Split the Data:
   Start by splitting your dataset into $k$ equal-sized folds. Each fold should ideally have the same distribution of classes or outcomes as the entire dataset. This can be done randomly or sequentially depending on the nature of your dataset.

2. Initialize Performance Metric:
   Choose a performance metric for evaluation, such as accuracy, precision, recall, F1-score, or others, depending on the nature of your problem.

3. Iterate Through Folds:
   For each fold:
   a. Set aside the current fold as the test set.
   b. Combine all other folds to create the training set.
   c. Train your model using the training set.
   d. Use the trained model to predict outcomes on the test set.
   e. Calculate the performance metric for this fold, using the predicted outcomes and the true outcomes from the test set.
   f. Store the performance metric obtained from this fold for later analysis.

4. Aggregate Results:
   Once all folds have been used as the test set, aggregate the performance metrics obtained from each fold to get an overall evaluation of the model's performance.

5. Report Final Result:
   Optionally, you can report the average performance metric across all folds as the final evaluation of your model.

6. Iterate Over Different Models/Parameters:
   If you're comparing multiple models or hyperparameters, repeat steps 3-5 for each model or parameter setting. This helps in selecting the best-performing model or parameter configuration.

**Remarks:**

- Leave-one-out cross-validation (LOOCV) is a special case of $k$-fold cross-validation where $k$ is set to the number of labeled data points. In LOOCV, each data point is used as the test set once, while the rest of the data points are used for training. This approach closely approximates the accuracy. However, LOOCV can be computationally expensive, especially for large datasets, since it requires training the model $n$ times, where $n$ is the number of data points. This makes LOOCV impractical for large datasets or models with high computational costs.
- While cross-validation methods are theoretically superior for estimating model performance, practical considerations such as computational cost often lead to compromises in their implementation, especially in the context of NNs and large datasets. Finding a balance between computational efficiency and model performance is crucial in such scenarios.
- Running each hyperparameter configuration until convergence can take a significant amount of time, especially when dealing with large datasets and complex models. To address this computational challenge, a common strategy is to limit the training time for each hyperparameter configuration. This approach involves running the training process for a fixed number of epochs rather than waiting for convergence. If a hyperparameter configuration does not show significant progress within the specified number of epochs, it is terminated, and the resources are reallocated to other configurations. In the end, only a few ensemble members are allowed to run to completion. One reason that such an approach works well is because the vast majority of the progress is often made in the early phases of the training.

## 6.5 Performance Measures

Performance metrics [56, 58,61, 167-172] in machine learning are crucial as they provide a quantifiable measurement of how well a model performs against a given dataset. These metrics are essential for several reasons:

1. Performance metrics help to assess the accuracy and efficacy of a machine learning model. By quantifying how well the model predicts new data, these metrics can indicate whether the model is performing as expected or needs improvement. This is critical in ensuring that the deployed model performs optimally in real-world scenarios.
2. Different models can be evaluated using the same metrics, making it possible to compare their performance objectively. This comparison is essential when deciding which model to deploy for a particular application.
3. Performance metrics can guide the process of feature selection by indicating how changes to the input variables affect the model's accuracy. This helps in optimizing the model to use only the most relevant features.
4. Many models require the adjustment of hyperparameters. Performance metrics can help determine the best set of hyperparameters that optimize a model's performance.
5. By identifying weaknesses in a model's performance (such as a high false positive rate), metrics can guide targeted improvements.
6. In many applications, particularly in fields like healthcare or finance, being able to trust a model's predictions is crucial. Performance metrics such as accuracy, precision, recall, and F1-score provide a foundation for establishing this trust.
7. Certain metrics can help in evaluating the cost-effectiveness of a model by balancing the trade-offs between various outcomes. For example, reducing the number of false negatives might be crucial in a medical diagnosis scenario, even if it increases the cost slightly.

The choice of the appropriate metric hinges on the specific application for which the NN is designed, such as classification, regression, or clustering. This section delves into a variety of metrics tailored for classification and regression tasks, each chosen to highlight different aspects of model performance.

### 6.5.1. Classification Metrics

In classification tasks, where the goal is to assign each input to one of several categories, accuracy might be the first metric that comes to mind. However, while accuracy measures the overall correctness of predictions, it can be misleading in cases of class imbalance. Therefore, more nuanced metrics such as precision, recall, and the F1-score are often more informative. Moreover, the confusion matrix provides a visual and quantitative way to measure how well a classification algorithm has performed by comparing the predicted labels with the actual true labels. The confusion matrix serves as the foundation from which other metrics are derived.

For a binary classification problem, the confusion matrix is a $2 \times 2$ table comprising four different components:

- True Positives (TP): These are the instances where the model correctly predicts the positive class.
- True Negatives (TN): These are the instances where the model correctly predicts the negative class.
- False Positives (FP): These are the instances where the model incorrectly predicts the positive class (also known as Type I error).
- False Negatives (FN): These are the instances where the model fails to predict the positive class when it is the true class (also known as Type II error).

The matrix is typically arranged as follows:

| | Predicted Positive (PP) | Predicted Negative (PN) |
|---|---|---|
| Actual Positive (P) | True Positives (TP) | False Negatives (FN) |
| Actual Negative (N) | False Positives (FP) | True Negatives (TN) |

To illustrate how a confusion matrix can be used to evaluate the performance of a binary classifier, let's consider a classifier designed to diagnose cancer. In this example, we have a sample of 12 individuals, 8 of whom have been diagnosed with cancer (positive class) and 4 who are cancer-free (negative class). The classifier was used to predict whether each individual has cancer. Out of the 12 individuals, the classifier made 9 correct predictions and 3 incorrect ones. Specifically, it failed to identify 2 individuals with cancer, labeling them as cancer-free, and it incorrectly identified 1 cancer-free individual as having cancer. The matrix is typically arranged as follows

| 8+4=12 | PP (7) | PN (5) |
|---|---|---|
| P (8) | TP: (6) | FN: (2) |
| N (4) | FP: (1) | TN: (3) |

A confusion matrix helps in visualizing the performance of the classifier.

- TP: Individuals who were correctly identified as having cancer. In this case, there were 6 true positives.
- FN: Individuals with cancer who were incorrectly identified as cancer-free. We had 2 false negatives.
- FP: Cancer-free individuals who were incorrectly identified as having cancer. There was 1 false positive.
- TN: Individuals who were correctly identified as being cancer-free. There were 3 true negatives.

In this matrix, the main diagonal (TP and TN) shows the number of correct predictions, and the off-diagonal (FN and FP) shows the number of incorrect predictions. For better visual clarity, correct predictions (TP and TN) can be highlighted in green. This visual enhancement helps quickly assess the areas where the classifier is performing well, and those where improvement is needed. By summing up the 2 rows of the confusion matrix, one can also deduce the total number of positive (P) and negative (N) samples in the original dataset, i.e. P=TP+FN and N=FP+TN.

For problems involving more than two classes, the confusion matrix can be expanded to include rows and columns for each class. In such matrices:

- Each row represents the instances in an actual class.
- Each column represents the instances in a predicted class.
- The diagonal elements represent the number of points for which the predicted label equals the true label.
- Off-diagonal elements are those that are mislabeled by the classifier.

The confusion matrix acts as a cornerstone, providing the essential data from which various other performance metrics are derived. Table 6.4 provides descriptions, definitions, and mathematical formulas for some classification metrics. In the following, we explore the most common performance metrics used in evaluating machine learning models.

1. Accuracy (ACC): Accuracy is the simplest and most intuitive performance measure. It is the proportion of correctly predicted observations (both true positives and true negatives) to the total number of observations. Mathematically,

$$\mathrm{ACC} = \frac{\mathrm{TP} + \mathrm{TN}}{\mathrm{TP} + \mathrm{TN} + \mathrm{FP} + \mathrm{FN}}. \tag{6.10}$$

While accuracy is straightforward, it can be misleading, especially in cases where the class distribution is imbalanced. For instance, in a dataset where 95% of the elements belong to one class, a model that naively predicts this majority class for all inputs will achieve a 95% accuracy, despite not having learned anything meaningful.

2. Precision (Positive Predictive Value (PPV)): Precision measures the accuracy of positive predictions. It is the ratio of correctly predicted positive observations to the total predicted positives. Mathematically,

$$\mathrm{PPV} = \frac{\mathrm{TP}}{\mathrm{PP}} = \frac{\mathrm{TP}}{\mathrm{TP} + \mathrm{FP}}. \tag{6.11}$$

Precision is particularly important in scenarios where the cost of a false positive is high. For example, in email spam detection, a false positive (marking a legitimate email as spam) can be more problematic than a false negative (failing to identify a spam email).

3. Recall (True Positive Rate (TPR), Sensitivity): Recall measures the ability of a model to find all the relevant cases within a dataset. It is the ratio of correctly predicted positive observations to all observations in actual class. Mathematically,

$$\mathrm{TPR} = \frac{\mathrm{TP}}{\mathrm{P}} = \frac{\mathrm{TP}}{\mathrm{TP} + \mathrm{FN}}. \tag{6.12}$$

Recall is critical in conditions where missing a positive instance is considerably worse than falsely predicting an instance as positive. For instance, in medical diagnostics for a serious disease, failing to detect a disease (a false negative) can have more severe consequences than falsely diagnosing it (a false positive).

5. Specificity (True Negative Rate (TNR)): Specificity measures the proportion of actual negatives. This is essentially the ability of the model to identify negative results correctly. Mathematically,

$$\mathrm{TNR} = \frac{\mathrm{TN}}{\mathrm{N}} = \frac{\mathrm{TN}}{\mathrm{TN} + \mathrm{FP}}. \tag{6.13}$$

Specificity is particularly critical in fields where the cost of a false positive is very high. For example, in cancer screening, a high specificity rate means that fewer healthy patients are misdiagnosed as having cancer, reducing unnecessary anxiety and invasive treatments.

4. F1-Score: The F1-Score is the harmonic mean of precision and recall. It is a single metric that balances both the concerns of catching as many positives as possible (high recall) and ensuring the positives caught are truly positive (high precision). Mathematically,

$$F1 = 2\left(\frac{\mathrm{PPV} \times \mathrm{TPR}}{\mathrm{PPV} + \mathrm{TPR}}\right). \tag{6.14}$$

It is a better measure than accuracy for cases where you need to balance false positives and false negatives.

### 6.5.2. Regression Metrics

For regression tasks, which involve predicting continuous values, different metrics are necessary to assess model accuracy. Mean squared error (MSE) is widely used for its emphasis on penalizing large prediction errors quadratically, making it sensitive to outliers. Conversely, mean absolute error (MAE) provides a straightforward average of absolute differences between predicted and actual values, offering a robust measure against outliers.

The most popular metric is MSE, which we discussed earlier in this chapter when explaining how regression works. However, we explained it as a function of the choice of hyperparameters. Here, we will redefine it in a general sense as follows:

$$\mathrm{MSE} = \frac{1}{N}\sum_{i=1}^{N}(\hat{y}_i - y_i)^2. \tag{6.15}$$

Another metric that is very similar to MSE is mean absolute error (MAE). While MSE penalizes big mistakes more (quadratically) and small errors much less, MAE penalizes everything in direct proportion to the absolute difference between what should be and what was predicted. This is a formal definition of MAE:

$$\mathrm{MAE} = \frac{1}{N}\sum_{i=1}^{N}|\hat{y}_i - y_i|. \tag{6.16}$$

Finally, out of the other measures for regression, the popular choice in deep learning is the $R^2$ score, also known as the coefficient of determination. This metric represents the proportion of variance, which is explained by the independent variables in the model. It measures how likely the model is to perform well on unseen data that follows the same statistical distribution as the training data. This is its definition:

$$\mathrm{R}^2 = 1 - \frac{\sum_{i=1}^{N}(\hat{y}_i - y_i)^2}{\sum_{i=1}^{N}(\hat{y}_i - \bar{y})^2}, \tag{6.17}$$

where $\bar{y}$ is the sample mean, $\bar{y} = \frac{1}{N}\sum_{i=1}^{N} y_i$.

**Table 6.4.** List of classification metrics using the confusion matrix.

| Acronym | Formulate | Description | Interpretation |
|---|---|---|---|
| PPV | $\frac{TP}{PP} = \frac{TP}{TP + FP}$ | Positive Predictive Value or Precision | This is the proportion of positive values that are predicted correctly out of all the values predicted to be positive. |
| NPV | $\frac{TN}{PN} = \frac{TN}{TN + FN}$ | Negative Predictive Value | This is the proportion of negative values that are predicted correctly out of all the values that are predicted to be negative. |
| FDR | $\frac{FP}{PP} = \frac{FP}{FP + TP}$ | False Discovery Rate | This is the proportion of incorrect predictions as false positives out of all the values that are predicted to be positive. |
| FOR | $\frac{FN}{PN} = \frac{FN}{FN + TN}$ | False Omission Rate | This is the proportion of incorrect predictions as false negatives out of all the values that are predicted to be negative. |
| TPR | $\frac{TP}{P} = \frac{TP}{TP + FN}$ | True Positive Rate, Sensitivity, Recall, Hit Rate | This is the proportion of predicted positives that are actually positives out of all that should be positives. |
| FPR | $\frac{FP}{N} = \frac{FP}{FP + TN}$ | False Positive Rate or Fall-Out | This is the proportion of predicted positives that are actually negatives out of all that should be negatives. |
| TNR | $\frac{TN}{N} = \frac{TN}{TN + FP}$ | True Negative Rate, Specificity, or Selectivity | This is the proportion of predicted negatives that are actually negatives out of all that should be negatives. |
| FNR | $\frac{FN}{P} = \frac{FN}{FN + TP}$ | False Negative Rate or Miss Rate | This is the proportion of predicted negatives that are actually positives out of all that should be positives. |
| ACC | $\frac{TP + TN}{TP + TN + FP + FN}$ | Accuracy | This is the rate of correctly predicting the positives and the negatives out of all the samples. |
| F1 | $2\left(\frac{PPV \times TPR}{PPV + TPR}\right)$ | F1-Score | This is the average of the precision and sensitivity. |
| MCC | $\frac{(TP \times TN) - (FP \times FN)}{PP \times P \times N \times PN}$ | Matthews | Correlation Coefficient This is the correlation between the desired and the predicted classes. |
| BER | $\frac{1}{2}(FPR + FNR)$ | Balanced Error Rate | This is the average error rate for cases where there is a class imbalance. |

## 6.6 Tuning Hyperparameter

Hyperparameters in NNs refer to the parameters that are set before the training process and determine the architecture or behavior of the model. They are distinct from the weights of the connections, which are learned during the training process through methods like backpropagation. Hyperparameter tuning is a critical step in optimizing the performance of a NN. Key hyperparameters include the learning rate, number of layers, number of neurons per layer, AFs, batch size, etc. Proper tuning of these hyperparameters can significantly impact the performance of the NN, including its accuracy and convergence speed.

The aim of hyperparameter optimization is indeed to find the hyperparameters that result in the best performance of the model, typically measured using a validation set or through cross-validation. By tuning hyperparameters effectively, one can improve the generalization and performance of the machine learning model on unseen data. Various techniques are used for hyperparameter optimization, including grid search, random search, and more advanced optimization algorithms like BO.

One common approach to formalize hyperparameter optimization is to define it as an optimization problem where the objective function represents the performance metric (e.g., accuracy, F1 score, loss function) of the model on a validation set, and the hyperparameters are the variables to be optimized. This can be represented as:

$$\boldsymbol{\theta}^* = \underset{\boldsymbol{\theta}\in\boldsymbol{\Theta}}{\operatorname{argmin}}\, f(\boldsymbol{\theta}) \quad \text{or} \quad \boldsymbol{\theta}^* = \underset{\boldsymbol{\theta}\in\boldsymbol{\Theta}}{\operatorname{argmax}}\, f(\boldsymbol{\theta}). \tag{6.18}$$

Where: $\boldsymbol{\theta}$ represents the vector of hyperparameters to be optimized. $f(\boldsymbol{\theta})$ represents the objective function— such as MSE or error rate— which evaluates the model's performance on the validation set given the hyperparameters $\boldsymbol{\theta}$. Different optimization techniques are then applied to search for the optimal hyperparameters $\boldsymbol{\theta}^*$ that minimize (or maximize) the objective function $f(\boldsymbol{\theta})$, thereby improving the model's performance. In simple terms, we want to find the model hyperparameters that yield the best score on the validation set metric.

The following is a general approach to hyperparameter tuning for a NN:

1. Define a parameter space:
   Determine which hyperparameters you want to tune and define a range or distribution for each hyperparameter. For example, you might specify a range for learning rates, batch sizes, number of neurons per layer, etc.
2. Choose a tuning method:
   There are several methods for hyperparameter tuning, including grid search, random search, BO, genetic algorithms, etc. Grid search and random search are simple and commonly used methods, while more advanced techniques like BO are more efficient but require more computational resources.
3. Set up cross-validation:
   Split your dataset into training, validation, and test sets. Use the training set for training your models, the validation set for evaluating their performance during tuning, and the test set for final evaluation after tuning.
4. Perform hyperparameter optimization:
   Train multiple models with different combinations of hyperparameters using your chosen tuning method. Evaluate each model's performance on the validation set.
5. Select the best hyperparameters:
   Choose the set of hyperparameters that yield the best performance on the validation set.
6. Evaluate on a holdout set:
   After selecting the best hyperparameters based on the validation set, evaluate the final model on a separate holdout test set that was not used during hyperparameter tuning. This gives you an unbiased estimate of the model's performance.
7. Iterate if necessary:
   Depending on the results, you might need to iterate on the tuning process, adjusting the parameter space or tuning method as needed.

Typically, when discussing hyperparameters, we tend to focus on numerical parameters, such as the learning rate or the number of neurons in a layer. However, it's important to remember that several other elements can also be adjusted to potentially improve your model's performance. These include:

- Number of epochs:
  Increasing the number of training epochs can allow the model to learn more from the data, potentially improving performance. However, it's essential to monitor for signs of overfitting.
- Choice of optimizer:
  Different optimizers like Adam, RMSprop, or SGD can have varying effects on training speed and final performance. Trying different optimizers can help find the one that works best for your specific problem.
- Varying the regularization method:

Exploring different regularization techniques such as $L_1$, $L_2$, or dropout can also influence model performance and generalization (regularization methods will be considered in Chapter 7).

- Choice of AF:
  While ReLU is commonly used, experimenting with alternative AFs like sigmoid, tanh, or newer options like Swish can yield improvements in certain scenarios.
- Number of layers and neurons in each layer:
  The architecture of the NN, including the depth and width of layers, can have a significant impact on its capacity to learn complex patterns. Experimenting with different architectures can lead to better performance.
- Learning rate decay methods:
  Adjusting the learning rate over the course of training can help balance between rapid learning in the beginning and fine-tuning towards the end. Different decay methods like exponential decay or step decay can be explored. Techniques like cyclic learning rates or learning rate warmup can improve convergence and generalization.
- Mini-batch size:
  Varying the size of mini-batches can affect the stability and convergence speed of the training process. Larger mini-batches may lead to faster convergence but can also result in increased memory usage.
- Weight initialization methods:
  The initial values of weights in the NN can influence the training dynamics and final performance. Trying different initialization methods such as random, Xavier, or He initialization can be beneficial.
- Dropout rate:
  Dropout (Chapter 7) is a regularization technique where randomly selected neurons are ignored during training, which helps prevent overfitting. You can experiment with different dropout rates to find the optimal level of regularization for your model.
- Batch normalization:
  Batch normalization is a technique that normalizes the activations of each layer to stabilize and accelerate the training process. You can vary the placement and parameters of batch normalization layers within your network architecture to improve performance.

Let's classify the parameters we can tune in our models in the following three categories:

Continuous real numbers:

- Learning rate
- Regularization parameter

Discrete but theoretically infinite values:

- Number of hidden layers
- Number of neurons in each layer
- Number of epochs

Discrete with finite possibilities:

- Optimizer (e.g., Adam, SGD, RMSprop)
- AF (e.g., ReLU, sigmoid, tanh)
- Learning rate decay method (e.g., exponential decay, step decay)
- Weight initialization method (e.g., random initialization, Xavier initialization)

These classifications help organize the parameters based on their nature, which can be useful for understanding the tuning process and implementing strategies for hyperparameter optimization.

Optimizing hyperparameters [56, 58, 61, 65, 173-181] is often done through techniques like grid search, random search, or more advanced methods like BO.

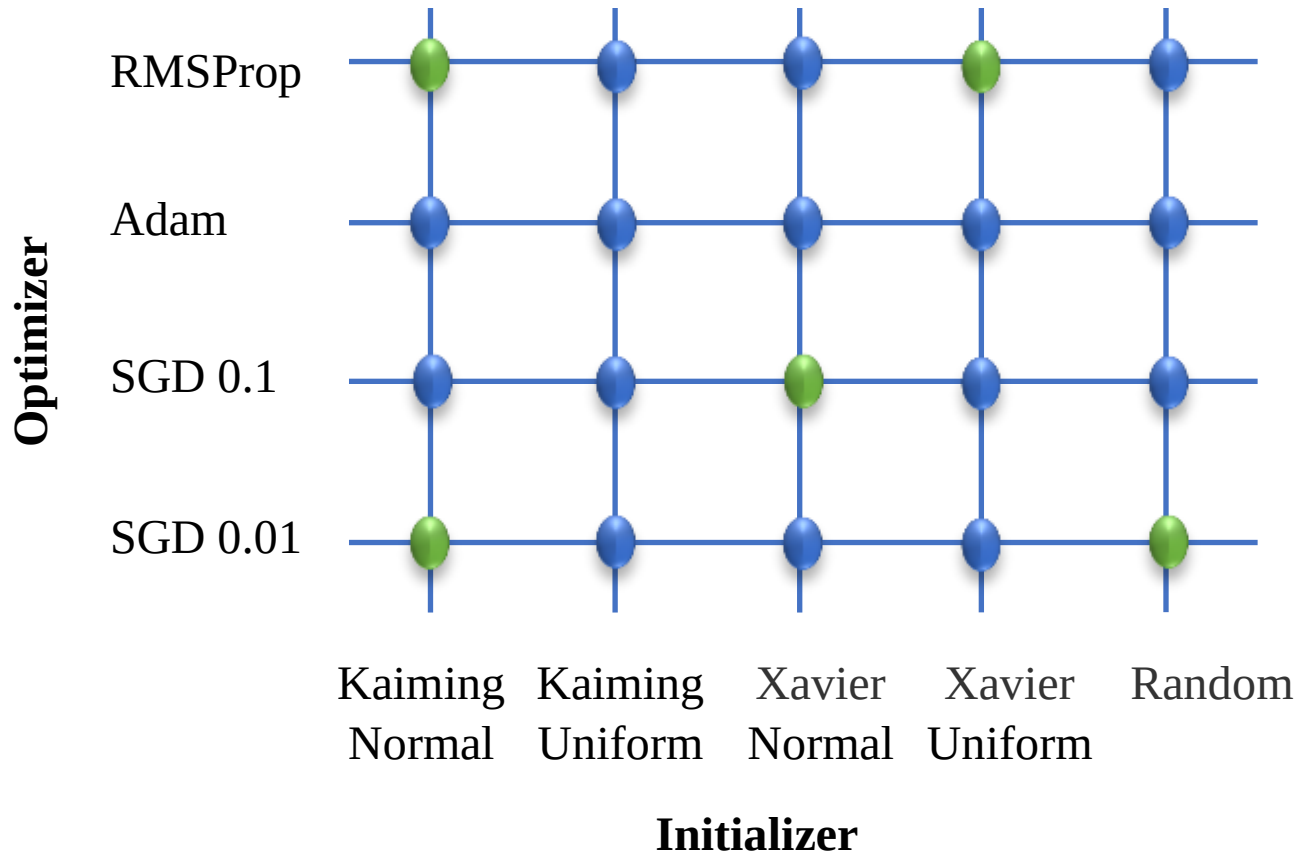

**Figure 6.14.** Grid search for two hyperparameters. An exhaustive grid search would simulate all combinations, whereas a random grid search might simulate only the combinations highlighted in green.

**Procedure 6.3:** Grid Search

- Start by defining the architecture of your NN, including the number of layers, types of layers (e.g., dense, convolutional, recurrent), AFs, etc.
- Create a grid of hyperparameters that you want to search over. This could include parameters like learning rate, batch size, number of neurons per layer, dropout rates, etc. Specify the range of values for each hyperparameter that you want to explore.
- Split your dataset into training, validation, and test sets.
- Choose an appropriate evaluation metric to assess the performance of the model during grid search. This could be accuracy, precision, recall, F1-score, or any other metric relevant to your problem.
- Use nested loops to iterate over all combinations of hyperparameters in the grid. For each combination:
    1. Train the NN on the training set.
    2. Evaluate its performance on the validation set using the chosen evaluation metric.
    3. Keep track of the performance for each combination.
- Once the grid search is complete, select the combination of hyperparameters that resulted in the best performance on the validation set.
- Finally, evaluate the model with the best hyperparameters on the test set to obtain an unbiased estimate of its performance.

### 6.6.1. Grid Search

Grid search is a hyperparameter optimization technique used to tune the parameters of a model to find the best combination of hyperparameters for optimal performance. When applied to NNs, grid search involves systematically searching through a predefined set of hyperparameters, such as learning rate, number of layers, number of neurons per layer, AFs, etc., and evaluating the model's performance on a validation set or through cross-validation for each combination. Grid search is illustrated in Figure 6.14 for the case of two hyperparameters (optimizer and initializer). We simply create a grid with each axis representing a single hyperparameter.

In the case of two hyperparameters, it becomes a 2D grid, as shown in the figure, but we can extend it to more dimensions, although we can only visualize, at most, three dimensions. Each intersection within the grid, depicted by a circle, signifies a distinct combination of hyperparameter values. Together, these circles encompass the entirety of possible configurations. Subsequently, we execute experiments for each grid point to determine the most optimal combination. Procedure 6.3 represents how grid search typically works.

**Procedure 6.4:** Random Grid Search

- Similar to grid search, you start by defining the hyperparameter space. This involves specifying the hyperparameters you want to tune and the range of values for each hyperparameter.
- Unlike grid search, where you define a grid with specific intervals for each hyperparameter, in random grid search, you specify the number of random combinations to evaluate. This number is typically much smaller than the total number of combinations in the entire hyperparameter space.
- Randomly select combinations of hyperparameters from the defined hyperparameter space according to the specified number of samples. Each combination is chosen independently of others, and there's no systematic order to how they're selected.
- For each randomly sampled combination of hyperparameters, run an experiment. This typically involves training a model using the specified hyperparameters on a training dataset and evaluating its performance on a validation dataset using a predefined evaluation metric.
- After running experiments for all randomly sampled combinations, analyze the results to determine which combination of hyperparameters yields the best performance according to the chosen evaluation metric. This could involve selecting the combination that achieves the highest accuracy, or other relevant metrics.

It's worth noting that grid search can be computationally expensive, especially for large datasets and complex NN architectures. As the number of hyperparameters increases, the size of the grid explodes exponentially, making the exhaustive search infeasible. For instance, if we have 5 hyperparameters and we're testing 10 values for each, the number of combinations to explore reaches 100,000 trials. Despite not necessarily running all trials, this sheer volume remains impractical for most scenarios, particularly those with even moderately sized datasets. In such cases, more advanced techniques like random search or BO may be more efficient.

### 6.6.2. Random Grid Search

Random grid search is a variation of grid search where instead of exhaustively evaluating all possible combinations of hyperparameters, a random subset of combinations is selected and evaluated. The main idea behind random grid search is to efficiently explore the hyperparameter space without the computational burden of exhaustively searching through all possible combinations. This alternative is illustrated in Figure 6.14 by the green dots that represent randomly chosen combinations. Note that, one must be careful when the optimal hyperparameter selected is at the edge of a grid range because one would need to test beyond the range to see if better values exist.

We can also do a hybrid approach in which we start with a random grid search to identify one or a couple of promising combinations, and then we can create a finer-grained grid around those combinations and do an exhaustive grid search in this zoomed-in part of the search space. Procedure 6.4 represents how random grid search typically works. However, it's worth noting that random grid search might not guarantee optimal hyperparameter values and may require more iterations to converge to an optimal solution compared to more sophisticated optimization techniques like BO. Nonetheless, it remains a popular and practical choice, particularly for initial hyperparameter tuning or in scenarios where computational resources are limited.

### 6.6.3. Coarse-to-Fine Optimization

There exists an optimization technique that can significantly enhance grid or random search methodologies. This method, known as coarse-to-fine optimization, proves to be an efficient strategy for refining search processes. The coarse-to-fine optimization is a variant of the random search algorithm also called "Iterative Random Search" or "Local Random Search." It involves iteratively refining the search space around the best-found solution to converge towards the optimal solution. To understand its application, let's consider an instance where our objective is to determine the maximum value of a function $f(x)$ within a specified interval, $R_1$, bounded by $x_{min}$ and $x_{max}$.

1- Random Search in $R_1$:

- Randomly sample points from the region $R_1 = (x_{min}, x_{max})$.
- Evaluate the objective function at each sampled point.
- Identify the point with the maximum objective function value, denoted as $(x_1,\ f_1)$.

**Procedure 6.5:** Coarse Grid Search

Full Grid Search:

- Start by performing a grid search over a wide range of hyperparameter values.
- Identify the combination of hyperparameters with the best performance.

Refinement around Optimal Parameters:

- Center a new set of grid ranges around the optimal hyperparameters found in the previous step.
- Define smaller grid ranges by geometrically decreasing the step size or spacing around each optimal parameter.
- Perform another grid search (or random search) within these refined grid ranges.
- Evaluate the performance metric for each combination of hyperparameters in the refined grid.

Iterative Refinement:

- Repeat the refinement process iteratively:
- Center new grid ranges around the optimal hyperparameters found in the previous step.
- Define even smaller grid ranges.
- Perform a grid search (or random search) within these refined ranges.
- Evaluate performance for each combination of hyperparameters.

2- Refinement around $x_1$:

- Define a smaller region $R_2 = (x_1 - \delta x_1, x_1 + \delta x_1)$ around $x_1$, where $\delta x_1$ is a predefined step size or radius.
- Randomly sample points from the region $R_2$.
- Evaluate the objective function at each sampled point.
- Identify the point with the maximum objective function value in $R_2$, denoted as $(x_2,\ f_2)$.

3- Iterative Refinement:

Repeat steps 2 and 3 iteratively:

- Define a smaller region $R_{i+1}$ around the point $(x_i,\ f_i)$ found in the previous step.
- Randomly sample points from the region $R_{i+1}$.
- Evaluate the objective function at each sampled point.
- Identify the point with the maximum objective function value in $R_{i+1}$, denoted as $(x_{i+1},\ f_{i+1})$.
- If the maximum no longer changes significantly, stop the iteration.

From the above discussion, this systematic method for hyperparameter tuning starts with a coarse grid search over a wide range of hyperparameter values and gradually refines the search space around the most promising regions. Procedure 6.5 represents how coarse grid search typically works.

Finally, it's crucial to acknowledge the challenge posed by large-scale settings, where running algorithms to completion can become a formidable task due to the extensive training times involved. For instance, a single run of a CNN for image processing may take a couple of weeks. Trying to execute the algorithm across different parameter combinations becomes impracticable under such circumstances. However, despite these constraints, it's still possible to gain valuable insights into the algorithm's behavior within a shorter timeframe. Again, it's common practice to run the algorithms for a certain number of epochs to monitor their progress. Runs that are obviously poor or diverge from convergence can be quickly killed.

The general problem of searching for optimal NN configurations is commonly referred to as neural architecture search, which extends far beyond the scope discussed in this chapter.

Advantages of Hyperparameter tuning:

- Improved model performance
- Reduced overfitting and underfitting
- Enhanced model generalizability

- Optimized resource utilization
- Improved model interpretability

Disadvantages of Hyperparameter tuning:

- Computational cost
- Time-consuming process
- Risk of overfitting
- No guarantee of optimal performance
- Requires expertise.

## 6.7 Gaussian Process

While grid search and random search are improvements over manual tuning because they automate the process, they are still relatively inefficient due to their lack of information usage from past evaluations. Grid search exhaustively evaluates predefined hyperparameter combinations from a grid, while random search randomly samples hyperparameters from predefined ranges. Both of these methods do not take into account any information gained from previous evaluations. As a result, they may spend a significant amount of time evaluating combinations of hyperparameters that are not promising, leading to suboptimal performance and inefficient resource utilization. To address this inefficiency, more advanced techniques such as BO can be employed. This method uses past evaluations to guide the selection of future hyperparameters, effectively learning from previous trials to focus on promising regions of the hyperparameter space. BO maintains a probabilistic model of the objective function and uses it to decide which hyperparameters to evaluate next, often leading to faster convergence to optimal or near-optimal solutions compared to grid search or random search. By leveraging information from past evaluations, this advanced technique can significantly reduce the number of evaluations needed to find good hyperparameter configurations, making hyperparameter optimization more efficient and effective.

GPs play a pivotal role at the core of BO [182-199]. In BO, the goal is to find the optimum of an objective function, (6.18), $f(\boldsymbol{\theta})$ where $\boldsymbol{\theta}$ typically belongs to some high-dimensional input space. GPs are used to model this objective function. Instead of explicitly defining the function, GPs provide a probabilistic model that captures our beliefs about the behavior of the objective function across the input space.

GPs are among the most powerful, flexible, elegant, and significant models in machine learning for regression, classification, and probabilistic modeling. In this section, we will delve deeply into the details of this method. This exploration extends beyond merely using them for tuning hyperparameters of machine learning models; it also includes a deep understanding of how GPs handle uncertainty, model complex data structures, and provide insightful predictions across various applications.

Why GP is a significant model in machine learning? It is important for several reasons:

- Unlike many machine learning methods that require a predefined form for the model (like linear regression or NNs), GPs are non-parametric. This means they don't assume a specific form for the underlying function they are trying to learn. Instead, they define a prior over functions and use observed data to update this prior to a posterior. This flexibility allows them to model complex and high-dimensional datasets with potentially infinite dimensions.
- Compared to deep learning models, GPs can often achieve good performance with much fewer data points, although they can become computationally expensive as the dataset grows. This makes them suitable for applications where data collection is expensive or difficult.

First, let us provide a straightforward intuition to demystify the concept of GPs. Let's consider the development of a person's life as a series of events that are interconnected and influence one another. Imagine we're trying to predict the trajectory of a person's life based on their experiences and choices. In our analogy, the mean function can be thought of as the "expected path" someone might take in their life. For simplicity, assume everyone starts with similar expectations (e.g., go to school, get a job). This is like assuming a zero mean in GPs, where we start with a baseline expectation before observing any specific data. The covariance function (Kernel) is akin to considering how life experiences at different times are related to each other. For example, the choice of college might influence career opportunities, just as early childhood development impacts learning abilities later in life. In GPs, the covariance function determines how much the function value at one point (say, life decision at age 20) informs us about the function value at another point (life decision at age 40). This function defines the "smoothness" of the life trajectory – whether someone's life changes dramatically over short periods or remains relatively stable. When we observe actual events in a person's life (data points), these help us refine our predictions about their future. In the context of GPs, when we get data about the function at certain points, we use this data to update our beliefs about the function's behavior at other points. This is done through a process involving:

- Prior Distribution: Our initial guess about the person's life trajectory, based on average expectations.
- Likelihood: How likely we are to see the observed life events given our model. This updates our understanding based on how the actual events compare to our initial expectations.
- Posterior Distribution: After observing the events, we update our predictions about the person's future. This combines our initial beliefs (prior) with the new insights gained from the observed data.

Using a GP, after observing some life events, we can predict future events with an associated level of confidence. For example, knowing someone's educational background and early career choices, we might predict their mid-career achievements and later life conditions. The GP provides not just the predictions but also the uncertainty in these predictions, acknowledging that many different life paths are possible from the same starting conditions. Just as life is unpredictable and influenced by a myriad of interconnected factors, GPs handle the uncertainty and variability in real-world data. They provide a flexible framework to model complex phenomena where predictions are uncertain and must be managed with care.

Before delving into the intricacies of GP, it's crucial to grasp the foundational mathematical concepts upon which they rely. As the term implies, GPs draw heavily from the Gaussian distribution. This distribution serves as the fundamental element from which GPs derive their properties. Of particular interest is the multivariate aspect of the Gaussian distribution. In this scenario, each random variable follows a normal distribution, and collectively, their joint distribution forms a Gaussian distribution as well.

### 6.7.1. Multivariate Normal Distribution

The multivariate normal distribution is a generalization of the one-dimensional (univariate) normal distribution to higher dimensions. For example, the bivariate normal distribution is an extension of the univariate normal distribution. It allows us to analyze and model relationships between two continuous RVs, taking into account both their individual characteristics and their dependence on each other. The bivariate normal distribution is characterized by its shape, mean vector, variance-covariance matrix, and correlation coefficient. Understanding these characteristics is crucial for comprehending the properties and applications of the distribution, see Figure 6.15.

- Shape: The bivariate normal distribution forms an elliptical shape in two dimensions. The contours of the distribution are typically centered around the mean vector and symmetrically spread along the axes. The elliptical shape can be stretched or compressed, depending on the correlation between the variables.
- Mean vector: The mean vector of the bivariate normal distribution represents the average values of the two variables. It is denoted as a column vector, $(\mu_1, \mu_2)^T$, where $\mu_1$ is the mean of the first variable and $\mu_2$ is the mean of the second variable. The mean vector determines the location of the center of the distribution.
- Variance-covariance matrix: The variance-covariance matrix of the bivariate normal distribution characterizes the variability and relationship between the two variables. It is a $2 \times 2$ symmetric matrix that contains variances, and covariances. The matrix has the form [1]:

$$\begin{pmatrix} \sigma_1^2 & \sigma_{12} \\ \sigma_{12} & \sigma_2^2 \end{pmatrix}, \tag{6.19}$$

where $\sigma_1^2$ and $\sigma_2^2$ represent the variances of the first and second variables, respectively, and $\sigma_{12}$ is the covariance between the two variables.

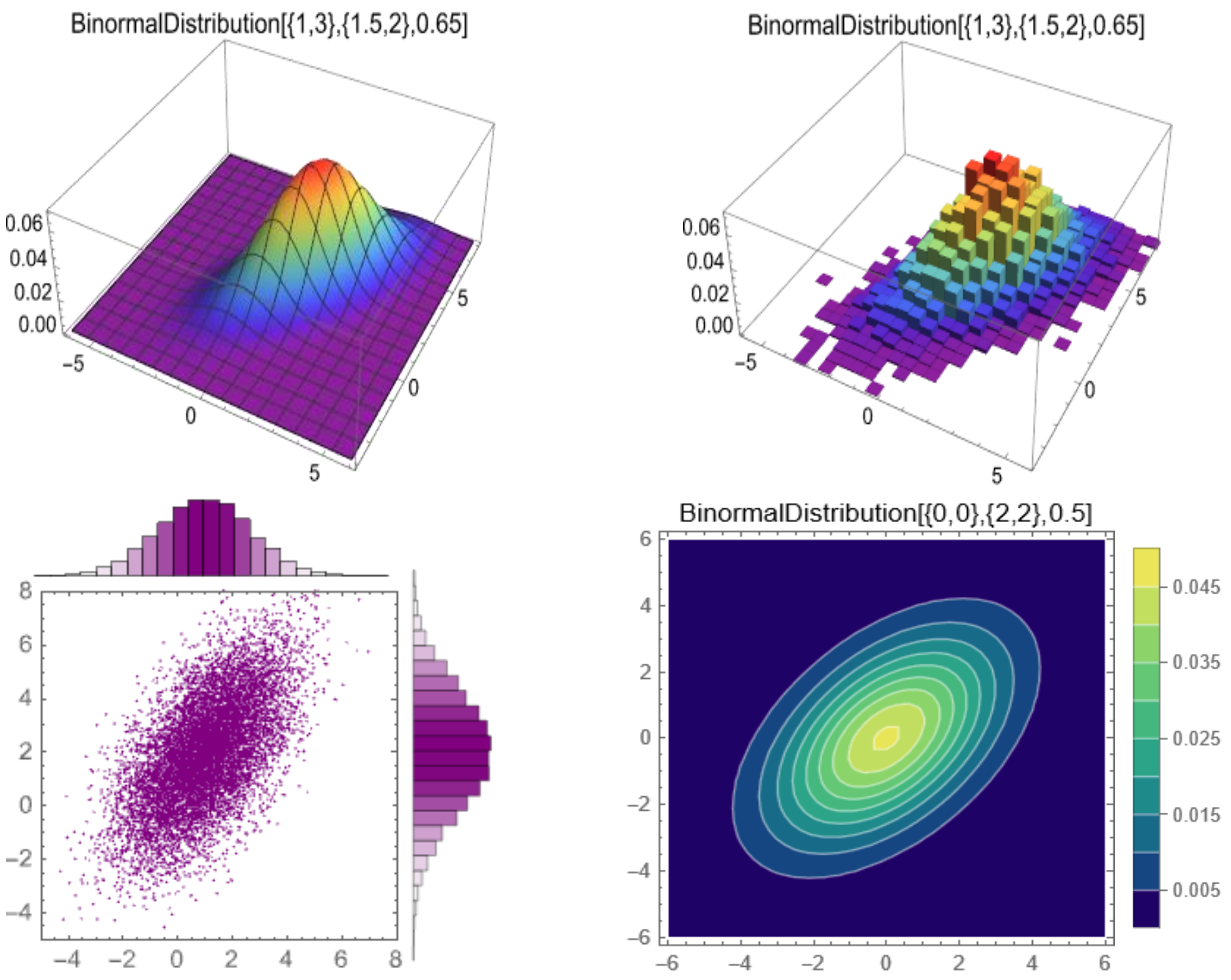

**Figure 6.15.** 3D plot (upper left panel), 3d histogram (upper right panel), scatter plot (lower left panel), and 2d contour plot (lower right panel) of bivariate normal distributions. The density function of bivariate normal distribution is a generalization of the familiar bell curve and graphs in three dimensions as a sort of bell-shaped hump. A bivariate normal density has elliptical contours. For each height $c > 0$ the set $\{(x, y): f_{XY}(x, y) = c\}$ is an ellipse. A scatter plot of $(x, y)$ pairs generated from a bivariate normal distribution will have a rough linear association, and the cloud of points will resemble an ellipse. If $X$ and $Y$ have a bivariate normal distribution, then the marginal distributions are also normal: $X$ has a normal $(\mu_X, \sigma_X)$ distribution and $Y$ has a Normal $(\mu_Y, \sigma_Y)$.

- Correlation coefficient: The correlation coefficient measures the linear relationship between the two variables in the bivariate normal distribution. It is denoted as $\rho$ and ranges between $-1$ and 1.

**Definition (Bivariate Normal PDF):** The PDF of a bivariate normal distribution, BVN $(\mu_1, \mu_2, \sigma_1^2, \sigma_2^2, \rho)$, is given by [1]

$$f_{XY}(x, y; \sigma_1, \sigma_2; \mu_1, \mu_2, \rho) = \frac{1}{2\pi\sigma_1\sigma_2\sqrt{1-\rho^2}} e^{-\frac{1}{2(1-\rho^2)}\left[\frac{(x-\mu_1)^2}{\sigma_1^2} - \frac{2\rho(x-\mu_1)(y-\mu_2)}{\sigma_1\sigma_2} + \frac{(y-\mu_2)^2}{\sigma_2^2}\right]}, \quad (6.20)$$

for $-\infty < x < \infty$ and $-\infty < y < \infty$, with parameters $\sigma_1 > 0$, $\sigma_2 > 0$, $-\infty < \mu_1 < \infty$, $-\infty < \mu_2 < \infty$, and $-1 < \rho < 1$.

In the context of multivariate distributions, it's crucial to distinguish between three main types of probability distributions: joint, conditional, and marginal probability distribution. A joint distribution refers to the probability distribution of multiple random variables considered together. A joint probability distribution describes the simultaneous behavior of multiple random variables. It provides the probabilities associated with all possible combinations of values for the variables. For example, if we have two random variables $X$ and $Y$, their joint distribution describes the probabilities associated with all possible combinations of values of $X$ and $Y$. Mathematically, if $X$ and $Y$ are discrete random variables, the joint distribution is denoted as $P(X = x, Y = y)$, and if they are continuous random variables, it's denoted as $f_{XY}(x, y)$. The joint distribution encapsulates the complete information about the relationship between the variables. On the other hand, conditional probability distributions arise when we consider the probability distribution of one or more variables given the knowledge or observation of other variables. For example, given a joint distribution $P(X,Y)$, we can compute the conditional distribution $P(X|Y)$ which represents the distribution of variable $X$ given a specific value or range of values for variable $Y$. The marginal distribution of a subset of random variables from a joint distribution is obtained by summing (or integrating, in the continuous case) over all other variables. In simpler terms, it represents the probability distribution of a subset of variables without considering the values of the other variables. For instance, if we have a joint distribution $P(X,Y)$, the marginal distribution of $X$ is obtained by summing $P(X,Y)$ over all possible values of $Y$, yielding $P(X)$. Similarly, for continuous random variables, we integrate over $Y$ to obtain $f_X(x)$, the marginal distribution of $X$.

Understanding these distinctions is crucial for modeling complex systems involving multiple variables, as it allows us to analyze dependencies, make predictions, and infer properties of interest based on observed data. Moreover, techniques such as Bayesian inference and Markov chain Monte Carlo methods heavily rely on manipulating these types of distributions to perform probabilistic reasoning and learning from data.

Gaussian distributions indeed possess the convenient property of being closed under conditioning and marginalization. This property is fundamental to GPs and plays a significant role in various statistical and machine-learning techniques. For example, bivariate normal distributions have the nice algebraic property of being closed under conditioning and marginalization.

**Theorem 6.2 (Marginal Distributions of Bivariate Normal Distribution):**
Let $(X,Y)\sim\text{BVN}\ (\mu_1,\mu_2,\sigma_1^2,\sigma_2^2,\rho)$, then the marginal PDFs of $X$ and $Y$ are also normal. The marginal PDFs of RVs $X$ and $Y$ are given by [1]

$$f_X(x) = \int_{-\infty}^{\infty} f_{XY}(x,y)dy = \frac{1}{\sqrt{2\pi}\sigma_1} e^{-\frac{1}{2}\left(\frac{x-\mu_1}{\sigma_1}\right)^2}, \tag{6.21.1}$$

$$f_Y(y) = \int_{-\infty}^{\infty} f_{XY}(x,y)dx = \frac{1}{\sqrt{2\pi}\sigma_2} e^{-\frac{1}{2}\left(\frac{y-\mu_2}{\sigma_2}\right)^2}. \tag{6.21.2}$$

In other words, $X\sim N(\mu_1,\sigma_1^2)$ and $Y\sim N(\mu_2,\sigma_2^2)$.

**Theorem 6.3 (Conditional Distributions of Bivariate Normal Distribution):**
Let $(X,Y)\sim\text{BVN}\ (\mu_1,\mu_2,\sigma_1^2,\sigma_2^2,\rho)$, then the conditional distribution of $X$ for a fixed $Y$, is given by

$$f_{X|Y}(x|y) = \frac{f_{XY}(x,y)}{f_Y(y)} = \frac{1}{\sqrt{2\pi}\sigma_1\sqrt{1-\rho^2}} e^{-\frac{1}{2(1-\rho^2)\sigma_1^2}\left[(x-\mu_1)-\rho\frac{\sigma_1}{\sigma_2}(y-\mu_2)\right]^2}, \tag{6.22.1}$$

and the conditional distribution of $Y$ for a fixed $X$, is given by

$$f_{Y|X}(y|x) = \frac{f_{XY}(x,y)}{f_X(x)} = \frac{1}{\sqrt{2\pi}\sigma_2\sqrt{1-\rho^2}} e^{-\frac{1}{2(1-\rho^2)\sigma_2^2}\left[(y-\mu_2)-\rho\frac{\sigma_2}{\sigma_1}(x-\mu_1)\right]^2}. \tag{6.22.2}$$

Hence, the conditional distribution of $X$ for fixed $Y$ is given by

$$(X|Y=y)\sim N\left[\mu_1+\rho\frac{\sigma_1}{\sigma_2}(y-\mu_2),\sigma_1^2(1-\rho^2)\right], \tag{6.22.3}$$

and the conditional distribution of $Y$ for fixed $X$ is given by

$$(Y|X=x)\sim N\left[\mu_2+\rho\frac{\sigma_2}{\sigma_1}(x-\mu_1),\sigma_2^2(1-\rho^2)\right]. \tag{6.22.4}$$

**Proof:**

The conditional distribution of $X$ for a fixed $Y$, is given by

$$\begin{aligned} f_{X|Y}(x|y) &= \frac{f_{XY}(x,y)}{f_Y(y)} \\ &= \frac{\sqrt{2\pi}\sigma_2}{2\pi\sigma_1\sigma_2\sqrt{1-\rho^2}} e^{-\frac{1}{2(1-\rho^2)}\left[\frac{(x-\mu_1)^2}{\sigma_1^2}-\frac{2\rho(x-\mu_1)(y-\mu_2)}{\sigma_1\sigma_2}+\frac{(y-\mu_2)^2}{\sigma_2^2}\right]+\frac{1}{2}\left(\frac{y-\mu_2}{\sigma_2}\right)^2} \\ &= \frac{1}{\sqrt{2\pi}\sigma_1\sqrt{1-\rho^2}} e^{-\frac{1}{2(1-\rho^2)}\left[\frac{(x-\mu_1)^2}{\sigma_1^2}-\frac{2\rho(x-\mu_1)(y-\mu_2)}{\sigma_1\sigma_2}+\frac{(y-\mu_2)^2}{\sigma_2^2}\right]+\frac{(1-\rho^2)}{2(1-\rho^2)}\left(\frac{y-\mu_2}{\sigma_2}\right)^2} \end{aligned}$$

$$= \frac{1}{\sqrt{2\pi}\sigma_1\sqrt{1-\rho^2}} e^{-\frac{1}{2(1-\rho^2)}\left[\frac{(x-\mu_1)^2}{\sigma_1^2} - \frac{2\rho(x-\mu_1)(y-\mu_2)}{\sigma_1\sigma_2} + \frac{(y-\mu_2)^2}{\sigma_2^2} - (1-\rho^2)\left(\frac{y-\mu_2}{\sigma_2}\right)^2\right]}$$

$$= \frac{1}{\sqrt{2\pi}\sigma_1\sqrt{1-\rho^2}} e^{-\frac{1}{2(1-\rho^2)}\left[\frac{(x-\mu_1)^2}{\sigma_1^2} - \frac{2\rho(x-\mu_1)(y-\mu_2)}{\sigma_1\sigma_2} + \frac{(y-\mu_2)^2}{\sigma_2^2}\{1-(1-\rho^2)\}\right]}$$

$$= \frac{1}{\sqrt{2\pi}\sigma_1\sqrt{1-\rho^2}} e^{-\frac{1}{2(1-\rho^2)}\left[\frac{(x-\mu_1)^2}{\sigma_1^2} - \frac{2\rho(x-\mu_1)(y-\mu_2)}{\sigma_1\sigma_2} + \frac{(y-\mu_2)^2}{\sigma_2^2}\rho^2\right]}$$

$$= \frac{1}{\sqrt{2\pi}\sigma_1\sqrt{1-\rho^2}} e^{-\frac{1}{2(1-\rho^2)}\left[\left(\frac{x-\mu_1}{\sigma_1}\right) - \rho\left(\frac{y-\mu_2}{\sigma_2}\right)\right]^2}$$

$$= \frac{1}{\sqrt{2\pi}\sigma_1\sqrt{1-\rho^2}} e^{-\frac{1}{2(1-\rho^2)\sigma_1^2}\left[(x-\mu_1) - \rho\frac{\sigma_1}{\sigma_2}(y-\mu_2)\right]^2}$$

$$= \frac{1}{\sqrt{2\pi}\sigma_1\sqrt{1-\rho^2}} e^{-\frac{1}{2(1-\rho^2)\sigma_1^2}\left[x-\left(\mu_1+\rho\frac{\sigma_1}{\sigma_2}(y-\mu_2)\right)\right]^2},$$

which is the probability function of a univariate normal distribution with mean and variance given by

$$\mathbb{E}(X|Y=y) = \mu_1 + \rho\frac{\sigma_1}{\sigma_2}(y-\mu_2),$$

$$\mathrm{Var}(X|Y=y) = \sigma_1^2(1-\rho^2).$$

Hence, the conditional distribution of $X$ for fixed $Y$ is given by

$$(X|Y=y) \sim N\left[\mu_1 + \rho\frac{\sigma_1}{\sigma_2}(y-\mu_2), \sigma_1^2(1-\rho^2)\right].$$

Similarly, the conditional distribution of RVs $Y$ for, a fixed $X$ is

$$f_{Y|X}(y|x) = \frac{f_{XY}(x,y)}{f_X(x)} = \frac{1}{\sqrt{2\pi}\sigma_2\sqrt{1-\rho^2}} e^{-\frac{1}{2(1-\rho^2)\sigma_2^2}\left[y-\left(\mu_2+\rho\frac{\sigma_2}{\sigma_1}(x-\mu_1)\right)\right]^2}.$$

Thus, the conditional distribution of $Y$ for fixed $X$ is given by

$$(Y|X=x) \sim N\left[\mu_2 + \rho\frac{\sigma_2}{\sigma_1}(x-\mu_1), \sigma_2^2(1-\rho^2)\right].$$

■

**Definition (Multivariate Normal PDF):** The multivariate normal distribution of a $k$-dimensional random vector $\mathbf{X} = (X_1, \ldots, X_k)^T$ can be written in the following notation:

$$\mathbf{X} \sim N(\boldsymbol{\mu}, \boldsymbol{\Sigma}), \quad (6.23.1)$$

with $k$-dimensional mean vector

$$\boldsymbol{\mu} = \mathbb{E}[\mathbf{X}] = (\mathbb{E}[X_1], \ldots, \mathbb{E}[X_k])^T = (\mu_1, \ldots, \mu_k)^T, \quad (6.23.2)$$

and $k \times k$ covariance matrix

$$\Sigma_{ij} = \mathbb{E}\left[(X_i-\mu_i)(X_j-\mu_j)\right] = \mathrm{Cov}[X_i, X_j], \quad (6.23.3)$$

such that $1 \le i \le k$ and $1 \le j \le k$.

The multivariate normal distribution is said to be "non-degenerate" when the symmetric covariance matrix $\boldsymbol{\Sigma}$ is positive definite. In this case, the distribution has a density

$$f_{\mathbf{x}}(x_1, \ldots, x_k) = \frac{1}{\sqrt{2\pi|\boldsymbol{\Sigma}|}} e^{-\frac{1}{2}(\mathbf{x}-\boldsymbol{\mu})^T\boldsymbol{\Sigma}^{-1}(\mathbf{x}-\boldsymbol{\mu})} = \frac{1}{\sqrt{2\pi|\boldsymbol{\Sigma}|}} e^{-\frac{1}{2}\langle \mathbf{x}-\boldsymbol{\mu}|\boldsymbol{\Sigma}^{-1}|\mathbf{x}-\boldsymbol{\mu}\rangle}, \quad (6.24)$$

where $\mathbf{x}$ is a real $k$-dimensional column vector and $|\boldsymbol{\Sigma}|$ is the determinant of $\boldsymbol{\Sigma}$, also known as the generalized variance.

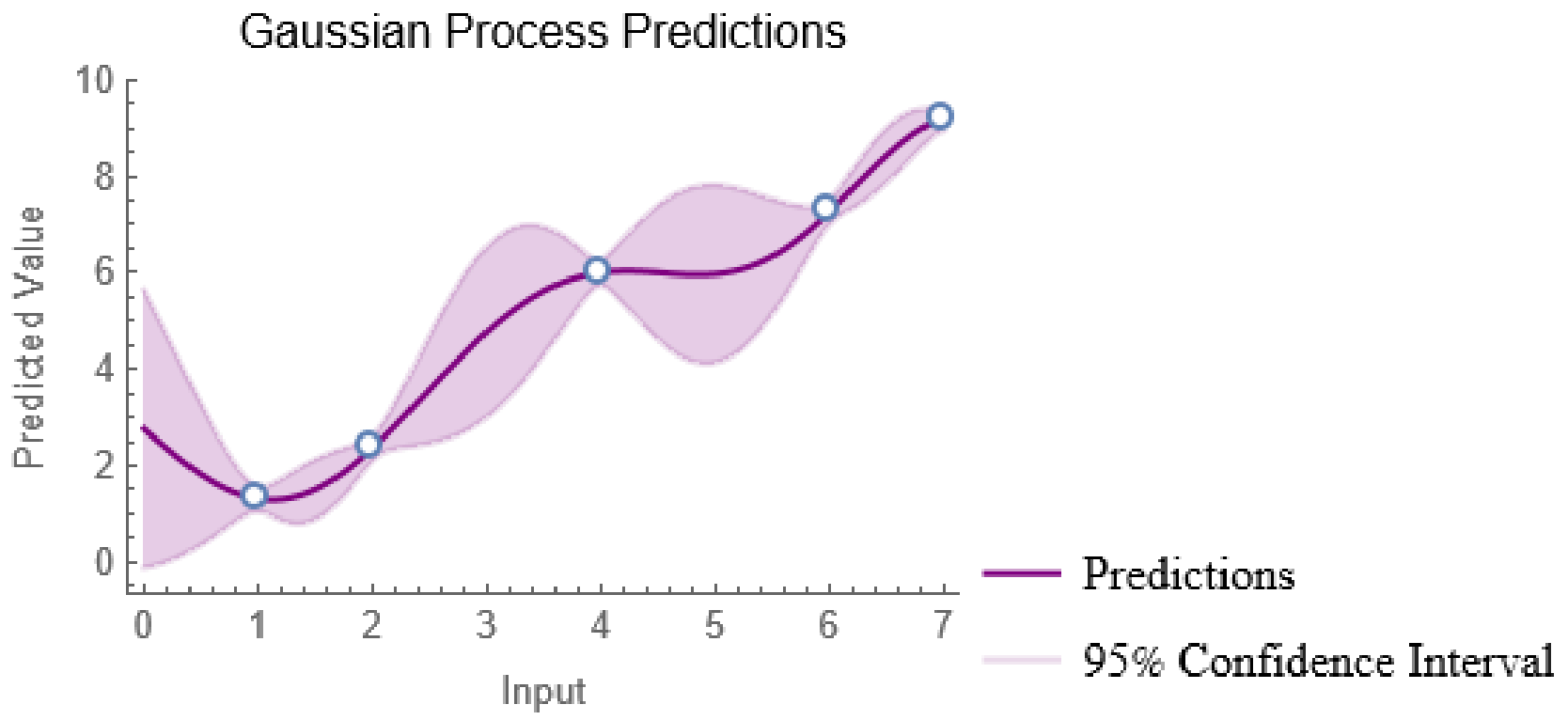

**Figure 6.16.** The figure illustrates GP predictions for a regression task, alongside their corresponding confidence intervals, over a range of input values. The training data points, shown as open markers, are used to train the GP model. The purple curve represents the GP predictions, which provide a smooth interpolation between the training examples. The shaded region around the predictions denotes the confidence interval, representing the uncertainty associated with the predictions. The width of the confidence interval varies across the input space, being narrower near the training examples and wider in between. The confidence level for the interval is set to 95%, indicating that there is a 95% probability that the true function values lie within the shaded region around the predictions.

### 6.7.2. Prior Distribution

Now that we have recalled some of the basic properties of multivariate Gaussian distributions, we will combine them together to define GPs and show how they can be used to tackle regression problems. But, before delving into the intricacies of this method, it's important to understand the nature of the model generated by the GP approach. Training a GP on a simple regression task provides a concrete example to illustrate the characteristics of this model. Given a set of training data consisting of input-output pairs, we train a GP model on this data. The GP learns the underlying pattern or trend in the data and captures it in the form of a probabilistic model. GPs provide not just point predictions but entire probability distributions over the possible values of the objective function at any given input point. Figure 6.16 shows the GP predictions as a smooth curve that passes through the data points, with shaded regions indicating the confidence intervals. This smoothness is inherent in GPs. This curve represents the mean of the predictive distribution generated by the GP model. It provides the central estimate of the predicted values. The shaded regions around the predictive curve represent the confidence intervals associated with the predictions. The upper and lower bounds of each shaded region indicate the range within which the true function values are likely to lie with a certain level of confidence. The level of confidence is determined by the specified confidence interval percentage (e.g., 95% confidence interval).

Unlike some other regression methods where confidence intervals have a constant size, GPs produce confidence intervals that vary in size across the input space. The uncertainty (or confidence interval) is small near the training examples where the model has observed data (the model has more information about the underlying function) and large in between, where the model has less information to make accurate predictions. This variability in confidence intervals reflects the model's ability to capture uncertainty caused by a lack of training data in regions far from the training examples. It is a key strength of GPs and is crucial for applications where reliable uncertainty estimation is essential. This property of GPs makes them particularly suitable for applications such as BO, where the goal is to optimize an expensive-to-evaluate objective function. By leveraging the uncertainty estimates, BO algorithms can intelligently select the next points to evaluate, balancing exploration (sampling in uncertain regions) and exploitation (sampling in regions likely to contain the optimum).

Now, let's take a closer look at the workings of the GP method to gain a deeper understanding. GP is a stochastic process (a collection of random variables), such that every finite collection of those random variables has a multivariate normal distribution.

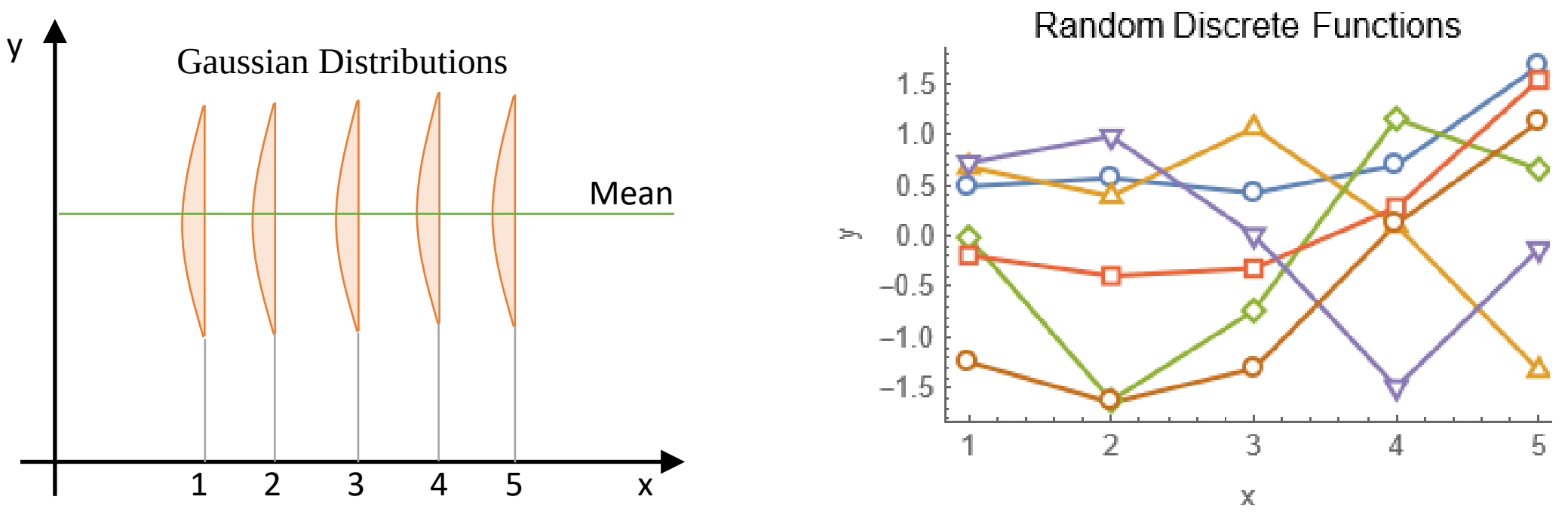

**Figure 6.17.** Left panel: The figure shows five Gaussian distributions positioned at distinct discrete input values ranging from 1 to 5. Right panel: The figure illustrates random discrete functions generated from a multivariate Gaussian distribution. The discrete functions are represented by lines, each indicating a different random sample drawn from the distribution. The x-axis denotes discrete input values ranging from 1 to 5, while the y-axis represents the associated function values.

**Table 6.5.** Random samples were generated from a multivariate Gaussian distribution. Each row represents a different sample.

| | 1 | 2 | 3 | 4 | 5 |
|---|---|---|---|---|---|
| Sample 1 | 0.485679 | 0.569254 | 0.421367 | 0.695165 | 1.672630 |
| Sample 2 | 0.676805 | 0.390511 | 1.062260 | 0.10210 | −1.32375 |
| Sample 3 | −0.01487 | −1.62062 | −0.73515 | 1.153100 | 0.656737 |
| Sample 4 | −0.19139 | −0.39519 | −0.32028 | 0.280233 | 1.53610 |
| Sample 5 | 0.719586 | 0.979797 | 0.008696 | −1.48790 | −0.12945 |
| Sample 6 | −1.24083 | −1.64066 | −1.30775 | 0.117652 | 1.124940 |

For example [182], let's establish a Gaussian distribution over five variables $y_1$, ..., $y_5$. Initially, we set the mean vector as: $\boldsymbol{\mu} = (0,0,0,0,0)^T$. Next, we define the covariance matrix, which dictates the correlation between pairs of variables:

$$\text{Cov} = \begin{pmatrix} 1 & 0.4 & 0.02 & 0 & 0 \\ 0.4 & 1 & 0.4 & 0.02 & 0 \\ 0.02 & 0.4 & 1 & 0.4 & 0.02 \\ 0 & 0.02 & 0.4 & 1 & 0.4 \\ 0 & 0 & 0.02 & 0.4 & 1 \end{pmatrix}.$$

The high positive correlations around the diagonal indicate that nearby variables have strong positive correlations, while correlations decrease as variables become more distant from each other. This setup allows for the generation of random samples from the specified Gaussian distribution, with the resulting samples reflecting the smoothness and correlations defined by the covariance matrix.

Now, let's consider each of these variables as the result of a function $f$ for discrete inputs: $f(1) = y_1$, $f(2) = y_2$, ..., $f(5) = y_5$. This transforms the multivariate distribution into a distribution over discrete functions. We can proceed by generating random samples from this distribution. Each sample drawn from this distribution represents a set of values for the variables $y_1$, ..., $y_5$ . These samples are generated based on the specified mean vector and covariance matrix, capturing the underlying relationships and variability among the variables. We can visualize these samples to gain insights into the behavior of the function $f$. Plotting the function values against the discrete inputs $x$ for each sample allows us to observe patterns, variability, and correlations in the function's behavior, see Figure 6.17, and Table 6.5. Random samples are drawn from the multivariate Gaussian distribution, with each sample representing a set of function values for the discrete inputs. These function values are plotted as lines on the graph, where the $x$-axis represents the discrete inputs ranging from 1 to 5, and the $y$-axis represents the corresponding function values $y$. Each line on the plot represents a different random sample.

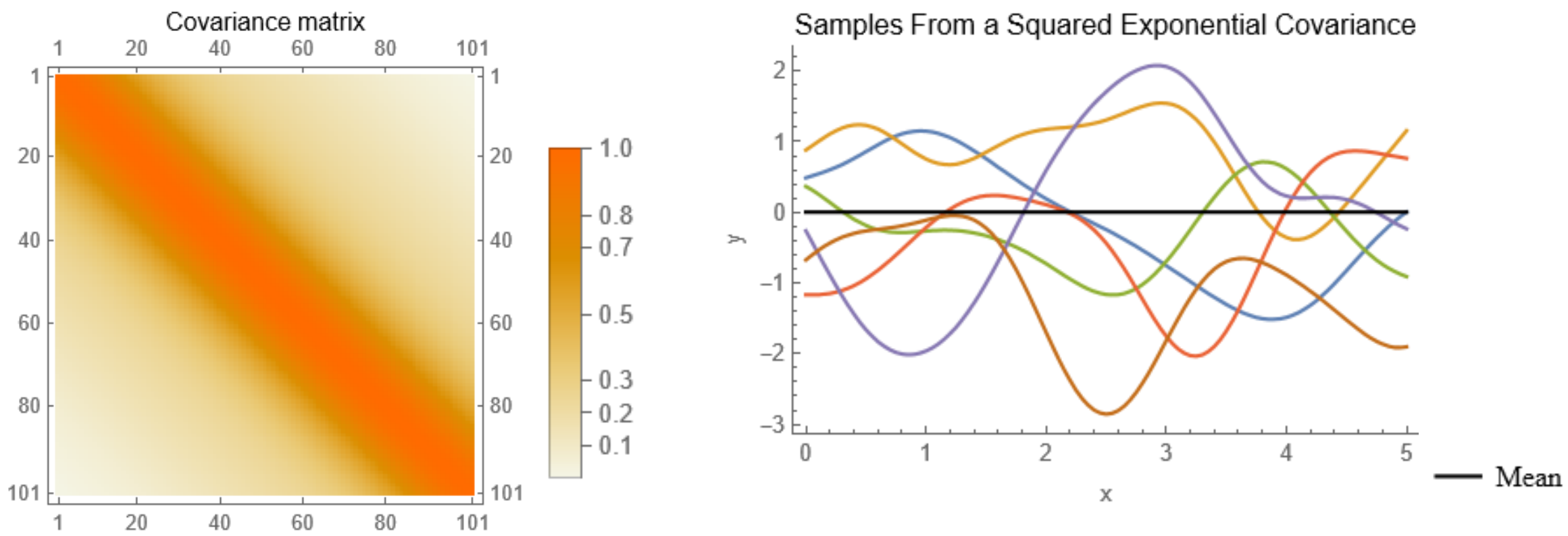

**Figure 6.18.** Visualization of the covariance matrix and sampled functions from a GP with squared exponential covariance.

By including more variables in the distribution, the gap between possible inputs decreases. This extension aims to capture a more continuous representation of the data. As the number of variables in a distribution approaches infinity, a crucial transformation occurs, discrete functions transition into continuous ones, and the distribution itself evolves into what we call a GP. In essence, a GP is defined as an infinite-dimensional multivariate Gaussian distribution. In a GP, there are infinitely many variables, each corresponding to a different point in the input space. These variables are interconnected through their covariance structure, which captures the relationships and dependencies between them.

In a GP with infinitely many variables, each corresponding to a point in the input space, specifying and storing the covariance between every pair of variables in a traditional covariance matrix becomes impractical due to the sheer number of variables. Instead, a covariance function is utilized. This function encapsulates the pairwise relationships between variables in a compact and computationally efficient manner, thereby overcoming the limitations associated with explicitly representing a covariance matrix. The covariance function quantifies the degree of similarity or correlation between function values $f(x)$ and $f(x')$ for any pair of input values $x$ and $x'$. By evaluating the covariance function for different pairs of input values, one can infer the level of correlation between the corresponding function values. Covariance functions come in various forms, each capturing different aspects of the relationship between input values and their corresponding function values. The choice of covariance function is crucial as it shapes the behavior and characteristics of the GP. Commonly used covariance functions include the squared exponential, Matérn, and rational quadratic kernels, each offering different degrees of smoothness and flexibility. Covariance functions play a central role in GPs, serving as the fundamental building blocks for modeling the correlations and dependencies between function values across the input space. By specifying an appropriate covariance function, practitioners can tailor the GP model to suit the specific characteristics and requirements of the data at hand. The problem of learning in GPs is exactly the problem of finding suitable properties for the covariance function. Note that this gives us a model of the data, and characteristics (such a smoothness, characteristic length-scale, etc.) which we can interpret.

The squared exponential covariance [183] is a commonly used covariance function in GPs. It characterizes smooth correlations between function values across the input space. This covariance function, denoted as $\Sigma(x,x') = k(x,x')$, assumes that nearby inputs will have similar function values, leading to smooth and continuous functions.

$$\Sigma(x,x') = k(x,x') = e^{-\frac{\|x-x'\|^2}{2l^2}}. \tag{6.25}$$

The parameter $l$ is the characteristic length-scale of the process (practically, "how close" two points $x$ and $x'$ have to be to influence each other significantly). Here covariance decays exponentially fast as $x$ and $x'$ become farther apart in the input, or $x$-space. In this specification, observe that $\Sigma(x,x) = 1$ and $\Sigma(x,x') < 1$ for $x' \neq x$.

Figure 6.18 showcases the generation of samples from a GP with a squared exponential covariance function. The left panel illustrates the covariance matrix (function), where each cell represents the covariance between function values at different input points. The right panel presents six sampled functions drawn at random from GP which favours smooth functions, along with the mean function shown in black. Each sampled function is depicted as a line plot, with the $x$-axis representing the input values and the $y$-axis displaying the corresponding function values. Note that, although the specific random functions drawn in Figure 6.18 (right panel) do not have a mean of zero, the mean of $f(x)$ values for any fixed $x$ would become zero, independent of $x$ as we kept on drawing more functions.

For now, we have not yet observed any training data. In the context of Bayesian inference, this is called the prior distribution. This prior is taken to represent our prior beliefs over the kinds of functions we expect to observe, before seeing any data.

**Definition (GP):** A GP is a (potentially infinite) collection of RVs such that the joint distribution of every finite subset of RVs is multivariate Gaussian:

$$f(\mathbf{x}) \sim GP\big(\mu(\mathbf{x}), k(\mathbf{x}, \mathbf{x}')\big), \tag{6.26.1}$$

where $\mu(\mathbf{x})$ and $k(\mathbf{x}, \mathbf{x}')$ are the mean and covariance functions,

$$\mu(\mathbf{x}) = \mathbb{E}[f(\mathbf{x})], \tag{6.26.2}$$

$$k(\mathbf{x}, \mathbf{x}') = \Sigma(\mathbf{x}, \mathbf{x}') = \mathbb{E}\big[\big(f(\mathbf{x}) - \mu(\mathbf{x})\big)\big(f(\mathbf{x}') - \mu(\mathbf{x}')\big)\big]. \tag{6.26.3}$$

With interest in modeling functions, we'll sometimes use the term mean function, thinking of $\mu(\mathbf{x})$, and covariance function, thinking of $\Sigma(\mathbf{x}, \mathbf{x}')$. Ultimately, we'll end up with vectors $\boldsymbol{\mu}$ and matrices $\boldsymbol{\Sigma}$ after evaluating those functions at specific input locations $\mathbf{x}_1, \dots, \mathbf{x}_n$.

### 6.7.3. Posterior Distribution with Noise-free Observations (Pointwise Case)

Suppose that we are given a dataset $D_{n=3} = \{(x_1, y_1), (x_2, y_2), (x_3, y_3)\}$ consisting of three observations, and we wish now to only consider functions that pass though these three data points exactly. This situation is illustrated in Figure 6.19 (bottom-right panel). The color lines show sample functions that are consistent with $D_{n=3}$, and the black line depicts the mean value of such functions. Notice how the uncertainty is reduced close to the observations. The combination of the prior and the data leads to the posterior distribution over functions. If more datapoints were added one would see the mean function adjust itself to pass through these points, and that the posterior uncertainty would reduce close to the observations. Notice that since the GP is not a parametric model, we do not have to worry about whether it is possible for the model to fit the data. Even when a lot of observations have been added, there may still be some flexibility left in the functions.

After observing data, the prior distribution is updated to a posterior distribution using Bayes' theorem. The posterior distribution incorporates both the prior beliefs and the information gained from the observed data. As expected, every sample crosses the known value because conditioning is equivalent to selecting the functions that agree with the data. We can see that the mean (predictions) has changed; it now crosses the known value as well. The intuition behind this step is that the training points constrain the set of functions to those that pass through the training points. The conditional distribution forces the set of functions to precisely pass through each training point. In many cases, this can lead to fitted functions that are unnecessarily complex.

Let's set aside interpretation—whether through Bayesian updating or a nuanced take on simple regression—for a moment and turn our attention to conditional distributions again, as they are the crux of the matter. Generating that posterior distribution (predictive distribution) is a straightforward process of deducing a conditional from a multivariate normal distribution. If an $N$-dimensional random vector $\boldsymbol{X}$ is partitioned as [184]

$$\boldsymbol{X} = \begin{pmatrix} \boldsymbol{X}_1 \\ \boldsymbol{X}_2 \end{pmatrix} \text{ with sizes } \begin{pmatrix} q \times 1 \\ (N-q) \times 1 \end{pmatrix}, \tag{6.27}$$

and accordingly, $\boldsymbol{\mu}$ and $\boldsymbol{\Sigma}$ are partitioned as,

$$\boldsymbol{\mu} = \begin{pmatrix} \boldsymbol{\mu}_1 \\ \boldsymbol{\mu}_2 \end{pmatrix} \text{ with sizes } \begin{pmatrix} q \times 1 \\ (N-q) \times 1 \end{pmatrix}, \tag{6.28}$$

and

$$\boldsymbol{\Sigma} = \begin{pmatrix} \boldsymbol{\Sigma}_{11} & \boldsymbol{\Sigma}_{12} \\ \boldsymbol{\Sigma}_{21} & \boldsymbol{\Sigma}_{22} \end{pmatrix} \text{ with sizes } \begin{pmatrix} q \times q & q \times (N-q) \\ (N-q) \times q & (N-q) \times (N-q) \end{pmatrix}, \tag{6.29}$$

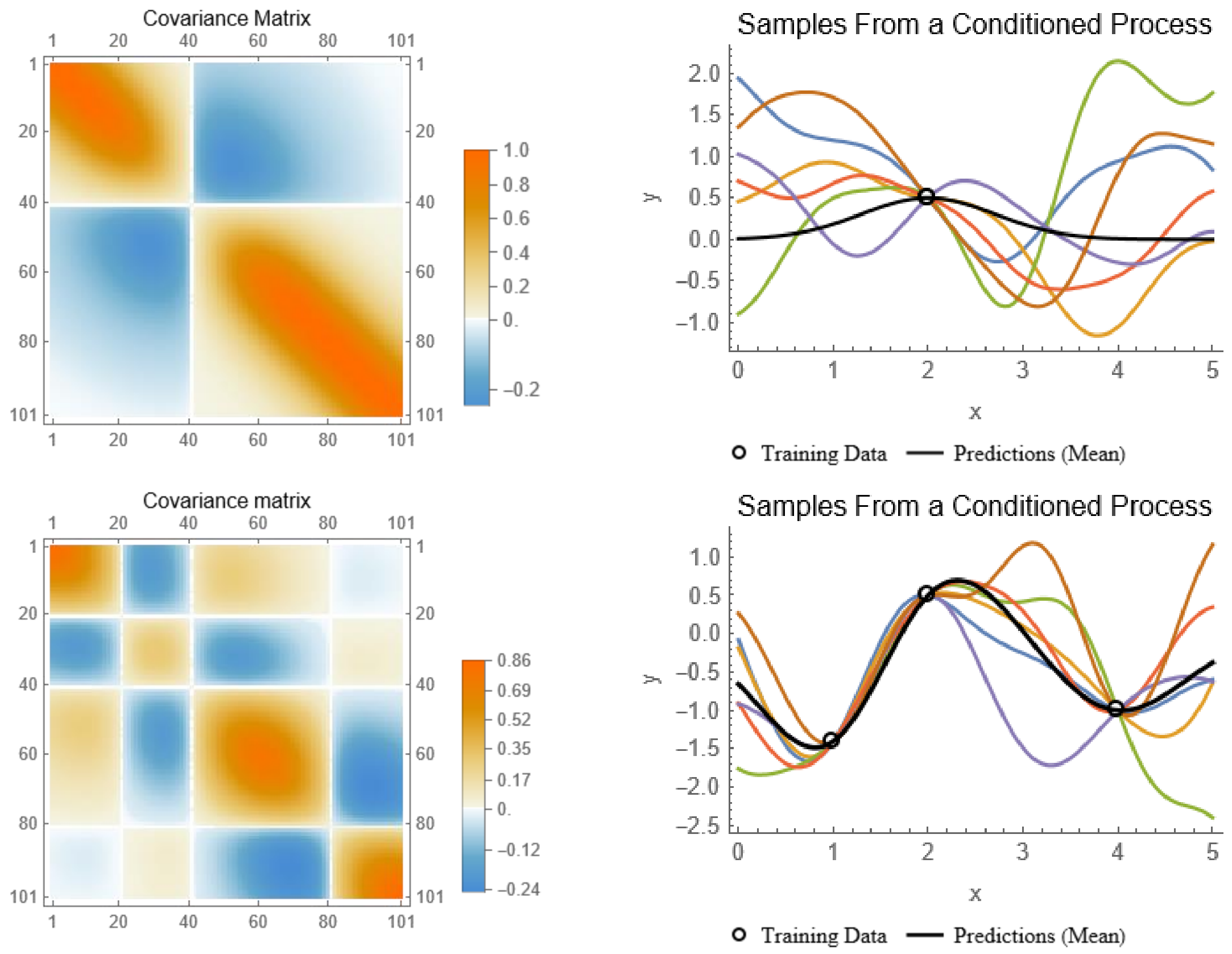

**Figure 6.19.** Visualization of conditioned GP with samples and predictions. The top-left panel displays the updated covariance matrix after conditioning on the training data. The heatmap visualization highlights changes in the correlations between input points, reflecting the impact of the observed data on the covariance structure. The top-right panel showcases six random samples generated from the conditioned GP. Each sampled function is depicted as a line plot, providing visualizations of different realizations of the conditioned process. Additionally, the open markers represent the observed training data points. In the top panels, only a single training data point is selected. However, in the bottom panels, three training data points are selected.

then the distribution of $\boldsymbol{X}_1$ conditional on $\boldsymbol{X}_2 = \mathbf{x}_2$ is the multivariate normal distribution [184]

$$(\boldsymbol{X}_1|\boldsymbol{X}_2 = \mathbf{x}_2) \sim N(\widehat{\boldsymbol{\mu}}, \widehat{\boldsymbol{\Sigma}}), \quad (6.30.1)$$

where

$$\widehat{\boldsymbol{\mu}} = \boldsymbol{\mu}_1 + \boldsymbol{\Sigma}_{12} \cdot \boldsymbol{\Sigma}_{22}^{-1} \cdot (\mathbf{x}_2 - \boldsymbol{\mu}_2), \quad (6.30.2)$$

$$\widehat{\boldsymbol{\Sigma}} = \boldsymbol{\Sigma}_{11} - \boldsymbol{\Sigma}_{12} \cdot \boldsymbol{\Sigma}_{22}^{-1} \cdot \boldsymbol{\Sigma}_{21}. \quad (6.30.3)$$

In the bivariate case (Theorem 6.3) where $\boldsymbol{X}$ is partitioned into $X_1$ and $X_2$, the conditional distribution is

$$(X_1|X_2 = \mathrm{x}_2) \sim N\left[\mu_1 + \rho\frac{\sigma_1}{\sigma_2}(\mathrm{x}_2 - \mu_2), \sigma_1^2(1-\rho^2)\right]. \quad (6.31)$$

This formulation beautifully highlights how knowledge of one set of variables (here, $\boldsymbol{X}_2$) influences our understanding and expectations about another set (here, $\boldsymbol{X}_1$), emphasizing the power of conditioning in reducing uncertainty in multivariate normal distributions.

**Remarks:**

- Note that knowing that $\boldsymbol{X}_2 = \mathbf{x}_2$ alters the variance, though the new variance does not depend on the specific value of $\mathbf{x}_2$. Reduction in variance when conditioning on data is a hallmark of statistical learning. Incorporating data into a statistical model helps us reduce uncertainty about the phenomena we're studying.

This process is fundamentally about updating our beliefs or predictions based on the evidence provided by data. Initially, without conditioning on any specific data, predictions can vary widely because the model hasn't learned any patterns yet. At this stage, there's high uncertainty. When we introduce data into the model, it "learns" from the specifics of that data. After the model has been trained with data, it generally becomes better at predicting outcomes because it now represents a refined synthesis of prior knowledge and empirical evidence. This refinement usually results in a reduction in the variance of predictions and thus less uncertainty.

- Perhaps more surprisingly, the mean is shifted by $\mathbf{\Sigma}_{12}\mathbf{\Sigma}_{22}^{-1}(\mathbf{x}_2 - \boldsymbol{\mu}_2)$. It's a direct illustration of how additional information (in this case, the value of $\boldsymbol{X}_2$) can refine our estimates. When we don't know the value of $\boldsymbol{X}_2$, the distribution of $\boldsymbol{X}_1$ is simply $N(\boldsymbol{\mu}_1, \mathbf{\Sigma}_{11})$. This represents our best guess about $\boldsymbol{X}_1$ based solely on its own statistics, ignoring any correlation it has with $\boldsymbol{X}_2$. However, when the value of $\boldsymbol{X}_2$ is known to be $\mathbf{x}_2$, the mean of $\boldsymbol{X}_1$ is adjusted to $\boldsymbol{\mu}_1 + \mathbf{\Sigma}_{12}\mathbf{\Sigma}_{22}^{-1}(\mathbf{x}_2 - \boldsymbol{\mu}_2)$. This shift in the mean reflects a more informed prediction about $\boldsymbol{X}_1$ based on the specific value of $\boldsymbol{X}_2$.
- When we condition on the value of $\boldsymbol{X}_2 = \mathbf{x}_2$, we are essentially restricting our analysis to instances where this condition holds true. This means that we are focusing only on a subset of the possible outcomes of the random vector $\boldsymbol{X}$.
- By conditioning on $\boldsymbol{X}_2 = \mathbf{x}_2$, we are incorporating additional information into our analysis. This additional information comes from the observed value of $\boldsymbol{X}_2$, which provides insights into the relationship between $\boldsymbol{X}_1$ and $\boldsymbol{X}_2$.
- In other words, conditioning on $\boldsymbol{X}_2 = \mathbf{x}_2$ allows us to update our beliefs about the distribution of $\boldsymbol{X}_1$. Prior to observing $\boldsymbol{X}_2 = \mathbf{x}_2$, we may have had a certain belief about the distribution of $\boldsymbol{X}_1$, characterized by its mean and variance. However, after observing $\boldsymbol{X}_2 = \mathbf{x}_2$, we update our beliefs about $\boldsymbol{X}_1$ based on this new information. This process of updating our beliefs leads to a reduction in uncertainty regarding $\boldsymbol{X}_1$. Prior to observing $\boldsymbol{X}_2 = \mathbf{x}_2$, there might have been some uncertainty about the possible values of $\boldsymbol{X}_1$. However, after conditioning on $\boldsymbol{X}_2 = \mathbf{x}_2$ and updating our beliefs, this uncertainty decreases. This reduction in uncertainty is reflected in the narrower spread of the conditional distribution of $\boldsymbol{X}_1$ given $\boldsymbol{X}_2 = \mathbf{x}_2$, compared to the original distribution of $\boldsymbol{X}_1$.

To derive the GP predictive distribution (posterior distribution) for a new observation $D_1 = (X, Y), X = x$ and $Y = y$, given existing observations, we begin by considering the training dataset $D_n = (\boldsymbol{X}_n, \boldsymbol{Y}_n)$. This dataset consists of: $\boldsymbol{X}_n = (x_1, \ldots, x_n)$: a set of training input points. $\boldsymbol{Y}_n = (\mathrm{y}_1, \ldots, \mathrm{y}_n)$: corresponding ground truth values at these points. Thus, the training data is represented as the set of pairs $\{(x_1, y_1), \ldots, (x_n, y_n)\}$. The key step in deriving the posterior distribution involves setting up the joint distribution between the new test point $Y(x)$ and the training data points $\boldsymbol{Y}_n$. This setup is crucial as it leverages the property of GPs, where any finite set of points (in this case $Y(x)$ along with $\boldsymbol{Y}_n$) follows a multivariate normal distribution. The result is a multivariate Gaussian distribution,

$$\begin{pmatrix} Y \\ \boldsymbol{Y}_n \end{pmatrix} \sim N\left[\mathbf{0}, \begin{pmatrix} \Sigma(x,x) & \mathbf{\Sigma}(x,\boldsymbol{X}_n) \\ \mathbf{\Sigma}(\boldsymbol{X}_n,x) & \mathbf{\Sigma}_n(\boldsymbol{X}_n,\boldsymbol{X}_n) \end{pmatrix}\right] = N\left[\mathbf{0}, \begin{pmatrix} \big(\Sigma(x,x)\big) & (\Sigma(x,x_1), \quad \ldots, \quad \Sigma(x,x_n)) \\ \begin{pmatrix} \Sigma(x,x_1) \\ \vdots \\ \Sigma(x,x_n) \end{pmatrix} & \begin{pmatrix} \Sigma(x_1,x_1) & \cdots & \Sigma(x_1,x_n) \\ \vdots & \ddots & \vdots \\ \Sigma(x_n,x_1) & \cdots & \Sigma(x_n,x_n) \end{pmatrix} \end{pmatrix}\right]. \tag{6.32.1}$$

Where:

- $\Sigma(x,x) = 1$ (for simplicity, assuming a unit variance in the squared exponential covariance function).
- $\mathbf{\Sigma}(x, \boldsymbol{X}_n)$ is a $1 \times n$ covariance matrix between $Y(x)$ and $\boldsymbol{Y}_n$,

$$\mathbf{\Sigma}(x,\boldsymbol{X}_n) = (\Sigma(x,x_1), \quad \ldots, \quad \Sigma(x,x_n)). \tag{6.32.2}$$

- $\mathbf{\Sigma}_n$ is the $n \times n$ covariance matrix for $\boldsymbol{Y}_n$,

$$\mathbf{\Sigma}_n(\boldsymbol{X}_n,\boldsymbol{X}_n) = \begin{pmatrix} \Sigma(x_1,x_1) & \cdots & \Sigma(x_1,x_n) \\ \vdots & \ddots & \vdots \\ \Sigma(x_n,x_1) & \cdots & \Sigma(x_n,x_n) \end{pmatrix}. \tag{6.32.3}$$

- $\mathbf{\Sigma}(\boldsymbol{X}_n, x) = \mathbf{\Sigma}(x, \boldsymbol{X}_n)^T$ is a $n \times 1$ covariance matrix between $\boldsymbol{Y}_n$ and $Y(x)$ (due to the symmetry of the covariance function),

$$\boldsymbol{\Sigma}(\boldsymbol{X}_n, x) = \begin{pmatrix} \Sigma(x, x_1) \\ \vdots \\ \Sigma(x, x_n) \end{pmatrix}. \tag{6.32.4}$$

The dimension of this joint Gaussian distribution is $n+1$, where $n$ represents the number of training data points, and the additional one accounts for the test point $Y(x)$. The $n+1$ dimensional covariance matrix encloses all information about how each point (both training and test points) is related to every other point under the GP's assumptions. The top-left element is the variance at the new point, the top-right and bottom-left blocks represent the covariance between the test point and training points, and the bottom-right block is the covariance matrix for the training points.

To get the posterior distribution over functions we need to restrict this joint prior distribution to contain only those functions which agree with the observed data points. The conditional distribution of $Y(x)$ given $\boldsymbol{Y}_n$ ($\boldsymbol{Y}_n$ denoted as $D_n$) is a normal distribution. Applying (6.30) yields the following predictive distribution

$$Y(x)|D_n \sim N\big(\mu(x), \sigma^2(x)\big). \tag{6.33.1}$$

The formulas for the mean $\mu(x)$ and variance $\sigma^2(x)$ of this conditional distribution are derived using the properties of the multivariate normal distribution:

$$\underbrace{\mu(x)}_{1\times 1} = \underbrace{\boldsymbol{\Sigma}(x, \boldsymbol{X}_n)}_{1\times n} \cdot \underbrace{\big(\boldsymbol{\Sigma}_n(\boldsymbol{X}_n, \boldsymbol{X}_n)\big)^{-1}}_{n\times n} \cdot \underbrace{\boldsymbol{Y}_n}_{n\times 1}. \tag{6.33.2}$$

Here, $\boldsymbol{\Sigma}(x, \boldsymbol{X}_n) \cdot \big(\boldsymbol{\Sigma}_n(\boldsymbol{X}_n, \boldsymbol{X}_n)\big)^{-1}$ represents the weights given to each observation in $\boldsymbol{Y}_n$ in predicting $Y(x)$.

$$\underbrace{\sigma^2(x)}_{1\times 1} = \underbrace{\Sigma(x, x)}_{1\times 1} - \underbrace{\boldsymbol{\Sigma}(x, \boldsymbol{X}_n)}_{1\times n} \cdot \underbrace{\big(\boldsymbol{\Sigma}_n(\boldsymbol{X}_n, \boldsymbol{X}_n)\big)^{-1}}_{n\times n} \cdot \underbrace{\boldsymbol{\Sigma}(\boldsymbol{X}_n, x)}_{n\times 1}. \tag{6.33.3}$$

The term $\boldsymbol{\Sigma}(x, \boldsymbol{X}_n) \cdot \big(\boldsymbol{\Sigma}_n(\boldsymbol{X}_n, \boldsymbol{X}_n)\big)^{-1} \cdot \boldsymbol{\Sigma}(\boldsymbol{X}_n, x)$ is the reduction in variance due to the information gained from $\boldsymbol{Y}_n$.

Thus, the GP predictive distribution for $Y(x)$ given the data $D_n$ is:

$$Y(x)|D_n \sim N\left(\boldsymbol{\Sigma}(x, \boldsymbol{X}_n) \cdot \big(\boldsymbol{\Sigma}_n(\boldsymbol{X}_n, \boldsymbol{X}_n)\big)^{-1} \cdot \boldsymbol{Y}_n, \Sigma(x, x) - \boldsymbol{\Sigma}(x, \boldsymbol{X}_n) \cdot \big(\boldsymbol{\Sigma}_n(\boldsymbol{X}_n, \boldsymbol{X}_n)\big)^{-1} \cdot \boldsymbol{\Sigma}(\boldsymbol{X}_n, x)\right). \tag{6.33.4}$$

### 6.7.4. Posterior Distribution with Noise-free Observations (General Case)

Indeed, the "pointwise" prediction method using GPs, often referred to as kriging in geostatistics, is effective for individual predictions. However, when predicting at multiple new locations $\boldsymbol{X}$ simultaneously, considering the joint distribution of all these points allows for a richer analysis that captures the correlations among predictions. This approach is especially important in fields like geospatial analysis, where spatial relationships significantly influence the predictions. Now, we have a training set $\mathfrak{D}_n$ of $n$ observations, $\mathfrak{D}_n = \{(\mathbf{x}_i, y_i) \mid i = 1, \dots, n\}$, where $\mathbf{x}$ denotes an input vector of dimension $d$ and $y$ denotes a scalar output or target (dependent variable); the column vector inputs for all $n$ cases are aggregated in the $d \times n$ design matrix $\boldsymbol{X}_n$, and the targets are collected in the vector $\boldsymbol{Y}_n$, so we can write $\mathfrak{D}_n = (\boldsymbol{X}_n, \boldsymbol{Y}_n)$. Similarly, we have a test set $\mathfrak{D}_m$ of $m$ observations, $\mathfrak{D}_m = \{(\bar{\mathbf{x}}_i, \bar{y}_i) \mid i = 1, \dots, m\} = (\bar{\boldsymbol{X}}_m, \bar{\boldsymbol{Y}}_m)$. In the regression setting the targets are real values. We are interested in making inferences about the relationship between inputs and targets, i.e. the conditional distribution of the targets given the inputs.

When extending GP predictions to multiple test locations simultaneously, you create a predictive distribution for a vector of outputs $\bar{\boldsymbol{Y}}_m$, where $\bar{\boldsymbol{X}}_m$ represents a set of $m$ new test input locations, given the data $\mathfrak{D}_n$. Graphically in Figure 6.19 you may think of generating functions from the prior, and rejecting the ones that disagree with the observations, although this strategy would not be computationally very efficient. Fortunately, in probabilistic terms this operation is extremely simple, corresponding to conditioning the joint Gaussian prior distribution on the observations.

The joint multivariate Gaussian distribution is,

$$\begin{pmatrix}\overline{\boldsymbol{Y}}_m\\ \boldsymbol{Y}_n\end{pmatrix} = f\begin{pmatrix}\overline{\boldsymbol{X}}_m\\ \boldsymbol{X}_n\end{pmatrix} \sim N\left[\mathbf{0}, \begin{pmatrix}\boldsymbol{\Sigma}_m(\overline{\boldsymbol{X}}_m, \overline{\boldsymbol{X}}_m) & \boldsymbol{\Sigma}_{mn}(\overline{\boldsymbol{X}}_m, \boldsymbol{X}_n)\\ \boldsymbol{\Sigma}_{nm}(\boldsymbol{X}_n, \overline{\boldsymbol{X}}_m) & \boldsymbol{\Sigma}_n(\boldsymbol{X}_n, \boldsymbol{X}_n)\end{pmatrix}\right]$$
$$= N\left[\mathbf{0}, \begin{pmatrix}\begin{pmatrix}\Sigma(\overline{\mathbf{x}}_1, \overline{\mathbf{x}}_1) & \cdots & \Sigma(\overline{\mathbf{x}}_1, \overline{\mathbf{x}}_m)\\ \vdots & \ddots & \vdots\\ \Sigma(\overline{\mathbf{x}}_m, \overline{\mathbf{x}}_1) & \cdots & \Sigma(\overline{\mathbf{x}}_m, \overline{\mathbf{x}}_m)\end{pmatrix} & \begin{pmatrix}\Sigma(\overline{\mathbf{x}}_1, \mathbf{x}_1) & \cdots & \Sigma(\overline{\mathbf{x}}_1, \mathbf{x}_n)\\ \vdots & \ddots & \vdots\\ \Sigma(\overline{\mathbf{x}}_m, \mathbf{x}_1) & \cdots & \Sigma(\overline{\mathbf{x}}_m, \mathbf{x}_n)\end{pmatrix}\\ \begin{pmatrix}\Sigma(\mathbf{x}_1, \overline{\mathbf{x}}_1) & \cdots & \Sigma(\mathbf{x}_1, \overline{\mathbf{x}}_m)\\ \vdots & \ddots & \vdots\\ \Sigma(\mathbf{x}_n, \overline{\mathbf{x}}_1) & \cdots & \Sigma(\mathbf{x}_n, \overline{\mathbf{x}}_m)\end{pmatrix} & \begin{pmatrix}\Sigma(\mathbf{x}_1, \mathbf{x}_1) & \cdots & \Sigma(\mathbf{x}_1, \mathbf{x}_n)\\ \vdots & \ddots & \vdots\\ \Sigma(\mathbf{x}_n, \mathbf{x}_1) & \cdots & \Sigma(\mathbf{x}_n, \mathbf{x}_n)\end{pmatrix}\end{pmatrix}\right]. \quad (6.34.1)$$

Note that, when utilizing vectors as inputs, the squared exponential covariance function can be expressed as

$$\Sigma(\mathbf{x}, \mathbf{x}') = k(\mathbf{x}, \mathbf{x}') = e^{-\frac{\|\mathbf{x}-\mathbf{x}'\|^2}{2l^2}}. \quad (6.34.2)$$

If there are $n$ training points and $m$ test points then $\boldsymbol{\Sigma}_{mn}(\overline{\boldsymbol{X}}_m, \boldsymbol{X}_n)$ denotes the $m \times n$ matrix of the covariances evaluated at all pairs of training and test points, and similarly for the other entries $\boldsymbol{\Sigma}_m(\overline{\boldsymbol{X}}_m, \overline{\boldsymbol{X}}_m)$, $\boldsymbol{\Sigma}_{nm}(\boldsymbol{X}_n, \overline{\boldsymbol{X}}_m)$ and $\boldsymbol{\Sigma}_n(\boldsymbol{X}_n, \boldsymbol{X}_n)$.

The predictive distribution is:

$$\overline{\boldsymbol{Y}}_m(\overline{\boldsymbol{X}})|\mathfrak{D}_n \sim N_m\big(\boldsymbol{\mu}(\overline{\boldsymbol{X}}_m), \boldsymbol{\sigma}^2(\overline{\boldsymbol{X}}_m)\big), \quad (6.35.1)$$

where

$$\underbrace{\boldsymbol{\mu}(\overline{\boldsymbol{X}}_m)}_{m\times 1} = \underbrace{\boldsymbol{\Sigma}(\overline{\boldsymbol{X}}_m, \boldsymbol{X}_n)}_{m\times n} \cdot \underbrace{\big(\boldsymbol{\Sigma}_n(\boldsymbol{X}_n, \boldsymbol{X}_n)\big)^{-1}}_{n\times n} \cdot \underbrace{\boldsymbol{Y}_n}_{n\times 1}, \quad (6.35.2)$$

$$\underbrace{\boldsymbol{\sigma}^2(\overline{\boldsymbol{X}}_m)}_{m\times m} = \underbrace{\boldsymbol{\Sigma}_m(\overline{\boldsymbol{X}}_m, \overline{\boldsymbol{X}}_m)}_{m\times m} - \underbrace{\boldsymbol{\Sigma}_{mn}(\overline{\boldsymbol{X}}_m, \boldsymbol{X}_n)}_{m\times n} \cdot \underbrace{\big(\boldsymbol{\Sigma}_n(\boldsymbol{X}_n, \boldsymbol{X}_n)\big)^{-1}}_{n\times n} \cdot \underbrace{\boldsymbol{\Sigma}_{nm}(\boldsymbol{X}_n, \overline{\boldsymbol{X}}_m)}_{n\times m}. \quad (6.35.3)$$

$\boldsymbol{\Sigma}_{mn}(\overline{\boldsymbol{X}}_m, \boldsymbol{X}_n) \cdot \big(\boldsymbol{\Sigma}_n(\boldsymbol{X}_n, \boldsymbol{X}_n)\big)^{-1} \cdot \boldsymbol{\Sigma}_{nm}(\boldsymbol{X}_n, \overline{\boldsymbol{X}}_m)$ represents the information from the existing data that influences the covariance among the new points.

Figure 6.19 illustrates the GP predictions conditioned on training data. The top right panel illustrates six random samples generated from a GP conditioned on a single training point ($n = 1$) with $m = 101$ test points. The open markers on the plot indicate the single training data point used for conditioning the GP. Each sample from the predictive distribution is a curve consisting of 101 points. The dimension of the vector $\boldsymbol{\mu}(\overline{\boldsymbol{X}}_m)$ is $101 \times 1$ and the covariance matrix $\boldsymbol{\sigma}^2(\overline{\boldsymbol{X}}_m)$ is $101 \times 101$, showcasing how the GP adapts to a minimal amount of training data. The bottom right panel of this figure shows six random samples generated from a GP conditioned on three training points ($n = 3$) with $m = 101$ test points. Similar to the top right panel, open markers indicate the training data points used for conditioning. Each sample from the predictive distribution, $\overline{\boldsymbol{Y}}_m(\overline{\boldsymbol{X}})|\mathfrak{D}_n \sim N_m\big(\boldsymbol{\mu}(\overline{\boldsymbol{X}}_m), \boldsymbol{\sigma}^2(\overline{\boldsymbol{X}}_m)\big)$, is again a curve of 101 points, detailed by the vector $\boldsymbol{\mu}(\overline{\boldsymbol{X}}_m)$ and covariance matrix $\boldsymbol{\sigma}^2(\overline{\boldsymbol{X}}_m)$, illustrating the impact of additional training data on the model's predictions and uncertainty. The mean of the GP predictions is plotted as a thick black line across the entire range of input values. This line represents the GP regression's best estimate of the underlying function after considering the training data. The spread and diversity of the sample paths around the mean prediction visually represent the uncertainty in the model's predictions.

### 6.7.5. Posterior Distribution Using Noisy Observations

Up until now, we have considered the training points $\boldsymbol{Y}_n$ to be perfect measurements, where $\mathfrak{D}_n = (\boldsymbol{X}_n, \boldsymbol{Y}_n)$ is a training set of examples and ground truth values. But in real-world scenarios, this is an unrealistic assumption, since most of our data is afflicted with measurement errors or uncertainty. We have assumed that the observed values $y$ differ from the function values $f(\mathbf{x})$ by additive noise [183],

$$y = f(\mathbf{x}) + \epsilon, \quad (6.36.1)$$

and we will further assume that this noise follows an independent, identically distributed Gaussian distribution with zero mean and variance $\sigma_n^2$

$$\epsilon \sim N(0, \sigma_n^2). \quad (6.36.2)$$

Similarly, the result is a joint multivariate Gaussian distribution,

$$\begin{aligned}
\begin{pmatrix}\overline{\boldsymbol{Y}}_m\\ \boldsymbol{Y}_n\end{pmatrix} &= f\begin{pmatrix}\overline{\boldsymbol{X}}_m\\ \boldsymbol{X}_n\end{pmatrix} + \begin{pmatrix}\mathbf{0}\\ \boldsymbol{\epsilon}_n\end{pmatrix}\\
&\sim N\left[\mathbf{0}, \begin{pmatrix}\boldsymbol{\Sigma}_m(\overline{\boldsymbol{X}}_m,\overline{\boldsymbol{X}}_m) & \boldsymbol{\Sigma}_{mn}(\overline{\boldsymbol{X}}_m,\boldsymbol{X}_n)\\ \boldsymbol{\Sigma}_{nm}(\boldsymbol{X}_n,\overline{\boldsymbol{X}}_m) & \boldsymbol{\Sigma}_n(\boldsymbol{X}_n,\boldsymbol{X}_n)\end{pmatrix}\right] + N\left[\mathbf{0}, \begin{pmatrix}\mathbf{0} & \mathbf{0}\\ \mathbf{0} & \sigma_n^2\,\mathbf{I}\end{pmatrix}\right]\\
&= N\left[\mathbf{0}, \begin{pmatrix}\boldsymbol{\Sigma}_m(\overline{\boldsymbol{X}}_m,\overline{\boldsymbol{X}}_m) & \boldsymbol{\Sigma}_{mn}(\overline{\boldsymbol{X}}_m,\boldsymbol{X}_n)\\ \boldsymbol{\Sigma}_{nm}(\boldsymbol{X}_n,\overline{\boldsymbol{X}}_m) & \boldsymbol{\Sigma}_n(\boldsymbol{X}_n,\boldsymbol{X}_n)+\sigma_n^2\,\mathbf{I}\end{pmatrix}\right]\\
&= N\left[\mathbf{0}, \begin{pmatrix}\begin{pmatrix}\Sigma(\bar{\mathbf{x}}_1,\bar{\mathbf{x}}_1) & \cdots & \Sigma(\bar{\mathbf{x}}_1,\bar{\mathbf{x}}_m)\\ \vdots & \ddots & \vdots\\ \Sigma(\bar{\mathbf{x}}_m,\bar{\mathbf{x}}_1) & \cdots & \Sigma(\bar{\mathbf{x}}_m,\bar{\mathbf{x}}_m)\end{pmatrix} & \begin{pmatrix}\Sigma(\bar{\mathbf{x}}_1,\mathbf{x}_1) & \cdots & \Sigma(\bar{\mathbf{x}}_1,\mathbf{x}_n)\\ \vdots & \ddots & \vdots\\ \Sigma(\bar{\mathbf{x}}_m,\mathbf{x}_1) & \cdots & \Sigma(\bar{\mathbf{x}}_m,\mathbf{x}_n)\end{pmatrix}\\ \begin{pmatrix}\Sigma(\mathbf{x}_1,\bar{\mathbf{x}}_1) & \cdots & \Sigma(\mathbf{x}_1,\bar{\mathbf{x}}_m)\\ \vdots & \ddots & \vdots\\ \Sigma(\mathbf{x}_n,\bar{\mathbf{x}}_1) & \cdots & \Sigma(\mathbf{x}_n,\bar{\mathbf{x}}_m)\end{pmatrix} & \begin{pmatrix}\Sigma(\mathbf{x}_1,\mathbf{x}_1)+\sigma_n^2 & \cdots & \Sigma(\mathbf{x}_1,\mathbf{x}_n)\\ \vdots & \ddots & \vdots\\ \Sigma(\mathbf{x}_n,\mathbf{x}_1) & \cdots & \Sigma(\mathbf{x}_n,\mathbf{x}_n)+\sigma_n^2\end{pmatrix}\end{pmatrix}\right].
\end{aligned} \tag{6.37}$$

In this case, the predictive distribution is:

$$\overline{\boldsymbol{Y}}_m(\overline{\boldsymbol{X}})|\mathfrak{D}_n \sim N_m\big(\boldsymbol{\mu}(\overline{\boldsymbol{X}}_m), \boldsymbol{\sigma}^2(\overline{\boldsymbol{X}}_m)\big), \tag{6.38.1}$$

where

$$\underbrace{\boldsymbol{\mu}(\overline{\boldsymbol{X}}_m)}_{m\times 1} = \underbrace{\boldsymbol{\Sigma}(\overline{\boldsymbol{X}}_m,\boldsymbol{X}_n)}_{m\times n}\cdot\underbrace{(\boldsymbol{\Sigma}_n(\boldsymbol{X}_n,\boldsymbol{X}_n)+\sigma_n^2\,\mathbf{I})^{-1}}_{n\times n}\cdot\underbrace{\boldsymbol{Y}_n}_{n\times 1}, \tag{6.38.2}$$

$$\underbrace{\boldsymbol{\sigma}^2(\overline{\boldsymbol{X}}_m)}_{m\times m} = \underbrace{\boldsymbol{\Sigma}_m(\overline{\boldsymbol{X}}_m,\overline{\boldsymbol{X}}_m)}_{m\times m} - \underbrace{\boldsymbol{\Sigma}_{mn}(\overline{\boldsymbol{X}}_m,\boldsymbol{X}_n)}_{m\times n}\cdot\underbrace{(\boldsymbol{\Sigma}_n(\boldsymbol{X}_n,\boldsymbol{X}_n)+\sigma_n^2\,\mathbf{I})^{-1}}_{n\times n}\cdot\underbrace{\boldsymbol{\Sigma}_{nm}(\boldsymbol{X}_n,\overline{\boldsymbol{X}}_m)}_{n\times m}. \tag{6.38.3}$$

Let's summarize the key points of this section. In regression, the goal is to predict a continuous output variable (dependent variable) based on one or more input variables (independent variables). In GPs, we model the relationship between input and output variables as a distribution over functions. This distribution serves as our prior belief about the true underlying function that generates the data. Instead of specifying a single functional form or parametric model, GPs consider a space of infinitely many functions. Each function in this space represents a possible relationship between the input and output variables. Before observing any data, we express our uncertainty about the true function by specifying a prior distribution over these functions. This prior captures our beliefs about the smoothness, regularity, and other properties of the underlying function. We then use Bayesian inference to update this prior belief based on observed training data. Bayesian inference involves combining our prior belief (prior distribution) with the likelihood of the data given the model (likelihood function) to obtain a posterior distribution, which represents our updated belief about the true function given the observed data. Learning in GPs is achieved by conditioning the prior distribution on the observed training data.

## 6.8 Tuning Hyperparameters with Bayesian Optimization

BO for tuning hyperparameters of a NN can be thought of as a very efficient and intelligent way to conduct a treasure hunt where you're trying to find the best settings for your NN that yield the highest performance (e.g., highest accuracy or lowest error). The process cleverly balances exploring new areas and exploiting known promising areas to efficiently find the treasure. Procedure 6.6 represents how tuning hyperparameters with BO typically works.

For a helpful analogy, imagine you have a map (the hyperparameter space), where each spot on the map can potentially hide a treasure (optimal performance of the NN). Some spots are more likely to have treasure than others, but you don’t know where to start digging. The hyperparameters might include things like the learning rate, the number of layers, or the number of neurons in each layer. You start by randomly choosing a few spots (initial set of hyperparameters) to dig (train and evaluate the NN). These initial guesses give you some preliminary information about where the treasure might be hidden — some spots might yield better results than others, indicating they are closer to the treasure. Using the results from your initial digs, you build a model (surrogate model, usually a GP) that represents your beliefs about where the treasure is likely to be hidden. This model makes predictions about untested spots on the map and estimates how uncertain these predictions are. Areas with high uncertainty are less explored, whereas areas with promising results are considered more likely to contain the treasure. To decide where to dig next, you use an ACF. This function helps you balance between places near where you found some treasure (exploitation, where the model predicts high performance) and places that are highly uncertain (exploration, where the outcome is not well known). This decision-making process is like choosing whether to dig near known treasures to find more or to explore new areas that might hide even greater treasures. Based on the ACF, you select the next spot (set of hyperparameters) to dig. You then train and evaluate the NN with these hyperparameters and observe the outcome. With every new digging result, you update your belief model. This update refines your map, making some areas more or less attractive based on the latest findings. You repeat this process, each time refining your map and getting closer to the best possible treasure (optimal hyperparameters). Eventually, after several iterations, you either find a spot that seems to be the best possible one, or you decide to stop because you've used up your resources (time, computation power) or the improvements become too small to justify further searches.

From the preceding discussion, it's clear that we face a fundamental trade-off between exploration and exploitation. To navigate this balance effectively, we can employ an ACF. This function acts as a strategic guide, helping us to optimize our approach. Let's delve deeper into these concepts to understand how they can be applied effectively.

### 6.8.1. Exploration

Exploration in BO is a strategy aimed at expanding the search into areas of the hyperparameter space that are less well-understood or where the surrogate model’s predictions (GP’s predictions) are highly uncertain. The goal of exploration is to discover new regions that might improve the overall optimization outcome, particularly in cases where the objective function has a complex landscape with multiple local minima or significant variability. In many optimization problems, especially with complex or non-convex functions such as those often found in NN training and hyperparameter tuning, there are multiple local minima. Without sufficient exploration, BO could converge prematurely to a suboptimal local minimum.

Exploration in BO is typically facilitated through the surrogate model and the choice of ACF. GPs, commonly used as surrogate models in BO, provide not only a prediction at each point in the hyperparameter space but also a measure of uncertainty (variance) about that prediction. Areas with high uncertainty are targets for exploration as the model is less certain about the function's behavior in these regions.

Challenges in Exploration

- As the number of hyperparameters increases, the volume of the hyperparameter space grows exponentially (the "curse of dimensionality"). This growth makes thorough exploration more challenging because the amount of data needed to reduce uncertainty across space increases dramatically.
- Exploration can be computationally expensive. Evaluating points in less promising regions might lead to higher computational overheads without immediate benefits.

### 6.8.2. Exploitation

Exploitation in BO refers to the strategy of focusing on areas of the hyperparameter space where the surrogate model predicts high performance, based on the data observed so far. This approach aims to refine and concentrate the search around promising regions to optimize the objective function more efficiently. Focusing on areas known to yield good results helps in achieving better outcomes with fewer function evaluations, which is especially crucial when each evaluation (such as training a NN with a specific set of hyperparameters) is costly or time-consuming. Moreover, exploitation uses the information gained from previous evaluations effectively, leveraging the accuracy of the surrogate model in regions of the space where the model is more confident about its predictions.

Exploitation in BO typically revolves around the surrogate model's predictions, particularly focusing on the mean prediction as a guide to potential performance. GPs are often used because they provide a predicted mean and standard deviation for each point in the hyperparameter space. For exploitation, the focus is on the predicted mean, with areas having higher predicted means being preferred for sampling.

**Procedure 6.6:** Tuning Hyperparameters with BO

1. Define the hyperparameter space: Identify and define the range of values for each hyperparameter you wish to optimize. Common hyperparameters in NNs include learning rate, number of layers, number of neurons per layer, batch size, AFs, etc.

2. Choose a surrogate model: Select a surrogate model to approximate the objective function. GPs are commonly used because they provide a probabilistic measure of uncertainty in predictions, which is helpful for exploration in hyperparameter space.

3. Select an ACF: Choose an ACF that will guide the optimization by determining the next set of hyperparameters to evaluate.

4. Initialization: Start with an initial set of hyperparameters. This can be random or based on some heuristic. Evaluate the NN's performance (e.g., validation loss) using these initial hyperparameters.

5. Update the surrogate model: Use the results from the initial evaluations to update the surrogate model. This involves fitting the model to the known hyperparameter-performance pairs.

6. Optimize the ACF: Use the updated surrogate model to optimize the ACF. This step determines the next set of hyperparameters to evaluate. The optimization of the ACF typically involves a balance between exploring hyperparameters in less certain regions and exploiting hyperparameters that are likely to yield better performance.

7. Evaluate the chosen Hyperparameters: Implement the NN using the hyperparameters suggested by the ACF and evaluate its performance.

8. Iterate: Update the surrogate model with the new results. Repeat Steps 6 and 7 until a stopping criterion is met, such as a maximum number of evaluations, a time limit, or convergence of the NN performance.

9. Post-optimization: Once the optimization process is complete, analyze the results to identify the best set of hyperparameters. Optionally, perform a final training of the NN using these hyperparameters to confirm their effectiveness.

Challenges in Exploitation

- A significant risk with a strong focus on exploitation is premature convergence to local minima. If the surrogate model's early predictions favor a particular region incorrectly or if the region only contains a local optimum, the optimization might settle there without exploring other potentially better regions.
- Properly balancing exploitation with exploration is critical. Overemphasis on exploitation can lead to under-exploration of the hyperparameter space, potentially missing out on globally optimal regions.
- Exploitation heavily relies on the accuracy and predictive power of the surrogate model. In regions where the model is incorrect or uncertain (but doesn't express high uncertainty), decisions based on such predictions can lead to suboptimal outcomes.

### 6.8.3. Balancing Exploration and Exploitation

The challenge in BO is to balance these two strategies effectively. Focusing too much on exploitation might lead to premature convergence on local minima, especially in complex landscapes with multiple optima. Conversely, excessive exploration can lead to inefficiency and wasted resources, as the optimization process might spend too much time evaluating less promising regions. This balance directly influences the efficiency and effectiveness of the search process. This balance is not static but should evolve over the course of the optimization process, guided by both the outcomes observed and the increasing accuracy of the surrogate model. Effective management of this balance can dramatically affect the success of the optimization process, making it one of the key skills in applying BO successfully.

**Procedure 6.7:** UCB

1. Start with some initial data points for which the objective function values are known.
2. Fit a GP to the known data points.
3. Use the UCB function to select the next point to evaluate by maximizing the UCB.
4. Evaluate the objective function at this new point.
5. Update the GP model with the new data point.
6. Repeat until a stopping criterion is met (e.g., a maximum number of evaluations).

Effective ACFs in BO are designed to strike a balance between these two strategies to optimize the objective function efficiently. You might consider using libraries such as Mathematic which provide implementations of GPs and various ACFs, making it easier to apply BO.

## 6.9 Acquisition Functions

ACFs [188,190] are used to decide where to sample next in the input space of the objective function. The choice of where to sample next is based on the predictions and uncertainties from the probabilistic model. Mathematically, we select the next evaluation point $\mathbf{x}_{t+1}$ by maximizing the ACF.

$$\mathbf{x}_{t+1} = \underset{\mathbf{x}\in\chi}{\operatorname{argmax}}\, \mathrm{ACF}(\mathbf{x}_t), \tag{6.39}$$

where $\chi$ is often a compact subset of $\mathbb{R}^d$ and represents the design space of interest (hyperparameter space), and $\mathbf{x}$ is a $d$-dimensional decision vector (hyperparameter combinations). In the design of the ACF, we need to balance local exploitation and global exploration reasonably. There are several types of ACFs, for example, EI, PI, and UCB functions, etc.

### 6.9.1. Upper Confidence Bound

UCB contains explicit exploitation $\mu(\mathbf{x})$ and exploration $\sigma(\mathbf{x})$ terms. The UCB is defined by [185,189-193]

$$\mathrm{UCB}(\mathbf{x};\kappa) = \mu(\mathbf{x}) + \kappa\sigma(\mathbf{x}), \tag{6.40}$$

where $\mu(\mathbf{x})$ is the predicted mean of the objective function at point $\mathbf{x}$ given by the GP. $\sigma(\mathbf{x})$ is the standard deviation (a measure of uncertainty or prediction confidence) at point $\mathbf{x}$. $\kappa$ is a tunable parameter that determines the trade-off between exploration and exploitation. A higher $\kappa$ encourages more exploration, potentially discovering better maxima at the risk of more function evaluations in less promising areas. A lower $\kappa$ promotes exploitation, focusing on areas already identified as potentially optimal. In BO, using the UCB function to select the next point for evaluating the objective function involves maximizing the UCB value across the domain of the function you're trying to optimize. Procedure 6.7 represents how UCB typically works.

The rationale behind maximizing the UCB function is to find the point $\mathbf{x}$ where the sum of predicted value and scaled uncertainty is highest. This strategy aims to identify locations where the potential for improvement is the greatest—either because the model predicts a high value for $\mu(\mathbf{x})$ or because the uncertainty $\sigma(\mathbf{x})$ is high, suggesting insufficient exploration.

- Exploration (high $\sigma(\mathbf{x})$): By including $\sigma(\mathbf{x})$ in the UCB, the algorithm is encouraged to explore regions with high uncertainty. These regions might hide better maxima than currently known but have been less sampled and thus the model is less certain about them.
- Exploitation (high $\mu(\mathbf{x})$): If a region has shown promising results (high mean), focusing more on that area might refine the model's understanding and inch closer to the local or global maximum.

When $\kappa = 0$, UCB becomes a purely exploitative scheme and therefore the solution with the best-predicted mean is evaluated expensively. Thus, it may rapidly converge to a local maximum prematurely. In contrast, when $\kappa$ is large, the optimization becomes purely exploratory, evaluating the location where the posterior uncertainty (variance) is largest, which is equivalent to maximally reducing the overall predictive entropy of the model. Consequently, it may eventually locate the global optima, but the rate of convergence may be very slow.

In practice, maximizing the UCB function involves a computational step where you evaluate the UCB function at various points in the domain (or use an optimization algorithm directly on the UCB function) to find the maximum UCB value. This point, where UCB is the maximum, is your next sampling point. After evaluating the objective function at this new point, the GP is updated with the new data (new $\mathbf{x}$ and its corresponding objective function outcome), which refines both the mean and uncertainty estimates for future iterations.

**Remarks:**

- Choosing the right value for $\kappa$ can be non-trivial and may require domain knowledge or experimentation.
- The performance of UCB can depend heavily on the quality of the underlying GP.
- In the context of minimization, the ACF would take the form

$$\text{LCB}(\mathbf{x};\kappa) = \mu(\mathbf{x}) - \kappa\sigma(\mathbf{x}), \tag{6.41}$$

where Cox and John [190] select the prediction site that has the smallest value of lower confidence bound (LCB) as the next evaluation point.

### 6.9.2. Probability of Improvement

The PI [189, 194] is another popular ACF used in BO. Like the UCB function, PI helps in making decisions about where to sample next in the function space. However, it does so by specifically targeting the probability that any given point will improve over the best objective value found so far.

In the context of maximizing a function $f(\mathbf{x})$, the improvement function $I(\mathbf{x})$ plays a pivotal role. It is defined as:

$$I(\mathbf{x}) = \max[f(\mathbf{x}) - f(\mathbf{x}^*), 0], \tag{6.42}$$

where $\mathbf{x}^*$ is the point yielding the highest value of $f(\mathbf{x})$ observed so far, $\mathbf{x}^* = \arg\max f(\mathbf{x}_i)$. This function distinguishes between scenarios where a new evaluation provides a better or worse outcome than the current best. When $f(\mathbf{x})$ is less than or equal to $f(\mathbf{x}^*)$, $I(\mathbf{x})$ equals zero. This indicates no gain from evaluating at point $\mathbf{x}$, as it does not surpass the best-known value. Conversely, if $f(\mathbf{x})$ exceeds $f(\mathbf{x}^*)$, $I(\mathbf{x})$ equals $f(\mathbf{x}) - f(\mathbf{x}^*)$. This positive result quantifies the exact improvement made over the current best, highlighting the benefit of sampling at $\mathbf{x}$.

BO leverages the improvement function through the PI function. PI assesses the likelihood that a new candidate point $\mathbf{x}$ will yield an outcome better than $f(\mathbf{x}^*)$:

$$\text{PI}(\mathbf{x}) = P(I(\mathbf{x}) > 0) = P\big(f(\mathbf{x}) > f(\mathbf{x}^*)\big). \tag{6.43}$$

We recall that given data $\mathcal{D}_n = \{(\mathbf{x}_1, y_1), \dots, (\mathbf{x}_n, y_n)\}$ the predictive distribution at $\mathbf{x}$ is the marginal GP $f(\mathbf{x})|\mathcal{D}_n \sim N(\mu(\mathbf{x}), \sigma^2(\mathbf{x}))$. To facilitate computation, we standardize $f(\mathbf{x})$ using the GP's predictive mean and standard deviation:

$$z(\mathbf{x}) = \frac{f(\mathbf{x}) - \mu(\mathbf{x})}{\sigma(\mathbf{x})} \sim N(0,1). \tag{6.44}$$

Using the standardized variable $z(\mathbf{x})$ and $z_0 = \frac{f(\mathbf{x}^*) - \mu(\mathbf{x})}{\sigma(\mathbf{x})}$, PI can be expressed as:

$$\begin{aligned}
\text{PI}(\mathbf{x}) &= P(I(\mathbf{x}) > 0) = P\big(f(\mathbf{x}) > f(\mathbf{x}^*)\big) \\
&= P\left(\underbrace{\frac{f(\mathbf{x}) - \mu(\mathbf{x})}{\sigma(\mathbf{x})}}_{z} > \underbrace{\frac{f(\mathbf{x}^*) - \mu(\mathbf{x})}{\sigma(\mathbf{x})}}_{z_0}\right) = P(z > z_0) \\
&= \int_{z_0}^{\infty} \phi(z)dz = 1 - \int_{-\infty}^{z_0} \phi(z)dz \\
&= 1 - \Phi(z_0) \\
&= \Phi(-z_0) \\
&= \Phi\left(\frac{\mu(\mathbf{x}) - f(\mathbf{x}^*)}{\sigma(\mathbf{x})}\right).
\end{aligned} \tag{6.45}$$

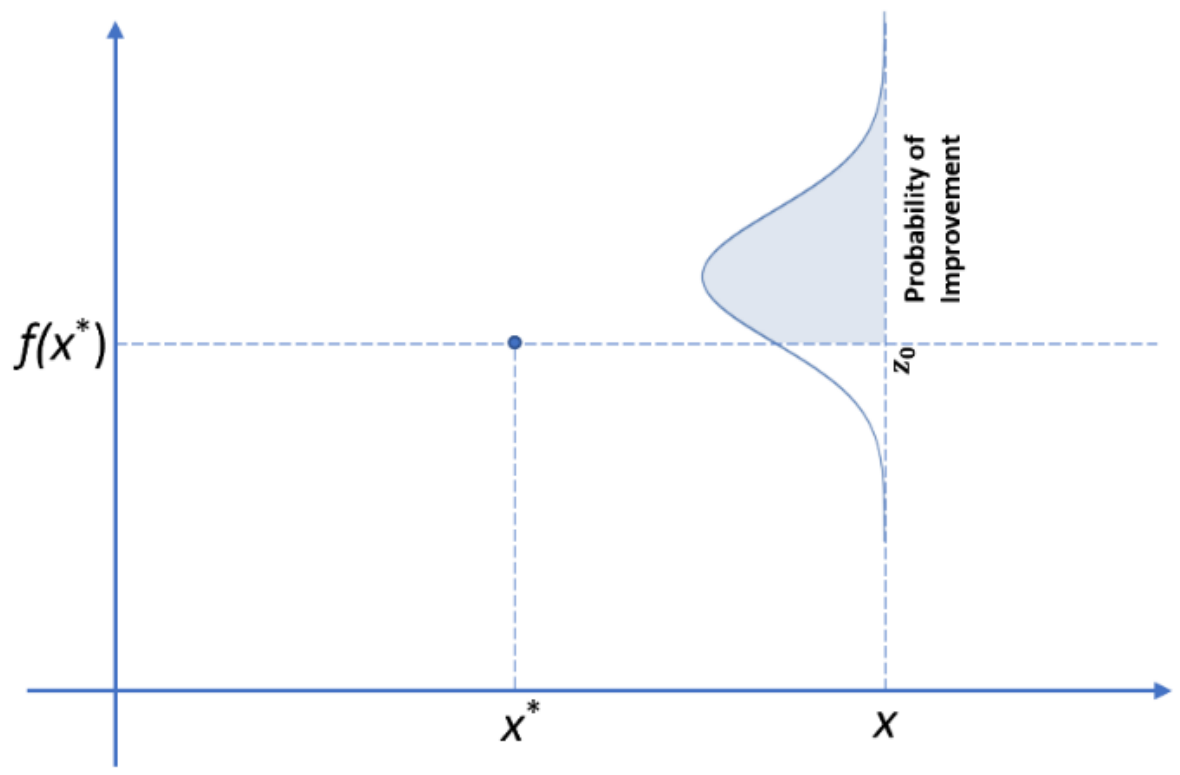

**Figure 6.20.** The figure illustrates the concept of PI in the context of BO using a GP model.

where $\phi(\mathbf{x};\mu,\sigma^2)$ and $\Phi(\mathbf{x};\mu,\sigma^2)$ represent the PDF and CDF (providing the probability that a normally distributed random variable is less than a given threshold) of a $N(\mu,\sigma^2)$ distribution evaluated at $\mathbf{x}$ respectively. When $\mu = 0$ and $\sigma^2 = 1$ we will simply write $\phi(\mathbf{x})$ and $\Phi(\mathbf{x})$ for brevity. If you look at Figure 6.20 above, it's clear that the probability of improvement is the shaded area under the Gaussian curve for $z > z_0$.

Figure 6.20 explains (6.45). The x-axis represents the input space of the function we want to optimize. It could be any parameter or set of parameters being tuned. The y-axis corresponds to the objective function value, denoted as $f(\mathbf{x})$. This is the function we are aiming to optimize, typically to maximize or minimize its value. The horizontal dashed line represents the current best-known value of the objective function, $f(\mathbf{x}^*)$, attained at point $\mathbf{x}^*$. This value serves as the benchmark for improvement. The point $\mathbf{x}^*$ on the x-axis directly below $f(\mathbf{x}^*)$ indicates the location in the input space where the current best objective function value has been found. An arbitrary point in the input space, $\mathbf{x}$, indicates a potential candidate for the next evaluation in the optimization process. The vertical axis on the right represents the PI at the new candidate point $\mathbf{x}$. The PI is a measure of the likelihood that the function value at $\mathbf{x}$ will be better (i.e., lower if minimizing, higher if maximizing) than $f(\mathbf{x}^*)$. The bell-shaped curve extending into the upper right quadrant illustrates the probability distribution for improvement at point $\mathbf{x}$. This distribution is derived from the predictive distribution of the GP model at $\mathbf{x}$, standardized to form a $\mathbf{z}$-distribution (a standard normal distribution). The shaded area under the curve to the left of the vertical dashed line passing through $\mathbf{x}$ represents the PI. Mathematically, this is the CDF evaluated at the standardized difference between $f(\mathbf{x}^*)$ and the GP's mean prediction at $\mathbf{x}$, scaled by the GP's prediction standard deviation.

The essence of PI is to use the model's predictive distribution at any candidate point $\mathbf{x}$ to compute the probability that evaluating the objective function at $\mathbf{x}$ will result in a better outcome than what has been observed so far. The decision to pick the next point to evaluate is based on maximizing the PI across the potential points.

**Remark:**

- PI can be particularly effective in scenarios where small improvements are significant. However, its performance can degrade if the global landscape of the function is multimodal or has several local maxima, as PI might get "stuck" near a local maximum.
- PI is also sometimes called MPI (for "Maximum Probability of Improvement") or "the P -algorithm" (since the utility is the probability of improvement).
- The drawback, intuitively, is that the formulation (6.45) is pure exploitation. Points that have a high probability of being infinitesimally greater than $f(x^*)$ will be drawn over points that offer larger gains but less certainty. As a result, a modification is to add a trade-off parameter $\xi \geq 0$ [185]:

$$\mathrm{PI}(\mathbf{x}) = \Phi\left(\frac{\mu(\mathbf{x}) - f(\mathbf{x}^*) - \xi}{\sigma(\mathbf{x})}\right) = \Phi\left(\frac{\mu(\mathbf{x}) - (f(\mathbf{x}^*) + \xi)}{\sigma(\mathbf{x})}\right). \tag{6.46}$$

- When $\xi$ is zero, the PI formula purely focuses on whether the predicted mean $\mu(\mathbf{x})$ at new points is better than the best-observed value $f(\mathbf{x}^*)$. This setting favors exploitation because it directly seeks improvements over the current best without considering the uncertainty of those predictions.
- With a positive $\xi$: The term $f(\mathbf{x}^*) + \xi$ becomes the new target for $\mu(\mathbf{x})$. This means that to be considered an improvement, $\mu(\mathbf{x})$ must not only exceed $f(\mathbf{x}^*)$, but also surpass it by at least $\xi$. This increment by $\xi$ effectively increases the difficulty of achieving an "improvement" status based purely on the mean prediction.
- The setting of $\xi$ enables the algorithm to control its tendency towards exploring uncertain areas versus exploiting known areas of high reward. By choosing $\xi$:
  - Higher $\xi$ values can be used when the algorithm needs to be discouraged from clustering around local maxima and encouraged to explore potentially better but uncertain regions.
  - Lower $\xi$ values make the algorithm focus more on areas around the current best observations, useful when fine-tuning around a known good solution or when the function landscape is expected to be smooth and well-behaved.
  - The exact choice of $\xi$ is left to the user, though Kushner [196] recommended a schedule for $\xi$, so that it started fairly high early in the optimization, to drive exploration, and decreased toward zero as the algorithm continued.

### 6.9.3. Expected Improvement

PI focuses solely on the probability of improving our current best estimate. It does not account for how much better the new evaluation might be. This characteristic can make PI somewhat limited in scenarios where the magnitude of improvement is critical, as it might lead to selecting points that offer very marginal gains but with a higher probability of being better than the best so far. On the other hand, EI [195-199] takes a more comprehensive approach by considering both the probability of improvement and the expected size of the improvement. This makes EI particularly useful in situations where you want to maximize the gains from each function evaluation, especially when function evaluations are costly or time-consuming. Mathematically, it is defined based on the posterior distribution obtained from a GP model used in BO. Instead of looking at the improvement $I(\mathbf{x})$, which is a random variable, we will instead calculate the "Expected Improvement", which is the expected value of $I(\mathbf{x})$:

$$\mathrm{EI}(\mathbf{x}) \equiv \mathbb{E}[I(\mathbf{x})] = \int_{-\infty}^{\infty} I(\mathbf{x})\phi(z)dz = \int_{-\infty}^{\infty} \underbrace{\max(f(\mathbf{x}) - f(\mathbf{x}^*), 0)}_{I(\mathbf{x})} \phi(z)dz, \tag{6.47}$$

where $\phi(z)$ is the PDF of the normal distribution $N(0,1)$, i.e., $\phi(z) = \frac{1}{\sqrt{2\pi}}\exp(-z^2/2)$ and $z(\mathbf{x}) = \frac{f(\mathbf{x})-\mu(\mathbf{x})}{\sigma(\mathbf{x})} \sim N(0,1)$.

**Theorem 6.4:** EI can be calculated analytically by

$$\mathrm{EI}(\mathbf{x}) \equiv= \sigma(\mathbf{x})\phi(Z) + \big(\mu(\mathbf{x}) - f(\mathbf{x}^*)\big)\Phi(Z), \tag{6.48.1}$$

where,

$$Z = \frac{\mu(\mathbf{x}) - f(\mathbf{x}^*)}{\sigma(\mathbf{x})}. \tag{6.48.2}$$

**Proof:**

Let's $z_0 = \frac{f(\mathbf{x}^*)-\mu(\mathbf{x})}{\sigma(\mathbf{x})}$,

$$\begin{aligned}
\mathrm{EI}(\mathbf{x}) &\equiv \int_{-\infty}^{\infty} \underbrace{\max(f(\mathbf{x}) - f(\mathbf{x}^*), 0)}_{I(\mathbf{x})} \phi(z)dz \\
&= \underbrace{\int_{-\infty}^{z_0} I(\mathbf{x})\varphi(z)dz}_{\text{Zero since } I(\mathbf{x})=0} + \int_{z_0}^{\infty} I(\mathbf{x})\phi(z)dz \\
&= \int_{z_0}^{\infty} \max(f(\mathbf{x}) - f(\mathbf{x}^*), 0)\, \phi(z)dz
\end{aligned}$$

$$= \int_{z_0}^{\infty} [f(\mathbf{x}) - f(\mathbf{x}^*)]\phi(z)dz$$
$$= \int_{z_0}^{\infty} [z(\mathbf{x})\sigma(\mathbf{x}) + \mu(\mathbf{x}) - f(\mathbf{x}^*)]\phi(z)dz$$
$$= \int_{z_0}^{\infty} [z(\mathbf{x})\sigma(\mathbf{x})\phi(\mathbf{z}) + \{\mu(\mathbf{x}) - f(\mathbf{x}^*)\}\phi(z)]dz\,.$$

Where we use $f(\mathbf{x}) = z(\mathbf{x})\sigma(\mathbf{x}) + \mu(\mathbf{x})$.

$$\begin{aligned}
\mathrm{EI}(\mathbf{x}) &= \int_{z_0}^{\infty} z(\mathbf{x})\sigma(\mathbf{x})\phi(z)dz + \int_{z_0}^{\infty} \{\mu(\mathbf{x}) - f(\mathbf{x}^*)\}\phi(z)dz \\
&= \int_{z_0}^{\infty} z(\mathbf{x})\sigma(\mathbf{x})\frac{1}{\sqrt{2\pi}}\mathrm{e}^{\left(-\frac{z^2}{2}\right)}dz + \{\mu(\mathbf{x}) - f(\mathbf{x}^*)\}\int_{z_0}^{\infty}\phi(z)dz \\
&= \frac{\sigma(\mathbf{x})}{\sqrt{2\pi}}\int_{z_0}^{\infty} z(\mathbf{x})\mathrm{e}^{\left(-\frac{z^2}{2}\right)}dz + \{\mu(\mathbf{x}) - f(\mathbf{x}^*)\}\big(1 - \Phi(z_0)\big).
\end{aligned}$$

Since $\frac{d}{dz}\mathrm{e}^{\left(-\frac{z^2}{2}\right)} = -z\mathrm{e}^{\left(-\frac{z^2}{2}\right)}$, we have $z(\mathbf{x})\mathrm{e}^{\left(-\frac{z^2}{2}\right)}dz = -d\mathrm{e}^{\left(-\frac{z^2}{2}\right)}$,

$$\begin{aligned}
\mathrm{EI}(\mathbf{x}) &= \frac{\sigma(\mathbf{x})}{\sqrt{2\pi}}\int_{z_0}^{\infty} -d\mathrm{e}^{\left(-\frac{z^2}{2}\right)} + \{\mu(\mathbf{x}) - f(\mathbf{x}^*)\}\big(1 - \Phi(z_0)\big) \\
&= -\frac{\sigma(\mathbf{x})}{\sqrt{2\pi}}\left[\mathrm{e}^{\left(-\frac{z^2}{2}\right)}\right]_{z_0}^{\infty} + \{\mu(\mathbf{x}) - f(\mathbf{x}^*)\}\big(1 - \Phi(z_0)\big) \\
&= -\sigma(\mathbf{x})\left(-\frac{1}{\sqrt{2\pi}}\mathrm{e}^{\left(-\frac{z_0^2}{2}\right)}\right) + \{\mu(\mathbf{x}) - f(\mathbf{x}^*)\}\big(1 - \Phi(z_0)\big) \\
&= -\sigma(\mathbf{x})\big(-\phi(z_0)\big) + \{\mu(\mathbf{x}) - f(\mathbf{x}^*)\}\big(1 - \Phi(z_0)\big) \\
&= \sigma(\mathbf{x})\phi(z_0) + \{\mu(\mathbf{x}) - f(\mathbf{x}^*)\}\Phi(-z_0) \\
&= \sigma(\mathbf{x})\phi(z_0) - z_0\sigma(\mathbf{x})\Phi(-z_0) \\
&= \sigma(\mathbf{x})\{\phi(z_0) - z_0\Phi(-z_0)\} \\
&= \sigma(\mathbf{x})\{\phi(-z_0) - z_0\Phi(-z_0)\} \\
&= \sigma(\mathbf{x})\left\{\phi\left(\frac{\mu(\mathbf{x}) - f(\mathbf{x}^*)}{\sigma(\mathbf{x})}\right) + \left(\frac{\mu(\mathbf{x}) - f(\mathbf{x}^*)}{\sigma(\mathbf{x})}\right)\Phi\left(\frac{\mu(\mathbf{x}) - f(\mathbf{x}^*)}{\sigma(\mathbf{x})}\right)\right\}.
\end{aligned}$$

Note that, we used the fact that the PDF of the normal distribution is symmetric, therefore $\phi(z_0) = \phi(-z_0)$.

Let $Z = \frac{\mu(\mathbf{x}) - f(\mathbf{x}^*)}{\sigma(\mathbf{x})}$, we have

$$\mathrm{EI}(\mathbf{x}) = \sigma(\mathbf{x})\phi(Z) + \big(\mu(\mathbf{x}) - f(\mathbf{x}^*)\big)\Phi(Z).$$

■

**Remarks:**

- EI is typically more effective than PI because it balances the probability of improvement with the expected size of the improvement. This balance helps the optimization algorithm avoid getting stuck in local optima, and it often leads to better overall optimization outcomes when dealing with functions that have multiple local optima.
- In practical use, one typically maximizes the EI across the search space to select the next sampling point. This process continues iteratively until a stopping condition is met, like a maximum number of iterations, time budget, or convergence to a satisfactory solution.

# CHAPTER 7

# REGULARIZATION TECHNIQUES

In machine learning, particularly in the field of deep learning, the complexity of models often leads to a significant challenge: overfitting. Overfitting occurs when a model learns the detail and noise in the training data to an extent that it negatively impacts the performance of the model on new data. The essence of regularization is to constrain our model in such a way that it avoids overfitting and thus performs better on unseen datasets. This chapter offers a comprehensive exploration of the strategies used to improve the generalization ability of NNs. By introducing regularization techniques into the training process, we aim to develop models that are not only accurate on the training data but also robust and adaptable to new, unseen data. This chapter will cover three principal regularization techniques: penalty-based regularization, early stopping, and ensemble methods (dropout), each addressing overfitting in a unique way.

**Penalty-Based Regularization:**

The first section of this chapter will delve into Penalty-Based Regularization, a method that involves modifying the loss function by adding a penalty term to it. This term penalizes certain values of model parameters to discourage complexity:

- $L_1$ regularization (Lasso) promotes sparsity by adding a penalty equivalent to the absolute value of the coefficients, which can reduce the number of features in the model.
- $L_2$ regularization (Ridge) adds a penalty equivalent to the square of the magnitude of the coefficients, encouraging the model parameters to be small and distributing the error among them.
- Elastic net combines both $L_1$ and $L_2$ penalties and is useful when there are correlations among features.

Through mathematical formulations, this section will illustrate how these techniques alter the model training dynamics and help in achieving lower complexity and better generalization.

**Early Stopping:**

The second section will explore Early Stopping, a different kind of regularization technique that does not modify the loss function but rather alters the training process itself. Early stopping involves monitoring the model's performance on a validation set and stopping the training as soon as the performance starts to degrade, despite improvements on the training set. This method not only helps in preventing overfitting but also saves computational resources by reducing unnecessary training iterations.

**Ensemble Methods (Dropout):**

The final section of this chapter will cover ensemble methods (with focus on dropout), a powerful regularization technique particularly popular in training DNNs. Unlike penalty-based methods, dropout works by randomly deactivating a subset of neurons in each training iteration. This randomness helps to break up situations where network layers co-adapt to correct mistakes from prior layers, thus making the model more capable of generalizing well. We will also look at variants like drop connect and other ensemble methods like bagging and subsampling.

By integrating these regularization techniques, machine learning practitioners can enhance model performance significantly. This chapter will equip you with the knowledge to understand these methods conceptually, ensuring that your NNs are not just powerful, but also robust and efficient.

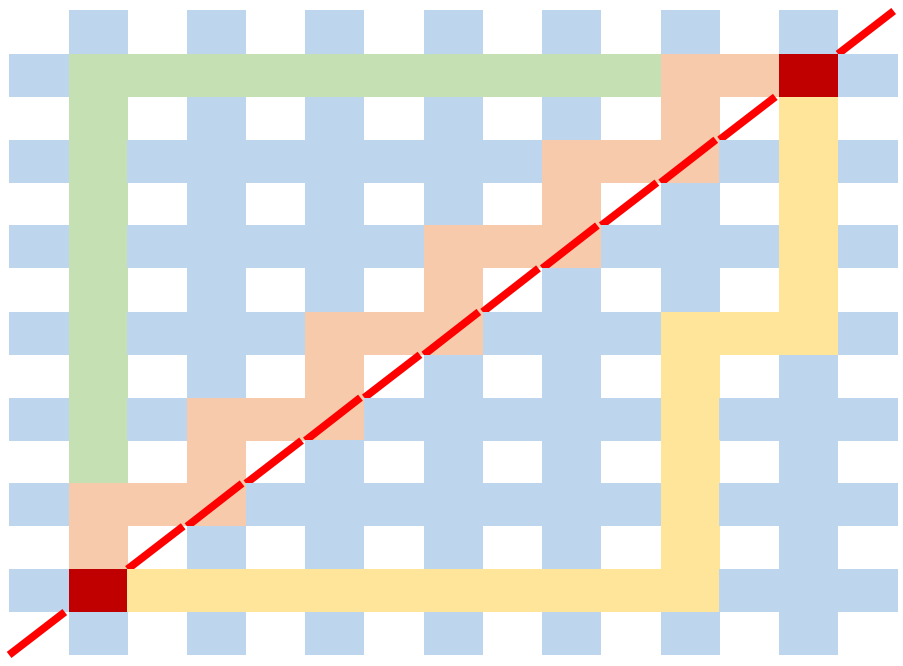

**Figure 7.1.** Taxicab or Manhattan distance versus Euclidean distance.

## 7.1 Penalty-Based Regularization

### 7.1.1 $L_p$-norm in Finite Dimensions

Before delving into the concepts of $L_1$ and $L_2$-regularization, it's important to introduce the notation of the $L_p$-norm. The length of a vector $|\mathbf{x}\rangle = (x_1, x_2, \ldots, x_n)^T$ in the $n$-dimensional real vector space $\mathbb{R}^n$ is usually given by the Euclidean norm [29]:

$$\|\mathbf{x}\| = (x_1^2 + x_2^2 + \cdots + x_n^2)^{\frac{1}{2}}. \tag{7.1}$$

It follows directly from the definition that $\|\mathbf{x}\| = \sqrt{(x_1, x_2, \ldots, x_n) \cdot (x_1, x_2, \ldots, x_n)^T} = \sqrt{\langle \mathbf{x}|\mathbf{x}\rangle}$. In 3-dimensional space, the formula for the Euclidean distance between two points $\mathbf{p}_1 = (x_1, y_1, z_1)^T$ and $\mathbf{p}_2 = (x_2, y_2, z_2)^T$ is

$$d(\mathbf{p}_1, \mathbf{p}_2) = ((x_1 - x_2)^2 + (y_1 - y_2)^2 + (z_1 - z_2)^2)^{\frac{1}{2}}. \tag{7.2}$$

The formula above, however, is not always the way we measure the distance between points laying in a 3-dimensional space. For example, we can think of the Earth as being a subset of three-dimensional space and the surface of the Earth as being the surface of a sphere. When measuring the distance between points on the surface of the Earth, say from New York City to Cairo. We do not draw a line through the two points and measure the distance along those points. This distance would correspond to the distance traveled if a tunnel was dug in a straight line through the Earth, starting at New York and ending at Cairo. This is not very practical; one usually does not travel between cities by tunneling from one to the other. Instead, we measure the distance between two points $\mathbf{p}_1$ and $\mathbf{p}_2$ on the Earth by measuring the short path starting at $\mathbf{p}_1$ and ending at $\mathbf{p}_2$ that stays on the surface of the Earth. In many situations, the Euclidean distance is insufficient for capturing the actual distances in a given space. An analogy to this is suggested by taxi drivers in a grid street plan who should measure distance not in terms of the length of the straight line to their destination but in terms of the rectilinear distance, which takes into account that streets are either orthogonal or parallel to each other (see Figure 7.1).

The class of $p$-norms generalizes these two examples and has an abundance of applications in many parts of mathematics, physics, and computer science. There are many useful measures of length (many different norms).

**Definition ($p$-Norm):** For a real number $p \geq 1$, the $p$-norm or $L_p$-norm of $|\mathbf{x}\rangle = (x_1, x_2, \ldots, x_n)^T$ is defined by

$$\|\mathbf{x}\|_p = (|x_1|^p + |x_2|^p + \cdots + |x_n|^p)^{\frac{1}{p}}. \tag{7.3}$$

Some norms are

- $L_1$-norm, 1-norm, or the Manhattan norm: $\|\mathbf{x}\|_1 = |x_1| + |x_2| + \cdots + |x_n|$,
- $L_2$-norm, 2-norm, or the Euclidean norm: $\|\mathbf{x}\|_2 = \sqrt{|x_1|^2 + |x_2|^2 + \cdots + |x_n|^2}$,
- $L_\infty$-norm, $\infty$-norm or the Chebyshev norm: $\|\mathbf{x}\|_\infty = \max\{|x_1|, |x_2|, \ldots |x_n|\}$.

Figure 7.2 gives a visual representation of these norms.

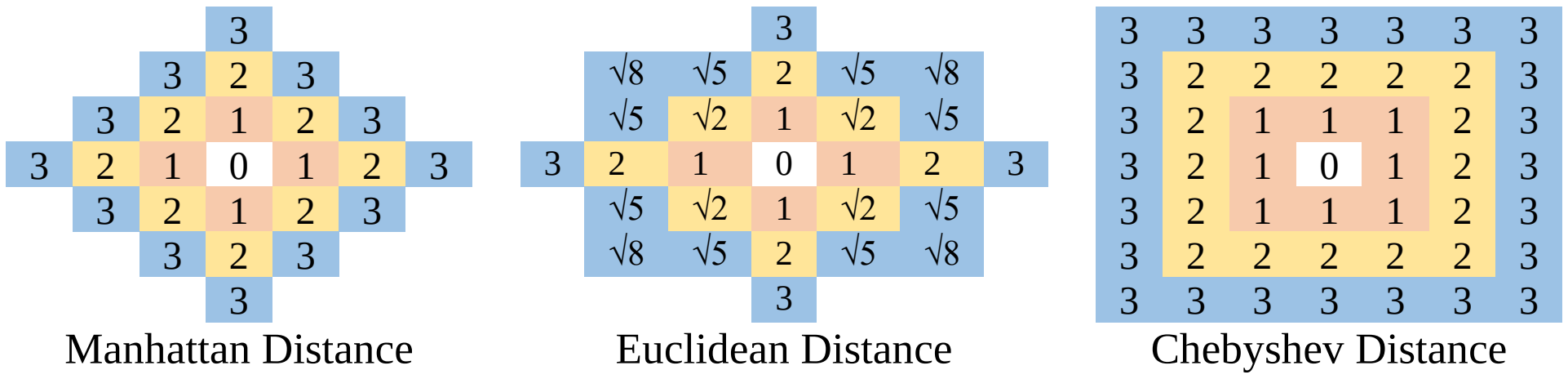

**Figure 7.2.** Manhattan, Euclidean, and Chebyshev Distance.

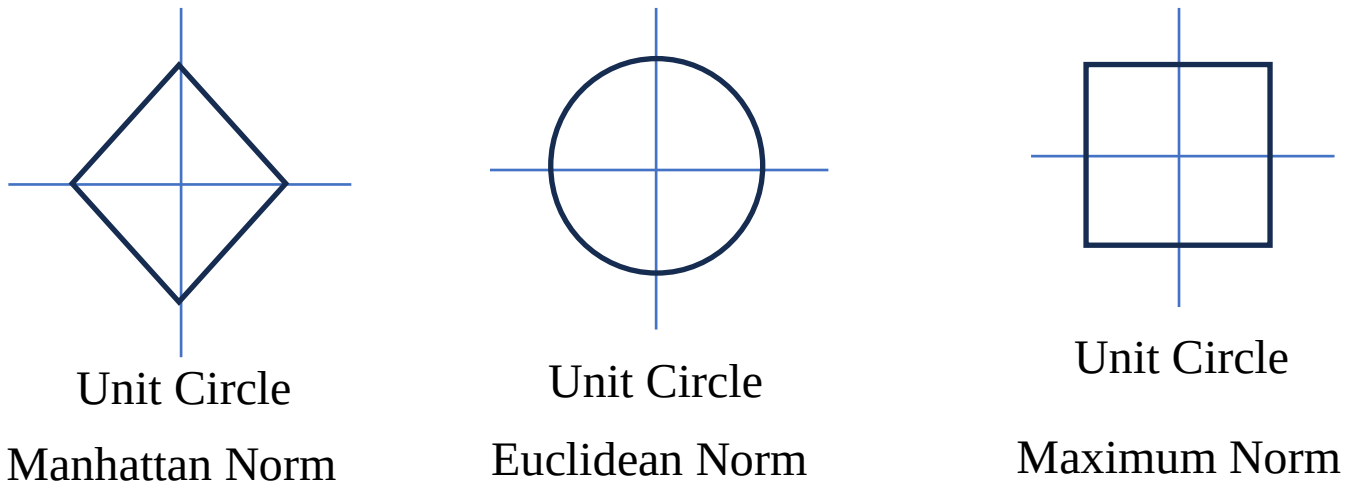

**Figure 7.3.** Unit circles in $\mathbb{R}^2$ based on different $p$-norms. Each vector from the origin to the unit circle has a length of one, calculated using the length formula of the corresponding $p$. The illustrations show how the shape of the unit circle varies with different $p$-values, including the Manhattan norm ($p = 1$), the Euclidean norm ($p = 2$), and the Maximum norm ($p = \infty$).

**Theorem 7.1:** For any vector $|\mathbf{x}\rangle = (x_1, x_2, \ldots, x_n)^T \in \mathbb{R}^n$, we have [29]

$$\|\mathbf{x}\|_\infty \leq \|\mathbf{x}\|_2 \leq \|\mathbf{x}\|_1. \tag{7.4}$$

**Proof:**

$$\begin{aligned}
\|\mathbf{x}\|_\infty &= \max\{|x_1|, |x_2|, \ldots |x_n|\} \\
&= \max\left\{\sqrt{x_1^2}, \sqrt{x_2^2}, \ldots, \sqrt{x_n^2}\right\} \\
&= \sqrt{x_k^2} \text{ for some } k \\
&\leq \sqrt{x_1^2 + x_2^2 + \cdots + x_n^2} \\
&= \|\mathbf{x}\|_2 \\
&= \sqrt{|x_1|^2 + |x_2|^2 + \cdots + |x_n|^2} \\
&\leq \sqrt{(|x_1| + |x_2| + \cdots + |x_n|)^2} \\
&= \|\mathbf{x}\|_1.
\end{aligned}$$

■

A unit circle in $\mathbb{R}^2$ is typically defined as the set of points that are a distance of 1 from the origin. The distance metric used to define this "distance of 1" can vary, leading to different unit circles based on different $p$-norms. Figure 7.3 gives a visual representation of these unit circles.

- $p = 1$ (Manhattan norm or $L_1$-norm): The Manhattan norm considers the sum of the absolute differences between the coordinates. For a point $(x, y)$ on the unit circle, we have $|x| + |y| = 1$. This equation represents a diamond-shaped figure centered at the origin, with its sides aligned with the diagonals of the coordinate axes (i.e., at 45-degree angles to the axes). It is symmetrical about both the $x$-axis and $y$-axis. Each side of this diamond intersects with one of the axes at the points $(0,1)$, $(0,-1)$, $(1,0)$, $(-1,0)$.

- $p = 2$ (Euclidean norm or $L_2$-norm): The Euclidean norm is the standard notion of distance, giving us the familiar circle equation $\sqrt{x^2 + y^2} = 1$. This equation represents a circle with radius 1 centered at the origin. This circle is smooth and has no corners.
- $p = \infty$ (Maximum norm or $L_\infty$-norm): The Maximum norm takes the maximum of the absolute values of the coordinates, giving us $\max(|x|, |y|) = 1$. This equation represents a square with sides aligned with the axes.

### 7.1.2 Penalty-Based Regularization

Penalty-based regularization [32, 49, 51, 58, 68-70] is a technique used in machine learning and statistical modeling to prevent overfitting and improve the generalization performance of models. In penalty-based regularization, a penalty term is added to the model's objective function (usually the loss function) to discourage overly complex models. This penalty term penalizes large parameter values or model complexity, encouraging the model to favor simpler solutions that are less likely to overfit.

Let's discuss how penalty-based regularization techniques can help address the trade-off between overfitting and oversimplification in the context of polynomial regression with NNs. To better understand this point, let's revisit the example involving a polynomial of degree $d$. In this case, the prediction $\hat{y}$ for a given value of $x$ is determined by the equation:

$$\hat{y} = \sum_{i=0}^{d} w_i \cdot x^i. \tag{7.5}$$

To model this prediction, we can utilize a single-layer network with $d$ inputs and a single bias neuron with weight $w_0$. Each input corresponds to a power of $x$, with the $i$-th input representing $x^i$. This NN employs linear activations. The squared loss function for a set of training instances $(x, y)$ from dataset $D$ is defined as:

$$\mathcal{L} = \frac{1}{2} \sum_{(x,y) \in D} (y - \hat{y})^2. \tag{7.6}$$

The high value of $d$ can lead to overfitting, where the model captures noise or random fluctuations in the training data, resulting in poor generalization to unseen data. One potential solution to mitigate this issue is to reduce the value of $d$. Essentially, employing a model with a parsimonious parameterization results in a simpler model. For instance, reducing $d$ to 1 creates a linear model with fewer degrees of freedom, which tends to fit the data in a similar way over different training samples. While reducing the degree of the polynomial $d$ can mitigate overfitting, it also reduces the model's expressiveness, particularly for complex data patterns. Oversimplification limits the NN's ability to adapt to diverse requirements across different datasets, potentially leading to underfitting and poor performance.

To retain some expressiveness without causing excessive overfitting, instead of reducing the number of parameters in a hard way, one can use a soft penalty on the use of parameters. This soft penalty approach is typically achieved through techniques such as $L_1$ and $L_2$-regularization, but with the regularization parameter carefully chosen to balance between model complexity and fitting the data. In this context, you can use a regularization term that penalizes the magnitude of the parameters in the model. This approach prevents any single parameter from having an excessively large magnitude, thus reducing the risk of overfitting. Additionally, by penalizing large absolute values of parameters more than small values, the regularization term can effectively control the model's complexity while retaining important features of the data. This is because small parameter values have less impact on the prediction, so penalizing them less allows the model to adapt to the data without being overly constrained.

There are two commonly used penalty-based regularization techniques: $L_2$ and $L_1$-regularization.

### 7.1.3 $L_2$-Regularization (Ridge)

In $L_2$-regularization, which is also referred to as Tikhonov regularization, Ridge regularization, or weight decay, the penalty term is the sum of the squared values of the model's coefficients multiplied by a regularization parameter ($\lambda$). Then, for the regularization parameter $\lambda > 0$, one can define the regularized objective function $J$ (including loss $\mathcal{L}$) as follows:

$$
\begin{aligned}
J &= \mathcal{L} + \frac{\lambda}{2}\|\mathbf{w}\|_2^2 \\
&= \mathcal{L} + \frac{\lambda}{2}\langle \mathbf{w}|\mathbf{w}\rangle \\
&= \mathcal{L} + \frac{\lambda}{2}\sum_{i=1}^{d} w_i^2 \\
&= \frac{1}{2}\sum_{(x,y)\in D}(y-\hat{y})^2 + \frac{\lambda}{2}\sum_{i=1}^{d} w_i^2,
\end{aligned}
\tag{7.7}
$$

where $\|\boldsymbol{w}\|_2^2$ is the squared $L_2$-norm of the parameter vector. The ridge weights minimize the regularized objective function, $J$,

$$
\begin{aligned}
\hat{\mathbf{w}}_{\text{Ridge}} &= \underset{\mathbf{w}}{\operatorname{argmin}}\, J \\
&= \underset{\mathbf{w}}{\operatorname{argmin}} \left\{ \frac{1}{2}\sum_{(x,y)\in D}(y-\hat{y})^2 + \frac{\lambda}{2}\sum_{i=1}^{d} w_i^2 \right\}.
\end{aligned}
\tag{7.8}
$$

The regularization term, $\frac{1}{2}\sum_{i=1}^{d} w_i^2$, is added to the error function. It penalizes large values of the weights $w_i$. The regularization parameter $\lambda$ controls the strength of this penalty. A higher $\lambda$ penalizes large weights more, encouraging the model to be simpler. This constraint limits the sum of squared weights, effectively limiting the complexity of the model. It prevents the model from fitting the training data too closely and helps it generalize better to new, unseen data.

Let us explain this point in more detail. During the training process of a machine learning model, the primary objective is to minimize the error between the predictions made by the model and the actual values in the training data. This is typically achieved by adjusting the model's parameters (such as weight in the NN) iteratively through optimization algorithms like GD. However, without any form of regularization, the model might become overly complex and prone to overfitting. In the context of linear regression, one way that overfitting can occur is if the model assigns overly large values to the coefficients of the features. Large coefficients can lead to large variations in the output for small variations in the input, which may capture noise rather than true patterns. When we introduce regularization, such as $L_2$-regularization, the optimization process gains an additional objective: to keep the coefficients small. In ordinary NNs, the optimization process solely focuses on minimizing the error. As a result, the model might end up with coefficients that are too large, which can lead to overfitting. In contrast, with $L_2$-regularization, the optimization process not only tries to minimize the error but also includes a penalty for having large coefficient values. This penalty term is proportional to the square of the coefficient values. So, during training, the optimization process strives to find the parameter values that not only make accurate predictions but also keep the coefficients small. By adding this extra objective of keeping the coefficients small, $L_2$ -regularization helps prevent overfitting by discouraging the model from fitting too closely to the training data. Instead, the model learns more generalizable patterns that are less likely to be influenced by noise and idiosyncrasies in the training set.

In other words, the regularization parameter, $\lambda$, in $L_2$-regularization controls the balance between two competing objectives:

- Fitting the training data well: This is achieved by minimizing the error between the predicted values and the actual values in the training dataset. The model aims to capture the underlying patterns and relationships present in the training data.
- Keeping the model simple: This is accomplished by penalizing large coefficient values. The penalty term discourages the model from becoming too complex or fitting too closely to the noise in the training data. Smaller coefficient values lead to a simpler model that is less likely to overfit.

Now, the regularization parameter, $\lambda$, comes into play. It determines the strength of regularization applied to the model.

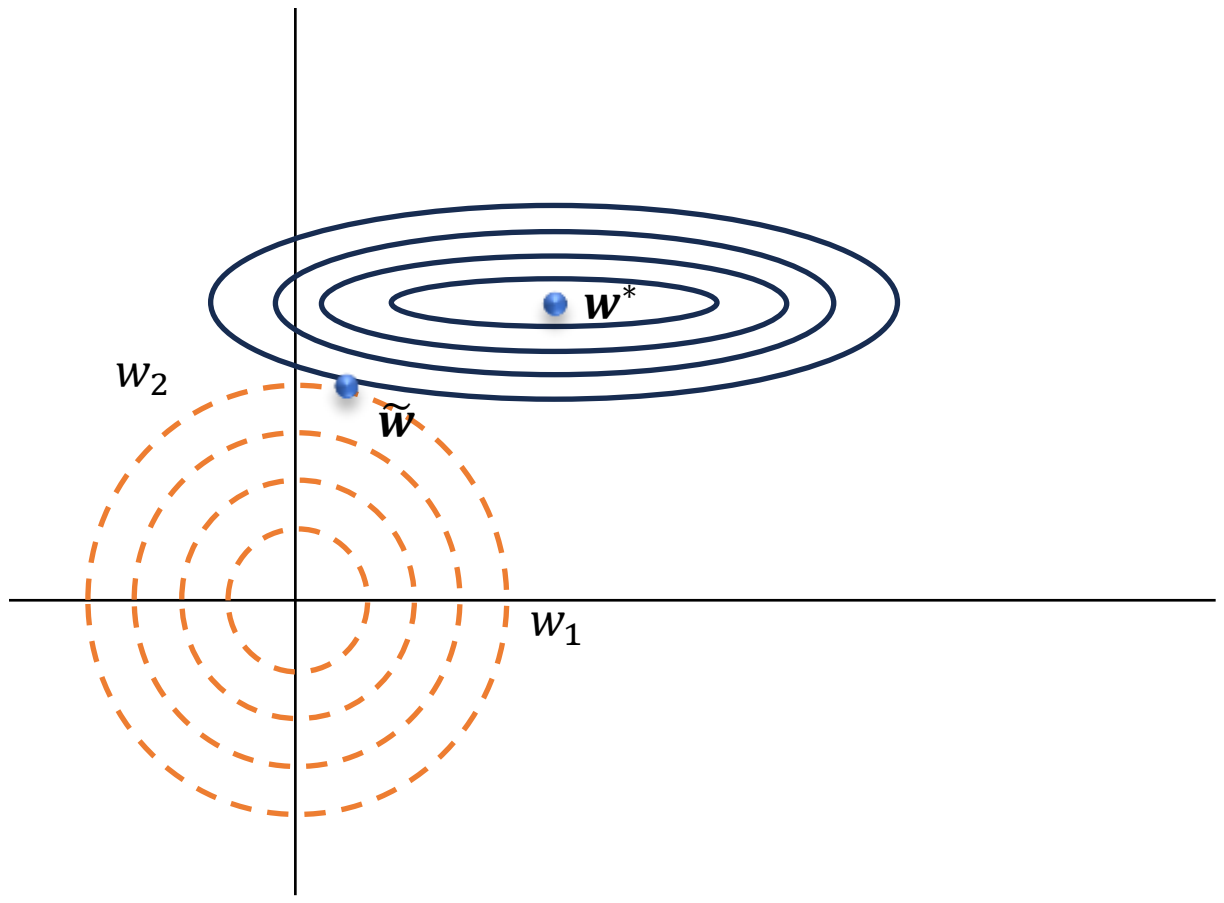

**Figure 7.4.** Contours of the error function (solid ellipses) and the regularization term (dotted circles). For $\lambda = 0$, the minimum error is indicated by $\boldsymbol{w}^*$. When $\lambda > 0$, the minimum of the regularized error function is shifted towards the origin. We see that the effect of the regularization term is to shrink the magnitudes of the weight parameters. However, the effect is much larger for the parameter $w_1$ because the unregularized error is much less sensitive to the value of $w_1$ compared to that of $w_2$. Intuitively only the parameter $w_2$ is 'active' because the output is relatively insensitive to $w_1$, and hence the regularizer pushes $w_1$ close to zero.

- When $\lambda$ is large, the penalty for large coefficients dominates the cost function. This encourages the model to prioritize simplicity over fitting the training data well. Consequently, the model tends to have smaller coefficient values, leading to a simpler model. This increased regularization can help prevent overfitting, especially when the dataset is small or noisy. However, if $\lambda$ is set too high, it can lead to underfitting, where the model is too simple to capture the underlying patterns in the data, resulting in poor performance on both the training and test datasets.
- Conversely, when $\lambda$ is small, the penalty for large coefficients has less impact on the cost function. As a result, the model focuses more on fitting the training data well, potentially allowing larger coefficient values. This may lead to a more complex model that can better capture intricate patterns in the training data. However, with lower regularization, there's an increased risk of overfitting, especially if the training dataset is small or noisy.

The general effect of a quadratic regularizer can be understood by examining a two-dimensional parameter space, coupled with an unregularized error function that takes the form of a quadratic function of $w$. This scenario corresponds to a simple linear regression model with a sum-of-squares error function. Figure 7.4 provides an illustration of this concept. The solid ellipses represent contours of equal value of the unregularized objective function. These contours show how the objective function changes as we move away from the optimal weight vector $\boldsymbol{w}^*$. The dotted circles represent contours of equal value of the $L_2$ regularizer, which penalizes large weights. At the point $\widetilde{\boldsymbol{w}}$, the competing objectives of minimizing the unregularized loss and minimizing the $L_2$-regularization term reaches an equilibrium. This means that the total objective function, which includes both the unregularized loss and the regularization term, is minimized at $\widetilde{\boldsymbol{w}}$. In the first dimension $w_1$, the eigenvalue of the Hessian (curvature) of the unregularized loss function is small. This indicates that the objective function does not increase much when moving horizontally away from $\boldsymbol{w}^*$ along this dimension. Because the objective function does not strongly prefer any particular value along this direction, the $L_2$ regularizer has a strong effect. It pulls the weight $w_1$ close to zero, as minimizing the $L_2$ term favors smaller weights. In the second dimension $w_2$, the objective function is very sensitive to movements away from $\boldsymbol{w}^*$, indicated by a large eigenvalue of the Hessian (high curvature). This high curvature means that the objective function changes significantly with small changes in $w_2$. As a result, the $L_2$-regularization term has relatively little effect on the position of $w_2$, as the penalty for large $w_2$ is outweighed by the importance of minimizing the unregularized loss along this direction.

On the other hand, it is important to note that minimizing the regularized objective function, $J$, is equivalent to minimizing the unregularized sum-of-squares error, $\mathcal{L}$, subject to the constraint $\sum_{i=1}^{d} w_i^2 \leq t$. Mathematically,

$$\widehat{\mathbf{w}}_{\text{Ridge}} = \operatorname*{argmin}_{\mathbf{w}} \mathcal{L}, \tag{7.9.1}$$

$$\text{subjet to } \Omega(\mathbf{w}) = \sum_{i=1}^{d} w_i^2 \leq t, \tag{7.9.2}$$

for an appropriate value of the parameter $t$, which makes explicit the size constraint on the parameters. The two approaches (7.8 and 7.9) can be related using Lagrange multipliers. We can thus think of a parameter norm penalty as imposing a constraint on the weights. In $L_2$ norm, then the weights are constrained to lie in an $L_2$ ball. While we do not know the exact size of the constraint region, we can control it roughly by increasing or decreasing $\lambda$ to grow or shrink the constraint region.

With an $L_2$ norm penalty, the weights are constrained to lie within an $L_2$ ball. This means that the sum of the squares of the weights must be less than or equal to a certain value (the radius of the $L_2$ ball). The regularization parameter $\lambda$ controls the strength of the penalty. By increasing or decreasing $\lambda$, we can roughly control the size of the constraint region, i.e., the $L_2$ ball in which the weights must lie. When we increase $\lambda$, the penalty for large weights becomes stronger. This results in a smaller constraint region (smaller $L_2$ ball), as the weights are more restricted in their magnitude. This can help prevent overfitting by discouraging the model from using overly complex solutions. Conversely, when we decrease $\lambda$, the penalty for large weights becomes weaker. This leads to a larger constraint region (larger $L_2$ ball), allowing the weights to take on larger values. This can be useful when we want the model to be more flexible and potentially fit the training data more closely.

Note that, sometimes, it might be preferable to use explicit constraints instead of penalties. In such cases, we can adapt algorithms like GD to first take a step downhill on the loss function, $\mathcal{L}$, and subsequently project the weight vector, $\mathbf{w}$, back to the closest point that satisfies the constraint $\sum_{i=1}^{d} w_i^2 \leq t$. This approach can be beneficial if we have a predetermined value of $t$ and want to avoid the time-consuming process of finding the corresponding $\lambda$ value. Another advantage of using explicit constraints and reprojection over penalties is that penalties can cause nonconvex optimization procedures to become trapped in local minima associated with small weight vectors, $\mathbf{w}$. In the context of training NNs, this often results in networks with several 'dead units'—units that contribute minimally to the network's learned function due to very small incoming or outgoing weights. When a penalty is applied to the norm of the weights, these configurations can be locally optimal. Even if increasing the weights could significantly reduce the loss function $\mathcal{L}$, the penalty prevents this from happening, keeping the weights small and the network suboptimal. In contrast, explicit constraints implemented through reprojection are more effective in such scenarios, as they do not encourage the weights to approach the origin.

Hence, $L_2$-regularization can be implemented in two main ways: through explicit constraints and penalties in the loss function.

- In explicit constraints, the $L_2$ -regularization is directly incorporated into the optimization problem as a constraint on the weights. The constraint takes the form of $\sum_{i=1}^{d} w_i^2 \leq t$, where $\sum_{i=1}^{d} w_i^2$ is the sum of squared weights and $t$ is a threshold value. The goal is to minimize the loss function (e.g., unregularized sum-of-squares error) subject to this constraint. This can be solved using techniques such as Lagrange multipliers to incorporate the constraint into the optimization problem.
- In penalties, the $L_2$ -regularization is added to the loss function as a penalty term. The penalty term is $\frac{1}{2}\sum_{i=1}^{d} w_i^2$, where $\lambda$ is the regularization parameter that controls the strength of the penalty. The combined loss function is the original loss function (e.g., sum-of-squares error) plus the penalty term. The goal is to minimize this combined loss function with respect to the weights. This is typically done using GD or other optimization algorithms.

Indeed, tuning the regularization parameter $\lambda$ provides greater flexibility in controlling the complexity of the model compared to fixing the model's economy upfront. For example, in polynomial regression, restricting the number of parameters upfront severely limits the flexibility of the learned polynomial, potentially leading to an oversimplified model that cannot capture complex data patterns effectively. For instance, limiting the polynomial degree to 1 (creating a linear model) may not be suitable for capturing non-linear relationships present in the data. On the other hand, using a soft penalty through regularization allows for more flexibility in controlling the shape of the learned polynomial. By adjusting the regularization parameter $\lambda$, the degree of regularization can be fine-tuned to balance between model complexity and fitting the data. This enables the model to adapt its complexity to the characteristics of the dataset in a more data-driven manner. Moreover, it has been empirically observed that using more complex models (e.g., larger NNs) with regularization tends to yield better performance compared to using simpler models without regularization. Complex models have more capacity to capture intricate patterns in the data, and regularization helps prevent overfitting by penalizing overly complex models. The regularization parameter serves as a tunable knob that allows practitioners to choose the appropriate level of regularization based on the data, rather than making a rigid decision on the model's size upfront.

For any given weight $w_i$ of the NN, the updates are defined by GD (or the batched version):

$$\begin{aligned} w_i &= w_i - \alpha \frac{\partial J}{\partial w_i} \\ &= w_i - \alpha \frac{\partial}{\partial w_i}\left(\mathcal{L} + \frac{\lambda}{2}\sum_{i=1}^{d} w_i^2\right) \\ &= w_i - \alpha\left(\frac{\partial \mathcal{L}}{\partial w_i} + \lambda w_i\right) \\ &= w_i(1-\alpha\lambda) - \alpha\frac{\partial \mathcal{L}}{\partial w_i}. \end{aligned} \tag{7.10}$$

Here, $\alpha$ is the learning rate. The inclusion of the weight decay term in the learning process alters the trajectory of learning. It modifies the learning rule, causing the weight vector to undergo a multiplicative reduction by a constant factor (the decay factor $(1-\alpha\lambda)$) before each standard gradient update.

For a helpful analogy, imagine the learning trajectory as driving down a hill towards a valley (global minimum of the loss function). Training undergoes significant changes with the inclusion of weight decay compared to when it is absent;

Without weight decay

- Picture a car at the top of a hill aiming straight for the deepest part of the valley below.
- The valley represents the global minimum of the loss function, while the car’s path is the learning trajectory.
- The car accelerates downwards, following the steepest gradient to reach the minimum quickly.
- Without any additional inward force, the car can veer off course due to rugged terrain (local minima) or noisy gradients.
- This may cause overfitting or convergence to suboptimal solutions.

Adding weight decay

- Adding weight decay introduces a slight inward force pulling the car towards the center of the road.
- This prevents the car from veering too far to the sides and keeps it on a smoother, controlled path.
- The car still heads towards the valley but follows a safer trajectory that:
    - Regularizes the path by keeping weights (car's position) closer to zero (middle of the road).
    - Avoids sharp turns or overly steep gradients, leading to smoother decision boundaries.
- The inward force (weight decay) results in smaller, more manageable weights that create smoother decision boundaries.
- The controlled descent ensures that the car (model) doesn't overfit by steering too aggressively, instead generalizing better to unseen terrain (data).

**Remark:**

Regularization allows complex models to be trained on data sets of limited size without severe overfitting, essentially by limiting the effective model complexity. However, the problem of determining the optimal model complexity is then shifted from one of finding the appropriate number of basis functions (degree of polynomial or number of neurons) to one of determining a suitable value of the regularization coefficient $\lambda$. Choosing the appropriate value of $\lambda$ is crucial for achieving the right balance between fitting the training data well and keeping the model simple. This is typically done through techniques such as cross-validation, where different values of $\lambda$ are tried, and the one that yields the best performance on a validation dataset is selected.

### 7.1.4 $L_1$-Regularization

In $L_1$-regularization, also known as Lasso regularization, the penalty term is the sum of the absolute values of the model's coefficients (parameters) multiplied by a regularization parameter $\lambda$. Therefore, the new objective function is as follows:

$$\begin{aligned} J &= \mathcal{L} + \lambda \cdot \sum_{i=1}^{d} |w_i| \\ &= \frac{1}{2} \sum_{(x,y)\in D} (y - \hat{y})^2 + \lambda \cdot \sum_{i=1}^{d} |w_i|. \end{aligned} \tag{7.11}$$

When using $L_1$ regularization, the absolute value term $|w_i|$ poses a challenge for gradient-based optimization methods, particularly when $w_i$ is exactly zero. This is because the absolute value function is not differentiable at $w_i = 0$. In instances where $w_i$ is non-zero, one can employ a direct update method derived from computing the partial derivative. By taking the derivative of the aforementioned objective function, we can formulate the update equation, specifically applicable when $w_i$ is not equal to 0:

$$\begin{aligned} w_i &= w_i - \alpha \frac{\partial J}{\partial w_i} \\ &= w_i - \alpha \frac{\partial}{\partial w_i} \left( \mathcal{L} + \lambda \cdot \sum_{i=1}^{d} |w_i| \right) \\ &= w_i - \alpha \lambda s_i - \alpha \frac{\partial \mathcal{L}}{\partial w_i}. \end{aligned} \tag{7.12}$$

The value of $s_i$, which is the partial derivative of $|w_i|$ (with respect to $w_i$), is as follows:

$$s_i = \begin{cases} -1, & w_i < 0, \\ +1, & w_i > 0. \end{cases} \tag{7.13}$$

Additionally, it's necessary to address the partial derivative of $|w_i|$ in cases where $w_i$ equals exactly 0. One approach is to employ the subgradient method, where the value of $w_i$ is stochastically (randomly choose a value) from the interval $[-1, +1]$.

In practice, due to the finite precision of computers, it's rare for a model coefficient $w_i$ to be exactly 0. Computational errors, especially in iterative optimization algorithms, generally prevent coefficients from reaching such precise values. As a result, the challenge posed by the non-differentiability of the absolute value function at $w_i = 0$ is typically mitigated by these computational errors. Furthermore, for the rare cases in which the value $w_i$ is exactly 0, one can omit the regularization and simply set $s_i$ to 0.

The $L_1$-regularization encourages sparsity [200-210] in the model, meaning it tends to drive some coefficients to zero, effectively performing feature selection by selecting only the most relevant features. This sparsity property makes $L_1$-regularization is particularly useful for feature selection. In the context of NNs, if the weight $w_i$ associated with a connection to the input layer is zero, it implies that the corresponding input feature has no impact on the model's prediction. Hence, those inputs can be dropped, simplifying the model and potentially improving its interpretability. In other words, $L_1$-regularization can lead to sparse connectivity patterns, where many connections have zero weights.

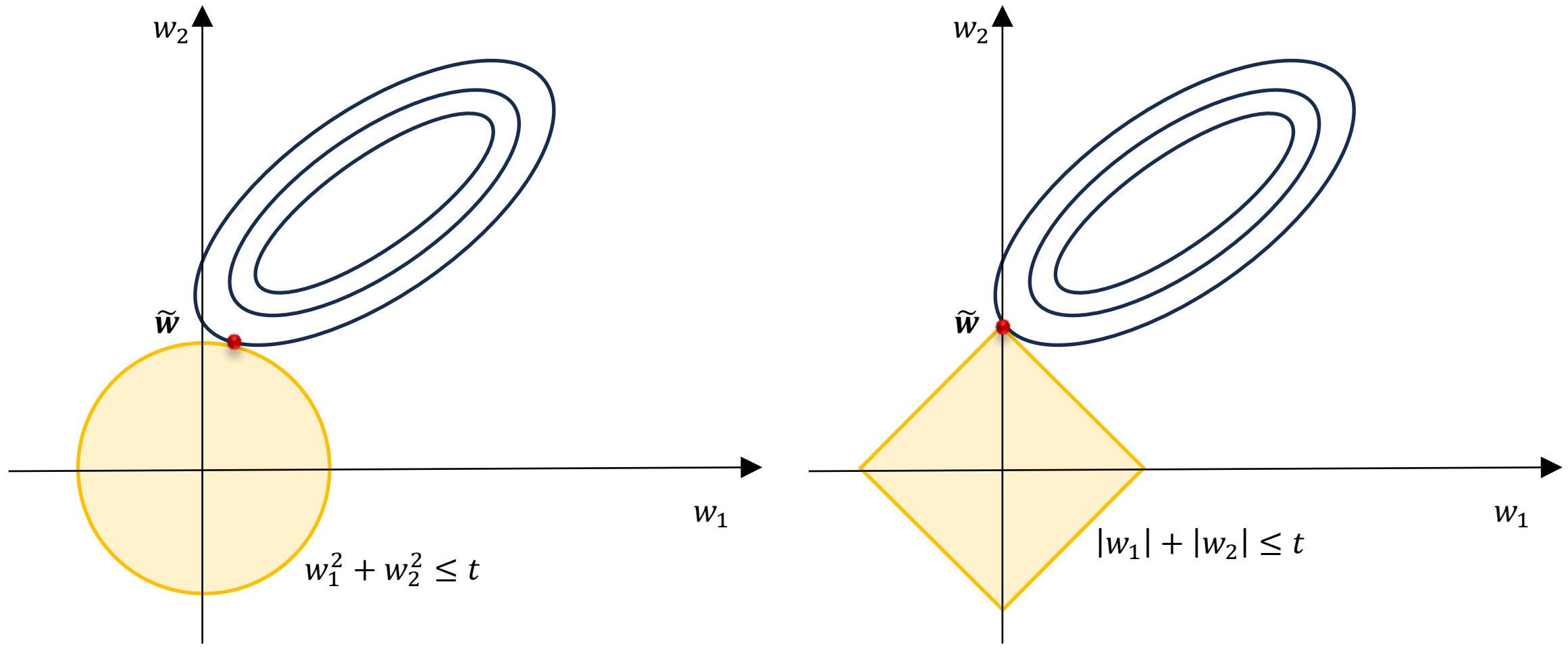

**Figure 7.5.** The solid areas represent the constraint regions $|w_1| + |w_2| \le t$ and $w_1^2 + w_2^2 \le t$, while the ellipses depict the contours of the least squares error function. On the left, the plot shows the contours of the unregularized error function alongside the constraint region for the ridge regularizer, and on the right, for the lasso regularizer, where the optimum value for the parameter vector $\boldsymbol{w}$ is denoted by $\widetilde{\boldsymbol{w}}$. The lasso produces a sparse solution with $\widetilde{w}_1 = 0$, while the ridge regularizer simply reduces $\widetilde{w}_1$ to a smaller value.

Connections with zero weights essentially act as dropout mechanisms, effectively removing those connections from the network architecture. This sparsity reduces the computational complexity of the network and can lead to more efficient training and inference.

Note that, again in $L_1$-regularization, minimizing the regularized objective function, $J$, is equivalent to minimizing the unregularized sum-of-squares error, $\mathcal{L}$, subject to the constraint $\sum_{i=1}^{d} |w_i| \le t$. Mathematically,

$$\widehat{\mathbf{w}}_{\text{Ridge}} = \underset{\mathbf{w}}{\operatorname{argmin}}\, \mathcal{L}, \tag{7.14.1}$$

$$\text{subjet to } \Omega(\mathbf{w}) = \sum_{i=1}^{d} |w_i| \le t, \tag{7.14.2}$$

for an appropriate value of the parameter $t$, which makes explicit the size constraint on the parameters.

**Remarks:**

- The regularization terms aim to penalize large coefficients by encouraging coefficients to approach zero. However, the mechanisms through which this is achieved differ: $L_2$-regularization introduces a continuous multiplicative decay, while $L_1$-regularization employs an additive update based on the sign of the coefficient. This difference leads to distinct behaviors in terms of sparsity and coefficient shrinkage, making each regularization technique suitable for different scenarios and preferences. In $L_2$-regularization, the term $(1 - \alpha\lambda)$ acts as the regularization term, and it tends to shrink the coefficients towards zero in a continuous, multiplicative manner. However, in $L_1$-regularization, the term $-\alpha\lambda s_i$ acts as the regularization term. It encourages coefficients to move towards zero by a fixed amount determined by the regularization strength $\lambda$ and the sign of the coefficient $s_i$.
- The sparsity-inducing effect of the $L_1$ -norm can also be interpreted by studying the geometry of the $L_1$-ball. The geometry of the ball B (constraint region) is directly related to the properties of the solutions $\widetilde{\boldsymbol{w}}$. The $L_2$-regularization adds a penalty term proportional to the square of the weights: $\frac{\lambda}{2}\sum_{i=1}^{d} w_i^2$. The contour lines of the unregularized error function are ellipses centered at the optimal weight vector $\boldsymbol{w}^*$, Figure 7.5. The constraint region for $L_2$-regularization is a circle around the origin, i.e., round ball that does not favor any specific direction of the space. The optimal weight vector $\widetilde{\boldsymbol{w}}$ lies at the intersection of the contour lines and the circle. On the other hand, the $L_1$ -regularization adds a penalty term proportional to the absolute values of the weights: $\sum_{i=1}^{d}|w_i|$. The contour lines of the unregularized error function are similar to the $L_2$ case. However, it is clear to see that the $L_1$ constraint, which corresponds to the diamond feasible region (with vertices on the axes at $(\pm t, 0)$ and $(0, \pm t)$.) or a polyhedron (in higher dimensions), is more likely to produce an intersection (the optimal weight vector $\tilde{\boldsymbol{w}}$) that has one component of the solution is zero (i.e., the sparse model) due to the geometric properties of ellipses, disks, and diamonds. It is simply because diamonds have corners (of which one component is zero) that are easier to intersect with the ellipses. The tangent point for $L_1$ cases is more likely to be achieved on the axis (some parameters be reduced to zero), while the tangent point for $L_2$ cases is more likely to be got on a none-axis point (none zero points), Figure 7.5.

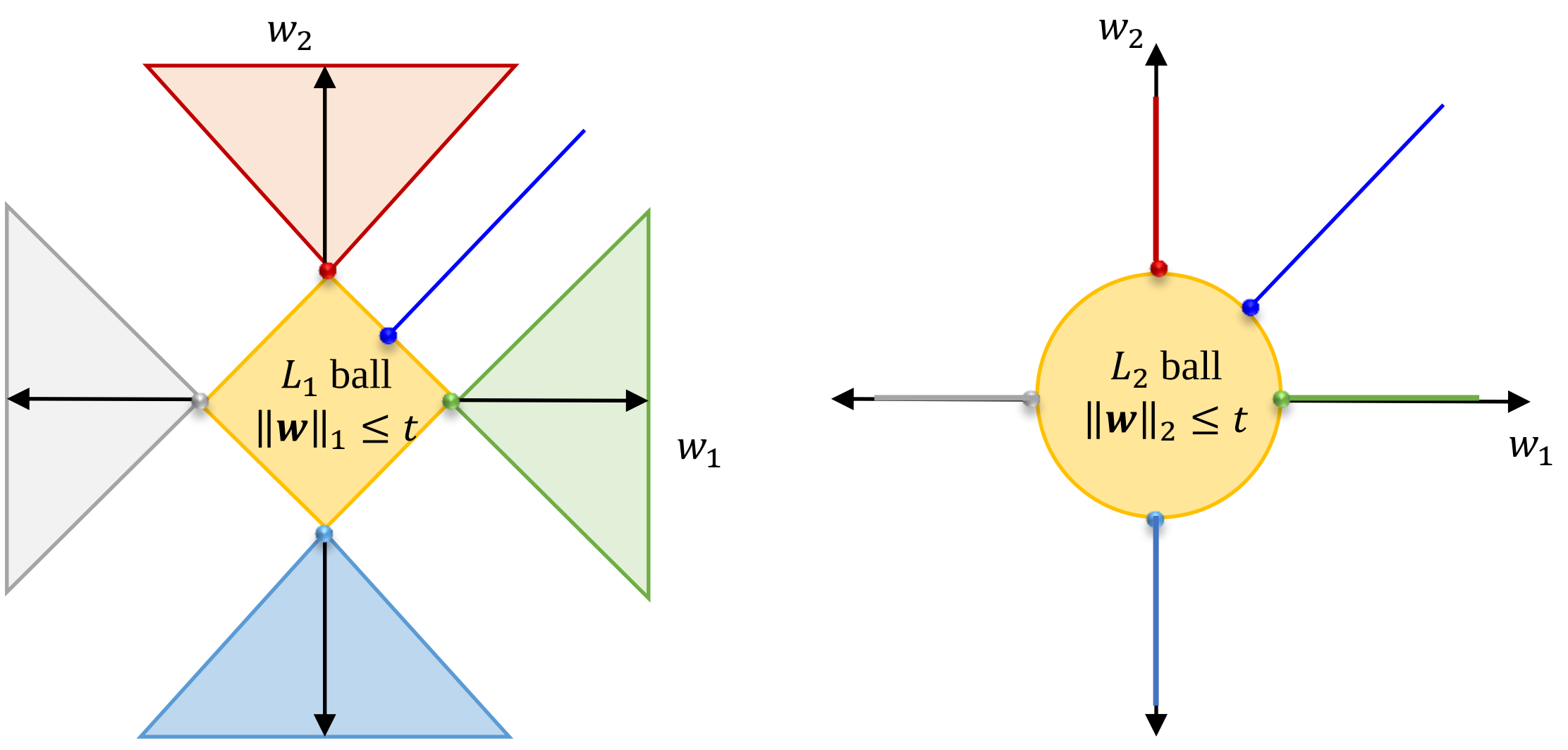

**Figure 7.6.** Illustration in two dimensions of the projection operator onto the $L_1$-ball in (left) and $L_2$ -ball in (right). Points within the red regions (for example) project onto a red dot located at coordinates $(0, t)$. Similarly, points from the other color regions project onto the corresponding color dots. In the $L_1$norm scenario, a significant portion of the diagram is occupied by the color regions, which project onto sparse solutions at the corners of the ball. In contrast, with the $L_2$-norm, projections do not favor sparsity; any non-sparse point—say, for instance on the blue line—is projected onto a non-sparse solution.

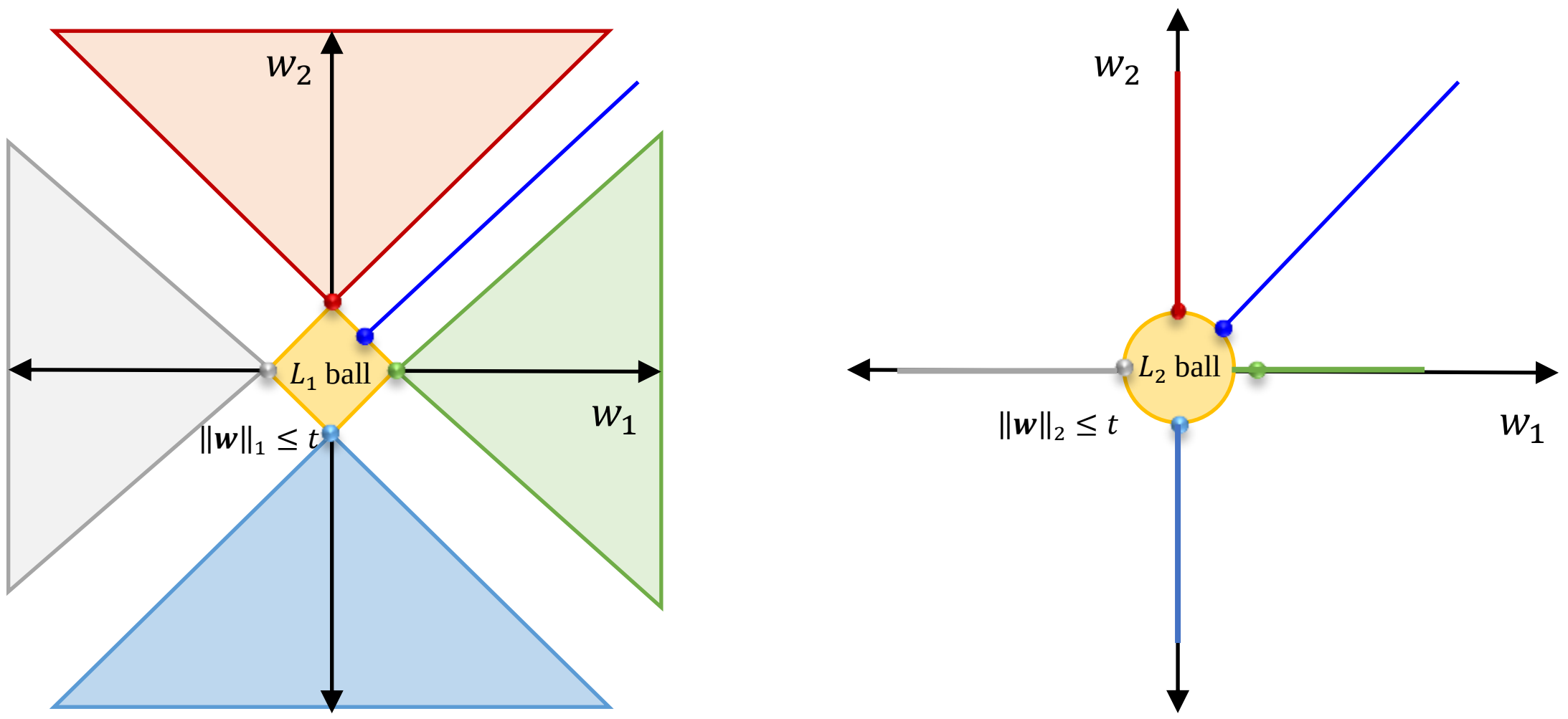

**Figure 7.7.** Illustration in two dimensions of the projection operator onto the $L_1$-ball in (left panel) and $L_2$-ball in (right panel). As the radius of the ball decreases, the size of the colored regions expands, leading to a higher likelihood of sparse solutions manifesting at the ball's corners.

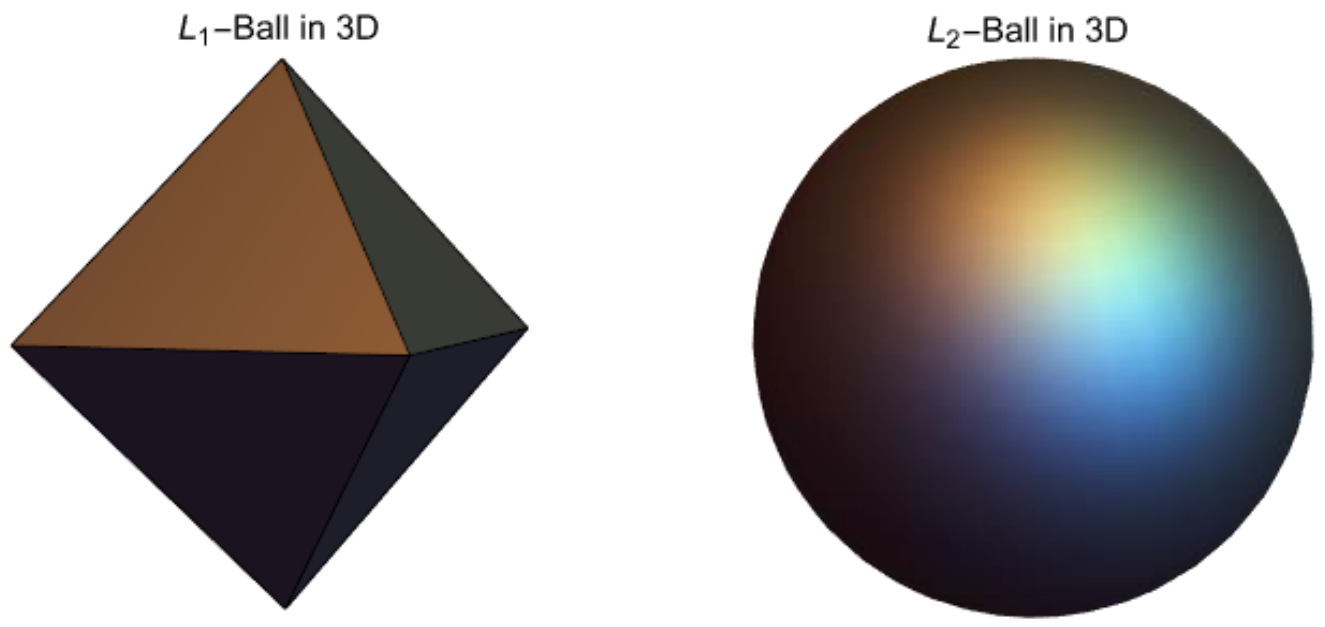

**Figure 7.8.** Representation in three dimensions of the $L_1$ - and $L_2$-balls.

- Figure 7.6 and Figure 7.7 illustrate the effect of the $L_1$ -norm projection and compare it to the case of the $L_2$-norm. The corners of the $L_1$-ball are on the main axes and correspond to sparse solutions. The four corners are represented by color dots, with respective coordinates $(\pm t, 0)$ and $(0, \pm t)$. Most strikingly, a large part of the space in the figure, represented by color regions, ends up on these corners after projection. In contrast, the set of points that is projected onto the blue dot, is simply the blue line. The blue dot corresponds in fact to a dense solution with coordinates $(t/2, t/2)$. Therefore, the figures illustrate that the $L_1$-ball in two dimensions encourages solutions to be on its corners. In the case of the $L_2$ -norm, the ball is isotropic and treats every direction equally.
- The mathematics underlying this involves understanding how the derivatives behave differently for the $L_1$ and $L_2$ norms. For the $L_1$ norm, represented by a diamond shape, the edges are linear, and thus the first-order derivative is constant along each edge, except at the vertices where the derivative is undefined due to the sharp corners. In contrast, the $L_2$ norm, represented by a circle, has a smooth curve with a well-defined first-order derivative at every point. To find the tangent point between the constraint area (diamond or circle) and the target function (ellipse), we look for points where the slopes (first-order derivatives) match. As we adjust the parameter $t$ to shrink the constraint area, the $L_2$ norm (circle) continuously provides points on its curve that meet our derivative-matching requirement, which might include points on the axes (where some weights are zero) or elsewhere (where weights are non-zero). On the other hand, the $L_1$ norm (diamond) behaves differently. When $t$ is relatively large, the edges of the diamond offer a wider range to find matching slopes along the contour lines of the ellipse. However, if $t$ is sufficiently small, the range becomes too narrow to find any matching points on the straight edges, Figure 7.7. The only possible choice points are then at the vertices, where the slope is undefined and can thus match any slope on the ellipse. This characteristic explains why the $L_1$ norm often results in tangent points at the vertices (or axes), leading to the dropout of coefficients by setting them to zero during regularization. In contrast, the $L_2$ norm typically results in non-zero but small coefficients.
- Figure 7.8 depicts two balls characterized by sparsity-inducing norms in three dimensions.

### 7.1.5 Elastic Net Regularization

Elastic net regularization [211] is a hybrid regularization technique that combines both $L_1$ (Lasso) and $L_2$ (Ridge) regularization penalties. It addresses some limitations of using either $L_1$ or $L_2$ regularization alone by leveraging their respective strengths. The elastic net regularization term is a linear combination of the $L_1$ and $L_2$ penalties, controlled by two hyperparameters: $\lambda_1$ and $\lambda_2$. The elastic net regularization term can be expressed as:

$$J_{\text{elastic}} = \mathcal{L} + \lambda_1 \cdot \sum_{i=1}^{d}|w_i| + \lambda_2 \cdot \sum_{i=1}^{d} w_i^2. \tag{7.15}$$

Where:

- $\mathcal{L}$ represents the data loss term (such as mean squared error).
- $\lambda_1$ and $\lambda_2$ are the regularization hyperparameters controlling the strengths of the $L_1$ and $L_2$ penalties, respectively.
- $|w_i|$ represents the absolute value of the $i$-th coefficient ($L_1$ penalty).
- $w_i^2$ represents the square of the $i$-th coefficient ($L_2$ penalty).

- $d$ is the total number of coefficients in the model.

The hyperparameters $\lambda_1$ and $\lambda_2$ control the trade-off between $L_1$ and $L_2$ regularization penalties. A higher $\lambda_1$ value encourages sparsity, while a higher $\lambda_2$ value encourages smaller but non-zero coefficients. Tuning these hyperparameters is crucial to achieve the desired balance between feature selection and coefficient shrinkage. The elastic net method includes the Lasso and Ridge regression: in other words, each of them is a special case where $\lambda_1 = \lambda, \lambda_2 = 0$ or $\lambda_1 = 0, \lambda_2 = \lambda$.

### 7.1.6 Sparse Representations (Activation Sparsity or Activation Regularization)

Both weight decay and activation sparsity aim to reduce overfitting, but they do so through different mechanisms. Weight decay directly simplifies the model by penalizing the magnitude of the weights, leading to smaller, more regularized weight values. Activation sparsity, however, encourages the model to use fewer active neurons, indirectly influencing the weight values through a more complex pathway that involves the AFs of the network. This can lead to a network that is effectively smaller and potentially more interpretable, as it reduces the number of active pathways through the network. In other words, even though the NN might be large and complex only a small part of it is used for predicting any given data instance.

Benefits of activation sparsity:

- By constraining the number of active neurons, it becomes easier to understand which parts of the NN are contributing to the output. This can be particularly useful in deep learning models where interpretability is often a challenge.
- Networks with sparse activations may require less computational resources during inference, as fewer computations are needed when many activations are zero.
- Similar to parameter sparsity, activation sparsity can help in reducing overfitting by limiting the capacity of the network to memorize training data, thus potentially improving generalization to new data.

While activation regularization can offer significant benefits, it also presents challenges:

- Like all regularization techniques, the strength of the regularization (i.e., how many activations you penalize) needs careful tuning to balance between too much sparsity (underfitting) and too little (overfitting).

In a typical NN, you have a sequence of layers, for each layer $l$, the activation values $\mathbf{a}^{(l)}$ are calculated using the weighted sum of the inputs followed by an AF:

$$\mathbf{z}^{(l)} = \mathbf{W}^{(l)} \cdot \mathbf{a}^{(l-1)} + \mathbf{b}^{(l)}, \tag{7.16.1}$$

$$\mathbf{a}^{(l)} = \sigma^{(l)}\left(\mathbf{z}^{(l)}\right), \tag{7.16.2}$$

where: $\mathbf{W}^{(l)}$ is the weight matrix for layer $l$, $\mathbf{b}^{(l)}$ is the bias vector for layer $l$, and $\sigma^{(l)}$ is the AF for layer $l$. The final output of the network is obtained after the activations pass through all layers. When incorporating activation regularization, you modify the loss function used during training by adding a regularization term $\Omega(\mathbf{a})$, which depends on the activations. The new loss function $J$ becomes:

$$J = \mathcal{L} + \lambda\Omega(\mathbf{a}). \tag{7.17}$$

Here, $\mathcal{L}$ is the original loss function (such as cross-entropy or mean squared error), $\lambda$ is a hyperparameter that controls the strength of the regularization, and $\Omega(\mathbf{a})$ is the regularization term. The choice of $\Omega(\mathbf{a})$ dictates the nature of the sparsity imposed, often using the $L_1$-norm,

$$\begin{aligned}\Omega(\mathbf{a}) &= \|\mathbf{a}\|_1 \\ &= \sum_{i=1}^{m} |a_i|.\end{aligned} \tag{7.18}$$

Here, $m$ is the total number of units in the network, and $a_i$ is the value of the $i$th hidden unit.

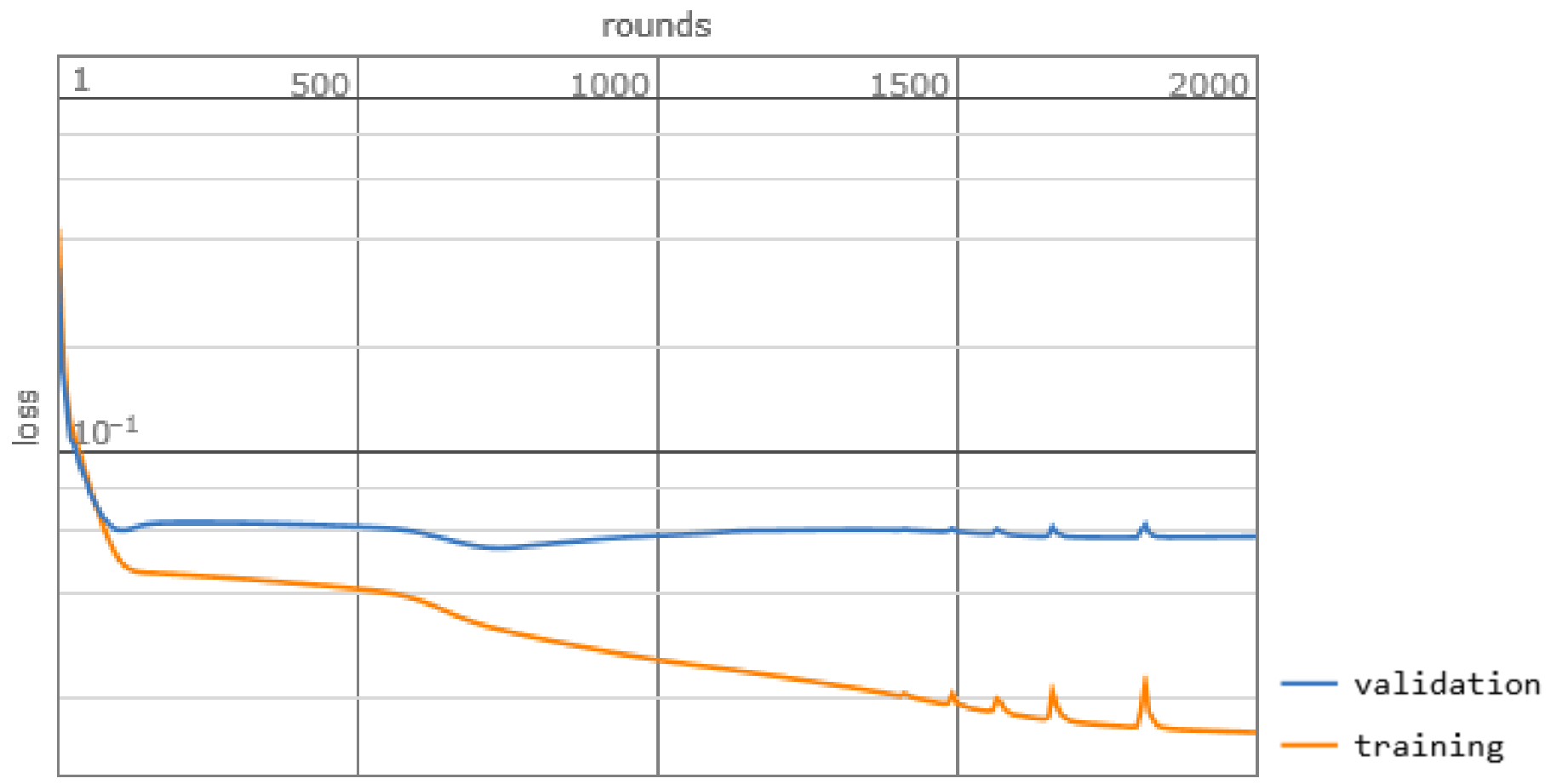

**Figure 7.9.** Learning curves depicting the evolution of loss over time, measured in terms of training iterations (epochs or rounds) across the dataset.

## 7.2 Early Stopping

We have already investigated how a model's generalization performance changes with variations in the number of parameters, the dataset size, and the coefficient of weight-decay regularization. Each of these factors introduces a trade-off between bias and variance, aiming to minimize the generalization error. Another critical element influencing this balance is the learning process itself. During the optimization of the error function via GD, training error typically decreases as model parameters are updated. In contrast, error on hold-out data might exhibit non-monotonic behavior. This phenomenon can be captured through learning curves, which chart performance metrics—like errors on the training and validation sets—across the iterations of an iterative learning process such as SGD. These curves not only offer valuable insights into the training progress but also provide a practical approach to optimizing model complexity and achieving optimal generalization performance, see Figure 7.9.

Early stopping is a technique used in training NNs to prevent overfitting. It is based on the observation that, during the training of a NN, the error (or loss) typically decreases on both the training set and the validation set initially. However, at some point, further training causes the error on the validation set to start increasing, even though the error on the training set continues to decrease. This phenomenon occurs because the model starts to overfit the training data, capturing noise and idiosyncrasies in the training set that do not generalize well to new data. Early stopping helps mitigate this by monitoring the model's performance on a separate validation dataset and stopping the training process when the performance on the validation set starts to degrade, indicating that the model is starting to overfit.

Keeping track of the best solution achieved so far on the validation set is crucial for effective early stopping. This helps prevent premature stopping due to minor fluctuations or noise in the validation error. By continuing to train and monitor the validation error, one can determine if the model is truly starting to overfit and if further training is unlikely to improve generalization performance. Early stopping, therefore, involves a balance between stopping early enough to prevent overfitting and continuing training long enough to ensure that the model has converged to a stable solution. The termination point is determined retrospectively during training. The model's performance on the validation set is continuously monitored after each epoch (or a certain number of iterations). If the performance on the validation set begins to deteriorate, such as an increase in validation loss or a decrease in accuracy, for a specified number of epochs (referred to as the patience parameter), the training process is halted prematurely. The model's weights from the epoch with the best performance on the validation set are then retained as the final model. It's important to use early stopping carefully, as stopping too early can lead to underfitting, where the model fails to capture the underlying patterns in the data. Procedure 7.1 represents how early stopping typically works.

**Procedure 7.1:** How does early stopping work in NN training?

1. Before training the NN, the available data is split into three sets:
   a. Training Set: Used to update the model's weights during training.
   b. Validation Set: Used to monitor the model's performance and decide when to stop training.
   c. Test Set: Held out and used only at the end to evaluate the final performance of the trained model.
2. During training, the NN is trained using the training set. After each epoch (or a certain number of iterations), the model's performance is evaluated on the validation set.
3. The performance metric used for monitoring can vary based on the task. For example, for classification tasks, metrics like accuracy, precision, recall, or F1 score can be used. For regression tasks, metrics like MSE or MAE can be used.
4. Early stopping involves specifying a criterion to determine when to stop training. One common criterion is to stop training when the performance on the validation set stops improving or starts to degrade. This can be detected by monitoring the validation metric and keeping track of the best value achieved so far.
5. To avoid stopping training prematurely due to temporary fluctuations in performance, a patience parameter is used. This parameter specifies the number of epochs to wait after the last time the best validation performance was achieved. If the performance does not improve within this patience period, training is stopped. Setting the right "patience" and "threshold" (the minimal change in validation performance to qualify as an improvement) is crucial for effective early stopping.
6. Once training is stopped, the model weights from the epoch with the best performance on the validation set are selected as the final model.
7. Finally, the selected model is evaluated on the test set to estimate its performance on new, unseen data.

Some key advantages of early stopping:

- By halting the training process before the model learns the noise in the training data, early stopping keeps the model generalized. This is crucial for ensuring that the model performs well on new, unseen data.
- Early stopping can lead to a reduction in computational resources and time because it stops the training process once no significant improvements are made, avoiding unnecessary computations.
- It helps in tuning other hyperparameters of the model more efficiently because the validation performance is a direct feedback mechanism during training.
- Compared to methods like weight decay that require trying different values of the regularization parameter $\lambda$, early stopping is relatively inexpensive computationally. It does not require additional hyperparameter tuning runs, which can save time and resources.
- Early stopping can be used in combination with other regularization techniques, such as weight decay or dropout, to further improve the generalization performance of the model. Its ease of integration makes it a valuable addition to the regularization toolbox.
- For a quadratic error function, Early stopping should exhibit similar behaviour to regularization using a simple weight-decay term [212].

How early stopping acts as regularization? There are two key perspectives to consider: The constraining optimization perspective and the variance perspective. By limiting the number of optimization steps in the training process, early stopping can be seen as imposing a constraint on the optimization process. This constraint effectively restricts the distance of the final solution from the initialization point, as the optimization process is terminated before the model has the opportunity to move too far from its initial parameters, see Figure 7.10. In the context of machine learning, adding constraints to the model is often a form of regularization. Regularization techniques are used to prevent overfitting by imposing additional constraints on the optimization problem. These constraints can take various forms, such as penalizing large weights (as in weight decay or $L_2$ regularization) or limiting the complexity of the model (as in early stopping).

On the other hand, by understanding early stopping from the variance perspective, we will gain a clearer view of how this technique effectively curbs overfitting and improves generalization. First, let's revisit the bias-variance trade-off. The empirical loss function, derived from a finite set of training data, is an approximation of the true loss function,

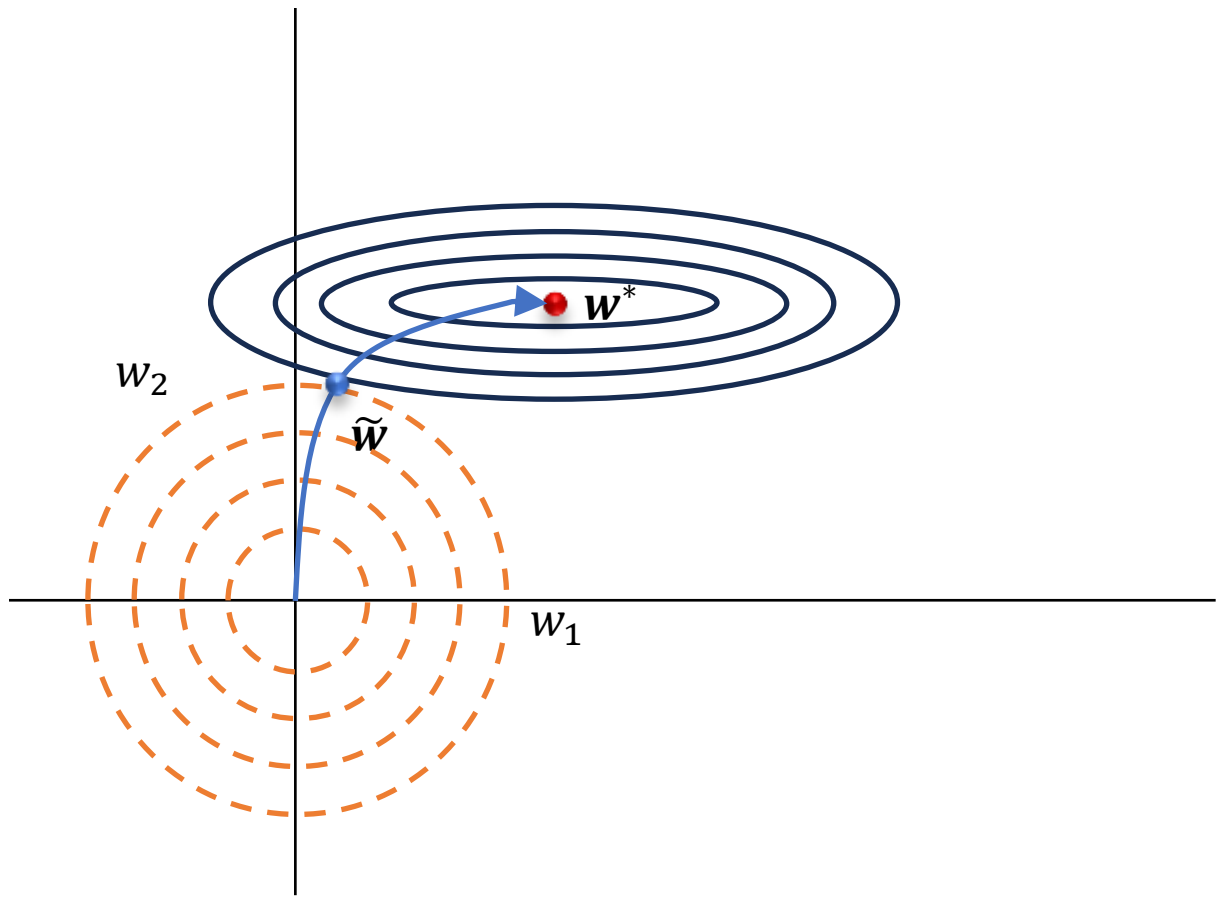

**Figure 7.10.** A schematic illustration of why early stopping yields results similar to weight decay for a quadratic error function. The solid ellipses represent the contours of the error function, while the dotted circles show the regularization term. For $\lambda = 0$, the minimum error is indicated by $\boldsymbol{w}^*$. When $\lambda > 0$, the minimum of the regularized error function is shifted towards the origin. The red point at the center of the contour ellipses represents the maximum likelihood solution, $\boldsymbol{w}^*$, corresponding to the minimum of the unregularized error function. The blue path illustrates the trajectory of the weight vector as it starts from the origin (the point where the blue path begins) and moves according to the negative gradient direction. If the weight vector continues training fully, it will converge to $\boldsymbol{w}^*$. The intermediate point marked on the path, $\widetilde{\boldsymbol{w}}$, represents the weight vector obtained by stopping training early. This solution, $\widetilde{\boldsymbol{w}}$, is qualitatively similar to the solution obtained using a weight-decay regularizer.

which could only be computed with an infinite amount of data. Let's expand on this with an illustrative example and explain the implications of this shifting:

Illustrative Example: Loss Function Contours

Imagine you have a model trying to fit data to predict outcomes based on inputs. You construct a loss function to measure the discrepancy between your model's predictions and the actual outcomes. If you had access to infinite data, you could perfectly map the true relationship, and your loss function would accurately reflect the true errors of your model across all possible inputs. However, with finite data, your constructed loss function is just an approximation. The true loss function represents the ideal scenario where the loss is calculated over an infinite dataset. The contours of this loss function (often visualized in 2D or 3D plots as concentric shapes like circles or ellipses) center around the optimal set of parameters (say $\boldsymbol{w}^*$). When you use a specific finite dataset (Set A) to train your model, the contours of the loss function might shift slightly due to the peculiarities and noise within Set A. The minimum of this loss function (say $\boldsymbol{w}_A$) will be different from $\boldsymbol{w}^*$. A different dataset (Set B) would lead to another variation of the loss function. Its contours shift differently compared to Set A, centering around a different set of parameters ($\boldsymbol{w}_{\mathrm{B}}$). If we could visualize these contours: The true loss function might be centered at a point in a graph. The empirical loss functions would show contours that are somewhat offset or distorted compared to the true loss function. These offsets represent the variance introduced by the specific training data used, see Figure 7.11.

The shifts in the loss function represent the variance of the model – how sensitive the model is to the specific set of training data. Different datasets lead to different parameters being optimal, indicating high variance if the shifts are significant. If all empirical loss functions tend to be consistently off from the true loss function in a similar way, this represents the bias of the model. It's an error introduced by a systematic deviation from the truth, often due to the model's inability to capture the underlying complexity. Reducing variance typically increases bias and vice versa. For example, a more complex model (like a high-degree polynomial) might fit each training dataset very closely (low bias), but its parameters might vary wildly with different datasets (high variance). Conversely, a simpler model might not fit the training data as closely (higher bias) but shows less variation in parameters across different datasets (lower variance).

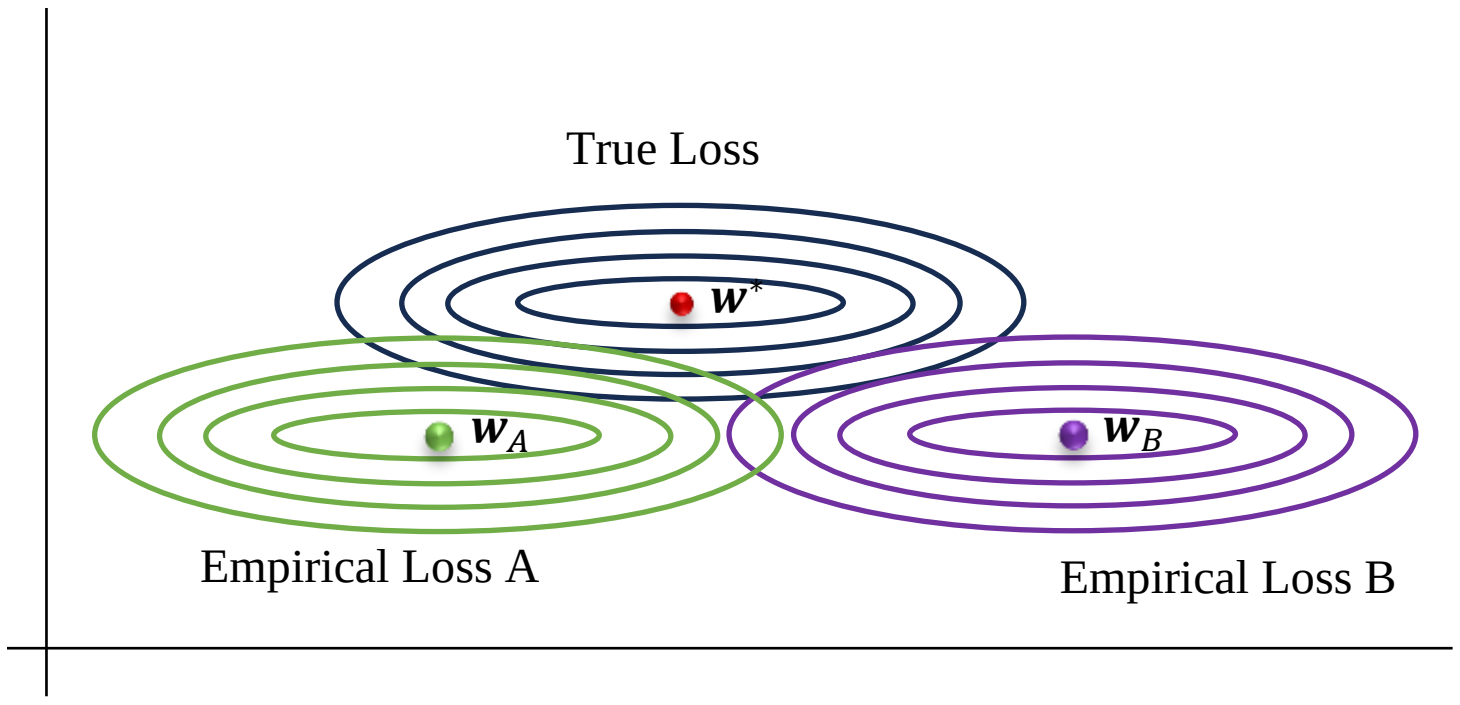

**Figure 7.11.** Shift in loss function caused by variance effects.

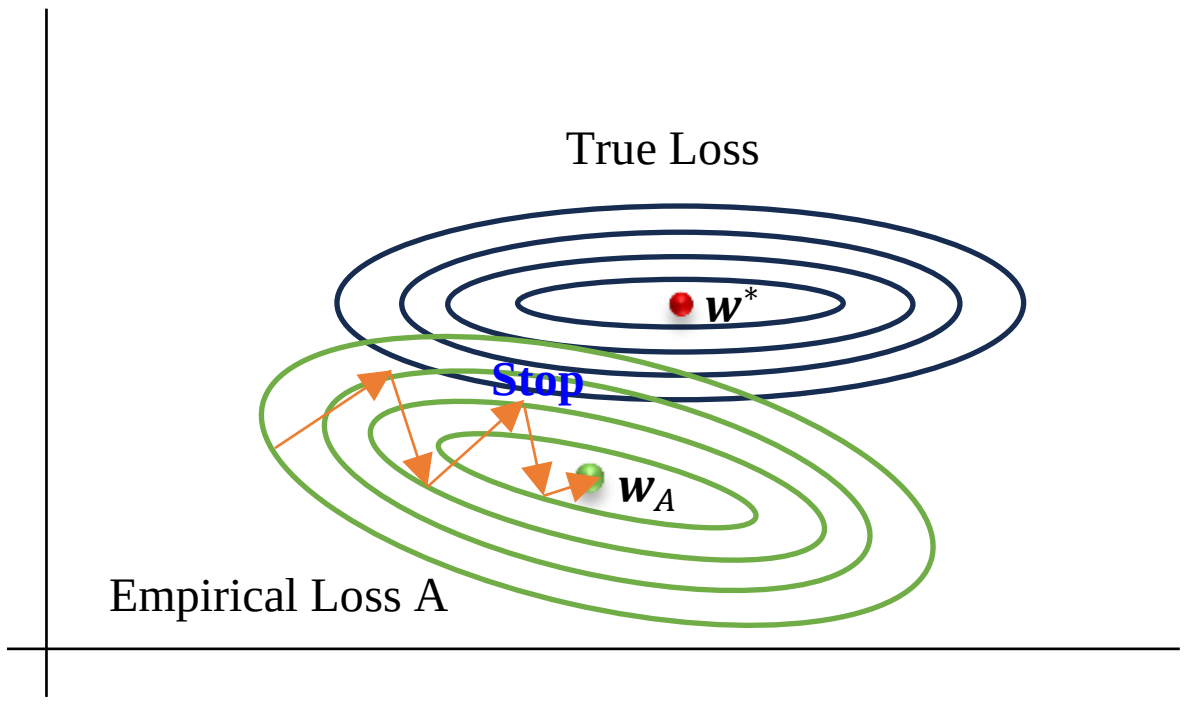

**Figure 7.12.** Because of the differences in the true loss function and that on the training data, the error will begin to rise if GD is continued beyond a certain point.

In Figure 7.11, we see two sets of contours: True loss contours represent the true loss function that we would have if we could evaluate our model on the entire distribution of the data, which is practically infinite. These contours are centered around the true optimum of the model parameters ($\boldsymbol{w}^*$). Training loss contours are the loss function contours obtained from the available finite training data. Because this training data is a sample from the entire data distribution, it contains noise and other variations that cause the contours to be shifted and possibly distorted compared to the true loss contours. The shift of the training loss contours from the true loss contours is a visual representation of the variance introduced by using a finite dataset. Different finite samples would shift the training loss in different ways, which reflects how variance affects our model training.

The 'STOP' sign, Figure 7.12, within the trajectory of the optimization path, indicates the point where early stopping would freeze the model's learning. If the optimization were to continue (following the path within the training loss contours), the model might find a minimum within the training loss contours, but this minimum would likely be far from the true minimum, leading to overfitting. This path shows the trajectory of GD optimization. Initially, the path seeks to minimize the training loss, getting closer to the training data's local minimum. However, without early stopping, it could continue to the point where the error with respect to the true loss function begins to rise, as the model starts to 'learn' the noise in the data rather than the true underlying patterns. The goal is to have our model generalize well to new, unseen data by finding a point on the optimization path where the training loss is low, but not so far along the path that we start fitting to noise. Early stopping helps to achieve this by stopping training at the point where the performance on a validation set starts to deteriorate.

In other words, the training loss function is the function that the learning algorithm can minimize directly because it is calculated from the training dataset. True loss function represents the ideal loss across the entire data distribution, which is unknown and can't be computed directly because we don't have access to all possible data. The path taken by GD is not direct. It is influenced by the shape of the loss function, learning rate, and other factors. This path is often "circuitous and oscillatory", meaning it can zigzag and loop around the true optimal point. The cause of this behavior includes the local geometry of the loss function (like the presence of saddle points or local minima) and the stochastic nature of the data samples. As it nears convergence on the training data, it might pass through several states that actually yield a better generalization (lower true loss) than the eventual training data optimum, see Figure 7.12, (the GD will often encounter better solutions with respect to the true loss function before it converges to the best solution with respect to the training data.). As the model trains, its accuracy on a separate validation set is monitored. This validation set is not used for training and thus serves as a proxy for the true loss function. A good early stopping point is one where the validation accuracy starts to decrease or cease improving, indicating potential overfitting if training continues and therefore provides a good termination point. The "good early stopping point" is essentially a point on the GD path that yields the best trade-off between bias and variance, providing the best generalization performance on unseen data. The concept acknowledges that while we can't directly minimize the true loss, we can infer its behavior by carefully observing the validation set performance during training. Therefore, early stopping relies on the validation set as an estimator of the true loss function to determine the best point to stop training.

## 7.3 Ensemble Methods

When multiple models are trained to solve the same problem, rather than selecting a single best model, improved generalization can often be achieved by averaging the predictions of these individual models. Such combinations are often referred to as committees or ensembles.

### 7.3.1 Bagging and Subsampling

When you have access to an infinite amount of data from the distribution that generates your samples, you can create an unlimited number of diverse training sets. Each of these training sets is an independent sample from the underlying data distribution. For each training dataset, a model can be trained independently. Because each dataset is a fresh sample from the distribution, the models will differ from one another in the details of what they learn, despite being trained to predict the same underlying phenomenon. The final prediction for any test instance can be obtained by averaging the predictions of all these independently trained models. If a sufficient number of training data sets is used, the variance of the prediction will be reduced to 0.

Indeed, while the theoretical model of using infinite data sets for training to minimize variance isn't practical due to the finite nature of real-world data, we can still apply a similar principle using the available data through methods like bagging and subsampling. These techniques attempt to simulate the effect of having multiple different training sets by creating variations from the single available dataset. The basic idea is to generate new training data sets from the single instance of the base data by sampling. The predictions for a specific test instance, derived from models trained on different training sets, are then averaged to produce the final prediction. The main difference between subsampling and bagging is the sampling technique (with or without replacement). An alternative approach to forming an ensemble is to use the original data set to train multiple different models having different architectures.

We can analyze the benefits of ensemble predictions by considering a regression problem with an input vector $\mathbf{x}$ and a single output variable $y$. Suppose we have a set of trained models $y_1(\mathbf{x})$,…, $y_M(\mathbf{x})$, and we form a committee prediction given by

$$y_{\mathrm{COM}}(\mathbf{x}) = \frac{1}{M}\sum_{m=1}^{M} y_m(\mathbf{x}). \tag{7.19}$$

If the true function that we are trying to predict is given by $h(\mathbf{x})$, then the output of each of the models can be written as the true value plus an error:

$$y_m(\mathbf{x}) = h(\mathbf{x}) + \epsilon_m(\mathbf{x}). \tag{7.20}$$

The average sum-of-squares error then takes the form

$$\mathbb{E}_{\mathbf{x}}[\{y_m(\mathbf{x}) - h(\mathbf{x})\}^2] = \mathbb{E}_{\mathbf{x}}[\{\epsilon_m(\mathbf{x})\}^2], \tag{7.21}$$

where $\mathbb{E}_{\mathbf{x}}[\cdot]$ denotes a frequentist expectation with respect to the distribution of the input vector $\mathbf{x}$. The average error made by the models acting individually is therefore

$$\mathbb{E}_{\text{AV}} = \frac{1}{M} \sum_{m=1}^{M} \mathbb{E}_{\mathbf{x}}[\{\epsilon_m(\mathbf{x})\}^2]. \tag{7.22}$$

Similarly, the expected error from the committee (7.19) is given by

$$\mathbb{E}_{\text{COM}} = \mathbb{E}_{\mathbf{x}}\left[\left\{\frac{1}{M}\sum_{m=1}^{M} y_m(\mathbf{x}) - h(\mathbf{x})\right\}^2\right] = \mathbb{E}_{\mathbf{x}}\left[\left\{\frac{1}{M}\sum_{m=1}^{M} \epsilon_m(\mathbf{x})\right\}^2\right]. \tag{7.23}$$

If we assume that the errors have zero mean and are uncorrelated, so that

$$\mathbb{E}_{\mathbf{x}}[\epsilon_m(\mathbf{x})] = 0, \tag{7.24.1}$$

$$\mathbb{E}_{\mathbf{x}}[\epsilon_m(\mathbf{x})\epsilon_l(\mathbf{x})] = 0, m \neq l, \tag{7.24.2}$$

then we obtain

$$\mathbb{E}_{\text{COM}} = \frac{1}{M}\mathbb{E}_{\text{AV}}. \tag{7.25}$$

This apparently dramatic result suggests that the average error of a model can be reduced by a factor of $M$ simply by averaging $M$ versions of the model. Unfortunately, it depends on the key assumption that the errors due to the individual models are uncorrelated. In practice, the errors are typically highly correlated, and the reduction in the overall error is generally much smaller. It can, however, be shown that the expected committee error will not exceed the expected error of the constituent models, so that $\mathbb{E}_{\text{COM}} \ll \mathbb{E}_{\text{AV}}$.

Now, we summarize these methods as follows:

### Bagging

- In bagging [213], multiple bootstrap samples are drawn from the original dataset. A bootstrap sample is created by randomly selecting data points with replacements, meaning each sample can have repeated elements and some original data points might be omitted. The sample size $S$ may be different from the size of the training data size $N$, although the classical approach is to set $S$ to $N$.
- A model is trained separately on each bootstrap sample. Because the data in each sample is slightly different due to the sampling with replacement, each model learns differently, capturing various aspects of the data.
- After training, predictions from each model are aggregated to form a final prediction. For regression problems, this might be the average of all model predictions. For classification, it could be the majority vote.
- Mathematically, let $X \equiv \{\mathbf{x}_1, \mathbf{x}_2, \ldots, \mathbf{x}_N\}$ be the original dataset containing $N$ samples. In bagging, we create $M$ new datasets $X_1$, $X_2$, …, $X_M$, each of size $N$ as well. Each $X_i$ is sampled from $X$ with replacement, which means each sample in $X$ can appear multiple times or not at all in each $X_i$. Let $f(\mathbf{x}; \boldsymbol{\theta}_i)$ represent a model trained on the dataset $X_i$ with parameters $\boldsymbol{\theta}_i$. The parameters $\boldsymbol{\theta}_i$ are estimated separately for each model based on its respective dataset. The final prediction for a new data point $\mathbf{x}$ is obtained by averaging the predictions from all models. For regression, the aggregated prediction is given by:

$$\hat{f}(\mathbf{x}) = \frac{1}{M}\sum_{i=1}^{M} f(\mathbf{x}; \boldsymbol{\theta}_i). \tag{7.26}$$

  For classification, the final prediction is often obtained by majority voting.

### Subsampling

- Subsampling is similar to bagging but with a key difference in the sampling method. Instead of bootstrap samples, subsampling involves drawing random samples from the original dataset without replacement. In this case, it is essential to choose $S < N$, because choosing $S = N$ yields the same training data set and identical results across different ensemble components. This means each data point can only appear once in each subset.

- Like in bagging, models are trained independently on each subsample. Since each subsample is different and none have overlapping data points, the models might have more varied learning outcomes compared to bagging.
- The final prediction is made by aggregating the outputs from all models, similar to bagging.
- Mathematically, unlike bagging, subsampling creates subsets $X_i$ by drawing samples from $X$ without replacement, usually less than $N$. Similar to bagging, we train models $f(\mathbf{x};\boldsymbol{\theta}_i)$ on each subset $X_i$. The final prediction method remains similar: $\hat{f}(\mathbf{x}) = \frac{1}{M}\sum_{i=1}^{M} f(\mathbf{x};\boldsymbol{\theta}_i)$.
- Subsampling can lead to more diversity among the models in the ensemble because each model sees a completely distinct set of data points.

Using these ensemble techniques, practitioners can achieve a more robust, stable, and accurate model performance on various tasks. These methods are particularly effective in settings prone to overfitting or when the dataset is not large enough to represent the entire complexity of the underlying distribution.

**Remarks:**

- When there's a large amount of data available, subsampling is often preferred because it can reduce the training time and computational resources needed while still providing diverse datasets for each model. Since each model sees a different subset, they can still learn varied aspects of the data, which helps in building a robust ensemble. Bagging can be particularly useful when the available data is limited because it maximizes the usage of this data by creating diverse training sets through resampling.
- The main challenge in directly using bagging for NNs is indeed the computational cost and inefficiency, as training multiple DNNs can be resource-intensive. The silver lining is that the training of these individual models can be fully parallelized. This means that if sufficient computational resources are available, such as multiple GPUs or a distributed computing system, the training can be carried out simultaneously across different models, which makes the process scalable.

### 7.3.2. Parametric Model Selection and Averaging

In deep learning, the primary challenge is the vast number of potential configurations one must consider. This includes decisions about the number of layers, the number of units in each layer, the choice of AFs, etc. This creates a huge parameter space that is impractical to explore exhaustively. Given the extensive combinations available, it's typically feasible only to explore a limited subset of all possible configurations. To enhance model reliability and reduce variance, one effective strategy is to select the top 'k' configurations based on their performance and average their predictions. This ensemble method, similar to bagging, tends to yield more robust predictions, particularly when the chosen configurations are diverse. Although each individual configuration might not be optimal, the aggregate outcome often proves to be quite resilient. Moreover, like bagging, the training process for multiple configurations can be executed in parallel, significantly speeding up the time to deployment and improving computational efficiency.

### 7.3.3. Randomized Connection Dropping

The random dropping of connections between different layers in a multilayer NN often leads to diverse models in which different combinations of features are used to construct the hidden variables. The dropping of connections imposes constraints on each model, potentially making each individual model less powerful or capable than a fully connected model. This is because each model has fewer paths for information flow and fewer parameters, which might limit its capacity to capture complex patterns in the data. Despite individual models being potentially less powerful, the ensemble's strength comes from the diversity of its models. By averaging the predictions of all models in the ensemble, one can often achieve higher accuracy and robustness. This is because errors in individual models are likely to be uncorrelated or differently biased, thus canceling out when averaged. Each model in the ensemble has its own set of weights, distinct from the others. This means that the learning process in each model is independent in terms of parameter adjustments. This independence is crucial for maintaining the diversity of the models within the ensemble. This technique is different from Dropout.

## 7.4 Dropout

Model combination (model ensemble) is known to significantly enhance machine learning performance due to its ability to reduce variance, leveraging the strengths of individual models. The challenge, however, lies in its computational cost, particularly with large NNs. Combining multiple models is most effective when the individual models vary significantly from one another. To achieve this diversity in NN models, they should either have distinct architectures or be trained on different datasets. However, finding optimal hyperparameters for each architecture is a challenging task, and training each large network requires significant computational resources. Additionally, large networks typically need extensive training data, which may not be available to train different networks on different data subsets. Even if one could train multiple distinct large networks, deploying all of them during inference would be impractical in scenarios where rapid response times are crucial. Dropout (Srivastava et al. [214]) is a technique that addresses these challenges. It prevents overfitting while offering an efficient way to approximate the combination of exponentially many diverse NN architectures. It is simple, yet highly effective in improving the generalization capabilities of NNs. It provides a computationally inexpensive but powerful method of regularizing a broad family of models.

Dropout uses node sampling instead of edge sampling in order to create a NN ensemble. The term "dropout" refers to the process of dropping out units, both hidden and visible, from a NN. Dropping out a unit means temporarily removing it from the network, along with all of its incoming and outgoing connections, as illustrated in Figure 7.13. In most modern NNs, which rely on a series of affine transformations and nonlinearities, we can effectively remove a unit from the network by multiplying its output value by zero. For simplicity, we present the dropout algorithm in terms of multiplication by zero, but it can be easily adapted to work with other operations that achieve the same effect.

The choice of which units to drop is random. In the simplest case, each unit is retained with a fixed probability $p$ that is independent of other units. In some cases, the probability of sampling input nodes differs from that of hidden nodes. This probability can be selected using a validation set or simply set at 0.5, which tends to be close to optimal for a wide range of networks and tasks. For input units, however, the optimal retention probability is usually closer to 1 than 0.5. For instance, if `$p = 0.5$`, each neuron has a 50% chance of being dropped out. The nodes are sampled exclusively from the input and hidden layers of the network. It's important to note that sampling the output node(s) would make it impossible to generate predictions and compute the loss function. Dropout can also be tailored for each layer. For example, in layers with many neurons, a small probability $p$ can be used. For layers with fewer neurons, $p$ can be set to 1.0, thereby retaining all neurons in those layers.

In dropout, the training process involves the following steps, repeated iteratively to cycle through all training points:

- Sample a subnetwork from the base network.
- Select a single training instance or a mini-batch of training instances.
- Update the weights of the retained edges in the sampled network using backpropagation on the sampled training instance or mini-batch.

To sample a subnetwork from the base network using dropout, you randomly deactivate a subset of neurons from the original network during training. The process involves generating a dropout mask and applying it to the base network to keep only active neurons. A dropout mask is a binary vector (for example [0,…,1,…,0,…,1…]) used to determine which neurons are active or inactive during the forward pass of a NN layer when applying dropout. Each element in the mask corresponds to a particular neuron, where 1 means the neuron is active, and 0 means the neuron is dropped out (inactive). The dropout mask is generated randomly using a specified dropout probability $(1 - p)$, where $p$ is the keep probability, i.e., each neuron has a probability of $p$ to be included in the mask and a probability of $1 - p$ to be excluded.

Let's say you're training a network for image classification with dropout applied to the hidden layers. In one batch, neurons 1, 3, and 5 might be active in a particular layer (consisting of 5 neurons), while neurons 2 and 4 are dropped. This configuration forms a specific sub-model. In the next batch, the dropout mask might deactivate neurons 3 and 5 while keeping neurons 1, 2, and 4 active, forming a different sub-model. This variability ensures that the network learns robust features, as it must perform well regardless of which specific neurons are active.

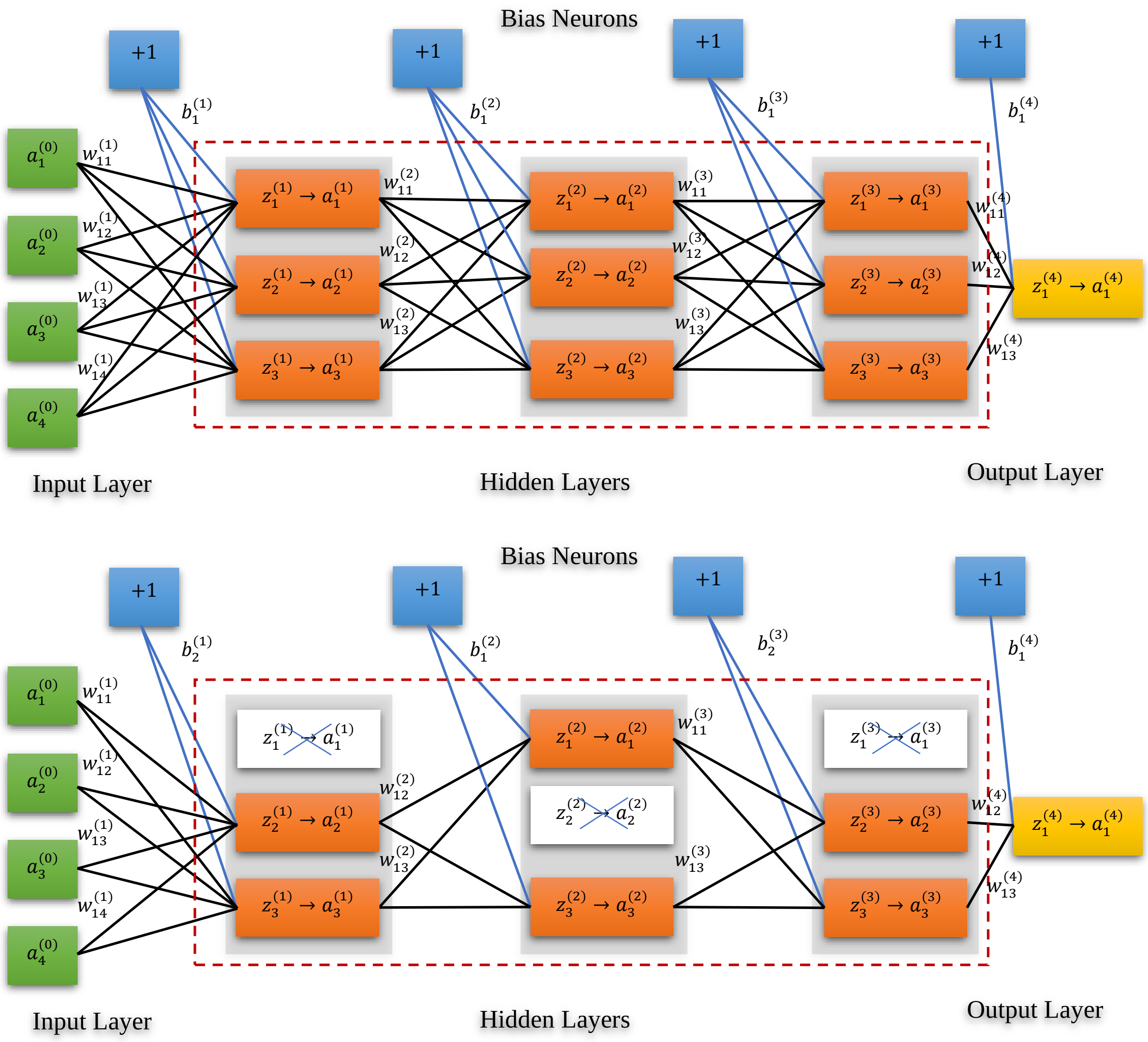

**Figure 7.13.** Dropout Neural Net Model. Upper panel: A standard neural net with 3 hidden layers. Lower panel: An example of a thinned net produced by applying dropout to the network on the upper panel. Crossed units have been dropped.

Each iteration of training (each epoch) involves multiple passes through the data (batches). For each batch during each epoch, a new dropout mask is generated independently. During each forward pass, a dropout mask is applied to deactivate a random subset of neurons in the layers where dropout is implemented. This mask determines which neurons are "dropped" and which remain active, effectively creating a new sub-model for that batch. The sub-model for one batch is independent of the sub-model for any previous or subsequent batch. Each sub-model (configured by its unique dropout mask) exists only for the duration of its specific forward and backward pass. Each training batch can be seen as a separate experiment with a different "thinned" version of the original network, helping to ensure that the model generalizes well and is robust against overfitting. Since each training batch potentially sees a different configuration of active and inactive neurons, the NN learns to generalize across a variety of network architectures.

Only the active neurons (those not dropped) contribute to the forward pass. The forward pass computes the output using the weights of the base network shared across subnetwork. The loss is computed based on the output of the current subnetwork. Gradients are calculated only for the active neurons. These gradients are computed with respect to the base network's weights. Gradients are used to update the base network's weights via an optimization algorithm (e.g., SGD). Each weight update integrates the gradient information from the current subnetwork. The base network's weights get updated for the next mini-batch.

Consider a NN with $L$ hidden layers. Let $l \in \{1, \ldots, L\}$ index the hidden layers of the network. Let $\mathbf{z}^{(l)}$ denote the vector of inputs into layer $l$, $\mathbf{a}^{(l)}$ denote the vector of outputs from layer $l$ ($\mathbf{a}^{(0)} = \mathbf{x}$ is the input). $\mathbf{W}^{(l)}$ and $\mathbf{b}^{(l)}$ are the weights and biases at layer $l$. The feed-forward operation of a standard NN can be described as (for $l \in \{0, \ldots, L-1\}$ and any hidden unit $i$)

$$z_i^{(l+1)} = \mathbf{w}_i^{(l+1)}\mathbf{a}^{(l)} + b_i^{(l+1)}, \tag{7.27.1}$$
$$a_i^{(l+1)} = \sigma\big(z_i^{(l+1)}\big), \tag{7.27.2}$$

where $\sigma$ is any AF. With dropout, the feed-forward operation becomes

$$r_i^{(l)} \sim \mathrm{Bernoulli}(p), \tag{7.28.1}$$
$$\tilde{\mathbf{a}}^{(l)} = \mathbf{r}^{(l)} \odot \mathbf{a}^{(l)}, \tag{7.28.2}$$
$$z_i^{(l+1)} = \mathbf{w}_i^{(l+1)}\tilde{\mathbf{a}}^{(l)} + b_i^{(l+1)}, \tag{7.28.3}$$
$$a_i^{(l+1)} = \sigma\big(z_i^{(l+1)}\big). \tag{7.28.4}$$

Here $\odot$ denotes an element-wise product. For any layer $l$, $\mathbf{r}^{(l)}$ is a vector of independent Bernoulli random variables each of which has probability $p$ of being 1. This vector is sampled and multiplied element-wise with the outputs of that layer, $\mathbf{a}^{(l)}$, to create the thinned outputs $\tilde{\mathbf{a}}^{(l)}$. The thinned outputs are then used as input to the next layer. This process is applied at each layer. This amounts to sampling a sub-network from a larger network. For learning, the derivatives of the loss function are backpropagated through the sub-network. Training dropout NNs involves a process similar to standard NNs using SGD. However, the key difference is that, for each training instance within a mini-batch, a "thinned" network is sampled by randomly dropping out some units. Forward and backward propagation for that specific training instance are conducted solely on this thinned network.

- For each training instance in a mini-batch, a subset of units is randomly selected to form a thinned network. The non-selected units are effectively "dropped out" and excluded from the network.
- The training instance is passed through the thinned network, and backpropagation is performed to compute gradients.
- The gradients for each parameter are averaged over the training instances in the mini-batch. Any training instance that does not use a particular parameter contributes a zero gradient for that parameter.

Note that dropout and bagging are both ensemble learning techniques but differ in several key aspects, particularly in how they handle model training. In bagging:

- Each model in the ensemble is trained independently using different random subsets of the training data.
- Each model maintains its own separate set of parameters.
- The ensemble contains a finite number of models.
- The predictions of all models are averaged (or majority-voted) to obtain the final prediction.
- Bagging trains multiple models explicitly.

However, in dropout training:

- Dropout trains a single model that implicitly represents an ensemble.
- During training, a dropout mask deactivates a random subset of neurons, creating a "sub-model" of the original NN. Each sub-model is a variation of the full NN, containing only a subset of the neurons.
- The concept of parameter sharing in dropout training is essential to understanding why dropout can approximate (simulate) a bagged ensemble. All models share parameters, as each "model" is essentially a subset of the original parent NN. In the case of dropout, the models share parameters, with each model inheriting a different subset of parameters from the parent NN. This parameter sharing allows for representing an exponential number of models with a manageable amount of memory. In other words, instead of training separate models, dropout trains a large number of sub-models using a single NN with shared parameters.

- It is noteworthy that a different NN is used for every small mini-batch of training examples. Therefore, the number of NNs sampled is rather large and depends on the size of the training data set. In the dropout method, thousands of NNs are sampled with shared weights, and a tiny training data set is used to update the weights in each case. Even though a large number of NNs is sampled, the fraction of NNs sampled out of the base number of possibilities is still minuscule. The network is typically large enough that sampling all possible subnetworks would be impractical, even over the lifetime of the universe. For a network with $M$ non-output nodes, there are $2^M$ pruned networks. Each node has two states: active or pruned. This is like flipping a coin for each node, leading to $2^M$ combinations.
- Another way to understand the power of dropout is to realize that a unique NN is generated at each training step. Once you have run 10,000 training steps, you have essentially trained 10,000 different NNs (each with just one training instance). These NNs are obviously not independent because they share many of their weights, but they are nevertheless all different. The resulting NN can be seen as an averaging ensemble of all these smaller NNs.
- Imagine a classroom of 15 students (representing 15 thinned NNs). Each day, a unique student attends class, but not all of them at once. Each student learns differently based on their participation in class. Eventually, when it's time for an exam (inference), all 15 students are called upon to solve the problems collectively (ensemble averaging). Although not everyone participated in every lesson, they collectively learned a comprehensive solution.

You can think of bagging as building many distinct houses from scratch, each with different architectural plans. In contrast, dropout is like rearranging the same modular furniture to create different layouts, allowing for countless configurations while maintaining the shared structure.

To create a prediction for an unseen test instance using an ensemble of NNs, one effective approach is to predict the instance with each NN in the ensemble and then combine their predictions. A common strategy is to compute the geometric mean of the probabilities predicted by the different networks. However, this method has a significant drawback: the ensemble often contains a large number of NNs, leading to inefficiency in computation and storage. A key insight of the dropout method is that it is not necessary to evaluate the prediction on all ensemble components. Rather, one can perform forward propagation on only the base network (with no dropping) after re-scaling the weights. The basic idea is to multiply the weights going out of each unit with the probability of sampling that unit. So, if a neuron had a 50% chance of being dropped during training, its influence in the final decision-making (inference) is halved. This adjustment simulates the average effect of the dropout process during training but does it in a way that is much more efficient for making predictions since it doesn't require simulating multiple different dropout configurations. By using this approach, the expected output of that unit from a sampled network is captured. This rule is referred to as the weight scaling inference rule. Using this rule also ensures that the input going into a unit is also the same as the expected input that would occur in a sampled network (i.e., during inference, all neurons are used, but their outputs are scaled down proportionally to the dropout probability to approximate the ensemble prediction).

The following intuition draws an analogy between the random attendance of employees at a company and the concept of dropout in NNs. Would a company perform better if its employees were instructed to toss a coin each morning to decide whether or not to come to work? Perhaps! The company would have to adapt its organization. It couldn't rely on any single person to handle the coffee machine or other critical tasks, so these skills would need to be spread across several people. Employees would have to cooperate with many coworkers, not just a select few. As a result, the company would become more resilient. If one person quit, it wouldn't make much of a difference. While this idea might be questionable for companies, it works exceptionally well for NNs. Neurons trained with dropout cannot co-adapt with neighboring neurons; they must be as useful as possible on their own. Furthermore, they can't rely too heavily on just a few input neurons and instead must pay attention to all their inputs. Consequently, they become less sensitive to small variations in the inputs, resulting in a more robust network that generalizes better.

**Remarks:**

- A key difference from the connection sampling approach discussed in the previous section is that the weights of the various sampled networks are shared. Thus, dropout combines node sampling with weight sharing.
- Dropout prevents the network from relying too much on a small number of nodes. Instead, the network is forced to use all the nodes.
- Equivalently, dropout encourages the training process to spread the weights to multiple nodes instead of putting much weight on a few nodes. This makes the effect somewhat similar to $L_2$-regularization.
- Dropout mitigates co-adaptation - a behavior whereby a group of nodes in the network behave in a highly correlated fashion, emitting similar outputs most of the time. This means the network could retain only one of them with no significant loss of accuracy.
- Another significant advantage of dropout is that it does not significantly limit the type of model or training procedure that can be used. It works well with nearly any model that uses a distributed representation and can be trained with SGD and other training procedures. Several enhancements to SGD are useful for dropout networks as well, including momentum, annealed learning rates, and $L_2$ weight decay.
- Even if your network doesn't show obvious signs of overfitting to the training data, incorporating some dropout into the network can improve validation accuracy, particularly in the later epochs of training.
- Applying dropout to every hidden layer in your network may be excessive. If your network is fairly deep, it could be sufficient to apply dropout only to the later layers (since the earliest layers might be harmlessly identifying features). To test this, start by applying dropout only to the final hidden layer and see if this effectively reduces overfitting. If not, gradually add dropout to the next deepest layer, test again, and repeat this process as needed.
- If your network is struggling to reduce validation cost or to recapitulate low validation costs attained when less dropout was applied, then you've added too much dropout—pare it back! As with other hyperparameters, there is a Goldilocks zone for dropout, too.
- When determining how much dropout to apply to a given layer, each network behaves differently, so some experimentation is necessary. In practice, dropping out 20% to 50% of the hidden-layer neurons in machine vision applications tends to yield the highest validation accuracies. In natural language applications, where individual words and phrases carry significant meaning, dropping out a smaller proportion—between 20% and 30% of the neurons in a given hidden layer—usually proves optimal.
- Start with a moderate dropout rate (e.g., 25% or 30%) and increase or decrease based on validation accuracy and overfitting trends.
- Tune dropout rates individually for each layer because layers at different depths may require different levels of regularization.
- Dropout introduces noise to the training process by randomly dropping out neurons during each forward and backward pass. Each mini-batch effectively uses a different network configuration, resulting in more variability (noise) in the gradients. As a result, individual parameter updates are noisier, making the training less stable and potentially prolonging convergence.
- The error function, or loss, also becomes noisy due to dropout. Because different neurons are dropped in each batch, the model's predictions vary more across batches, leading to fluctuations in the loss value. This noise makes it challenging to confirm that the optimization algorithm is working correctly simply by observing a decreasing error function.
- Since dropout is a regularization method, it reduces the expressive power of the network. Therefore, one needs to use larger models and more units in order to gain the full advantages of dropout.

**A Practical Guide for Training Dropout Networks:**

- Network Size:
  Dropout randomly deactivates some units in a NN during training. For any unit in a layer, the probability of it being retained (i.e., remaining active) is $p$, where $0 < p \leq 1$. If a layer has $n$ hidden units and each unit has a probability $p$ of being retained, the expected number of units active during training is $pn$. This dynamic behavior helps prevent overfitting by reducing the ability of units to develop strong co-adaptations. When designing a NN with dropout, it's important to account for this reduction in the expected number of active units. The rule of thumb is that if $n$ units are optimal for a layer in a standard network, then a dropout layer should have approximately $n/p$ units to maintain equivalent capacity. Suppose a fully connected layer in a standard network has $100$ units, and we want to apply dropout with $p = 0.5$ (i.e., retain half of the units). To maintain equivalent capacity, the dropout layer should have 100/0.5=200 units. In each training iteration, approximately $100$ units will remain active after dropout is applied. This heuristic aids in determining the optimal number of hidden units for both convolutional and fully connected networks.

- Learning Rate and Momentum:
  Dropout introduces a significant amount of noise in the gradients compared to the standard training process (SGD) because the network is learning with only a subset of the units active during any given iteration. As a result, the gradients calculated have higher variance than in a standard NN. This requires adjustments to the learning rate and momentum to compensate for the increased noise and variance. The increased noise due to dropout leads to gradients partially canceling each other out. In order to make up for this, a dropout net should typically use 10-100 times the learning rate that was optimal for a standard neural net. The higher learning rate helps amplify the gradients, compensating for the noise introduced by dropout. This allows the model to converge faster and not get stuck in suboptimal regions due to noisy gradients. Suppose the optimal learning rate for a standard network is $0.001$. With dropout, the learning rate should be increased to around $0.01$ to $0.1$. The noise in gradients can lead to unstable updates and slower convergence. Another way to reduce the effect the noise is to use a high momentum. Higher momentum helps smooth out the noisy gradients by accumulating past gradient directions, thus reducing the effect of noise and providing a more consistent update direction. This improves convergence stability and accelerates training. Standard networks typically use a momentum value of $0.9$. For dropout networks, increase momentum to the range of $0.95$ to $0.99$. By increasing both the learning rate and momentum, dropout networks can counter the noise introduced by randomly deactivating units. This leads to more efficient learning and helps prevent the network from stagnating due to noisy gradients.
- Max-norm Regularization:
  Though large momentum and learning rate speed up learning, they sometimes cause the network weights to grow very large. To prevent this, we can use max-norm regularization. In max-norm regularization, the Euclidean norm of each weight vector is limited to a predefined maximum value.

  $$\|w_i\|_2 \leq c, \tag{7.29}$$

  where $w_i$ is weight vector associated with the $i$-th neuron. $\|w_i\|_2$ is Euclidean norm ($L_2$ norm) of the weight vector. $c$ is maximum allowable norm (hyperparameter). This constrains the norm of the vector of incoming weights at each hidden unit to be bound by a constant $c$. After each weight update, rescale any weight vector exceeding the maximum norm:

  $$w_i = \frac{w_i}{\|w_i\|_2} c \quad \text{if} \quad \|w_i\|_2 > c. \tag{7.30}$$

  By limiting the norm, max-norm regularization helps prevent individual weight vectors from growing too large and thus reduces overfitting. Typical values of $c$ range from 3 to 4.
- Dropout Rate:
  Dropout introduces an additional hyperparameter—the probability of retaining a unit, $p$. $1 - p$ is the probability of dropping a unit. This hyperparameter controls the intensity of dropout. When $p = 1$, no dropout is applied, while lower values of $p$ increase the amount of dropout. For hidden layers, the selection of $p$ is closely related to the choice of the number of hidden units, $n$. Lower $p$ requires a larger $n$, which can slow down training and lead to underfitting. Conversely, higher $p$ may not apply sufficient dropout to effectively prevent overfitting.
- In the years since, a wide range of stochastic techniques inspired by the original dropout method have been proposed for use with deep learning models. We use the term dropout methods to refer to them in general. They include dropconnect [215], standout [216], fast dropout [217], variational dropout [218], Monte Carlo dropout [219] and many others, and there are many comprehensive surveys available [220-223].

# CHAPTER 8

# ADVANCED ACTIVATION FUNCTIONS

The AFs play a very crucial role in NNs by learning the abstract features through non-linear transformations. Some common properties of the AFs are as follows [50]: a) it should add the non-linear curvature in the optimization landscape to improve the training convergence of the network; b) it should not increase the computational complexity of the model extensively; c) it should not hamper the gradient flow during training. Several AFs have been explored in recent years for deep learning to achieve the above-mentioned properties. This chapter delves into the various types of AFs, their properties, and taxonomy, providing an overview of the most widely used functions and their variants.

Sigmoid-Based AFs: In order to introduce non-linearity into the NNs, the Logistic Sigmoid, and Tanh AFs have been used in the early days. The firing of biological neurons was the motivation for using the Logistic Sigmoid and Tanh AFs with artificial neurons. The Logistic Sigmoid and Tanh AFs majorly suffer from vanishing gradients. Several improvements have been proposed based on the Logistic Sigmoid AF. The most common Sigmoid-based/related functions are Tanh, HardSigmoid, and HardTanh, Penalized Tanh, Soft-Root-Sign, and Sigmoid-Weighted Linear.

ReLU-Based AFs: The saturated output and increased complexity are the key limitations of the above-mentioned Logistic Sigmoid-based AFs. The ReLU has become the state-of-the-art AF due to its simplicity and improved performance. Various variants of ReLU have been investigated by tackling its drawbacks, such as non-utilization of negative values, limited non-linearity, and unbounded output. The most common ReLU-based/related functions are: Leaky Rectified Linear Unit, Parametric ReLU, Randomized ReLU, Random Translation ReLU, Elastic ReLU, Elastic Parametric ReLU, Linearized Sigmoidal Activation, Rectified Linear Tanh, Shifted ReLU, Displaced ReLU and Multi-bin Trainable Linear Unit.

ELU-Based AFs: The major problem faced by the Logistic Sigmoid-based AFs is with its saturated output for large positive and negative input. Similarly, the major problem with ReLU-based AFs is the under-utilization of negative values leading to a vanishing gradient. In order to cope up with these limitations the ELU-based AFs have been used in the literature. The ELU-based AF utilizes the negative values with the help of the exponential function. Several AFs have been introduced in the literature as ELU variants. The most common ELU-based/related functions are Scaled ELU, Parametric ELU, Rectified Exponential Unit, Parametric Rectified Exponential Unit, and Elastic ELU.

Non-Standard AFs: Non-standard AFs include those that combine multiple standard functions or operate on different principles. The common non-standard AFs are Maxout and Softmax.

Miscellaneous AFs: Such as Swish-based/related AFs (Swish, E-Swish, HardSwish), SoftPlus-based/related AFs (SoftPlus, SoftPlus Linear Unit, Mish), Probabilistic AF (Gaussian Error Linear Unit, and Symmetrical Gaussian Error Linear Unit).

Combining AFs: Most of the Sigmoid, Tanh, ReLU, and ELU-based AFs are designed manually which might not be able to exploit the data complexity. Combining AFs are the recent trends. The most common AF are Mixed, Gated, and Hierarchical AFs, Adaptive Piecewise Linear Units, Mexican ReLU, Look-up Table Unit, and Bi-Modal Derivative Sigmoidal AFs. Unlike traditional AFs such as ReLU, Sigmoid, or ELU, which have fixed functional forms, adaptive AFs learn their parameters from the data, allowing the network to adapt more flexibly to different tasks.

This chapter will explore each of these AFs in detail, providing mathematical definitions, properties, and practical applications in NN architectures. By understanding the strengths and limitations of various AFs, practitioners can make informed choices to optimize their models for specific tasks and datasets.

## 8.1 AF Properties and Taxonomy

The AFs are needed in order to increase the complexity of the NN, without them it would just be a linear sandwich of linear functions (multiplying linear functions = linear function). Although the NN becomes simpler, learning any complex task is impossible, and our model would be just a linear regression model. The choice of AF should be guided by the specific requirements and characteristics of the problem you are trying to solve, as well as empirical testing to determine which AF works best for your NN architecture and data. AFs used in NNs have several common properties that define their behavior and suitability for different tasks. The following are some of the common properties [50]:

- Non-Linearity: One of the primary properties of AFs is non-linearity. They introduce non-linear transformations to the input data. This non-linearity is crucial for NNs to model complex, non-linear relationships in data.
- Continuity: AFs are typically continuous functions, which means they do not have abrupt jumps or discontinuities. This continuity is essential for ensuring that the gradients can be calculated and used for gradient-based optimization during training.
- Differentiability: Many AFs are differentiable, which allows for the calculation of gradients during backpropagation. This property is vital for updating the weights of the NN using optimization algorithms like GD.
- Smoothness: Smooth AFs ensure a smooth gradient and help in avoiding issues during optimization.
- Range: Understanding the range of an AF is important when designing NN architectures. Unbounded AFs allow the output to range from negative to positive infinity. This can help capture a wide range of information and facilitate learning complex patterns. On the other hand, bounded AFs restrict the output to a specific range. This can be useful in scenarios where you want to ensure that the output remains within a specific range. The choice between unbounded and bounded AFs depends on the specific requirements of your NN and the task at hand. Unbounded functions are often preferred in hidden layers for their ability to capture diverse patterns, while bounded functions are commonly used in the output layer.
- Monotonicity: A function is monotonic if it is either entirely non-increasing or non-decreasing. Monotonic AFs are preferred in NNs because they ensure that the output of the model changes consistently with changes in input, making it easier to understand and optimize.
- Computational Efficiency: AFs with simpler mathematical formulations and fewer parameters can be easier to implement and computationally more efficient. This simplicity is valuable, especially when deploying models in real-world applications. Efficient computation of the function and its derivative is important for training large NNs. AFs with computationally simple derivatives are favored in practice.
- Memory Efficiency: AFs that require less memory for computation can be advantageous, especially when dealing with large-scale NNs or resource-constrained environments. Functions with simple calculations or that do not store a significant amount of intermediate values can be more memory-efficient.
- Parameterization: Some AFs have parameters that can be learned during training which can be advantageous in certain scenarios. Understanding how changes in these parameters affect the behavior of the function allows for more fine-grained control over the network's learning dynamics. However, it is crucial to balance this flexibility with the risk of overfitting to the training data.
- **Sparsity:** Some AFs encourage sparsity in the NN, which can lead to more efficient representations and reduced computational requirements.

The design of AFs is an extremely active area of research and does not yet have many definitive guiding theoretical principles [32]. Several AFs have been explored in recent years for deep learning to achieve the above-mentioned properties. It can be difficult to determine when to use which kind. Predicting in advance which will work best is usually impossible. The design process consists of trial and error, intuiting that a kind of hidden unit may work well, and then training a NN with that kind of hidden unit and evaluating its performance on a validation set.

A taxonomy of AFs was proposed in [224], and a survey was conducted in [225, 226]. The classification is primarily based on whether the shape of the AF can be modified during the training phase. There are three main categories:

1. Fixed-shape AFs: These are AFs with a fixed shape.

- Examples include classic AFs used in NN literature, such as Sigmoid, Tanh, and ReLU.
- ReLU function has become significant in literature, marking a turning point and contributing to improved NN performance.
- Further subcategories within fixed-shape AFs include:
  - ReLU function: This includes all functions belonging to the rectifier family, such as ReLU, LReLU, etc.
  - Classic AF: This includes functions that are not in the rectifier family, such as Sigmoid, Tanh, and step functions.

2. Trainable AFs: This category includes AFs whose shape is learned during the training phase. The idea behind this kind of function is to search for a good function shape using knowledge given by the training data.

Two families of trainable AFs are described:

- Parameterized standard functions: These are trainable functions derived from standard fixed AFs with the addition of a set of trainable parameters. Essentially, they are a parameterized version of a standard fixed function, and the parameter values are learned from data. However, the trained AF shape turns out to be very similar to its corresponding non-trainable version. In the end, the general function shape remains substantially bounded to assume the basic function(s) shape on which it has been built.
- Functions based on ensemble methods: These functions are defined by mixing several distinct functions. A common approach is to combine them linearly, meaning the final AFs are modeled as linear combinations of one-variable functions. These one-variable functions may have additional parameters.

3. Trainable non-standard neuron: These functions can be considered as a different type of computational neuron unit compared with the original computational neuron model.

There are several ways for the taxonomy of AFs. It is possible also to define the taxonomy of AFs based on their scientific applications [], where the output field states are real or complex-valued.

- The real-valued activations are well-known in the literature.
- The complex-valued AFs are another set of AFs whose output is a complex number. These AFs are often required due to the complex-valued output, which has many applications in science and engineering.
- The process of mapping continuous values from an infinite set to discrete finite values is known as quantization. Both real and complex-valued AFs can be quantized in order to reduce memory requirements. The output of the quantized activation is integers rather than floating point values.
- Just like real-valued AFs, both complex-valued as well as quantized activations can be made adaptive by introducing the tunable variables.

Before discussing AFs in detail, it is important to note that it can be challenging to make a comparison, in terms of network performances, among the AFs that have been proposed in literature, since the experiments are often conducted using different experimental setups. AFs play a crucial role in the training of NNs by introducing non-linearity. However, they are not the sole determinants of a network's performance.

- The architecture and hyperparameters of a NN, such as the number of neurons, layer arrangement, weight initialization, and hyperparameter values required by the learning algorithms are equally important. The choice of these parameters can significantly influence the network's learning ability and generalization.
- Differences in the pre-processing of datasets can have a substantial impact on the performance of a NN. Varying pre-processing techniques, such as normalization, augmentation, or handling missing data, can lead to different experimental outcomes.
- The choice of performance metrics is also critical. Different metrics may emphasize different aspects of model performance (e.g., accuracy, precision, etc.), and the relevance of these metrics depends on the specific task.
- The performance of AFs may vary depending on the nature of the task (classification, regression, etc.)

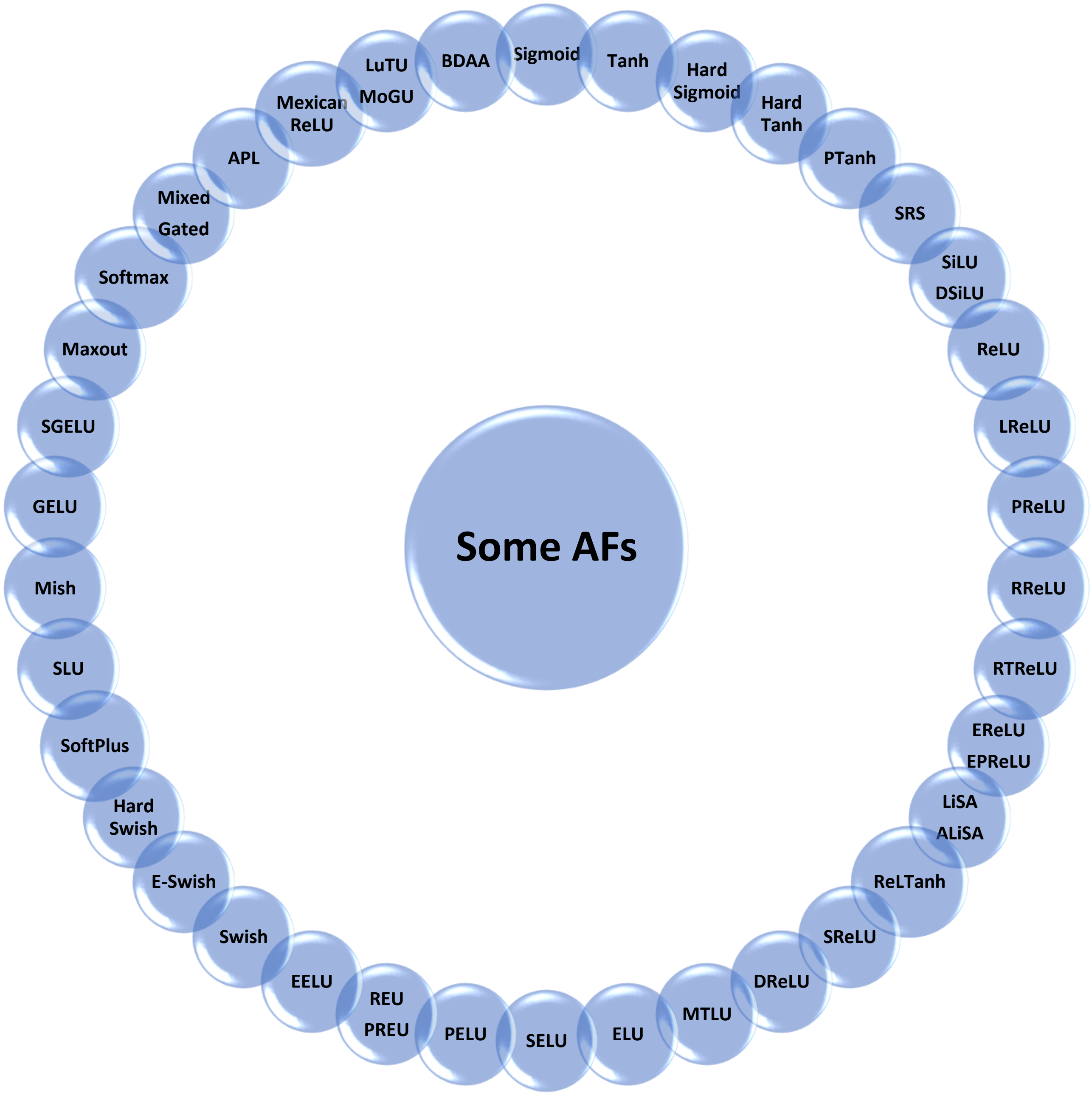

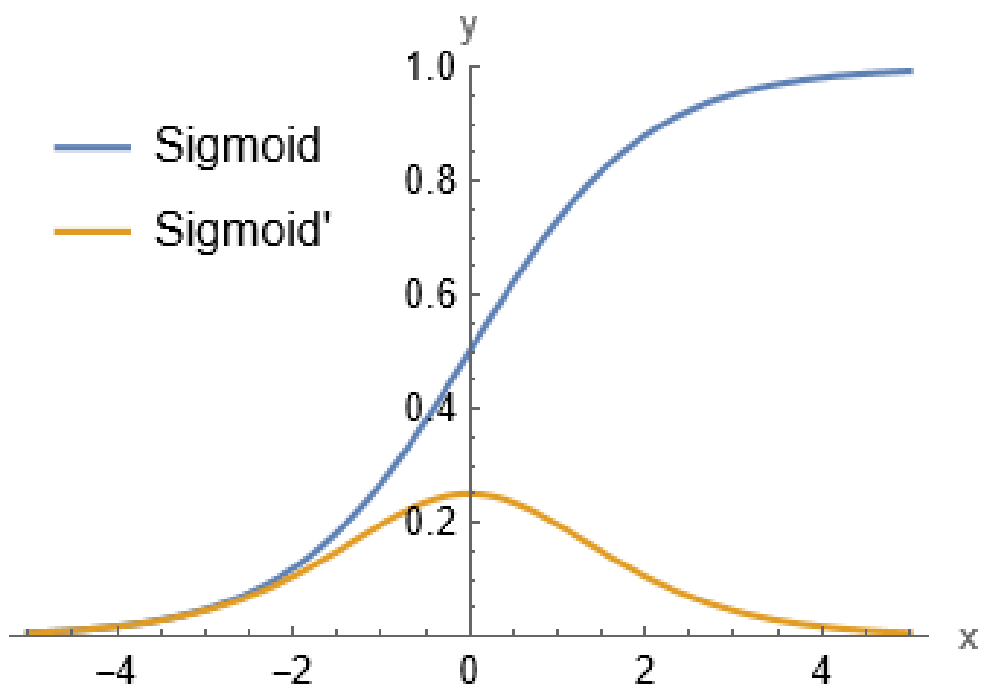

**Figure 8.1.** Sigmoid AF and its derivative. The plot illustrates the Sigmoid AF and its derivative, showcasing the characteristic $S$-shape of the Sigmoid function along with the bell-shaped curve of its derivative. The plot ranges from $x = -5$ to $x = 5$ with the function values constrained between 0 and 1.

## 8.2 Sigmoid-Based AFs

This section investigates the area of Sigmoid-based AFs, exploring the fundamental concepts behind the Sigmoid function and its variants. The Logistic-Sigmoid function is particularly popular in binary classification problems, where the goal is to predict outcomes as either 0 or 1. However, as the field of deep learning has evolved, researchers and practitioners have sought to enhance and modify the Logistic-Sigmoid function to address certain limitations and improve performance in different scenarios.

The section begins by providing a comprehensive overview of the standard Logistic-Sigmoid AF, examining its mathematical formulation and properties. Subsequently, it explores the motivations and considerations that led to the development of various variants of the Sigmoid function. These variants aim to overcome issues such as vanishing gradients and extend the applicability of Sigmoid-based activations to different types of NN architectures. This exploration will provide valuable knowledge to elevate your understanding of this crucial aspect of ANNs.

### 8.2.1 Sigmoid function

Richards [227] developed the Sigmoid AF family that spans the S-shaped curves like the Tanh function [228] and the Logistic-Sigmoid function. Subsequently, the first NN [229] used the Sigmoid AF for modeling biological neuron firing. The Logistic-Sigmoid AF is given by

$$\begin{aligned}\sigma_{\text{Logistic}}(x) = \frac{1}{1+e^{-x}} &= \frac{e^x}{e^x+1} \\ &= \frac{-1+(1+e^x)}{1+e^x} \\ &= 1-\sigma_{\text{Logistic}}(-x),\end{aligned} \tag{8.1.1}$$

$$\begin{aligned}\frac{d}{dx}\sigma_{\text{Logistic}}(x) = \frac{d}{dx}\frac{e^x}{1+e^x} &= \frac{e^x(1+e^x)-e^x e^x}{(1+e^x)^2} \\ &= \frac{e^x[(1+e^x)-e^x]}{(1+e^x)^2} \\ &= \frac{1}{(1+e^x)}\frac{e^x}{(1+e^x)} \\ &= \left[1-\frac{e^x}{(1+e^x)}\right]\frac{e^x}{(1+e^x)} \\ &= \left(1-\sigma_{\text{Logistic}}(x)\right)\sigma_{\text{Logistic}}(x).\end{aligned} \tag{8.1.2}$$

Figure 8.1 depicts the plot of the Logistic-Sigmoid function and its derivative.

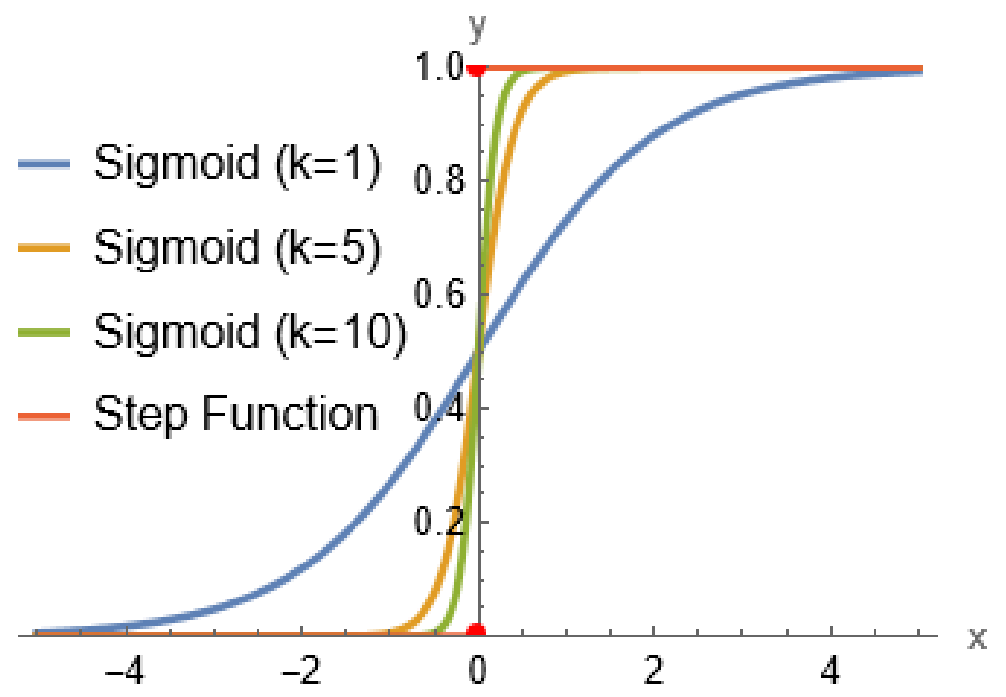

**Figure 8.2.** The plot compares the behavior of the step function and Sigmoid functions with varying steepness parameters ($k = 1$, $k = 5$, $k = 10$) over a range of $x$ values from $-5$ to 5. It illustrates how increasing the steepness parameter $k$ in the Sigmoid function makes it more closely approximate the sharp transition of the step function at $x = 0$. The Sigmoid curves transition smoothly from 0 to 1, with higher $k$ values resulting in steeper transitions. The step function, depicted as a binary switch, activates sharply at $x = 0$.

The Logistic-Sigmoid function has several important properties:

- The graph of the Logistic-Sigmoid function resembles the letter "$S$," with the curve starting near 0.5, rising gradually, and then approaching 1 as the input becomes more positive. Similarly, as the input becomes more negative, the function approaches 0. Hence, a Logistic-Sigmoid function is constrained by a pair of horizontal asymptotes as $x \to \pm\infty$, i.e., the function approaches 0 as $x$ approaches negative infinity and approaches 1 as $x$ approaches positive infinity.
- The Logistic-Sigmoid function is a monotonically increasing function, meaning that as the input $x$ increases, the output $\sigma_{\text{Logistic}}(x)$ also increases.
- The function is infinitely smooth.
- The Logistic-Sigmoid function is differentiable for all values of $x$, which makes it useful for optimization algorithms that require derivatives, such as GD. In general, a Logistic-Sigmoid function has a first derivative that is bell-shaped.
- Since the derivative quickly decays to zero away from $x = 0$, this AF can lead to slow convergence of the network while training.
- No zero-centered output.
- The Logistic-Sigmoid function involves the computation of the exponential function, $e^{-x}$, which can be computationally expensive. Exponential calculations are more resource-intensive compared to simpler operations like additions or multiplications. This can slow down the training process, especially in DNNs with a large number of parameters.
- The output of the Logistic-Sigmoid function is always in the range $(0,1)$, i.e., the function is bounded. This property makes it useful for modeling probabilities and binary classification problems. When $\sigma_{\text{Logistic}}$ is close to 1, it indicates a high probability that the input $x$ belongs to the positive class. Conversely, when $\sigma_{\text{Logistic}}$ is close to 0, it indicates a high probability that the input $x$ belongs to the negative class. The midpoint of the Logistic-Sigmoid curve is at $x = 0$, and $\sigma(0)$ equals 0.5. The predicted output is as follows:
$$\text{Output} = \begin{cases} 1, & \text{Result} > 0.5, \\ 0, & \text{Result} \leq 0.5. \end{cases} \quad (8.2)$$
Hence, if the resultant value is more than 0.5, the forecasted output is 1; otherwise, the forecasted output is 0.
- The Logistic-Sigmoid function can be seen as a continuous and smooth approximation of the step function (discontinuous at the threshold and lack of smoothness). The Logistic-Sigmoid function has a parameter that controls its steepness or slope, often denoted as $k$ in the function $\sigma_{\text{Logistic}}(x) = \frac{1}{1+e^{-kx}}$, Figure 8.2. The parameter $k$ adjusts how quickly the Sigmoid function transitions from 0 to 1 as $x$ passes through zero. As $k$ increases, the transition zone of the Sigmoid function, where it moves from near 0 to near 1, becomes narrower and sharper. In the limit, as $k \to \infty$, the Sigmoid function effectively becomes a step function. Mathematically, this can be expressed as:

$$\lim_{k\to\infty} \sigma_{\text{Logistic}}(x) = \begin{cases} 0, & x < 0, \\ 1, & x > 0. \end{cases} \tag{8.3}$$

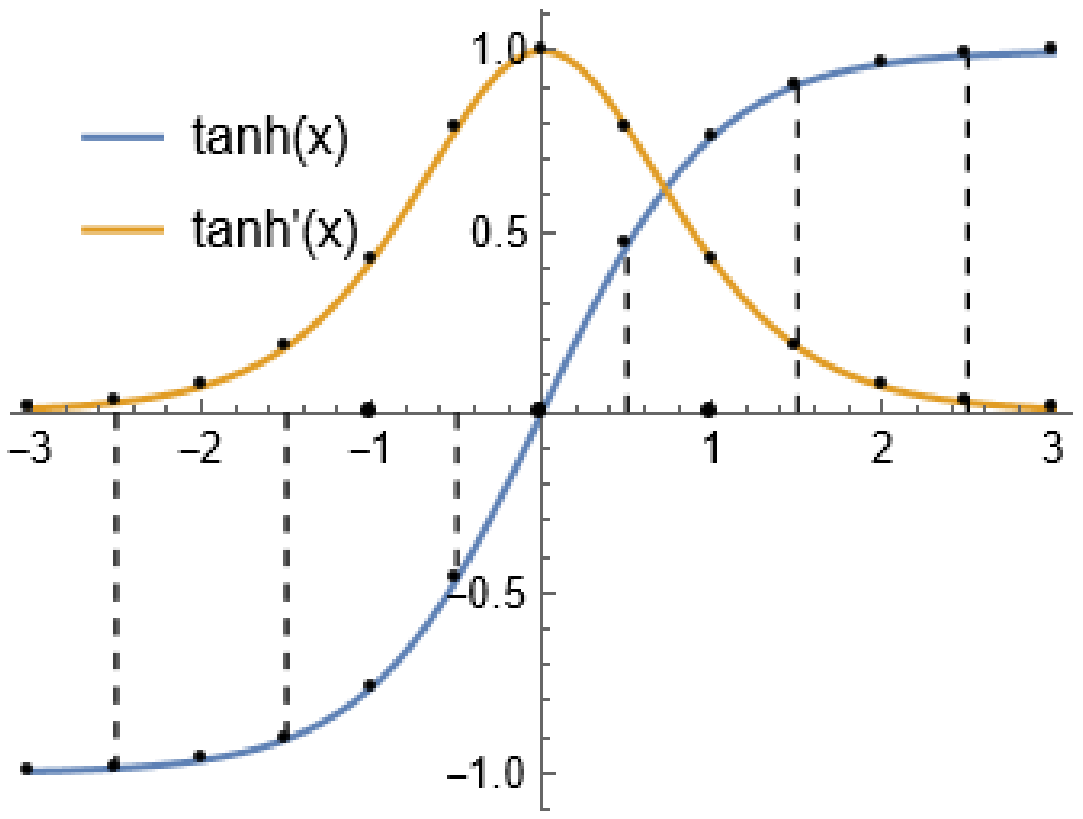

**Figure 8.3.** Tanh AF and its derivative. The plot presents the Tanh AF and its derivative plotted over a range of $x$ values from $-3$ to 3. The main plot shows the Tanh function, which smoothly transitions between $-1$ and 1, and its derivative, which illustrates the rate of change of Tanh and peaks at zero crossings of $x$.

As depicted in Figure 8.1 and demonstrated by executing $\sigma_{\text{Logistic}}(0.00001)$, near-0 inputs into the Logistic-Sigmoid function will lead it to return values near 0.5. Increasingly large positive inputs will result in values that approach 1. As an extreme example, an input of 10000 results in an output of 1.0. Moving more gradually with our inputs—this time in the negative direction—we obtain outputs that gently approach 0: As examples, $\sigma_{\text{Logistic}}(-1)= 0.2689$, while $\sigma_{\text{Logistic}}(-10) = 4.5398 \cdot 10^{-5}$. Any artificial neuron that features the Logistic-Sigmoid function as its AF is called a Logistic-Sigmoid neuron, and the advantage of these over the perceptron should now be tangible: Small, gradual changes in a given Logistic-Sigmoid neuron's parameters **w** or $b$ cause small, gradual changes in pre-activation $x = z$, thereby producing similarly gradual changes in the neuron's activation, $a$. Large negative or large positive values of $z$ illustrate an exception: At extreme $z$ values, Logistic-Sigmoid neurons—like perceptrons—will output 0's (when $z$ is negative) or 1's (when $z$ is positive). As with the perceptron, this means that subtle updates to the weights and biases during training will have little to no effect on the output, and thus learning will stall. This situation is neuron saturation.

### 8.2.2 Tanh AF

Standard Tanh or tangent hyperbolic function is basically the mathematically shifted kind of Logistic-Sigmoid AF. Both are analogous and can be derived using each other. The Tanh function has a shape similar to that of the Logistic-Sigmoid function, except that it is horizontally stretched and vertically translated/re-scaled to $[-1, 1]$:

$$\begin{aligned} \sigma_{\text{Tanh}}(x) &= \text{Tanh}(x) \\ &= \frac{e^x - e^{-x}}{e^x + e^{-x}} \\ &= \frac{1 - e^{-2x}}{1 + e^{-2x}} \\ &= \frac{2}{1 + e^{-2x}} - 1, \end{aligned} \tag{8.4.1}$$

$$\frac{d}{dx}\sigma_{\text{Tanh}}(x) = 1 - \text{Tanh}^2(x). \tag{8.4.2}$$

Figure 8.3 shows the curve for the standard Tanh AF and its derivative.

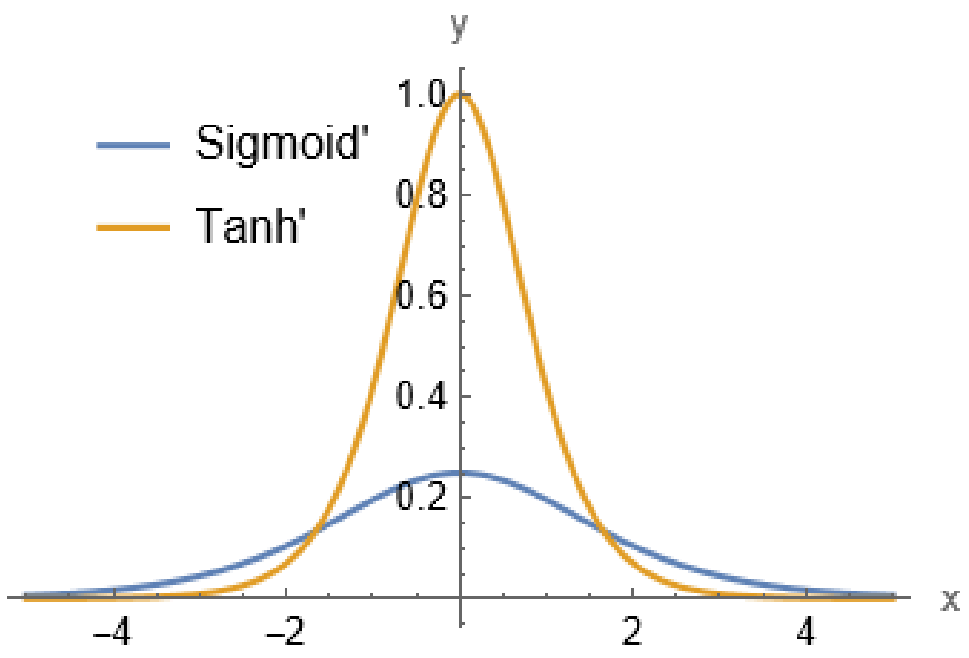

**Figure 8.4.** The figure displays the derivatives of the Sigmoid and hyperbolic tangent (Tanh) functions over a range of $x$ from $-5$ to $5$. The Sigmoid derivative, labeled as "Sigmoid'", shows a bell-shaped curve peaking at $x = 0$ and approaching zero as $x$ moves away from zero, reflecting its maximum rate of change at the midpoint of the Sigmoid function. The derivative of the Tanh function, labeled "Tanh'", also presents a similar bell-shaped curve but starts and ends at higher values compared to the Sigmoid derivative, highlighting differences in how these AFs propagate gradients in NNs. The plot effectively visualizes these derivatives, providing insights into the behavior of these commonly used AFs in deep learning.

The standard Tanh and Logistic-Sigmoid functions are related as follows:

$$\begin{aligned}
\sigma_{\text{Tanh}}(x) &= \frac{e^{x} - e^{-x}}{e^{x} + e^{-x}} \\
&= \frac{1 - e^{-2x}}{1 + e^{-2x}} \\
&= \frac{1}{1 + e^{-2x}} - \frac{e^{-2x}}{1 + e^{-2x}} \\
&= \sigma_{\text{Logistic}}(2x) - \frac{e^{-2x}}{1 + e^{-2x}} \\
&= \sigma_{\text{Logistic}}(2x) - \frac{e^{-2x} + 1 - 1}{1 + e^{-2x}} \\
&= \sigma_{\text{Logistic}}(2x) - \left(1 - \frac{1}{1 + e^{-2x}}\right) \\
&= \sigma_{\text{Logistic}}(2x) - \left(1 - \sigma_{\text{Logistic}}(2x)\right) \\
&= 2\sigma_{\text{Logistic}}(2x) - 1.
\end{aligned} \tag{8.5}$$

The Tanh function has the following properties:

- The Logistic-Sigmoid and the standard Tanh functions have been the historical tools of choice for incorporating nonlinearity in the NN.
- The Tanh function is infinitely smooth, differentiable, and monotonic, which is important for training NNs using gradient-based optimization algorithms like SGD.
- The range of the function is $[-1,1]$, i.e., the function is bounded. Note that it maps zeros input to zero while pushing positive (negative) inputs to $+1$ $(-1)$. This helps the network to keep its weights bounded and prevents the exploding gradient problem where the value of the gradients becomes very large.
- The Tanh function is symmetric around the origin, meaning that it outputs negative values for negative input values, and positive values for positive input values.
- Similar to the Logistic-Sigmoid function, the derivative of Tanh quickly decays to zero away from $x = 0$ and can thus lead to slow convergence while training networks.
- The negative $x$ inputs corresponding to negative $\sigma_{\text{Tanh}}(x)$ output, $x = 0$ corresponding to $\sigma_{\text{Tanh}}(0) = 0$, and positive $x$ corresponding to positive $\sigma_{\text{Tanh}}(x)$ output, the output from Tanh neurons tends to be centered near $0$. These $0$-centered outputs usually serve as the inputs $x$ to other artificial neurons in a network, and such 0-centered inputs make neuron saturation less likely, thereby enabling the entire network to learn more efficiently.

- An important difference between the Logistic-Sigmoid and the Tanh functions is the behavior of their gradient: $\frac{d}{dx}\sigma_{\text{Logistic}}(x) = \left(1 - \sigma_{\text{Logistic}}(x)\right)\sigma_{\text{Logistic}}(x)$, $\frac{d}{dx}\sigma_{\text{Tanh}}(x) = 1 - \text{Tanh}^2(x)$. Figure 8.4 shows the curve for the gradient of the Logistic-Sigmoid and the Tanh AFs. The Logistic-Sigmoid function has a slower rate of change than the Tanh function, which means that the derivative of the Logistic-Sigmoid function will be smaller than the derivative of the Tanh function for a given input. This can influence the convergence rate (speed of converge) of a NN trained with the Logistic-Sigmoid AF compared to one trained using the Tanh function.
- In other words, for Tanh AF the gradient is stronger as compared to the Logistic-Sigmoid function. When we are using these AFs in a NN, our data are usually centered around zero. So, we should focus our attention on the behavior of each gradient in the region near zero. We observe that the gradient of Tanh is four times greater than the gradient of the Logistic-Sigmoid function. This means that using the Tanh AF results in higher values of gradient during training and higher updates in the weights of the network. So, if we want strong gradients and big learning steps, we should use the Tanh AF.
- The limitation of the Tanh function is that it suffers from the vanishing gradient problem. This is usually a big problem in DNNs having many layers since the gradients can become extremely small by the time they reach the earlier layers. This leads to slow convergence and poor performance.
- The Logistic-Sigmoid function is often used in the output layer of a binary classification network, while the Tanh function is often used in the hidden layers of a network. The Tanh function is particularly useful when the input data is zero-centered or when negative values are meaningful in the context of the problem, as it can capture both positive and negative features in the data.
- The traditional AFs such as Logistic-Sigmoid and Tanh were used very extensively in the early days of NNs. However, these AFs had shown the hurdle to train the DNNs due to their saturated output. Several attempts have also been made to improve these AFs for different networks.

### 8.2.3 HardSigmoid and HardTanh

**Definition (Hard and Soft Saturation):** Let $h(x)$ be AF with derivative $h'(x)$, and $c$ be a constant. We say that $h(\cdot)$ right hard saturates when $x > c$ implies $h'(x) = 0$ and left hard saturates when $x < c$ implies $h'(x) = 0$, $\forall x$. We say that $h(\cdot)$ hard saturates (without qualification) if it is both left and right hard saturates.
An AF that saturates but achieves zero gradient only in the limit is said to be soft saturate.

We can construct hard-saturating versions of soft-saturating AFs [230] by taking a first-order Taylor expansion about zero and clipping the results to an appropriate range. For example, expanding Tanh and Sigmoid around 0, with $x \approx 0$, we obtain linearized functions $u^t$ and $u^s$ of Tanh and Sigmoid respectively:

$$\sigma_{\text{Sigmoid}}(x) \approx u^s(x) = 0.25x + 0.5, \tag{8.6}$$

$$\sigma_{\text{Tanh}}(x) \approx u^t(x) = x. \tag{8.7}$$

Clipping the linear approximations results in,

$$\sigma_{\text{HardSigmoid}}(x) = \max(\min(u^s(x), 1), 0), \tag{8.8}$$

$$\sigma_{\text{HardTanh}}(x) = \max(\min(u^t(x), 1), -1). \tag{8.9}$$

The motivation behind this construction is to introduce linear behavior around zero to allow gradients to flow easily when the unit is not saturated while providing a crisp decision in the saturated regime.

The hard Sigmoid is a non-smooth function used in place of a Sigmoid function. These retain the basic shape of a Sigmoid, rising from 0 to 1, but using simpler functions, especially piecewise linear functions or piecewise constant functions. Hard Sigmoid is also defined as

$$\begin{aligned}\sigma_{\text{HardSigmoid}}(x) &= \text{Clip}(0.2x + 0.5, 0, 1)\\ &= \max(0, \min(1, (0.2x + 0.5)))\\ &= \begin{cases} 0, & x < -2.5,\\ 1, & x > 2.5,\\ 0.2x + 0.5, & -2.5 \le x \le 2.5.\end{cases}\end{aligned} \tag{8.10}$$

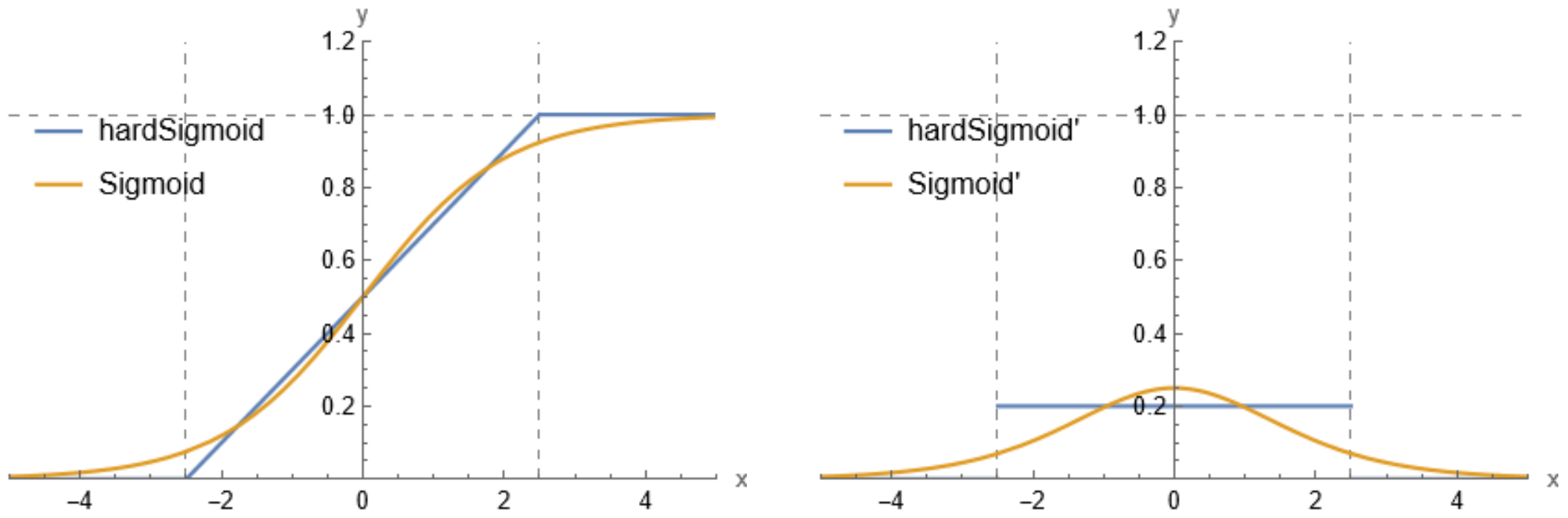

**Figure 8.5.** Left panel: The plot visually illustrates the Sigmoid AF, which exhibits a smooth, S-shaped curve, with the HardSigmoid AF, a piecewise linear approximation. HardSigmoid is defined with flat segments at the extremes and a linear segment in the middle, clearly delineated by grid lines at $x = -2.5$ and $x = 2.5$. Right panel: This plot illustrates the derivatives of both the Sigmoid and HardSigmoid functions across a range of $x$ from $-5$ to 5. The derivative of the Sigmoid function forms a bell-shaped curve, highlighting its smooth and continuous nature. In contrast, the derivative of the HardSigmoid is piecewise constant, zero for most of the range except in the middle transition region between $x = -2.5$ and $x = 2.5$, where it is constant (non-zero).

In this definition, the function is linear in the "transition" region between $-2.5$ and 2.5, and it quickly saturates to 0 or 1 outside of this range (see Figure 8.5).

One of the main advantages of the Hard Sigmoid function is its computational efficiency. Unlike the standard Sigmoid function, which involves exponentiation and can be computationally expensive, the Hard Sigmoid function relies on simple linear operations (addition and multiplication), making it faster to compute. This efficiency can be particularly beneficial in scenarios where computational resources are limited or when training large NNs.

The "HardTanh" AF is a piecewise linear approximation of the Tanh function, (see Figure 8.6). It is often used in NNs as an alternative to the Tanh. It is a cheaper and more computationally efficient version of the Tanh activation. HardTanh is defined as [230]

$$\begin{aligned}\sigma_{\text{HardTanh}}(x) &= \text{Clip}[x, -1, 1]\\ &= \max(-1, \min(1, x))\\ &= \begin{cases} -1, & x < -1,\\ 1, & x > 1,\\ x, & -1 \le x \le 1.\end{cases}\end{aligned} \tag{8.11}$$

The derivative can also be expressed in a piecewise functional form (see Figure 8.6):

$$\frac{\partial}{\partial x}\sigma_{\text{HardTanh}} = \begin{cases} 1, & -1 \le x \le 1,\\ 0, & \text{otherwise}.\end{cases} \tag{8.12}$$

The following are some additional points about the Hard Tanh AF:

- The Hard Tanh function is piecewise linear and consists of three linear segments: one for inputs less than $-1$, one for inputs between $-1$ and 1, and one for inputs greater than 1. This linearity simplifies the training process.
- The primary advantage of the Hard Tanh function is that it bounds the output within a specific range, typically between $-1$ and 1. This bounded output range can be beneficial in certain applications where the network needs to produce values within a specific interval, such as in reinforcement learning or when modeling probabilities.

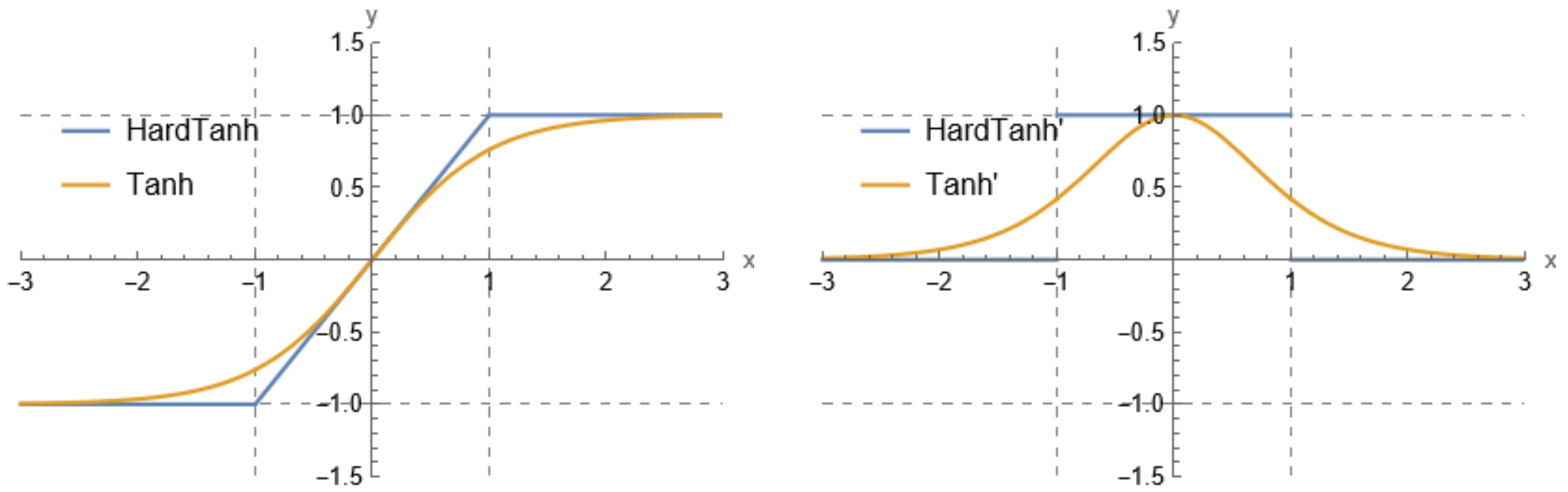

**Figure 8.6.** Left panel: The plot compares the Tanh AF with the HardTanh AF over a range of $x$ from $-3$ to $3$. Tanh smoothly transitions between $-1$ and $1$, while HardTanh abruptly clips values outside this range, creating a piecewise linear effect. The grid lines at $x = -1$ and $x = 1$ highlight these transition points. Right panel: The plot displays the derivatives of the Tanh and HardTanh AFs over the same range. The derivative of Tanh, "Tanh'", shows a bell-shaped curve, illustrating its non-zero gradient everywhere except the asymptotes. In contrast, "HardTanh'" has a derivative of zero almost everywhere except at $x = -1$ and $x = 1$, where it is undefined.

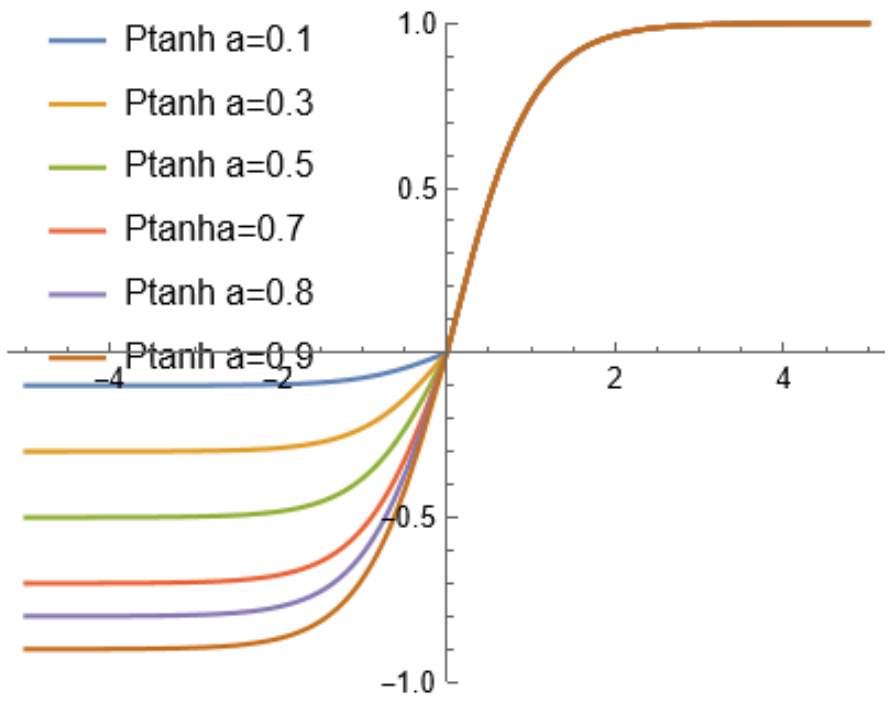

**Figure 8.7.** The figure illustrates the PTanh AF for different values of the parameter $a$ (0.1, 0.3, 0.5, 0.7, 0.8, and 0.9) over a range of $x$ from $-5$ to $5$. The PTanh function is piecewise-defined: it behaves like the standard $\text{Tanh}(x)$ function for non-negative $x$ values, and as $a \cdot \text{Tanh}(x)$ for negative $x$ values, introducing an adjustable scaling factor $a$. The plot shows how varying $a$ affects the curve's shape, particularly in the negative $x$ region. Lower values of $a$ result in flatter curves for $x < 0$, while higher values make the curve steeper and closer to the standard Tanh function.

- The gradient of the Hard Tanh function is well-behaved almost everywhere, except at the points where the function changes from one linear segment to another (i.e., at $-1$ and $1$).

### 8.2.4 Penalized Tanh AF

The Penalized Tanh (PTanh) is defined as,

$$\sigma_{\text{PTanh}}(x) = \begin{cases} \text{Tanh}(x), & x > 0, \\ a\,\text{Tanh}(x), & \text{otherwise}, \end{cases} \tag{8.13}$$

with the output range in $(-a, 1)$ where $a \in (0, 1)$.

In this definition: For positive values of $x$, the function uses the regular $\tanh(x)$ function. In this case, the Tanh function squashes the input values between 0 and 1. However, for non-positive values of $x$, the function applies a scaled version of the Tanh function with a scaling factor $a$. This means that for values of $x$ less than or equal to 0, the output is $a$ times the result of $\text{Tanh}(x)$. This allows you to control the behavior of the activation for negative values. In other words, $a$ is a hyperparameter that controls the scaling of the Tanh function for non-positive values of $x$. Figure 8.7 compares the PTanh for different values of $a$.

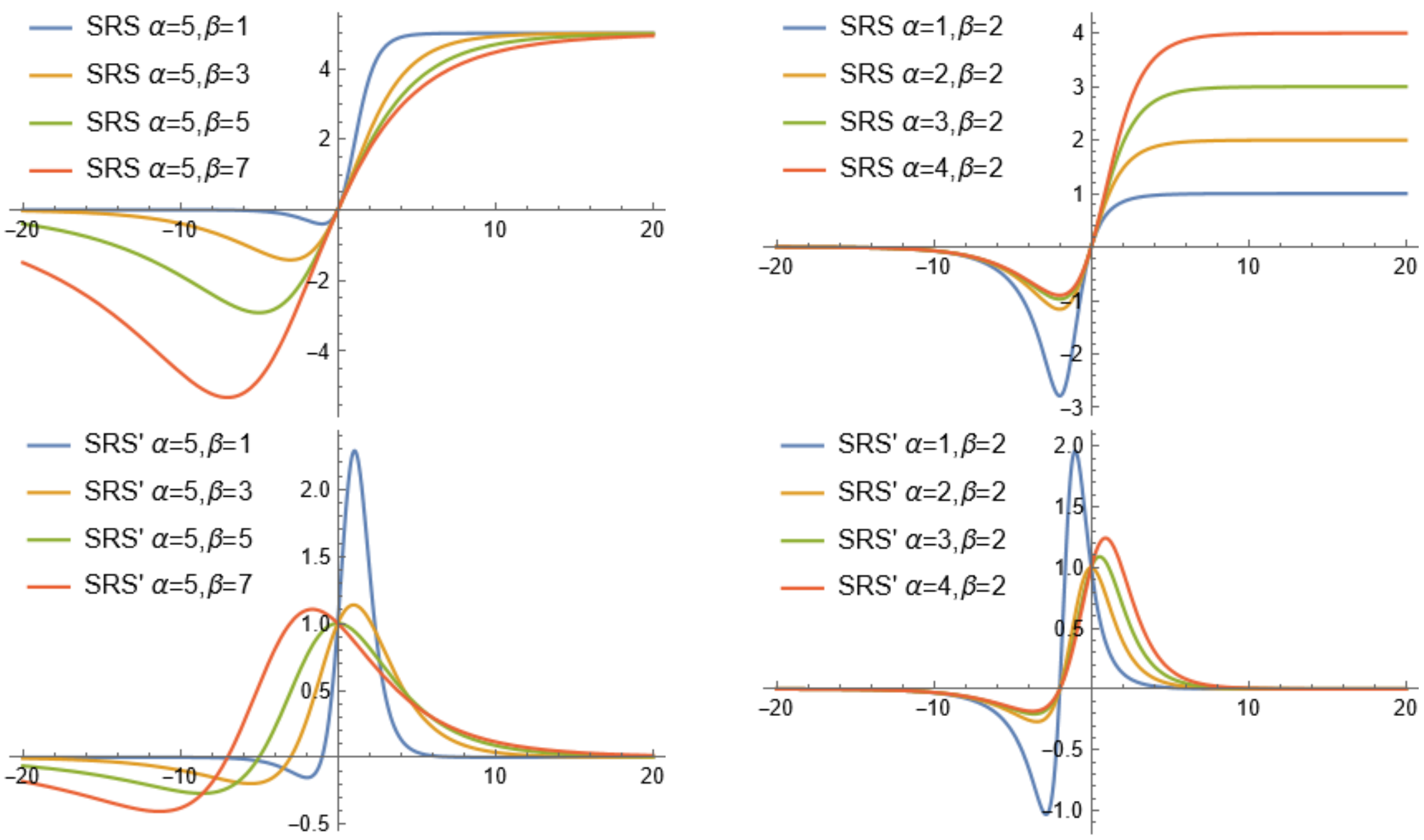

**Figure 8.8.** The figure consists of four plots illustrating the SRS AF and its derivatives for various parameter values over a range of $x$ from $-20$ to $20$. Top left panel: The first plot shows the AF for a fixed $\alpha = 5$ with different $\beta$ values $(1, 3, 5, 7)$, highlighting how $\beta$ affects the function's smoothness and steepness. Top right panel: The second plot presents the AF for a fixed $\beta = 2$ with varying $\alpha$ values $(1, 2, 3, 4)$, demonstrating the impact of $\alpha$ on the function's scaling and shape. Bottom left panel: The third plot displays the derivatives of the AF for $\alpha = 5$ and $\beta$ values $(1, 3, 5, 7)$, showing how the gradient changes across the input range. Bottom right panel: The fourth plot shows the derivatives for $\beta = 2$ and $\alpha$ values $(1, 2, 3, 4)$, illustrating the effect of $\alpha$ on the gradient behavior.

**Remarks:**

- PTanh saturates to $-a$ and $1$ when moving away from $0$. It is saturated outside its linear regime.
- $\sigma_{\text{PTanh}}(x)$ AF also suffers from the vanishing gradient problem.
- PTanh is comparable and even outperforms the state-of-the-art non-saturated functions including ReLU on convolution DNNs [231].
- The PTanh can be viewed as a saturated version of LReLU (next section). These two functions have similar values near $0$ since both functions share the same Taylor expansion up to the first order.
- This kind of piecewise AF can be useful in scenarios where you want to apply different transformations to the positive and non-positive regions of the input. The choice of $a$ would depend on the specific requirements of your problem and would likely require experimentation and tuning.

### 8.2.5 Soft-Root-Sign AF

An effective AF is required to have:

- Negative and positive values for controlling the mean toward zero to speed up learning,
- Saturation regions (derivatives approaching zero) to ensure a noise-robust state,
- A continuous-differential curve that helps with effective optimization and generalization.

Based on the above insights, the Soft-Root-Sign (SRS) AF [232] was defined as,

$$\sigma_{\text{SRS}}(x) = \frac{x}{\frac{x}{\alpha} + e^{-\frac{x}{\beta}}}, \tag{8.14.1}$$

where $\alpha$ and $\beta$ are a pair of trainable non-negative parameters. Figure 8.8 shows the graph of the $\sigma_{\text{SRS}}$ AF. The derivative of SRS is defined as

$$\frac{\partial}{\partial x}\sigma_{\text{SRS}}(x) = \frac{\left(1 + \frac{x}{\beta}\right) e^{-\frac{x}{\beta}}}{\left(\frac{x}{\alpha} + e^{-\frac{x}{\beta}}\right)^2}. \tag{8.14.2}$$

Figure 8.8 illustrates the first derivative of $\sigma_{\text{SRS}}(x)$, which gives nice continuity and effectiveness.

**Remarks:**

- SRS is smooth, non-monotonic, and bounded.
- As shown in Figure 8.8, the $\sigma_{\text{SRS}}(x)$ AF is bounded output with a range $[\frac{\alpha\,\beta}{\beta-\alpha e}, \alpha)$. Specifically, the minimum of $\sigma_{\text{SRS}}(x)$ is observed to be at $x = -\beta$ with a magnitude of $\frac{\alpha\,\beta}{\beta-\alpha e}$; and the maximum of $\sigma_{\text{SRS}}(x)$ is $\alpha$ when the network input $x \to +\infty$.
- The maximum value and slope of the function can be controlled by changing the parameters $\alpha$ and $\beta$, respectively. Through further setting $\alpha$ and $\beta$ as trainable, $\sigma_{\text{SRS}}(x)$ can not only control how fast the first derivative asymptotes to saturation, but also adaptively adjust the output to a suitable distribution, which avoids gradient-based problems and ensures fast as well as robust learning for multiple layers.
- The choice of AF is essential for building state-of-the-art NNs. At present, the most widely used AF with effectiveness is ReLU. However, ReLU has the problems of non-zero mean, negative missing, and unbounded output, thus it has potential disadvantages in the optimization process. In contrast to ReLU, SRS has a non-monotonic region when $x < 0$ which helps capture negative information and provides zero-mean property. Meanwhile, SRS is bounded output when $x > 0$ which avoids and rectifies the output distribution to be scattered in the non-negative real number space.
- Moreover, SRS adaptively adjusts a pair of independent trainable parameters to provide a zero-mean output, resulting in better generalization performance and faster learning speed.

### 8.2.6 Sigmoid-Weighted Linear Unit

The output of the Sigmoid function is multiplied by its input in the Sigmoid-weighted linear unit (SiLU) [233] as

$$\sigma_{\text{SiLU}}(x) = x\,\sigma_{\text{Sigmoid}}(x), \tag{8.15}$$

in the output range of $(-0.5, \infty)$.

**Remarks:**

- The $\sigma_{\text{SiLU}}(x)$ AF looks like a continuous and "undershooting" version of the ReLU AF (see Figure 8.9). For $x$-values of large magnitude, the activation of the SiLU is approximately equal to the activation of the ReLU, i.e., the activation is approximately equal to zero for large negative $x$-values and approximately equal to $x$ for large positive $x$-values.
- Unlike the ReLU (and other commonly used activation units such as Sigmoid and Tanh units), the activation of the SiLU is not monotonically increasing, meaning it does not strictly increase or decrease across its entire domain. This non-monotonic behavior has been suggested to improve learning dynamics in certain cases. It has a global minimum value of approximately $-0.28$ for $x \approx -1.28$.
- The SiLU function is smooth and differentiable everywhere. This smoothness can be beneficial for gradient-based optimization algorithms used in training NNs, such as SGD.
- SiLU has been proposed as a way to mitigate the vanishing gradient problem. SiLU's non-linearity and the fact that it allows information to flow through the network more freely than traditional AFs like Sigmoid or hyperbolic tangent can help alleviate this issue.

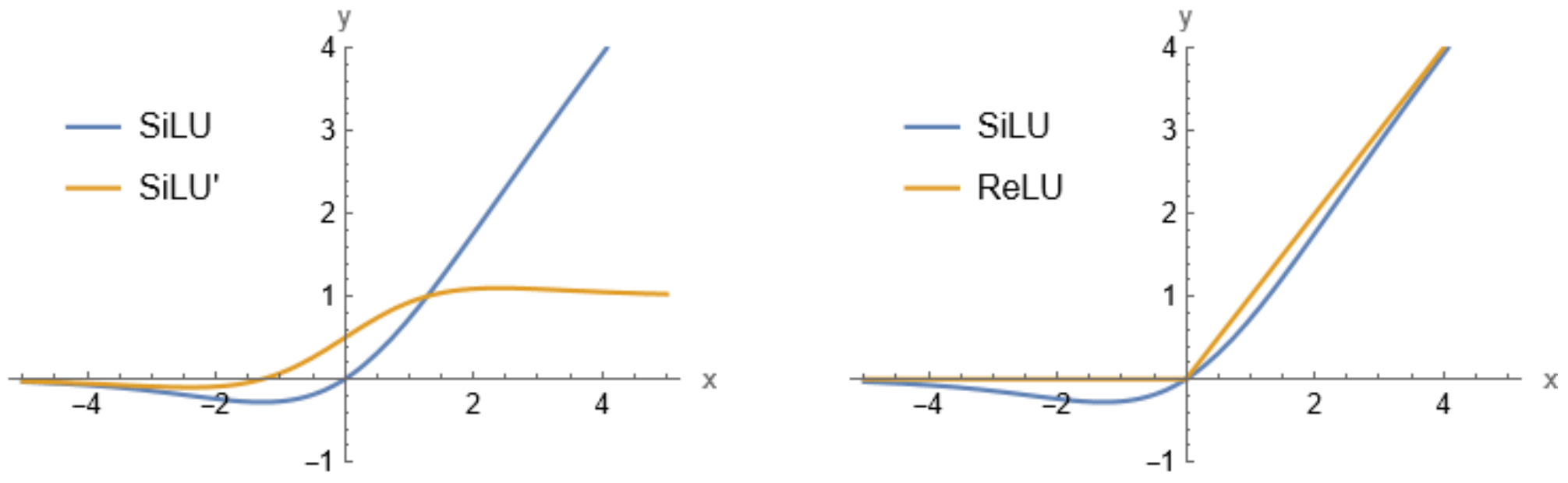

**Figure 8.9.** Left panel: This plot shows the SiLU AF and its derivative. The SiLU function smoothly transitions between linear behavior for large positive values and exponential decay for negative values. The derivative curve demonstrates how the gradient varies, providing insights into the function's smooth gradient properties. Right panel: This plot compares the SiLU function with the ReLU function. While ReLU outputs zero for negative inputs and increases linearly for positive inputs, SiLU offers a smooth, continuous alternative that avoids the abrupt transition at zero, blending linearity and Sigmoid behavior.

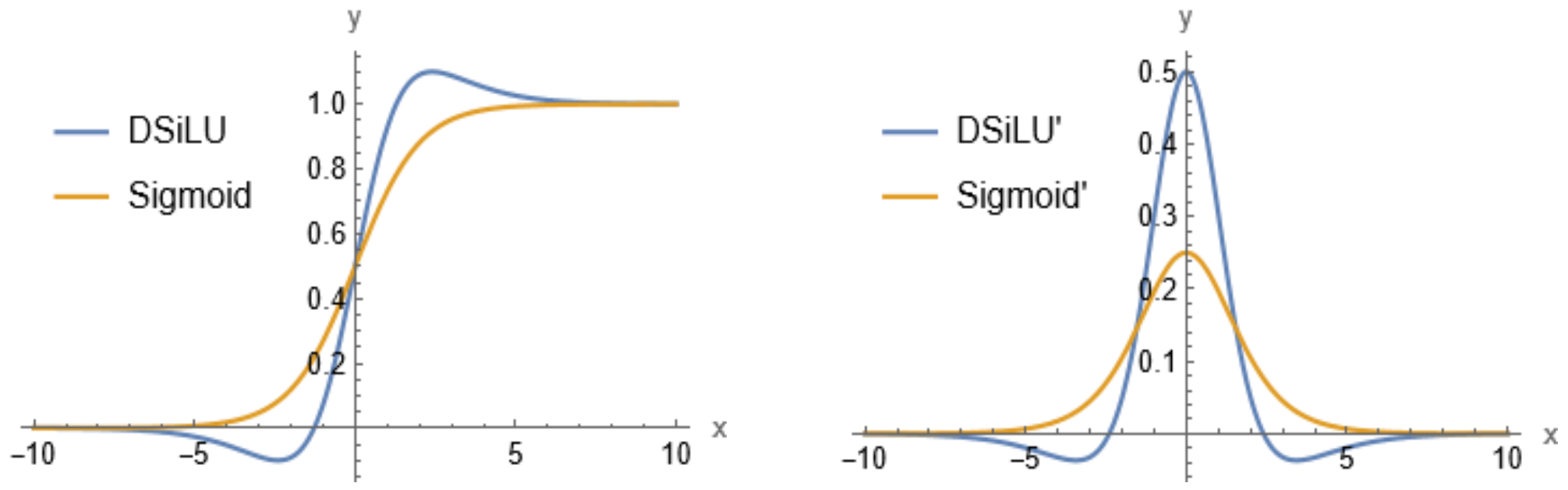

**Figure 8.10.** Left panel: This plot compares the derivative of the SiLU function (DSiLU) with the Logistic Sigmoid function itself. Right panel: This plot shows the second derivative of SiLU (DSiLU') compared with the first derivative of the Logistic Sigmoid function. The curves illustrate the more complex gradient behavior of DSiLU, providing a deeper understanding of its smooth and adaptive nature compared to the simpler gradient of the Sigmoid function.

- SiLU has been compared with other popular AFs like ReLU and has shown competitive performance in various scenarios [233]. It has been reported to achieve similar or better accuracy in certain deep-learning tasks.

The Derivative SiLU (DSiLU) AF is computed by the derivative of the SiLU [233]:

$$\sigma_{\text{DSiLU}}(x) = \sigma_{\text{Sigmoid}}(x)\left[1 + x\left(1 - \sigma_{\text{Sigmoid}}(x)\right)\right]. \tag{8.16}$$

The AF of the $\sigma_{\text{DSiLU}}(x)$ looks like a steeper and "overshooting" Sigmoid function (see Figure 8.10). The DSiLU has a maximum value of approximately 1.1 and a minimum value of approximately $-0.1$ for $x \approx \pm 2.4$.

Note that, most of the variants of Sigmoid/Tanh AFs have tried to overcome the vanishing gradient issue. However, this issue is still present in most of these AFs.

## 8.3 ReLU Based AFs

In the realm of NNs, the ReLU AF stands as a fundamental AF known for its ability to efficiently handle the vanishing gradient problem and expedite learning through non-saturation properties. However, the need for variant AFs arises from the shortcomings of standard ReLU, such as the "dying ReLU" problem, where neurons can become inactive and stop learning during training. Additionally, the unbounded nature of ReLU can lead to issues like exploding gradients, making it challenging to optimize DNNs effectively. These challenges prompt researchers to explore modifications and alternatives to enhance the performance and stability of NNs.

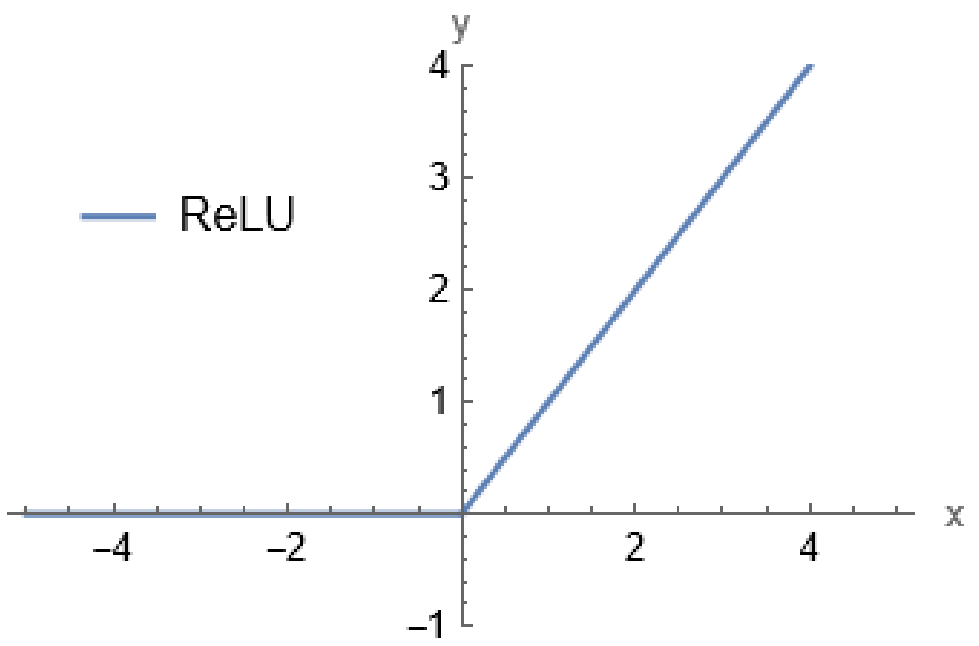

**Figure 8.11.** The plot illustrates the ReLU AF plotted over a range of $x$ from $-5$ to $5$. ReLU is defined as $\max(0, x)$, which results in a graph where all negative values are clamped to zero, and positive values remain unchanged. The function transitions sharply from 0 to a linear increase at $x = 0$.

One such adaptation is the LReLU, which mitigates the dying ReLU effect by incorporating a fixed negative slope, ensuring the preservation of small negative signals through non-zero gradients. Taking the concept further, the PReLU emerges as a dynamic alternative, allowing the learning of negative slopes from training samples. The flexibility of PReLU lies in its ability to share parameters across all channels or assign them independently to each channel within a hidden layer. The exploration doesn't stop there, as the introduction of randomness into ReLU-based architectures introduces new dimensions of adaptability. RReLU adds an element of unpredictability by altering the slope of the negative part during training, drawing from a uniform distribution.

This section provides an overview of the key issues associated with standard ReLU and introduces the reader to the main objectives of exploring its variants. Each variant introduces unique characteristics, advantages, and potential drawbacks, contributing to the rich landscape of AFs in modern deep-learning.

### 8.3.1 ReLU AF

The ReLU AF is an AF defined as the positive part of its argument [234]:

$$\begin{aligned}\sigma_{\mathrm{ReLU}}(x) &= \max(0, x) \\ &= \frac{x + |x|}{2} \\ &= \begin{cases} x, & x \geq 0, \\ 0, & x < 0, \end{cases}\end{aligned} \tag{8.17.1}$$

$$\frac{\partial}{\partial x}\sigma_{\mathrm{ReLU}}(x) = \begin{cases} 1, & x > 0, \\ 0, & x < 0, \end{cases} \tag{8.17.2}$$

where $x$ is the input to a neuron. This is also known as a ramp function. This AF was introduced by Kunihiko Fukushima in 1969 [234]. In 2011 [235], it was found to enable better training of deeper networks, compared to the widely used AFs before 2011, e.g., the Sigmoid and the hyperbolic tangent. The ReLU function is one of the simplest functions to imagine that is nonlinear. That is, like the Sigmoid and Tanh functions, its output does not vary uniformly linearly across all values of $x$. The ReLU is in essence two distinct linear functions combined (one at negative $x$ values returning 0, and the other at positive $x$ values returning $x$, as is visible in Figure 8.11), to form a straightforward, nonlinear function overall. It provides output as $x$ if $x$ is positive and 0 if the value of $x$ is negative.

Some features of the ReLU function are:

- ReLU introduces non-linearity to the network, allowing it to model complex, non-linear relationships in the data. This is crucial for the network's ability to learn and represent a wide range of functions.

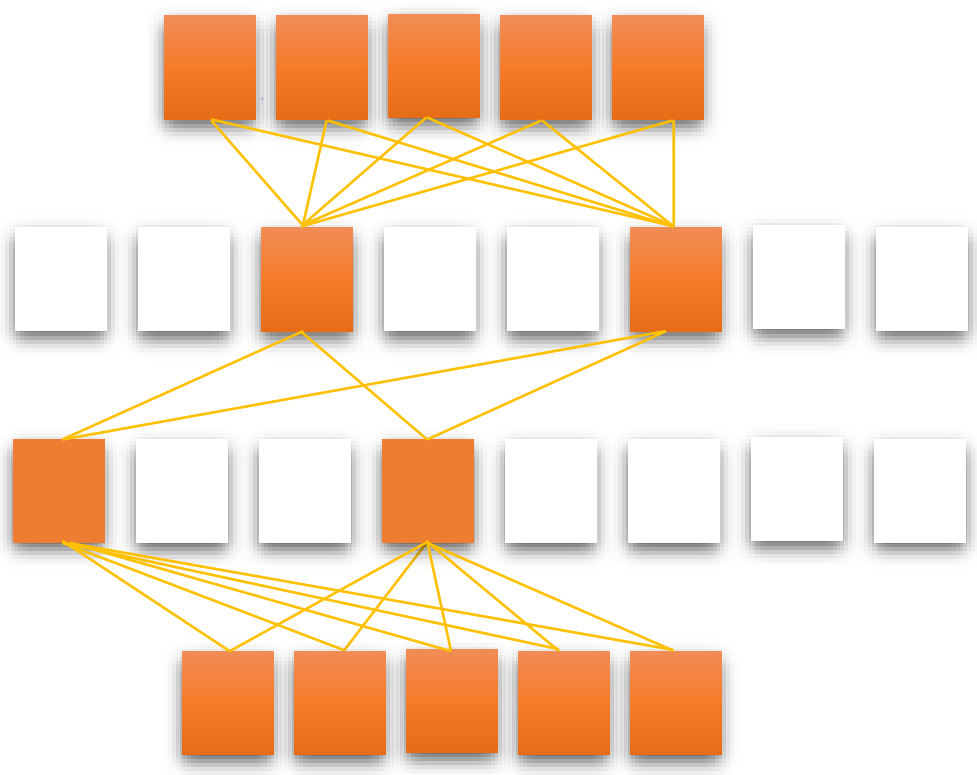

**Figure 8.12.** In a network comprised of ReLU units, the process of activation and gradient propagation occurs sparsely. This means that only a subset of neurons, determined by the input, becomes active, and the computational load remains linear within this selected subset.

- Unlike the linear and Sigmoid AFs, which are implemented at the output layer of the NN, ReLU is implemented at the hidden layers of the NN.
- ReLU involves simpler mathematical operations compared to Tanh and Sigmoid, thereby boosting its computational performance further. It is computationally efficient and easy to implement, making it a popular choice in deep learning models.
- The range of the function is $[0, \infty)$.
- ReLU does not saturate (for positive input values).
- ReLU does not have the problem of vanishing gradient for positive inputs, which is suffered by other AFs like Tanh or Sigmoid.
- Converges much faster than Sigmoid/Tanh in practice.
- ReLU is unbounded AF. One may thus need to use a regularizer to prevent potential numerical problems.
- ReLU does not have zero-centered output. The hyperbolic tangent absolute value non-linearity $|\mathrm{Tanh}(x)|$ enforces sign symmetry. A $\mathrm{Tanh}(x)$ non-linearity enforces sign antisymmetry. ReLU is one-sided and therefore does not enforce a sign symmetry or antisymmetry, instead, the response to the opposite of an excitatory input pattern is 0 (no response). However, we can obtain symmetry or antisymmetry by combining two rectifier units sharing parameters.
- In order to efficiently represent symmetric/antisymmetric behavior in the data, a ReLU network would need twice as many hidden units as a network of symmetric/antisymmetric AFs.
- The function is continuous, while its derivative will be piecewise constant with a jump at $x = 0$. The second derivative will be a Dirac function concentrated at $x = 0$. In other words, the higher-order derivates (greater than 1) are not well-defined.

### Network Sparse Representation

Studies on brain energy expense suggest that neurons encode information in a sparse and distributed way, estimating the percentage of neurons active at the same time to be between 1 and 4%. This corresponds to a trade-off between the richness of representation and small action potential energy expenditure. Without additional regularization, such as an $L_1$ penalty, ordinary FFNNs do not have this property. For example, the Sigmoid activation has a steady state regime around $1/2$, therefore, after initializing with small weights, all neurons fire at half their saturation regime. This is biologically implausible and hurts gradient-based optimization. ReLU units are sparsely activated, which can lead to more efficient training and generalization, meaning that they are active (output a non-zero value) for positive inputs and inactive (output zero) for negative inputs, see Figure 8.12. This can be seen as a form of regularization that helps prevent overfitting in some cases. For example, after uniform initialization of the weights, around 50% of hidden units' continuous output values are real zeros, and this fraction can easily increase with sparsity-inducing regularization.

As illustrated in Figure 8.12, the only non-linearity in the network comes from the path selection associated with individual neurons being active or not. For a given input only a subset of neurons is active. Computation is linear on this subset: once this subset of neurons is selected, the output is a linear function of the input (although a large enough change can trigger a discrete change of the active set of neurons). The function computed by each neuron or by the network output in terms of the network input is thus linear by parts. We can see the model as an exponential number of linear models that share parameters [236]. Because of this linearity, gradients flow well on the active paths of neurons (there is no gradient vanishing effect due to activation non-linearities of Sigmoid or Tanh units), and mathematical investigation is easier.

### Dying ReLU problem

ReLU neurons can sometimes be pushed into states in which they become inactive for essentially all inputs. Neurons that output zero for all inputs are considered "dead" and do not contribute to the learning process. This can happen when the input to a ReLU neuron is consistently negative, causing the gradient during backpropagation to be zero, preventing weight updates. In this state, no gradients flow backward through the neuron, and so the neuron becomes stuck in a perpetually inactive state and "dies". This is a form of the vanishing gradient problem. In some cases, large numbers of neurons in a network can become stuck in dead states, effectively decreasing the model capacity. The "dying ReLU" problem can be exacerbated by high learning rates and large negative biases. Let us examine how each of these factors can contribute to the issue:

High Learning Rates:

- In NNs, the weights are updated using the following equation:

$$\left(\mathrm{W}_{ij}^{(k)}\right)_{new} = \left(\mathrm{W}_{ij}^{(k)}\right)_{old} - \alpha \frac{\partial \mathcal{L}}{\partial \mathrm{W}_{ij}^{(k)}}. \tag{8.18}$$

  When you use a high learning rate, $\alpha$, in training a NN, the weight updates during backpropagation are more substantial, and this can lead to rapid changes in the model's parameters. i.e., if the $\alpha$ is too high when setting the learning rate, the new weights can have a negative range.
- If a ReLU neuron's output is pushed into a negative region (i.e., below zero) due to the weight updates (these negative values become the new inputs to the ReLU), it may become inactive (outputting zero) for many inputs, effectively leading to the dying ReLU problem.
- To address this, it's common to reduce the learning rate or apply techniques like learning rate scheduling, which gradually reduces the learning rate as training progresses. Lower learning rates allow for more stable weight updates and can help prevent the dying ReLU problem.

Large Negative Bias:

- Neurons in a NN often have biases associated with them. A large negative bias term can shift the entire input to a neuron to the left on the ReLU AF's curve. This can make it more likely for the neuron to be pushed into a region where it always outputs zero (i.e., if a neuron's bias is set too negatively, it effectively forces the neuron to be inactive for most inputs, and it contributes to the dying ReLU problem).
- To mitigate the impact of a large negative bias, you can either initialize biases closer to zero or use AFs like LReLU, PReLU, or ELU, which are less sensitive to bias values far from zero.

In practice, it is often recommended to fine-tune the learning rate, use proper weight initialization techniques (e.g., He initialization), and avoid large negative biases to ensure smoother training and avoid the dying ReLU problem. Moreover, techniques like BN can also help stabilize activations during training and mitigate the issues associated with high learning rates and biases.

### Understanding the Derivative of ReLU

Some of the AF units are not actually differentiable at some input points. For example, the ReLU AF is not differentiable at $x = 0$. This may seem like it invalidates AF for use with a gradient-based learning algorithm. In practice, GD still performs well enough for these models to be used for machine learning tasks.

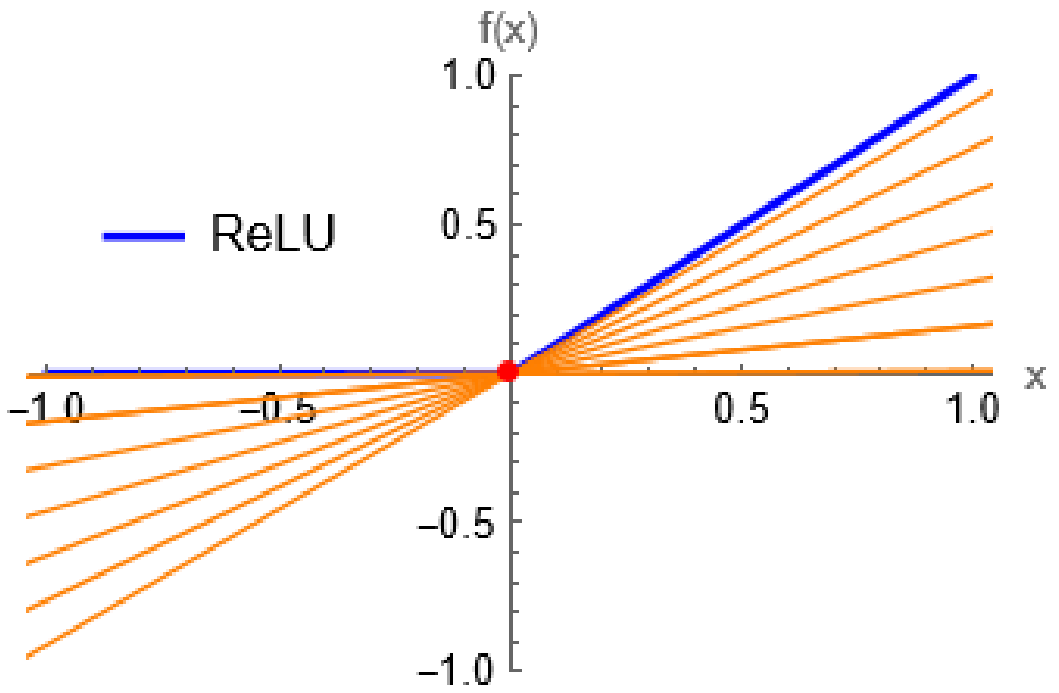

**Figure 8.13.** The figure showcases the ReLU function plotted over a range of $x$ from $-1$ to 1, with the function $\text{ReLU}(x) = \max(0, x)$. It highlights the ReLU function in blue, which sharply transitions from 0 to a linear increase at $x = 0$. Additionally, the plot includes seven possible slopes, visualized in orange, representing different elements of the subdifferential at $x = 0$. These lines illustrate the concept that at the point of non-differentiability (where $x = 0$), multiple tangent slopes could potentially describe the function. A red point marks the exact location $x = 0$ on the function to draw attention to this critical point of non-differentiability.

ReLU is not differentiable at $x = 0$. In mathematical terms, it means that the derivative of the ReLU function is undefined at $x = 0$. This can be seen when you consider the slope of the function at $x = 0$. The slope on the left side is 0 (since the function is flat at that point), and the slope on the right side is 1 (since the function is a straight line with a slope of 1 for $x > 0$). So, the function has a sharp corner at $x = 0$, and the derivative does not exist at this point.

Now, when it comes to gradient-based learning algorithms, like SGD, backpropagation, and various optimization techniques used for training NNs, the differentiability of the AFs plays a crucial role. The gradient of an AF is used to compute the gradient of the loss function with respect to the parameters of the model. This gradient is essential for adjusting the weights of the model during training to minimize the loss and improve the performance of the model. In the case of ReLU, since it is not differentiable at $x = 0$, you might wonder how gradient-based learning algorithms work with it. In practice, a common approach is to consider the subderivative or subgradient of ReLU at $x = 0$. The subgradient is a generalization of the derivative that can handle non-differentiable points. To understand the subgradient, let us delve into the mathematics and properties of the ReLU function and its subdifferential. At $x = 0$, the ReLU function has a sharp corner, which is the point of non-differentiability. The subdifferential of a function at a point is a set of slopes or gradients that describe the behavior of the function around that point, see Figure 8.13. In the case of the ReLU function at $x = 0$, the subdifferential contains all possible values between 0 and 1 because there is a range of possible slopes at this point. Mathematically, the subdifferential of the ReLU function at $x = 0$ can be expressed as:

$$\partial f(0) = [0,1]. \tag{8.19}$$

Here, the square brackets denote a closed interval, meaning that the subdifferential includes both 0 and 1. This means that any value between 0 and 1 is a valid subgradient for the ReLU function at $x = 0$. In practice, the subgradient can be any value between 0 and 1, and the choice of a specific subgradient value is often based on the specific requirements of the optimization algorithm being used.

Common choices for the subgradient at $x = 0$ include:

- Subgradient = 0: This choice simplifies the calculation and is often used in practice. It effectively means that the ReLU function is treated as a straight line with a slope of 0 for $x \leq 0$. This makes the derivative 0 at $x = 0$, and the subgradient is also set to 0. In certain platforms of deep learning, when $x = 0$, the derivative of ReLU with respect to $x$ is computed as 0. Their justification is that we should favor more sparse outputs for a better training experience.
- Subgradient = 1: This choice also simplifies the calculation. It assumes that the ReLU function is a straight line with a slope of 1 for $x \geq 0$, treating it as if it were a continuous function with a derivative of 1 at $x = 0$.

- Some would choose the exact value of 0.5 as the gradient of ReLU with respect to $x$ at $x = 0$. Because the expected value of the sub-gradient over an infinite number of sub-gradients is 0.5
- Random subgradient: In some cases, for added robustness or exploration in optimization algorithms, a random subgradient within the interval $[0,1]$ may be chosen. This can introduce some stochasticity into the training process.

The choice of subgradient at $x = 0$ can impact the convergence and behavior of gradient-based optimization algorithms. It is often chosen based on the specific problem and the needs of the training process. Regardless of the choice, it allows gradient-based optimization algorithms to continue making updates to the weights of the model, even in the presence of non-differentiable points like $x = 0$ in the ReLU function. So, while ReLU is not differentiable at $x = 0$, it can still be used in gradient-based learning algorithms by considering the subgradient or by approximating it as 0 or 1. This has proven to be effective in training DNNs and has contributed to the widespread use of ReLU in modern deep-learning architectures.

Moreover, hidden units that are not differentiable are usually nondifferentiable at only a small number of points. The functions used in the context of NNs usually have defined left derivatives and defined right derivatives. In the case of ReLU AF, the left derivative at $x = 0$ is 0, and the right derivative is 1. Software implementations of NN training usually return one of the one-sided derivatives rather than reporting that the derivative is undefined or raising an error. Moreover, this may be heuristically justified by observing that gradient-based optimization on a digital computer is subject to numerical error anyway. When a function is asked to evaluate $\sigma_{\text{ReLU}}(0)$, it is very unlikely that the underlying value truly was 0. Instead, it was likely to be some small value $\epsilon$ that was rounded to 0. For a given neuron with ReLU AF, the chances of the pre-activation value $x$ becoming exactly 0 is infinitesimally low (0.0000000000 is seldom reached).

### 8.3.2 Leaky Rectified Linear Unit

The Leaky Rectified Linear Unit (LReLU) [237] is defined as,

$$\sigma_{\text{LReLU}}(x) = \begin{cases} x, & x \geq 0, \\ \alpha x, & x < 0, \end{cases} \tag{8.20.1}$$

$$\frac{\partial}{\partial x}\sigma_{\text{LReLU}}(x) = \begin{cases} 1, & x > 0, \\ \alpha, & x < 0. \end{cases} \tag{8.20.2}$$

The amount of leak is determined by the value of hyper-parameter $\alpha$. Its value is small and generally varies between 0.01 to 0.1

LReLU is an attempt to fix the Dying ReLU problem. Instead of the function being zero when $x < 0$, a LReLU will instead have a small negative slope of 0.01, as $0.01x$, $x < 0$. This small slope for negative values allows the neurons to have some gradient even when the input is less than zero. This means that during backpropagation, the gradients are non-zero for both positive and negative inputs, making it easier for the network to learn. Figure 8.14 shows the plot for the LReLU AF.

In a standard ReLU, the output is zero for all negative inputs, effectively "turning off" certain neurons. The sparsity property can be beneficial in some scenarios, as it reduces the overall computational load and makes the network more efficient. On the other hand, the LReLU allows a small, non-zero output for negative inputs, preventing neurons from being completely turned off. While this helps mitigate the dying ReLU problem and allows for the flow of gradients during backpropagation, it comes at the cost of losing the sparsity property.

The loss of sparsity might lead to a denser representation in the network, potentially increasing the computational requirements. However, the impact of this loss depends on the specific characteristics of the data and the goals of the model. In some cases, the benefits of avoiding dead neurons and maintaining a continuous flow of gradients during training outweigh the loss of sparsity.

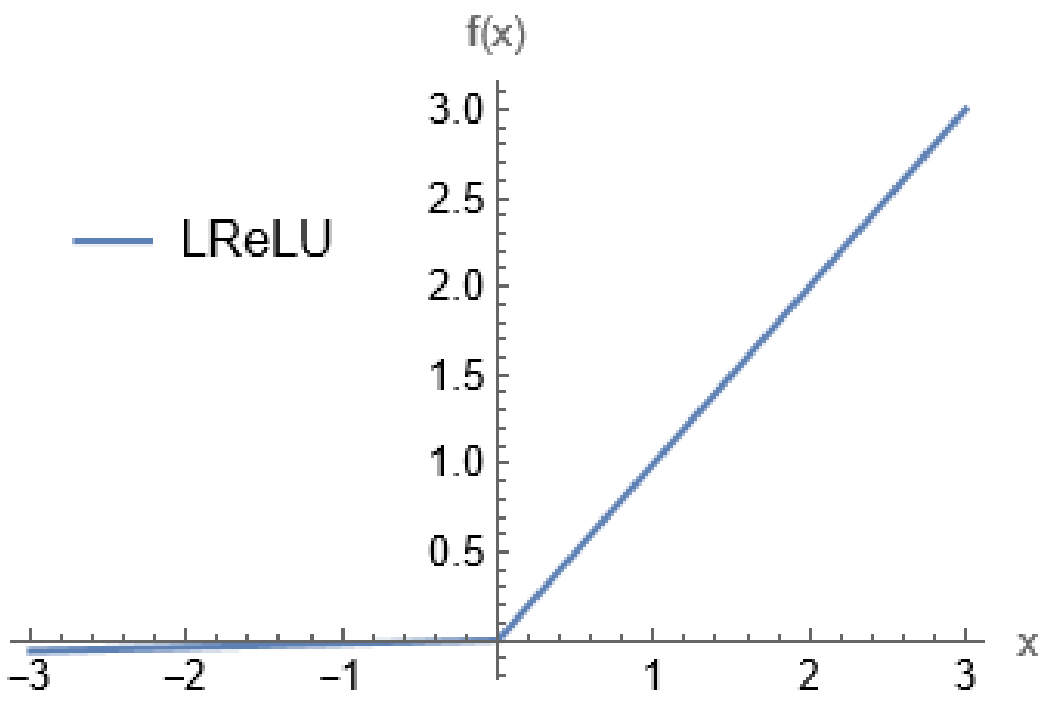

**Figure 8.14.** The figure illustrates the LReLU AF, plotted over a range of $x$ from $-3$ to 3. LReLU is designed to address the "dying ReLU" problem by allowing a small, non-zero gradient ($\alpha = 0.01$) when the input is less than zero. This is achieved by using the function LReLU, resulting in a linear relationship with a very gentle slope for negative values and a normal linear relationship for positive values.

Some features of the LReLU AF include:

- By allowing a small gradient for negative inputs, LReLU helps to prevent neurons from becoming inactive during training.
- Like the standard ReLU, LReLU is simple to implement and computationally efficient.
- The range of the function is $(-\infty, \infty)$.
- It does not saturate for positive values of input and hence does not run into problems related to exploding/vanishing gradients during GD.
- Converges much faster than Sigmoid/Tanh in practice.
- Not zero-centered output.
- In the traditional LReLU AF, the value of $\alpha$ is indeed a fixed hyperparameter that needs to be chosen prior to the training process. This fixed value of $\alpha$ may not be optimal for all scenarios, and finding the right value often involves a degree of trial and error.
- At $x = 0$, the left-hand derivative of LReLU is 0.01 while the right-hand derivative is 1. Since the left-hand derivative and the right-hand derivative are not equal at $x = 0$, the LReLU function is not differentiable at $x = 0$. In the positive part of the LReLU function where the gradient is always 1, there is no vanishing gradient problem. But on the negative part, the gradient is always 0.01 which is close to zero. It leads to a risk of vanishing gradient problem.

### 8.3.3 Parametric ReLU

One major problem associated with LReLU is the finding of the right slope in linear function for negative inputs. Different slopes might be suited for different problems and different networks. Thus, it is extended to Parametric ReLU (PReLU) by considering the slope for negative input as a trainable parameter [238]. The PReLU is given as,

$$\sigma_{\text{PReLU}}(x) = \max(0, x) + \alpha \min(0, x) = \begin{cases} x, & x \geq 0, \\ \alpha\, x, & x < 0, \end{cases} \tag{8.21.1}$$

$$\frac{\partial}{\partial x}\sigma_{\text{PReLU}}(x) = \begin{cases} 1, & x > 0, \\ \alpha, & x < 0, \end{cases} \tag{8.21.2}$$

in the output range of $(-\infty, \infty)$ where $\alpha$ is a learnable (trainable) parameter that determines the slope for negative values. Figure 8.15 depicts the plot of the PReLU AF.

**Remarks:**

- PReLU provides a form of adaptive activation.
- When $\alpha = 0$, it becomes ReLU. If $\alpha$ is small and fixed, PReLU becomes LReLU ($\alpha = 0.01$).

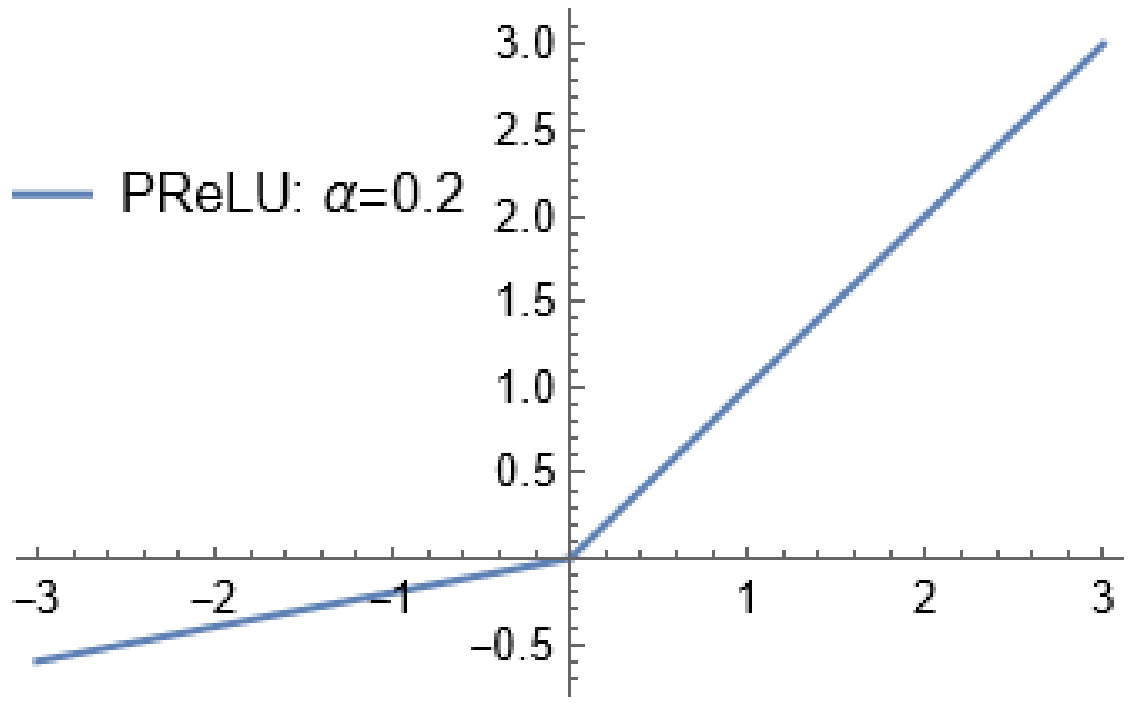

**Figure 8.15.** The figure depicts the PReLU AF, plotted over a range of $x$ from $-3$ to 3 with the parameter $\alpha = 0.2$. PReLU enhances the ReLU function by introducing a learnable parameter $\alpha$ that scales the negative part of the input, thereby allowing the gradient to flow through and avoid dead neurons for negative inputs. The function transitions from a gentle negative slope ($y = 0.2x$ for $x < 0$) to a positive linear relationship ($y = x$ for $x \geq 0$), clearly showing the adaptive nature of PReLU.

- The PReLU AF aims to address some of the limitations of the standard ReLU, such as the "dying ReLU" problem, where neurons can become inactive (output zero) for all inputs during training, leading to dead paths and potential loss of information. By allowing a small negative slope, PReLU seeks to mitigate this issue and enhance the learning capability of NNs.
- During the training process, the parameter $\alpha$ is adjusted using GD or other optimization algorithms, allowing the network to learn an optimal value for the negative slope. This adaptability can be beneficial in scenarios where different parts of the input space might require different slopes for effective learning.
- PReLU introduces a very small number of extra parameters. The number of extra parameters is equal to the total number of channels, which is negligible when considering the total number of weights. There is no extra risk of overfitting. It uses a channel-shared variant: $\sigma_{\text{PReLU}}(x) = \max(0, x) + \alpha \min(0, x)$ where the coefficient is shared by all channels of one layer. This variant only introduces a single extra parameter into each layer. Hence, the computational cost is usually not significantly higher compared to standard ReLU.
- The added flexibility can lead to better model performance, especially in scenarios where the data distribution varies across different dimensions.

### 8.3.4 Randomized ReLU

The Randomized ReLU (RReLU) introduces randomness into the AF by using a random slope for negative inputs during training. This randomness can act as a form of regularization and help prevent overfitting. The RReLU considers the slope of LReLU randomly during training sampled from a uniform distribution $U(l, u)$ [239]. The RReLU is defined as,

$$\sigma_{\text{RReLU}}(x) = \begin{cases} x, & x \geq 0, \\ r\,x, & x < 0. \end{cases} \tag{8.22}$$

Here, "$r$" is a random number sampled from a uniform distribution within a specified range, typically [$l$ =lower, $u$ =upper], where both "lower" and "upper" are small positive constants. These values are usually set before training and remain fixed throughout the training process.

$$r \sim U(l, u), \qquad l < u \quad \text{and} \quad l, u \in [0, 1). \tag{8.23}$$

It uses a deterministic value $\frac{x}{\frac{l+u}{2}}$ during test time. The random nature of "$r$" introduces variability into the AF. The output range is $(-\infty, \infty)$. Figure 8.16 depicts the plot of the RReLU AF.

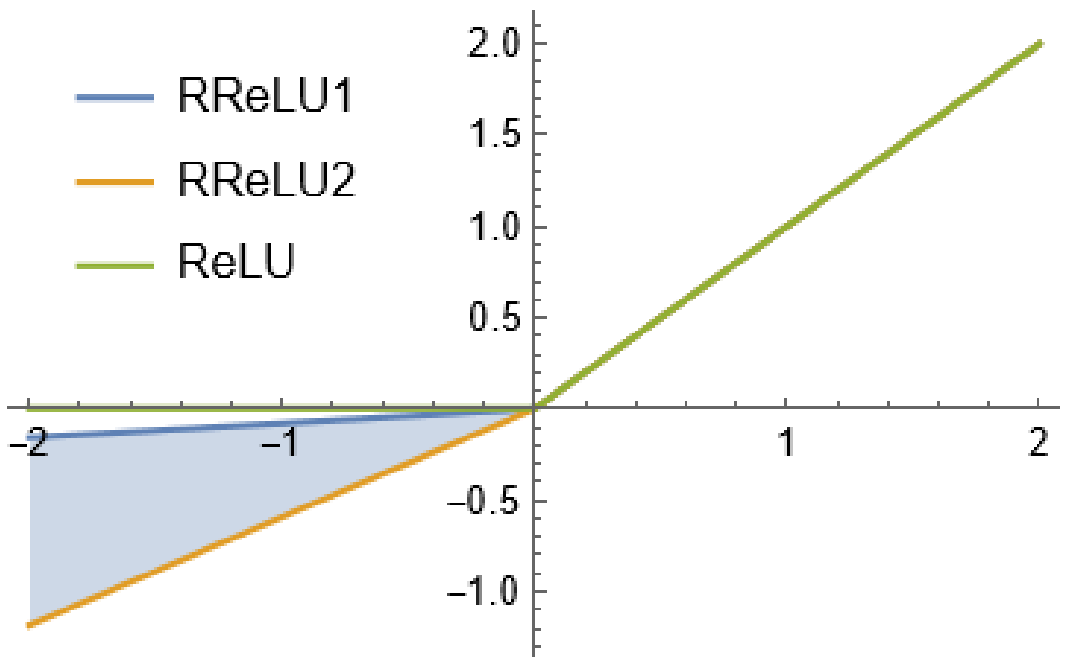

**Figure 8.16.** The figure visualizes two instances of the RReLU AF alongside the standard ReLU function, plotted over a range of $x$ from $-2$ to $2$. Each RReLU curve is generated with a specific, randomly chosen slope parameter $r$, resulting in two unique AFs (denoted as "RReLU1" and "RReLU2") that apply different linear transformations for negative input values. The standard ReLU function, represented here as "ReLU", acts as a reference by clamping all negative inputs to zero.

**Remarks:**

- The choice of $l$ and $u$ defines the range from which $r$ is sampled. The specific values for these parameters depend on the characteristics of the problem and should be chosen through experimentation. As suggested by the Kaggle National Data Science Bowl (NDSB) competition winner, $r$ is sampled from $U(3,8)$.
- The random slope adds robustness to the model by preventing neurons from shutting down completely. This ensures that even if a neuron receives consistently negative inputs, it can still contribute to the learning process.
- Introducing randomness encourages neurons to learn more diverse features, as they are not confined to a strict threshold of zero for negative inputs. This can be beneficial for capturing subtle patterns in the data.
- RReLU does not significantly increase computational complexity compared to standard ReLU. The additional randomness in the slope is usually a lightweight operation in terms of computation.
- The introduction of randomness in the AF can be viewed as a form of regularization. It adds noise to the learning process, which can help prevent overfitting and improve generalization to unseen data.

### 8.3.5 Random Translation ReLU

A similar arrangement is also followed by Random Translation ReLU (RTReLU) [240] by utilizing an offset, sampled from a Gaussian distribution, given as,

$$\sigma_{\text{RTReLU}}(x) = \begin{cases} x + a, & x + a > 0, \\ 0, & x + a \leq 0. \end{cases} \tag{8.24}$$

The output range is $[0, \infty)$ where $a$ is a random number. Figure 8.17 depicts the plot of the RTReLU AF.

In the training stage, the offset of RT-ReLU (i.e., $a$ ) is randomly sampled from Gaussian distribution at each iteration. Namely,

$$a \sim N(0, \sigma), \tag{8.25}$$

where $\sigma$ is the standard deviation of Gaussian distribution.

In the test stage, the average value of $a$ is used for RT-ReLU. Because $a$ is sampled from Gaussian distribution, the average value of $a$ is $0$. Thus, RT-ReLU in the test stage is the same as ReLU.

Based on PReLU, the randomly translational non-linear activation is called RT-PReLU, which can be expressed as:

$$\sigma_{\text{RTReLU}}(x) = \begin{cases} x + a, & x + a > 0, \\ k(x + a), & x + a \leq 0, \end{cases} \tag{8.26}$$

where $a$ is the offset of RT-PReLU on $x$ -axis, which is sampled from Gaussian distribution.

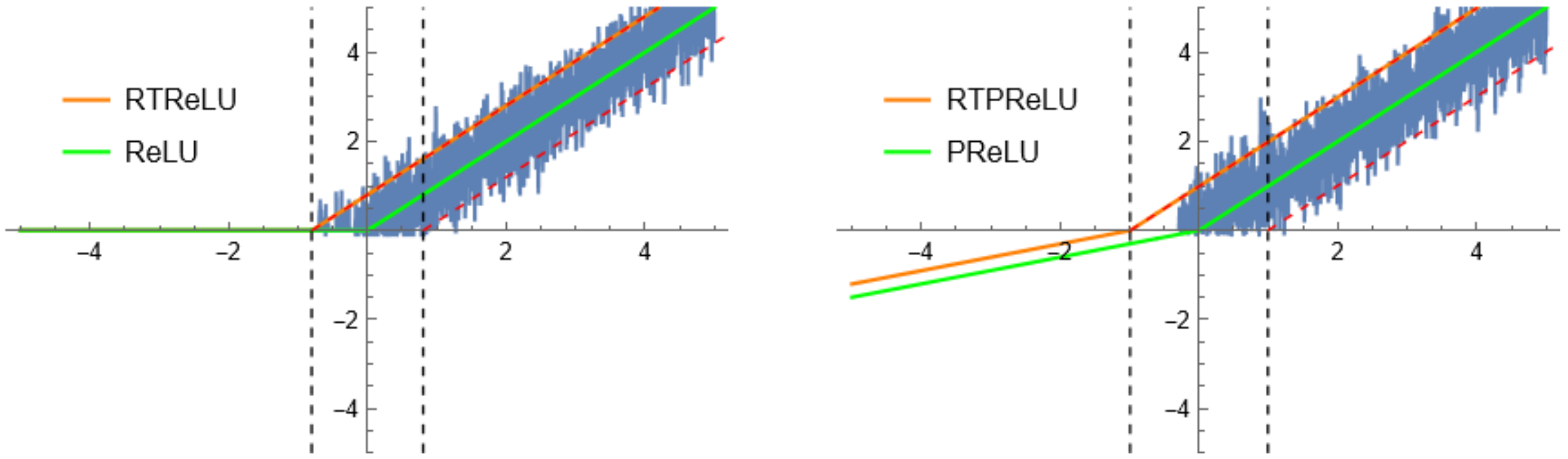

**Figure 8.17.** Left panel: The figure illustrates the behavior of the RTReLU AF compared with the standard ReLU function over a range of $x$ from $-5$ to 5. RTReLU introduces a translation based on a random variable $a$, sampled from a normal distribution with standard deviation $\sigma = 0.5$. The RTReLU curve demonstrates how the random translation affects the activation, introducing variability compared to the fixed threshold of the standard ReLU. Dashed lines indicate the regions of translation, marking the boundaries at $-a$ and $a$, and extending these lines to show the impact on the activation thresholds. Right panel: The figure compares two AF: RTPReLU, and PReLU over a range of $x$ from $-5$ to 5. The RTReLU function incorporates a random translation sampled from a normal distribution with $\sigma = 0.5$, demonstrating variability in its activation threshold. The RTPReLU function further modifies RTReLU by applying a parameterized linear function with a slope of 0.3 for negative inputs. The PReLU function, shown with $\alpha = 0.3$, serves as a reference, applying a fixed slope for negative inputs and linear for positive. Dashed lines indicate the regions of translation, showing the boundaries at $-a$ and $a$, and extending these lines to illustrate the impact on activation thresholds. The plot highlights the flexible behavior of RTReLU and RTPReLU, compared to the more deterministic PReLU, showcasing the potential benefits of incorporating randomness and parameterization in AFs for NNs.

The advantages of RT-ReLU can be summarized as follows:

(1) In the training stage, incorporating the random translation near zero (i.e., $a$) will result in the output of non-linear activation being sensitive to the hard threshold zero if the input is close to zero. However, if the input is far from zero, the output is not sensitive to the hard threshold zero.
(2) Randomly translational non-linear activation can be seen as the regularization of ReLU to reduce the overfitting in the training stage.
(3) Because the randomly translational non-linear activation is the same as the original non-linear activation in the test stage, it can improve accuracy with no increase in computation cost.

### 8.3.6 Elastic ReLU and Elastic Parametric ReLU

An Elastic ReLU (EReLU) considers a slope randomly drawn from a uniform distribution during the training for the positive inputs to control the amount of non-linearity [241]. The EReLU is defined as,

$$\sigma_{\text{EReLU}}(x) = \max(R\,\text{x}, 0) = \begin{cases} R\,\text{x}, & x > 0, \\ 0, & x \le 0, \end{cases} \tag{8.27}$$

in the output range of $[0, \infty)$ where $R$ is randomly selected from a uniform distribution denoted by $R \sim U(1-\alpha, 1+\alpha)$ with $\alpha \in (0,1)$. And $\alpha$ is a variable representing the degree of response fluctuation. It is noted that $R$ only changes during training stage. Figure 8.18 depicts the plot of the EReLU AF. At test time, $R$ equal to the expectation value of $R$, $\mathbb{E}(R)$. Obviously, $\mathbb{E}(R)$ equals to 1 given $R \sim U(1-\alpha, 1+\alpha)$. Therefore, at test time, EReLU becomes ReLU.

An illustration of EReLU is shown in Figure. 8.18. It is seen that the positive outputs of EReLU do not lie in one line. Instead, they lie in the region determined by two rays from the origin, which are denoted by $\sigma_{\text{EReLU }1}(x) = (1-\alpha)x$ and $\sigma_{\text{EReLU }2} = (1+\alpha)x$, respectively. Given an input $x > 0$, ReLU outputs an exact value $\sigma_{\text{ReLU}}(x) = x$ whereas EReLU generates a random value selected uniformly from $[(1-\alpha)x, (1+\alpha)x]$.

**Remarks:**

- EReLU improves model fitting with no extra parameters and little overfitting risk.
- EReLU makes the positive part fluctuate within a moderate range in each epoch during training stage, which strengthens the robustness of the network model.

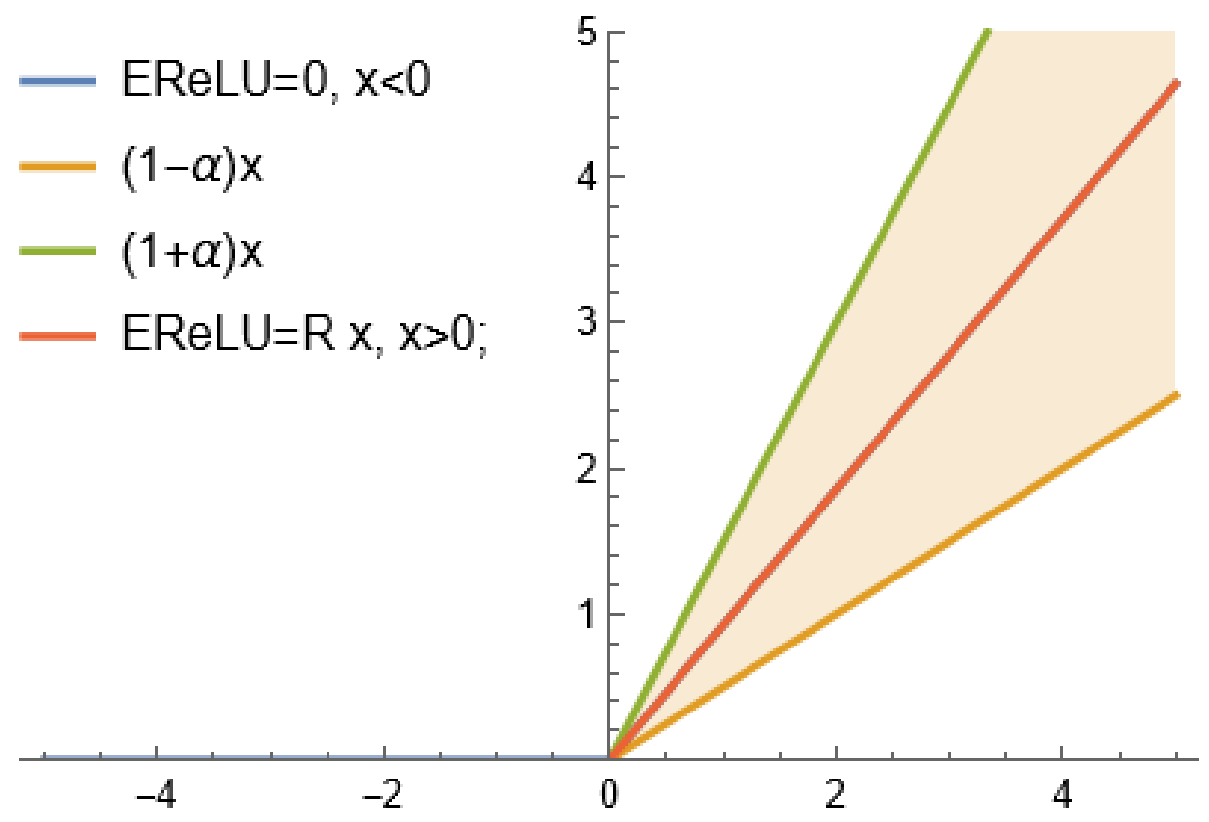

**Figure 8.18.** The figure illustrates the EReLU AF over a range of $x$ from $-5$ to 5, highlighting its adaptive nature by introducing a random scaling factor $R$ for positive inputs, sampled from a uniform distribution between $(1-\alpha)$ and $(1+\alpha)$, $\alpha = 0.5$. The plot displays the EReLU function (depicted as $R\ x$), along with the lower bound $(1-\alpha)x$ and upper bound $(1+\alpha)x$, filling the area between these bounds to emphasize the range of possible values.

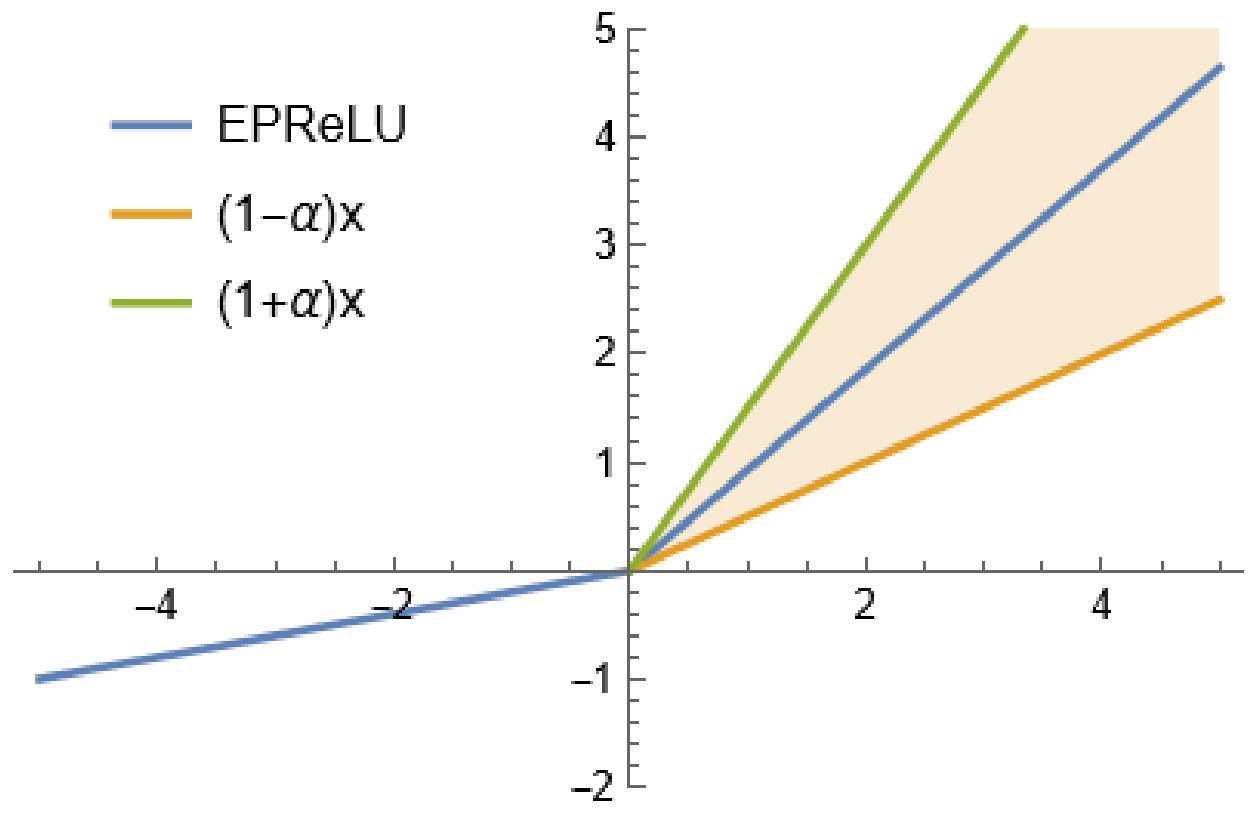

**Figure 8.19.** The figure illustrates the EPReLU AF over a range of $x$ from $-5$ to 5, showcasing its flexible behavior by incorporating a random scaling factor $R$ for positive inputs and a fixed scaling factor $a = 0.2$ for negative inputs. The plot includes the EPReLU function, with $R$ sampled from $[1-\alpha, 1+\alpha]$ where $\alpha = 0.5$, and also displays the lower bound $(1-\alpha)\ x$ and upper bound $(1+\alpha)\ x$, filling the area between these bounds to emphasize the range of possible values.

PReLU and EReLU are two quite different techniques. PReLU mainly improves the negative part of the AF while EReLU reforms the positive part of the AF. Elastic Parametric ReLU (EPReLU) combines the advantages of EReLU and PReLU. Formally, the EPReLU was defined as:

$$\sigma_{\text{EPReLU}}(x) = \max(R\ \text{x}, a\ x) = \begin{cases} R\ x, & x > 0, \\ a\ x, & x \leq 0, \end{cases} \tag{8.28.1}$$

$$\frac{\partial}{\partial x}\sigma_{\text{EPReLU}}(x) = \begin{cases} R, & x > 0, \\ a, & x \leq 0. \end{cases} \tag{8.28.2}$$

It is seen that EPReLU utilizes EReLU and PReLU to process the positive part and negative part, respectively. Figure 8.19 depicts the plot of the EPReLU AF. It is important to note that the two variables $R$ and $a$ are processed quite differently. $R$ is randomly chosen from a uniform distribution whereas $a$ needs to be updated using backpropagation. Both $R$ and $a$ have an effect on the updating of the weights $W$ of networks during training stage.

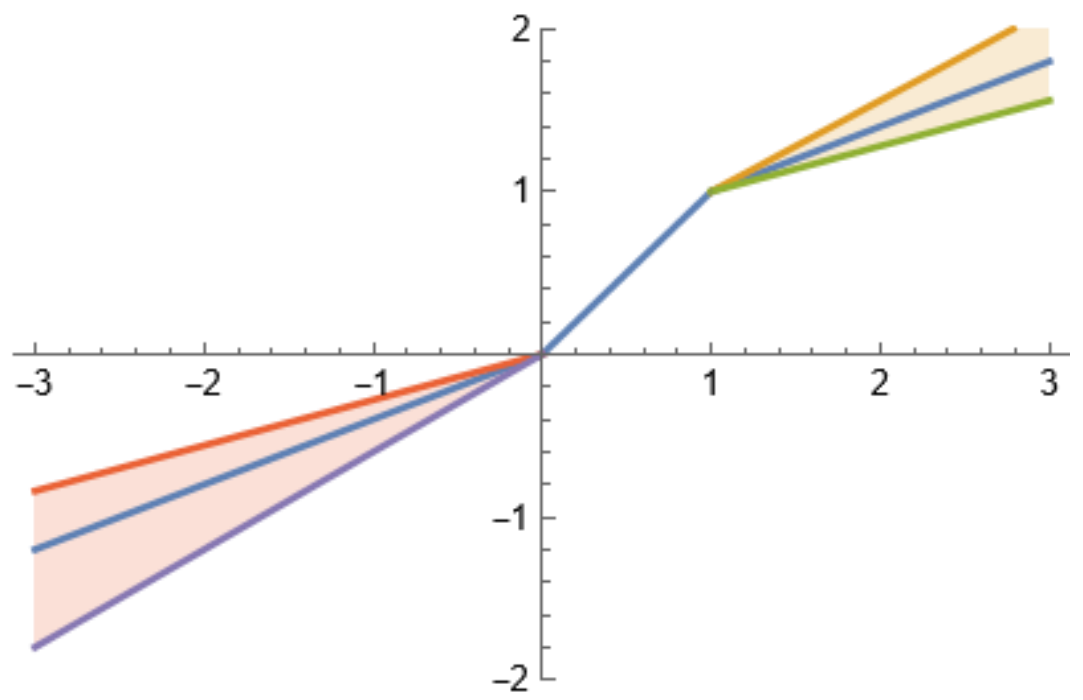

**Figure 8.20.** The figure visualizes the LiSA and ALiSA functions over a range of $x$ from $-3$ to 3, illustrating their piecewise linear behavior across different input regions. The LiSA and ALiSA functions are defined with three segments: a scaled linear function for negative inputs, a linear region for inputs between 0 and 1, and another scaled linear function for positive inputs greater than 1. The plot includes the LiSA and ALiSA functions with parameters $\alpha_1 = 0.4$ and $\alpha_2 = 0.4$, as well as upper and lower bounds for the positive and negative segments. The areas between the bounds are filled to emphasize the range of variability.

### 8.3.7 Linearized Sigmoidal Activation

The Linearized Sigmoidal Activation (LiSA) [242] was defined as:

$$\sigma_{\text{LiSA}}(x) = \begin{cases} \alpha_1 x - \alpha_1 + 1, & 1 < x < \infty, \\ x, & 0 \le x \le 1, \\ \alpha_2 x, & -\infty < x < 0, \end{cases} \tag{8.29.1}$$

$$\frac{\partial}{\partial x}\sigma_{\text{LiSA}}(x) = \begin{cases} \alpha_1, & 1 < x < \infty, \\ 1, & 0 \le x \le 1, \\ \alpha_2, & -\infty < x < 0. \end{cases} \tag{8.29.2}$$

The function considers input data range in three activity regions (Figure 8.20):

- Positive activity region ($1 < x < \infty$),
- Linear activity region ($0 \le x \le 1$), and
- Negative activity region ($-\infty < x < 0$)

Instead of considering data into only two segments (positive or negative), LiSA function divides the data range into multiple segments. Output in negative activity region is controlled by negative slope coefficient ($\alpha_2$) and output for positive activity region is controlled by positive slope coefficient ($\alpha_1$). The linear activity region has direct input to output mapping. The data within a single range segment hold a linear relation, whereas the data points from different range segments hold a non-linear relationship. In other words, all the data points falling in the single range segment (e.g., within linear, positive, or negative activity regions) will hold a linear association. Hence, the model can exploit a wide range of linear and non-linear structural associations in data. Above mentioned behavior brings qualities of saturating and non-saturating AFs into LiSA. Similar to non-saturating AF, LiSA provides a smooth gradient flow even in models with much higher depth and hence does not suffer from vanishing gradient problem. On the other hand, LiSA is able to model higher-order non-linear data associations because all the segments exhibit different activation behavior. This non-linearity is of paramount importance while modeling higher-complexity data. The LiSA function considers three linear functions to increase the non-linearity characteristics.

If the coefficients $\alpha_1$ and $\alpha_2$ are set as follows: $\alpha_1 = 1$ and $\alpha_2 = 0$, the LiSA function transforms into ReLU AF.

$$\begin{aligned} \sigma_{\text{LiSA}}(x) &= \begin{cases} x, & 1 < x < \infty \\ x, & 0 \le x \le 1 \\ 0, & -\infty < x < 0 \end{cases} \\ &= \begin{cases} x, & 0 \le x \le \infty \\ 0, & -\infty < x < 0 \end{cases} \\ &= \sigma_{\text{ReLU}}(x). \end{aligned} \tag{8.30}$$

Similarly, if slope coefficients of LiSA function are set as: $\alpha_1 = 1$ and $\alpha_2 = a$. Then, the LiSA function takes the shape of LReLU AF.

$$\sigma_{\text{LiSA}}(x) = \begin{cases} x, & 1 < x < \infty \\ x, & 0 \le x \le 1 \\ ax, & -\infty < x < 0 \end{cases} = \begin{cases} x, & 0 \le x \le \infty \\ ax, & -\infty < x < 0 \end{cases} = \sigma_{\text{LReLU}}(x). \tag{8.31}$$

The Adaptive Linearized Sigmoidal Activation (ALiSA) function uses trainable slope parameters that can adjust themselves to the required value according to the task at hand. These trainable slope parameters are trained alongside the model parameters during training process. The basic structure of ALiSA remains the same as LiSA. Hence, the same figure can be used to represent the function (Figure 8.20).

$$\sigma_{\text{ALiSA}}(x) = \begin{cases} \alpha_1 x - \alpha_1 + 1, & 1 < x < \infty, \\ x, & 0 \le x \le 1, \\ \alpha_2 x, & -\infty < x < 0. \end{cases} \tag{8.32.1}$$

$$\frac{\partial}{\partial x}\sigma_{\text{ALiSA}}(x) = \begin{cases} \alpha_1, & 1 < x < \infty, \\ 1, & 0 \le x \le 1, \\ \alpha_2, & -\infty < x < 0. \end{cases} \tag{8.32.2}$$

$$\frac{\partial}{\partial \alpha_1}\sigma_{\text{ALiSA}}(x) = \begin{cases} x - 1, & 1 < x < \infty, \\ 0, & 0 \le x \le 1, \\ 0, & -\infty < x < 0. \end{cases} \tag{8.32.3}$$

$$\frac{\partial}{\partial \alpha_2}\sigma_{\text{ALiSA}}(x) = \begin{cases} 0, & 1 < x < \infty, \\ 0, & 0 \le x \le 1, \\ x, & -\infty < x < 0. \end{cases} \tag{8.32.4}$$

### 8.3.8 Rectified Linear Tanh

In order to diminish the vanishing gradient problem that perplexes Tanh and reduce bias shift and noise-sensitiveness that torments the ReLU family, Rectified Linear Tanh (ReLTanh) AF was proposed [243]. The AF ReLTanh is defined as follows.

$$\sigma_{\text{ReLTanh}}(x) = \begin{cases} \text{Tanh}'(\lambda^+)\,(x - \lambda^+) + \text{Tanh}(\lambda^+), & x \ge \lambda^+, \\ \text{Tanh}(x), & \lambda^- < x < \lambda^+, \\ \text{Tanh}'(\lambda^-)\,(x - \lambda^-) + \text{Tanh}(\lambda^-), & x \le \lambda^-. \end{cases} \tag{8.33.1}$$

$$\sigma'_{\text{ReLTanh}}(x) = \begin{cases} \text{Tanh}''(\lambda^+), & x \ge \lambda^+, \\ \text{Tanh}'(x), & \lambda^- < x < \lambda^+, \\ \text{Tanh}''(\lambda^+), & x \le \lambda^-, \end{cases} \tag{8.33.2}$$

where $\lambda^+_{\text{lower}} \le \lambda^+ \le \lambda^+_{\text{upper}}$ and $\lambda^-_{\text{lower}} \le \lambda^- \le \lambda^-_{\text{upper}}$. Figure 8.21 depicts the plot of the ReLTanh AF.

- ReLTanh is constructed by replacing Tanh's saturated waveforms in positive and negative inactive regions with two straight lines, and the slopes of the lines are calculated by Tanh's derivatives at two learnable thresholds, so ReLTanh consists of the nonlinear Tanh in the center and two linear parts on both ends.
- The positive line is steeper than the negative one, and both lines start at two learnable thresholds.
- The middle Tanh waveform provides ReLTanh with the ability of nonlinear fitting, and the linear parts contribute to the relief of the vanishing gradient problem.
- $\lambda^+$ and $\lambda^-$ are respectively the positive and negative thresholds that determine the start positions and slopes of the straight lines.
- Besides, both $\lambda^+$ and $\lambda^-$ can be trained by the BP algorithm, so it can tolerate the variation of inputs and help to minimize the cost function and maximize the data fitting performance.

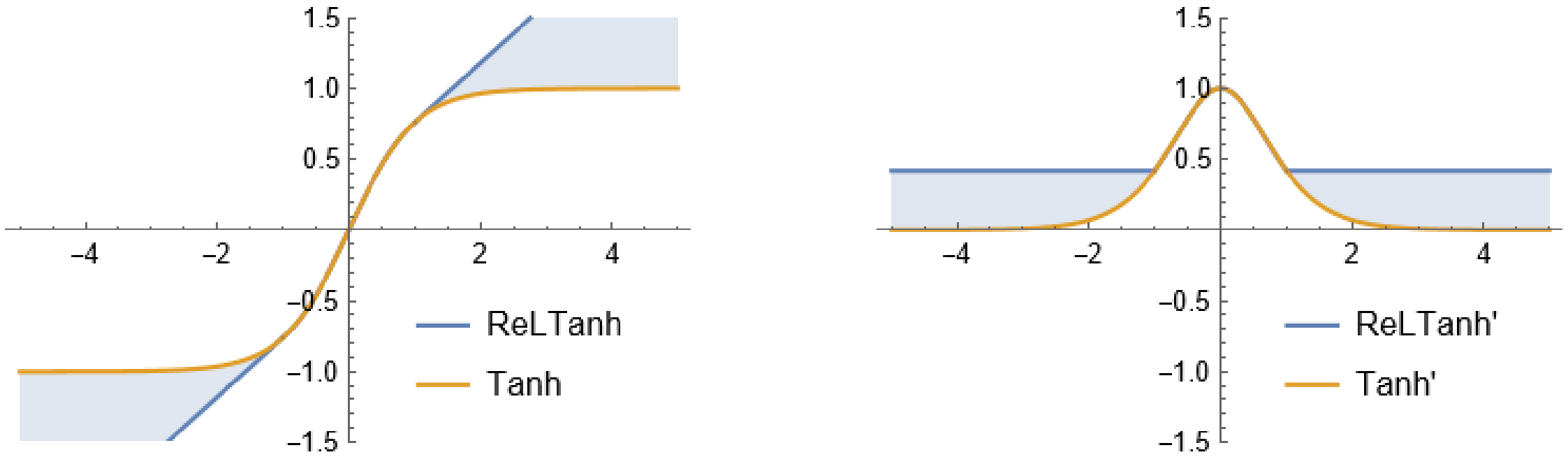

**Figure 8.21.** The figure includes two plots comparing the ReLTanh AF and its derivative with the standard Tanh AF and its derivative over a range of $x$ from $-5$ to 5. Left panel: This plot shows the ReLTanh function, which combines three regions: a right linear region for $x \geq \lambda^+$ (with $\lambda^+ = 1$), a hyperbolic tangent region for $\lambda^- < x < \lambda^+$ (with $\lambda^- = -1$), and a left linear region for $x \leq \lambda^-$. The Tanh function is included for comparison, and the area between the ReLTanh and Tanh functions is filled to highlight their differences. Right panel: This plot presents the derivatives of the ReLTanh and Tanh functions. The ReLTanh derivative consists of constant values in the linear regions $\text{Tanh}'(\lambda^+)$ and $\text{Tanh}'(\lambda^-)$ and the derivative of the hyperbolic tangent $\text{sech}[x]^2$ in the middle region. The derivative of Tanh is shown as $1 - \text{Tanh}[x]^2$, and the area between the ReLTanh and Tanh derivatives is filled to emphasize the differences in their gradient behaviors.

- It is worth noting there are extra limiting conditions for both $\lambda^+$ and $\lambda^-$: $\lambda^+_{\text{lower}} \leq \lambda^+ \leq \lambda^+_{\text{upper}}$ and $\lambda^-_{\text{lower}} \leq \lambda^- \leq \lambda^-_{\text{upper}}$, and they are mainly used to constrain the learnable range of slopes to avoid unreasonable waveform and guarantee the gradient-vanishing proof capacity.
- These thresholds can be trained and updated along the GD direction of the cost function. So ReLTanh can improve the vanishing gradient problem just like the ReLU family, and its mean outputs are closer to zero so that it is affected less bias shift than ReLU family.

The advantages of ReLTanh compared against other common AFs are listed as follows.

- ReLTanh has better derivative performance compared to Tanh, and it can diminish the vanishing gradient problem as ReLU family does.
- For mean activation, the outputs of ReLTanh are closer to zero, thus it is affected less by bias shift than ReLU family. Compared to ReLTanh, ReLU is absolutely non-negative, and LReLU has negligible negative outputs compared to their positive ones. With lighter bias shift influence, ReLTanh can speed up and smooth training process.
- The advantage of learnable thresholds helps ReLTanh to approach more closely to the global minimum. As training goes on, ReLTanh can adjust automatically the learnable thresholds in the descending direction of loss. It is worth noting that ReLTanh has a similar waveform to ELU, but ReLTanh can outperform ELU, not only because ReLTanh can update thresholds to help to search the minimize of the cost function, but also because ELU still suffers from vanishing gradient problem in the negative interval.

### 8.3.9 Shifted ReLU and Displaced ReLU

The ReLU AFs are commonly used in NNs due to their simplicity and effectiveness in mitigating the vanishing gradient problem. The BN is another technique that is often employed to improve the training stability and convergence of DNNs. While both ReLU and BN have been successful in practice, there are certain interactions and complications when using them together. Here are some considerations:

- In ML, normalizing the distribution of the input data decreases the training time and improves test accuracy. Consequently, normalization also improves NNs performance. A standard approach to normalizing input data distributions is the mean standard technique. The input data is transformed to present zero mean and standard deviation of one.
- However, if instead of working with shallow NNs, we are dealing with DNNs; the problem becomes more sophisticated. Indeed, in a DNN, the output of a layer works as input data to the next. Therefore, in this sense, each layer of a deep model has his own "input data" that is composed of the previous layer output. The only exception is the first layer, for which the input is the original data.

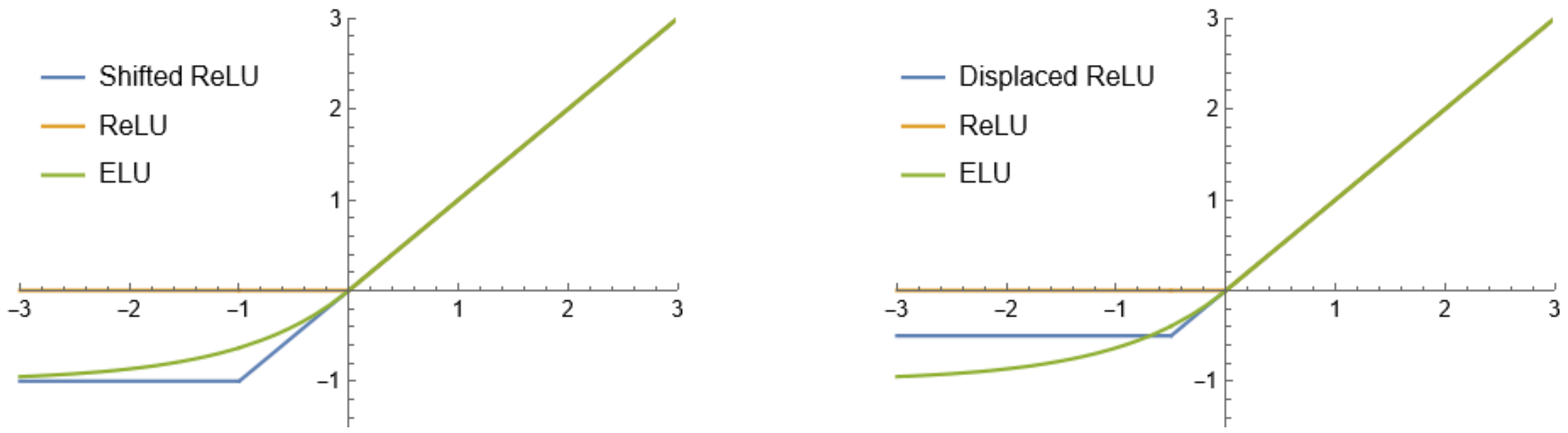

**Figure 8.22.** The figure consists of two plots comparing different AFs: SReLU, ReLU, and ELU in the first plot, and DReLU, ReLU, and ELU in the second plot, all over a range of $x$ from $-3$ to 3. Left panel: This plot visualizes the SReLU, ReLU, and ELU AFs. The SReLU function is defined as $\max(-1, x)$, shifting the ReLU function downward by 1. The ReLU function outputs zero for negative inputs and linearly increases for positive inputs. The ELU function, with $\alpha = 1$, exponentially decays for negative inputs and linearly increases for positive inputs. The plot ranges from $-1.5$ to 3 on the $y$-axis, showing how each function behaves across the input range. Right panel: This plot shows the DReLU function, defined as $\max(x, -\delta)$ with $\delta = 0.5$, alongside the standard ReLU and ELU ($\alpha = 1$) functions. The DReLU introduces a displacement, shifting the threshold to $-0.5$, compared to the standard ReLU and ELU functions.

- Considering each layer has its own "input data" (the output of the previous layer), normalizing only the actual input data of a DNN produces a limited effect in enhancing learning speed and test accuracy. Moreover, during the training process, the distribution of the input of each layer changes, which makes training even harder. Indeed, the parameters of a layer are updated while its input (the activations of the previous layer) is modified.
- This phenomenon is called internal covariant shift, which is a major factor that hardens the training of DNNs. In fact, while the data of shallow NNs are normalized and static during training, the input of a deep model layer, which is the output of the previous one, is neither a priori normalized nor static throughout training. BN is an effective method to mitigate the internal covariant shift. This approach, which significantly improves training speed and test accuracy, proposes normalizing the inputs of the layers when training DNNs.
- Despite BN, in the case of ReLU AF, the activations are not perfectly normalized since these outputs present neither zero mean nor unit variance. Consequently, regardless of the presence of a BN layer, after the ReLU, the inputs passed to the next composed layer have neither mean of zero nor variance of one that was the objective in the first place. In this sense, ReLU skews an otherwise previous normalized output. In other words, ReLU reduces the correction of the internal covariance shift promoted by the BN layer. The ReLU bias shift effect is directly related to the drawback ReLU generates to the BN procedure.

Consequently, Shifted ReLU (SReLU) [244] and Displaced ReLU (DReLU) [245], aiming to mitigate the mentioned problem, which is essentially a diagonally displaced ReLU were proposed. SReLU can be written as,

$$\sigma_{\text{SReLU}}(x) = \max(-1, x). \tag{8.34}$$

Figure 8.22 depicts the plot of the SReLU AF. The DReLU is designed as a generalization of SReLU. The DReLU displaces the rectification point to consider the negative values, given as,

$$\sigma_{\text{DReLU}}(x) = \begin{cases} x, & x \geq -\delta, \\ -\delta, & x < -\delta, \end{cases} \tag{8.35}$$

having the output range in $[-\delta, \infty]$, see Figure 8.22.

- DReLU is essentially a ReLU diagonally displaced into the third quadrant. Different from LReLU, PReLU, and ELU AFs, the inflection of DReLU does not happen at the origin, but in the third quadrant.
- DReLU generalizes both ReLU and SReLU by allowing its inflection to move diagonally from the origin to any point of the form $(-\delta, -\delta)$. If $\delta = 0$, DReLU becomes ReLU. If $\delta = 1$, DReLU becomes SReLU. Therefore, the slope zero component of the AF provides negative activations, instead of null ones. Unlike ReLU, in DReLU learning can happen for negative inputs since the gradient is not necessarily zero.

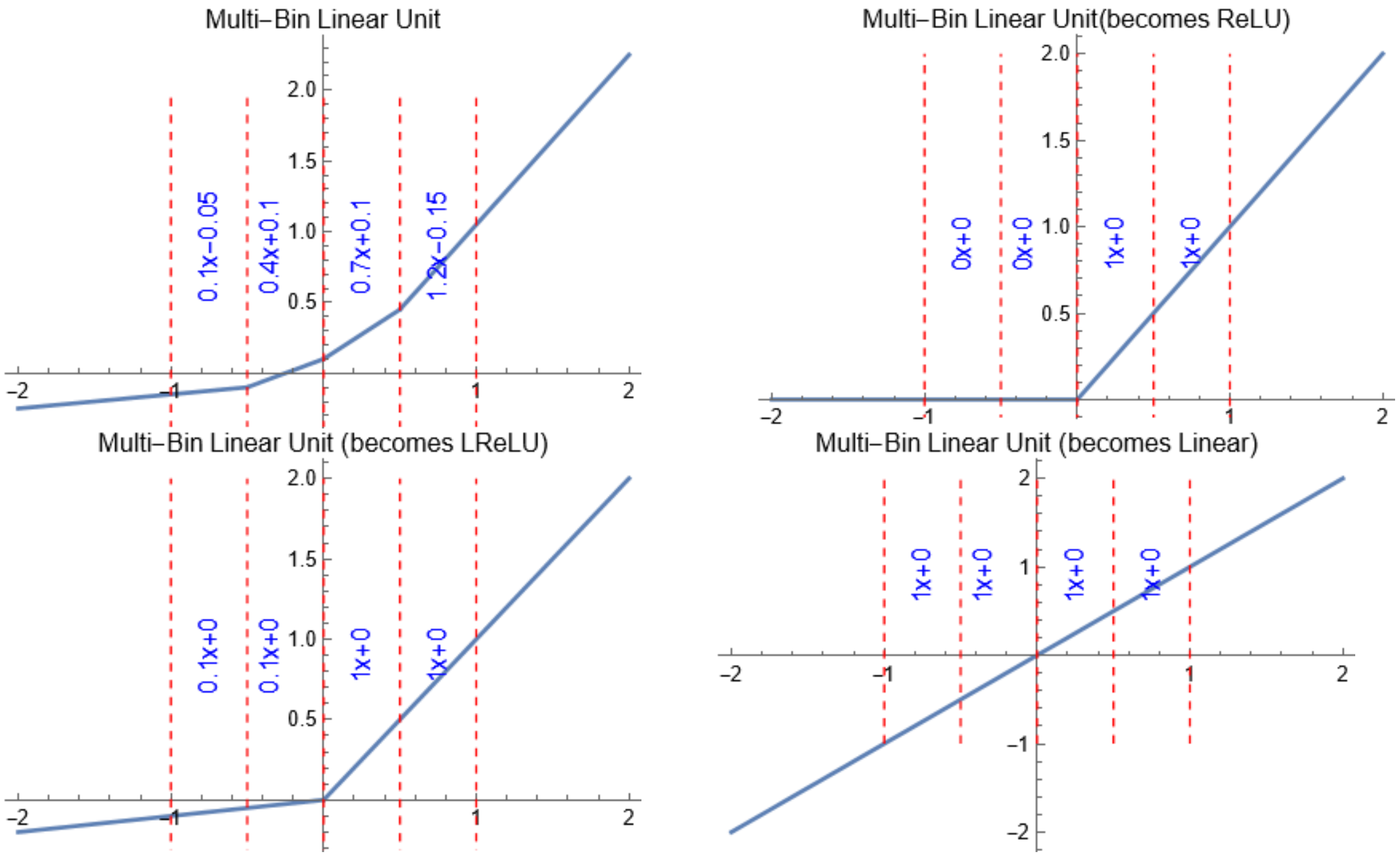

**Figure 8.23.** The figure showcases four plots of the MBLU AF, illustrating its versatility in mimicking different types of AFs over a range of $x$ from $-2$ to 2. Top left panel: The first plot presents the MBLU with piecewise linear segments having varying slopes and intercepts. Top right panel: The second plot demonstrates how the MBLU can resemble a ReLU function, outputting zero for negative inputs and linearly increasing for positive inputs. Bottom left panel: The third plot shows the MBLU configured like a LReLU, introducing a small slope for negative inputs. Bottom right panel: The fourth plot displays the MBLU as a purely linear function across all input values. Each plot includes dashed lines and labels to indicate the specific equations for each segment, highlighting the adaptable nature of the MBLU in NN applications.

- DReLU is less computationally complex than LReLU, PReLU, and ELU. In fact, since DReLU has the same shape as ReLU, it essentially has the same computational complexity.

### 8.3.10 Multi-bin Trainable Linear Unit

The nonlinearity of NNs comes from the non-linear AFs, by stacking some simple operations, e.g. ReLU, the networks are able to model any nonlinear function. However, in many real applications, the computational resources are limited and, thus, we are not able to deploy very deep models to fully capture the nonlinearity. The Multi-bin Trainable Linear Unit (MTLU) is used to improve the capacity of AFs for better nonlinearity modeling.

Instead of designing fixed AFs, MTLU was proposed to parameterize the AFs and learn optimal functions for different stages of networks [246]. The MTLU AF simply divides the activation space into multiple equidistant bins and uses different linear functions to generate activations in different bins. The MTLU can be written as,

$$\sigma_{\mathrm{MTLU}}(x) = \begin{cases} a_0 x + b_0 & x \le c_0, \\ a_k x + b_k & c_{k-1} < x \le c_k, \\ a_K x + b_K & c_{K-1} < x. \end{cases} \tag{8.36}$$

having the output range in $(-\infty, \infty)$. The number of bins and the range of bins are the hyperparameters, whereas the linear function of a bin is trainable (i.e., $a_0, \ldots, a_K$ $b_0, \ldots, b_K$ are the learnable parameters). Since the anchor points $c_k$ in the model are uniformly assigned, they are defined by the number of bins, $K$, and the bin width. Furthermore, given the input value $x$, a simple dividing and flooring function can be utilized to find its corresponding bin-index. Having the bin-index, the activation output can be achieved by an extra multiplication and addition function.

Figure 8.23 presents 4-bin MTLU AF and, for reference, some other commonly used AFs are also included. One can see that in this simple case, MTLU divides the activation space into 4 parts, $(-\infty, -0.5]$, $(-0.5, 0]$, $(0, 0.5]$ and $(0.5, \infty)$, and adopts different linear functions in different parts to form the non-linear AF. PReLU can be seen as a special case of the proposed parameterization in which the input space is divided into two bins $(-\infty, 0]$ and $(0, \infty)$ and only the parameter $a_0$ is learned, the other parameters $b_0$, $a_1$ and $b_1$ being fixed to 0, 1, and 0, respectively. You can initialize MTLU as a ReLU function, see Figure 8.23. With other initializations, such as random initialization of $\{a_k, b_k\}, k = 0, \ldots, K$ and initialization MTLU as identity mapping function $\sigma(x) = x$, Figure 8.23, MTLU is still trainable.

**Remarks:**

- Activations are carefully parameterized within the range of $[-1, 1]$, as most inputs to MTLU fall within this interval.
- After fixing the parameterization range, the choice of bin-width is crucial. The number of bins can be determined as 2 divided by the bin-width.
- There is a trade-off between bin-width and the number of bins. A smaller bin-width is intuitively expected to enhance the parameterization accuracy of MTLU, thereby improving the nonlinearity modeling capacity of the network. However, this improvement comes at the cost of introducing more parameters (a larger number of bins) to cover the range between $(-1, 1)$.

## 8.4 ELU Based AFs

### 8.4.1 Exponential Linear Unit

- Units with non-zero mean activations effectively act as biases for the next layer. If these non-zero means are not canceled out or balanced, it can lead to a bias shift effect as learning progresses through the network.
- The more correlated the units with non-zero mean activations are, the higher the potential for bias shift. Correlated activations can amplify each other's impact, leading to a cumulative effect on the bias shift in subsequent layers.
- AFs play a crucial role in shaping the distribution of activations. Good AFs aim to push activation means closer to zero. This property helps in reducing bias shift effects and contributes to more stable learning.
- Centering activations around zero is known to speed up learning in NNs. This is because weight updates during training are influenced by the gradients of the AFs. When activations are centered, gradients are more balanced, and the optimization process is generally more effective.
- BN is a technique that aims to address issues related to internal covariate shift. It normalizes the activations in a mini-batch, centering them around zero and scaling them to have a certain variance. This normalization can contribute to mitigating bias shift effects and accelerating training.

Using an AF that naturally pushes the mean activation toward zero is an alternative approach to achieve the benefits of centered activations. AFs that inherently maintain activations around zero can help mitigate bias shift and facilitate more stable learning in NNs. For example:

1. Tanh has been preferred over logistic functions.
2. LReLUs that replace the negative part of the ReLU with a linear function have been shown to be superior to ReLUs.
3. PReLUs generalize LReLUs by learning the slope of the negative part which yielded improved learning behavior on large image benchmark data sets.
4. RReLUs which randomly sample the slope of the negative part raised the performance on image benchmark datasets and convolutional networks.

The Exponential Linear Unit (ELU) has negative values to allow for mean activations close to zero, but it saturates to a negative value with smaller arguments.

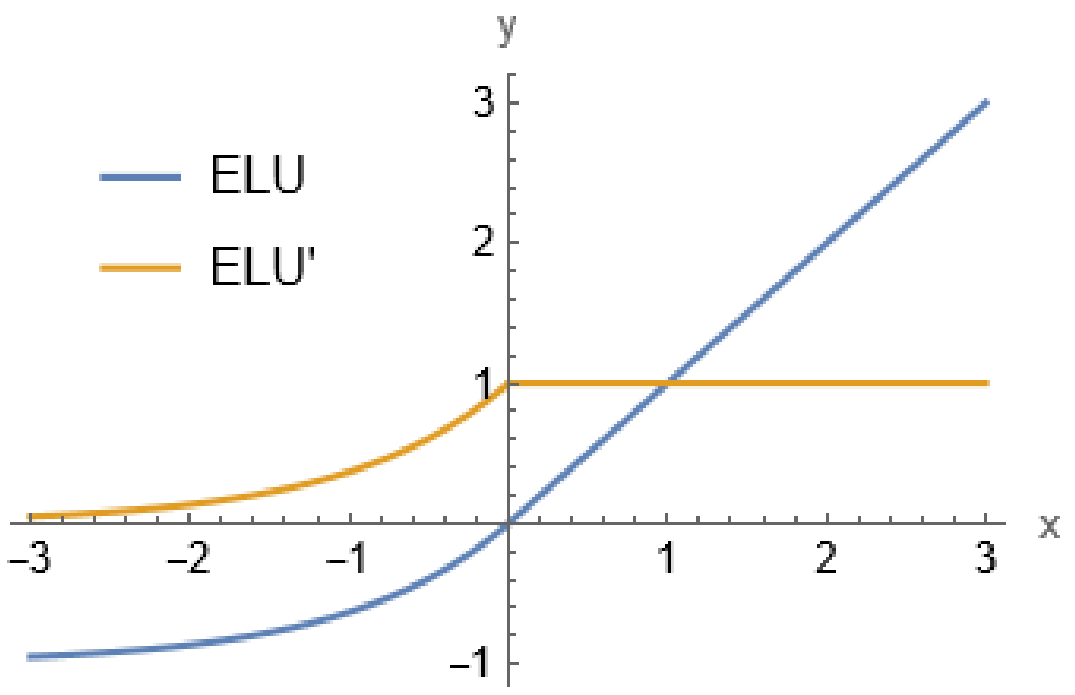

**Figure 8.24.** The figure illustrates the ELU AF alongside its derivative, plotted over a range of $x$ from $-3$ to $3$. The ELU function is designed to improve the convergence rate during training and reduce the vanishing gradient problem compared to traditional ReLU. It transitions from an exponential growth for negative values (where $\sigma_{\text{ELU}}(x) = \alpha(\exp(x) - 1)$ for $x < 0$) to a linear relationship for non-negative values (where $\sigma_{\text{ELU}}(x) = x$ for $x \geq 0$). The derivative of ELU shows how the rate of change adapts across different values of $x$.

ELU with $0 < \alpha$ [244] is given as,

$$\sigma_{\text{ELU}}(x) = \begin{cases} x, & x > 0, \\ \alpha(e^x - 1), & x \leq 0. \end{cases} \tag{8.37.1}$$

$$\frac{\partial}{\partial x}\sigma_{\text{ELU}}(x) = \begin{cases} 1, & x > 0 \\ \alpha e^x, & x \leq 0 \end{cases} = \begin{cases} 1, & x > 0, \\ \sigma_{\text{ELU}}(x) + \alpha, & x \leq 0, \end{cases} \tag{8.37.2}$$

having the output range in $[-1, \infty)$ where $\alpha$ is a learnable parameter. The ELU hyperparameter $\alpha$ controls the value to which an ELU saturates for negative net inputs (see Figure. 8.24). If $x$ keeps reducing past zero, eventually, the output of the ELU will be capped at $-1$, as the limit of $e^x$ as $x$ approaches negative infinity is $0$. The value for $\alpha$ is chosen to control what we want this cap to be regardless of how low the input gets. This is called the saturation point. At the saturation point, and below, there is very little difference in the output of this function, and hence there's little to no variation (differential) in the information delivered from this node to the other node in the forward propagation.

The formulation of ELU involves an exponential term for negative inputs, which allows the network to capture information from both positive and negative sides of the input space, contributing to its ability to maintain mean activations close to zero.

Key characteristics and benefits of the ELU AF include:

- Like ReLUs, LReLUs, and PReLUs, ELUs alleviate the vanishing gradient problem via the identity for positive values.
- However, ELUs have improved learning characteristics compared to the units with other AFs. In contrast to ReLUs, ELUs have negative values which allows them to push mean unit activations closer to zero like BN but with lower computational complexity. Mean shifts toward zero speed up learning.
- While LReLUs and PReLUs have negative values, too, they do not ensure a noise-robust deactivation state. ELUs saturate to a negative value with smaller inputs and thereby decrease the forward propagated variation and information.
- ELU can help mitigate the "dying ReLU" problem, which occurs when ReLU units always output zero for certain inputs, causing neurons to become inactive and impeding learning. ELU neurons do not suffer from this problem because they can produce non-zero outputs for negative inputs.
- ELU can handle outliers better than ReLU, as the exponential term allows it to capture extreme values without saturating (Robust to outliers).
- ELU can lead to faster convergence and potentially better generalization performance compared to some other AFs, especially when dealing with complex datasets.

- The $\alpha$ parameter in the ELU equation controls the slope of the function for negative inputs. It's often set to a small positive value like 1.0, but it can be tuned to suit the specific problem. The choice of $\alpha$ can impact the behavior of ELU, and in practice, it's often chosen through experimentation.

### 8.4.2 Scaled ELU

Self-Normalizing Neural Networks (SNNs) are a type of ANN designed to address the challenges of training DNNs. They were introduced to overcome some of the limitations associated with traditional AFs and help stabilize the training process in DNNs. The key idea behind SNNs is the use of AFs that promote self-normalization of the network's activations. The concept of self-normalization refers to the ability of the network to maintain a stable distribution of activations across layers during training, which can lead to more efficient and faster convergence. For a NN with AF $\sigma$, let us consider two consecutive layers that are connected by a weight matrix $\mathbf{W}$.

- Since the input to a NN is a random variable, the activations in the lower layer $\mathbf{x}$, the network inputs $\mathbf{z} = \mathbf{W}\mathbf{x}$, and activations in the higher layer $\mathbf{y} = \sigma(\mathbf{z})$ are treated as random variables due to the variability in the data and weights.
- The mean $\mu$ and variance $\nu$ of activations $x_i$ in the lower layer are denoted by $\mathbb{E}(x_i)$ and $\mathrm{Var}(x_i)$, respectively.
- The mean $\tilde{\mu}$ and variance $\tilde{\nu}$ of activations $y$ in the higher layer are denoted by $\mathbb{E}(y)$ and $\mathrm{Var}(y)$, respectively.
- The net input $z$ for a single activation $y = \sigma(z)$ is given by $z = \mathbf{w}^T\mathbf{x}$, $\mathbf{w} \in \mathbb{R}^n$.
- For $n$ units in the lower layer with activations $x_i$, $1 \leq i \leq n$, the mean of the weight vector $\mathbf{w}$ is defined as $\omega = \sum_{i=1}^{n} w_i$ and the second moment as $\tau = \sum_{i=1}^{n} w_i^2$.
- The mapping function $g$ is introduced, which maps the mean and variance of activations from one layer to the mean and variance of activations in the next layer. Formally, $g: (\mu, \nu) \rightarrow (\tilde{\mu}, \tilde{\nu})$.
- Normalization techniques (batch, layer, or weight normalization) ensure that $g$ keeps $(\mu, \nu)$ and $(\tilde{\mu}, \tilde{\nu})$ close to predefined values, typically $(0,1)$.

**Definition (SNN):** A NN is considered SNN if it possesses a mapping $g: \Omega \rightarrow \Omega$ for each activation $y$ that maps mean and variance from one layer to the next, and has a stable and attracting fixed point depending on $(\omega, \tau)$ in $\Omega$, where $\Omega = \{(\mu, \nu): \mu \in [\mu_{\min}, \mu_{\max}], \nu \in [\nu_{\min}, \nu_{\max}]\}$. When iteratively applying the mapping $g$, each point within $\Omega$ converges to this fixed point.

**Remarks:**

- Activations of a NN are considered normalized if both their mean and their variance across samples are within predefined intervals.
- If the mean and variance of $\mathbf{x}$ are already within these intervals, then the mean and variance of $\mathbf{y}$ also remain in these intervals, indicating transitivity across layers.
- Within these intervals, both the mean and variance converge to a fixed point if the mapping $g$ is applied iteratively.
- Therefore, SNNs keep normalization of activations when propagating them through layers of the network.
- This convergence property of SNNs allows to train DNNs with many layers, employs strong regularization schemes, and makes learning highly robust.

SNNs cannot be derived with (scaled) ReLUs, Sigmoid units, Tanh units, and LReLUs. Requirements for the AF used in SNNs are:

1. The AF must have both negative and positive values to control the mean of the activations.
2. It should have saturation regions where derivatives approach zero. This helps dampen the variance if it is too large in the lower layer.
3. The AF should have a slope larger than one to increase the variance if it is too small in the lower layer.
4. It should exhibit a continuous curve to ensure the existence of a fixed point, where variance damping is balanced by variance increasing.

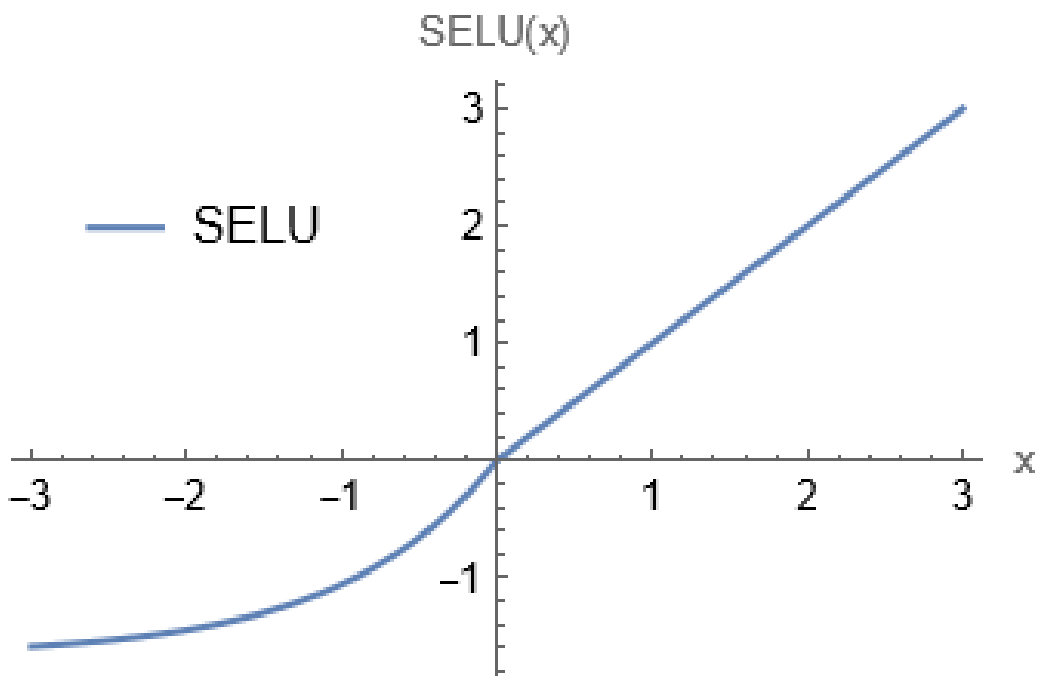

**Figure 8.25.** The figure displays the SELU AF, plotted over a range of $x$ from $-3$ to 3. SELU is an adaptation of the ELU function, scaled by the parameters $\lambda$ and $\alpha$ to ensure self-normalizing properties in NNs. It behaves linearly for positive inputs ($x > 0$, where $\sigma_{\text{SELU}}(x) = \lambda x$) and transitions to an exponential growth minus one for negative inputs ($x \leq 0$, where $\sigma_{\text{SELU}}(x) = \lambda\alpha(\exp(x) - 1)$), designed to keep the mean and variance of the outputs close to zero and one, respectively, across layers.

The ELU is extended to Scaled ELU (SELU) [247] by using a scaling hyperparameter to make the slope larger than one for positive inputs. The SELU AF has specific properties that encourage self-normalization. The SELU can be defined as,

$$\sigma_{\text{SELU}}(x) = \lambda \begin{cases} x, & x > 0, \\ \alpha(e^x - 1), & x \leq 0, \end{cases} \tag{8.38}$$

having the output range in $[-\lambda, \infty)$ where $\alpha$ is a hyperparameter. $\lambda$ is a scaling factor, typically set to values close to 1 (e.g., 1.0507) (see Figure. 8.25).

Choice of AF -ELU:

- The ELU is chosen as the base AF.
- ELU is a type of ReLU variant that allows negative values, addressing the requirement for both negative and positive values.
- The ELU is multiplied by a scaling factor $\lambda > 1$.
- This modification ensures that the slope of the AF is larger than one for positive net inputs, satisfying the requirement for a slope larger than one.
- The modification with $\lambda > 1$ is introduced to enhance the ability of the AF to increase the variance when needed.
- It helps achieve a balance between damping and increasing the variance, promoting stability during the training of DNNs.

When properly initialized, SELU neurons transform their inputs in such a way that the activations tend to have a mean of approximately zero and a standard deviation of approximately one during training. This self-normalization property helps in addressing the vanishing and exploding gradient problems, making it easier to train DNNs.

### 8.4.3 Parametric ELU

The standard ELU is defined as the identity for positive arguments and $a(e^x - 1)$ for negative arguments ($x < 0$). Although the parameter $a$ can be any positive value, we need $a = 1$ to have a fully differentiable function. For other values $a \neq 1$, the function is non-differentiable at $x = 0$. Directly learning parameter $a$ would break differentiability at $x = 0$, which could impede back-propagation. For this reason, let us first start by adding two additional parameters to ELU:

$$\sigma_{\text{PELU}}(x) = \begin{cases} c\, x, & x \geq 0, \\ a\left(e^{\frac{x}{b}} - 1\right), & x < 0, \end{cases} \tag{8.39}$$

where $a, b, c > 0$. The original ELU can be recovered when $a = b = c = 1$.

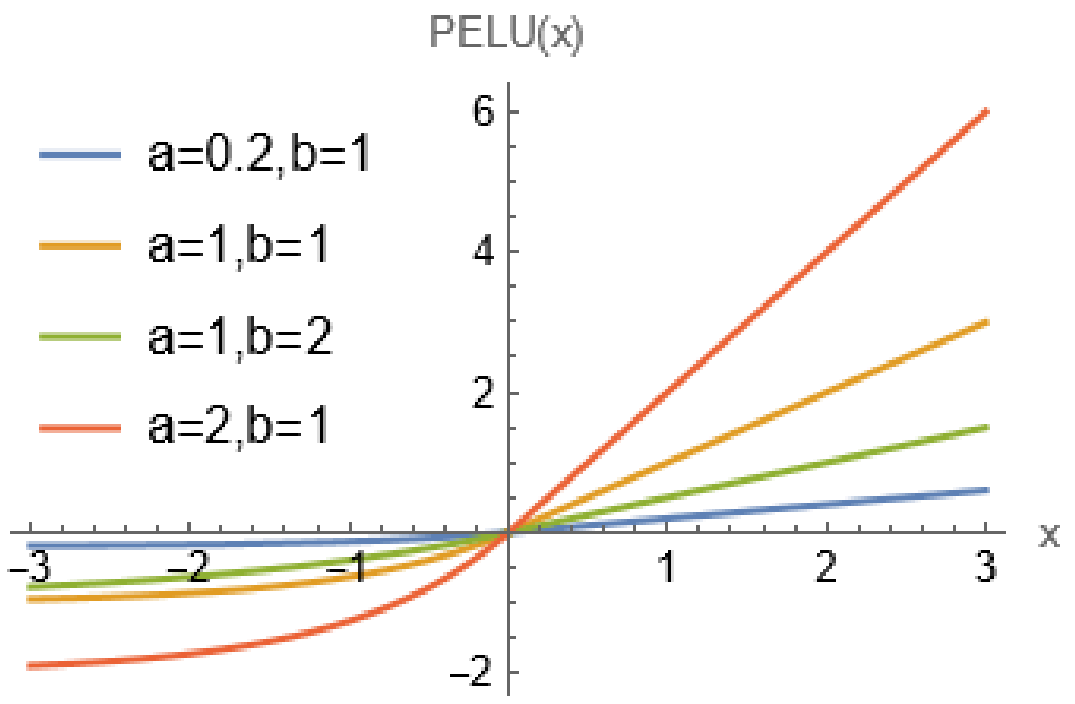

**Figure 8.26.** The figure showcases the PELU AF with various parameter settings, plotted over a range of $x$ from $-3$ to 3. The PELU function generalizes the ELU by introducing parameters $a$ and $b$, which control the output for negative inputs more flexibly. The plot includes four curves, each representing different combinations of the parameters $a$ and $b$: ($a = 0.2$, $b = 1$), ($a = 1$, $b = 1$), ($a = 1$, $b = 2$), and ($a = 2$, $b = 1$). These variations illustrate how adjusting $a$ and $b$ affects the slope and curvature of the PELU function for negative $x$, demonstrating its adaptability to different activation characteristics.

Each parameter controls different aspects of the activation.

- Parameter $c$ changes the slope of the linear function in the positive quadrant (the larger $c$, the steeper the slope).
- Parameter $b$ affects the scale of the exponential decay (the larger $b$, the smaller the decay).
- While $a$ acts on the saturation point in the negative quadrant (the larger $a$, the lower the saturation point).

Constraining the parameters in the positive quadrant forces the activation to be a monotonic function, such that reducing the weight magnitude during training always lowers the neuron contribution.

Using this parameterization, the network can control its non-linear behavior throughout the training phase. It may increase the slope with $c$, the decay with $b$, or lower the saturation point with $a$. However, a standard gradient update on parameters $a$, $b$, $c$ would make the function non-differentiable at $x = 0$ and impair back-propagation. By equaling the derivatives on both sides of zero, solving for $c$ gives $c = a/b$ as a solution. The Parametric ELU (PELU) is defined as follows [248]:

$$\sigma_{\text{PELU}}(x) = \begin{cases} \frac{a}{b}\,x, & x \geq 0, \\ a\left(e^{\frac{x}{b}} - 1\right), & x < 0. \end{cases} \tag{8.40.1}$$

$$\frac{\partial}{\partial x}\sigma_{\text{PELU}}(x) = \begin{cases} \frac{a}{b}, & x \geq 0, \\ \frac{a}{b}e^{\frac{x}{b}}, & x < 0. \end{cases} \tag{8.40.2}$$

$$\frac{\partial}{\partial a}\sigma_{\text{PELU}}(x) = \begin{cases} \frac{x}{b}, & x \geq 0, \\ e^{\frac{x}{b}} - 1, & x < 0. \end{cases} \tag{8.40.3}$$

$$\frac{\partial}{\partial b}\sigma_{\text{PELU}}(x) = \begin{cases} -\frac{a}{b^2}\,x, & x \geq 0, \\ -\frac{a}{b^2}e^{\frac{x}{b}}, & x < 0. \end{cases} \tag{8.40.4}$$

With this parameterization, in addition to changing the saturation point and exponential decay respectively, both $a$ and $b$ adjust the slope of the linear function in the positive part to ensure differentiability at $x = 0$, (see Figure. 8.26). PELU is trained simultaneously with all the network parameters during back-propagation.

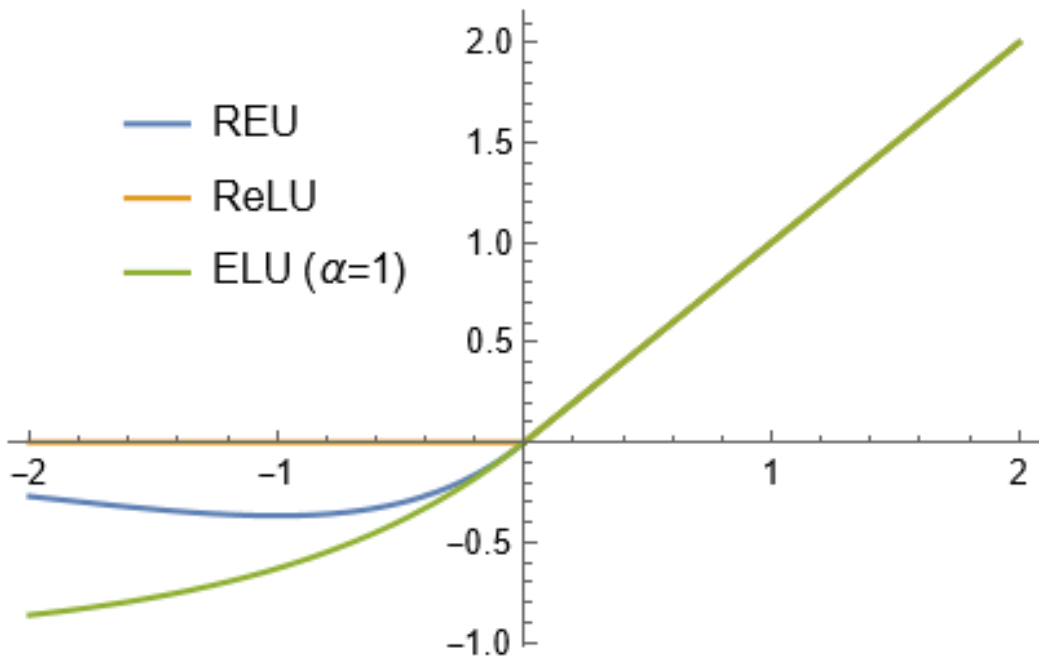

**Figure 8.27.** The figure compares three AFs: REU, ReLU, and ELU with $\alpha = 1$, plotted over a range of $x$ from $-2$ to 2. The REU function is defined as $x$ for $x > 0$ and $xe^x$ for $x \leq 0$, blending linear and exponential behaviors. The ReLU function outputs zero for negative inputs and linearly increases for positive inputs. The ELU function, with $\alpha = 1$, exponentially decays for negative inputs and linearly increases for positive inputs.

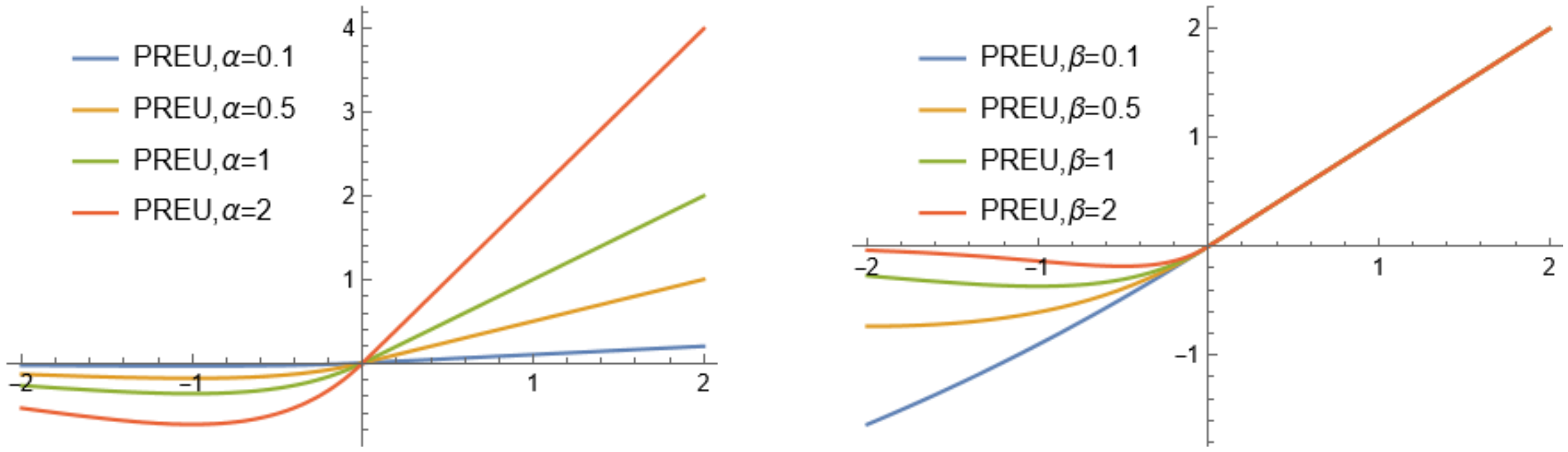

**Figure 8.28.** The figure contains two plots illustrating the behavior of the PREU AF over a range of $x$ from $-2$ to 2, highlighting the effects of varying parameters $\alpha$ and $\beta$. Left panel: This plot shows the PREU function for a fixed $\beta = 1$ and different values of $\alpha$ (0.1, 0.5, 1, 2). The PREU function is defined as $\alpha x$ for $x > 0$ and $\alpha x e^{\beta x}$ for $x \leq 0$. The plot demonstrates how increasing $\alpha$ affects the slope of the function for positive inputs and the curvature for negative inputs. Right panel: This plot presents the PREU function for a fixed $\alpha = 1$ and varying values of $\beta$ (0.1, 0.5, 1, 2). Here, the PREU function's behavior for positive inputs remains linear, while the curvature for negative inputs changes with different $\beta$ values. The plot shows how increasing $\beta$ intensifies the exponential growth for negative inputs.

The PELU AF is related to other parametric approaches in the literature. For example, PReLU learns a parameterization of the LReLU activation. PReLU learns a leak parameter in order to find a proper positive slope for negative inputs. This prevents negative neurons from dying. Based on the empirical evidence, learning the leak parameter $\alpha$ rather than setting it to a pre-defined value (as done in LReLU) improves performance.

### 8.4.4 Rectified Exponential and Parametric Rectified Exponential Units

The design choices in Rectified Exponential Unit (REU) [249] are motivated by the desire to combine the positive aspects of identity mapping for positive values with a non-monotonic behavior in the negative values. The definition of REU is given by (see Figure 8.27):

$$\sigma_{\text{REU}}(x) = \begin{cases} x, & x > 0, \\ xe^x, & x \leq 0. \end{cases} \tag{8.41}$$

**Remarks:**

- Similar to other popular AFs like ReLU, LReLU, PReLU, and ELU, REU maintains an identity mapping for positive input values ($x > 0$). This means that if the input is positive, the output is the same as the input, allowing the positive information to pass through unchanged.

- The introduction of a non-monotonic property in the negative part of the function is highlighted as a key feature. Non-monotonic functions have points where the slope changes direction, providing a more complex behavior compared to monotonic functions.
- REU employs an exponential function $e^x$ in the negative part $x \leq 0$. The rationale behind this choice is to retain more information in the negative part of the input. Exponential functions grow rapidly, and this characteristic may capture and amplify certain patterns or features in the negative range.

The Parametric Rectified Exponential Unit (PREU) [249] is a flexible AF with adjustable parameters $\alpha$ and $\beta$, providing control over the slope in the positive quadrant and the saturation in the negative quadrant. A PREU is designed as,

$$\sigma_{\text{PREU}}(x) = \begin{cases} \alpha\, x, & x > 0, \\ \alpha\, x e^{\beta x}, & x \leq 0, \end{cases} \tag{8.42.1}$$

having the output range in $[-1, \infty)$. Here $\alpha$ and $\beta$ can be fixed constants or trainable parameters. Each parameter controls a different aspect of PREU, as shown in Figure 8.28:

- $\alpha$ mainly controls the slope in the positive quadrant, influencing how quickly the function increases for positive values of $x$.
- $\beta$ controls the saturation in the negative quadrant, determining how quickly the function approaches zero for negative values of $x$.

The derivatives of PREU with respect to $x$, $\alpha$, and $\beta$ are given by

$$\frac{\partial}{\partial x}\sigma_{\text{PREU}}(x) = \begin{cases} \alpha, & x > 0, \\ \alpha(1+\beta x)e^{\beta x}, & x \leq 0. \end{cases} \tag{8.42.2}$$

$$\frac{\partial}{\partial \alpha}\sigma_{\text{PREU}}(x) = \begin{cases} x, & x > 0, \\ x e^{\beta x}, & x \leq 0. \end{cases} \tag{8.42.3}$$

$$\frac{\partial}{\partial \beta}\sigma_{\text{PREU}}(x) = \begin{cases} 0, & x > 0, \\ \alpha x^2 e^{\beta x}, & x \leq 0. \end{cases} \tag{8.42.4}$$

### 8.4.5 Elastic ELU

AFs play important roles in determining the depth and non-linearity of deep learning models. Since the ReLU was introduced, many modifications have been proposed to avoid overfitting. ELU and their variants, with trainable parameters, have been proposed to reduce the bias shift effects which are often observed in ReLU-type AFs. The Elastic Exponential Linear Unit (EELU) [250] combines the advantages of both types of AFs in a generalized form. EELU changes the positive slope to prevent overfitting, as do EReLU and RReLU and also preserves the negative signal to reduce the bias shift effect, as does ELU. However, the positive slope of EELU is modified from the Gaussian distribution with a randomized standard deviation, instead using a simple uniform distribution to determine the scale of random noise, like EReLU and RReLU. The EELU is defined as,

$$\sigma_{\text{EELU}}(x) = \begin{cases} k\, x, & x > 0, \\ \alpha\left(e^{\beta x} - 1\right), & x \leq 0, \end{cases} \tag{8.43}$$

having the output range in $[-\alpha, \infty)$ where $\alpha$ and $\beta$ are the trainable parameters.

**Remarks:**

- The coefficient $k$ plays an important role in EELU. In the training stage, the coefficient is sampled from a Gaussian distribution with a random standard deviation and a fixed mean; the sampled coefficient is truncated from 0 to 2 because the Gaussian distribution ranges from $-\infty$ to $+\infty$ . The standard deviation $\sigma$ of the Gaussian distribution is randomly chosen from a uniform distribution, instead of using a fixed parameter. It is denoted by:

$$k = \max(0, \min(s, 2)), \quad s \sim N(1, \sigma), \tag{8.44.1}$$

$$\sigma \sim U(0, \epsilon), \quad \epsilon \in (0,1]. \tag{8.44.2}$$

Here, $\epsilon$ is a hyperparameter.

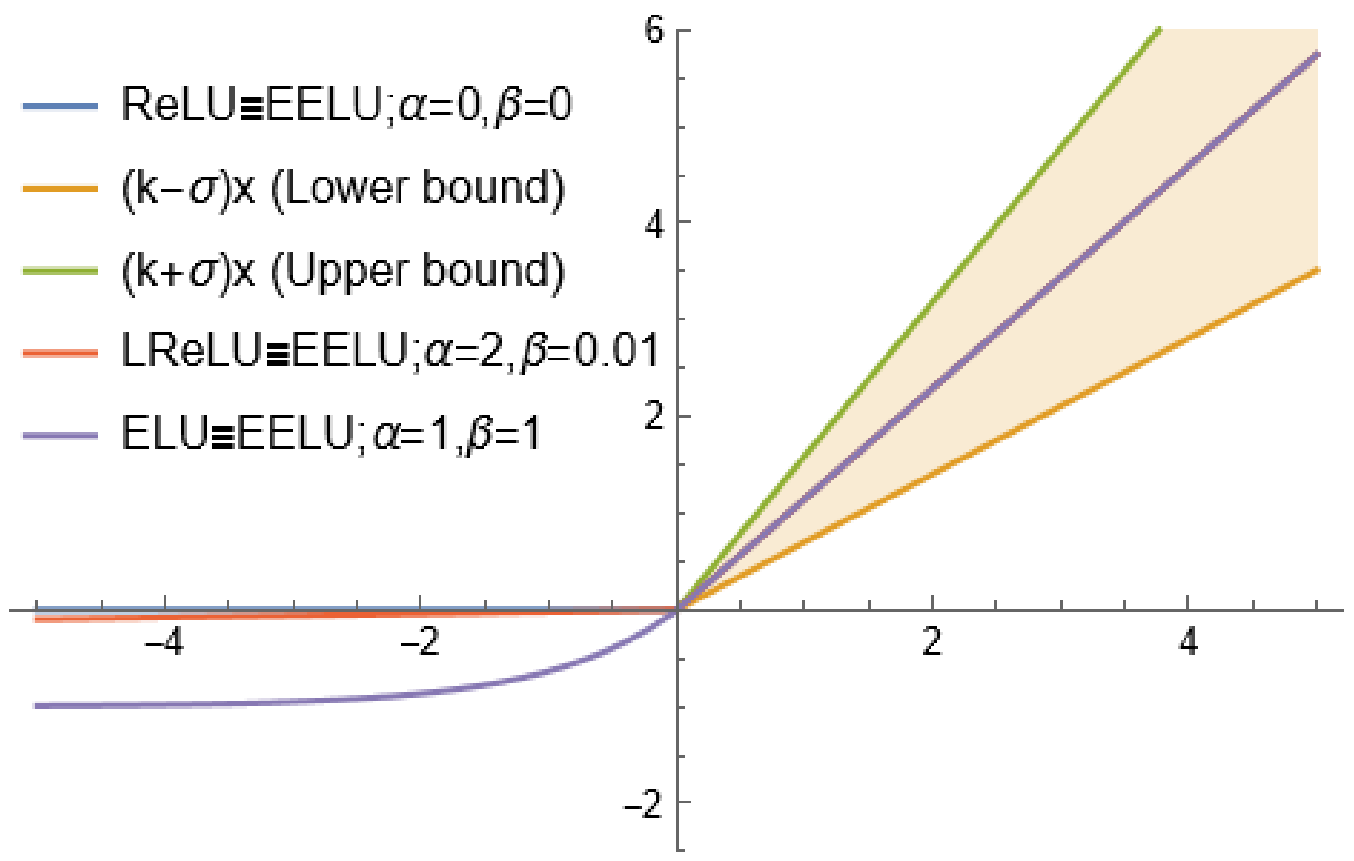

**Figure 8.29.** The figure depicts the EELU AF, illustrating its versatility through five different curves across an $x$ range of $-5$ to 5. The plot demonstrates the function's linear behavior similar to ReLU for positive $x$ values, modifiable through parameters $\alpha$, $\beta$, and $k$, and includes bounds to show the variability due to the sampled $k$ value. Additionally, the figure showcases variations that mimic LReLU and ELU for negative $x$ values, thereby highlighting the function's adaptability to different neuron behaviors in a NN. The filled area between the lower and upper bounds emphasizes the potential variability in the linear portion.

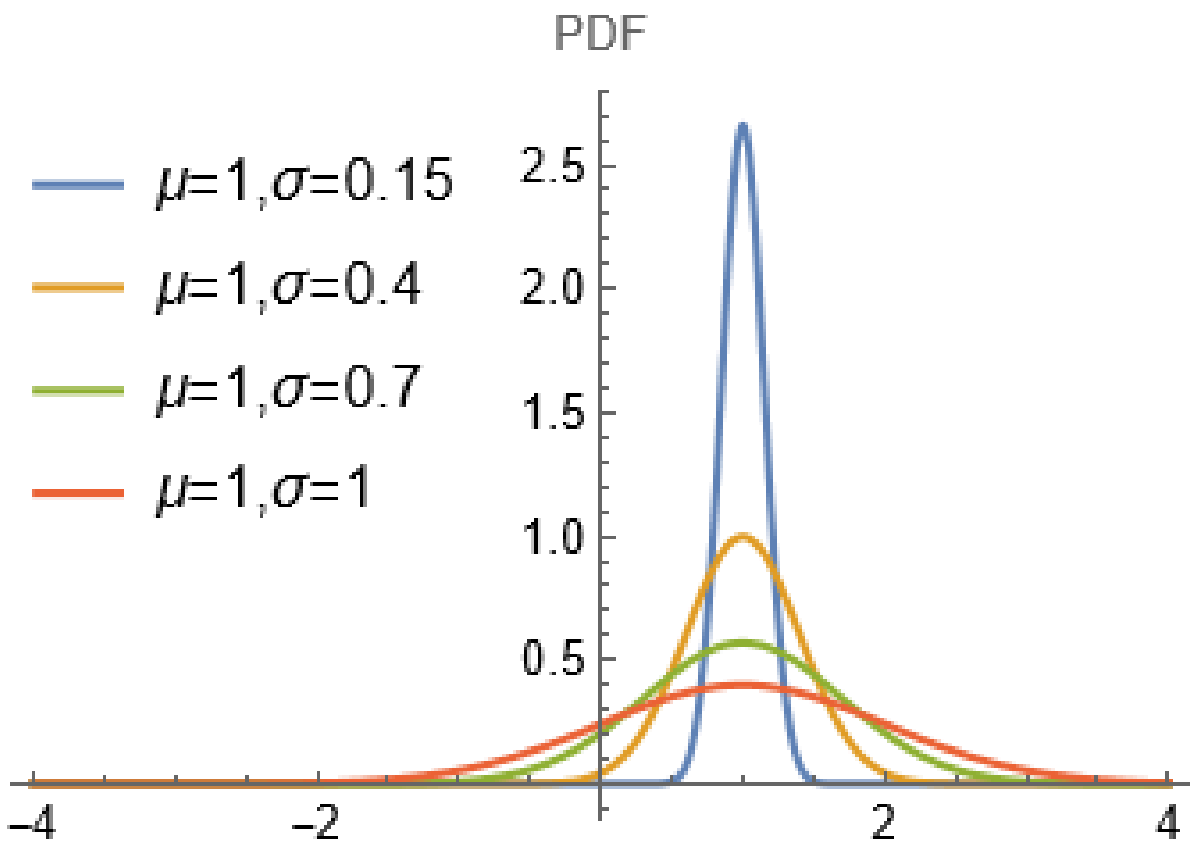

**Figure 8.30.** The figure displays the PDFs of four normal distributions, each with a mean ($\mu$) of 1 but different standard deviations ($\sigma$) of 0.15, 0.4, 0.7, and 1, plotted over a range of $x$ from $-4$ to 4. As $\sigma$ increases, the curves become wider and flatter, indicating greater variability around the mean.

- In the test stage, EELU replaces $k$ with $\mathbb{E}(k)$ (the expectation of $k$) in the positive part.
- Figure 8.29 shows how EELU works in the training stage and the test stage. EELU is divided into two parts, one for positive and one for negative inputs. The slope of the positive part is varied using a Gaussian distribution with a randomized standard deviation. This approach is similar to inserting noise because it outputs similar input features to various output features. The negative input values are determined by $\alpha$ and $\beta$. $\alpha$ and $\beta$ are the learning parameters determined from the training samples and are constrained to be greater than zero.
- EELU can represent various AFs such as ReLU, LReLU, PReLU, and ELU.
- The advantage of EELU is that it can represent various output features with random noise. The random noise is independent of the inputs and can confer sensitivity of the NNs to a wide variety of inputs.

**Table 8.1.** Probability of a value of less than zero from the Gaussian distribution

| Standard Deviation | Probability |
|---|---|
| 0.1 | 0.00% |
| 0.2 | 0.00% |
| 0.3 | 0.04% |
| 0.4 | 0.62% |
| 0.5 | 2.28% |
| 0.6 | 4.78% |
| 0.7 | 7.66% |
| 0.8 | 10.56% |
| 0.9 | 13.33% |
| 1.0 | 15.87% |

- Gaussian distributions with different standard deviations are shown in Figure 8.30. The probabilities of a negative $k$ with different standard deviations are shown in Table 8.1. The larger the standard deviation, the greater the probability that the sampled parameter is less than zero. If the sampled value is negative, input signals are discarded and the hidden unit is deactivated. The neuronal noise, which is simulated by various scales and random deactivation, works to regularize DNNs to learn various latent representations from training samples.

## 8.5 Miscellaneous AFs

### 8.5.1 Swish

The introduction of Swish was part of a broader effort to explore and discover new AFs. The authors of the paper "Searching for activation functions" [251] used a neural architecture search approach to automatically discover AFs that performed well on certain tasks (a combination of exhaustive and reinforcement learning-based search). The paper contributed to the ongoing research in the deep learning community by highlighting the potential benefits of automatic methods for discovering NN architectures and components. The introduction of Swish and similar approaches demonstrated that there might be room for improvement beyond traditional AFs, and automatic search methods could be valuable in this exploration.

Swish [251] is defined as

$$\sigma_{\text{Swish}}(x) = x \cdot \sigma_{\text{Sigmoid}}(\beta x), \tag{8.45.1}$$

where $\sigma_{\text{Sigmoid}}(x) = (1 + \exp(-x))^{-1}$ is the Sigmoid function and $\beta$ is either a constant or a trainable parameter. The output range of Swish is $(-\infty, \infty)$. Swish is a smooth and differentiable AF. Smoothness and differentiability are desirable properties in NN AFs because they facilitate gradient-based optimization during the training process. Figure 8.31 plots the graph of Swish for different values of $\beta$.

- If $\beta = 1$, Swish is equivalent to the SiLU (8.15).
- If $\beta = 0$, Swish becomes the scaled linear function $\sigma(x) = \frac{x}{2}$.
- As $\beta \to \infty$, the Sigmoid component approaches a $0 - 1$ function, so Swish becomes like the ReLU function.

This suggests that Swish can be loosely viewed as a smooth function that nonlinearly interpolates between the linear function and the ReLU function. The degree of interpolation can be controlled by the model if $\beta$ is set as a trainable parameter.

Like ReLU, Swish is unbounded above and bounded below. Unlike ReLU, Swish is smooth and non-monotonic. In fact, the non-monotonicity property of Swish distinguishes it from most common AFs. The derivative of Swish is

$$\begin{aligned}\frac{\partial}{\partial x}\sigma_{\text{Swish}}(x) &= \sigma_{\text{Sigmoid}}(\beta x) + x\,\beta\,\sigma_{\text{Sigmoid}}(\beta x)\left(1-\sigma_{\text{Sigmoid}}(\beta x)\right)\\ &= \sigma_{\text{Sigmoid}}(\beta x) + x\,\beta\,\sigma_{\text{Sigmoid}}(\beta x) - x\,\beta\left[\sigma_{\text{Sigmoid}}(\beta x)\right]^2\\ &= x\,\beta\,\sigma_{\text{Sigmoid}}(\beta x) + \sigma_{\text{Sigmoid}}(\beta x)\left(1 - x\,\beta\,\sigma_{\text{Sigmoid}}(\beta x)\right)\\ &= \beta\,\sigma_{\text{Swish}}(x) + \sigma_{\text{Sigmoid}}(\beta x)\left(1-\sigma_{\text{Swish}}(x)\right).\end{aligned} \quad (8.45.2)$$

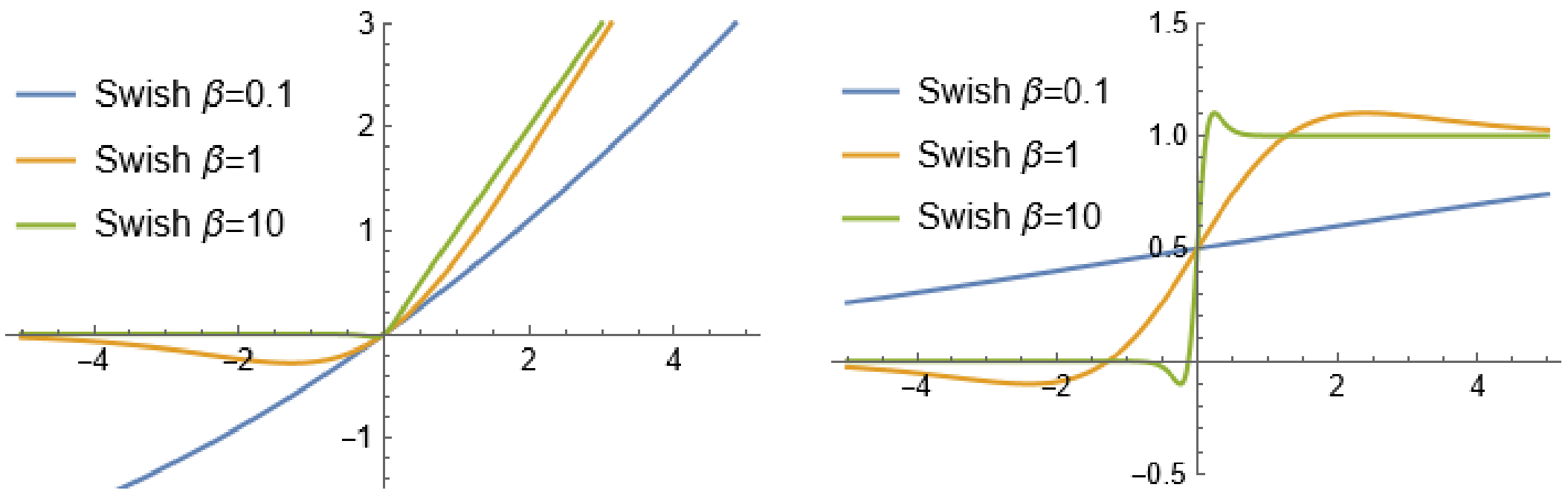

**Figure 8.31.** Left panel: This plot visualizes the Swish AF, a smooth, non-monotonic function, for three different values of the parameter $\beta$ (0.1, 1, and 10) over a range of $x$ from $-5$ to 5. The function is defined as $x\,\sigma_{\text{Sigmoid}}(\beta x)$. As $\beta$ increases, the Swish function transitions from a nearly linear behavior ($\beta = 0.1$) to a more pronounced Sigmoid-like curve ($\beta = 10$), illustrating its adaptability and the impact of $\beta$ on the activation output. Right panel: The plot shows the first derivatives of the Swish function for the same $\beta$ values (0.1, 1, and 10). Each derivative curve illustrates the rate of change of the Swish function at different points, highlighting how the responsiveness of the function varies with changes in $\beta$. For lower $\beta$, the derivative remains close to constant, while for higher $\beta$, the curve exhibits sharper changes, especially near zero.

The first derivative of Swish is shown in Figure 8.31 for different values of $\beta$. The scale of $\beta$ controls how fast the first derivative asymptotes to 0 and 1.

When $\beta = 1$, the derivative has a magnitude less than 1 for inputs that are less than around 1.25. Thus, the success of Swish with $\beta = 1$ implies that the gradient-preserving property of ReLU (i.e., having a derivative of 1 when $x > 0$) may no longer be a distinct advantage in modern architectures.

**Remarks:**

- The most striking difference between Swish and ReLU is the non-monotonic "bump" of Swish when $x < 0$. The shape of the bump can be controlled by changing the $\beta$ parameter. The smaller and higher values of $\beta$ lead toward the linear and ReLU functions, respectively. Thus, it can control the amount of non-linearity based on the dataset and network complexity.
- While Swish showed promise in terms of performance improvements, it is important to consider the computational cost. Swish involves the computation of the Sigmoid function, which might be more computationally expensive compared to simpler AFs like ReLU.

### 8.5.2 E-Swish

Swish is also extended to E-Swish by multiplying the Swish with a learnable parameter to control the slope in the positive direction [252]. The E-Swish is defined as,

$$\sigma_{\text{E-Swish}}(x) = \beta \cdot x \cdot \sigma_{\text{Sigmoid}}(x), \quad (8.46.1)$$

having the output the range in $(-\infty, \infty)$ and $\beta$ is trainable parameter and $1 \le \beta \le 2$. The properties of E-swish are very similar to the ones of Swish, see Figure 8.32. In fact, when $\beta = 1$, E-swish reverts to Swish. Like both ReLU and Swish, E-swish is unbounded above and bounded below. Like Swish, it is also smooth and non-monotonic. The property of non-monotonicity is almost exclusive of Swish and E-swish. Another exclusive feature of both, Swish and E-swish, is that there is a region where the derivative is greater than 1.

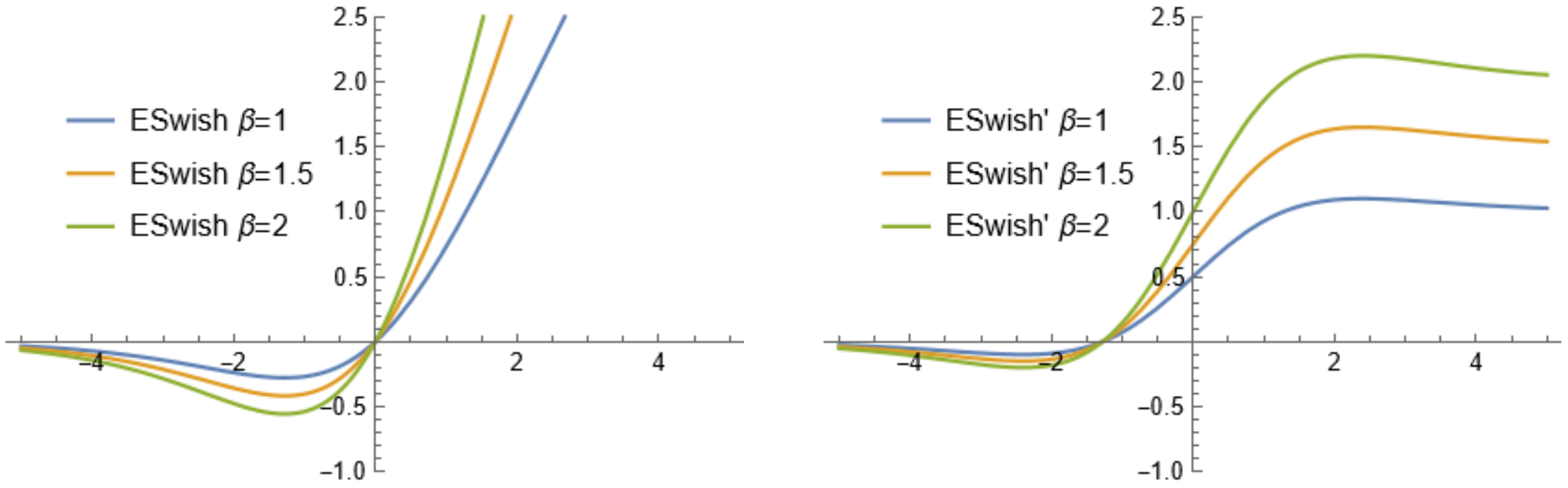

**Figure 8.32.** Left panel: This plot displays the E-Swish AF, which is a variant of the Swish function enhanced by a scaling parameter $\beta$, plotted for three values of $\beta$ (1, 1.5, and 2) over a range of $x$ from $-5$ to 5. The E-Swish function is defined as $\beta\, x\, \sigma_{\text{Sigmoid}}(x)$. As $\beta$ increases, the amplitude of the E-Swish function also increases, showing a more pronounced curve that enhances the activation's capability to manage different signal strengths in NNs. Right panel: This plot illustrates the first derivatives of the E-Swish function for the same set of $\beta$ values. These derivatives highlight how the rate of change of the E-Swish function varies with $\beta$, indicating the function's sensitivity at different points. For higher $\beta$, the derivative shows more pronounced peaks, suggesting a more responsive behavior at certain ranges of $x$.

The derivative of E-swish is:

$$\begin{aligned}\frac{\partial}{\partial x}\sigma_{\text{E-Swish}}(x) &= \beta\sigma_{\text{Sigmoid}}(x) + x\,\beta\,\sigma_{\text{Sigmoid}}(x)\left(1-\sigma_{\text{Sigmoid}}(x)\right)\\ &= \beta\sigma_{\text{Sigmoid}}(x) + x\,\beta\,\sigma_{\text{Sigmoid}}(x) - x\,\beta\left[\sigma_{\text{Sigmoid}}(x)\right]^2\\ &= \beta\sigma_{\text{Sigmoid}}(x) + \sigma_{\text{E-Swish}}(x) - \sigma_{\text{E-Swish}}(x)\sigma_{\text{Sigmoid}}(x)\\ &= \sigma_{\text{E-Swish}}(x) + \sigma_{\text{Sigmoid}}(x)\left(\beta - \sigma_{\text{E-Swish}}(x)\right).\end{aligned} \quad (8.46.2)$$

The fact that the gradients for the negative part of the function approaches zero can also be observed in Swish and ReLU activations. However, the particular shape of the curve described in the negative part, which gives both Swish and E-swish the non-monotonicity property, improves performance since they can output small negative numbers, unlike ReLU.

### 8.5.3 HardSwish

Hard-Swish [253] is closely related to AF Swish. It is defined as

$$\sigma_{\text{HardSigmoid}}(x) = \max(0, \min(1, (0.2x + 0.5)), \quad (8.47.1)$$

$$\begin{aligned}\sigma_{\text{HardSwish}}(x) &= 2x\,\sigma_{\text{HardSigmoid}}(\beta x)\\ &= 2x\ \max(0, \min(1, (0.2\beta x + 0.5))\\ &= x \min\frac{[\max[x+3,0], 6]}{6}\\ &= \begin{cases} 0, & x \le -3,\\ x, & x \ge 3,\\ \dfrac{x(x+3)}{6}, & \text{otherwise,}\end{cases}\end{aligned} \quad (8.47.2)$$

where $\beta$, is either a trainable parameter or a constant. As $\beta \to \infty$, the hard-Sigmoid component approaches $0-1$, and Hard-Swish will act like the ReLU AF. This indicates that Hard-Swish interpolates non-linearly between the ReLU function and linear function smoothly. Setting $\beta$, as a trainable parameter can be used to control the degree of interpolation in the model.

The properties of Hard-Swish are similar to Swish because both are unbounded above and bounded below. It is non-monotonic. It is faster in computation compared to swish because it doesn't involve any exponential calculation. The particular shape of the curve in the negative part improves performance as it can output small negative numbers.

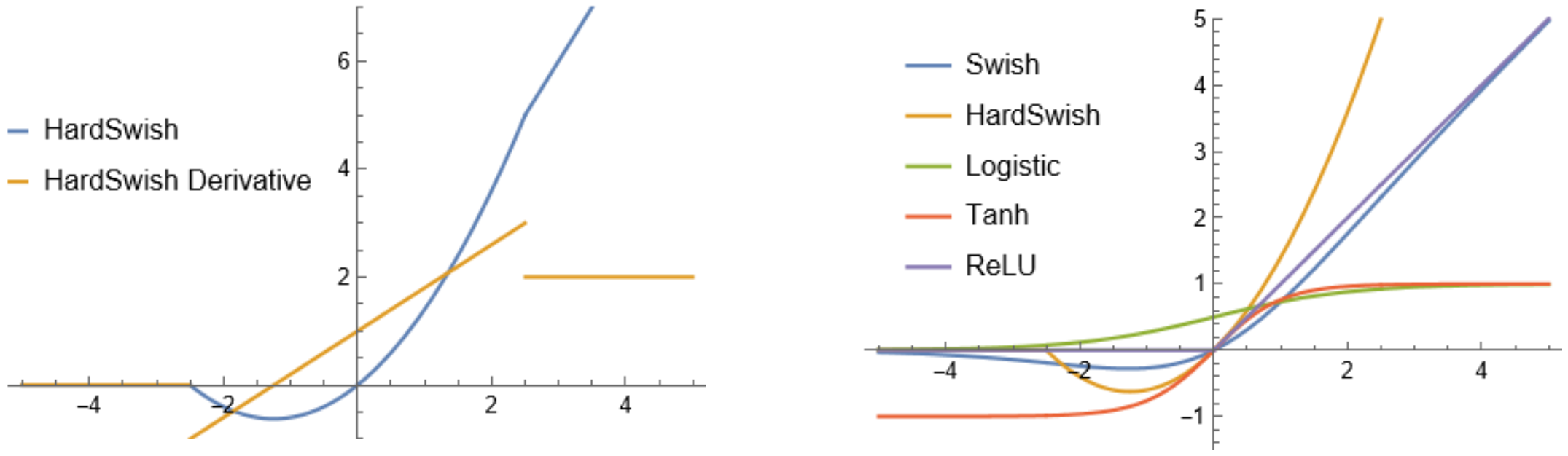

**Figure 8.33.** Left panel: The figure displays the HardSwish AF and its derivative, plotted over a range of $x$ from $-5$ to 5. HardSwish is a piecewise linear approximation of the Swish function, designed to provide computational benefits while retaining non-linear characteristics beneficial for deep learning models. The AF (HardSwish) shows a smoother transition compared to traditional ReLU, incorporating a non-linear region around zero, and it smoothly transitions to linear regions for positive and negative values. Its derivative, plotted alongside, highlights how the gradient varies across different values of $x$, with distinct segments showing where the function is actively adjusting its slope. Right panel: The figure provides a comprehensive comparison of five popular AFs used in NNs: Swish, HardSwish, Logistic Sigmoid, Tanh, and ReLU, plotted over a range of $x$ from $-5$ to 5. Each function demonstrates distinct characteristics: Swish shows a smooth, Sigmoid-like response; HardSwish offers a piecewise linear, computationally efficient approximation of Swish; Logistic Sigmoid illustrates a classic S-shaped curve; Tanh presents a symmetric Sigmoid centered around zero; and ReLU activates linearly for positive inputs while clamping negative values to zero.

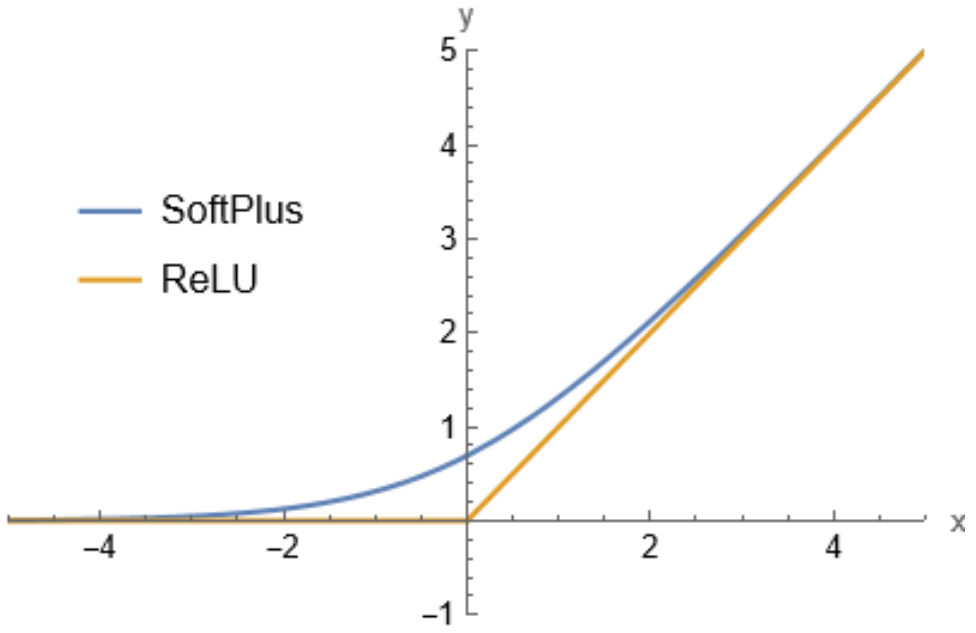

**Figure 8.34.** The figure compares the SoftPlus and ReLU AFs, plotted over a range of $x$ from $-5$ to 5. SoftPlus, a smooth approximation of ReLU, gradually transitions from near zero to a linearly increasing function, displaying a continuous and differentiable curve. In contrast, ReLU features a sharp threshold at $x = 0$, where it transitions from zero for negative inputs to a linear response for positive inputs.

The non-monotonic bump is the most striking difference between Hard-Swish and other AF when $x$ is less than 0 as shown in Figure 8.33. Inside the domain of the bump ($2.5 \leq x \leq 0$), a large percentage of preactivations fall leading to a better convergence and improvement on benchmarks.

### 8.5.4 SoftPlus

The SoftPlus function [254] was proposed in 2001 and is mostly used in statistical applications. SoftPlus unit-based AF is also used in DNNs [255]. As a smoothing version of the ReLU function, Figure 8.34, the SoftPlus function is defined as:

$$\sigma_{\text{SoftPlus}}(x) = \log(e^x + 1). \tag{8.48.1}$$

The SoftPlus has a number of advantages.

- The SoftPlus function is smooth and differentiable everywhere, which makes it well-suited for gradient-based optimization techniques like backpropagation. This property makes the SoftPlus function more stable no matter when being estimated from the positive and negative directions, while ReLU has a discontinuous gradient at point 0.

- The SoftPlus function maps its input to a range between 0 and positive infinity, which can be useful in certain situations.
- The SoftPlus function is a monotonic function, meaning that as the input $x$ increases, the output also increases.
- Unlike some other AFs, such as the Sigmoid or Tanh, the SoftPlus function does not saturate for large positive inputs. Saturated AFs can lead to vanishing gradients and slow down the learning process.
- Another advantage is that the SoftPlus unit has a non-zero gradient while the input of the unit is negative. Unlike ReLU propagates no gradient in $x < 0$, the SoftPlus function can propagate gradients throughout all real inputs.
- Unlike the ReLU AF, the SoftPlus function does not suffer from the "dying ReLU" problem, where neurons can become inactive and not update their weights during training.
- The derivative of the SoftPlus function with respect to its input $x$ can be computed as follows:

$$\frac{\partial}{\partial x}\sigma_{\text{SoftPlus}}(x) = \frac{e^x}{1+e^x} = \sigma_{\text{Sigmoid}}(x). \tag{8.48.2}$$

  This derivative can help train NNs using gradient-based optimization algorithms like backpropagation.
- The SoftPlus unit also outperforms the Sigmoid unit in the following aspects. The derivative of SoftPlus is a Sigmoid function. It means that the gradient of the SoftPlus unit approaches 1 when the input increases, which largely reduces the bad effects of the vanishing gradient problem.
- It is important to note that, while there may be concerns about the hard saturation of ReLU and the resulting zero gradients, experimental evidence [235] suggests that this characteristic might not hinder supervised training as much as initially thought. The use of SoftPlus as an alternative does not necessarily provide a clear advantage in all cases, and the benefits of hard zeros in ReLU might outweigh the drawbacks in certain scenarios, particularly when some hidden units remain active during training. The hard non-linearities do not hurt so long as the gradient can propagate along some paths, i.e., that some of the hidden units in each layer are non-zero.

### 8.5.5 SoftPlus Linear Unit (SLU)

A SoftPlus Linear Unit (SLU) was also proposed by considering SoftPlus with a rectified unit [256]. It is known that a zero gradient problem and a bias shift exist when ReLU is used in networks. Based on the theory that “zero mean activations improve learning ability”, the SLU was proposed as an adaptive AF. The SLU AF is defined as,

$$\sigma_{\text{SLU}}(x) = \begin{cases} \beta x, & x \geq 0, \\ \alpha \log(e^x + 1) + \gamma, & x < 0, \end{cases} \tag{8.49.1}$$

where $\beta$ is the slope factor in the positive part. The larger the slope factor $\beta$, the steeper the slope of the positive part is. The parameter $\alpha$ relates to the location of the saturation point in the negative quadrant. The larger the $\alpha$, the lower the saturation point is. The parameter $\gamma$ denotes the distance to the horizontal axis in the negative part, and the bigger the value of $\gamma$, the larger the distance is. Obviously, the negative part of the SLU is always less than zero. The average value of the SLU reduces the bias shift compared to the ReLU.

The derivative of SLU can be calculated as:

$$\frac{\partial}{\partial x}\sigma_{\text{SLU}}(x) = \begin{cases} \beta, & x \geq 0, \\ \dfrac{\alpha}{e^{-x}+1}, & x < 0. \end{cases} \tag{8.49.2}$$

The negative part is the Sigmoid function multiplied by a constant $\alpha$. In order to ensure that the function is continuous and differentiable at zero, the parameters were constrained as follows:

$$\lim_{x\to 0^+} \sigma_{\text{SLU}}(x) = 0, \tag{8.50.1}$$

$$\lim_{x\to 0^-} \sigma_{\text{SLU}}(x) = \alpha \log 2 - \gamma, \tag{8.50.2}$$

$$\lim_{x\to 0^+} \sigma_{\text{SLU}}(x) = \lim_{x\to 0^-} \sigma_{\text{SLU}}(x), \tag{8.50.3}$$

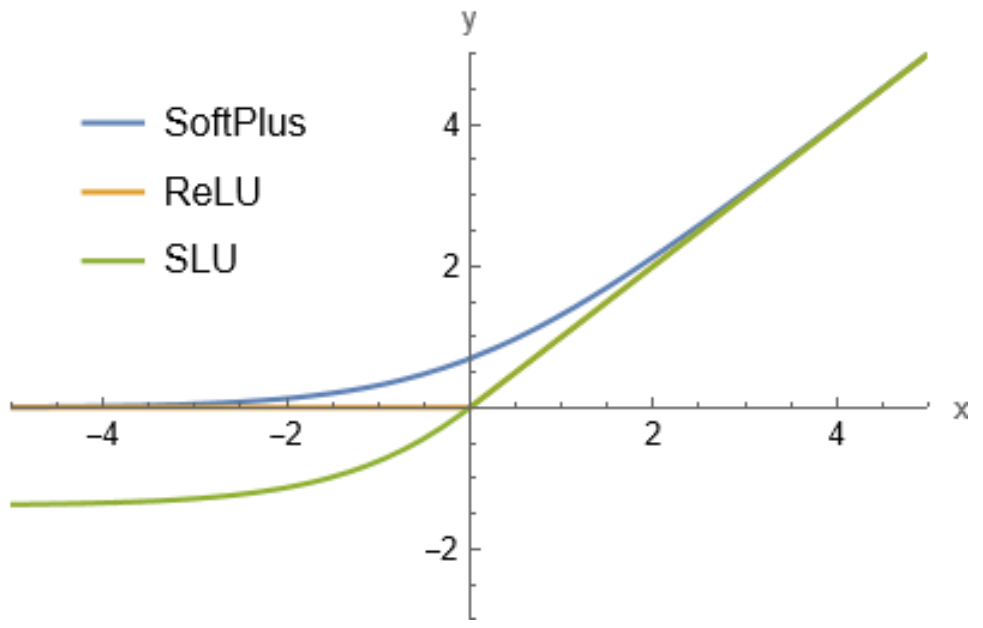

**Figure 8.35.** The figure presents a comparative view of three AFs: SoftPlus, ReLU, and SLU, plotted over a range of $x$ from $-5$ to 5. SoftPlus offers a smooth, gradual transition, acting as a continuously differentiable approximation of the ReLU function, which directly clamps all negative values to zero and linearly passes positive values. SLU introduces a unique behavior where it behaves linearly for non-negative inputs and transitions to a Sigmoid-logarithmic form for negative inputs, blending properties of the Sigmoid and logarithmic functions for a softer response compared to ReLU.

$$\lim_{x\to 0^+} \sigma'_{\text{SLU}}(x) = \beta, \tag{8.50.4}$$

$$\lim_{x\to 0^-} \sigma'_{\text{SLU}}(x) = \frac{\alpha}{2}, \tag{8.50.5}$$

$$\lim_{x\to 0^+} \sigma'_{\text{SLU}}(x) = \lim_{x\to 0^-} \sigma'_{\text{SLU}}(x). \tag{8.50.6}$$

By solving (8.50.3) and (8.50.6), the following is obtained:

$$\gamma = \alpha \log 2, \tag{8.51.1}$$

$$\beta = \frac{\alpha}{2}. \tag{8.51.2}$$

In order to avoid a vanishing or exploding gradient during back propagation, the slope factor $\beta$ should be constrained by: $\beta = 1$. Hence, $\alpha = 2$ and $\gamma = 2\log 2$. Therefore, the precise definition of SLU, Figure 8.35, is:

$$\sigma_{\text{SLU}}(x) = \begin{cases} x, & x \geq 0, \\ 2\log\left(\dfrac{e^x + 1}{2}\right), & x < 0. \end{cases} \tag{8.52}$$

### 8.5.6 Mish

The SoftPlus function is also used with Tanh function in Mish AF [257]. Mish, as visualized in Figure 8.36, is a smooth, continuous, self-regularized, non-monotonic AF mathematically defined as:

$$\sigma_{\text{Mish}}(x) = x\,\text{Tanh}(\sigma_{\text{SoftPlus}}(x)) = x\,\text{Tanh}(\ln(1 + e^x)). \tag{8.53.1}$$

Similar to Swish, Mish is bounded below and unbounded above with a range of $[\approx -0.31, \infty)$. The first derivative of Mish, as shown in Figure 8.36, can be defined as:

$$\frac{\partial}{\partial x}\sigma_{\text{Mish}}(x) = \frac{\omega e^x}{\delta}, \tag{8.53.2}$$

where, $\omega = 4(x+1) + 4e^{2x} + e^{3x} + e^x(4x+6)$ and $\delta = 2e^x + e^{2x} + 2$.

The Mish has a number of advantages.

- Due to the preservation of a small amount of negative information, Mish eliminated by design the preconditions necessary for the Dying ReLU phenomenon. This property helps in better expressivity and information flow.
- Being unbounded above, Mish avoids saturation, which generally causes training to slow down due to near-zero gradients drastically.
- Being bounded below is also advantageous since it results in strong regularization effects.
- Unlike ReLU, Mish is continuously differentiable, a property that is preferable because it avoids singularities and, therefore, undesired side effects when performing gradient-based optimization.

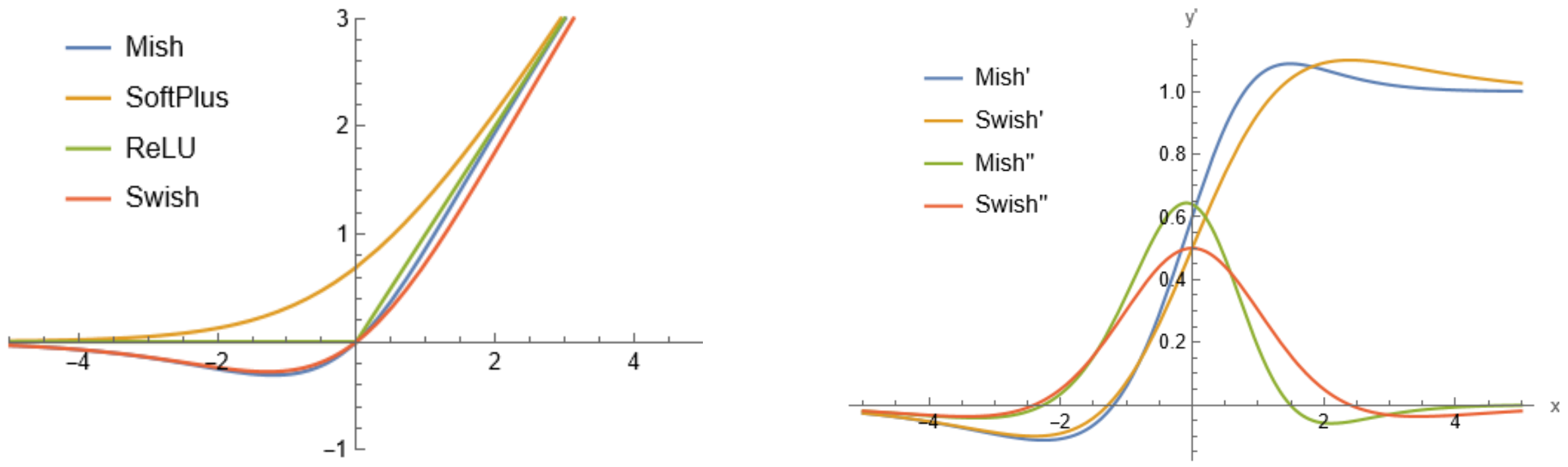

**Figure 8.36.** Left panel: The figure illustrates a comparison of four advanced AFs: Mish, SoftPlus, ReLU, and Swish, each plotted over a range of $x$ from $-5$ to $5$. Mish, defined as $x\,\mathrm{Tanh}(\ln(1+e^{x}))$, showcases a smooth and non-monotonic curve that closely resembles the behavior of Swish but with subtle differences in negative input handling. SoftPlus, a smooth approximation of ReLU, shows a gentle logistic-like transition, emphasizing its continuous and differentiable nature. ReLU remains the simplest, with a direct zero-clamping behavior for negative inputs and linear response for positive values. Swish, defined as $x \cdot \sigma_{\mathrm{Sigmoid}}(\beta x)$ with $\beta = 1$, combines aspects of linear and sigmoidal responses, providing a flexible shape that varies with the input value. Right panel: The figure presents a detailed comparison of the first and second derivatives of the Mish and Swish AFs, plotted over a range of $x$ from $-5$ to $5$. The first derivatives, labeled as "Mish'" and "Swish'", show the rate of change of the respective AFs, illustrating how they respond to different input values. The second derivatives, labeled as "Mish''" and "Swish''", indicate the curvature or the rate of change of the slope of the AFs, providing deeper insights into their dynamic behavior.

- The Mish function is non-monotonic, meaning it does not follow a simple increasing or decreasing pattern. This can help capture complex and non-linear relationships in data.
- Like the SoftPlus function, the Mish function is smooth and differentiable everywhere, making it suitable for gradient-based optimization methods such as backpropagation.
- The Mish function has a built-in self-regularization property, meaning that it resists very large input values, helping to avoid exploding gradients during training.
- Mish has shown competitive performance in DNNs compared to other popular AFs, such as ReLU and LReLU, in some experimental settings.
- However, the increased complexity in Mish due to the multiple functions can be a limitation for the DNNs.

### 8.5.7 Gaussian Error Linear Unit

The motivation behind Gaussian Error Linear Unit (GELU) [258] is to bridge stochastic regularizers, such as dropout, with non-linearities.

- Dropout is a regularization technique used in NNs to prevent overfitting. The idea behind dropout is to randomly "drop out" (i.e., set to zero) a fraction of the units/neurons in a layer during training. This prevents individual neurons from becoming overly specialized and encourages more robust learning. When dropout is applied stochastically during training, it means that for each training example, a random subset of neurons is dropped out. In other words, dropout regularization stochastically multiplies a neuron's inputs with $0$, randomly rendering them inactive. This suggests a more probabilistic view of a neuron's output.
  - On the other hand, ReLU activation deterministically multiplies inputs with $0$ or $1$ dependent upon the input's value.

GELU merges both functionalities by multiplying inputs by a value from $0$ to $1$. However, the value of this zero-one mask, while stochastically determined, is also dependent upon the input's value. The GELU AF is

$$\sigma_{\mathrm{GELU}}(x) = x\,\Phi(x) = x\,P(X \le x), \tag{8.54.1}$$

where $\Phi(x)$ the standard Gaussian cumulative distribution function. The GELU AF is defined as $x$ times the standard Gaussian cumulative distribution function of $x$, which can be written as $x$ times the probability that a random variable from a normal distribution with mean $0$ and variance $1$ is less than or equal to $x$.

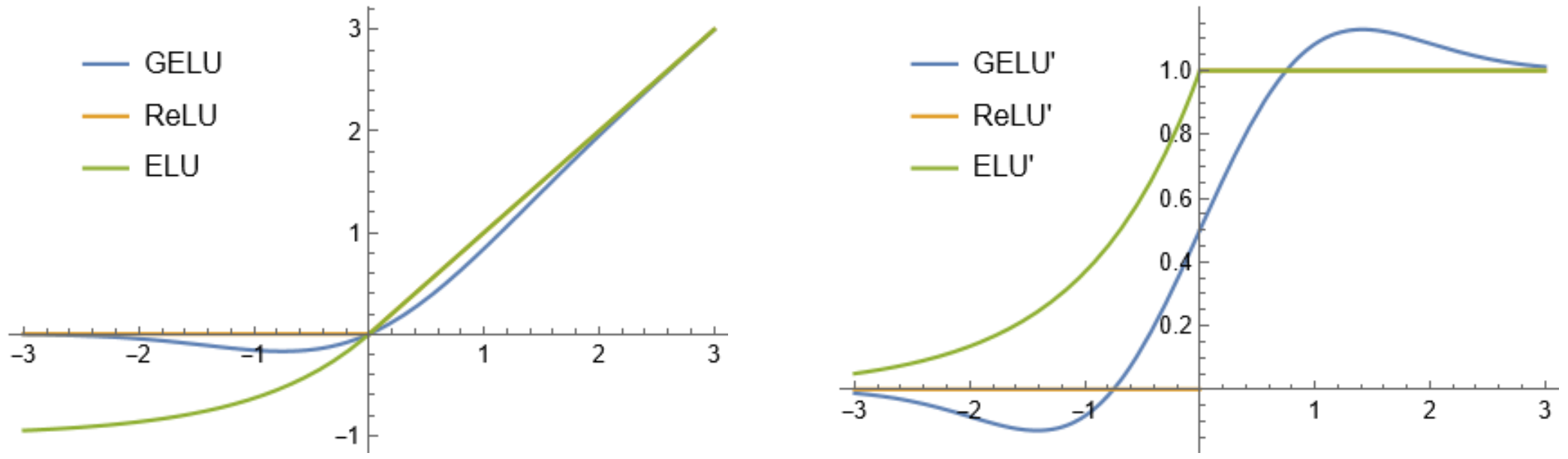

**Figure 8.37.** Left panel: The figure provides a comparative visualization of three AFs: GELU, ReLU, and ELU, plotted over a range of $x$ from $-3$ to 3. The GELU function, defined as $\frac{x}{2}\left(1+\text{erf}\left(\frac{x}{\sqrt{2}}\right)\right)$, offers a smooth, probabilistic approach to activation by incorporating the error function (erf). ReLU, the simplest function, activates linearly for positive $x$ values and clamps negative values to zero, illustrating a sharp transition at zero. ELU, with $\alpha = 1$, combines linear behavior for positive $x$ values and exponential decay for negative $x$, providing a smooth transition that aims to mitigate the dying ReLU problem. Right panel: The figure displays the first derivatives of three AFs: GELU, ReLU, and ELU, plotted over a range of $x$ from $-3$ to 3. The derivative of the GELU function, labeled as "GELU'", demonstrates a smooth, S-shaped curve reflecting its probabilistic nature and gradual transitions. The ReLU derivative, labeled as "ReLU'", is a piecewise function that is zero for negative $x$ and one for positive $x$, showing a sharp change at zero. The ELU derivative, labeled as "ELU'" with $\alpha = 1$, combines a constant positive slope for $x > 0$ and an exponential rise for $x \leq 0$, offering a smoother gradient flow compared to ReLU.

In this setting, inputs have a higher probability of being "dropped" as $x$ decreases, so the transformation applied to $x$ is stochastic yet depends upon the input. The GELU nonlinearity weights inputs by their value, rather than gates inputs by their sign as in ReLUs. So the transformation applied by GELU is stochastic, yet it depends upon the input's value through $\Phi(x)$.

In fact, the GELU can be viewed as a way to smooth a ReLU. To see this, recall that $\sigma_{\text{ReLU}}(x) = \max(x, 0) = x\mathbf{1}(x > 0)$ (where $\mathbf{1}$ is the indicator function), while the GELU is $\sigma_{\text{GELU}}(x) = x\,\Phi(x)$ if $\mu = 0, \sigma = 1$. Then the CDF is a smooth approximation to the binary function the ReLU uses, like how the Sigmoid smoothed binary threshold activations. Unlike the ReLU, the GELU and ELU can be both negative and positive. In Figure 8.37, observe how $\sigma_{\text{GELU}}(x)$ starts from zero for small values of $x$ since the CDF is almost equal to 0. However, around the value of $-2$, CDF starts increasing. Hence, we see $\sigma_{\text{GELU}}(x)$ deviating from zero. For the positive values, since CDF moves closer to a value of 1, $\sigma_{\text{GELU}}(x)$ starts approximating $\sigma_{\text{ReLU}}(x)$.

Since the cumulative distribution function of a Gaussian is often computed with the error function, we define the GELU as

$$\sigma_{\text{GELU}}(x) = x\,\Phi(x) = \frac{x}{2}\left(1+\text{erf}\left(\frac{x}{\sqrt{2}}\right)\right). \tag{8.54.2}$$

We can approximate the GELU with

$$\sigma_{\text{GELU}}(x) \approx 0.5\,x\left(1+\text{Tanh}\left[\sqrt{\frac{2}{\pi}}(x+0.044715x^3)\right]\right). \tag{8.55.1}$$

Alternatively, the function can be approximated using the Sigmoid function and a scaling parameter, as shown below:

$$\sigma_{\text{GELU}}(x) \approx\; x\sigma_{\text{Sigmoid}}(1.702x). \tag{8.55.2}$$

Both are sufficiently fast, easy-to-implement approximations.

The choice of the GELU AF is motivated by several desirable properties that make it suitable for NNs:

- GELU introduces a non-linearity in the network, allowing it to capture complex relationships in the data. The cubic term in the GELU function contributes to this non-linearity and the combination of different elements

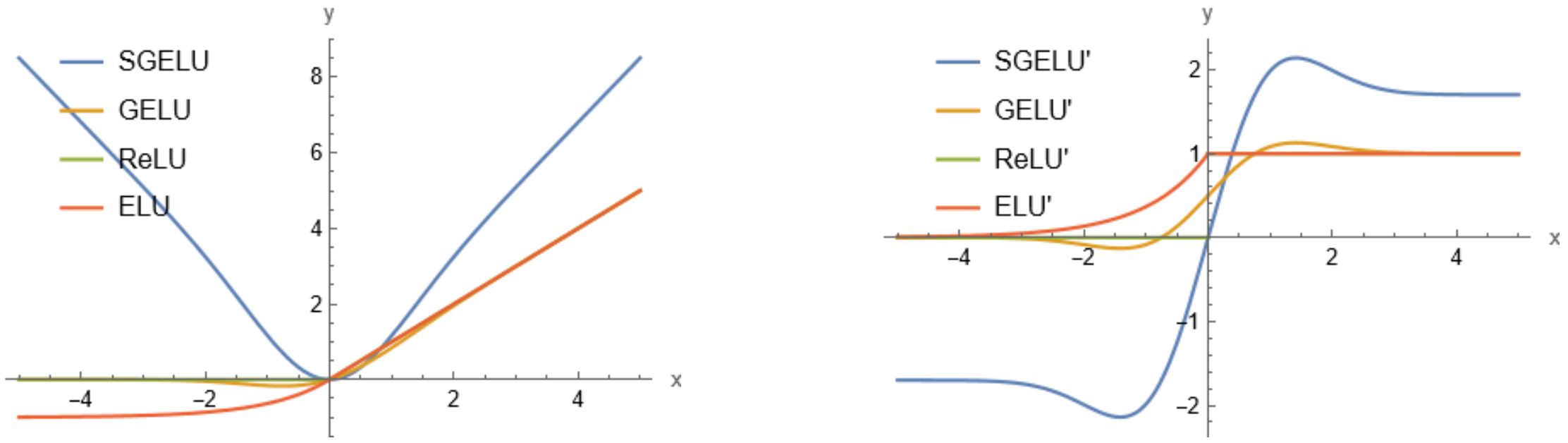

**Figure 8.38.** Left panel: The figure compares four different AFs: SGELU, GELU, ReLU, and ELU, plotted over a range of $x$ from $-5$ to 5. The SGELU function, scaled by a parameter $\alpha$ (default 1.702), shows a smooth curve enhanced by the error function, providing a flexible and smooth transition. The GELU function, with its probabilistic nature, also exhibits a smooth transition similar to SGELU but without the scaling factor. ReLU is characterized by its linear increase for positive values and zero output for negative values, indicating a sharp change at zero. The ELU function, with $\alpha = 1$, combines an exponential approach for negative inputs with a linear response for positive inputs, providing a more gradual transition compared to ReLU. Right panel: The figure presents the first derivatives of four AFs: SGELU, GELU, ReLU, and ELU, plotted over a range of $x$ from $-5$ to 5.

helps shape the AF's curve.

- GELU is a smooth function, meaning that it has continuous derivatives across its entire domain. This smoothness can be advantageous for optimization algorithms that rely on gradients, such as GD. GELU is differentiable everywhere, which is crucial for backpropagation during the training of NNs. The ability to compute gradients allows optimization algorithms to update the model parameters in the direction that minimizes the loss function.
- Unlike some other AFs (e.g., ReLU), GELU is zero-centered. This property can help the optimization process by reducing the risk of weights getting updated in a consistently positive or negative direction. The zero-centeredness contributes to more stable learning dynamics.
- The incorporation of the Gaussian cumulative distribution function $\Phi(x)$ in the GELU formulation aligns its behavior with certain aspects of the standard normal distribution. This can be beneficial for capturing statistical properties in the data and improving generalization.
- However, the GELU has several notable differences. This non-convex, non-monotonic function is not linear in the positive domain and exhibits curvature at all points. Meanwhile, ReLUs and ELUs, which are convex and monotonic activations, are linear in the positive domain and thereby can lack curvature. As such, increased curvature and non-monotonicity may allow GELUs to more easily approximate complicated functions than ReLUs or ELUs.
- Also, since $\sigma_{\text{ReLU}}(x) = x\mathbf{1}(x > 0)$ and $\sigma_{\text{GELU}}(x) = x\Phi(x)$ if $\mu = 0$, $\sigma = 1$, we can see that the ReLU gates the input depending upon its sign, while the GELU weights its input depending upon how much greater it is than other inputs.
- In addition and significantly, the GELU has a probabilistic interpretation given that it is the expectation of a stochastic regularizer.

### 8.5.8 Symmetrical Gaussian Error Linear Unit

Since the GELU function represents the nonlinearity using the stochastic regularizer on an input, which is the cumulative distribution function derived from the Gaussian error function, it has shown the advantage over other functions, e.g., ReLU, and ELU. However, most AFs do not fully exploit the negative value. Taking this into account, Symmetrical Gaussian Error Linear Unit (SGELU) [259] was proposed to combine the advantage of stochastic regularizer on the input and exploit the negative value, which can be represented by

$$\sigma_{\text{SGELU}}(x) = \alpha \cdot x \cdot \text{erf}\left(\frac{x}{\sqrt{2}}\right), \tag{8.56}$$

in which $\alpha$ represents the hyper-parameter that can be tuned in the computation to obtain the optimum solution. The properties of different AFs have been shown in Figure 8.38. In Figure 8.38, the SGELU AF is different from GELU, ReLU and ELU, which show the symmetrical characteristics together with Gaussian regularizer. For better illustration, the derivatives of ReLU, ELU, GELU, and SGELU are plotted as shown in Figure 8.38.

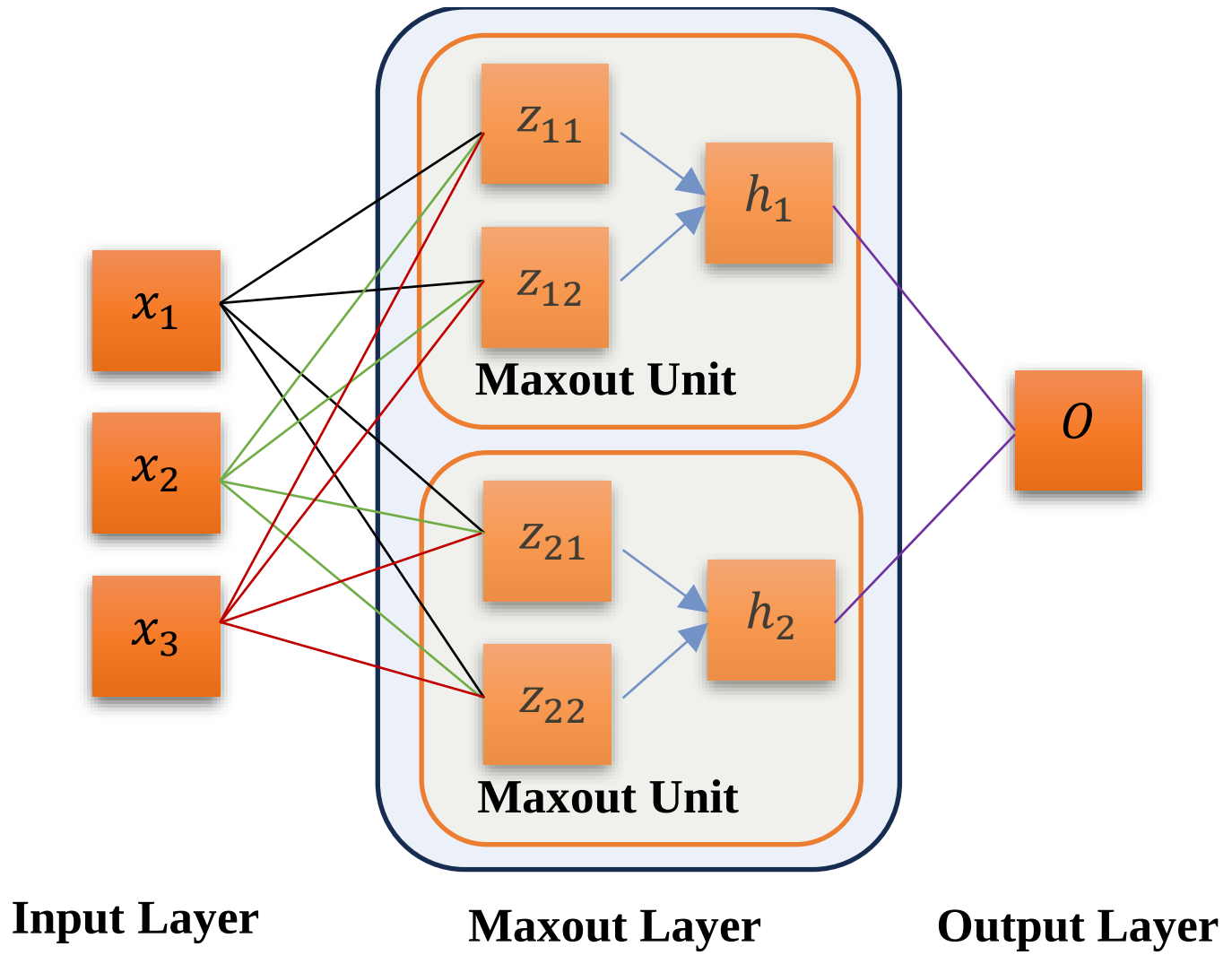

**Figure 8.39.** NN Architecture with Maxout Layer. The figure illustrates a NN architecture that includes a Maxout layer. The network consists of an input layer, a hidden Maxout layer, and an output layer. The input layer receives three inputs, $x_1$, $x_2$, and $x_3$. The Maxout layer contains two Maxout units. Each Maxout unit takes two inputs, $z_{11}$ and $z_{12}$ for the first unit, and $z_{21}$ and $z_{22}$ for the second unit, and computes the maximum value between them, denoted as $h_1$ and $h_2$ respectively. The outputs from the Maxout layer are then fed into the output layer to produce the final output $O$. This configuration enables the network to learn more complex functions by implementing a piecewise linear AF through the Maxout units.

The biggest difference of SGELU from other AFs is what happens in the negative half-axis. Instead of forcing the output to be zero like ReLU, deviating the output from a true value like GELU until it stops converging, or dragging the output from negative to positive like ELU, SGELU can update its weight symmetrically towards to two directions in both positive and negative half axis. In other words, the function of SGELU is a two-to-one mapping between the input and the output, while the others are a one-to-one mapping.

## 8.6 Non-Standard AFs

So far, we have taken care of AFs in the classic meaning given by literature, i.e., a function that builds the output of the neuron using as input the value returned by the internal transformation $\mathbf{w}^T\mathbf{x} + b$ made by the classic computational neuron model. In this section, we will review works which change the standard definition of neuron computation but are considered as NN models with trainable AFs in the literature. In other terms, these functions can be considered as a different type of computational neuron unit compared with the original computational neuron model.

### 8.6.1 Maxout

In NNs, the Maxout activation [260] takes the maximum value of the pre-activations. Figure 8.39 shows two pre-activations per Maxout unit, each of these pre-activations has a different set of weights from the inputs denoted as "**x**". Each hidden unit takes the maximum value over the $j$ units of a group:

$$h_i = \max_j z_{ij}, \tag{8.57}$$

where $z$ is the linear pre-activation value, $i$ is the number of maxout units, and $j$ the number of pre-activation values.

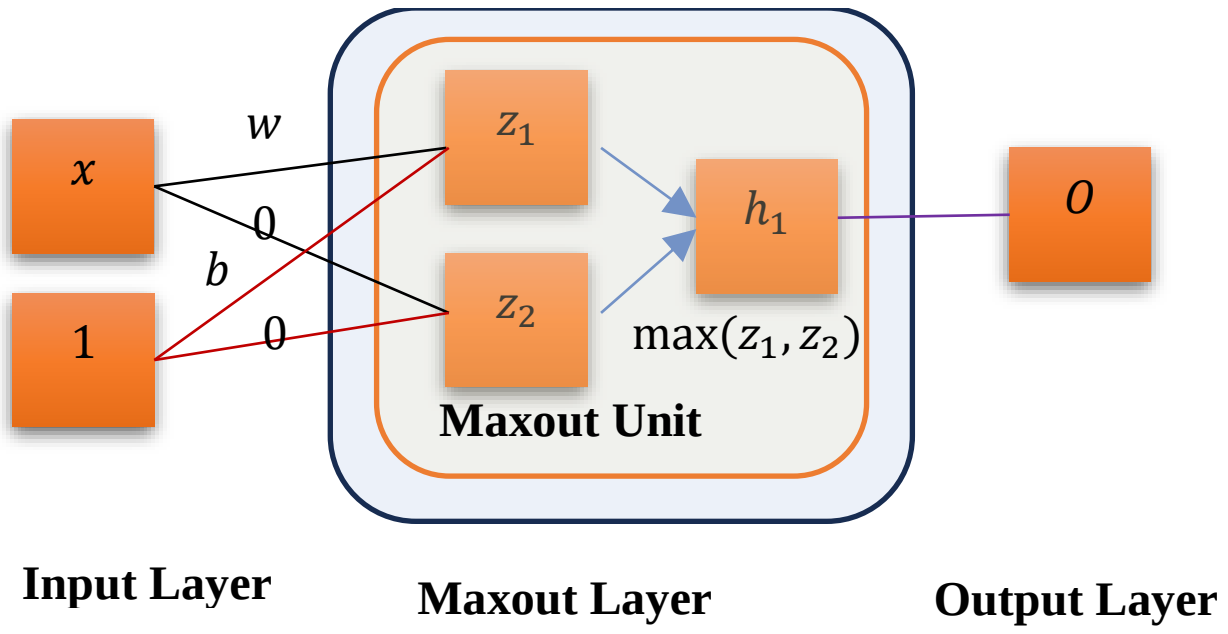

**Figure 8.40.** ReLU as a Special Case of Maxout in a NN. This figure demonstrates how the ReLU AF can be seen as a special case of the Maxout AF within a NN. The network consists of an input layer, a Maxout layer, and an output layer. The input layer receives a single input, $x$, which is processed into two linear combinations, $z_1$ and $z_2$. The Maxout layer includes a Maxout unit that computes the maximum value between $z_1$ and $z_2$, represented as $h_1 = \max(z_1, z_2)$. This value is then passed to the output layer to produce the final output $O$. By setting one of the linear combinations, $z_2$, to zero, and the bias term $b$ to zero, the Maxout unit effectively implements the ReLU function, $\max(0, z_1)$, showing that ReLU is a specific form of the more general Maxout function.

In other words, a layer of linear nodes is added to every hidden unit, each connected to all the input nodes, see Figure 8.39. These nodes use the AF $y(x) = x$ and send a signal forward to the hidden node, which uses the maxout unit. The hidden node simply chooses the largest of the values produced by the linear nodes and sends it forward to the output node. Hence, the Maxout AF works as follows:

Input:

- Maxout takes multiple inputs (usually two or more) instead of a single scalar value like other AFs.
- Each input represents the weighted sum of the inputs to a neuron before the AF is applied.

Computation:

- For each neuron, Maxout computes the maximum value of its input values.
- In other words, it takes the maximum value among the inputs and returns that as the output.
- Mathematically, for a Maxout unit with '$k$' inputs, the output is the maximum of these inputs: $\max(z_1, z_2, \ldots, z_k)$.

Maxout chooses the maximum of $n$ input features to produce each output feature in a network, the simplest case of maxout is the Max-Feature-Map (MFM), where $n = 2$.

The MFM maxout computes the function

$$\sigma_{\text{Maxout}}(\mathbf{x}) = \max(\mathbf{w}_1^T\mathbf{x} + b_1, \mathbf{w}_2^T\mathbf{x} + b_2), \tag{8.58}$$

and both the ReLU and LReLU are a special case of this form. When specific weight values $\mathbf{w}_1$, $b_1$, $\mathbf{w}_2$ and $b_2$ of the MFM inputs are learned, MFM can emulate ReLU and other rectified linear variants, Figure 8.40. The maxout unit is helpful for tackling the problem of vanishing gradients because the gradient can flow through every maxout unit. In the ReLU, LReLU, and PReLU, there must be two sets of points: one lying in the negative side and another on the positive side of the AF domain (see region 1 covering the negative side and region 2 on the positive side in PReLU/LReLU cases of Figure 8.41). Moreover, in the Maxout activation units, there must be $k$ sets of points - each set lying in one of the $k$ regions of the AF domain (see Maxout case ($k = 4$) in Figure 8.41).

Maxout units take the maximum value over a subspace of $k$ trainable linear functions of the same input $\mathbf{x}$, obtaining a piece-wise linear approximator capable of approximating any convex function. The AF created by the maxout nodes will always be a convex function, however, with enough maxout nodes it can approximate any convex function arbitrary well. In theory, maxout can approximate any convex function, but a large number of extra parameters introduced by the $k$ linear functions of each hidden maxout unit result in large RAM storage memory cost and a considerable increase in training time, which affect the training efficiency of very DNNs.

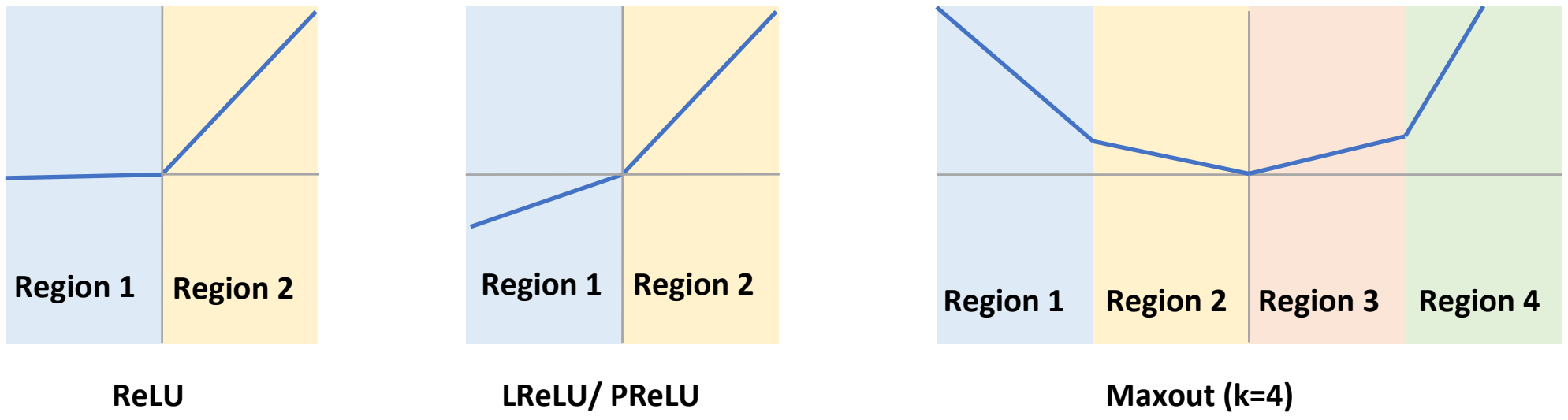

**Figure 8.41.** Piecewise Linear AFs: ReLU, LReLU/PReLU, and Maxout. The figure demonstrates how the Maxout AF can generalize ReLU and LReLU/PReLU. In the ReLU case, the Maxout function reduces to $\max(z_1, 0)$ by setting $z_2$ to zero and bias $b$ to zero, creating two regions (Region 1 and Region 2). For LReLU/PReLU, different slopes for the negative part of the activation are introduced, still dividing the function into two regions. The Maxout function with $k = 4$ shows the division into multiple regions (Regions 1-4), indicating its flexibility in approximating various piecewise linear functions by taking the maximum of multiple linear combinations. This versatility makes Maxout a powerful tool for learning complex activation patterns in NNs.

### 8.6.2 Softmax

The Softmax AF [32] is a widely used AF in NNs, particularly in multi-class classification problems. It takes as input a vector of real numbers and transforms them into a probability distribution. Given an input vector $\mathbf{z} = (z_1, z_2, \ldots, z_n)$, the softmax function calculates the output vector $\mathbf{a} = (a_1, a_2, \ldots, a_n)$ as follows:

$$\sigma_{\text{Softmax}}(z_i) = a_i = \frac{e^{z_i}}{\sum_j e^{z_j}} \quad \text{for } i = 1,2,\ldots,n. \tag{8.59}$$

The softmax AF works by taking a vector of real numbers as input and transforming these numbers into a probability distribution. It assigns probabilities to each element in the input vector, such that the values are in the range [0,1], and they sum up to 1. This makes it suitable for multi-class classification problems where you want to determine the likelihood of an input belonging to each class.

The following is a step-by-step explanation of how the softmax AF works, see Figures. 8.42 and 8.43:

- The final hidden layer preceding the softmax layer might use linear (identity) activations. The identity activation means that the output of the layer is simply a linear combination of its inputs, without applying any non-linear AF. “Logits” is defined as the numerical output of the final linear layer of a multi-class NN.
- You start with an input vector $\mathbf{z} = (z_1, z_2, \ldots, z_n)$, where each $z_i$ is a real number. These values are typically the raw scores or logits obtained from the previous layer of a NN. The goal is to convert these scores into a probability distribution.
- For each element $z_i$ in the input vector, you compute the exponential ($e^{z_i}$) of that element. This step amplifies the differences between the values in the input vector, emphasizing the larger values and diminishing the smaller ones.
- Next, you sum up all the exponentiated values to get the denominator of the softmax formula. This step involves calculating the sum of all $e^{z_j}$ for $j$ in the range from 1 to $n$, $\sum_j e^{z_j}$.
- For each element in the input vector, you divide the exponential of that element by the denominator calculated in the above step. This calculates the probability of each element being the most likely class. The result is an output vector $\mathbf{a}$, where each $a_i$ is a probability.

$$a_i = \frac{e^{z_i}}{\sum_{j=1}^{n} e^{z_j}} \quad \text{for } i = 1,2,\ldots,n. \tag{8.60}$$

- The output vector $\mathbf{a}$ now contains the probabilities for each class. Each element, $a_i$, represents the probability that the input belongs to class $i$. The class with the highest probability is considered the predicted class, and the softmax function ensures that all probabilities sum up to 1.

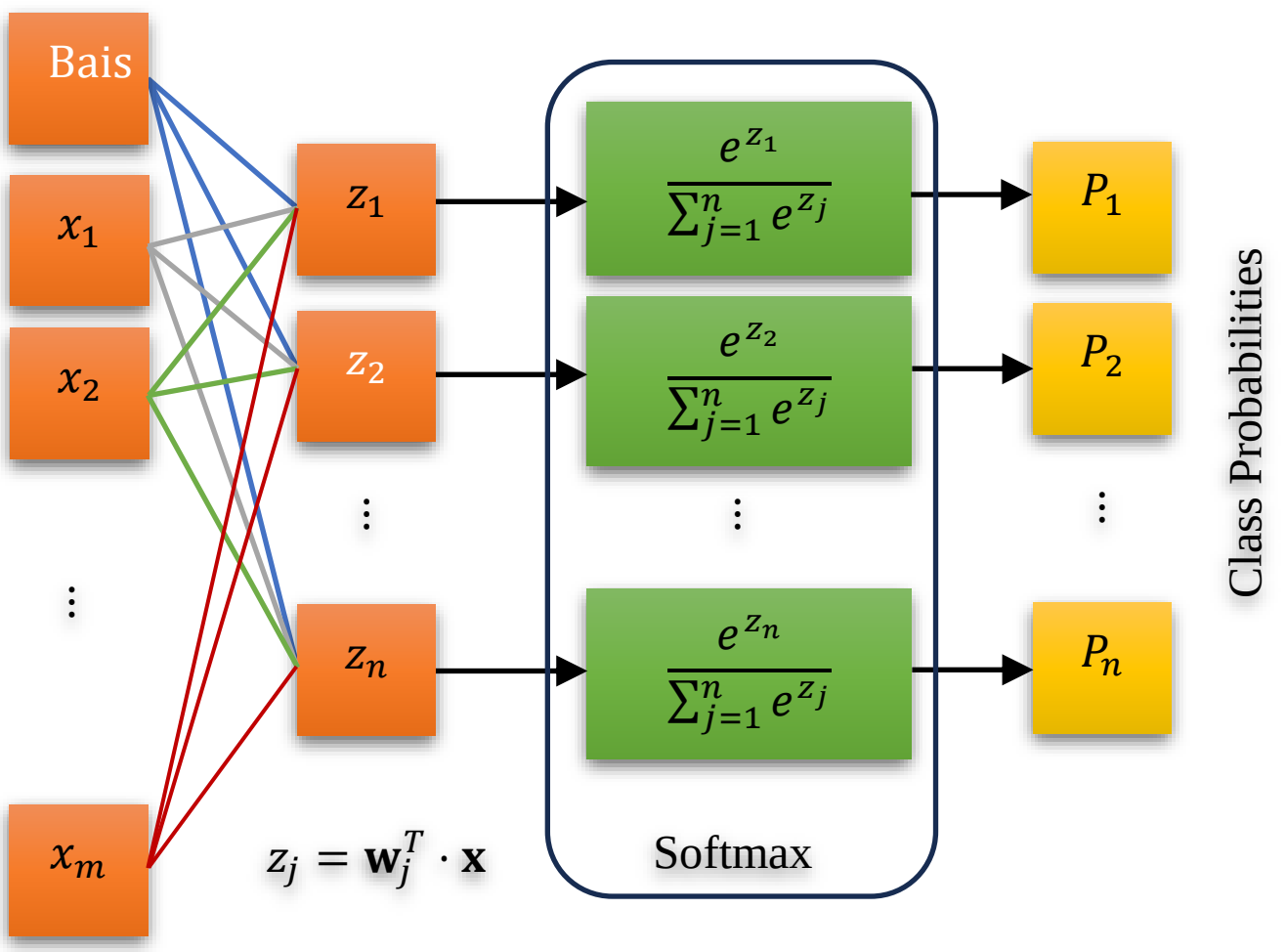

**Figure 8.42.** Softmax AF in a NN. The figure illustrates the softmax AF used in the output layer of a NN for multi-class classification. The network consists of an input layer, a fully connected layer, and a softmax output layer. The input layer receives $m$ input features $x_1, x_2, \ldots, x_m$, which are linearly combined with weights and biases to compute the logits $z_1, z_2, \ldots, z_n$ for each class $j$. The logits are calculated as $z_j = \mathbf{w}_j^T \cdot \mathbf{x} + b_j$, where $\mathbf{w}_j$ is the weight vector for class $j$, and $\mathbf{x}$ is the input feature vector. The softmax function then transforms these logits into class probabilities. For each class $i$, the probability $P_i$ is computed using the softmax formula: $P_i = e^{z_i} / \sum_j e^{z_j}$. This formula ensures that the probabilities for all classes sum to 1, allowing the network to assign a probability distribution over the possible classes. In the figure, the computation for each class probability is depicted, showing the exponential of each logit $e^{z_1}, e^{z_2}$, … , $e^{z_n}$ and the normalization by the sum of all exponentials. The resulting class probabilities $P_1, P_2, \ldots, P_n$ indicate the likelihood of the input belonging to each respective class. This softmax layer is commonly used in the final layer of NNs for classification tasks to provide a probabilistic interpretation of the model's predictions.

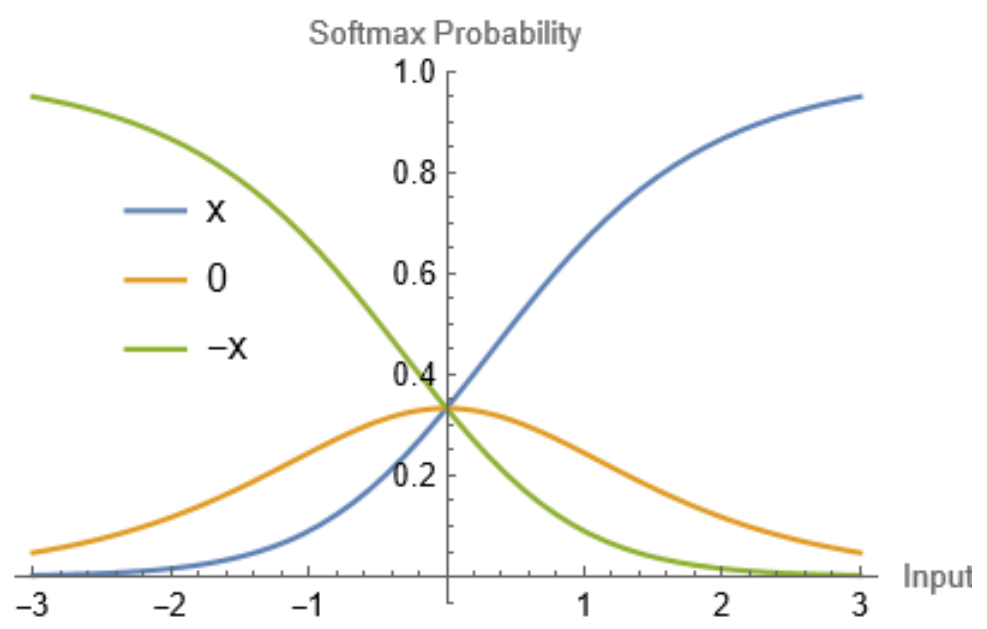

**Figure 8.43.** The figure visualizes the Softmax function applied to a range of inputs, showcasing how it transforms raw input values into probabilities that sum to 1. The plot demonstrates the Softmax probabilities for three inputs $(x, 0, -x)$ as $x$ varies from $-3$ to 3. The Softmax function, which is often used in classification tasks in NNs, converts these inputs into a set of probabilities. As $x$ increases, the probability associated with the input $x$ rises, while the probabilities for the inputs 0 and $-x$ adjust accordingly.

- For example, if you have an input vector $\mathbf{z} = (2.0, 1.0, 0.1)$, applying the softmax function would yield an output vector $\mathbf{a} = (0.659001, 0.242433, 0.0985659)$. This means that the first class is the most likely class for the given input, with a probability of approximately $0.659$.
- Before applying the softmax function, class labels are typically represented using one-hot encoding, where each class is represented as a binary vector with a 1 at the index corresponding to the class and 0s everywhere else. Softmax then provides probabilities for each class.

**Lemma 8.1:** The derivative of the softmax function with respect to the logit ($z_i = \mathbf{w}_i^T \cdot \mathbf{x}$) is

$$\frac{\partial}{\partial z_i}\sigma_{\text{Softmax}}(z_j) = \sigma_{\text{Softmax}}(z_j)\left(\delta_{ij} - \sigma_{\text{Softmax}}(z_i)\right). \tag{8.61}$$

**Proof:**

Let

$$\begin{aligned}\sigma_{\text{Softmax}}(z_j) &= \frac{e^{\mathbf{w}_j^T\cdot\mathbf{x}}}{\sum_{k=1}^{n} e^{\mathbf{w}_k^T\cdot\mathbf{x}}}\\ &= \frac{e^{z_j}}{\sum_{k=1}^{n} e^{z_k}}.\end{aligned}$$

Computing the

$$\frac{\partial}{\partial z_i}\sigma_{\text{Softmax}}(z_j) = \frac{\partial}{\partial z_i}\frac{e^{z_j}}{\sum_{k=1}^{n} e^{z_k}}.$$

The derivative of $\sum_{k=1}^{n} e^{z_k} = e^{z_1} + \cdots + e^{z_i} + \cdots + e^{z_n}$ with respect to any $z_i$ will be $e^{z_i}$.

If $i = j$, and using the quotient rule,

$$\begin{aligned}\frac{\partial}{\partial z_i}\frac{e^{z_j}}{\sum_{k=1}^{n} e^{z_k}} &= \frac{e^{z_j}\sum_{k=1}^{n} e^{z_k} - e^{z_i}e^{z_j}}{[\sum_{k=1}^{n} e^{z_k}]^2}\\ &= \frac{e^{z_j}(\sum_{k=1}^{n} e^{z_k} - e^{z_i})}{\sum_{k=1}^{n} e^{z_k}\sum_{k=1}^{n} e^{z_k}}\\ &= \frac{e^{z_j}}{\sum_{k=1}^{n} e^{z_k}}\frac{\sum_{k=1}^{n} e^{z_k} - e^{z_i}}{\sum_{k=1}^{n} e^{z_k}}\\ &= \frac{e^{z_j}}{\sum_{k=1}^{n} e^{z_k}}\left(1 - \frac{e^{z_i}}{\sum_{k=1}^{n} e^{z_k}}\right)\\ &= \sigma_{\text{Softmax}}(z_j)\left(1 - \sigma_{\text{Softmax}}(z_i)\right).\end{aligned}$$

If on the other hand, $i \neq j$:

$$\begin{aligned}\frac{\partial}{\partial z_i}\frac{e^{z_j}}{\sum_{k=1}^{n} e^{z_k}} &= \frac{0 - e^{z_i}e^{z_j}}{[\sum_{k=1}^{n} e^{z_k}]^2}\\ &= \frac{-e^{z_i}e^{z_j}}{\sum_{k=1}^{n} e^{z_k}\sum_{k=1}^{n} e^{z_k}}\\ &= -\frac{e^{z_j}}{\sum_{k=1}^{n} e^{z_k}}\frac{e^{z_i}}{\sum_{k=1}^{n} e^{z_k}}\\ &= -\sigma_{\text{Softmax}}(z_j)\sigma_{\text{Softmax}}(z_i)\\ &= \sigma_{\text{Softmax}}(z_j)\left(0 - \sigma_{\text{Softmax}}(z_i)\right).\end{aligned}$$

These two scenarios can be brought together as

$$\frac{\partial}{\partial z_i}\sigma_{\text{Softmax}}(z_j) = \sigma_{\text{Softmax}}(z_j)\left(\delta_{ij} - \sigma_{\text{Softmax}}(z_i)\right).$$

where

$$\delta_{ij} = \begin{cases}1, & i = j,\\ 0, & i \neq j.\end{cases}$$

■

## 8.7 Combining AFs

### 8.7.1 Mixed, Gated, and Hierarchical AFs

First, let us consider the mixed activation and gated activation strategies [261].

1. The mixed activation strategy involves combining basic AFs linearly. This means taking a weighted sum of different AFs. The combination coefficients are learned from the data, meaning the NN adapts and adjusts these weights during training.

2. In a gated activation strategy, basic AFs are combined nonlinearly. This often involves the use of gating mechanisms to control the flow of information. Similar to the mixed strategy, the coefficients for combining AFs are learned from the data.

3. Let us explain these two strategies from the perspective of information change:

- Linear-type (LReLU) AFs $\sigma_{\text{LReLU}}(x) = (x \text{ if } x > 0, \alpha x \text{ if } x \leq 0)$ (considering the negative part) do not become saturated (i.e., they do not reach extreme values) regardless of how small the input is.
- Exponential-type (ELU) AFs $\sigma_{\text{ELU}}(x) = (x \text{ if } x > 0, \beta(e^x - 1) \text{ if } x \leq 0)$ saturate to a negative value when the input is small.
- Both types of AFs can change forward and backward propagated information, but they do so in different ways. The variables $\alpha$ and $\beta$ are used to represent the degree of information change, and they provide a way to quantify how much the information is altered by the AFs.

The mixed activation is:

$$\sigma_{\text{Mix}}(x) = \rho\sigma_{\text{LReLU}}(x) + (1-\rho)\sigma_{\text{ELU}}(x), \tag{8.62}$$

where $\rho \in [0,1]$ is a combination coefficient specifying the specific combination of LReLU and ELU. The specific combination coefficient $\rho$ is learned from the data. However, once each combination coefficient is learned in a mixed activation strategy, then the mixed activation will be kept constant. In other words, it is not adapted to the specific inputs after learning because it keeps fixed no matter what characteristics of the inputs appear.

Instead of directly learning a fixed combination coefficient, a gating mask is learned during training. The combination coefficient is not fixed; it depends on both the learned gating mask and the input values. The product of the gating mask and inputs is passed through a Sigmoid function, providing a dynamic and adaptive way to generate the combination coefficient between different AFs. The AF for the gated strategy can be formulated as follows. Let, $x$ be the input, $\omega$ be the gating mask, and $\sigma_{\text{Sigmoid}}(.)$ be the Sigmoid function. The combination coefficient $\tau$ is defined using the Sigmoid function applied to the product of $\omega$ and $x$:

$$\tau = \sigma_{\text{Sigmoid}}(\omega x) = \frac{1}{1+\exp(-\omega x)}. \tag{8.63.1}$$

Now, the AF for the gated strategy can be expressed as a weighted combination of different AFs (LReLU and ELU, in this case) using the learned combination coefficient $\tau$:

$$\begin{aligned}\sigma_{\text{Gate}}(x) &= \tau\, \sigma_{\text{LReLU}}(x) + (1-\tau)\sigma_{\text{ELU}}(x)\\ &= \sigma_{\text{Sigmoid}}(\omega x)\cdot \sigma_{\text{LReLU}}(x) + \left(1-\sigma_{\text{Sigmoid}}(\omega x)\right)\sigma_{\text{ELU}}(x).\end{aligned} \tag{8.63.2}$$

This formulation allows the AF to dynamically adjust its behavior based on both the input $x$ and the learned gating mask $\omega$. The Sigmoid function ensures that the combination coefficient $\tau$ is within the range of [0,1], providing a weighted combination of the two AFs.

These strategies make the designed AFs more flexible, enhancing the NN's ability to learn non-linear transformations. The designed AFs still retain the characteristics of information change, and they model the degree of information change in both qualitative and quantitative terms.

The mixed strategy and the gated strategy adopt linear and nonlinear methods, respectively. By comparing the above two strategies, the main difference is that the mixed strategy is not adaptive to the specific inputs, but the gated strategy is adaptive in adjusting the mixture of LReLU and ELU for the specific inputs. From the perspective of information change as described above, the degree of information change can be further quantitatively adjusted by $\rho$ or $\tau$. Specifically, the gated strategy learns $\omega$ which indicates the degree of information change for the whole dataset. After learning $\omega$, $\tau$ determines the specific proportion of information change for the specific $x$. In addition, when $x$ varies from zero to a smaller value in $\tau = \sigma_{\text{Sigmoid}}(\omega x)$, the degree of information change will tend to be determined by one of the AF types instead of both.

The two strategies detailed above focus on the ways of combining basic AFs with predefined parameters. In order to further improve the ability to learn non-linear transformation and has the adaptability to the inputs, the basic AFs being combined are organized in a more complex hierarchical structure. The hierarchical structure involves three levels, where low-level nodes are associated with learnable AFs, middle-level nodes combine pairs of low-level nodes, and high-level nodes integrate outputs from all middle-level nodes using the winner-take-all principle.

In other words, the hierarchical structure introduces a more complex organization of basic AFs with learnable parameters, aiming to enhance the network's capacity for learning non-linear transformations and improving adaptability to diverse inputs. The winner-take-all integration principle at the high level suggests a mechanism for selecting the most relevant information from the lower levels.

To be more specific in the hierarchical structure, for a low-level node with index $n$, the basic AF is denoted by $\sigma_{\text{low}}^{n}(x)$. For middle-level nodes, we proceed in the same way to the gated strategy. At each middle-level node with index $m$, a pair of child low-level nodes are combined into a single parent value $\sigma_{\text{mid}}(x)$ with a learned gating mask denoted by $\omega_m$. From the perspective of the middle-level node, the pair of low-level nodes can also be denoted by $\sigma_{\text{mid}}^{m\,\text{left}}(x)$ and $\sigma_{\text{mid}}^{m\,\text{right}}(x)$. The combination of child nodes into a parent value is given by:

$$\sigma_{\text{mid}}^{m}(x) = \tau\sigma_{\text{mid}}^{m\,\text{left}}(x) + (1-\tau)\sigma_{\text{mid}}^{m\,\text{right}}(x). \tag{8.64.1}$$

The overall activation operation $\sigma_{\text{high}}(x) = \sigma_{\text{Hierarchical}}(x)$ involves selecting the maximum value across $k$ middle-level nodes based on the winner-take-all principle. Neurons in the high-level node compete with each other, with only the neuron having the highest activation being activated while others are inhibited. Overall, the activation result at each level node is

$$\sigma_{\text{Hierarchical}}(x) = \begin{cases} \sigma_{\text{low}}^{n}(x), & \text{low}-\text{level nodes}, \\ \sigma_{\text{mid}}^{m}(x), & \text{middle}-\text{level nodes}, \\ \sigma_{\text{high}}(x), & \text{high}-\text{level node}, \end{cases} \tag{8.64.2}$$

where

$$\sigma_{\text{high}}(x) = \max_{m\in[1,k]} \sigma_{\text{mid}}^{m}(x). \tag{8.64.3}$$

Inspired by Maxout, the winner-take-all principle, $\max_{m\in[1,k]} \sigma_{\text{mid}}^{m}(x)$, introduces competition among activation neurons in a layer. Only the neuron with the highest activation is allowed to be activated, enhancing nonlinearity and adaptability to specific inputs.

### 8.7.2 Adaptive Piecewise Linear Units

While the type of AF can have a significant impact on learning, the space of possible functions has hardly been explored. One way to explore this space is to learn the AF during training. The Adaptive Piecewise Linear (APL) [262] activation unit uses summation of ReLU-like units to increase the capacity of the AF. The method formulates the AF $\sigma_{\text{APL}}(x)$ of an APL unit $i$ as a sum of hinge-shaped functions,

$$\sigma_{\text{APL}}(x) = \max(0, x) + \sum_{s=1}^{S} (a_i^s \max(0, -x + b_i^s)). \tag{8.65}$$

Figure 8.44: Sample AFs obtained from changing the parameters, $S = 1$ for all plots.

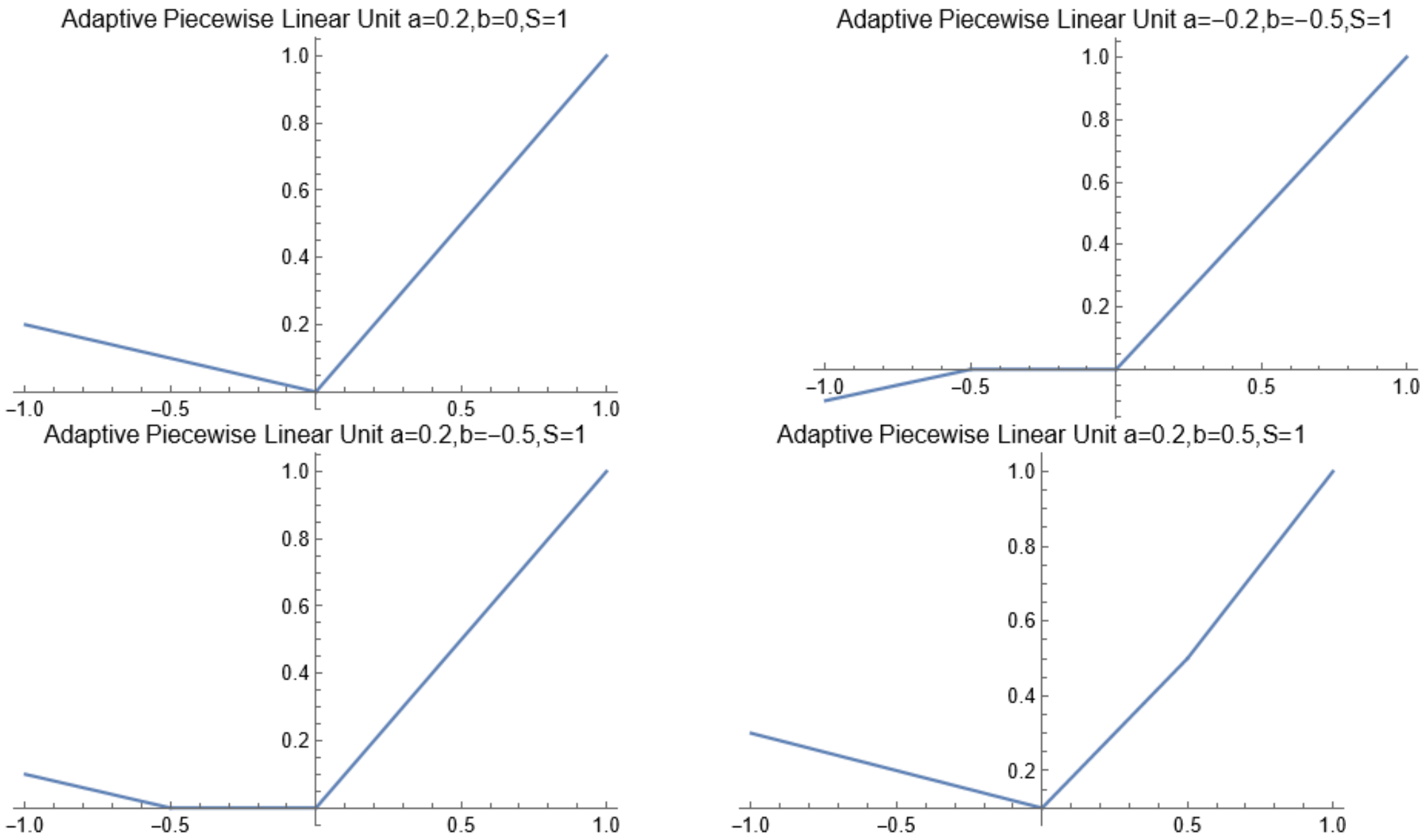

**Figure 8.44.** The figure contains four plots illustrating the behavior of the APL AF over a range of $x$ from $-1$ to $1$, with different parameter settings for $a$ and $b$ with $S = 1$. APL ($a = 0.2, b = 0, S = 1$): The function combines a ReLU component $\max(0, x)$ with a linear adjustment term, resulting in a slight modification for negative inputs. APL ($a = -0.2, b = -0.5, S = 1$): The negative $a$ value introduces a downward adjustment for inputs less than $b$, significantly modifying the activation for negative inputs. APL ($a = 0.2, b = -0.5, S = 1$): This configuration introduces a positive adjustment for inputs less than $b$, altering the AF's slope in the negative region. APL ($a = 0.2, b = 0.5, S = 1$): The positive $b$ value shifts the adjustment term to a different input range, affecting the slope for inputs around $b$.

**Remarks:**

- The result is a piecewise linear AF.
- $\max(0, x)$: This is the ReLU AF, which returns the maximum of zero and $x$. It introduces non-linearity by zeroing out negative values.
- $\max(0, -x + b_i^s)$: This is another ReLU activation applied to the quantity $-x + b_i^s$. It also introduces non-linearity by zeroing out negative values.
- The number of hinges, $S$, is a hyperparameter set in advance, while the variables $a_i^s$, $b_i^s$ for $i \in 1, \dots, S$ are learned using standard GD during training.
- The $a_i^s$ variables control the slopes of the linear segments, while the $b_i^s$ variables determine the locations of the hinges.
- The number of additional parameters that must be learned when using these APL units is $2SM$, where $M$ is the total number of hidden units in the network. This number is small compared to the total number of weights in typical networks.
- This parametrized, piecewise linear AF is learned independently for each neuron and can represent both convex and non-convex functions of the input.

### 8.7.3 Mexican ReLU

The "Mexican hat type" function is defined as follows:

$$\phi_{a,\lambda}(x) = \max(\lambda - |x - a|, 0). \tag{8.66}$$

Here, $a$ and $\lambda$ are real numbers, and $x$ is the variable. The parameter $a$ determines the center of the function along the $x$-axis. It represents the location where the function has its peak or central point resembling the top of a Mexican hat. Shifting the value of $a$ will move the entire function horizontally along the $x$-axis. A larger $a$ shifts the function to the right, and a smaller $a$ shifts it to the left, see Figure 8.45. The parameter $\lambda$ controls the width of the function. It is associated with the spread or width of the "Mexican hat" shape. A larger $\lambda$ results in a broader "Mexican hat," meaning that the function will have a wider range of values for which it is non-zero. Conversely, a smaller $\lambda$ leads to a narrower hat with a more localized impact. The width of the function is related to the scale of features that the function can detect or represent. Smaller values of $\lambda$ are associated with higher-frequency components, while larger values correspond to lower-frequency components.

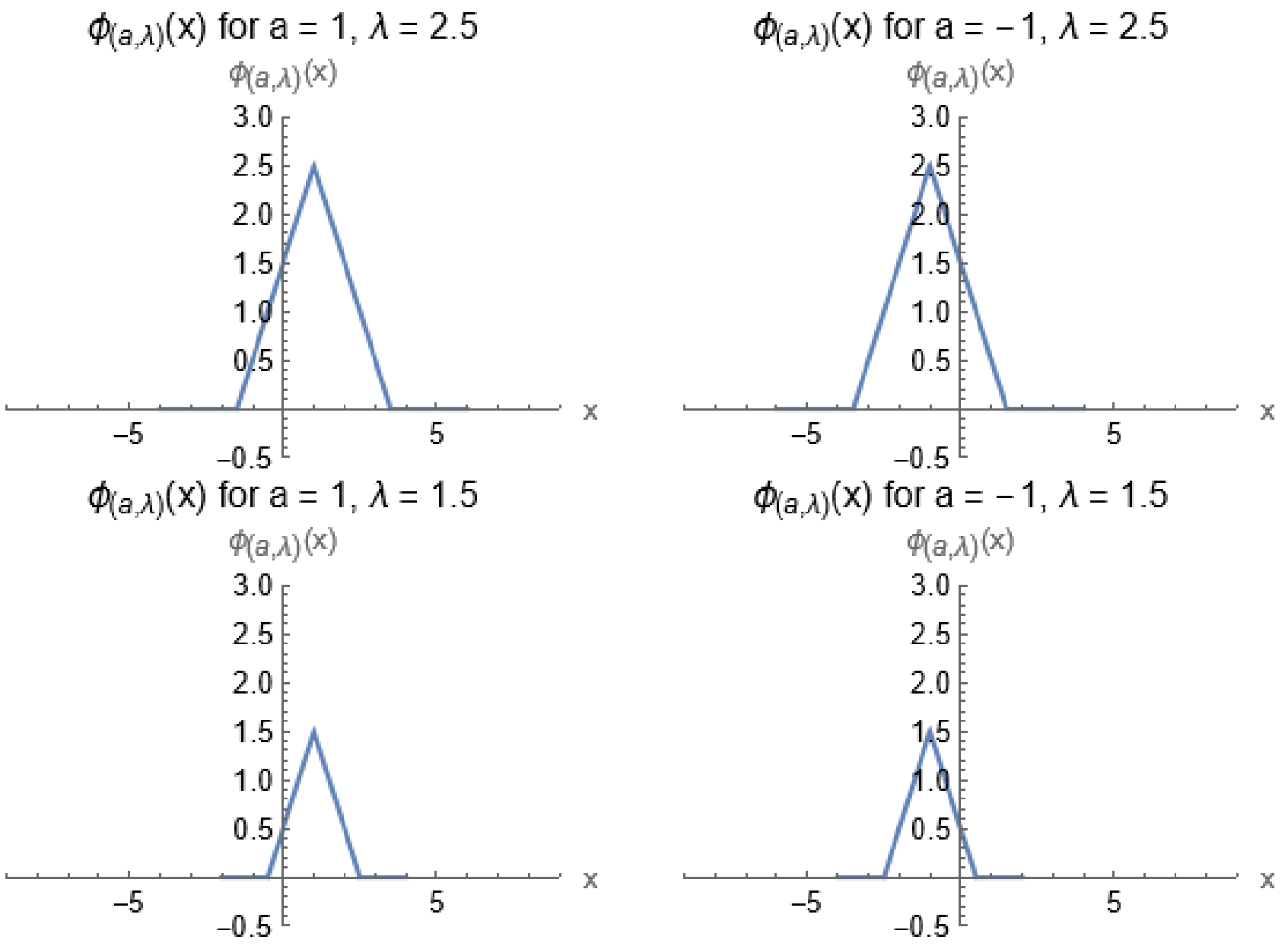

**Figure 8.45.** The figure includes four plots illustrating the Mexican hat-type function $\phi_{a,\lambda}(x) = \max(\lambda - |x - a|,0)$ for various parameter settings. Each plot demonstrates how the function changes based on different values of $a$ and $\lambda$.

The function has a shape resembling a Mexican hat, which is a term often used in mathematics to describe functions with a characteristic peaked shape. The function is null (equal to 0) when the absolute difference between $x$ and $a$ is greater than $\lambda$, i.e., $|x - a| > \lambda$. It increases with a derivative of 1 in the interval $(a - \lambda, a)$. This means that in this interval, the function rises steadily, and its rate of increase is constant. It decreases with a derivative of $-1$ in the interval $(a, a + \lambda)$. In this interval, the function decreases steadily, and its rate of decrease is constant. These functions are the building blocks of Mexican ReLU (MeLU) AF [263]. MeLU is defined as

$$\sigma_{\mathrm{MeLU}}(x) = \sigma_{\mathrm{PReLU}}(x) + \sum_{j=1}^{k-1} c_j\, \phi_{a_j,\lambda_j}(x), \tag{8.67}$$

for each channel of the hidden layer. $c_j$ is a learnable parameter associated with the $j$-th term in the sum. The parameters $a_j$ and $\lambda_j$ are chosen recursively. This means that the values of $a_j$ and $\lambda_j$ for each term in the sum depends on previous terms, likely through a recursive relationship. The total number of learnable parameters in every channel is given by $k$. This includes the $k - 1$ coefficients $c_j$ associated with the "Mexican hat type" functions, and one learnable parameter in the PReLU AF.

MeLU is a flexible AF that combines the piecewise linearity of PReLU with the localized and potentially non-linear effects of "Mexican hat type" functions. This combination can enhance the model's ability to capture and represent both linear and non-linear features in the data. The model will learn the values of the parameters during the training process to optimize the network for a specific task.

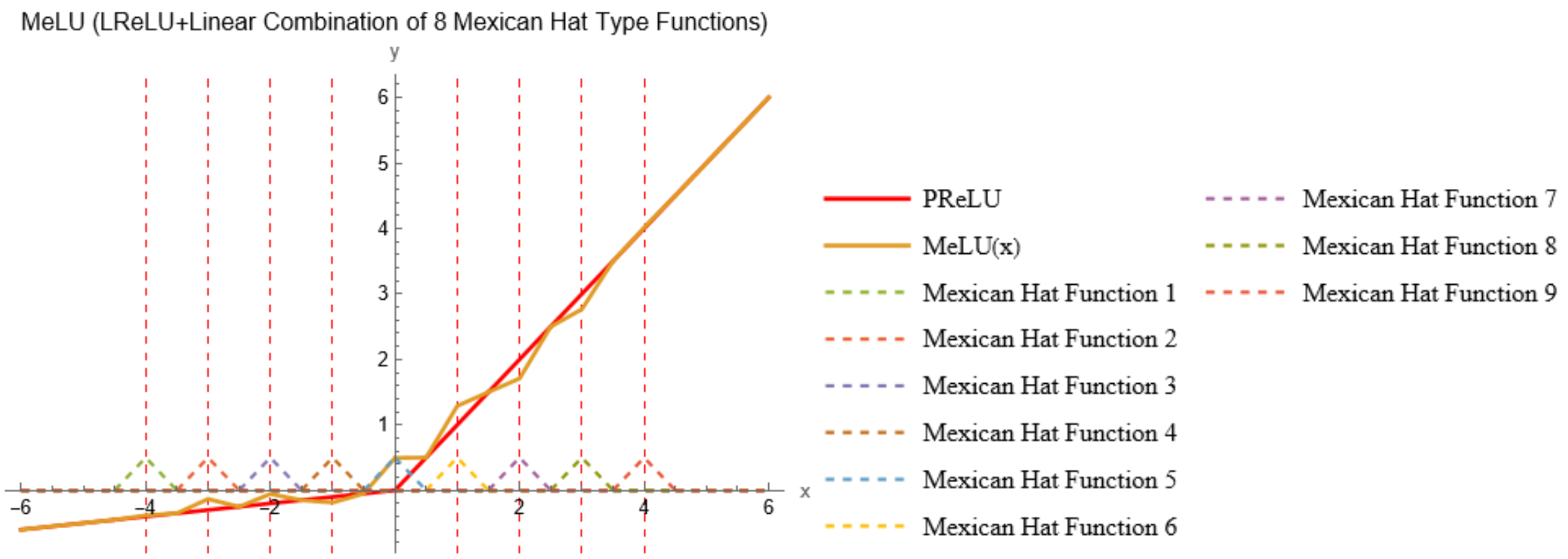

**Figure 8.46.** The figure illustrates the MeLU AF, which is a combination of a PReLU and a series of nine Mexican Hat type functions, plotted over a range of $x$ from $-6$ to 6. The PReLU function is shown as a red, thick line, representing the baseline AF. The MeLU function, plotted as a thick solid line, incorporates the PReLU function and adds a linear combination of the Mexican Hat type functions, each centered at different points ($x = -4$ to 4) with widths of 0.5 and randomly generated coefficients. The individual Mexican Hat-type functions are displayed as dashed lines, showing their contributions to the overall MeLU function.

**Remarks:**

- The Mexican hat functions are continuous and piecewise differentiable. MeLU inherits these properties because it is defined as a combination of PReLU and the sum of "Mexican hat type" functions.
- If all the coefficients $c_i$ in the MeLU AF are initialized to zero, MeLU reduces to the PReLU AF. This is an interesting observation, as it implies that MeLU can smoothly transition from PReLU behavior by adjusting the values of the $c_i$ coefficients during training.
- The same holds for networks trained with LReLU or PReLU.
- It can exhibit different AF behaviors based on initialization and adapt its characteristics during training. Since it has more parameters, it has a higher representation power, but it might overfit easily, see Figure 8.46.
- The derivatives of "Mexican hat type" functions form a Hilbert basis on a compact set with the $L^2$ norm. This is a notable mathematical property indicating that the derivatives of these functions can be used to approximate every function in $L^2$ space on a compact set, as $k$ goes to infinity.
- The structure of a hidden layer in a NN is defined as $\sigma(\mathbf{W}^T.\mathbf{x} + b)$, where $\mathbf{x}$ is the input of the hidden layer, $\mathbf{W}$ is the weight matrix, $\mathbf{b}$ is the bias and $\sigma$ is the activation. Through joint optimization of weights ($\mathbf{W}$), biases ($\mathbf{b}$), and activation parameters (parameters of the MeLU AF), (i.e., simultaneous optimization of all these parameters during the training process), NNs with MeLU activation can approximate any continuous function on a compact set.
- APL and MeLU look very similar. Both APL and MeLU have the ability to approximate the same set of functions: piecewise linear functions that behave like the identity for sufficiently large $x$. While APL and MeLU can both approximate the same set of functions, they do so in distinct ways. APL achieves its flexibility through learnable points of non-differentiability. The positions of these points are adjusted during training. MeLU, on the other hand, achieves its expressive power through the joint optimization of the weights matrix and biases. MeLU is noted to use only half of the parameters of APL while having the same representation power. This suggests that MeLU may offer a more parameter-efficient way to achieve similar function approximation capabilities.

### 8.7.4 Look-up Table Unit AF

AFs introduce non-linearity to the NN, allowing it to model complex relationships. Instead of using pre-defined AFs (e.g., Sigmoid or hyperbolic tangent), the Look-up Table Unit AF (LuTU) [264] suggests learning the shape of the AF itself. This implies that the NN will determine the optimal AF during the learning process, offering a high degree of flexibility.

This approach aims to strike a balance between allowing the model to learn the optimal AF shape and ensuring some level of control or constraint over the range of possible shapes. This may lead to more adaptive and data-driven AFs tailored to the specific characteristics of the problem at hand. Moreover, this approach aligns with the broader philosophy of leveraging the power of deep learning to automatically discover complex patterns and representations in the input data.

- The LuTU method employs a look-up table style structure to store a set of anchor points. This structure is likely a data structure, possibly an array or a table, where each entry corresponds to an anchor point for the AF.
- The AF is interpolated from the anchor points using either simple linear interpolation or cosine smoothing. Interpolation is a method of estimating values between two known values. Linear interpolation is a straightforward method, while cosine smoothing suggests a more sophisticated interpolation technique that likely aims to produce a smoother transition between anchor points.
- When the distance between adjacent anchor points is small enough, the proposed function can approximate any univariate function. This implies that the look-up table, along with the chosen interpolation method, is capable of capturing a wide range of AF shapes.
- Visualization of the learned functions demonstrates high diversity, indicating that the LuTU method is capable of capturing a variety of AF shapes. This diversity is considered different from previous AFs, suggesting that the learned functions may have unique characteristics.

The LuTU AF is a type of AF for NNs. Instead of being defined by a mathematical expression, LuTU is controlled by a look-up table. The look-up table consists of a set of anchor points $\{x_i, y_i\}$, where $i = 0,1,\ldots,n$. The $x$-values $\{x_i\}$ are pre-defined and uniformly spaced with a step size of $s$, while the $y$-values $\{y_i\}$ are learnable parameters. These anchor points are crucial as they determine the rough shape of the LuTU AF. The anchor points $\{x_i, y_i\}$ are configured in such a way that

$$x_i = x_0 + s * i, \tag{8.68}$$

where $x_0$ is a starting point, $s$ is the step size, and $i$ is the index. This configuration ensures that the $x$-values are uniformly spaced with the specified step size. The AF is generated from the anchor points using two methods: linear interpolation and cosine smoothing.

**Linear Interpolation:**

This method likely involves connecting the anchor points with straight line segments. The LuTU AF is then interpolated between these points using linear interpolation. Linear interpolation is a simple method that estimates values between two known points based on a linear equation. The AF based on linear interpolation is defined as

$$\sigma_{\text{LulU}-\mathbf{Interp}}(x) = \frac{1}{s}\left(y_i(x_{i+1} - x) + y_{i+1}(x - x_i)\right), \qquad \text{if } x_i \le x \le x_{i+1}. \tag{8.69}$$

For any input value between $x_i$ and $x_{i+1}$, the output is linearly interpolated from $y_i$ and $y_{i+1}$. When $s \to 0$, $x_0 \to -\infty$ and $n \to \infty$, this AF can approximate any univariate function, see Figure 8.47.

The derivative of $\sigma_{\text{LulU}-\mathbf{Interp}}(x)$ over $y_i$ and input $x$ are straightforward:

$$\frac{\partial}{\partial y_i}\sigma_{\text{LulU}-\mathbf{Interp}}(x) = \frac{x_{i+1} - x}{s}, \tag{8.70.1}$$

$$\frac{\partial}{\partial y_{i+1}}\sigma_{\text{LulU}-\mathbf{Interp}}(x) = \frac{x - x_i}{s}, \tag{8.70.2}$$

$$\frac{\partial}{\partial x}\sigma_{\text{LulU}-\mathbf{Interp}}(x) = \frac{y_{i+1} - y_i}{s}, \qquad x_i \le x \le x_{i+1}. \tag{8.70.3}$$

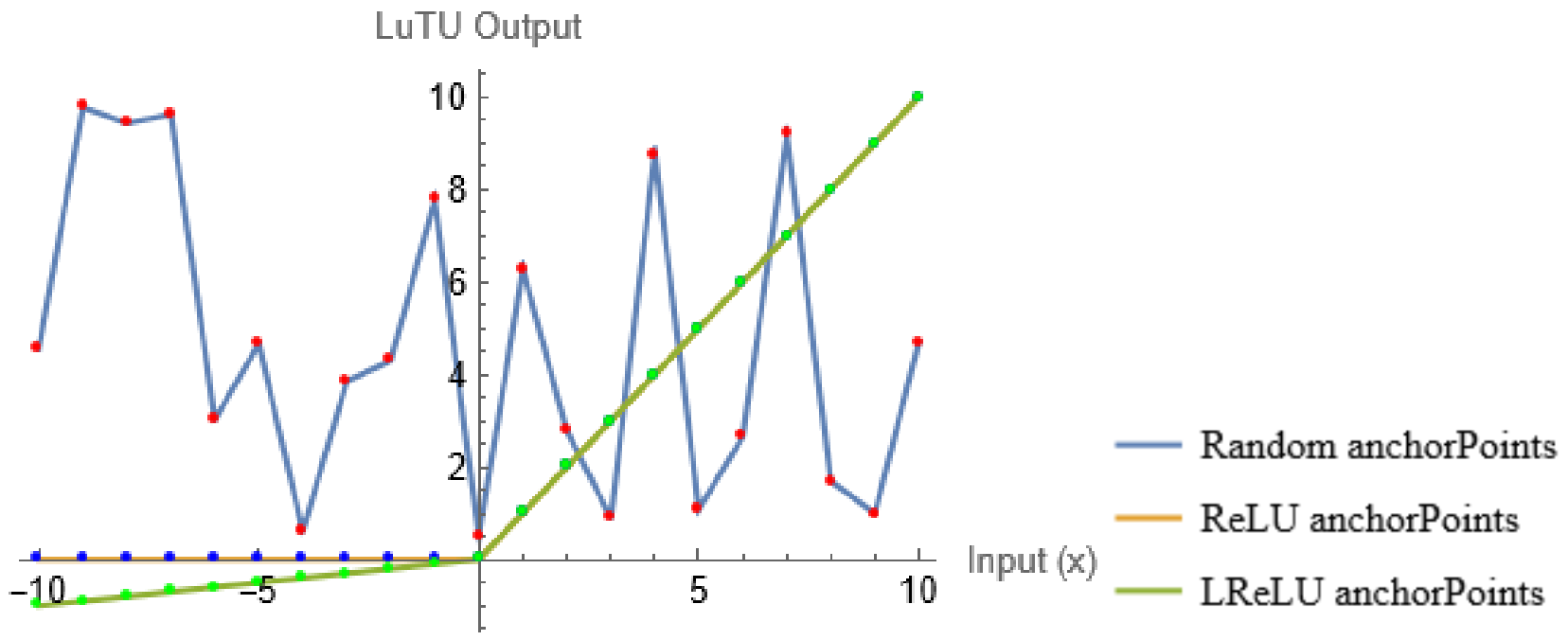

**Figure 8.47.** The figure displays the LuTU AF with linear interpolation applied to three sets of anchor points (21 anchor points): randomly generated points (red), ReLU points (blue), and LReLU points (green). Each set of anchor points defines a distinct LuTU function through linear interpolation, calculated as $\sigma_{\text{LulU}-\textbf{Interp}}(x) = \frac{1}{s}\left(y_i(x_{i+1} - x) + y_{i+1}(x - x_i)\right)$, for $x$ values between consecutive anchor points. The plot shows how the LuTU activation varies based on the different anchor point definitions, with the $x$-axis representing input values from $-10$ to $10$ and the $y$-axis showing the corresponding LuTU output. The legends distinguish the three types of anchor points.

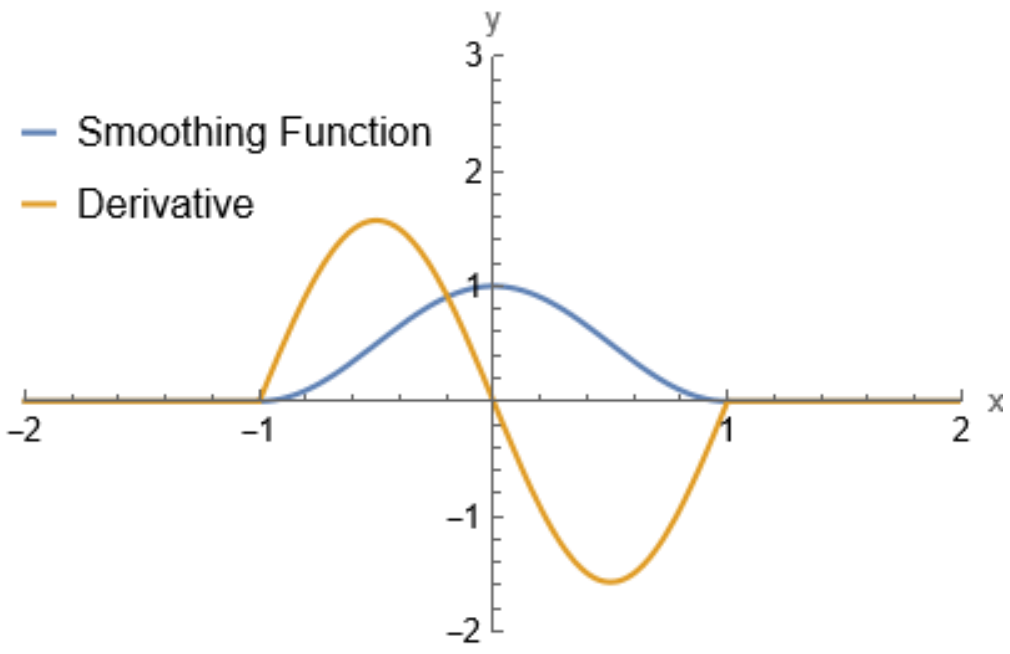

**Figure 8.48.** The figure shows a plot of a smoothing function and its derivative for $\tau = 1$, highlighting their behavior over the range $x = -2$ to $x = 2$. The smoothing function is defined as $\frac{1}{2\tau}\left(1 + \cos\left(\frac{\pi}{\tau}x\right)\right)$ within $[-\tau, \tau]$ and zero elsewhere, while the derivative captures the rate of change of the smoothing function.

**Cosine Smoothing:**

This method involves a more sophisticated interpolation technique known as cosine smoothing. Cosine smoothing suggests that the transition between anchor points is made smoother, possibly to avoid abrupt changes and improve the continuity of the LuTU AF. The smoothing function is defined as:

$$r(x,\tau) = \begin{cases} \frac{1}{2\tau}\left(1 + \cos\left(\frac{\pi}{\tau}x\right)\right), & -\tau \leq x \leq \tau, \\ 0, & \text{otherwise}. \end{cases} \tag{8.71}$$

The function is defined piecewise. For values of $x$ within the interval $[-\tau, \tau]$, it is given by $\frac{1}{2\tau}\left(1 + \cos\left(\frac{\pi}{\tau}x\right)\right)$, and for values outside this interval, the output is zero, Figure 8.48.

- The smoothing function is constructed by shifting and scaling one period of a cosine function. The hyper-parameter $\tau$ controls the period ($2\,\tau$) of the cosine function. This allows flexibility in adjusting the width of the smoothing function.
- The smoothing function is differentiable in all its input domain. This property is advantageous when used as part of AFs in NNs, as differentiability is essential for gradient-based optimization algorithms like backpropagation.

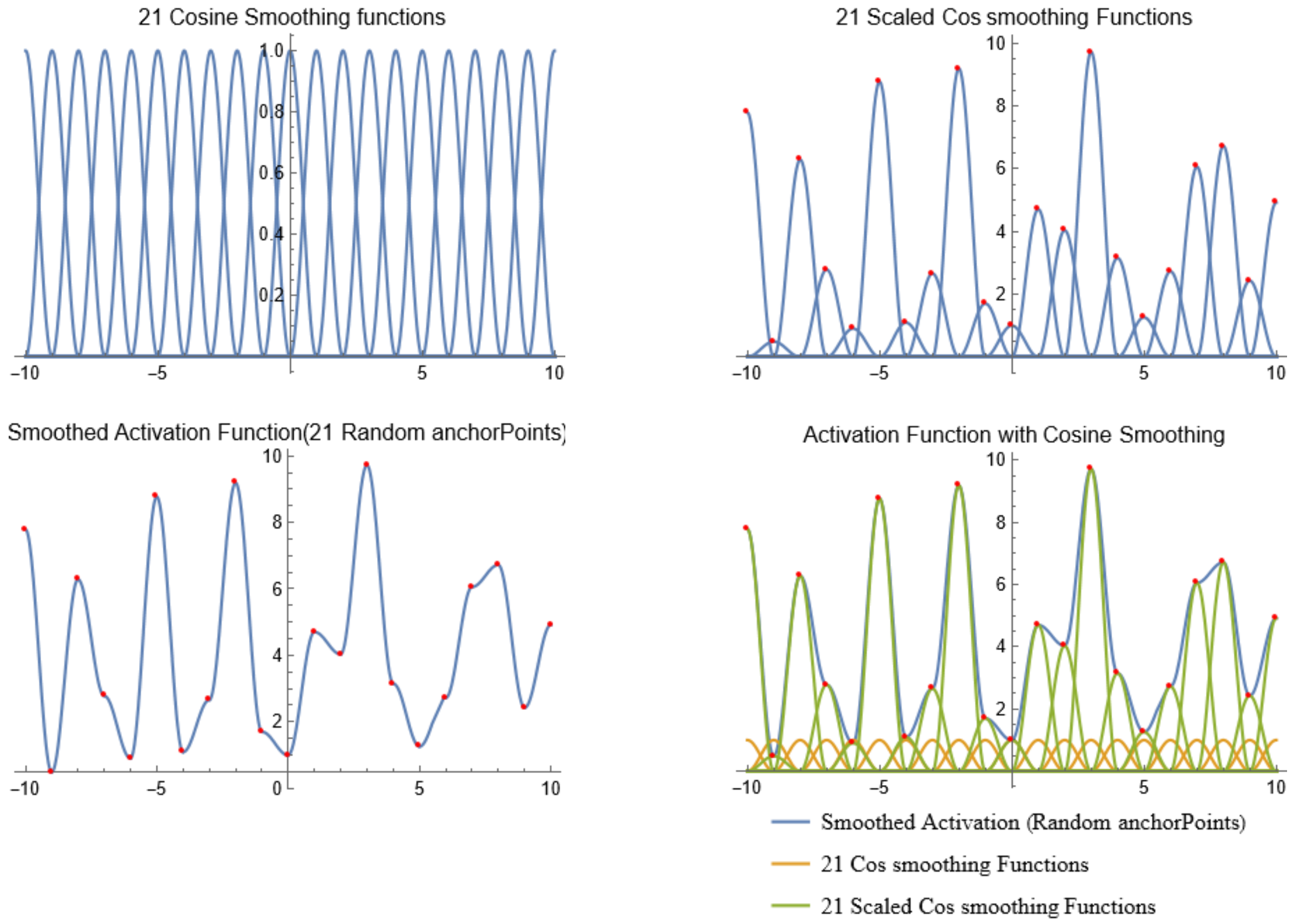

**Figure 8.49.** The figures illustrate the construction of the LuTU AF using cosine smoothing. Top left panel: The figure shows 21 individual cosine smoothing functions centered around anchor points from $-10$ to $10$. Top right panel: The figure scales these functions by random values associated with each anchor point, highlighting their contribution to the overall activation. Bottom left panel: The figure sums these scaled functions to form the smoothed AF, with red points indicating the anchor locations. Bottom right panel: The figure combines all components, demonstrating the integration process of individual and scaled smoothing functions into the final LuTU AF.

- The valid input domain for the smoothing function is limited to one period of the cosine function, i.e., $[-\tau, \tau]$. Outside this interval, the output is zero. This limitation reduces the computational workload when calculating the output of the smoothed AF, as only values within the specified interval contribute to the result.
- The integral of this function over its entire domain is equal to one. This property is important when using the smoothing function to convolve or filter other functions. The integral being one means that smoothing a discrete function with this mask will not change its scale, preserving the overall magnitude of the function.

The overall AF $\sigma_{\text{LulU}-\mathbf{Cos}}(x)$ is defined as the sum of individual terms, each of which involves multiplying a learnable parameter $y_i$ with a smoothing mask $r(x - x_i, t\,s)$. The function is defined as follows:

$$\sigma_{\text{LulU}-\mathbf{Cos}}(x) = \sum_{i=0}^{n} y_i\, r(x - x_i, t\,s). \tag{8.72}$$

The smoothing mask $r(x - x_i, t\,s)$ is generated using the cosine smoothing function, see Figure 8.49. It is centered at $x_i$, and the parameter $t$ determines the ratio between $\tau$ and $s$. The input domain of the smoothing mask is truncated, limited to the interval $[x_i - t\,s, x_i + t\,s]$. This means that for each input value $x$, the calculation of the smoothing masks only involves those masks whose input domain includes $x$. This is a computational optimization, as it reduces the number of masks that need to be evaluated for a given input. The parameter $t$ is an integer that defines the ratio between $\tau$ (the hyper-parameter controlling the period of the cosine function) and $s$ (the step size between anchor points). It influences the width of the smoothing mask and can be adjusted to control the smoothness of the overall AF.

In summary, the AF is a weighted sum of smoothing masks, where each mask is generated by multiplying a learnable parameter with a cosine smoothing function centered at a specific anchor point. The truncated input domain of the smoothing mask allows for efficient computation, considering only the relevant masks for a given input value $x$. The integer parameter $t$ provides a means of controlling the width and smoothness of the individual smoothing masks.

**Mixture of Gaussian Unit:**

Based on the observed learned shapes, the Mixture of Gaussian unit (MoGU) AF is designed to capture shapes similar to those learned with LuTU cosine smoothing but with a reduced number of parameters. This can lead to more efficient learning and potentially faster convergence during training. The AF is defined as the sum of individual Gaussian functions:

$$\sigma_{\text{MoGU}}(x) = \sum_{i=1}^{n} \frac{\lambda_i}{\sqrt{2\pi\sigma_i^2}} e^{-\frac{(x-\mu_i)^2}{2\sigma_i^2}}, \tag{8.73}$$

where,

$\lambda_i$: Learnable parameter controlling the scale of the i-th Gaussian.

$\mu_i$: Learnable parameter representing the mean of the i-th Gaussian.

$\sigma_i$: Learnable parameter controlling the standard deviation of the i-th Gaussian.

Each term in the summation corresponds to a Gaussian function. Gaussian functions are bell-shaped curves determined by their mean $\mu_i$ and standard deviation $\sigma_i$. The scale of each Gaussian is controlled by $\lambda_i$.

### 8.7.5 Bi-Modal Derivative Sigmoidal AFs

A Bi-modal Derivative Adaptive Activation (BDAA) function uses twin maxima derivative sigmoidal function [265] by controlling the maxima's position with an adaptive parameter. There are four types of BDAA AF.

The BDAA1 is given as,

$$\sigma_{\text{BDAA1}}(x) = \frac{1}{2}\left(\sigma_{\text{Logistic}}(x) + \sigma_{\text{Logistic}}(x+a)\right), \tag{8.74.1}$$

$$\frac{\partial}{\partial x}\sigma_{\text{BDAA1}}(x) = \frac{1}{2}\left(\sigma_{\text{Logistic}}(x)\left(1-\sigma_{\text{Logistic}}(x)\right) + \sigma_{\text{Logistic}}(x+a)\left(1-\sigma_{\text{Logistic}}(x+a)\right)\right), \tag{8.74.2}$$

$$\frac{\partial}{\partial a}\sigma_{\text{BDAA1}}(x) = \frac{\sigma_{\text{Logistic}}(x+a)\left(1-\sigma_{\text{Logistic}}(x+a)\right)}{2}, \tag{8.74.3}$$

where $a \in \mathbb{R}$ and $\sigma_{\text{Logistic}} = 1/(1+e^{-x})$ is the Logistic-Sigmoid function. The Sigmoid function has an $S$-shaped curve. It maps any real-valued number to the range $(0,1)$. The term $\sigma_{\text{Logistic}}(x+a)$ implies a horizontal shift of the Sigmoid function by $a$ units. The function $\sigma_{\text{BDAA1}}$ takes the average of the original Sigmoid function $\sigma_{\text{Logistic}}(x)$ and the shifted Sigmoid function $\sigma_{\text{Logistic}}(x+a)$. The scaling factor $1/2$ ensures that the resulting function stays within the range $(0,1)$ since each Sigmoid function individually is in that range.

The function, $\sigma_{\text{BDAA1}}$, is not symmetric in $a$, thus, the possible values of $a$ to be considered lie in the interval $(-\infty,\infty)$. The function and its derivatives are shown in Figure. 8.50 for a few values of $a$, as seen from the figure, the derivatives are not symmetric about the $y$-axis, except for the value $a = 0$.

The derivative of the Logistic Sigmoid function, $\frac{\partial}{\partial x}\sigma_{\text{Logistic}}(x)$, has a maximum value at the points where $\sigma_{\text{Logistic}}(x)$ is 0.5, and it approaches 0 as $x$ goes to $-\infty$ or $+\infty$. When you combine this with the derivative of the shifted Sigmoid function $\sigma_{\text{Logistic}}(x+a)$, you introduce another set of points where the derivative has a maximum value. The derivative of the shifted Sigmoid also approaches 0 as $x$ goes to $-\infty$ or $+\infty$, but its maximum value occurs at the points where $\sigma_{\text{Logistic}}(x+a)$ is 0.5. As a result, you get two peaks or modes in the derivative of $\sigma_{\text{BDAA1}}(x)$ with respect to $x$. The positions of these modes depend on the values of $a$ and the locations of the points where the Sigmoid functions have a value of 0.5. Hence, the derivatives of the AF in this case are bi-modal.

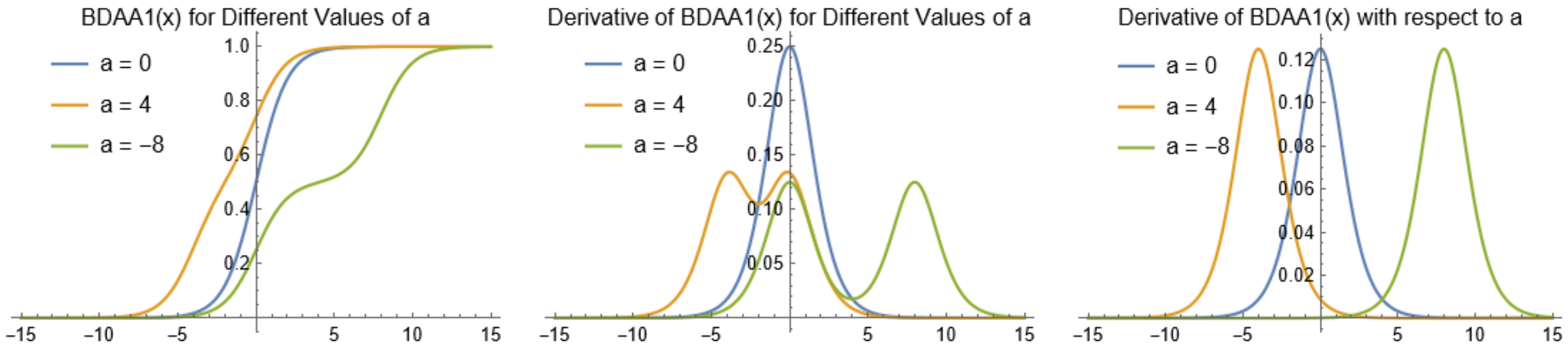

**Figure 8.50.** Left panel: The figure compares the BDAA1 function for different values of the parameter $a$ (0, 4, and $-8$) over a range of $x$ from $-15$ to 15. The BDAA1 function combines two logistic Sigmoid functions shifted by $a$. The plot demonstrates how varying $a$ shifts and blends the sigmoidal curves, influencing the function's output. When $a = 0$, the curve resembles a standard Logistic Sigmoid, while positive and negative $a$ values create asymmetry, modifying the steepness and position of the transition. Middle panel: The figure displays the derivatives of the BDAA1 function for different values of the parameter $a$ (0, 4, and $-8$) over a range of $x$ from $-15$ to 15. The BDAA1 derivative combines the derivatives of two Logistic Sigmoid functions shifted by $a$. The plot illustrates how the value of $a$ affects the derivative's shape and position. When $a = 0$, the derivative resembles the standard Logistic Sigmoid derivative, while positive and negative values of $a$ create variations in the peaks and transition points. Right panel: The figure displays the derivatives of the BDAA1 function with respect to the parameter $a$, plotted for three different values of $a$ (0, 4, and $-8$) over a range of $x$ from $-15$ to 15. The plot illustrates the behavior of these derivatives, showing how the curves change as $a$ varies, with $a = 0$ representing the standard derivative and other values introducing shifts and changes in the peak positions.

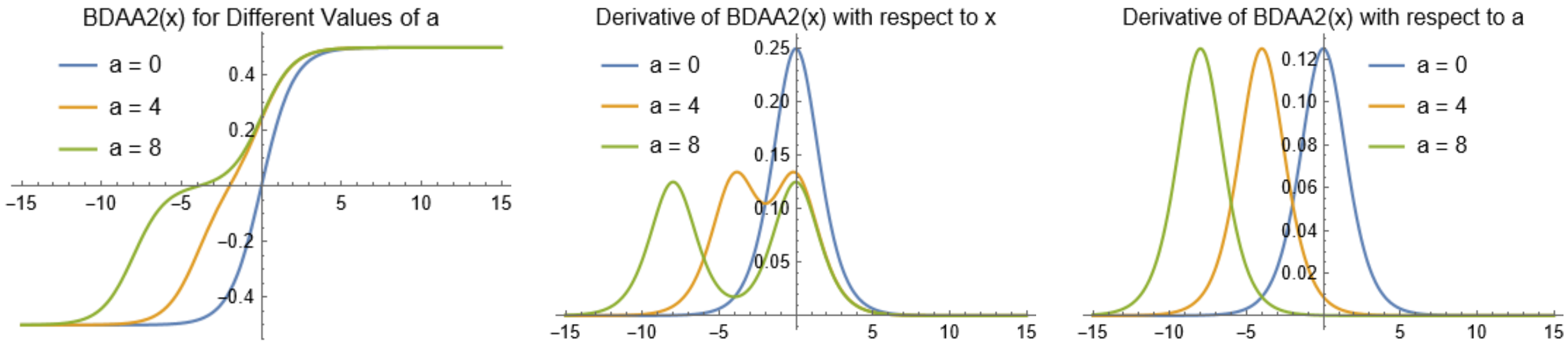

**Figure 8.51.** Similar to Figure 8.50 but using the BDAA2 function.

This bi-modal behavior can have interesting implications in the context of NNs, especially in terms of learning and convergence dynamics. It indicates that there are two regions in the input space where the network's weights are updated more significantly. Understanding such behaviors can provide insights into the learning dynamics of NNs with this specific AF.

The BDAA2 is given as,

$$\sigma_{\text{BDAA2}}(x) = \sigma_{\text{BDAA1}} - \frac{1}{2} = \frac{1}{2}\left(\sigma_{\text{Logistic}}(x) + \sigma_{\text{Logistic}}(x+a) - 1\right), \quad (8.75.1)$$

$$\frac{\partial}{\partial x}\sigma_{\text{BDAA2}}(x) = \frac{\partial}{\partial x}\sigma_{\text{BDAA1}}(x), \quad (8.75.2)$$

$$\frac{\partial}{\partial a}\sigma_{\text{BDAA2}}(x) = \frac{\partial}{\partial a}\sigma_{\text{BDAA1}}(x). \quad (8.75.3)$$

The subtracted constant $1/2$ shifts the function downwards, affecting the baseline. The subtraction does not affect the shape of the Sigmoid functions themselves but influences their interaction and the overall position of the function. The entire curve of $\sigma_{\text{BDAA1}}(x)$ is shifted downward by $1/2$. This means that the function's values are lowered, and the new baseline (the level at which the function approaches for large negative or positive $x$ is set to $\pm$½). Before the subtraction, the baseline was 1.

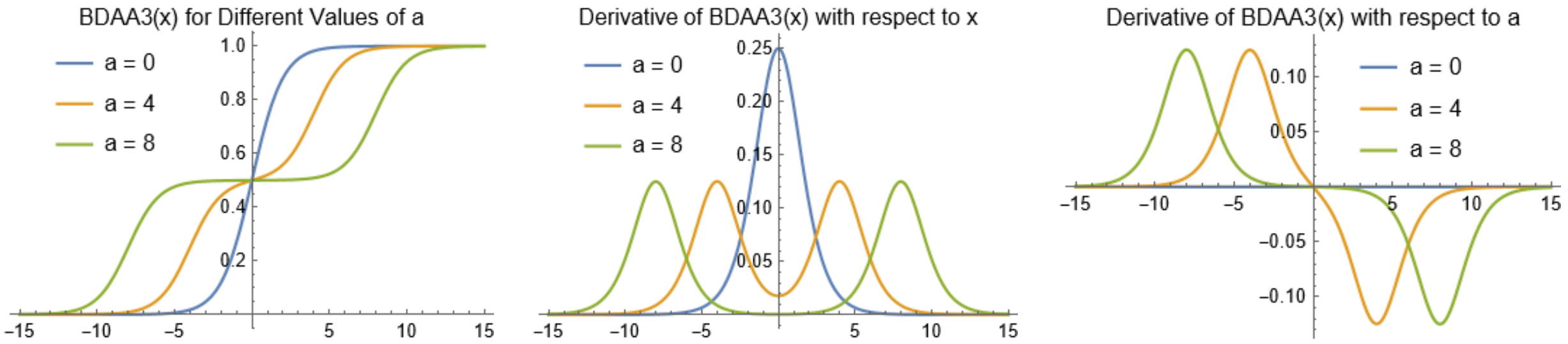

**Figure 8.52.** Similar to Figure 8.50 but using the BDAA3 function.

This function is not symmetric in $a$, thus, the values of $a$ to be considered lie in the interval $(-\infty, \infty)$. The function and its derivatives are shown in Figure. 8.51 for a few values of $a$, as seen from the figure, the derivatives are asymmetric about the $y$-axis, and except for $a = 0$, the function $\sigma_{\text{BDAA2}}(x)$ is not antisymmetric.

The BDAA3 is a combination of two Sigmoid functions with positive and negative shifts, resulting in a function that is influenced by both shifts with equal weight. The symmetry of the Sigmoid terms plays a role in determining the overall symmetry of $\sigma_{\text{BDAA3}}(x)$. The BDAA3 is given as,

$$\sigma_{\text{BDAA3}}(x) = \frac{1}{2}\Big(\sigma_{\text{Logistic}}(x+a) + \sigma_{\text{Logistic}}(x-a)\Big), \tag{8.76.1}$$

$$\begin{aligned}\frac{\partial}{\partial x}\sigma_{\text{BDAA3}}(x) = \frac{1}{2}\Big(&\sigma_{\text{Logistic}}(x+a)\left(1-\sigma_{\text{Logistic}}(x+a)\right) \\ &+ \sigma_{\text{Logistic}}(x-a)\left(1-\sigma_{\text{Logistic}}(x-a)\right)\Big),\end{aligned} \tag{8.76.2}$$

$$\begin{aligned}\frac{\partial}{\partial a}\sigma_{\text{BDAA3}}(x) = \frac{1}{2}\Big(&\sigma_{\text{Logistic}}(x+a)\left(1-\sigma_{\text{Logistic}}(x+a)\right) \\ &- \sigma_{\text{Logistic}}(x-a)\left(1-\sigma_{\text{Logistic}}(x-a)\right)\Big).\end{aligned} \tag{8.76.3}$$

The positive shift $(x + a)$ and negative shift $(x - a)$ in the Sigmoid functions introduce translations along the $x$-axis. The value of $a$ determines the extent of these shifts. This function is symmetric in $a$, hence the set of values to be considered for $a$ may be taken as $[0, \infty)$. The function and its derivatives are shown in Figure. 8.52 for a few values of $a$, as seen from the figure, the derivatives are symmetric about the $y$-axis.

The BDAA4 is given as,

$$\sigma_{\text{BDAA4}}(x) = \sigma_{\text{BDAA3}}(x) - \frac{1}{2}, \tag{8.77.1}$$

$$\frac{\partial}{\partial x}\sigma_{\text{BDAA4}}(x) = \frac{\partial}{\partial x}\sigma_{\text{BDAA3}}(x), \tag{8.77.2}$$

$$\frac{\partial}{\partial a}\sigma_{\text{BDAA4}}(x) = \frac{\partial}{\partial a}\sigma_{\text{BDAA3}}(x). \tag{8.77.3}$$

This function is symmetric in $a$ , thus, the possible values of $a$ to be considered lie in the interval $[0, \infty)$. In this case, the derivatives are symmetric about the $y$-axis. The function and its derivatives are shown in Figure. 8.53 for a few values of $a$, as seen from the figure, the derivatives are symmetric about the $y$-axis, and the function $\sigma_{\text{BDAA4}}(x)$ is anti-symmetric for all values of $a$.

The functions $\sigma_{\text{BDAA1}}(x)$, $\sigma_{\text{BDAA2}}(x)$, $\sigma_{\text{BDAA3}}(x)$ and $\sigma_{\text{BDAA4}}(x)$ are bounded, continuous, differentiable, and sigmoidal functions satisfying the relations:

$$\lim_{x\to\infty}\sigma_{\text{BDAA1}}(x) = \lim_{x\to\infty}\sigma_{\text{BDAA3}}(x) = 1, \lim_{x\to-\infty}\sigma_{\text{BDAA1}}(x) = \lim_{x\to-\infty}\sigma_{\text{BDAA3}}(x) = 0, \tag{8.78.1}$$

$$\lim_{x\to\infty}\sigma_{\text{BDAA2}}(x) = \lim_{x\to\infty}\sigma_{\text{BDAA4}}(x) = \frac{1}{2}, \lim_{x\to-\infty}\sigma_{\text{BDAA2}}(x) = \lim_{x\to-\infty}\sigma_{\text{BDAA4}}(x) = -\frac{1}{2}. \tag{8.78.2}$$

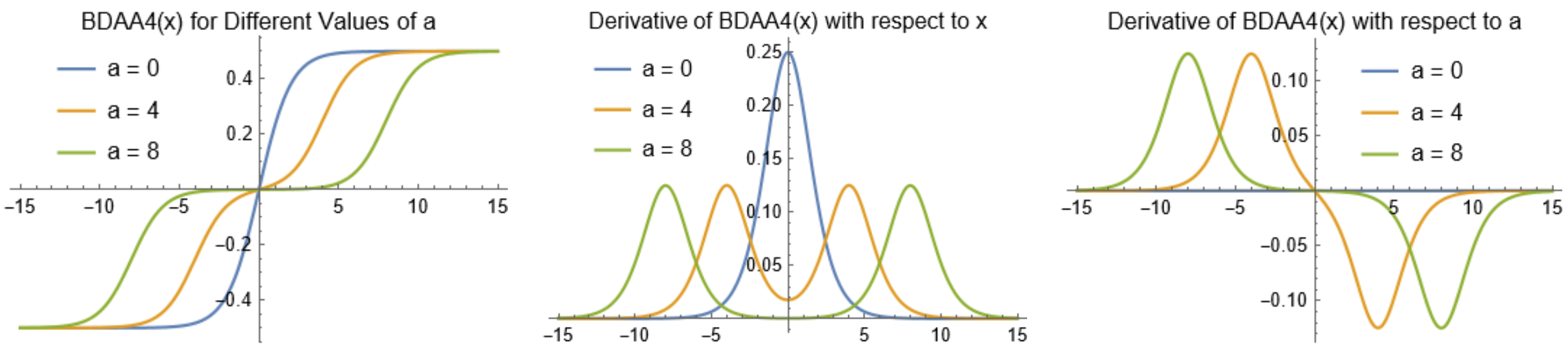

**Figure 8.53.** Similar to Figure 8.50 but using the BDAA4 function.

**Remarks:**

- The functions $\sigma_{\text{BDAA1}}(x)$, $\sigma_{\text{BDAA2}}(x)$, $\sigma_{\text{BDAA3}}(x)$ and $\sigma_{\text{BDAA4}}(x)$ are monotonically increasing functions for any value of $a$.
- The derivative functions $\sigma'_{\text{BDAA1}}(x)$, $\sigma'_{\text{BDAA2}}(x)$, $\sigma'_{\text{BDAA3}}(x)$ and $\sigma'_{\text{BDAA4}}(x)$ have two local maxima for $a \neq 0$, that is, these functions are bi-modal. The functions $\sigma'_{\text{BDAA1}}(x)$, and $\sigma'_{\text{BDAA2}}(x)$ have local maxima at the point $x = 0$ and $x = -a$, while the functions $\sigma'_{\text{BDAA3}}(x)$, and $\sigma'_{\text{BDAA4}}(x)$ have the local maxima at $x = \pm a$.

## 8.8 Performance Comparison and Analysis

Comparing the performance of AFs in NNs requires a systematic and comprehensive approach to ensure fairness and reliability. The following best practices are recommended for making such comparisons:

1. Use a standardized experimental setup by employing the same network architecture across all experiments. This includes keeping the number of layers, number of neurons per layer, and the type of layers used consistent. Additionally, maintain constant hyperparameters like learning rate, batch size, and number of epochs. Furthermore, apply the same weight initialization method to ensure that initial conditions do not bias the results.
2. Utilize the same training and testing datasets for all experiments and apply identical pre-processing techniques such as normalization, augmentation, and handling of missing data. This ensures uniformity in the data input across experiments.
3. Implement $k$-fold cross-validation to ensure that the results are not dependent on a particular train-test split. This provides a more robust measure of performance by averaging results over multiple data partitions.
4. Evaluate the models using multiple metrics. For classification tasks, use (for example) accuracy, precision, and F1-score, and for regression tasks, use mean squared error and $R^2$. It is crucial to choose metrics that are most relevant to the specific task being addressed.
5. Apply statistical analysis using tests such as T-tests or ANOVA to determine if differences in performance are statistically significant. Report confidence intervals for performance metrics to provide a range within which the true performance likely falls.
6. Measure and compare the computational efficiency of each AF by assessing the time taken to train the network and the time required for inference.
7. Conduct multiple runs of each experiment to ensure consistency of results and eliminate the possibility of outcomes being influenced by random variation. Test with different random initializations to confirm the robustness of the results.
8. Visualize the training process by plotting learning curves to illustrate the training and validation performance over epochs. This comprehensive approach will ensure that the comparisons of AFs are fair, reliable, and insightful.

In order to compare the AFs, two experiments are conducted in this chapter, one for regression and another for classification. The Mathematica framework is used in all the experiments. All experiments are performed over a desktop system.

### 8.8.1 Regression Experiment

*Objective:*

The primary objective of this experiment is to systematically evaluate the performance of various AFs in a NN. This will be achieved by closely monitoring several key metrics throughout the training process:

- Gradients Root Mean Square (RMS) per Batch for Each Layer: This metric will provide insights into the stability and magnitude of gradients at each layer during individual training batches.
- Gradients RMS per Epoch (or Round) for Each Layer: By examining this metric, we can observe the overall trend and stability of gradient magnitudes across all layers over each training epoch (or round).
- Weights RMS per Epoch for Each Layer: This will allow us to track the evolution and scaling of weights for each layer as training progresses, offering insights into weight dynamics.
- Round Loss: This metric reflects the loss incurred during each training round, helping us gauge the immediate impact of different AFs on training efficacy.
- Validation Loss: Monitoring validation loss will enable us to assess how well the network generalizes to unseen data, providing a measure of the practical performance of each AF.

By systematically analyzing these metrics, we aim to gain a comprehensive understanding of how different AFs influence the training dynamics and overall performance of the NN.

*Experiment Design:*

- Generating Training Data: The synthetic training data is generated using a Gaussian-modulated exponential function, $f(x) = e^{-x^2}$, which produces a smooth, bell-shaped curve. To simulate real-world data variability, Gaussian noise is added to the function values. The noise level is set to $0.15$, meaning the noise is sampled from a normal distribution with a mean of $0$ and a standard deviation of $0.15$. The training data is generated for $x$ values ranging from $-3$ to $3$ with a step size of $0.01$, resulting in a dense set of data points.
- AFs: The experiment evaluates 12 different AFs: ReLU, ELU, SELU, GELU, Swish, HardSwish, Mish, SoftPlus, HardTanh, HardSigmoid, Sigmoid, and Tanh. Each AF has unique properties affecting gradient flow and training stability.
- NN Architecture: The NN consists of three hidden layers, each followed by an AF from the set being evaluated. Each hidden layer has 10 neurons followed by the specified AF. The output layer is a linear layer with one neuron, suitable for regression tasks.
- Training Process: Mean Squared Loss (MSE) is used as the training objective, which measures the average squared difference between the predicted and actual values. The Adam optimizer is used for training, known for its adaptive learning rate capabilities and efficiency in handling sparse gradients. The learning rate is set to 0.01, a commonly used value for the Adam optimizer. The training data is processed in batches of 64 samples, balancing computational efficiency and gradient estimation accuracy. The training runs for a maximum of 50 epochs (rounds).
- Monitoring Metrics: During training, the Gradients RMS per batch, Figure 8.54, Gradients RMS per epoch, Figure 8.55, weights RMS per epoch, Figure 8.56, round loss and validation loss, Figure 8.57, are monitored to ensure proper training dynamics. The evolutions of metrics are plotted for each AF, showing how the metrics change over the training iterations.

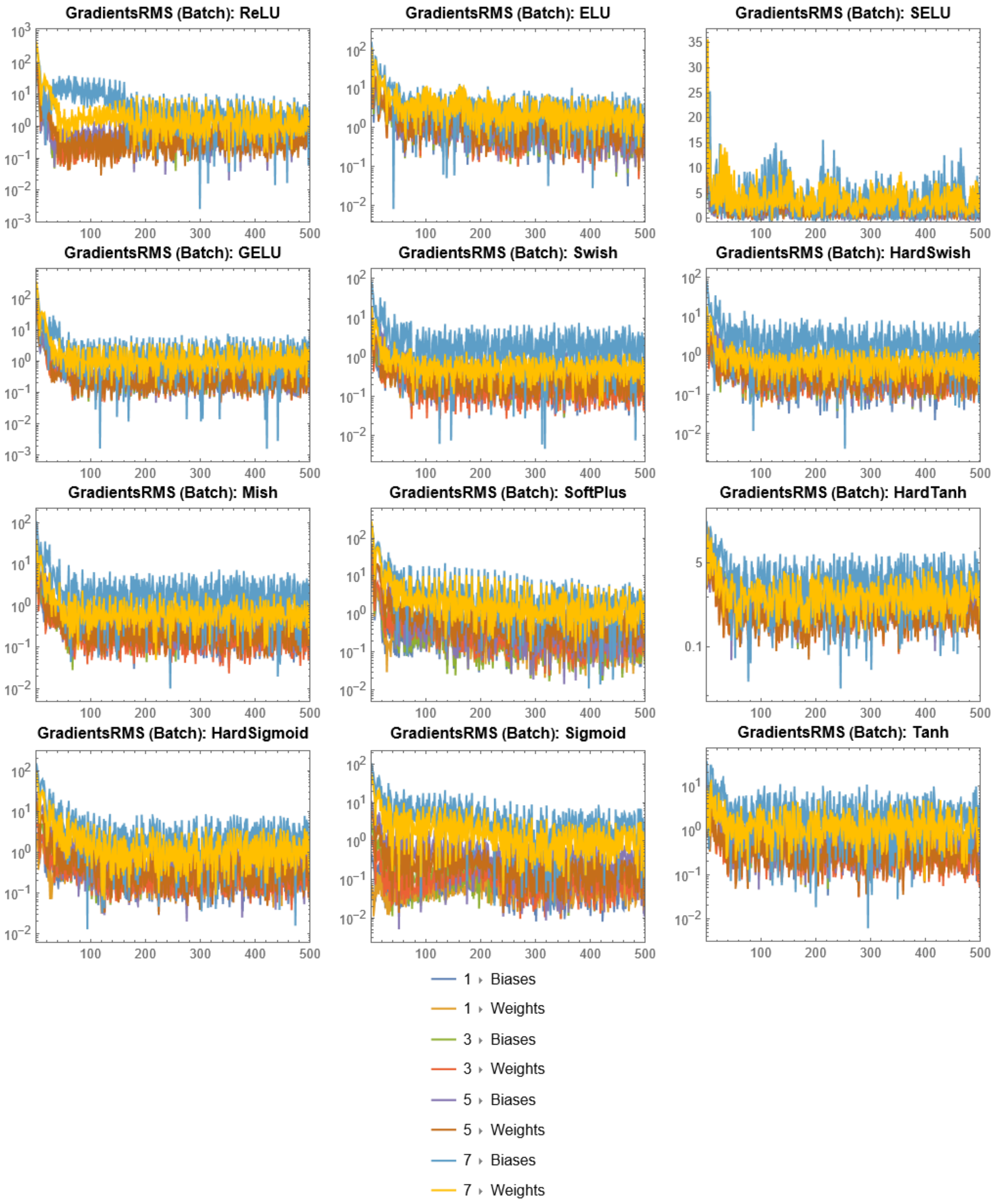

**Figure 8.54.** **(Regression Experiment)** This set of subplots illustrates the Gradients RMS evolution for NNs trained with different AFs: ReLU, ELU, SELU, GELU, Swish, HardSwish, Mish, SoftPlus, HardTanh, HardSigmoid, Sigmoid, and Tanh. Each subplot presents the Gradients RMS monitored at each training batch over a maximum of 50 training rounds (epochs) for each layer. The $x$-axis represents the training batches, while the $y$-axis shows the Gradients RMS values. Monitoring GradientsRMS helps understand the stability and convergence of the training process.

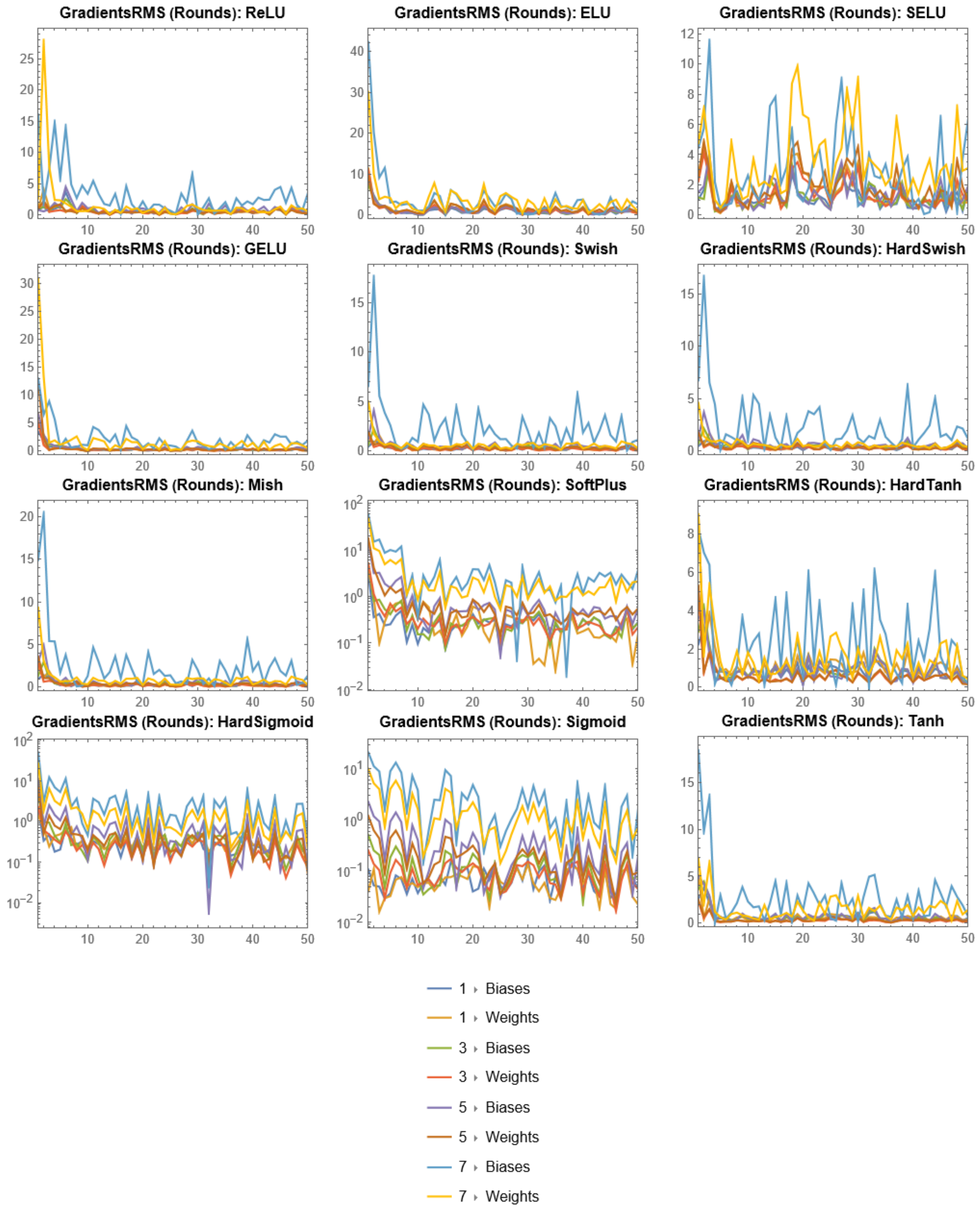

**Figure 8.55. (Regression Experiment)** Similar to Figure.8.54 but each subplot presents the Gradients RMS monitored at each training round (epoch) over a maximum of 50 training rounds for each layer.

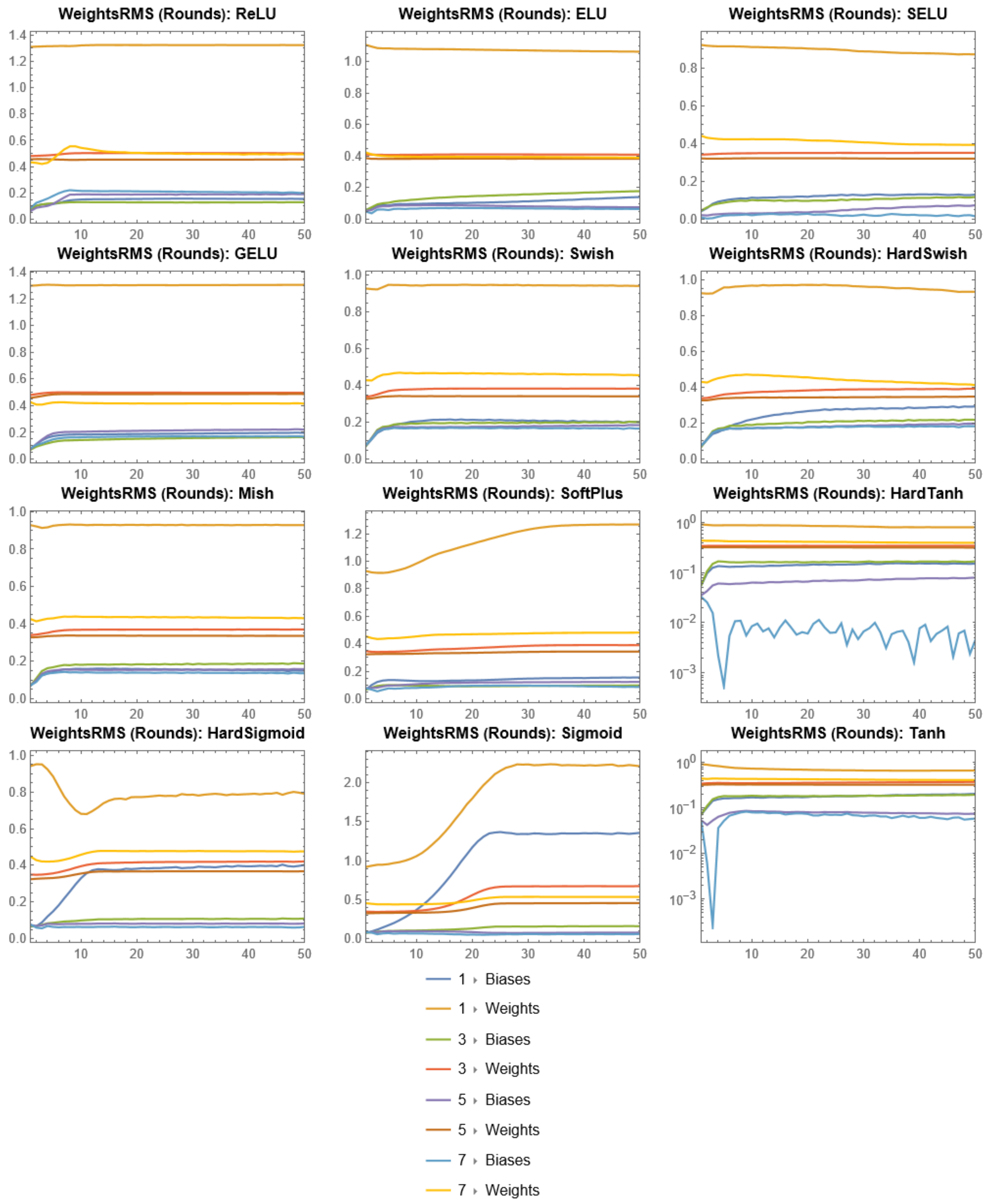

**Figure 8.56.** **(Regression Experiment)** This set of subplots illustrates the weight RMS evolution for NNs trained with different AFs: ReLU, ELU, SELU, GELU, Swish, HardSwish, Mish, SoftPlus, HardTanh, HardSigmoid, Sigmoid, and Tanh. Each subplot presents the weight RMS monitored at each training round over a maximum of 50 training rounds (epochs) for each layer. The $x$-axis represents the training rounds, while the $y$-axis shows the weight RMS values.

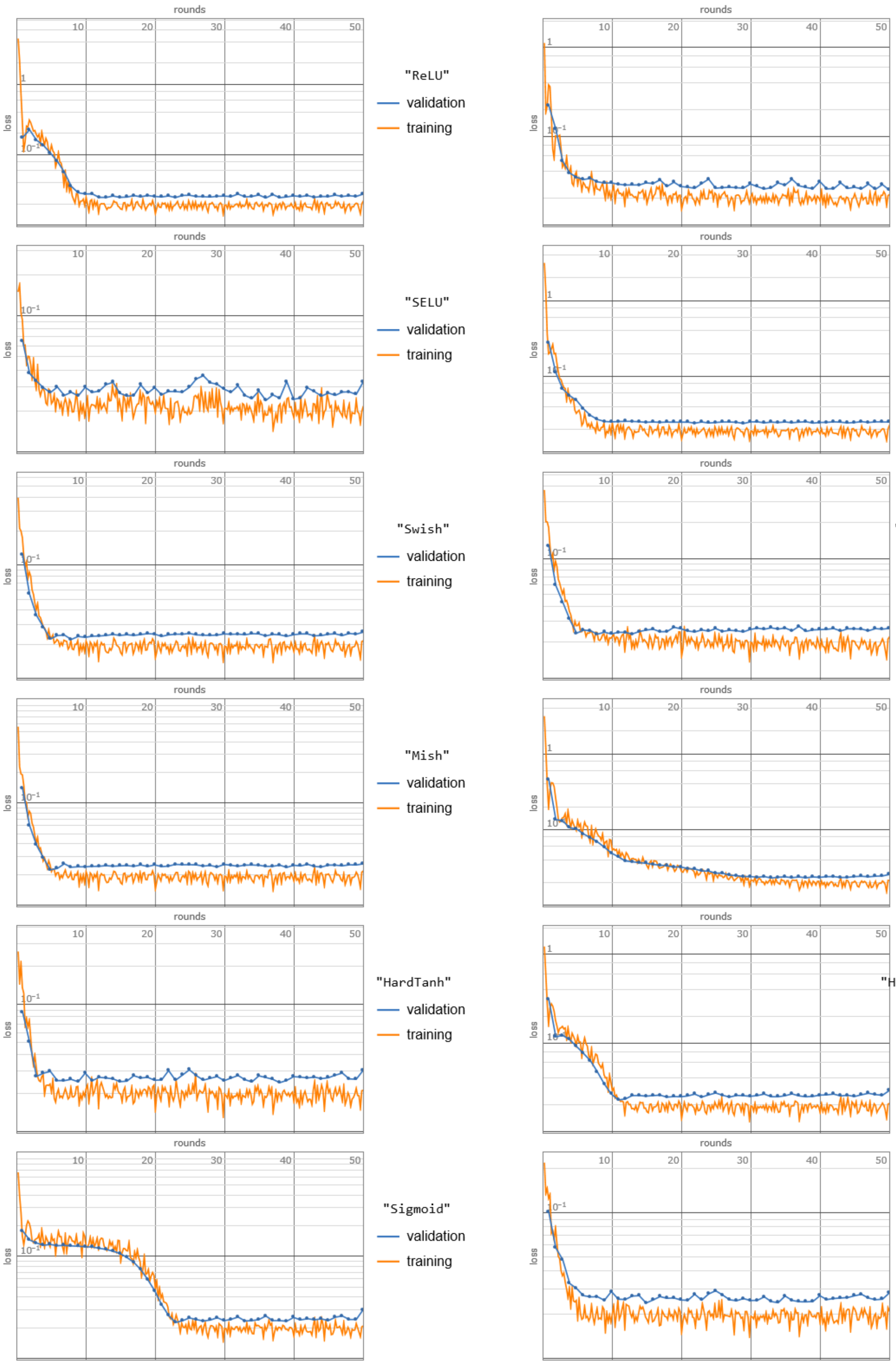

**Figure 8.57. (Regression Experiment)** These plots show the training and validation loss (mean squared error) over 50 training rounds for NNs using various AFs. This visual representation allows for the comparison of training dynamics, highlighting how quickly and effectively each AF minimizes the training and validation loss.

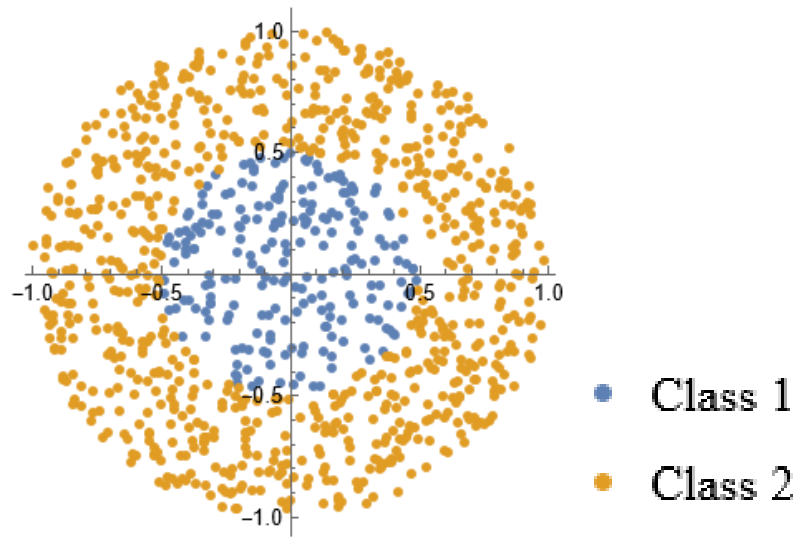

**Figure 8.58.** The synthetic training data generated within a unit disk, labeled based on the distance from the origin. Points are colored differently to represent two classes: Class 1 (inside radius 0.5) and Class 2 (outside radius 0.5 but within the unit disk).

### 8.8.2 Classification Experiment

*Objective*:
The primary objective of this experiment is to generate synthetic training data points within a unit disk, label them based on their distance from the origin, and evaluate the performance of various AFs in a NN. This experiment aims to provide insights into how different AFs impact the training dynamics and classification performance of the NN.

*Experiment Design*:

- Generating Training Data: 1000 random points are generated within a unit disk centered at the origin. Each point is labeled based on its distance from the origin. Points within a radius of 0.5 are labeled as class 1, and points outside this radius but within the unit disk are labeled as class 2. The synthetic training data is visualized using a scatter plot, Figure 8.58, where points from class 1 and class 2 are shown in different colors. This visualization helps inspect the data distribution and verify the labeling process.
- AFs: The experiment evaluates 12 different AFs: ReLU, ELU, SELU, GELU, Swish, HardSwish, Mish, SoftPlus, HardTanh, HardSigmoid, Sigmoid, and Tanh. Each AF has unique properties affecting gradient flow, training stability, and classification performance.
- NN Architecture: The NN consists of two hidden layers, each with 5 neurons followed by one of the specified AFs. The output layer uses a Logistic Sigmoid AF for binary classification. The output layer is followed by a NetDecoder set to "Boolean" to decode the network output into binary labels.
- Training Process: The Adam optimizer is used for training, known for its adaptive learning rate capabilities and efficiency in handling sparse gradients. The learning rate is set to 0.01. The training data is processed in batches of 64 samples. The training runs for a maximum of 50 iterations (rounds).
- Monitoring Metrics: During training, the Gradients RMS per batch, Figure 8.59, Gradients RMS per epoch, Figure 8.60, weights RMS per epoch, Figure 8.61, round loss and validation loss, Figure 8.62, are monitored to ensure proper training dynamics. The evolutions of metrics are plotted for each AF, showing how the metrics change over the training iterations. Moreover, the experiment monitors training progress through confusion matrix plots, Figure 8.63, compares the results, and visualizes the decision boundaries, Figure 8.64, to understand the influence of each AF on training dynamics and classification performance.

This experiment provides a comprehensive analysis of the impact of different AFs on NN training and classification performance. By systematically evaluating and comparing the validation accuracy, confusion matrices, loss plots, and decision boundaries, the experiment highlights the strengths and weaknesses of each AF. These insights guide the selection of appropriate AFs for specific NN architectures and tasks, aiding in the design of more effective NNs.

We also visualized the output landscape of a 3-layer randomly initialized NN with various AFs such as ReLU, ELU, SELU, GELU, Swish, HardSwish, Mish, SoftPlus, HardTanh, HardSigmoid, Sigmoid, and Tanh for visual clarity, as depicted in Figure 8.65. Specifically, we input the 2-dimensional coordinates of each position in a grid into the network and plotted the scalar network output for each grid point. Our observations indicate that AFs significantly influence the smoothness of output landscapes.

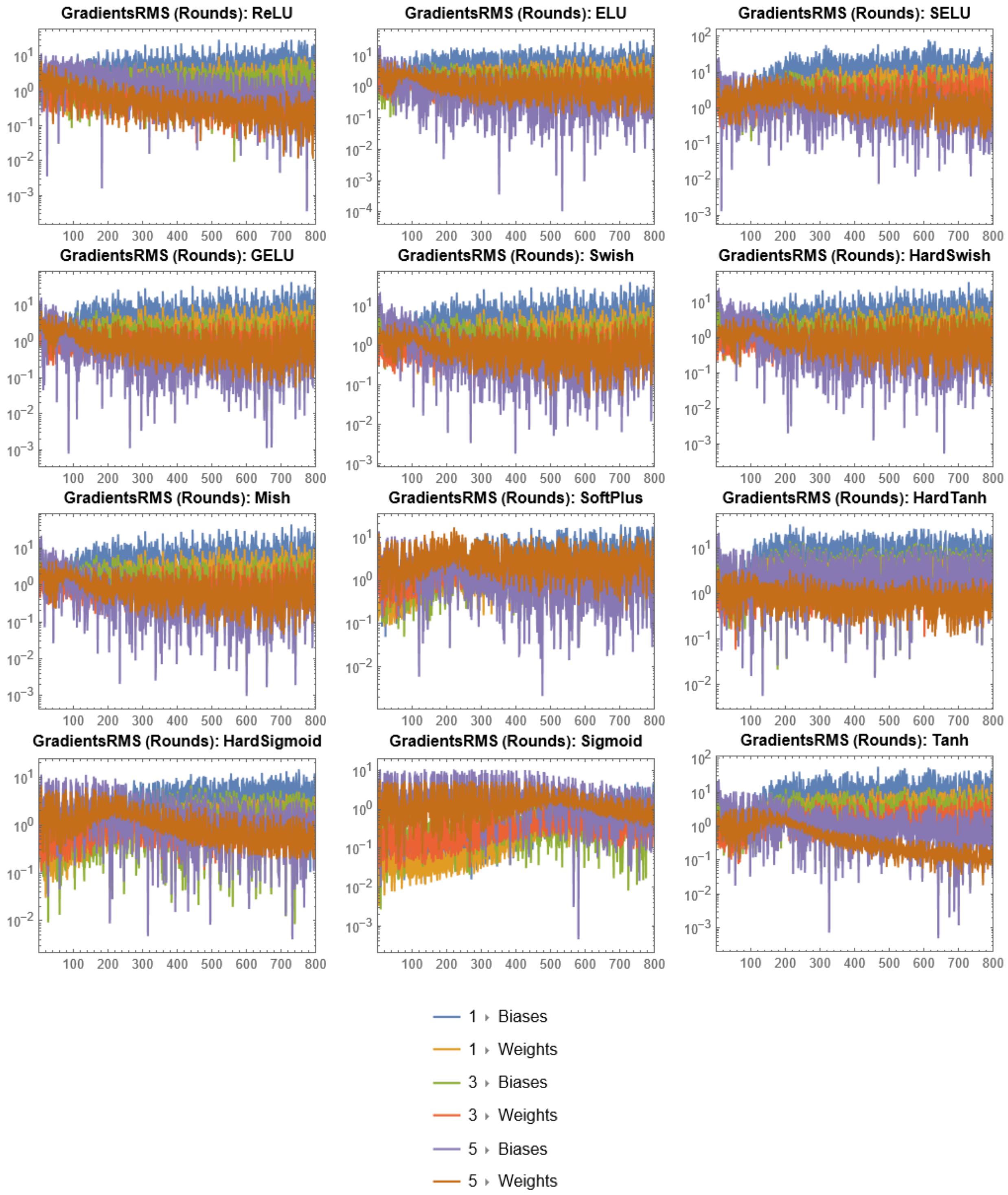

**Figure 8.59.** **(Classification Experiment)** The figure consists of a series of 12 subplots, each depicting the Gradients RMS evolution for a NN trained with a distinct AF: ReLU, ELU, SELU, GELU, Swish, HardSwish, Mish, SoftPlus, HardTanh, HardSigmoid, Sigmoid, and Tanh. Each subplot presents the Gradients RMS monitored at each training batch over a maximum of 50 training rounds (epochs) for each layer. The $x$-axis represents the training batches, while the $y$-axis shows the Gradients RMS values. Monitoring Gradients RMS helps understand the stability and convergence of the training process.

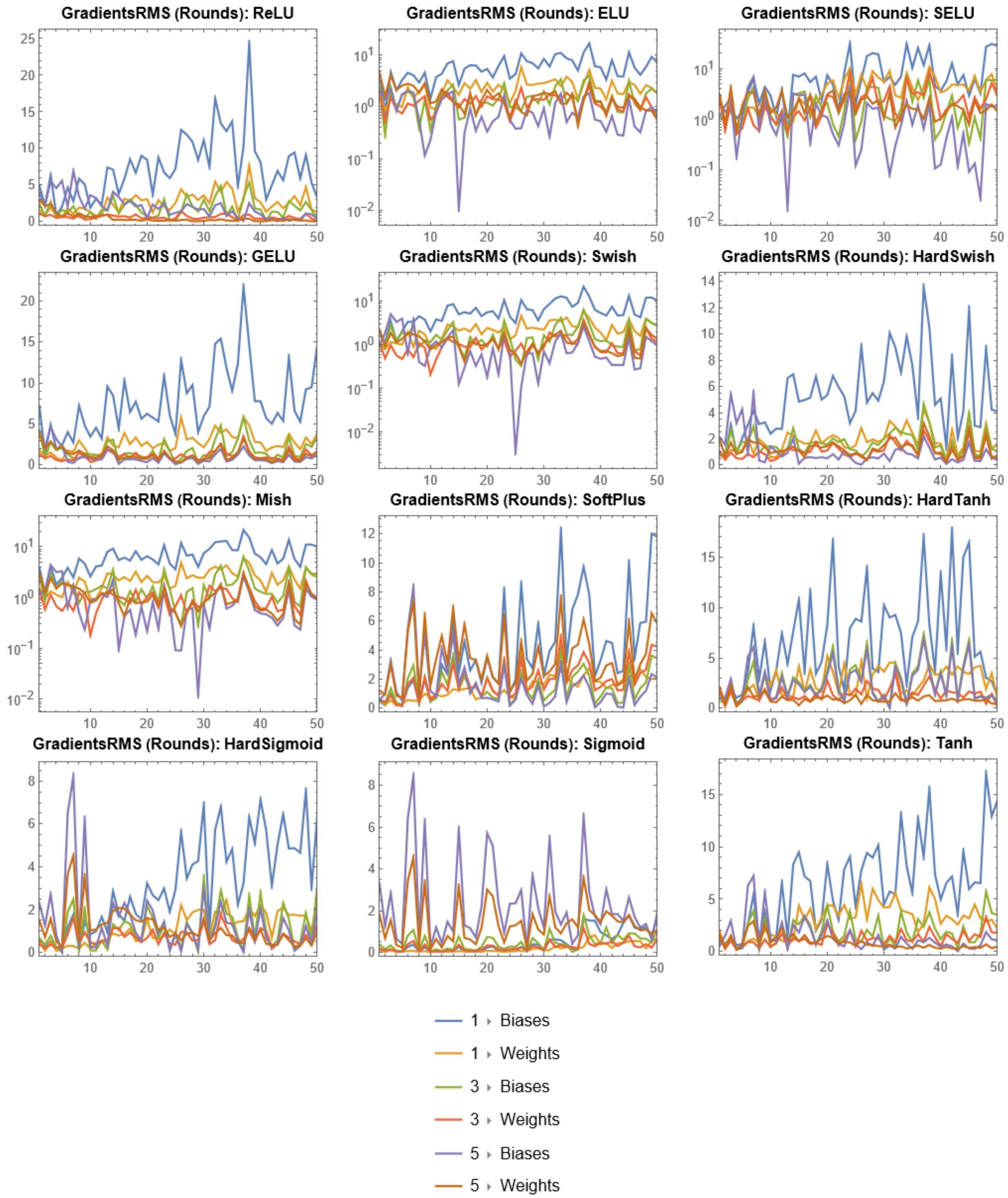

**Figure 8.60. (Classification Experiment)** Similar to Figure 8.59 but each subplot presents the Gradients RMS monitored at each training round over a maximum of 50 training rounds for each layer.

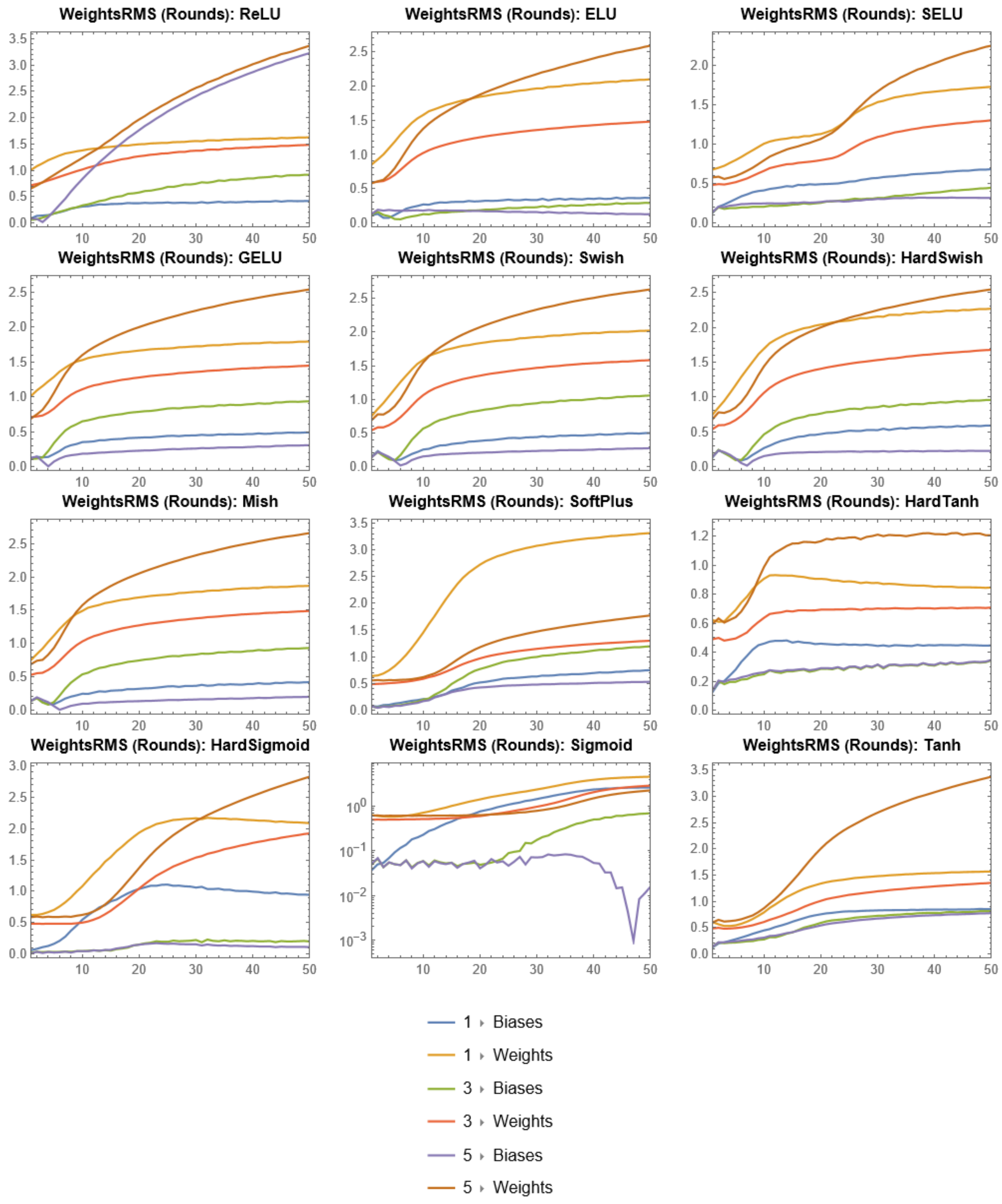

**Figure 8.61. (Classification Experiment)** This set of subplots illustrates the weight RMS evolution for NNs trained with different AFs: ReLU, ELU, SELU, GELU, Swish, HardSwish, Mish, SoftPlus, HardTanh, HardSigmoid, Sigmoid, and Tanh. Each subplot presents the weight RMS monitored at each training round over a maximum of 50 training rounds (epochs) for each layer. The $x$-axis represents the training rounds, while the $y$-axis shows the weight RMS values.

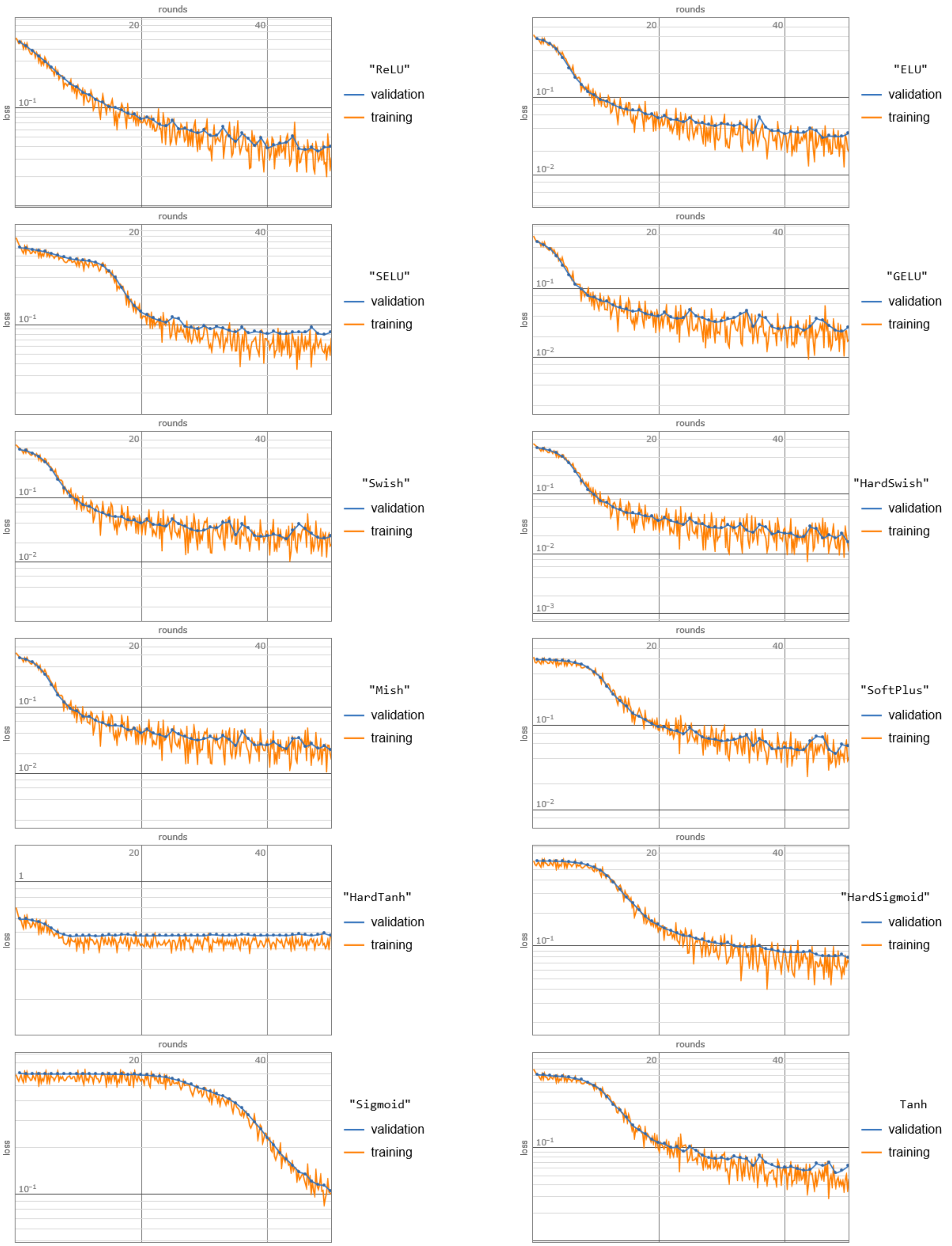

**Figure 8.62.** **(Classification Experiment)** The figure consists of 12 plots, each depicting the training and validation loss evolution for a NN trained with a distinct AF for classification. The loss plots are essential for understanding how quickly and effectively each AF minimizes the loss during training and validation. Comparing these plots helps identify which AFs provide the fastest and most stable convergence. Functions that show consistent and rapid loss reduction are preferred for their efficiency in optimizing the NN.

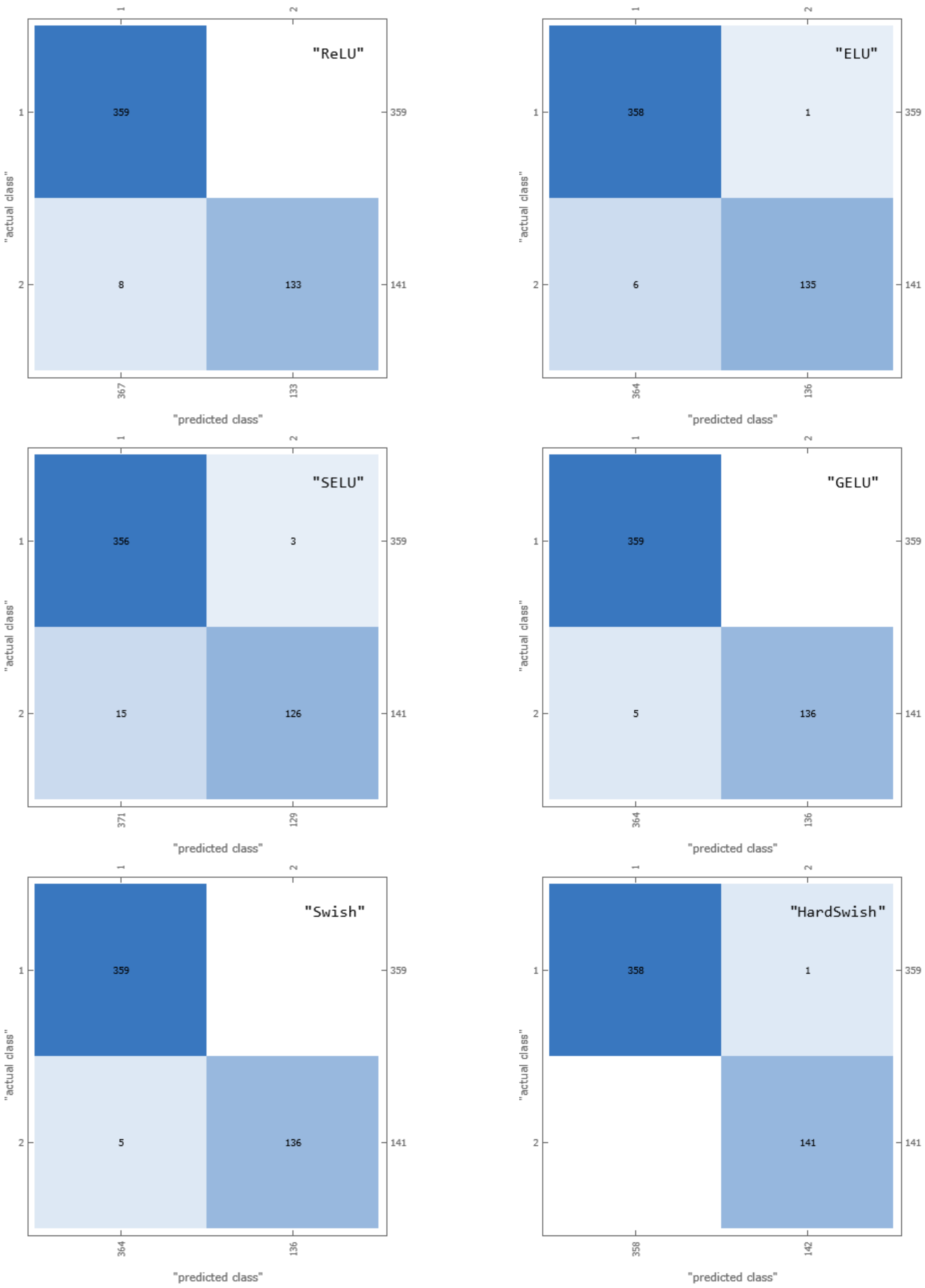

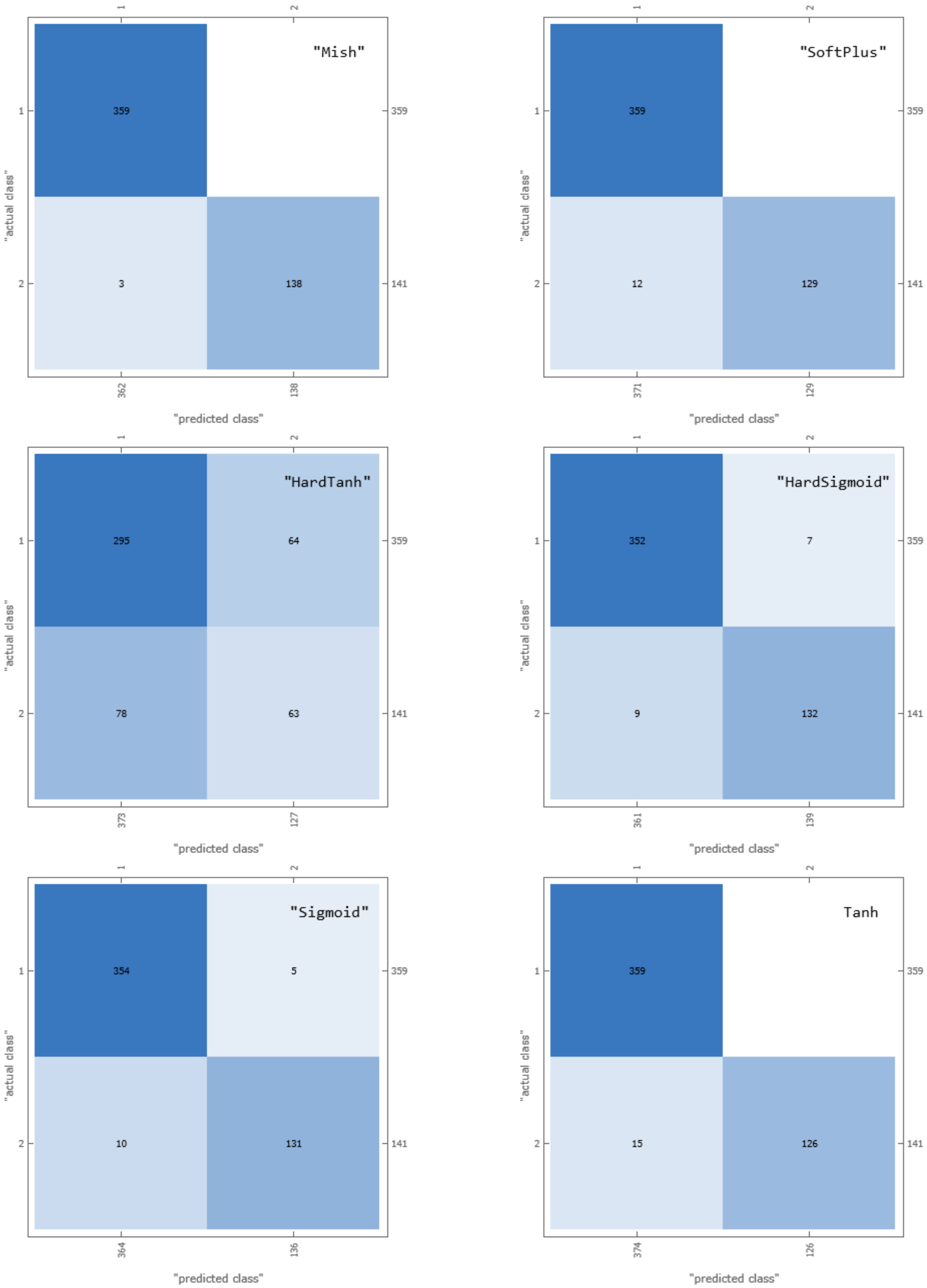

**Figure 8.63. (Classification Experiment)** The figure consists of 12 plots, each depicting the confusion matrix for a NN trained with a distinct AF. The confusion matrix provides detailed insights into the classification performance. The $x$-axis and $y$-axis represent the predicted and actual classes, respectively. The plot title includes the AF used. The entries in the matrix show the counts of true positives, true negatives, false positives, and false negatives. A balanced and diagonal-heavy confusion matrix indicates good classification performance. Comparing these plots helps identify which AFs result in better discrimination between classes.

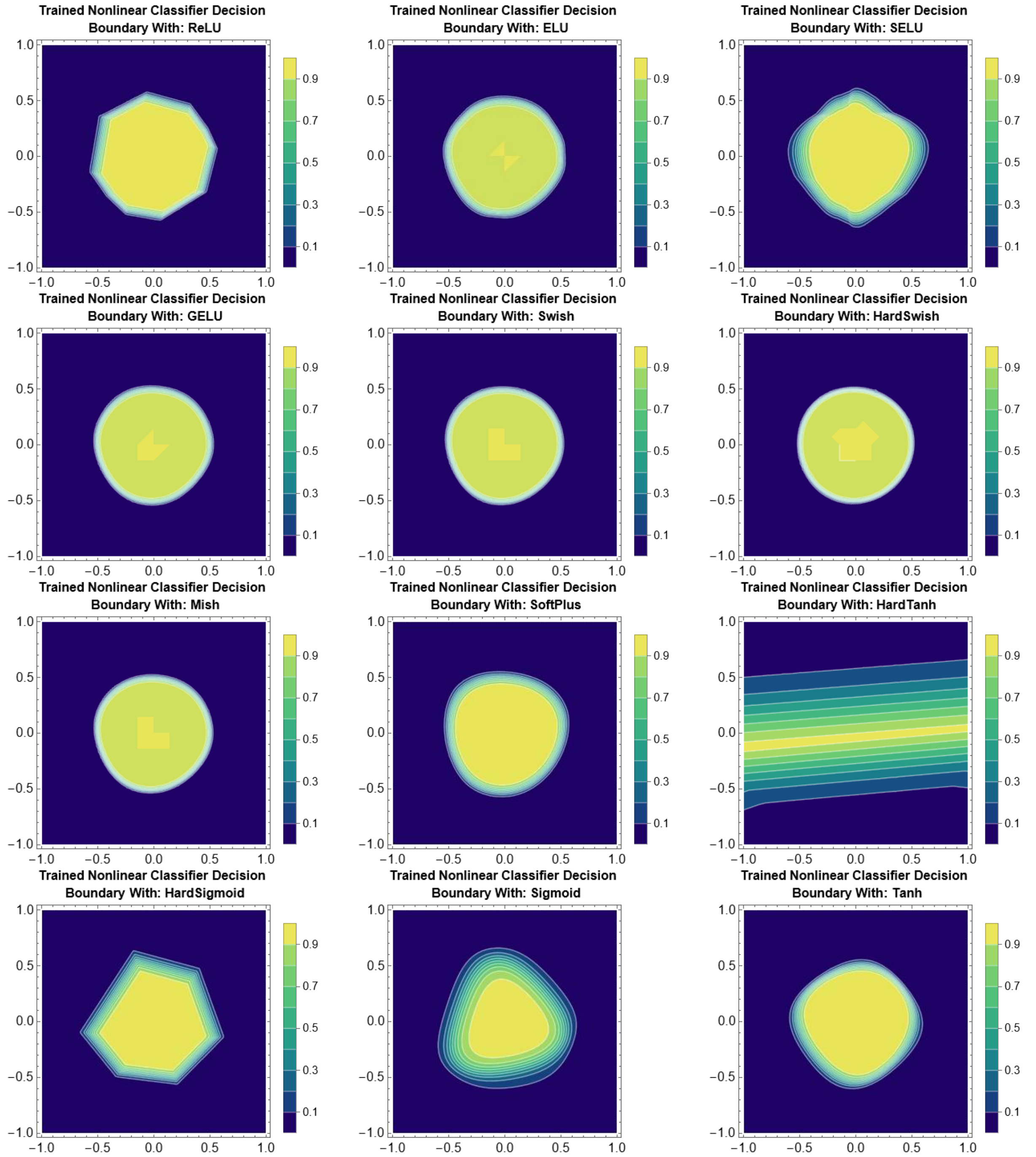

**Figure 8.64. (Classification Experiment)** The figure consists of 12 plots, each depicting the decision boundary for a NN trained with a distinct AF. The decision boundary illustrates how the network classifies different regions of the input space. The $x$-axis and $y$-axis represent the coordinates of the input space within the unit disk. The plot title includes the AF used. Different colors in the plot represent different classes. A well-defined and smooth decision boundary indicates effective learning and classification by the network. Comparing these plots helps understand the influence of each AF on the network's decision-making process. The figure provides a comprehensive visualization of how different AFs affect the classification regions, aiding in the selection of suitable AFs for future NN designs.

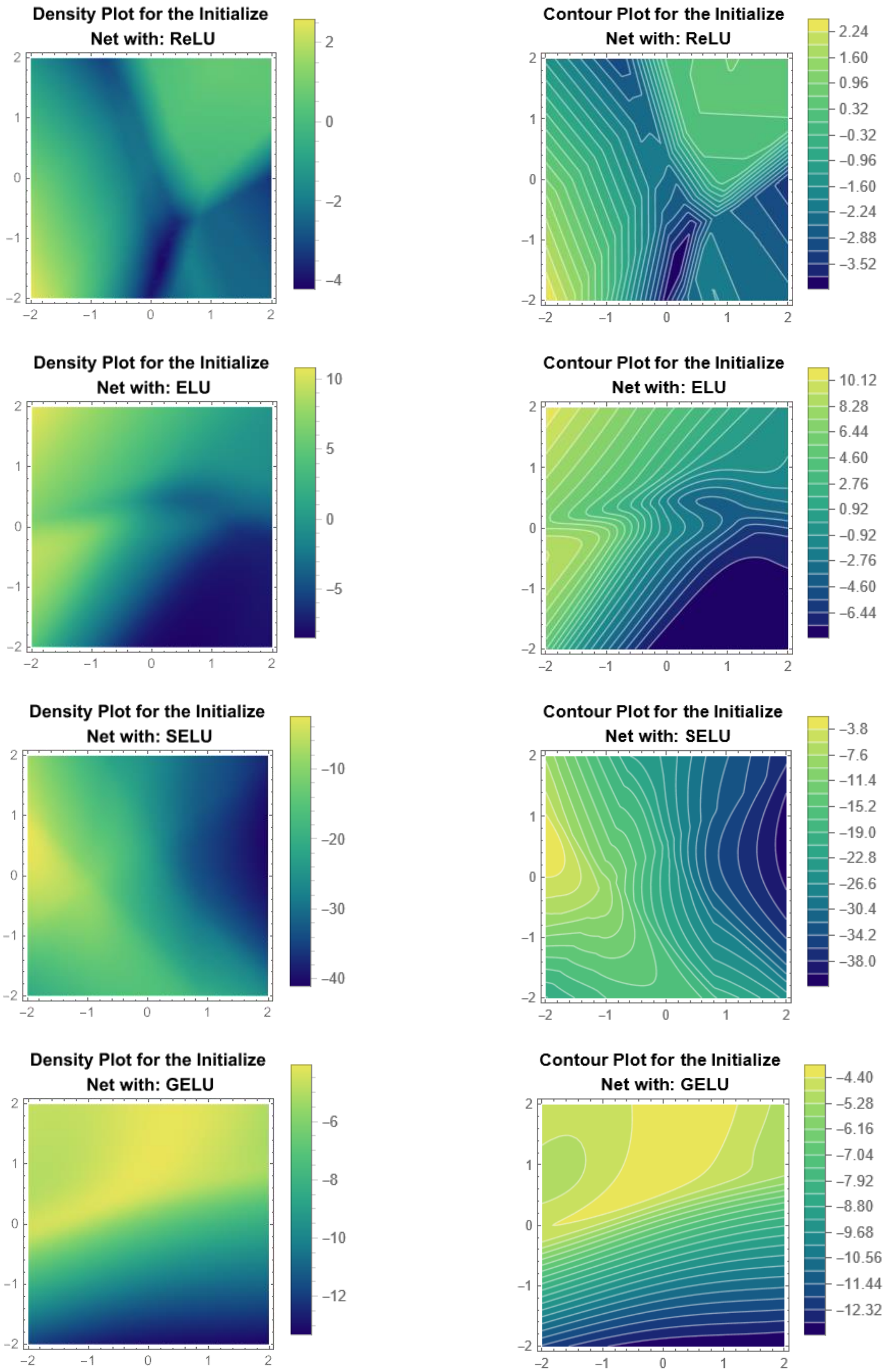
Density Plot for the Initialize Net with: ReLU
2
0
−2
−4
Contour Plot for the Initialize Net with: ReLU
2.24
1.60
0.96
0.32
−0.32
−0.96
−1.60
−2.24
−2.88
−3.52
Density Plot for the Initialize Net with: ELU
10
5
0
−5
Contour Plot for the Initialize Net with: ELU
10.12
8.28
6.44
4.60
2.76
0.92
−0.92
−2.76
−4.60
−6.44
Density Plot for the Initialize Net with: SELU
−10
−20
−30
−40
Contour Plot for the Initialize Net with: SELU
−3.8
−7.6
−11.4
−15.2
−19.0
−22.8
−26.6
−30.4
−34.2
−38.0
Density Plot for the Initialize Net with: GELU
−6
−8
−10
−12
Contour Plot for the Initialize Net with: GELU
−4.40
−5.28
−6.16
−7.04
−7.92
−8.80
−9.68
−10.56
−11.44
−12.32
−2 −1 0 1 2

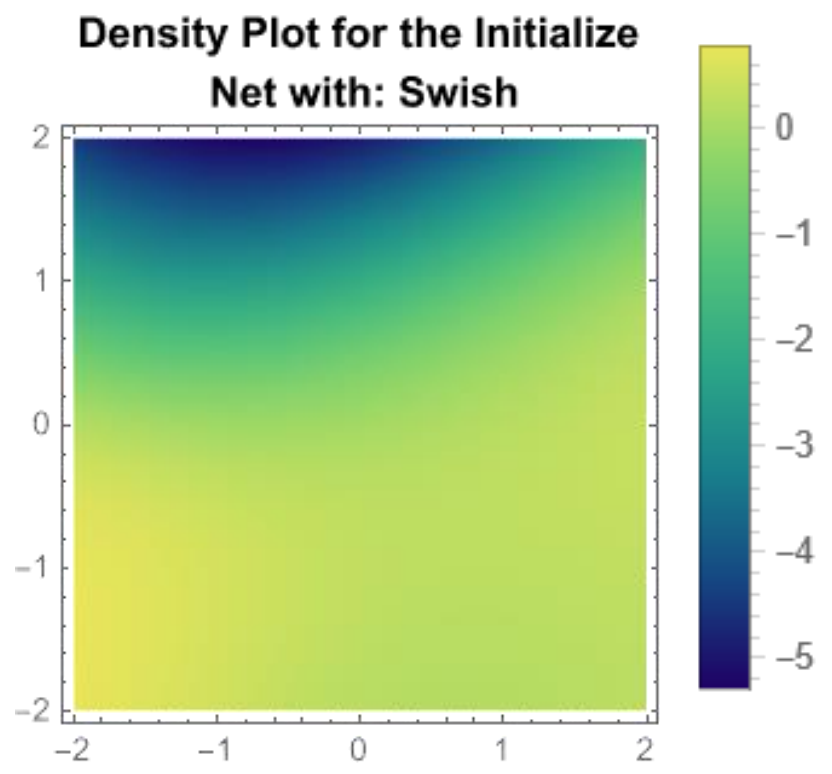
Density Plot for the Initialize
Net with: Swish

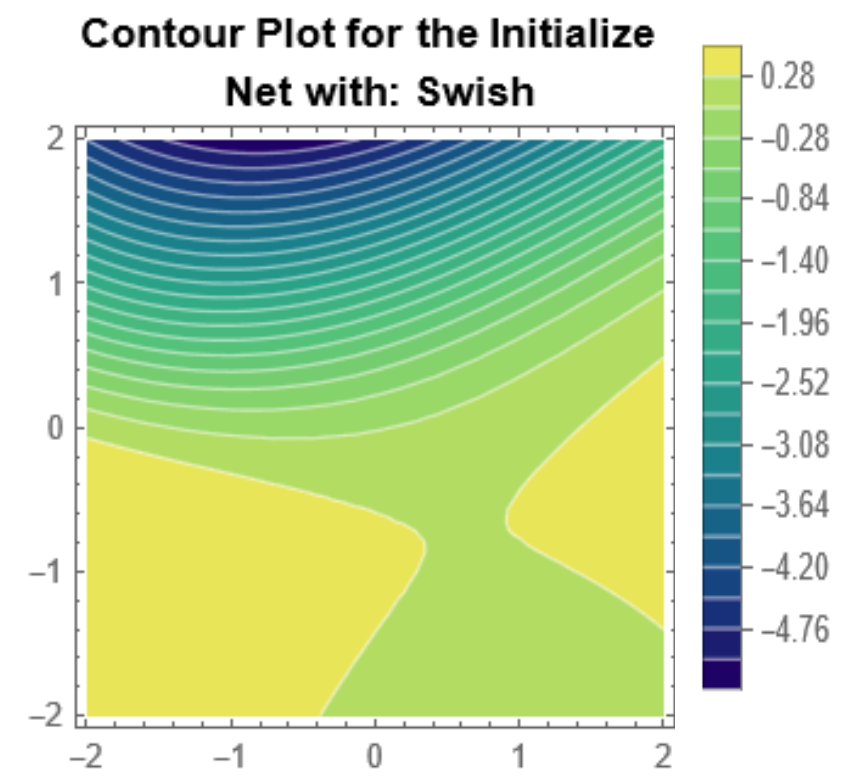
Contour Plot for the Initialize
Net with: Swish

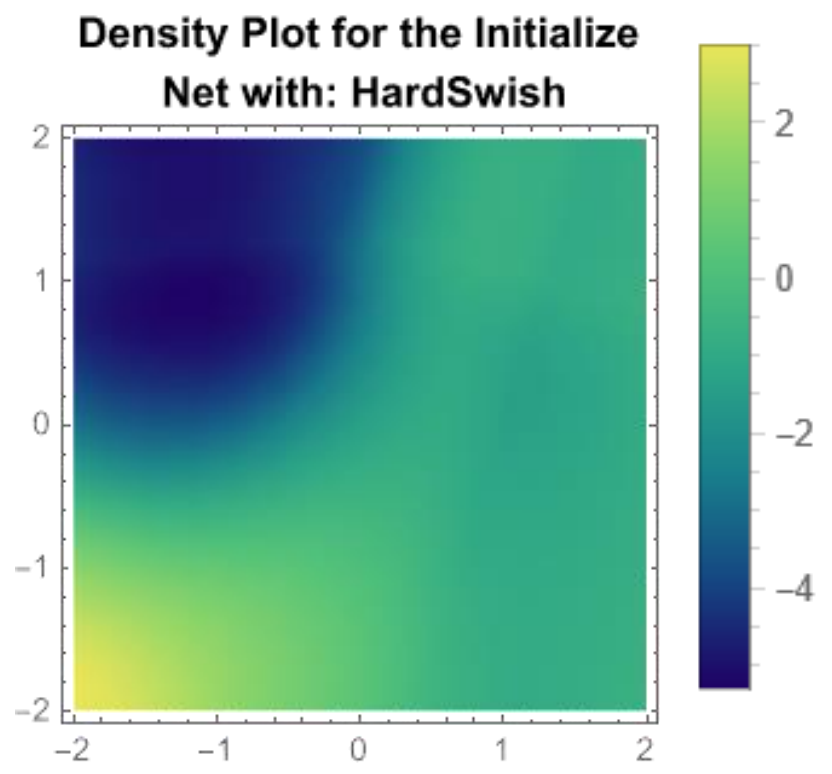
Density Plot for the Initialize
Net with: HardSwish

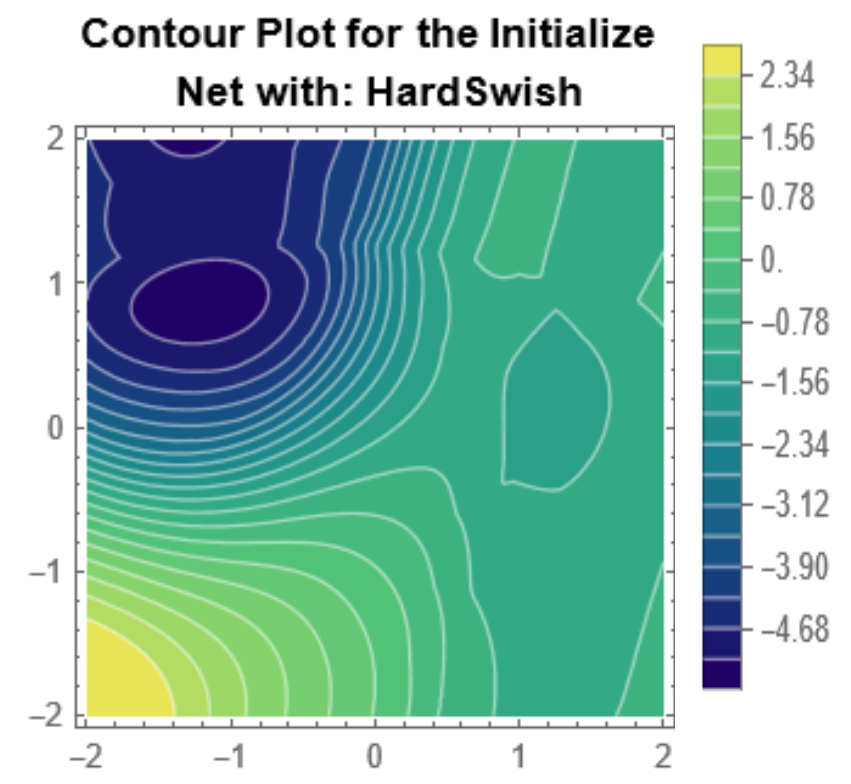
Contour Plot for the Initialize
Net with: HardSwish

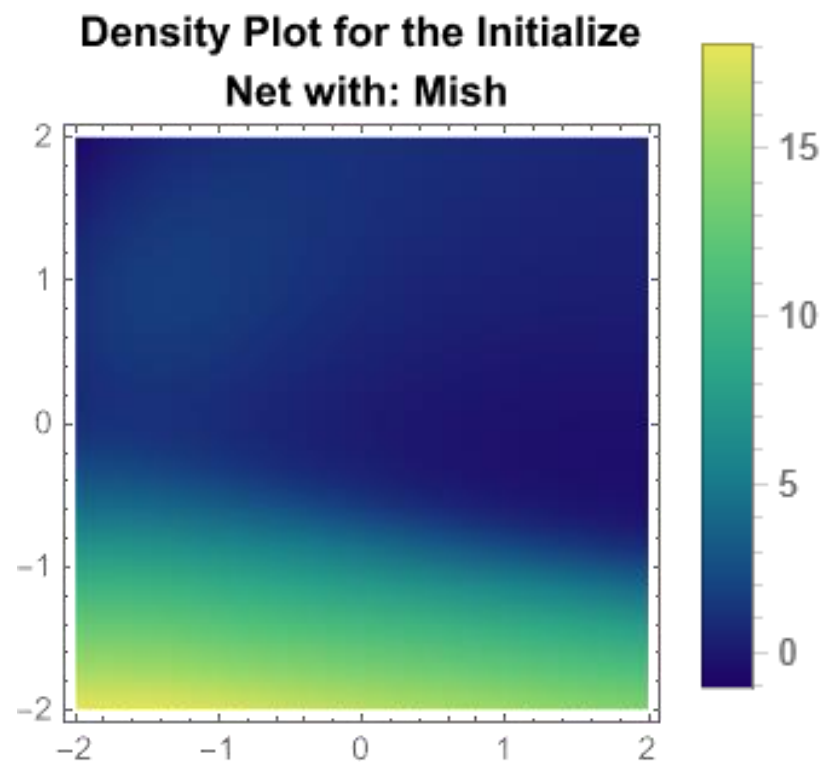
Density Plot for the Initialize
Net with: Mish

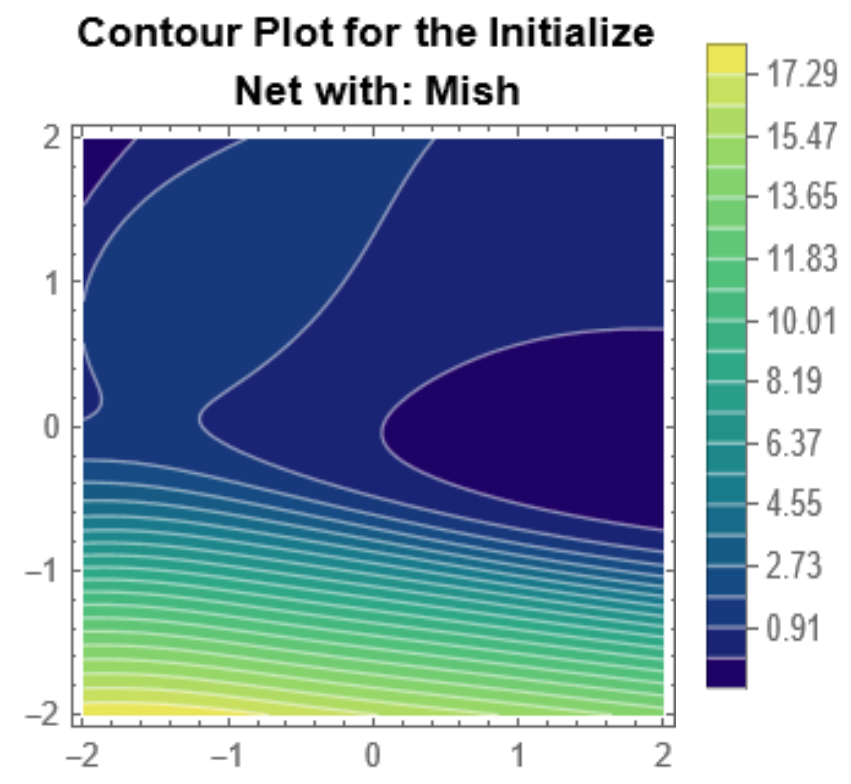
Contour Plot for the Initialize
Net with: Mish

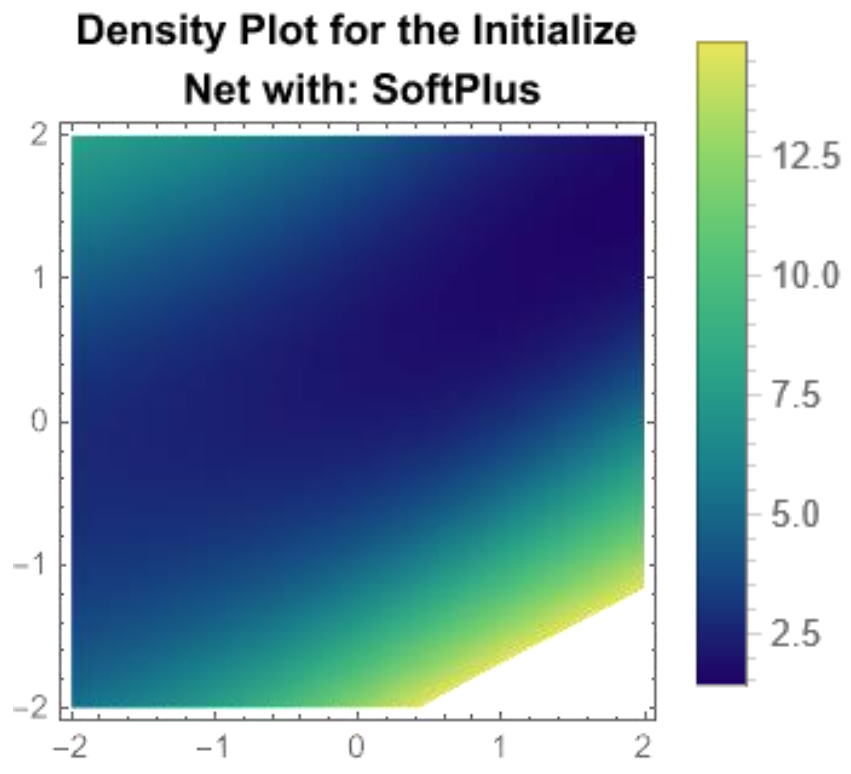
Density Plot for the Initialize
Net with: SoftPlus

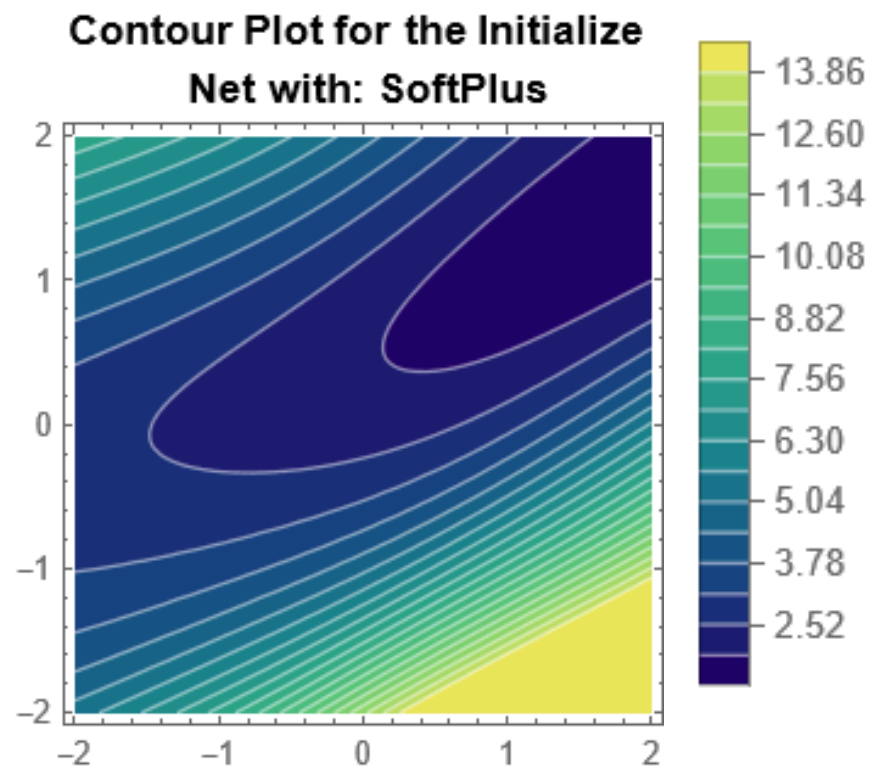
Contour Plot for the Initialize
Net with: SoftPlus

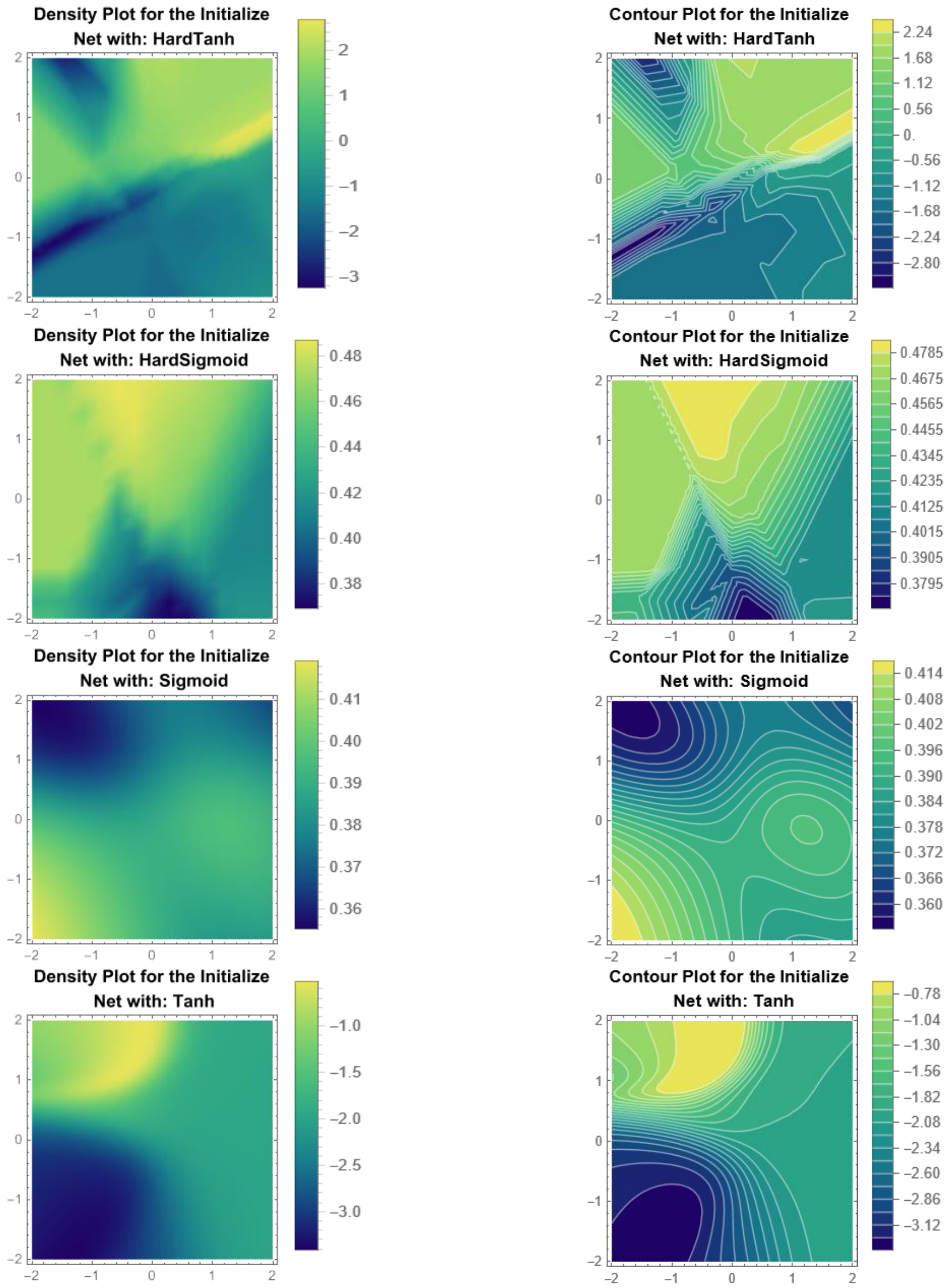

**Figure 8.65.** **(Smooth Output Landscape)** Visualization of Output Landscapes for Various AFs. This figure showcases the output landscapes of a 3-layer randomly initialized NN, visualized using density and contour plots, for different AFs. The AFs tested include ReLU, ELU, SELU, GELU, Swish, HardSwish, Mish, SoftPlus, HardTanh, HardSigmoid, Sigmoid, and Tanh. Each subplot represents the network's scalar output evaluated over a 2-dimensional grid ranging from $-2$ to $2$ in both $x$ and $y$ dimensions. Left panels: (Density Plots) These plots display the network's output values using a gradient color scheme (Blue-Green-Yellow), highlighting regions of varying output intensity. Right panels: (Contour Plots) These plots present the network's output as contour lines, providing a clear view of the smoothness and sharpness in the output landscape. The visualizations illustrate the dramatic effect that different AFs have on the smoothness and structure of the output landscape. The smoother landscapes observed with functions like Swish contrast with the sharp and chaotic regions produced by ReLU, underscoring the importance of AF selection in NN optimization and generalization.

ReLU, HardTanh, HardSigmoid, and HardSwish each uniquely affect NN performance. ReLU enhances sparsity and efficiency by producing zero output for negative inputs, but its abrupt transition at zero causes sharp regions in the output landscape (see Figure 8.65), posing challenges for gradient-based optimization. HardTanh, which approximates the Tanh function, provides some sparsity with constant gradients within the range of $-1$ to 1, but introduces sharp transitions at $-1$ and 1, potentially affecting optimization. HardSigmoid, a piecewise linear approximation of the Sigmoid function, offers smoother transitions with a constant gradient within the range of $-2.5$ to 2.5, yet still has sharp regions at the boundaries, impacting optimization. HardSwish, a piecewise linear approximation of Swish, maintains smoothness within $[-3, 3]$, promoting better gradient flow and mitigating the vanishing gradient problem while being computationally efficient. These four functions can create non-smooth regions in the output landscape (and loss landscape), potentially leading to issues in gradient-based optimization. Conversely, networks using ELU, SELU, GELU, Swish, Mish, SoftPlus, Sigmoid, and Tanh exhibit considerably smoother output landscapes. Smoother output landscapes lead to smoother loss landscapes, which are easier to optimize and result in improved training and test accuracy.

# CHAPTER 9

# COMPLEX VALUED NEURAL NETWORKS

This chapter provides a thorough introduction to Complex-Valued Neural Networks (CVNNs), from the foundational concepts of complex numbers and functions to the advanced techniques of complex BP and the design of complex AFs. By extending NNs into the complex domain, we unlock new possibilities for innovation and performance in various applications.

We begin our journey with a study of complex numbers and functions. This study lays the foundation by introducing the basic concepts of complex numbers, including their arithmetic and geometric representations. We explore the complex plane, also known as the Argand plane, where complex numbers are visualized as points or vectors. Next, we delve into complex functions, examining how these functions map complex numbers to other complex numbers and the profound implications of these mappings. Visualization techniques play a crucial role in understanding complex functions, so we will use tools like domain coloring to bring these abstract concepts to life.

Building on the foundation of complex functions, we explore complex calculus, starting with the definition and properties of analytic functions. We will introduce the Cauchy-Riemann equations, which provide necessary and sufficient conditions for a function to be analytic. These equations link the partial derivatives of the real and imaginary parts of a complex function, offering a bridge between real and complex analysis. Furthermore, we explore Wirtinger derivatives, a convenient formalism for handling derivatives in complex calculus.

One of the most exciting applications of complex numbers in NNs is the development of complex BP algorithms [63]. We examine three distinct cases for deriving these algorithms. Case 1: The derivation assumes that the complex derivative $\sigma'(z) = d\sigma(z)/dz$ of the AF $\sigma(z)$ exists. This scenario explores the straightforward application of complex differentiation in NN training. Case 2: The derivation is based on the assumption that the partial derivatives $\frac{\partial u}{\partial x}, \frac{\partial u}{\partial y}, \frac{\partial v}{\partial x}$, and $\frac{\partial v}{\partial y}$ of the AF $\sigma(z) = u(x, y) + iv(x, y)$ exists. This assumption is not sufficient that $\sigma'(z)$ exists. Case 3: This scenario incorporates the Cauchy-Riemann equations to derive a complex BP algorithm.

Moreover, we explore two main types of CVNNs:

- Fully-CVNNs: These networks operate entirely in the complex domain, using complex-valued weights, activations, and operations. We discussed their architecture, training algorithms, and potential benefits.
- Split-CVNNs: These networks split the real and imaginary parts of complex numbers, processing them separately while maintaining the connections between them.

Finally, we examine the properties of complex AFs. Liouville's theorem in complex analysis states that any bounded entire function must be constant. We discuss its implications for the design of AFs in CVNNs. Moreover, we compare split AFs, which handle real and imaginary parts separately, with fully complex AFs that operate on complex numbers as a whole. Understanding these properties helps in selecting and designing effective AFs for specific applications. Additionally, we provided an extensive catalog of Complex-Valued Activation Functions (CVAFs), detailing their definitions, properties, and applications. These functions include: Split-Step Function, Split-Sigmoid, Split-Parametric Sigmoid, Split-Tanh, Split-Sigmoid Tanh, Split-Hard Tanh, Split-CReLU, Split-QAM, Amplitude-Phase-Type Function, Amplitude-Phase Sigmoidal Function, Complex Cardioid, modReLU, Fully Complex Tanh, Fully Complex Logistic-Sigmoidal, Fully Complex Elementary Transcendental Function (ETF), and zReLU. Each AF is discussed in terms of its mathematical formulation, and graphical representation. Moreover, this chapter not only offers a comprehensive overview of the current state of CVNNs but also contributes to ongoing research and development by introducing a new set of CVAFs (fully complex, split and complex amplitude-phase AFs).

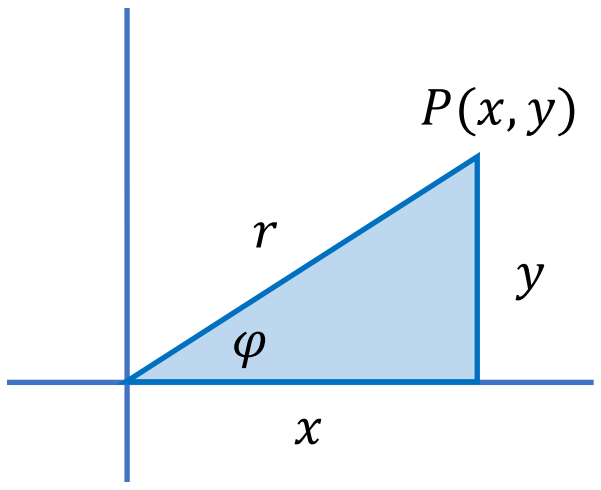

**Figure 9.1.** Complex Numbers as Position Vectors in the Complex Plane. The complex number $z = x + iy$ is shown as a vector originating from the origin $(0,0)$ and terminating at the point $P(x, y)$. Here, $x$ represents the real part and $y$ represents the imaginary part of the complex number. The length of the vector, denoted as $r$, is the magnitude of the complex number and is calculated as $r = \sqrt{x^2 + y^2}$. The angle $\varphi$ represents the argument of the complex number, which is the angle between the positive real axis and the vector, measured in the counterclockwise direction. This polar representation helps in visualizing and performing operations on complex numbers geometrically.

## 9.1 Complex Numbers, Complex Functions, and Their Visualizations

A complex number [266-268] is of the form $a + ib$, where $a$ and $b$ are real numbers, and $i$ is the imaginary unit satisfying $i^2 = -1$, or $i^2 + 1 = 0$. For example, $2 + 3i$ is a complex number. The relation $i^2 + 1 = 0$ induces several equalities for any integer $k$:

$$i^{4k} = 1,\ i^{4k+1} = i,\ i^{4k+2} = -1, \text{ and } i^{4k+3} = -i. \tag{9.1}$$

The real part of a complex number $z = a + ib$ is denoted by $\text{Re}(z)$, or $\Re(z)$, and the imaginary part is denoted by $\text{Im}(z)$ or $\Im(z)$. For example, $\text{Re}(3 + 2i) = 3$ and $\text{Im}(3 + 2i) = 2$.

**Remarks:**

- A complex number $z$ can be represented as an ordered pair $(\text{Re}(z), \text{Im}\,(z))$ of real numbers. This ordered pair can be interpreted as the coordinates of a point in a two-dimensional space. The complex plane is a two-dimensional plane where the horizontal axis represents the real part, $\text{Re}(z)$, and the vertical axis represents the imaginary part, $\text{Im}\,(z)$. The complex plane is also referred to as the Argand diagram. To each complex number, there corresponds one and only one point in the plane, and conversely to each point in the plane there corresponds one and only one complex number.
- A complex number $z = a + ib$ can be viewed as a position vector in the complex plane, starting from the origin $(0,0)$ and ending at the point $(a, b)$. The vector represents both the magnitude (distance from the origin) and the direction (angle with the positive real axis), see Figure 9.1.

When performing mathematical operations on complex numbers, you treat the real and imaginary parts separately. Here are the basic operations:

- Two complex numbers $a = x + iy$ and $b = u + iv$ are most easily added by separately adding their real and imaginary parts. That is to say:

$$a + b = (x + iy) + (u + iv) = (x + u) + i(y + v). \tag{9.2}$$

- This geometrically corresponds to moving from the origin to the point $a$, and then from there, moving to the point $b$. The resulting complex number $a + b$ is the diagonal of the parallelogram formed by the vectors representing $a$ and $b$. This method of adding complex numbers aligns with the geometric interpretation of complex addition, where the real parts represent movement along the horizontal axis, and the imaginary parts represent movement along the vertical axis in the complex plane, see Figure 9.2.

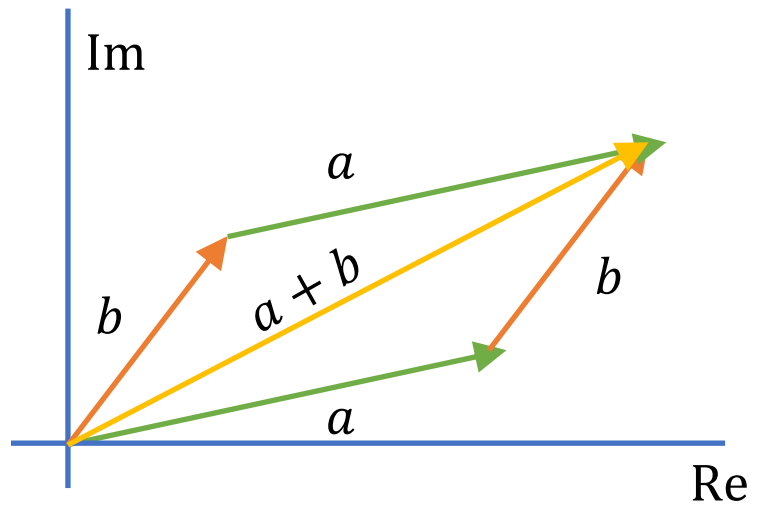

**Figure 9.2.** Geometric Addition of Complex Numbers Using a Parallelogram. The complex numbers $a$ and $b$ are represented as vectors originating from the origin, with $a$ terminating at $(x, y)$ and $b$ terminating at $(u, v)$. To add these two complex numbers geometrically, a parallelogram is constructed with $a$ and $b$ as adjacent sides. The vector representing $a$ is drawn from the origin, and the vector representing $b$ is drawn starting from the endpoint of $a$. Similarly, a vector equal to $a$ is drawn starting from the endpoint of $b$. The opposite sides of the parallelogram are parallel and equal in length. The resultant vector, which represents the sum $a + b$, is the diagonal of the parallelogram originating from the origin. It terminates at the point where the opposite corners of the parallelogram meet. This point has coordinates $((x + u), (y + v))$, which corresponds to the sum of the real and imaginary parts of the two complex numbers.

- Similarly, subtraction can be performed as

$$a - b = (x + iy) - (u + iv) = (x - u) + i(y - v). \tag{9.3}$$

- Multiplication of a complex number $a = x + iy$ and a real number $r$ can be done similarly by multiplying separately $r$ and the real and imaginary parts of $a$:

$$r\, a = r(x + iy) = rx + iry. \tag{9.4}$$

- The multiplication of two complex numbers can be performed by applying the distributive property, the commutative properties of addition and multiplication, and the defining property $i^2 = -1$. Consequently, we have:

$$(x + iy)(u + iv) = (xu - yv) + i(xv + yu). \tag{9.5}$$

  In particular,

$$(x + iy)^2 = x^2 - y^2 + 2ixy. \tag{9.6}$$

An alternative to Cartesian coordinates in the complex plane is the polar coordinate system. Let $P$ be a point in the complex plane corresponding to the complex number $(x, y)$ or $x + iy$. Then we see from Figure 9.1 that

$$x = r\cos\varphi \quad \text{and} \quad y = r\sin\varphi. \tag{9.7}$$

- In polar coordinates, a complex number $z = x + iy$ is represented by its distance from the origin ($r = |z|$) and the angle $\varphi$ it makes with the positive real axis in a counterclockwise direction. The polar form of a complex number $z$ is given by

$$z = x + iy = r(\cos\varphi + i\sin\varphi), \tag{9.8}$$

  where $r$ is the absolute value of $z$, and $\varphi$ is the argument (angle) of $z$.
- For any complex number $z \neq 0$, there corresponds only one value of $\varphi$ in $0 \leq \varphi < 2\pi$. However, any other interval of length $2\pi$, for example $-\pi < \varphi \leq \pi$, can be used. Any particular choice, decided upon in advance, is called the principal range, and the value of $\varphi$ is called its principal value.
- The absolute value of a complex number $z = x + iy$ is given by

$$r = |z| = \sqrt{x^2 + y^2}. \tag{9.9}$$

  The absolute value of a complex number, as given by Pythagoras' theorem, represents the distance from the origin to the point representing the complex number in the complex plane.

- The argument of a complex number $z = x + iy$ is the angle $\varphi$ that the line segment from the origin to the point $(x, y)$ makes with the positive real axis. To calculate the argument, you can use the arctangent function. The argument (written as $\arg z$) can be found using the formula:

$$\arg z = \varphi = \tan^{-1}\left(\frac{y}{x}\right), \tag{9.10}$$

with

$$\varphi = \arg(x+iy) = \begin{cases} \tan^{-1}\left(\frac{y}{x}\right), & \text{x} > 0, \text{right half plane} \\ \tan^{-1}\left(\frac{y}{x}\right) + \pi, & x < 0, y \geq 0, \text{upper left half plane} \\ \tan^{-1}\left(\frac{y}{x}\right) - \pi, & x < 0, y < 0, \text{lower left half plane} \\ \frac{\pi}{2}, & x = 0\,, y > 0, +i \text{ axis} \\ -\frac{\pi}{2}, & x = 0\,, y < 0, -i \text{ axis} \\ \text{undefined} & x = 0, y = 0, \text{origin} \end{cases}. \tag{9.11}$$

- De Moivre's Theorem:
  Let $z_1 = x_1 + iy_1 = r_1(\cos\varphi_1 + i\sin\varphi_1)$ and $z_2 = x_2 + iy_2 = r_2(\cos\varphi_2 + i\sin\varphi_2)$, because of the trigonometric identities

$$\cos a \cos b - \sin a \sin b = \cos(a+b), \tag{9.12.1}$$
$$\cos a \sin b + \sin a \cos b = \sin(a+b), \tag{9.12.2}$$

we may derive

$$z_1 z_2 = r_1 r_2\{\cos(\varphi_1 + \varphi_2) + i\sin(\varphi_1 + \varphi_2)\}, \tag{9.13.1}$$
$$\frac{z_1}{z_2} = \frac{r_1}{r_2}\{\cos(\varphi_1 - \varphi_2) + i\sin(\varphi_1 - \varphi_2)\}. \tag{9.13.2}$$

- By assuming that the infinite series expansion $e^x = 1 + x + \frac{x^2}{2!} + \frac{x^3}{3!} + \frac{x^4}{4!} + \cdots$ of elementary calculus holds when $x = i\varphi$, we can arrive at the result

$$\begin{aligned} e^{i\varphi} &= 1 + (i\varphi) + \frac{(i\varphi)^2}{2!} + \frac{(i\varphi)^3}{3!} + \frac{(i\varphi)^4}{4!} + \frac{(i\varphi)^5}{5!} + \frac{(i\varphi)^6}{6!} + \frac{(i\varphi)^7}{7!} \ldots \\ &= 1 + i\varphi - \frac{\varphi^2}{2!} - i\frac{\varphi^3}{3!} + \frac{\varphi^4}{4!} + i\frac{\varphi^5}{5!} - \frac{\varphi^6}{6!} - i\frac{\varphi^7}{7!} \ldots \\ &= \left(1 - \frac{\varphi^2}{2!} + \frac{\varphi^4}{4!} - \frac{\varphi^6}{6!} + \cdots\right) + i\left(\varphi - \frac{\varphi^3}{3!} + \frac{\varphi^5}{5!} - \frac{\varphi^7}{7!} \ldots\right), \\ &= \cos\varphi + i\sin\varphi, \end{aligned} \tag{9.14}$$

which is called Euler's formula, for any real number $\varphi$. The functional equation implies thus that, if $x$ and $y$ are real, one has

$$\begin{aligned} e^z &= e^{x+iy} \\ &= e^x e^{iy} \\ &= e^x(\cos y + i\sin y) \\ &= e^x \cos y + ie^x \sin y, \end{aligned} \tag{9.15}$$

which is the decomposition of the exponential function into its real and imaginary parts.

- For each $z \neq 0$, there are infinitely many possible values for $\arg z$, which all differ from each other by an integer multiple of $2\pi$.

$$\begin{aligned} e^{i(\varphi+2k\pi)} &= \cos(\varphi + 2k\pi) + i\sin(\varphi + 2k\pi) \\ &= \cos(\varphi) + i\sin(\varphi) \\ &= e^{i\varphi}, \qquad k = 0, +1, +2, \ldots. \end{aligned} \tag{9.16}$$

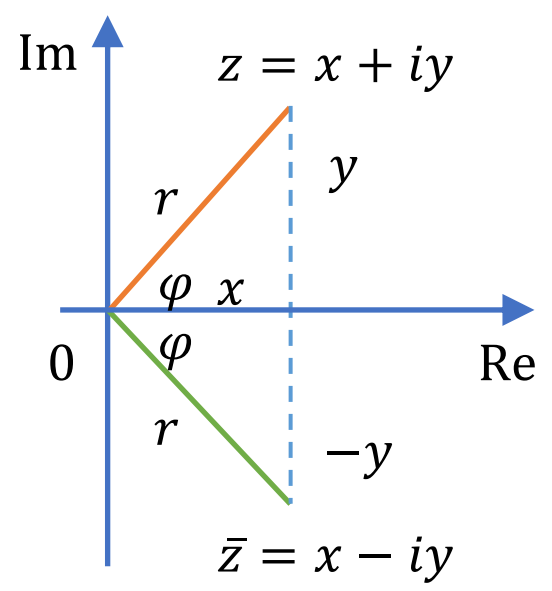

**Figure 9.3.** Geometric Representation of $z$ and Its Conjugate $\bar{z}$ in the Complex Plane. The complex number $z = x + iy$ is depicted as a vector originating from the origin (0,0) and terminating at the point $(x, y)$, where $x$ is the real part and $y$ is the imaginary part of $z$. The magnitude of $z$ is represented by $r$, which is calculated as $r = \sqrt{x^2 + y^2}$. The angle $\varphi$ is the argument of $z$, measured counterclockwise from the positive real axis. The conjugate of $z$, denoted as $\bar{z} = x - iy$, is also represented as a vector originating from the origin and terminating at the point $(x, -y)$. This reflects the fact that the conjugate has the same real part $x$ but the opposite sign for the imaginary part, $-y$. The magnitude $r$ and the angle $\varphi$ remain the same for both $z$ and $\bar{z}$, as the magnitudes are equal and the angles are reflections over the real axis.

The conjugate of a complex number $z = x + iy$ is denoted by $\bar{z}$ or $z^*$, and it is obtained by changing the sign of the imaginary part. Mathematically, the conjugate is given by: $x - iy$. Similarly, if $z = x - iy$, then $\bar{z} = x + iy$, see Figure 9.3. Properties of the complex conjugate include:

$$\overline{(\bar{z})} = z, \tag{9.17.1}$$
$$\overline{z_1 \pm z_2} = \bar{z}_1 \pm \bar{z}_2, \tag{9.17.2}$$
$$\overline{z_1 \cdot z_2} = \bar{z}_1 \cdot \bar{z}_2, \tag{9.17.3}$$
$$\overline{z_1 / z_2} = \bar{z}_1 / \bar{z}_2. \tag{9.17.4}$$

The conjugate is useful in various mathematical operations, including finding the modulus (absolute value) of a complex number, dividing complex numbers, and simplifying expressions involving complex conjugates. Here are the basic operations:

- The reflection leaves both the real part and the magnitude of $z$ unchanged, that is

$$\mathrm{Re}(\bar{z}) = \mathrm{Re}\,(z) \qquad \text{and} \qquad |\bar{z}| = |z|. \tag{9.18}$$

- The imaginary part and the argument of a complex number $z$ change their sign under conjugation

$$\mathrm{Im}(\bar{z}) = -\mathrm{Im}(z) \qquad \text{and} \qquad \arg(\bar{z}) = -\arg(z). \tag{9.19}$$

- The product of a complex number $z = x + iy$ and its conjugate is known as the absolute square. It is always a non-negative real number and equals the square of the magnitude of each:

$$z \cdot \bar{z} = x^2 + y^2 = |z|^2 = |\bar{z}|^2. \tag{9.20}$$

- The real and imaginary parts of a complex number $z$ can be expressed using the conjugate as:

$$\mathrm{Re}(z) = \frac{z + \bar{z}}{2} \qquad \text{and} \qquad \mathrm{Im}(z) = \frac{\bar{z} - z}{2i}. \tag{9.21}$$

- Moreover, a complex number is real if and only if it equals its own conjugate.
- Using the conjugate, the reciprocal of a nonzero complex number $z = x + iy$ can be broken into real and imaginary components

$$\frac{1}{z} = \frac{\bar{z}}{z\bar{z}} = \frac{\bar{z}}{|z|^2} = \frac{x - iy}{x^2 + y^2} = \frac{x}{x^2 + y^2} - i\frac{y}{x^2 + y^2}. \tag{9.22}$$

This can be used to express a division of an arbitrary complex number $w = u + iv$ by a non-zero complex number $z = x + iy$ as

$$\frac{w}{z} = \frac{w\bar{z}}{|z|^2} = \frac{(u+iv)(x-iy)}{x^2+y^2} = \frac{ux+vy}{x^2+y^2} + i\frac{vx-uy}{x^2+y^2}. \tag{9.23}$$

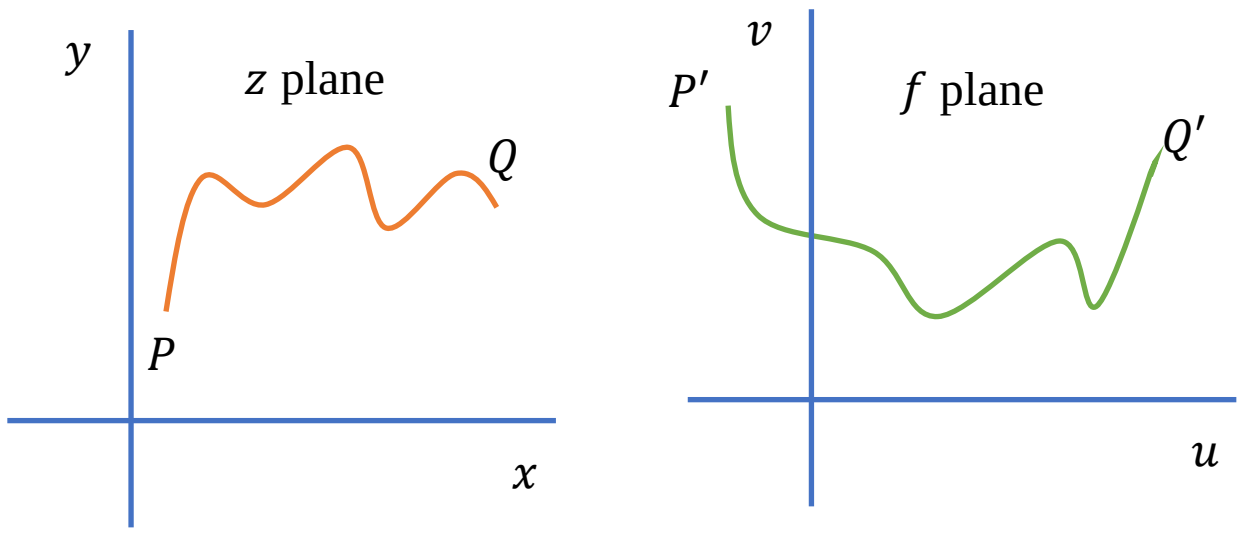

**Figure 9.4.** Mapping and Transformation from $z$ Plane to $f$ Plane. Given a point $P$ in the $z$ plane with coordinates $(x, y)$, there is a corresponding point $P'$ in the $f$ plane with coordinates $(u, v)$. The set of equations defining this relationship is given by $f(z)$, which is called a transformation. Point $P$ is mapped or transformed into point $P'$ by this transformation, and $P'$ is referred to as the image of $P$. This figure visually captures the relationship between the original points and their images under a given transformation, illustrating how the transformation affects the positions of points and curves in the complex plane.

A complex function is a function from complex numbers to complex numbers. In other words, it is a function that has a subset of the complex numbers as a domain and the complex numbers as a codomain. For any complex function, the values $z$ from the domain and their images $f(z)$ in the range may be separated into real and imaginary parts:

$$z = x + iy \quad \text{and} \quad f(z) = f(x+iy) = u(x,y) + iv(x,y), \tag{9.24}$$

where $x$, $y$, $u(x,y)$, $v(x,y)$ are all real-valued. Hence, a complex function $f: \mathbb{C} \to \mathbb{C}$ may be decomposed into

$$u: \mathbb{R}^2 \to \mathbb{R} \quad \text{and} \quad v: \mathbb{R}^2 \to \mathbb{R}, \tag{9.25}$$

i.e., into two real-valued functions $(u, v)$ of two real variables $(x, y)$.

If only one value of $f$ corresponds to each value of $z$, we say that $f$ is a single-valued function of $z$ or that $f(z)$ is single-valued. If more than one value of $f$ corresponds to each value of $z$, we say that $f$ is a multiple-valued or many-valued function of $z$. Whenever we speak of function, we shall, unless otherwise stated, assume a single-valued function.

For example, consider the complex function $f(z) = z^2$. If we let $z = x + iy$, then $f(z) = (x+iy)^2$. Expanding this expression yields both real and imaginary parts:

$$f(z) = x^2 - y^2 + 2ixy. \tag{9.26}$$

In this case,

$$u(x,y) = x^2 - y^2, \tag{9.27}$$

and

$$v(x,y) = 2xy, \tag{9.28}$$

making $f(z) = u + iv$. Thus, given a point $(x, y)$ in the $z$ plane, such as $P$ in Figure 9.4, there corresponds to a point $(u, v)$ in the $f$ plane, say $P'$ in Figure 9.4. The set of equations (9.24) [or the equivalent, $f(z)$] is called a transformation. We say that point $P$ is mapped or transformed into point $P'$ by means of the transformation and call $P'$ the image of $P$. For example, the image of a point $(1,2)$ in the $z$ plane is the point $(-3,4)$ in the $f(z) = z^2$ plane. In general, under a transformation, a set of points such as those on curve $PQ$ of Figure 9.4 and Figure 9.5 is mapped into a corresponding set of points, called the image, such as those on curve $P'Q'$. The particular characteristics of the image depend of course on the type of function $f(z)$, which is sometimes called a mapping function.

Now, in the above example, we will define new variables $c_1$ and $c_2$ such that $u = c_1$ and $v = c_2$. Therefore, the equations become:

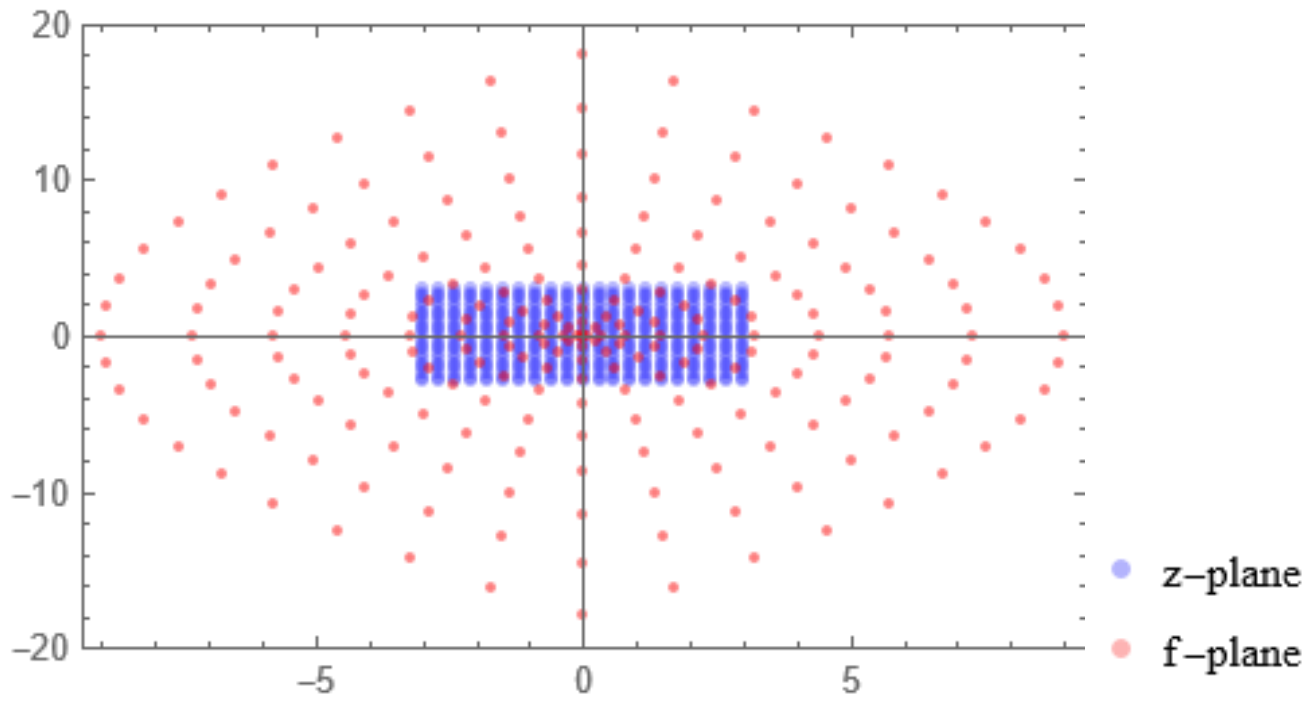

**Figure 9.5.** Mapping of Points in the Complex Plane Using $f(z) = z^2$. The plot shows two sets of points: the original points in the $z$-plane and their corresponding transformed points in the $f$-plane. The points in the $z$-plane are displayed in blue with a larger point size and lower opacity, while the points in the $f$-plane are displayed in red with a smaller point size and lower opacity.

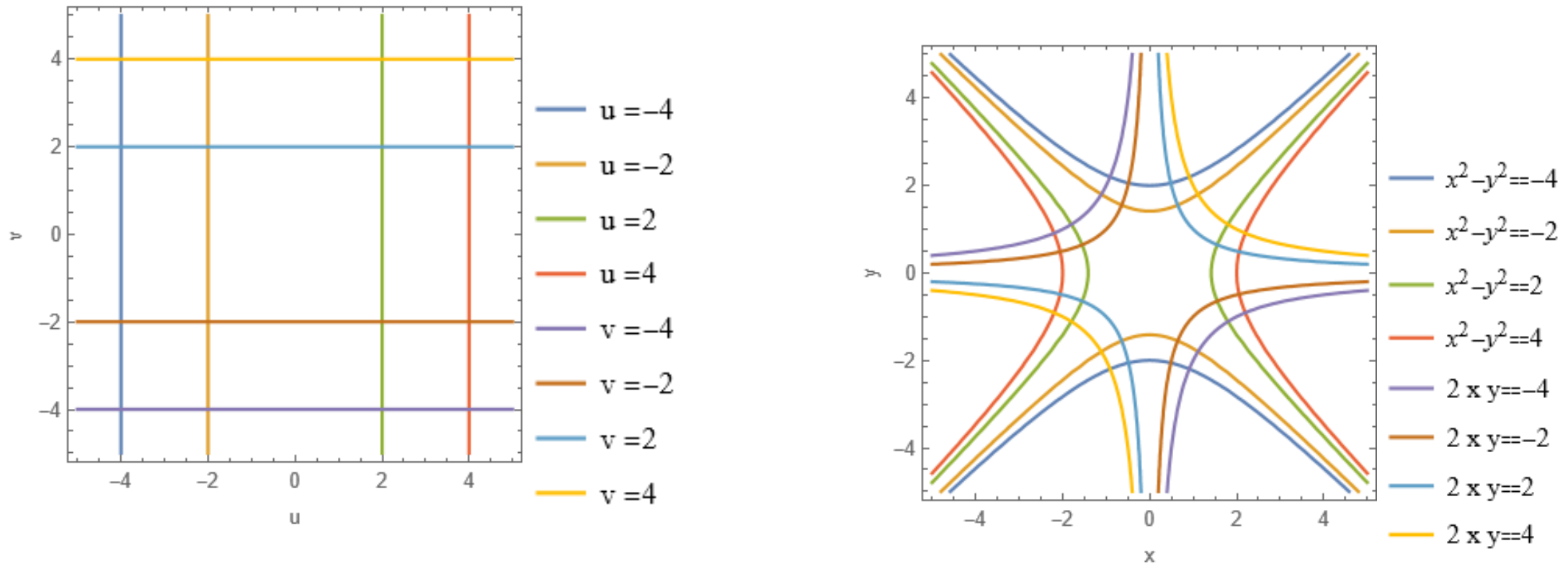

**Figure 9.6.** Contour plots illustrate the real and imaginary parts of a squared complex function in the $uv$-plane, and hyperbolas defined by quadratic equations in the $xy$-plane. Left panel: The figure displays contour lines representing the real part $u$ and imaginary part $v$ of the function $f(x, y) = (x + iy)^2$. The contour lines are plotted for the values $u = -4, -2, 2, 4$ and $v = -4, -2, 2, 4$ within the range of $[-5,5]$ on both axes. Each contour line represents a constant value of $u$ or $v$, highlighting the relationship between the real and imaginary parts of the complex function in the $uv$-plane. Right panel: The figure illustrates hyperbolas defined by the equations $x^2 - y^2 = c_1$ and $2xy = c_2$ in the $xy$-plane. The contour lines represent these hyperbolas for the values $c_1 = -4, -2, 2, 4$ and $c_2 = -4, -2, 2, 4$, plotted within the range of $[-5,5]$ on both axes. Each contour line corresponds to a specific hyperbola, demonstrating the geometric shapes formed by these quadratic equations.

$$u(x, y) = x^2 - y^2 = c_1, \qquad v(x, y) = 2xy = c_2. \tag{9.29}$$

These equations represent curves in the $xy$-plane, see Figure 9.6. The first equation $x^2 - y^2 = c_1$ represents a family of hyperbolas, and the second equation $2xy = c_2$ represents another family of curves. These curves in the $xy$-plane correspond to lines in the $uv$-plane, where $u$ is on the $c_1$ line and $v$ is on the $c_2$ line.

A graph of a real function can be drawn in two dimensions because there are two represented variables, $x$ and $y$. When visualizing complex functions, both a complex input and output are needed. Complex numbers are represented by two variables and therefore two dimensions; this means that representing a complex function (more precisely, a complex-valued function of one complex variable $f: \mathbb{C} \to \mathbb{C}$) requires the visualization of four dimensions, which is possible only in projections. Because of this, other ways of visualizing complex functions have been designed.

It is possible to add variables that keep the four-dimensional process without requiring a visualization of four dimensions. In this case, the two added variables are visual inputs such as color and brightness because they are naturally two variables easily processed and distinguished by the human eye. This assignment is called a "color function".

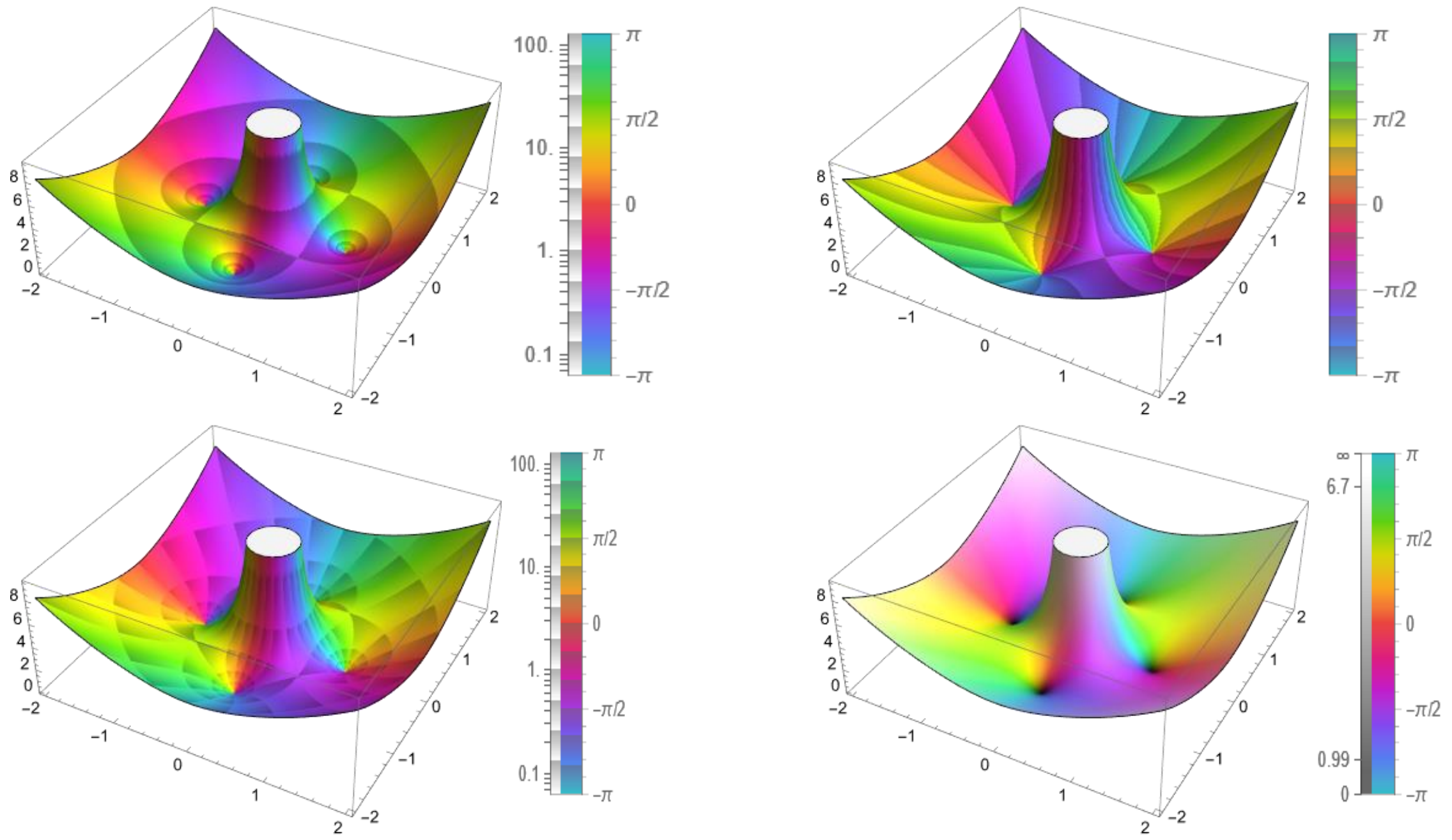

**Figure 9.7.** Complex Plots of $(z^4 - 1)/z^2$ with Different Shading Functions. These figures are 3D plots of the complex function $(z^4 - 1)/z^2$ over the complex rectangle defined by the corners $-2 - 2i$ and $2 + 2i$. Each plot uses a different shading function to highlight various features of the function. The top-left panel uses CyclicLogAbs shading: the height of the surface represents the magnitude of the function, and colors are shaded cyclically based on the logarithm of the magnitude log(Abs[$f$]), creating contour-like appearances for constant magnitudes. Features visible include zeros (points where $f = 0$), poles (points where $f$ tends to infinity), and lines of constant magnitude. The top-right panel uses CyclicArg shading: the height of the surface represents the magnitude of the function, and colors are shaded cyclically based on the argument (phase) of the function Arg[$f$], highlighting regions of constant phase. Features visible include zeros, poles, and phase contours, making it easier to analyze the phase behavior of the function. The bottom-left panel uses CyclicLogAbsArg shading: the height of the surface represents the magnitude of the function, and colors are shaded cyclically based on both the logarithm of the magnitude log(Abs[$f$]) and the argument (phase) of the function Arg[$f$]. This combined shading technique highlights regions of constant magnitude and constant phase simultaneously, providing detailed contour-like appearances. The bottom-right panel uses QuantileAbs shading: the height of the surface represents the magnitude of the function, and colors are shaded from dark to light based on quantiles of Abs[$f$], with darker colors representing smaller magnitudes and lighter colors representing larger magnitudes. This shading technique emphasizes the relative magnitudes of the function values, making it easier to distinguish between low and high-magnitude regions. Each plot includes a legend that explains color mapping.

In domain coloring the output dimensions are represented by color and brightness, respectively. Each point in the complex plane as domain is ornated, typically with color representing the argument of the complex number, and brightness representing the magnitude. Dark spots mark moduli near zero, brighter spots are farther away from the origin, and the gradation may be discontinuous, but is assumed as monotonous. Domain coloring allows one to see how the function behaves across the entire complex plane, revealing patterns and structures that might not be immediately apparent in other types of plots.

Mathematica stands as a powerful tool that empowers users to explore and visualize complex numbers and functions with unprecedented ease and precision. In this survey, we generate the figures using Mathematica. In Mathematica, there are numerous functions available to represent complex functions:

- `ComplexPlot3D[f,{z,zmin,zmax}]` generates a 3D plot of Abs[f] colored by Arg[f] over the complex rectangle with corners zmin and zmax (See Figure 9.7).
- `ComplexPlot[f,{z,zmin,zmax}]` generates a plot of Arg[f] over the complex rectangle with corners zmin and zmax (See Figure 9.8).

Moreover, different color shading techniques can be used, such as CyclicLogAbs shading, CyclicArg shading, CyclicLogAbsArg shading, and QuantileAbs shading (see Figures 9.7 and 9.8).

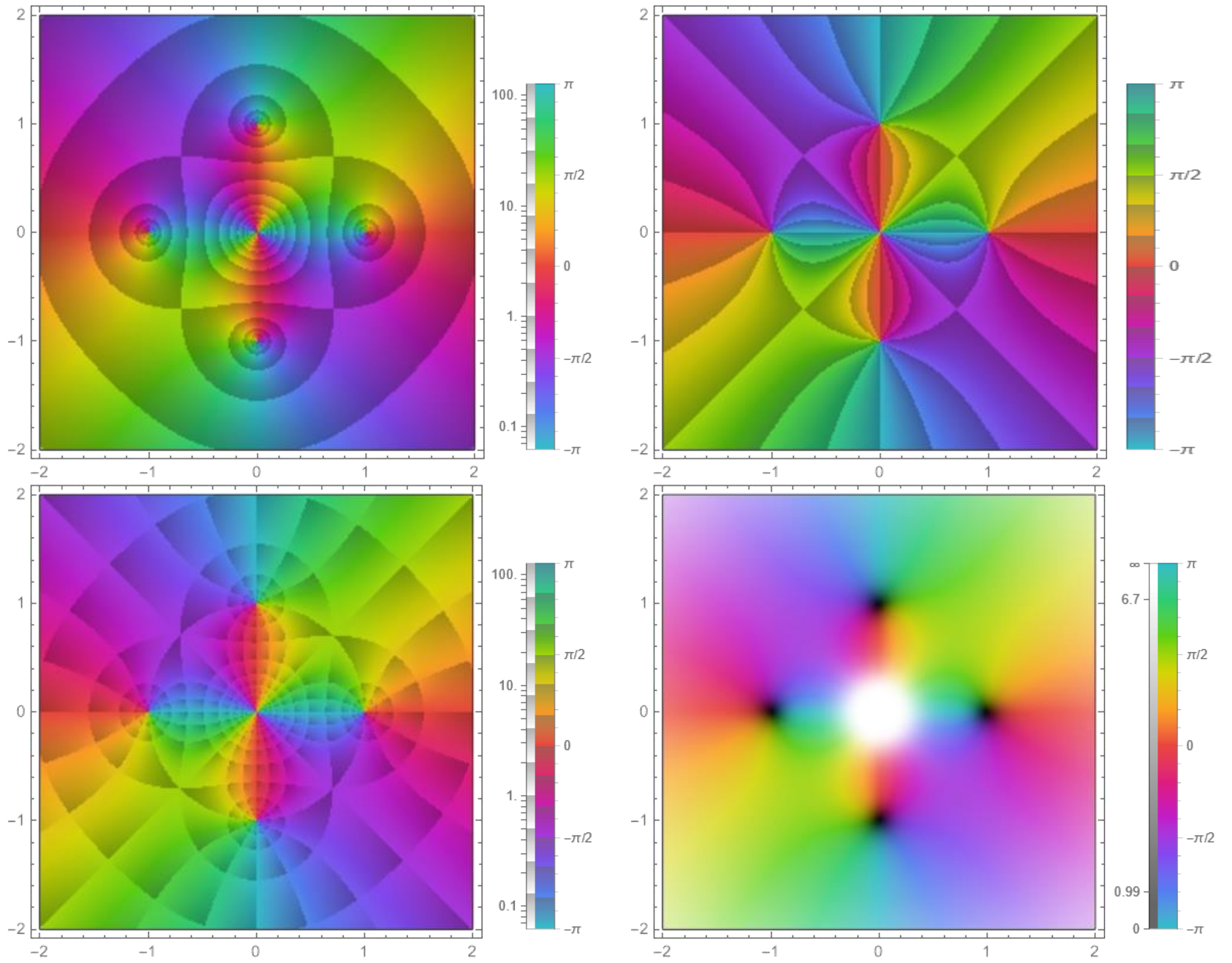

**Figure 9.8.** Complex Plots of $(z^4 - 1)/z^2$ with Different Shading Functions. These figures are complex plots of the function $(z^4 - 1)/z^2$ over the complex rectangle defined by the corners $-2 - 2i$ and $2 + 2i$. Each plot uses a different shading function to visualize the argument (phase) of the function, highlighting various features such as zeros, poles, and essential singularities. The top-left panel uses CyclicLogAbs shading: the plot represents the argument of the function with colors shaded cyclically based on the logarithm of the magnitude log(Abs[$f$]). This shading creates contour-like appearances for constant magnitudes, helping to identify zeros (points where $f = 0$) and poles (points where $f$ tends to infinity). The top-right panel uses CyclicArg shading: the plot represents the argument of the function with colors shaded cyclically based on the argument itself. This shading highlights regions of constant phase, making it easier to analyze the phase behavior and locate zeros and poles. The bottom-left panel uses CyclicLogAbsArg shading: the plot combines cyclic shading based on both the logarithm of the magnitude log(Abs[$f$]) and the argument, Arg[$f$]. This combined shading provides a detailed visualization of both constant magnitude and constant phase regions, enhancing the identification of zeros, poles, and essential singularities. The bottom-right panel uses QuantileAbs shading: the plot represents the argument of the function with colors shaded from dark to light based on quantiles of the magnitude Abs[$f$]. Darker colors represent smaller magnitudes, and lighter colors represent larger magnitudes, emphasizing the relative magnitudes of the function values.

## 9.2 Complex Calculus

**Definition (Complex Differentiable):** Let $\mathbb{A} \subset \mathbb{C}$ be an open set. The function $f: \mathbb{A} \to \mathbb{C}$ is said to be (complex) differentiable at $z_0 \in \mathbb{A}$ if the limit

$$f'(z_0) = \lim_{z \to z_0} \frac{f(z) - f(z_0)}{z - z_0}, \tag{9.30}$$

exists independent of the manner in which $z \to z_0$. This limit is then denoted by $f'(z_0) = \frac{\partial f(z)}{\partial z}\Big|_{z=z_0}$ and is called the derivative of $f$ with respect to $z$ at the point $z_0$.

**Definition (Holomorphic or Analytic):** Let $\mathbb{U} \subseteq \mathbb{A}$ be a nonempty open set. The function $f: \mathbb{A} \to \mathbb{C}$ is called holomorphic (or analytic) in $\mathbb{U}$ , if $f$ is differentiable in $z_0$ for all $z_0 \in \mathbb{U}$.

Basic properties for the derivative of a sum, product, and composition of two functions known from real-valued analysis remain inherently valid in the complex domain. Assume that $f(z)$ and $g(z)$ are differentiable at $z_0$. Then, the following propositions hold:

The sum $f + g$ is differentiable at $z_0$ and

$$(f+g)'(z_0) = f'(z_0) + g'(z_0). \tag{9.31.1}$$

The product $fg$ is differentiable at $z_0$ and

$$(fg)'(z_0) = f'(z_0)g(z_0) + f(z_0)g'(z_0). \tag{9.31.2}$$

If $g(z_0) \neq 0$, the quotient $f/g$ is differentiable at $z_0$ and

$$\left(\frac{f}{g}\right)'(z_0) = \frac{f'(z_0)g(z_0) - f(z_0)g'(z_0)}{g^2(z_0)}. \tag{9.31.3}$$

If $w = f(\xi)$ where $\xi = g(z)$ ((chain rule)) then

$$\frac{dw}{dz} = \frac{dw}{d\xi}\frac{d\xi}{dz} = f'(\xi)\frac{d\xi}{dz} = f'\big(g(z)\big)g'(z). \tag{9.31.4}$$

**Theorem 9.1:** A necessary condition that $w = f(z) = u(x,y) + iv(x,y)$, $f: \mathbb{C} \to \mathbb{C}$, with $u, v \in \mathbb{R}$, and $z = x + iy$ with $x, y \in \mathbb{R}$ be analytic in a region $\mathbb{U}$ is that, in $\mathbb{U}$, $u$ and $v$ satisfy the Cauchy–Riemann equations [266-271]

$$\frac{\partial u}{\partial x} = \frac{\partial v}{\partial y}, \qquad \frac{\partial u}{\partial y} = -\frac{\partial v}{\partial x}. \tag{9.32}$$

If the partial derivatives in (9.32) are continuous in $\mathbb{U}$, then the Cauchy–Riemann equations are sufficient conditions that $f(z)$ be analytic in $\mathbb{U}$.

**Proof:**

**Necessary Condition:**

In order for $f(z)$ to be analytic, the limit

$$\begin{aligned} f'(z) &= \lim_{\Delta z \to 0} \frac{f(z+\Delta z) - f(z)}{\Delta z} \\ &= \lim_{\substack{\Delta x \to 0 \\ \Delta y \to 0}} \frac{[u(x+\Delta x, y+\Delta y) + iv(x+\Delta x, y+\Delta y)] - [u(x,y) + iv(x,y)]}{\Delta x + i\Delta y}, \end{aligned}$$

must exist independent of the manner in which $\Delta z$ (or $\Delta x$ and $\Delta y$) approaches zero. We consider two possible approaches.

**Case 1:** $\Delta y$ =0, $\Delta x \to 0$. In this case, $\lim_{\Delta z \to 0} \frac{f(z+\Delta z) - f(z)}{\Delta z} = f'(z)$ becomes

$$\begin{aligned} &\lim_{\Delta x \to 0} \frac{[u(x+\Delta x, y) + iv(x+\Delta x, y)] - [u(x,y) + iv(x,y)]}{\Delta x} \\ &= \lim_{\Delta x \to 0} \frac{[u(x+\Delta x, y) - u(x,y)] + i[v(x+\Delta x, y) - v(x,y)]}{\Delta x} \\ &= \lim_{\Delta x \to 0} \left( \frac{u(x+\Delta x, y) - u(x,y)}{\Delta x} + i\frac{v(x+\Delta x, y) - v(x,y)}{\Delta x} \right) \\ &= \frac{\partial u}{\partial x} + i\frac{\partial v}{\partial x}, \end{aligned}$$

provided the partial derivatives exist.

**Case 2:** $\Delta x = 0$, $\Delta y \to 0$. In this case, $\lim_{\Delta z \to 0} \frac{f(z+\Delta z)-f(z)}{\Delta z} = f'(z)$ becomes

$$\begin{aligned}
&\lim_{\Delta y \to 0} \frac{[u(x,y+\Delta y) + iv(x,y+\Delta y)] - [u(x,y) + iv(x,y)]}{i\Delta y} \\
&= \lim_{\Delta y \to 0} \frac{[u(x,y+\Delta y) - u(x,y)] + i[v(x,y+\Delta y) - v(x,y)]}{i\Delta y} \\
&= \lim_{\Delta y \to 0} \left( \frac{u(x,y+\Delta y) - u(x,y)}{i\Delta y} + \frac{i[v(x,y+\Delta y) - v(x,y)]}{i\Delta y} \right) \\
&= \lim_{\Delta y \to 0} \left( -i \frac{u(x,y+\Delta y) - u(x,y)}{\Delta y} + \frac{[v(x,y+\Delta y) - v(x,y)]}{\Delta y} \right) \\
&= -i \frac{\partial u}{\partial y} + \frac{\partial v}{\partial y}.
\end{aligned}$$

Now $f(z)$ cannot possibly be analytic unless these two limits are identical. Thus, a necessary condition that $f(z)$ be analytic is

$$\frac{\partial u}{\partial x} + i\frac{\partial v}{\partial x} = -i\frac{\partial u}{\partial y} + \frac{\partial v}{\partial y},$$

or

$$\frac{\partial u}{\partial x} = \frac{\partial v}{\partial y}, \qquad \frac{\partial v}{\partial x} = -\frac{\partial u}{\partial y}.$$

**Sufficient Condition:**

Since $\frac{\partial u}{\partial x}$ and $\frac{\partial u}{\partial y}$ are supposed to be continuous, we have

$$\begin{aligned}
\Delta u &= u(x+\Delta x, y+\Delta y) - u(x,y) \\
&= \{u(x+\Delta x, y+\Delta y) - u(x, y+\Delta y)\} + \{u(x,y+\Delta y) - u(x,y)\} \\
&= \left\{\frac{\partial u}{\partial x} + \Delta\epsilon_1\right\}\Delta x + \left\{\frac{\partial u}{\partial y} + \Delta\eta_1\right\}\Delta y \\
&= \frac{\partial u}{\partial x}\Delta x + \Delta\epsilon_1\Delta x + \frac{\partial u}{\partial y}\Delta y + \Delta\eta_1\Delta y \\
&= \frac{\partial u}{\partial x}\Delta x + \frac{\partial u}{\partial y}\Delta y + \Delta\epsilon_1\Delta x + \Delta\eta_1\Delta y,
\end{aligned}$$

where $\Delta\epsilon_1 \to 0$ and $\Delta\eta_1 \to 0$ as $\Delta x \to 0$ and $\Delta y \to 0$.

Similarly, since $\frac{\partial v}{\partial x}$ and $\frac{\partial v}{\partial y}$ are supposed to be continuous, we have

$$\begin{aligned}
\Delta v &= v(x+\Delta x, y+\Delta y) - v(x,y) \\
&= \{v(x+\Delta x, y+\Delta y) - v(x, y+\Delta y)\} + \{v(x,y+\Delta y) - v(x,y)\} \\
&= \left\{\frac{\partial v}{\partial x} + \Delta\epsilon_2\right\}\Delta x + \left\{\frac{\partial v}{\partial y} + \Delta\eta_2\right\}\Delta y \\
&= \frac{\partial v}{\partial x}\Delta x + \Delta\epsilon_2\Delta x + \frac{\partial v}{\partial y}\Delta y + \Delta\eta_2\Delta y \\
&= \frac{\partial v}{\partial x}\Delta x + \frac{\partial v}{\partial y}\Delta y + \Delta\epsilon_2\Delta x + \Delta\eta_2\Delta y,
\end{aligned}$$

where $\Delta\epsilon_2 \to 0$ and $\Delta\eta_2 \to 0$ as $\Delta x \to 0$ and $\Delta y \to 0$.

Then

$$
\begin{aligned}
\Delta w &= \Delta u + i\Delta v \\
&= \frac{\partial u}{\partial x}\Delta x + \frac{\partial u}{\partial y}\Delta y + \Delta\epsilon_1\Delta x + \Delta\eta_1\Delta y + i\left(\frac{\partial v}{\partial x}\Delta x + \frac{\partial v}{\partial y}\Delta y + \Delta\epsilon_2\Delta x + \Delta\eta_2\Delta y\right) \\
&= \frac{\partial u}{\partial x}\Delta x + i\frac{\partial v}{\partial x}\Delta x + \frac{\partial u}{\partial y}\Delta y + i\frac{\partial v}{\partial y}\Delta y + \Delta\epsilon_1\Delta x + \Delta\eta_1\Delta y + i\Delta\epsilon_2\Delta x + i\Delta\eta_2\Delta y \\
&= \left(\frac{\partial u}{\partial x} + i\frac{\partial v}{\partial x}\right)\Delta x + \left(\frac{\partial u}{\partial y} + i\frac{\partial v}{\partial y}\right)\Delta y + (\Delta\epsilon_1 + i\Delta\epsilon_2)\Delta x + (\Delta\eta_1 + i\Delta\eta_2)\Delta y \\
&= \left(\frac{\partial u}{\partial x} + i\frac{\partial v}{\partial x}\right)\Delta x + \left(\frac{\partial u}{\partial y} + i\frac{\partial v}{\partial y}\right)\Delta y + \Delta\epsilon\Delta x + \Delta\eta\Delta y,
\end{aligned}
$$

or

$$
\Delta w = \left(\frac{\partial u}{\partial x} + i\frac{\partial v}{\partial x}\right)\Delta x + \left(\frac{\partial u}{\partial y} + i\frac{\partial v}{\partial y}\right)\Delta y + \Delta\epsilon\Delta x + \Delta\eta\Delta y,
$$

where $\Delta\epsilon = \Delta\epsilon_1 + i\Delta\epsilon_2 \to 0$ and $\Delta\eta = \Delta\eta_1 + i\Delta\eta_2 \to 0$ as $\Delta x \to 0$ and $\Delta x \to 0$.

By the Cauchy–Riemann equations, $\Delta w$ can be written

$$
\begin{aligned}
\Delta w &= \left(\frac{\partial u}{\partial x} + i\frac{\partial v}{\partial x}\right)\Delta x + \left(-\frac{\partial v}{\partial x} + i\frac{\partial u}{\partial x}\right)\Delta y + \Delta\epsilon\Delta x + \Delta\eta\Delta y \\
&= \frac{\partial u}{\partial x}\Delta x + i\frac{\partial v}{\partial x}\Delta x - \frac{\partial v}{\partial x}\Delta y + i\frac{\partial u}{\partial x}\Delta y + \Delta\epsilon\Delta x + \Delta\eta\Delta y \\
&= \frac{\partial u}{\partial x}\Delta x + i\frac{\partial u}{\partial x}\Delta y + i\frac{\partial v}{\partial x}\Delta x - \frac{\partial v}{\partial x}\Delta y + \Delta\epsilon\Delta x + \Delta\eta\Delta y \\
&= \frac{\partial u}{\partial x}(\Delta x + i\Delta y) + i\frac{\partial v}{\partial x}\left(\Delta x - \frac{1}{i}\Delta y\right) + \Delta\epsilon\Delta x + \Delta\eta\Delta y \\
&= \frac{\partial u}{\partial x}(\Delta x + i\Delta y) + i\frac{\partial v}{\partial x}(\Delta x + i\Delta y) + \Delta\epsilon\Delta x + \Delta\eta\Delta y \\
&= \left(\frac{\partial u}{\partial x} + i\frac{\partial v}{\partial x}\right)(\Delta x + i\Delta y) + \Delta\epsilon\Delta x + \Delta\eta\Delta y.
\end{aligned}
$$

Then, on dividing by $\Delta z = \Delta x + i\Delta y$ and taking the limit as $\Delta z \to 0$, we see that

$$
\frac{dw}{dz} = f'(z) = \lim_{\Delta z \to 0}\frac{\Delta w}{\Delta z} = \frac{\partial u}{\partial x} + i\frac{\partial v}{\partial x},
$$

so that the derivative exists and is unique, i.e., $f(z)$ is analytic in $\mathbb{U}$.

■

The geometric interpretation of the complex derivative is analogous to the interpretation of the derivative for real functions but with some differences due to the fact that complex functions map from the complex plane to itself. Let $z_0$, Figure 9.9, be a point $P$ in the $z$ plane and let $w_0$ be its image $P'$ in the $w$ plane under the transformation $w = f(z)$. Since we suppose that $f(z)$ is single-valued, the point $z_0$ maps into only one point $w_0$. If we give $z_0$ an increment $\Delta z$, we obtain the point $Q$ of Figure 9.9. This point has image $Q'$ in the $w$ plane. Thus, from Figure 9.9, we see that $P'Q'$ represents the complex number $\Delta w = f(z_0 + \Delta z) - f(z_0)$. It follows that the derivative at $z_0$ (if it exists) is given by

$$
\lim_{\Delta z \to 0}\frac{\Delta w}{\Delta z} = \lim_{\Delta z \to 0}\frac{f(z_0 + \Delta z) - f(z_0)}{\Delta z} = \lim_{\Delta z \to 0}\frac{P'Q'}{PQ}, \tag{9.33}
$$

that is, the limit of the ratio $P'Q'$ to $PQ$ as point $Q$ approaches point $P$. This interpretation clearly holds when $z_0$ is replaced by any point $z$.

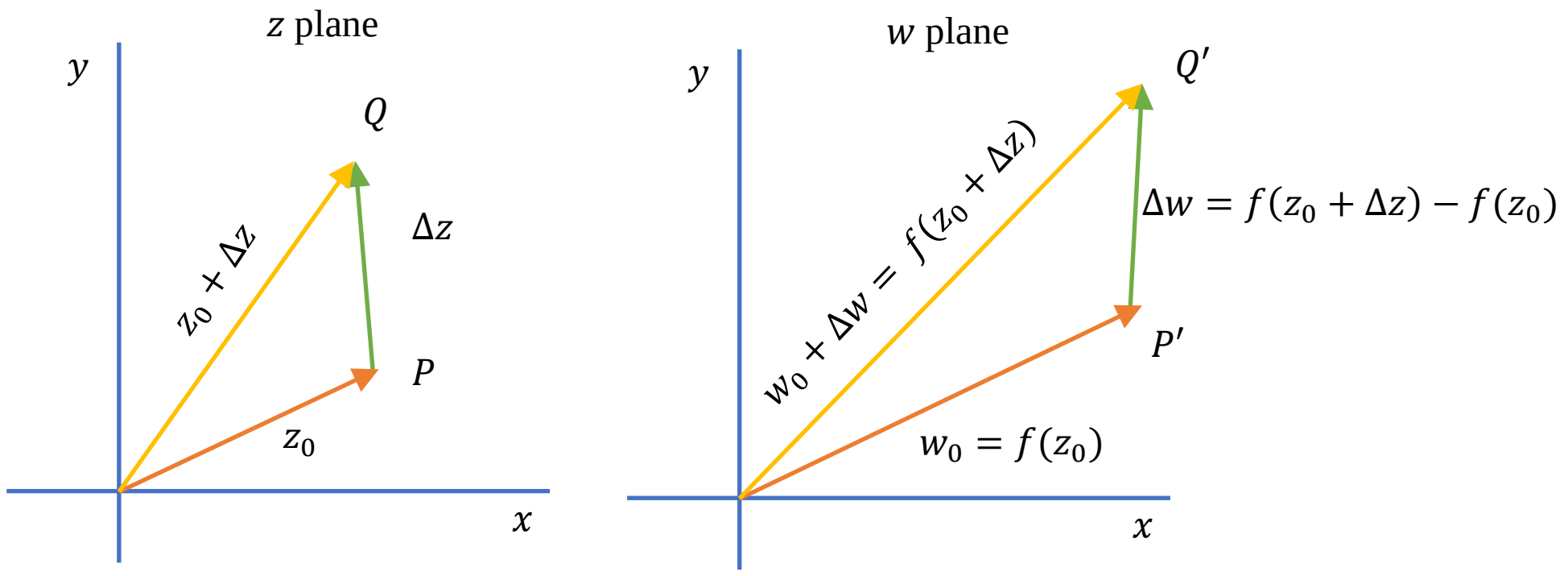

**Figure 9.9.** Geometric Interpretation of the Complex Derivative. In the $z$-plane, point $P$ corresponds to $z_0$, and point $Q$ corresponds to $z_0 + \Delta z$. The change in the complex number is denoted by $\Delta z$. In the $w$-plane, point $P'$ corresponds to $w_0 = f(z_0)$, and point $Q'$ corresponds to $f(z_0 + \Delta z)$. The change in the function value is denoted by $\Delta w = f(z_0 + \Delta z) - f(z_0)$. The interpretation showcases how a small perturbation $\Delta z$ in the $z$-plane maps to a perturbation $\Delta w$ in the $w$-plane, illustrating the derivative's role in describing this transformation.

Let $\Delta z = dz$ be an increment given to $z$. Then $\Delta w = f(z + \Delta z) - f(z)$ is called the increment in $w = f(z)$. If $f(z)$ is continuous and has a continuous first derivative in a region, then

$$\Delta w = f'(z)\Delta z + \epsilon\Delta z = f'(z)\mathrm{d}z + \epsilon\mathrm{d}z, \tag{9.34}$$

where $\epsilon \to 0$ as $\Delta z \to 0$. The expression

$$\mathrm{d}w = f'(z)\mathrm{d}z, \tag{9.35}$$

is called the differential of $w$ or $f(z)$, or the principal part of $\Delta w$.

**Theorem 9.2:** The differential $\mathrm{d}f$ of a complex-valued function $f(z): \mathbb{A} \to \mathbb{C}$ with $\mathbb{A} \subset \mathbb{C}$ can be expressed as

$$\mathrm{d}f = \frac{\partial f(z)}{\partial z} dz + \frac{\partial f(z)}{\partial z^*} dz^*. \tag{9.36}$$

**Proof:**

The total differential of the bivariate function $F(x,y): \mathbb{R}^2 \to \mathbb{C}$ and $u, v : \mathbb{R}^2 \to \mathbb{R}$ associated with the univariate function $f(z)$ via

$$z = x + iy, \qquad F(x,y) = u(x,y) + iv(x,y) = f(z) = f(z = x + iy),$$

reads as

$$\mathrm{d}F = \frac{\partial F(x,y)}{\partial x}\mathrm{d}x + \frac{\partial F(x,y)}{\partial y}\mathrm{d}y.$$

Of course, the differentiability of $F(x,y)$ with respect to $x$ and $y$ in the real sense has to be imposed for the existence of the differential $\mathrm{d}F$. This implies the differentiability of the real-valued functions $u(x,y)$ and $v(x,y)$ with respect to $x$ and $y$. Rewriting $\mathrm{d}F$ by means of $F(x,y) = u(x,y) + iv(x,y)$ yields

$$\begin{aligned}\mathrm{d}F &= \frac{\partial F(x,y)}{\partial x}\mathrm{d}x + \frac{\partial F(x,y)}{\partial y}\mathrm{d}y \\ &= \frac{\partial}{\partial x}[u(x,y) + iv(x,y)]\mathrm{d}x + \frac{\partial}{\partial y}[u(x,y) + iv(x,y)]\mathrm{d}y \\ &= \frac{\partial u(x,y)}{\partial x}\mathrm{d}x + i\frac{\partial v(x,y)}{\partial x}\mathrm{d}x + \frac{\partial u(x,y)}{\partial y}\mathrm{d}y + i\frac{\partial v(x,y)}{\partial y}\mathrm{d}y.\end{aligned}$$

Making use of

$$\mathrm{d}z = \mathrm{d}x + i\mathrm{d}y,$$

$$\mathrm{d}z^* = \mathrm{d}x - i\mathrm{d}y,$$

the two differentials $dx$ and $dy$ can be expressed via

$$\mathrm{d}x = \frac{1}{2}(\mathrm{d}z + \mathrm{d}z^*), \qquad \mathrm{d}y = \frac{1}{2i}(\mathrm{d}z - \mathrm{d}z^*).$$

Inserting $\mathrm{d}x$ and $\mathrm{d}y$ into the differential expression $\mathrm{d}F$ and reordering the result leads to

$$\begin{aligned}
\mathrm{d}F &= \frac{\partial u(x,y)}{\partial x}\mathrm{d}x + i\frac{\partial v(x,y)}{\partial x}\mathrm{d}x + \frac{\partial u(x,y)}{\partial y}\mathrm{d}y + i\frac{\partial v(x,y)}{\partial y}\mathrm{d}y \\
&= \frac{\partial u(x,y)}{\partial x}\frac{1}{2}(\mathrm{d}z + \mathrm{d}z^*) + i\frac{\partial v(x,y)}{\partial x}\frac{1}{2}(\mathrm{d}z + \mathrm{d}z^*) \\
&+ \frac{\partial u(x,y)}{\partial y}\frac{1}{2i}(\mathrm{d}z - \mathrm{d}z^*) + i\frac{\partial v(x,y)}{\partial y}\frac{1}{2i}(\mathrm{d}z - \mathrm{d}z^*) \\
&= \left(\frac{1}{2}\frac{\partial u(x,y)}{\partial x}\mathrm{d}z + \frac{1}{2}\frac{\partial u(x,y)}{\partial x}\mathrm{d}z^*\right) + \left(\frac{i}{2}\frac{\partial v(x,y)}{\partial x}\mathrm{d}z + \frac{i}{2}\frac{\partial v(x,y)}{\partial x}\mathrm{d}z^*\right) \\
&+ \left(\frac{1}{2i}\frac{\partial u(x,y)}{\partial y}\mathrm{d}z - \frac{1}{2i}\frac{\partial u(x,y)}{\partial y}\mathrm{d}z^*\right) + \left(\frac{1}{2}\frac{\partial v(x,y)}{\partial y}\mathrm{d}z - \frac{1}{2}\frac{\partial v(x,y)}{\partial y}\mathrm{d}z^*\right) \\
&= \left(\frac{1}{2}\frac{\partial u(x,y)}{\partial x}\mathrm{d}z + \frac{i}{2}\frac{\partial v(x,y)}{\partial x}\mathrm{d}z + \frac{1}{2i}\frac{\partial u(x,y)}{\partial y}\mathrm{d}z + \frac{1}{2}\frac{\partial v(x,y)}{\partial y}\mathrm{d}z\right) \\
&+ \left(\frac{1}{2}\frac{\partial u(x,y)}{\partial x}\mathrm{d}z^* + \frac{i}{2}\frac{\partial v(x,y)}{\partial x}\mathrm{d}z^* - \frac{1}{2i}\frac{\partial u(x,y)}{\partial y}\mathrm{d}z^* - \frac{1}{2}\frac{\partial v(x,y)}{\partial y}\mathrm{d}z^*\right) \\
&= \frac{1}{2}\left(\frac{\partial u(x,y)}{\partial x} + i\frac{\partial v(x,y)}{\partial x} - i\frac{\partial u(x,y)}{\partial y} + \frac{\partial v(x,y)}{\partial y}\right)\mathrm{d}z \\
&+ \frac{1}{2}\left(\frac{\partial u(x,y)}{\partial x} + i\frac{\partial v(x,y)}{\partial x} + i\frac{\partial u(x,y)}{\partial y} - \frac{\partial v(x,y)}{\partial y}\right)\mathrm{d}z^* \\
&= \frac{1}{2}\left(\frac{\partial}{\partial x}[u(x,y) + iv(x,y)] - i\frac{\partial}{\partial y}[u(x,y) + iv(x,y)]\right)\mathrm{d}z \\
&+ \frac{1}{2}\left(\frac{\partial}{\partial x}[u(x,y) + iv(x,y)] + i\frac{\partial}{\partial y}[u(x,y) + iv(x,y)]\right)\mathrm{d}z^* \\
&= \frac{1}{2}\left(\frac{\partial}{\partial x}F(x,y) - i\frac{\partial}{\partial y}F(x,y)\right)\mathrm{d}z + \frac{1}{2}\left(\frac{\partial}{\partial x}F(x,y) + i\frac{\partial}{\partial y}F(x,y)\right)\mathrm{d}z^* \\
&= \frac{1}{2}\left(\frac{\partial}{\partial x} - i\frac{\partial}{\partial y}\right)F(x,y)\mathrm{d}z + \frac{1}{2}\left(\frac{\partial}{\partial x} + i\frac{\partial}{\partial y}\right)F(x,y)\mathrm{d}z^*.
\end{aligned}$$

Finally,

$$\mathrm{d}F = \frac{1}{2}\left(\frac{\partial}{\partial x} - i\frac{\partial}{\partial y}\right)F(x,y)\mathrm{d}z + \frac{1}{2}\left(\frac{\partial}{\partial x} + i\frac{\partial}{\partial y}\right)F(x,y)\mathrm{d}z^*.$$

According to the total differential for real-valued multivariate functions, the introduction of the two operators

$$\frac{\partial}{\partial z} = \frac{1}{2}\left(\frac{\partial}{\partial x} - i\frac{\partial}{\partial y}\right),$$

$$\frac{\partial}{\partial z^*} = \frac{1}{2}\left(\frac{\partial}{\partial x} + i\frac{\partial}{\partial y}\right),$$

is reasonable as it leads to the very nice description of the differential $\mathrm{d}f$, where the real-valued partial derivatives are hidden.

$$\mathrm{d}f = \frac{\partial f(z)}{\partial z}\mathrm{d}z + \frac{\partial f(z)}{\partial z^*}\mathrm{d}z^*.$$

■

Although many important complex functions are holomorphic, including the functions $z^n$, $e^z$, $\ln(z)$, $\sin(z)$, and $\cos(z)$, and hence differentiable in the standard complex variables sense, there are commonly encountered useful functions which are not:

- The function $f(z) = z^*$, fails to satisfy the Cauchy-Riemann conditions.
- The functions $f(z) = \mathrm{Re}(z) = \frac{z+z^*}{2} = x$ and $g(z) = \mathrm{Im}(z) = i\frac{z-z^*}{2} = y$ fail to satisfy the Cauchy-Riemann conditions.

Note that the nonholomorphic (nonanalytic in the complex variable $z$) functions, given in the above examples, can all be written in the form $f(z, z^*)$, where they are holomorphic in $z = x + iy$ for fixed $z^*$ and holomorphic in $z^* = x - iy$ for fixed $z$. That is, if we make the substitution $w = z^*$, they are analytic in $w$ for fixed $z$, and analytic in $z$ for fixed $w$. It can be shown that this fact is true in general for any complex-valued function

$$f(z) = f(z, z^*) = f(x, y) = u(x, y) + iv(x, y). \tag{9.37}$$

Non-holomorphic functions, however, can be dealt with conjugate coordinates, which are related to the real coordinates by

$$z = x + iy, \qquad z^* = x - iy. \tag{9.38}$$

Then, one can write the two real-variables as,

$$x = \frac{z + z^*}{2}, \qquad y = i\frac{z - z^*}{2}. \tag{9.39}$$

**Definition (Wirtinger Derivatives):** The two 'partial derivative' operators $\frac{\partial}{\partial z}$ and $\frac{\partial}{\partial z^*}$ are defined by [269-271]

$$\frac{\partial}{\partial z} = \frac{1}{2}\left(\frac{\partial}{\partial x} - i\frac{\partial}{\partial y}\right), \tag{9.40.1}$$

$$\frac{\partial}{\partial z^*} = \frac{1}{2}\left(\frac{\partial}{\partial x} + i\frac{\partial}{\partial y}\right), \tag{9.40.2}$$

and are often referred to as the Wirtinger derivatives (sometimes also called Wirtinger operators). The derivatives, (9.40.1) and (9.40.2) are called ℝ-derivative and conjugate ℝ-derivative, respectively.

Consequently,

$$\frac{\partial}{\partial x} = \frac{\partial}{\partial z} + \frac{\partial}{\partial z^*}, \qquad \frac{\partial}{\partial y} = i\left(\frac{\partial}{\partial z} - \frac{\partial}{\partial z^*}\right). \tag{9.41}$$

In the evaluation of $\frac{\partial f}{\partial z}$, $z^*$ is considered as a constant and vice versa. For example, with $f(z, z^*) = z^2 z^*$, we have $\frac{\partial f}{\partial z} = 2zz^*$, $\frac{\partial f}{\partial z^*} = z^2$.

Wirtinger calculus (Wirtinger operators) extends standard calculus rules to the complex domain. The attractiveness of Wirtinger calculus is that it enables us to perform all computations directly in the complex domain, and the derivatives obey all rules of conventional calculus, including the chain rule, differentiation of products, and quotients. If $\alpha$ and $\beta$ are complex numbers, we have

$$\frac{\partial}{\partial z}(\alpha f + \beta g) = \alpha\frac{\partial f}{\partial z} + \beta\frac{\partial g}{\partial z}, \tag{9.42.1}$$

$$\frac{\partial}{\partial z^*}(\alpha f + \beta g) = \alpha\frac{\partial f}{\partial z^*} + \beta\frac{\partial g}{\partial z^*}, \tag{9.42.2}$$

$$\frac{\partial}{\partial z}(f \cdot g) = \frac{\partial f}{\partial z}\cdot g + f\cdot\frac{\partial g}{\partial z}, \tag{9.42.3}$$

$$\frac{\partial}{\partial z^*}(f \cdot g) = \frac{\partial f}{\partial z^*} \cdot g + f \cdot \frac{\partial g}{\partial z^*}, \tag{9.42.4}$$

$$\left(\frac{\partial f}{\partial z}\right)^* = \frac{\partial f^*}{\partial z^*}, \tag{9.42.5}$$

$$\left(\frac{\partial f}{\partial z^*}\right)^* = \frac{\partial f^*}{\partial z}. \tag{9.42.6}$$

An elegant approach due to Wirtinger relaxes the strong requirement for differentiability (Cauchy-Riemann equations), and defines a less stringent form for the complex domain. Wirtinger derivatives allow you to treat $z$ and $z^*$ as independent variables. This is particularly useful when dealing with functions that depend on both a complex variable and its conjugate. This independence is reflected in the fact that

$$\frac{\partial z}{\partial z} = \frac{\partial z^*}{\partial z^*} = 1, \qquad \frac{\partial z}{\partial z^*} = \frac{\partial z^*}{\partial z} = 0. \tag{9.42.7}$$

**Theorem 9.3:** Given $f\colon \mathbb{C} \to \mathbb{C}$ holomorphic with $f(z) = f(z = x + iy) = u(x, y) + iv(x, y)$ where $u, v : \mathbb{R}^2 \to \mathbb{R}$ real differentiable functions. Then

$$\frac{\partial f}{\partial z^*} = 0. \tag{9.43}$$

It is easy to see that the Cauchy-Riemann equations are equivalent to $\frac{\partial f}{\partial z^*} = 0$.

**Proof:**

Using Wirtinger calculus

$$\frac{\partial f}{\partial z^*} = \frac{1}{2}\left(\frac{\partial}{\partial x} + i\frac{\partial}{\partial y}\right) f.$$

By definition, then:

$$\begin{aligned}
\frac{\partial f}{\partial z^*} &= \frac{1}{2}\left(\frac{\partial f}{\partial x} + i\frac{\partial f}{\partial y}\right) \\
&= \frac{1}{2}\left(\frac{\partial}{\partial x}[u(x, y) + iv(x, y)] + i\frac{\partial}{\partial y}[u(x, y) + iv(x, y)]\right) \\
&= \frac{1}{2}\left(\frac{\partial}{\partial x}u(x, y) + i\frac{\partial}{\partial x}v(x, y) + i\frac{\partial}{\partial y}u(x, y) - \frac{\partial}{\partial y}v(x, y)\right) \\
&= \frac{1}{2}\left(\left[\frac{\partial}{\partial x}u(x, y) - \frac{\partial}{\partial y}v(x, y)\right] + i\left[\frac{\partial}{\partial x}v(x, y) + \frac{\partial}{\partial y}u(x, y)\right]\right) \\
&= \frac{1}{2}\left(\left[\frac{\partial}{\partial x}u(x, y) - \frac{\partial}{\partial x}u(x, y)\right] + i\left[\frac{\partial}{\partial x}v(x, y) - \frac{\partial}{\partial x}v(x, y)\right]\right) \\
&= 0.
\end{aligned}$$

Because $f$ is holomorphic then the Cauchy-Riemann equations $\frac{\partial u}{\partial x} = \frac{\partial v}{\partial y}, \frac{\partial u}{\partial y} = -\frac{\partial v}{\partial x}$ applies, making $\frac{\partial f}{\partial z^*}$ equal to zero.

■

The condition $\frac{\partial f}{\partial z^*} = 0$ is true for an $\mathbb{R}$-differentiable function $f$ if and only the Cauchy-Riemann conditions are satisfied. Thus, a function $f$ is holomorphic (complex-analytic in $z$) if and only if it does not depend on the complex conjugate variable $z^*$. This obviously provides a simple and powerful characterization of holomorphic and nonholomorphic functions and shows the elegance of the Wirtinger calculus formulation based on the use of conjugate coordinates $(z, z^*)$. Note that the two Cauchy-Riemann conditions are replaced by the single condition $\frac{\partial f}{\partial z^*} = 0$.

Optimizations in machine learning are targeted on the minimization of a cost. The cost functions are real-valued. On account of this, we focus on functions $f(z)\colon \mathbb{U} \to \mathbb{R}$ having complex-valued arguments $z \in \mathbb{U} \subset \mathbb{C}$ that are mapped to real-valued scalars $f(z) \in \mathbb{R}$.

First of all, it is obvious that the only possibility of a real-valued function $f(z)$ with complex argument $z$ for being analytic is that $f(z)$ is constant for all $z$ of its domain. This follows from the Cauchy-Riemann equations, since $v(x,y) = 0$ for real-valued $f(z)$. This leads to the following proposition.

**Lemma 9.1:** All non-trivial (not constant) real-valued functions $f(z)$ mapping $z \in \mathbb{A} \subset \mathbb{C}$ onto $\mathbb{R}$ are non-analytic functions and therefore not complex differentiable.

**Theorem 9.4** Let $\mathbb{A} \subset \mathbb{C}$, and $f\colon \mathbb{A} \to \mathbb{R}$ be a real-valued function. The total differential of $f$ is given by

$$\mathrm{d}f = 2\mathrm{Re}\left(\frac{\partial f(z)}{\partial z} dz\right) = 2\mathrm{Re}\left(\frac{\partial f(z)}{\partial z^*} dz^*\right). \tag{9.44}$$

**Proof:**

From the definition of the Wirtinger differentials and partial derivatives we obtain

$$\begin{aligned}\frac{\partial f(z)}{\partial z} dz &= \frac{1}{2}\left(\frac{\partial f}{\partial x} - i\frac{\partial f}{\partial y}\right)(\mathrm{d}x + i\mathrm{d}y) \\ &= \frac{1}{2}\frac{\partial f}{\partial x}\mathrm{d}x + \frac{i}{2}\frac{\partial f}{\partial x}\mathrm{d}y - \frac{i}{2}\frac{\partial f}{\partial y}\mathrm{d}x + \frac{1}{2}\frac{\partial f}{\partial y}\mathrm{d}y \\ &= \frac{1}{2}\left(\frac{\partial f}{\partial x}\mathrm{d}x + \frac{\partial f}{\partial y}\mathrm{d}y\right) + \frac{i}{2}\left(\frac{\partial f}{\partial x}\mathrm{d}y - \frac{\partial f}{\partial y}\mathrm{d}x\right).\end{aligned}$$

Hence,

$$2\mathrm{Re}\left(\frac{\partial f(z)}{\partial z} dz\right) = \frac{\partial f}{\partial x}\mathrm{d}x + \frac{\partial f}{\partial y}\mathrm{d}y = \mathrm{d}f.$$

The analog statement holds for the conjugates

$$\begin{aligned}\frac{\partial f(z)}{\partial z^*} dz^* &= \frac{1}{2}\left(\frac{\partial f}{\partial x} + i\frac{\partial f}{\partial y}\right)(\mathrm{d}x - i\mathrm{d}y) \\ &= \frac{1}{2}\frac{\partial f}{\partial x}\mathrm{d}x - \frac{i}{2}\frac{\partial f}{\partial x}\mathrm{d}y + \frac{i}{2}\frac{\partial f}{\partial y}\mathrm{d}x + \frac{1}{2}\frac{\partial f}{\partial y}\mathrm{d}y \\ &= \frac{1}{2}\left(\frac{\partial f}{\partial x}\mathrm{d}x + \frac{\partial f}{\partial y}\mathrm{d}y\right) + \frac{i}{2}\left(\frac{\partial f}{\partial y}\mathrm{d}x - \frac{\partial f}{\partial x}\mathrm{d}y\right).\end{aligned}$$

Hence,

$$2\mathrm{Re}\left(\frac{\partial f(z)}{\partial z^*} dz^*\right) = \frac{\partial f}{\partial x}\mathrm{d}x + \frac{\partial f}{\partial y}\mathrm{d}y = \mathrm{d}f.$$

■

## 9.3 Complex Backpropagation Algorithms

Real-Valued Neural Networks (RVNNs) are the standard form of NNs where all parameters, inputs, and outputs are real numbers. They are extensively used due to their simplicity, ease of implementation, and well-established theoretical foundation. The real-valued nature simplifies the mathematical operations involved, making RVNNs computationally efficient. Applications of RVNNs span various domains, including:

- Image recognition and classification
- Natural language processing
- Autonomous driving
- Financial modeling
- Medical diagnosis

CVNNs, in contrast, operate with complex numbers for their inputs, parameters, and potentially their outputs [63]. This added complexity allows CVNNs to capture information and relationships that RVNNs might miss, particularly in domains where the data naturally resides in the complex plane. CVNNs are particularly advantageous in applications where phase information is crucial or where the underlying processes are inherently complex-valued. Key applications include:

- MRI signal processing
- Speech enhancement
- Wind prediction
- Image classification and segmentation

While CVNNs offer advantages in specific applications, they are often more computationally intensive than RVNNs due to the need to handle complex arithmetic. This increased complexity can make CVNNs more challenging to implement and require more computational resources. However, recent advancements in hardware and optimization techniques are gradually mitigating these challenges, leading to the broader adoption of CVNNs in various fields.

**Case 1:**

**The derivation is based on the assumption that the (complex) derivative $\sigma'(z) = d\sigma(z)/dz$ of the AF $\sigma(z)$ exists.**

A multilayer perceptron consists of many adaptive linear combiners, each of which has a nonlinearity at its output as shown in Figure 9.10. The input/output relationship of such a unit is characterized by the nonlinear recursive difference equation

$$a_i^{(l+1)} = \sigma\left(\sum_{j=1}^{N_l} w_{ij}^{(l)} a_j^{(l)} + b_i^{(l)}\right), \tag{9.45}$$

and this relation is generalized to all units in multilayer perceptron as listed in Figure 9.10.

The error signal $\epsilon_j$, required for adaptation is defined as the difference between the desired response and the output of the perceptron:

$$\epsilon_j(n) = d_j(n) - o_j(n), \qquad j = 1,2, \dots, N_L, \tag{9.46}$$

where $d_j(n)$ is the desired response at the $j$-th node of the output layer at time $n$; $o_j(n) = a_j^{(L)}(n)$ is the output at the $j$-th node of the output layer, and $N_L$ is the number of nodes in the $L$-th layer or the output layer. Hence, the sum of error squares produced by the network is [272]

$$\mathcal{L}(n) = \sum_{j=1}^{N_L} \epsilon_j(n)\epsilon_j^*(n). \tag{9.47}$$

The BP algorithm minimizes the cost functional $\mathcal{L}(n)$ by recursively altering the coefficient $\{w_{ij}^{(l)}, b_i^{(l)}\}$ based on the gradient search technique. Thus, finding the gradient vector of $\mathcal{L}(n)$ is the main idea of deriving the BP algorithm. We first find the partial derivative of $\mathcal{L}(n)$ with respect to the coefficients of the output layer, and then extend to the coefficient of all hidden units. Since $\mathcal{L}(n)$ is a real-valued function which is not analytic, we need to derive the partial derivative of $\mathcal{L}(n)$ with respect to the real and imaginary parts of the coefficients separately. Writing $w_{ij}^{(l)}(n)$ as

$$w_{ij}^{(l)}(n) = w_{ij}^{\Re,(l)}(n) + i w_{ij}^{\Im,(l)}(n), \tag{9.48}$$

our purpose is to obtain $\partial\mathcal{L}(n)/\partial w_{ij}^{\Re,(l)}$ and $\partial\mathcal{L}(n)/\partial w_{ij}^{\Im,(l)}$. First, let us consider the update rule of the output layer

$$w_{ij}^{\Re,(L-1)}(n+1) = w_{ij}^{\Re,(L-1)}(n) - \frac{1}{2}\alpha \frac{\partial\mathcal{L}(n)}{\partial w_{ij}^{\Re,(L-1)}(n)}, \tag{9.49.1}$$

$$w_{ij}^{\Im,(L-1)}(n+1) = w_{ij}^{\Im,(L-1)}(n) - \frac{1}{2}\alpha \frac{\partial\mathcal{L}(n)}{\partial w_{ij}^{\Im,(L-1)}(n)}. \tag{9.49.2}$$

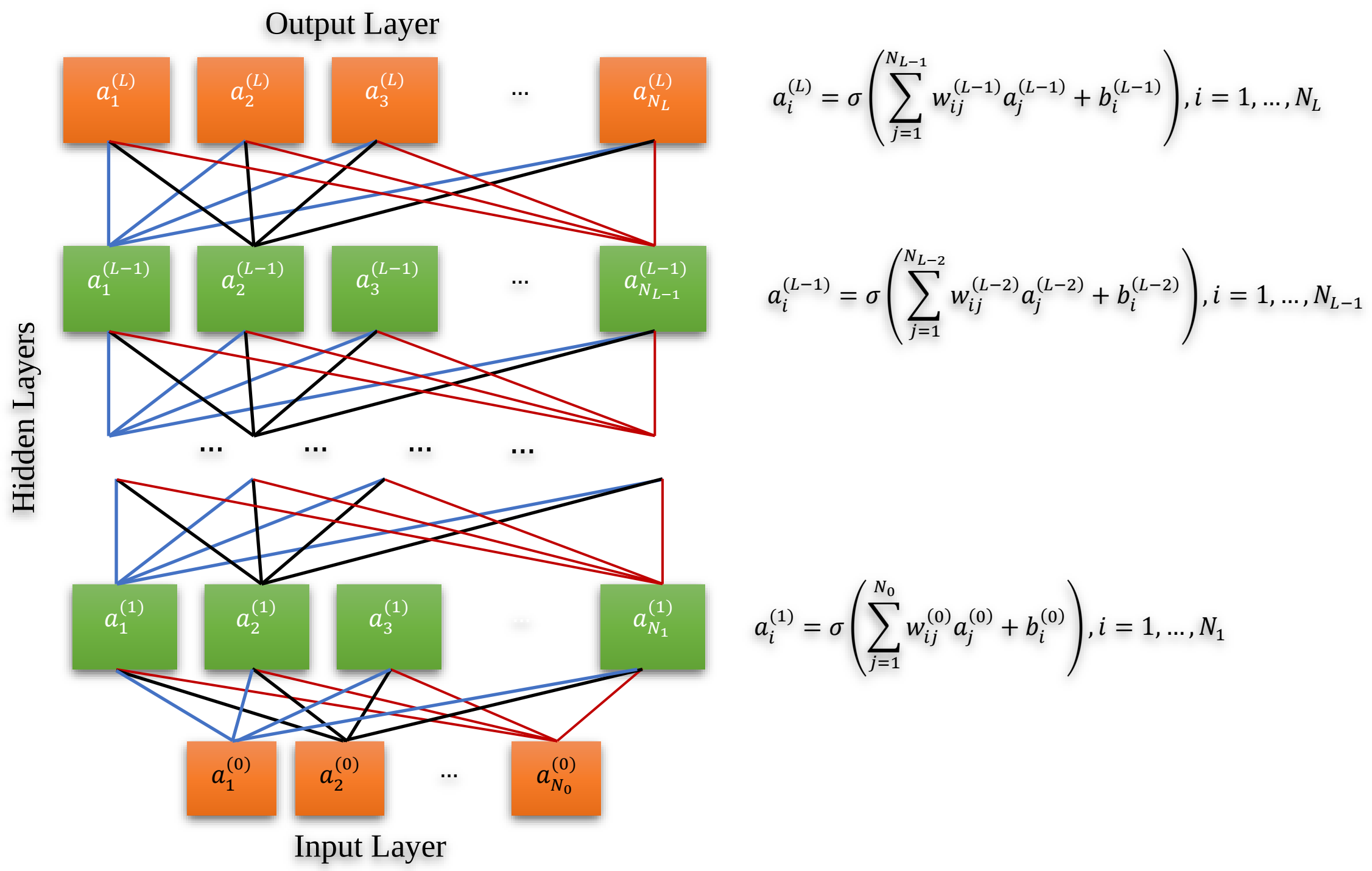

**Figure 9.10.** Structure of a MLP. The diagram represents a MLP with one input layer, several hidden layers, and one output layer. The input layer consists of nodes $a_1^{(0)}, a_2^{(0)}, \ldots, a_{N_0}^{(0)}$, which receive the input features. Each hidden layer $m$ has nodes $a_1^{(m)}, a_2^{(m)}, \ldots, a_{N_m}^{(m)}$, where each node applies a weighted sum of the inputs from the previous layer, adds a bias term, and passes the result through an AF ($\sigma$). The output layer nodes $a_1^{(L)}, a_2^{(L)}, \ldots, a_{N_L}^{(L)}$ produce the final output of the network. This structure enables the MLP to learn and model complex patterns in the data through a series of linear transformations and non-linear activations.

Combining (9.49.1) and (9.49.2), we have

$$w_{ij}^{\Re,(L-1)}(n+1) + i w_{ij}^{\Im,(L-1)}(n+1) = w_{ij}^{\Re,(L-1)}(n) + i w_{ij}^{\Im,(L-1)}(n) - \frac{1}{2}\alpha\left(\frac{\partial\mathcal{L}(n)}{\partial w_{ij}^{\Re,(L-1)}(n)} + i\frac{\partial\mathcal{L}(n)}{\partial w_{ij}^{\Im,(L-1)}(n)}\right), \tag{9.50.1}$$

or

$$w_{ij}^{(L-1)}(n+1) = w_{ij}^{(L-1)}(n) - \frac{1}{2}\alpha\left(\frac{\partial\mathcal{L}(n)}{\partial w_{ij}^{\Re,(L-1)}(n)} + i\frac{\partial\mathcal{L}(n)}{\partial w_{ij}^{\Im,(L-1)}(n)}\right). \tag{9.50.2}$$

Thus, the next step is to find some expressions for the partial derivative:

$$\frac{\partial \mathcal{L}(n)}{\partial w_{ij}^{\Re,(L-1)}} = \frac{\partial \mathcal{L}(n)}{\partial o_i} \frac{\partial o_i}{\partial z_i^{(L)}} \frac{\partial z_i^{(L)}}{\partial w_{ij}^{\Re,(L-1)}} + \frac{\partial \mathcal{L}(n)}{\partial o_i^*} \frac{\partial o_i^*}{\partial z_i^{(L)^*}} \frac{\partial z_i^{(L)^*}}{\partial w_{ij}^{\Re,(L-1)}}, \tag{9.51.1}$$

where

$$o_i = \sigma\left(z_i^{(L)}\right), \tag{9.51.2}$$

$$z_i^{(L)} = \sum_{j=1}^{N_{L-1}} w_{ij}^{(L-1)} a_j^{(L-1)} + b_i^{(L-1)}, \tag{9.51.3}$$

and $\sigma$ can be any nonlinear function in principle. We take $\sigma$ as the sigmoidal function $1/1 + \exp(-z_i)$. The exponential function is now used in its complex version, and the principal values are used. One point to stress is that if we use the exponential function, we need to take care of the problem of singularities. We avoid this problem by scaling the input data to some region in the complex plane.

One point that should be noted is the absence of the partial derivatives $\partial o_i/\partial z_i^*$ and $\partial o_i^*/\partial z_i$, in (9.51.1). More precisely, there should be two more terms in (9.51.1) when the chain rule is applied. When we look at the definition of $o_i$, the function $\sigma$ is a sigmoidal function which does not contain any complex operation. That is, $\sigma$ maps a real number to a real number and a complex number to a complex number, but it will not map a real number to a complex number. Thus

$$a_i^{*(l+1)} = \sigma\left(z_i^{*(l+1)}\right), \qquad l = 0,1, \ldots, L-1, \tag{9.52}$$

and hence those two derivatives are equal to zero.

Evaluating all the partial derivatives in (9.51.1), we get

$$\frac{\partial \mathcal{L}(n)}{\partial w_{ij}^{\Re,(L-1)}} = -(d_i^* - o_i^*)\sigma'(z_i)a_j^{(L-1)} - (d_i - o_i)\sigma'(z_i^*)a_j^{*(L-1)}. \tag{9.53.1}$$

Similarly

$$\frac{\partial \mathcal{L}(n)}{\partial w_{ij}^{\Im,(L-1)}} = -i(d_i^* - o_i^*)\sigma'(z_i)a_j^{(L-1)} + i(d_i - o_i)\sigma'(z_i^*)a_j^{*(L-1)}. \tag{9.53.2}$$

Combining (9.53.1) and (9.53.2), we have

$$\frac{\partial \mathcal{L}(n)}{\partial w_{ij}^{\Re,(L-1)}} + i\frac{\partial \mathcal{L}(n)}{\partial w_{ij}^{\Im,(L-1)}} = -2(d_i - o_i)\sigma'(z_i^*)a_j^{*(L-1)}. \tag{9.54}$$

Substituting (9.54) into (9.50.2), we obtain the adaptation rule for the output layer:

$$w_{ij}^{(L-1)}(n+1) = w_{ij}^{(L-1)}(n) + \alpha\big(d_i(n) - o_i(n)\big)\sigma'\big(z_i^*(n)\big)a_j^{*(L-1)}(n),$$
$$i = j, \ldots, N_L;$$
$$j = 1,2, \ldots, N_{L-1}, \tag{9.55}$$

where $N_L$ and $N_{L-1}$, are the numbers of nodes in layer $L$ (output layer) and layer $L-1$ (the hidden layer converted to the output layer), respectively. We now apply the same procedure to the $(L-1)$-th layer. The generalization to other hidden layers is trivial and will be given at the end of this section. The adaptation rule for this hidden layer is the same as (9.50.2). Thus, what we need to do is to find the partial derivatives

$$\frac{\partial \mathcal{L}(n)}{\partial w_{ij}^{\Re,(L-2)}} = \frac{\partial \mathcal{L}(n)}{\partial a_i^{(L-1)}} \frac{\partial a_i^{(L-1)}}{\partial z_i^{(L-1)}} \frac{\partial z_i^{(L-1)}}{\partial w_{ij}^{\Re,(L-2)}} + \frac{\partial \mathcal{L}(n)}{\partial a_i^{*(L-1)}} \frac{\partial a_i^{*(L-1)}}{\partial z_i^{*(L-1)}} \frac{\partial z_i^{*(L-1)}}{\partial w_{ij}^{\Re,(L-2)}}. \tag{9.56}$$

Since $\mathcal{L}(n)$ is not related to $a_i^{(L-1)}$ or $a_i^{*(L-1)}$ explicitly, the evaluation of the above equation requires the use of the chain rule again. Consider

$$\frac{\partial \mathcal{L}(n)}{\partial a_i^{(L-1)}} = -\sum_k (d_k^* - o_k^*) \frac{\partial o_k}{\partial a_i^{(L-1)}}, \tag{9.57}$$

where

$$o_k = \sigma\left(z_k^{(L)}\right) = \sigma\left(\sum_{l=1}^{N_{L-1}} w_{kl}^{(L-1)} a_i^{(L-1)} + b_k^{(L-1)}\right). \tag{9.58}$$

Hence

$$\frac{\partial \mathcal{L}(n)}{\partial a_i^{(L-1)}} = -\sum_k (d_k^* - o_k^*)\sigma'\left(z_k^{(L)}\right) w_{ki}^{(L-1)}. \tag{9.59.1}$$

Similarly,

$$\frac{\partial \mathcal{L}(n)}{\partial a_i^{*(L-1)}} = -\sum_k (d_k - o_k)\sigma'\left(z_k^{*(L)}\right) w_{ki}^{*(L-1)}. \tag{9.59.2}$$

Substituting (9.59.1) and (9.59.2) into (9.56) we have

$$\begin{aligned}\frac{\partial \mathcal{L}(n)}{\partial w_{ij}^{\Re,(L-2)}} &= \left[-\sum_k (d_k^* - o_k^*)\sigma'\left(z_k^{(L)}\right) w_{ki}^{(L-1)}\right]\sigma'\left(z_i^{(L-1)}\right)a_j^{(L-2)} \\ &+ \left[-\sum_k (d_k - o_k)\sigma'\left(z_k^{*(L)}\right) w_{ki}^{*(L-1)}\right]\sigma'\left(z_i^{*(L-1)}\right)a_j^{*(L-2)}.\end{aligned} \tag{9.60.1}$$

The partial derivative with respect to the imaginary part is then

$$\begin{aligned}\frac{\partial \mathcal{L}(n)}{\partial w_{ij}^{\Im,(L-2)}} &= i\left[-\sum_k (d_k^* - o_k^*)\sigma'\left(z_k^{(L)}\right) w_{ki}^{(L-1)}\right]\sigma'\left(z_i^{(L-1)}\right)a_j^{(L-2)} \\ &- i\left[-\sum_k (d_k - o_k)\sigma'\left(z_k^{*(L)}\right) w_{ki}^{*(L-1)}\right]\sigma'\left(z_i^{*(L-1)}\right)a_j^{*(L-2)}.\end{aligned} \tag{9.60.2}$$

Combining (9.60.1) and (9.60.2) into complex form yields the adaptation rule

$$w_{ij}^{(L-2)}(n+1) = w_{ij}^{(L-2)}(n) + \alpha\left[\sum_k (d_k - o_k)\sigma'\left(z_k^{*(L)}\right)w_{ki}^{*(L-1)}\right]\sigma'\left(z_i^{*(L-1)}\right)a_j^{*(L-2)}. \tag{9.61}$$

Equations (9.61) and (9.55) are the two basic update equations for the BP algorithm. The logic of deriving the update formula for other hidden units is exactly the same.

**Case 2:**

**The derivation is based on the assumption that the partial derivatives $\frac{\partial u}{\partial x}, \frac{\partial u}{\partial y}, \frac{\partial v}{\partial x}$, and $\frac{\partial v}{\partial y}$ of the AF $\sigma(z) = u(x,y) + iv(x,y)$ exists. This assumption is not sufficient that $\sigma'(z)$ exists.**

Let the neuron AF be

$$\sigma(z) = u(x,y) + iv(x,y), \qquad z = x + iy. \tag{9.62}$$

The functions $u$ and $v$ are the real and imaginary parts of $\sigma$, respectively. Likewise, $x$ and $y$ are the real and imaginary parts of $z$. For now, it is sufficient to assume that the partial derivatives $u_x = \partial u/\partial x$, $u_y = \partial u/\partial y$, $v_x = \partial v/\partial x$, and $v_y = \partial v/\partial y$ exist for all $z \in \mathbb{C}$. This assumption is not sufficient that $\sigma'(z)$ exists. If it exists, the partial derivatives also have to satisfy the Cauchy-Riemann equations.

The error signal $\epsilon_k$, required for adaptation is defined as the difference between the desired output at the $k$-th neuron of the output layer, $d_k$, and the actual output of the $k$-th output neuron, $o_k$:

$$\epsilon_k = d_k - o_k. \tag{9.63}$$

Hence, the sum of error squares produced by the network is

$$\begin{aligned}\mathcal{L} &= \frac{1}{2}\sum_{k=1}|\epsilon_k|^2 \\ &= \frac{1}{2}\sum_{k=1}\epsilon_k\epsilon_k^* \\ &= \frac{1}{2}\sum_{k=1}(d_k - o_k)(d_k - o_k)^* \\ &= \frac{1}{2}\sum_{k=1}(d_k - o_k)(d_k^* - o_k^*) \\ &= \frac{1}{2}\sum_{k=1} d_k d_k^* - o_k d_k^* - d_k o_k^* + o_k o_k^* \\ &= \frac{1}{2}\sum_{k=1}(d_k d_k^* + o_k o_k^*) - (o_k d_k^* + d_k o_k^*).\end{aligned} \tag{9.64}$$

It should be noted that the error $\mathcal{L}$ is a real scalar function. In what follows, the subscripts $\Re$ and $\Im$ indicate the real and imaginary parts, respectively. The output $o_j$ of neuron $j$ in the network is [273]

$$o_j = \sigma(z_j) = u_j + iv_j, \tag{9.65}$$

and

$$z_j = x_j + iy_j = \sum_{l=1} w_{jl}a_{jl}, \tag{9.66.1}$$

where the $w_{jl}$'s are the (complex) weights of neuron $j$

$$w_{jl} = w_{jl}^{\Re} + iw_{jl}^{\Im}, \tag{9.66.2}$$

and $a_{jl}$'s their corresponding (complex) inputs

$$a_{jl} = a_{jl}^{\Re} + ia_{jl}^{\Im}. \tag{9.66.3}$$

Hence,

$$\begin{aligned}z_j &= x_j + iy_j \\ &= \sum_{l=1}(w_{jl}^{\Re} + iw_{jl}^{\Im})(a_{jl}^{\Re} + ia_{jl}^{\Im}) \\ &= \sum_{l=1} w_{jl}^{\Re}a_{jl}^{\Re} + iw_{jl}^{\Im}a_{jl}^{\Re} + iw_{jl}^{\Re}a_{jl}^{\Im} - w_{jl}^{\Im}a_{jl}^{\Im} \\ &= \sum_{l=1}(w_{jl}^{\Re}a_{jl}^{\Re} - w_{jl}^{\Im}a_{jl}^{\Im}) + i(w_{jl}^{\Im}a_{jl}^{\Re} + w_{jl}^{\Re}a_{jl}^{\Im}).\end{aligned} \tag{9.67}$$

We get,

$$x_j = \sum_{l=1} w_{jl}^{\Re}a_{jl}^{\Re} - w_{jl}^{\Im}a_{jl}^{\Im}, \tag{9.68.1}$$

$$y_j = \sum_{l=1} w_{jl}^{\Im}a_{jl}^{\Re} + w_{jl}^{\Re}a_{jl}^{\Im}. \tag{9.68.2}$$

A bias weight, having permanent input (1,0), may also be added to each neuron. We note the following partial derivatives:

$$\frac{\partial x_j}{\partial w_{jl}^{\Re}} = a_{jl}^{\Re}, \quad \frac{\partial y_j}{\partial w_{jl}^{\Re}} = a_{jl}^{\Im}, \quad \frac{\partial x_j}{\partial w_{jl}^{\Im}} = -a_{jl}^{\Im}, \quad \frac{\partial y_j}{\partial w_{jl}^{\Im}} = a_{jl}^{\Re}. \tag{9.69}$$

In order to use the chain rule to find the gradient of the error function $\mathcal{L}$ with respect to the real part of $w_{jl}$, we have to observe the variable dependencies: The real function $\mathcal{L}$ is a function of both $u_j(x_j, y_j)$ and $v_j(x_j, y_j)$, and $x_j$ and $y_j$ are both functions of $w_{jl}^{\Re}$ (and $w_{jl}^{\Im}$). Thus, the gradient of the error function with respect to the real part of $w_{jl}$ can be written as

$$\begin{aligned}\frac{\partial \mathcal{L}}{\partial w_{jl}^{\Re}} &= \frac{\partial \mathcal{L}}{\partial u_j}\left(\frac{\partial u_j}{\partial x_j}\frac{\partial x_j}{\partial w_{jl}^{\Re}} + \frac{\partial u_j}{\partial y_j}\frac{\partial y_j}{\partial w_{jl}^{\Re}}\right) + \frac{\partial \mathcal{L}}{\partial v_j}\left(\frac{\partial v_j}{\partial x_j}\frac{\partial x_j}{\partial w_{jl}^{\Re}} + \frac{\partial v_j}{\partial y_j}\frac{\partial y_j}{\partial w_{jl}^{\Re}}\right)\\ &= \frac{\partial \mathcal{L}}{\partial u_j}\left(\frac{\partial u_j}{\partial x_j}a_{jl}^{\Re} + \frac{\partial u_j}{\partial y_j}a_{jl}^{\Im}\right) + \frac{\partial \mathcal{L}}{\partial v_j}\left(\frac{\partial v_j}{\partial x_j}a_{jl}^{\Re} + \frac{\partial v_j}{\partial y_j}a_{jl}^{\Im}\right)\\ &= -\delta_j^{\Re}\left(\frac{\partial u_j}{\partial x_j}a_{jl}^{\Re} + \frac{\partial u_j}{\partial y_j}a_{jl}^{\Im}\right) - \delta_j^{\Im}\left(\frac{\partial v_j}{\partial x_j}a_{jl}^{\Re} + \frac{\partial v_j}{\partial y_j}a_{jl}^{\Im}\right),\end{aligned} \tag{9.70}$$

with $\delta_j^{\Re} = -\partial\mathcal{L}/\partial u_j$ and $\delta_j^{\Im} = -\partial\mathcal{L}/\partial v_j$ and consequently

$$\begin{aligned}\delta_j &= \delta_j^{\Re} + i\delta_j^{\Im}\\ &= -\frac{\partial \mathcal{L}}{\partial u_j} - i\frac{\partial \mathcal{L}}{\partial v_j}.\end{aligned} \tag{9.71}$$

Likewise, the gradient of the error function with respect to the imaginary part of $w_{jl}$ is

$$\begin{aligned}\frac{\partial \mathcal{L}}{\partial w_{jl}^{\Im}} &= \frac{\partial \mathcal{L}}{\partial u_j}\left(\frac{\partial u_j}{\partial x_j}\frac{\partial x_j}{\partial w_{jl}^{\Im}} + \frac{\partial u_j}{\partial y_j}\frac{\partial y_j}{\partial w_{jl}^{\Im}}\right) + \frac{\partial \mathcal{L}}{\partial v_j}\left(\frac{\partial v_j}{\partial x_j}\frac{\partial x_j}{\partial w_{jl}^{\Im}} + \frac{\partial v_j}{\partial y_j}\frac{\partial y_j}{\partial w_{jl}^{\Im}}\right)\\ &= \frac{\partial \mathcal{L}}{\partial u_j}\left(\frac{\partial u_j}{\partial x_j}(-a_{jl}^{\Im}) + \frac{\partial u_j}{\partial y_j}a_{jl}^{\Re}\right) + \frac{\partial \mathcal{L}}{\partial v_j}\left(\frac{\partial v_j}{\partial x_j}(-a_{jl}^{\Im}) + \frac{\partial v_j}{\partial y_j}a_{jl}^{\Re}\right)\\ &= -\delta_j^{\Re}\left(\frac{\partial u_j}{\partial x_j}(-a_{jl}^{\Im}) + \frac{\partial u_j}{\partial y_j}a_{jl}^{\Re}\right) - \delta_j^{\Im}\left(\frac{\partial v_j}{\partial x_j}(-a_{jl}^{\Im}) + \frac{\partial v_j}{\partial y_j}a_{jl}^{\Re}\right).\end{aligned} \tag{9.72}$$

Combining (9.70) and (9.72), we can write the gradient of the error function $\mathcal{L}$ with respect to the complex weight $w_{jl}$ as [273]

$$\begin{aligned}\nabla_{w_{jl}}\mathcal{L} &= \frac{\partial \mathcal{L}}{\partial w_{jl}^{\Re}} + i\frac{\partial \mathcal{L}}{\partial w_{jl}^{\Im}}\\ &= -\delta_j^{\Re}\left(\frac{\partial u_j}{\partial x_j}a_{jl}^{\Re} + \frac{\partial u_j}{\partial y_j}a_{jl}^{\Im}\right) - \delta_j^{\Im}\left(\frac{\partial v_j}{\partial x_j}a_{jl}^{\Re} + \frac{\partial v_j}{\partial y_j}a_{jl}^{\Im}\right)\\ &\quad -i\delta_j^{\Re}\left(\frac{\partial u_j}{\partial x_j}(-a_{jl}^{\Im}) + \frac{\partial u_j}{\partial y_j}a_{jl}^{\Re}\right) - i\delta_j^{\Im}\left(\frac{\partial v_j}{\partial x_j}(-a_{jl}^{\Im}) + \frac{\partial v_j}{\partial y_j}a_{jl}^{\Re}\right)\\ &= -\delta_j^{\Re}\frac{\partial u_j}{\partial x_j}a_{jl}^{\Re} - \delta_j^{\Re}\frac{\partial u_j}{\partial y_j}a_{jl}^{\Im} - \delta_j^{\Im}\frac{\partial v_j}{\partial x_j}a_{jl}^{\Re} - \delta_j^{\Im}\frac{\partial v_j}{\partial y_j}a_{jl}^{\Im}\\ &\quad +i\delta_j^{\Re}\frac{\partial u_j}{\partial x_j}a_{jl}^{\Im} - i\delta_j^{\Re}\frac{\partial u_j}{\partial y_j}a_{jl}^{\Re} + i\delta_j^{\Im}\frac{\partial v_j}{\partial x_j}a_{jl}^{\Im} - i\delta_j^{\Im}\frac{\partial v_j}{\partial y_j}a_{jl}^{\Re}\\ &= \delta_j^{\Re}\left(-\frac{\partial u_j}{\partial x_j}a_{jl}^{\Re} - \frac{\partial u_j}{\partial y_j}a_{jl}^{\Im} + i\frac{\partial u_j}{\partial x_j}a_{jl}^{\Im} - i\frac{\partial u_j}{\partial y_j}a_{jl}^{\Re}\right)\\ &\quad +\delta_j^{\Im}\left(-\frac{\partial v_j}{\partial x_j}a_{jl}^{\Re} - \frac{\partial v_j}{\partial y_j}a_{jl}^{\Im} + i\frac{\partial v_j}{\partial x_j}a_{jl}^{\Im} - i\frac{\partial v_j}{\partial y_j}a_{jl}^{\Re}\right)\\ &= \delta_j^{\Re}\Theta_1 + \delta_j^{\Im}\Theta_2,\end{aligned}$$

i.e.,

$$\nabla_{w_{jl}}\mathcal{L} = \delta_j^{\Re}\Theta_1 + \delta_j^{\Im}\Theta_2, \tag{9.73.1}$$

where,

$$\begin{aligned}\Theta_1 &= \left(-\frac{\partial u_j}{\partial x_j}a_{jl}^{\Re} - \frac{\partial u_j}{\partial y_j}a_{jl}^{\Im} + i\frac{\partial u_j}{\partial x_j}a_{jl}^{\Im} - i\frac{\partial u_j}{\partial y_j}a_{jl}^{\Re}\right)\\ &= \left(-\left[\frac{\partial u_j}{\partial x_j} + i\frac{\partial u_j}{\partial y_j}\right]a_{jl}^{\Re} + \frac{1}{i}\left[-i\frac{\partial u_j}{\partial y_j} - \frac{\partial u_j}{\partial x_j}\right]a_{jl}^{\Im}\right)\\ &= \left(-\left[\frac{\partial u_j}{\partial x_j} + i\frac{\partial u_j}{\partial y_j}\right]a_{jl}^{\Re} - \frac{1}{i}\left[\frac{\partial u_j}{\partial x_j} + i\frac{\partial u_j}{\partial y_j}\right]a_{jl}^{\Im}\right)\end{aligned}$$

$$= \left[\frac{\partial u_j}{\partial x_j} + i\frac{\partial u_j}{\partial y_j}\right]\left(-a_{jl}^{\Re} + i a_{jl}^{\Im}\right)$$
$$= -\left[\frac{\partial u_j}{\partial x_j} + i\frac{\partial u_j}{\partial y_j}\right]\left(a_{jl}^{\Re} - i a_{jl}^{\Im}\right)$$
$$= -\left[\frac{\partial u_j}{\partial x_j} + i\frac{\partial u_j}{\partial y_j}\right]\bar{a}_{jl},$$

i.e.,

$$\Theta_1 = -\left[\frac{\partial u_j}{\partial x_j} + i\frac{\partial u_j}{\partial y_j}\right]\bar{a}_{jl}, \tag{9.73.2}$$

and

$$\Theta_2 = \left(-\frac{\partial v_j}{\partial x_j}a_{jl}^{\Re} - \frac{\partial v_j}{\partial y_j}a_{jl}^{\Im} + i\frac{\partial v_j}{\partial x_j}a_{jl}^{\Im} - i\frac{\partial v_j}{\partial y_j}a_{jl}^{\Re}\right)$$
$$= \left(-\left[\frac{\partial v_j}{\partial x_j} + i\frac{\partial v_j}{\partial y_j}\right]a_{jl}^{\Re} + \frac{1}{i}\left[-i\frac{\partial v_j}{\partial y_j} - \frac{\partial v_j}{\partial x_j}\right]a_{jl}^{\Im}\right)$$
$$= \left(-\left[\frac{\partial v_j}{\partial x_j} + i\frac{\partial v_j}{\partial y_j}\right]a_{jl}^{\Re} - \frac{1}{i}\left[\frac{\partial v_j}{\partial x_j} + i\frac{\partial v_j}{\partial y_j}\right]a_{jl}^{\Im}\right)$$
$$= -\left[\frac{\partial v_j}{\partial x_j} + i\frac{\partial v_j}{\partial y_j}\right]\left(a_{jl}^{\Re} + \frac{1}{i}a_{jl}^{\Im}\right)$$
$$= -\left[\frac{\partial v_j}{\partial x_j} + i\frac{\partial v_j}{\partial y_j}\right]\left(a_{jl}^{\Re} - i a_{jl}^{\Im}\right)$$
$$= -\left[\frac{\partial v_j}{\partial x_j} + i\frac{\partial v_j}{\partial y_j}\right]\bar{a}_{jl}.$$

i.e.,

$$\Theta_2 = -\left[\frac{\partial v_j}{\partial x_j} + i\frac{\partial v_j}{\partial y_j}\right]\bar{a}_{jl}. \tag{9.73.3}$$

Finally, the gradient of the error function $\mathcal{L}$ with respect to the complex weight becomes

$$\nabla_{w_{jl}}\mathcal{L} = -\left[\frac{\partial u_j}{\partial x_j} + i\frac{\partial u_j}{\partial y_j}\right]\bar{a}_{jl}\delta_j^{\Re} - \left[\frac{\partial v_j}{\partial x_j} + i\frac{\partial v_j}{\partial y_j}\right]\bar{a}_{jl}\delta_j^{\Im}$$
$$= -\bar{a}_{jl}\left\{\left[\frac{\partial u_j}{\partial x_j} + i\frac{\partial u_j}{\partial y_j}\right]\delta_j^{\Re} + \left[\frac{\partial v_j}{\partial x_j} + i\frac{\partial v_j}{\partial y_j}\right]\delta_j^{\Im}\right\}. \tag{9.74}$$

To minimize the error $\mathcal{L}$, each complex weight $w_{jl}$ should be changed by a quantity $\Delta w_{jl}$ proportional to the negative gradient:

$$\Delta w_{jl} = w_{jl,\text{new}} - w_{jl,\text{old}}$$
$$= -\alpha\nabla_{w_{jl}}\mathcal{L}$$
$$= \alpha\bar{a}_{jl}\left\{\left[\frac{\partial u_j}{\partial x_j} + i\frac{\partial u_j}{\partial y_j}\right]\delta_j^{\Re} + \left[\frac{\partial v_j}{\partial x_j} + i\frac{\partial v_j}{\partial y_j}\right]\delta_j^{\Im}\right\}, \tag{9.75}$$

where $\alpha$ is the learning rate, a real positive constant. A momentum term may be added to the above learning equation. When the weight $w_{jl}$ belongs to an output neuron, then $\delta_j^{\Re}$, and $\delta_j^{\Im}$ in (9.74) have the values

$$\delta_j^{\Re} = -\frac{\partial\mathcal{L}}{\partial u_j} = \epsilon_j^{\Re} = d_j^{\Re} - u_j \quad \text{and} \quad \delta_j^{\Im} = -\frac{\partial\mathcal{L}}{\partial v_j} = \epsilon_j^{\Im} = d_j^{\Im} - v_j, \tag{9.76}$$

or more compactly:

$$\begin{aligned}\delta_j &= \epsilon_j \\ &= \epsilon_j^{\Re} + i\epsilon_j^{\Im} \\ &= d_j^{\Re} - u_j + i\left(d_j^{\Im} - v_j\right) \\ &= \left(d_j^{\Re} + i d_j^{\Im}\right) - \left(u_j + i v_j\right) \\ &= d_j - o_j. \end{aligned} \tag{9.77}$$

When weight $w_{jl}$ belongs to a hidden neuron, i.e., when the output of the neuron is fed to other neurons in subsequent layers, in order to compute $\delta_j$, or equivalently $\delta_j^{\Re}$ and $\delta_j^{\Im}$, we have to use the chain rule. Let the index $k$ indicate a neuron that receives input from neuron $j$. Then the net input $z_k$ to neuron $k$ is

$$\begin{aligned} z_k &= x_k + i y_k \\ &= \sum_l (u_l + i v_l)\left(w_{kl}^{\Re} + i w_{kl}^{\Im}\right) \\ &= \sum_l \left(u_l w_{kl}^{\Re} - v_l w_{kl}^{\Im}\right) + i\left(v_l w_{kl}^{\Re} + u_l w_{kl}^{\Im}\right), \end{aligned} \tag{9.78}$$

where the index $l$ runs through the neurons from which neuron $k$ receives input. Hence

$$x_k = \sum_l u_l w_{kl}^{\Re} - v_l w_{kl}^{\Im}, \tag{9.79.1}$$

$$y_k = \sum_l v_l w_{kl}^{\Re} + u_l w_{kl}^{\Im}. \tag{9.79.2}$$

Thus, we have the following partial derivatives:

$$\frac{\partial x_k}{\partial u_j} = w_{kj}^{\Re}, \qquad \frac{\partial y_k}{\partial u_j} = w_{kj}^{\Im}, \qquad \frac{\partial x_k}{\partial v_j} = -w_{kj}^{\Im}, \qquad \frac{\partial y_k}{\partial v_j} = w_{kj}^{\Re}. \tag{9.80}$$

Using the chain rule we compute $\delta_j^{\Re}$ [273]:

$$\begin{aligned} \delta_j^{\Re} &= -\frac{\partial \mathcal{L}}{\partial u_j} \\ &= -\sum_k \frac{\partial \mathcal{L}}{\partial u_k}\left(\frac{\partial u_k}{\partial x_k}\frac{\partial x_k}{\partial u_j} + \frac{\partial u_k}{\partial y_k}\frac{\partial y_k}{\partial u_j}\right) - \sum_k \frac{\partial \mathcal{L}}{\partial v_k}\left(\frac{\partial v_k}{\partial x_k}\frac{\partial x_k}{\partial u_j} + \frac{\partial v_k}{\partial y_k}\frac{\partial y_k}{\partial u_j}\right) \\ &= -\sum_k \frac{\partial \mathcal{L}}{\partial u_k}\left(\frac{\partial u_k}{\partial x_k} w_{kj}^{\Re} + \frac{\partial u_k}{\partial y_k} w_{kj}^{\Im}\right) - \sum_k \frac{\partial \mathcal{L}}{\partial v_k}\left(\frac{\partial v_k}{\partial x_k} w_{kj}^{\Re} + \frac{\partial v_k}{\partial y_k} w_{kj}^{\Im}\right) \\ &= \sum_k \delta_k^{\Re}\left(\frac{\partial u_k}{\partial x_k} w_{kj}^{\Re} + \frac{\partial u_k}{\partial y_k} w_{kj}^{\Im}\right) + \sum_k \delta_k^{\Im}\left(\frac{\partial v_k}{\partial x_k} w_{kj}^{\Re} + \frac{\partial v_k}{\partial y_k} w_{kj}^{\Im}\right), \end{aligned} \tag{9.81}$$

where the index $k$ runs through the neurons that receive input from neuron $j$. In a similar manner, we can compute $\delta_j^{\Im}$:

$$\begin{aligned} \delta_j^{\Im} &= -\frac{\partial \mathcal{L}}{\partial v_j} \\ &= -\sum_k \frac{\partial \mathcal{L}}{\partial u_k}\left(\frac{\partial u_k}{\partial x_k}\frac{\partial x_k}{\partial v_j} + \frac{\partial u_k}{\partial y_k}\frac{\partial y_k}{\partial v_j}\right) - \sum_k \frac{\partial \mathcal{L}}{\partial v_k}\left(\frac{\partial v_k}{\partial x_k}\frac{\partial x_k}{\partial v_j} + \frac{\partial v_k}{\partial y_k}\frac{\partial y_k}{\partial v_j}\right) \\ &= -\sum_k \frac{\partial \mathcal{L}}{\partial u_k}\left(\frac{\partial u_k}{\partial x_k}\left(-w_{kj}^{\Im}\right) + \frac{\partial u_k}{\partial y_k} w_{kj}^{\Re}\right) - \sum_k \frac{\partial \mathcal{L}}{\partial v_k}\left(\frac{\partial v_k}{\partial x_k}\left(-w_{kj}^{\Im}\right) + \frac{\partial v_k}{\partial y_k} w_{kj}^{\Re}\right) \\ &= \sum_k \delta_k^{\Re}\left(\frac{\partial u_k}{\partial x_k}\left(-w_{kj}^{\Im}\right) + \frac{\partial u_k}{\partial y_k} w_{kj}^{\Re}\right) + \sum_k \delta_k^{\Im}\left(\frac{\partial v_k}{\partial x_k}\left(-w_{kj}^{\Im}\right) + \frac{\partial v_k}{\partial y_k} w_{kj}^{\Re}\right). \end{aligned} \tag{9.82}$$

Combining (9.81) and (9.82), we arrive at the following expression for $\delta_j$:

$$\delta_j = \delta_j^{\Re} + i\delta_j^{\Im}$$

$$= \sum_k \delta_k^{\Re}\left(\frac{\partial u_k}{\partial x_k} w_{kj}^{\Re} + \frac{\partial u_k}{\partial y_k} w_{kj}^{\Im}\right) + \sum_k \delta_k^{\Im}\left(\frac{\partial v_k}{\partial x_k} w_{kj}^{\Re} + \frac{\partial v_k}{\partial y_k} w_{kj}^{\Im}\right)$$

$$+i \sum_k \delta_k^{\Re}\left(\frac{\partial u_k}{\partial x_k}(-w_{kj}^{\Im}) + \frac{\partial u_k}{\partial y_k} w_{kj}^{\Re}\right) + i \sum_k \delta_k^{\Im}\left(\frac{\partial v_k}{\partial x_k}(-w_{kj}^{\Im}) + \frac{\partial v_k}{\partial y_k} w_{kj}^{\Re}\right)$$

$$= \sum_k \left(\delta_k^{\Re} \frac{\partial u_k}{\partial x_k} w_{kj}^{\Re} + \delta_k^{\Re} \frac{\partial u_k}{\partial y_k} w_{kj}^{\Im} + \delta_k^{\Im} \frac{\partial v_k}{\partial x_k} w_{kj}^{\Re} + \delta_k^{\Im} \frac{\partial v_k}{\partial y_k} w_{kj}^{\Im}\right)$$

$$+ \sum_k \left(-i\delta_k^{\Re} \frac{\partial u_k}{\partial x_k} w_{kj}^{\Im} + i\delta_k^{\Re} \frac{\partial u_k}{\partial y_k} w_{kj}^{\Re} - i\delta_k^{\Im} \frac{\partial v_k}{\partial x_k} w_{kj}^{\Im} + i\delta_k^{\Im} \frac{\partial v_k}{\partial y_k} w_{kj}^{\Re}\right)$$

$$= \sum_k \left(\left[\delta_k^{\Re} \frac{\partial u_k}{\partial x_k} w_{kj}^{\Re} - i\delta_k^{\Re} \frac{\partial u_k}{\partial x_k} w_{kj}^{\Im}\right] + \left[\delta_k^{\Re} \frac{\partial u_k}{\partial y_k} w_{kj}^{\Im} + i\delta_k^{\Re} \frac{\partial u_k}{\partial y_k} w_{kj}^{\Re}\right] + \left[\delta_k^{\Im} \frac{\partial v_k}{\partial x_k} w_{kj}^{\Re} - i\delta_k^{\Im} \frac{\partial v_k}{\partial x_k} w_{kj}^{\Im}\right]\right.$$

$$\left. + \left[\delta_k^{\Im} \frac{\partial v_k}{\partial y_k} w_{kj}^{\Im} + i\delta_k^{\Im} \frac{\partial v_k}{\partial y_k} w_{kj}^{\Re}\right]\right)$$

$$= \sum_k \left(\delta_k^{\Re} \frac{\partial u_k}{\partial x_k}\left[w_{kj}^{\Re} - i w_{kj}^{\Im}\right] + \delta_k^{\Re} \frac{\partial u_k}{\partial y_k}\left[w_{kj}^{\Im} + i w_{kj}^{\Re}\right] + \delta_k^{\Im} \frac{\partial v_k}{\partial x_k}\left[w_{kj}^{\Re} - i w_{kj}^{\Im}\right] + \delta_k^{\Im} \frac{\partial v_k}{\partial y_k}\left[w_{kj}^{\Im} + i w_{kj}^{\Re}\right]\right)$$

$$= \sum_k \left(\delta_k^{\Re} \frac{\partial u_k}{\partial x_k}\left[w_{kj}^{\Re} - i w_{kj}^{\Im}\right] + i\delta_k^{\Re} \frac{\partial u_k}{\partial y_k}\left[w_{kj}^{\Re} - i w_{kj}^{\Im}\right] + \delta_k^{\Im} \frac{\partial v_k}{\partial x_k}\left[w_{kj}^{\Re} - i w_{kj}^{\Im}\right] + i\delta_k^{\Im} \frac{\partial v_k}{\partial y_k}\left[w_{kj}^{\Re} - i w_{kj}^{\Im}\right]\right)$$

$$= \sum_k \left[w_{kj}^{\Re} - i w_{kj}^{\Im}\right]\left(\left[\frac{\partial u_k}{\partial x_k} + i \frac{\partial u_k}{\partial y_k}\right]\delta_k^{\Re} + \left[\frac{\partial v_k}{\partial x_k} + i \frac{\partial v_k}{\partial y_k}\right]\delta_k^{\Im}\right)$$

$$= \sum_k \bar{w}_{kj}\left(\left[\frac{\partial u_k}{\partial x_k} + i \frac{\partial u_k}{\partial y_k}\right]\delta_k^{\Re} + \left[\frac{\partial v_k}{\partial x_k} + i \frac{\partial v_k}{\partial y_k}\right]\delta_k^{\Im}\right),$$

$$\delta_j = \sum_k \bar{w}_{kj}\left(\left[\frac{\partial u_k}{\partial x_k} + i \frac{\partial u_k}{\partial y_k}\right]\delta_k^{\Re} + \left[\frac{\partial v_k}{\partial x_k} + i \frac{\partial v_k}{\partial y_k}\right]\delta_k^{\Im}\right). \tag{9.83}$$

Training of a feed-forward network with the complex BP algorithm is done in a similar manner as in the usual real BP.

- First, the weights are initialized to small random complex values.
- Until an acceptable output error level is arrived at, each input vector is presented to the network.
- The corresponding output and output error are calculated (forward pass).
- Then the error is backpropagated to each neuron in the network and the weights are adjusted accordingly (backward pass).
- More precisely, for the input pattern $a_j$, $\delta_j$ for neuron $j$ is computed by starting at the neurons in the output layer using (9.77) and then for neurons in hidden layers by recursively using (9.83). As soon as $\delta_j$ is computed for neuron $j$, its weights are changed according to (9.75).

### Case 3:

### Cauchy-Riemann equations and the fully complex backpropagation algorithm

Cauchy-Riemann equations can be used to simplify the fully complex backpropagation algorithm derived in case 2. Cauchy-Riemann equations are the necessary condition for a complex function to be analytic at a point $z \in \mathbb{C}$ and can be written by noting that the partial derivatives of $\sigma(z) = u(x, y) + iv(x, y)$, where $z = x + iy$, should be equal along the real and imaginary axes:

$$\sigma'(z) = \frac{\partial u}{\partial x} + i \frac{\partial v}{\partial x} = \frac{\partial v}{\partial y} - i \frac{\partial u}{\partial y}. \tag{9.84}$$

Equating the real and imaginary parts in (9.84), we obtain the Cauchy-Riemann equations: $\frac{\partial u}{\partial x} = \frac{\partial v}{\partial y}, \frac{\partial v}{\partial x} = -\frac{\partial u}{\partial y}$. Also note that this enables (9.84) to be expressed more concisely as (using Wirtinger derivatives) [274]

$$\begin{aligned}
\sigma'(z) &= \frac{\partial \sigma}{\partial z} \\
&= \frac{1}{2}\left(\frac{\partial \sigma}{\partial x} - i\frac{\partial \sigma}{\partial y}\right) \\
&= \frac{1}{2}\left(\frac{\partial}{\partial x}[u(x,y) + iv(x,y)] - i\frac{\partial}{\partial y}[u(x,y) + iv(x,y)]\right) \\
&= \frac{1}{2}\left(\left[\frac{\partial}{\partial x}u(x,y) + i\frac{\partial}{\partial x}v(x,y)\right] - i\left[\frac{\partial}{\partial y}u(x,y) + i\frac{\partial}{\partial y}v(x,y)\right]\right) \\
&= \frac{1}{2}\left(\left[\frac{\partial}{\partial x}u(x,y) + i\frac{\partial}{\partial x}v(x,y)\right] - i\left[-\frac{\partial}{\partial x}v(x,y) + i\frac{\partial}{\partial x}u(x,y)\right]\right) \\
&= \frac{1}{2}\left(\frac{\partial}{\partial x}u(x,y) + i\frac{\partial}{\partial x}v(x,y) + i\frac{\partial}{\partial x}v(x,y) + \frac{\partial}{\partial x}u(x,y)\right) \\
&= \frac{1}{2}\left(2\frac{\partial}{\partial x}u(x,y) + 2i\frac{\partial}{\partial x}v(x,y)\right) \\
&= \frac{\partial}{\partial x}\big(u(x,y) + iv(x,y)\big) = \frac{\partial \sigma}{\partial x},
\end{aligned}$$

and

$$\begin{aligned}
\sigma'(z) &= \frac{\partial \sigma}{\partial z} \\
&= \frac{1}{2}\left(\frac{\partial \sigma}{\partial x} - i\frac{\partial \sigma}{\partial y}\right) \\
&= \frac{1}{2}\left(\frac{\partial}{\partial x}[u(x,y) + iv(x,y)] - i\frac{\partial}{\partial y}[u(x,y) + iv(x,y)]\right) \\
&= \frac{1}{2}\left(\left[\frac{\partial}{\partial x}u(x,y) + i\frac{\partial}{\partial x}v(x,y)\right] - i\left[\frac{\partial}{\partial y}u(x,y) + i\frac{\partial}{\partial y}v(x,y)\right]\right) \\
&= \frac{1}{2}\left(\left[\frac{\partial}{\partial y}v(x,y) - i\frac{\partial}{\partial y}u(x,y)\right] - i\left[\frac{\partial}{\partial y}u(x,y) + i\frac{\partial}{\partial y}v(x,y)\right]\right) \\
&= \frac{1}{2}\left(\frac{\partial}{\partial y}v(x,y) - i\frac{\partial}{\partial y}u(x,y) - i\frac{\partial}{\partial y}u(x,y) + \frac{\partial}{\partial y}v(x,y)\right) \\
&= \frac{1}{2}\frac{\partial}{\partial y}\big(2v(x,y) - 2iu(x,y)\big) \\
&= -i\frac{\partial}{\partial y}\left(-\frac{1}{i}v(x,y) + u(x,y)\right) \\
&= -i\frac{\partial}{\partial y}\big(u(x,y) + iv(x,y)\big) = -i\frac{\partial \sigma}{\partial y}.
\end{aligned}$$

Hence, we have

$$\sigma'(z) = \frac{\partial \sigma}{\partial x} = -i\frac{\partial \sigma}{\partial y}. \tag{9.85.1}$$

Similarly,

$$\begin{aligned}
\bar{\sigma}'(z) &= \overline{\left(\frac{\partial \sigma}{\partial z}\right)} \\
&= \frac{1}{2}\left(\frac{\partial \sigma}{\partial x} - i\frac{\partial \sigma}{\partial y}\right)^* \\
&= \frac{1}{2}\left(\frac{\partial}{\partial x}[u(x,y) + iv(x,y)]^* + i\frac{\partial}{\partial y}[u(x,y) + iv(x,y)]^*\right)
\end{aligned}$$

$$
\begin{aligned}
&= \frac{1}{2}\left(\left[\frac{\partial}{\partial x}u(x,y) - i\frac{\partial}{\partial x}v(x,y)\right] + i\left[\frac{\partial}{\partial y}u(x,y) - i\frac{\partial}{\partial y}v(x,y)\right]\right) \\
&= \frac{1}{2}\left(\left[\frac{\partial}{\partial x}u(x,y) - i\frac{\partial}{\partial x}v(x,y)\right] + i\left[-\frac{\partial}{\partial x}v(x,y) - i\frac{\partial}{\partial x}u(x,y)\right]\right) \\
&= \frac{1}{2}\left(\frac{\partial}{\partial x}u(x,y) - i\frac{\partial}{\partial x}v(x,y) - i\frac{\partial}{\partial x}v(x,y) + \frac{\partial}{\partial x}u(x,y)\right) \\
&= \frac{1}{2}\left(2\frac{\partial}{\partial x}u(x,y) - 2i\frac{\partial}{\partial x}v(x,y)\right) \\
&= \frac{\partial}{\partial x}\big(u(x,y) - iv(x,y)\big) = \frac{\partial\bar{\sigma}}{\partial x},
\end{aligned}
$$

and

$$
\begin{aligned}
\bar{\sigma}'(z) &= \overline{\left(\frac{\partial\sigma}{\partial z}\right)} \\
&= \frac{1}{2}\left(\frac{\partial\sigma}{\partial x} - i\frac{\partial\sigma}{\partial y}\right)^{*} \\
&= \frac{1}{2}\left(\frac{\partial}{\partial x}[u(x,y) + iv(x,y)]^{*} + i\frac{\partial}{\partial y}[u(x,y) + iv(x,y)]^{*}\right) \\
&= \frac{1}{2}\left(\left[\frac{\partial}{\partial x}u(x,y) - i\frac{\partial}{\partial x}v(x,y)\right] + i\left[\frac{\partial}{\partial y}u(x,y) - i\frac{\partial}{\partial y}v(x,y)\right]\right) \\
&= \frac{1}{2}\left(\left[\frac{\partial}{\partial y}v(x,y) + i\frac{\partial}{\partial y}u(x,y)\right] + i\left[\frac{\partial}{\partial y}u(x,y) - i\frac{\partial}{\partial y}v(x,y)\right]\right) \\
&= \frac{1}{2}\left(\frac{\partial}{\partial y}v(x,y) + i\frac{\partial}{\partial y}u(x,y) + i\frac{\partial}{\partial y}u(x,y) + \frac{\partial}{\partial y}v(x,y)\right) \\
&= \frac{1}{2}\frac{\partial}{\partial y}\big(2v(x,y) + 2iu(x,y)\big) \\
&= i\frac{\partial}{\partial y}\left(\frac{1}{i}v(x,y) + u(x,y)\right) \\
&= i\frac{\partial}{\partial y}\big(u(x,y) - iv(x,y)\big) = i\frac{\partial\bar{\sigma}}{\partial y}.
\end{aligned}
$$

Hence, we have

$$
\bar{\sigma}'(z) = \frac{\partial\bar{\sigma}}{\partial x} = i\frac{\partial\bar{\sigma}}{\partial y}. \tag{9.85.2}
$$

Further applying the Cauchy-Riemann equations to (9.74), a more compact representation for the gradient of the error function is obtained using the simple derivative form given in (9.85.1)

$$
\begin{aligned}
\nabla_{w_{jl}}\mathcal{L} &= -\bar{a}_{jl}\left\{\left[\frac{\partial u_j}{\partial x_j} + i\frac{\partial u_j}{\partial y_j}\right]\delta_j^{\Re} + \left[\frac{\partial v_j}{\partial x_j} + i\frac{\partial v_j}{\partial y_j}\right]\delta_j^{\Im}\right\} \\
&= -\bar{a}_{jl}\left\{\left[\frac{\partial u_j}{\partial x_j} - i\frac{\partial v_j}{\partial x_j}\right]\delta_j^{\Re} + \left[-\frac{\partial u_j}{\partial y_j} + i\frac{\partial v_j}{\partial y_j}\right]\delta_j^{\Im}\right\} \\
&= -\bar{a}_{jl}\left\{\frac{\partial}{\partial x_j}[u_j - iv_j]\delta_j^{\Re} - \frac{\partial}{\partial y_j}[u_j - iv_j]\delta_j^{\Im}\right\} \\
&= -\bar{a}_{jl}\left\{\frac{\partial\bar{\sigma}}{\partial x_j}\delta_j^{\Re} - \frac{\partial\bar{\sigma}}{\partial y_j}\delta_j^{\Im}\right\} \\
&= -\bar{a}_{jl}\left\{\frac{\partial\bar{\sigma}}{\partial x_j}\delta_j^{\Re} + \left(i\frac{\partial\bar{\sigma}}{\partial y_j}\right)\left(i\delta_j^{\Im}\right)\right\}
\end{aligned}
$$

$$= -\bar{a}_{jl}\left\{\frac{\partial\bar{\sigma}}{\partial x_j}\delta_j^{\Re} + \frac{\partial\bar{\sigma}}{\partial x_j}\left(i\delta_j^{\Im}\right)\right\}$$
$$= -\bar{a}_{jl}\frac{\partial\bar{\sigma}}{\partial x_j}\{\delta_j^{\Re} + i\delta_j^{\Im}\}$$
$$= -\bar{a}_{jl}\bar{\sigma}'\delta_j,$$

where, $\bar{\sigma}' = \frac{\partial\bar{\sigma}}{\partial x_j} = i\frac{\partial\bar{\sigma}}{\partial y_j}$. So, we have [274]

$$\nabla_{w_{jl}}\mathcal{L} = -\bar{a}_{jl}\bar{\sigma}'(z)\delta_j. \tag{9.86}$$

Complex weight update $\Delta w_{jl}$ is proportional to the negative gradient:

$$\Delta w_{jl} = \alpha\bar{a}_{jl}\bar{\sigma}'(z)\delta_j, \tag{9.87}$$

where $\alpha$ is a real positive learning rate. When the complex weight belongs to an output neuron:

$$\delta_j = \epsilon_j = d_j - o_j. \tag{9.88}$$

When weight $w_{jl}$ belongs to a hidden neuron, i.e., when the output of the neuron is fed to other neurons in subsequent layers, in order to compute $\delta_j$, or equivalently $\delta_j^{\Re}$ and $\delta_j^{\Im}$, we have to use the chain rule. Let the index $k$ indicate a neuron that receives input from neuron $j$. Using (9.85.1) with (9.83), the following expression is obtained similarly for the weight update function (9.87) using (9.86), [274]:

$$\delta_j = \delta_j^{\Re} + i\delta_j^{\Im}$$
$$= \sum_k \bar{w}_{kj}\left(\left[\frac{\partial u_k}{\partial x_k} + i\frac{\partial u_k}{\partial y_k}\right]\delta_k^{\Re} + \left[\frac{\partial v_k}{\partial x_k} + i\frac{\partial v_k}{\partial y_k}\right]\delta_k^{\Im}\right)$$
$$= \sum_k \bar{w}_{kj}\left(\left[\frac{\partial u_k}{\partial x_k} - i\frac{\partial v_k}{\partial x_k}\right]\delta_k^{\Re} + \left[-\frac{\partial u_k}{\partial y_k} + i\frac{\partial v_k}{\partial y_k}\right]\delta_k^{\Im}\right)$$
$$= \sum_k \bar{w}_{kj}\left(\frac{\partial}{\partial x_k}[u_k - iv_k]\delta_k^{\Re} - \frac{\partial}{\partial y_k}[u_k - iv_k]\delta_k^{\Im}\right)$$
$$= \sum_k \bar{w}_{kj}\left(\frac{\partial\bar{\sigma}}{\partial x_k}\delta_k^{\Re} - \frac{\partial\bar{\sigma}}{\partial y_k}\delta_k^{\Im}\right)$$
$$= \sum_k \bar{w}_{kj}\left(\frac{\partial\bar{\sigma}}{\partial x_k}\delta_k^{\Re} + \left(i\frac{\partial\bar{\sigma}}{\partial y_k}\right)\left(i\delta_k^{\Im}\right)\right)$$
$$= \sum_k \bar{w}_{kj}\left(\frac{\partial\bar{\sigma}}{\partial x_k}\delta_k^{\Re} + \frac{\partial\bar{\sigma}}{\partial x_j}\left(i\delta_k^{\Im}\right)\right)$$
$$= \sum_k \bar{w}_{kj}\frac{\partial\bar{\sigma}}{\partial x_k}\left(\delta_k^{\Re} + i\delta_k^{\Im}\right) = \sum_k \bar{w}_{kj}\bar{\sigma}'\delta_k,$$

where, $\bar{\sigma}' = \frac{\partial\bar{\sigma}}{\partial x_j} = i\frac{\partial\bar{\sigma}}{\partial y_j}$. So, we have

$$\delta_j = \sum_k \bar{w}_{kj}\bar{\sigma}'(z)\delta_k. \tag{9.89}$$

Compared to the fully complex activation representation as $\sigma(z) = u(x,y) + iv(x,y)$, the split complex AF is a special case and can be represented as $\sigma(z) = u(x) + iv(y)$. This indicates that $u_y = v_x = 0$ for the split complex backpropagation algorithm. Removing these zero terms from (9.74), we obtain the following gradient and (complex) weight updates:

$$\nabla_{w_{jl}}\mathcal{L} = -\bar{a}_{jl}\left\{\left[\frac{\partial u_j}{\partial x_j} + i\frac{\partial u_j}{\partial y_j}\right]\delta_j^{\Re} + \left[\frac{\partial v_j}{\partial x_j} + i\frac{\partial v_j}{\partial y_j}\right]\delta_j^{\Im}\right\} = -\bar{a}_{jl}\left\{\frac{\partial u_j}{\partial x_j}\delta_j^{\Re} + i\frac{\partial v_j}{\partial y_j}\delta_j^{\Im}\right\}, \tag{9.90}$$

and

$$\Delta w_{jl} = \alpha \bar{a}_{jl} \left\{ \frac{\partial u_j}{\partial x_j} \delta_j^{\Re} + i \frac{\partial v_j}{\partial y_j} \delta_j^{\Im} \right\}. \tag{9.91}$$

As before, for output layer neurons, $\delta_j = \epsilon_j = d_j - o_j$. For the input and hidden layer, (9.83) becomes

$$\begin{aligned} \delta_j &= \sum_k \bar{w}_{kj} \left( \left[ \frac{\partial u_k}{\partial x_k} + i \frac{\partial u_k}{\partial y_k} \right] \delta_k^{\Re} + \left[ \frac{\partial v_k}{\partial x_k} + i \frac{\partial v_k}{\partial y_k} \right] \delta_k^{\Im} \right) \\ &= \sum_k \bar{w}_{kj} \left( \frac{\partial u_k}{\partial x_k} \delta_k^{\Re} + i \frac{\partial v_k}{\partial y_k} \delta_k^{\Im} \right). \end{aligned} \tag{9.92}$$

For further details on complex backpropagation algorithms, refer to [275-283].

## 9.4 Classification of Complex-Valued Neural Networks

CVNNs have emerged as a powerful tool in the realm of NN architectures, introducing a unique approach to processing complex-valued inputs. There are two types of CVNNs [284-287]: fully complex CVNNs and split-CVNNs.

**A. Fully-CVNNs**

To deal with complex-valued signals, fully-CVNNs have been proposed by [288] where both of the weights and AFs are in the complex domain. Fully CVNNs preserve the information carried by phase components. However, fully CVNNs increase the computation complexity and face difficulty in obtaining complex-differentiable (or holomorphic) AF [273].

**B. Split-CVNNs**

Split-CVNNs are commonly used to avoid the singularity problems in complex-valued functions and their derivatives. The fundamental idea behind split-CVNNs involves the division of complex-valued inputs, comprising real and imaginary parts, into distinct real-valued components. This split, occurring in either rectangular (real and imaginary) or polar coordinates (phase and magnitude), as shown in (9.93.1) and (9.93.2), respectively, forms the basis for subsequent processing in RVNNs.

$$\{z_1, z_2\} = \{(x_1, y_1), (x_2, y_2)\}, \tag{9.93.1}$$

$$\{z_1, z_2\} = \{(r_1, \varphi_1), (r_2, \varphi_2)\}, \tag{9.93.2}$$

where $z_1, z_2 \in \mathbb{C}$, $z = x + iy$, in which $x$ and $y$ are the real and imaginary components of the complex-valued input, respectively, while $r$ and $\varphi$ are the magnitude and phase components of the complex-valued input, respectively. However, split-CVNNs are categorized into two distinct types: those with both real-valued weights and a real-valued AF, and those with complex-valued weights and a real-valued AF.

**B.1. Split-CVNNs: Real-valued weights and real-valued AFs**

The complex inputs (containing both real and imaginary parts) and targets (desired outputs) are separated into two sets of real-valued data.

This separation leads to an increase in dimensionality. For instance, a problem with two complex-valued inputs and one complex-valued output becomes a problem with four real-valued inputs and two real-valued outputs (doubling the number of data points). This approach allows us to utilize existing RVNN architectures like the multi-layer perceptron with backpropagation. However, it's important to consider:

- Computational cost: This split-CVNN, while conceptually straightforward, presents challenges such as increased input dimension, learnable parameters, and network size. The increased dimensionality can lead to higher computational demands.
- Potential loss of information: Splitting complex data might discard some of the inherent relationships encoded within the complex numbers. It introduces phase distortion and accuracy issues during network updates, as real-valued gradients may not accurately represent the true complex gradient.

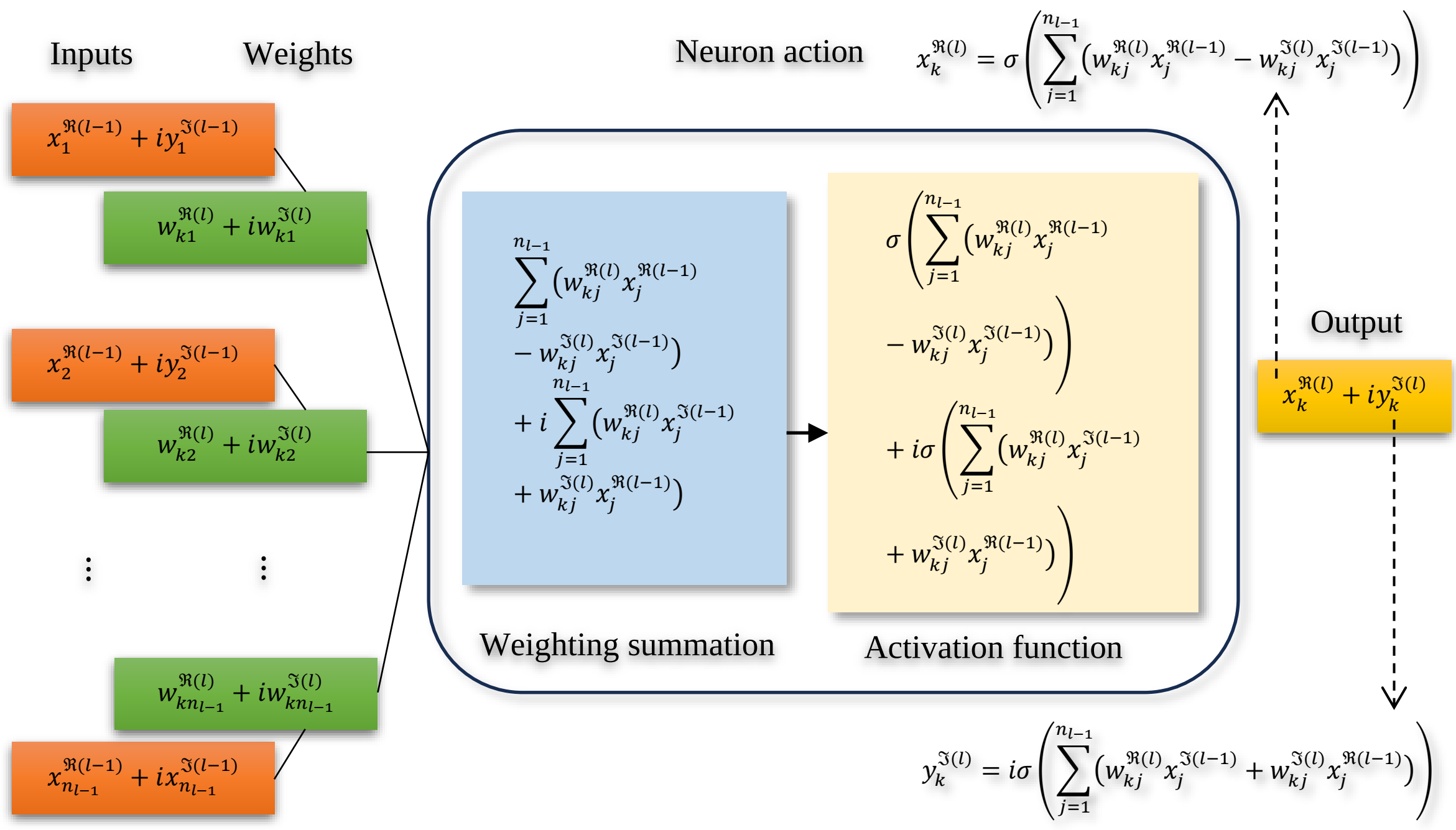

**Figure 9.11.** Structure of a neuron in a split CVNN.

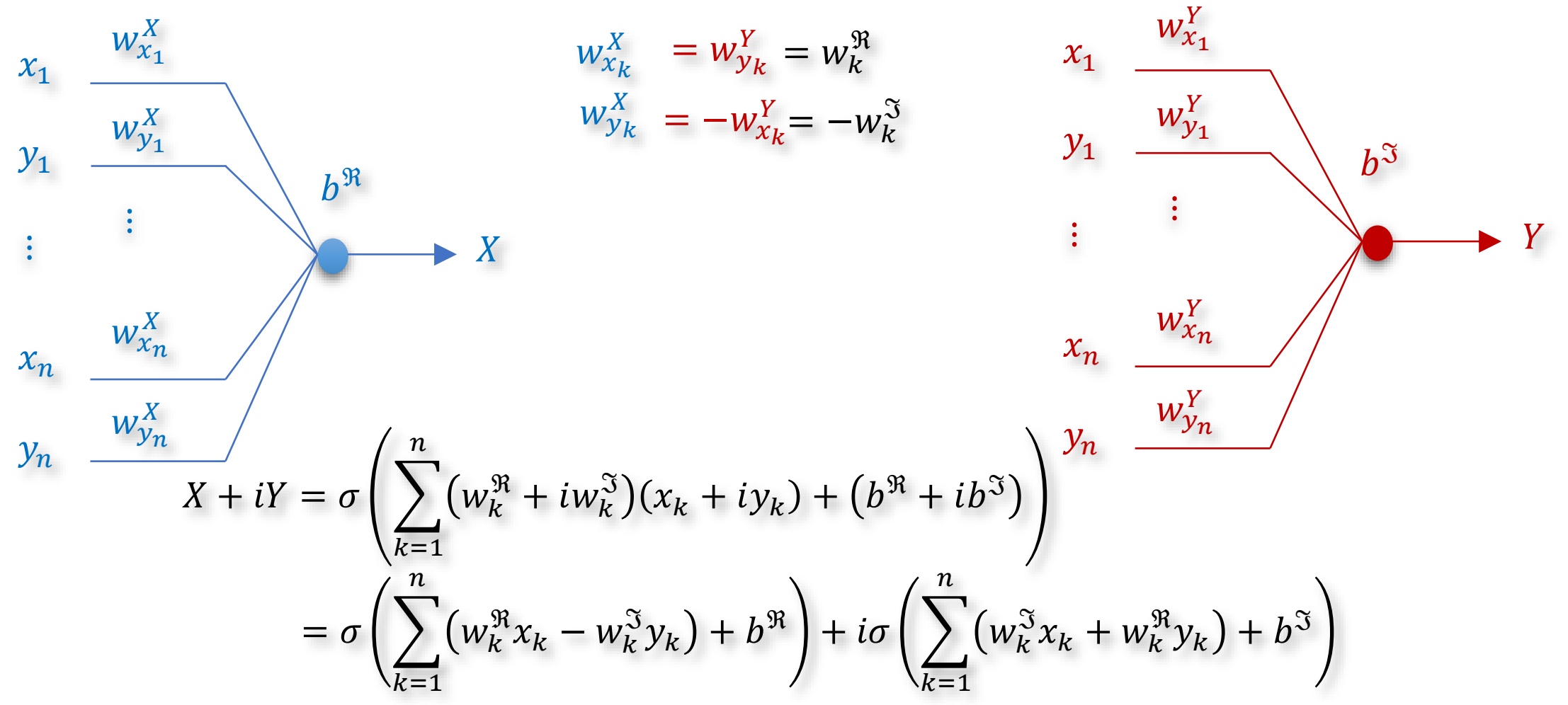

**Figure 9.12.** Two real-valued neurons which are equivalent to a complex-valued neuron.

### B.2. Split-CVNNs: Complex-valued weights and real-valued AFs

In contrast, CVNNs with complex-valued weights and a real-valued AF offer a solution to the phase distortion challenge. This approach, exemplified in [275], has found widespread application in various domains due to its compatibility with popular real-valued AFs, including ReLU, Sigmoid, and Tanh. The ease of computation and avoidance of phase distortion make this variant of CVNNs particularly attractive, as evidenced in applications ranging from [288-292]. Figure 9.11 shows the structure of a single neuron in the split-CVNN with complex-valued weights and a real-valued AF. It shows that the output of a neuron in the split-CVNN is equal to the output of the AF with the complex weighting summation of all its inputs being argument. In Figure 9.12, a complex-valued neuron with $n$-inputs is equivalent to two real-valued neurons with $2n$-inputs. We shall refer to a real-valued neuron corresponding to the real part $X$ of an output of a complex-valued neuron as a Real-part Neuron, and a real-valued neuron corresponding to the imaginary part $Y$ as an Imaginary-part Neuron.

**Remarks:**

- In NNs, input signals are subjected to summation processes where connection weights are multiplied and added up. To unveil potential distinctions between real and complex representations, let us examine the summation of four real numbers ($x_1$, $x_2$, $y_1$, and $y_2$), i.e., sum on $x$ $(x_1 + x_2)$, sum on $y$ $(y_1 + y_2)$ and compare it with the summation of two complex numbers ($z_1 = x_1 + iy_1$ and $z_2 = x_2 + iy_2$), i.e., $(x_1 + x_2) + i(y_1 + y_2)$. Upon evaluating these summations, we observe that the basis of summation remains unchanged between the real and complex domains. The real and imaginary components ($x_n$ and $y_n$) are summed separately in both cases, highlighting a commonality in this fundamental operation.
- The crucial divergence emerges when we explore the weighting processes, particularly the multiplication of signals by connection weights. In the context of real numbers, the product is straightforward, involving the multiplication of individual real values (multiplication on $x$ $(x_1x_2)$ and multiplication on $y$ $(y_1y_2)$). However, in the complex domain, the multiplication of two complex numbers $(z_1z_2)$ yields a more intricate result:

$$z_1z_2 = (x_1 + iy_1)(x_2 + iy_2) = (x_1x_2 - y_1y_2) + i(x_1y_2 + x_2y_1). \tag{9.94}$$

  This complex product involves a combination of real and imaginary components, deviating significantly from the simplicity of real-number multiplication [283].
- To comprehend the intricacies of complex multiplication, we can reinterpret the process in terms of amplitude and phase. If we represent complex numbers as $z_1 = r_1e^{i\,\theta_1}$ and $z_2 = r_2e^{i\,\theta_2}$, the complex multiplication $z_1z_2$ can be expressed as:

$$z_1z_2 = r_1r_2e^{i\,(\theta_1+\theta_2)}. \tag{9.95}$$

  This representation demonstrates that complex multiplication is equivalent to combining amplitude multiplication and phase addition. This fact suggests that CVNNs can be more meaningful in applications where they should adopt the amplitude-phase-type AF.

## 9.5 Properties of Complex AFs

As with any innovative approach, challenges arise during the development and optimization of the CVNNs. One such critical challenge revolves around the choice of AFs within the complex-valued context. AFs play a pivotal role in shaping the non-linear transformations applied to the input data, enabling NNs to learn complex patterns and representations. The standard practice of extending real-valued AFs to the complex domain has been questioned, particularly when applying the Sigmoid function to create a complex-valued saturation function for CVNNs.

The crux of the issue lies in the non-analytic nature of certain CVAF. A complex function is considered analytic (or holomorphic) if it is differentiable at every point within its domain. If $\sigma(z)$ is analytic at all points $z \in \mathbb{C}$, is called entire. The concept of differentiability at each point categorizes a function as regular at that point, while points, where differentiability fails, are termed singular points. The significance of analyticity lies in its implications for the analytical study of neural dynamics, such as learning, self-organization, and processing. One of the most challenging parts of CVNN is to design a suitable AF for CVNNs as Liouville's theorem.

**Theorem 9.5 (Liouville Theorem):** If $\sigma(z)$ is entire and bounded on the complex plane, then $\sigma(z)$ is a constant function.

Since a suitable $\sigma(z)$ must be bounded, it follows from Liouville's theorem that if in addition $\sigma(z)$ is entire, then $\sigma(z)$ is constant, clearly not a suitable AF. In other words, a function that is bounded and analytic everywhere in the complex plane is not a suitable CVAF. In fact, the requirement that $\sigma(z)$ be entire alone imposes considerable structure on it, e.g., the Cauchy-Riemann equations should be satisfied. Therefore, most of the AFs in CVNNs are designed by selecting one property (i.e., boundedness or analytic).

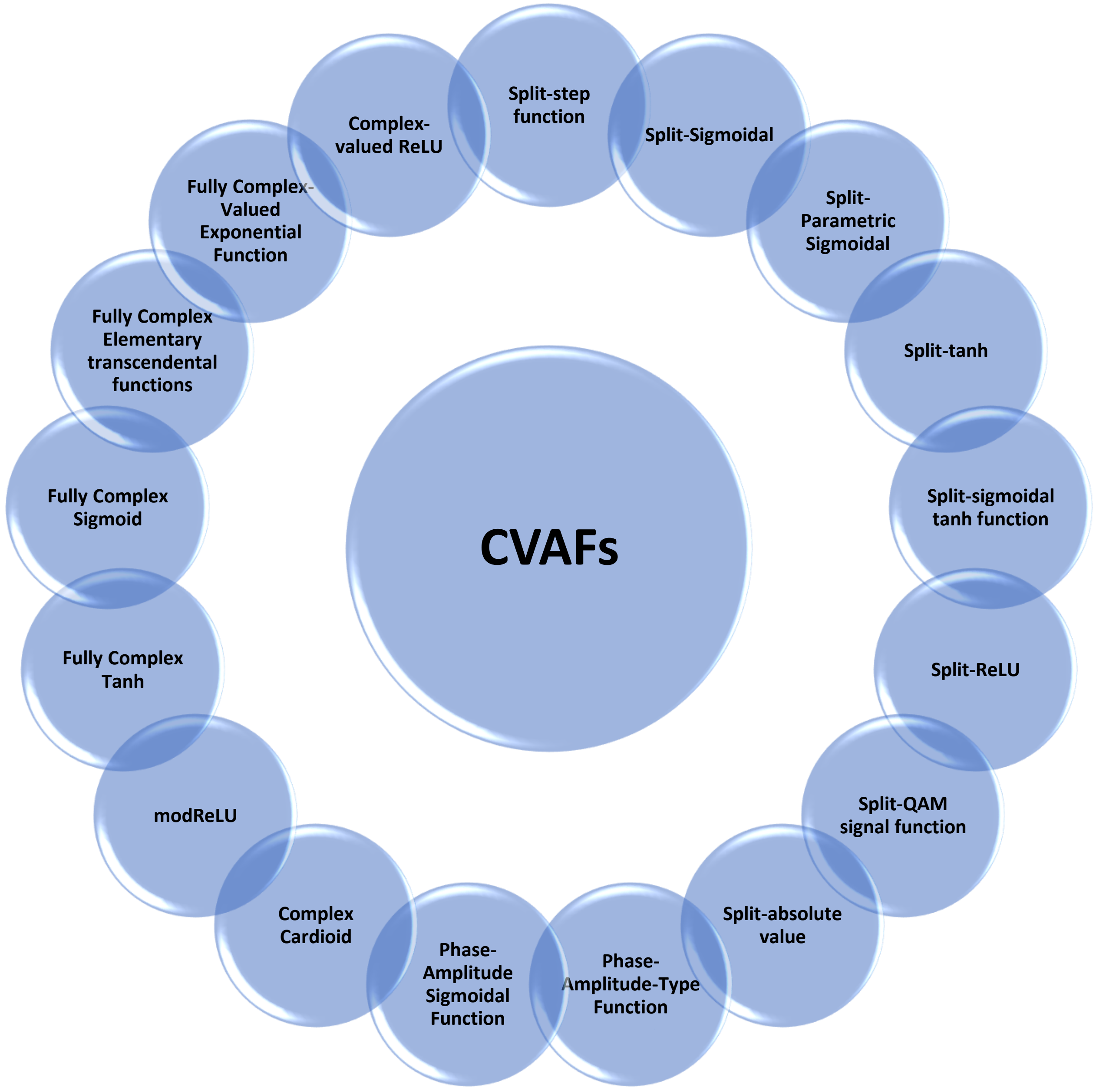

A suitable AF $\sigma(z)$ must possess two properties [273]:

i) $\sigma(z)$ is bounded, and
ii) $\sigma(z)$ is such that $\delta_j \neq (0,0)$ and $x_{jl} \neq (0,0)$ imply $\nabla_{w_{jl}}\mathcal{L} \neq (0,0)$.

Non-compliance with the latter condition is undesirable since it would imply that even in the presence of both non-zero input $x_{jl} \neq (0,0)$ and non-zero error $\delta_j \neq (0,0)$, it is still possible that $\Delta w_{jl} = 0$, i.e., no learning takes place.

**Lemma 9.2:** If $\sigma(z) = u + iv$, and

$$\frac{\partial u}{\partial x}\frac{\partial v}{\partial y} = \frac{\partial v}{\partial x}\frac{\partial u}{\partial y}, \tag{9.96}$$

then $\sigma(z)$ is not a suitable AF.

**Proof:**

We would like to show that an AF with the above property violates condition ii. Suppose that $x_{jl} \neq (0,0)$. We will show that there exists $\delta_j \neq (0,0)$ such that $\nabla_{w_{jl}}\mathcal{L} = 0$. From (9.74) we see that $\nabla_{w_{jl}}\mathcal{L} = 0$ when

$$\left\{\left[\frac{\partial u_j}{\partial x_j} + i\frac{\partial u_j}{\partial y_j}\right]\delta_j^{\Re} + \left[\frac{\partial v_j}{\partial x_j} + i\frac{\partial v_j}{\partial y_j}\right]\delta_j^{\Im}\right\} = 0,$$

or equivalently, when the real and imaginary parts of the equation equal to zero:

$$\frac{\partial u_j}{\partial x_j}\delta_j^{\Re} + \frac{\partial v_j}{\partial x_j}\delta_j^{\Im} = 0,$$

$$\frac{\partial u_j}{\partial y_j}\delta_j^{\Re} + \frac{\partial v_j}{\partial y_j}\delta_j^{\Im} = 0.$$

A nontrivial solution of the above homogeneous system of equations in $\delta_j^{\Re}$ and $\delta_j^{\Im}$ , i.e., $\delta_j^{\Re}$ and $\delta_j^{\Im}$ not both zero, can be found if and only if the determinant of the coefficient matrix is zero: $\frac{\partial u_j}{\partial x_j}\frac{\partial v_j}{\partial y_j} - \frac{\partial u_j}{\partial y_j}\frac{\partial v_j}{\partial x_j} = 0$ or $\frac{\partial u_j}{\partial x_j}\frac{\partial v_j}{\partial y_j} = \frac{\partial u_j}{\partial y_j}\frac{\partial v_j}{\partial x_j}$.

■

The Sigmoid function, $\sigma_{\text{Sigmoid}}(z) = 1/(1 + e^{-z})$, as well as other commonly used AFs like $\text{Tanh}(z)$ and $e^{-z^2}$, are indeed unbounded when their domain is extended from the real line to the complex plane.

- It can be easily verified that when $z$ approaches any value in the set $\{(0 \pm i(2n+1)\pi$: $n$ is any integer$\}$, then $|\sigma_{\text{Sigmoid}}(z)| \to \infty$, and thus $\sigma_{\text{Sigmoid}}(z)$ is unbounded.
- One can verify that $|\text{Tanh}(z)| \to \infty$ as $z$ approaches a value in the set $\{(0 \pm i((2n+1)/2)\pi$: $n$ is any integer$\}$.
- $|e^{-z^2}| \to \infty$ when $z = 0 + iy$ and $y \to \infty$.

In order to avoid the problem of singularities in the Sigmoid function $\sigma_{\text{Sigmoid}}(z)$, it was suggested in [272] to scale "the input data to some region in the complex plane. Scaling the input data to some region in the complex plane could potentially control the magnitude of $z$. However, backpropagation, being a weak optimization procedure, may not enforce constraints on the weights effectively. The weights can still assume arbitrary values, and the combination of inputs and weights, $z$, can result in values that cause numerical issues.

It is clear that some other AF must be found for CDBP. In the derivation of CDBP we only assumed that the partial derivatives $\frac{\partial u}{\partial x}$, $\frac{\partial v}{\partial y}$, $\frac{\partial v}{\partial x}$, and $\frac{\partial u}{\partial y}$ exist. Other important properties the AF $\sigma(z) = u(x,y) + iv(x,y)$ should possess are the following [273]:

1. The AF $\sigma(z) = u(x, y) + iv(x, y)$ should be nonlinear in both $x$ and $y$. This is a fundamental requirement for NNs to have the capacity to learn and represent complex, nonlinear relationships in the data.
2. The AF should be bounded. Boundedness is crucial to avoid numerical instability during training. If the AF is unbounded, it might lead to issues like exploding gradient, making the training process difficult. This is true if and only if both $u$ and $v$ bounded. Since both $u$ and $v$ are used during training (forward pass), they must be bounded. If either one was unbounded, then a software overflow could occur.
3. The partial derivatives $\frac{\partial u}{\partial x}, \frac{\partial v}{\partial y}, \frac{\partial v}{\partial x}$, and $\frac{\partial u}{\partial y}$ must exist and be bounded. This is necessary for the backpropagation algorithm to compute gradients during training. Bounded derivatives contribute to stable learning.
4. $\sigma(z)$ is not entire. See Theorem 9.5.
5. The relationship $\frac{\partial u}{\partial x}\frac{\partial v}{\partial y} \neq \frac{\partial v}{\partial x}\frac{\partial u}{\partial y}$ should hold. See Lemma 9.2.

Considering all these properties helps in defining an AF that is suitable for complex domain backpropagation.

Note that by Louiville's theorem, the second and the fourth conditions are redundant, i.e., a bounded nonlinear function in $\mathbb{C}$ cannot be an entire function. Using this fact, You and Hong [293] reduced the above conditions into the four conditions given below to introduce a split complex AF:

- The AF $\sigma(z) = u(x, y) + iv(x, y)$ is nonlinear in $x$ and $y$.
- For the stability of a system, $\sigma(z)$ should have no singularities and be bounded for all $z$ in a bounded set.
- The partial derivatives $\frac{\partial u}{\partial x}, \frac{\partial v}{\partial y}, \frac{\partial v}{\partial x}$, and $\frac{\partial u}{\partial y}$ should be continuous and bounded.
- The relationship $\frac{\partial u}{\partial x}\frac{\partial v}{\partial y} \neq \frac{\partial v}{\partial x}\frac{\partial u}{\partial y}$ if not, then $\sigma(z)$ is not a suitable AF except in the following cases:

$$\frac{\partial u}{\partial x} = \frac{\partial v}{\partial x} = 0, \text{and } \frac{\partial u}{\partial y} \neq 0, \frac{\partial v}{\partial y} \neq 0, \tag{9.97.1}$$

$$\frac{\partial u}{\partial y} = \frac{\partial v}{\partial y} = 0, \text{and } \frac{\partial u}{\partial x} \neq 0, \frac{\partial v}{\partial x} \neq 0. \tag{9.97.2}$$

Note that both sets of conditions above emphasize the boundedness of an AF and its partial derivatives, even when the function is defined in a local domain of interest. By Louisville theorem, however, the cost of this restriction is that a bounded AF cannot be analytic. The $\mathrm{Tanh}(z)$ function violates the second and third boundedness requirements of both sets of conditions above. Instead of boundedness, $\mathrm{Tanh}(z)$ function has well-defined but not necessarily bounded first order derivatives almost everywhere in $\mathbb{C}$. Since they are bounded almost everywhere, the rare existence of singular points hardly poses a problem in learning, and the singular points can be handled separately. Therefore, the above boundedness requirements are unnecessary for the fully complex AFs that are almost everywhere bounded.

Several complex AFs have been proposed in the literature. These AFs are mainly classified into the following two categories.

**A- Split AF**

The initial idea of split AF is proposed by [275, 294], where they adopt the real-valued AF (for example, Tanh and Sigmoid function) as shown in (9.98.1) and (9.98.2).

- Split real–imaginary AF

$$\sigma_{\text{Real-imaginary}}(z) = \sigma^{\Re}(\Re[z]) + i\sigma^{\Re}(\Im[z]). \tag{9.98.1}$$

- Split phase-amplitude AF

$$\sigma_{\text{Phase-amplitude}}(z) = \sigma^{\Re}(|z|)\exp(i\arg(z)). \tag{9.98.2}$$

Where $z = x + iy \in \mathbb{C}$, $\sigma^{\Re}(\cdot)$ is the real-valued AF (e.g., Sigmoid, Tanh), and $\sigma_{\text{Real-imaginary}}(z) \in \mathbb{C}$, $\sigma_{\text{Phase-amplitude}}(z) \in \mathbb{C}$.

Such AF is bounded and not analytic. In contrast to RVNNs, where analyticity has been a cornerstone for rigorous investigations, CVNNs adopt a more pragmatic approach. The attention has shifted towards constructing the dynamics of learning and self-organization based on meaningful partial derivatives. This departure from the requirement of analyticity does not diminish the significance of the analyses but rather opens up avenues for more flexible and creative network design.

An interesting consequence of this shift in thinking is the emergence of coordinate-dependent neural dynamics. Unlike the general principle that mechanics, including NNs, should be coordinate-independent, CVNNs embrace a certain level of coordinate dependence. This choice is not a drawback but a deliberate strategy to enhance the CVNNs. Two widely used AFs exemplify this departure from traditional thinking – one that focuses on the real and imaginary components (9.98.1) and another that manipulates amplitude and phase (9.98.2). The coordinate dependence of CVNN dynamics becomes an asset when interacting with the real world. Considerations of amplitude, phase, and frequency become crucial in applications involving wave control or periodic motion. A CVNN is uniquely positioned to utilize this coordinate dependence as an advantage, aligning its dynamics with the specific properties it directly deals with in real-world interactions.

**B- Fully Complex AF**

Where real and imaginary parts are treated as a single entity. There are only a few works of literature on fully complex AF compared to split-type due to the computation complexity and the difficulty in fulfilling Livoullie's theorem.

$$\sigma_{\text{Fully Complex}}(z) = \sigma^{\mathbb{C}}(z), \tag{9.99}$$

where $z = x + iy \in \mathbb{C}$, $\sigma^{\mathbb{C}}(\cdot)$ denotes, for example, a complex Sigmoid.

In [274, 276], ETFs (for instance, Tanh(z), Sinh(z), Sin(z), ArcSin(z), Tan(z), ArcTan(z), ArcTanh(z), ArcCos(z)) are studied where the functions are entire and either bounded in a desired domain or bounded almost everywhere (i.e., unbounded on a set of points at complex plane).

## 9.6 Complex Activation Functions

### 9.6.1 Split-Step Function

The Split-Step function [295] is a CVAF defined by complex combining a real-valued step function applied separately to the real and imaginary components of a complex variable $z = x + iy$. The mathematical expression for the Split-Step Function is given by:

$$\begin{aligned}\sigma_{\text{Split-StepF}}(z) &= \sigma^{\Re}_{\text{Split-StepF}}(x) + i\sigma^{\Re}_{\text{Split-StepF}}(y)\\ &= \sigma^{\Re}_{\text{Split-StepF}}\big(\Re(z)\big) + i\sigma^{\Re}_{\text{Split-StepF}}\big(\Im(z)\big),\end{aligned} \tag{9.100.1}$$

where $z = x + iy$ and $\sigma^{\Re}_{\text{Split-StepF}}$ is a real-valued step function defined on $\mathbb{R}$; that is,

$$\sigma^{\Re}_{\text{Split-StepF}}(u) = \begin{cases}1, & u \geq 0\\ 0, & \text{otherwise}'\end{cases} \tag{9.100.2}$$

for any $u \in \mathbb{R}$.

**Remarks:**

- The Split-Step function introduces a threshold activation behavior, see Figure 9.13. If the real part or imaginary part of the complex variable is non-negative, the real part or imaginary part of the AF becomes 1, respectively; otherwise, it becomes 0.
- The AF acts independently on the real and imaginary components of the complex variable. This separation allows the function to capture different aspects of information.
- The output of each component of the Split-Step function is binary, either 0 or 1, making it suitable for scenarios where a binary decision or representation is desired.

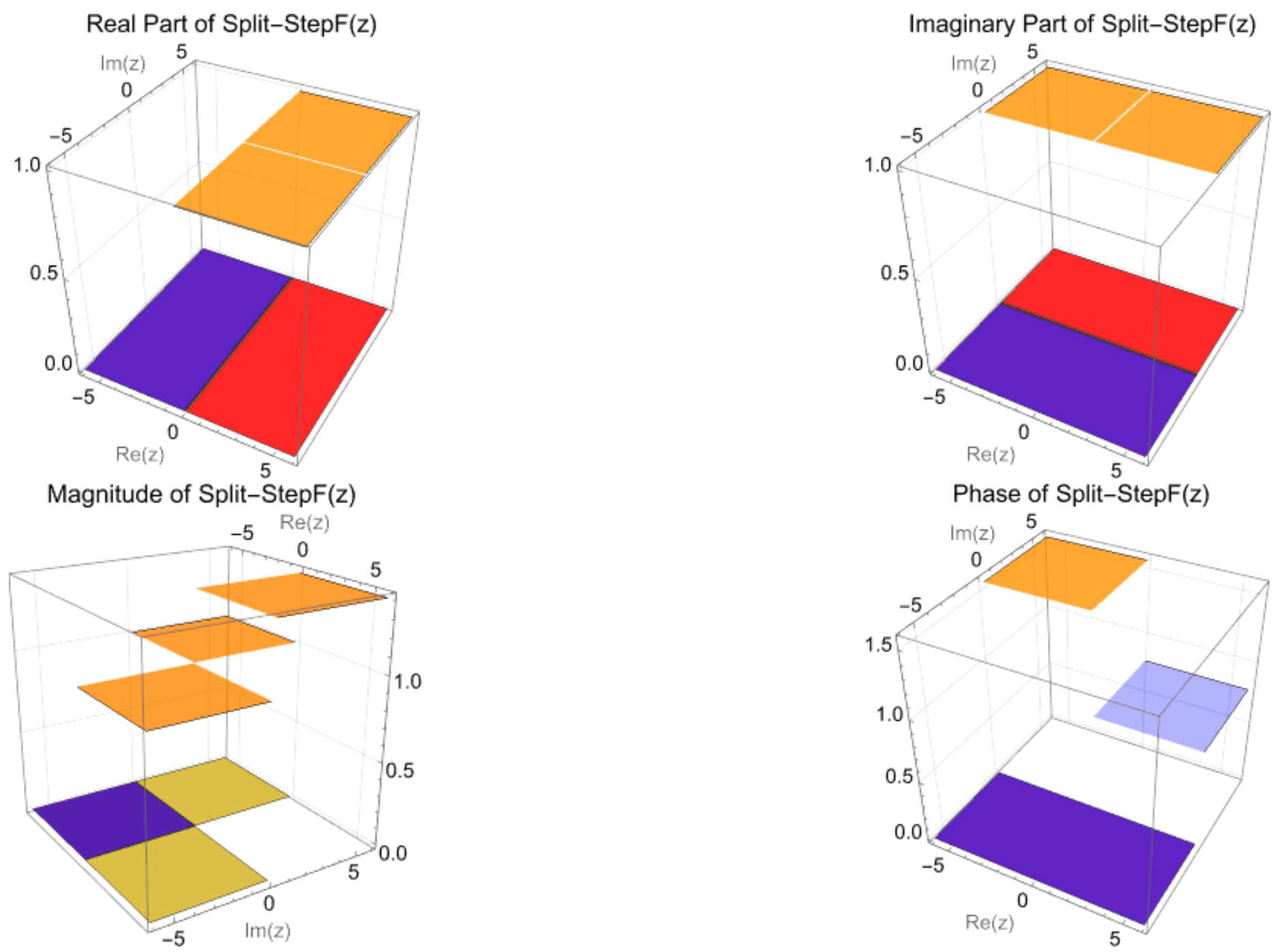

**Figure 9.13.** Visualizations of the real, imaginary, magnitude, and phase parts of the split-step function. These figures provide a comprehensive visualization of the Split-Step function $\sigma_{\text{Split-StepF}}(z)$, where $z = x + iy$, illustrating its real part, imaginary part, magnitude, and phase over a 3D grid with $x$ and $y$ ranging from $-6$ to 6. Each plot uses blue and orange mesh shading for clarity and includes a slice contour plot at $z = 0$ to enhance understanding of the function's behavior at this specific plane.

- The XOR problem and the detection of symmetry problem which cannot be solved with a single real-valued neuron (i.e. a two-layered RVNN), can be solved with a single complex-valued neuron (i.e. a two-layered CVNN using Split-Step Function) with the orthogonal decision boundaries, which reveals the potent computational power of complex-valued neurons.

### 9.6.2 Split-Sigmoid Function

The Split-Sigmoid function [296-306] is a CVAF defined by the expression:

$$\begin{aligned}\sigma_{\text{Split-SigmoidF}}(z) &= \sigma^{\Re}_{\text{Split-SigmoidF}}(x) + i\sigma^{\Re}_{\text{Split-SigmoidF}}(y)\\ &= \frac{1}{1+e^{-\Re(z)}} + i\frac{1}{1+e^{-\Im(z)}},\end{aligned} \tag{9.101.1}$$

where $z = x + iy$ and $\sigma^{\Re}_{\text{Split-SigmoidF}}$ is a real-valued function. This AF exhibits sigmoidal behavior on the real axis, as described by:

$$\sigma^{\Re}_{\text{Split-SigmoidF}}(u) = \frac{1}{1+e^{-u}}, \tag{9.101.2}$$

for any $u \in \mathbb{R}$. Remarkably, this AF is a unique fusion of real-valued Sigmoid functions applied separately to the real and imaginary components of the complex variable $z$.

**Remarks:**

- The real- and imaginary- parts of the Split-Sigmoid function are bounded for any complex number, a direct consequence of the Sigmoid function's properties used in their definitions, see Figure 9.14. The Sigmoid function ensures that both the imaginary and real parts of the Split-Sigmoid CVAF remain bounded between 0 and 1. This characteristic is crucial in NNs as it contributes to stability during forward propagation, preventing exploding activations that can impede training.

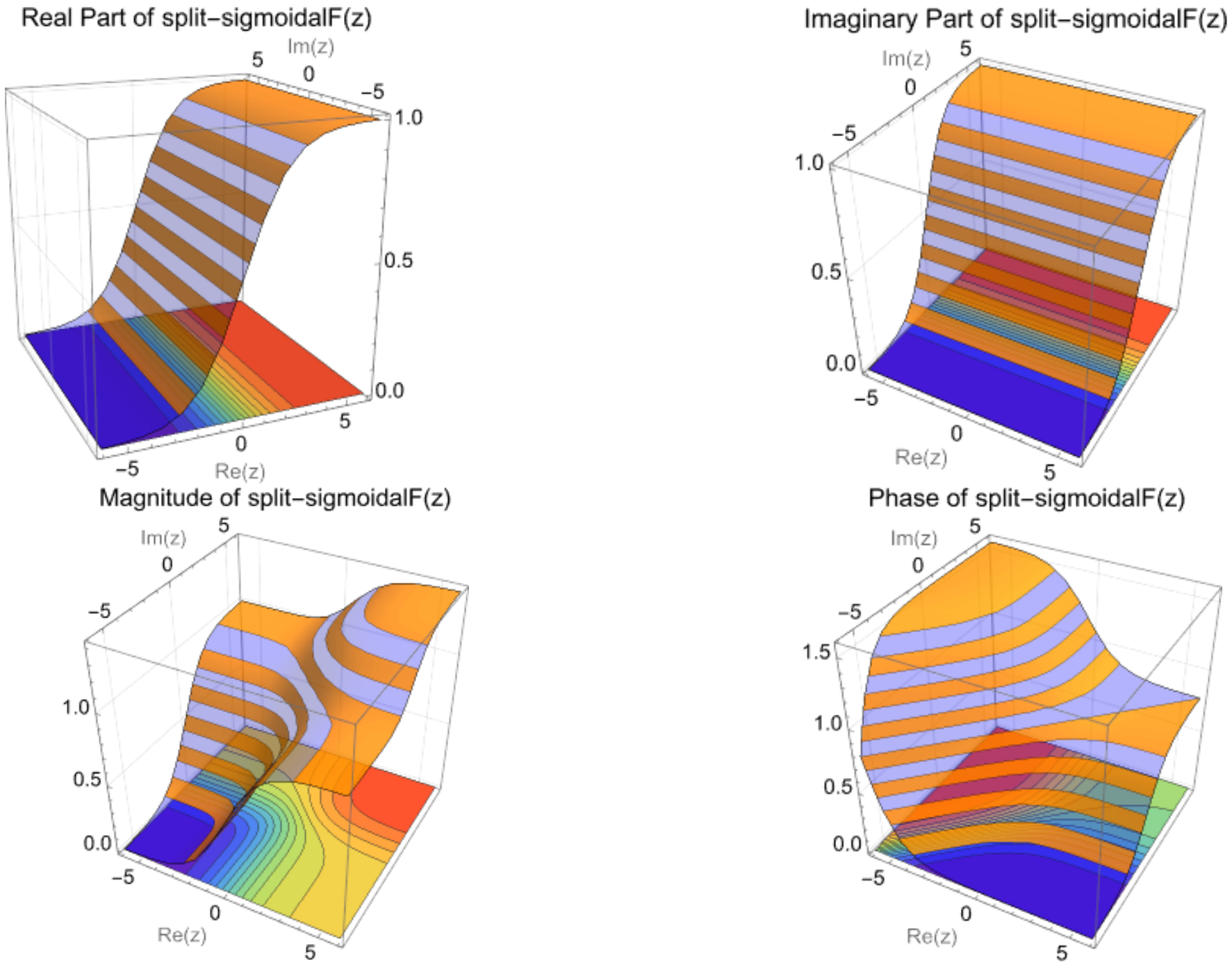

**Figure 9.14.** Visualizations of the real, imaginary, magnitude, and phase parts of the Split-Sigmoid function.

- The Split-Sigmoid CVAF exhibits line symmetry with respect to both the real and imaginary axes, see Figure 9.14. This symmetry implies that replacing a complex number $z_1 = (a, b)$ with $z_2 = (a, -b)$ leaves the real part unchanged and replacing $z_1 = (a, b)$ with $z_2 = (-a, b)$ leaves the imaginary part unchanged. This simplifies the analysis of the function, enhancing its interpretability and reducing complexity.
- The line symmetry exhibited by the Split-Sigmoid CVAF is a key feature with implications for neural dynamics. This symmetry simplifies the analysis by establishing a relationship between values on one side of the real or imaginary axes and their mirrored counterparts.
- The Split-Sigmoid CVAF's characteristics make it particularly suitable for scenarios where NNs encounter information with inherent symmetries. The line symmetry along the real and imaginary axes implies that the function excels in processing complex information and displaying specific symmetrical patterns.
- The analogy drawn between the Split-Sigmoid CVAF and real-valued NNs processing $2n$-dimensional real-valued information provides valuable insights into the neural dynamics associated with this CVAF. The separate processing of real and imaginary components parallels the operations in RVNNs, offering a bridge for understanding and potentially leveraging existing knowledge from real-valued domains.

### 9.6.3 Split-Parametric Sigmoid Function

The Split-Parametric Sigmoid function was introduced as a CVAF [275, 276], expressed mathematically as:

$$\sigma_{\text{Split-PSigmoidF}}(z) = \sigma^{\Re}_{\text{Split-PSigmoidF}}(x) + i\sigma^{\Re}_{\text{Split-PSigmoidF}}(y)$$
$$= \frac{2c_1}{1 + e^{-c_2 \Re(z)}} - c_1 + i\left(\frac{2c_1}{1 + e^{-c_2 \Im(z)}} - c_1\right), \tag{9.102.1}$$

where $z = x + iy$ and $\sigma^{\Re}_{\text{Split-PSigmoid}}$ is a real-valued function. It has sigmoidal behavior defined on $\mathbb{R}$; that is,

$$\sigma^{\Re}_{\text{Split-PSigmoidF}}(u) = \frac{2c_1}{1 + e^{-c_2 u}} - c_1, \tag{9.102.2}$$

for any $u \in \mathbb{R}$, with $c_1$, and $c_2$ suitable real parameters. The derivative of $\sigma^{\Re}_{\text{Split-PSigmoidF}}(u)$ has a simple expression:

$$\frac{\partial}{\partial u}\sigma^{\Re}_{\text{Split-PSigmoidF}}(u) = \frac{c_2}{2c_1}\left[c_1^2 - \left(\sigma^{\Re}_{\text{Split-PSigmoidF}}(u)\right)^2\right]. \tag{9.102.3}$$

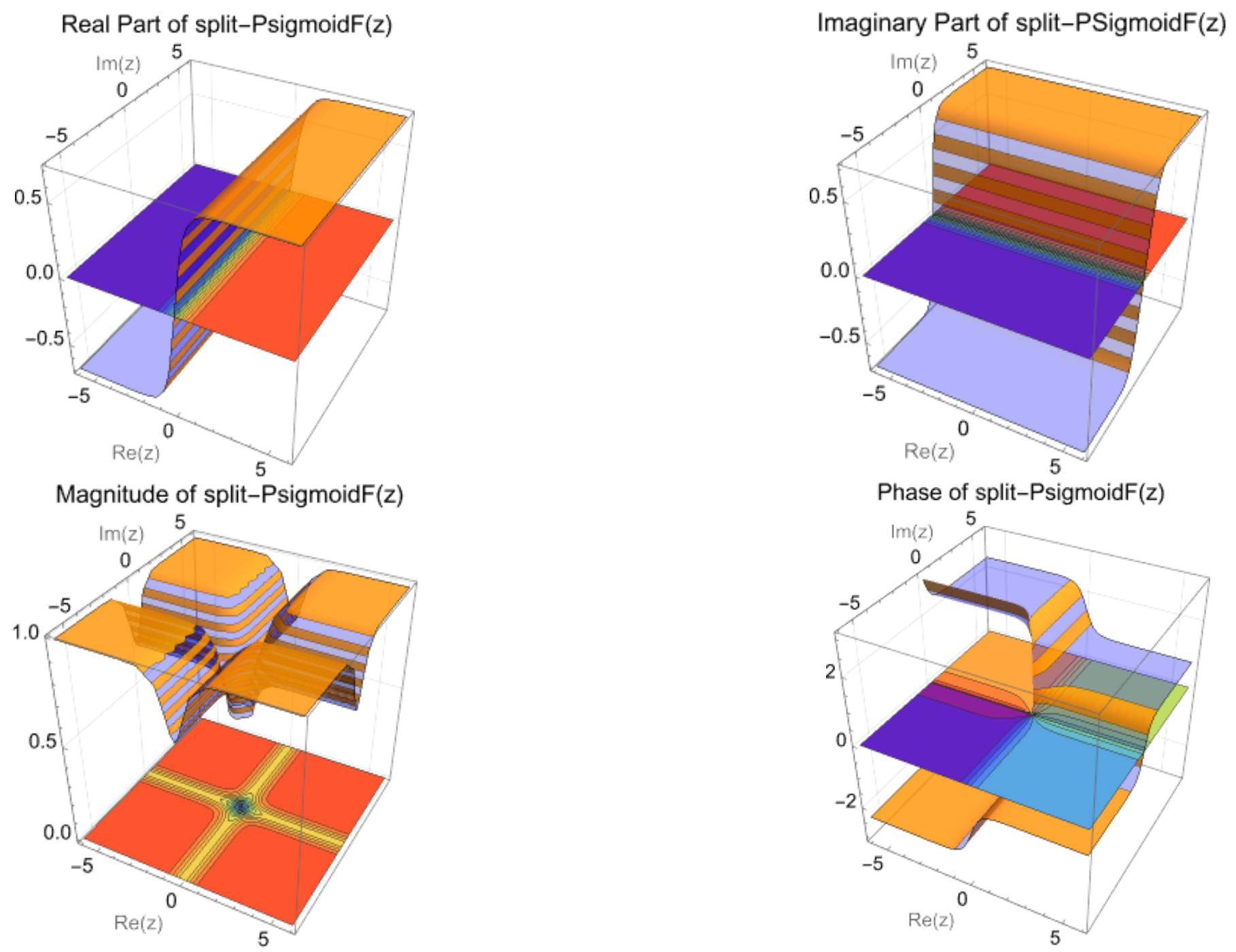

**Figure 9.15.** Visualizations of the real, imaginary, magnitude, and phase parts of the Split-Parametric Sigmoid function.

**Remarks:**

1. A notable feature of the Split-PSigmoidF is its inherent boundedness, see Figure 9.15. The AF's behavior is a two-dimensional extension of the Sigmoid function applied to the real axis. The Sigmoid behavior ensures that the Split-PSigmoidF remains within certain limits, contributing to stability during NN training.
2. The parameters $c_1$ and $c_2$ give practitioners control over the amplitude, shape, sensitivity, and convergence properties of the Split-Parametric Sigmoidal function, allowing for fine-tuning and adaptation to specific NN requirements, Figure 9.15. Adjusting these parameters may involve a balance between achieving convergence, avoiding saturation, and capturing intricate patterns in the data.
   - $c_1$ is a real-valued parameter that scales the output of the sigmoidal function in the real- and imaginary- part of the Split-PSigmoidF. Increasing $c_1$ amplifies the output of the real- and imaginary- part, effectively increasing the amplitude of the Split-PSigmoidF.
   - $c_2$ is a real-valued parameter that controls the rate of the exponential function in the sigmoidal part of the Split-PSigmoidF. Higher values of $c_2$ lead to a steeper sigmoidal curve, causing the AF to saturate more quickly. Lower values result in a more gradual saturation.
   - $c_2$ affects the sensitivity of the CVAF to changes in the input. A higher $c_2$ makes the function more sensitive, while a lower $c_2$ makes it less sensitive. The $c_2$ parameter directly appears in the derivative expression, influencing the rate of change of the AF.

### 9.6.4 Split-Tanh

One of the widely used CVAFs [307-310] is Split-Tanh. Split-Tanh CVAF is defined as

$$\begin{aligned}\sigma_{\text{Split-Tanh}}(z) &= \sigma^{\Re}_{\text{Split-Tanh}}(x) + i\sigma^{\Re}_{\text{Split-Tanh}}(y) \\ &= \text{Tanh}\big(\Re(z)\big) + i\,\text{Tanh}\big(\Im(z)\big),\end{aligned} \tag{9.103.1}$$

where $z = x + iy$ and $\sigma^{\Re}_{\text{Split-Tanh}}$ is a real-valued function,

$$\sigma^{\Re}_{\text{Split-Tanh}}(u) = \text{Tanh}(u), \tag{9.103.2}$$

for any $u \in \mathbb{R}$.

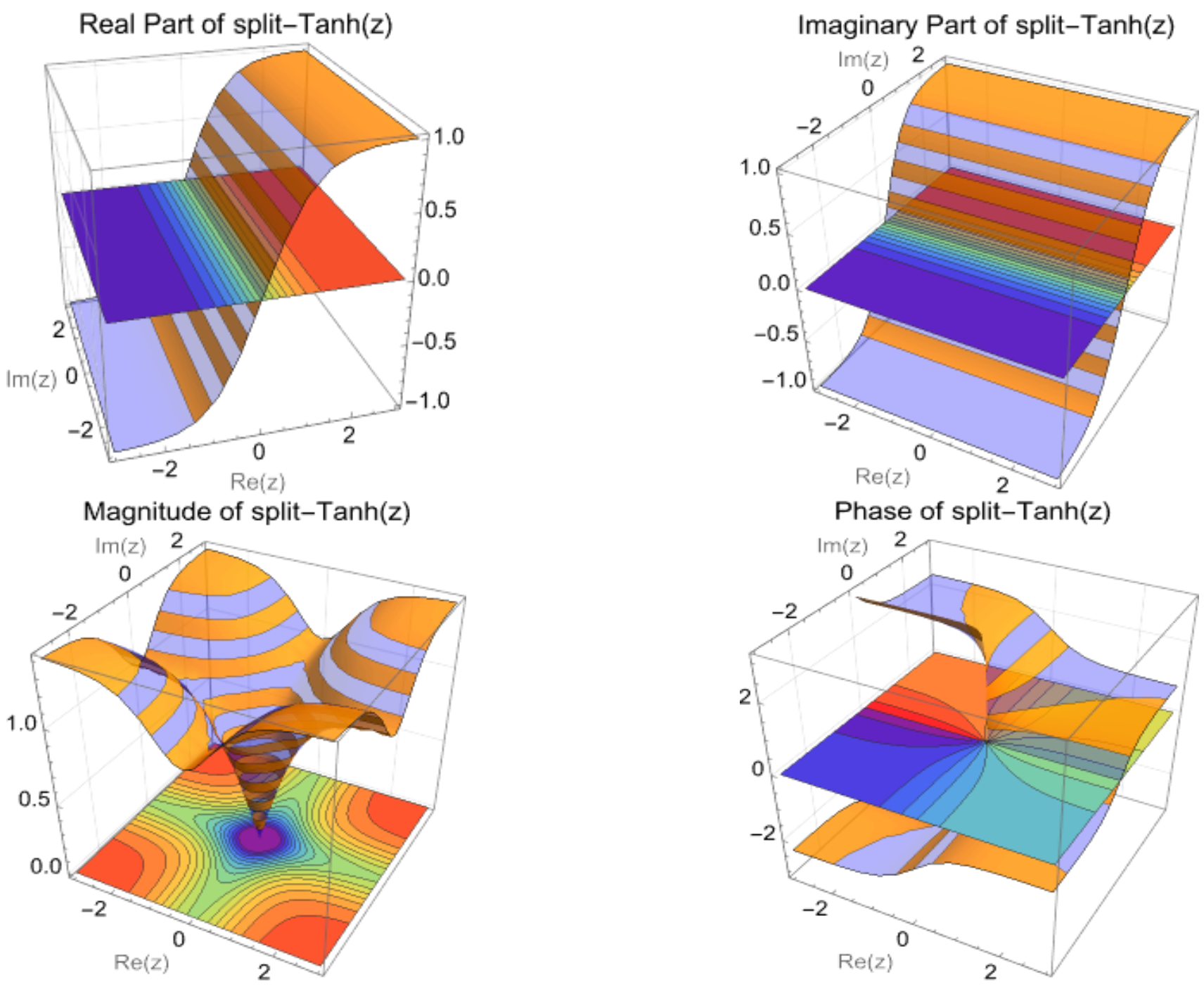

**Figure 9.16.** Visualizations of the real, imaginary, magnitude, and phase parts of the Split-Tanh function.

**Remarks:**

- This CVAF is a unique fusion of real-valued hyperbolic tangent functions applied separately to the real and imaginary components of the complex variable.
- The real- and imaginary- part of the Split-Tanh CVAF is bounded for any complex number, see Figure 9.16. This boundedness is a result of the properties of the hyperbolic tangent function that is used in the definition of the Split-Tanh CVAF. The hyperbolic tangent function has a range between $-1$ and $1$. Specifically: $-1 < \mathrm{Tanh}(u) < 1$. In the case of the real- and imaginary- part of the Split-Tanh AF, where $\mathrm{Tanh}(\Re(z))$ and $\mathrm{Tanh}(\Im(z))$ are involved, the real- and imaginary- parts are also bounded between $-1$ and $1$.
- This boundedness is important in the context of NNs because it helps in controlling the scale of values during the forward propagation of information through the network. It contributes to the stability of the network and prevents the activation values from growing too large, a phenomenon known as exploding activations, which can make training difficult.
- This AF exhibits line symmetry with respect to both the real and imaginary axes. The real- and imaginary- parts of Split-Tanh AF are visually depicted in Figure 9.16, showcasing its characteristic shape.
- The symmetry of the real part of the split-Tanh function with respect to the real axis ($\Im(\mathrm{z}) = 0$) implies that if you replace $z_1 = (a, b)$ with $z_2 = (a, -b)$ the real part of the split-Tanh function value remains the same.
- The symmetry of the imaginary part of the split-Tanh function with respect to the imaginary axis ($\Re(z) = 0$) implies that if you replace $z_1 = (a, b)$ with $z_2 = (-a, b)$ the imaginary part of the split-Tanh function value remains the same.
- This symmetry implies that the neural dynamics associated with this AF hold special significance along the axes of $\Im(z) = 0$ and $\Re(z) = 0$, i.e., the real and imaginary parts. Line symmetry simplifies the analysis of the function, making it easier to understand and work with. It reduces the complexity of studying the behavior of the function by providing a clear relationship between the values on one side of the real axis or imaginary axis and their mirrored counterparts on the other side.
- We can also observe characteristic changes on and around the real and imaginary axes in figures of the amplitude and the phase of the Split-Tanh AF, Figure 9.16. The symmetry gives the two axes a special meaning in neural dynamics.

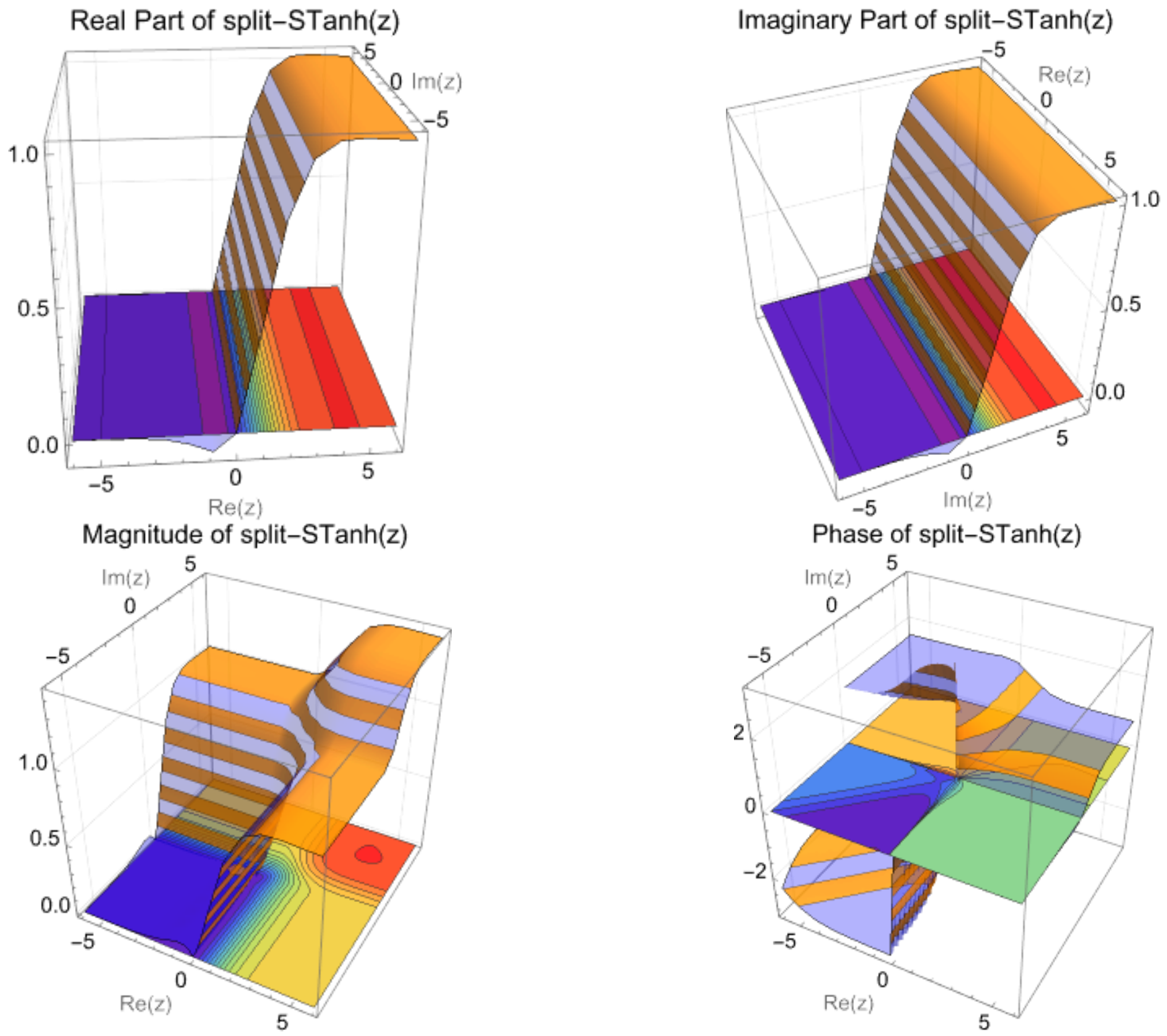

**Figure 9.17.** Visualizations of the real, imaginary, magnitude, and phase parts of the Split-STanh function.

- The line symmetry of the real-imaginary-type CVAF suggests that it excels when dealing with complex information that possesses symmetry or holds specific meanings along the real and imaginary axes.
- This symmetry implies that the neural dynamics of a network using this AF may resemble, to some extent, those of a RVNN processing $2n$-dimensional real-valued information. This analogy arises from the separate and independent processing of real and imaginary components, as demonstrated in (9.103.1) and (9.103.2).

### 9.6.5 Split- Sigmoid Tanh Function

The Split- Sigmoidal Tanh (Split-STanh) CVAF [311] was defined as follows:

$$\sigma_{\text{Split-STanh}}(z) = \sigma^{\Re}_{\text{Split-STanh}}(x) + i\sigma^{\Re}_{\text{Split-STanh}}(y) = \frac{\text{Tanh}(\Re(z))}{1-(\Re(z)-3)e^{-\Re(z)}} + i\frac{\text{Tanh}(\Im(z))}{1-(\Im(z)-3)e^{-\Im(z)}}, \tag{9.104.1}$$

where $z = x + iy$ and $\sigma^{\Re}_{\text{Split-STanh}}$ is a real-valued function,

$$\sigma^{\Re}_{\text{Split-STanh}}(u) = \frac{\text{Tanh}(u)}{1-(u-3)e^{-u}}, \tag{9.104.2}$$

for any $u \in \mathbb{R}$.

**Remarks:**

- The Split-STanh CVAF is designed to ensure boundedness for both its real and imaginary parts. The maximum value of the real part of the Split-STanh AF is approximately 1.01802 and occurs when the input $x$ is around $4.06725$, and the minimum value of the real part is approximately $-0.0715838$ and occurs when the input $x$ is around $-0.67288$. Similarly, the maximum and minimum values of the imaginary part of the Split-STanh AF approximately $1.01802$ and $-0.0715838$ occur when the input $y$ is around $4.06725$, and $-0.67288$, respectively.

- The AF exhibits line symmetry with respect to both the real and imaginary axes. The real- and imaginary-parts of Split-STanh AF, as well as the amplitude and the phase of the AF, are visually depicted in Figure 9.17, showcasing its characteristic shape.

### 9.6.6 Split-Hard Tanh (Split-absolute value)

The Split-Hard Tanh [312] is a non-smooth function used in place of a Split-Tanh function. The Split-Hard Tanh retains the basic shape of Split-Tanh but uses simpler functions. Split-Hard Tanh AF is defined as

$$\begin{aligned}\sigma_{\text{Split-Hard Tanh}}(z) &= \sigma^{\Re}_{\text{Split-Hard Tanh}}(x) + i\sigma^{\Re}_{\text{Split-Hard Tanh}}(y)\\ &= \frac{1}{2}(|\Re(z)+1| - |\Re(z)-1|) + i\frac{1}{2}(|\Im(z)+1| - |\Im(z)-1|),\end{aligned} \quad (9.105.1)$$

where $z = x + iy$ and $\sigma^{\Re}_{\text{Split-Hard Tanh}}$ is a real-valued function,

$$\sigma^{\Re}_{\text{Split-Hard Tanh}}(u) = \frac{1}{2}(|u+1| - |u-1|), \quad (9.105.2)$$

for any $u \in \mathbb{R}$.

The function, given by $\sigma^{\Re}_{\text{Split-Hard Tanh}}(u) = \frac{1}{2}(|u+1| - |u-1|)$, is particularly interesting due to its connection with HardTanh AF. Let us explore its behavior across various ranges of real numbers.

- Case $(u > 1)$: $u+1$ will be positive. $u-1$ will be positive. The function becomes $\sigma^{\Re}_{\text{Split-Hard Tanh}}(u) = \frac{1}{2}\big(u+1-(u-1)\big) = \frac{1}{2}2 = 1$.
- Case $(u < -1)$: $u+1$ will be negative. $u-1$ will also be negative. The function becomes $\sigma^{\Re}_{\text{Split-Hard Tanh}}(u) = \frac{1}{2}(-(u+1) - [-(u-1)]) = \frac{1}{2}(-2) = -1$.
- Case $(0 < u < 1)$ : $u+1$ will be positive. $u-1$ will be negative. The function becomes $\sigma^{\Re}_{\text{Split-Hard Tanh}}(u) = \frac{1}{2}\ ((u+1) - (-(u-1))) = 1/2\ \cdot 2u = u$.
- Case $(-1 < u < 0)$: $u+1$ will be positive. $u-1$ will be negative. The function becomes $\sigma^{\Re}_{\text{Split-Hard Tanh}}(u) = \frac{1}{2}\ ((u+1) - (-(u-1))) = 1/2\ \cdot 2u = u$.
- So, for both $-1 < u < 0$ and $0 < u < 1$, the function $\sigma^{\Re}_{\text{Split-Hard Tanh}}(u)$ evaluates to $u$. Therefore, within the interval $(-1,1)$, the function simplifies to $u$. This specific behavior showcases a linear relationship between the function's output and the real variable $u$ in this range. Mathematically, the $\sigma^{\Re}_{\text{Split-Hard Tanh}}(u)$ function can be defined as follows:

$$\sigma^{\Re}_{\text{Split-Hard Tanh}}(u) = \begin{cases} -1, & u < -1,\\ 1, & u > 1,\\ u, & -1 \le u \le 1.\end{cases} \quad (9.106)$$

- The HardTanh function is a modified version of the standard Tanh function that clips its output values to a certain range, typically $[-1,1]$. Mathematically, the hard Tanh function can be defined as follows:

$$\sigma_{\text{Hard Tanh}}(u) = \begin{cases} -1, & u < -1,\\ 1, & u > 1,\\ u, & -1 \le u \le 1.\end{cases} \quad (9.107)$$

- As a result, the $\sigma_{\text{Split-Hard Tanh}}(z)$ function is a modified version of the $\sigma_{\text{Split-Tanh}}(z)$ function, Figure 9.18.

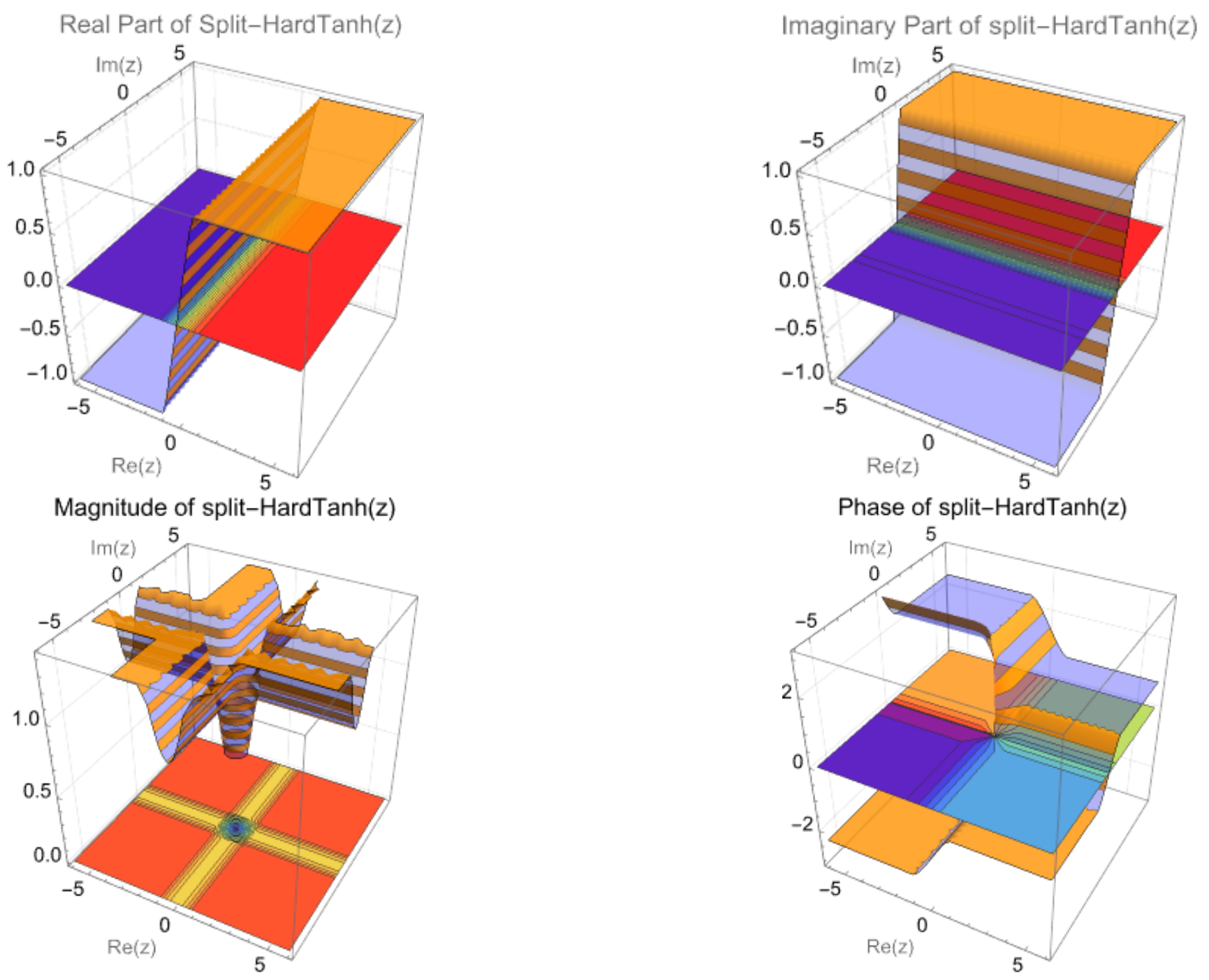

**Figure 9.18.** Visualizations of the real, imaginary, magnitude, and phase parts of the Split-Hard Tanh function.

- One of the main advantages of the Split-Hard Tanh function is its computational efficiency. Unlike the Split-Tanh function, which can be computationally expensive, the Split-Hard Tanh function relies on simple linear operations (addition and absolute value), making it faster to compute. This efficiency can be particularly beneficial in scenarios where computational resources are limited or when training large NNs.

### 9.6.7 Split-CReLU

The Split-CReLU [313, 314] AF is an extension of the ReLU AF for real numbers, but it is applied separately to the real and imaginary parts of a complex number. The general form of the Split-CReLU AF is given by:

$$\begin{aligned}\sigma_{\text{Split-CReLU}}(z) &= \sigma^{\Re}_{\text{Split-CReLU}}(x) + i\sigma^{\Re}_{\text{Split-CReLU}}(y) \\ &= \max(\Re(z), 0) + i\max(\Im(z), 0),\end{aligned} \tag{9.108.1}$$

where $z = x + iy$ and $\sigma^{\Re}_{\text{Split-CReLU}}$ is a real-valued function,

$$\sigma^{\Re}_{\text{Split-CReLU}}(u) = \text{ReLU}(u) = \max(u, 0), \tag{9.108.2}$$

for any $u \in \mathbb{R}$.

**Remarks:**

- The result of applying the Split-ReLU to a complex input is a complex number with potentially different real and imaginary components.
- Like the traditional ReLU, the Split-ReLU induces sparsity in the representation. Any negative values in either the real or imaginary part are set to zero, effectively activating only positive components, see Figure 9.19. This sparsity can be beneficial in certain scenarios, such as feature selection or reducing the risk of vanishing gradients during training.
- Split-CReLU satisfies the Cauchy-Riemann equations when both the real and imaginary parts are at the same time either strictly positive or strictly negative. This means that Split-CReLU satisfies the Cauchy-Riemann equations when $\theta_z \in (0, \pi/2)$ or $\theta_z \in (\pi, 3\pi/2)$.

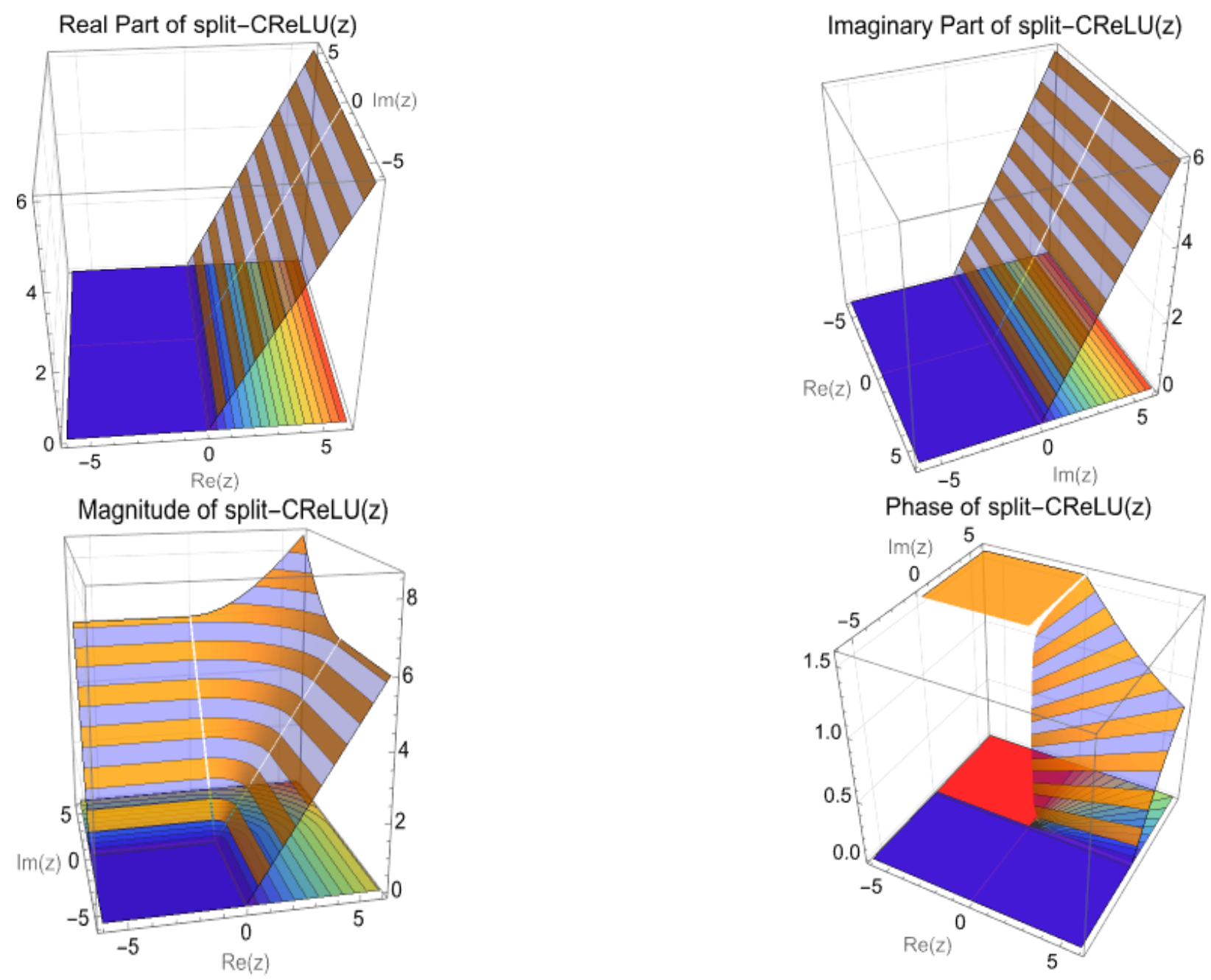

**Figure 9.19.** Visualizations of the real, imaginary, magnitude, and phase parts of the Split-CReLU function.

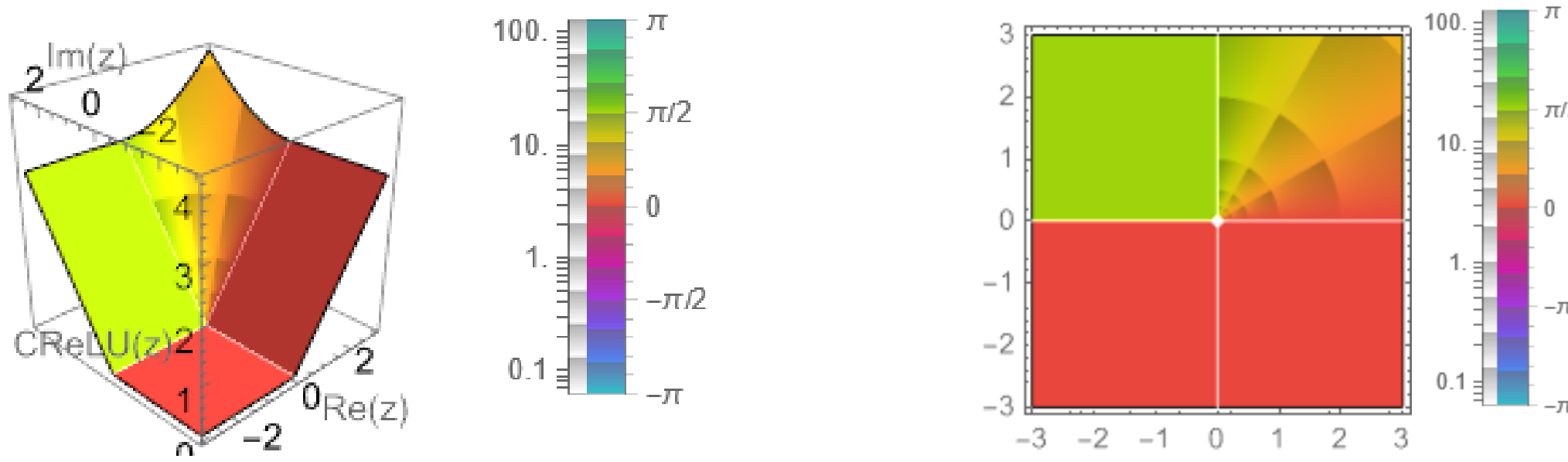

**Figure 9.20.** Left panel: ComplexPlot3D generates a 3D plot of Abs[Split-CReLU] colored by arg[Split-CReLU] over the complex rectangle with corners $-3-3i$ and $3+3i$. Using "CyclicLogAbsArg" to cyclically shade colors to give the appearance of contours of constant Abs[Split-CReLU] and constant arg[Split-CReLU]. Right panel: ComplexPlot generates a plot of arg[Split-CReLU] over the complex rectangle with corners $-3-3i$ and $3+3i$. Using "CyclicLogAbsArg" to cyclically shade colors to give the appearance of contours of constant Abs[Split-CReLU] and constant arg[Split-CReLU].

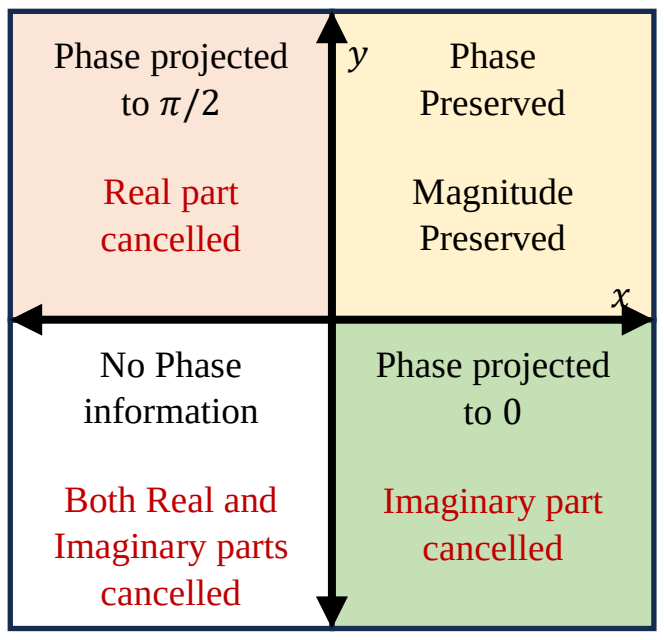

**Figure 9.21.** Phase information encoding for Split-CReLU Function. The $x$-axis represents the real part and the $y$-axis axis represents the imaginary part. Split-CReLU discriminates the complex information into 4 regions where in two of which, phase information is projected to 0 and $\pi/2$ and not canceled.

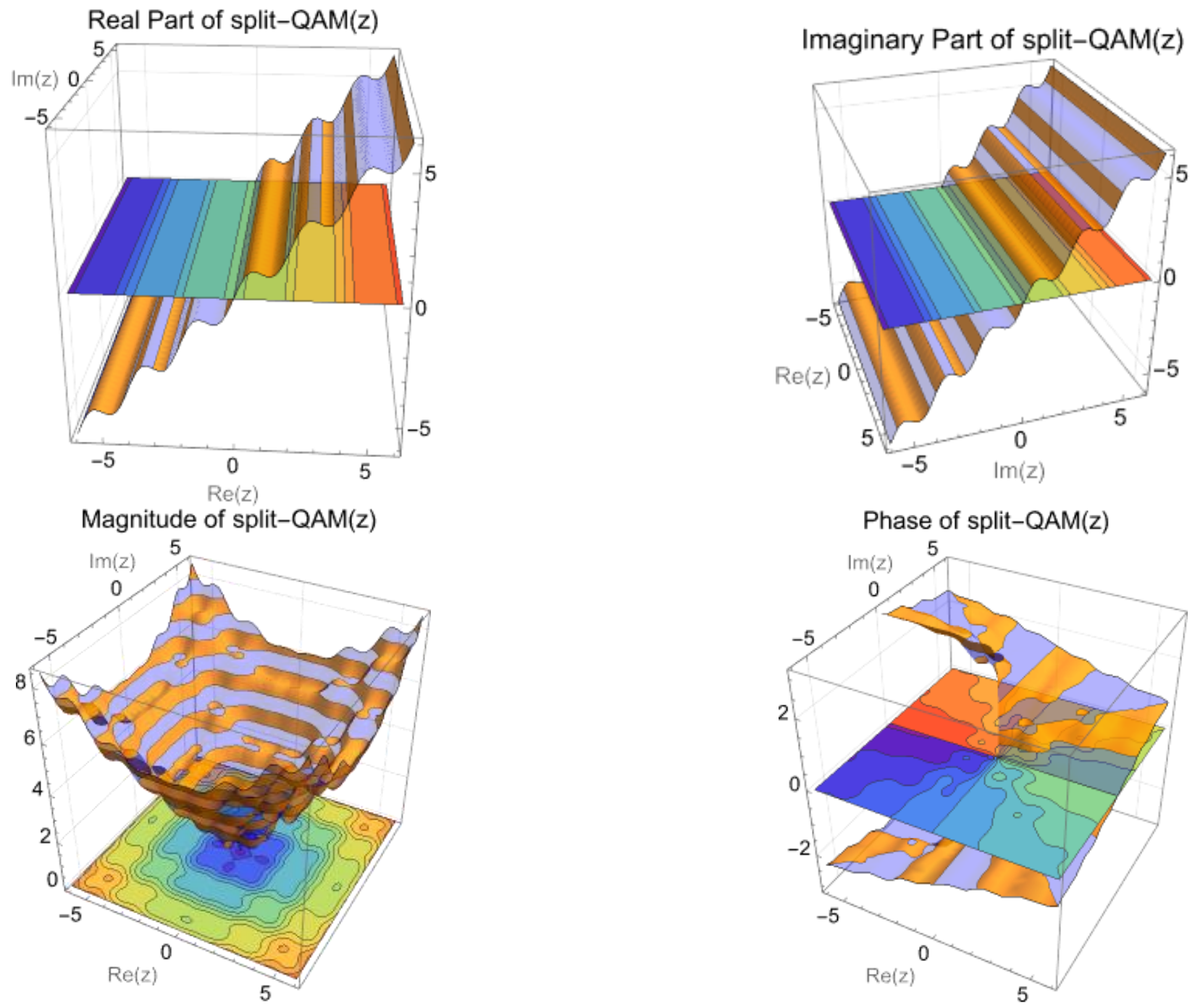

**Figure 9.22.** Visualizations of the real, imaginary, magnitude, and phase parts of the Split-QAM function.

- Split-CReLU discriminates the complex information into 4 regions where in two of which, phase information is projected and not canceled. This allows Split-CReLU to discriminate information easier with respect to phase information than the other AFs, see Figures 9.20 and 9.21.
- CReLU has flexibility manipulating phase as it can either set it to zero or $\pi/2$, or even delete the phase information (when both real and imaginary parts are canceled) at a given level of depth in the network.

### 9.6.8 Split-QAM Function

Split-Quadrature Amplitude-Modulation (Split-QAM) [293] AF was defined by

$$\begin{aligned}\sigma_{\text{Split-QAM}}(z) &= \sigma^{\Re}_{\text{Split-QAM}}(x) + i\sigma^{\Re}_{\text{Split-QAM}}(y) \\ &= \Re(z) + \alpha \sin(\pi\, \Re(z)) + i[\Im(z) + \alpha \sin(\pi\, \Im(z))],\end{aligned} \tag{9.109.1}$$

where $z = x + iy$ and $\sigma^{\Re}_{\text{Split-QAM}}$ is a real-valued function,

$$\sigma^{\Re}_{\text{Split-QAM}}(u) = u + \alpha \sin(\pi u), \tag{9.109.2}$$

for any $u \in \mathbb{R}$ and the slope parameter $\alpha$ that determines the degree of nonlinearity is a positive real constant.

**Remarks:**

- Split-QAM CVAF was developed to deal with Quadrature Amplitude-Modulation (QAM) signals of any constellation sizes.
- The multi-saturation characteristic of $S$-shape that the output of this function shows is pertinent to $M$-ary QAM signals with discrete amplitudes and makes the network robust to noise because $\Delta y$ is small for large $\Delta x$, see Figure 9.22.
- The Split-QAM AF modulates both the real and imaginary parts of the complex input using sinusoidal terms. The parameter $\alpha$ controls the strength of this modulation. The parameter $\alpha$ requires careful tuning based on the task at hand.

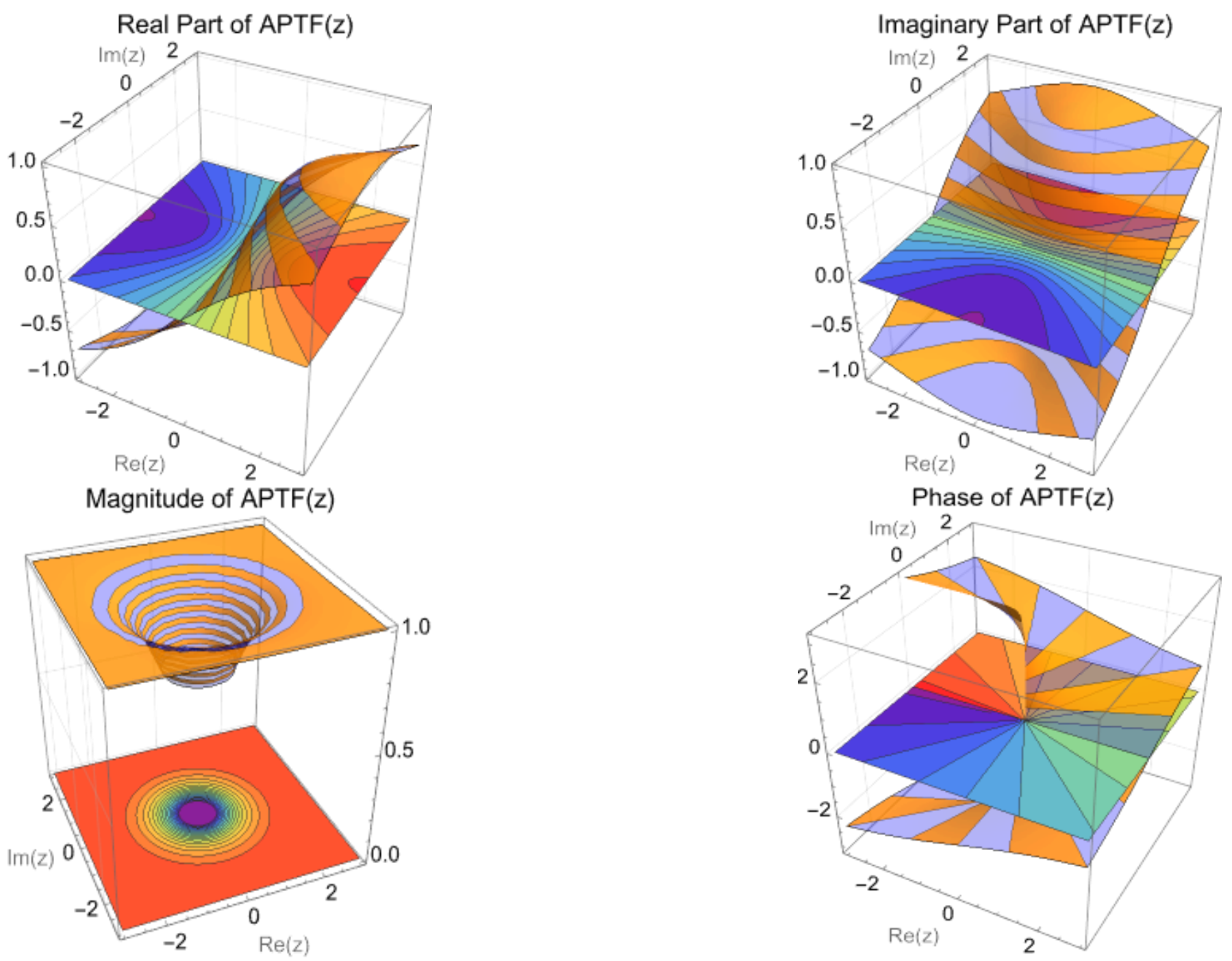

**Figure 9.23.** Visualizations of the real, imaginary, magnitude, and phase parts of the Amplitude-phase-type function.

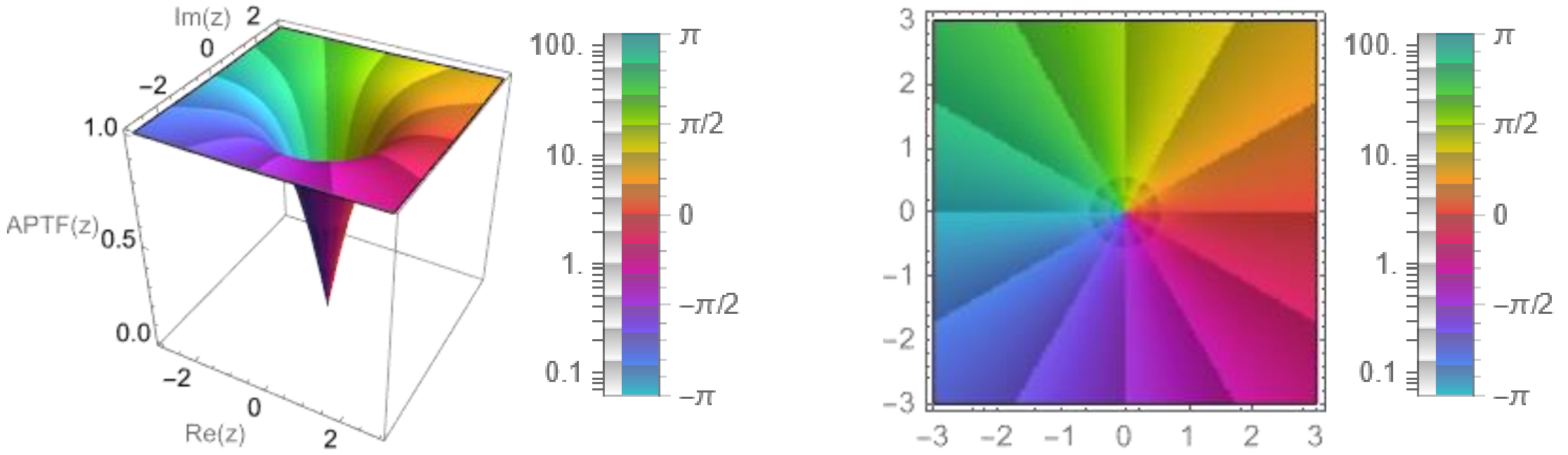

**Figure 9.24.** Left panel: The ComplexPlot3D generates a 3D plot of Abs[$\sigma_{\mathrm{APTF}}(z)$] colored by arg[$\sigma_{\mathrm{FATAF}}(z)$] over the complex rectangle with corners $z_{min} = -3 - 3i$ and $z_{max} = 3 + 3i$. Using "CyclicLogAbsArg" to cyclically shade colors to give the appearance of contours of constant Abs[$\sigma_{\mathrm{APTF}}(z)$] and constant arg[$\sigma_{\mathrm{APTF}}(z)$]. Right panel: ComplexPlot generates a plot of arg[$\sigma_{\mathrm{APTF}}(z)$] over the complex rectangle with corners $-3 - 3i$ and $3 + 3i$. Using "CyclicLogAbsArg" to cyclically shade colors to give the appearance of contours of constant Abs[$\sigma_{\mathrm{APTF}}(z)$] and constant arg[$\sigma_{\mathrm{APTF}}(z)$].

- This type of AF could be used in NNs for communication systems where the modulation of the AF plays a role in signal processing.
- Depending on the application, the additional computational cost introduced by the sinusoidal terms may need to be considered.

### 9.6.9 Amplitude-Phase-Type Function

The Amplitude-Phase-Type function (APTF) [315-321] is expressed as

$$\sigma_{\mathrm{APTF}}(z) = \mathrm{Tanh}(|z|)\ e^{i\,\arg(z)}, \tag{9.110}$$

where $|z|$ denotes the magnitude (amplitude) and $\arg(z)$ represents the phase of the complex variable $z$.

**Remarks:**

- Unlike the real-imaginary-type AF, this formulation emphasizes saturation in amplitude while keeping the phase unchanged.
- The term $\tanh(|z|)$ in the APTF emphasizes saturation in amplitude. The hyperbolic tangent function squashes the input values between $-1$ and $1$, promoting saturation when the amplitude of $z$ is high.
- The visual representation in Figures 9.23 and 9.24 illustrates the distinctive shape of the amplitude-phase-type AF, emphasizing its point symmetry about the origin $(0, i0)$, which is clearly observed in amplitude and phase figures.
- The symmetry around the origin implies that the function's behavior remains consistent regardless of how the coordinate axes are oriented. In other words, if you rotate the entire coordinate system, the APTF will still exhibit the same shape and properties (amplitude).
- In scenarios where information processing involves rotation around the origin, the independence from the orientation of the axes becomes advantageous. The APTF is not affected by the angle of rotation, making it particularly suitable for tasks where the input data or features may undergo rotational transformations.
- The amplitude-phase-type AF finds its niche in the processing of wave-related information. By associating the wave amplitude with the amplitude of the complex variable and the wave phase with the phase of the neural variable, this CVAF becomes a valuable tool for applications involving electromagnetic waves, light waves, sonic waves, ultrasonic waves, quantum waves, and other wave-related phenomena.
- Beyond its utility in wave processing, the amplitude-phase-type AF is well-suited for applications where point symmetry concerning the origin is essential.

### 9.6.10 Amplitude- Phase Sigmoidal Function

Another common class of non-analytic CVAFs is the Amplitude-Phase Sigmoidal Function (APSF) [273] popularized by

$$\sigma_{\text{APSF}}(z) = \frac{z}{a + \frac{1}{b}|z|} = \left(\frac{b}{ab + |z|}\right) z, \tag{9.111}$$

where $a$ and $b$ are real positive constants.

**Remarks:**

- This function has the property of mapping a point $z = x + iy = (x, y)$ on the complex plane to a unique point $\sigma_{\text{APSF}}(z) = (\frac{x}{a+|z|/b}, \frac{y}{a+|z|/b})$ on the open disc $\{z: |z| < b\}$.
- The division by $(ab + |z|)$ in the denominator ensures that the magnitude is scaled such that the output lies in the open disc $\{z: |z| < b\}$. This is because $ab + |z|$ is always greater than $|z|$ due to the positivity of both $a$ and $b$. The inequality $ab + |z| > |z|$ guarantees that $1 > \frac{|z|}{ab+|z|}$, which implies $b > \left(\frac{b}{ab+|z|}\right)|z| = |\sigma_{\text{APSF}}(z)|$. Therefore, the resulting magnitude is always scaled down to be less than $b$. The APSF is designed to map complex numbers to points within a specific open disc, and its magnitude is scaled to ensure that the output always lies within that disc.
- The APSF has the property of monotonically squashing the magnitude $|z|$ to a value $|\sigma_{\text{APSF}}(z)|$ within the interval $[0, b)$. This means that as the magnitude of $z$ increases, the output magnitude $|\sigma_{\text{APSF}}(z)|$ increases as well, but it is always constrained to be less than $b$. The Sigmoid function maps a real number $x$ to a point in the interval $(0,1)$. The hyperbolic tangent function maps a real number $x$ to a point in the interval $(-1,1)$. In a similar vein, the APSF maps complex numbers to a point with magnitude in the interval $[0, b)$, exhibiting a squashing behavior akin to the Sigmoid and hyperbolic tangent functions. So, the APSF can be seen as the natural generalization of real-valued squashing functions such as the Sigmoid.

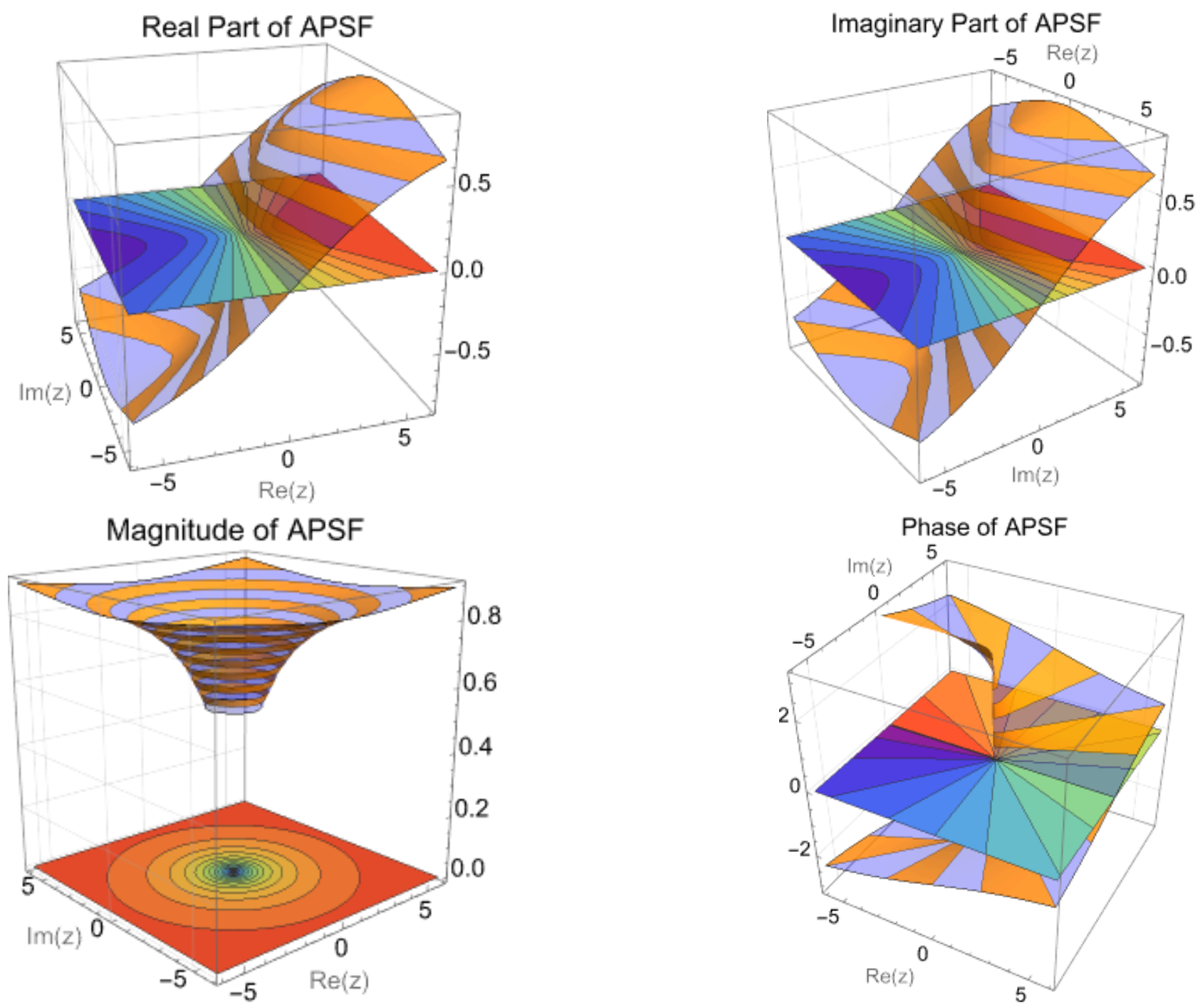

**Figure 9.25.** Visualizations of the real, imaginary, magnitude, and phase parts of the Amplitude-Phase Sigmoidal function.

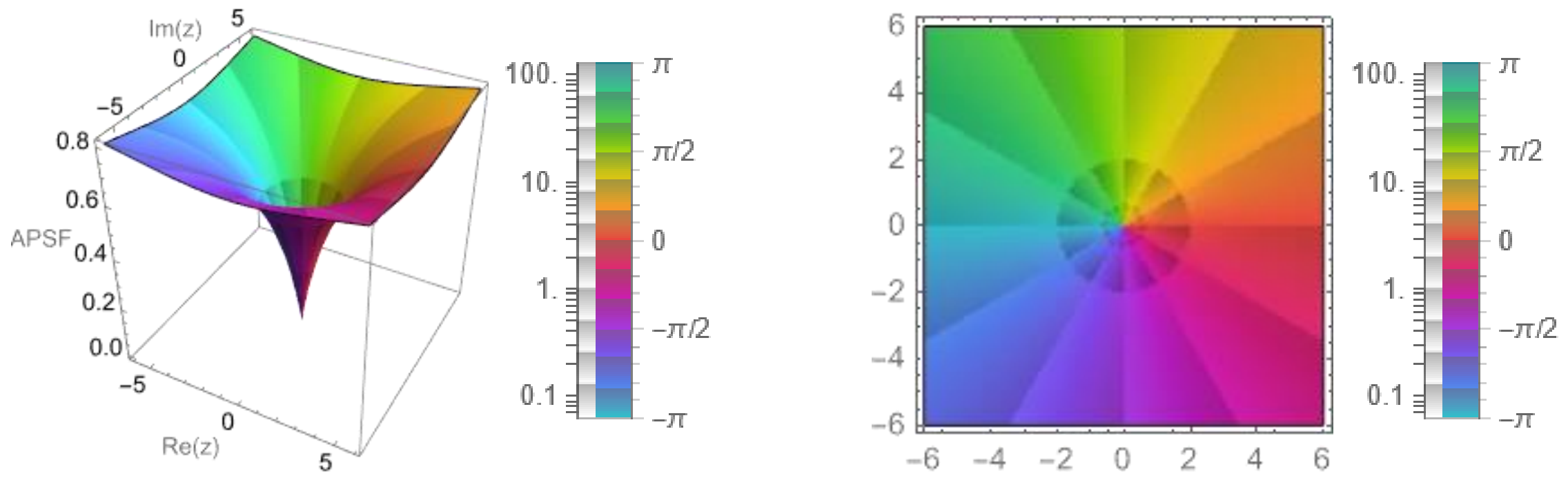

**Figure 9.26.** Left panel: The ComplexPlot3D generates a 3D plot of Abs[$\sigma_{\text{APSF}}(z)$] colored by arg[$\sigma_{\text{APSF}}(z)$] over the complex rectangle with corners $z_{min} = -6 - 6i$ and $z_{max} = 6 + 6i$. Using "CyclicLogAbsArg" to cyclically shade colors to give the appearance of contours of constant Abs[$\sigma_{\text{APSF}}(z)$] and constant arg[$\sigma_{\text{APSF}}(z)$]. Right panel: ComplexPlot generates a plot of arg[$\sigma_{\text{APSF}}(z)$] over the complex rectangle with corners $-6 - 6i$ and $6 + 6i$. Using "CyclicLogAbsArg" to cyclically shade colors to give the appearance of contours of constant Abs[$\sigma_{\text{APSF}}(z)$] and constant arg[$\sigma_{\text{APSF}}(z)$].(with parameters; $a = 2$ and $b = 1$)

- The visual representation in Figures 9.25 and 9.26 illustrates the distinctive shape of the APSF, emphasizing its point symmetry about the origin $(0, i0)$, which is clearly observed in amplitude and phase figures.
- The parameter $a$ in the APSF controls the steepness of the function. Specifically, it influences how quickly the function approaches its limiting value of $b$ as the magnitude of $z$ increases.
  - For larger values of $a$, the function approaches its limiting value more gradually, resulting in a smoother transition.
  - For smaller values of $a$, the function approaches its limiting value more rapidly, leading to a steeper transition.

- The steepness parameter $a$ provides a way to adjust the sensitivity of the function to changes in the input, allowing for customization based on the specific characteristics desired for a given application or task in a NN or mathematical model.
- The term $z$ at the end of the APSF expression plays a crucial role in preserving the phase of the input. Multiplying $z$ by $\frac{b}{ab+|z|}$ scales the magnitude without changing the angle $\theta$, preserving the phase information.
- The partial derivatives $\frac{\partial u}{\partial x}, \frac{\partial u}{\partial y}, \frac{\partial v}{\partial x}$ and $\frac{\partial v}{\partial y}$ are

$$\frac{\partial u}{\partial x} = \begin{cases} \frac{b(y^2 + ab|z|)}{|z|(ab+|z|)^2}, & |z| \neq 0 \\ \frac{1}{a}, & |z| = 0 \end{cases}, \tag{9.112.1}$$

$$\frac{\partial u}{\partial y} = \begin{cases} -\frac{bxy}{|z|(ab+|z|)^2}, & |z| \neq 0 \\ 0, & |z| = 0 \end{cases}, \tag{9.112.2}$$

$$\frac{\partial v}{\partial x} = \begin{cases} \frac{bxy}{|z|(ab+|z|)^2}, & |z| \neq 0 \\ 0, & |z| = 0 \end{cases}, \tag{9.112.3}$$

$$\frac{\partial v}{\partial y} = \begin{cases} \frac{b(x^2 + ab|z|)}{|z|(ab+|z|)^2}, & |z| \neq 0 \\ \frac{1}{a}, & |z| = 0 \end{cases}. \tag{9.112.4}$$

- The special definitions of the partial derivatives when $|z| = 0$, i.e., at $z = (0,0)$, correspond to their limits as $z \to (0,0)$. Being thus defined, the singularities at the origin are removed and all partial derivatives exist and are continuous for all $z \in \mathbb{C}$.

### 9.6.11 Complex Cardioid

In their 1992 paper [273], Georgiou and Koutsougeras presented the activation, $\sigma_{\text{APSF}}(z)$, that attenuates the magnitude of the signal while preserving the phase. This CVAF was designed to be particularly useful when it is important to maintain the phase information of the input signal while adjusting the magnitude in a controlled manner. In [322], the authors refer to this CVAF as SigLog as it modifies the magnitude by applying the Sigmoid of the log of the magnitude:

$$\sigma_{\text{SigLog}}(z) = \frac{z}{1+|z|} = \sigma_{\text{Sigmoid}}(\log(|z|))e^{i \arg z}, \tag{9.113.1}$$

$$\sigma_{\text{Sigmoid}}(z) = \frac{1}{1+e^{-z}}. \tag{9.113.2}$$

**Remarks:**

- $\sigma_{\text{SigLog}}(z)$ is a special case from $\sigma_{\text{APSF}}(z)$ with $a = 1$ and $b = 1$.
- $|z|$ represents the magnitude of the complex number $z$, which is calculated as the distance of $z$ from the origin in the complex plane.
- The division by $(1 + |z|)$ ensures that the magnitude is scaled such that the output lies in the unit circle. This is because the denominator $(1 + |z|)$ is always greater than $|z|$, so the resulting magnitude is always scaled down to be less than 1.
- When $|z|$ is small, the attenuation is minimal, and when $|z|$ is large, the attenuation becomes more significant. This implies that the magnitude of the complex number is attenuated in a manner that depends on its original magnitude.
- The combination of the logarithmic function and the Sigmoid activation serves to control the magnitude attenuation. Let us prove the equality, $\frac{z}{1+|z|} = \sigma_{\text{sigmoid}}(\log(|z|))e^{i \arg z}$,

$$\begin{aligned}\sigma_{\text{sigmoid}}(\log(|z|))e^{i\arg z} &= \frac{1}{1+e^{-\log(|z|)}}e^{i\arg z} \\ &= \frac{1}{1+\frac{1}{|z|}}e^{i\arg z} \\ &= \frac{|z|}{|z|+1}e^{i\arg z} = \frac{z}{(|z|+1)}. \end{aligned} \quad (9.114)$$

Complex Cardioid AF [322] is CVAF sensitive to the input phase rather than the input magnitude. The Complex Cardioid is defined as:

$$\sigma_{\text{CCardioid}}(z) = \frac{1}{2}(1+\cos[\arg z])z, \quad (9.115.1)$$

$$\frac{\partial}{\partial z}\sigma_{\text{CCardioid}}(z) = \frac{1}{2} + \frac{1}{2}\cos[\arg z] + \frac{i}{4}\sin[\arg z]. \quad (9.115.2)$$

**Remarks:**

- The function $\cos[\arg z]$ introduces phase sensitivity. The cosine function oscillates between $-1$ and $1$, meaning it varies based on the angle of $z$. This phase sensitivity is a key feature of Complex Cardioid activation.
- The factor $\frac{1}{2}(1+\cos[\arg z])$ is responsible for amplitude modulation. When $\cos[\arg z] = 1$, the factor becomes 1, resulting in no attenuation of the magnitude. When $\cos[\arg z] = -1$, the factor becomes 0, leading to complete attenuation. The modulation is controlled by the cosine term, which depends on the phase of the input complex number.
- The $z$ term at the end ensures that the output phase remains the same as the input phase.
- Multiplying $z$ does not alter the phase but contributes to scaling the final output.
- The overall effect is that the output magnitude is attenuated based on the input phase ($\arg(z)$), while the output phase remains equal to the input phase.
- The modulation of magnitude based on phase sensitivity might offer benefits in certain types of data where the relationship between magnitude and phase is significant. This CVAF could be useful in NNs where preserving phase information is crucial, such as in signal processing tasks or applications involving waveforms.
- The real- and imaginary- part of $\sigma_{\text{CCardioid}}(z)$ AF, as well as the amplitude and the phase of the AF are visually depicted in Figures 9.27 and 9.28 showcasing its characteristic shape.
- Input values that lie on the positive real axis are scaled by one. This means there is no attenuation; the magnitude remains unchanged. For the positive real axis, $z$ is a real number with zero imaginary part, i.e., $z = x$ where $x > 0$. The argument $\arg(z)$ for a positive real number is zero because it lies directly along the positive real axis. This means that $\cos[\arg z] = \cos(0) = 1$. The amplitude modulation factor is 1 for the positive real axis. The Complex Cardioid AF preserves the magnitude of input values on the positive real axis, resulting in an output that is equal to the input for these specific values.
- Input values that lie on the negative real axis are scaled by zero. This implies complete attenuation; the magnitude is reduced to zero. For the negative real axis, $z$ is a real number with zero imaginary part, i.e., $z = -x$ where $x > 0$. The argument $\arg(z)$ for a negative real number is $\pi$ because it lies directly along the negative real axis. This means that $\cos[\arg z] = \cos(\pi) = -1$. The amplitude modulation factor is 0 for the negative real axis. The Complex Cardioid AF effectively "turns off" or attenuates the magnitude for input values on the negative real axis, resulting in an output of zero for these specific values.
- For input values with nonzero imaginary components, the scaling factor varies gradually from one to zero. As the complex number rotates in phase from the positive real axis towards the negative real axis, the cosine term ($\cos[\arg(z)]$) changes accordingly, leading to a gradual reduction in the scaling factor.
- When the input values are restricted to real values, the Complex Cardioid function is simply the ReLU AF. When the input values are restricted to real numbers (zero imaginary component), the argument $\arg(z)$ becomes 0 for positive real numbers and $\pi$ for negative real numbers. $\sigma_{\text{cardioid}}(z) = z,\ z > 0$ and $\sigma_{\text{cardioid}}(z) = 0,\ z < 0$. Hence, Complex Cardioid AF is a phase sensitive complex extension of ReLU.

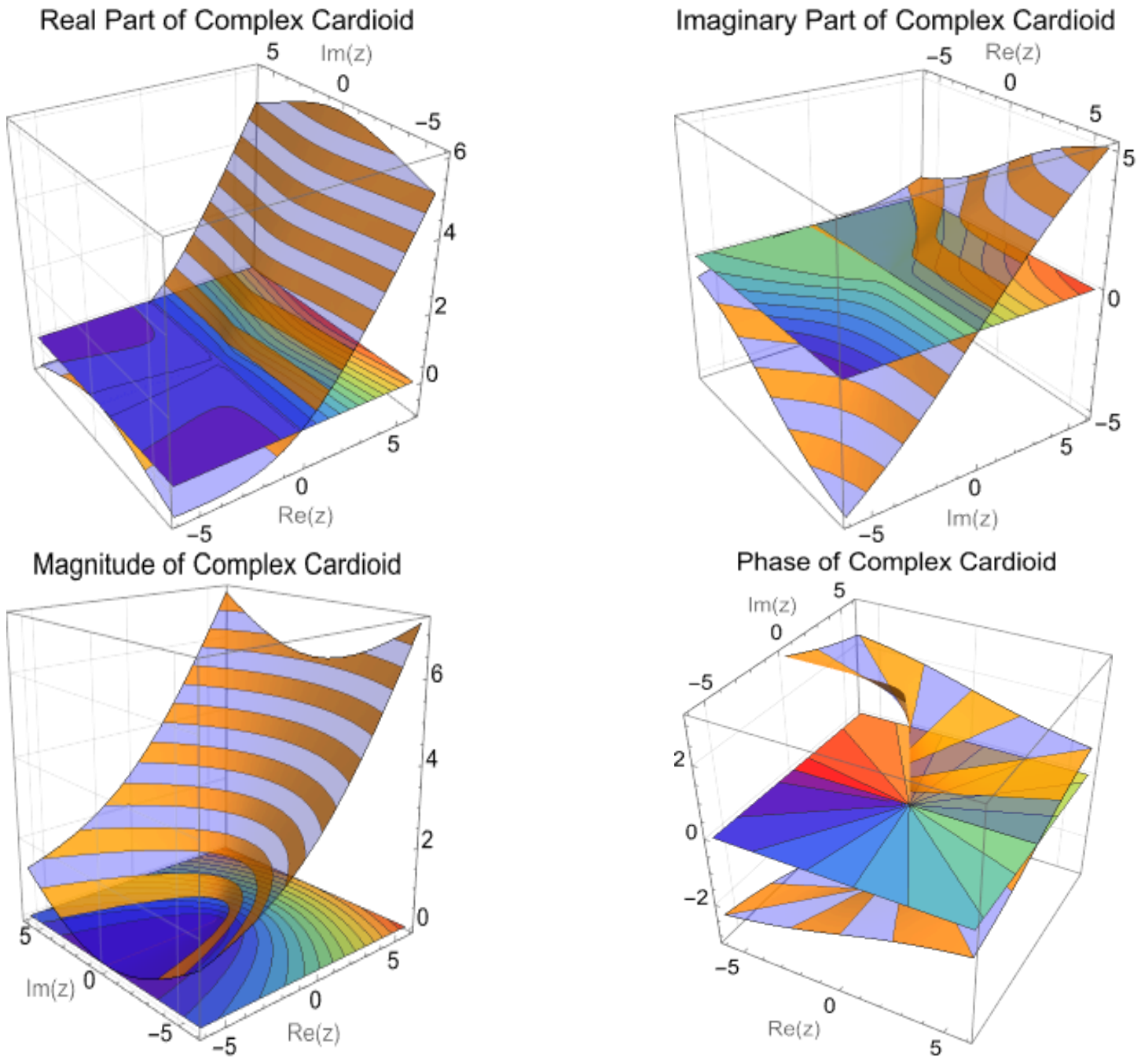

**Figure 9.27.** Real part, imaginary part, amplitude, and phase of the $\sigma_{\text{CCardioid}}(z)$ AF.

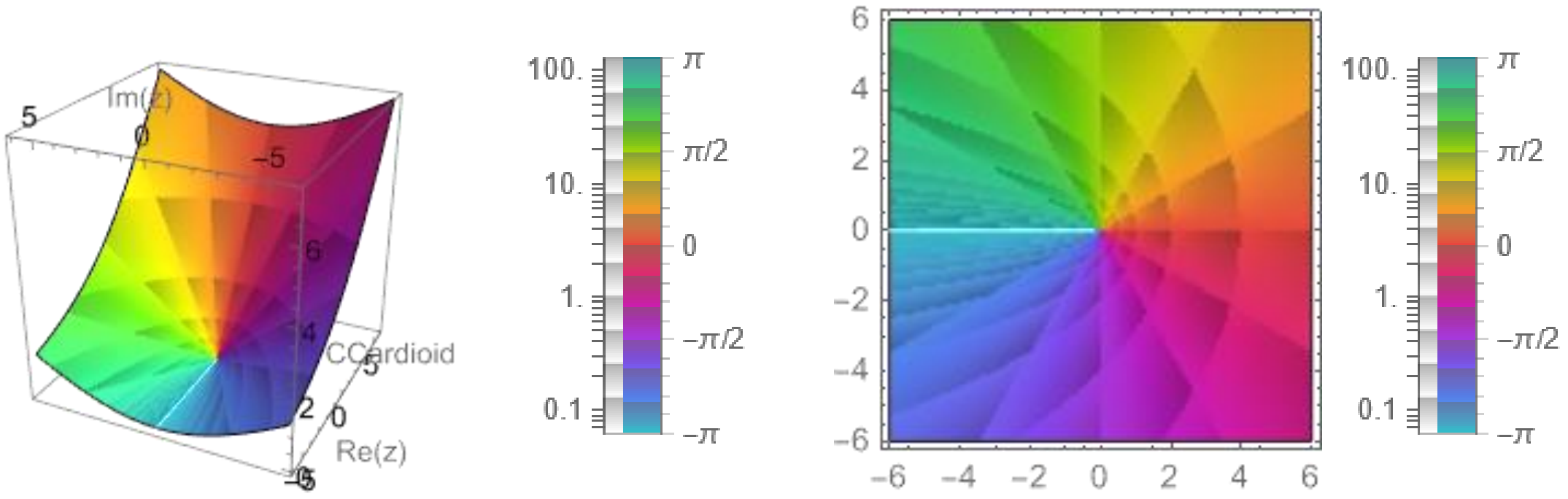

**Figure 9.28.** Left panel: The ComplexPlot3D generates a 3D plot of Abs[$\sigma_{\text{CCardioid}}(z)$] colored by arg[$\sigma_{\text{CCardioid}}(z)$] over the complex rectangle with corners $z_{min} = -6 - 6i$ and $z_{max} = 6 + 6i$. Using "CyclicLogAbsArg" to cyclically shade colors to give the appearance of contours of constant Abs[$\sigma_{\text{CCardioid}}(z)$] and constant arg[$\sigma_{\text{CCardioid}}(z)$]. Right panel: ComplexPlot generates a plot of arg[$\sigma_{\text{CCardioid}}(z)$] over the complex rectangle with corners $-6 - 6i$ and $6 + 6i$. Using "CyclicLogAbsArg" to cyclically shade colors to give the appearance of contours of constant Abs[$\sigma_{\text{CCardioid}}(z)$] and constant arg[$\sigma_{\text{CCardioid}}(z)$].

### 9.6.12 modReLU

modReLU [313, 314, 323-325] is a pointwise nonlinearity, $\sigma_{\text{modReLU}}(z)\colon \mathbb{C} \to \mathbb{C}$, which affects only the absolute value of a complex number, defined as

$$\sigma_{\text{modReLU}}(z) = \begin{cases} \left(\dfrac{|z| + b}{|z|}\right) z, & |z| + b \geq 0, \\ 0, & |z| + b < 0, \end{cases} \tag{9.116.1}$$

where $b \in \mathbb{R}$ is a bias parameter of the nonlinearity. Note that the modReLU is similar to the ReLU in spirit, in fact more concretely,

$$\sigma_{\text{modReLU}}(z) = \sigma_{\text{ReLU}}(|z| + b)\frac{z}{|z|} = \max(|z| + b, 0)\frac{z}{|z|} = \sigma_{\text{ReLU}}(|z| + b)e^{i\theta}. \tag{9.116.2}$$

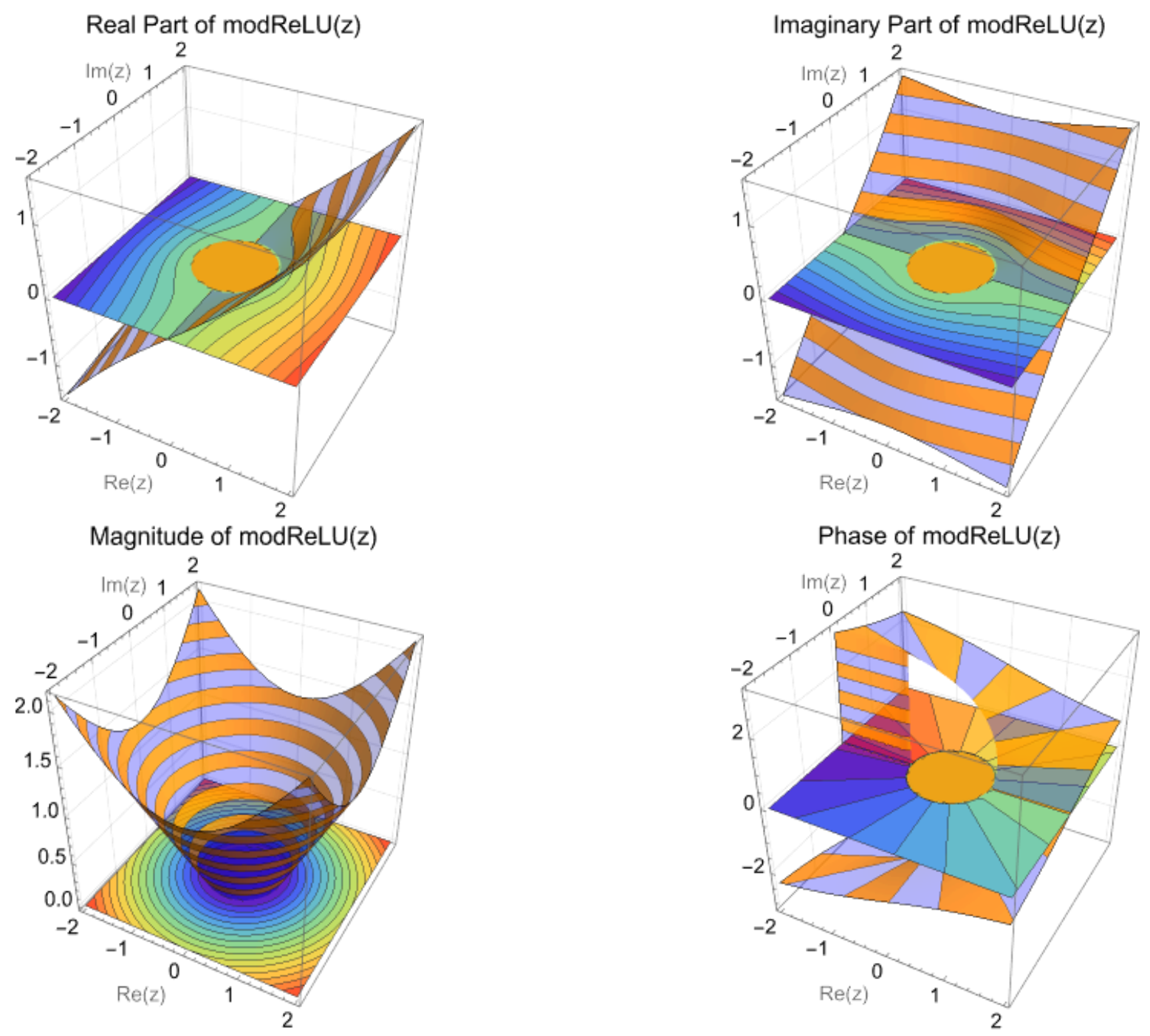

**Figure 9.29.** Real part, imaginary part, amplitude, and phase of the $\sigma_{\text{modReLU}}(z)$ AF. ($b = -0.7$)

**Remarks:**

- The $z$ term at the end of $\left(\frac{|z|+b}{|z|}\right)z$ is responsible for preserving the phase of the complex number. Since $z$ is multiplied back, it doesn't alter the phase of the complex number. This ensures that the output phase remains the same as the input phase.
- The term $\frac{|z|+b}{|z|}$ is responsible for amplitude modulation. This part scales the magnitude of the complex number while introducing a bias $b$.
- The ReLU operation $\sigma_{\text{ReLU}}(|z| + b) = \max(|z| + b, 0)$ ensures that the function becomes zero for negative inputs ($|z| + b < 0$) and retains the amplitude-modulated $z$ for non-negative inputs ($|z| + b \geq 0$).
- As $|z|$ is always positive, a bias $b$ is introduced in order to create a "dead zone" of radius $b$ around the origin 0 where the neuron is inactive, and outside of which it is active.
- modReLU does not satisfy the Cauchy-Riemann equations and thus is not holomorphic.
- Case $b < 0$, see Figure 9.29: In this case, the conditions are as follows:
  - $|z| + b \geq 0$: This condition is satisfied when $|z| \geq -b$.
  - $|z| + b < 0$: This condition is satisfied when $|z| < -b$.

  Therefore, for $b < 0$, the modReLU function is:

$$\sigma_{\text{modReLU}}(z) = \begin{cases} \left(\dfrac{|z| + b}{|z|}\right)z, & |z| \geq -b, \\ 0, & |z| < -b. \end{cases} \tag{9.117}$$

  This means that when $b < 0$, the modReLU function scales the complex number $z$ by a factor determined by the modified magnitude only when the original magnitude $|z|$ is greater than $-b$. Otherwise, the output is 0.
- Case $b > 0$: In this case, the condition $|z| + b \geq 0$ is always satisfied, as $|z|$ is always non-negative, and adding a positive value $b$ to it results in a non-negative quantity. Therefore, the modReLU function simplifies to: $\sigma_{\text{modReLU}}(z) = \big((|z| + b)/|z|\big)\, z$. This means that when $b > 0$, the modReLU function scales the complex number $z$ by a factor determined by the modified magnitude $(|z| + b/|z|)$. The "dead zone" has disappeared.
- Figure 9.30 represents the phase information encoding for the modReLU AF. Figure 9.31 compares the amplitudes of modReLU in two cases $b < 0$ and $b > 0$.

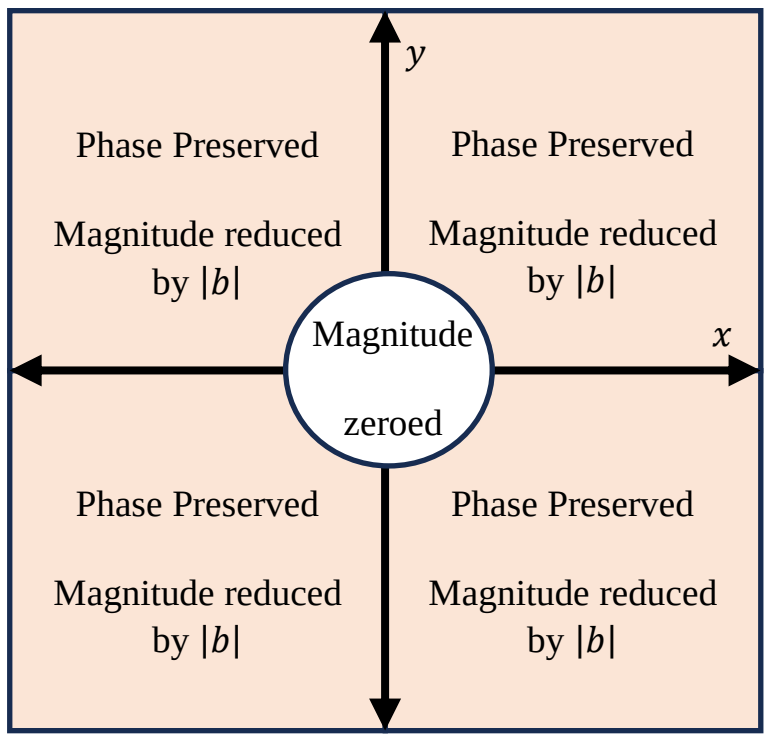

**Figure 9.30.** Phase information encoding for the modReLU AF. The $x$-axis represents the real part and the $y$-axis axis represents the imaginary part. The case where $b < 0$ for modReLU. The radius of the white circle is equal to $|b|$. In the case where $b \geq 0$, the whole complex plane would be preserving both phase and magnitude information and the whole plane would have been colored with orange. We can see the modReLU, the complex representation is discriminated into two regions, i.e., the one that preserves the whole complex information (colored in orange) and the one that cancels it (colored in white).

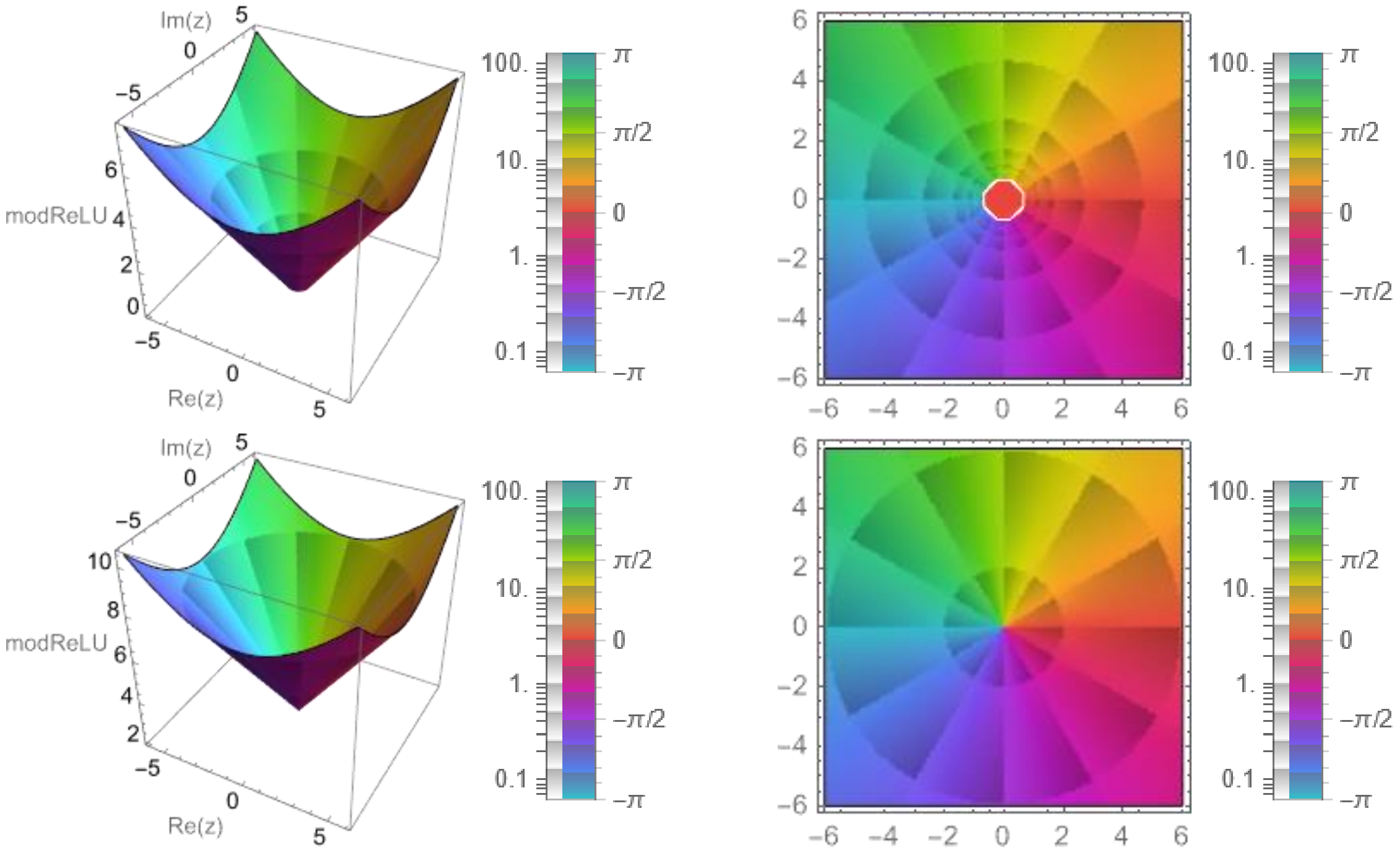

**Figure 9.31.** Upper panel left: The ComplexPlot3D generates a 3D plot of Abs[$\sigma_{\text{modReLU}}(z)$] with $b = -0.7$ colored by arg[$\sigma_{\text{modReLU}}(z)$] over the complex rectangle with corners $z_{min} = -6 - 6i$ and $z_{max} = 6 + 6i$. Using "CyclicLogAbsArg" to cyclically shade colors to give the appearance of contours of constant Abs[$\sigma_{\text{modReLU}}(z)$] and constant arg[$\sigma_{\text{modReLU}}(z)$]. Upper panel right: ComplexPlot generates a plot of arg[$\sigma_{\text{modReLU}}(z)$] with $b = -0.7$ over the complex rectangle with corners $-6 - 6i$ and $6 + 6i$. Using "CyclicLogAbsArg" to cyclically shade colors to give the appearance of contours of constant Abs[$\sigma_{\text{modReLU}}(z)$] and constant arg[$\sigma_{\text{modReLU}}(z)$]. Note that, the bias $b$ is introduced in order to create a “dead zone” of radius $|b|$ around the origin 0 where the neuron is inactive, and outside of which it is active. Lower panel: The 3D and 2D plots show the Abs[$\sigma_{\text{modReLU}}(z)$] with $b = 2$. Note that, the “dead zone” has disappeared.

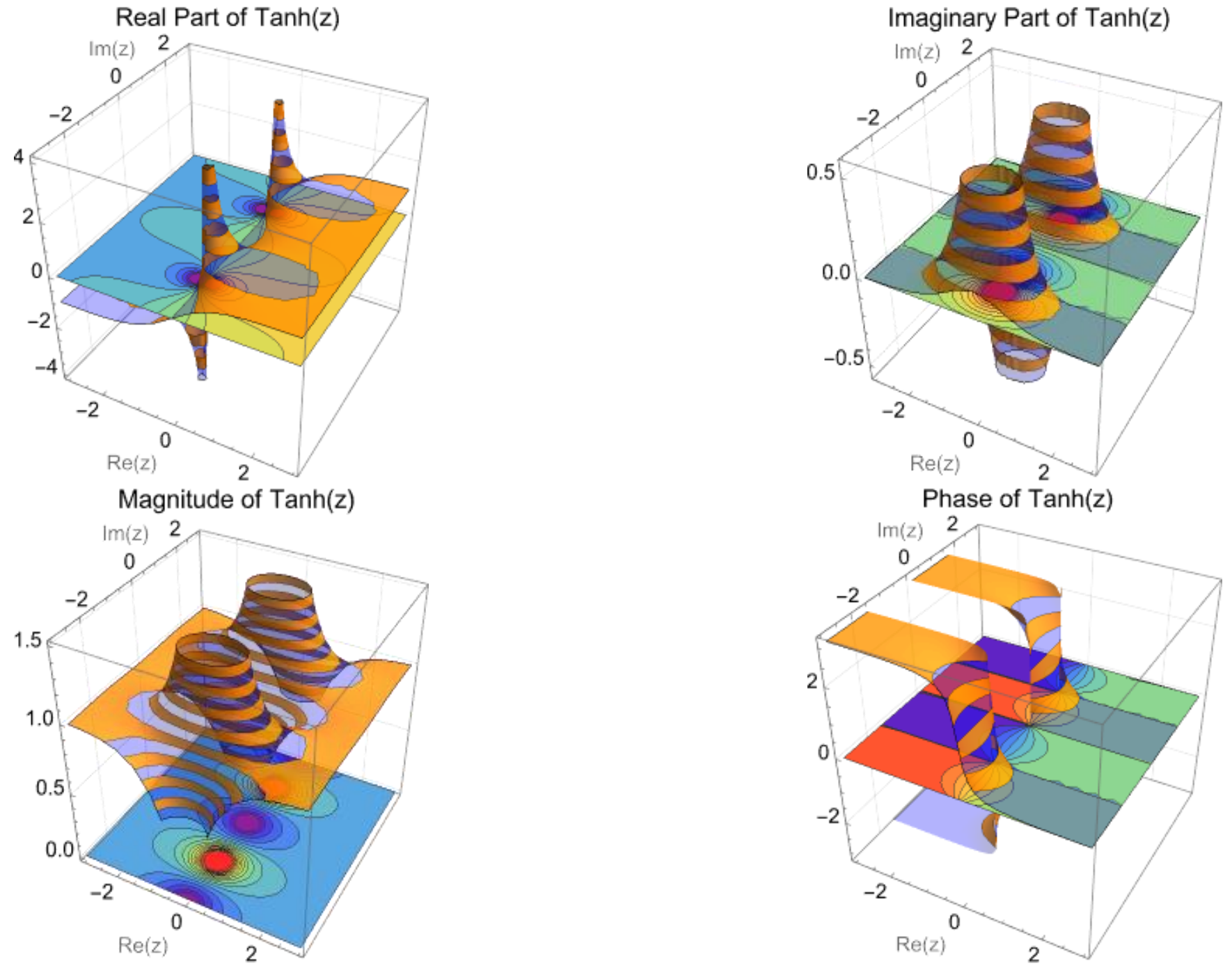

**Figure 9.32.** Complex $\text{Tanh}(z)$ as a function of complex variables, real-part, imaginary-part, amplitude, and phase.

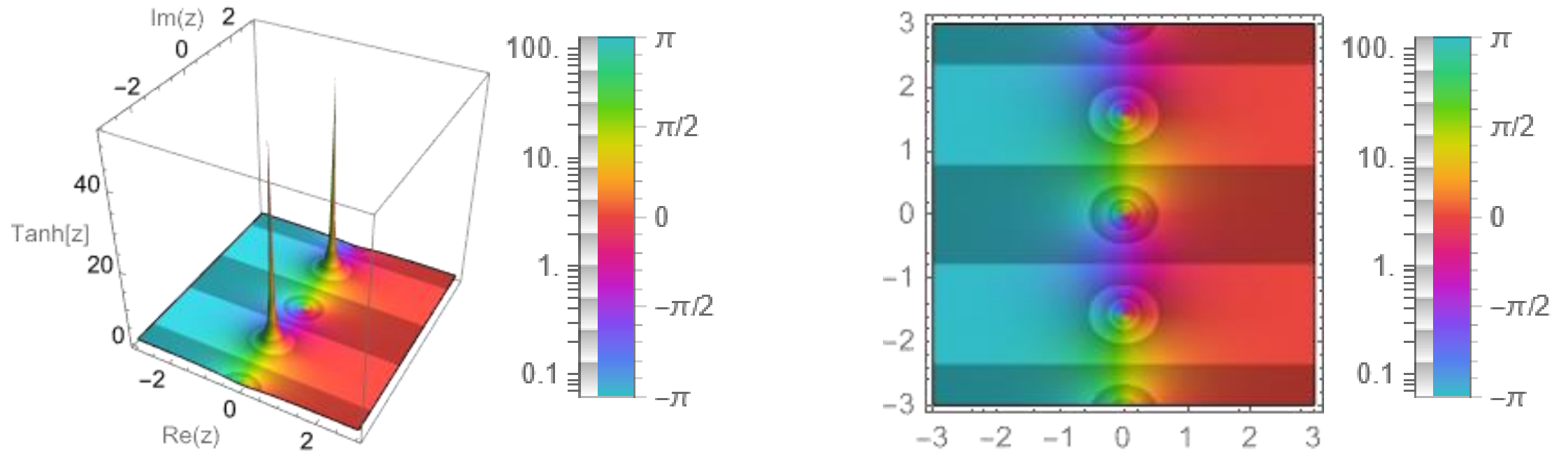

**Figure 9.33.** Left panel: The ComplexPlot3D generates a 3D plot of Abs[$\sigma_{\text{FCTanh}}(z)$] colored by arg[$\sigma_{\text{FCTanh}}(z)$] over the complex rectangle with corners $z_{min} = -3 - 3i$ and $z_{max} = 3 + 3i$. Using "CyclicLogAbs" to cyclically shade colors to give the appearance of contours of constant Abs[$\sigma_{\text{FCTanh}}(z)$]. Right panel: ComplexPlot generates a plot of arg[$\sigma_{\text{FCTanh}}(z)$] over the complex rectangle with corners $-3 - 3i$ and $3 + 3i$.

### 9.6.13 Fully Complex Tanh Function

The hyperbolic tangent function, $\text{Tanh}(z)$, is easily defined as the ratio between the hyperbolic sine and the cosine functions (or expanded, as the ratio of the half-difference and half-sum of two exponential functions in the points $z$ and $-z$). The formula for the hyperbolic tangent of a complex number $z$ is given by:

$$\sigma_{\text{FCTanh}}(z) = \text{Tanh}(z) = \frac{e^z - e^{-z}}{e^z + e^{-z}} = \frac{\text{Sinh}\, z}{\text{Cosh}\, z}, \tag{9.118.1}$$

$$\frac{\partial}{\partial z}\sigma_{\text{FCTanh}}(z) = \text{Sech}^2(z). \tag{9.118.2}$$

Complex hyperbolic tangent function was proposed as a “Fully Complex” AF [274, 276].

**Remarks:**

- Note that singular points do exist at $z = (n + 1/2)\pi i$, $\forall n \in \mathbb{N}$, where $\cosh z$ has zeros, the denominator of the last formula equals zero.
- Figures 9.32 and 9.33 show the shape of the function, which is far away from the "saturation" feature. While this function is differentiable at almost every point, it diverges to infinity, deviating significantly from the expected saturation behavior.
- Instead of boundedness, $\mathrm{Tanh}(z)$ function has well-defined but not necessarily bounded first order derivatives almost everywhere in $\mathbb{C}$. Since they are bounded almost everywhere, the rare existence of singular points hardly poses a problem in learning, and the singular points can be handled separately.
- Since $\mathrm{Tanh}(z)$ is analytic and bounded almost everywhere in the complex plane, when trained by backpropagation, it can easily outperform the non-analytic split complex AF in convergence speed and achievable minimum squared error when the domain is bounded around the unit circle.
- The non-saturating behavior of CVAFs poses a fundamental challenge to the development of CVNNs. This issue was the most serious reason that CVNNs were considered difficult to develop before. This observation challenges the intuition derived from real-valued Tanh functions and underscores the need for careful consideration when extending AFs into the complex domain.
- Understanding and resolving this issue are crucial for unlocking the full potential of CVNNs and leveraging their advantages in various applications.

### 9.6.14 Fully Complex Logistic-Sigmoidal Function

The traditional logistic Sigmoid function is commonly used in NNs to introduce non-linearity and map real-valued inputs to a range between 0 and 1. The Fully Complex Logistic-Sigmoid AF [272] extends this concept to complex numbers, allowing for the incorporation of complex-valued inputs and weights in NNs. The function is defined as:

$$\sigma_{\mathrm{FCLogistic-Sigmoid}}(z) = \frac{1}{1 + e^{-z}}, \tag{9.119}$$

here, $z$ is a complex number.

**Remarks:**

- Figure 9.34 shows $\sigma_{\mathrm{FCLogistic-Sigmoid}}(z)$ in complex domain.
- The use of the exponential function $e^{-z}$ in the $\sigma_{\mathrm{FCLogistic-Sigmoid}}(z)$ involves potential issues with singularities when $z$ takes certain values. The denominator $1 + e^{-z}$ becomes zero when $e^{-z} = -1$. Solving for $z$, we find that this occurs when: $z = (2n + 1)i\pi$, where $n$ is an integer.
- To mitigate the singularity problem, practitioners often resort to scaling the input data to a manageable region in the complex plane. This scaling can be achieved by normalizing or restricting the input values to a certain range. By doing so, you effectively limit the range of possible values for the real and imaginary parts of the complex input $z$.
- This scaling helps in maintaining numerical stability during training and avoids situations where the exponential function produces extreme values. It is a common practice in CVNN implementations to preprocess the input data to ensure that it falls within a reasonable range.

### 9.6.15 Fully Complex ETFs

A single-valued function is said to have a singularity at a point if the function is not analytic at the point [266, 267]. The behavior of a complex function near a singular point can be classified into three main types: removable singularities, poles, and essential singularities. Here are some key concepts and details related to classification of singularity points:

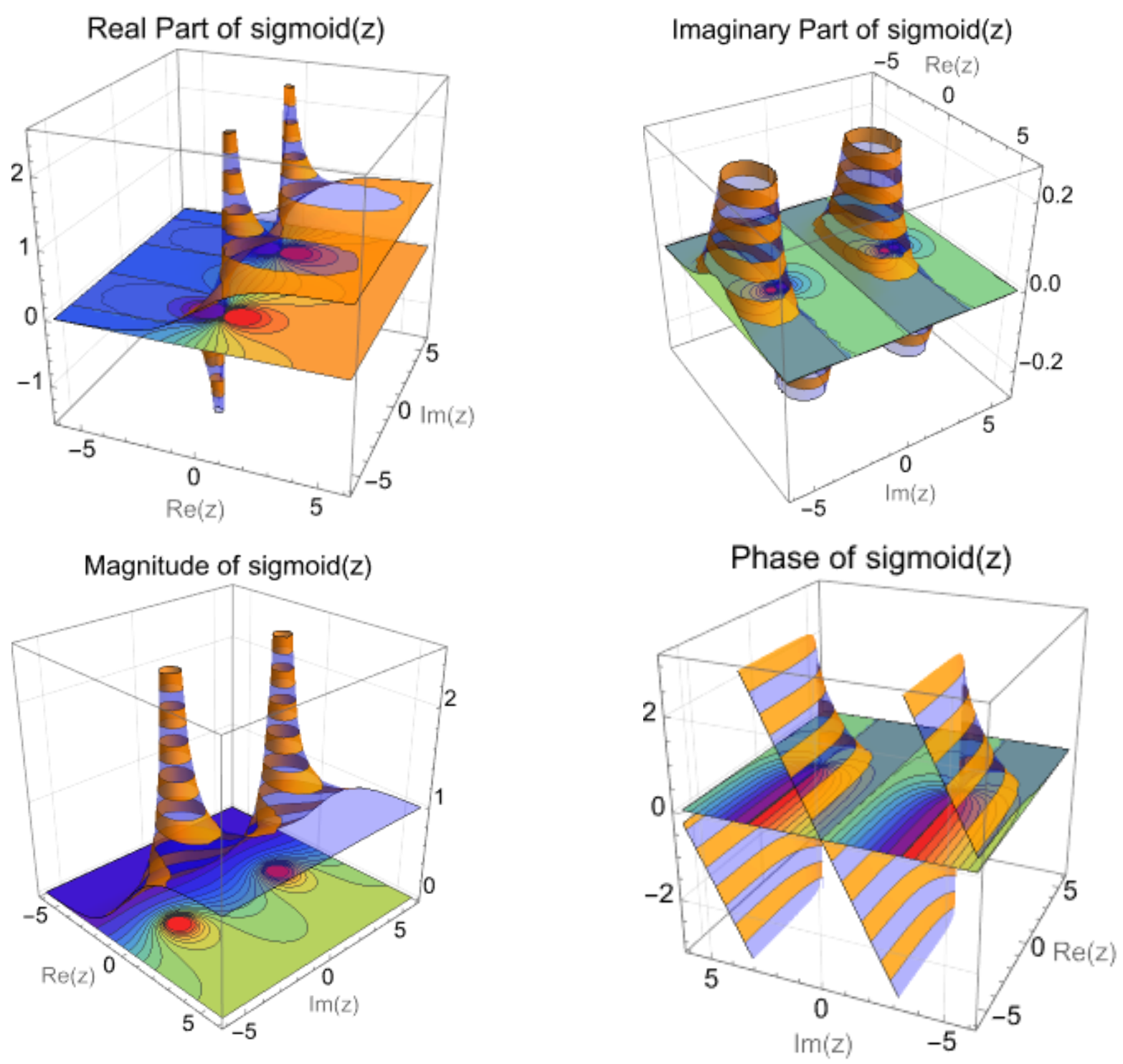

**Figure 9.34.** Fully Complex Logistic-Sigmoidal Function as a function of a complex variable, real part, imaginary part, amplitude, and phase.

- An isolated singularity is a type of singularity for a complex function that is surrounded by a neighborhood where the function is analytic. In other words, an isolated singularity is a point at which the function is not analytic, but there exists a deleted neighborhood (a punctured disk or an open set excluding the singularity itself) where the function is analytic. Mathematically, let $U \in \mathbb{C}$ be an open set and $z_0 \in U$. Suppose that $f: U \backslash \{z_0\} \to \mathbb{C}$ is holomorphic. In this situation, we say that $f$ has an isolated singular point (or isolated singularity) at $z_0$. The implication of the phrase is usually just that $f$ is defined and holomorphic on some such "deleted neighborhood" of $z_0$.
- A removable singularity is a type of singularity that can be "removed" or "filled in" by redefining the function at that point in such a way that the function becomes analytic at that point. In other words, a function has a removable singularity at a point if the singularity can be eliminated by assigning a value to the function at that point or by modifying the function in a small neighborhood of that point. Mathematically, if $f(z)$ has a removable singularity at $z_0$, it means that there exists a continuous function $g(z)$ such that $g(z) = f(z)$ for all $z$ in some deleted neighborhood of $z_0$, except possibly at $z_0$ itself. The function $g(z)$ is then analytic at $z_0$.
- Graphically, if you were to plot a function with a removable singularity, it might appear to have a "hole" or a point where the function is not defined. However, by redefining the function at that point or making a modification, you can smooth out the graph, and the function becomes well-behaved and analytic in the entire neighborhood.
- Example: Consider the function $f(z) = \frac{\sin(z)}{z}$, which has a singularity at $z = 0$. This singularity is removable because we can define a new function $g(z)$ as:

$$g(z) = \begin{cases} \dfrac{\sin(z)}{z}, & z \neq 0, \\ 1, & z = 0, \end{cases} \tag{9.120}$$

(by defining $f(0)$ to be 1 which is the limit of $f(z)$ as $z$ tends to 0). The function $g(z)$ is defined and continuous at $z = 0$ and is equal to $f(z)$ in its domain. Thus, the singularity at $z = 0$ is removable.
Note that, if $f(z)$ has an isolated singularity at $z_0$, the singularity is said to be removable if $\lim_{z \to z_0} f(z)$ exists.

- Riemann Removable Singularities Theorem [266]: Let $f: U \backslash \{z_0\} \to \mathbb{C}$ be holomorphic and bounded. Then $\lim_{z \to z_0} f(z)$ exists. The function $g: U \to \mathbb{C}$ defined by

$$g(z) = \begin{cases} f(z), & z \neq 0, \\ \lim_{t \to z_0} f(t), & z = 0, \end{cases} \tag{9.121}$$

is holomorphic.

- A pole is a type of singularity that occurs when a function approaches infinity at a certain point. If $\lim_{z \to z_0} f(z) \to \infty$, while $f(z)$ is analytic in a deleted neighborhood of $z = z_0$, then $f(z)$ is not removable but has a pole at $z = z_0$. Poles are characterized by the fact that the function becomes unbounded as the variable approaches a specific complex number. Equivalently, $f$ has a pole of order $n$ at $z_0$ if $n$ is the smallest positive integer for which $(z - z_0)^n f(z) = g(z)$ is holomorphic at $z_0$.
Example: The basic example of a pole is $f(z) = 1/z^n$, which has a single pole of order $n$ at $z = 0$.
- An isolated singularity that is neither removable nor a pole is said to be an isolated essential singularity.
- This classification helps us understand the local behavior of a complex function near a singular point. Removable singularities are "fixable," poles represent a certain kind of divergence, and essential singularities indicate more complex and intricate behaviors.
- Furthermore, A branch point [267] is a point in the complex plane where a multi-valued function ceases to be single-valued. In other words, a branch point of a function is a point in the complex plane whose complex argument can be mapped from a single point in the domain to multiple points in the range. Suppose that we are given the function $w = z^{1/2}$. Suppose further that we allow $z$ to make a complete circuit (counter-clockwise) around the origin starting from point $A$. We have $z = re^{i\theta}$, $w = \sqrt{r}e^{i\theta/2}$ so that at $A$, $\theta = \theta_1$ and $w = \sqrt{r}e^{i\theta_1/2}$. After a complete circuit back to $A$, $\theta = \theta_1 + 2\pi$ and $w = \sqrt{r}e^{i(\theta_1+2\pi)/2} = -\sqrt{r}e^{i\theta_1/2}$. Thus, we have not achieved the same value of $w$ with which we started. However, by making a second complete circuit back to $A$, i.e., $\theta = \theta_1 + 4\pi$, $w = \sqrt{r}e^{i(\theta_1+4\pi)/2} = \sqrt{r}e^{i\theta_1/2}$ and we then do obtain the same value of $w$ with which we started. We can describe the above by stating that if $0 \leq \theta < 2\pi$, we are on one branch of the multiple-valued function $z^{1/2}$, while if $2\pi \leq \theta < 4\pi$, we are on the other branch of the function. It is clear that each branch of the function is single-valued. In order to keep the function single-valued, we set up an artificial barrier such as $OB$ where $B$ is at infinity [although any other line from $O$ can be used], which we agree not to cross. This barrier is called a branch line or branch cut. Hence, a branch cut is a curve or a line in the complex plane along which a multi-valued function is modified to make it single-valued. By introducing a branch cut, we exclude certain paths from the domain of the function, allowing us to define a single-valued function.

A number of ETFs derivable from the entire exponential function $e^Z$ that are analytic are defined as “Fully Complex” AFs [276, 326, 327] and are shown to provide a parsimonious structure for processing data in the complex domain, and address most of the shortcomings of the traditional approach. The following ETFs are identified to provide adequate non-linearity as an AF.

**Circular functions**

$$\mathrm{Tan}(z) = \sigma_{\mathrm{FCTan}}(z) = \frac{e^{iz} - e^{-iz}}{i(e^{iz} + e^{-iz})}, \tag{9.122}$$

$$\mathrm{Sin}(z) = \sigma_{\mathrm{FCSin}}(z) = \frac{e^{iz} - e^{-iz}}{2i}. \tag{9.123}$$

**Inverse circular functions**

$$\mathrm{ArcTan}(z) = \sigma_{\mathrm{FCArcTan}}(z) = \int_0^z \frac{dt}{1 + t^2}, \tag{9.124}$$

$$\mathrm{ArcSin}(z) = \sigma_{\mathrm{FCArcSin}}(z) = \int_0^z \frac{dt}{(1-t)^{\frac{1}{2}}}, \tag{9.125}$$

$$\mathrm{ArcCos}(z) = \sigma_{\mathrm{FCArcCos}}(z) = \int_z^1 \frac{dt}{(1-t^2)^{\frac{1}{2}}}. \tag{9.126}$$

**Hyperbolic functions**

$$\mathrm{Tanh}(z) = \sigma_{\mathrm{FCTanh}}(z) = \frac{\mathrm{Sinh}(z)}{\mathrm{Cosh}(z)} = \frac{e^z - e^{-z}}{e^z + e^{-z}}, \qquad \mathrm{Sinh}(z) = \sigma_{\mathrm{FCSinh}}(z) = \frac{e^z - e^{-z}}{2}. \tag{9.127}$$

**Inverse hyperbolic functions**

$$\mathrm{ArcTanh}(z) = \sigma_{\mathrm{FCArcTanh}}(z) = \int_0^z \frac{dt}{1-t^2}, \qquad \mathrm{ArcSinh}(z) = \sigma_{\mathrm{FCArcSinh}}(z) = \int_0^z \frac{dt}{(1+t^2)^{\frac{1}{2}}}. \tag{9.128}$$

**Remarks:**

- Figures 9.35 through 9.43 show the magnitude of these ETFs in the vicinity of unit circle.
- These figures show two distinctive patterns of inevitable unboundedness property when real-valued ETFs are generalized into complex-valued ETFs. The ETFs divide into two classes, based on their types of singularity: a set of bounded squashing AFs over the bounded domain and a set of AFs with unbounded isolated singularities.
- First category of unboundedness is observable in the tangent function family ($\mathrm{Tan}(z)$, $\mathrm{Tanh}(z)$, $\mathrm{ArcTan}(z)$, and $\mathrm{ArcTanh}\, z$), Figures 9.35 through 9.38, that possesses finite number of point set discontinuities in a bounded domain. Note that $\mathrm{ArcTan}(z)$ has isolated singularity at $\pm i$ and $\mathrm{ArcTanh}(z)$ has isolated singularity at $\pm 1$. Moreover, $\mathrm{Tanh}(z)$ has isolated essential singularities at every $(1/2+n)\pi i$, $n \in \mathbb{N}$, since it is asymptotically $+\infty$ as $(1/2+n)\pi i$ is approached from below and to $-\infty$ as $(1/2+n)\pi i$ is approached from above along the imaginary axis. Similarly, $\mathrm{Tan}(z)$ has isolated essential singularities at every $(1/2+n)\pi$.
- Usually, these singular points and discontinuities at non-zero points do not pose a problem in training when the domain of interest is bounded within a circle of radius $\pi/2$. If the domain is larger including these irregular points, the training process tends to become more sensitive to the size of the learning rate. In this case, the initial random weights need to be bounded in a small radius, typically below 0.1 to avoid oscillation in the gradient-based updates.
- Second category of unboundedness is observable in the inverse sine and cosine family including $\mathrm{ArcSin}(z)$, $\mathrm{ArcCos}(z)$, and $\mathrm{ArcSinh}(z)$ functions that exhibit unbounded but decreasing rate of magnitude growth as they move away from the origin, as shown in Figures 9.39 through 9.41. The inverse sine and cosines have removable singularities represented as ridges along the real or imaginary axis outside the unit circle, which is defined as the branch cuts, Figures 9.39 through 9.41. This branch-type discontinuity results from the definition of the integral where the branch should not be crossed.
- However, $\mathrm{ArcSin}(z)$ and $\mathrm{ArcSinh}(z)$ functions are the most radially symmetric functions in magnitude. $\mathrm{ArcCos}(z)$ function is asymmetric along the real axis and has a discontinuous zero point at $z=1$. When the domain is bounded, $\mathrm{ArcSin}(z)$, $\mathrm{ArcCos}(z)$, and $\mathrm{ArcSinh}(z)$ are naturally bounded with squashing function characteristics while providing more discriminating nonlinearity than the split-Tanh function. It has been observed that the radial symmetricity as well as the nonlinear phase response of $\mathrm{ArcSinh}(z)$ and $\mathrm{ArcSin}(z)$ functions tend to provide efficient nonlinear approximation capability.
- $\mathrm{Sin}(z)$ and $\mathrm{Sinh}(z)$ functions exhibit continuous magnitude but are unbounded functions with convex behavior along the imaginary and real axes, respectively. Figures 9.42 and 9.43 represent $\mathrm{Sin}(z)$ and $\mathrm{Sinh}(z)$ functions, respectively. Also, note that $\mathrm{Sin}(z)$ is equivalent to $\mathrm{Sin}(x)$ along the real axis, while $\mathrm{Sinh}(z)$ is equivalent to $\mathrm{Sin}(x)$ along the imaginary axis. Therefore, the bounded squashing property for $\mathrm{Sin}(z)$ and $\mathrm{Sinh}(z)$ functions is available within a radius of $\pi/2$ from the origin. Therefore, these functions, when bounded in a radius of $\pi/2$ can be effective nonlinear approximators. Consequently, $\mathrm{Sin}(z)$, $\mathrm{Sinh}(z)$, $\mathrm{ArcSin}(z)$, $\mathrm{ArcCos}(z)$, and $\mathrm{ArcSinh}(z)$ functions (the sine family) can be classified as complex squashing functions within a bounded domain.

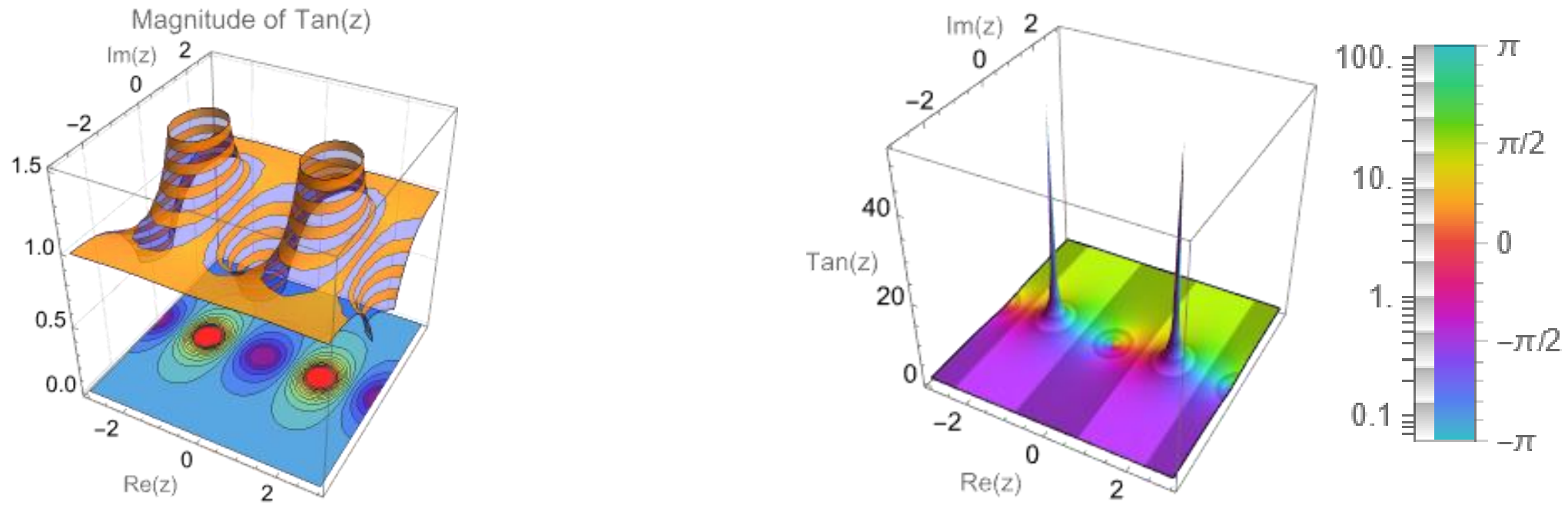

**Figure 9.35.** Left panel: 3D and contour plot of the magnitude of $\sigma_{\text{FCTan}}(z)$. Right panel: The ComplexPlot3D generates a 3D plot of Abs[$\sigma_{\text{FCTan}}(z)$] colored by arg[$\sigma_{\text{FCTan}}(z)$] over the complex rectangle with corners $z_{min} = -3 - 3i$ and $z_{max} = 3 + 3i$. Additionally, the color shading is achieved using the "CyclicLogAbs" function to cyclically shade colors and give the appearance of contours of constant Abs[$\sigma_{\text{FCTan}}(z)$].

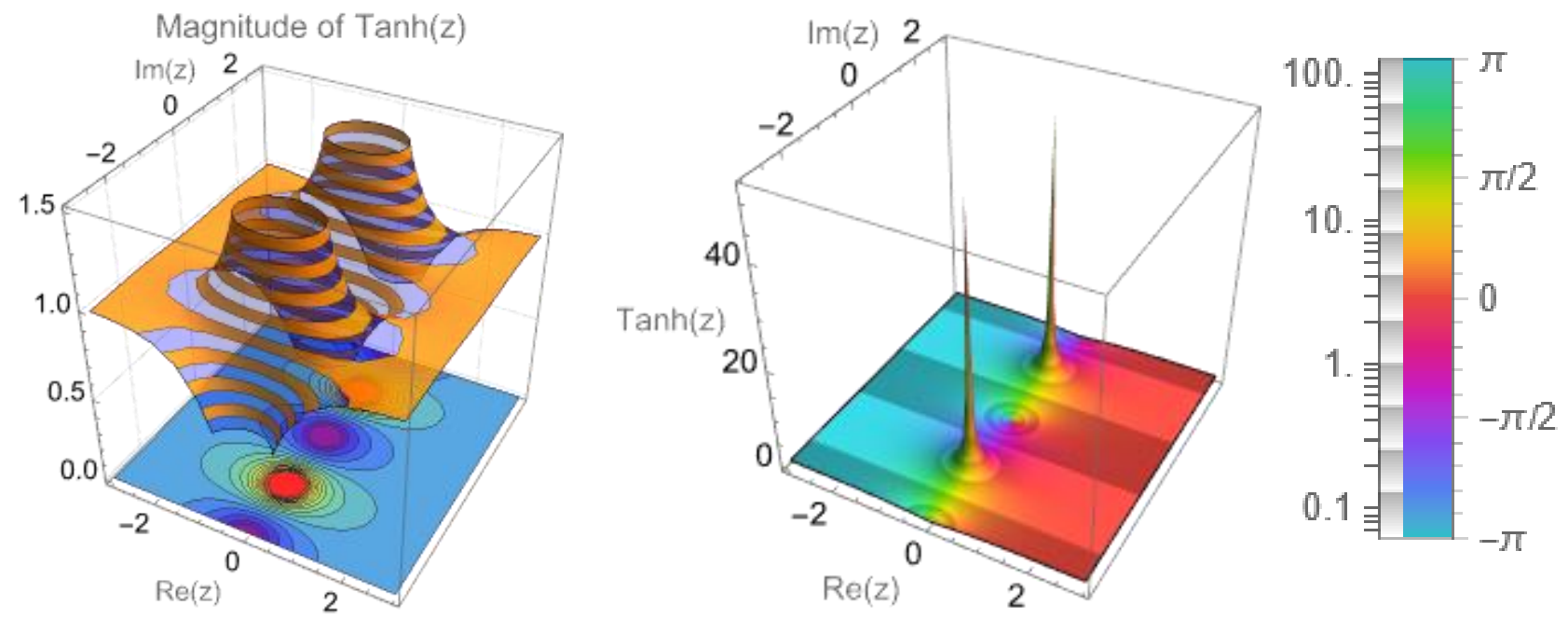

**Figure 9.36.** Same as Figure 9.35 but for the function $\sigma_{\text{FCTanh}}(z)$.

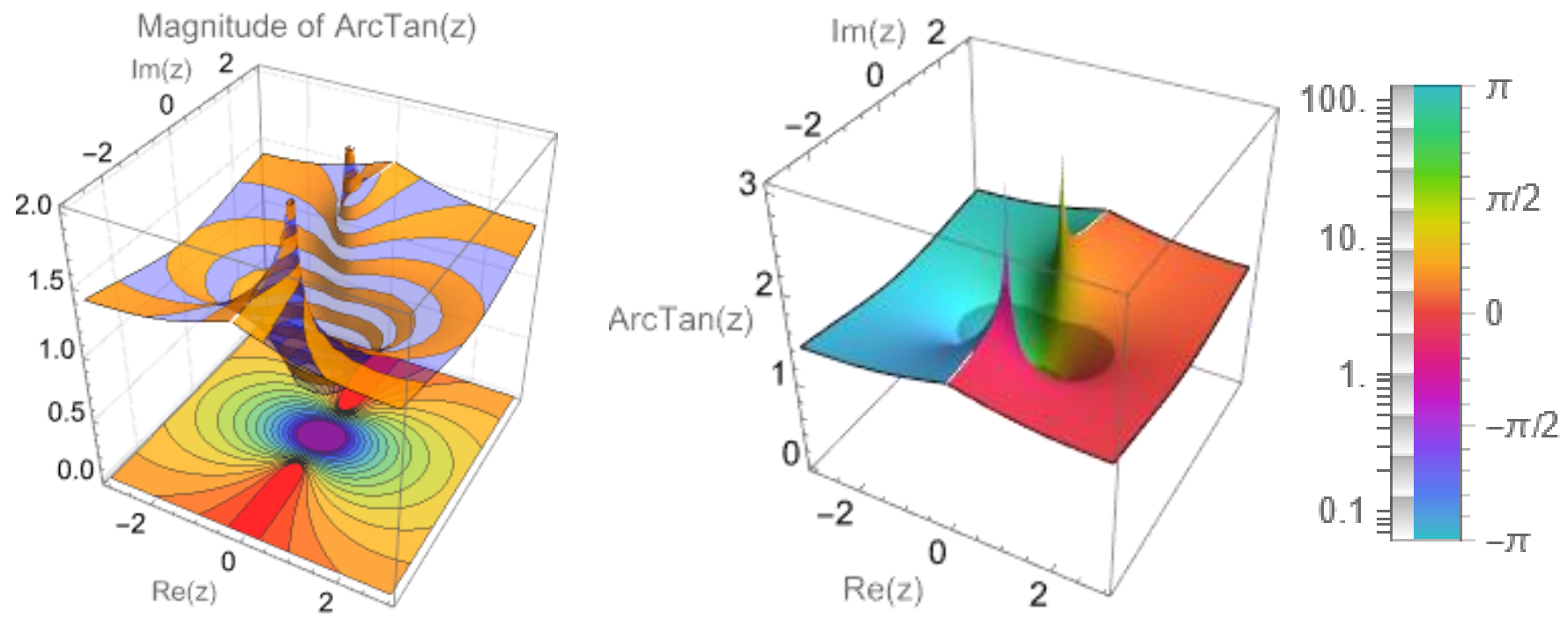

**Figure 9.37.** Same as Figure 9.35 but for the function $\sigma_{\text{FCArcTan}}(z)$.

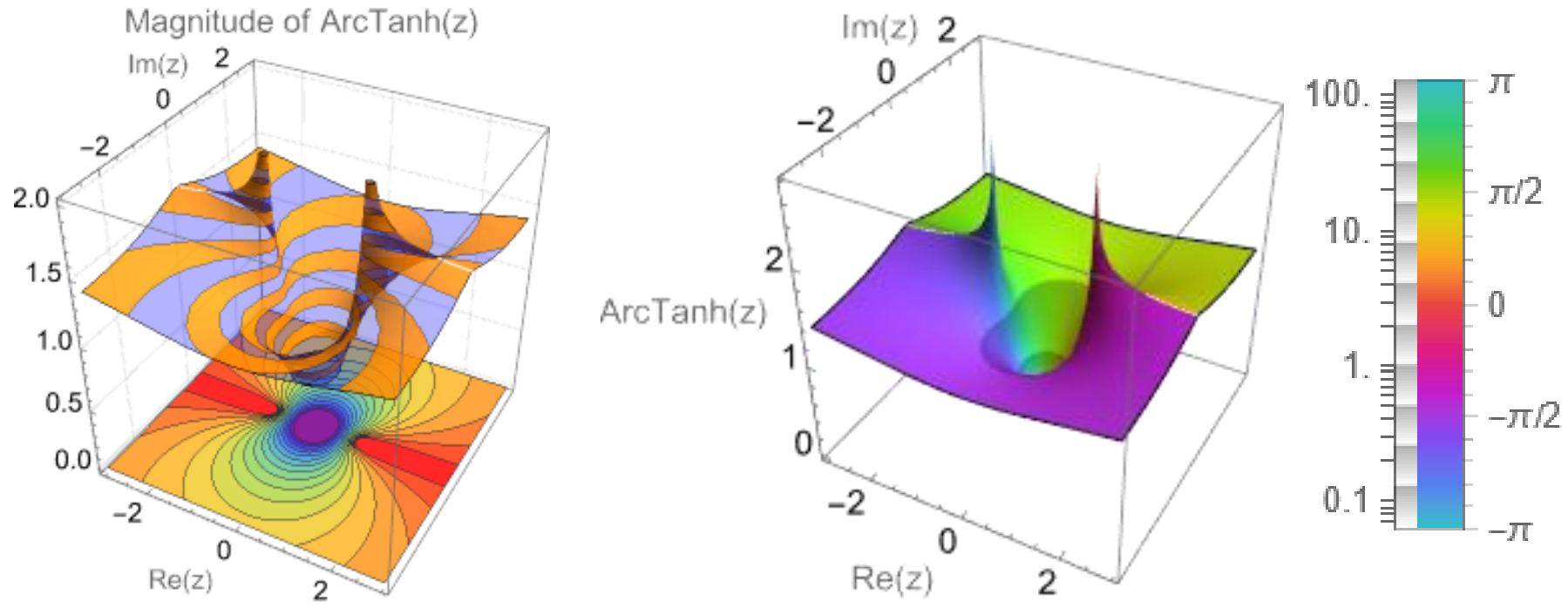

**Figure 9.38.** Same as Figure 9.35 but for the function $\sigma_{\text{FCArcTanh}}(z)$.

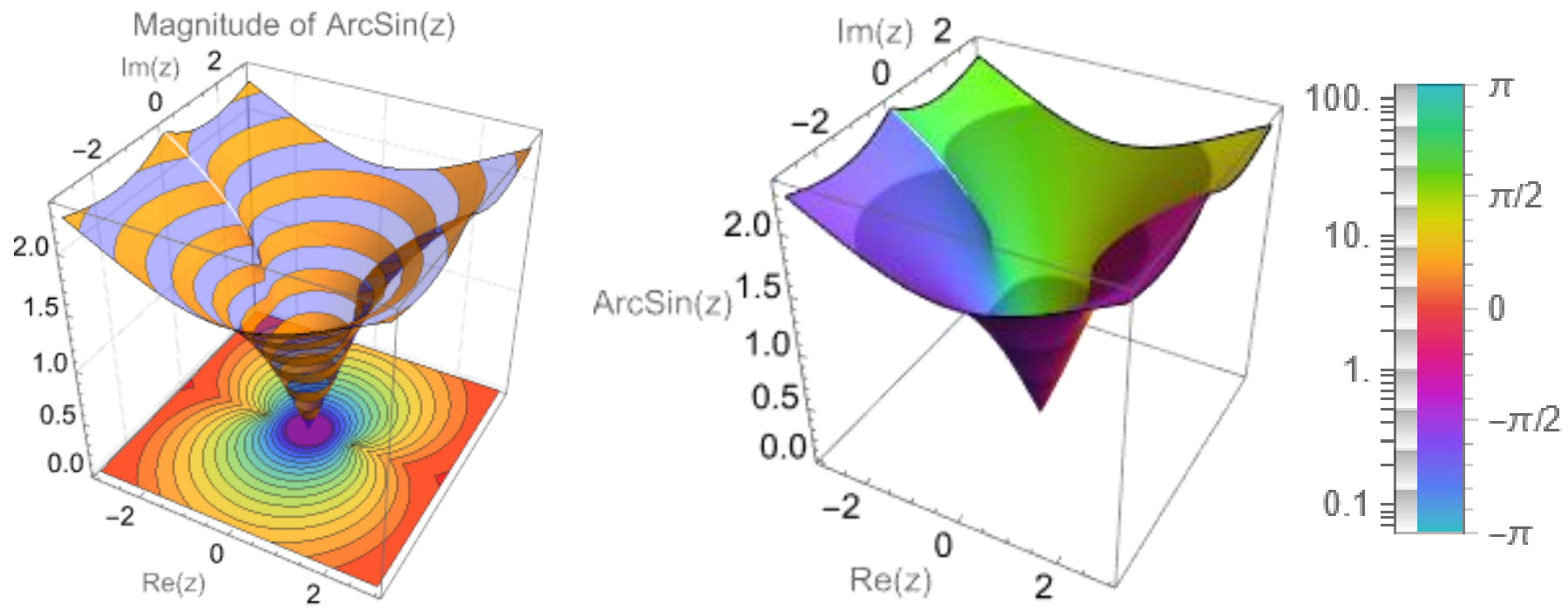

**Figure 9.39.** Same as Figure 9.35 but for the function $\sigma_{\mathrm{FCArcSin}}(z)$.

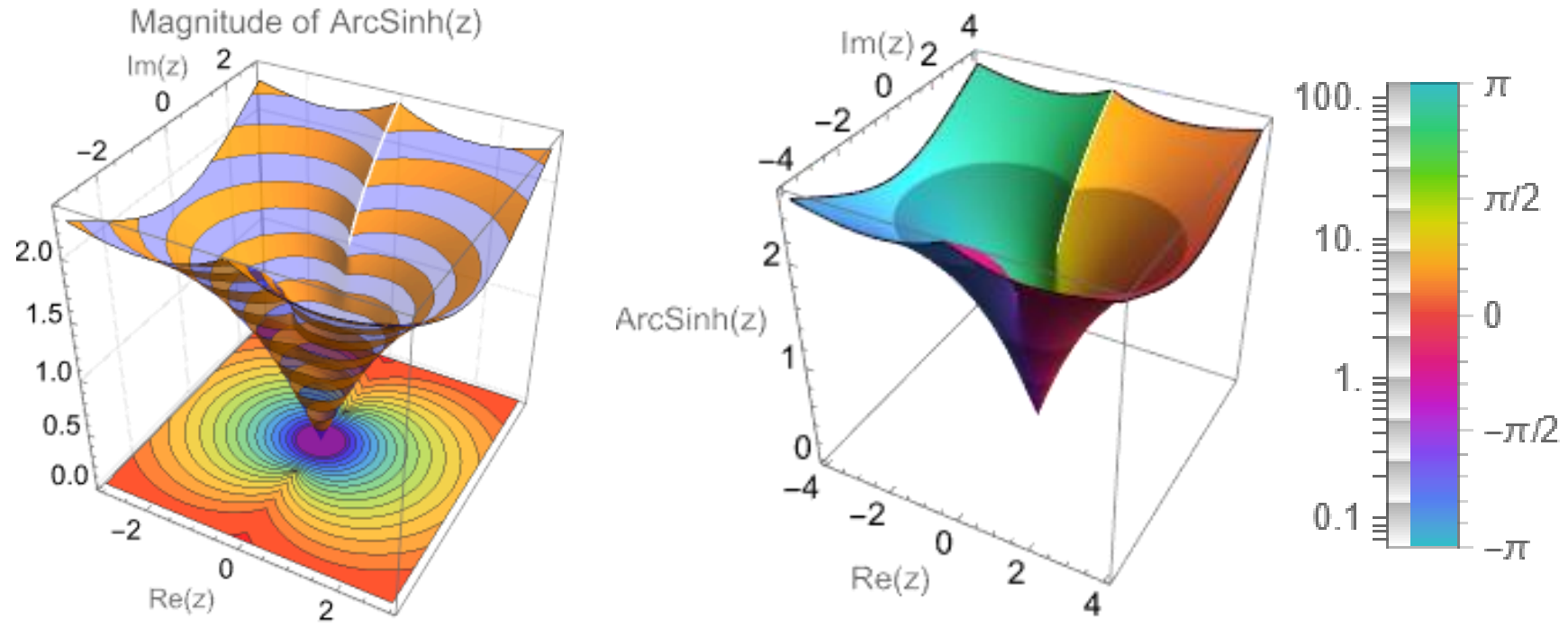

**Figure 9.40.** Same as Figure 9.35 but for the function $\sigma_{\mathrm{FCArcSinh}}(z)$.

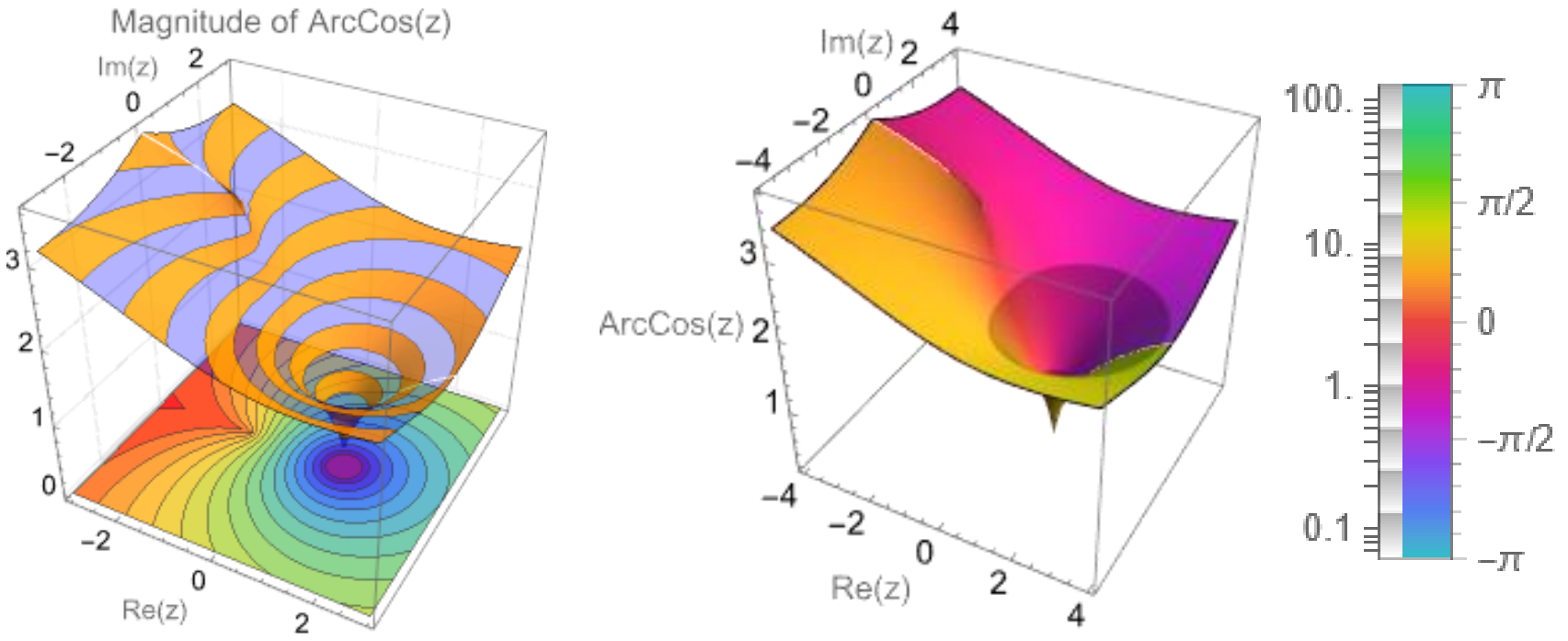

**Figure 9.41.** Same as Figure 9.35 but for the function $\sigma_{\mathrm{FCArcCos}}(z)$.

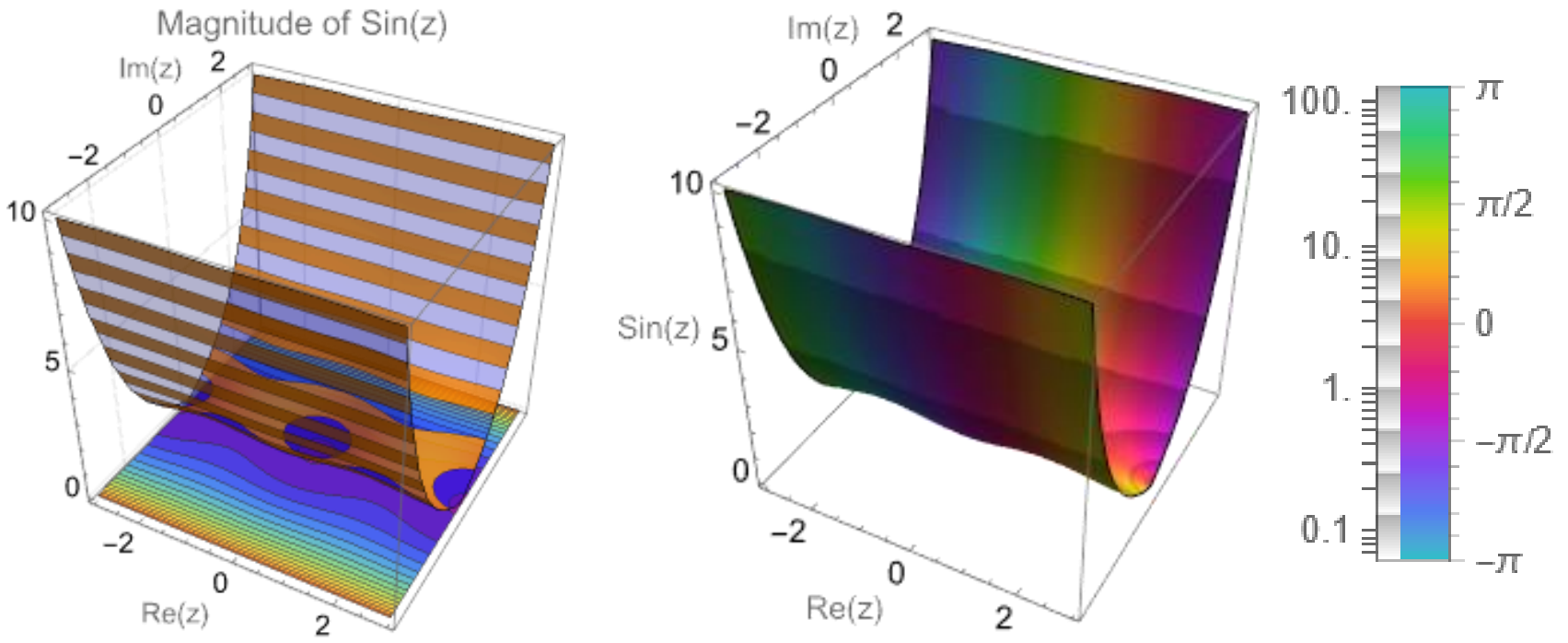

**Figure 9.42.** Same as Figure 9.35 but for the function $\sigma_{\mathrm{FCSin}}(z)$.

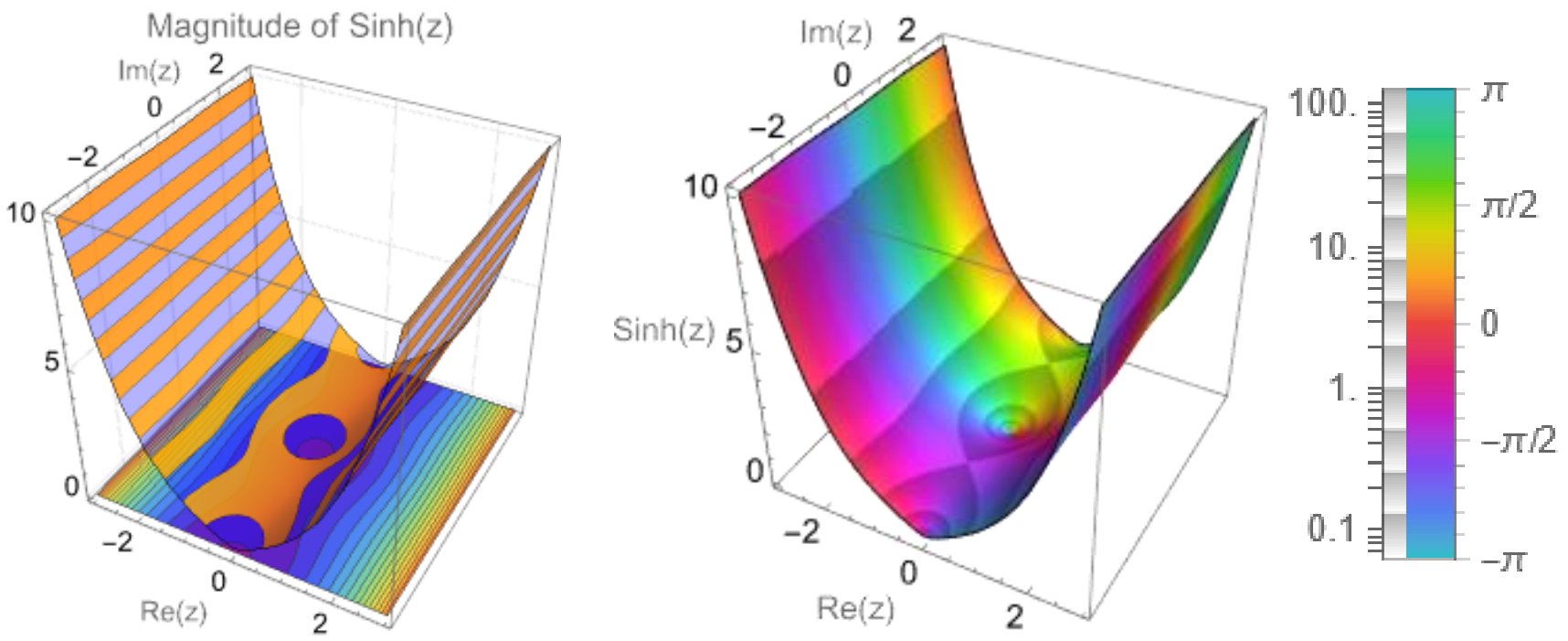

**Figure 9.43.** Same as Figure 9.35 but for the function $\sigma_{\text{FCSinh}}(z)$.

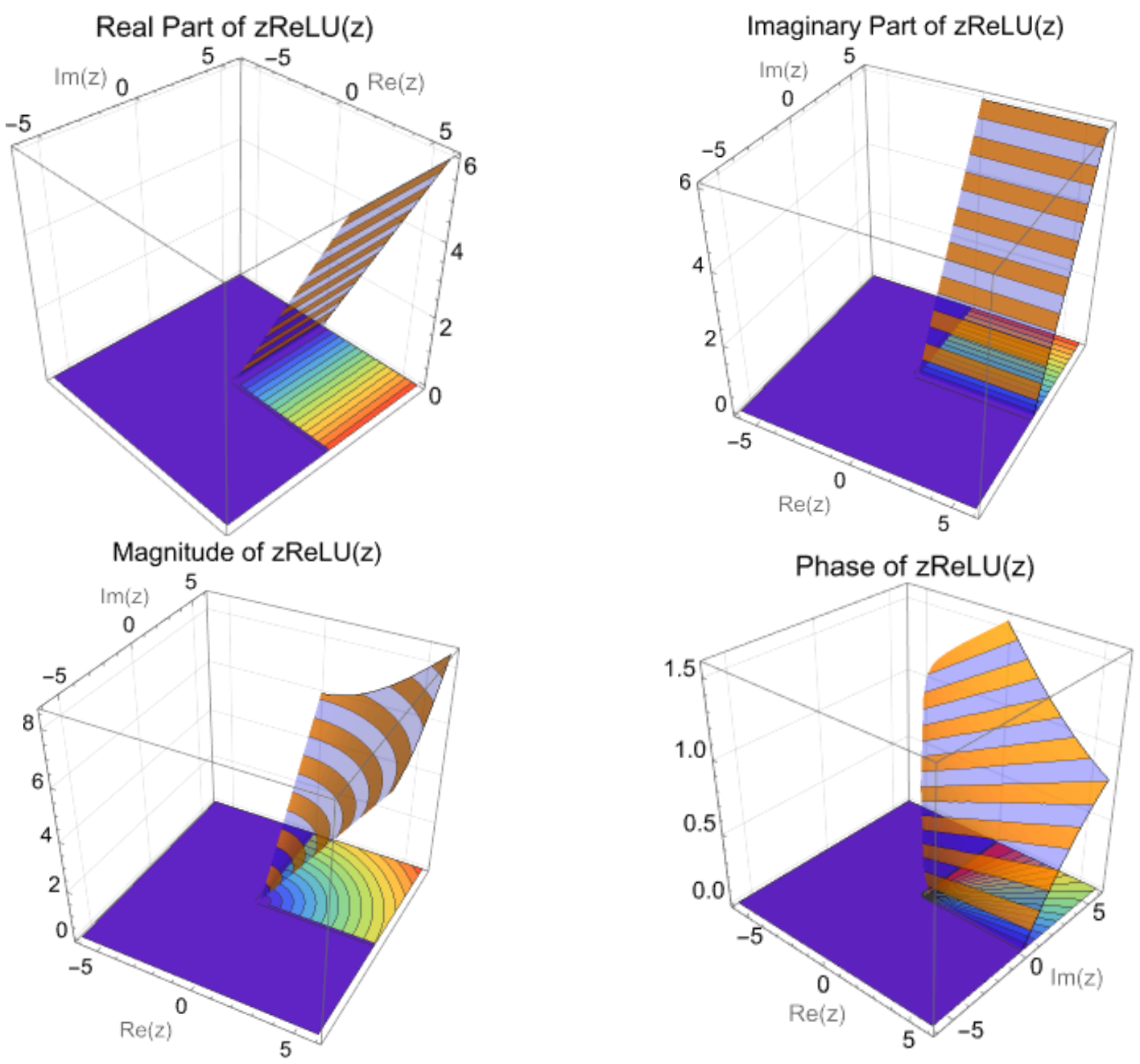

**Figure 9.44.** Complex $\sigma_{\text{zReLU}}(z)$: real and imaginary parts, amplitude, and phase.

### 9.6.16 zReLU, z3ReLU, zPReLU and z3PReLU

The traditional ReLU function used in RVNNs is popular for introducing non-linearity by outputting the input if it is positive; otherwise, it outputs zero. Extending this concept to complex numbers allows for the incorporation of complex-valued inputs and weights in NNs. [270] proposed a complex-valued ReLU as:

$$\sigma_{\text{zReLU}}(z) = \begin{cases} z, & \text{Re}(z) > 0 \text{ and } \text{Im}(z) > 0 \\ 0, & \text{otherwise} \end{cases} = \begin{cases} z, & \arg(z) \in [0, \pi/2], \\ 0, & \text{otherwise.} \end{cases} \quad (9.129)$$

Figures 9.44 and 9.45 represent the real and imaginary parts, amplitude, and phase of $\sigma_{\text{zReLU}}(z)$. [328] proposed z3 Rectified Linear Unit (z3ReLU). The formula for the z3ReLU is given by:

$$\sigma_{\text{z3ReLU}}(z) = \begin{cases} z, & \text{Re}(z) > 0 \text{ or } \text{Im}(z) > 0, \\ 0, & \text{otherwise.} \end{cases} \quad (9.130)$$

When compared to similar CVAFs such as zReLU, the $\sigma_{\text{z3ReLU}}$ AF preserves more of the input magnitude and phase but still provides nonlinearity by rectifying inputs that have negative-valued real or imaginary segments, see Figure 9.46.

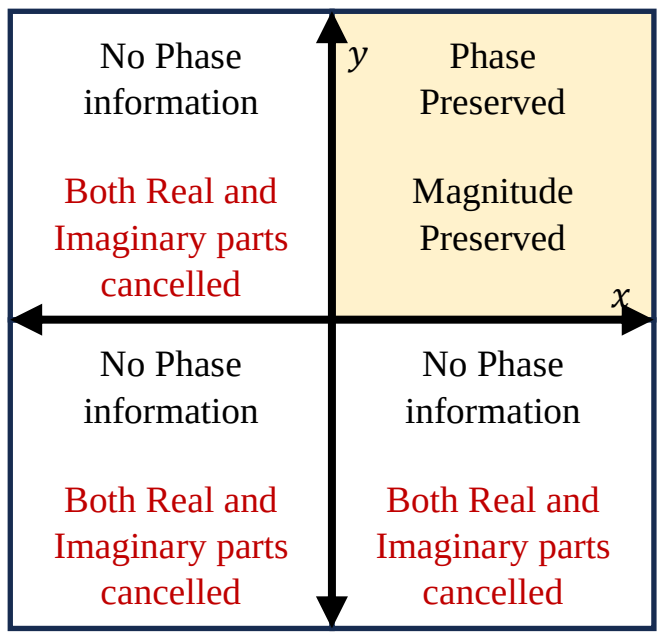

**Figure 9.45.** Phase information encoding for the zReLU AF. The $x$-axis represents the real part and the $y$-axis axis represents the imaginary part. The complex representation is discriminated into two regions, i.e., the one that preserves the whole complex information (colored in orange) and the one that cancels it (colored in white).

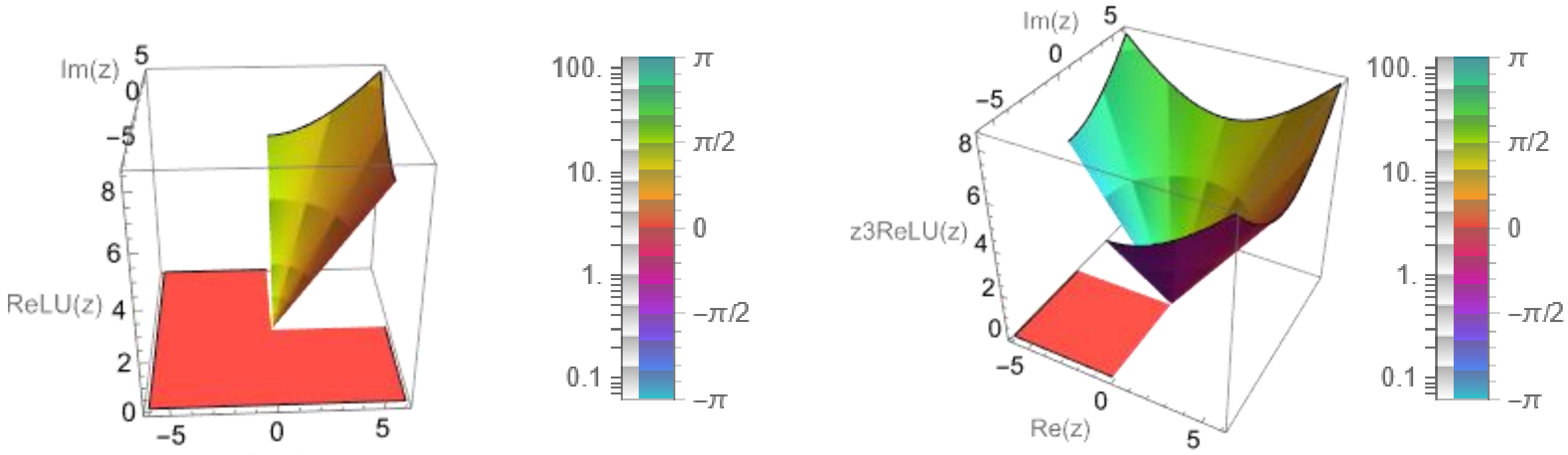

**Figure 9.46.** Left panel: The ComplexPlot3D generates a 3D plot of Abs[$\sigma_{\text{zReLU}}(z)$] colored by arg[$\sigma_{\text{zReLU}}(z)$] over the complex rectangle with corners $z_{min} = -6 - 6i$ and $z_{max} = 6 + 6i$. Additionally, the color shading is achieved using the "CyclicLogAbsArg" function to cyclically shade colors and give the appearance of contours of constant Abs[$\sigma_{\text{zReLU}}(z)$] and constant arg[$\sigma_{\text{zReLU}}(z)$]. Right panel: Same as left figure but for the function $\sigma_{\text{z3ReLU}}(z)$.

Moreover, z Parametric Rectified Linear Unit (zPReLU) [329] CVAF was defined as follows:

$$\sigma_{\text{zPReLU}}(z) = \begin{cases} z, & \arg(z) \in [0, \pi/2], \\ \alpha z, & \text{otherwise.} \end{cases} \tag{9.131}$$

This nonlinearity is similar to zReLU. It has one complex-valued trainable parameter $\alpha$. The input, $z$, is multiplied with $\alpha$ if it does not lie in the first quadrant. Additionally, the z3 Parametric Rectified Linear Unit (z3PReLU) [329] AF is an extension of the previous $\sigma_{\text{zPReLU}}$, and it is defined as follows:

$$\sigma_{\text{z3PReLU}}(z) = \begin{cases} z, & \arg(z) \in [0, \pi/2], \\ \alpha_1 z, & \arg(z) \in (\pi/2, \pi], \\ \alpha_2 z, & \arg(z) \in (\pi, 3\pi/2], \\ \alpha_3 z, & \arg(z) \in (3\pi/2, 2\pi]. \end{cases} \tag{9.132}$$

This nonlinearity has three complex-valued trainable parameters. The input, $z$, is multiplied with different complex numbers depending on its quadrant.

### 9.6.17 Fully Complex Exponential Function

The ETFs used in fully complex NNs may exhibit sensitivity to weight initializations and learning rates. Additionally, these functions may suffer from singularities in the finite region of the complex plane, which can complicate the training process. An exponential function was proposed as a CVAF for the nonlinear processing of complex-valued data [269, 330].

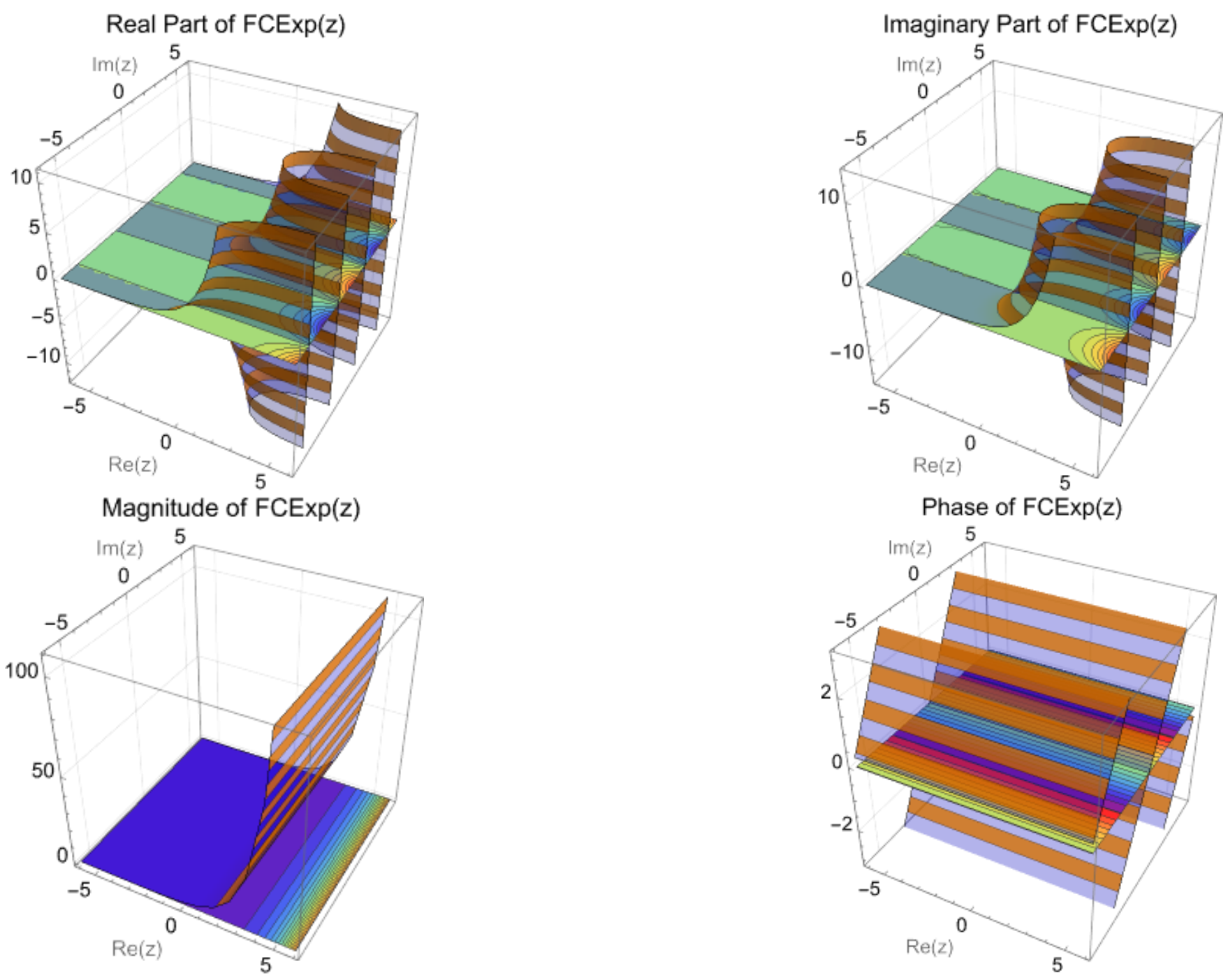

**Figure 9.47.** Complex $\sigma_{\mathrm{FCExp}}(z)$ as a function of complex variables, real-part, imaginary-part, amplitude, and phase.

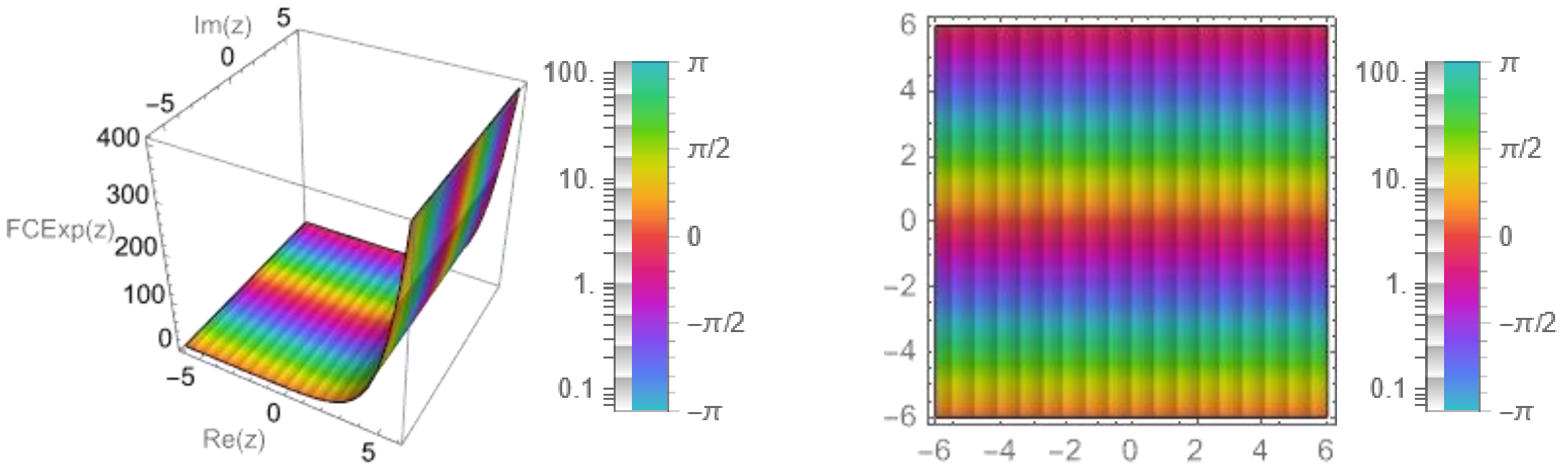

**Figure 9.48.** Left panel: The ComplexPlot3D generates a 3D plot of Abs[$\sigma_{\mathrm{FCExp}}(z)$] colored by arg[$\sigma_{\mathrm{FCExp}}(z)$] over the complex rectangle with corners $z_{min} = -6 - 6i$ and $z_{max} = 6 + 6i$. Additionally, the color shading is achieved using the "CyclicLogAbs" function to cyclically shade colors and give the appearance of contours of constant Abs[$\sigma_{\mathrm{FCExp}}(z)$]. Right panel: ComplexPlot generates a plot of arg[$\sigma_{\mathrm{FCExp}}(z)$] over the complex rectangle with corners $-6 - 6i$ and $6 + 6i$, with using the "CyclicLogAbs" function.

The exponential function,

$$\sigma_{\mathrm{FCExp}}(z) = \mathrm{e}^{z}, \tag{9.133}$$

is entire in $\mathbb{C}$ since $\sigma_{\mathrm{Exp}}(z) = \sigma'_{\mathrm{ECExp}}(z) = \mathrm{e}^{z}$.

**Remarks:**

- The complex-valued exponential function has an essential singularity at $\pm\infty$. To address the challenges posed by singularities, a strategic decision involves constraining the weights of the fully complex NN. By limiting the weights to a small ball of radius $r$ and controlling the number of hidden neurons, the network achieves a bounded behavior, ensuring numerical stability during the training process.

- A graphical representation (see Figures 9.47 and 9.48) provides a visual understanding of the complex-valued exponential function's behavior. The plot showcases the real and imaginary parts, amplitude, and phase of the activation function. Notably, the amplitude is illustrated as continuous, increasing, and bounded within a specific region of the complex domain, emphasizing its suitability for NN applications.

## 9.7 New Complex Activation Functions

This section explores the extension of popular real-valued AFs—Exponential Linear Unit (ELU), Mish, SoftPlus, and Swish—to their complex counterparts, introducing Split-ELU, Split-Mish, Split-SoftPlus, and Split-Swish. These CVAFs operate by separately applying the real-valued AFs to the real and imaginary parts of the complex input, then recombining them into a complex output. This approach preserves the essential properties of the original AFs while enhancing control over phase and amplitude modifications, which are crucial in many complex-valued applications. We further generalize the real-valued AFs Mish and Swish to their fully complex counterparts, introducing Fully Complex Mish (FCMish), and Fully Complex Swish (FCSwish). Additionally, we propose a novel set of CVAFs designed to modulate the amplitude of complex inputs while preserving their phase, ensuring the output remains a valid complex number. The following sections provide a detailed explanation of these extendeds and novel CVAFs, their mathematical formulations, properties, and visual representations.

### 9.7.1 Split: ELU, Mish, SoftPlus, and Swish

Now, we extend the popular real-valued AFs ELU, Mish, SoftPlus, and Swish to their complex counterparts: Split-ELU, Split-Mish, Split-SoftPlus, and Split-Swish. These functions split the real and imaginary parts of the complex input, apply the AF separately to each part, and then combine them back into a complex number.

$$\sigma_{\text{Split-ELU}}(z) = \sigma_{\text{ELU}}[\Re(z)] + i\sigma_{\text{ELU}}[\Im(z)], \tag{134.1}$$
$$\sigma_{\text{Split-Mish}}(z) = \sigma_{\text{Mish}}[\Re(z)] + i\sigma_{\text{Mish}}[\Im(z)], \tag{134.2}$$
$$\sigma_{\text{Split-SoftPlus}}(z) = \sigma_{\text{SoftPlus}}[\Re(z)] + i\sigma_{\text{SoftPlus}}[\Im(z)], \tag{134.3}$$
$$\sigma_{\text{Split-Swish}}(z) = \sigma_{\text{Swish}}[\Re(z)] + i\sigma_{\text{Swish}}[\Im(z)]. \tag{134.4}$$

The split approach offers several benefits, including enhanced control over phase and amplitude modifications, crucial for domains where these elements carry significant information. For instance, Split-ELU (Figures 9.49 and 9.50) uses the ELU function to provide a response for negative inputs, aiding in gradient stability and learning efficiency. Split-Mish (Figures 9.51 and 9.52), with its non-monotonic smoothness, manages extreme values better, enhancing the network's ability to handle outliers. Split-SoftPlus (Figures 9.53 and 9.54) offers a continuously differentiable approximation of ReLU, making it ideal for maintaining smooth gradient flow in complex-valued calculations. Lastly, Split-Swish (Figures 9.55 and 9.56) incorporates a self-gating mechanism that regulates gradient flow, adding a form of light regularization beneficial for complex data.

The Split-ELU, Split-Mish, Split-SoftPlus, and Split-Swish CVAFs exhibit line symmetry with respect to both the real and imaginary axes. This symmetry means that these functions' behavior can be mirrored across these axes, simplifying their analysis and understanding. The real part of these AFs is symmetric with respect to the real axis ($\Im(z) = 0$). This means if you replace a complex number $z_1 = (a, b)$ with $z_2 = (a, -b)$, where $a$ and $b$ are the real and imaginary parts, respectively, the real part of the function remains unchanged. Mathematically, if $\sigma$ is the CVAF, then: $\Re(\sigma(a, b)) = \Re(\sigma(a, -b))$. This implies that the behavior of the real part of the function does not change when the imaginary part of the input is negated. This property can be particularly useful when analyzing or visualizing the real part of these functions because it reduces the problem to considering only one half of the imaginary plane. On the other hand, the symmetry of the imaginary part of these functions with respect to the imaginary axis ($\Re(z) = 0$) implies that if you replace $z_1 = (a, b)$ with $z_2 = (-a, b)$ the imaginary part of these functions value remains the same. The symmetry in these CVAFs implies that the neural dynamics associated with these CVAFs have special significance along the real and imaginary axes, $\Im(z) = 0$ and $\Re(z) = 0$. Along these axes, the behavior of the CVAFs is more predictable and structured. By knowing that the function behaves symmetrically with respect to both the real and imaginary axes, the complexity of studying the function's behavior is reduced. This simplification can be particularly beneficial in both theoretical analysis and practical applications, such as NN training and optimization, where understanding the behavior of AFs is crucial.

### 9.7.2 Fully Complex: Swish and Mish

We define Fully Complex Swish (FCSwish) (Figure 9.57) as

$$\sigma_{\text{FCSwish}}(z) = z\,\sigma_{\text{Sigmoid}}(z), \tag{135}$$

where $\sigma_{\text{Sigmoid}}(z) = (1 + \exp(-z))^{-1}$ is the Sigmoid function. The singularities of the FCSwish function occur at specific points in the complex plane. To understand these singularities, we first examine the conditions under which the Logistic Sigmoid function becomes undefined or singular. The singularities of $\sigma_{\text{Sigmoid}}(z)$ arise when its denominator equals zero, which happens when $1 + \exp(-z) = 0$. Solving for $z$, we get $\exp(-z) = -1$, leading to $-z = \ln(-1)$. In the complex plane, $\ln(-1) = i\pi$ for the principal branch, but due to the periodic nature of the complex exponential function, the logarithm can also be expressed as $i\pi + 2i\pi n$, where $n$ is any integer. Therefore, the singularities of the Sigmoid function occur at $z = i\pi + 2i\pi n$. Applying this to the FCSwish function, which includes the Logistic Sigmoid function as a factor, we find that the singularities of the FCSwish function also occur at these points. Thus, the singularities of the FCSwish function can be expressed as $z = i\pi + 2i\pi n$ for $n \in \mathbb{Z}$. This means that the FCSwish function has an infinite number of singularities, periodically spaced along the imaginary axis at intervals of $2i\pi$.

Similarly, we define Fully Complex Mish (FCMish) (Figure 9.57) as

$$\sigma_{\text{FCMish}}(z) = z\,\text{Tanh}(\ln(1 + e^{z})). \tag{136}$$

The singularities of the FCMish function occur at specific points in the complex plane, $z = 2i\pi n + \log[-1 \pm i]$, for $n \in \mathbb{Z}$.

While the singular points of the FCSwish and FCMish functions theoretically exist, they do not typically interfere with the training of NNs. This is because the domain of interest is usually bounded within a small circle, making the occurrence of these problematic points rare and manageable. Moreover, during the initial stages of training, the weights and biases are usually initialized to small random values, ensuring that the initial activations are also small. As training progresses, regularization techniques and learning rate schedules help keep the activations within a desirable range, further minimizing the risk of encountering singularities. If the network's parameters do start to produce inputs near these singular points, mechanisms like gradient clipping, normalization, and careful monitoring of the training process can be employed to mitigate potential issues.

### 9.7.3 New Amplitude-Phase-Type Functions

In addition to extending existing real-valued AFs to complex domains, we introduce a set of novel CVAFs designed to enhance the performance and stability of CVNNs. These functions modulate the amplitude of the input complex number while preserving its phase, ensuring the output remains a valid complex number.

#### *9.7.3.1 Complex Amplitude-Phase Piecewise Linear Scaling (CAP-PLS)*

The CAP-PLS AF is defined as (Figure 9.58):

$$\sigma_{\text{CAPPLS}}(z) = \min[|z|, a]\,\frac{z}{|z|}, \tag{137.1}$$

where $a$ is positive parameter. For a complex number $z$ such that $|z| < a$, the function $\sigma_{\text{CAPPLS}}(z)$ returns the original complex number $z$. However, for a complex number $z$ such that $|z| > a$, the function scales the complex number $z$ to have an absolute value of $a$ while maintaining its direction (phase). This behavior can be interpreted as a "cap" on the magnitude of $z$. If the magnitude of $z$ exceeds a certain threshold $a$, the function reduces the magnitude to $a$, preserving the direction. One of the main advantages of the CAP-PLS function is its computational efficiency. Unlike the APTF function, which can be computationally expensive (using Tanh), the CAP-PLS function relies on simple linear operations (min and absolute value), making it faster to compute. This efficiency can be particularly beneficial in scenarios where computational resources are limited or when training large NNs.

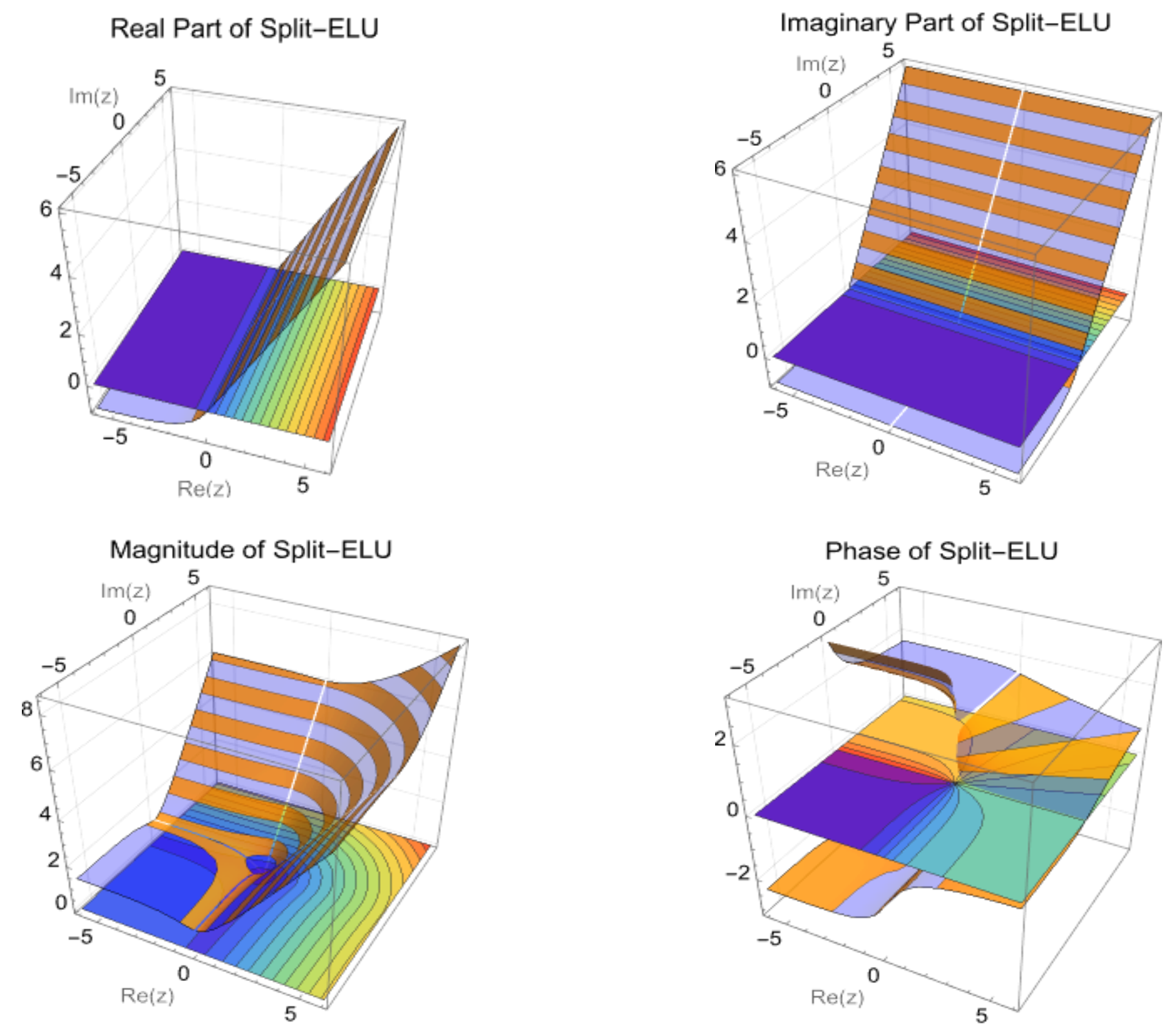

**Figure 9.49.** Visualizations of the real, imaginary, magnitude, and phase parts of the Split-ELU function.

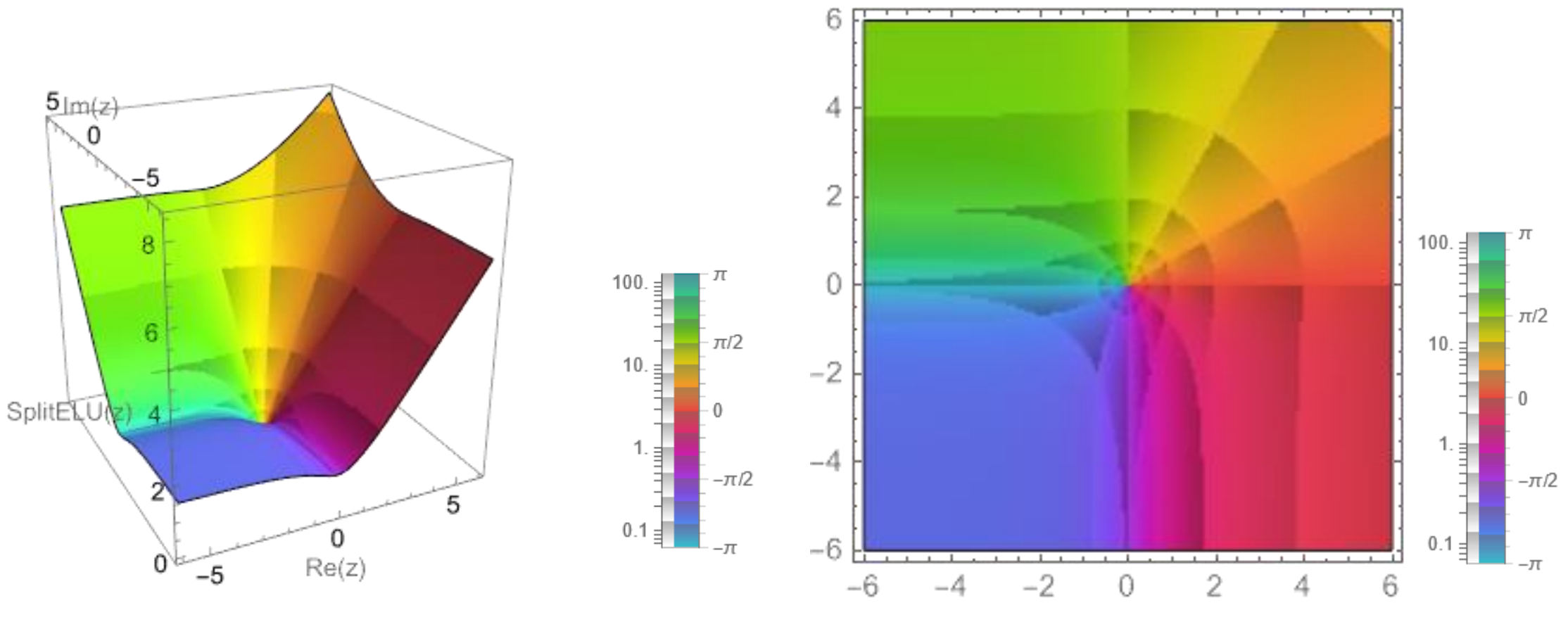

**Figure 9.50.** Left panel: The ComplexPlot3D generates a 3D plot of Abs[$\sigma_{\text{Split-ELU}}(z)$] colored by arg[$\sigma_{\text{Split-ELU}}(z)$] over the complex rectangle with corners $z_{min} = -6 - 6i$ and $z_{max} = 6 + 6i$. Using "CyclicLogAbsArg" to cyclically shade colors to give the appearance of contours of constant Abs[$\sigma_{\text{Split-ELU}}(z)$] and constant arg[$\sigma_{\text{Split-ELU}}(z)$]. Right panel: ComplexPlot generates a plot of arg[$\sigma_{\text{Split-ELU}}(z)$] over the complex rectangle with corners $-6 - 6i$ and $6 + 6i$. Using "CyclicLogAbsArg" to cyclically shade colors to give the appearance of contours of constant Abs[$\sigma_{\text{Split-ELU}}(z)$] and constant arg[$\sigma_{\text{Split-ELU}}(z)$].

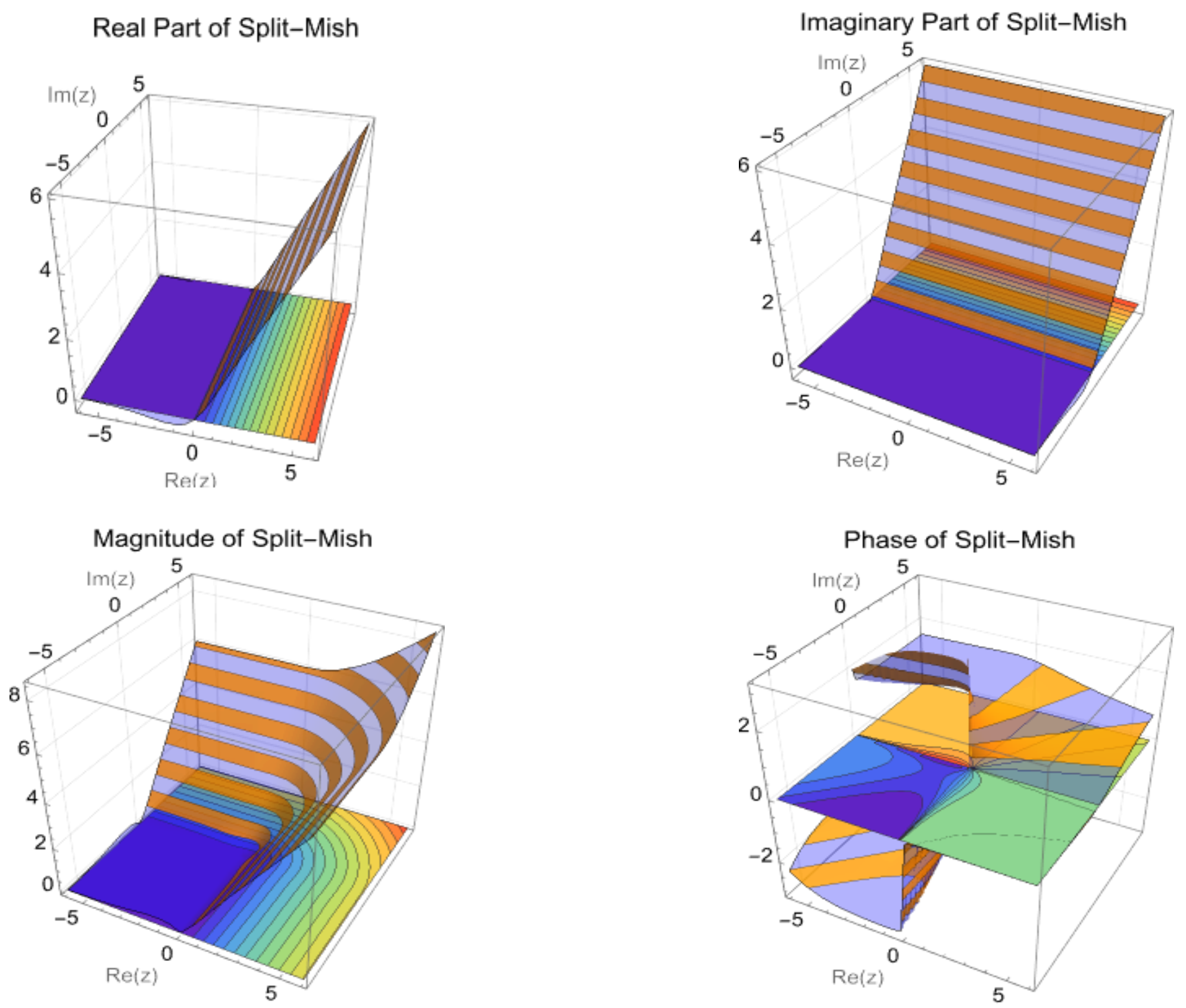

**Figure 9.51.** Visualizations of the real, imaginary, magnitude, and phase parts of the Split-Mish function.

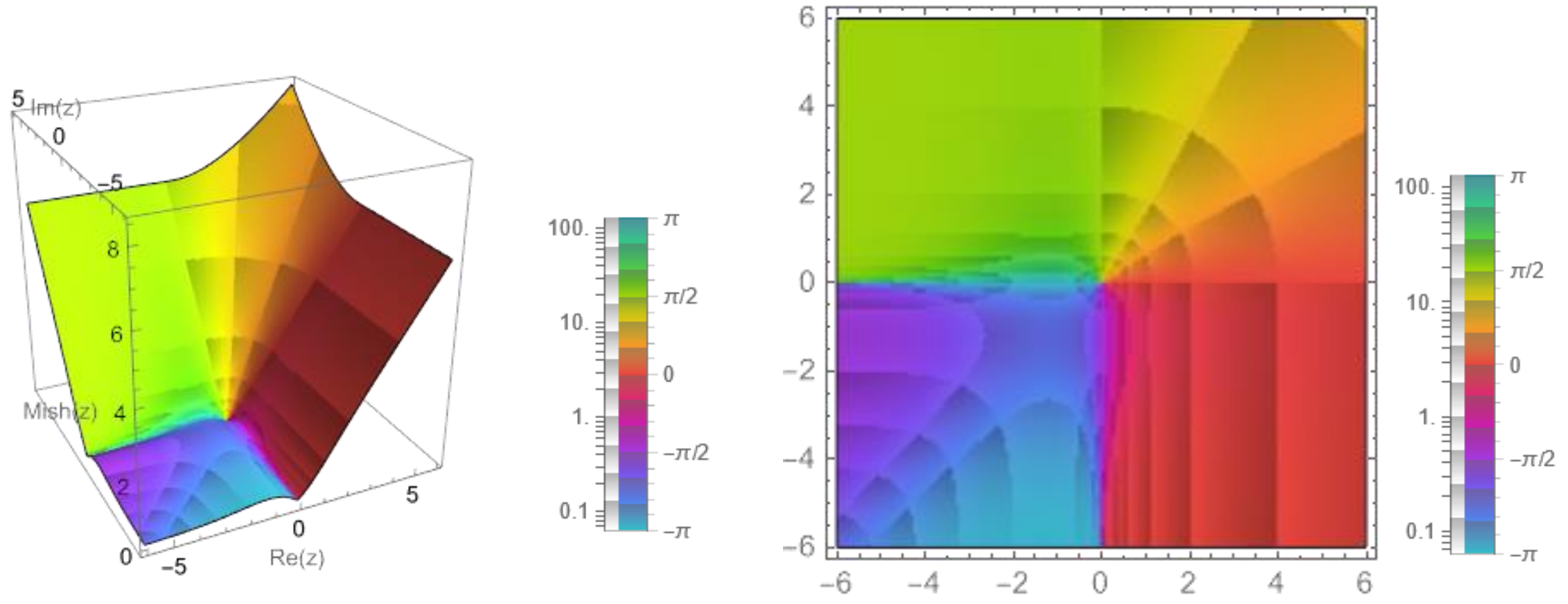

**Figure 9.52.** Left panel: The ComplexPlot3D generates a 3D plot of Abs[$\sigma_{\text{Split-Mish}}(z)$] colored by arg[$\sigma_{\text{Split-Mish}}(z)$] over the complex rectangle with corners $z_{min} = -6 - 6i$ and $z_{max} = 6 + 6i$. Using "CyclicLogAbsArg" to cyclically shade colors to give the appearance of contours of constant Abs[$\sigma_{\text{Split-Mish}}(z)$] and constant arg[$\sigma_{\text{Split-Mish}}(z)$]. Right panel: ComplexPlot generates a plot of arg[$\sigma_{\text{Split-Mish}}(z)$] over the complex rectangle with corners $-6 - 6i$ and $6 + 6i$. Using "CyclicLogAbsArg" to cyclically shade colors to give the appearance of contours of constant Abs[$\sigma_{\text{Split-Mish}}(z)$] and constant arg[$\sigma_{\text{Split-Mish}}(z)$].

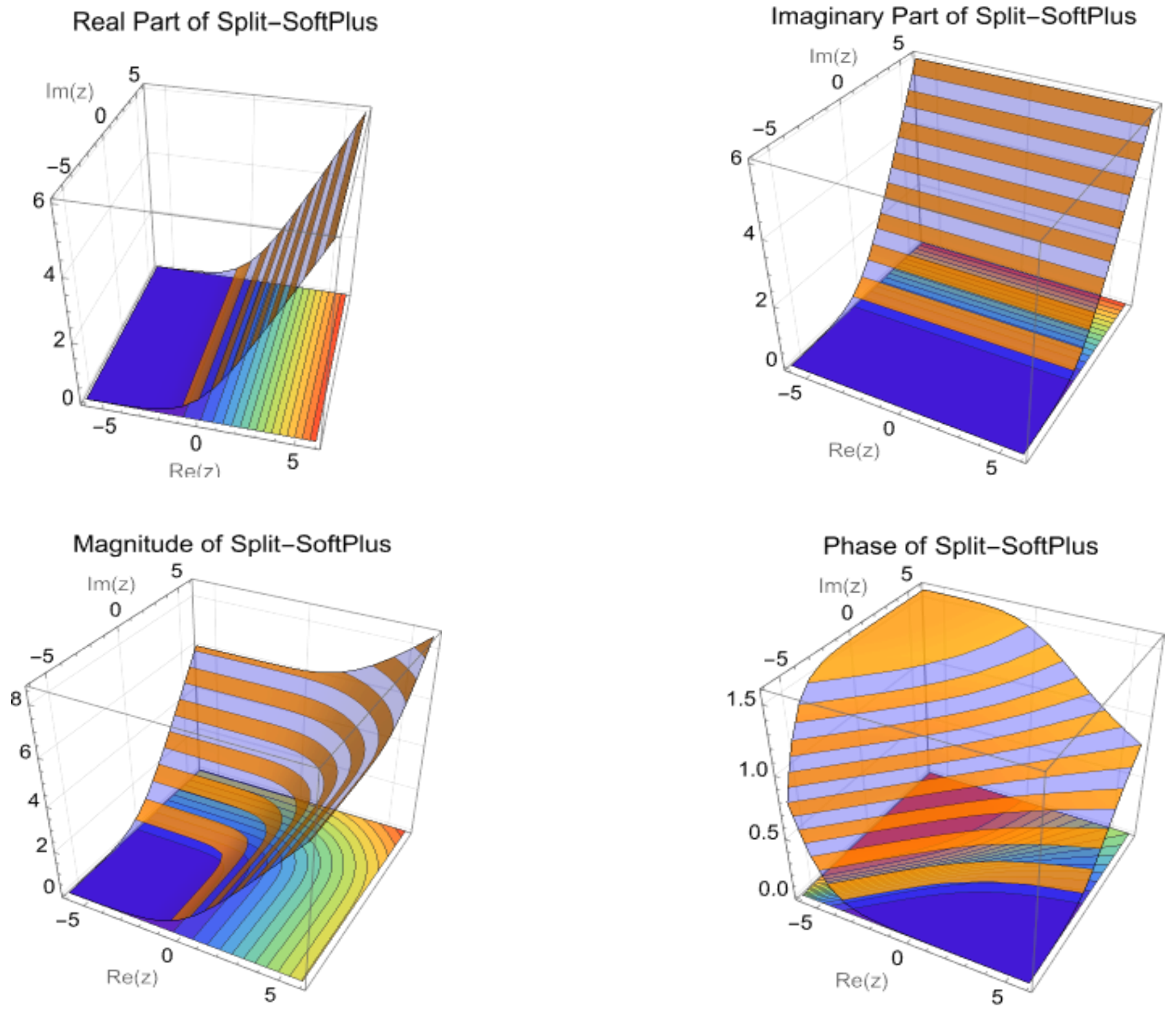

**Figure 9.53.** Visualizations of the real, imaginary, magnitude, and phase parts of the Split-SoftPlus function.

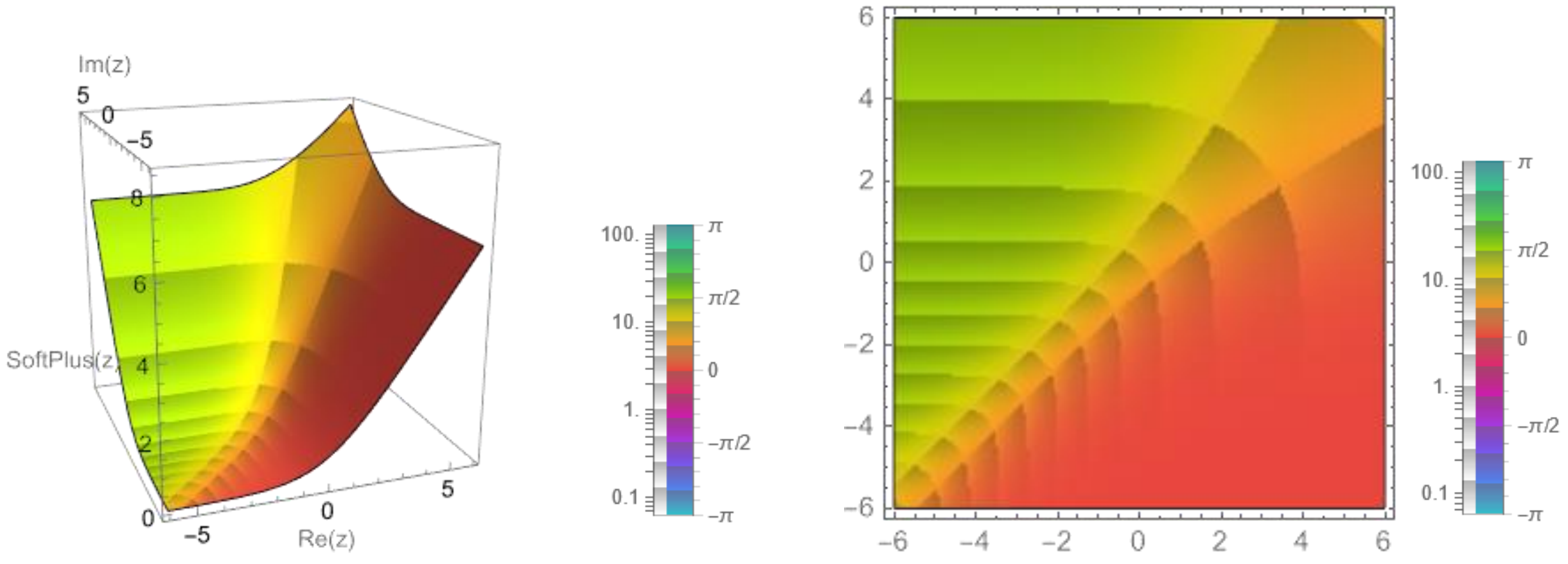

**Figure 9.54.** Left panel: The ComplexPlot3D generates a 3D plot of Abs[$\sigma_{\text{Split-SoftPlus}}(z)$] colored by arg[$\sigma_{\text{Split-SoftPlus}}(z)$] over the complex rectangle with corners $z_{min} = -6 - 6i$ and $z_{max} = 6 + 6i$. Using "CyclicLogAbsArg" to cyclically shade colors to give the appearance of contours of constant Abs[$\sigma_{\text{Split-SoftPlus}}(z)$] and constant arg[$\sigma_{\text{Split-SoftPlus}}(z)$]. Right panel: ComplexPlot generates a plot of arg[$\sigma_{\text{Split-SoftPlus}}(z)$] over the complex rectangle with corners $-6 - 6i$ and $6 + 6i$. Using "CyclicLogAbsArg" to cyclically shade colors to give the appearance of contours of constant Abs[$\sigma_{\text{Split-SoftPlus}}(z)$] and constant arg[$\sigma_{\text{Split-SoftPlus}}(z)$].

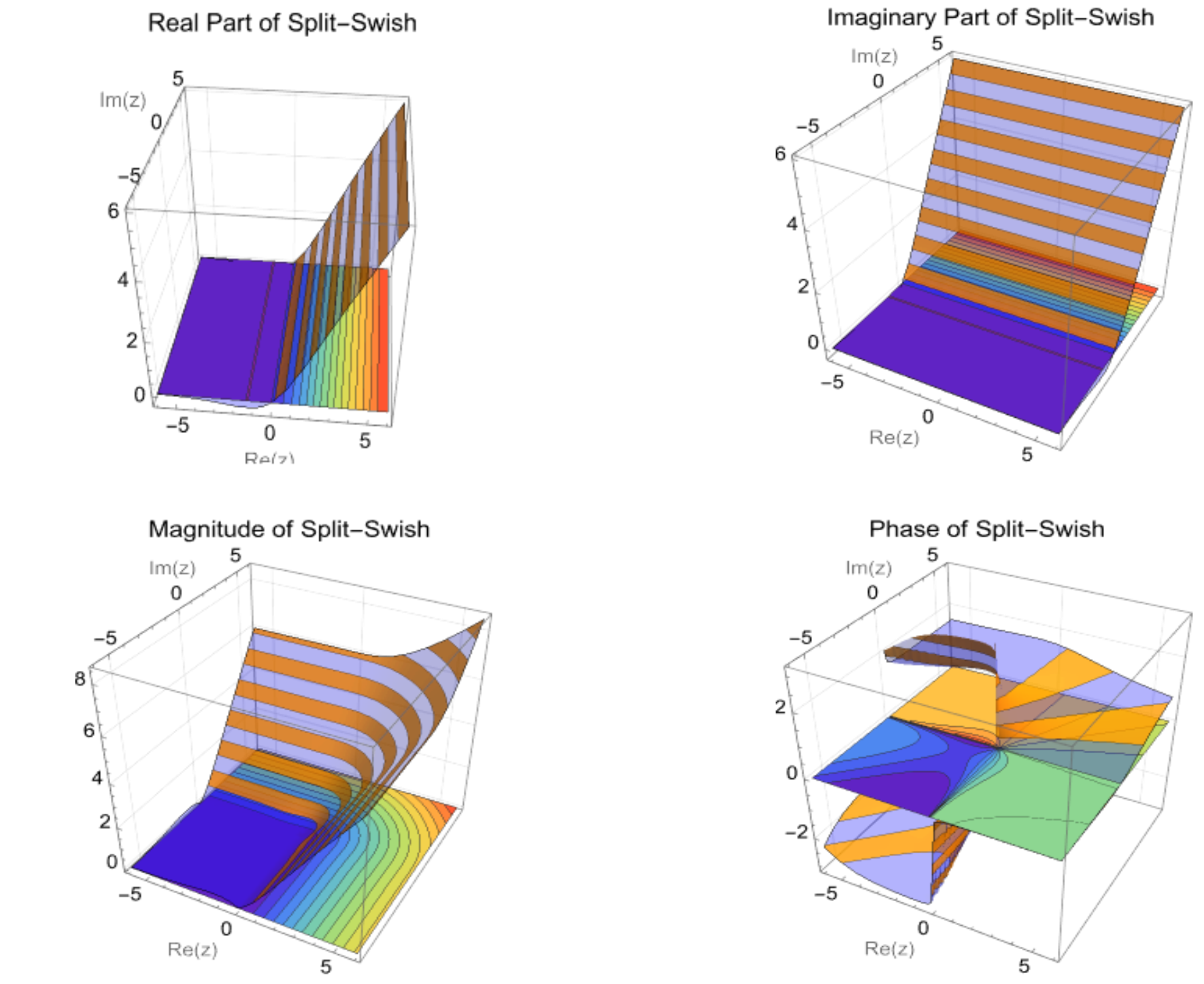

**Figure 9.55.** Visualizations of the real, imaginary, magnitude, and phase parts of the Split-Swish function.

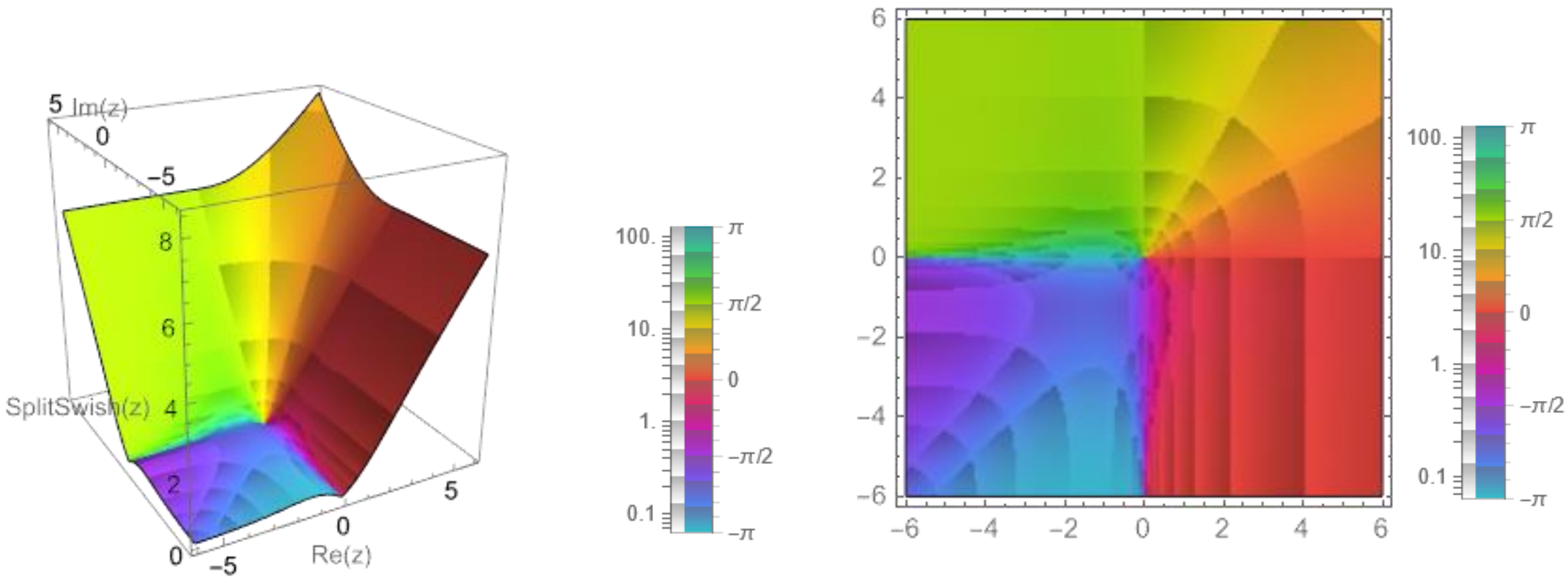

**Figure 9.56.** Left panel: The ComplexPlot3D generates a 3D plot of Abs[$\sigma_{\text{Split-Swish}}(z)$] colored by arg[$\sigma_{\text{Split-Swish}}(z)$] over the complex rectangle with corners $z_{min} = -6 - 6i$ and $z_{max} = 6 + 6i$. Using "CyclicLogAbsArg" to cyclically shade colors to give the appearance of contours of constant Abs[$\sigma_{\text{Split-Swish}}(z)$] and constant arg[$\sigma_{\text{Split-Swish}}(z)$]. Right panel: ComplexPlot generates a plot of arg[$\sigma_{\text{Split-Swish}}(z)$] over the complex rectangle with corners $-6 - 6i$ and $6 + 6i$. Using "CyclicLogAbsArg" to cyclically shade colors to give the appearance of contours of constant Abs[$\sigma_{\text{Split-Swish}}(z)$] and constant arg[$\sigma_{\text{Split-Swish}}(z)$].

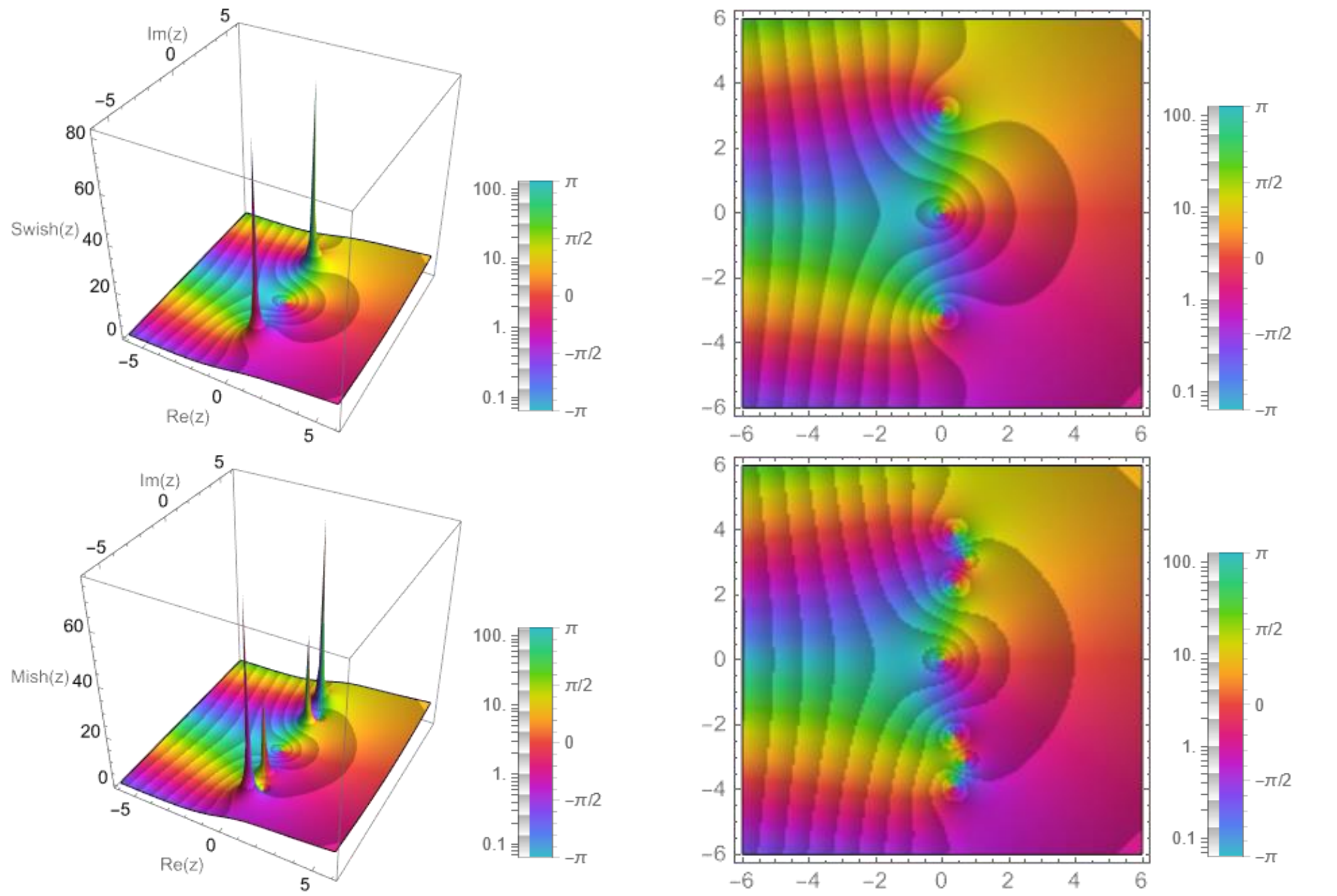

**Figure 9.57.** Top panel (left): The ComplexPlot3D generates a 3D plot of Abs[$\sigma_{\text{FCSwish}}(z)$] colored by arg[$\sigma_{\text{FCSwish}}(z)$] over the complex rectangle with corners $z_{min} = -6 - 6i$ and $z_{max} = 6 + 6i$. Using "CyclicLogAbsArg" to cyclically shade colors to give the appearance of contours of constant Abs[$\sigma_{\text{FCSwish}}(z)$] and constant arg[$\sigma_{\text{FCSwish}}(z)$]. Top panel (right): ComplexPlot generates a plot of arg[$\sigma_{\text{FCSwish}}(z)$] over the complex rectangle with corners $-6 - 6i$ and $6 + 6i$. Using "CyclicLogAbsArg" to cyclically shade colors to give the appearance of contours of constant Abs[$\sigma_{\text{FCSwish}}(z)$] and constant arg[$\sigma_{\text{FCSwish}}(z)$]. Bottom panel: Same as top panel but for the function $\sigma_{\text{FCMish}}(z)$.

### *9.7.3.2 Complex Amplitude-Phase Exponential Scaling (CAP-ES)*

The CAP-ES AF is formulated as (Figure 9.59):

$$\sigma_{\text{CAPES}}(z) = \left(1 - e^{-|z|}\right)\frac{z}{|z|}. \tag{137.2}$$

For any complex number $z$, the function $\sigma_{\text{CAPES}}(z)$ modulates the magnitude of $z$ based on the exponential decay while preserving its direction. When $|z|$ is small, $e^{-|z|}$ is close to 1, making $1 - e^{-|z|}$ close to 0, and thus $\sigma_{\text{CAPES}}(z)$ is also small. As $|z|$ increases, $e^{-|z|}$ approaches 0, making $1 - e^{-|z|}$ approach 1, and $\sigma_{\text{CAPES}}(z)$ approximates $z$ normalized to unit magnitude. This formulation emphasizes saturation in amplitude while keeping the phase unchanged.

### *9.7.3.3 Complex Amplitude-Phase ArcTan Scaling (CAP-ArcTanS)*

The CAP-ArcTanS AF is represented as (Figure 9.60):

$$\sigma_{\text{CAPArcTanS}}(z) = \text{ArcTan}(|z|)\frac{z}{|z|}. \tag{137.3}$$

The function $\sigma_{\text{CAPArcTanS}}(z)$ modulates the magnitude of $z$ through ArcTan($|z|$) while keeping the direction unchanged. When $|z|$ is small, ArcTan($|z|$) $\approx |z|$, making $\sigma_{\text{CAPArcTanS}}(z)$ approximately equal to $z$. As $|z|$ increases, $\sigma_{\text{CAPArcTanS}}(z)$ asymptotically approaches $\pi/2$, causing the magnitude of $\sigma_{\text{CAPArcTanS}}(z)$ to grow more slowly and ultimately saturate at this value. This behavior ensures controlled amplitude growth, which can be useful in preventing the explosion of values during processing while maintaining the phase information intact.

*9.7.3.4 Complex Amplitude-Phase Erf Attenuation (CAP-ErfA)*

The CAP-CErfA AF is (Figure 9.61):

$$\sigma_{\text{APCErfA}}(z) = \text{erf}(|z|)\frac{z}{|z|}. \tag{137.4}$$

The function $\sigma_{\text{APCErfA}}(z)$ modulates the magnitude of $z$ through $\text{erf}(|z|)$ while preserving its direction. For small values of $|z|$, $\text{erf}(|z|) = |z|$, making $\sigma_{\text{APCErfA}}(z)$ approximately equal to $z$. As $|z|$ increases, $\text{erf}(|z|)$ approaches 1, causing the magnitude of $\sigma_{\text{APCErfA}}(z)$ to saturate and preventing it from growing indefinitely. This characteristic ensures that the amplitude of $\sigma_{\text{APCErfA}}(z)$ remains within a bounded range, thus controlling the growth of the magnitude while maintaining the phase information.

*9.7.3.5 Complex Amplitude-Phase Softplus (CAP-Softplus or modSoftplus)*

The CAP-Softplus AF or modSoftplus is mathematically specified as (Figure 9.62):

$$\sigma_{\text{CAPSoftplus}}(z) = \sigma_{\text{modSoftplus}}(z) = \text{Log}\left[1 + e^{(|z|-a)}\right]\frac{z}{|z|}. \tag{137.5}$$

The function $\sigma_{\text{CAPSoftplus}}(z)$ modifies the magnitude of $z$ through $\text{Log}\left[1 + e^{(|z|-a)}\right]$ while keeping the direction unchanged. This transformation ensures that the amplitude grows smoothly and never becomes negative, which is particularly useful for stability in NN training. For values of $|z|$ close to $a$, $e^{(|z|-a)}$ is small, and thus $\text{Log}\left[1 + e^{(|z|-a)}\right] \approx \text{Log}[2]$, leading to a moderated amplitude. For large values of $|z|$, the term $\text{Log}\left[1 + \text{e}^{(|z|-a)}\right]$ asymptotically approaches $|z| - a$. It is a smoothing version of the modReLU function. While modReLU creates a hard threshold where the function output is zero for negative values of $|z| + b$ and retains the adjusted amplitude for non-negative values, CAP-Softplus smooths this transition. CAP-Softplus avoids the sharp transitions and potential instability of modReLU by using the logarithmic function to provide a smooth and continuous growth of the amplitude.

*9.7.3.6 Complex Amplitude-Phase Exponential Linear Unit (CAP-ELU or modELU)*

The CAP-ELU AF or modELU is defined as (Figure 9.63):

$$\sigma_{\text{CAPELU}}(z) = \sigma_{\text{modELU}}(z) = \begin{cases} (|z| + b)\dfrac{z}{|z|}, & |z| < -\text{b}, \\ \alpha(e^{|z|+b} - 1)\dfrac{z}{|z|}, & \text{otherwise}, \end{cases} \tag{137.6}$$

where $b$ is negative parameter, $b < 0$. This function combines linear and exponential terms to modulate the magnitude of $z$. For $|z| < -b$, the function modifies the amplitude by simply adding $b$ to $|z|$. This ensures that for small magnitudes, the amplitude is shifted by $b$ but the phase of $z$ is preserved. For $|z| > -b$, the function applies an exponential transformation. This allows the amplitude to grow exponentially with $|z|$, modulated by the parameter $\alpha$, while still preserving the phase of $z$. As $|z|$ is always positive, a bias $b$ is introduced to create a "Modulation Region" of radius $b$ around the origin 0 where the amplitude is shifted by $b$, and outside of which the amplitude is grow exponentially with $|z|$. This behaver like modReLU AF where the ReLU operation $\sigma_{\text{ReLU}}(|z| + b) = \max(|z| + b, 0)$ ensures that the function becomes zero for inputs ($|z| + b < 0$) and retains the amplitude-modulated $z$ for inputs ($|z| + b \geq 0$). In modReLU AF, as $|z|$ is always positive, a bias $b$ is introduced in order to create a "dead zone" of radius $b$ around the origin 0 where the neuron is inactive, and outside of which it is active. The CAP-ELU AF, however, goes a step further by combining linear and exponential terms. This combination allows the network to leverage the benefits of both linear adjustments for small magnitudes and exponential growth for larger magnitudes. The formulation of CAP-ELU allows the network to capture information from both $|z| < -\text{b}$ and $|z| > -\text{b}$ of the input space, contributing to its ability to maintain mean activations close to zero.

*9.7.3.7 Complex Amplitude-Phase Swish (CAP-Swish or modSwish)*

The CAP-Swish AF or modSwish is given by (Figure 9.64):

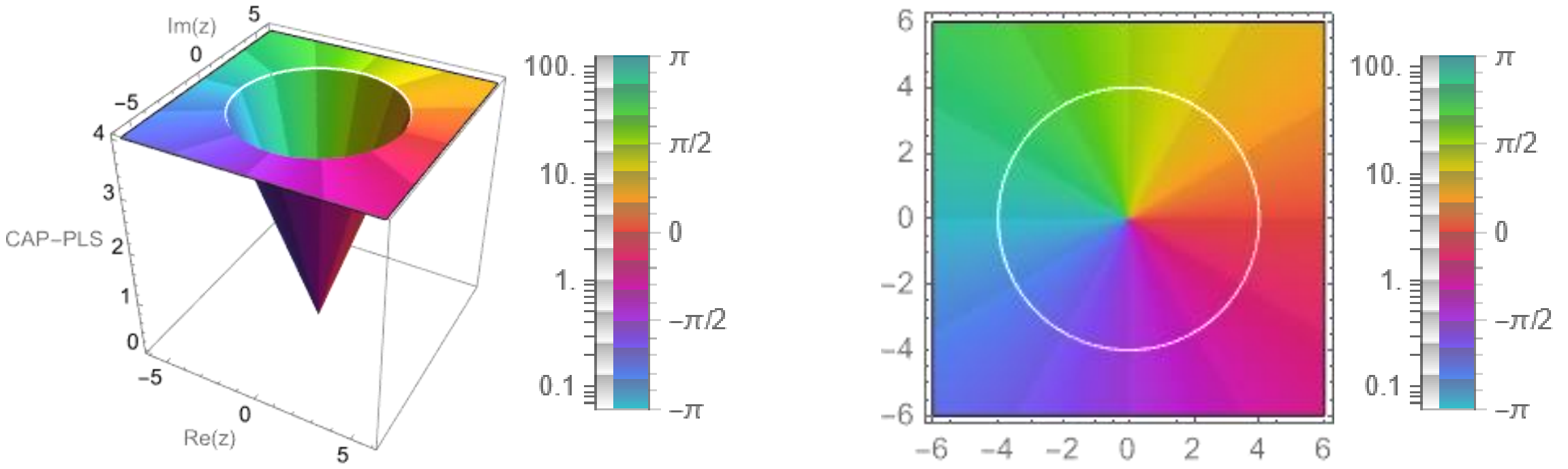

**Figure 9.58.** Top panel: The ComplexPlot3D generates a 3D plot of Abs[$\sigma_{\text{CAP-PLS}}(z)$] colored by arg[$\sigma_{\text{CAP-PLS}}(z)$] over the complex rectangle with corners $z_{min} = -6 - 6i$ and $z_{max} = 6 + 6i$. Using "CyclicLogAbsArg" to cyclically shade colors to give the appearance of contours of constant Abs[$\sigma_{\text{CAP-PLS}}(z)$] and constant arg[$\sigma_{\text{CAP-PLS}}(z)$]. Top panel (right): ComplexPlot generates a plot of arg[$\sigma_{\text{CAP-PLS}}(z)$] over the complex rectangle with corners $-6 - 6i$ and $6 + 6i$. Using "CyclicLogAbsArg" to cyclically shade colors to give the appearance of contours of constant Abs[$\sigma_{\text{CAP-PLS}}(z)$] and constant arg[$\sigma_{\text{CAP-PLS}}(z)$].

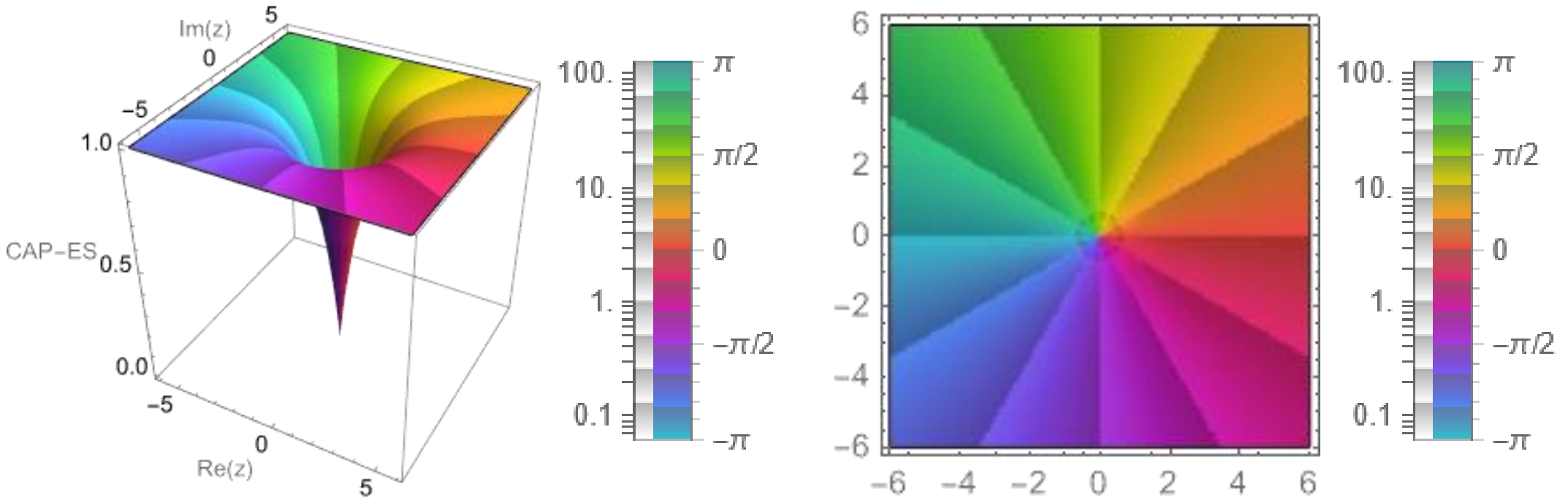

**Figure 9.59.** Same as Figure 9.58 but for the function $\sigma_{\text{CAP-ES}}(z)$.

$$\sigma_{\text{CAPSwish}}(z) = \frac{|z|}{1 + e^{-|z|+b}} \frac{z}{|z|}. \tag{137.7}$$

When $|z|$ is small, the term $e^{-|z|+b}$ is close to $e^b$, making $\frac{|z|}{1+e^{-|z|+b}} \approx \frac{|z|}{1+e^b}$, thus moderating the amplitude for small magnitudes. As $|z|$ increases, $e^{-|z|+b}$ approaches zero, and $\frac{|z|}{1+e^{-|z|+b}}$ approximates $|z|$, allowing the amplitude to grow smoothly and without abrupt changes. As $|z|$ is always positive, a bias $b$ is introduced to create a "Modulation Region" of radius $b$ around the origin 0 where the amplitude is moderating, and outside of which the amplitude is grow linearly with $|z|$. This behaver like modReLU AF where the ReLU operation $\sigma_{\text{ReLU}}(|z| + b) = \max(|z| + b, 0)$ ensures that the function becomes zero for inputs ($|z| + b < 0$) and retains the amplitude-modulated $z$ for inputs ($|z| + b \geq 0$).

CAP-Swish (or modSwish) is a versatile AF that combines the benefits of smoothness, non-linearity, and directional consistency. Its design, incorporating a Sigmoid-like modulation around zero and linear growth for larger inputs, makes it suitable for various CVNN architectures and training scenarios.

In general, the above proposed AFs share several important properties that make them well-suited for use in CVNNs. All functions preserve the phase of the input $z$ by multiplying the modulated magnitude with $z/|z|$, maintaining the complex nature of the input. Most functions ensure that the output is bounded or grows slowly with increasing $|z|$, preventing the propagation of excessively large values and contributing to the stability of the training process.

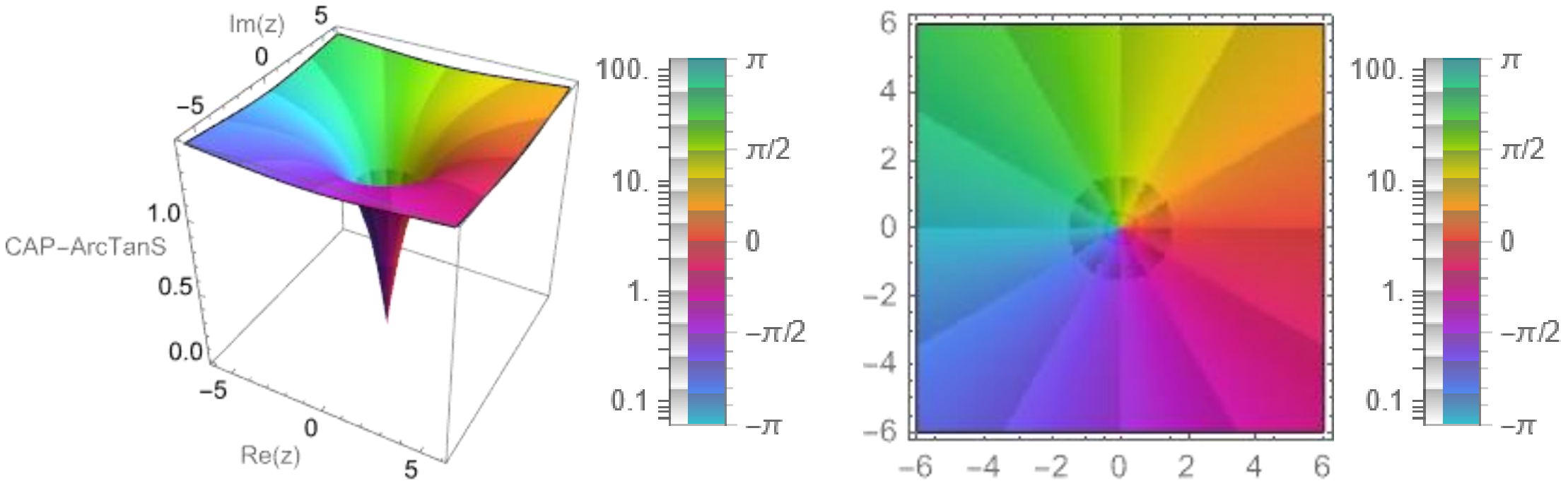

**Figure 9.60.** Same as Figure 9.58 but for the function $\sigma_{\text{CAP-ArcTanS}}(z)$.

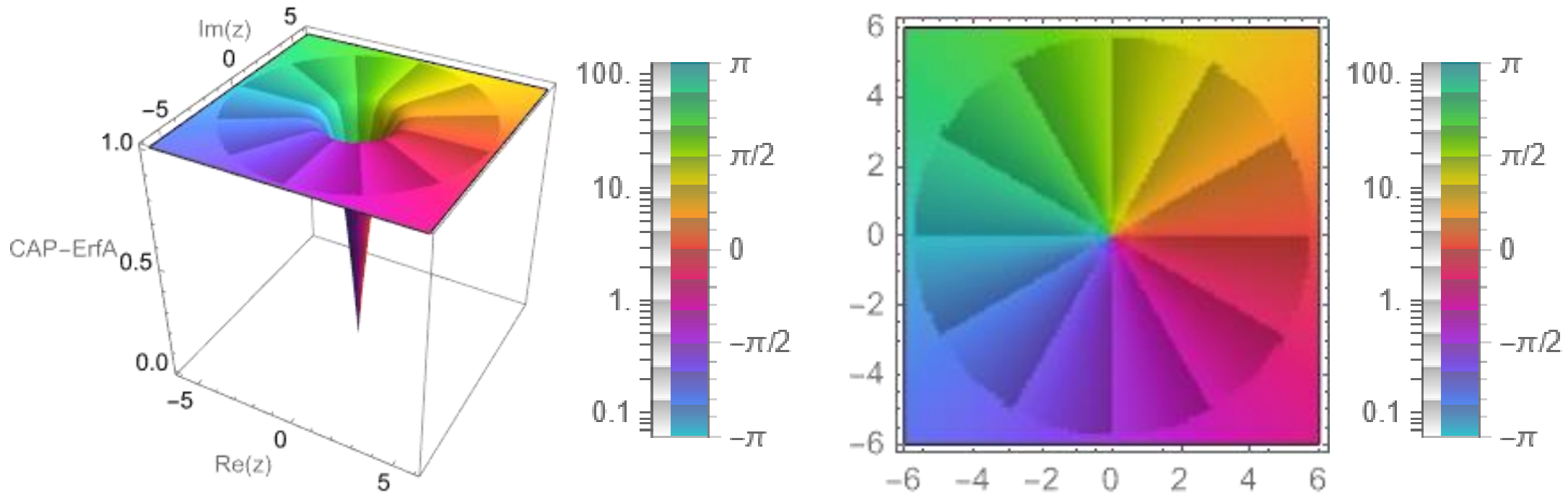

**Figure 9.61.** Same as Figure 9.58 but for the function $\sigma_{\text{CAP-ErfA}}(z)$.

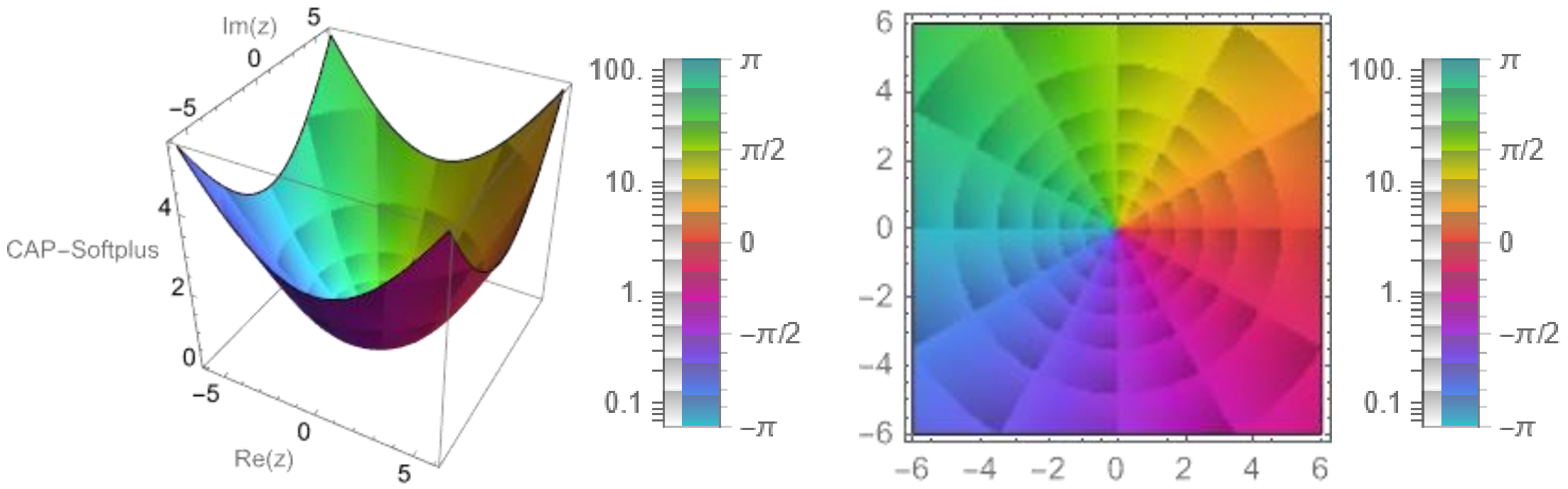

**Figure 9.62.** Same as Figure 9.58 but for the function $\sigma_{\text{CAP-SoftPlus}}(z)$.

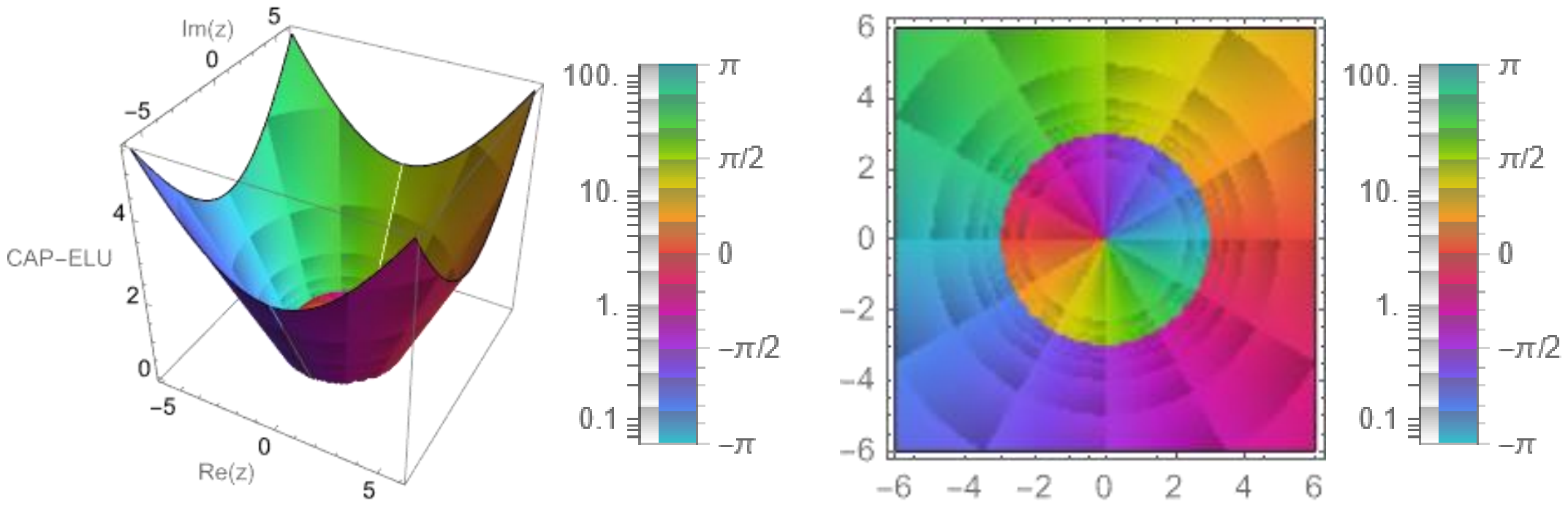

**Figure 9.63.** Same as Figure 9.58 but for the function $\sigma_{\text{CAP-ELU}}(z)$.

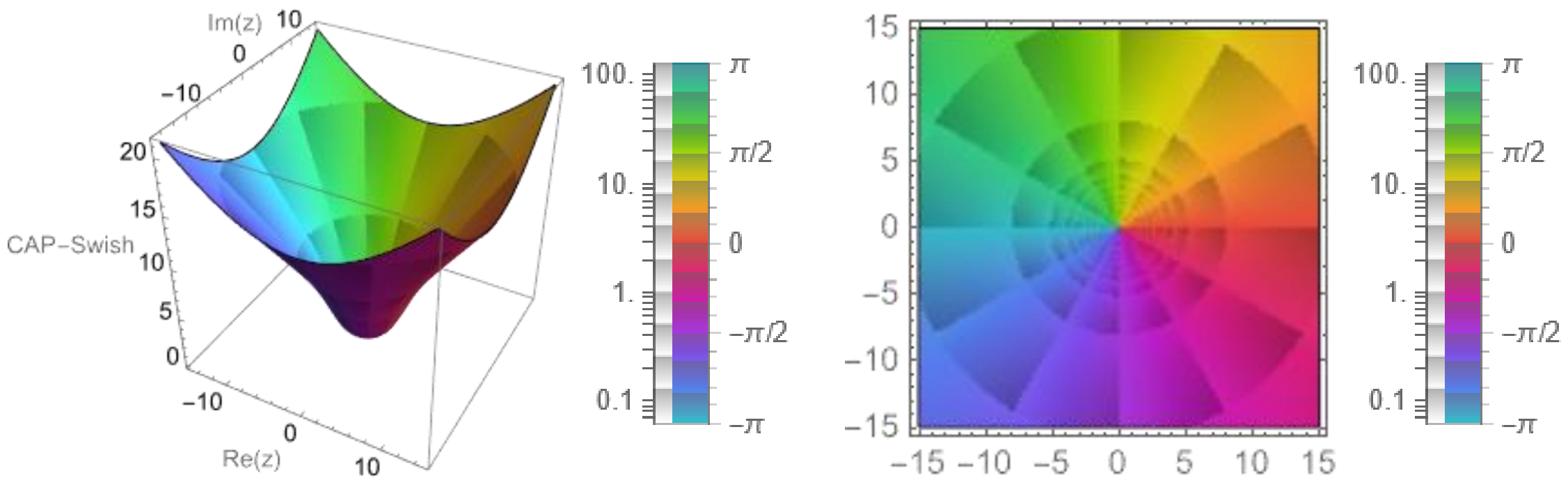

**Figure 9.64.** Same as Figure 9.58 but for the function $\sigma_{\text{CAP-Swish}}(z)$.